\documentclass[final]{cvpr}

\usepackage{times}
\usepackage{epsfig}
\usepackage{graphicx}
\usepackage{amsmath}
\usepackage{amssymb}

\usepackage[pagebackref=true,breaklinks=true,colorlinks,bookmarks=false]{hyperref}

\def\cvprPaperID{1234} % *** Enter the CVPR Paper ID here
\def\confYear{CVPR 2021}
\usepackage{pifont}% http://ctan.org/pkg/pifont
\usepackage{acro}
\usepackage{booktabs}
\usepackage{comment}

\usepackage{array}
\newcolumntype{P}[1]{>{\centering\arraybackslash}p{#1}}

\usepackage{caption} 

\usepackage[export]{adjustbox}

\DeclareAcronym{vae}{
  short = VAE,
  long = Variational Autoencoder
}
\DeclareAcronym{gan}{
  short = GAN,
  long = Generative Adversarial Network
}
\DeclareAcronym{nf}{
  short = NF ,
  long = Normalizing Flow
}
\DeclareAcronym{nlp}{
  short = NLP,
  long = Natural Language Processing
}

\DeclareAcronym{cnn}{
  short = CNN,
  long = Convolutional Neural Network
}

\DeclareAcronym{rnn}{
  short = RNN,
  long = Recurrent Neural Network
}

\DeclareAcronym{lstm}{
  short = LSTM,
  long = Long Short-Term Memory
}

\DeclareAcronym{ocr}{
  short=OCR,
  long=Optical Character Recognition
}

\DeclareAcronym{rvnn}{
    short=RvNN,
    long=Recursive Neural Network
}

\DeclareAcronym{tts}{
    short=TTS,
    long=Text-to-Speech
}

\DeclareAcronym{elbo}{
    short=ELBO,
    long=evidence lower bound
}

\DeclareAcronym{ui}{
    short=UI,
    long=User Interface,
    long-plural-form=User Interfaces
}

\DeclareAcronym{pcb}{
    short=PCB,
    long=printed circuit board
}

\DeclareAcronym{gcn}{
    short=GCN,
    long=Graph Convolution Network
}

\DeclareAcronym{iou}{
    short=IoU,
    long=Intersection over Union
}

\DeclareAcronym{vtn}{
    short=VTN,
    long=Variational Transformer Network
}

\begin{document}

%%%%%%%%% TITLE
\title{Variational Transformer Networks for Layout Generation}

\author{Diego Martin Arroyo\textsuperscript{1}\\
{\tt\small martinarroyo@google.com}\\
\textsuperscript{1}Google, Inc\\
\and
Janis Postels\textsuperscript{2}\\
{\tt\small jpostels@vision.ee.ethz.ch}\\
\textsuperscript{2}ETH Z\"urich\\
\and
Federico Tombari\textsuperscript{1,3}\\
{\tt\small tombari@google.com}\\
\textsuperscript{3}Technische Universit\"at M\"unchen\\
}

\maketitle
\thispagestyle{empty}

%%%%%%%%% ABSTRACT
% The ABSTRACT is to be in fully-justified italicized text, at the top
% of the left-hand column, below the author and affiliation
% information. Use the word ``Abstract'' as the title, in 12-point
% Times, boldface type, centered relative to the column, initially
% capitalized. The abstract is to be in 10-point, single-spaced type.
% Leave two blank lines after the Abstract, then begin the main text.
% Look at previous CVPR abstracts to get a feel for style and length.
\begin{abstract}
% Generative models able to synthesize layouts of different kinds (\eg documents, user interfaces or furniture arrangements) are a useful tool to aid design processes and as a first step in the generation of synthetic data, among other tasks. We focus on this problem to drive our investigation into the capabilities of self-attention layers as building blocks of the well-known \ac{vae} formulation. Specifically, we propose a novel VAE model based on transformer networks and show the benefits of this combination for the task of layout generation. An extensive evaluation on publicly available benchmarks demonstrates the benefits in terms of variability and quality of our approach with respect to the state of the art.
Generative models able to synthesize layouts of different kinds (\eg documents, user interfaces or furniture arrangements) are a useful tool to aid design processes and as a first step in the generation of synthetic data, among other tasks. We exploit the properties of self-attention layers to capture high level relationships between elements in a layout, and use these as the building blocks of the well-known \ac{vae} formulation. Our proposed \acf{vtn} is capable of learning margins, alignments and other global design rules without explicit supervision. Layouts sampled from our model have a high degree of resemblance to the training data, while demonstrating appealing diversity. In an extensive evaluation on publicly available benchmarks for different layout types \acp{vtn} achieve state-of-the-art diversity and perceptual quality. Additionally, we show the capabilities of this method as part of a document layout detection pipeline.
\end{abstract}
\section{Introduction}

Layouts, \ie the abstract positioning of elements in a scene or document, constitute an essential tool for various downstream tasks. Consequently, the ability to flexibly render novel, realistic layouts has the potential to yield significant improvements in many tasks, such as neural scene synthesis \cite{wang2018deep}, graphic design or in data synthesis pipelines. Even though the task of synthesizing novel layouts has recently started to gain the attention of the deep learning community \cite{DBLP:conf/iclr/LiYHZX19, DBLP:conf/iccv/JyothiDHSM19, 2020-Lee-NDNGLGWC, DBLP:conf/cvpr/PatilBPA20}, it is still a sparsely explored area and provides unique challenges to generative models based on neural networks, namely a non-sequential data structure consisting of varying length samples with discrete (classes) and continuous (coordinates) elements simultaneously.

Generative models based on neural networks have received a significant share of attention in recent years, as they proved capable of learning complex, high-dimensional distributions. Common formulations such as \acp{gan} \cite{2014arXiv1406.2661G} and \acfp{vae} \cite{DBLP:journals/corr/KingmaW13} have shown impressive results in tasks such as image translation \cite{CycleGAN2017}, image synthesis \cite{DBLP:conf/cvpr/KarrasLA19}, and text generation \cite{DBLP:conf/conll/BowmanVVDJB16}.
A \ac{gan} is comprised of an arrangement of generator-discriminator neural networks in a zero-sum configuration, while a \ac{vae} learns a lower bound of the data distribution using an encoder-decoder neural network with a regularized bottleneck. Since these are general frameworks, they leave room for adapting the underlying neural architectures to exploit the properties of the data. For example, the weight sharing strategy of \acp{cnn} renders them the most common building block for image processing, while for sequential data (\eg, text), \acp{rnn} or attention modules are often the architecture of choice. In particular, the attention mechanism has recently demonstrated strong performance on a variety of tasks, such as language translation \cite{DBLP:conf/nips/VaswaniSPUJGKP17} and object detection \cite{DBLP:conf/eccv/CarionMSUKZ20}, proving its superiority over \acp{rnn} regarding modeling long-term relationships.

\begin{figure}[t]
    \centering
    \setlength{\tabcolsep}{1pt}
    \includegraphics[width=\linewidth]{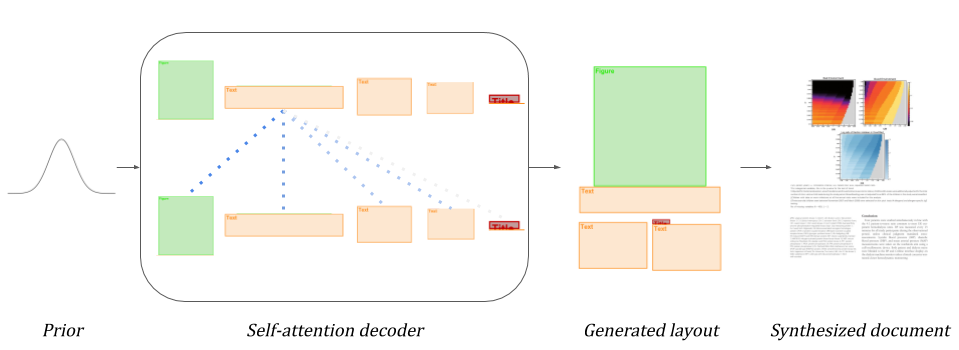}
    \begin{tabular}{cccc}
    \raisebox{0.0\height}{\includegraphics[width=0.25\linewidth,frame]{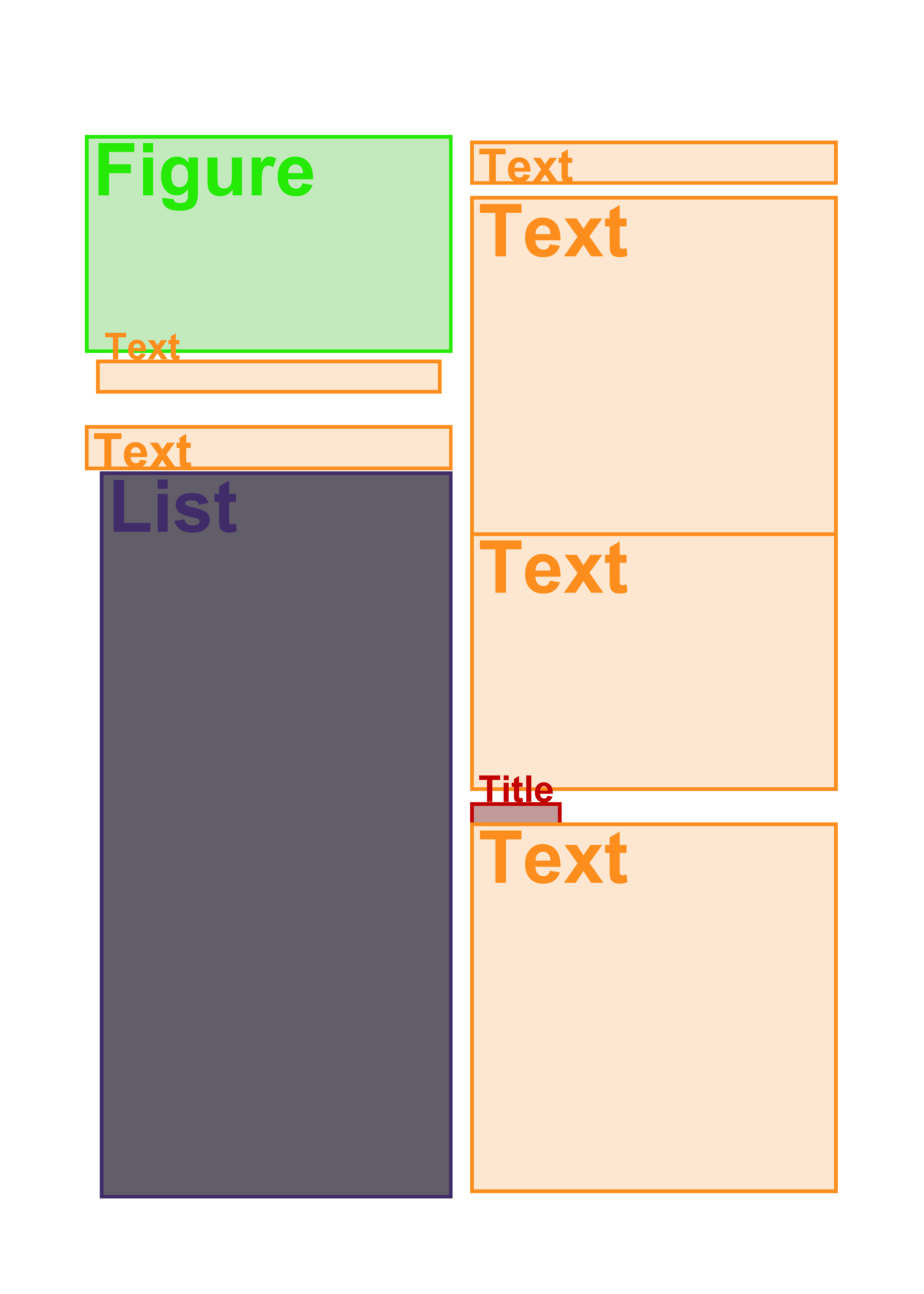}} &
    \includegraphics[width=0.198\linewidth]{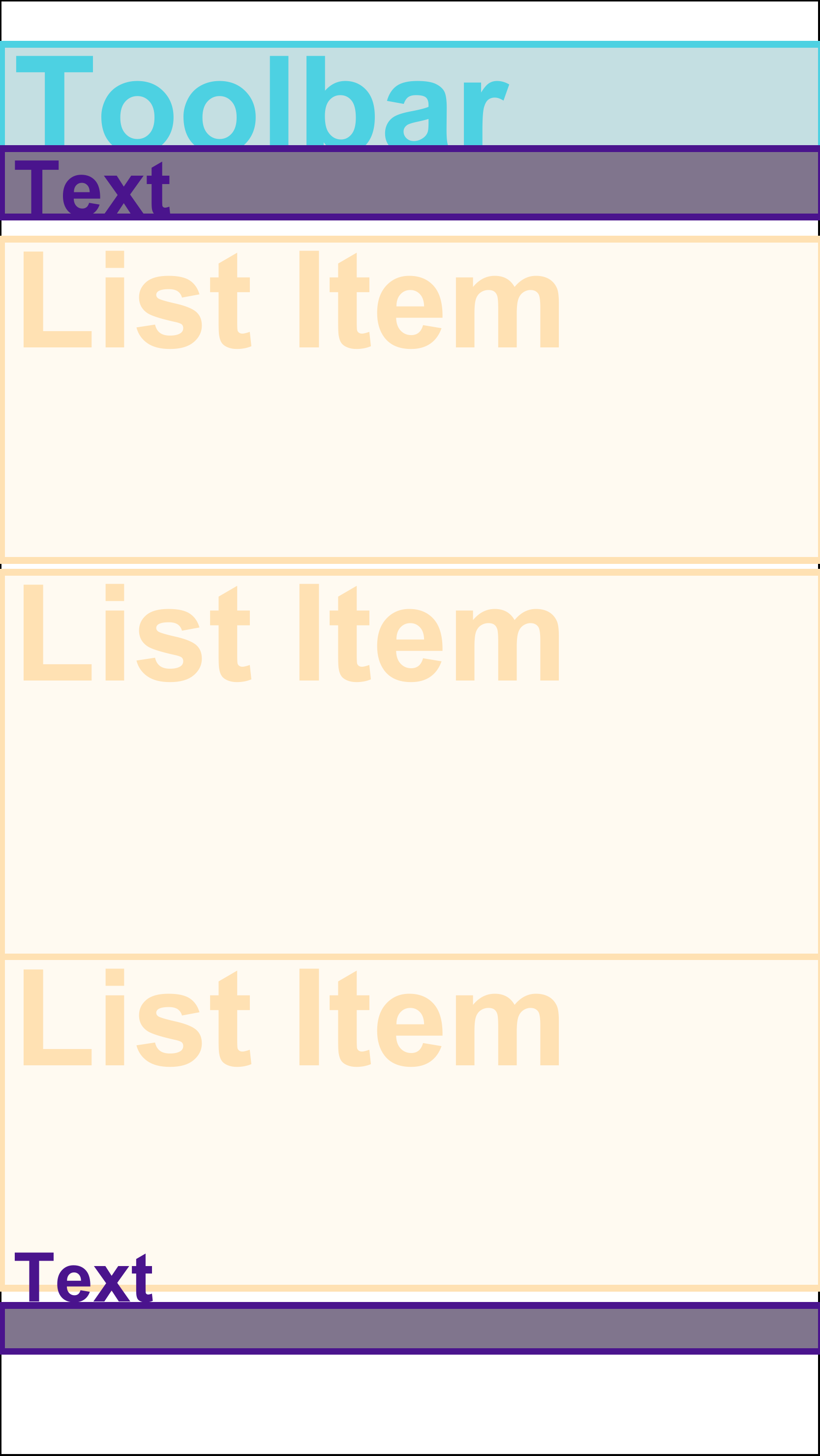} &
    \raisebox{0.25\height}{\includegraphics[width=0.24\linewidth]{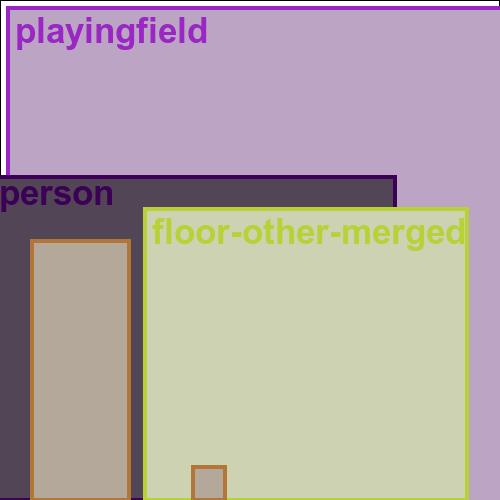}} &
    \raisebox{0.25\height}{\includegraphics[width=0.24\linewidth]{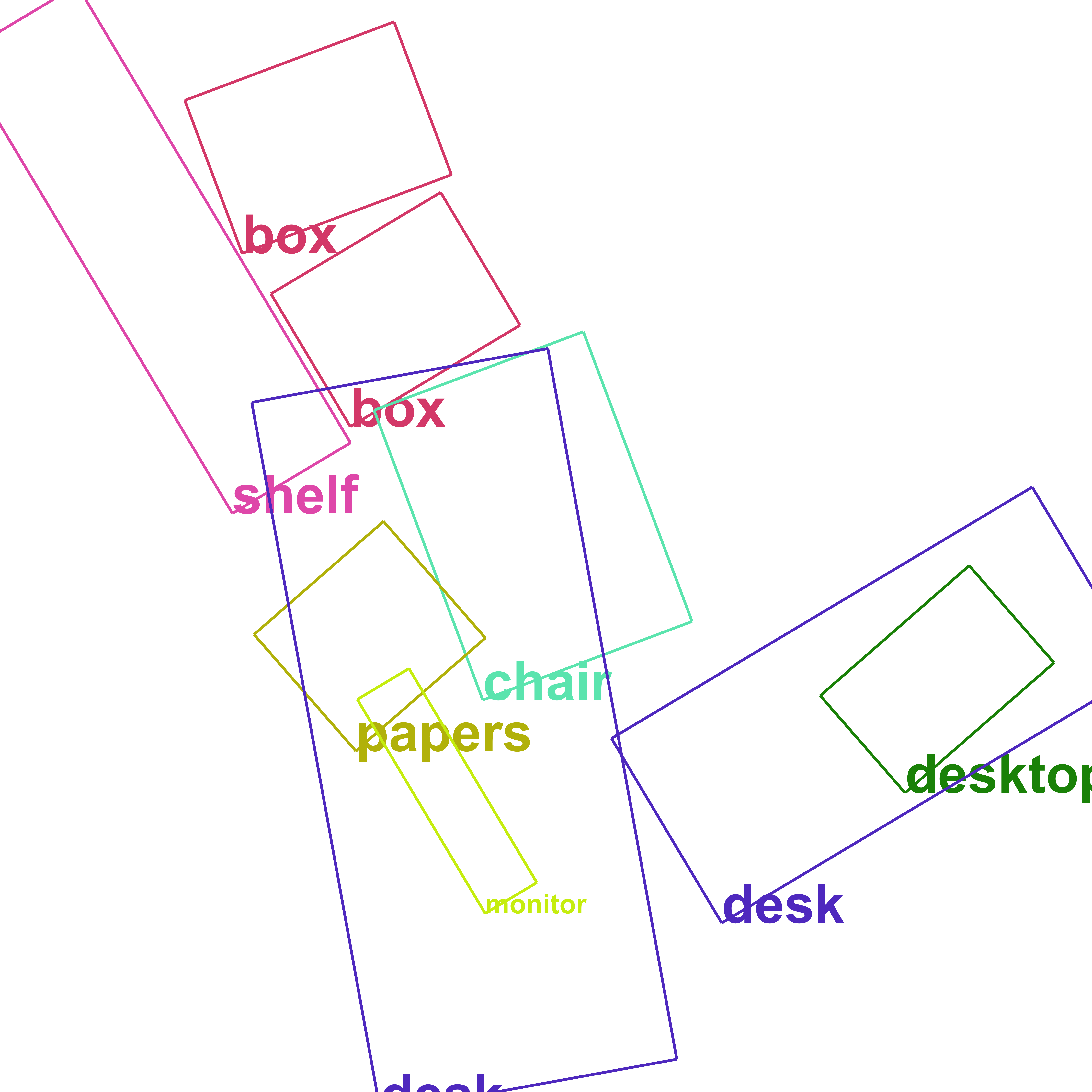}}
    \\
    Document & UI design & Natural Scene & Room layout
    \end{tabular}
    
    \caption{Given a random vector $z$, our novel transformer VAE model produces layouts that follow the design constraints of the training data. It can generate various layouts types, from documents to objects and scenes.}
    \label{fig:teaser}
\vspace{-5mm}
\end{figure}

Prior work has built the foundation by proving the effectiveness of deep learning to generate novel documents \cite{2020-Lee-NDNGLGWC,DBLP:conf/cvpr/PatilBPA20,2020arXiv200614615G}, natural scenes \cite{DBLP:conf/iccv/JyothiDHSM19} and \ac{ui} designs \cite{2020-Lee-NDNGLGWC}.
Mostly, the location and size of a given element depends not only on the particularities of its type (\eg titles tend to be small and at the top of a document, while figures or tables usually occupy a significant amount of space), but also on their \textit{relationship} to other elements. One way to incorporate this knowledge into modeling a layout distribution is to define handcrafted rules, (\eg enforcing margins, alignment, the allowed number of elements in a document\dots). However, such rules are subjective, hard to define unambiguously and certainly do not generalize to arbitrary layout distributions. Consequently, we refrain from modeling any prior knowledge by \ie enforcing heuristics, and instead equip the neural architecture itself with an inherent bias towards learning the relationship between elements in a layout. This makes the attention mechanism a suitable fundamental architectural component, since it naturally models many-to-many relationships and is, thus, particularly suitable for discovering relationships in a given layout distribution in an unsupervised manner.

By instantiating the \ac{vae} framework with an attention-based architecture, this work investigates an important gap in literature. We explore relevant design choices in great detail - \eg autoregressive vs. non-autoregressive decoder, learned vs. non-learned prior. Furthermore, we tailor our novel approach to the yet under-explored task of layout generation, where we demonstrate state-of-the-art performance across various metrics on several publicly available datasets.
To summarize, our main contributions are:
\begin{itemize}
    \item A novel generative model specialized in layout generation that incorporates an inductive bias towards high-level relationships between a large number of elements in a layout without annotations.
    \item Exploration of strategies for creating a variational bottleneck on sequences with varying lengths.
\end{itemize}
\section{Related work}

\paragraph{Layout synthesis}
\label{ssec:layout_synthesis}
The task of layout synthesis has not yet been exhaustively covered by literature, but fueled increasing interest in the research community in recent years. \textbf{LayoutGAN} \cite{DBLP:conf/iclr/LiYHZX19} is, to the best of our knowledge, the first paper to apply generative models (in particular \acp{gan}) to this task. The authors use a generator network to synthesize bounding box annotations. In order to use a \ac{cnn} as discriminator, LayoutGAN applies a novel differential render module to turn a collection of bounding boxes into an image.
Similarly to our approach, it uses self-attention to  model many-to-many relationships. However, the authors only evaluate single-column documents with at most nine elements, which corresponds to much sparser layouts than provided by common publicly available datasets.

\textbf{LayoutVAE} \cite{DBLP:conf/iccv/JyothiDHSM19} proposes an autoregressive model based on a conditional \ac{vae} with a conditional prior (conditioned on the number and type of elements in the layout). The authors use an \ac{lstm} \cite{hochreiter_lstm} to aggregate information over time. Additionally they propose using a second conditional \ac{vae} to model the distribution of category counts which is used as conditional information during layout generation. Their underlying neural architecture is comprised of fully-connected layers and \acp{lstm}. Consequently, it is expected that LayoutVAE struggles to model layouts with a large number of elements, since \acp{lstm} do not explicitly model the relationships of all components. Unlike LayoutVAE, our work explicitly biases the underlying neural network towards learning the relationships between elements in a layout, and only makes the decoder autoregressive (reducing the computational costs). Further, we only train a single \ac{vae} for learning the layout distribution instead of resorting to two separate \acp{vae}. 

In \textbf{Neural Design Networks} the authors \cite{2020-Lee-NDNGLGWC} generate document layouts with an optional set of design constraints. Initially, a complete graph for modeling the relationships between elements is built. The distribution of these relationships is learned using a \ac{vae} based on \acp{gcn}, where the labels of the relationships are based on heuristics. The actual layout is subsequently generated by a separate \ac{gcn}. The resulting raw layout is then polished by an additional refinement network. In contrast to Neural Design Networks, this work does not rely on labels extracted using heuristics on the training data for learning a layout distribution, which is prone to introduce ambiguities and unlikely to generalize across datasets. Moreover, our approach learns the layout distribution end-to-end without relying on training three separate neural networks.

Similarly, \textbf{READ} \cite{DBLP:conf/cvpr/PatilBPA20} also uses heuristics to determine the relationships between elements and then trains a \ac{vae} which is based on \acp{rvnn} \cite{DBLP:conf/icnn/GollerK96} to learn the layout distribution.

\textbf{Content-aware Generative Modeling of Graphic Design Layouts} \cite{zheng-sig19} trains a VAEGAN conditioned on images, keywords, attributes of the layout and corresponding coordinates. However, the authors focus on learning the layout distribution \textit{conditioned} on additional user input.

\textbf{Layout Generation and Completion with Self-attention} \cite{2020arXiv200614615G} is most relevant to this work. The authors perform self-supervised training (\ie layout completion) using an autoregressive decoder motivated by Transformers \cite{DBLP:conf/nips/VaswaniSPUJGKP17}. Subsequently, novel layouts are synthesized using beam search \cite{DBLP:journals/ai/MedressCFGKONNRRSWW77}. While this generation approach can yield strong results, it requires optimizing additional hyperparameters (\eg beam size) and, more importantly, it does not have any theoretical guarantees for learning the actual data distribution. The resulting distribution rather depends on finding the right level of regularization at training time. Only if the model is regularized appropriately beam search will yield outcomes of sufficient diversity. Since this generation process lacks theoretical guarantees for capturing the full diversity of the layout distribution and heavily relies on heuristics, we directly approximate the distribution using a attention-based \ac{vae} instead. 

Some works have been proposed with particular focus on furniture arrangement \cite{wang2018deep, 2017arXiv171110939H}. In the method of Wang \etal \cite{wang2018deep}, one \ac{cnn} places elements in a room by estimating the likelihood of each possible location, while a second \ac{cnn} determines when the scene is complete. \cite{DBLP:conf/cvpr/Ritchie0L19} extends this to model orientations and room dimensions. Moreover, Henderson \etal \cite{2017arXiv171110939H} propose to learn a distribution for each element type and model high-order relationships between objects using a direct acyclical graph. Since all of these methods use the now unavailable SUNCG dataset \cite{DBLP:conf/cvpr/SongYZCSF17} for training, establishing a comparison with them is difficult.

Additionally, tab. \ref{tab:model_comparison} provides a high-level comparison between this work and the most relevant adjacent methods. We differentiate existing works along four important dimensions: 1) Are models equipped with inductive biases towards learning the relationships between elements? 2) Are these relationships learned without supervision or are additional labels, using \eg heuristics, necessary? 3) Can layouts contain an arbitrary number of elements? 4) Does the learning approach provide guarantees for learning the underlying distribution by applying probabilistic methods?

\textbf{Attention-based \acp{vae}} are a recent development in the \ac{nlp} literature. The common goal is to learn the distribution of real data more accurately than with deterministic self-supervised approaches \cite{DBLP:journals/corr/abs-2003-12738, DBLP:conf/ijcnn/LiuL19a, DBLP:conf/ijcai/Wang019b}. To combine Transformers and \acp{vae} \cite{DBLP:conf/ijcnn/LiuL19a} uses self-attention layers for the encoder and decoder components. The encoder turns a sentence into a collection of high-dimensional vectors of the same length as the input. These constitute the \ac{vae} bottleneck, and are passed after re-parameterization to the decoder to reconstruct the sentence. By feeding a set of vectors sampled from the prior, a sentence of the same length can be generated. Further, \cite{DBLP:journals/corr/abs-2003-12738} implements a conditional \ac{vae} (conditioned on the context of a conversation) based on the Transformer to improve diversity on the task of response generation. \cite{DBLP:conf/ijcai/Wang019b} develops a Transformer-based \ac{vae} to enhance variability on the task of story completion. Their encoder and decoder share weights while the bottleneck of their \ac{vae} is fed into the penultimate layer of the decoder.

\begin{table}
\setlength{\tabcolsep}{0.07cm}
\begin{tabular}{lP{1.20cm}P{1.62cm}P{1.20cm}P{1.47cm}}
 & \small{Inductive Bias} & \small{Unsupervised Relationship} & \small{Arbitrary Size} & \small{Distribution Learning}\\\hline
LayoutGAN \cite{DBLP:conf/iclr/LiYHZX19} &  \ding{51}  & \ding{51} & \ding{55} & \ding{51} \\
LayoutVAE \cite{DBLP:conf/iccv/JyothiDHSM19} &  \ding{55} & \ding{51} & \small{Practically difficult} & \ding{51} \\
READ \cite{DBLP:conf/cvpr/PatilBPA20} & \ding{51} & \ding{55} & \ding{51} & \ding{51} \\
NDN \cite{2020-Lee-NDNGLGWC} & \ding{51} & \ding{55} & \ding{55} & \ding{51} \\
Gupta \etal \cite{2020arXiv200614615G} & \ding{51} & \ding{51} & \ding{51} & \ding{55} \\
Ours & \ding{51} & \ding{51} & \ding{51} & \ding{51}\\\bottomrule
\end{tabular}
\caption{Comparison with existing methods. We consider whether methods 1) equip their models with inductive biases towards learning the relationships between elements, 2) learn relationships unsupervised, 3) allow layouts of arbitrary size  and 4) have guarantees for learning the underlying distribution by applying probabilistic methods.}\label{tab:model_comparison}
\end{table}

\section{Variational Transformer Networks}

This section illustrates the proposed  \aclp{vtn}. From a high-level perspective \acp{vtn} are an instance of the \ac{vae} framework tailored to the task of layout synthesis, where the main building blocks of the neural networks parameterizing the encoder and decoder are attention layers. Firstly, we briefly revisit the concept of \acp{vae}. Subsequently, we explain how \acp{vtn} exploit the data format of layouts, their architecture and how to train them.

\subsection{\aclp{vae}}

\acp{vae} are a family of latent variable models that approximate a data distribution $P(X)$ by maximizing the \ac{elbo} \cite{DBLP:journals/corr/KingmaW13}
\begin{align} \label{eq:elbo}
    \mathcal{L}(\theta, \phi) =\mathop{\mathbb{E}}_{z\sim q_{\theta}\left(z|x\right)}\left[ \log\left(p_{\phi}\left(x|z\right)\right) \right]
    - KL\left( q_{\theta}\left(z|x\right)|| p\left(z\right) \right)
\end{align}
where $p_{\phi}(x|z)$ denotes a decoder parameterized by a neural network with parameter $\phi$, $q_{\theta}(z|x)$ is the approximate posterior distribution, similarly parameterized by a neural network with weights $\theta$, and $p(z)$ the prior distribution.

\subsection{Exploiting the Data Format of Layouts}
The central aspect of layout generation is its unique underlying data format. Layouts are sets of elements of variable size, where each element can be described by both discrete and continuous features. More formally, each layout $\textbf{x}$ in a given dataset $\textbf{X}$ consists of a variable number $l$ of bounding boxes. Further, each bounding box $x_i$ with $i\in [1, \dots, l]$ contains information about its class (for documents, \eg text, image\dots), location and dimension.

Another important characteristic of layout datasets is that there exists a high degree of correlation between the individual elements in a layout. For example, in case of document layouts, titles tend to be positioned at the top of a text. It is therefore essential to bias an approach for learning layout distributions towards exploiting the relationships between elements. While some methods introduce additional features, such as annotations for the relationships between elements \cite{2020-Lee-NDNGLGWC,DBLP:conf/cvpr/PatilBPA20}, our approach instead relies solely on bounding box annotations, since additional features are expensive to create, prone to ambiguity and fail to generalize across datasets. Therefore, we introduce an inductive bias to learn from the relationships by using an attention-based neural network. Notably, the attention mechanism is an ideal candidate for exploiting pairwise relationships, since it leverages the pairwise dot product as its fundamental computational block for guiding the information flow.

Moreover, the attention mechanism also helps modeling another aspect of the data - namely a varying and large number of elements. To mitigate this problem other works have restricted the maximum number of elements occurring in one layout \cite{DBLP:conf/iclr/LiYHZX19, 2020-Lee-NDNGLGWC}. However, attention-based architectures are well suited for learning the relationships of a large number of elements, since this is one of the reasons for their success in the \ac{nlp} literature \cite{DBLP:conf/nips/VaswaniSPUJGKP17}. Notably, \acp{rnn}, as used by LayoutVAE \cite{DBLP:conf/iccv/JyothiDHSM19}, are also capable of modeling a varying number of elements in a layout. However, they struggle with long-term dependencies, \ie a large number of elements in a layout. This follows from results in the \ac{nlp} literature and is also observed by us (see section \ref{sec:qualitative_results}).

\begin{figure*}
\centering
\includegraphics[width=0.95\linewidth]{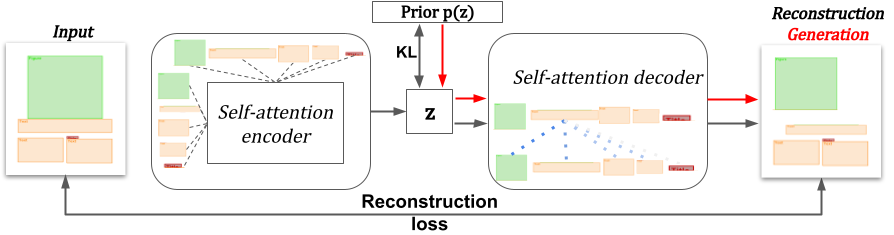}
\caption{\ac{vtn}. The encoder and decoder are parameterized by attention-based neural networks. This biases the network to learn relationships between the layout elements and enables processing layouts of arbitrary size. During training (black arrow) the reconstruction loss and the KL-divergence between the prior $p(z)$ and the approximate posterior distribution are minimized. During inference (red arrow) we sample latent representations $z$ from the prior and transform those into layouts using our self-attention-based decoder.} \label{fig:architecture}
\vspace{-5mm}
\end{figure*}

\subsection{Architecture of \acp{vtn}}

The architecture of \acp{vtn} is based on Transformers  \cite{DBLP:conf/nips/VaswaniSPUJGKP17}, which are sequence models that consist of an encoder-decoder architecture, where both encoder and decoder use attention layers as their fundamental building blocks. We refer to fig. \ref{fig:architecture} for a schematic overview of our approach. 

The encoder of the Transformer architecture parameterizes the posterior distribution $q_{\theta}(z|x)$  in the \ac{vae} framework. In particular, $q_{\theta}(z|x)$ is parameterized as a multivariate normal distribution with diagonal covariance matrix, whose parameters are determined by the output of the encoder network. To train the encoder using backpropagation, we apply the local re-parameterization trick \cite{kingma2015variational}. The original Transformer is a highly specialized language model, which is usually trained on vast quantities of text data. Therefore, it is necessary to adjust the hyper-parameters. It is essential to keep the number of attention heads large (here $n_\text{heads}=8$) to average out outliers from individual attention heads \cite{DBLP:conf/nips/VaswaniSPUJGKP17}. Similarly, we keep a large model dimensionality ($d_\text{model}=512$) and size of the point-wise feed-forward layers ($d_{ff}=2048$). However, we find that the number of attention-blocks (see  \cite{DBLP:conf/nips/VaswaniSPUJGKP17}) can be reduced to four without performance loss. This hints that relationships between elements in a layout are less complex than between words in language. We further omit the positional encodings used in the context of \ac{nlp} since the features of bounding boxes already contain positional information.

The decoder $p_{\phi}(x|z)$ in \acp{vtn} is a mirrored version of the encoder. Note that this breaks with \cite{DBLP:conf/nips/VaswaniSPUJGKP17} which adds additional attention-layers whose keys and queries are the output of the encoder. We empirically find that feeding the output of the encoder as an input to the first layer of the decoder yields better results. We further experiment with another major architectural choice: the autoregressive decoder, \ie $p_{\phi}(x|z) = \prod_{i=1}^l p_{\phi}(x_i|x_{i-1}, z)$, where $l$ denotes the number of bounding boxes in a layout, and a non-autoregressive variant. While the former has more representational power, since theoretically any distribution can be modeled as an autoregressive one, it is also more prone to posterior collapse due to the expressive decoder \cite{DBLP:conf/conll/BowmanVVDJB16} and requires more computational resources.

Furthermore, we consider two distinct prior distributions. First, we use the common choice of a fixed multivariate zero-mean normal distribution. However, this often proves too restrictive for learning the true posterior distribution \cite{chen2016variational}. In principle there are two avenues to mitigate this issue: use a more expressive parameterization of the posterior \cite{kingma2016improved} or the prior distribution \cite{chen2016variational}. In this work we attempt to extend the expressiveness of the prior distribution by learning the parameters of the multivariate normal distribution with a diagonal covariance matrix. Since layouts consist of a varying number of bounding boxes, we parameterize the distribution with an \ac{lstm} \cite{hochreiter1997long}. 

Importantly, while an autoregressive decoder enables sampling of layouts with varying number of elements - \eg by introducing symbols for start/end of the layout - the non-autoregressive decoder requires incorporating this into the prior distribution. Therefore, we model the prior in this case as $p(z, s) = p(z|s)p(s)$ where $s$ denotes the number of bounding boxes. We learn $p(s)$ during training by counting the number of occurrences of each sequence length. 

Finally, we note that in the case of the autoregressive decoder, we find empirically that aggregating the latent representations $z$ across all elements in a layout yields better perceptual quality. This corresponds to parameterizing the posterior distribution with the output of the encoder aggregated along the dimension of the layout elements. To this end we follow BERT \cite{DBLP:conf/naacl/DevlinCLT19} where the final hidden state of the encoder for the first token is used to represent the entire sequence, and used as the first element in the the decoder input. In the case of the non-autoregressive decoder we do not aggregate the latent representations, but feed them directly with variable dimensionality to the decoder.

\subsection{Optimizing \acp{vtn}}

Since we are learning the layout distribution using a \ac{vae}, we optimize the \ac{elbo} defined in eq. \eqref{eq:elbo}. However, a practical optimization challenge of \acp{vae} is the so-called posterior collapse \cite{DBLP:conf/conll/BowmanVVDJB16, he2019lagging}. The decoder ignores the information in the latent representation and collapses onto modes of the data distribution. At the same time the posterior distribution parameterized by the encoder can perfectly match the prior distribution, since it does not need to transmit information to the decoder. Therefore, this work follows a common heuristic by optimizing the $\beta$-\ac{vae} objective instead of eq. \eqref{eq:elbo}
\begin{align} \label{eq:betaelbo}
    \mathcal{L}(\theta, \phi) = \mathop{\mathbb{E}}_{z\sim q_{\theta}\left(z|x\right)}\left[ \log(p_{\phi}(x|z)) \right] - \beta KL\left( q_{\theta}(z|x)|| p(z) \right)
\end{align}
To optimize eq. \ref{eq:betaelbo} we use Adam \cite{DBLP:journals/corr/KingmaB14} and follow the learning rate schedule in \cite{DBLP:conf/nips/VaswaniSPUJGKP17}. To further reduce the risk of posterior collapse, it is common to increase $\beta$ at the beginning of training from zero to the desired value. Specifically, we implement the exponential beta schedule proposed by \cite{DBLP:conf/conll/BowmanVVDJB16,DBLP:conf/ijcnn/LiuL19a}. In all our experiments we use $\beta = 1$ with the autoregressive decoder and $\beta = 0.5$ with the non-autoregressive decoder.

Moreover, we follow \cite{2020arXiv200614615G} in discretizing the location, width and height of the bounding boxes. Thus each bounding box is represented by a feature vector containing a one-hot encoding of the class concatenated with the one-hot encodings representing the above discretization. We use categorical cross-entropy as a reconstruction loss.

\paragraph{Implementation Details}
We implement our method using Tensorflow 2 \cite{tensorflow2015-whitepaper} and a NVIDIA V100 GPU for acceleration. We train using the Adam optimizer with a batch size of 64 for 30 epochs in the case of the autoregressive decoder and 50 epochs using the non-autoregressive version.

\section{Experiments}

\subsection{Datasets}

We evaluate our method on the following publicly available datasets of layouts for documents, natural scenes, furniture arrangements and mobile phone \acp{ui}. \newline
\textbf{PubLayNet} \cite{zhong2019publaynet} contains 330K samples of machine-annotated scientific documents crawled from the Internet. It has the categories \textit{text}, \textit{title}, \textit{figure}, \textit{list}, \textit{table}. \newline
\textbf{RICO} \cite{Deka:2017:Rico} is a dataset of user interface designs for mobile applications. It contains 91K entries with 27 element categories (\textit{button}, \textit{toolbar}, \textit{list item}\dots). Due to memory constraints we omit layouts with more than 100 elements\footnote{Note that this restriction originates from \textit{memory constraints} and does not imply that our approach is not capable of learning larger layouts given sufficient memory.}, in total removing $0.031\%$ of the data. \newline
\textbf{COCO} \cite{DBLP:conf/eccv/LinMBHPRDZ14} Contains $\sim$100K images of natural scenes. We use the \textit{Stuff} variant, which contains 80 \textit{thing} and 91 \textit{stuff} categories, removing small bounding boxes ($\leq 2\%$ image area), as well as instances that are tagged as ``\textit{iscrowd}''.\newline
\textbf{SUN RGB-D}\cite{DBLP:conf/cvpr/SongLX15} is a scene understanding dataset with 10000 samples, including scenes from \cite{DBLP:conf/eccv/SilbermanHKF12}, \cite{DBLP:conf/iccvw/JanochKJBFSD11} and \cite{DBLP:conf/iccv/XiaoOT13}. The annotations comprise different household objects. We compute the 2D bounding boxes of the semantic regions from a top-down perspective.

\subsection{Evaluation methodology}

It is important to evaluate layouts along two high-level dimensions - perceptual quality and diversity. Note that in the case of layouts perceptual quality is prone to subjectivity and different aspects must be considered from dataset to dataset. It is thus difficult to define a single metric that entirely covers both aspects. We therefore resort to a set of metrics where each aims at representing an individual aspect of either perceptual quality or diversity. 

\textbf{Alignment and overlap.} Some datasets, such as PubLayNet or RICO, consist of entries with strictly defined alignments and small overlaps between bounding boxes. Consequently, these properties are an indicator of the perceptual quality of synthesized layouts. We 
follow LayoutGAN \cite{DBLP:conf/iclr/LiYHZX19} in  measuring overlaps using the total overlapping area among any two bounding boxes inside the whole page (\textit{overlap index}) and the average \ac{iou} between elements.
Additionally, we quantify alignment using the alignment loss proposed by \cite{2020-Lee-NDNGLGWC}.

\textbf{Unique matches under DocSim metric.} We use the number of unique matches between real sets of layouts and synthesized layouts as a proxy for diversity. We use the DocSim metric \cite{DBLP:conf/cvpr/PatilBPA20} as a similarity metric. Note that, while the number of unique matches primarily analyzes diversity, it also partially reflects perceptual quality.

\textbf{Wasserstein distance.} A rigorous approach to evaluate diversity would be computing the Wasserstein distance between the real and learned data distributions. Unfortunately this is infeasible. However, we can approximate the Wasserstein distance between real and generated data for two marginal distributions - the class distribution (discrete) and the bounding box distribution (continuous, 4-d vectors ($x_\text{center}$, $y_\text{center}$, \textit{width}, \textit{height})). In practice, we compute these Wasserstein distances from a finite set of samples.

\subsection{Quantitative results}\label{sec:quantitative_results}
\paragraph{Comparison to state of the art} A comparison to any of the methods described in section \ref{ssec:layout_synthesis} is difficult, since, to the best of our knowledge, none has a publicly available implementation\footnote{Though the LayoutGAN authors recently released an implementation, they only did so for a toy example on MNIST: \href{https://github.com/JiananLi2016/LayoutGAN-Tensorflow}{https://github.com/JiananLi2016/LayoutGAN-Tensorflow}}. Similar to the authors of LayoutVAE \cite{DBLP:conf/iccv/JyothiDHSM19}, we were unable to reproduce the results of LayoutGAN \cite{DBLP:conf/iclr/LiYHZX19} on documents. We reimplement LayoutVAE and the approach of Gupta \etal \cite{2020arXiv200614615G}.
In the LayoutVAE case, we follow \cite{2020arXiv200614615G} and sample category counts from the test dataset.
For Gupta \etal, we use a mixture of nucleus sampling with $p=0.9$ and top-$k$ sampling with $k=30$. As suggested by the authors, we found nucleus sampling to improve the diversity of the synthesized layouts. We, further, compare against NDN \cite{DBLP:conf/cvpr/PatilBPA20} on RICO using their proposed alignment metric.

In tab. \ref{tab:publaynet_ablation} we ablate our model on the prior type and the decoding strategy. We observe that, while the autoregressive decoder slightly decreases diversity (Wasserstein distance class/bounding box and number of unique matches), it yields large improvements regarding perceptual quality (IoU, overlap index and alignment). Moreover, in the case of the non-autoregressive decoder a learned prior yields improvements regarding perceptual quality. However, when using an autoregressive decoder the learned and non-learned priors yield similar results. Therefore, we apply an autoregressive decoder with a non-learned prior in the remaining experiments, since it strikes the optimal balance between diversity, perceptual quality and model simplicity.

We report quantitative results for the aforementioned metrics on different datasets. Unless explicitly stated, all metrics are computed on 1000 samples, and the value is averaged across 5 trainings with different random initialization.
In tabs. \ref{tab:publaynet} and \ref{tab:tab2_extended} we show the results of our method in comparison to existing art.
We show that our method produces a large number of distinct layouts that have similar alignment metrics as the real data. Furthermore, we clearly outperform LayoutVAE \cite{DBLP:conf/iccv/JyothiDHSM19} across all metrics and demonstrate improved diversity at similar perceptual quality compared to \cite{2020arXiv200614615G}, as expected since our method explicitly approximates the layout distribution. Given that both LayoutVAE and Gupta \etal generate layouts autoregressively and considering our ablation in tab. \ref{tab:publaynet_ablation}, we note that autoregressive modeling denotes an important element of learning layout distributions. 

Furhtermore, in tab. \ref{tab:alignment} we also compare our approach to Lee \etal \cite{2020-Lee-NDNGLGWC} on the RICO dataset using their proposed alignment metric. We demonstrate superior results when no explicit design constraints are given (NDN-none), showing that our method is better at discovering relationships without supervision. Even in the NDN-all case, where all relationships are given to the network, we show similar performance despite not relying on this information.

\begin{table*}[t]
\centering
\begin{tabular}{lcccccc}
\toprule
 & IoU & Overlap & Alignment & W class $\downarrow$ & W bbox $\downarrow$ & \# unique matches $\uparrow$ \\

LayoutVAE \cite{DBLP:conf/iccv/JyothiDHSM19} & 0.171 &0.321 & 0.472 & - & 0.045 & 241\\

Gupta \etal \cite{2020arXiv200614615G} & 0.039 & 0.006 & 0.361 & \textbf{0.018} & 0.012 & 546 \\

Ours (autoregressive) & 0.031 & 0.017 & 0.347 & 0.022 & 0.012 & \textbf{697} \\

 &  &  &  &  &  & \\ \hline
Real data & 0.048 & 0.007 & 0.353 & - & - & - \\ \bottomrule
\end{tabular}
\caption{Quantitative evaluation on PubLayNet. We generate 1000 layouts with each method and compare them regarding average \ac{iou}, overlap index \cite{DBLP:conf/iclr/LiYHZX19}, alignment \cite{2020-Lee-NDNGLGWC}, Wasserstein (W) distance of the classes and bounding boxes to the real data and the number of unique matches according to the DocSim. Ours (autoregressive) denotes using an autoregressive decoder.}
\label{tab:publaynet}
\end{table*}

\begin{table*}[t]
\setlength{\tabcolsep}{5.4pt}
\centering
\begin{tabular}{P{1.75cm}P{1cm}|lllllP{1.6cm}}
\toprule
 Autoregressive decoder & Learned prior & IoU $\downarrow$ & Overlap $\downarrow$ & Alignment $\downarrow$ & W class $\downarrow$ & W bbox $\downarrow$ & \# unique matches $\uparrow$ \\
\ding{55} & \ding{55} & 0.259 $\pm$ 0.114 & 0.178 $\pm$ 0.122 & 0.364 $\pm$ 0.080 & \textbf{0.011 $\pm$ 0.007} & 0.018 $\pm$ 0.012 & \textbf{813 $\pm$ 51} \\
\ding{55} & \ding{51} & 0.243 $\pm$ 0.027 & 0.097 $\pm$ 0.040 & 0.381 $\pm$ 0.010 & 0.013 $\pm$ 0.007 & \textbf{0.011 $\pm$ 0.001} & 794 $\pm$ 34  \\
\ding{51} & \ding{55} & \textbf{0.031 $\pm$ 0.004} & 0.017 $\pm$ 0.006 & \textbf{0.347 $\pm$ 0.005} & 0.022 $\pm$ 0.002 & 0.012 $\pm$ 0.001 & 697 $\pm$ 13\\
\ding{51} & \ding{51} & 0.032 $\pm$ 0.002 & \textbf{0.015 $\pm$ 0.004} & 0.353 $\pm$ 0.004 & 0.022 $\pm$ 0.005 & 0.013 $\pm$ 0.001 & 677 $\pm$ 16\\ \bottomrule
\end{tabular}
\caption{Quantitative ablation study on PubLayNet. We generate 1000 layouts and compare them in average \ac{iou}, overlap index, alignment, Wasserstein (W) distance of the classes and bounding boxes to the real data and the number of unique matches according to DocSim. We compare our model w/wo autoregressive decoder and with learned/non-learned prior.}
\label{tab:publaynet_ablation}
\vspace{-3mm}
\end{table*}

\begin{table}[b]
\setlength{\tabcolsep}{1pt}
\centering
\resizebox{\columnwidth}{!}{
\begin{tabular}{l|cccccc}
\toprule
\multicolumn{7}{c}{\small RICO}\\\hline
 & IoU & Overlap & Alignment & $\text{W}_{\text{class}}$ $\downarrow$ & $\text{W}_{\text{bbox}}$ $\downarrow$ & \# unique m. $\uparrow$ \\

\cite{DBLP:conf/iccv/JyothiDHSM19} & 0.193 & 0.400 & 0.416 & - & 0.045 & 496\\

\cite{2020arXiv200614615G} & 0.086 & 0.145 & 0.366 & \textbf{0.004} & 0.023 & 604 \\

Ours & 0.115 & 0.165 & 0.373 & 0.007 & \textbf{0.018} & \textbf{680} \\
\hline
Real & 0.084 & 0.175 & 0.410 & - & - & - \\ \bottomrule
\multicolumn{7}{c}{\small COCO}\\
\toprule

\cite{DBLP:conf/iccv/JyothiDHSM19} & 0.325 & 2.819 & 0.246 & - & 0.062 & 700 \\

\cite{2020arXiv200614615G} & 0.194 & 1.709 & 0.334 & 0.001 & 0.016 & 601 \\

Ours & 0.197 & 2.384 & 0.330 & \textbf{0.0005} & \textbf{0.013} & \textbf{776} \\

\hline
Real & 0.192 & 1.724 & 0.347 & - & - & - \\ \bottomrule
\end{tabular}
}
\caption{Extension of Tab. \ref{tab:publaynet} for RICO, COCO}
\label{tab:tab2_extended}
\end{table}

\begin{table}[t]
    \centering
    \begin{tabular}{c|c}
        Method & Alignment \\\toprule
        NDN-none \cite{2020-Lee-NDNGLGWC} & $0.91\pm0.030$ \\
        NDN-all \cite{2020-Lee-NDNGLGWC} & $0.32\pm0.020$ \\
        Ours & $0.37 \pm 0.009$ \\ \hline
        Real data & 0.0012\\\bottomrule
    \end{tabular}
    \caption{Comparison between Neural Design Network \cite{2020-Lee-NDNGLGWC} and our approach using their proposed alignment metric on RICO.}
    \label{tab:alignment}
\end{table}

\subsection{Qualitative results}\label{sec:qualitative_results}

\begin{figure}[t]
    \newlength{\publaynetSamplesWidth}
    \setlength{\publaynetSamplesWidth}{0.24\linewidth}
    \centering
    \includegraphics[width=\publaynetSamplesWidth,frame=0.1pt]{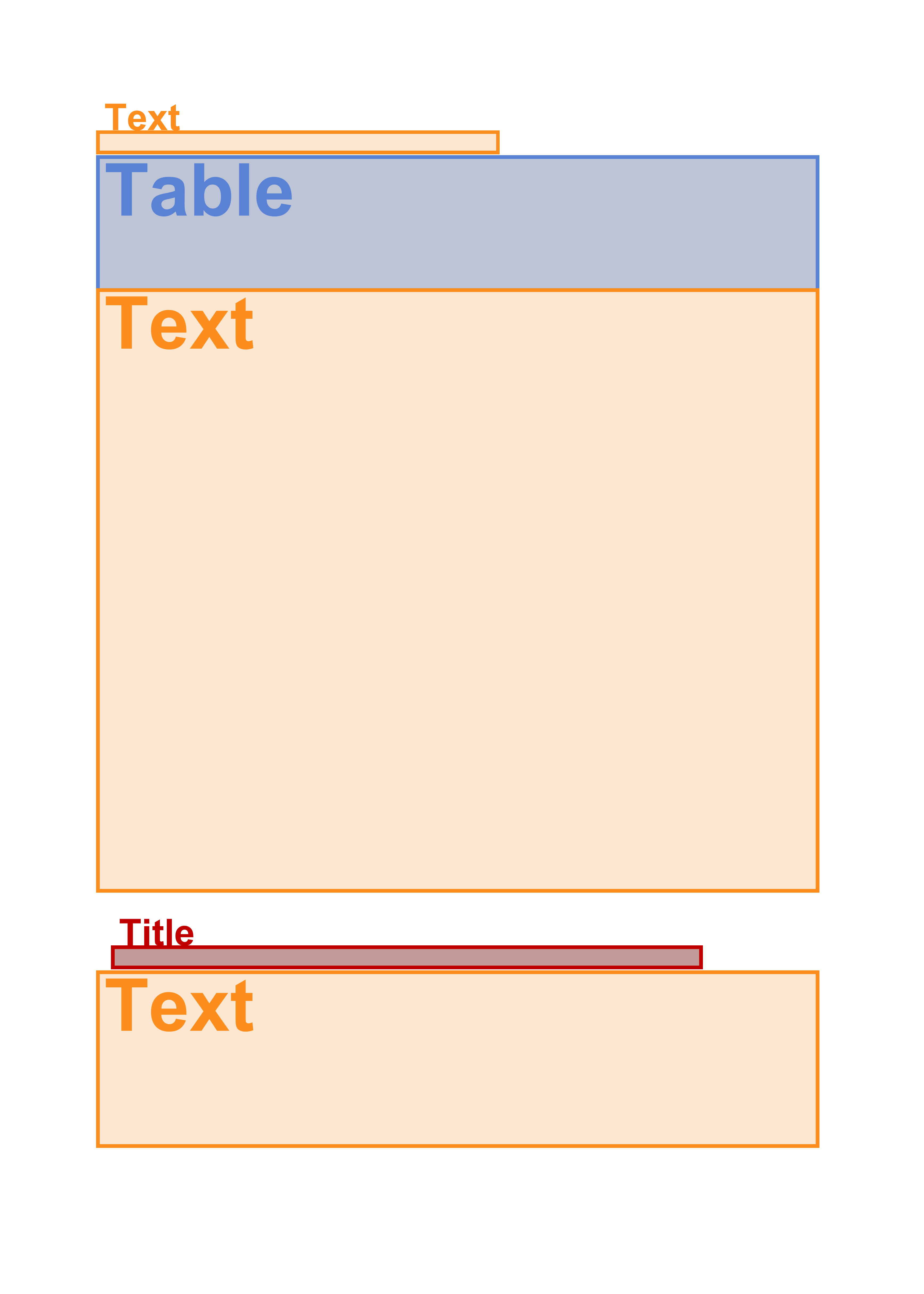}
    \includegraphics[width=\publaynetSamplesWidth,frame=0.1pt]{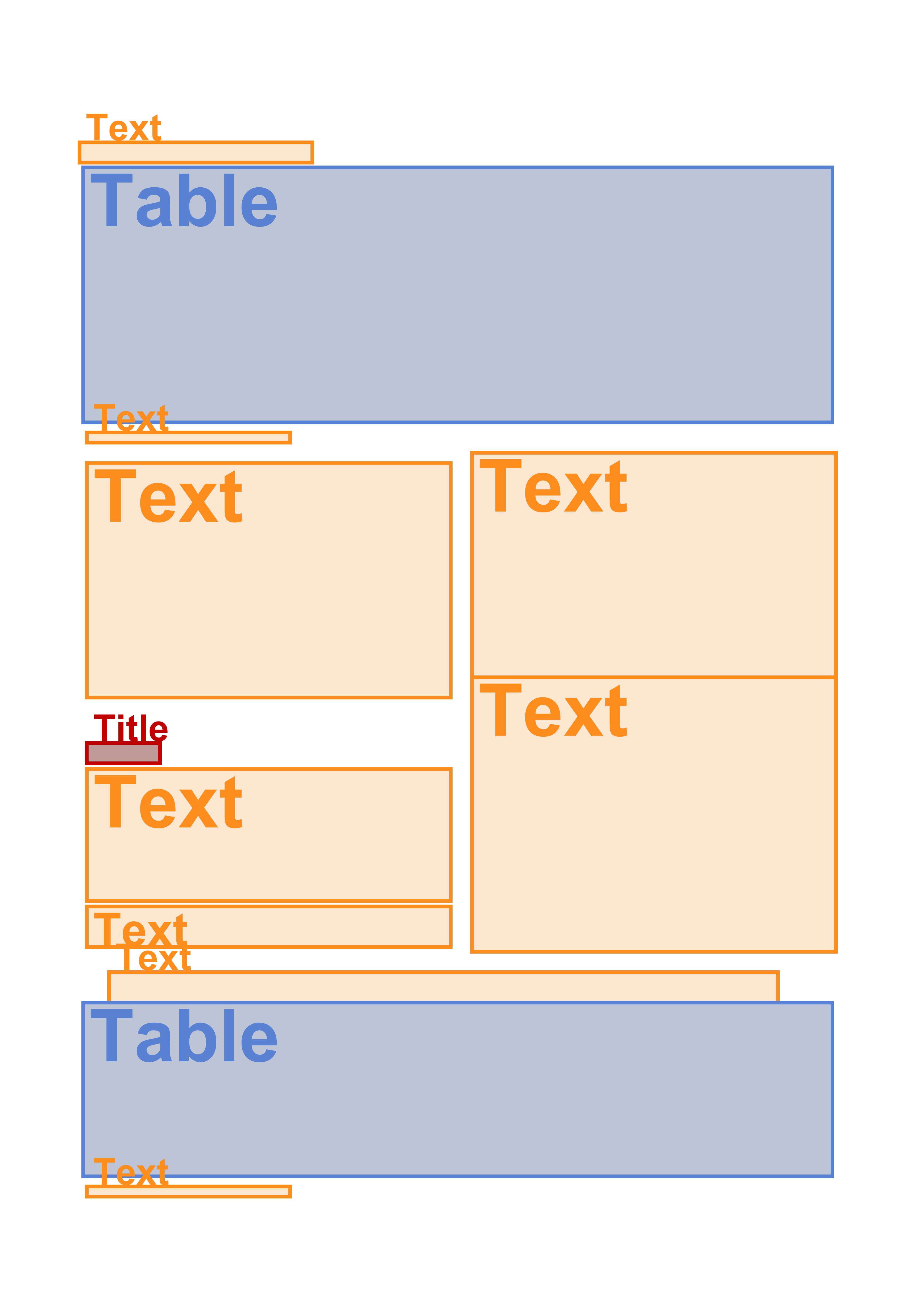}
    \includegraphics[width=\publaynetSamplesWidth,frame=0.1pt]{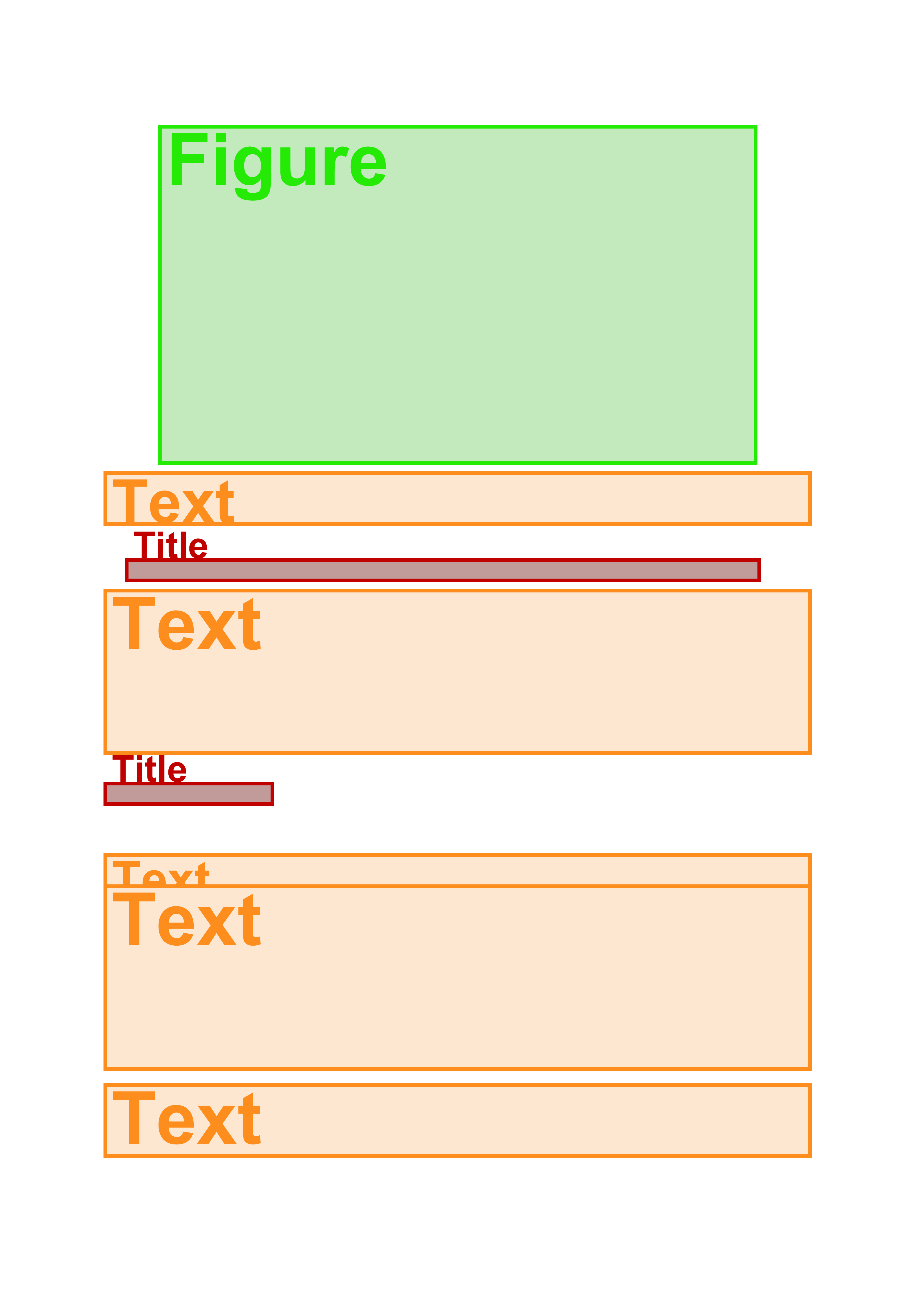}
    \includegraphics[width=\publaynetSamplesWidth,frame=0.1pt]{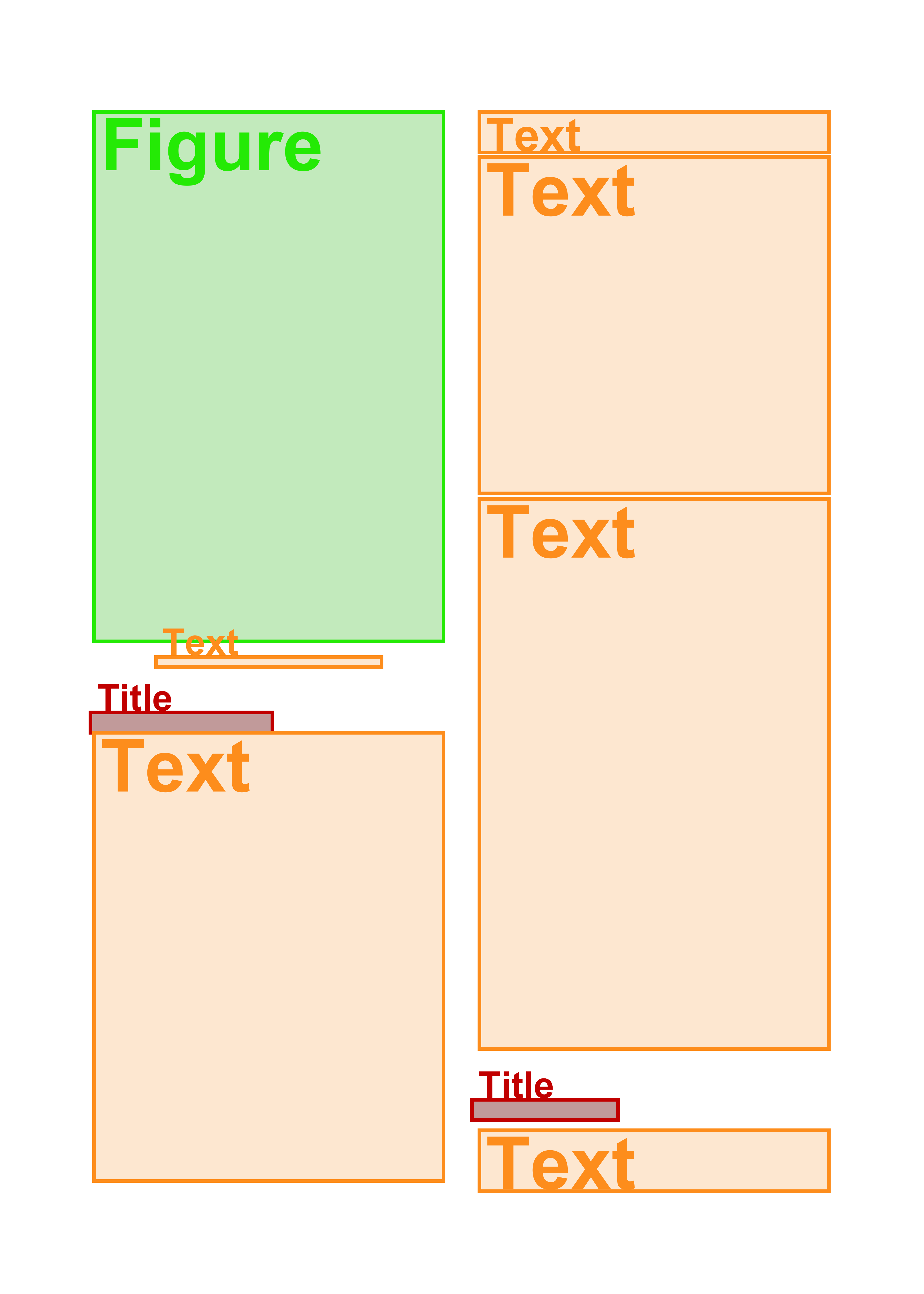}
    \includegraphics[width=\publaynetSamplesWidth,frame=0.1pt]{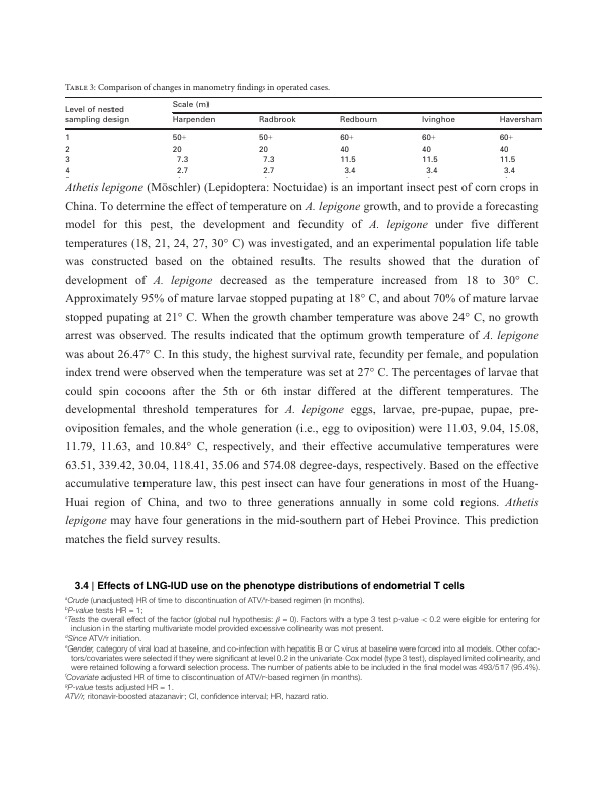}
    \includegraphics[width=\publaynetSamplesWidth,frame=0.1pt]{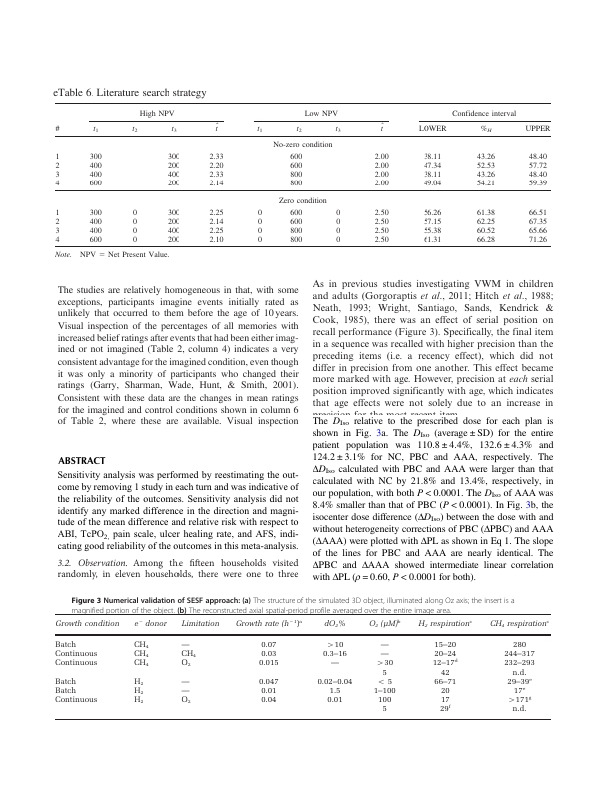}
    \includegraphics[width=\publaynetSamplesWidth,frame=0.1pt]{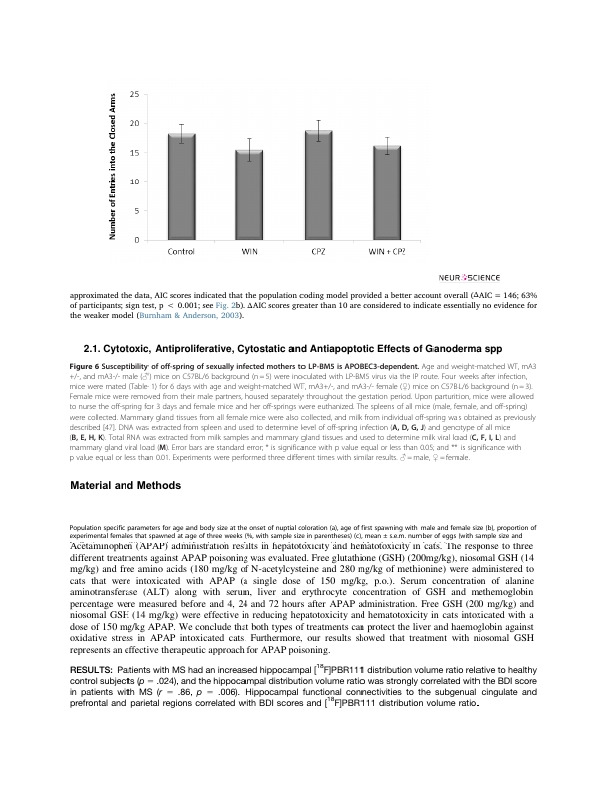}
    \includegraphics[width=\publaynetSamplesWidth,frame=0.1pt]{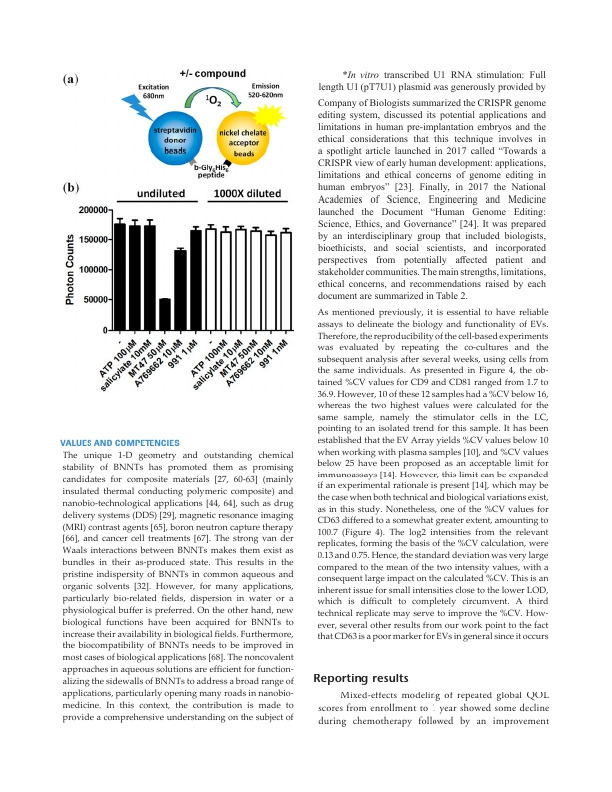}\\
    \caption{Top: Generated layouts from our autoregressive model for PubLayNet. Bottom: Renderings of the layouts. The supplementary material shows more samples.}
    \label{fig:publaynet_samples}
\end{figure}

\begin{figure}
    \setlength{\tabcolsep}{3pt}
    \newlength{\qualitativeComparisonWidth}
    \setlength{\qualitativeComparisonWidth}{0.21\linewidth}
    \centering
    \begin{tabular}{ccccc}
\rotatebox{90}{LayoutVAE \cite{DBLP:conf/iccv/JyothiDHSM19}} &
\includegraphics[width=\qualitativeComparisonWidth]{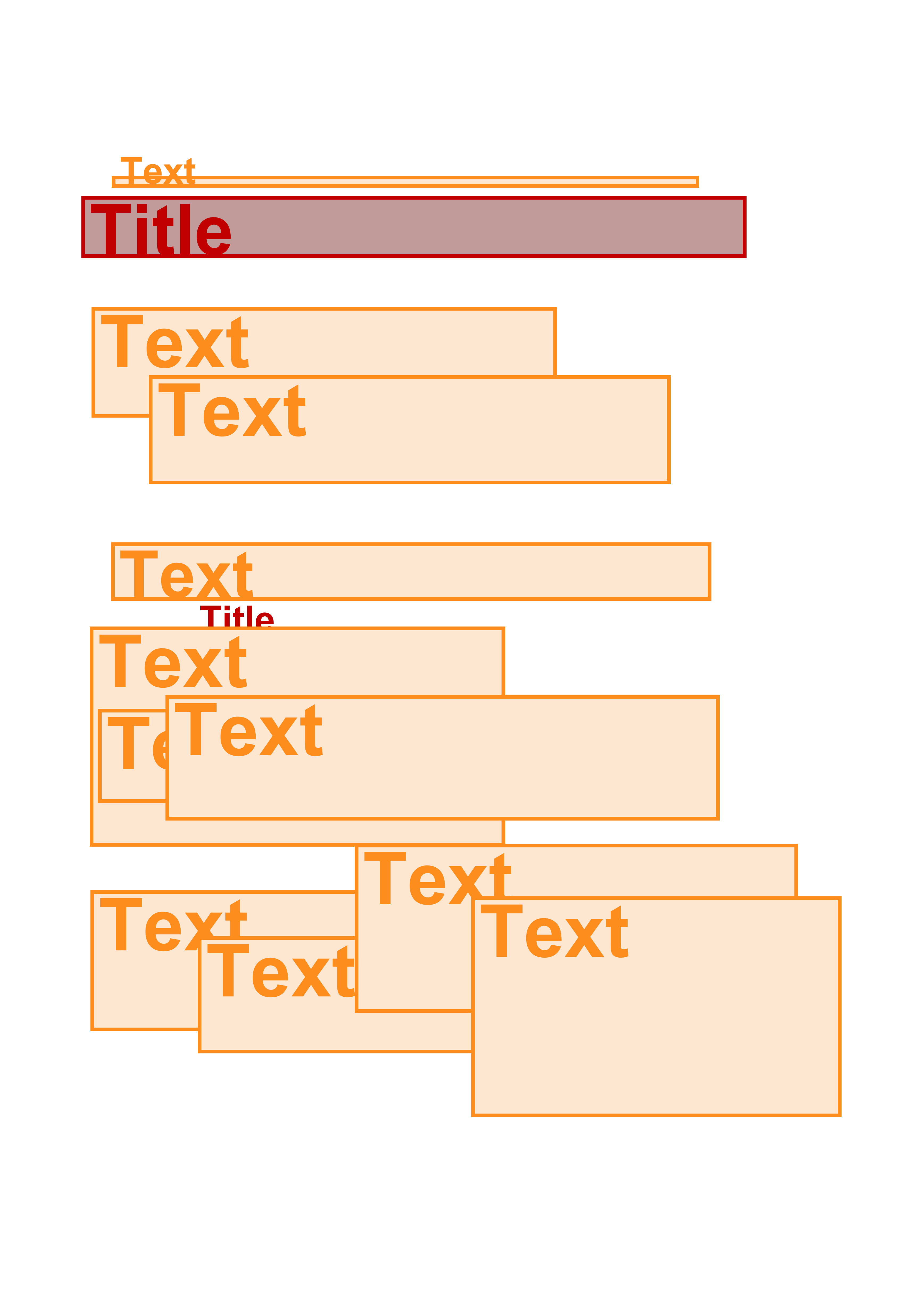} &
\includegraphics[width=\qualitativeComparisonWidth]{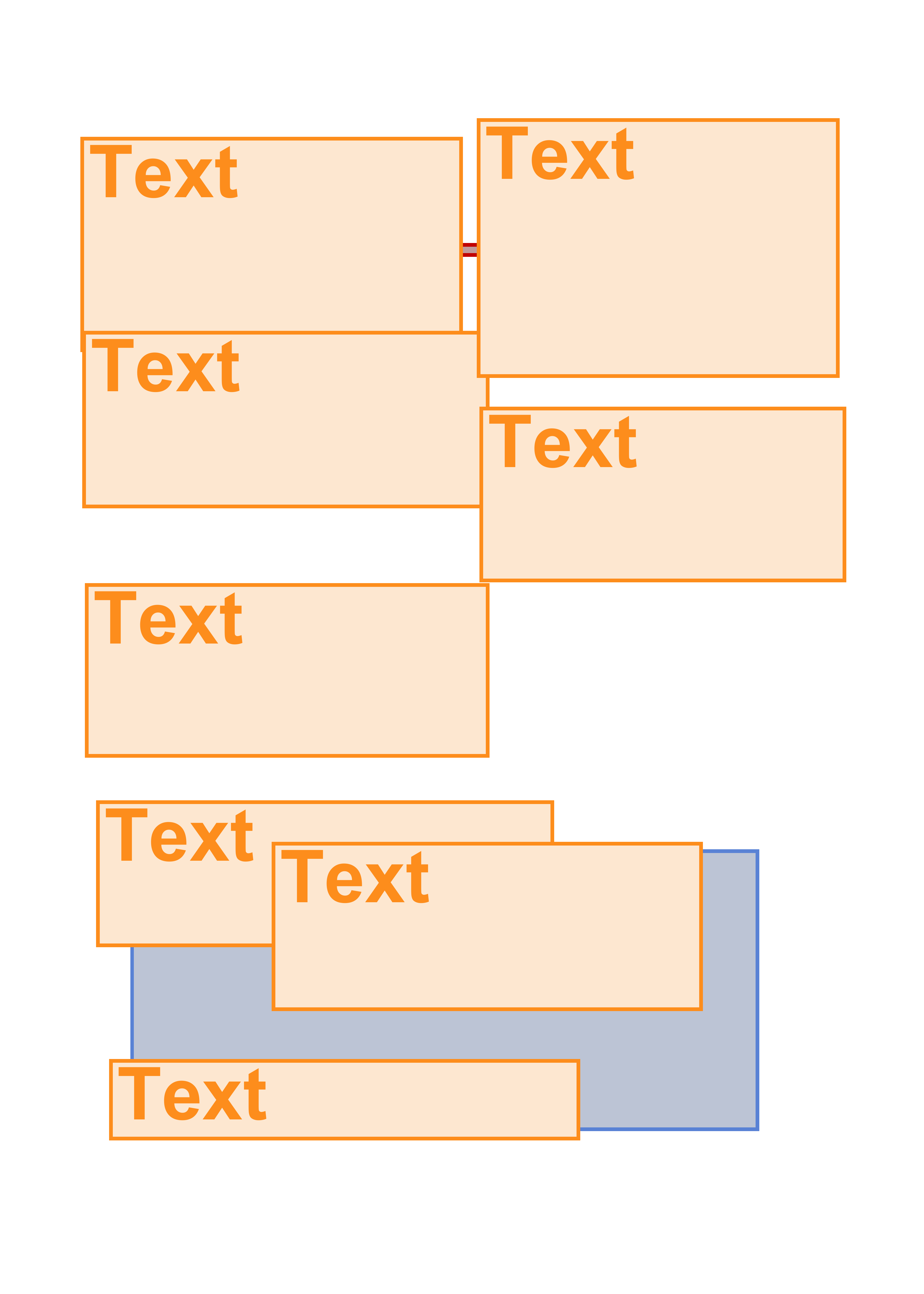} &
\includegraphics[width=\qualitativeComparisonWidth]{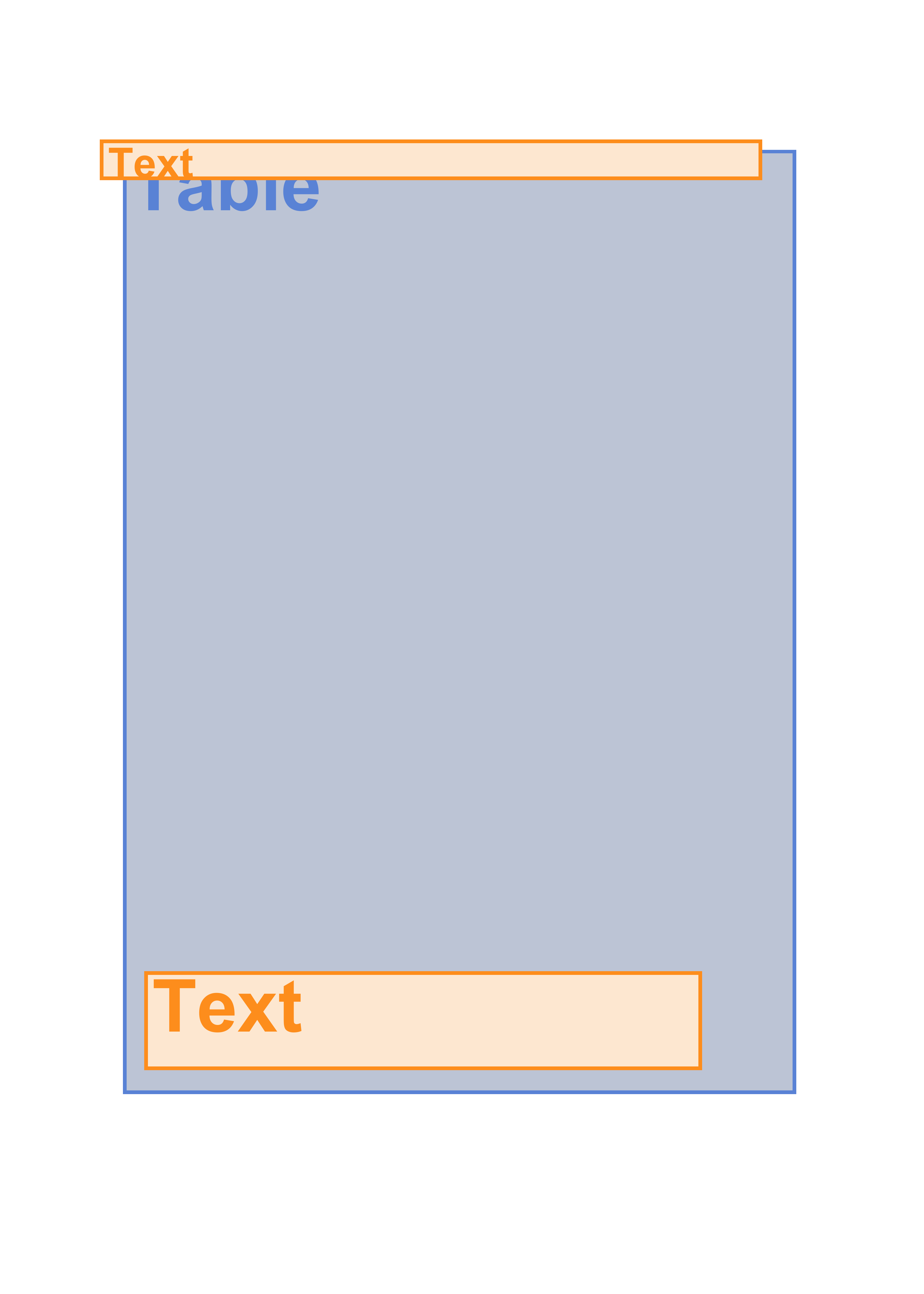} &
\includegraphics[width=\qualitativeComparisonWidth]{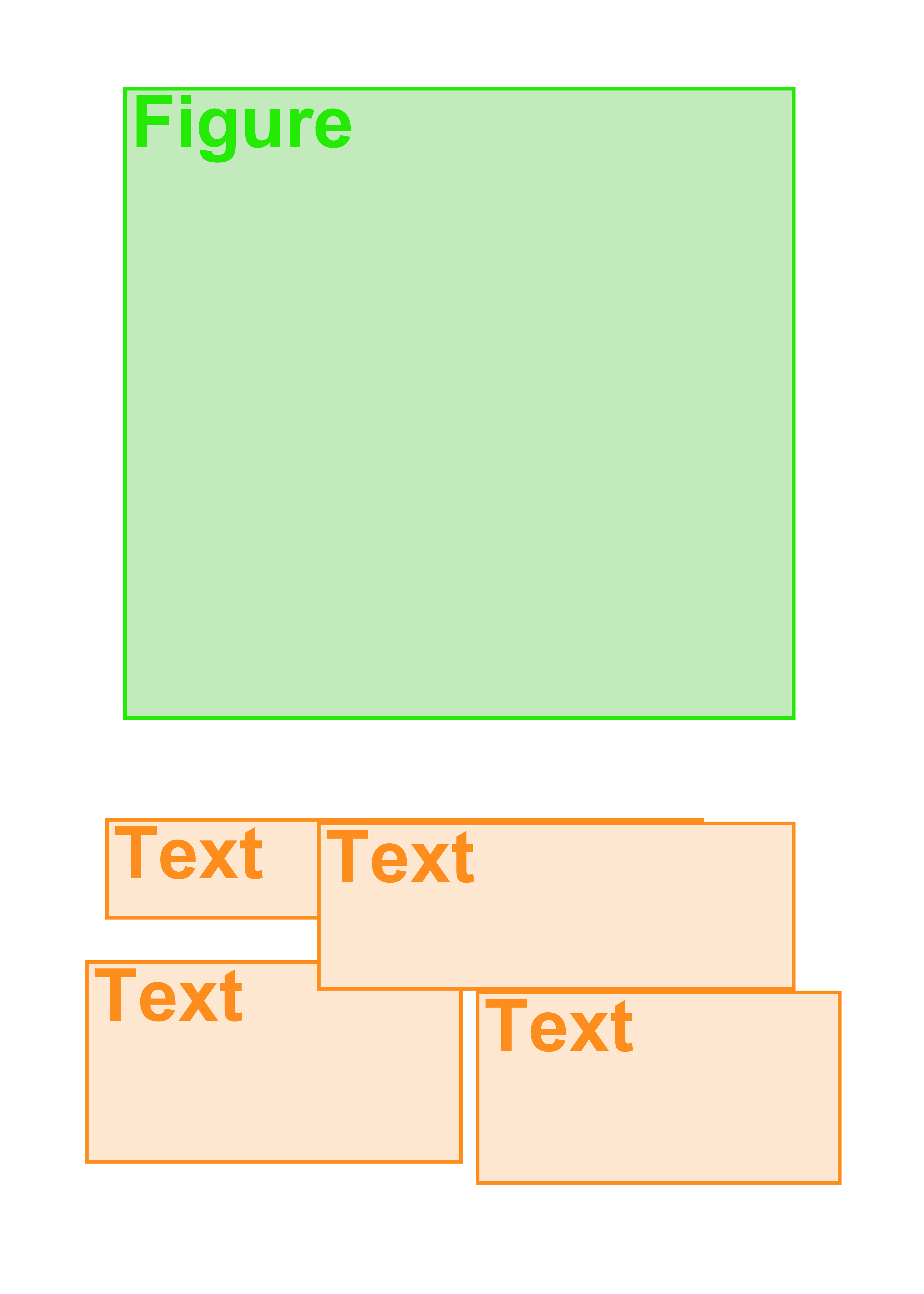}\\

\rotatebox{90}{Gupta \etal \cite{2020arXiv200614615G}} &
\includegraphics[width=\qualitativeComparisonWidth]{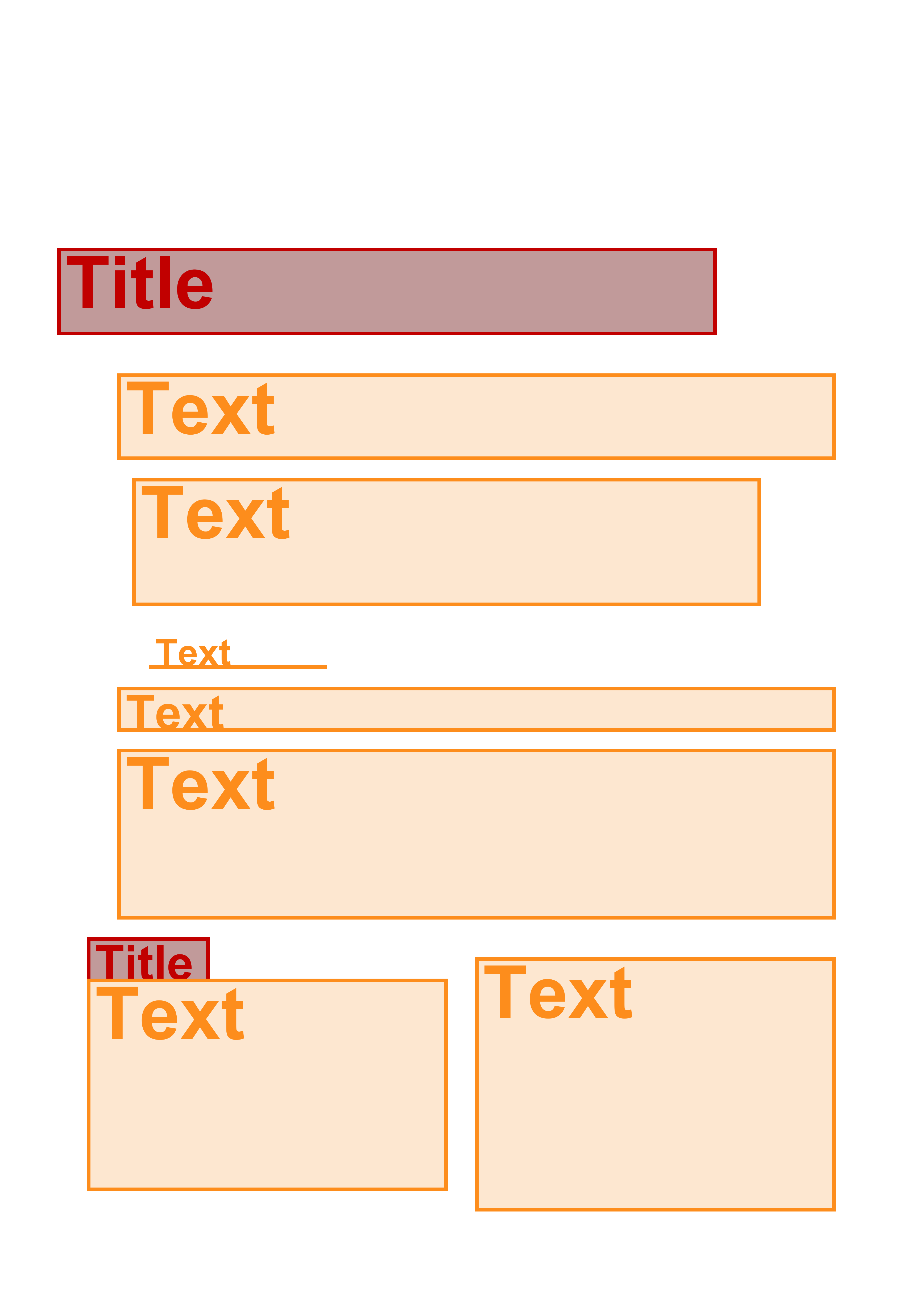} &
\includegraphics[width=\qualitativeComparisonWidth]{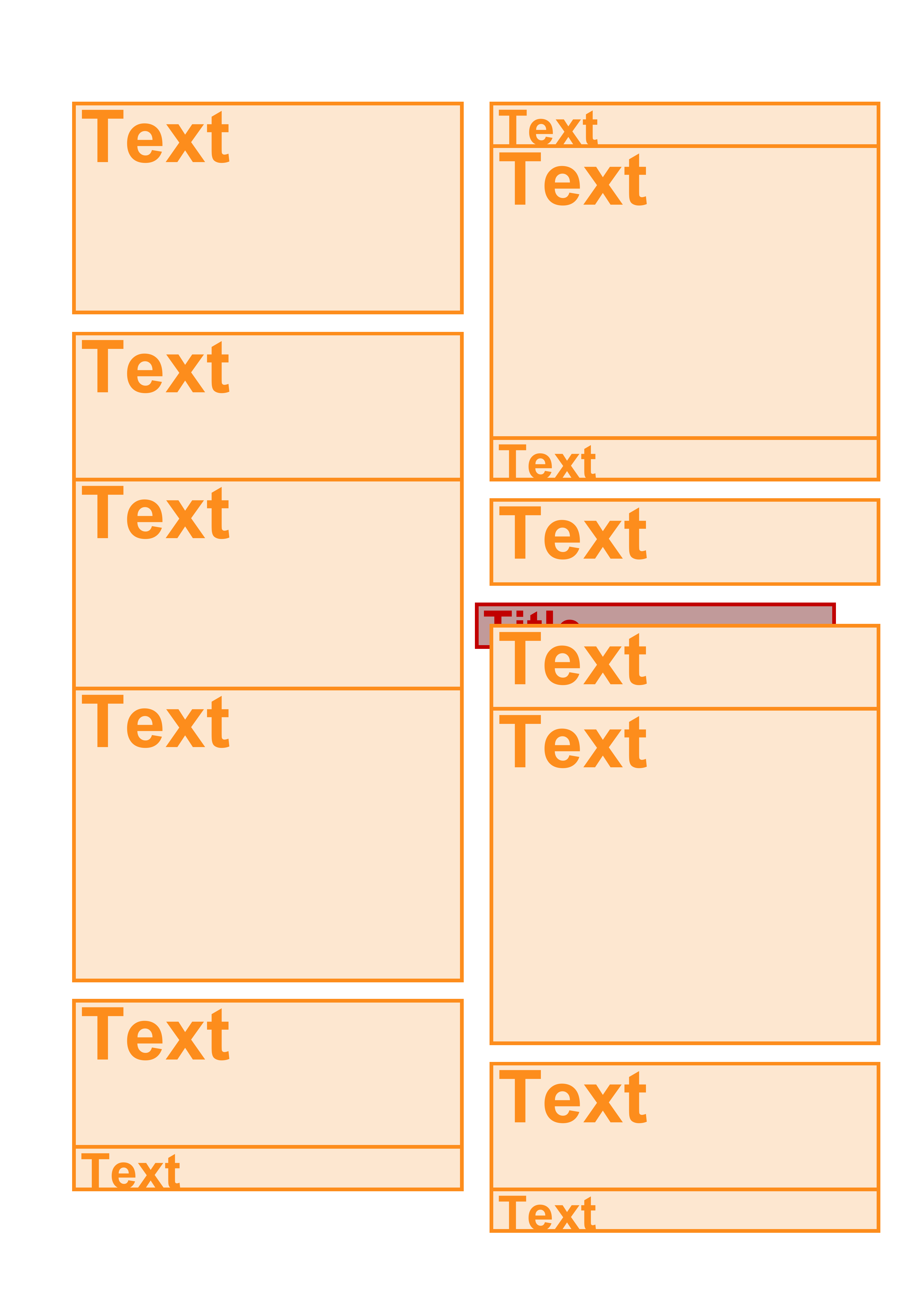} &
\includegraphics[width=\qualitativeComparisonWidth]{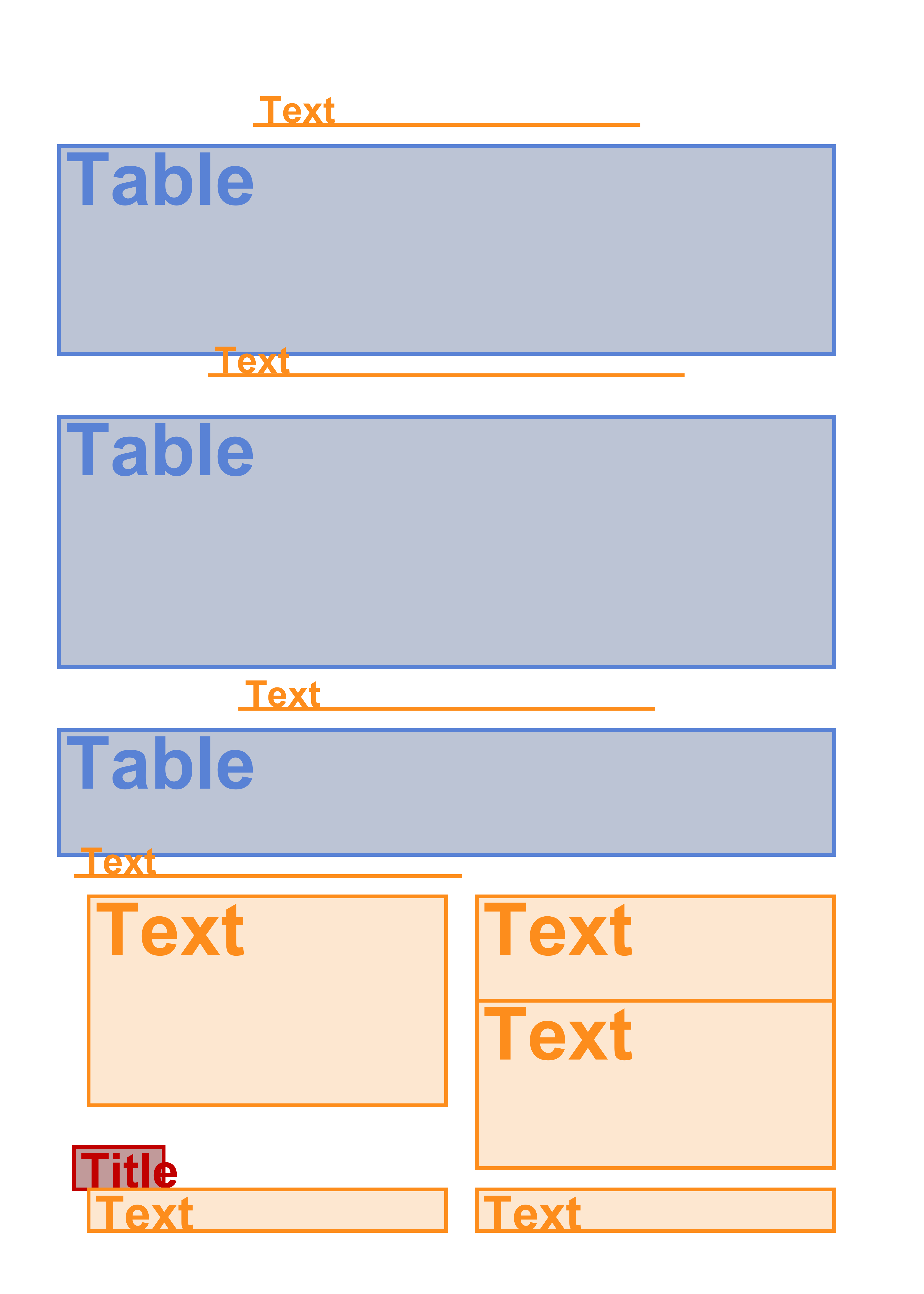} &
\includegraphics[width=\qualitativeComparisonWidth]{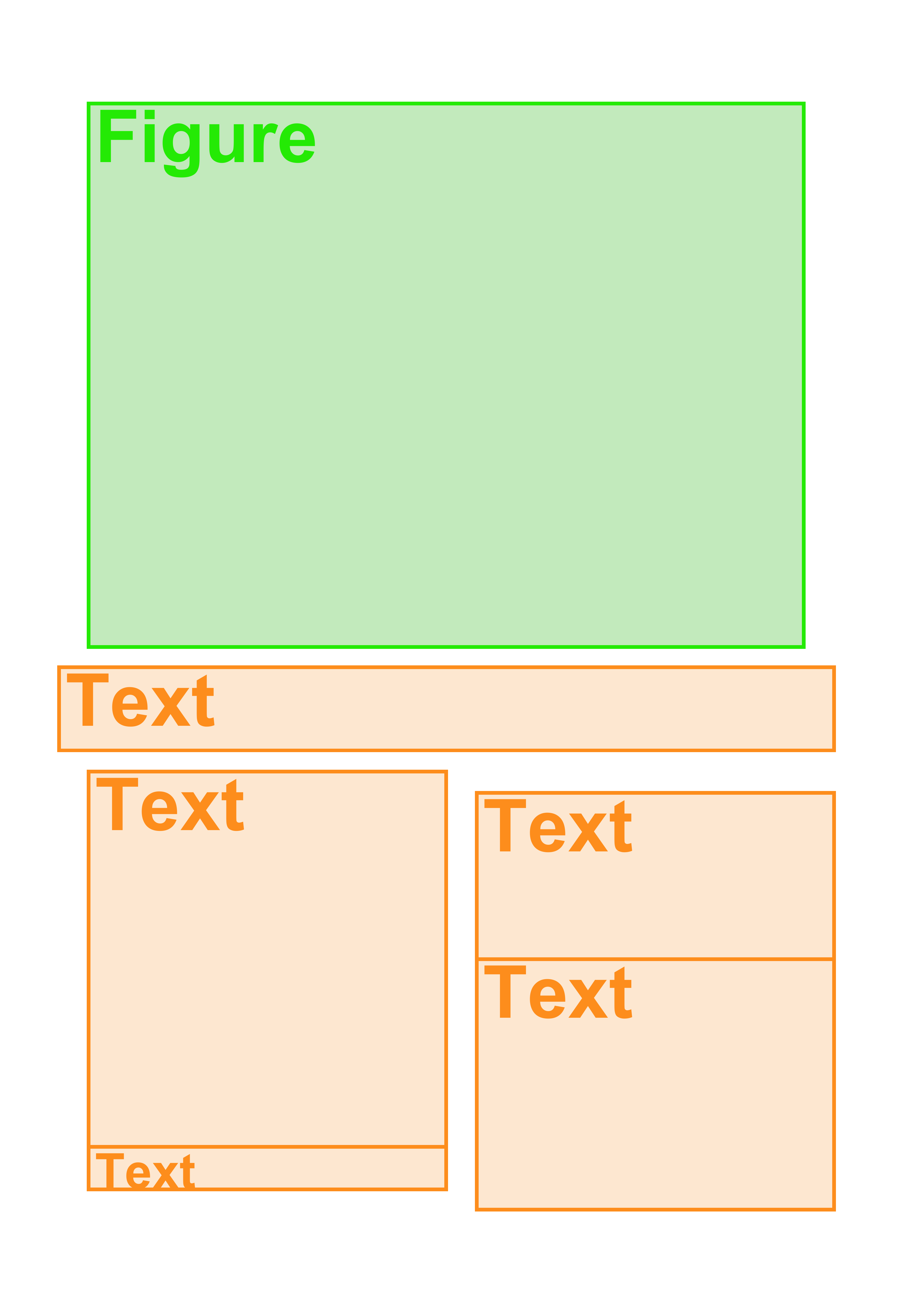} \\

\rotatebox{90}{\hspace{0.7cm} Ours} &
\includegraphics[width=\qualitativeComparisonWidth]{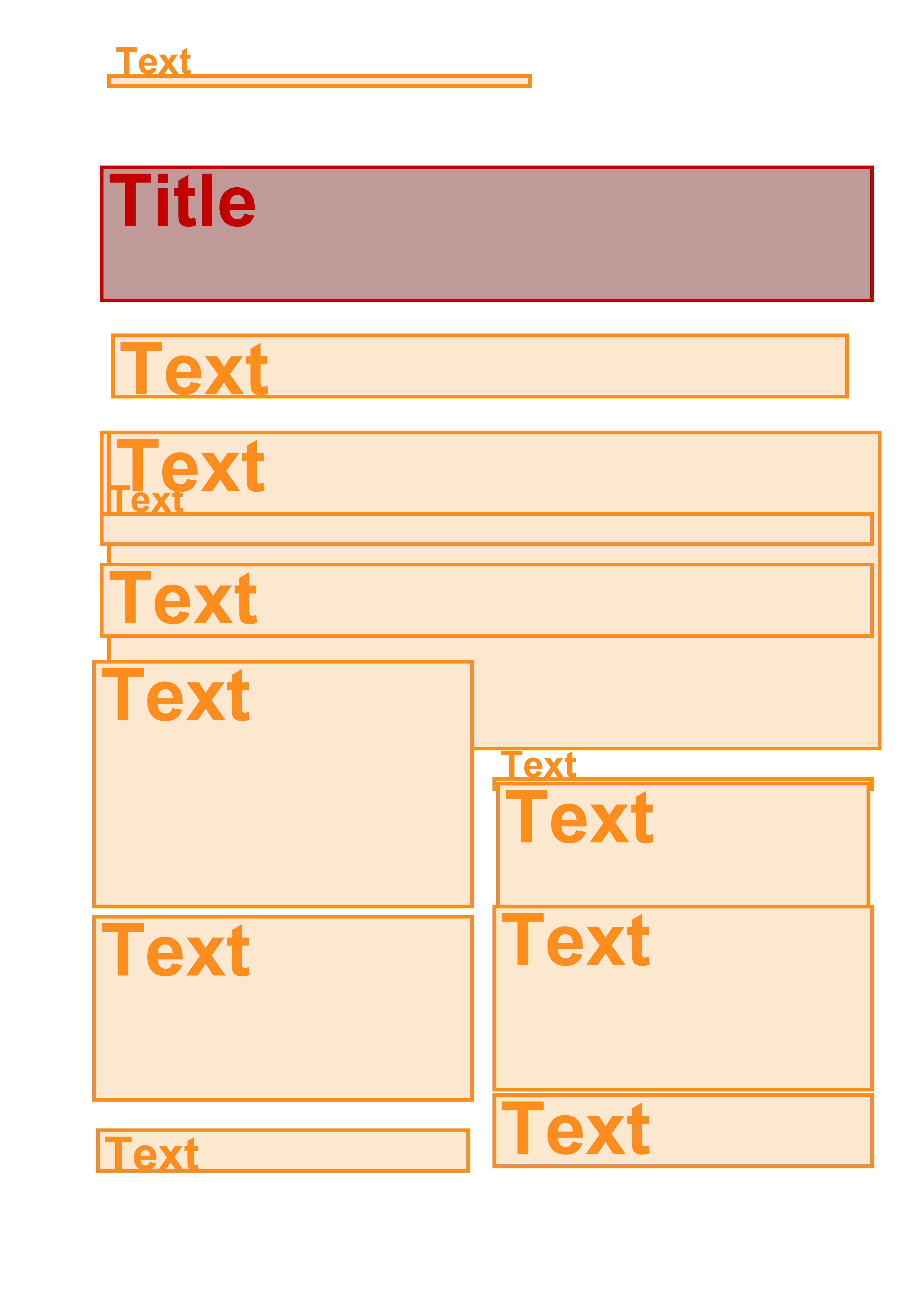} &
\includegraphics[width=\qualitativeComparisonWidth]{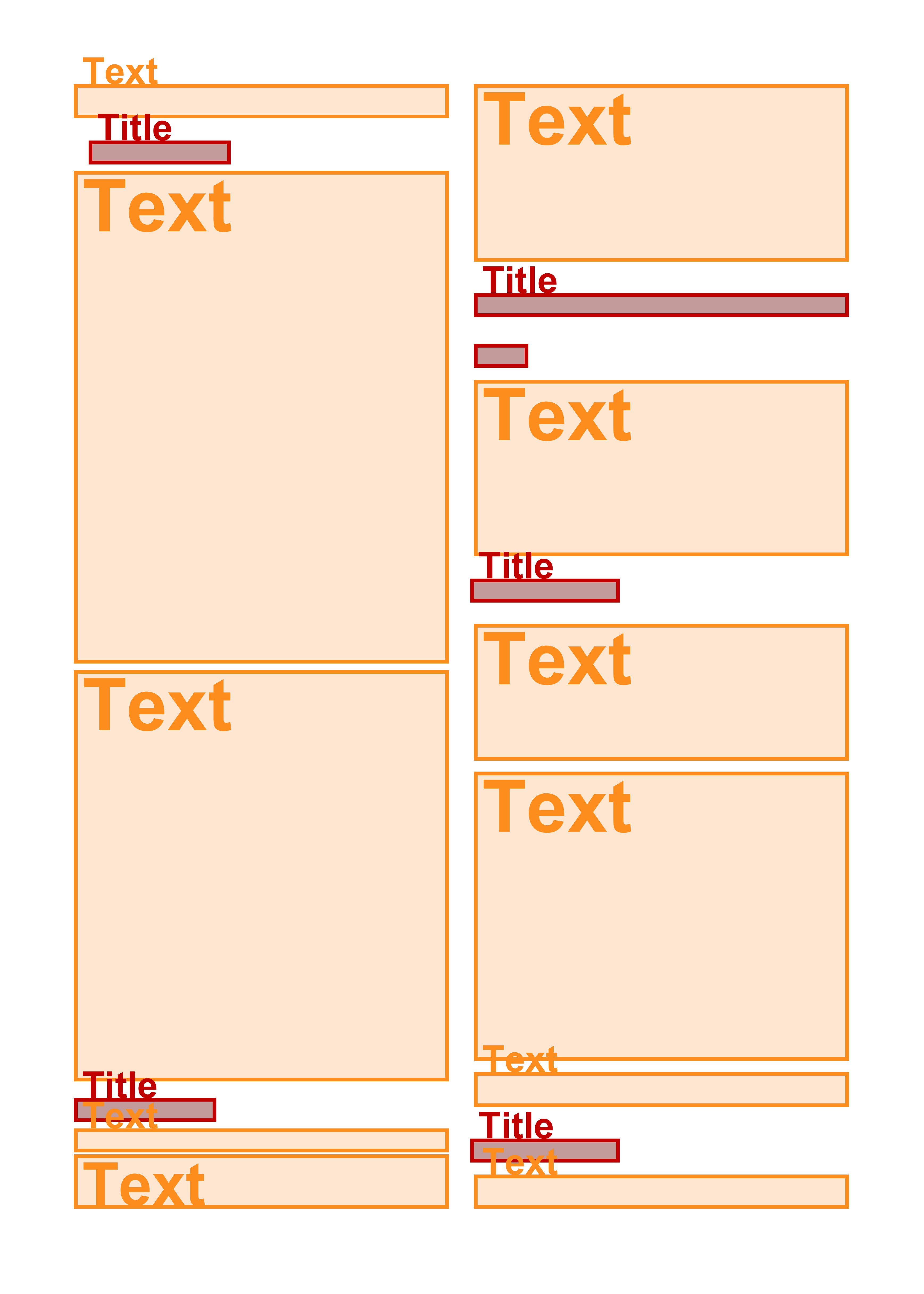} &
\includegraphics[width=\qualitativeComparisonWidth]{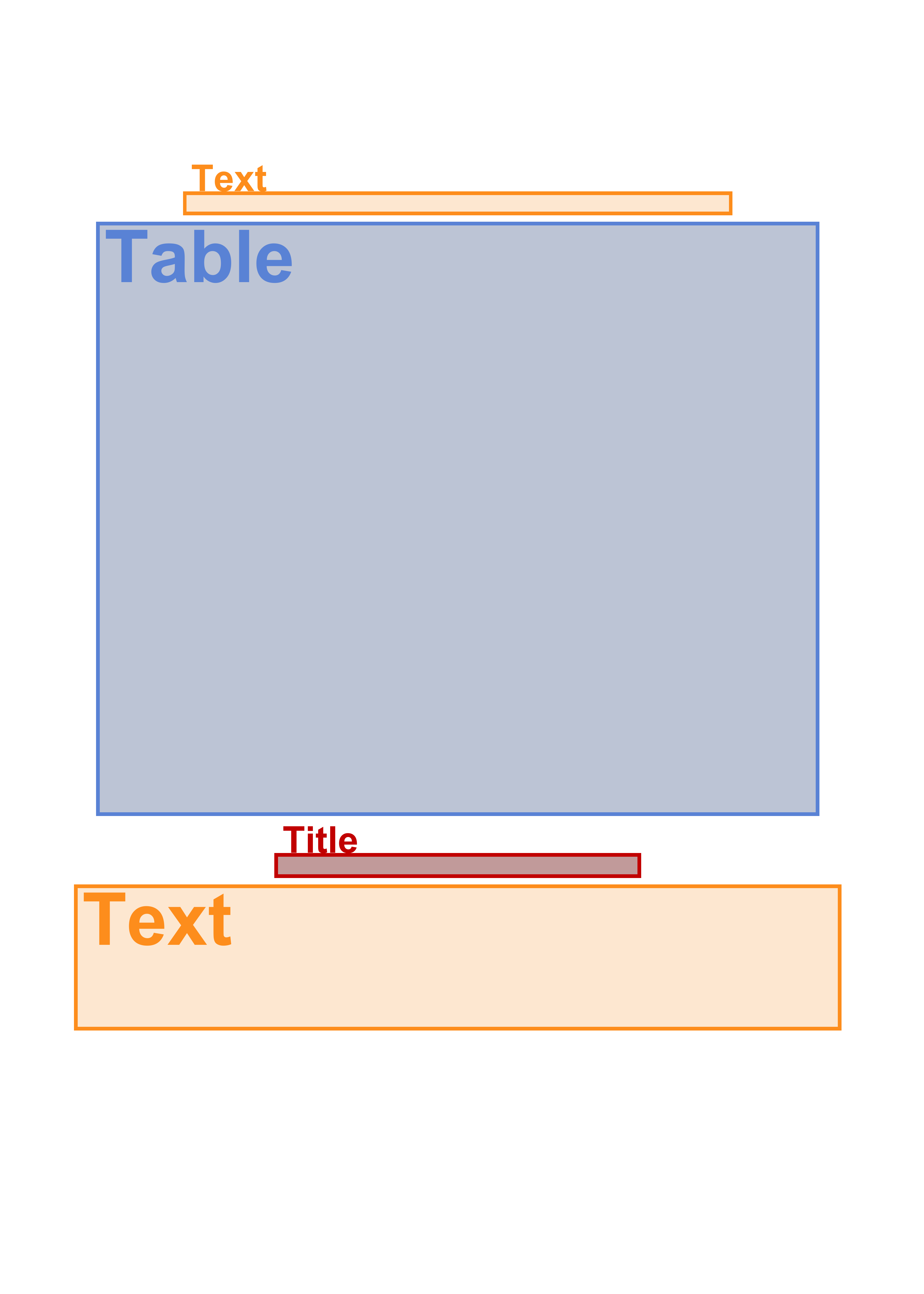} &
\includegraphics[width=\qualitativeComparisonWidth]{figures/samples/publaynet/layouts/0025} \\

\end{tabular}
    \caption{Qualitative comparison between LayoutVAE, Gupta \etal and our method on PubLayNet. The \ac{rnn} of LayoutVAE struggles with a large number of elements.}
    \label{fig:qualitative_comparision_publaynet}
\end{figure}

\begin{figure}
\setlength{\tabcolsep}{2pt}
\newlength{\docsimRICOWidth}
\setlength{\docsimRICOWidth}{0.178\linewidth}
    \centering
        \begin{tabular}{cccccc}
        \rotatebox{90}{\hspace{0.8cm}\small Synthetic} & \includegraphics[width=\docsimRICOWidth]{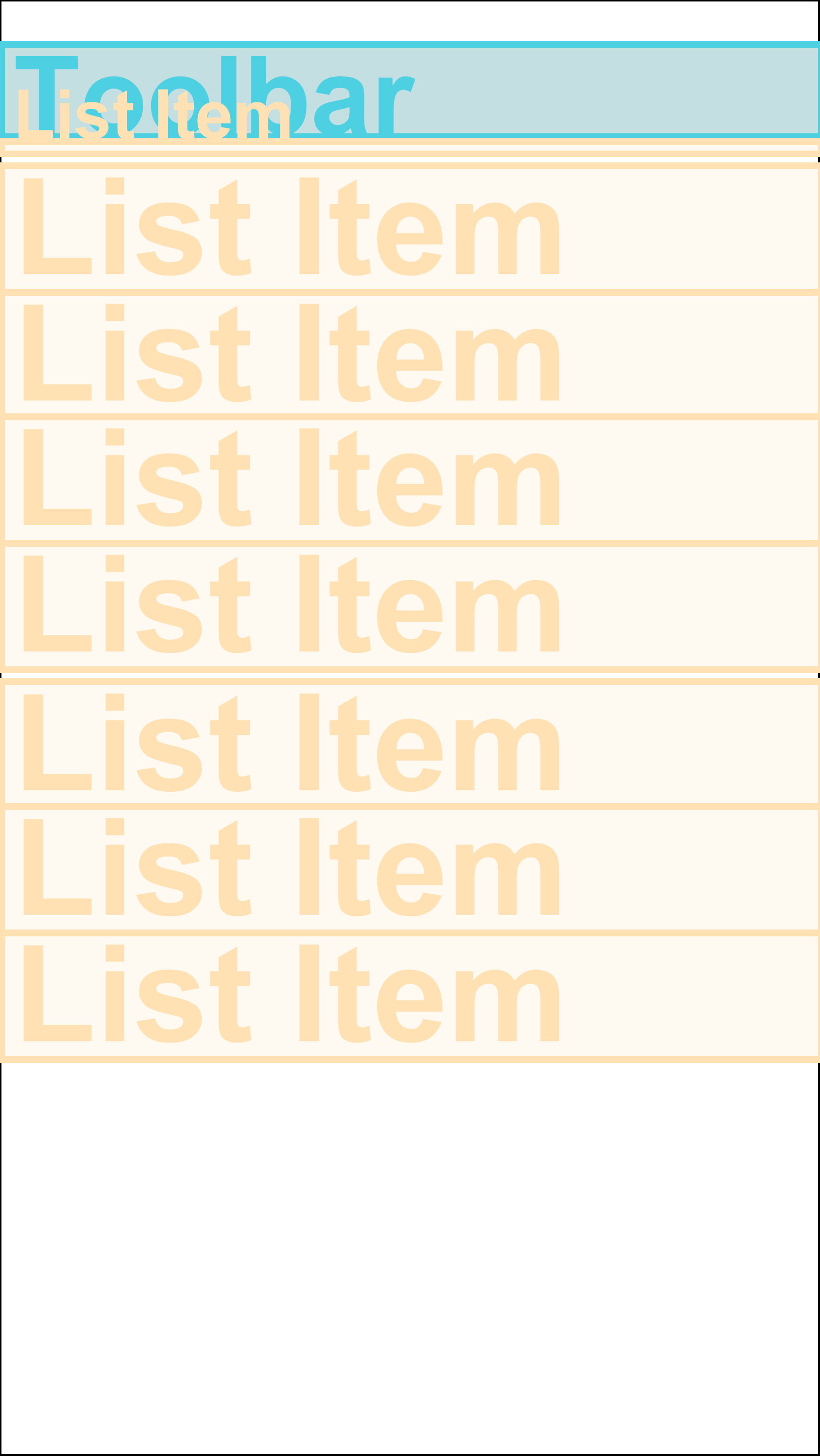} &
        \includegraphics[width=\docsimRICOWidth]{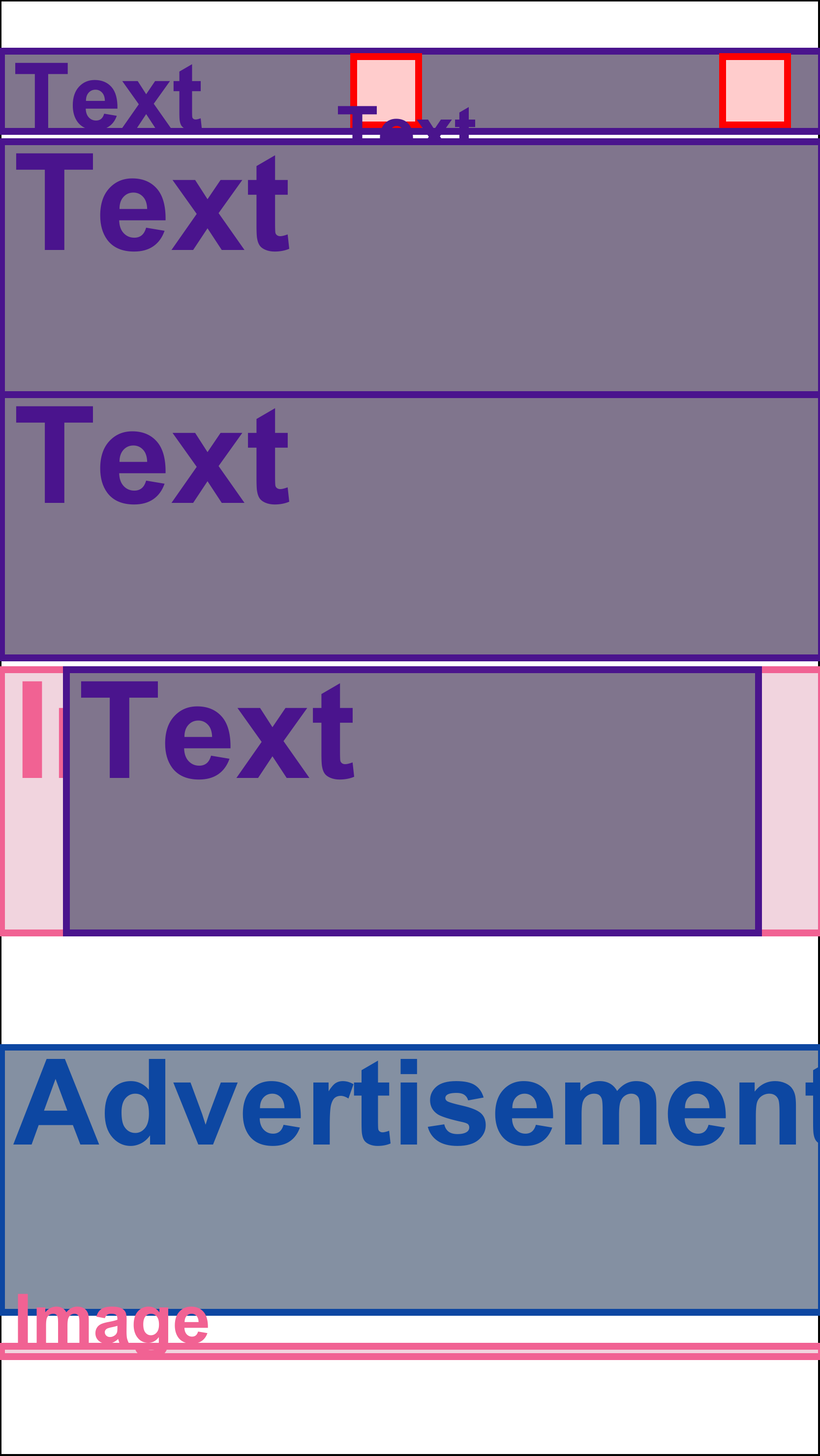} & \includegraphics[width=\docsimRICOWidth]{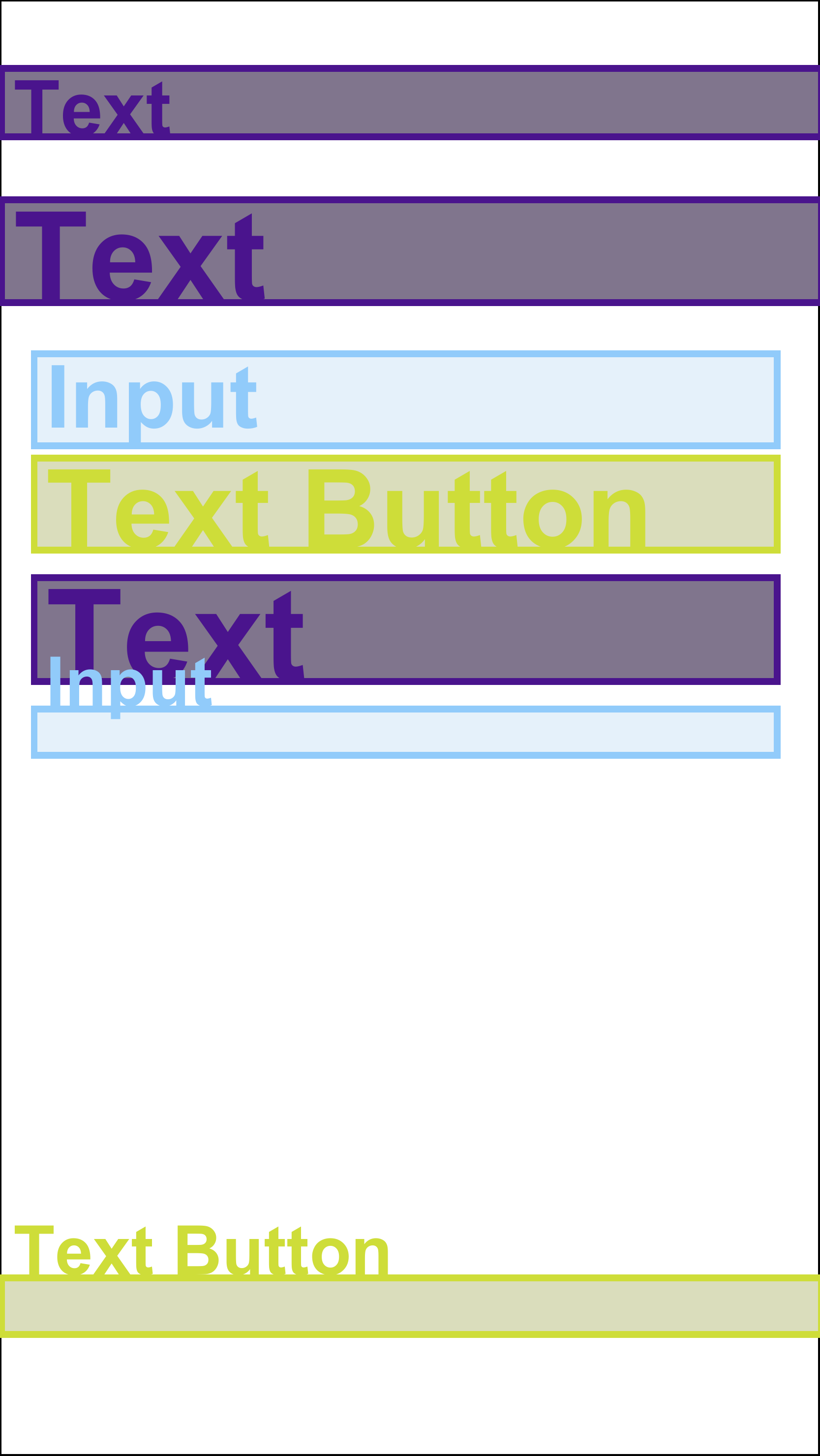} & \includegraphics[width=\docsimRICOWidth]{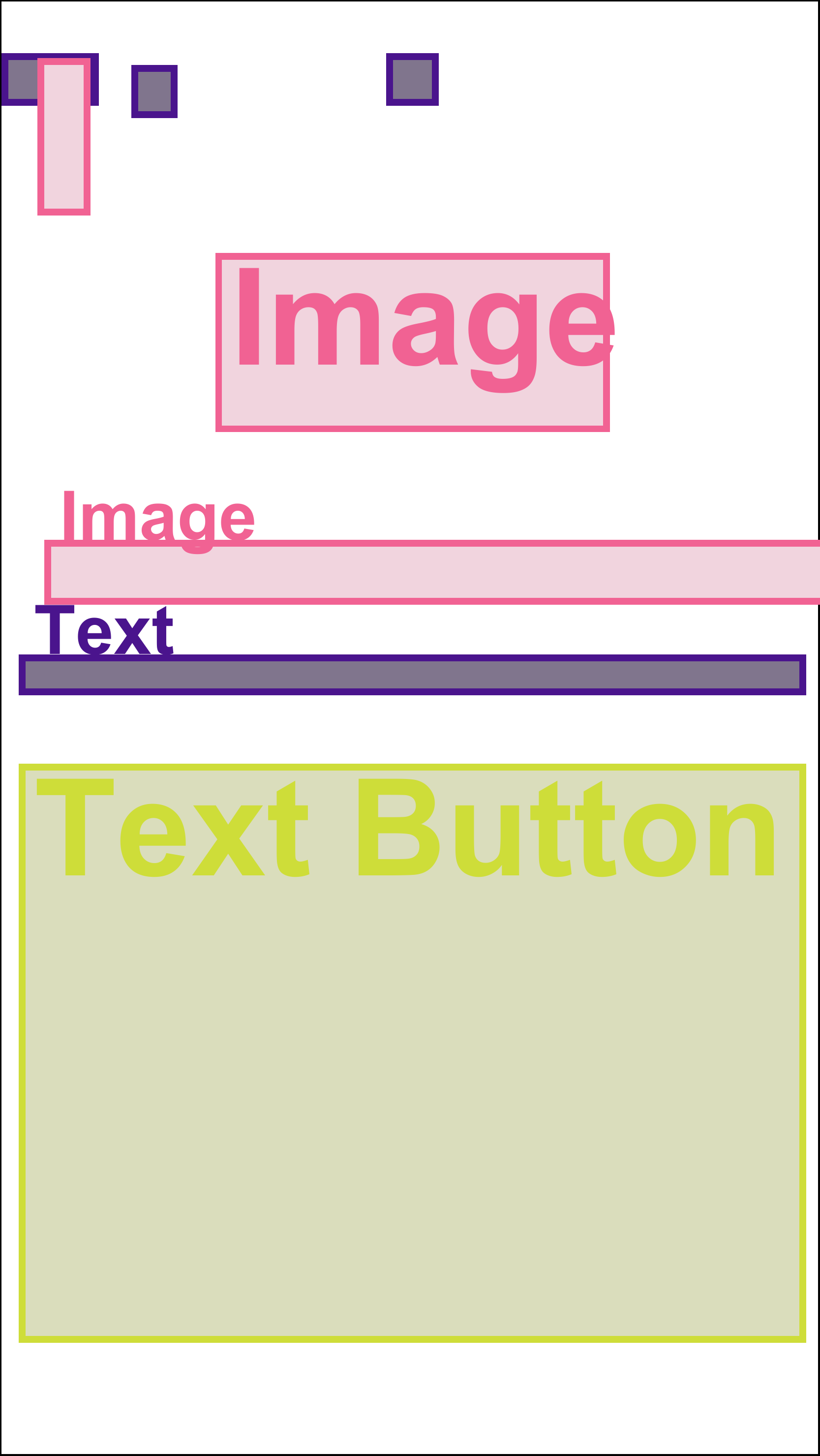} & \includegraphics[width=\docsimRICOWidth]{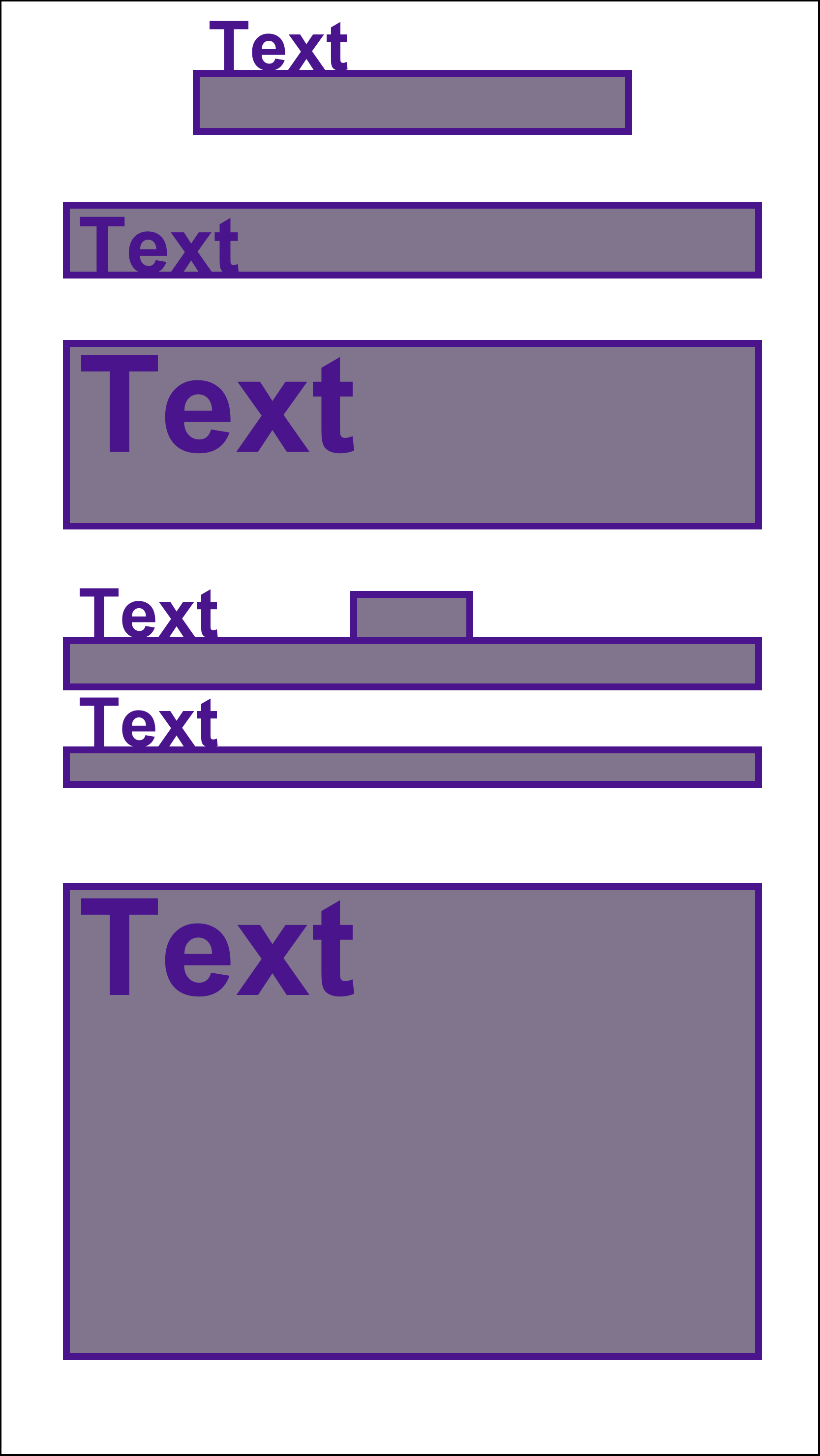}\\
        \rotatebox{90}{\hspace{0.6cm}\small Real layout} & \includegraphics[width=\docsimRICOWidth]{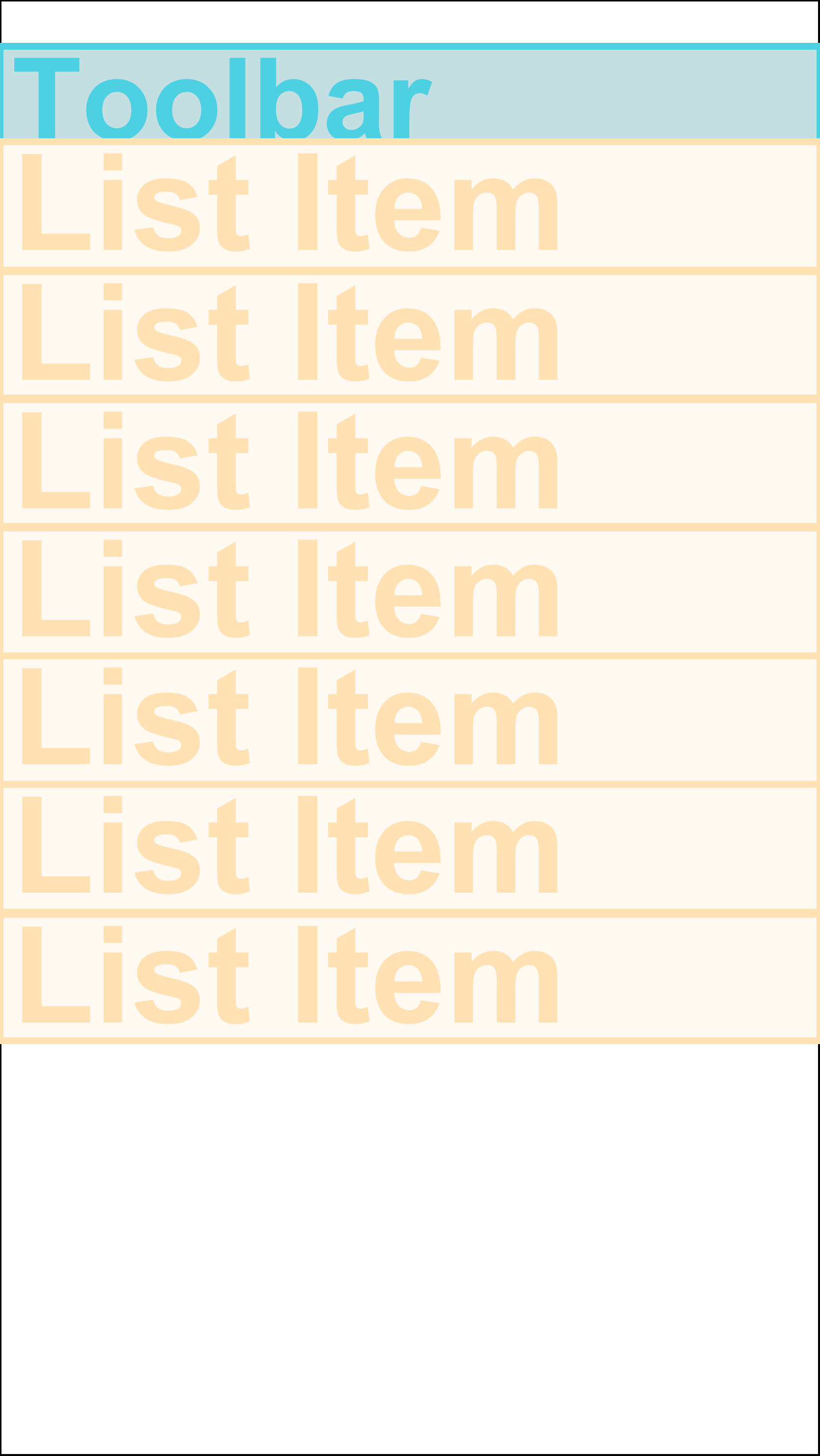} & \includegraphics[width=\docsimRICOWidth]{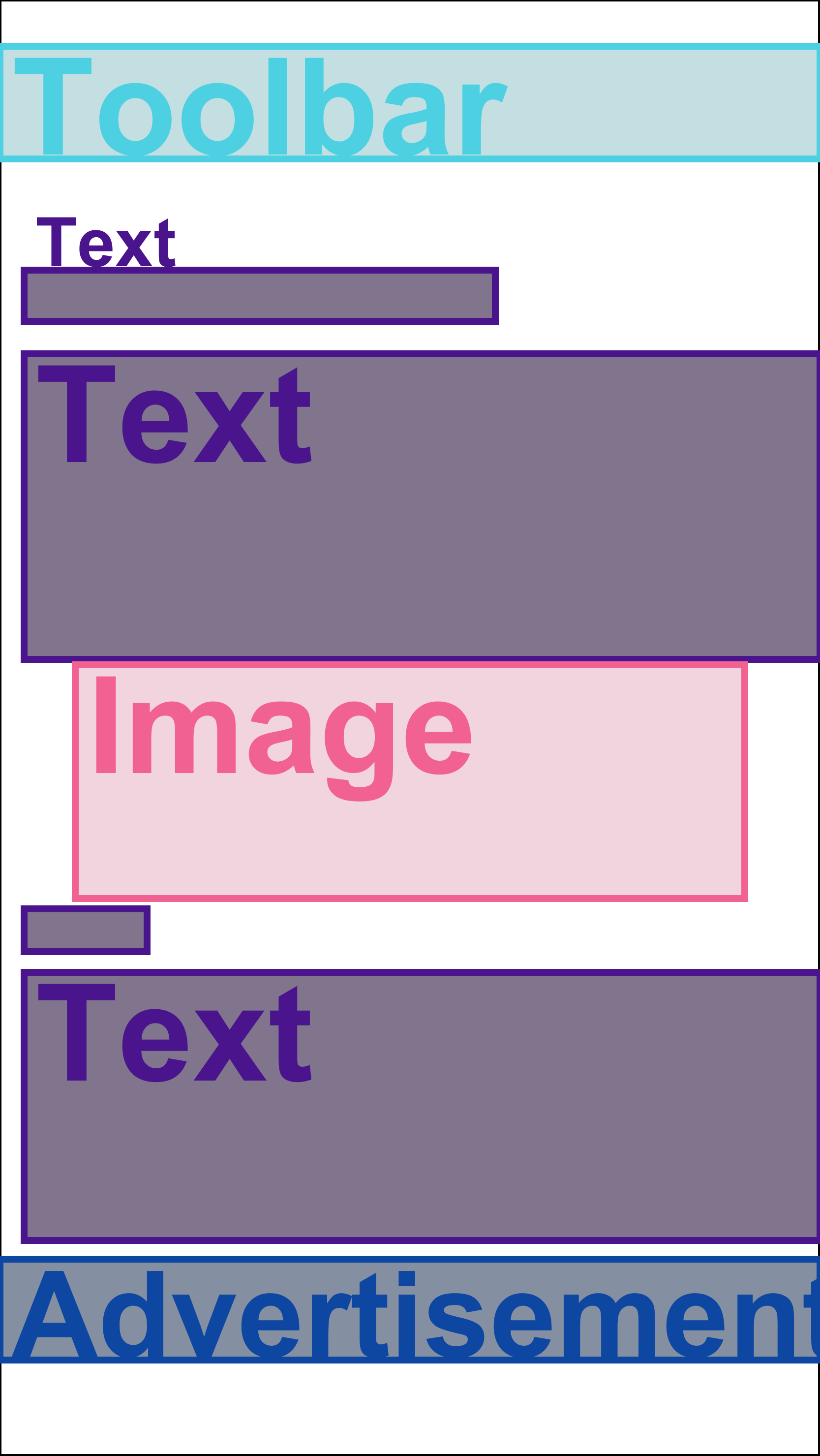} & \includegraphics[width=\docsimRICOWidth]{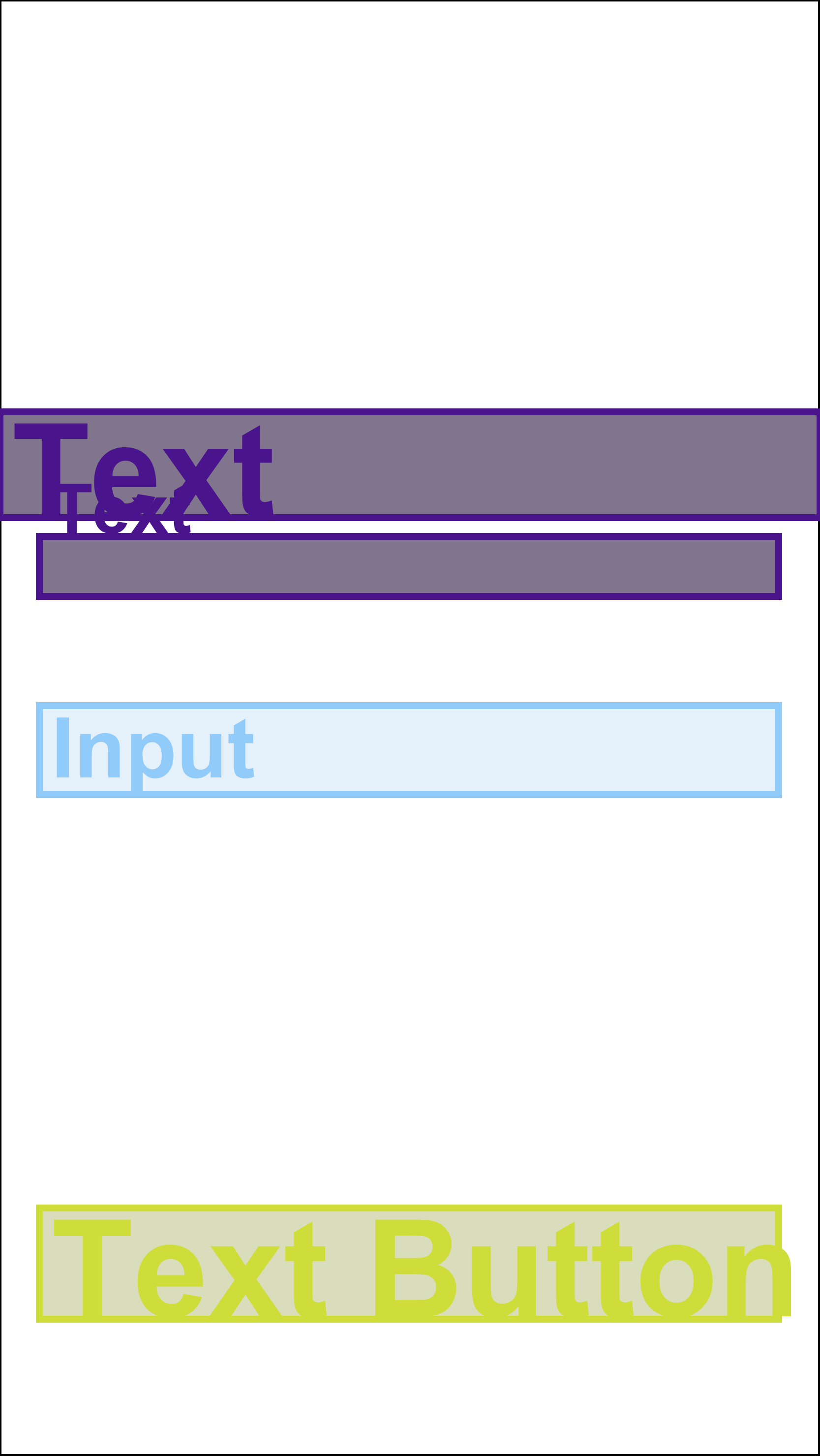} & \includegraphics[width=\docsimRICOWidth]{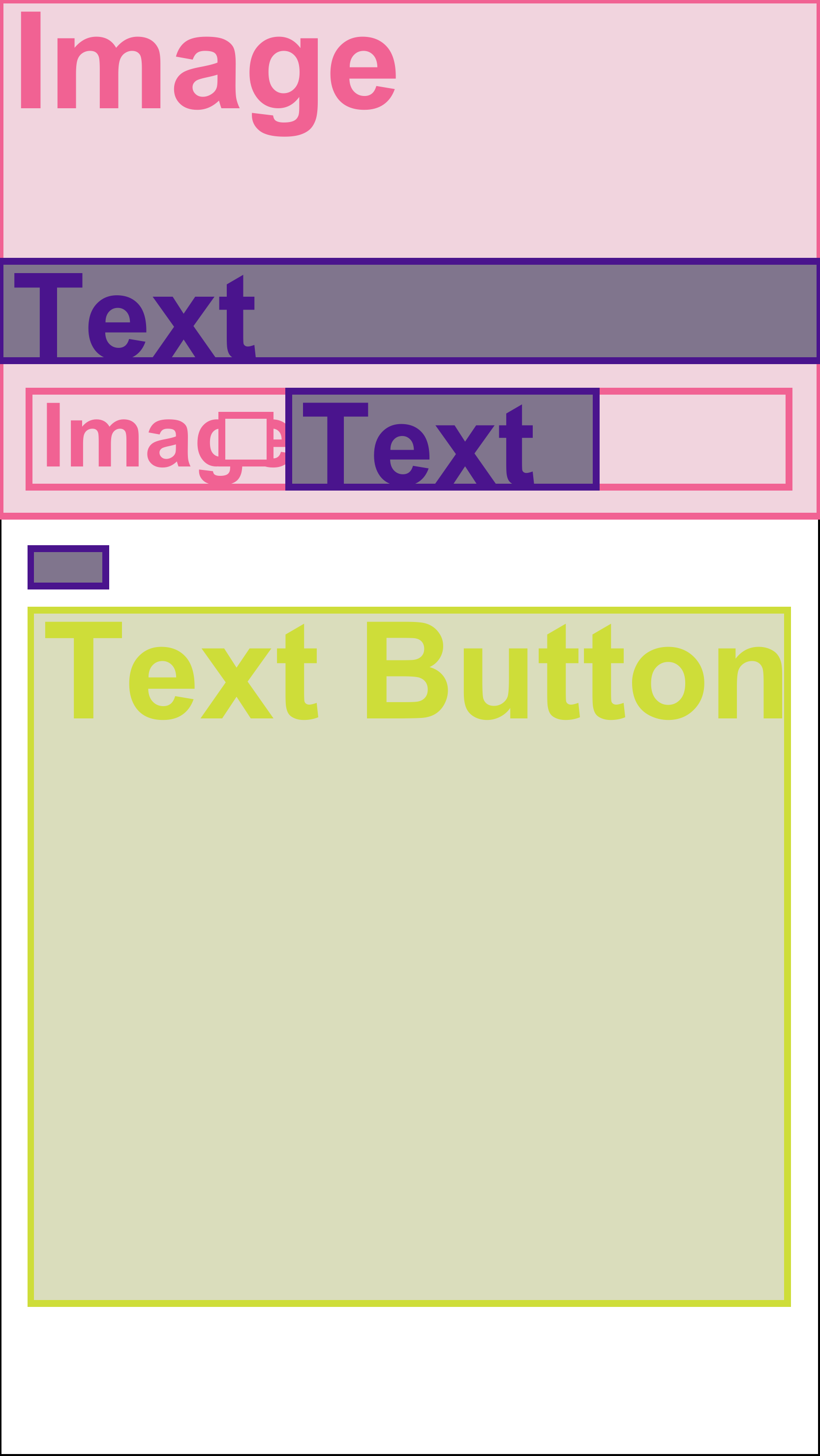} & \includegraphics[width=\docsimRICOWidth]{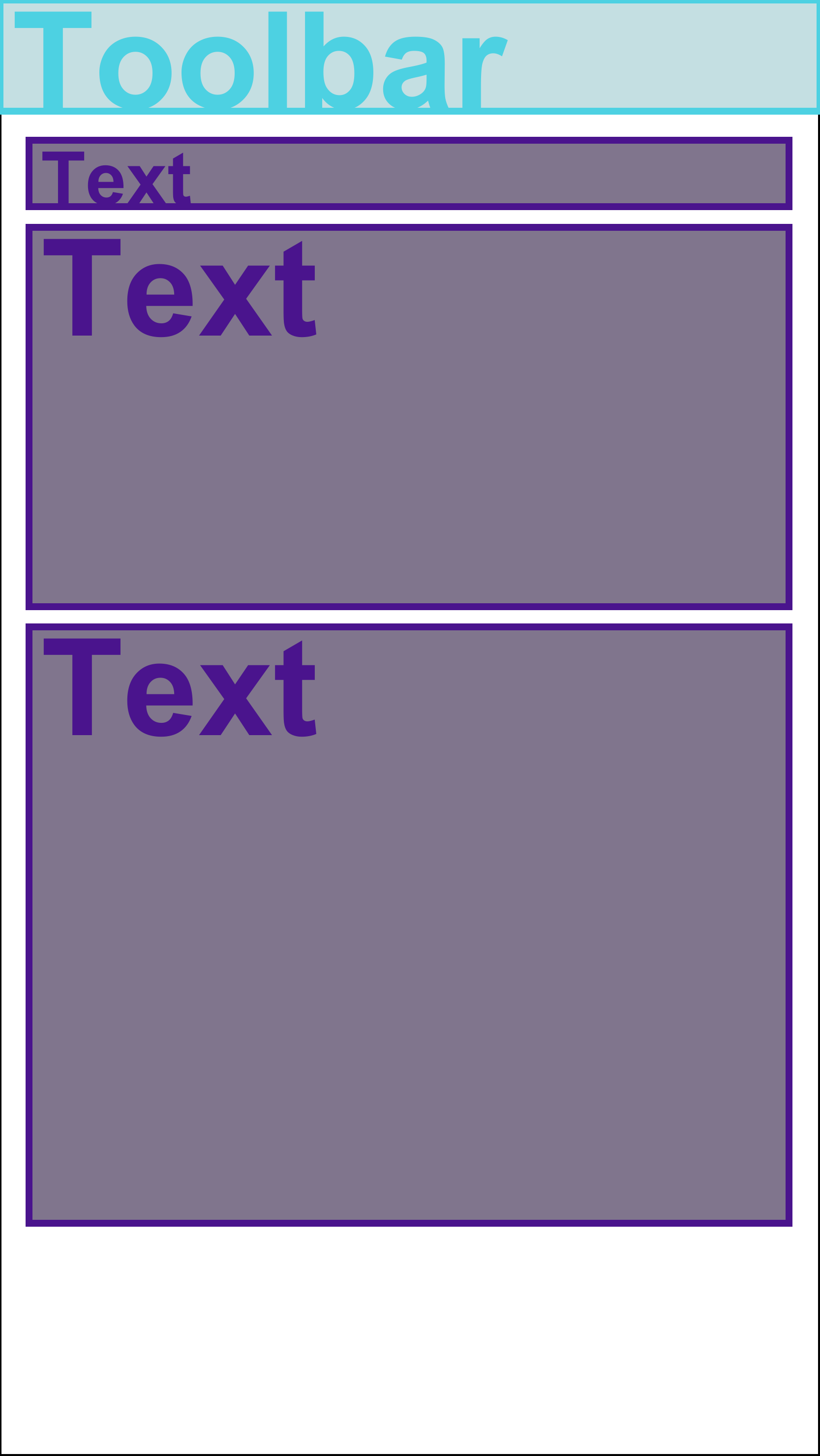}\\
        \rotatebox{90}{\hspace{0.7cm}\small Real image} & \includegraphics[width=\docsimRICOWidth]{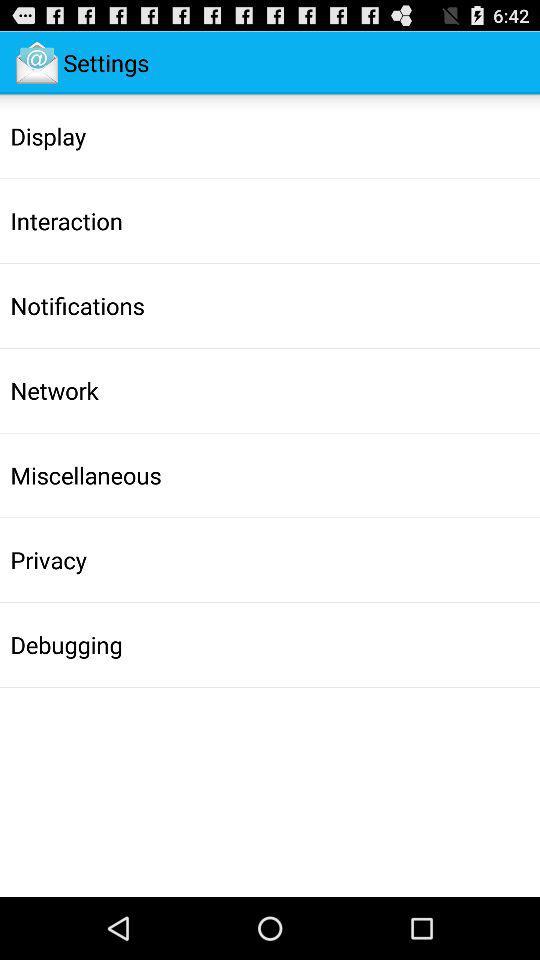} &
        \includegraphics[width=\docsimRICOWidth]{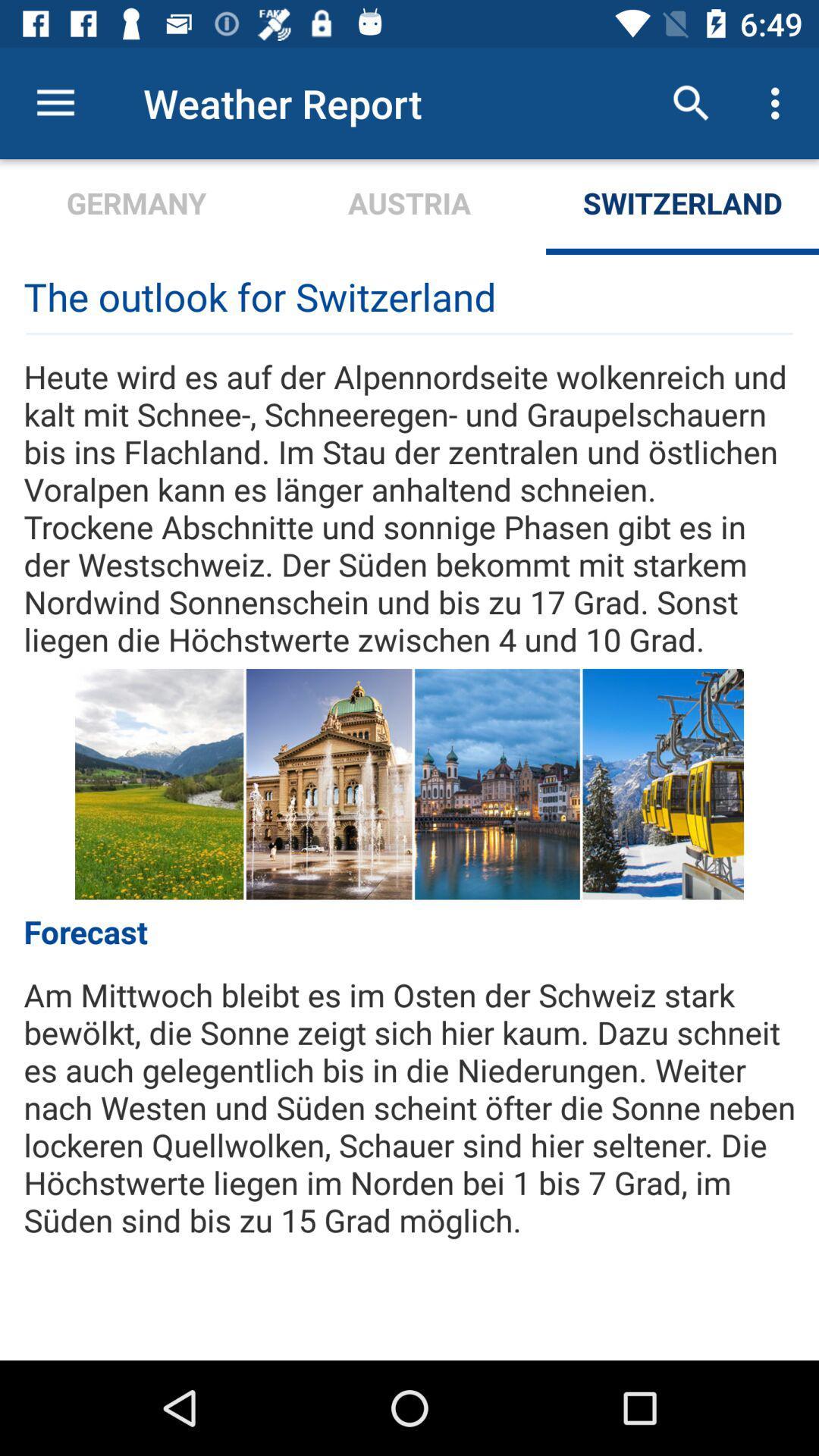} & \includegraphics[width=\docsimRICOWidth]{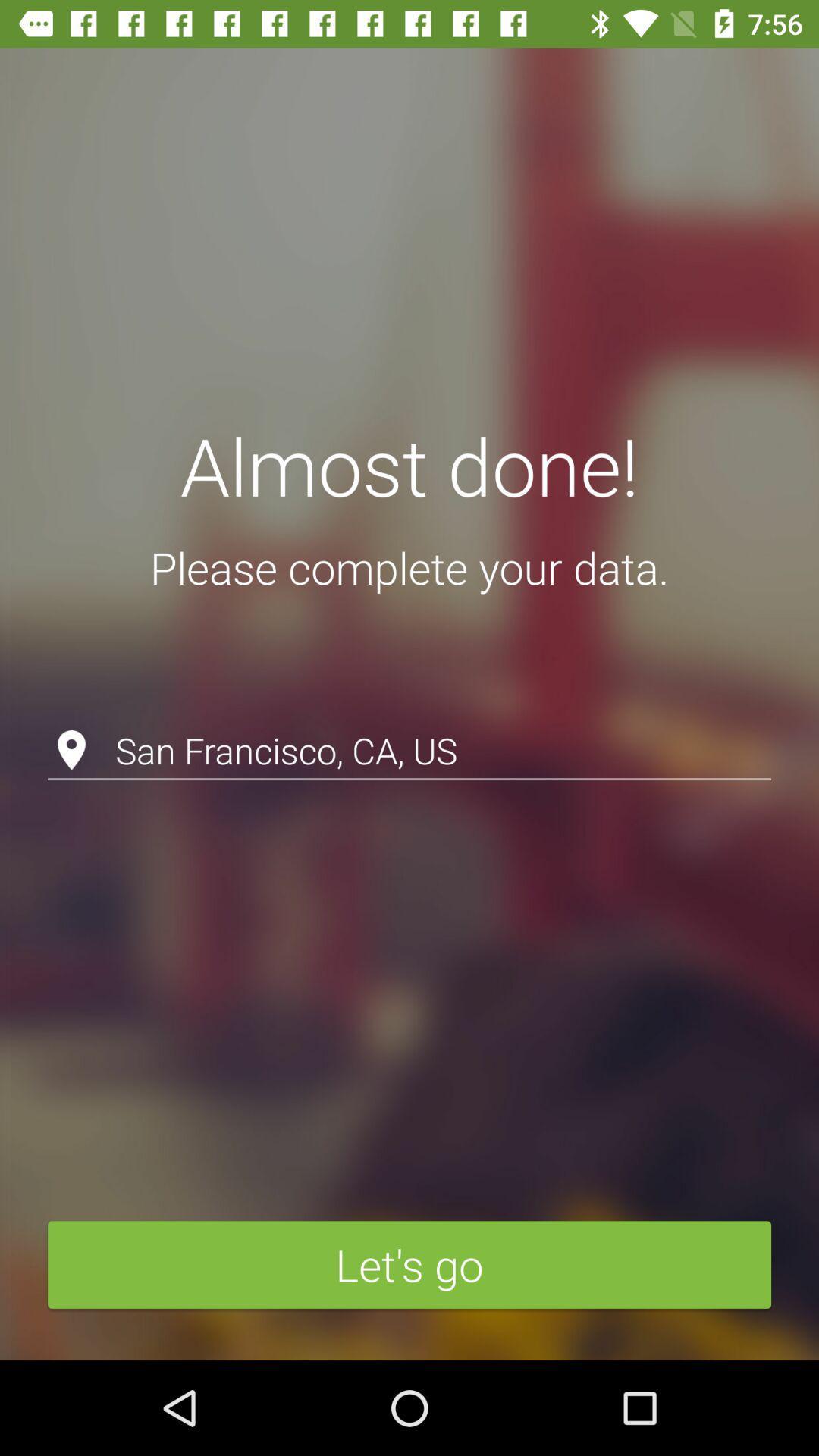} & \includegraphics[width=\docsimRICOWidth]{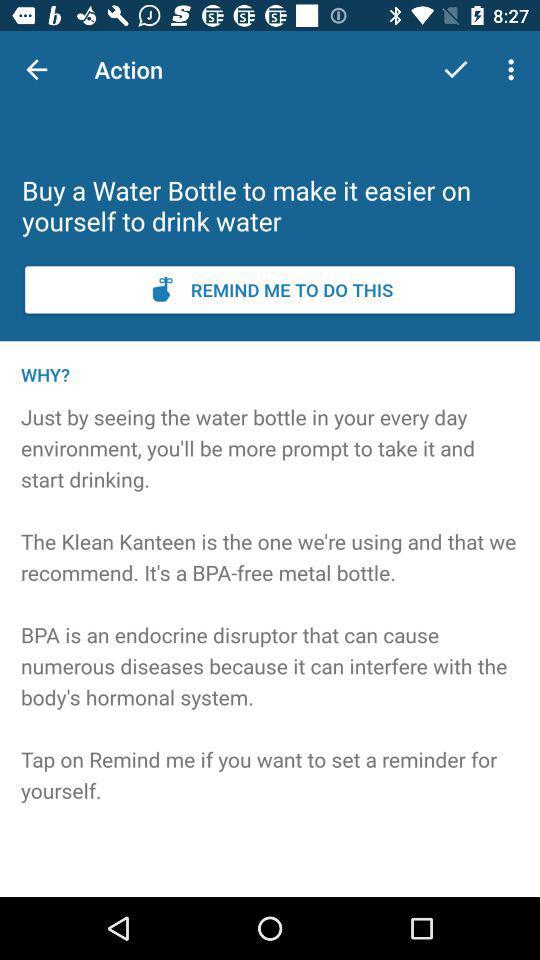} & \includegraphics[width=\docsimRICOWidth]{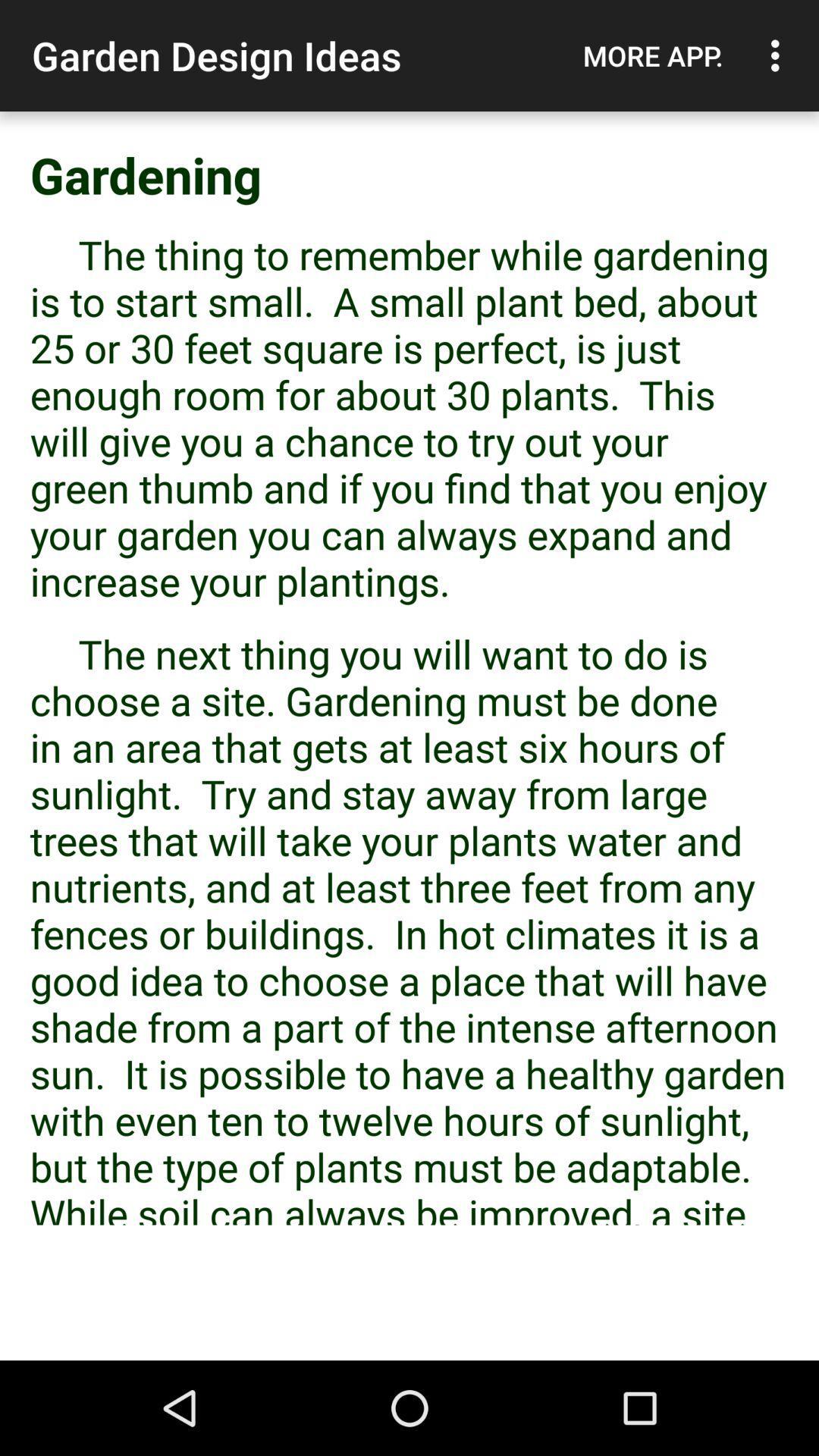}\\
    \end{tabular}
    
    \caption{Generated layouts for RICO and their associated DocSim match. The supplement shows more samples.}
    \label{fig:rico_samples}
\end{figure}

\begin{figure}
    \newlength{\lostganWidth}
    \setlength{\lostganWidth}{0.19\linewidth}
    \centering
\includegraphics[width=\lostganWidth]{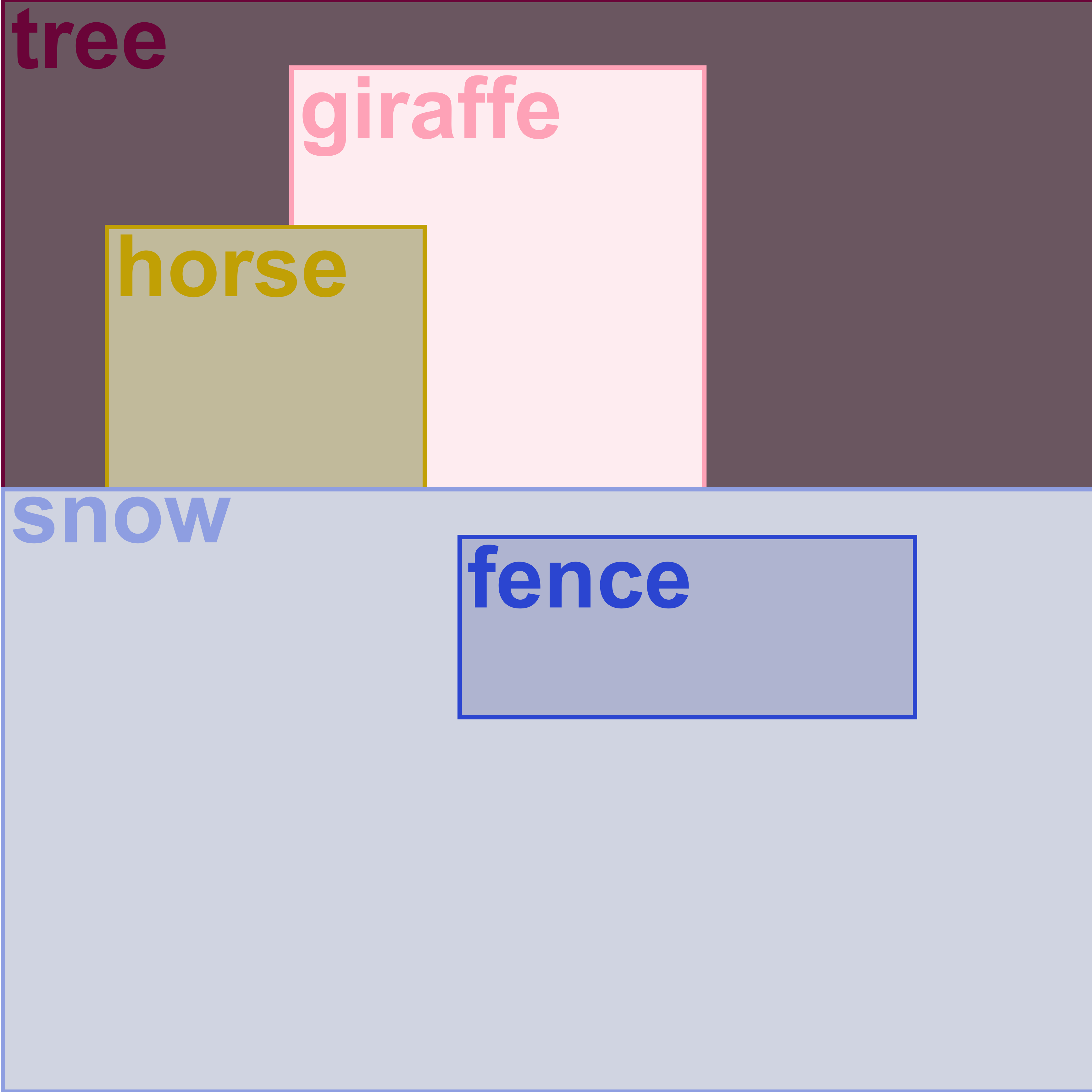}
\includegraphics[width=\lostganWidth]{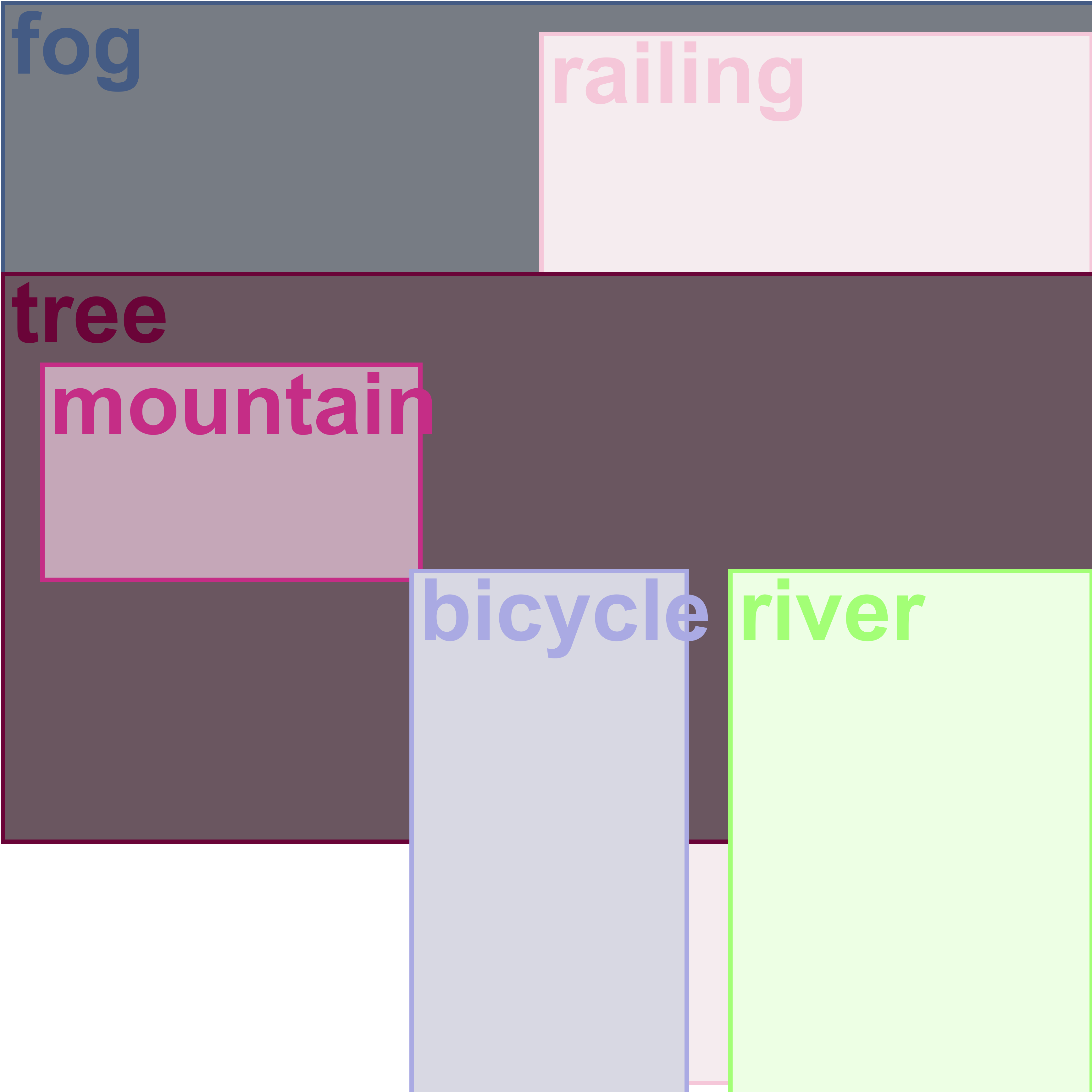}
\includegraphics[width=\lostganWidth]{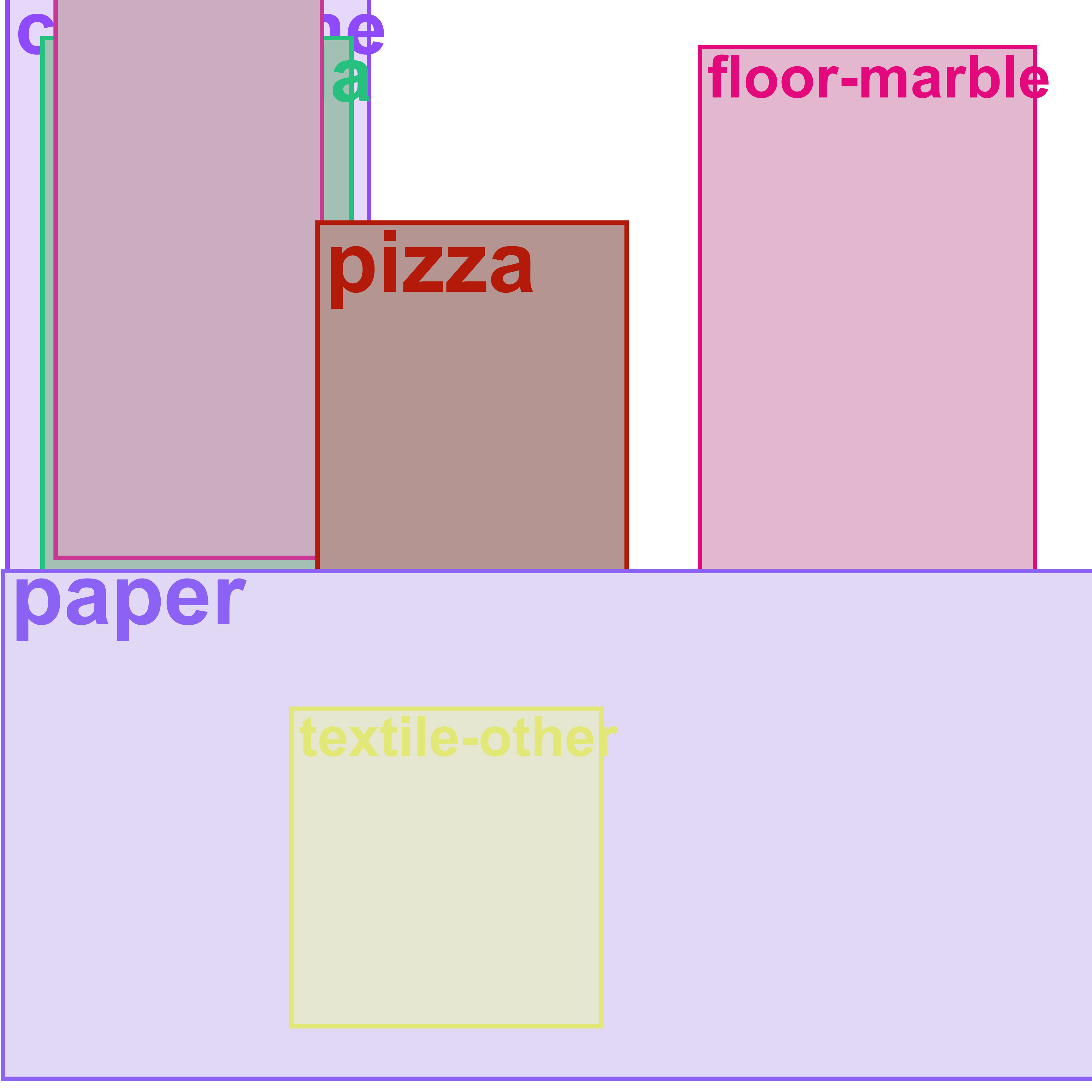}
\includegraphics[width=\lostganWidth]{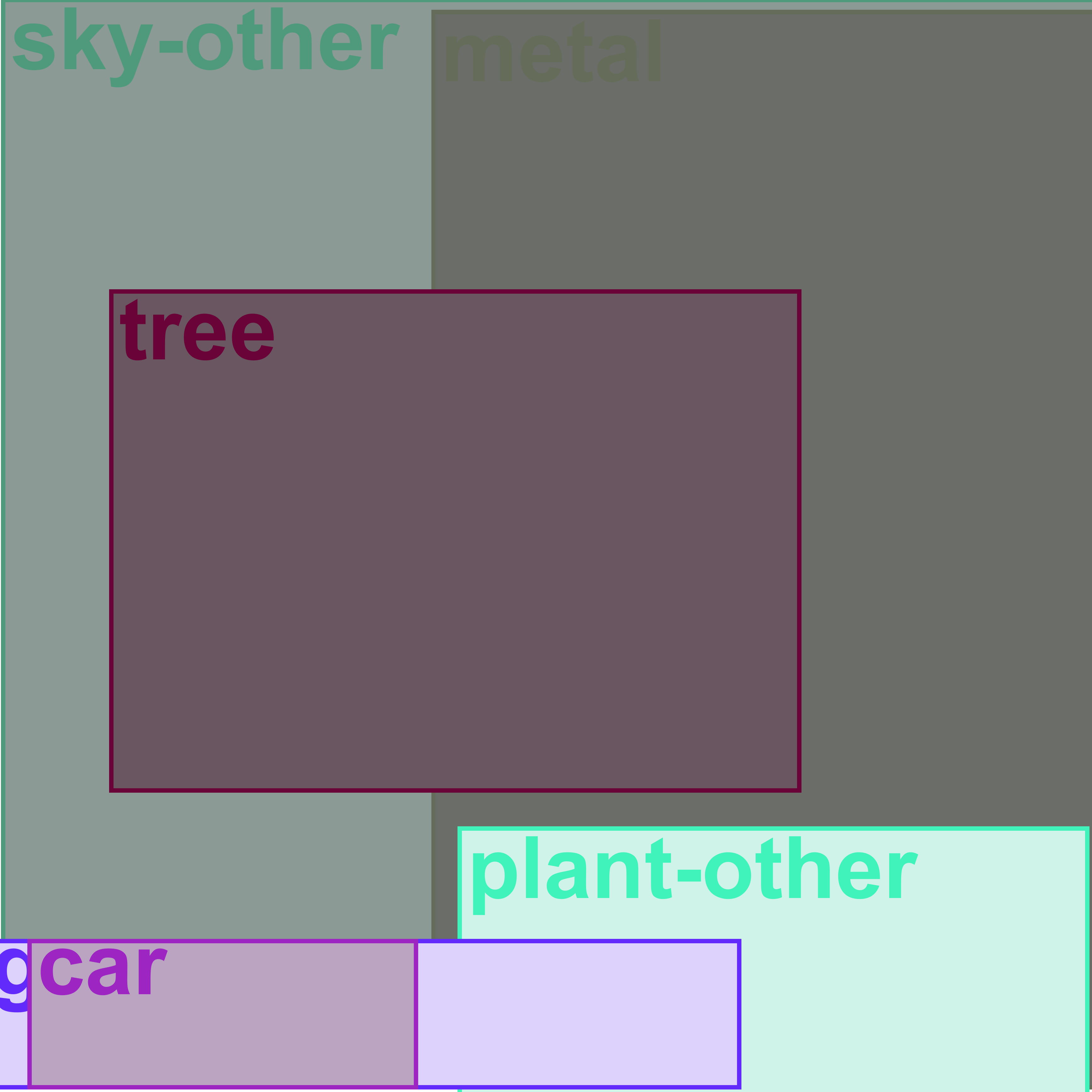}
\includegraphics[width=\lostganWidth]{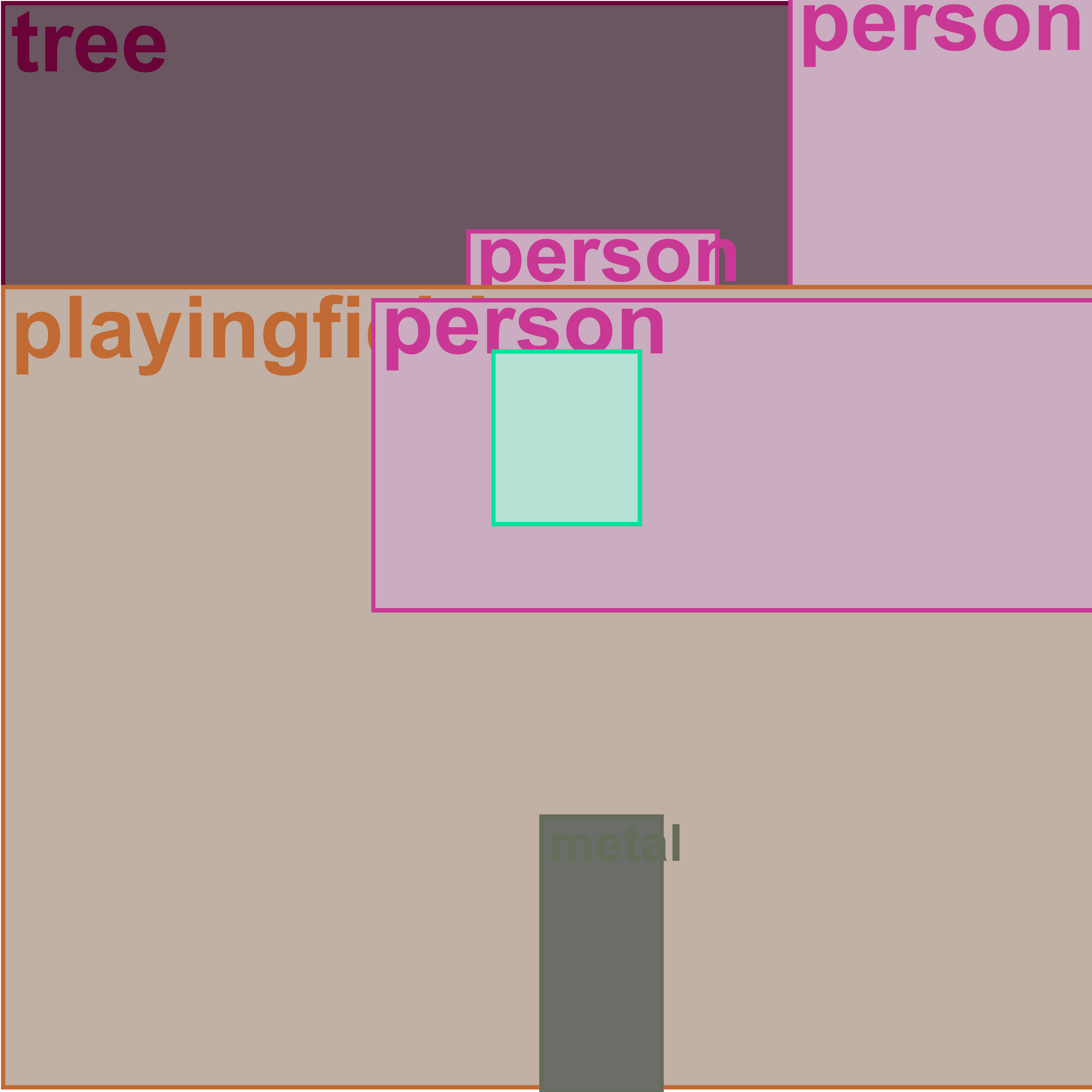}
\\

\includegraphics[width=\lostganWidth]{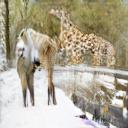}
\includegraphics[width=\lostganWidth]{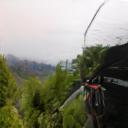}
\includegraphics[width=\lostganWidth]{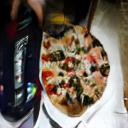}
\includegraphics[width=\lostganWidth]{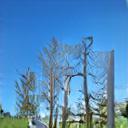}
\includegraphics[width=\lostganWidth]{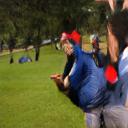}
    \caption{Generated layouts on COCO-Stuff (top) and images generated by LostGAN based on these layouts (bottom). The supplementary material shows more samples.}
    \label{fig:coco_stuff_samples}
\end{figure}

\begin{figure}
    \newlength{\sunrgbdWidth}
    \setlength{\sunrgbdWidth}{0.21\linewidth}
    \setlength{\tabcolsep}{2pt}
    \centering
    \begin{tabular}{ccccc}
        \rotatebox{90}{\hspace{.3cm}\small Synthetic} & 
        \includegraphics[width=\sunrgbdWidth,frame=0.1pt]{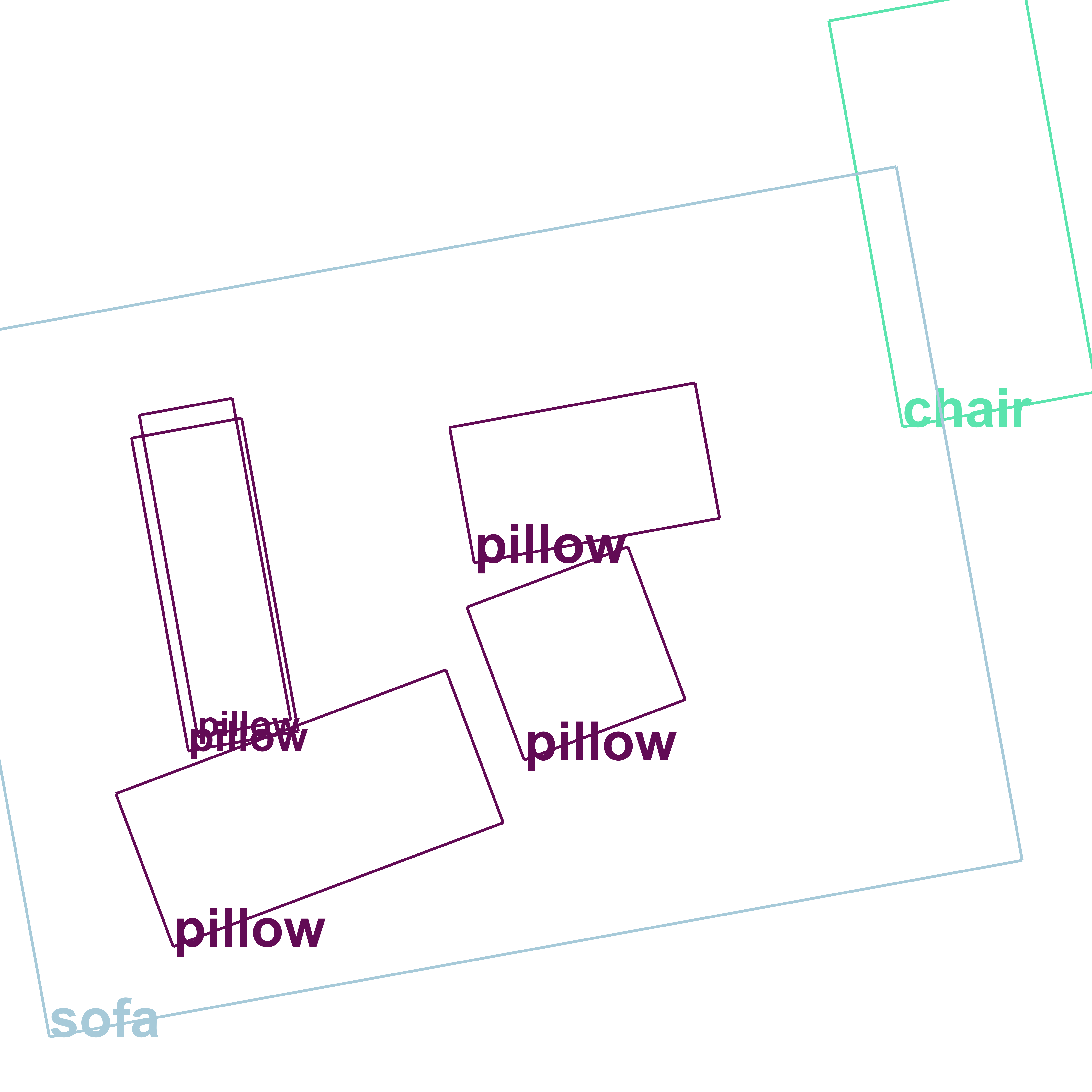} & 
        \includegraphics[width=\sunrgbdWidth,frame=0.1pt]{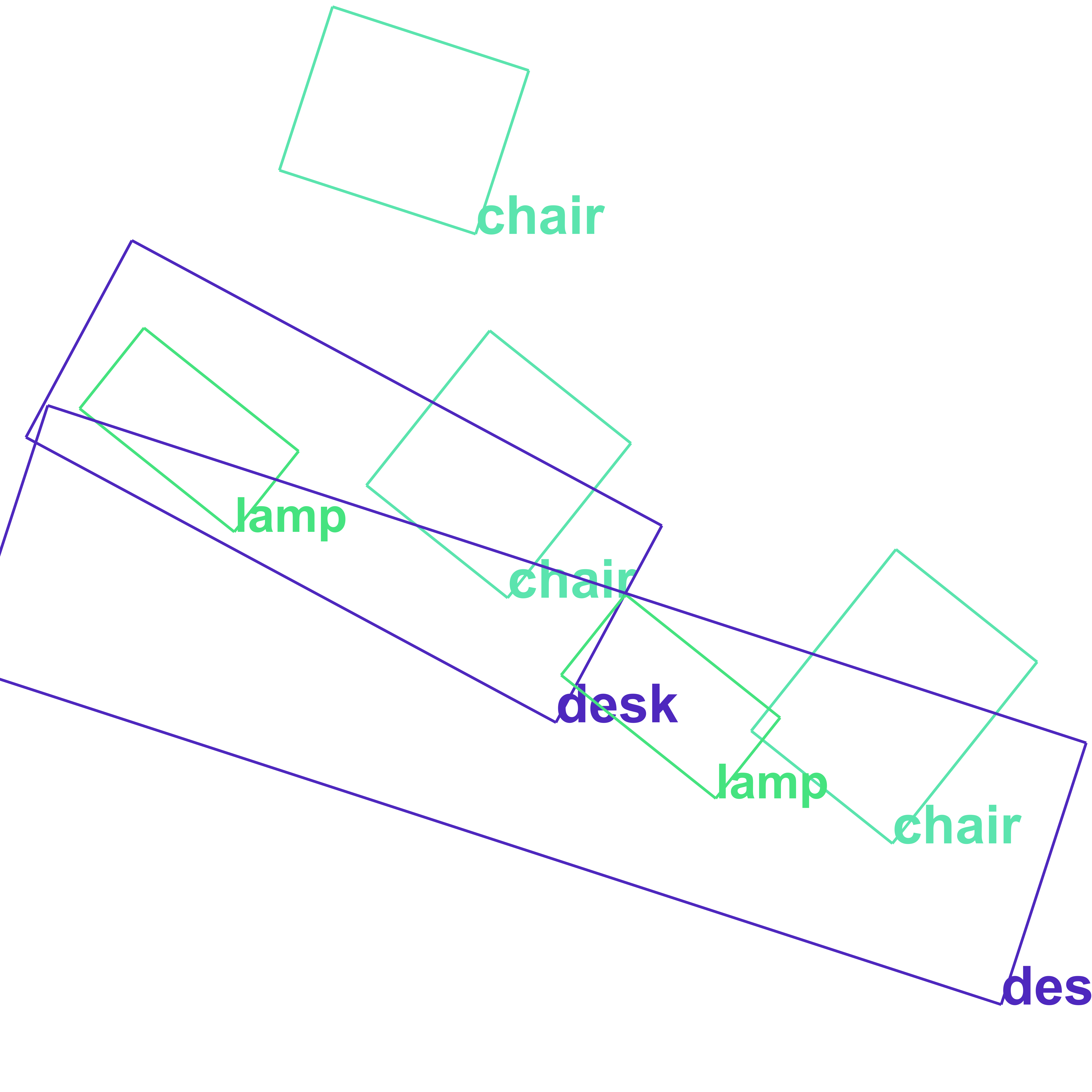} &
        \includegraphics[width=\sunrgbdWidth,frame=0.1pt]{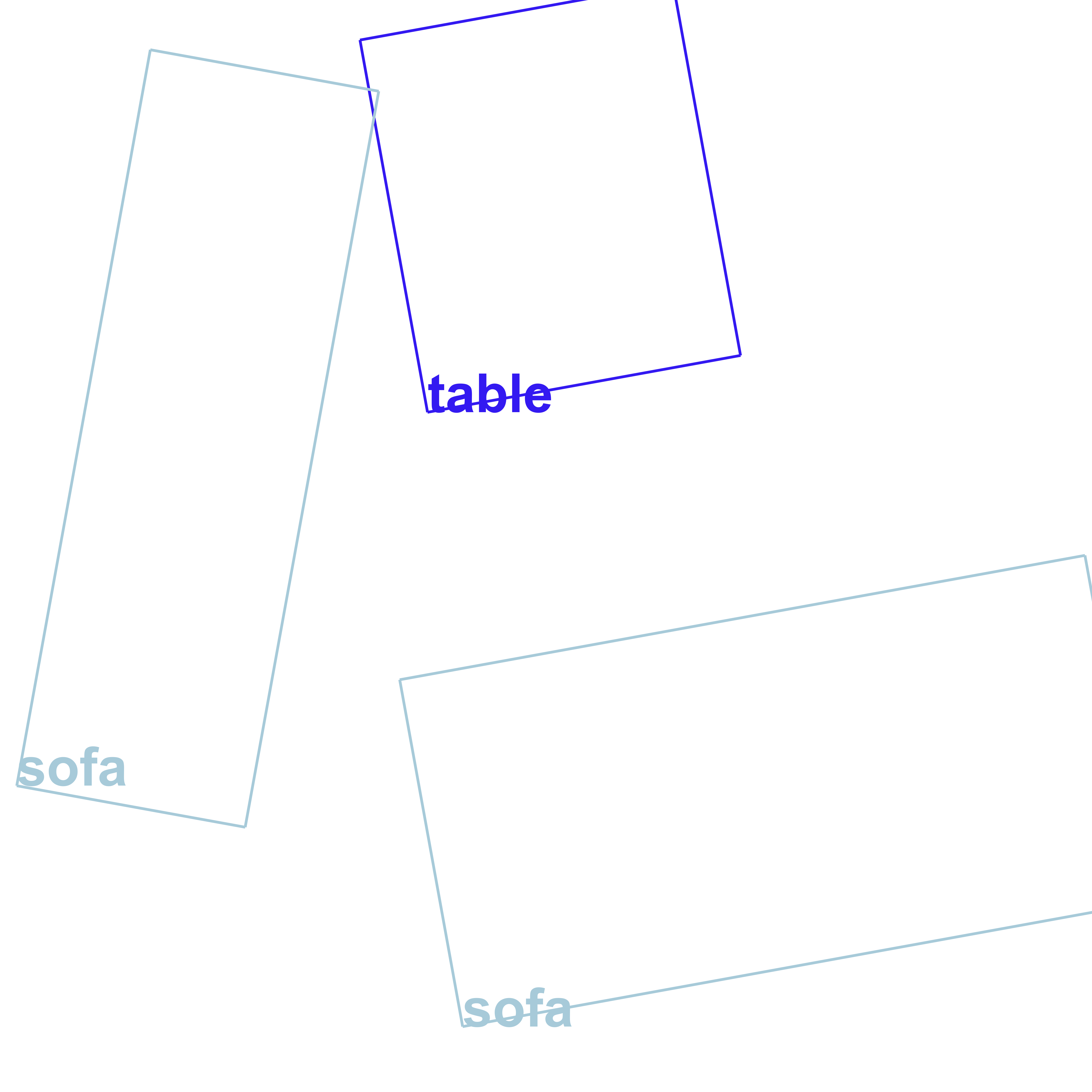} & 
        \includegraphics[width=\sunrgbdWidth,frame=0.1pt]{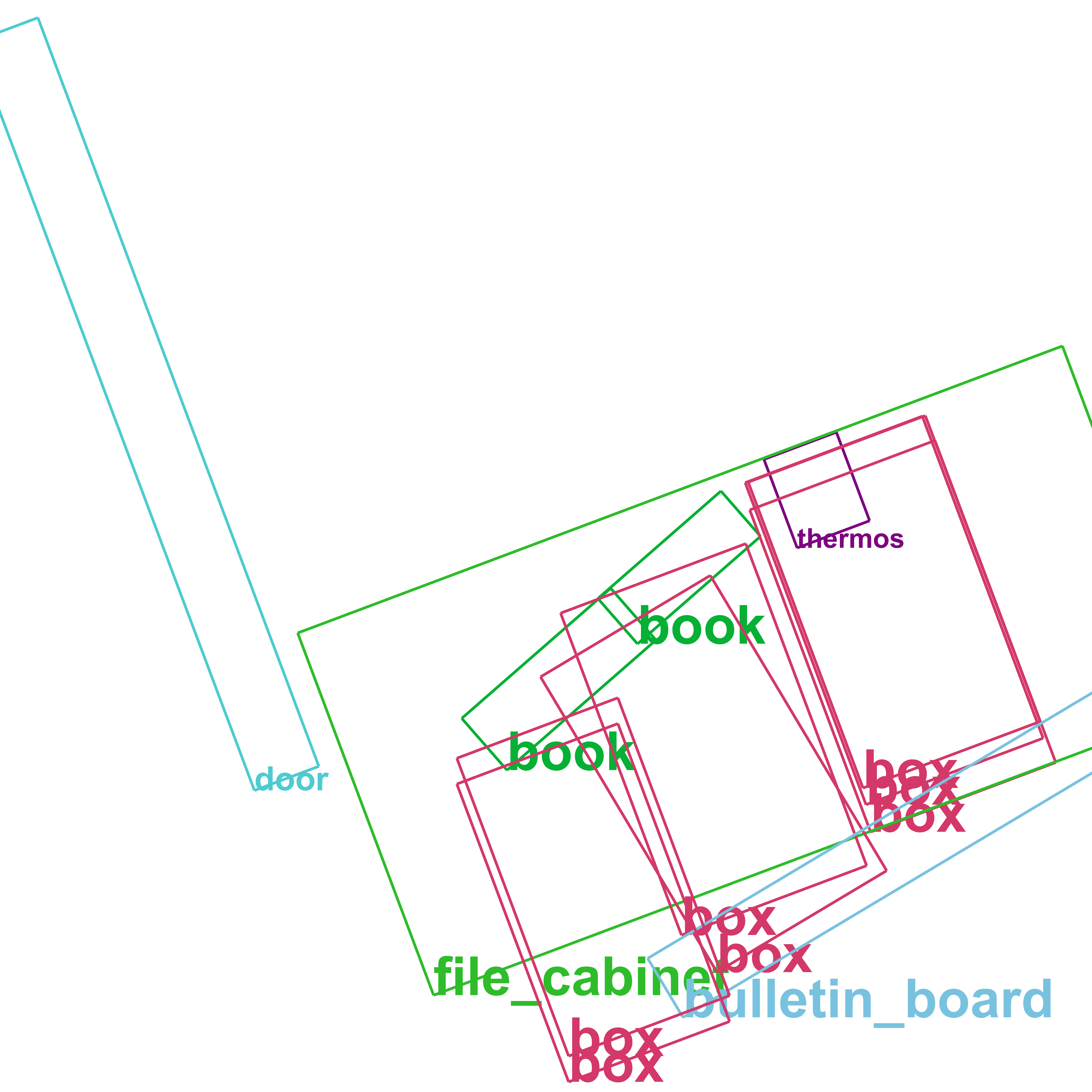}\\
        
        \rotatebox{90}{\hspace{.2cm}\small Real layout} & 
        \includegraphics[width=\sunrgbdWidth,frame=0.1pt]{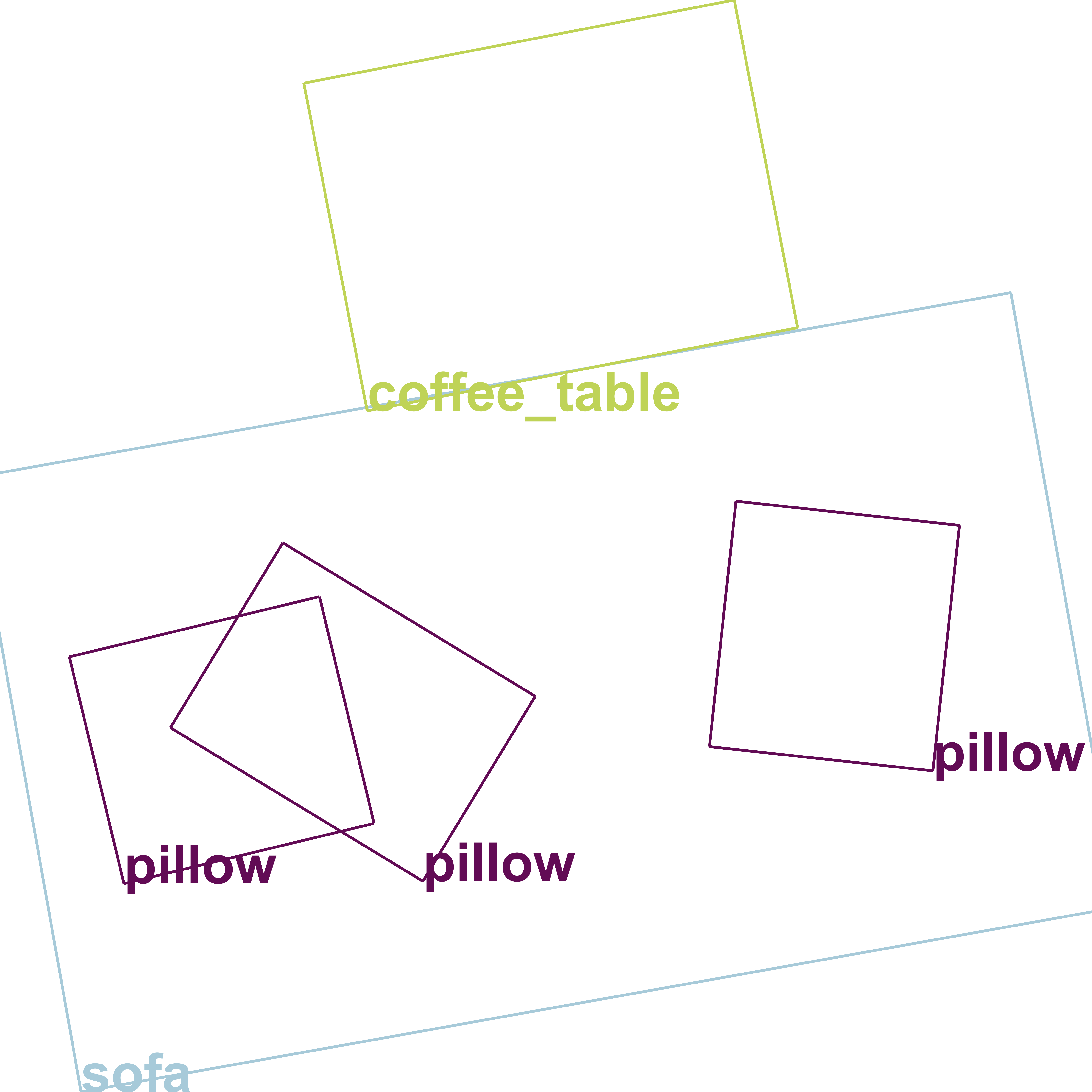} & 
        \includegraphics[width=\sunrgbdWidth,frame=0.1pt]{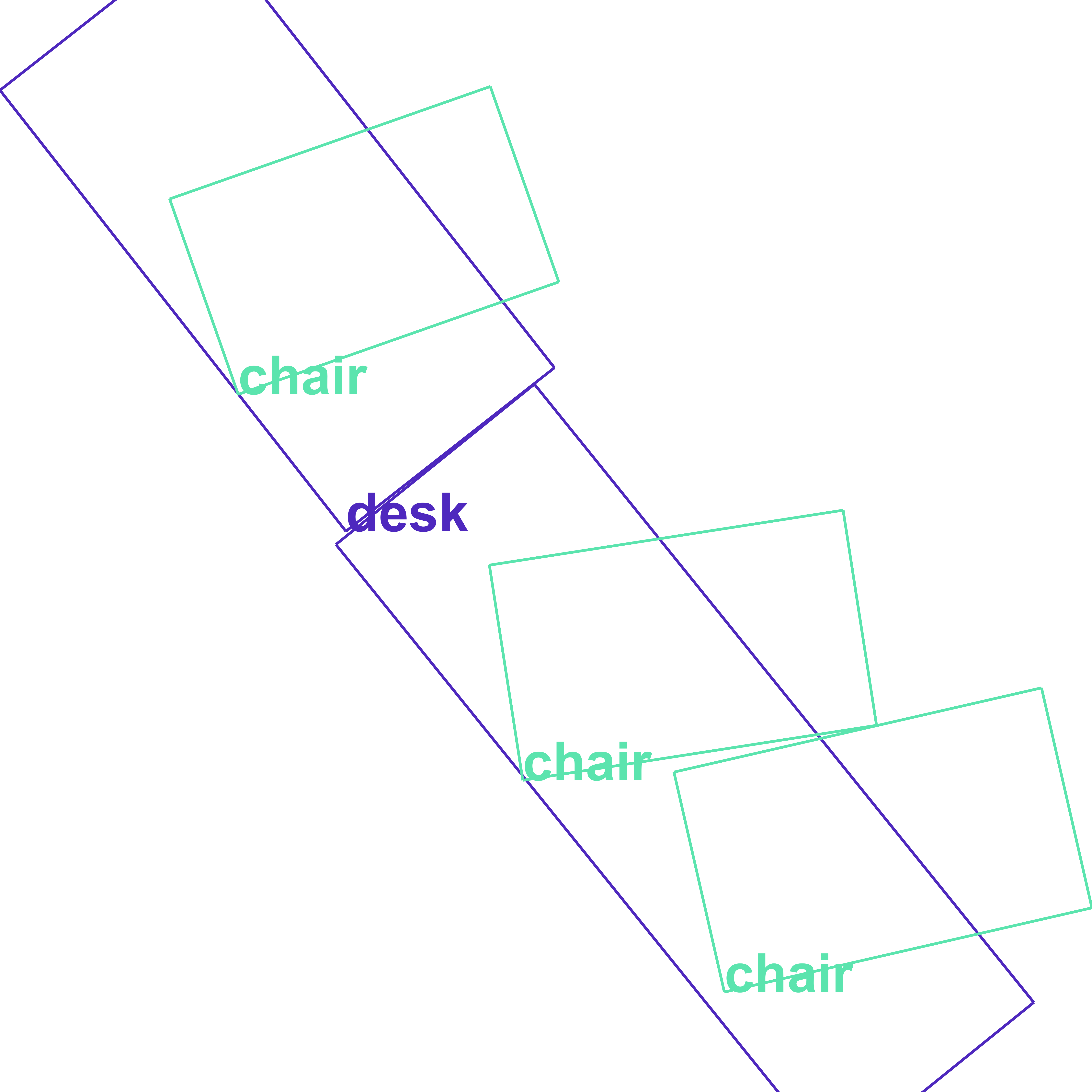} &
        \includegraphics[width=\sunrgbdWidth,frame=0.1pt]{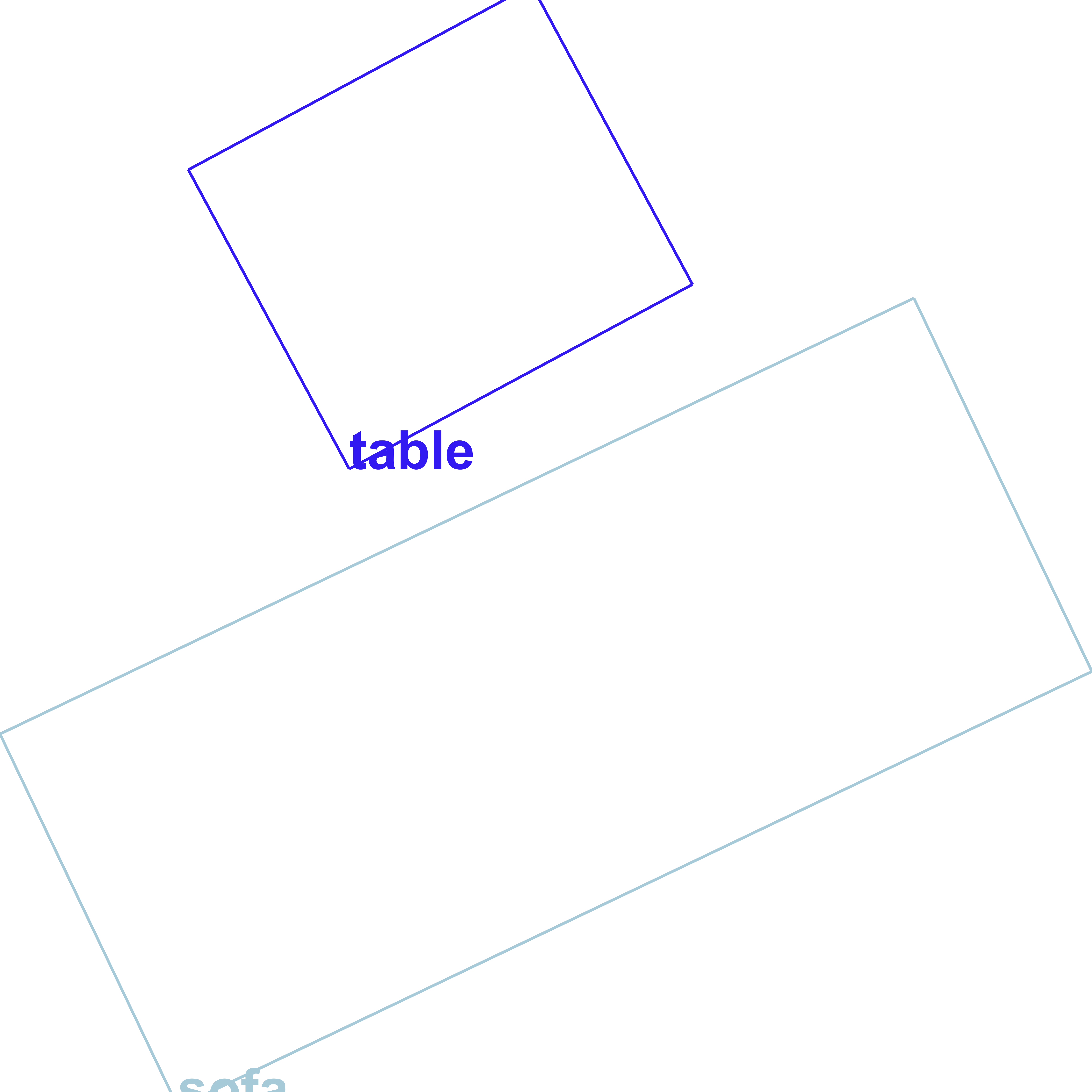} & 
        \includegraphics[width=\sunrgbdWidth,frame=0.1pt]{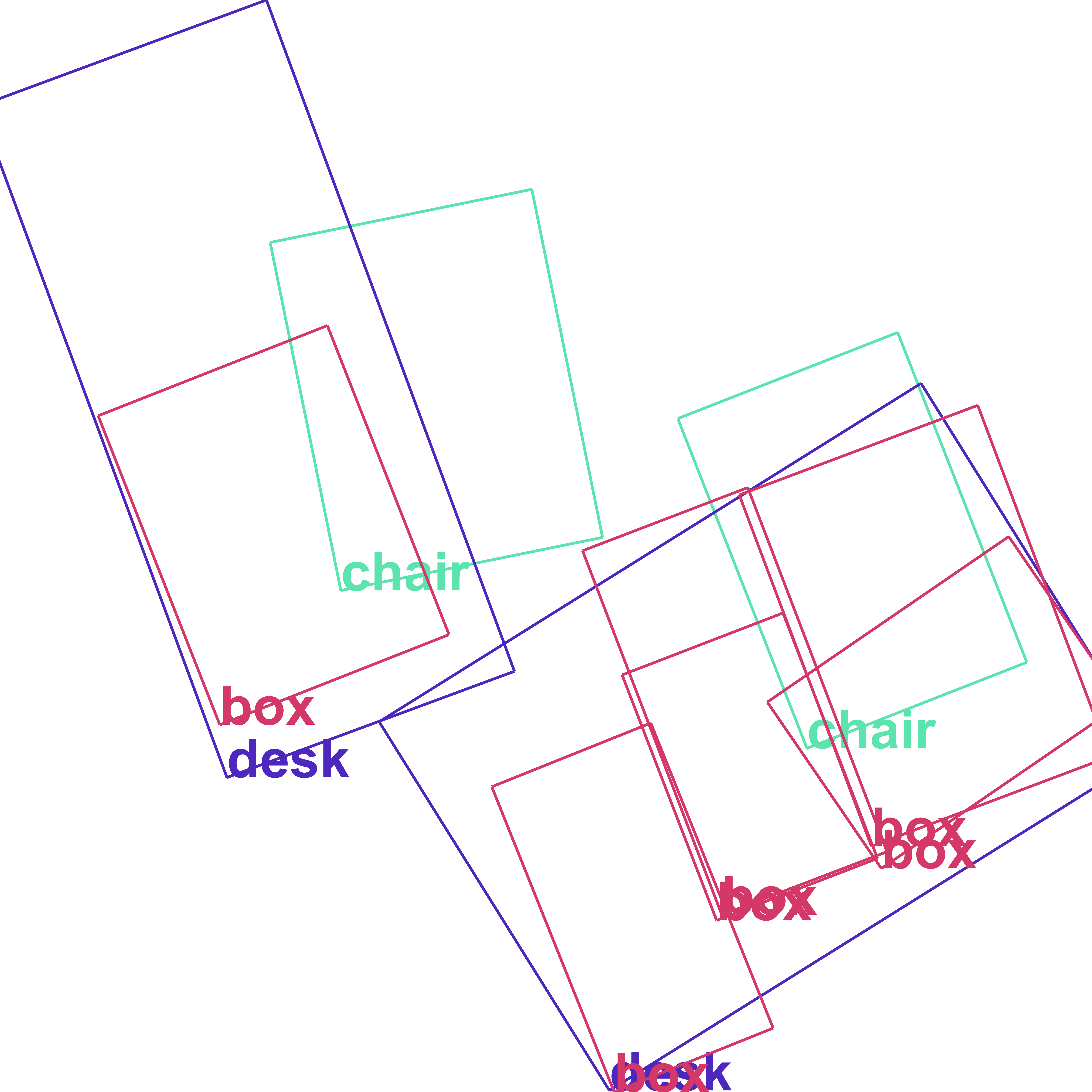}\\
        
        \rotatebox{90}{\small Real image} & 
        \includegraphics[width=\sunrgbdWidth]{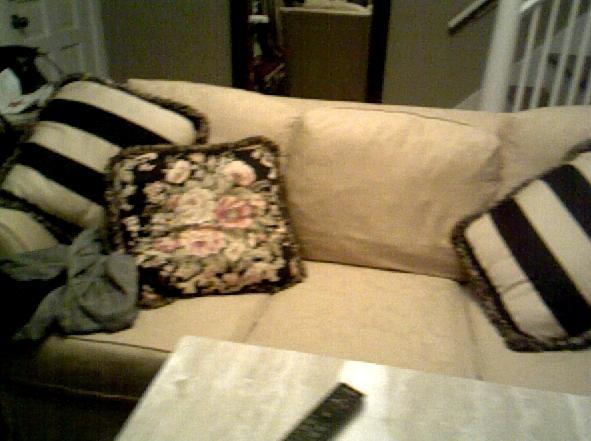} & 
        \includegraphics[width=\sunrgbdWidth]{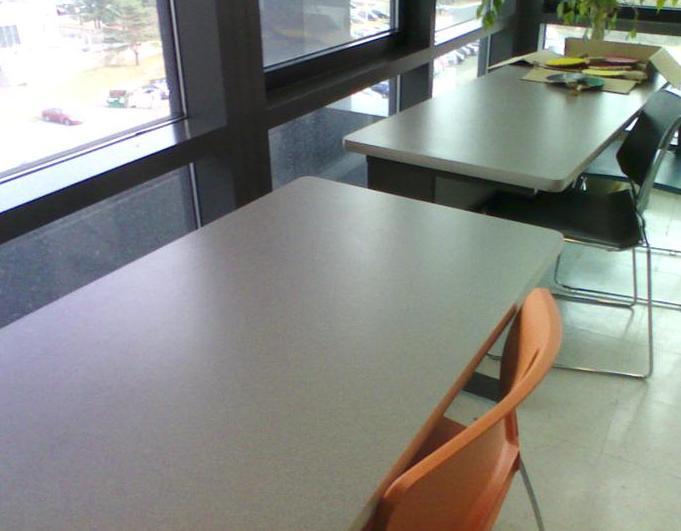} &
        \includegraphics[width=\sunrgbdWidth]{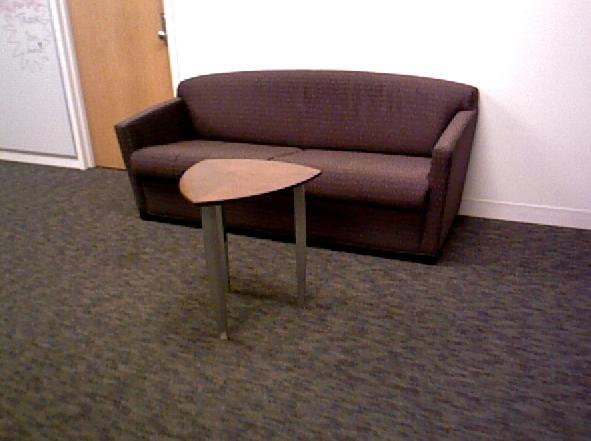} & 
        \includegraphics[width=\sunrgbdWidth]{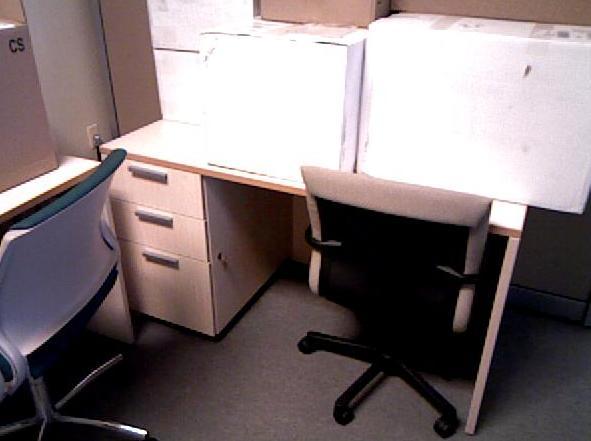}\\
    \end{tabular}
    \caption{Generated layouts for SUN RGB-D and their associated DocSim matches with the corresponding image.}
    \label{fig:samples_sunrgbd}
\end{figure}

We show qualitative results for PubLayNet in fig. \ref{fig:publaynet_samples}, as well as a qualitative comparison with existing methods in fig. \ref{fig:qualitative_comparision_publaynet}. In alignment with the quantitative results in section \ref{sec:quantitative_results}, we observe that our approach and \cite{2020arXiv200614615G} yield similar perceptual quality. Furthermore, LayoutVAE \cite{DBLP:conf/iccv/JyothiDHSM19} struggles to model layouts with a large number of elements. As previously discussed, this results from the application of \acp{rnn}, which are inferior at modeling the relationships between a large number of elements compared with the attention mechanism. In fig. \ref{fig:rico_samples} we show synthetic samples for RICO as well as the closest DocSim match in the real dataset. We show similar results for SUN RGB-D in fig. \ref{fig:samples_sunrgbd}. In order to show the capabilities of our method on the task of natural scene generation, we train our model on the COCO-Stuff dataset. In fig. \ref{fig:coco_stuff_samples} we show samples from our network. For better understanding we feed our generations to a pretrained instance of LostGAN \cite{DBLP:conf/iccv/SunW19}\footnote{\href{https://github.com/iVMCL/LostGANs}{https://github.com/iVMCL/LostGANs}}.
These results show that our method is capable of capturing relationships between elements regardless of their distance or position in the input sequence. This is observed by the strict margins modeled by our network, which resemble those of the real data. In the case of COCO or SUN RGB-D, we show how the network identifies joint occurrences of different elements (\eg \textit{giraffe} and \textit{tree}, \textit{person} and \textit{playingfield} or \textit{table} and \textit{chair}).

\subsection{Layout detection}\label{sec:layout_detection}

This experiment demonstrates the benefit of our approach regarding data augmentation for a downstream task. Document understanding comprises multiple tasks that go beyond simple \ac{ocr}. Understanding the arrangement of different pieces of text and images and their boundaries (the document \textit{layout}) is also necessary for applications such as text extraction or to determine the reading order in a complex document. While \ac{ocr}-annotated data is quite abundant, this is not the case for layout detection. Annotating documents is a tedious process which is prone to ambiguity, as the rules that define \eg what a paragraph is are often subjective. This ambiguity also makes automatic annotators based on heuristics fail or be constrained to specific domains \cite{zhong2019publaynet}. Most works, such as PubLayNet \cite{zhong2019publaynet}, LayoutLM \cite{DBLP:conf/kdd/XuL0HW020} or \cite{DBLP:conf/cvpr/YangYAKKG17} are based on object detection backbones using \acp{cnn}.
Here, we use our method to create a training dataset for a layout detector on the PubLayNet dataset. We use the bounding boxes generated by our method to guide the rendering. Obtaining realistic text, images, tables or lists for a given domain is labor-intensive, therefore, we crop these from the original dataset guided by the ground truth annotations and use the most appropriate one for a particular box according to its class and dimensionality. This approach ensures that the aspect ratio is preserved. In fig. \ref{fig:publaynet_samples} we show several examples of this approach.
We sample 240000 layouts from our model to train a Faster R-CNN model \cite{DBLP:conf/nips/RenHGS15} with a Resnet-50 backbone \cite{DBLP:conf/cvpr/HeZRS16} and evaluate the performance on the test set of PubLayNet in tab. \ref{tab:layout_detection_results}. We do not perform any postprocessing on the sampled layouts. For comparison, we run the same experiment with renderings created from real bounding box annotations (``Real layouts''), as well as with actual training images (``Real PubLayNet'').
We compare the mean average precision (mAP) at 0.5 IoU. Our synthesized layouts alone are capable of achieving a good accuracy score.

\begin{table}[h]
    \setlength{\tabcolsep}{4pt}
    \centering
    \begin{tabular}{ccc|c}
    &Ours & Real layouts & Real PubLayNet \\ \toprule
 mAP @ 0.5 IoU & 0.769 & 0.883 & 0.9646 \\
    \end{tabular}
    \caption{Detection accuracy scores for a layout detection model trained with synthetic and real data.}
    \label{tab:layout_detection_results}
\end{table}
\section{Conclusion and future work}

This work proposes self-attention layers as fundamental building blocks of a \ac{vae} and develops a solution tailored to layout synthesis, evaluating it on a diverse set of public datasets. Our approach yields state-of-art quantitative performance across all our metrics (see section \ref{sec:quantitative_results}) and layout samples of appealing perceptual quality (see section \ref{sec:qualitative_results}). We observe that autoregressive decoding constitutes an important ingredient to obtain high quality layouts. We also demonstrate its applicability as a data synthesizer for the downstream task of layout detection (see section \ref{sec:layout_detection}). However, we also note that our proposal can still be improved in promising future research directions.
Namely, learning to generate additional properties (\eg font or text size) or the dimensions of the layout, which could be useful for documents with varying size, (\eg, leaflets). Moreover, it could be interesting to incorporate an end-to-end approach for layout synthesis, such as ours, into a scene synthesis pipeline.

\clearpage
{\small
\bibliographystyle{ieee_fullname}
\bibliography{egbib}

\begin{thebibliography}{10}\itemsep=-1pt

\bibitem{tensorflow2015-whitepaper}
Mart\'{\i}n Abadi, Ashish Agarwal, Paul Barham, Eugene Brevdo, Zhifeng Chen,
  Craig Citro, Greg~S. Corrado, Andy Davis, Jeffrey Dean, Matthieu Devin,
  Sanjay Ghemawat, Ian Goodfellow, Andrew Harp, Geoffrey Irving, Michael Isard,
  Yangqing Jia, Rafal Jozefowicz, Lukasz Kaiser, Manjunath Kudlur, Josh
  Levenberg, Dandelion Man\'{e}, Rajat Monga, Sherry Moore, Derek Murray, Chris
  Olah, Mike Schuster, Jonathon Shlens, Benoit Steiner, Ilya Sutskever, Kunal
  Talwar, Paul Tucker, Vincent Vanhoucke, Vijay Vasudevan, Fernanda Vi\'{e}gas,
  Oriol Vinyals, Pete Warden, Martin Wattenberg, Martin Wicke, Yuan Yu, and
  Xiaoqiang Zheng.
\newblock {TensorFlow}: Large-scale machine learning on heterogeneous systems,
  2015.
\newblock Software available from tensorflow.org.

\bibitem{DBLP:conf/conll/BowmanVVDJB16}
Samuel~R. Bowman, Luke Vilnis, Oriol Vinyals, Andrew~M. Dai, Rafal
  J{\'{o}}zefowicz, and Samy Bengio.
\newblock Generating sentences from a continuous space.
\newblock In Yoav Goldberg and Stefan Riezler, editors, {\em Proceedings of the
  20th {SIGNLL} Conference on Computational Natural Language Learning, CoNLL
  2016, Berlin, Germany, August 11-12, 2016}, pages 10--21. {ACL}, 2016.

\bibitem{DBLP:conf/eccv/CarionMSUKZ20}
Nicolas Carion, Francisco Massa, Gabriel Synnaeve, Nicolas Usunier, Alexander
  Kirillov, and Sergey Zagoruyko.
\newblock End-to-end object detection with transformers.
\newblock In Andrea Vedaldi, Horst Bischof, Thomas Brox, and Jan{-}Michael
  Frahm, editors, {\em Computer Vision - {ECCV} 2020 - 16th European
  Conference, Glasgow, UK, August 23-28, 2020, Proceedings, Part {I}}, volume
  12346 of {\em Lecture Notes in Computer Science}, pages 213--229. Springer,
  2020.

\bibitem{chen2016variational}
Xi Chen, Diederik~P Kingma, Tim Salimans, Yan Duan, Prafulla Dhariwal, John
  Schulman, Ilya Sutskever, and Pieter Abbeel.
\newblock Variational lossy autoencoder.
\newblock {\em International Conference on Learning Representations}, 2017.

\bibitem{Deka:2017:Rico}
Biplab Deka, Zifeng Huang, Chad Franzen, Joshua Hibschman, Daniel Afergan, Yang
  Li, Jeffrey Nichols, and Ranjitha Kumar.
\newblock Rico: A mobile app dataset for building data-driven design
  applications.
\newblock In {\em Proceedings of the 30th Annual Symposium on User Interface
  Software and Technology}, UIST '17, 2017.

\bibitem{DBLP:conf/naacl/DevlinCLT19}
Jacob Devlin, Ming{-}Wei Chang, Kenton Lee, and Kristina Toutanova.
\newblock {BERT:} pre-training of deep bidirectional transformers for language
  understanding.
\newblock In Jill Burstein, Christy Doran, and Thamar Solorio, editors, {\em
  Proceedings of the 2019 Conference of the North American Chapter of the
  Association for Computational Linguistics: Human Language Technologies,
  {NAACL-HLT} 2019, Minneapolis, MN, USA, June 2-7, 2019, Volume 1 (Long and
  Short Papers)}, pages 4171--4186. Association for Computational Linguistics,
  2019.

\bibitem{DBLP:conf/icnn/GollerK96}
Christoph Goller and Andreas K{\"{u}}chler.
\newblock Learning task-dependent distributed representations by
  backpropagation through structure.
\newblock In {\em Proceedings of International Conference on Neural Networks
  (ICNN'96), Washington, DC, USA, June 3-6, 1996}, pages 347--352. {IEEE},
  1996.

\bibitem{2014arXiv1406.2661G}
Ian~J. {Goodfellow}, Jean {Pouget-Abadie}, Mehdi {Mirza}, Bing {Xu}, David
  {Warde-Farley}, Sherjil {Ozair}, Aaron {Courville}, and Yoshua {Bengio}.
\newblock {Generative Adversarial Networks}.
\newblock {\em Conference on Neural Information Processing Systems}, June 2014.

\bibitem{2020arXiv200614615G}
Kamal {Gupta}, Alessandro {Achille}, Justin {Lazarow}, Larry {Davis}, Vijay
  {Mahadevan}, and Abhinav {Shrivastava}.
\newblock {Layout Generation and Completion with Self-attention}.
\newblock {\em arXiv e-prints}, page arXiv:2006.14615, June 2020.

\bibitem{he2019lagging}
Junxian He, Daniel Spokoyny, Graham Neubig, and Taylor Berg-Kirkpatrick.
\newblock Lagging inference networks and posterior collapse in variational
  autoencoders.
\newblock {\em International Conference on Learning Representations}, 2019.

\bibitem{DBLP:conf/cvpr/HeZRS16}
Kaiming He, Xiangyu Zhang, Shaoqing Ren, and Jian Sun.
\newblock Deep residual learning for image recognition.
\newblock In {\em 2016 {IEEE} Conference on Computer Vision and Pattern
  Recognition, {CVPR} 2016, Las Vegas, NV, USA, June 27-30, 2016}, pages
  770--778. {IEEE} Computer Society, 2016.

\bibitem{2017arXiv171110939H}
Paul {Henderson}, Kartic {Subr}, and Vittorio {Ferrari}.
\newblock {Automatic Generation of Constrained Furniture Layouts}.
\newblock {\em arXiv e-prints}, page arXiv:1711.10939, Nov. 2017.

\bibitem{hochreiter_lstm}
Sepp Hochreiter and Jürgen Schmidhuber.
\newblock Long short-term memory.
\newblock {\em Neural computation}, 9:1735--80, 12 1997.

\bibitem{hochreiter1997long}
Sepp Hochreiter and J{\"u}rgen Schmidhuber.
\newblock Long short-term memory.
\newblock {\em Neural computation}, 9(8):1735--1780, 1997.

\bibitem{DBLP:conf/iccvw/JanochKJBFSD11}
Allison Janoch, Sergey Karayev, Yangqing Jia, Jonathan~T. Barron, Mario Fritz,
  Kate Saenko, and Trevor Darrell.
\newblock A category-level 3-d object dataset: Putting the kinect to work.
\newblock In {\em {IEEE} International Conference on Computer Vision Workshops,
  {ICCV} 2011 Workshops, Barcelona, Spain, November 6-13, 2011}, pages
  1168--1174. {IEEE} Computer Society, 2011.

\bibitem{DBLP:conf/iccv/JyothiDHSM19}
Akash~Abdu Jyothi, Thibaut Durand, Jiawei He, Leonid Sigal, and Greg Mori.
\newblock Layoutvae: Stochastic scene layout generation from a label set.
\newblock In {\em 2019 {IEEE/CVF} International Conference on Computer Vision,
  {ICCV} 2019, Seoul, Korea (South), October 27 - November 2, 2019}, pages
  9894--9903. {IEEE}, 2019.

\bibitem{DBLP:conf/cvpr/KarrasLA19}
Tero Karras, Samuli Laine, and Timo Aila.
\newblock A style-based generator architecture for generative adversarial
  networks.
\newblock In {\em {IEEE} Conference on Computer Vision and Pattern Recognition,
  {CVPR} 2019, Long Beach, CA, USA, June 16-20, 2019}, pages 4401--4410.
  Computer Vision Foundation / {IEEE}, 2019.

\bibitem{DBLP:journals/corr/KingmaB14}
Diederik~P. Kingma and Jimmy Ba.
\newblock Adam: {A} method for stochastic optimization.
\newblock In Yoshua Bengio and Yann LeCun, editors, {\em 3rd International
  Conference on Learning Representations, {ICLR} 2015, San Diego, CA, USA, May
  7-9, 2015, Conference Track Proceedings}, 2015.

\bibitem{kingma2016improved}
Durk~P Kingma, Tim Salimans, Rafal Jozefowicz, Xi Chen, Ilya Sutskever, and Max
  Welling.
\newblock Improved variational inference with inverse autoregressive flow.
\newblock In {\em Advances in neural information processing systems}, pages
  4743--4751, 2016.

\bibitem{kingma2015variational}
Durk~P Kingma, Tim Salimans, and Max Welling.
\newblock Variational dropout and the local reparameterization trick.
\newblock In {\em Advances in neural information processing systems}, pages
  2575--2583, 2015.

\bibitem{DBLP:journals/corr/KingmaW13}
Diederik~P. Kingma and Max Welling.
\newblock Auto-encoding variational bayes.
\newblock In Yoshua Bengio and Yann LeCun, editors, {\em 2nd International
  Conference on Learning Representations, {ICLR} 2014, Banff, AB, Canada, April
  14-16, 2014, Conference Track Proceedings}, 2014.

\bibitem{2020-Lee-NDNGLGWC}
Hsin-Ying Lee, Weilong Yang, Lu Jiang, Madison Le, Irfan Essa, Haifeng Gong,
  and Ming-Hsuan Yang.
\newblock Neural design network: Graphic layout generation with constraints.
\newblock In {\em {Proceedings of European Conference on Computer Vision
  (ECCV)}}, August 2020.

\bibitem{DBLP:conf/iclr/LiYHZX19}
Jianan Li, Jimei Yang, Aaron Hertzmann, Jianming Zhang, and Tingfa Xu.
\newblock Layoutgan: Generating graphic layouts with wireframe discriminators.
\newblock In {\em 7th International Conference on Learning Representations,
  {ICLR} 2019, New Orleans, LA, USA, May 6-9, 2019}. OpenReview.net, 2019.

\bibitem{DBLP:conf/eccv/LinMBHPRDZ14}
Tsung{-}Yi Lin, Michael Maire, Serge~J. Belongie, James Hays, Pietro Perona,
  Deva Ramanan, Piotr Doll{\'{a}}r, and C.~Lawrence Zitnick.
\newblock Microsoft {COCO:} common objects in context.
\newblock In David~J. Fleet, Tom{\'{a}}s Pajdla, Bernt Schiele, and Tinne
  Tuytelaars, editors, {\em Computer Vision - {ECCV} 2014 - 13th European
  Conference, Zurich, Switzerland, September 6-12, 2014, Proceedings, Part
  {V}}, volume 8693 of {\em Lecture Notes in Computer Science}, pages 740--755.
  Springer, 2014.

\bibitem{DBLP:journals/corr/abs-2003-12738}
Zhaojiang Lin, Genta~Indra Winata, Peng Xu, Zihan Liu, and Pascale Fung.
\newblock Variational transformers for diverse response generation.
\newblock {\em CoRR}, abs/2003.12738, 2020.

\bibitem{DBLP:conf/ijcnn/LiuL19a}
Danyang Liu and Gongshen Liu.
\newblock A transformer-based variational autoencoder for sentence generation.
\newblock In {\em International Joint Conference on Neural Networks, {IJCNN}
  2019 Budapest, Hungary, July 14-19, 2019}, pages 1--7. {IEEE}, 2019.

\bibitem{DBLP:journals/ai/MedressCFGKONNRRSWW77}
Mark~F. Medress, Franklin~S. Cooper, James~W. Forgie, C.~C. Green, Dennis~H.
  Klatt, Michael~H. O'Malley, Edward~P. Neuburg, Allen Newell, Raj Reddy,
  H.~Barry Ritea, J.~E. Shoup{-}Hummel, Donald~E. Walker, and William~A. Woods.
\newblock Speech understanding systems.
\newblock {\em Artif. Intell.}, 9(3):307--316, 1977.

\bibitem{DBLP:conf/cvpr/PatilBPA20}
Akshay~Gadi Patil, Omri Ben{-}Eliezer, Or Perel, and Hadar Averbuch{-}Elor.
\newblock {READ:} recursive autoencoders for document layout generation.
\newblock In {\em 2020 {IEEE/CVF} Conference on Computer Vision and Pattern
  Recognition, {CVPR} Workshops 2020, Seattle, WA, USA, June 14-19, 2020},
  pages 2316--2325. {IEEE}, 2020.

\bibitem{DBLP:conf/nips/RenHGS15}
Shaoqing Ren, Kaiming He, Ross~B. Girshick, and Jian Sun.
\newblock Faster {R-CNN:} towards real-time object detection with region
  proposal networks.
\newblock In Corinna Cortes, Neil~D. Lawrence, Daniel~D. Lee, Masashi Sugiyama,
  and Roman Garnett, editors, {\em Advances in Neural Information Processing
  Systems 28: Annual Conference on Neural Information Processing Systems 2015,
  December 7-12, 2015, Montreal, Quebec, Canada}, pages 91--99, 2015.

\bibitem{DBLP:conf/cvpr/Ritchie0L19}
Daniel Ritchie, Kai Wang, and Yu{-}An Lin.
\newblock Fast and flexible indoor scene synthesis via deep convolutional
  generative models.
\newblock In {\em {IEEE} Conference on Computer Vision and Pattern Recognition,
  {CVPR} 2019, Long Beach, CA, USA, June 16-20, 2019}, pages 6182--6190.
  Computer Vision Foundation / {IEEE}, 2019.

\bibitem{DBLP:conf/eccv/SilbermanHKF12}
Nathan Silberman, Derek Hoiem, Pushmeet Kohli, and Rob Fergus.
\newblock Indoor segmentation and support inference from {RGBD} images.
\newblock In Andrew~W. Fitzgibbon, Svetlana Lazebnik, Pietro Perona, Yoichi
  Sato, and Cordelia Schmid, editors, {\em Computer Vision - {ECCV} 2012 - 12th
  European Conference on Computer Vision, Florence, Italy, October 7-13, 2012,
  Proceedings, Part {V}}, volume 7576 of {\em Lecture Notes in Computer
  Science}, pages 746--760. Springer, 2012.

\bibitem{DBLP:conf/cvpr/SongLX15}
Shuran Song, Samuel~P. Lichtenberg, and Jianxiong Xiao.
\newblock {SUN} {RGB-D:} {A} {RGB-D} scene understanding benchmark suite.
\newblock In {\em {IEEE} Conference on Computer Vision and Pattern Recognition,
  {CVPR} 2015, Boston, MA, USA, June 7-12, 2015}, pages 567--576. {IEEE}
  Computer Society, 2015.

\bibitem{DBLP:conf/cvpr/SongYZCSF17}
Shuran Song, Fisher Yu, Andy Zeng, Angel~X. Chang, Manolis Savva, and Thomas~A.
  Funkhouser.
\newblock Semantic scene completion from a single depth image.
\newblock In {\em 2017 {IEEE} Conference on Computer Vision and Pattern
  Recognition, {CVPR} 2017, Honolulu, HI, USA, July 21-26, 2017}, pages
  190--198. {IEEE} Computer Society, 2017.

\bibitem{DBLP:conf/iccv/SunW19}
Wei Sun and Tianfu Wu.
\newblock Image synthesis from reconfigurable layout and style.
\newblock In {\em 2019 {IEEE/CVF} International Conference on Computer Vision,
  {ICCV} 2019, Seoul, Korea (South), October 27 - November 2, 2019}, pages
  10530--10539. {IEEE}, 2019.

\bibitem{DBLP:conf/nips/VaswaniSPUJGKP17}
Ashish Vaswani, Noam Shazeer, Niki Parmar, Jakob Uszkoreit, Llion Jones,
  Aidan~N. Gomez, Lukasz Kaiser, and Illia Polosukhin.
\newblock Attention is all you need.
\newblock In Isabelle Guyon, Ulrike von Luxburg, Samy Bengio, Hanna~M. Wallach,
  Rob Fergus, S.~V.~N. Vishwanathan, and Roman Garnett, editors, {\em Advances
  in Neural Information Processing Systems 30: Annual Conference on Neural
  Information Processing Systems 2017, 4-9 December 2017, Long Beach, CA,
  {USA}}, pages 5998--6008, 2017.

\bibitem{wang2018deep}
Kai Wang, Manolis Savva, Angel~X Chang, and Daniel Ritchie.
\newblock Deep convolutional priors for indoor scene synthesis.
\newblock {\em ACM Transactions on Graphics (TOG)}, 37(4):70, 2018.

\bibitem{DBLP:conf/ijcai/Wang019b}
Tianming Wang and Xiaojun Wan.
\newblock {T-CVAE:} transformer-based conditioned variational autoencoder for
  story completion.
\newblock In Sarit Kraus, editor, {\em Proceedings of the Twenty-Eighth
  International Joint Conference on Artificial Intelligence, {IJCAI} 2019,
  Macao, China, August 10-16, 2019}, pages 5233--5239. ijcai.org, 2019.

\bibitem{DBLP:conf/iccv/XiaoOT13}
Jianxiong Xiao, Andrew Owens, and Antonio Torralba.
\newblock {SUN3D:} {A} database of big spaces reconstructed using sfm and
  object labels.
\newblock In {\em {IEEE} International Conference on Computer Vision, {ICCV}
  2013, Sydney, Australia, December 1-8, 2013}, pages 1625--1632. {IEEE}
  Computer Society, 2013.

\bibitem{zheng-sig19}
Ying~Cao Xinru~Zheng, Xiaotian~Qiao and Rynson~W.H. Lau.
\newblock Content-aware generative modeling of graphic design layouts.
\newblock {\em ACM Transactions on Graphics (Proc. of SIGGRAPH 2019)}, 38,
  2019.

\bibitem{DBLP:conf/kdd/XuL0HW020}
Yiheng Xu, Minghao Li, Lei Cui, Shaohan Huang, Furu Wei, and Ming Zhou.
\newblock Layoutlm: Pre-training of text and layout for document image
  understanding.
\newblock In Rajesh Gupta, Yan Liu, Jiliang Tang, and B.~Aditya Prakash,
  editors, {\em {KDD} '20: The 26th {ACM} {SIGKDD} Conference on Knowledge
  Discovery and Data Mining, Virtual Event, CA, USA, August 23-27, 2020}, pages
  1192--1200. {ACM}, 2020.

\bibitem{DBLP:conf/cvpr/YangYAKKG17}
Xiao Yang, Ersin Yumer, Paul Asente, Mike Kraley, Daniel Kifer, and C.~Lee
  Giles.
\newblock Learning to extract semantic structure from documents using
  multimodal fully convolutional neural networks.
\newblock In {\em 2017 {IEEE} Conference on Computer Vision and Pattern
  Recognition, {CVPR} 2017, Honolulu, HI, USA, July 21-26, 2017}, pages
  4342--4351. {IEEE} Computer Society, 2017.

\bibitem{zhong2019publaynet}
Xu Zhong, Jianbin Tang, and Antonio~Jimeno Yepes.
\newblock Publaynet: largest dataset ever for document layout analysis.
\newblock In {\em 2019 International Conference on Document Analysis and
  Recognition (ICDAR)}, pages 1015--1022. IEEE, Sep. 2019.

\bibitem{CycleGAN2017}
Jun-Yan Zhu, Taesung Park, Phillip Isola, and Alexei~A Efros.
\newblock Unpaired image-to-image translation using cycle-consistent
  adversarial networks.
\newblock In {\em Computer Vision (ICCV), 2017 IEEE International Conference
  on}, 2017.

\end{thebibliography}


\begin{thebibliography}{1}\itemsep=-1pt

\bibitem{ni-etal-2019-justifying}
Jianmo Ni, Jiacheng Li, and Julian McAuley.
\newblock Justifying recommendations using distantly-labeled reviews and
  fine-grained aspects.
\newblock In {\em Proceedings of the 2019 Conference on Empirical Methods in
  Natural Language Processing and the 9th International Joint Conference on
  Natural Language Processing (EMNLP-IJCNLP)}, pages 188--197, Hong Kong,
  China, Nov. 2019. Association for Computational Linguistics.

\end{thebibliography}
}

\clearpage

\end{document}

%%%%%%%%% TITLE
\title{Supplementary: Variational Transformer Networks for Layout Generation}
\author{Diego Martin Arroyo\textsuperscript{1}\\
{\tt\small martinarroyo@google.com}\\
\textsuperscript{1}Google, Inc\\
\and
Janis Postels\textsuperscript{2}\\
{\tt\small jpostels@vision.ee.ethz.ch}\\
\textsuperscript{2}ETH Z\"urich\\
\and
Federico Tombari\textsuperscript{1,3}\\
{\tt\small tombari@google.com}\\
\textsuperscript{3}Technische Universit\"at M\"unchen\\
}

\onecolumn
\maketitle

%\tableofcontents

\section{Attention analysis}
The main claim for the effectiveness of our method is its inductive bias towards the relationships between elements. The self-attention layers in our network weigh the relevance of each component regardless of their distance in the input sequence. In this section we analyze the validity of this claim by observing the attention maps on each self-attention layer.

\subsection{Encoder}
\begin{figure}[h]
\setlength{\tabcolsep}{1pt}
\newlength{\attentionVisDimmedWidth}
\setlength{\attentionVisDimmedWidth}{0.061\linewidth}
    \centering
    \begin{tabular}{lccccccccccccccc}
&\multicolumn{15}{c}{Final encoding result}\\
&\multicolumn{15}{c}{\includegraphics[width=0.0822\linewidth,frame=.1pt]{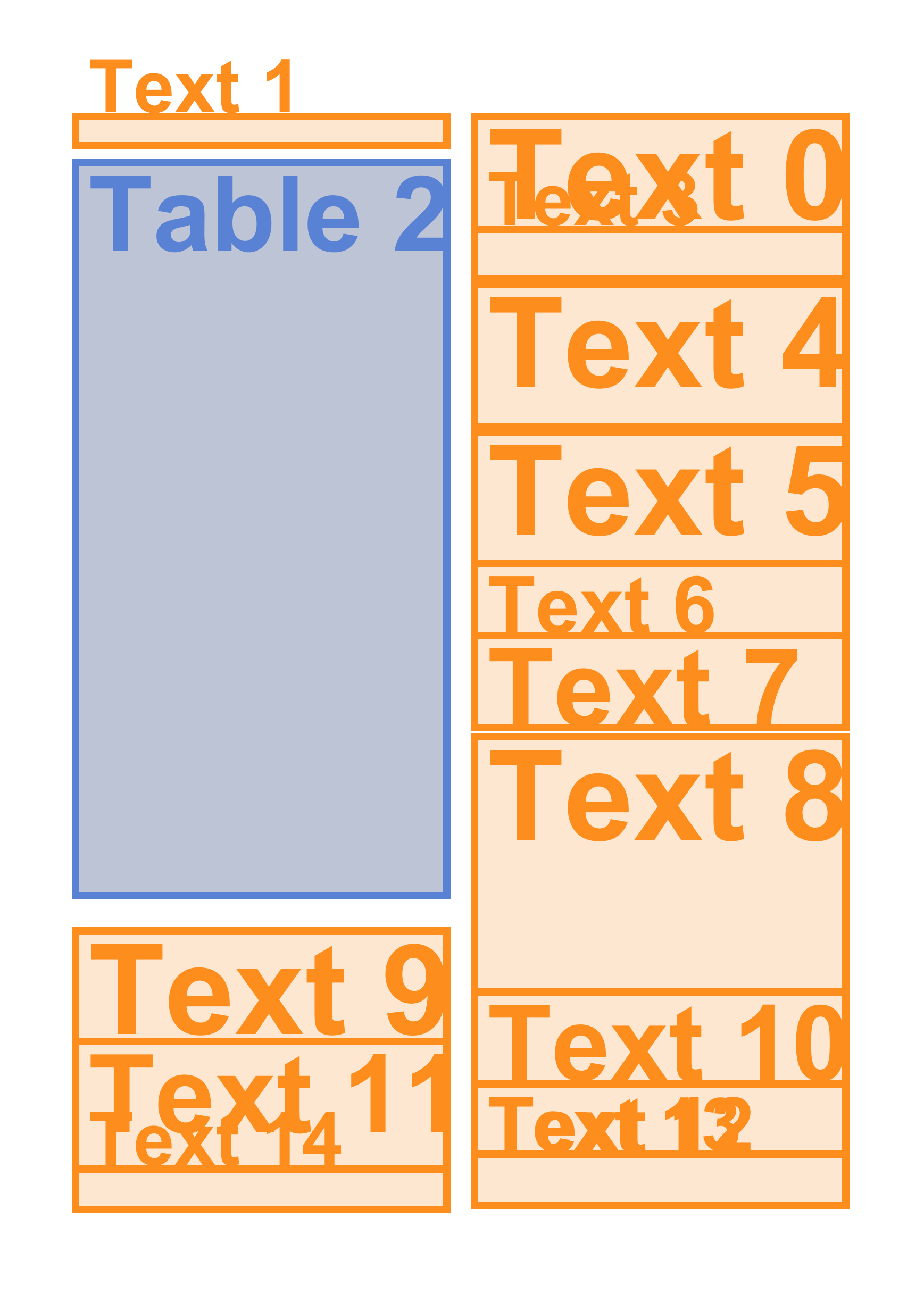}}\\ \toprule
&\multicolumn{15}{c}{Attention}\\ \toprule
    & 
    \footnotesize Text 0 &
    \footnotesize Text 1 &
    \footnotesize List 2 &
    \footnotesize Text 3 &
    \footnotesize Text 4 &
    \footnotesize Text 5 &
    \footnotesize Text 6 &
    \footnotesize Text 7 &
    \footnotesize Text 8 &
    \footnotesize Text 9 &
    \footnotesize Text 10 &
    \footnotesize Text 11 &
    \footnotesize Text 12 &
    \footnotesize Text 13 &
    \footnotesize Text 14\\
    \rotatebox{90}{\hspace{0.3cm}\footnotesize Layer 1} & \includegraphics[width=\attentionVisDimmedWidth,frame=0.1pt]{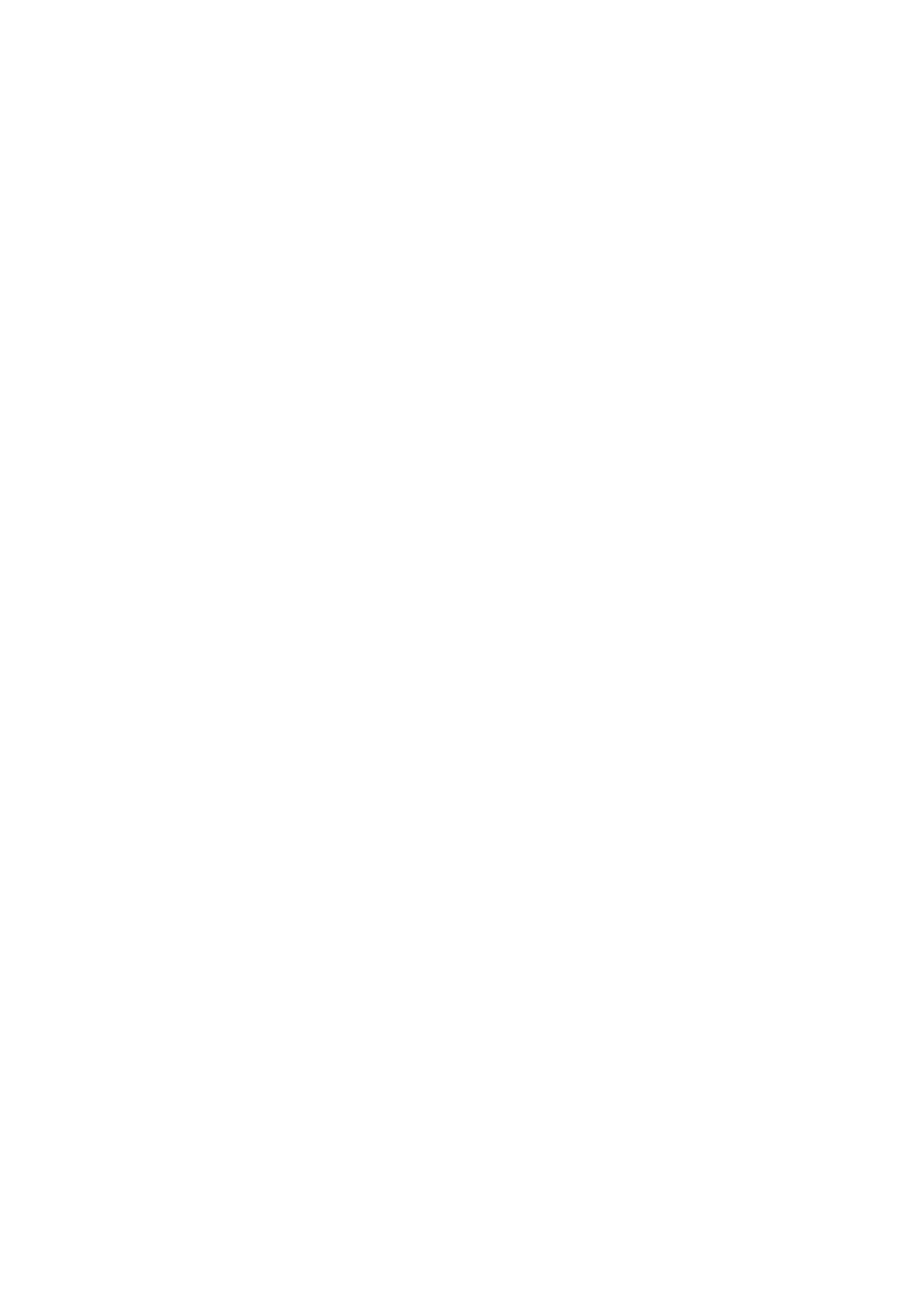} &
    \includegraphics[width=\attentionVisDimmedWidth,frame=0.1pt]{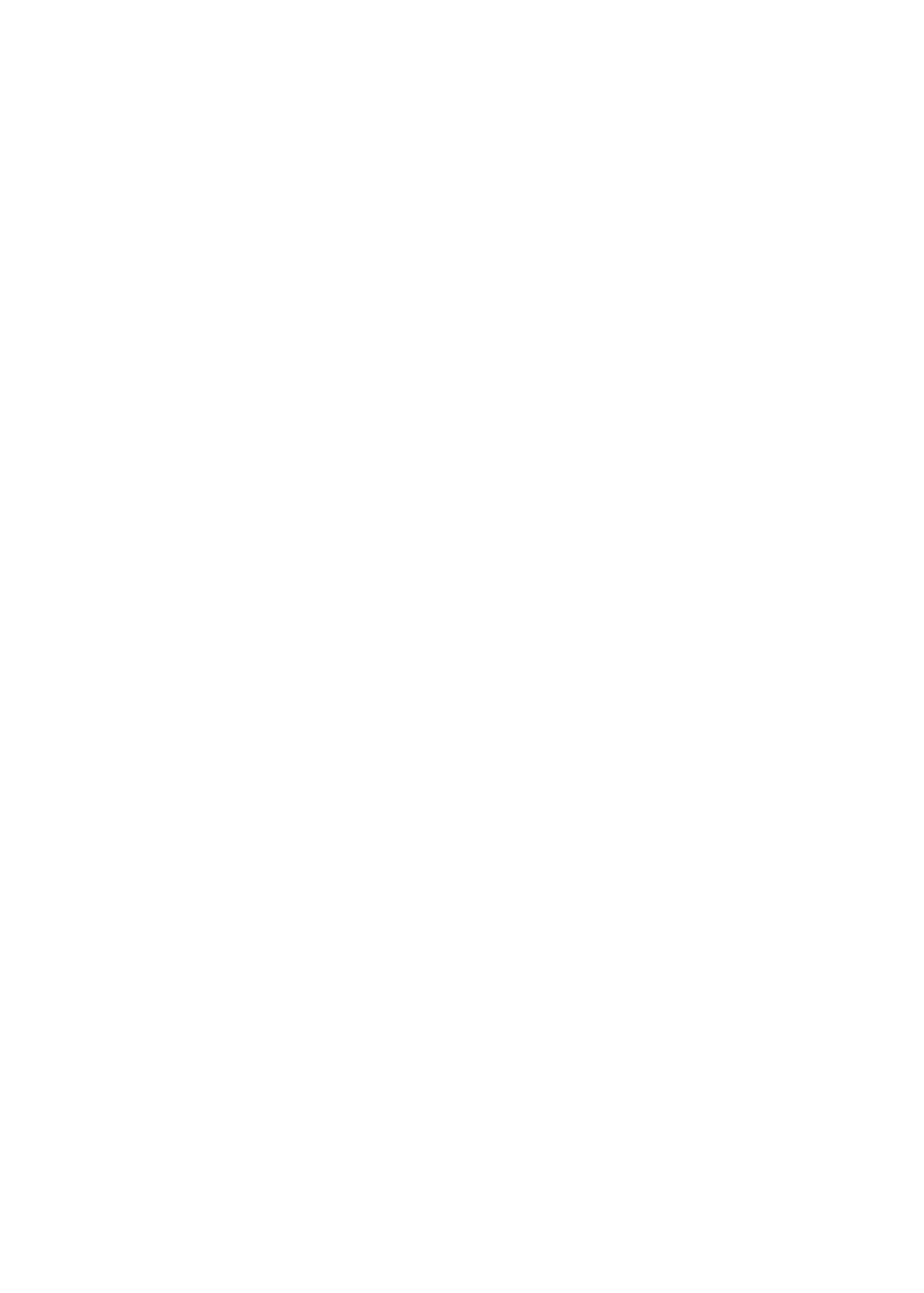} &
    \includegraphics[width=\attentionVisDimmedWidth,frame=0.1pt]{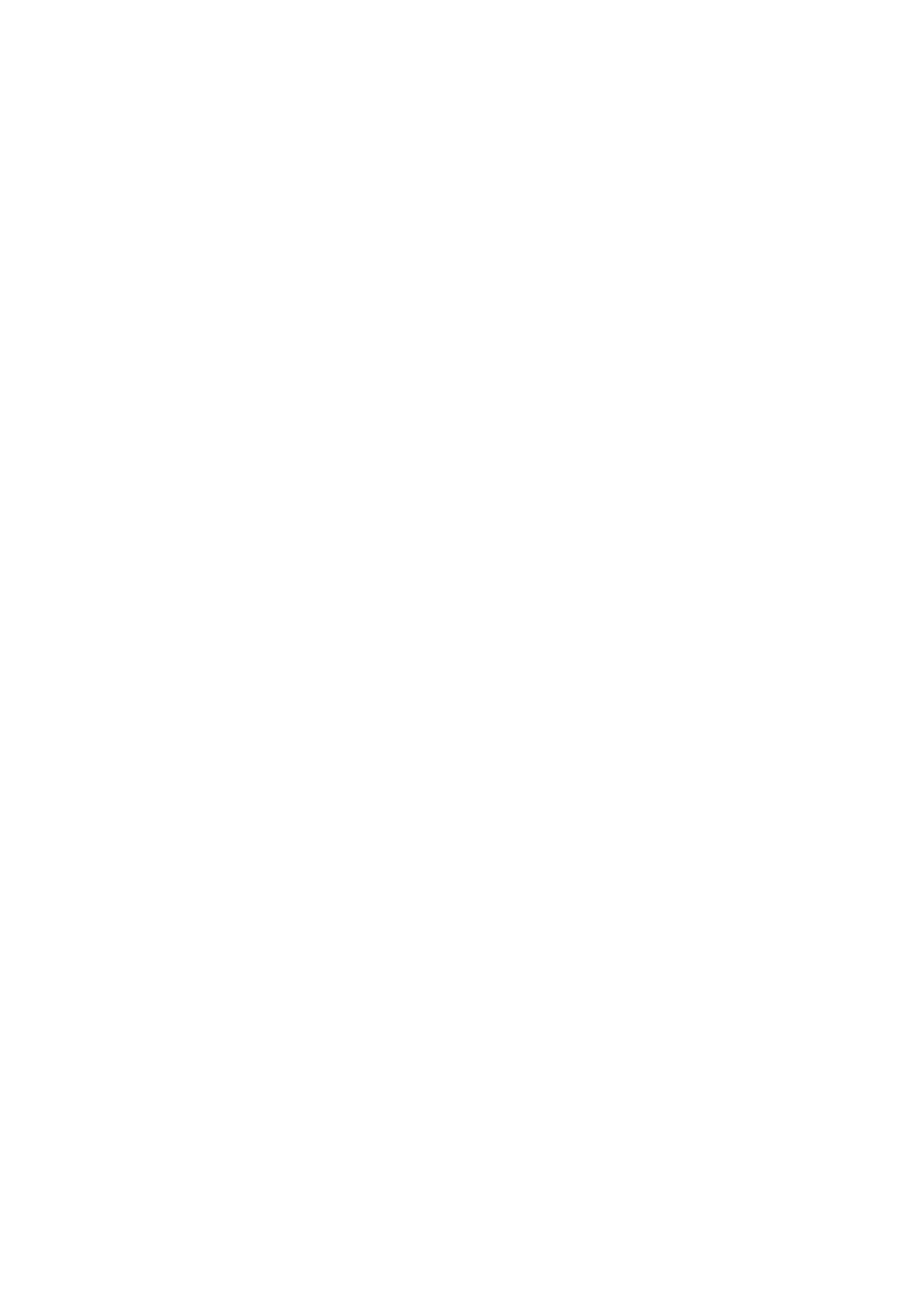} &
    \includegraphics[width=\attentionVisDimmedWidth,frame=0.1pt]{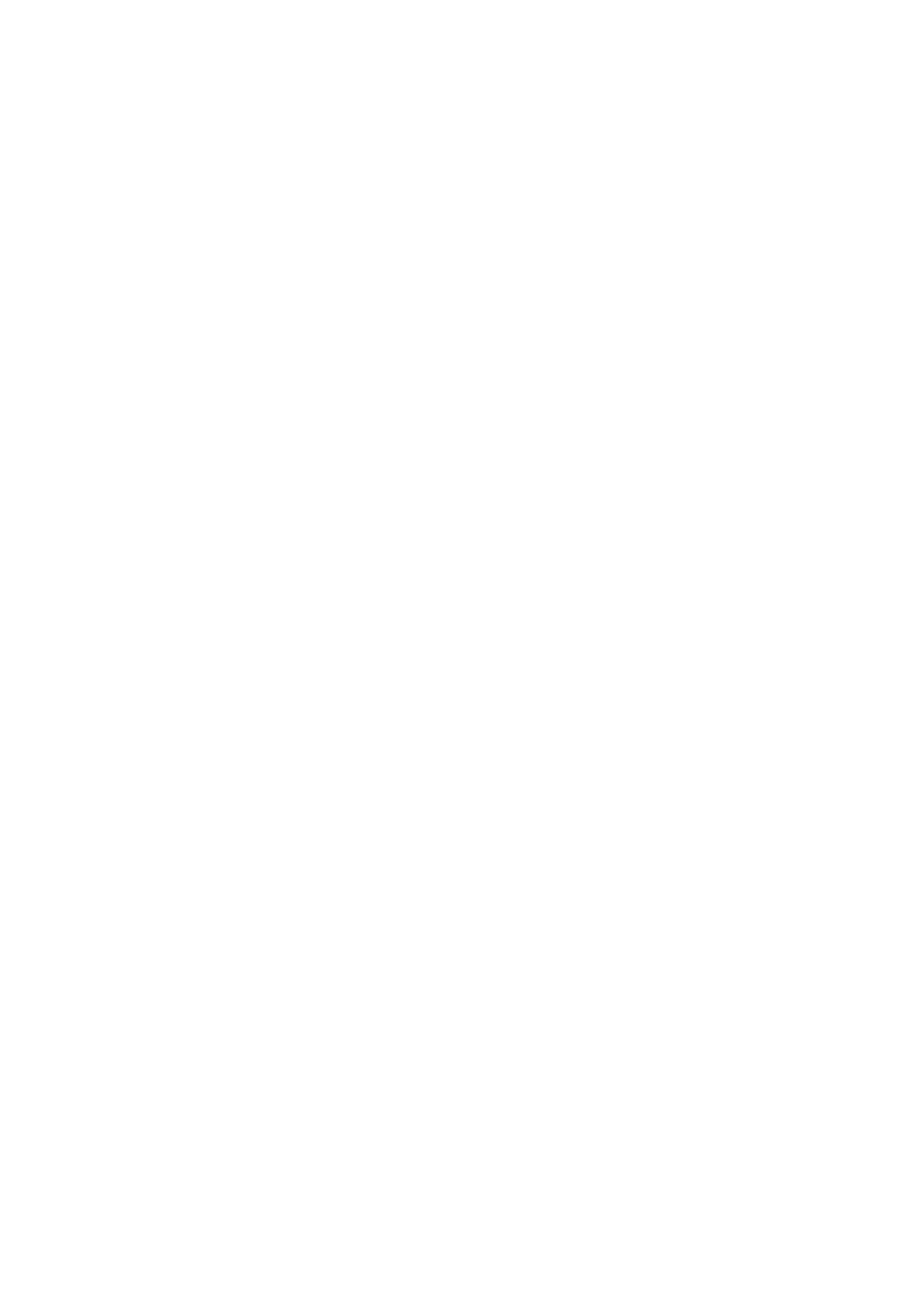} &
    \includegraphics[width=\attentionVisDimmedWidth,frame=0.1pt]{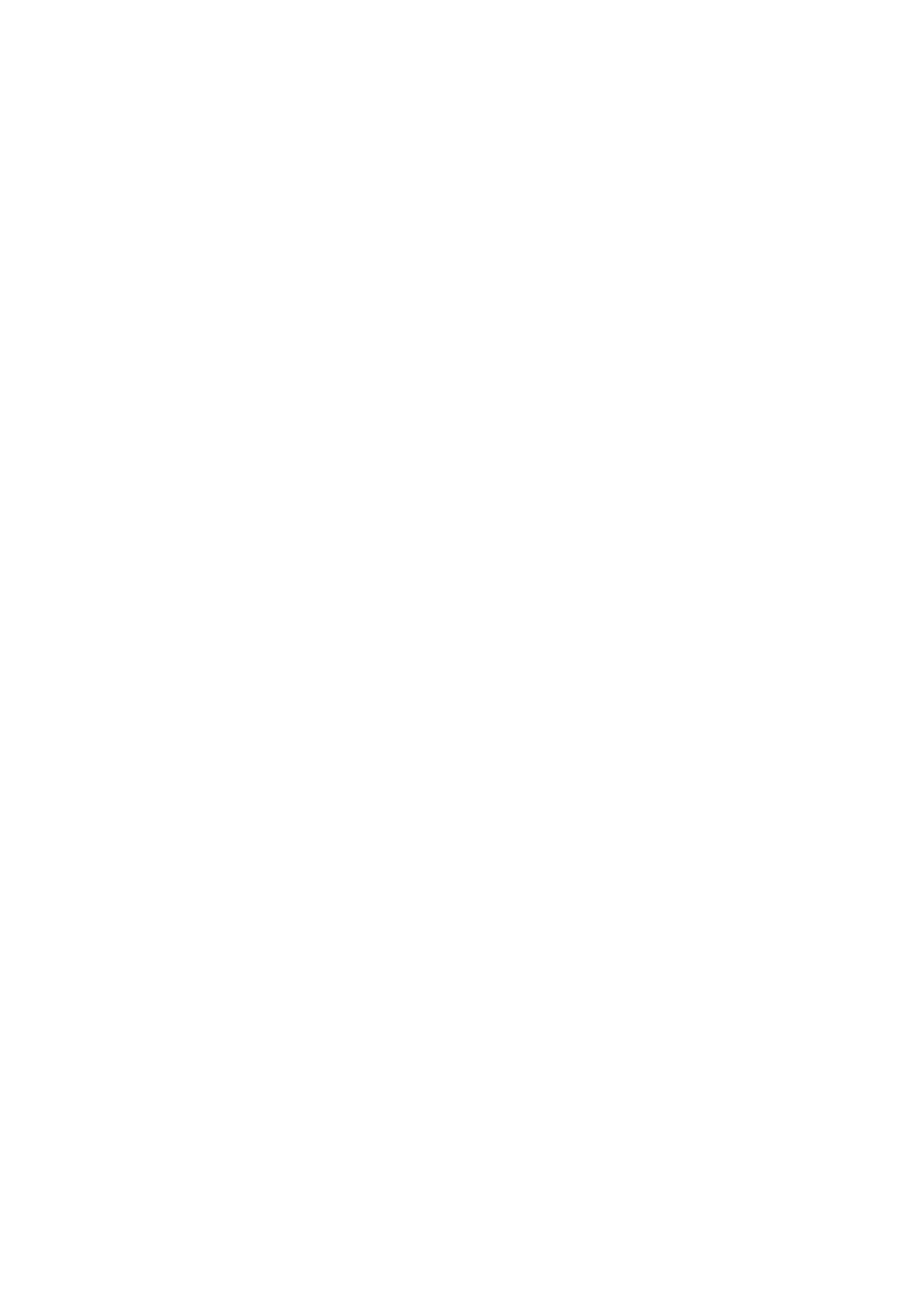} &
    \includegraphics[width=\attentionVisDimmedWidth,frame=0.1pt]{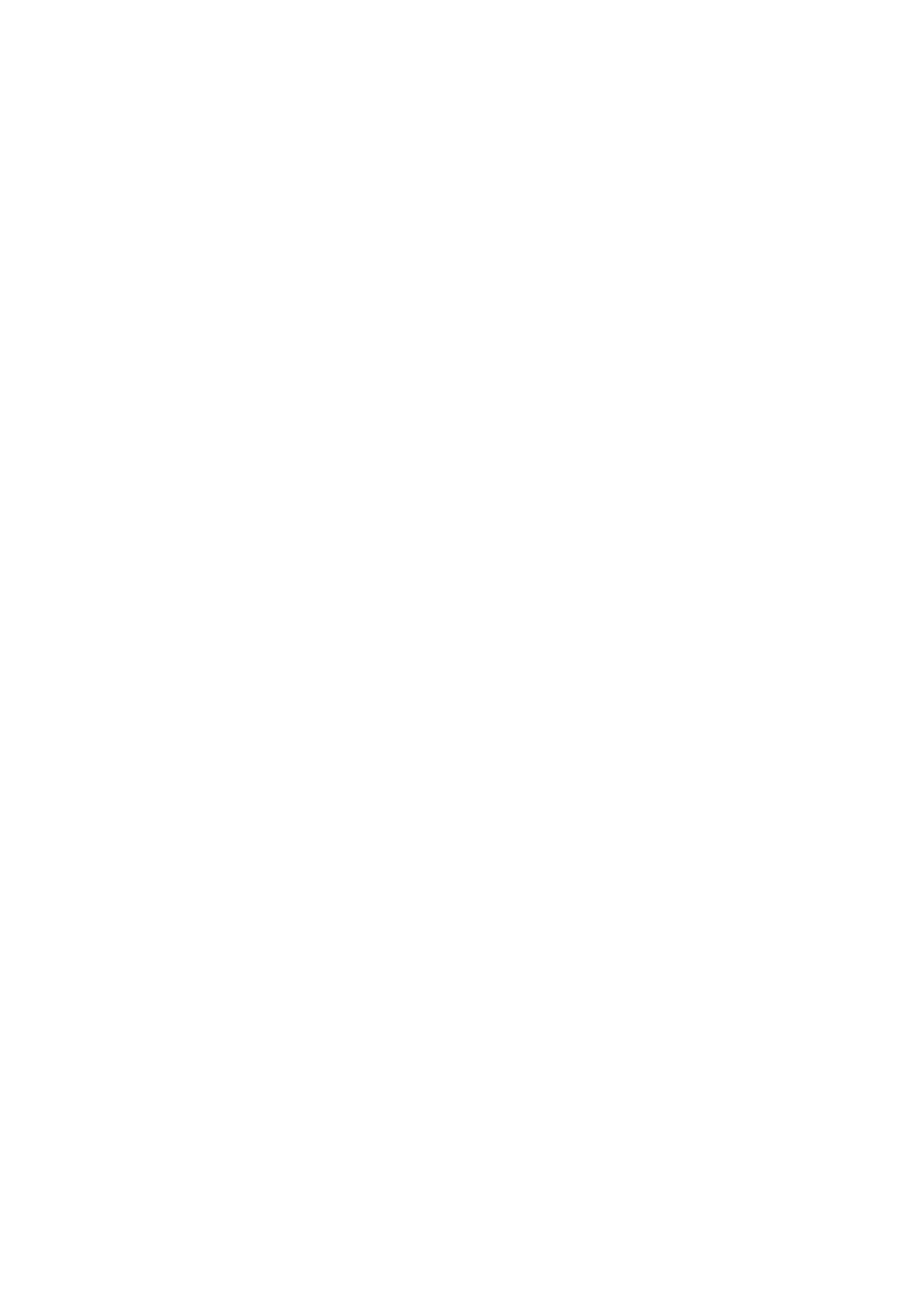} &
    \includegraphics[width=\attentionVisDimmedWidth,frame=0.1pt]{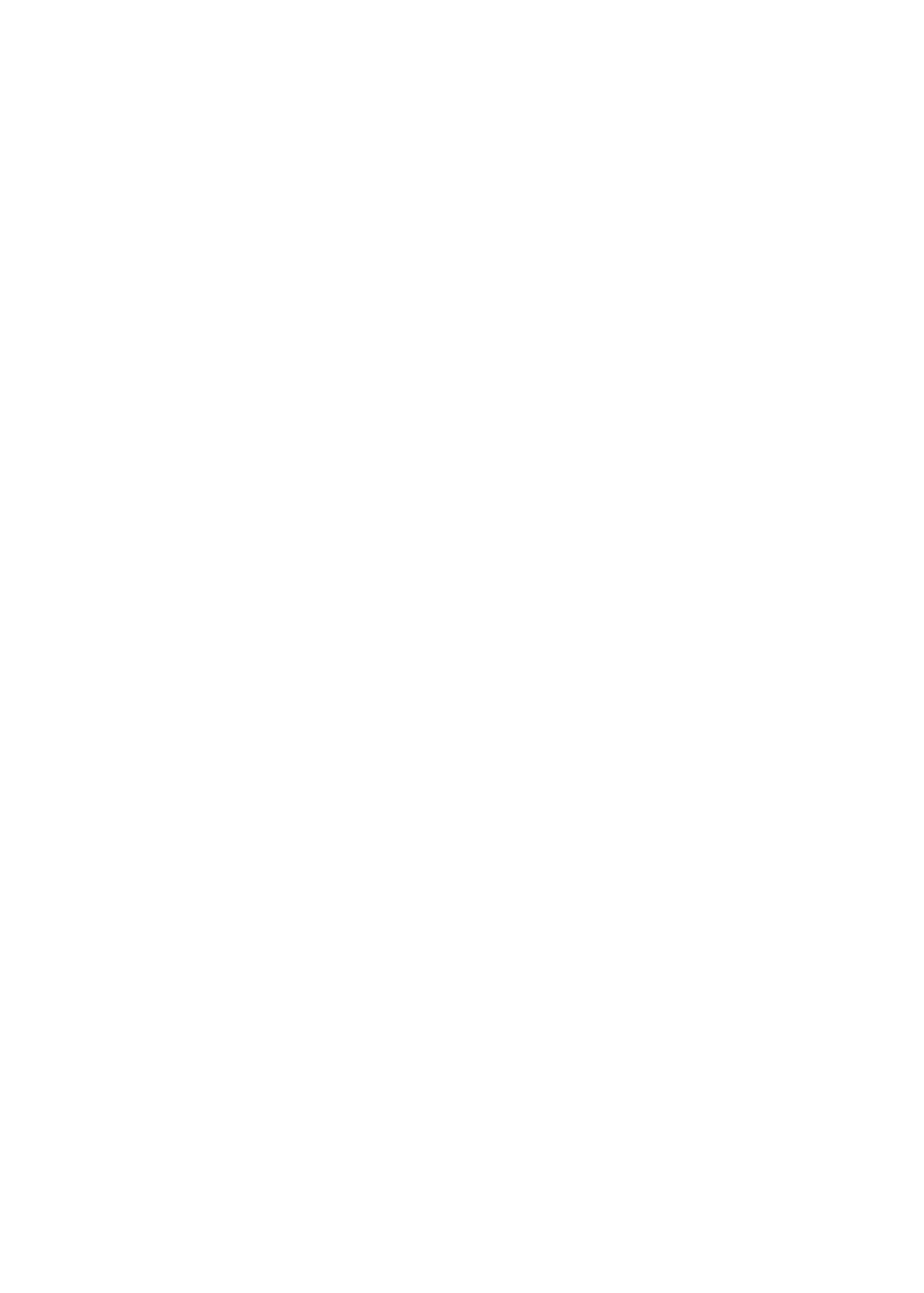} &
    \includegraphics[width=\attentionVisDimmedWidth,frame=0.1pt]{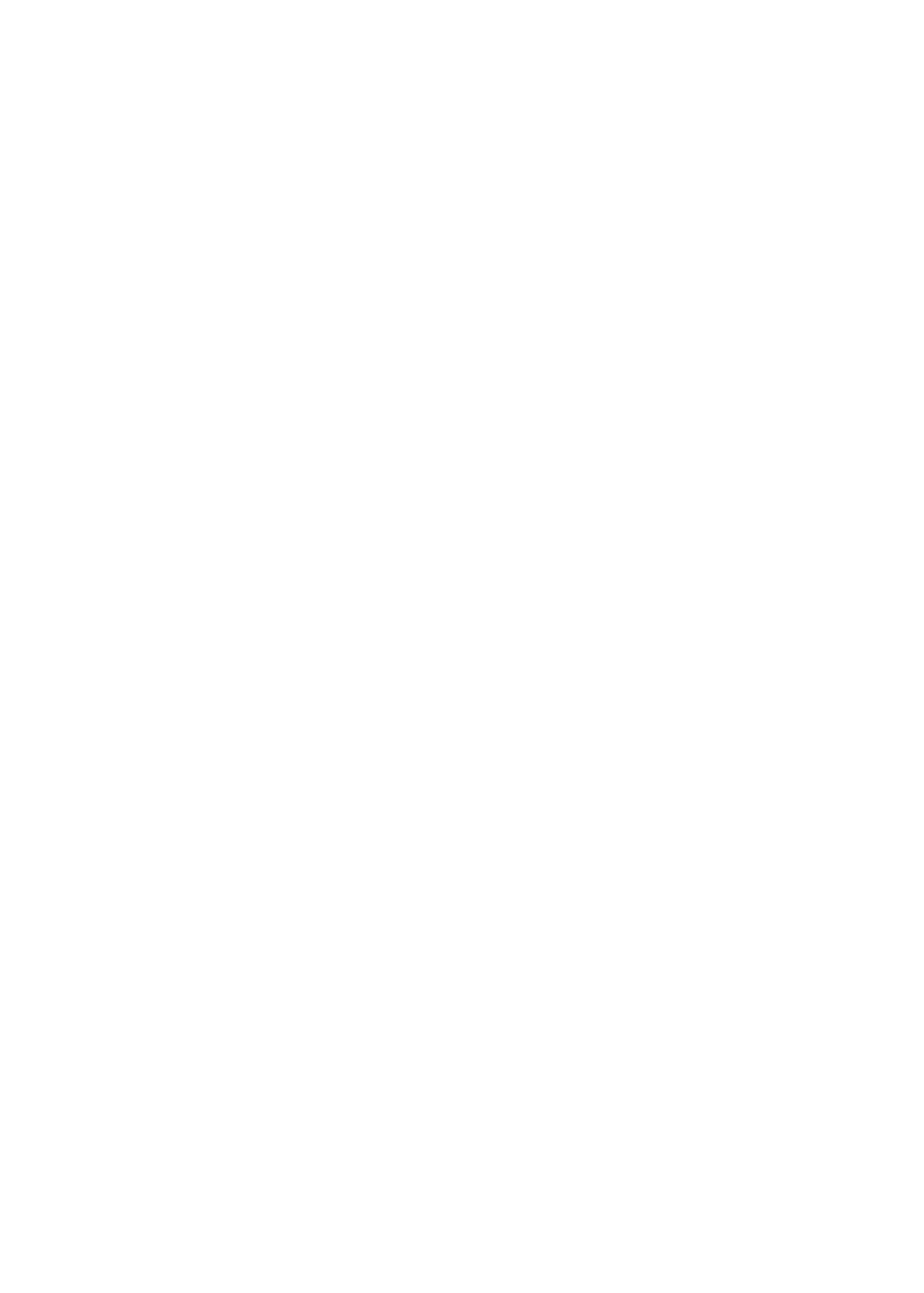} &
    \includegraphics[width=\attentionVisDimmedWidth,frame=0.1pt]{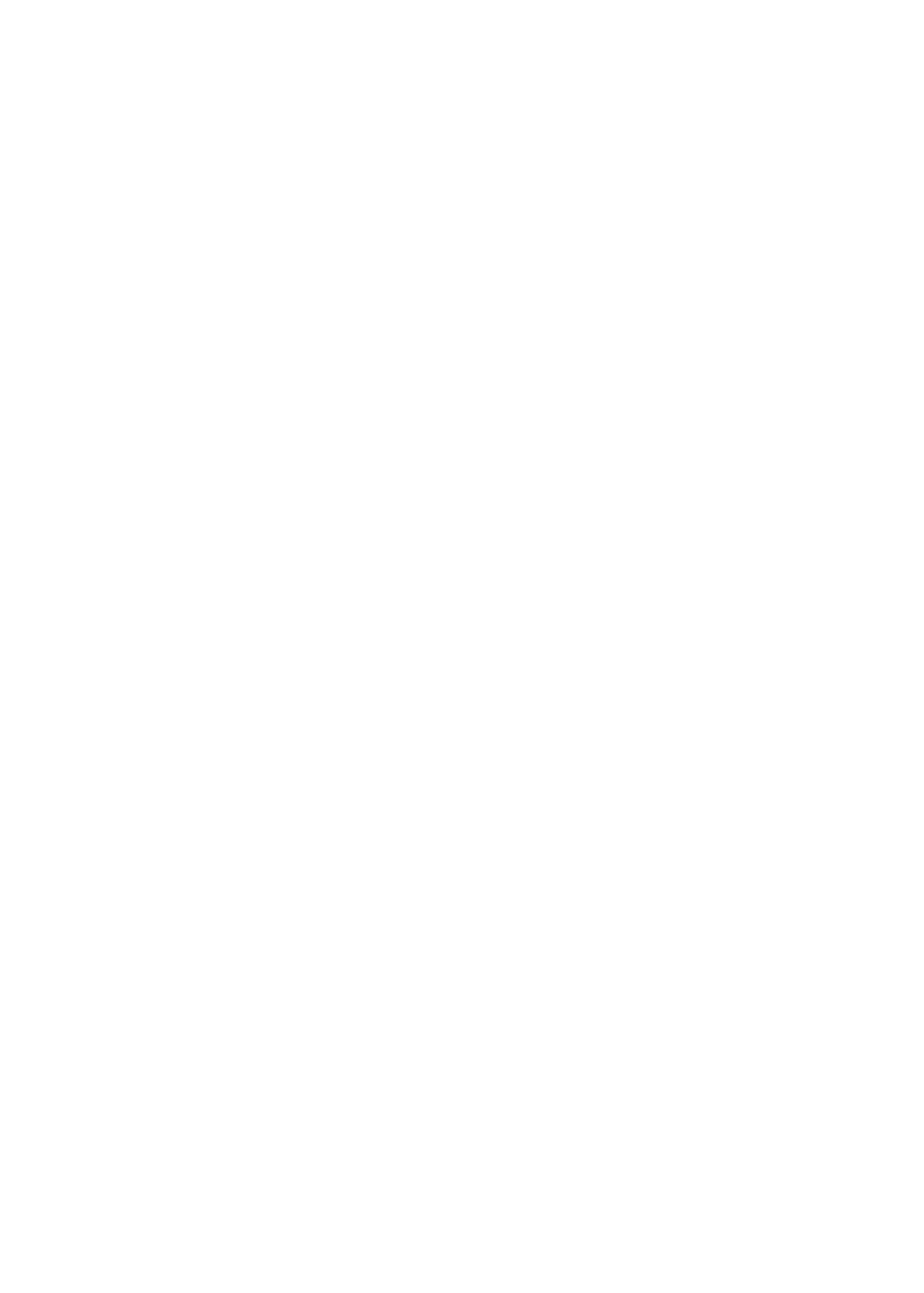} &
    \includegraphics[width=\attentionVisDimmedWidth,frame=0.1pt]{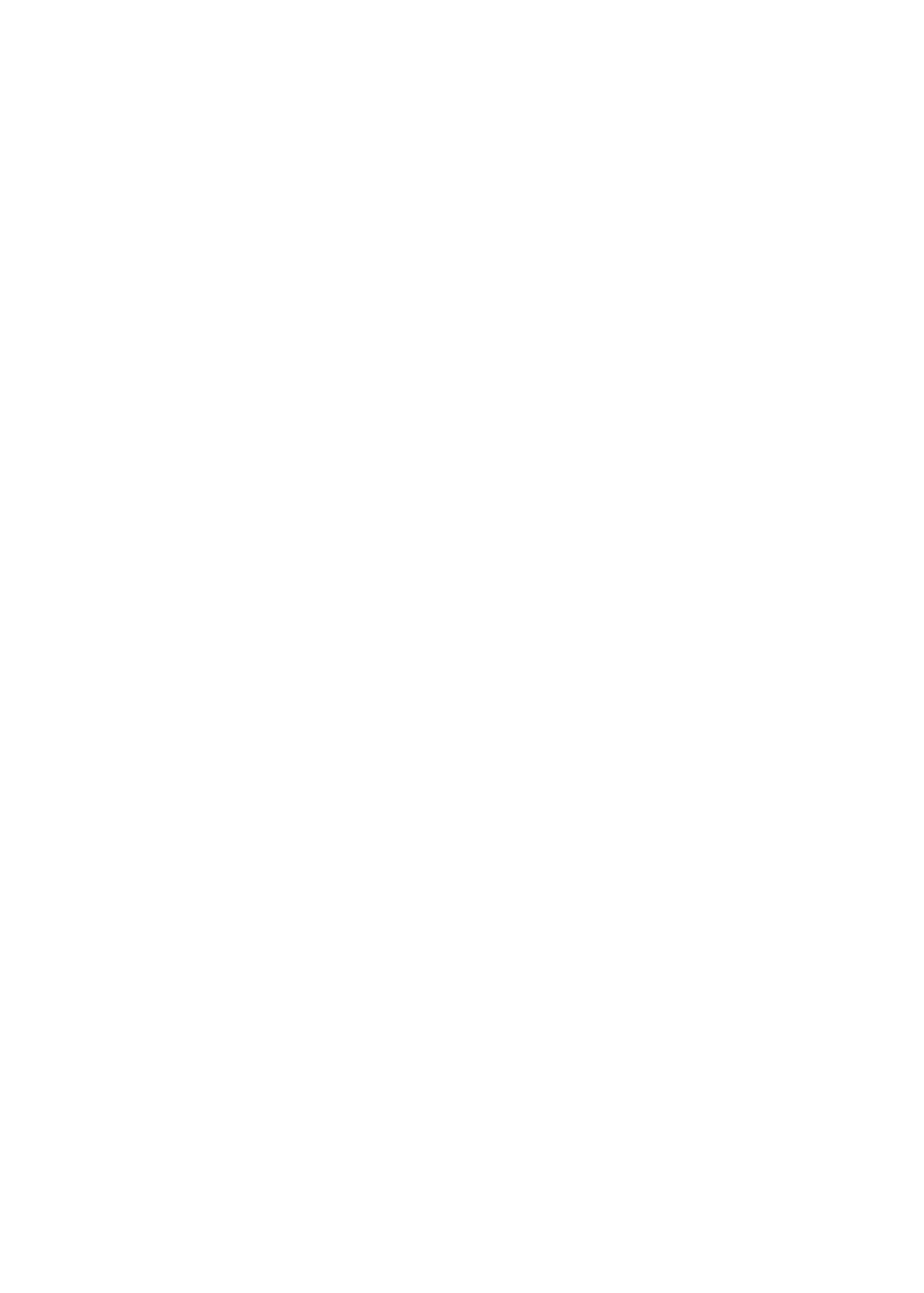} &
    \includegraphics[width=\attentionVisDimmedWidth,frame=0.1pt]{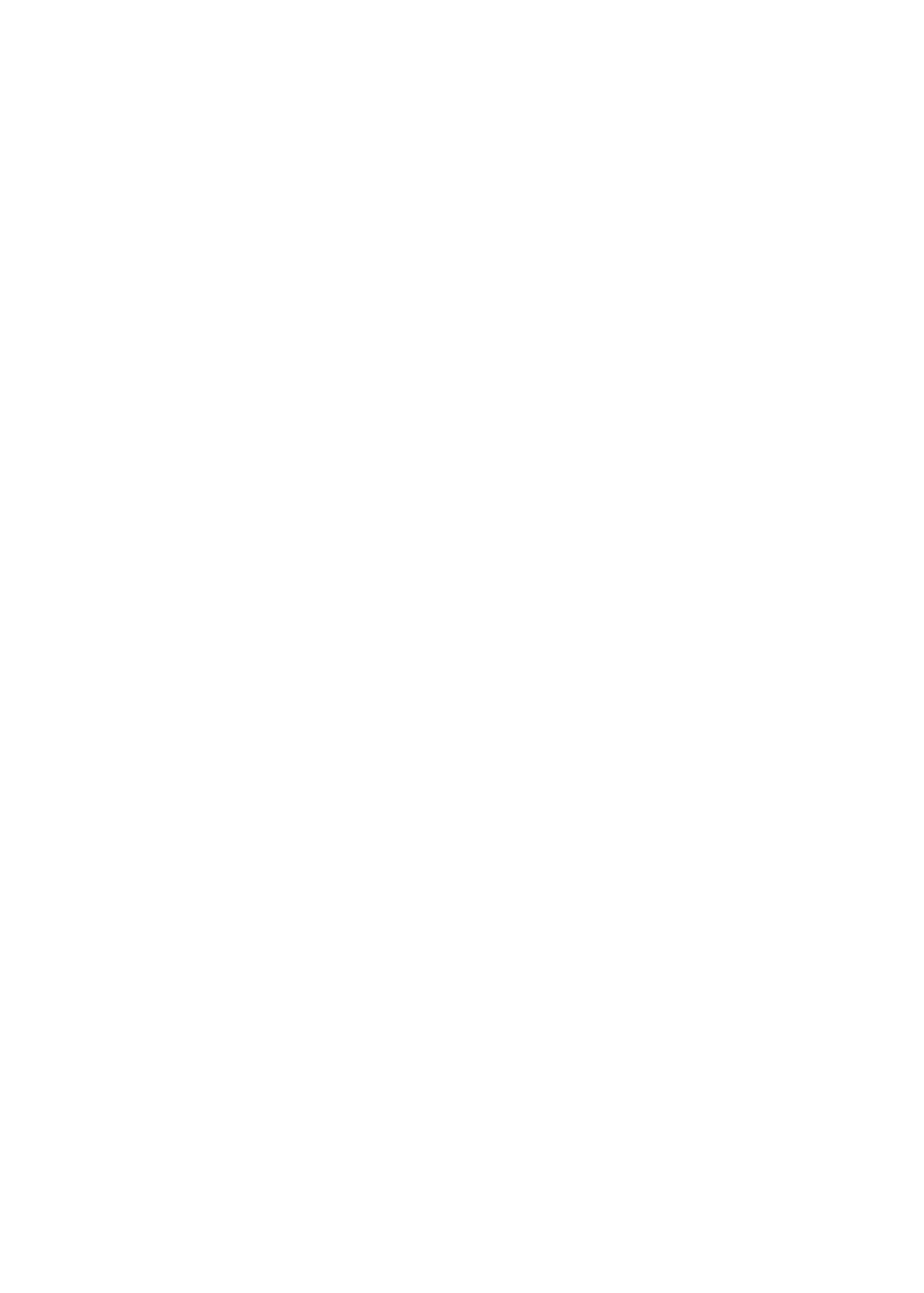} &
    \includegraphics[width=\attentionVisDimmedWidth,frame=0.1pt]{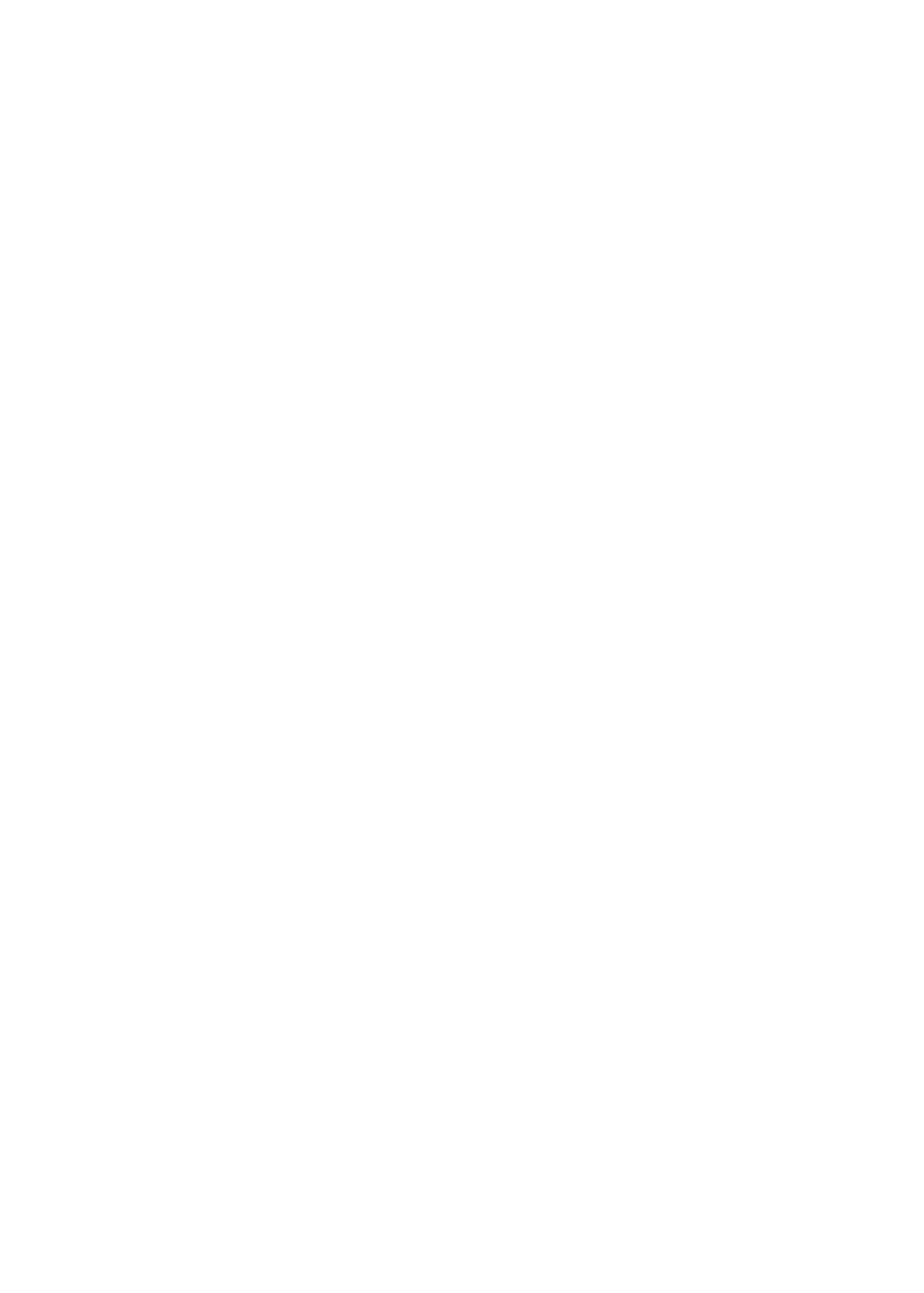} &
    \includegraphics[width=\attentionVisDimmedWidth,frame=0.1pt]{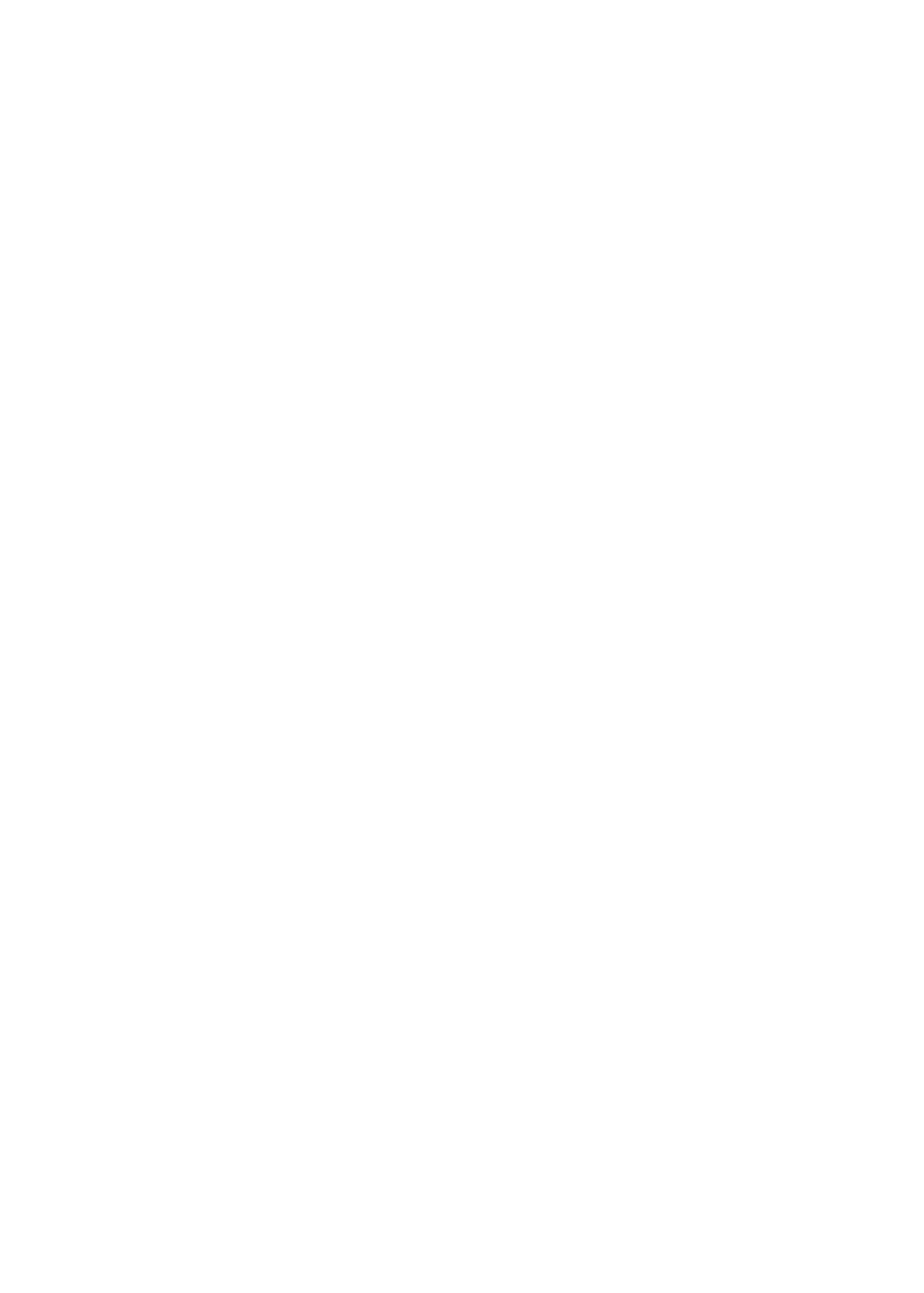} &
    \includegraphics[width=\attentionVisDimmedWidth,frame=0.1pt]{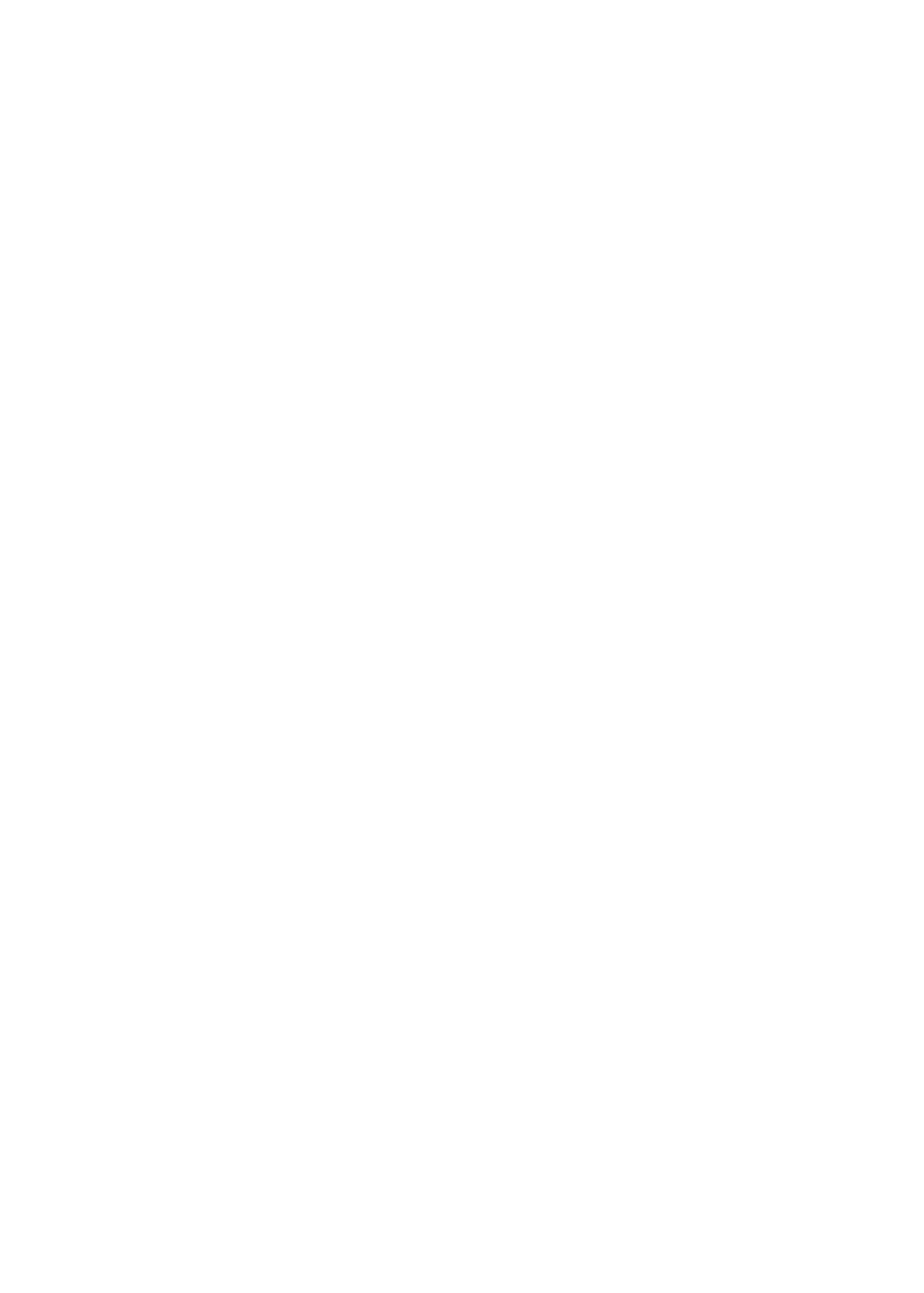} &
    \includegraphics[width=\attentionVisDimmedWidth,frame=0.1pt]{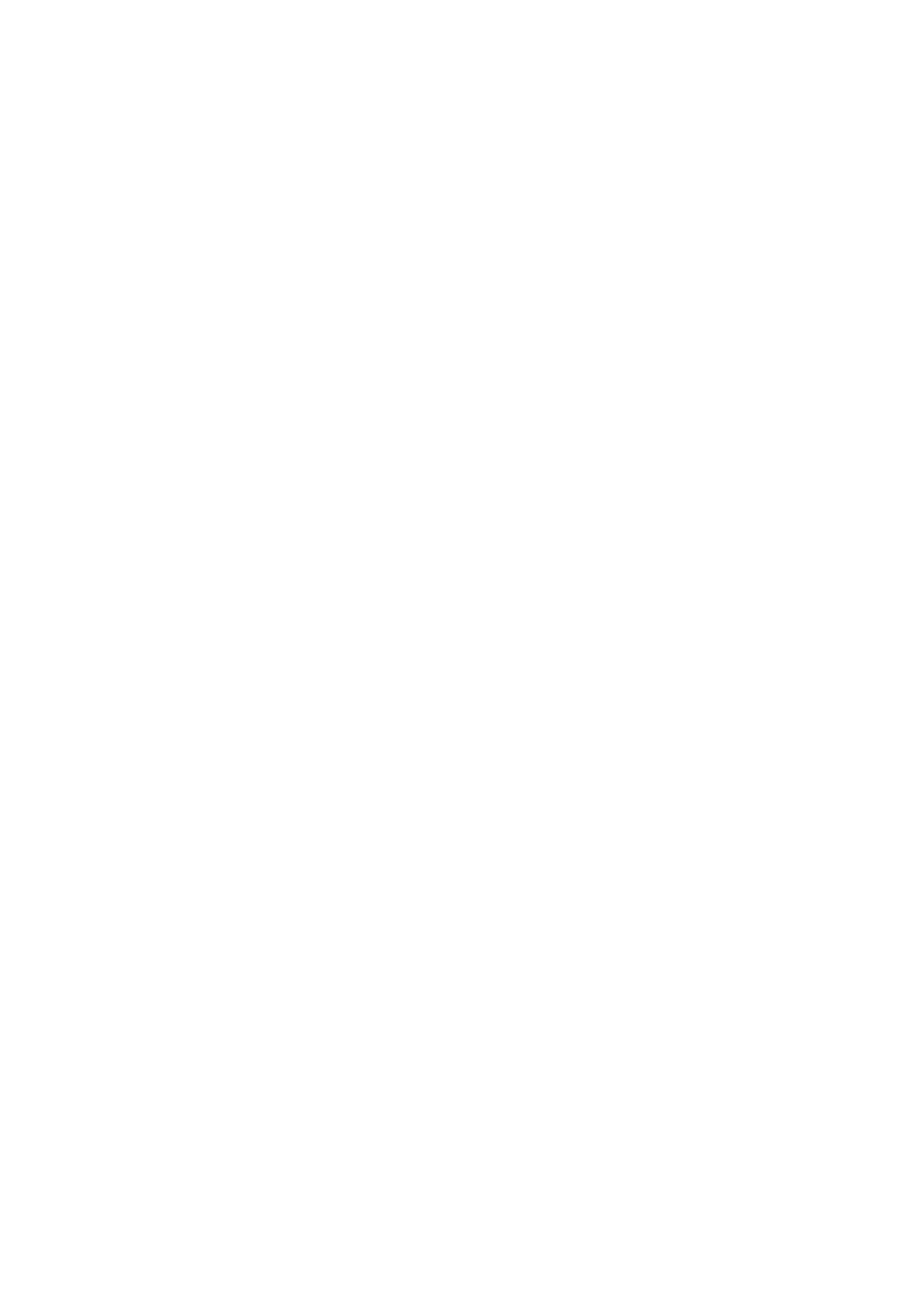} \\
    
    \rotatebox{90}{\hspace{0.3cm}\footnotesize Layer 2} &
    \includegraphics[width=\attentionVisDimmedWidth,frame=0.1pt]{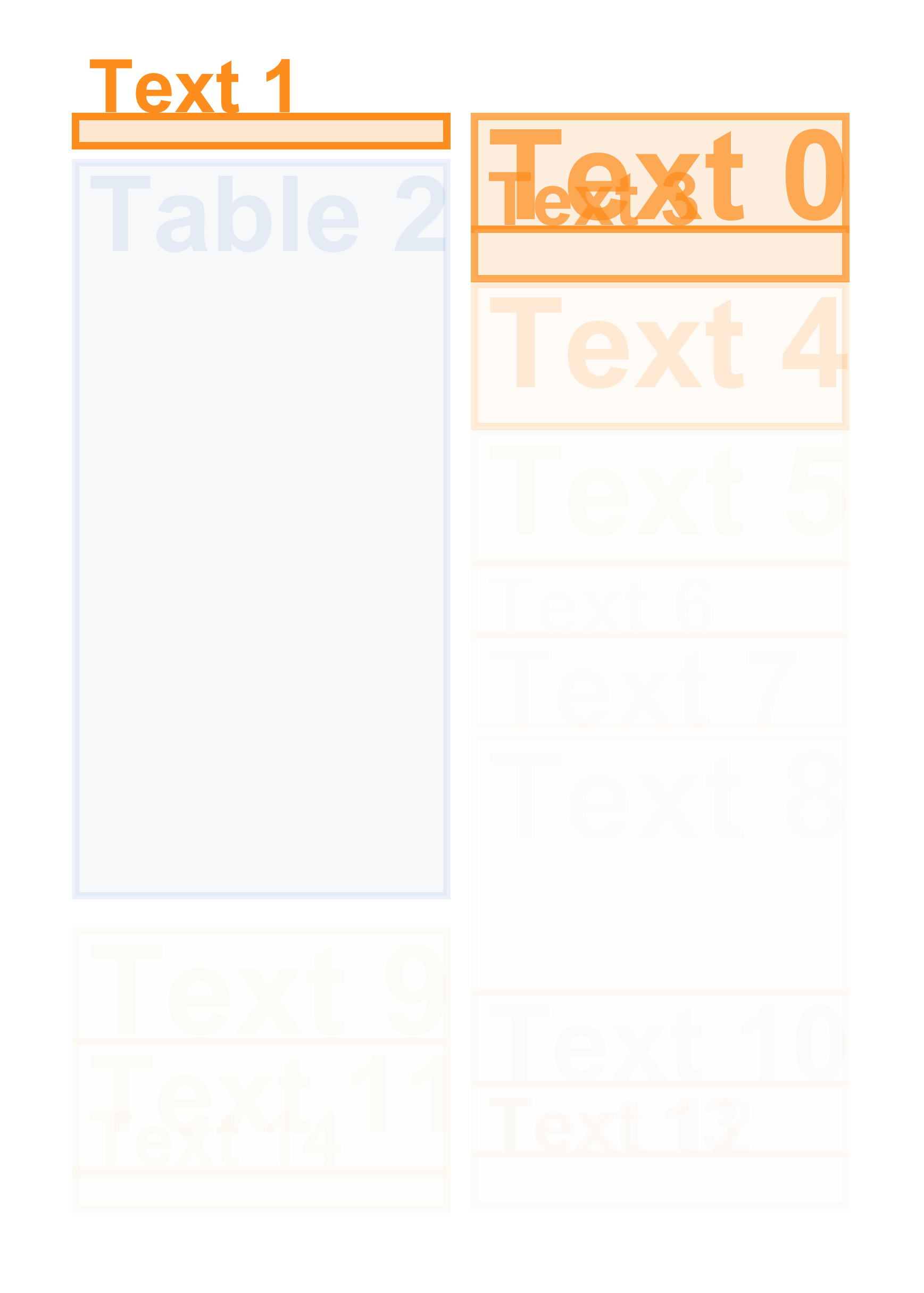} &
    \includegraphics[width=\attentionVisDimmedWidth,frame=0.1pt]{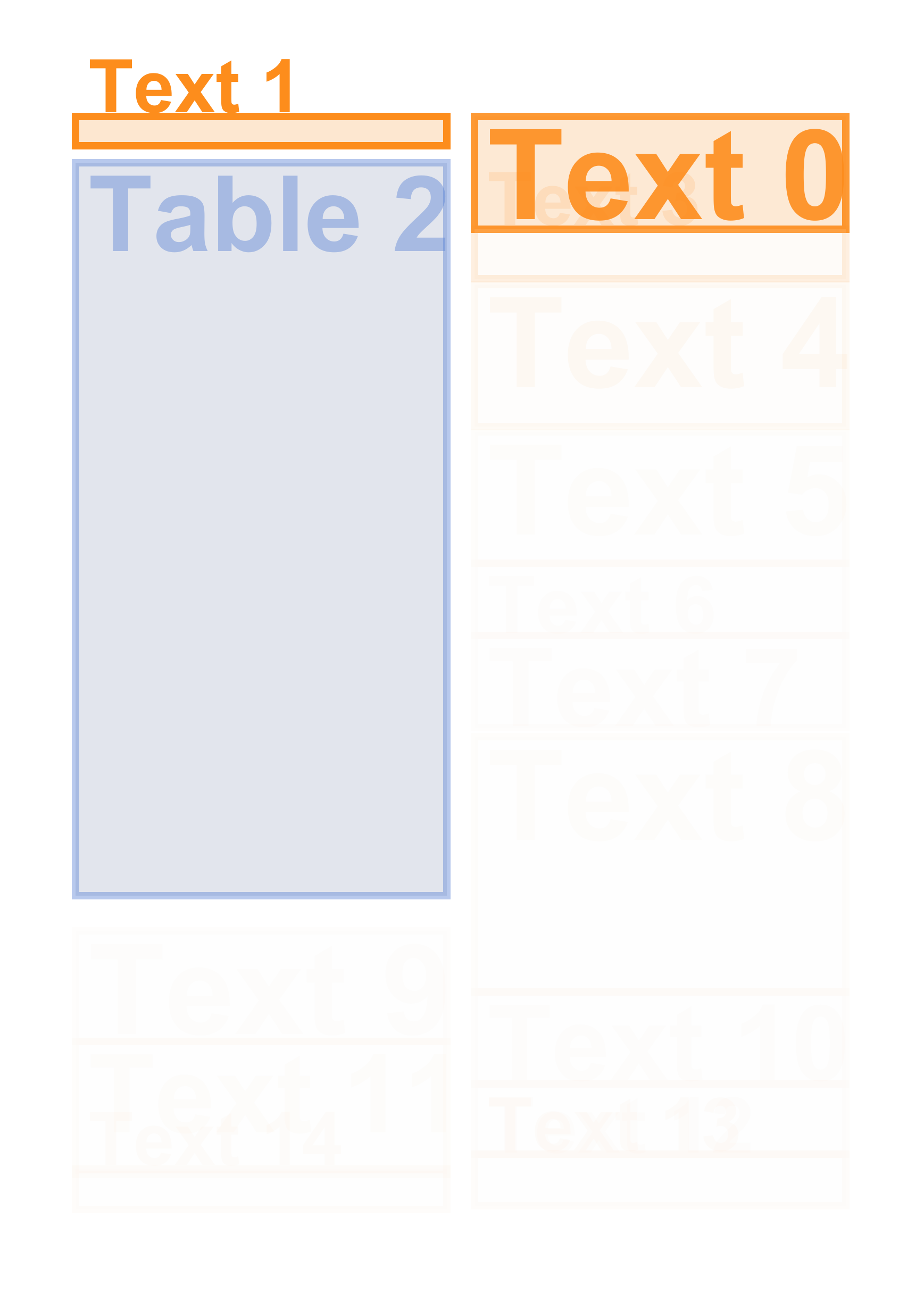} &
    \includegraphics[width=\attentionVisDimmedWidth,frame=0.1pt]{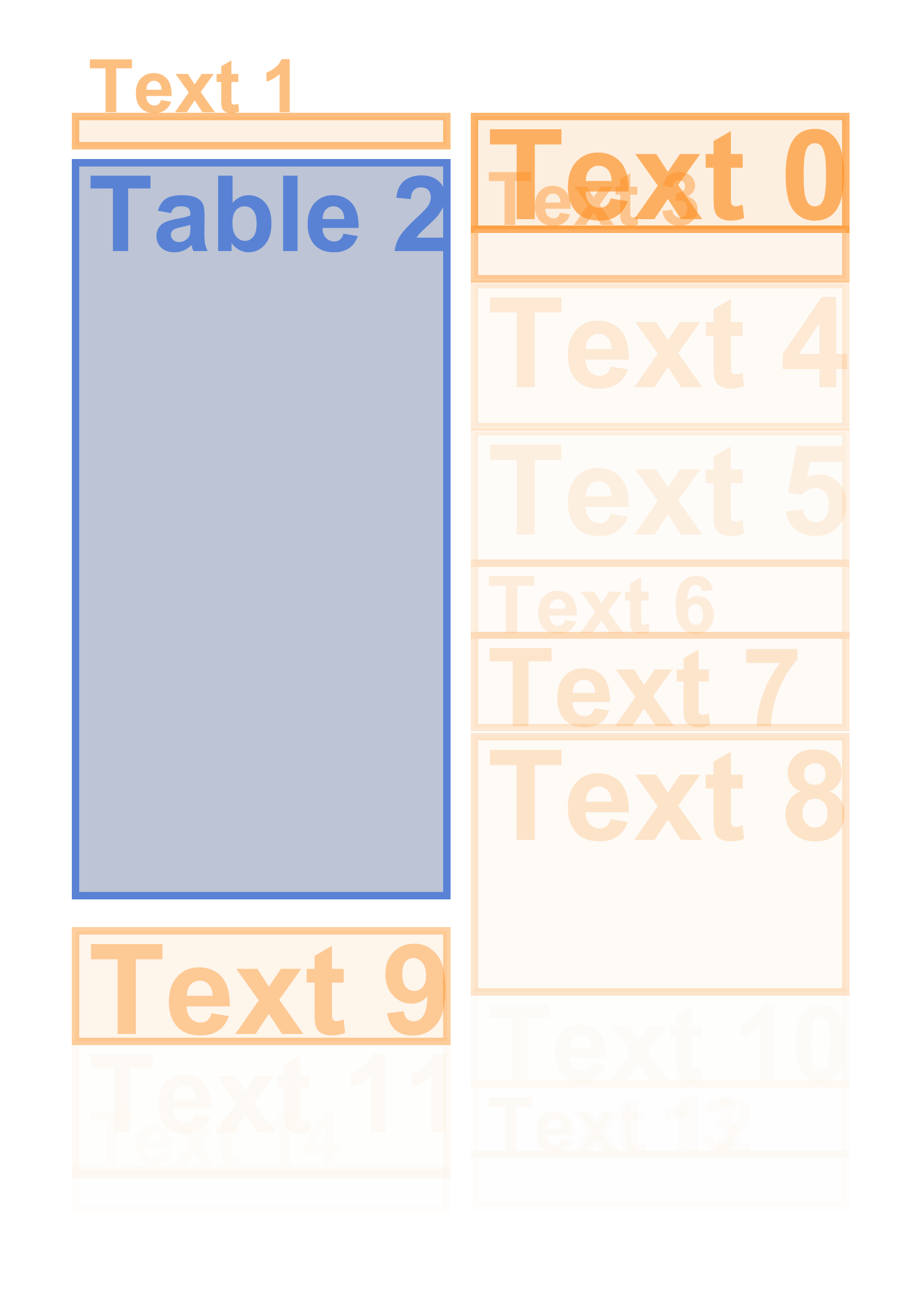} &
    \includegraphics[width=\attentionVisDimmedWidth,frame=0.1pt]{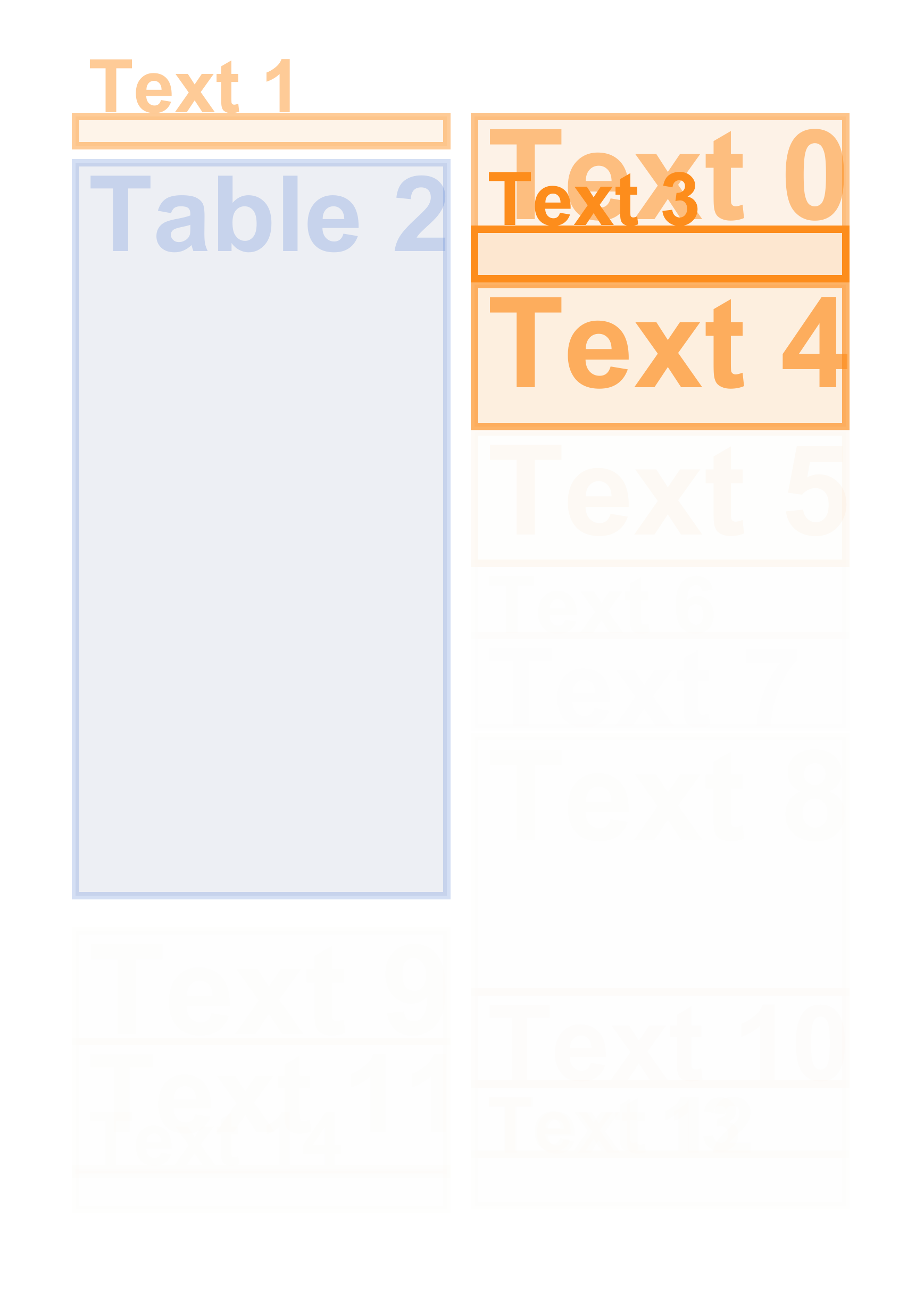} &
    \includegraphics[width=\attentionVisDimmedWidth,frame=0.1pt]{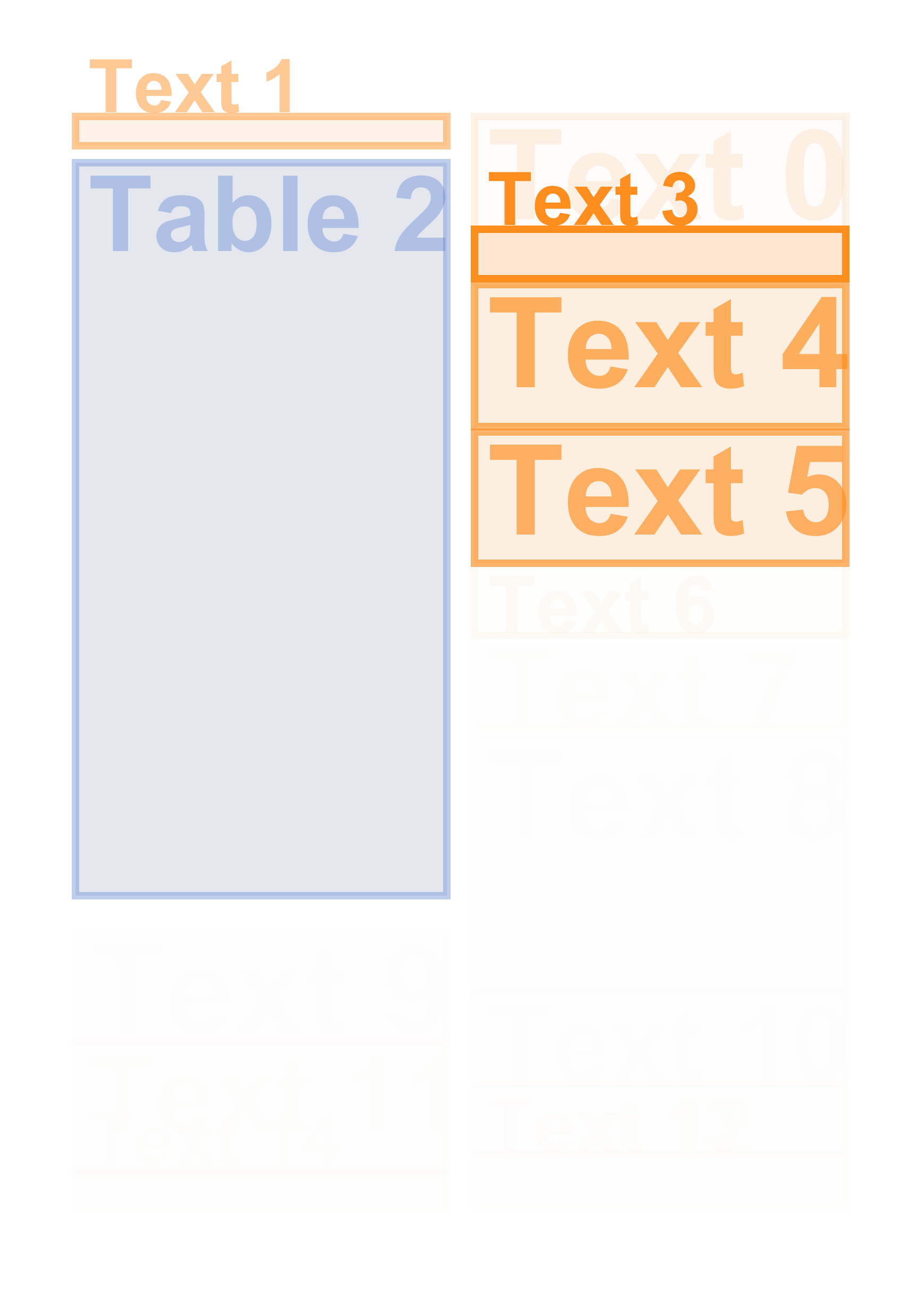} &
    \includegraphics[width=\attentionVisDimmedWidth,frame=0.1pt]{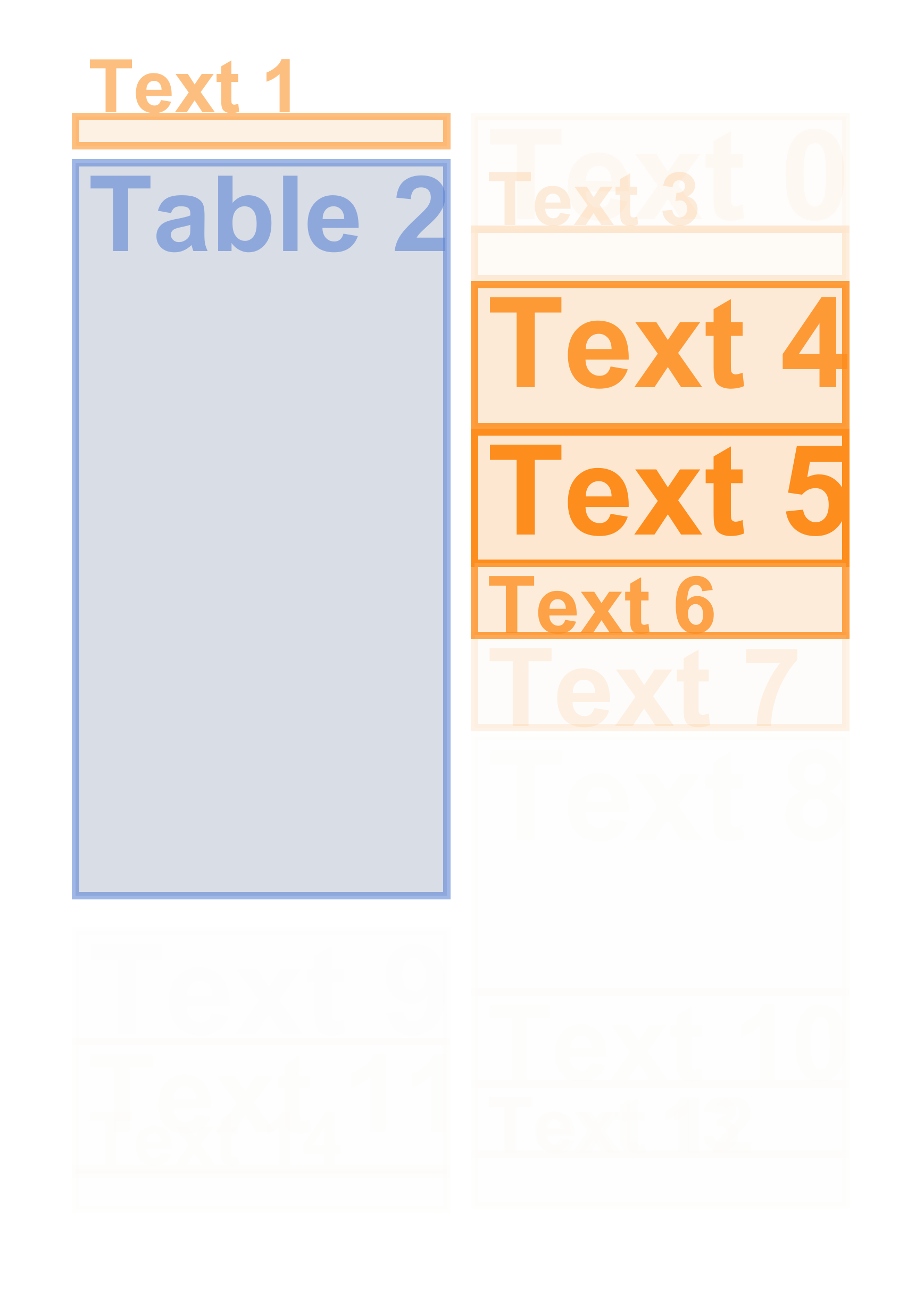} &
    \includegraphics[width=\attentionVisDimmedWidth,frame=0.1pt]{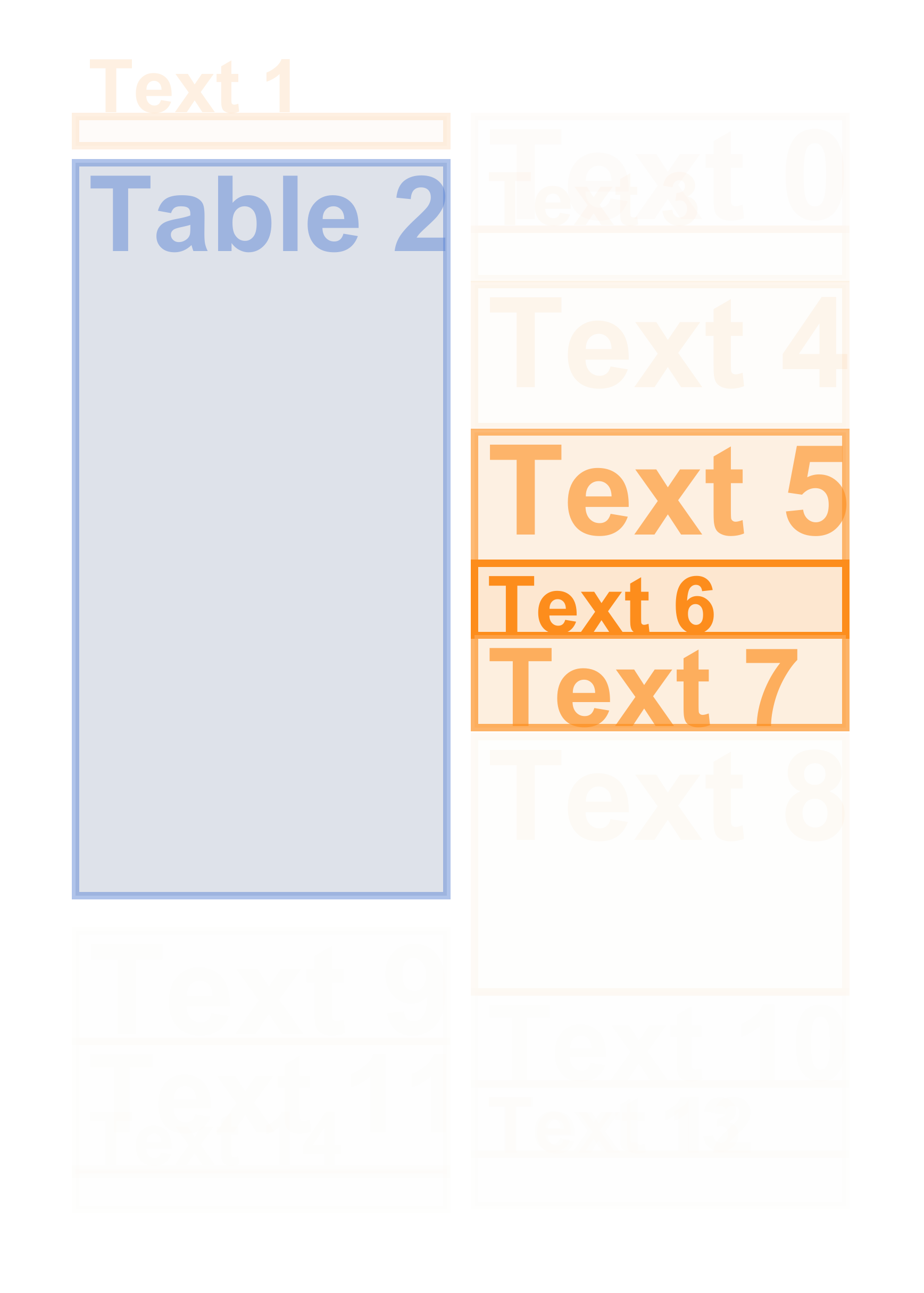} &
    \includegraphics[width=\attentionVisDimmedWidth,frame=0.1pt]{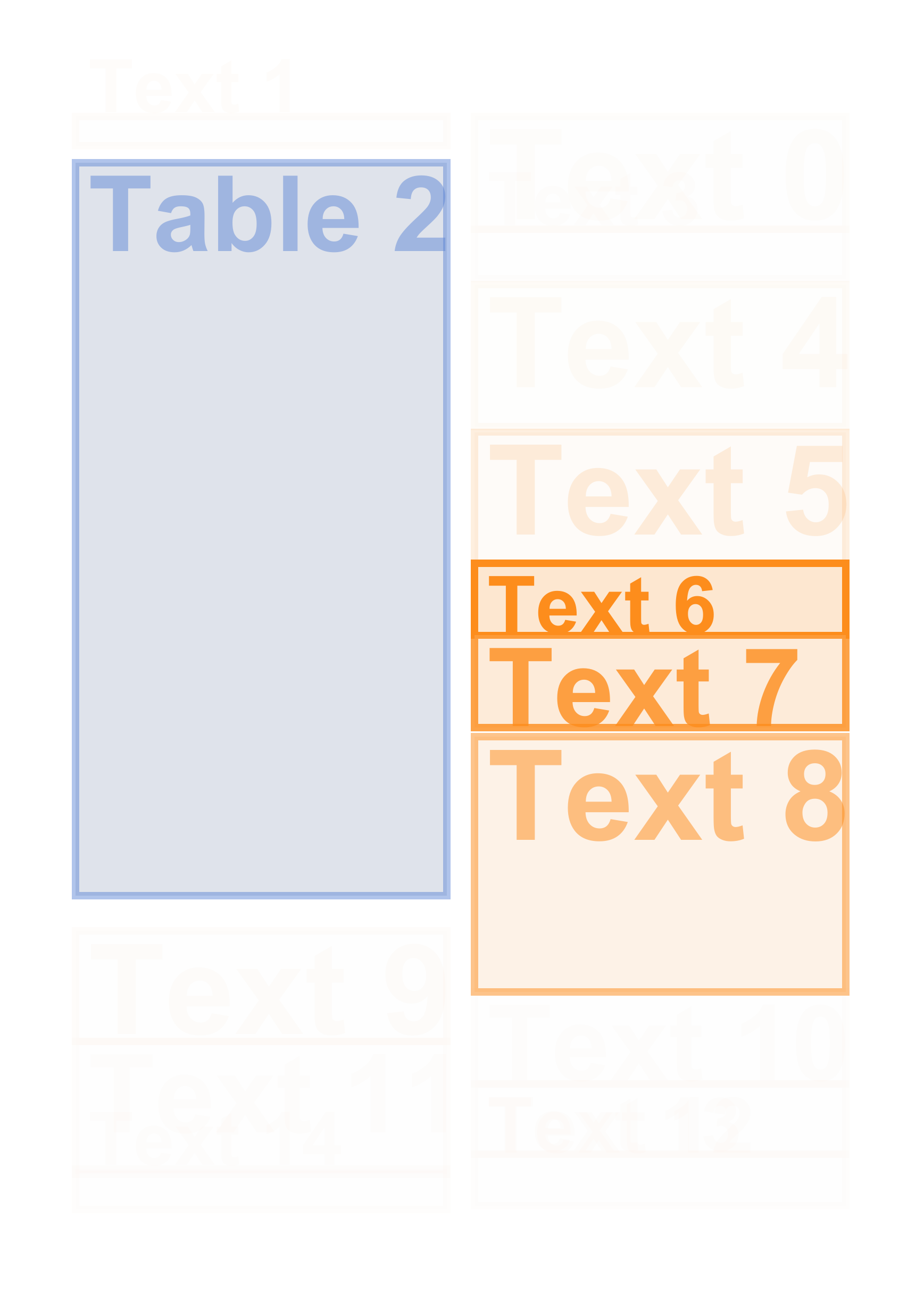} &
    \includegraphics[width=\attentionVisDimmedWidth,frame=0.1pt]{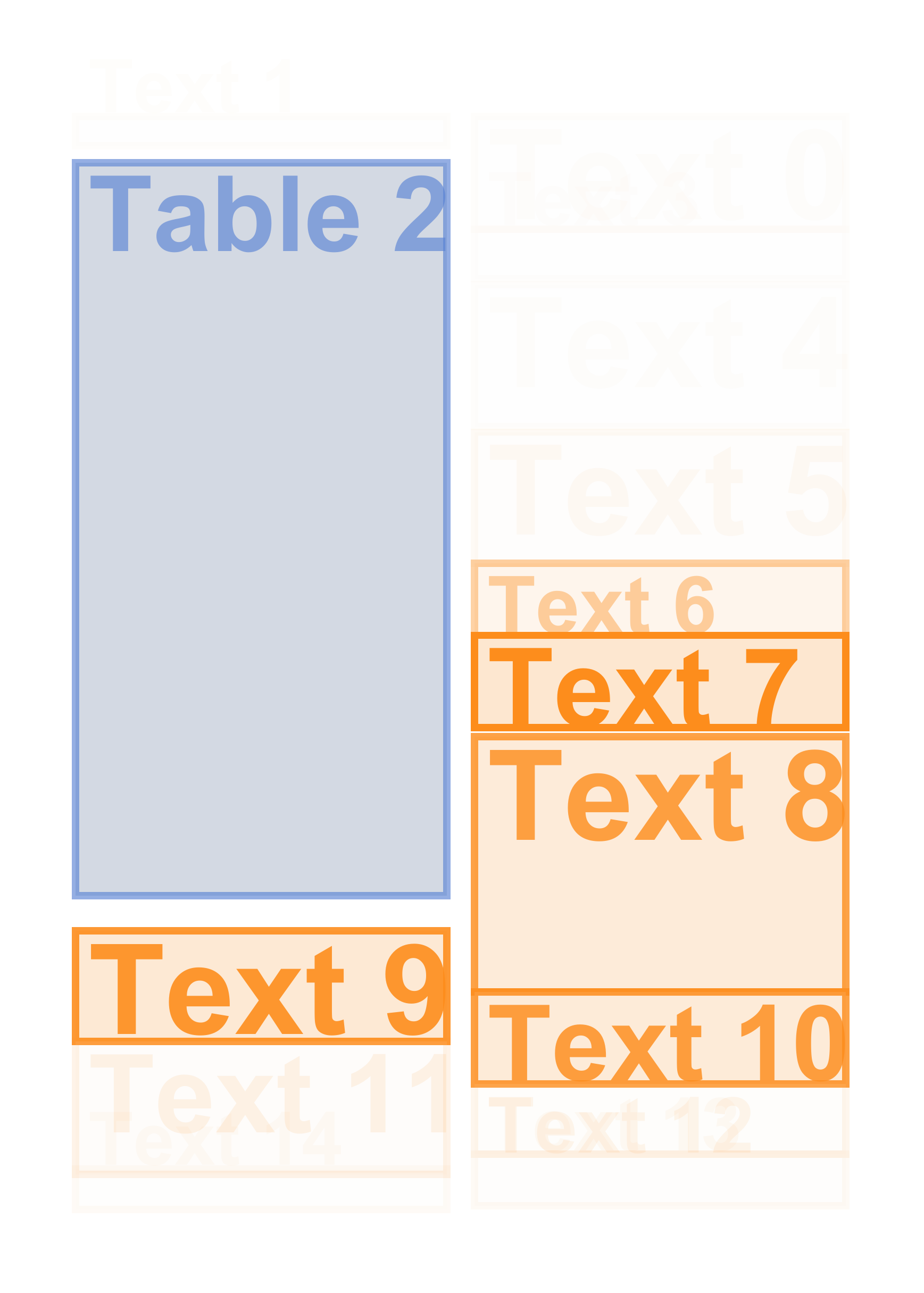} &
    \includegraphics[width=\attentionVisDimmedWidth,frame=0.1pt]{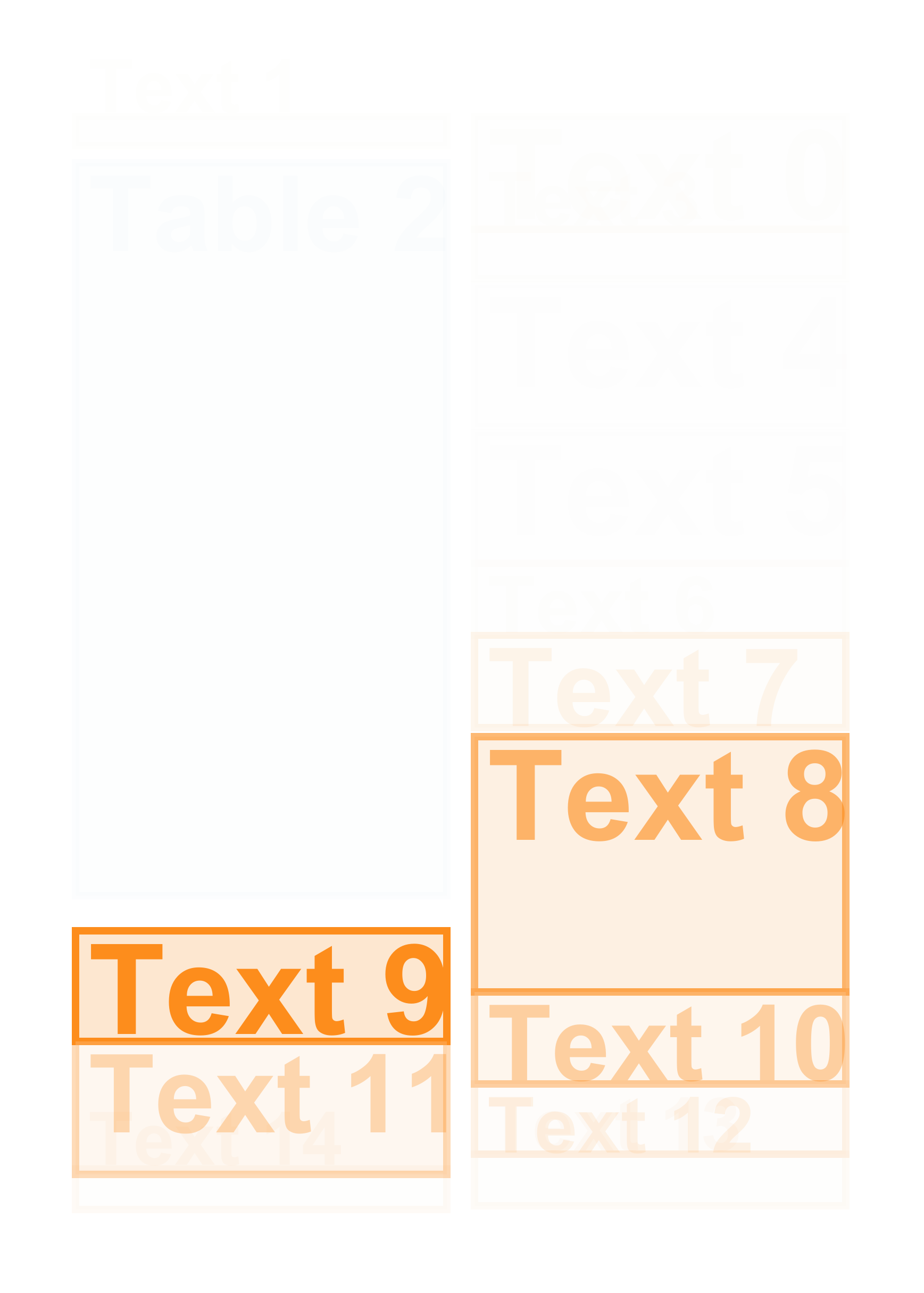} &
    \includegraphics[width=\attentionVisDimmedWidth,frame=0.1pt]{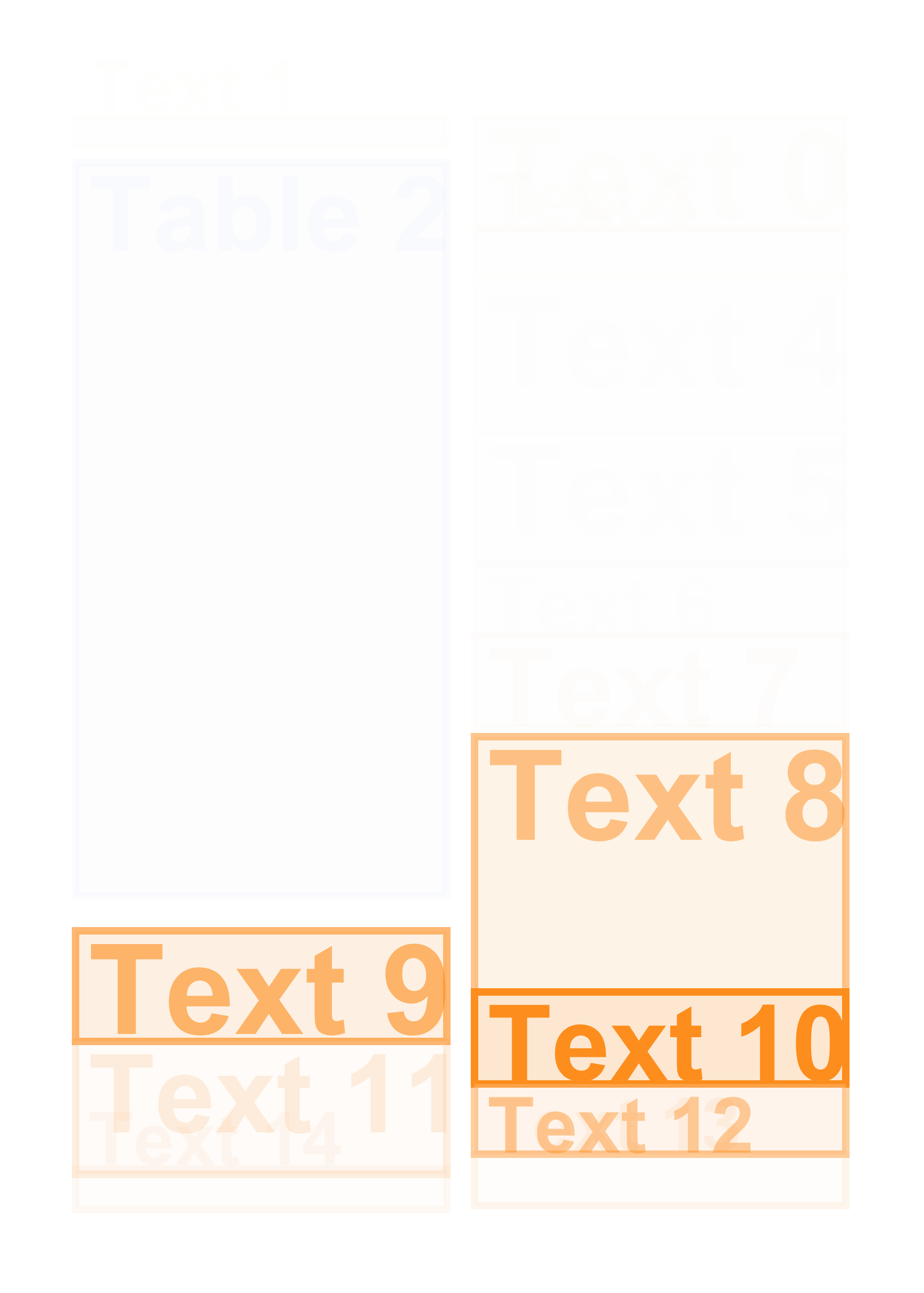} &
    \includegraphics[width=\attentionVisDimmedWidth,frame=0.1pt]{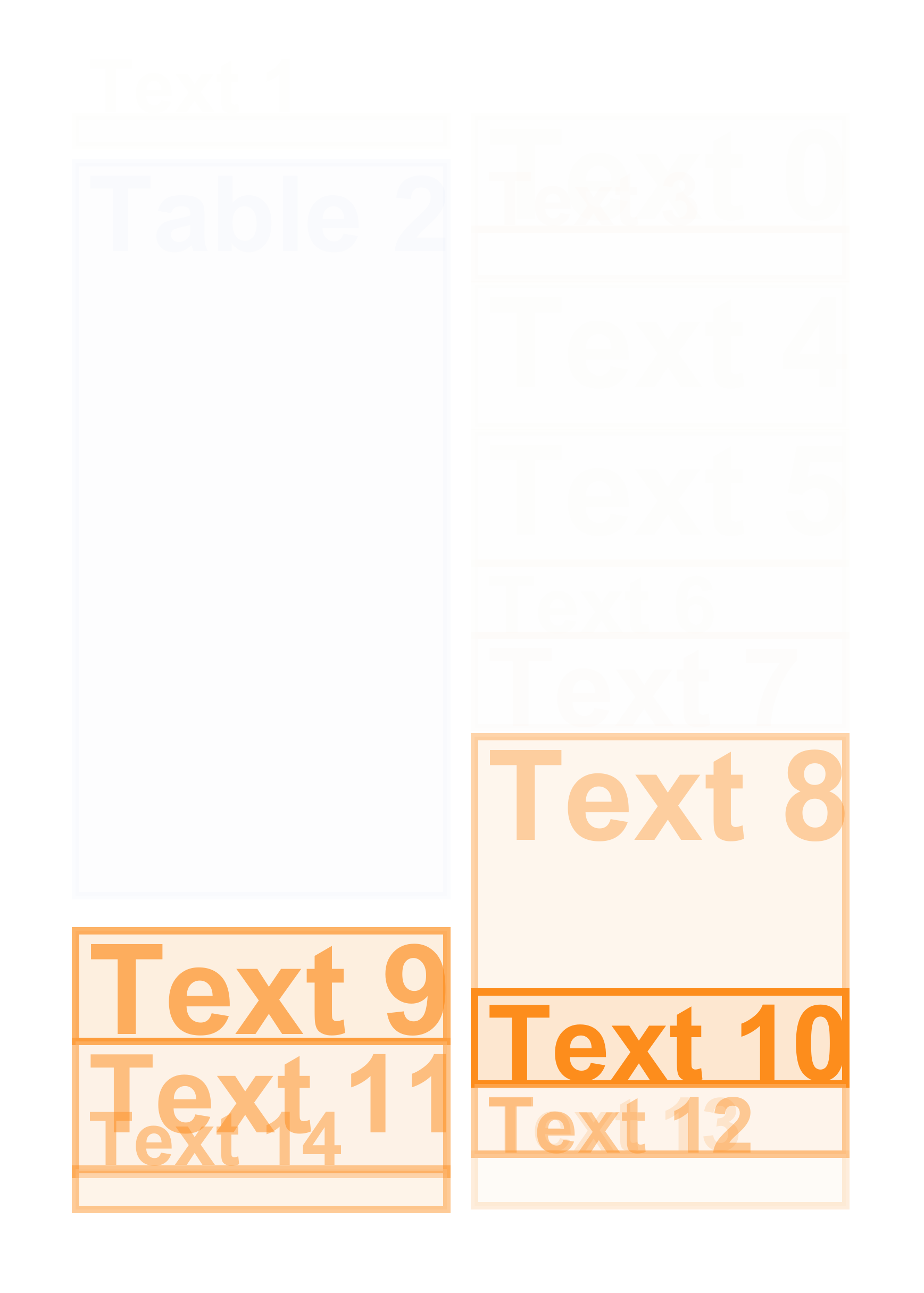} &
    \includegraphics[width=\attentionVisDimmedWidth,frame=0.1pt]{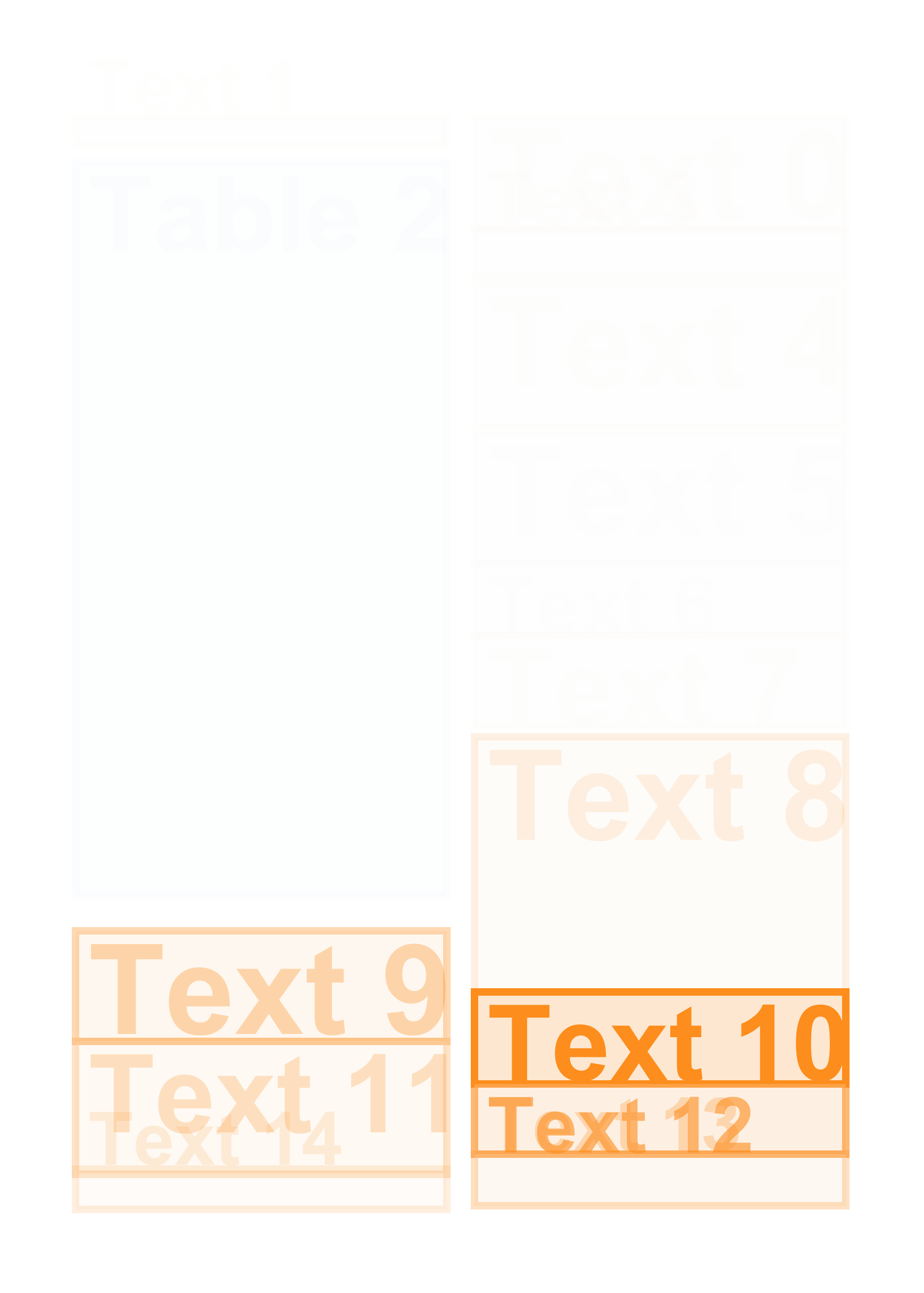} &
    \includegraphics[width=\attentionVisDimmedWidth,frame=0.1pt]{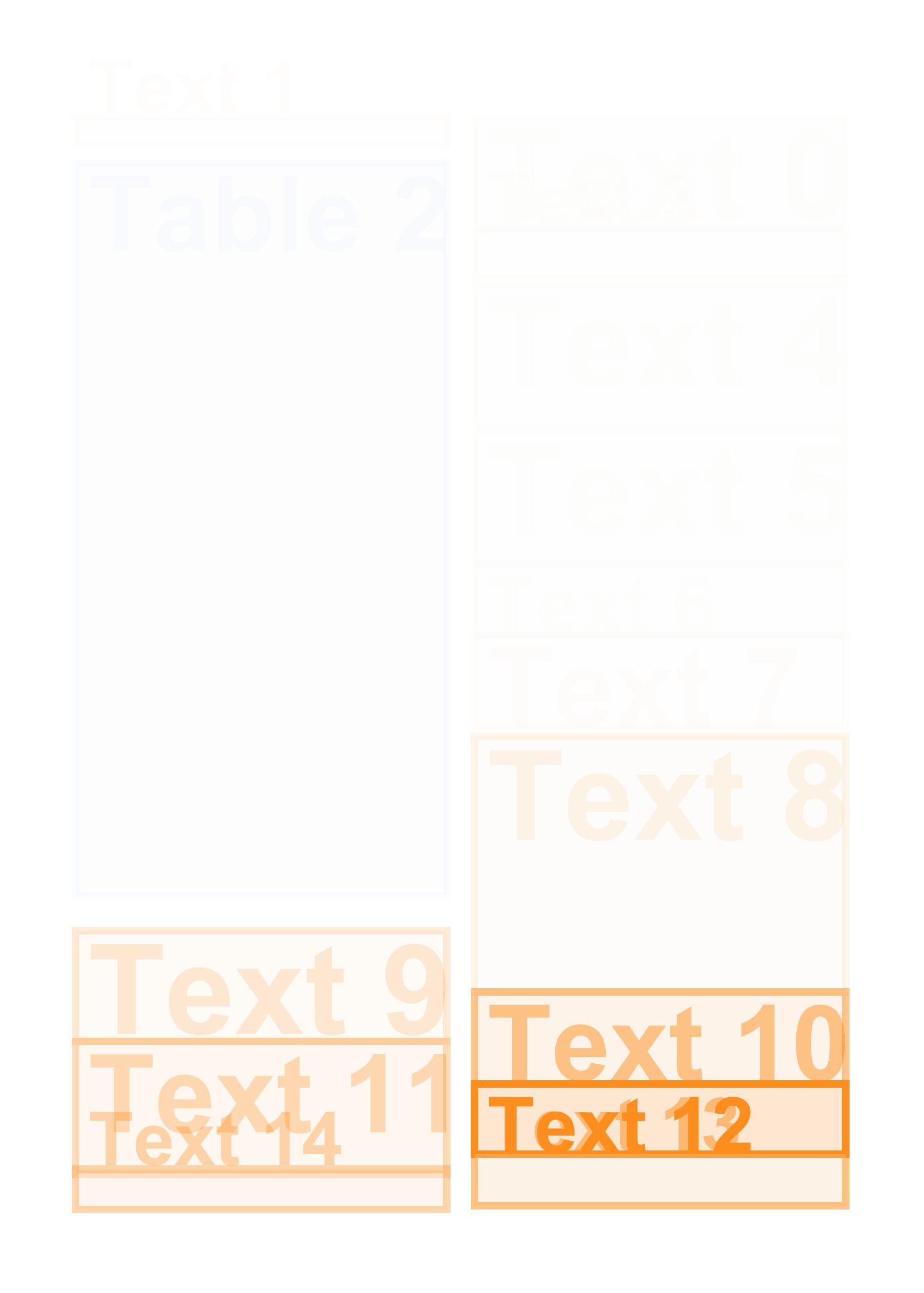} &
    \includegraphics[width=\attentionVisDimmedWidth,frame=0.1pt]{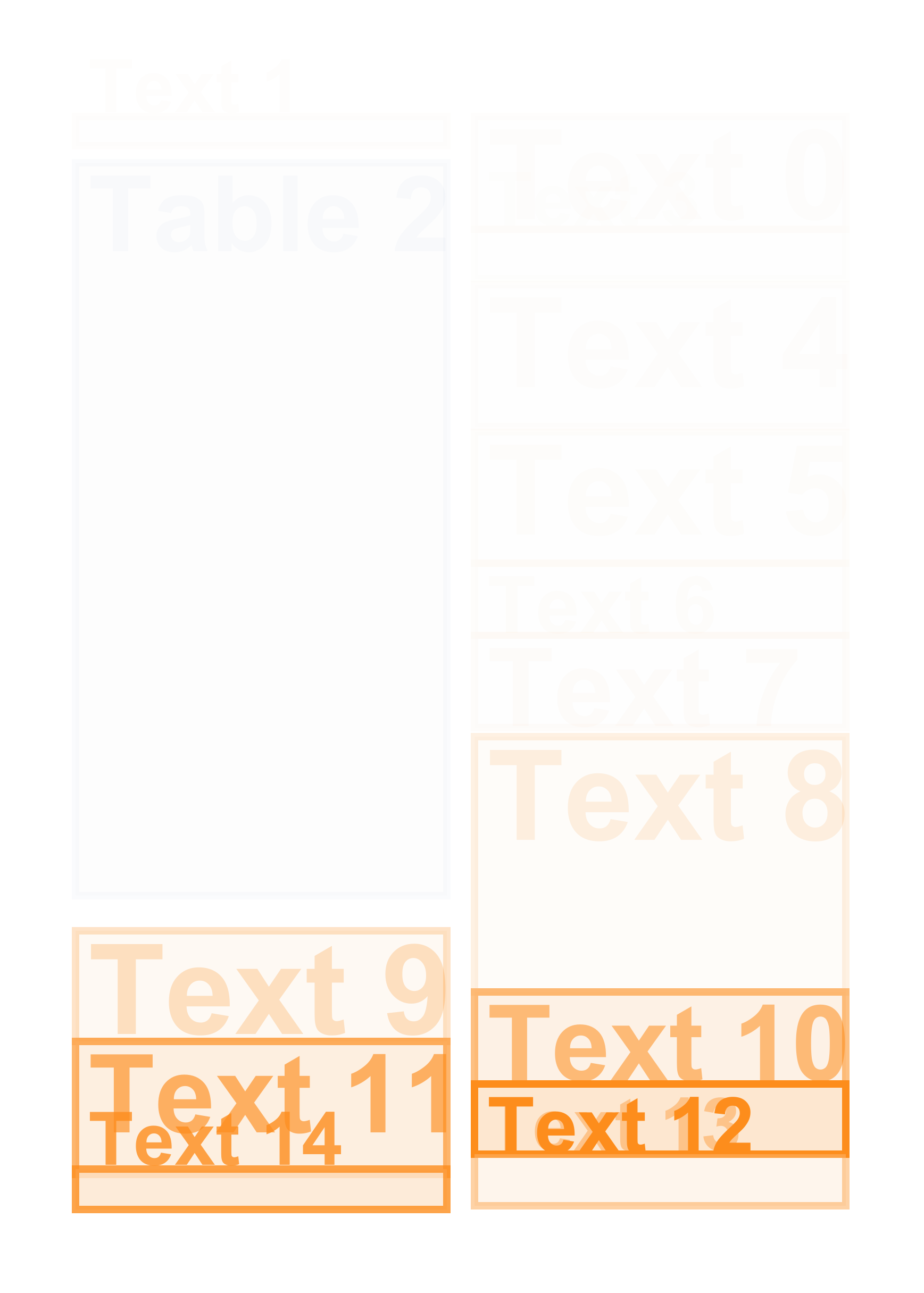} \\

    \rotatebox{90}{\hspace{0.3cm}\footnotesize Layer 3}  &
    \includegraphics[width=\attentionVisDimmedWidth,frame=0.1pt]{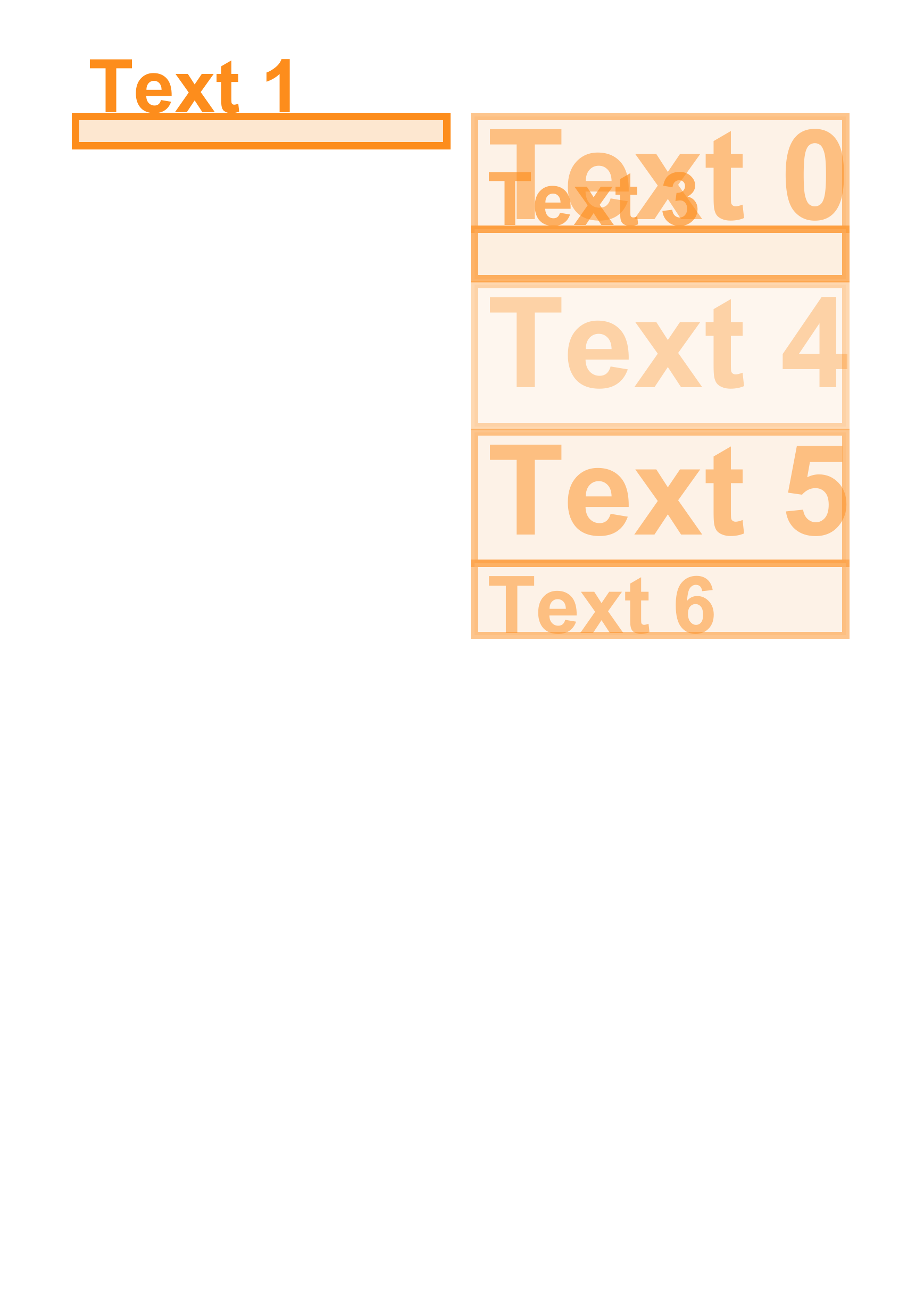} &
    \includegraphics[width=\attentionVisDimmedWidth,frame=0.1pt]{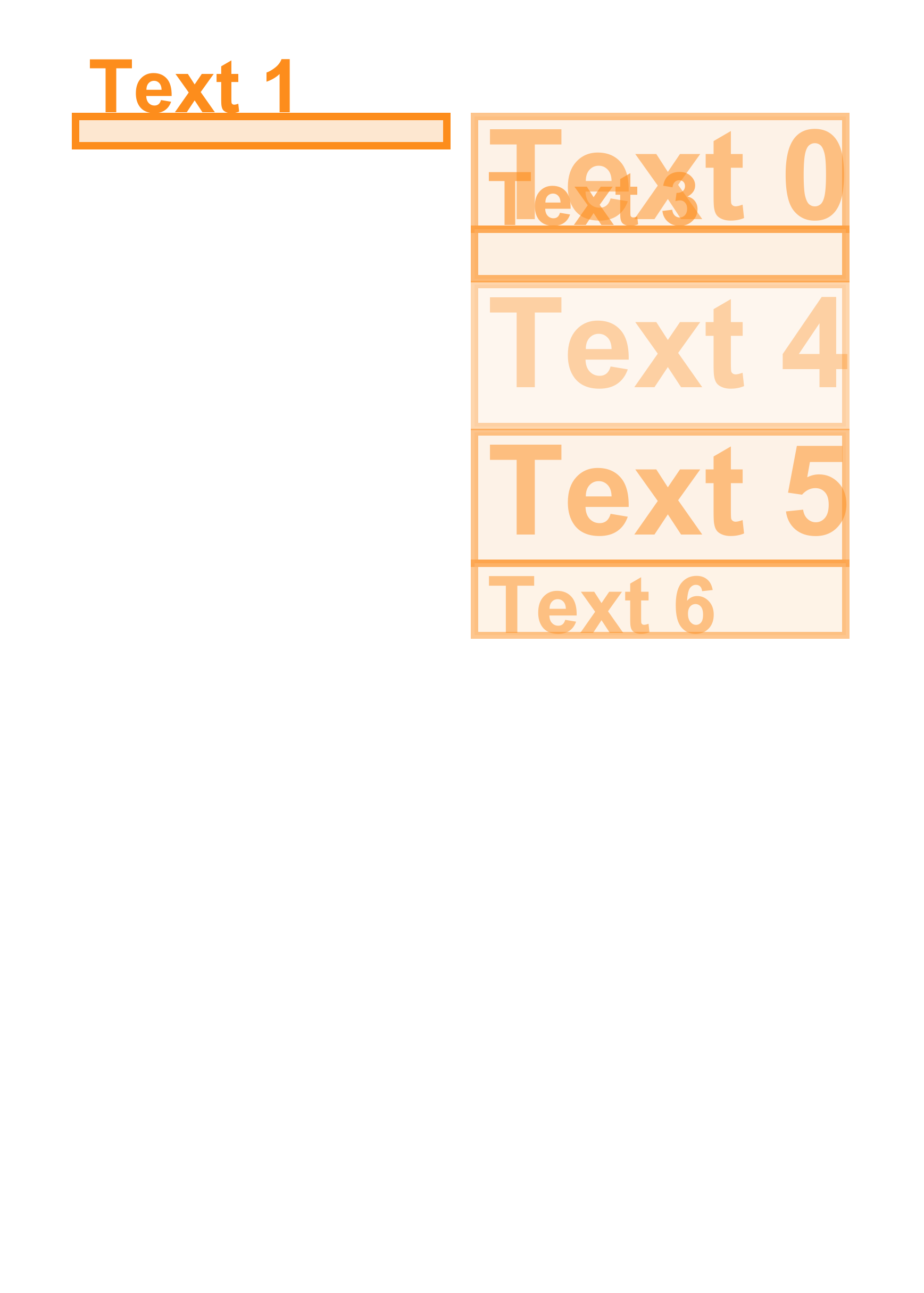} &
    \includegraphics[width=\attentionVisDimmedWidth,frame=0.1pt]{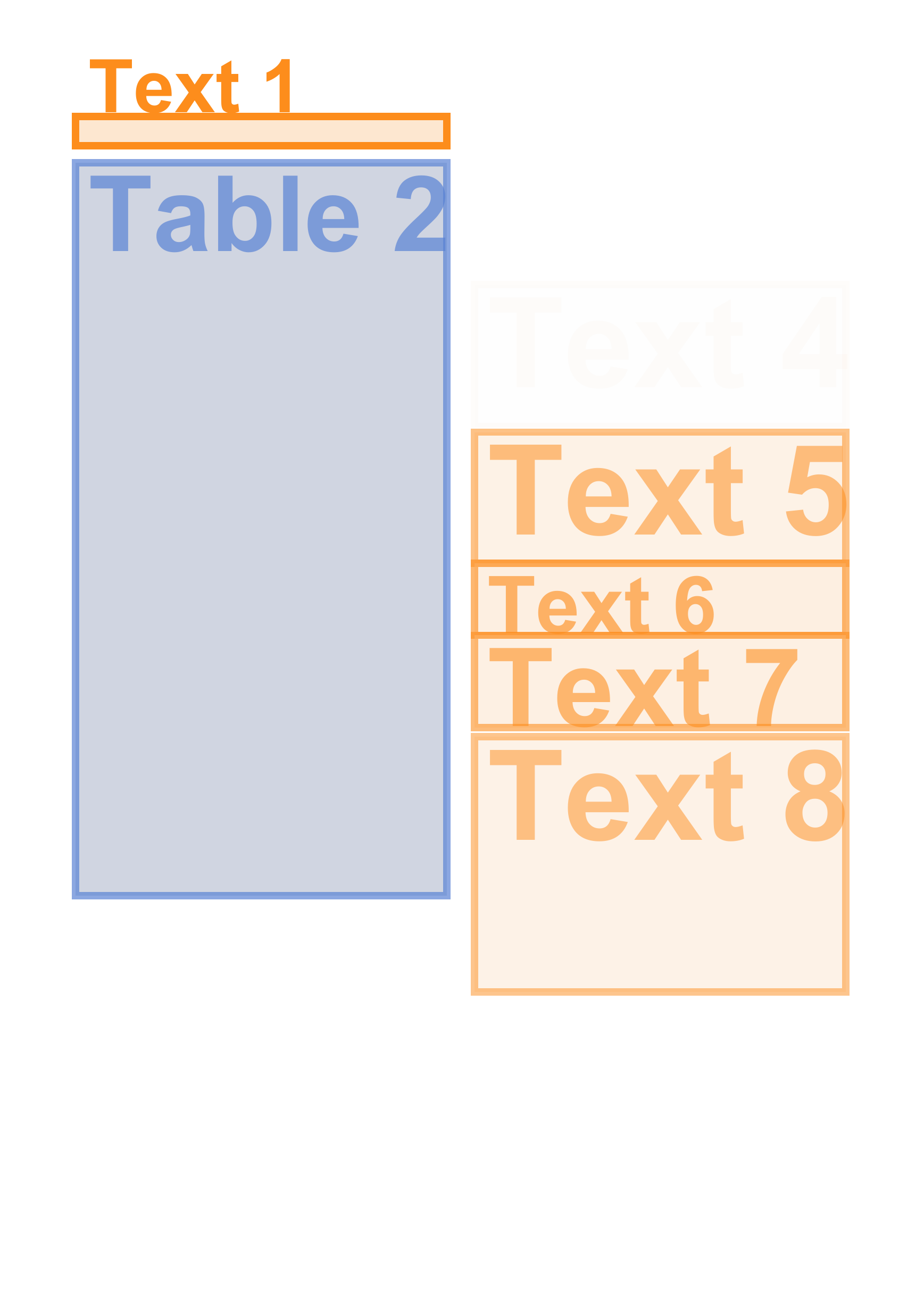} &
    \includegraphics[width=\attentionVisDimmedWidth,frame=0.1pt]{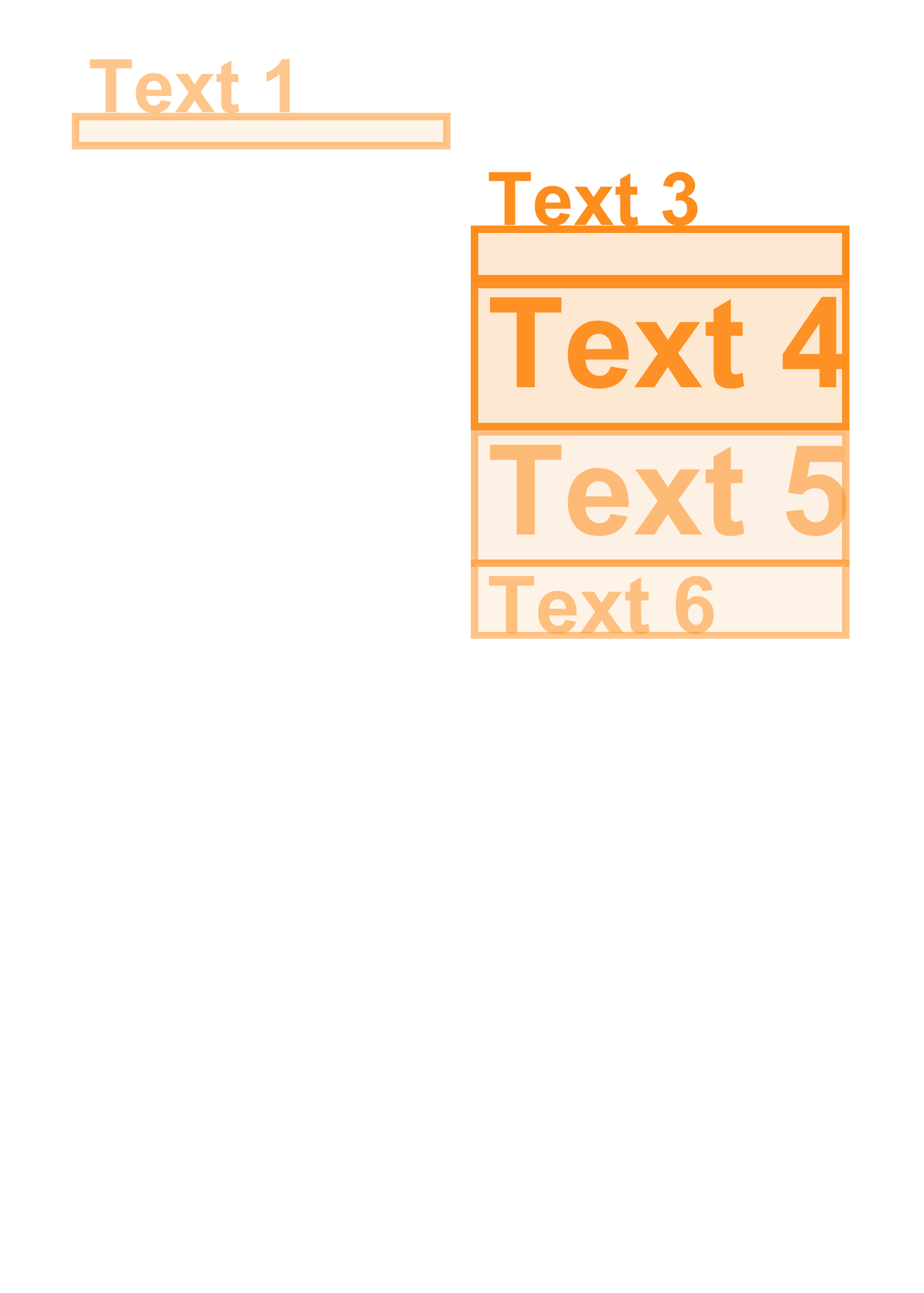} &
    \includegraphics[width=\attentionVisDimmedWidth,frame=0.1pt]{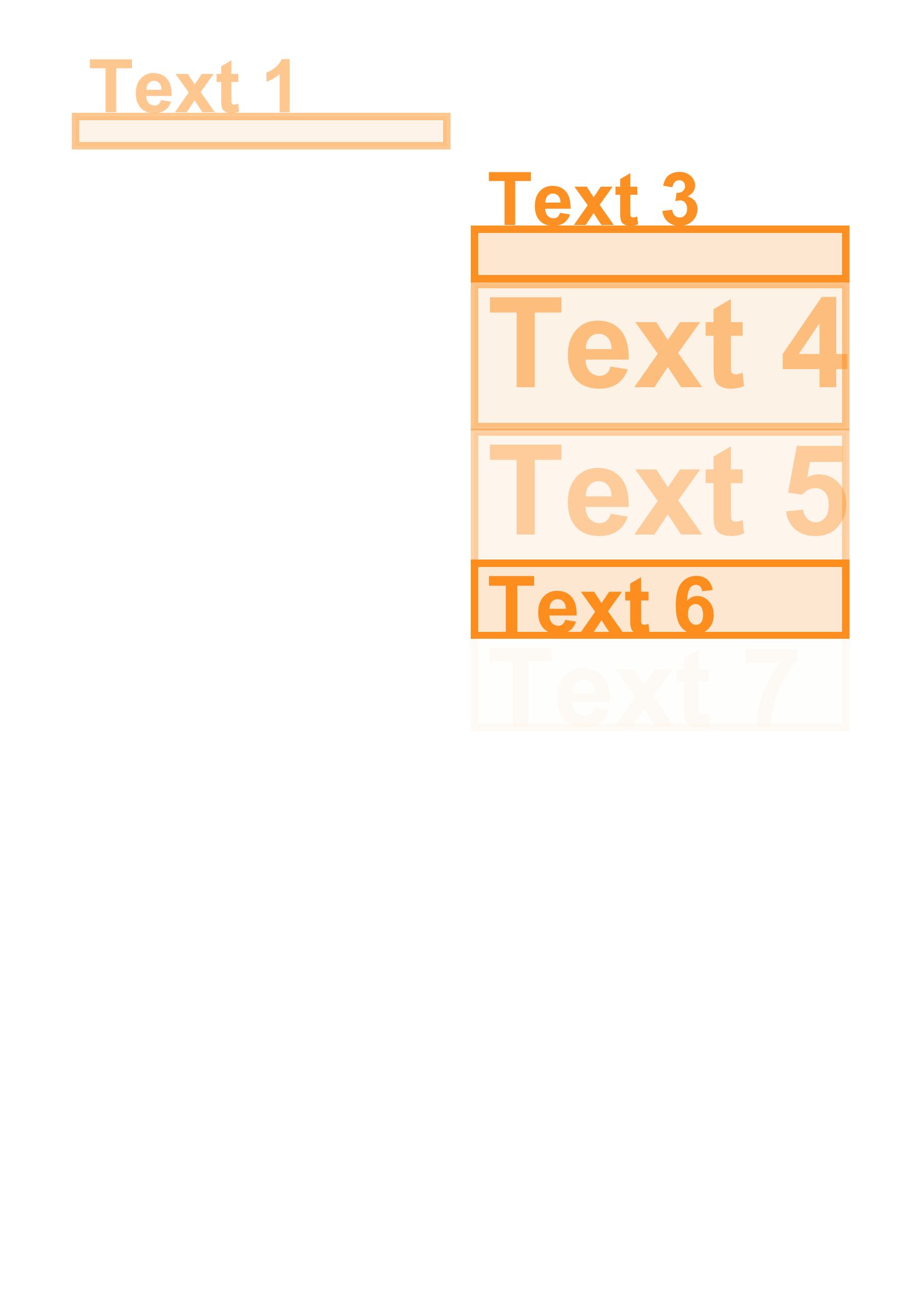} &
    \includegraphics[width=\attentionVisDimmedWidth,frame=0.1pt]{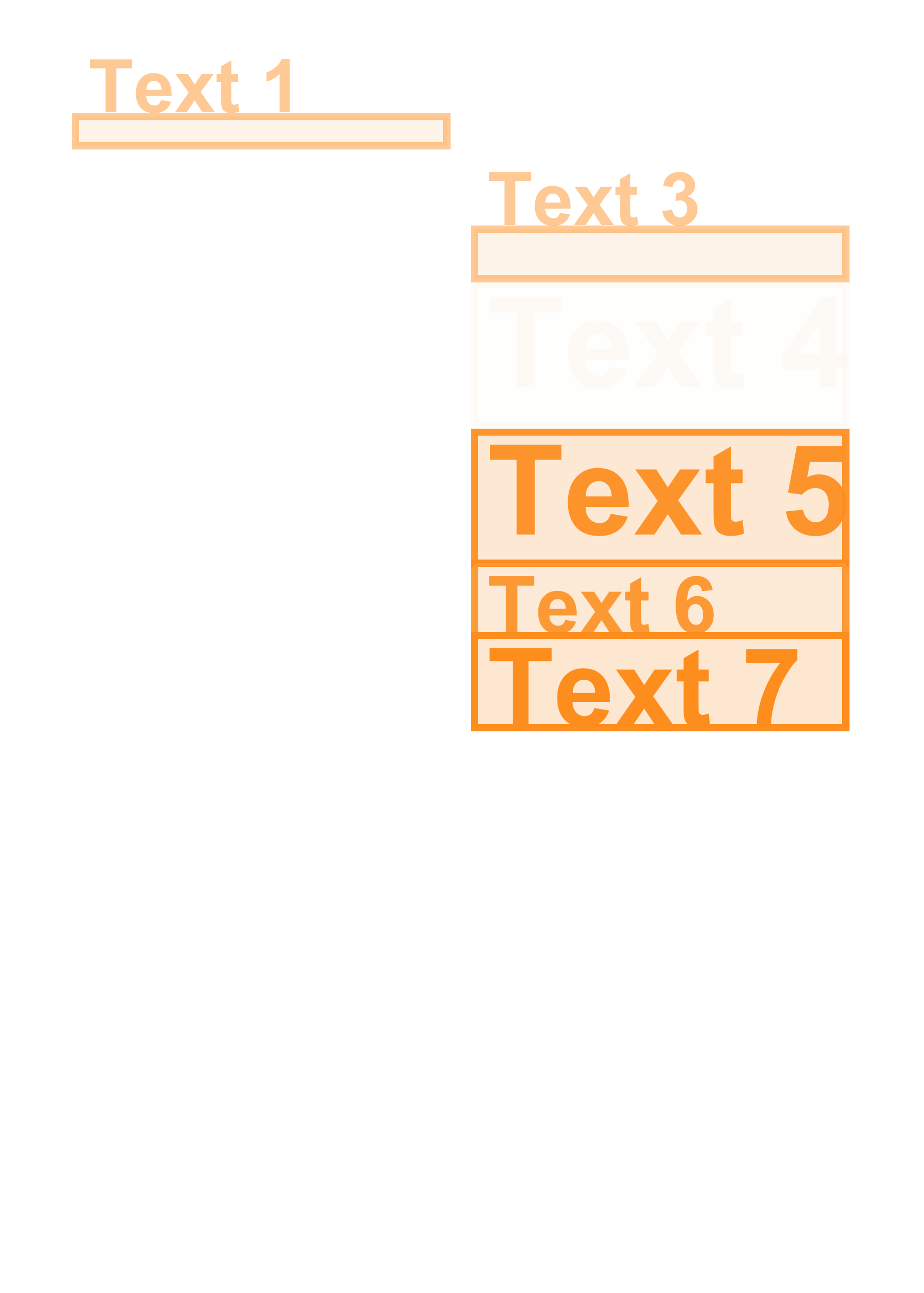} &
    \includegraphics[width=\attentionVisDimmedWidth,frame=0.1pt]{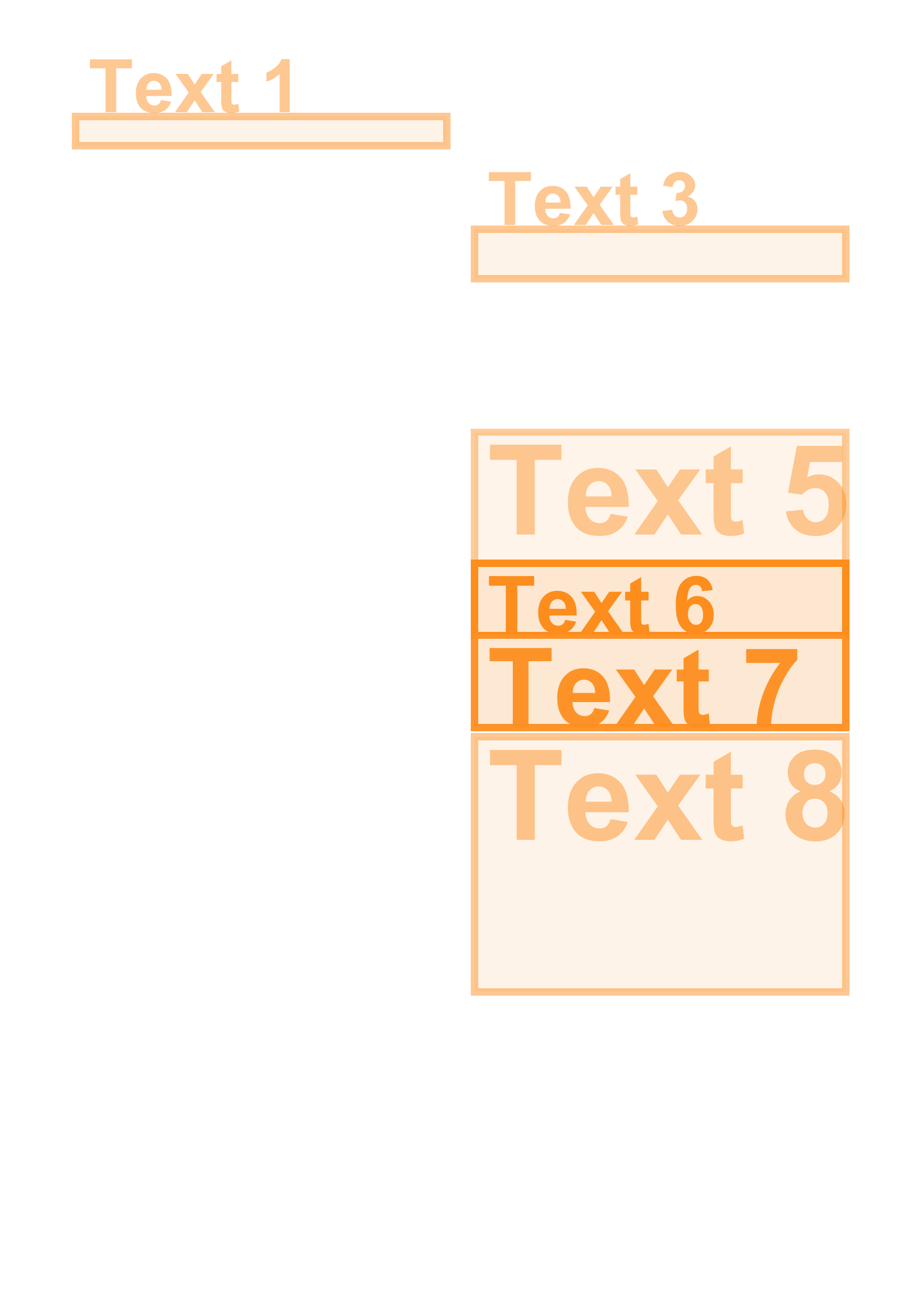} &
    \includegraphics[width=\attentionVisDimmedWidth,frame=0.1pt]{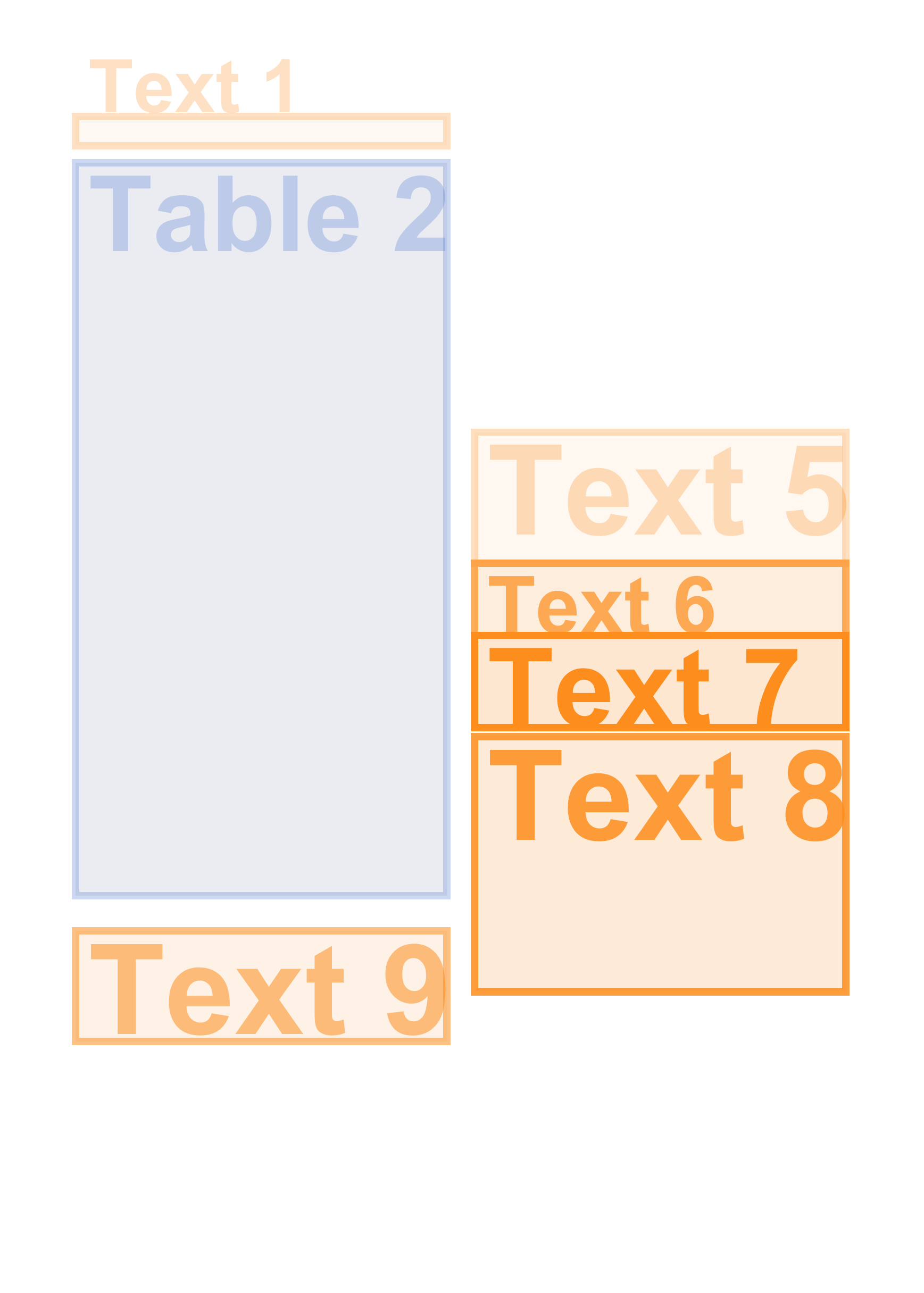} &
    \includegraphics[width=\attentionVisDimmedWidth,frame=0.1pt]{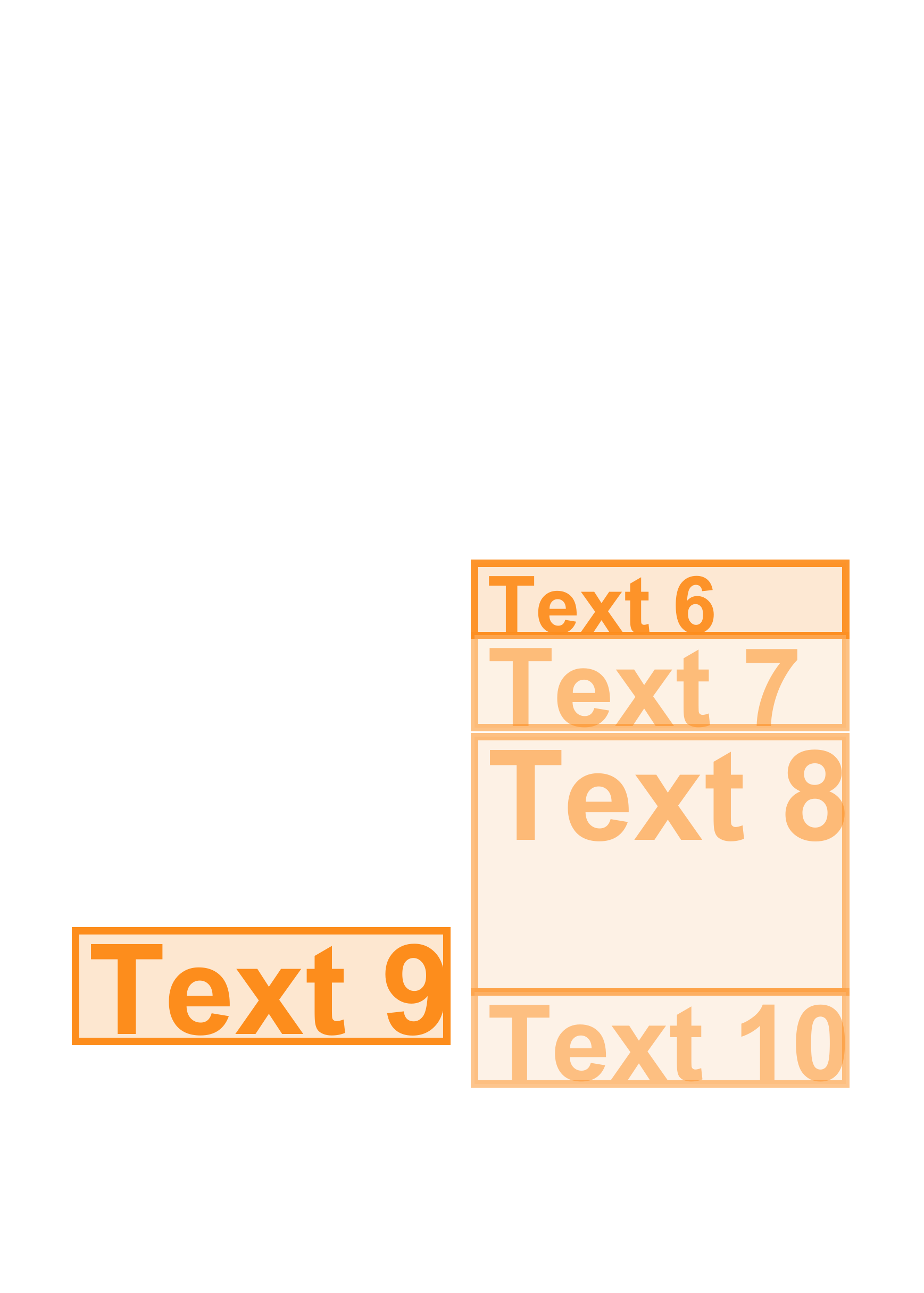} &
    \includegraphics[width=\attentionVisDimmedWidth,frame=0.1pt]{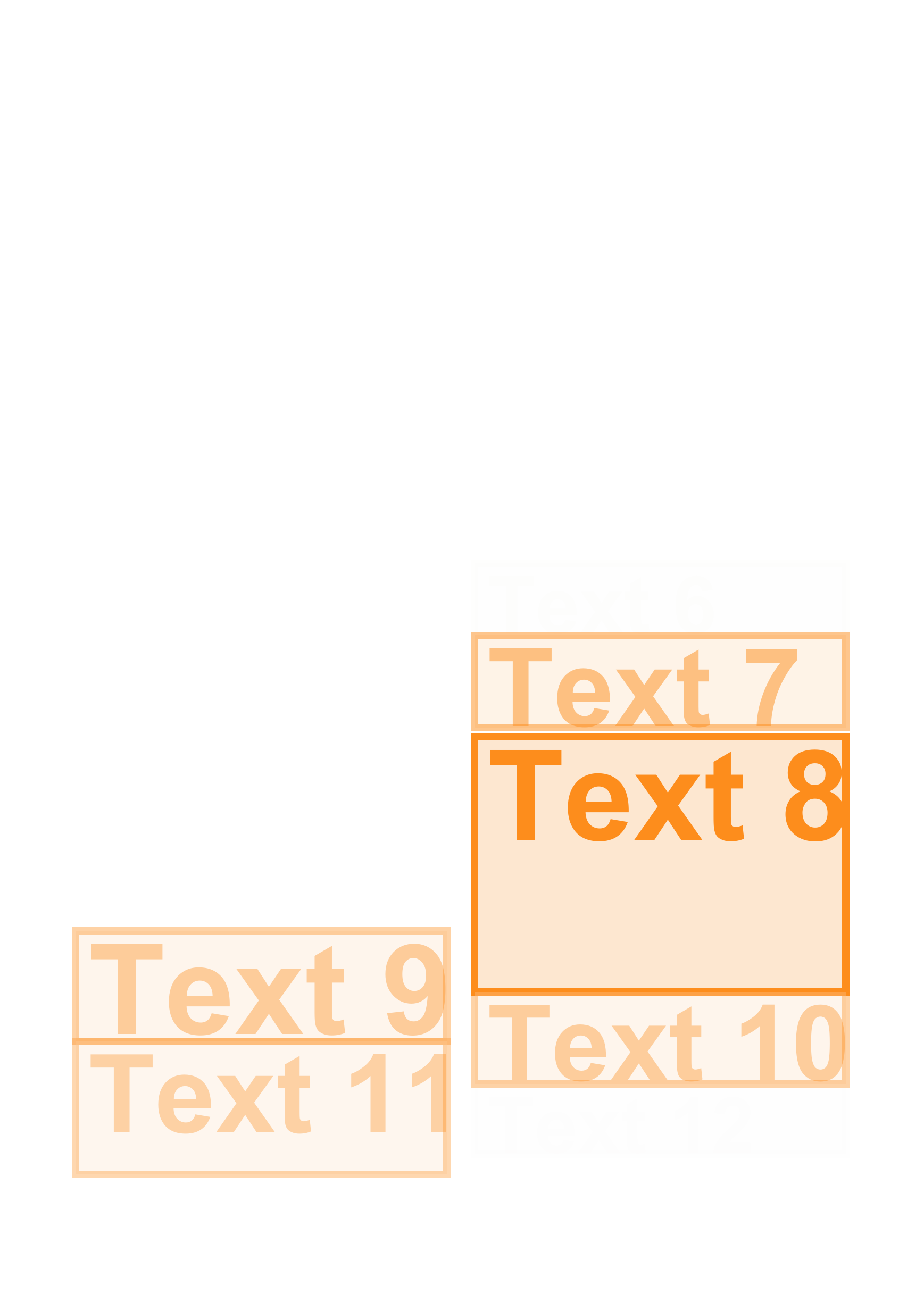} &
    \includegraphics[width=\attentionVisDimmedWidth,frame=0.1pt]{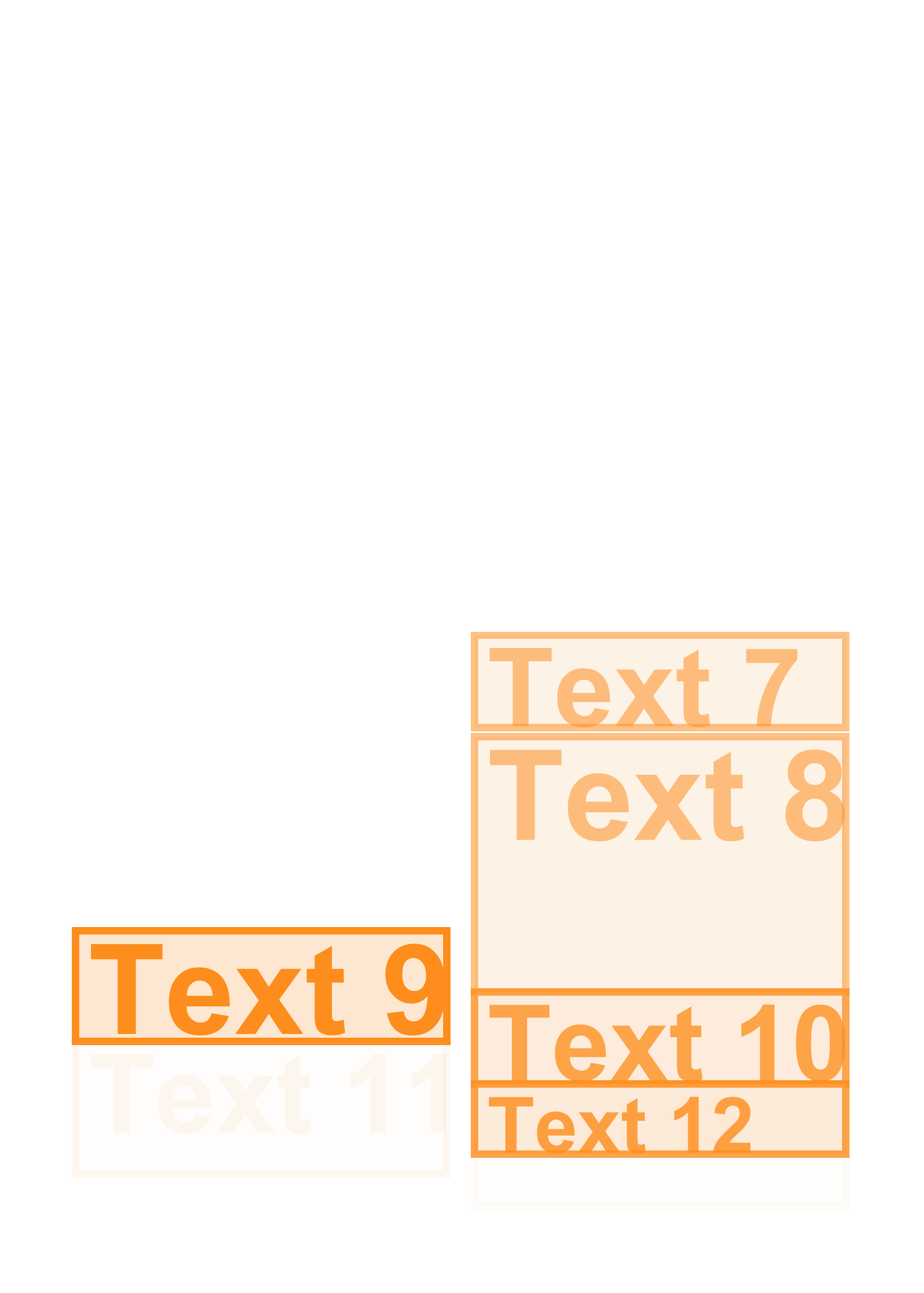} &
    \includegraphics[width=\attentionVisDimmedWidth,frame=0.1pt]{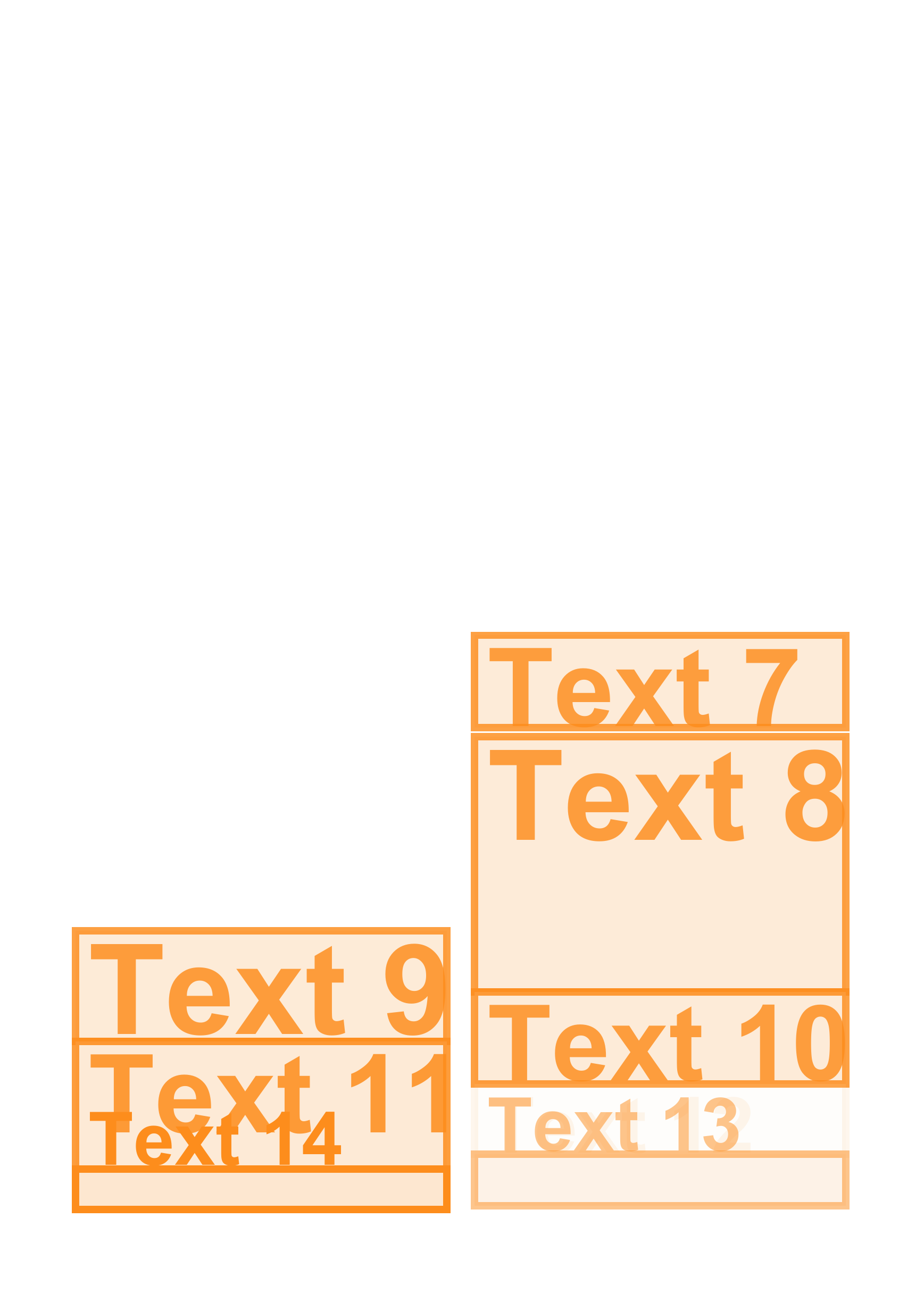} &
    \includegraphics[width=\attentionVisDimmedWidth,frame=0.1pt]{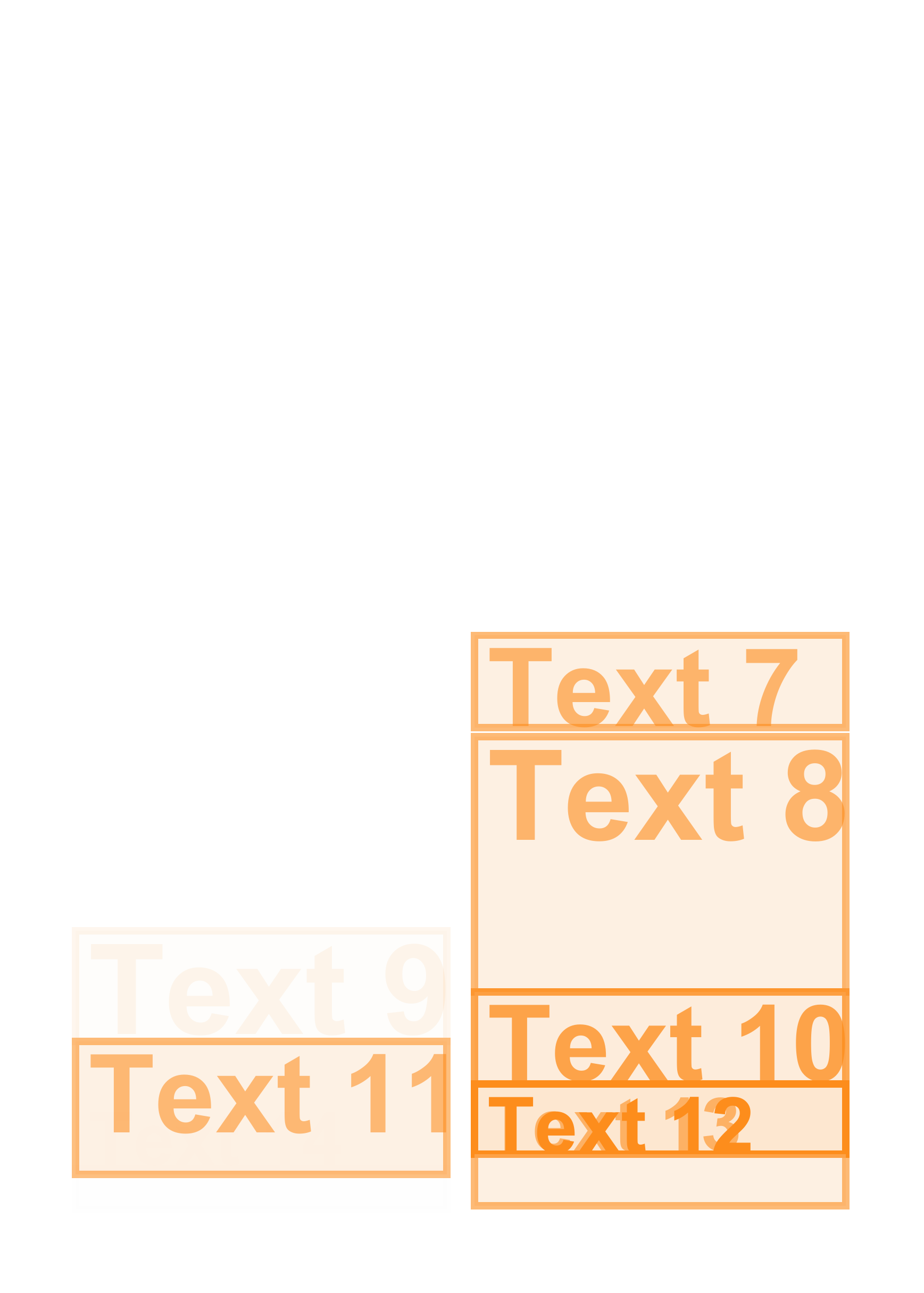} &
    \includegraphics[width=\attentionVisDimmedWidth,frame=0.1pt]{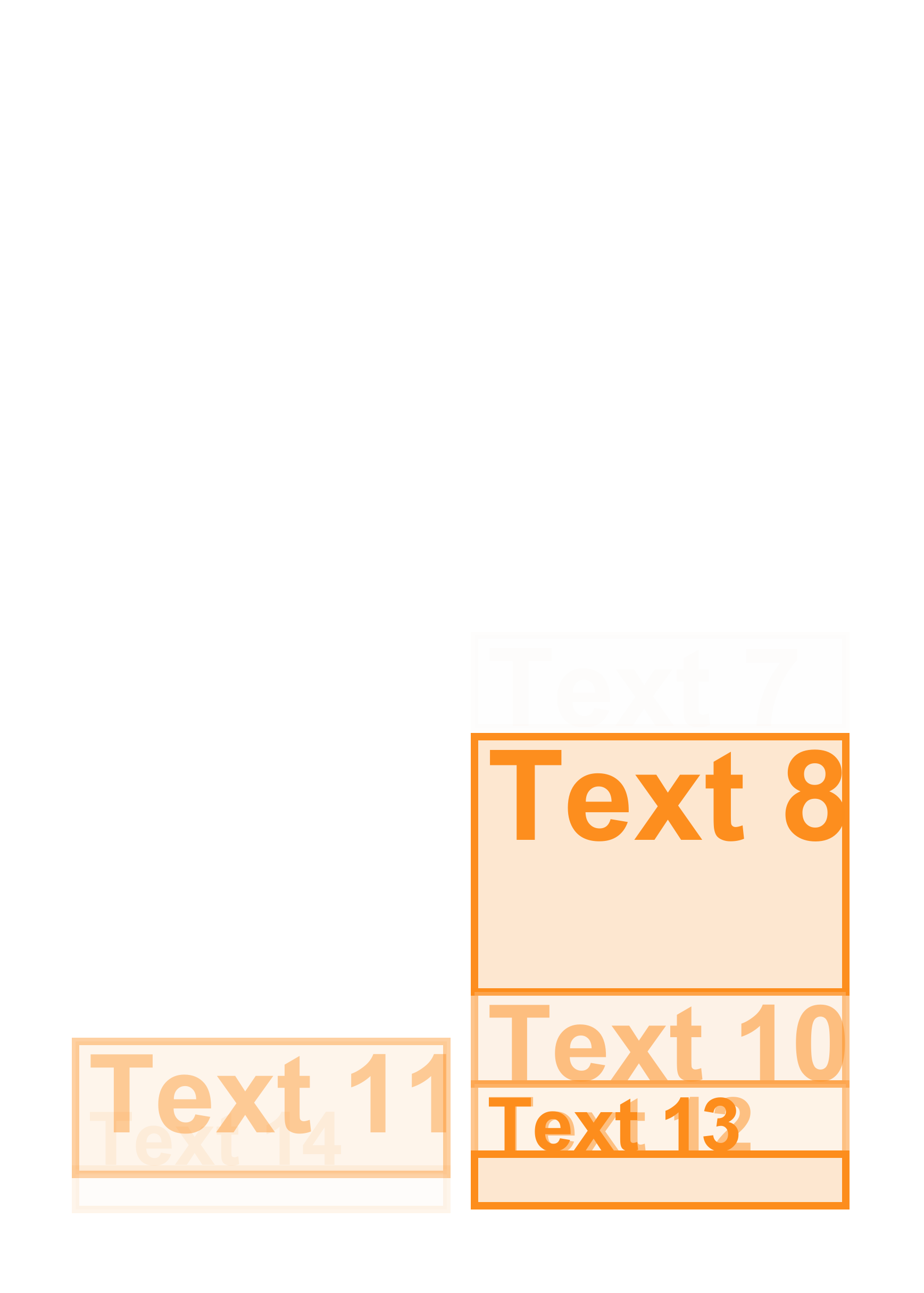} &
    \includegraphics[width=\attentionVisDimmedWidth,frame=0.1pt]{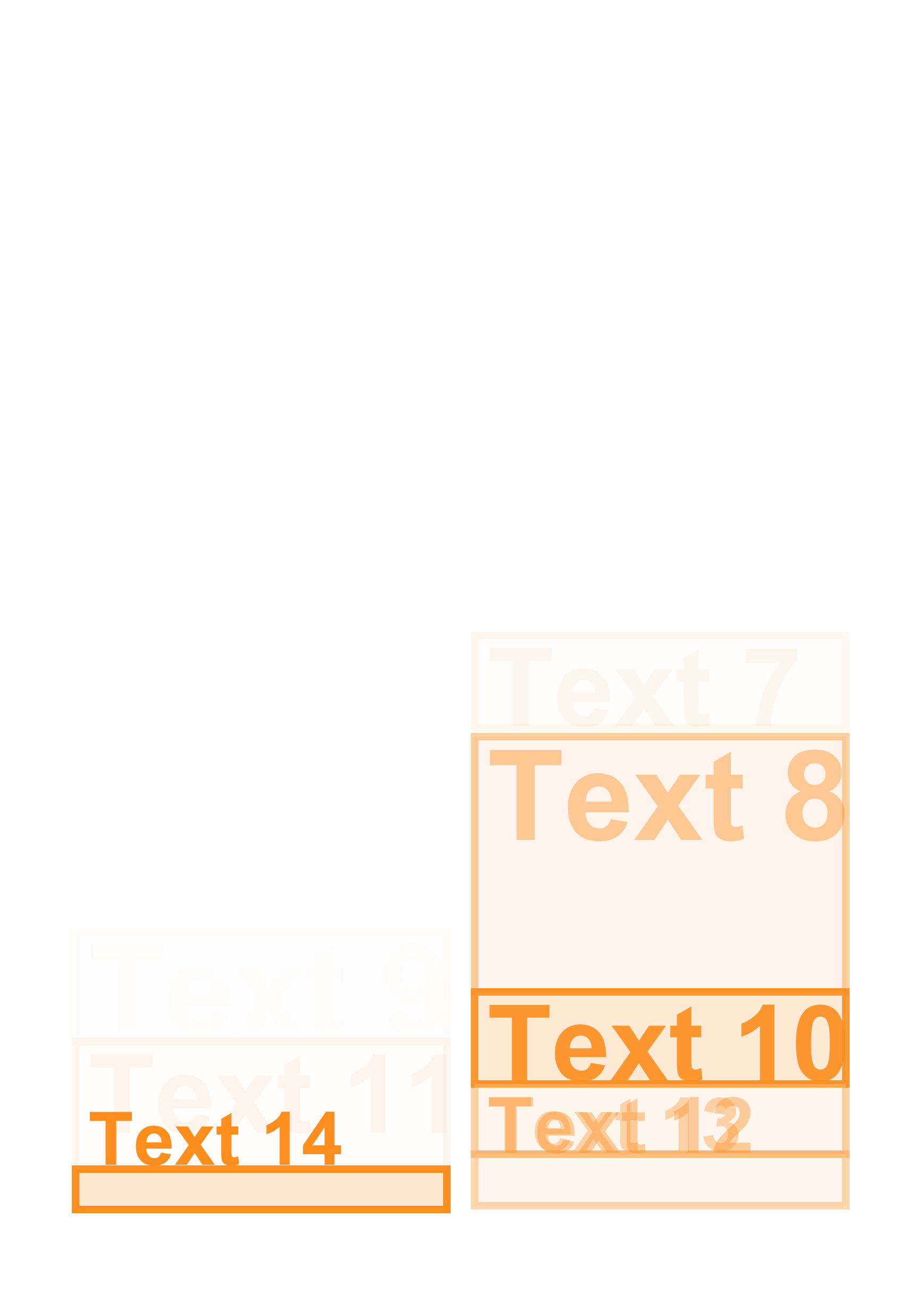} \\

    \rotatebox{90}{\hspace{0.3cm}\footnotesize Layer 4}  & \includegraphics[width=\attentionVisDimmedWidth,frame=0.1pt]{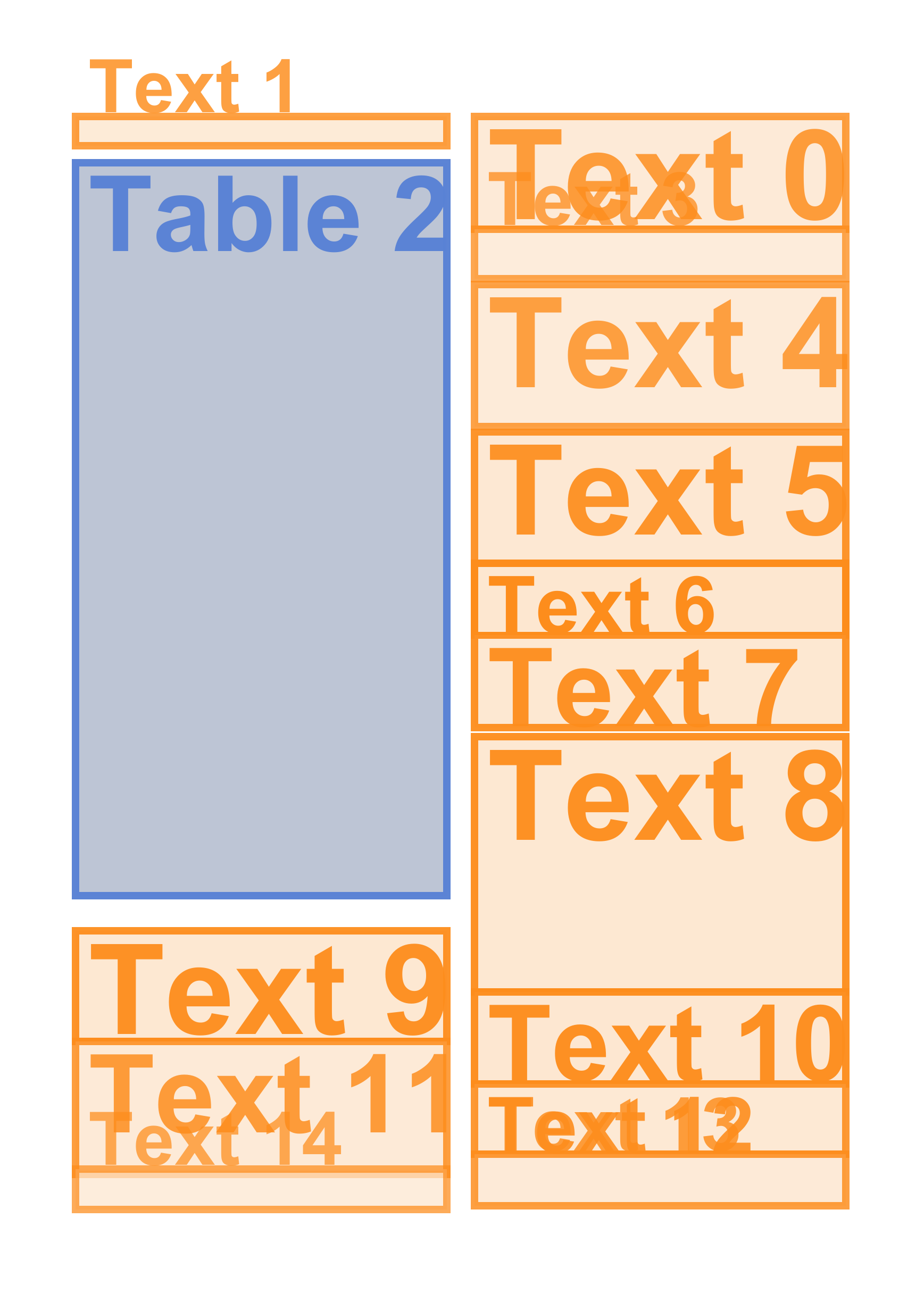} &
    \includegraphics[width=\attentionVisDimmedWidth,frame=0.1pt]{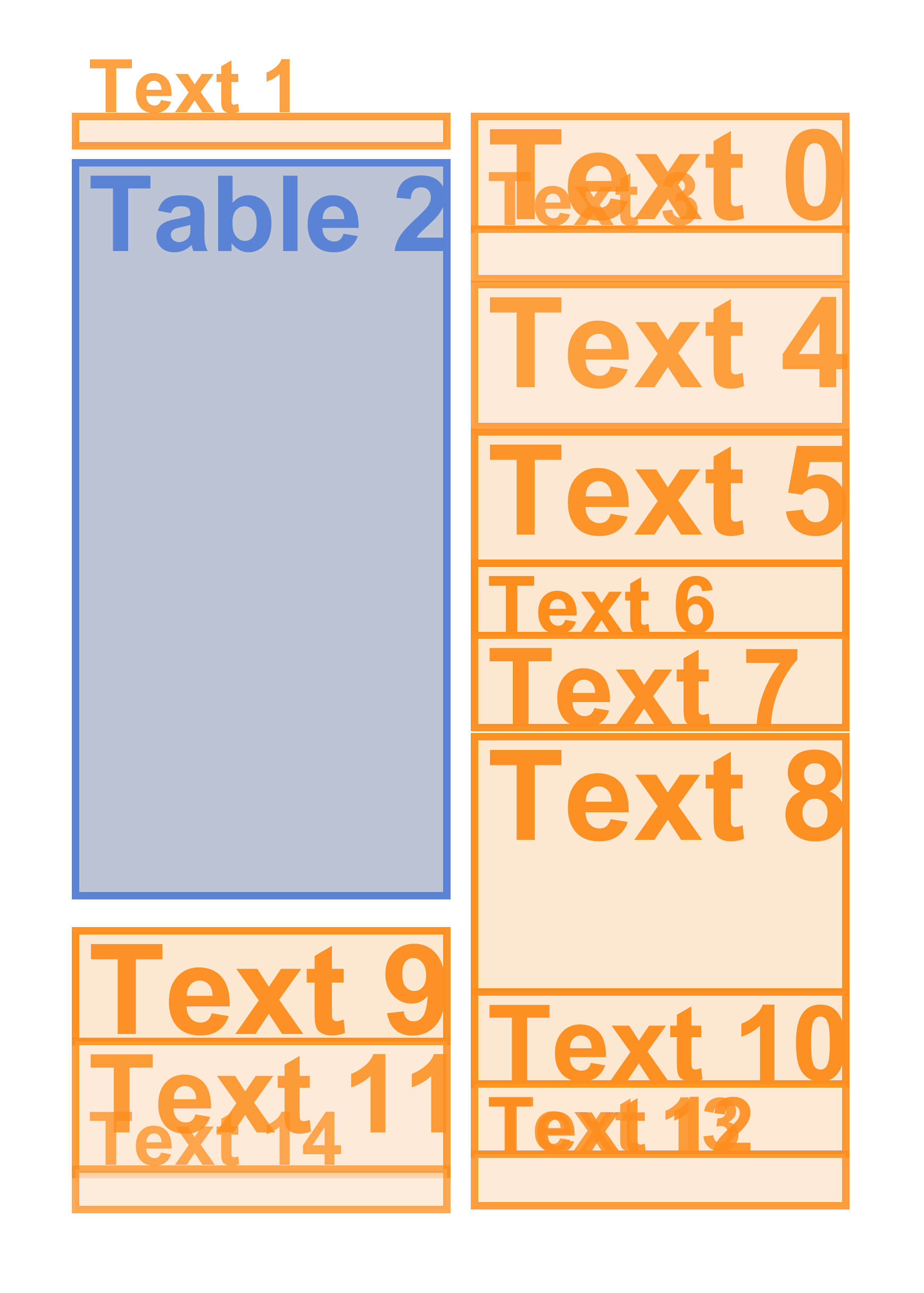} &
    \includegraphics[width=\attentionVisDimmedWidth,frame=0.1pt]{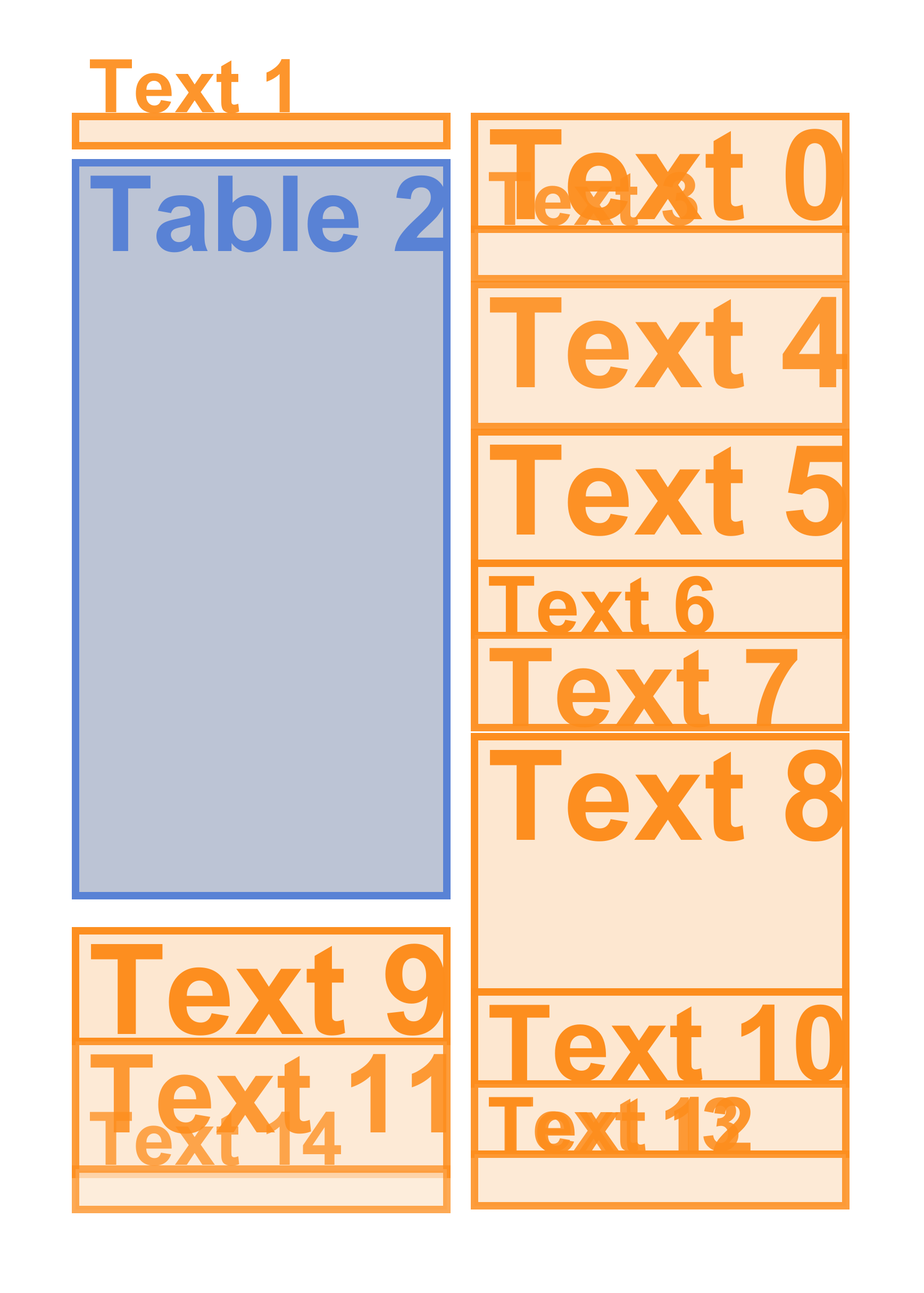} &
    \includegraphics[width=\attentionVisDimmedWidth,frame=0.1pt]{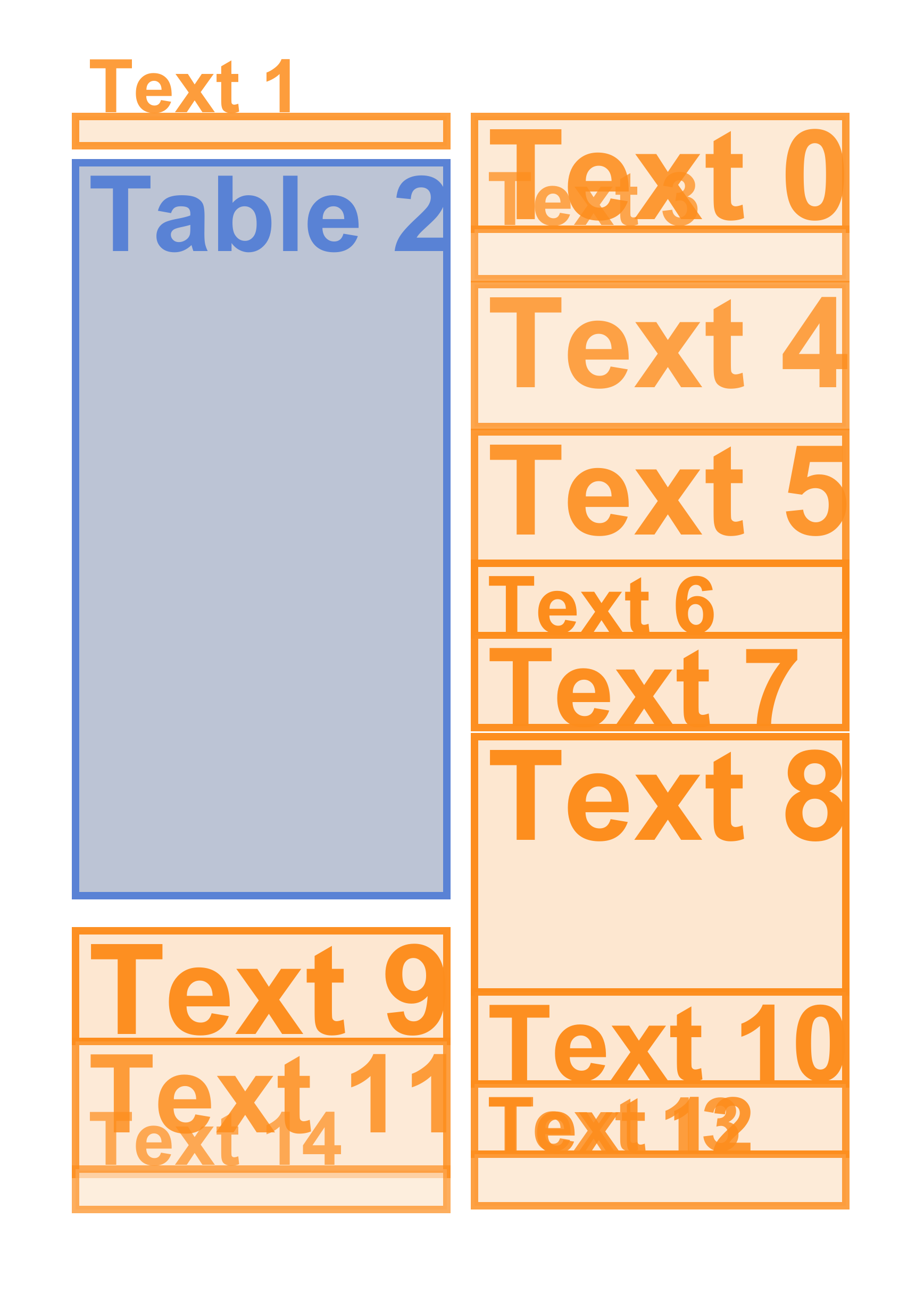} &
    \includegraphics[width=\attentionVisDimmedWidth,frame=0.1pt]{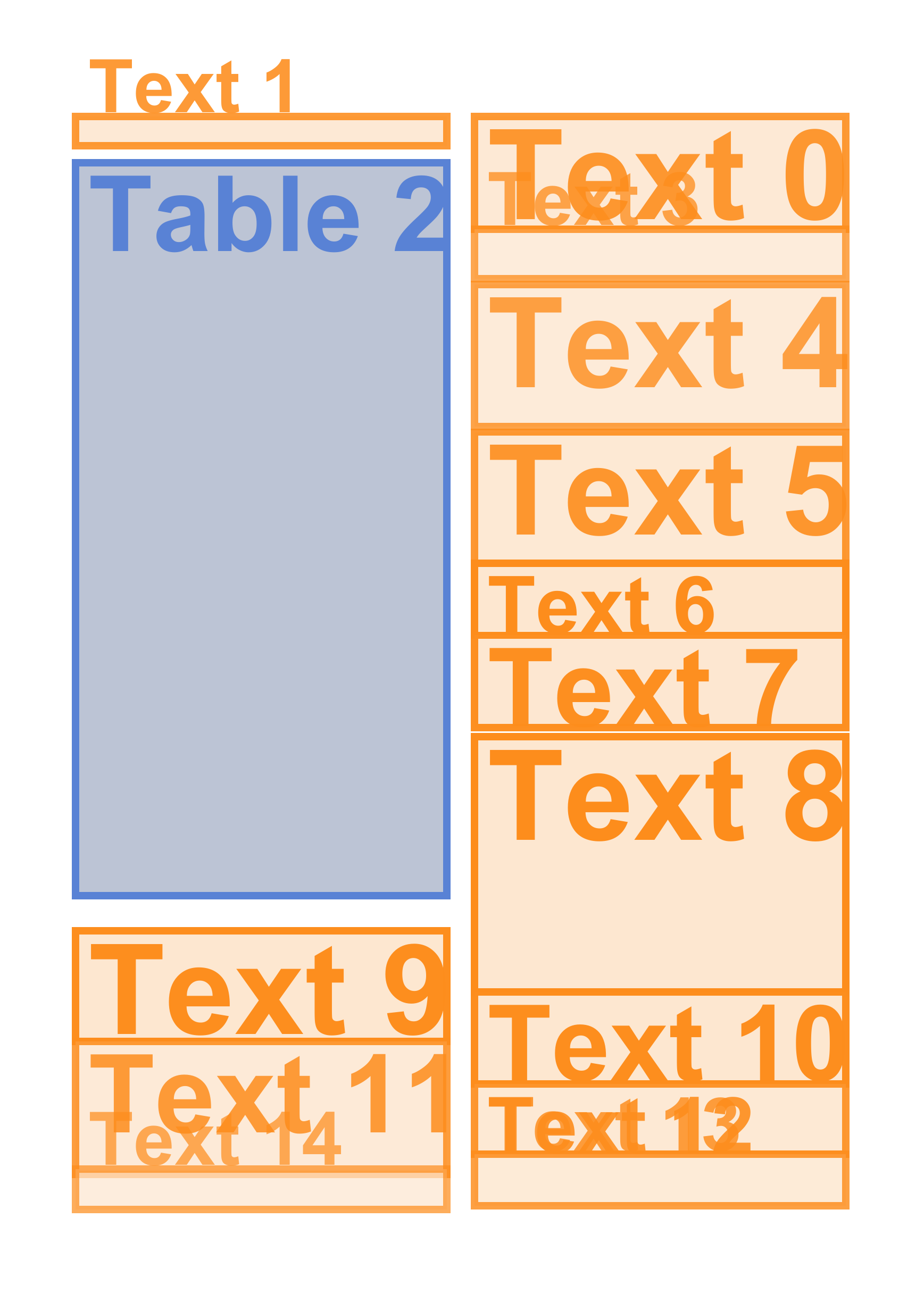} &
    \includegraphics[width=\attentionVisDimmedWidth,frame=0.1pt]{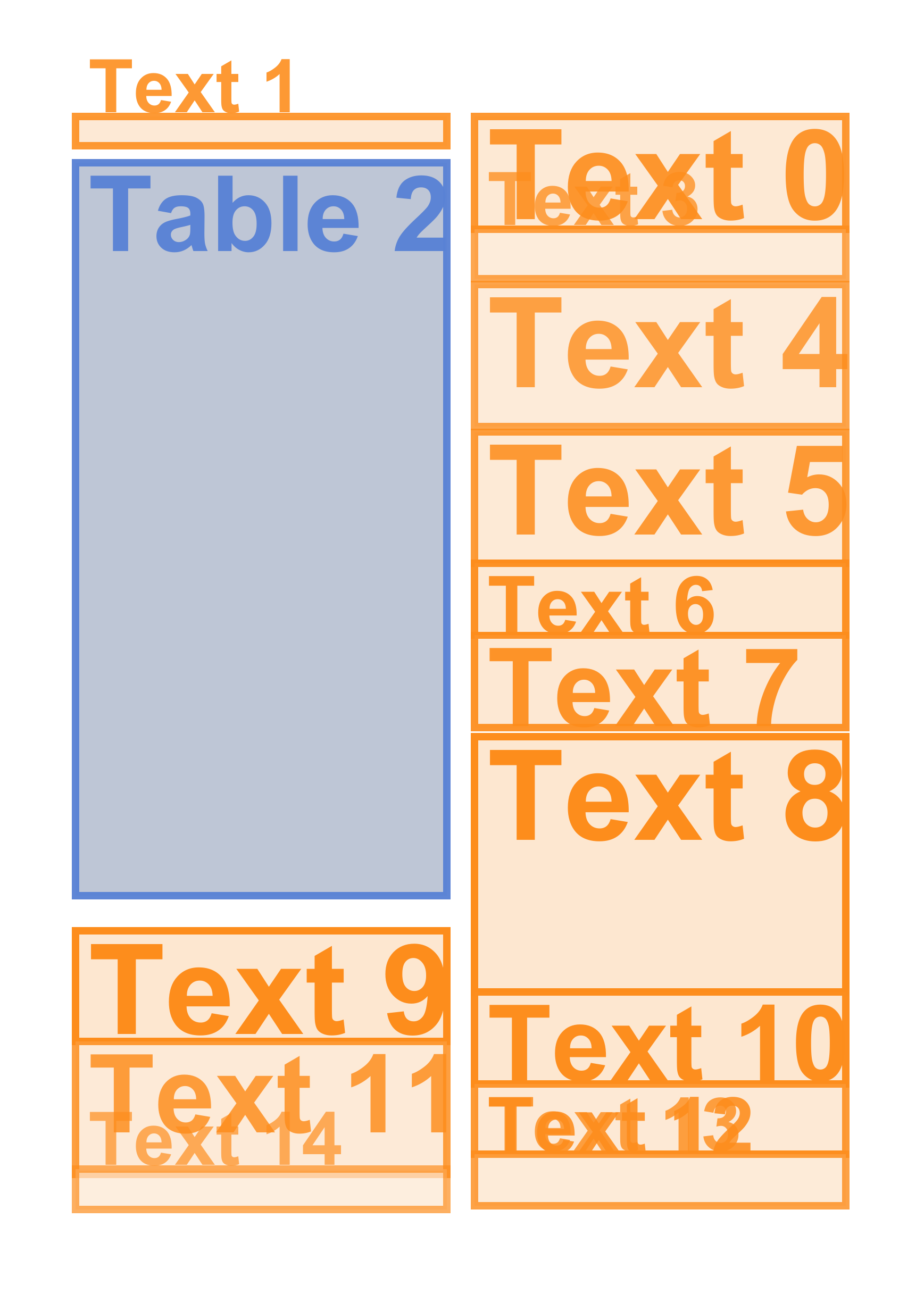} &
    \includegraphics[width=\attentionVisDimmedWidth,frame=0.1pt]{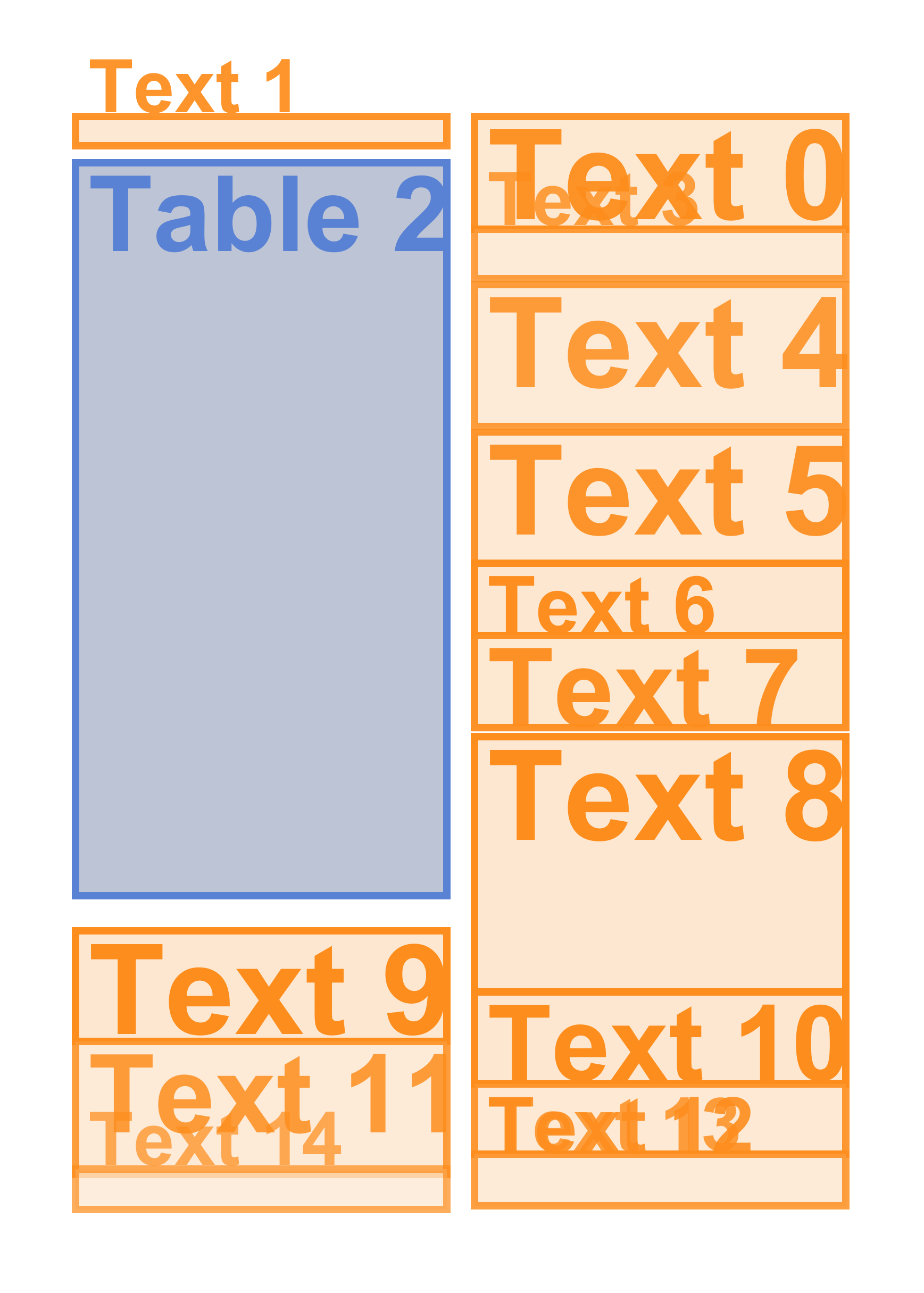} &
    \includegraphics[width=\attentionVisDimmedWidth,frame=0.1pt]{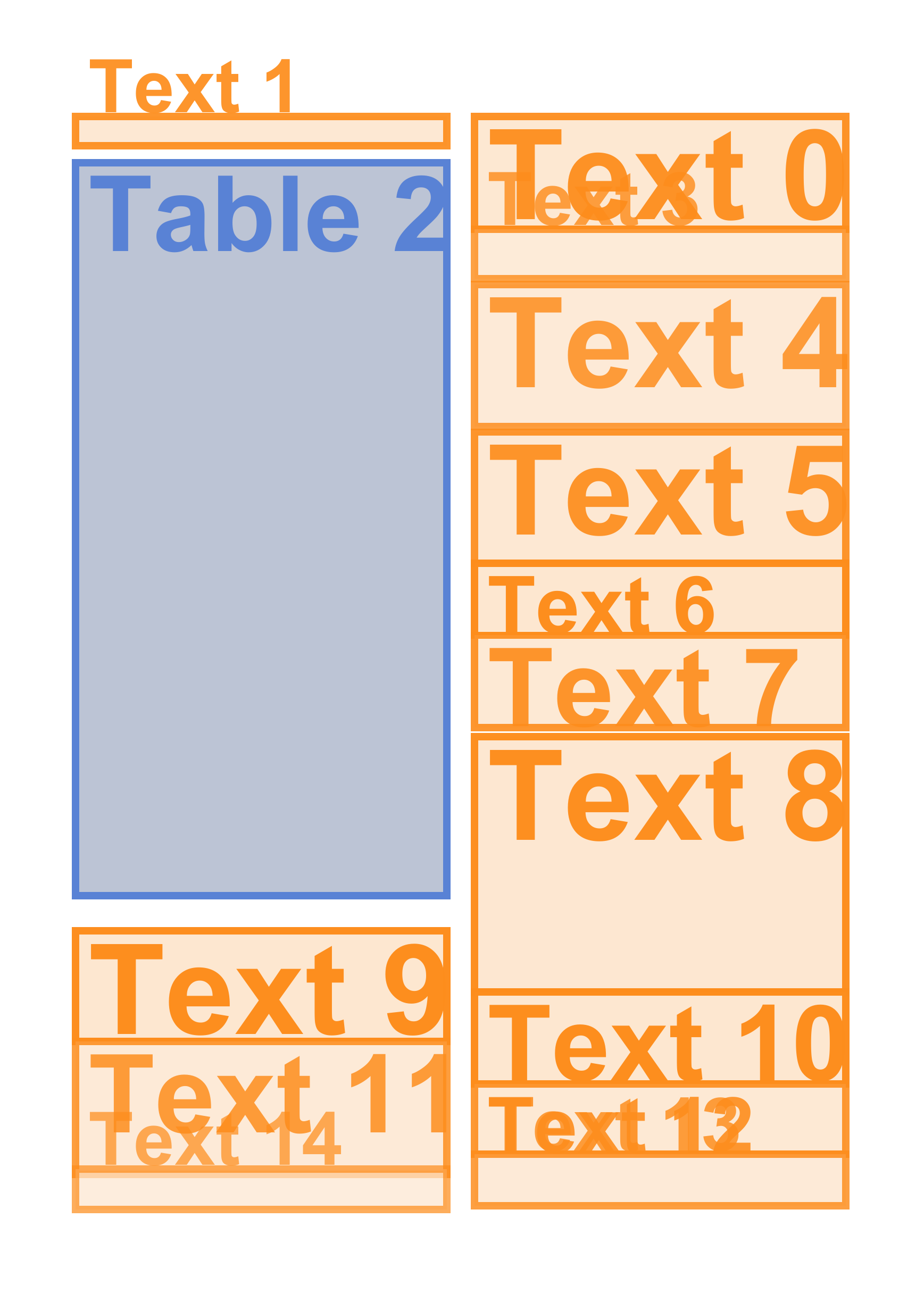} &
    \includegraphics[width=\attentionVisDimmedWidth,frame=0.1pt]{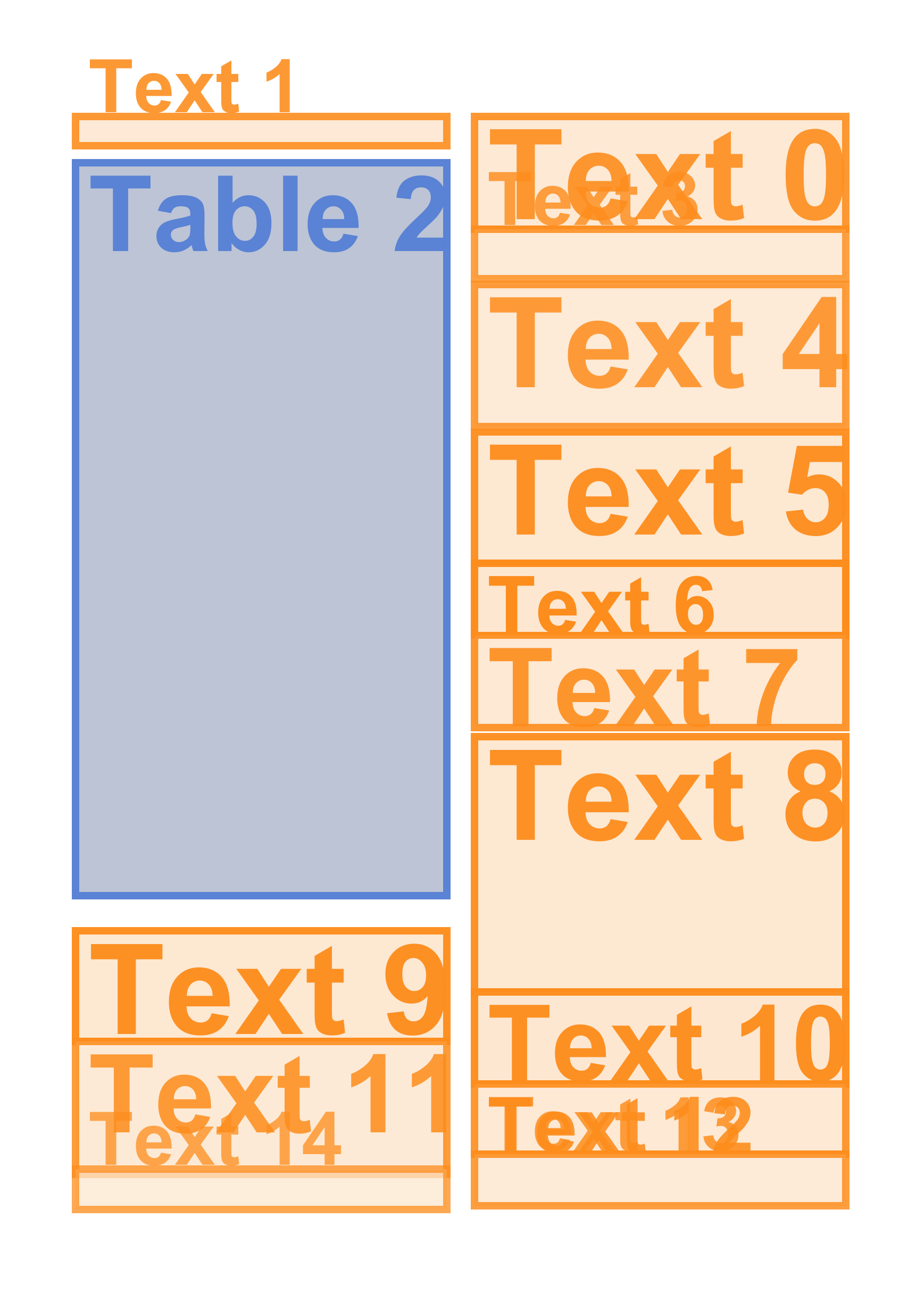} &
    \includegraphics[width=\attentionVisDimmedWidth,frame=0.1pt]{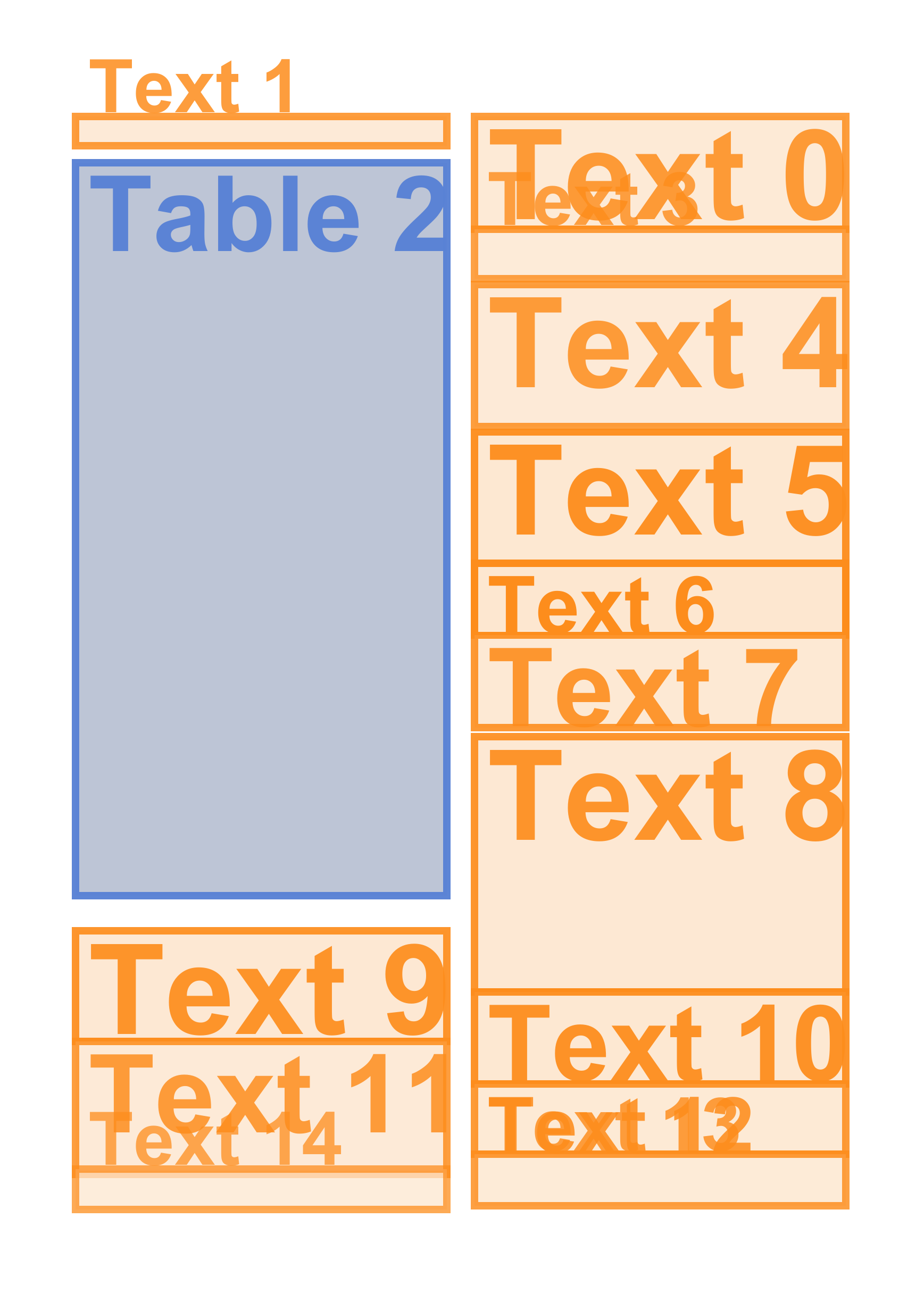} &
    \includegraphics[width=\attentionVisDimmedWidth,frame=0.1pt]{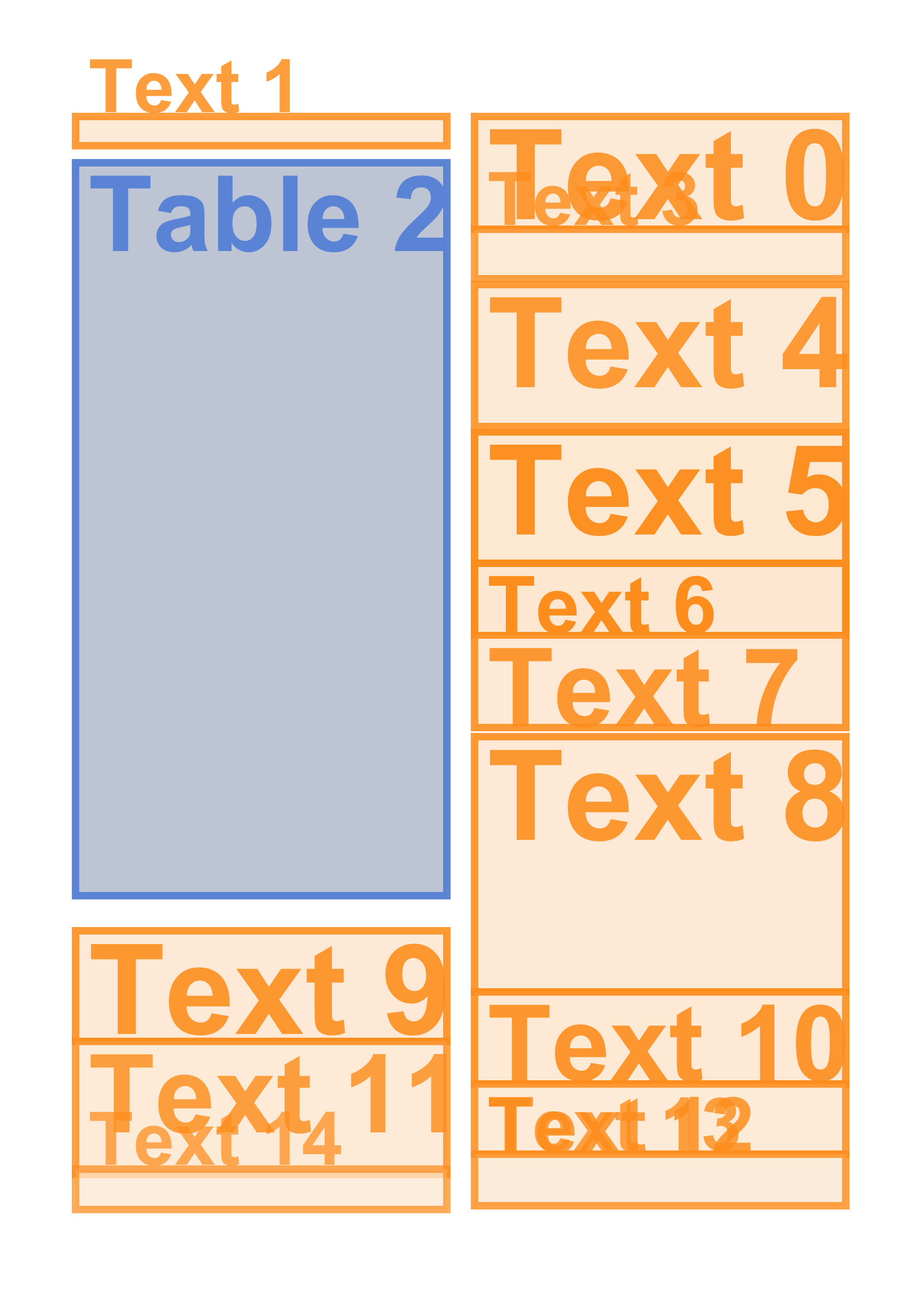} &
    \includegraphics[width=\attentionVisDimmedWidth,frame=0.1pt]{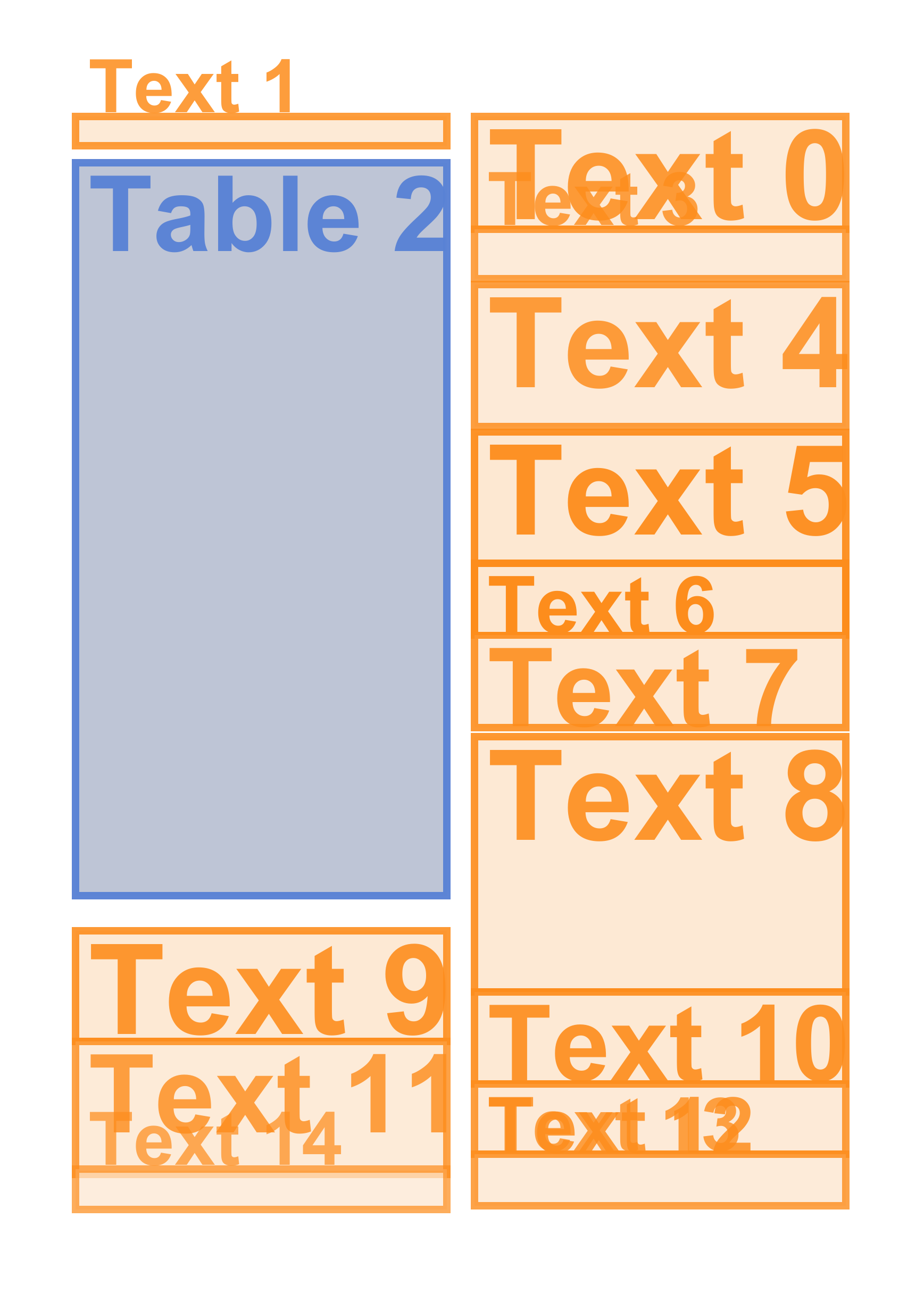} &
    \includegraphics[width=\attentionVisDimmedWidth,frame=0.1pt]{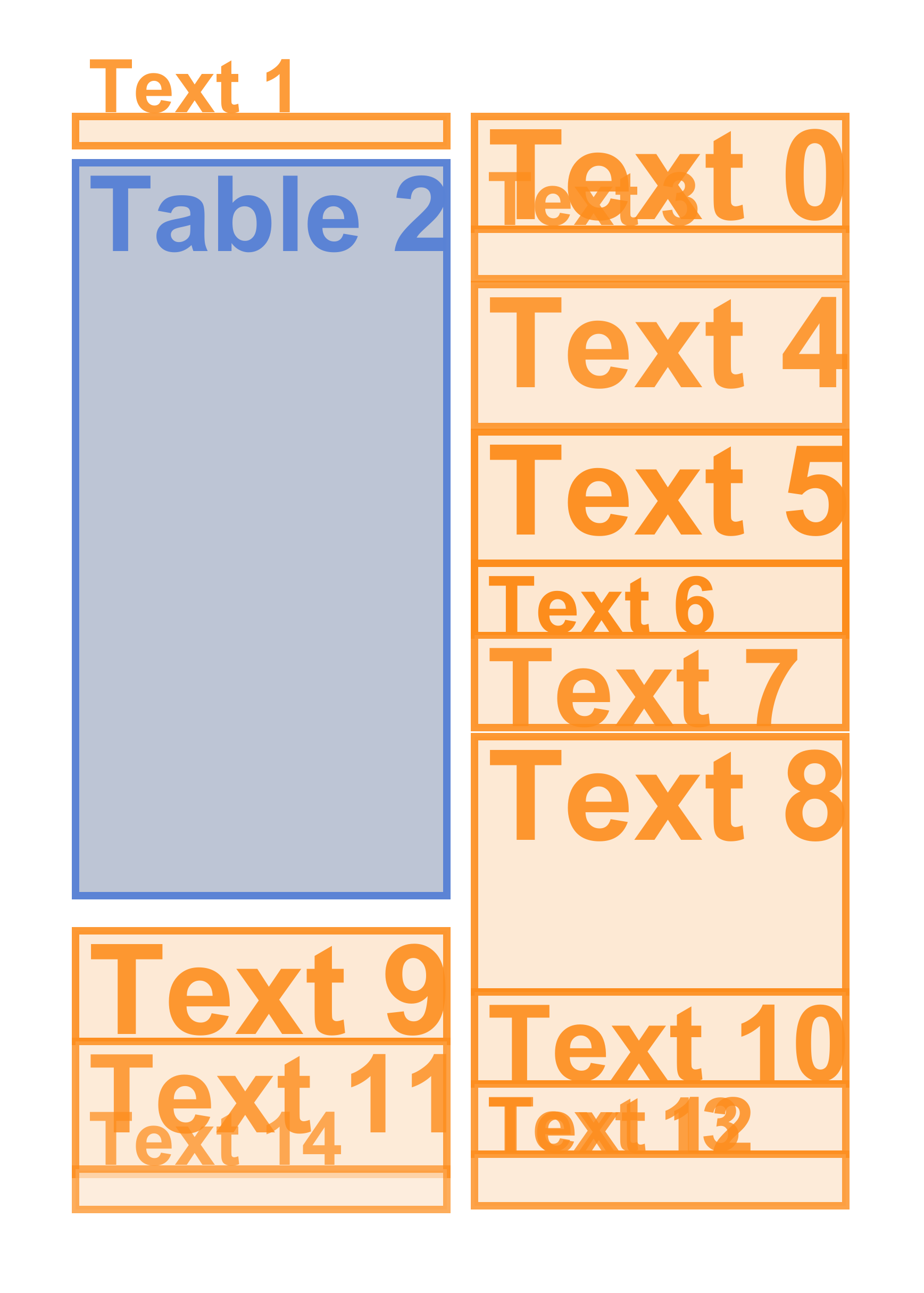} &
    \includegraphics[width=\attentionVisDimmedWidth,frame=0.1pt]{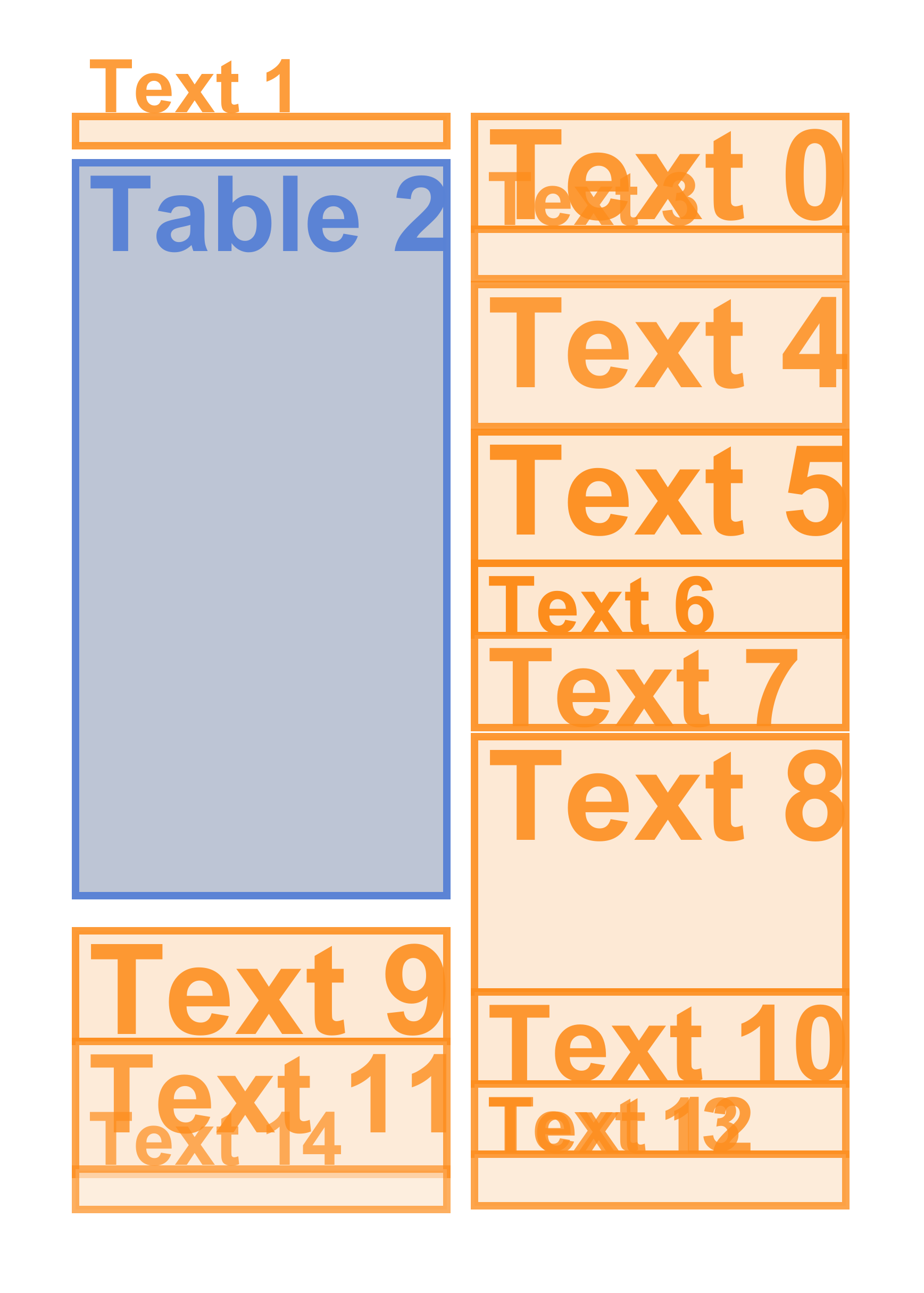} &
    \includegraphics[width=\attentionVisDimmedWidth,frame=0.1pt]{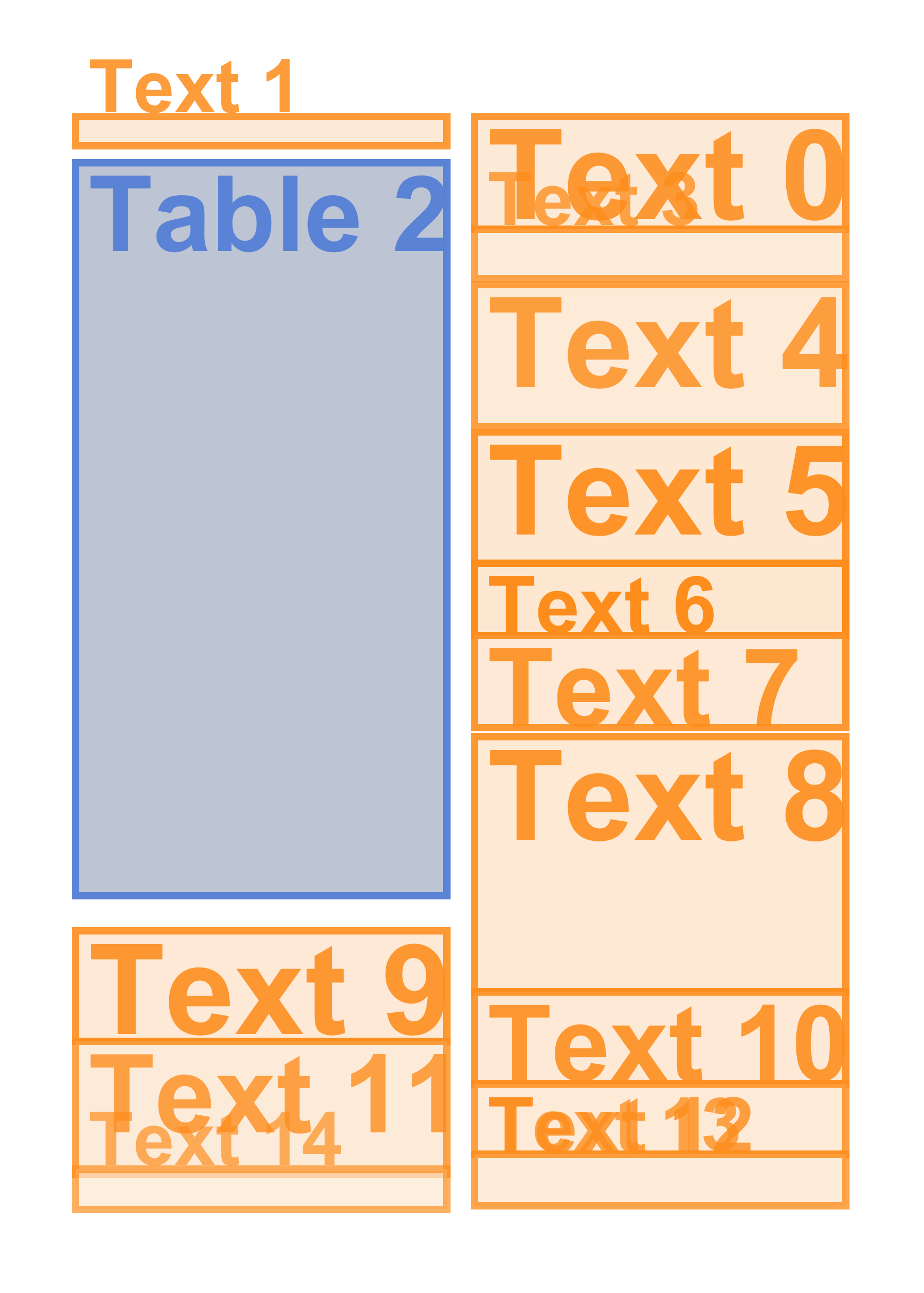} \\
    \end{tabular}
    \caption{Attention visualization. A higher color intensity reflects a higher attention weight. In the first layer, no attention is paid to any element. In subsequent layers, other elements are considered regardless of their distance in the sequence. This is particularly useful to model two-column documents: for example, in the attention map for Text 8, Table 2 has a significant weight, despite their distance in the input sequence. The network correctly identifies it as a relevant element to consider.
    }
    \label{fig:attention_vis_document}
\end{figure}

The encoder processes the entire document in a single pass. In the case of PubLayNet, where $n_\text{heads}=4$, we observe that in the first layer elements are independent of each other, since no element receives any attention. In subsequent layers, elements start to consider others in the computation. In fig. \ref{fig:attention_vis_document} we show a visualization of this process.

\clearpage
\subsection{Decoder}

\begin{figure}[h]
    \centering
    \setlength{\tabcolsep}{0.5pt}
    \newlength{\attentionDecoderVisDimmed}
    \setlength{\attentionDecoderVisDimmed}{0.0707\linewidth}
    \begin{tabular}{lcccccccccccccc}
&\multicolumn{13}{c}{Final decoding result}\\
&\multicolumn{13}{c}{\includegraphics[width=0.10\linewidth,frame=.1pt]{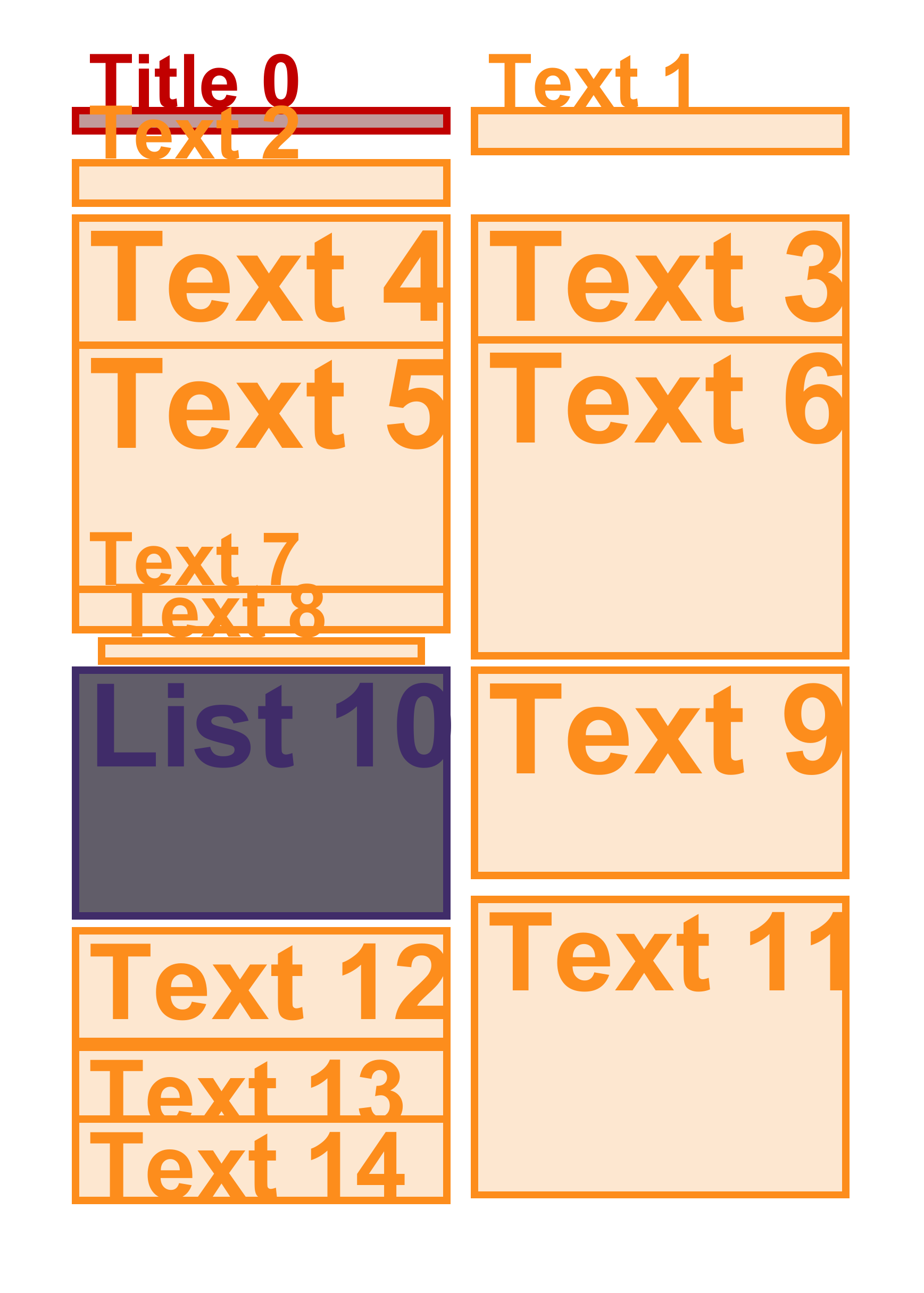}}\\ \toprule
&\multicolumn{13}{c}{Attention}\\ \toprule
& 
\footnotesize Title 1 &
\footnotesize Text 2 &
\footnotesize Text 3 &
\footnotesize Text 4 &
\footnotesize Text 5 &
\footnotesize Text 6 &
\footnotesize Text 7 &
\footnotesize Text 8 &
\footnotesize Text 9 &
\footnotesize List 10 &
\footnotesize Text 11 &
\footnotesize Text 12 &
\footnotesize Text 13 &
\footnotesize Text 14 \\
\rotatebox{90}{\hspace{0.3cm}\footnotesize Layer 1}  &  
\includegraphics[width=\attentionDecoderVisDimmed,frame=0.1pt]{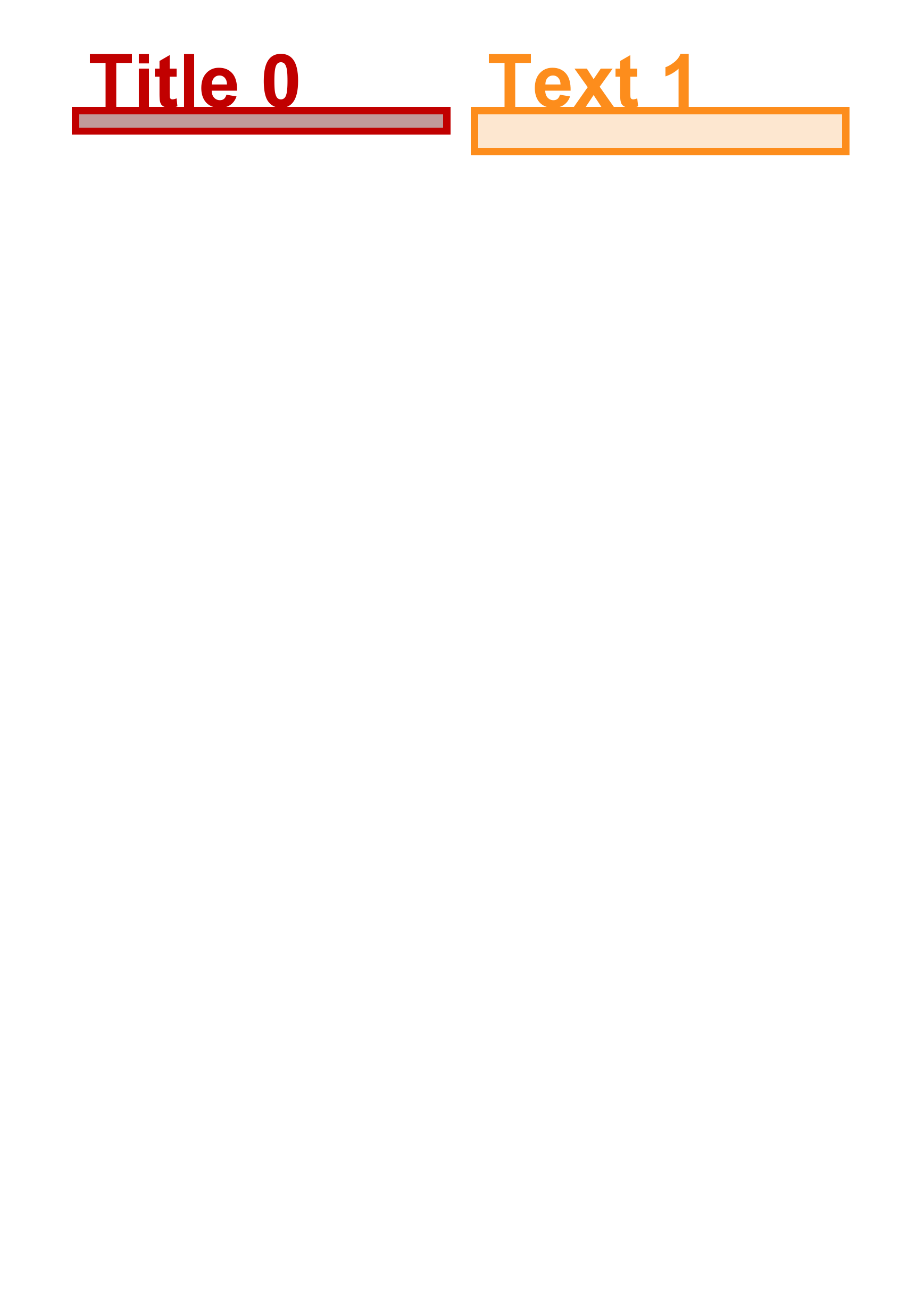} &
\includegraphics[width=\attentionDecoderVisDimmed,frame=0.1pt]{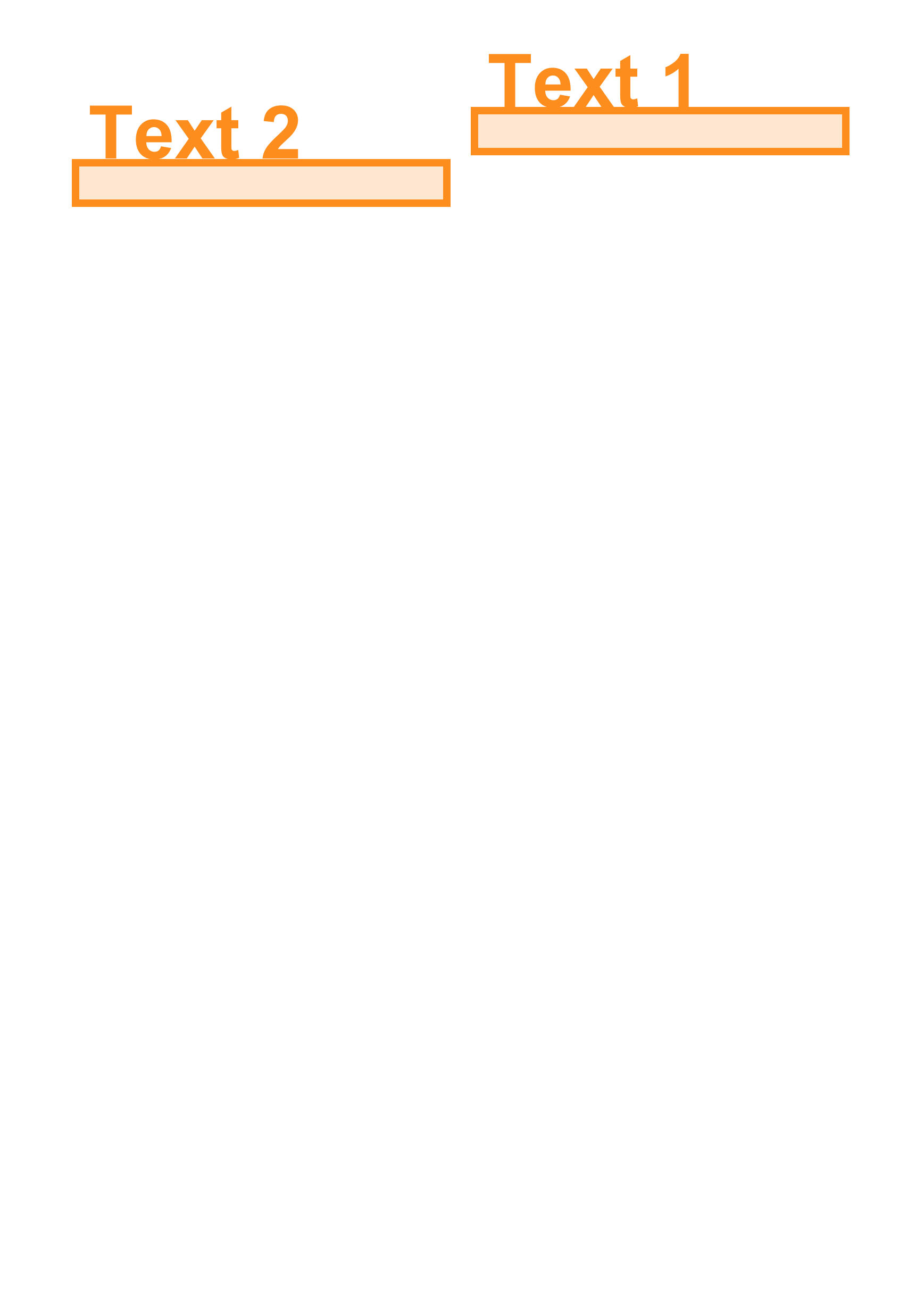} &
\includegraphics[width=\attentionDecoderVisDimmed,frame=0.1pt]{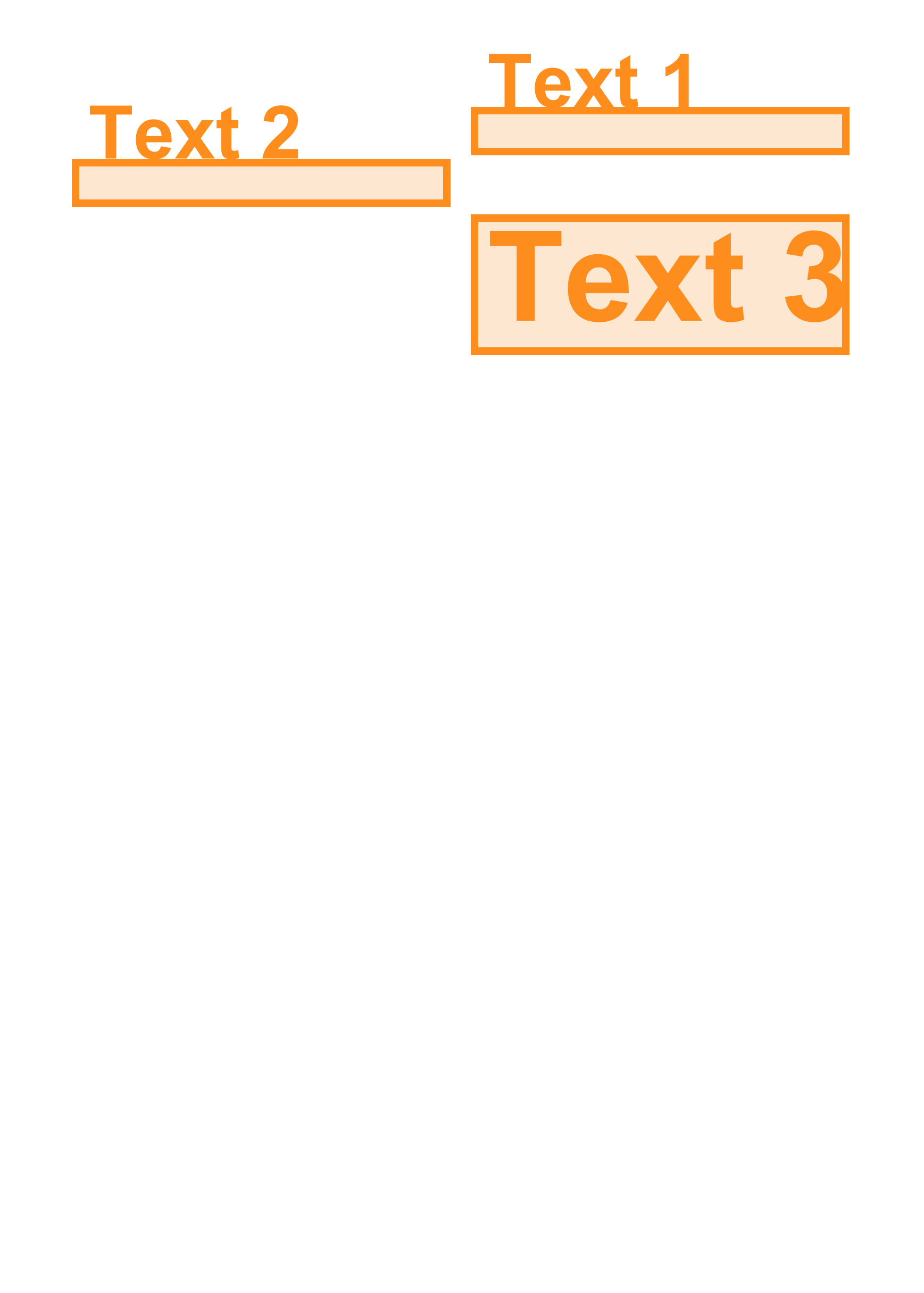} &
\includegraphics[width=\attentionDecoderVisDimmed,frame=0.1pt]{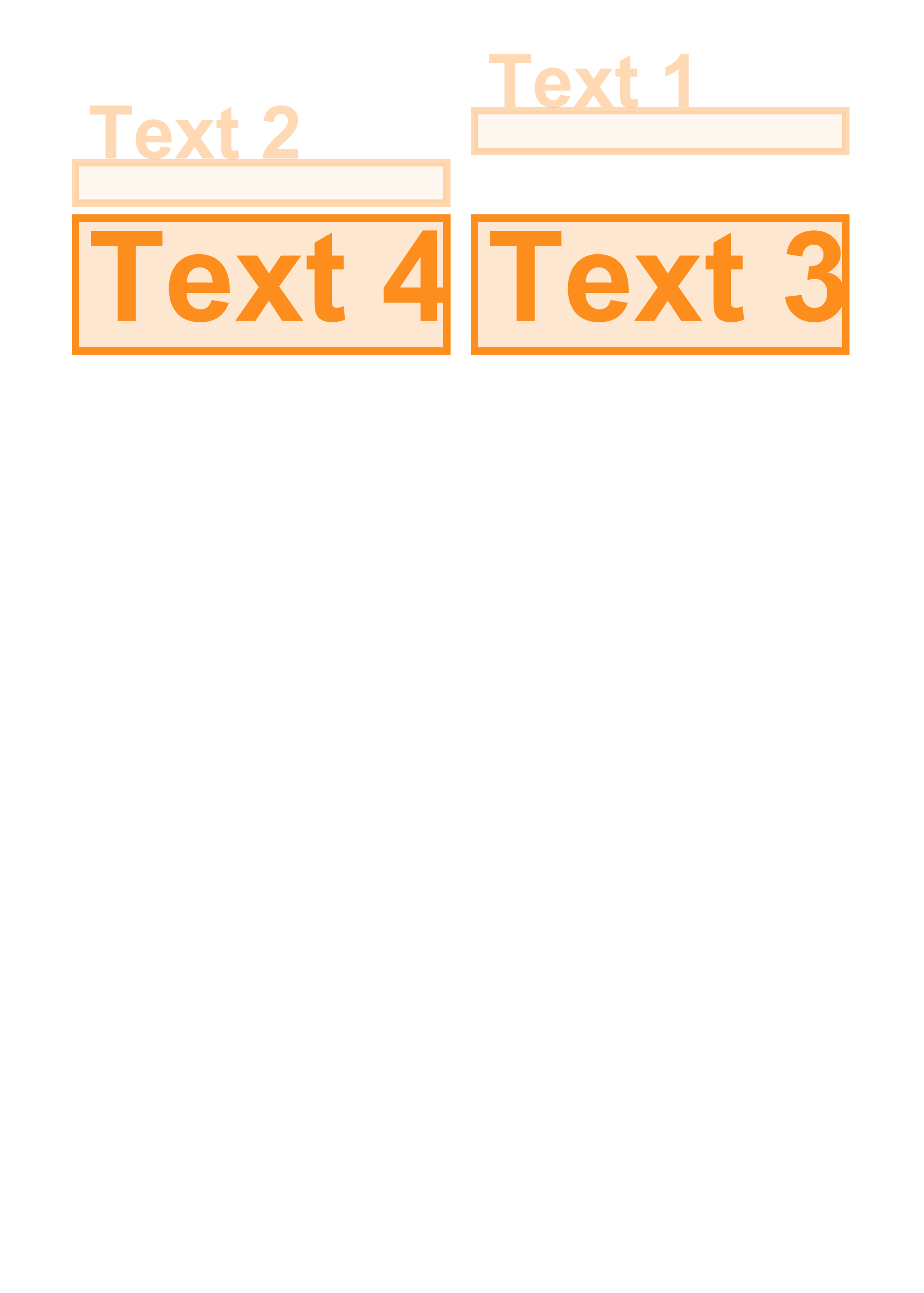} &
\includegraphics[width=\attentionDecoderVisDimmed,frame=0.1pt]{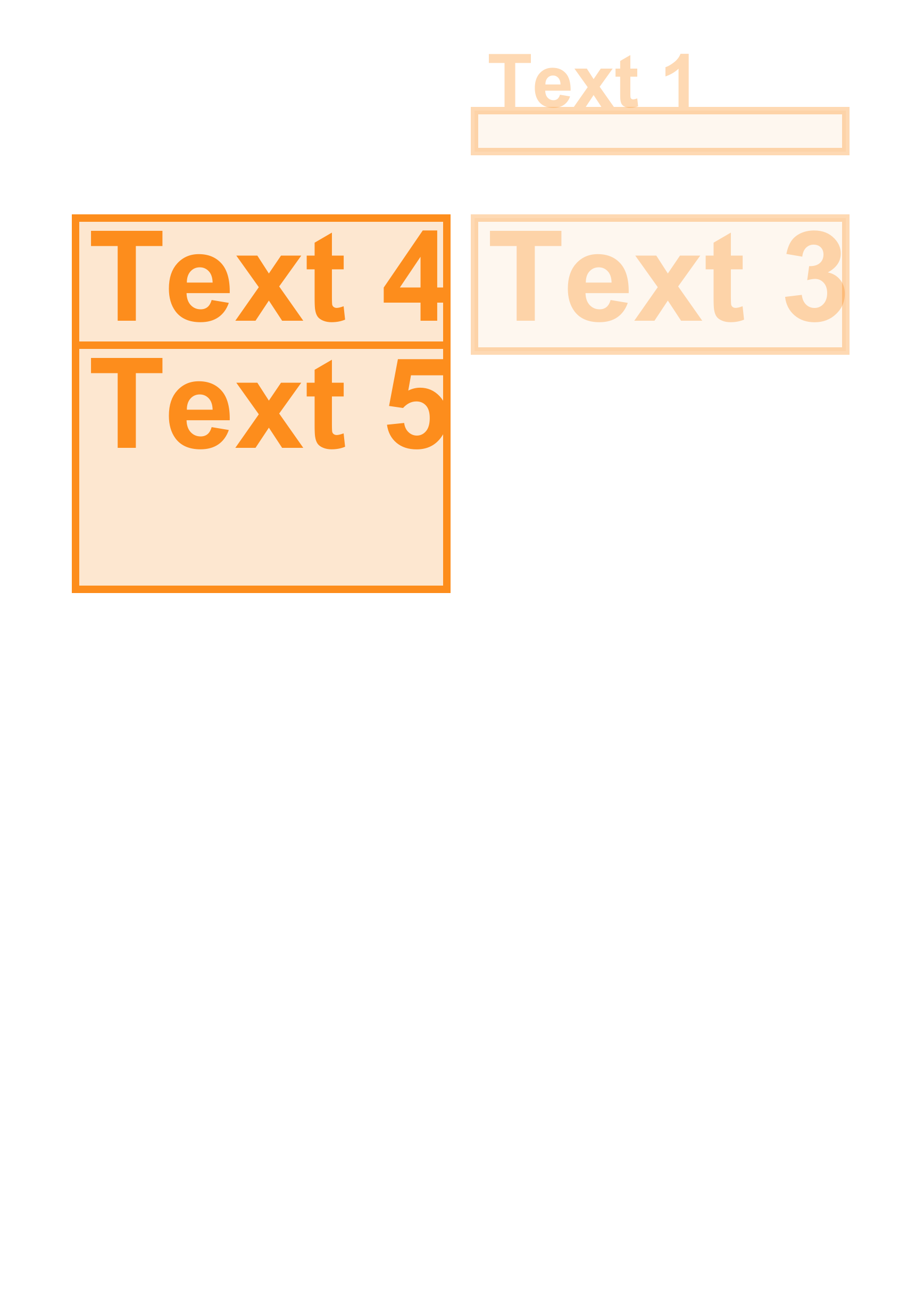} &
\includegraphics[width=\attentionDecoderVisDimmed,frame=0.1pt]{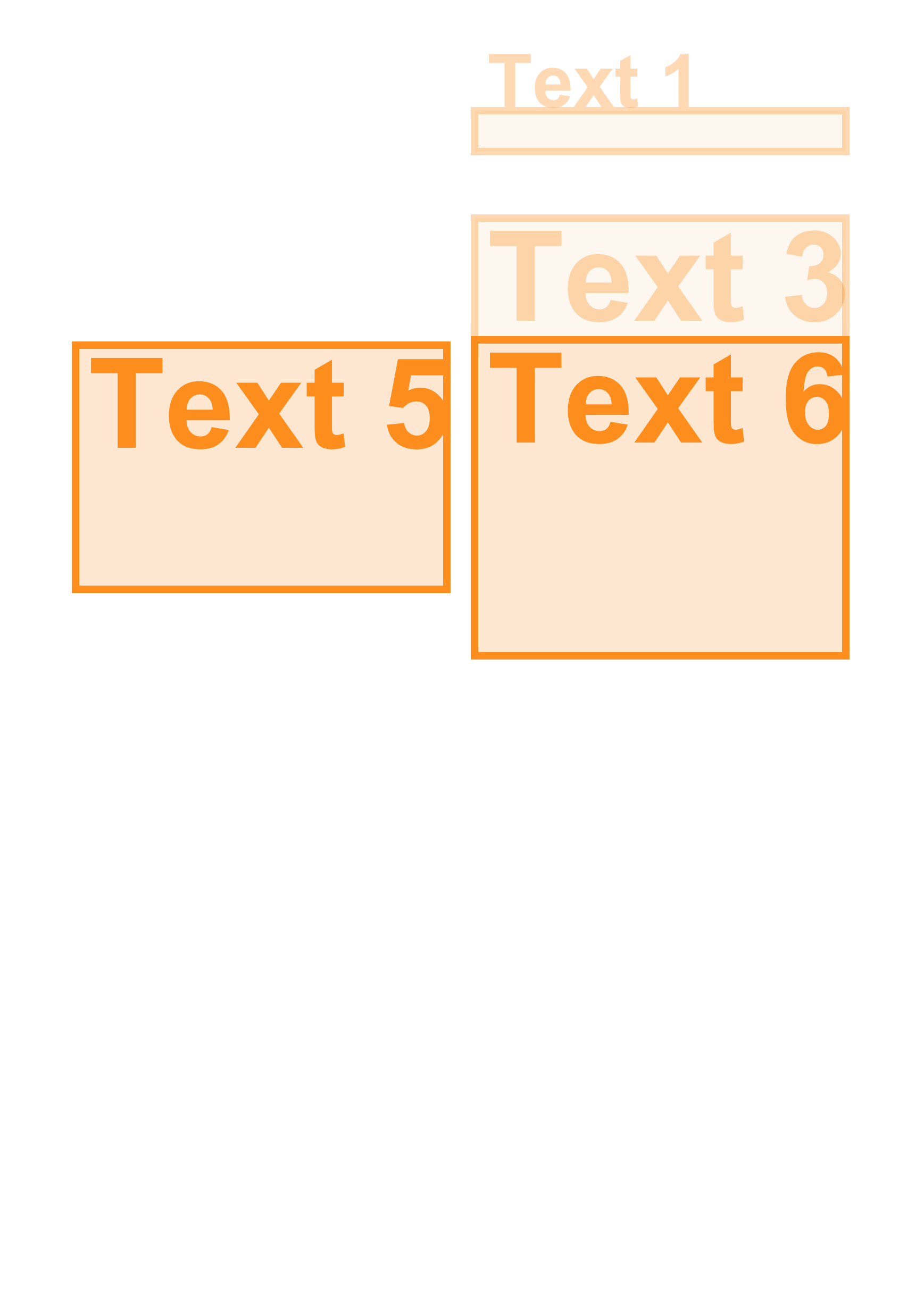} &
\includegraphics[width=\attentionDecoderVisDimmed,frame=0.1pt]{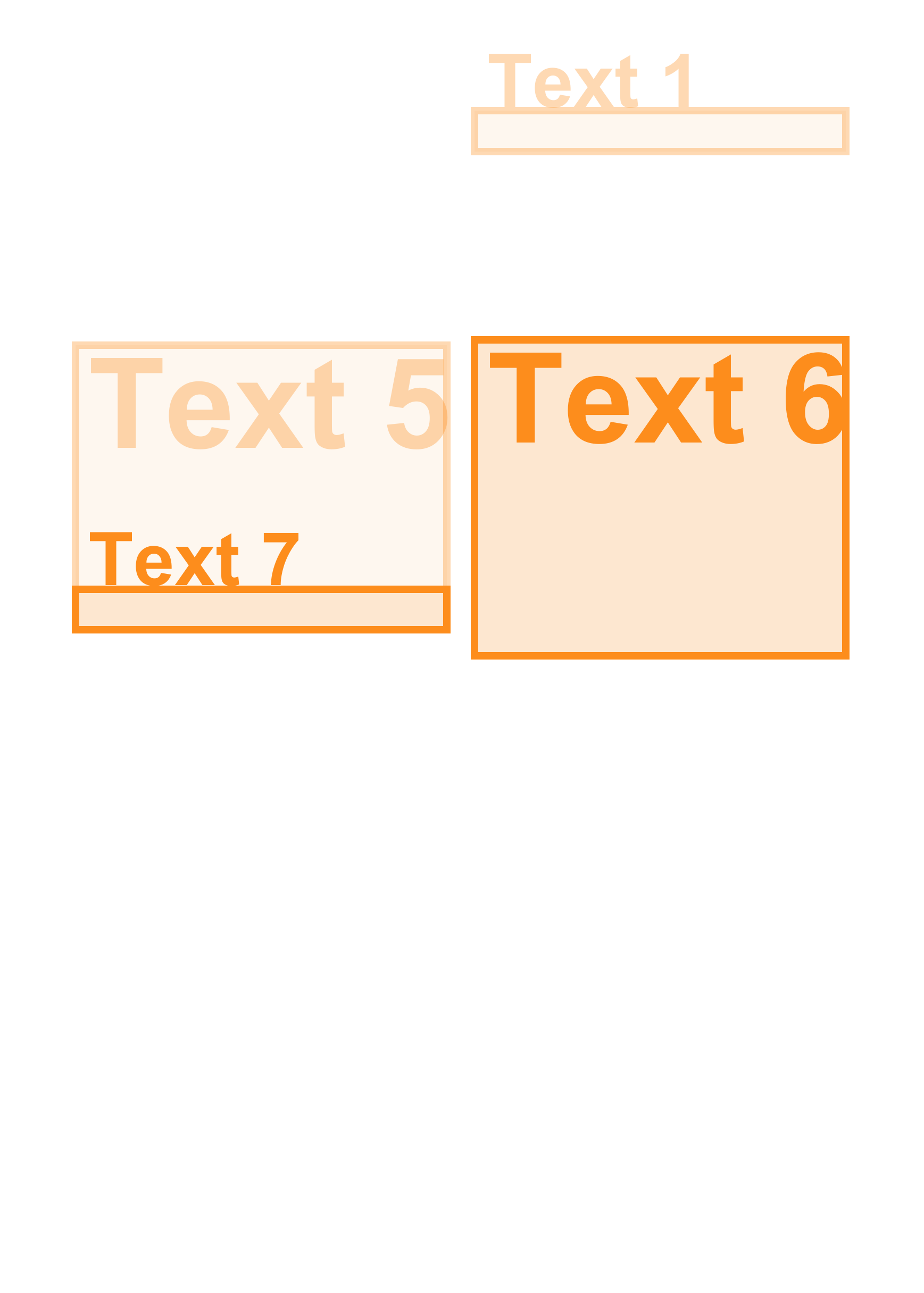} &
\includegraphics[width=\attentionDecoderVisDimmed,frame=0.1pt]{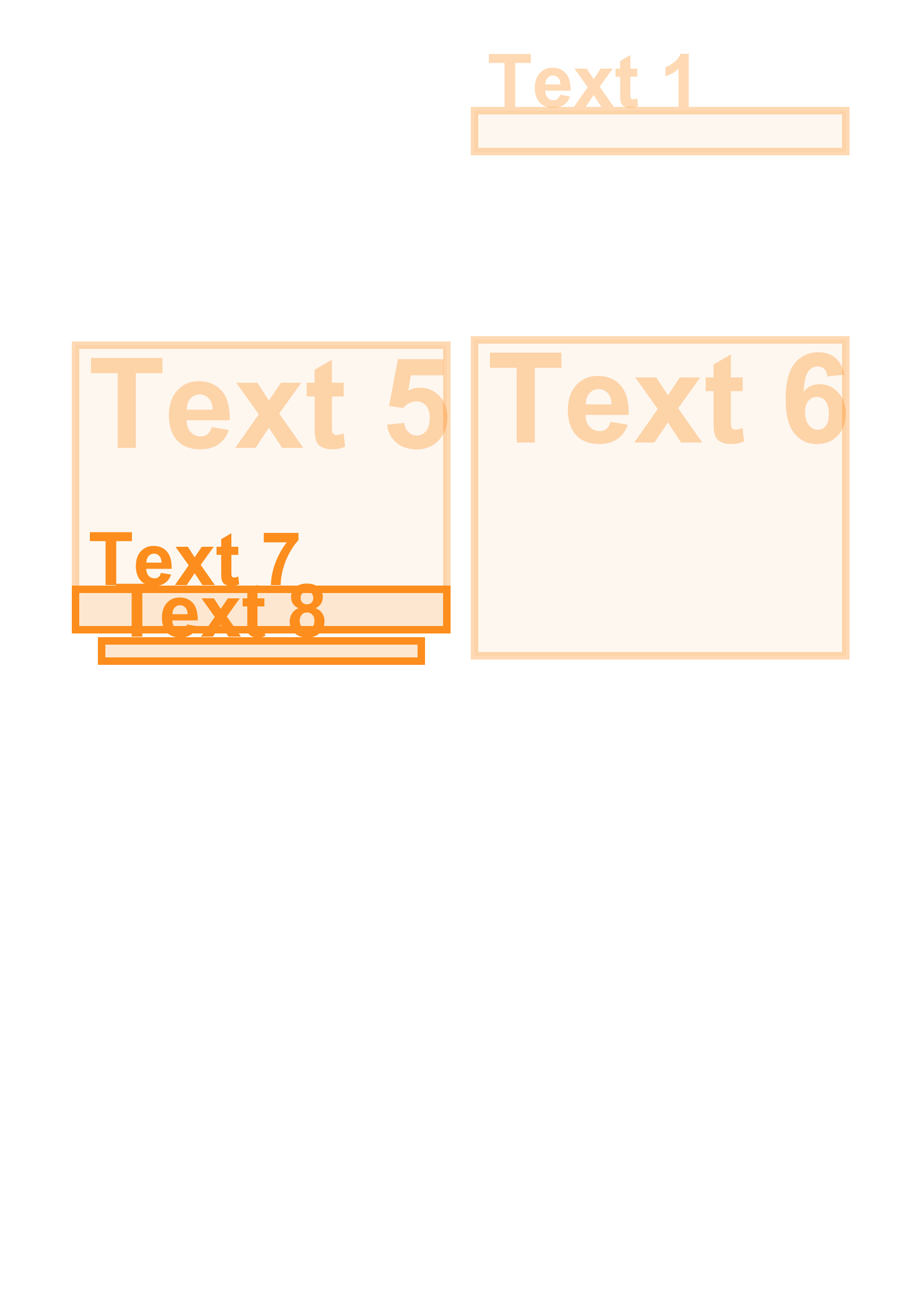} &
\includegraphics[width=\attentionDecoderVisDimmed,frame=0.1pt]{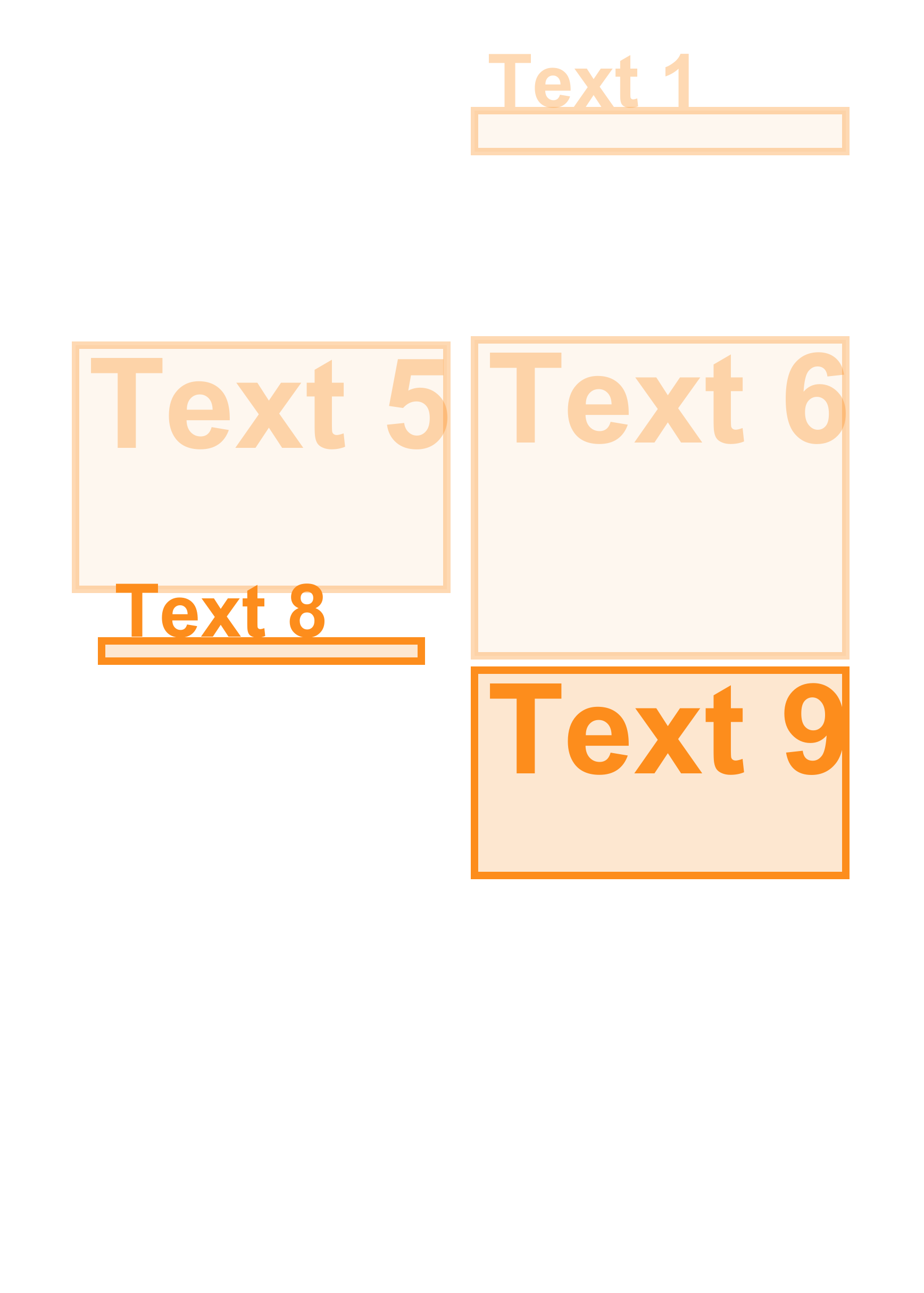} &
\includegraphics[width=\attentionDecoderVisDimmed,frame=0.1pt]{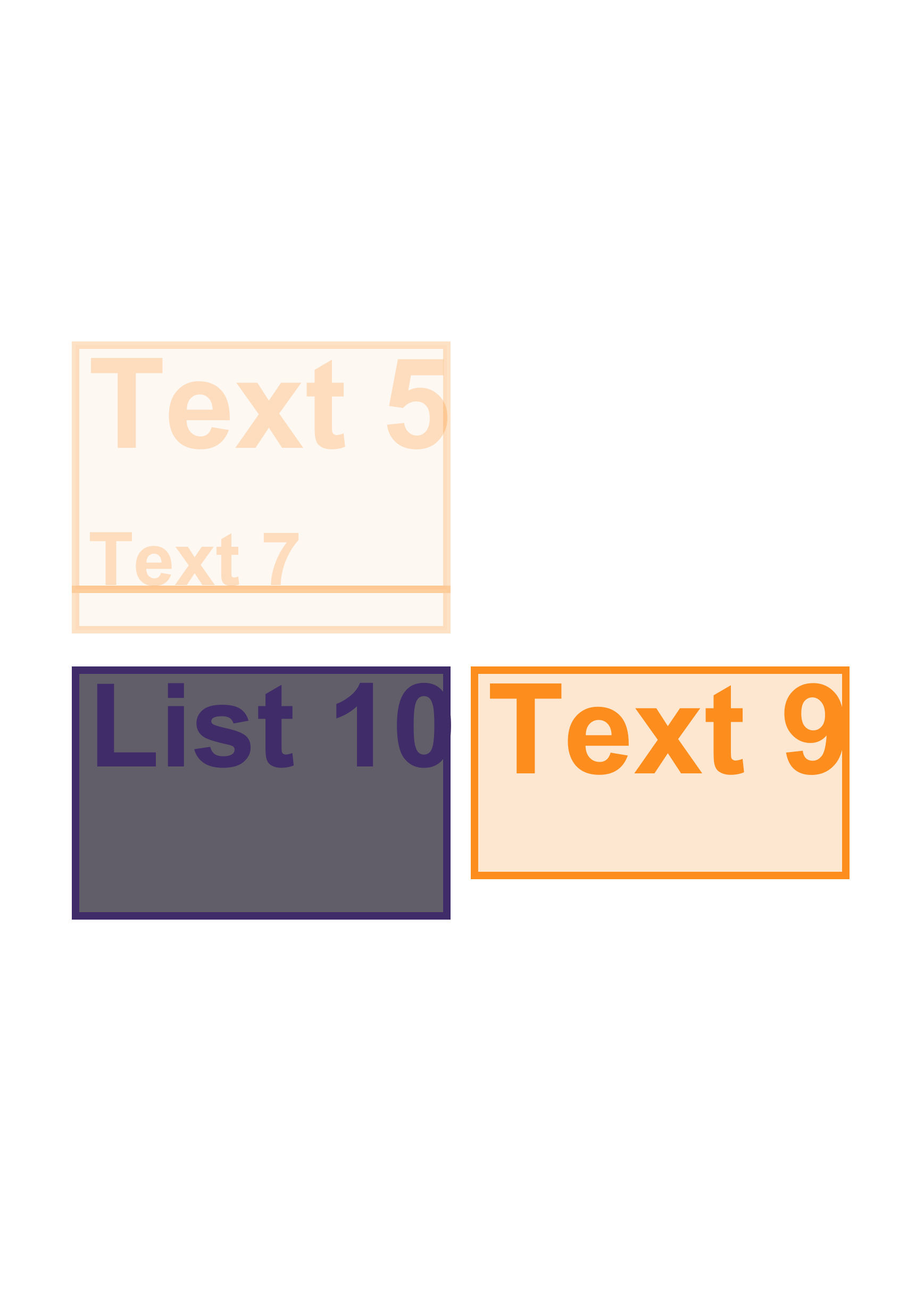} &
\includegraphics[width=\attentionDecoderVisDimmed,frame=0.1pt]{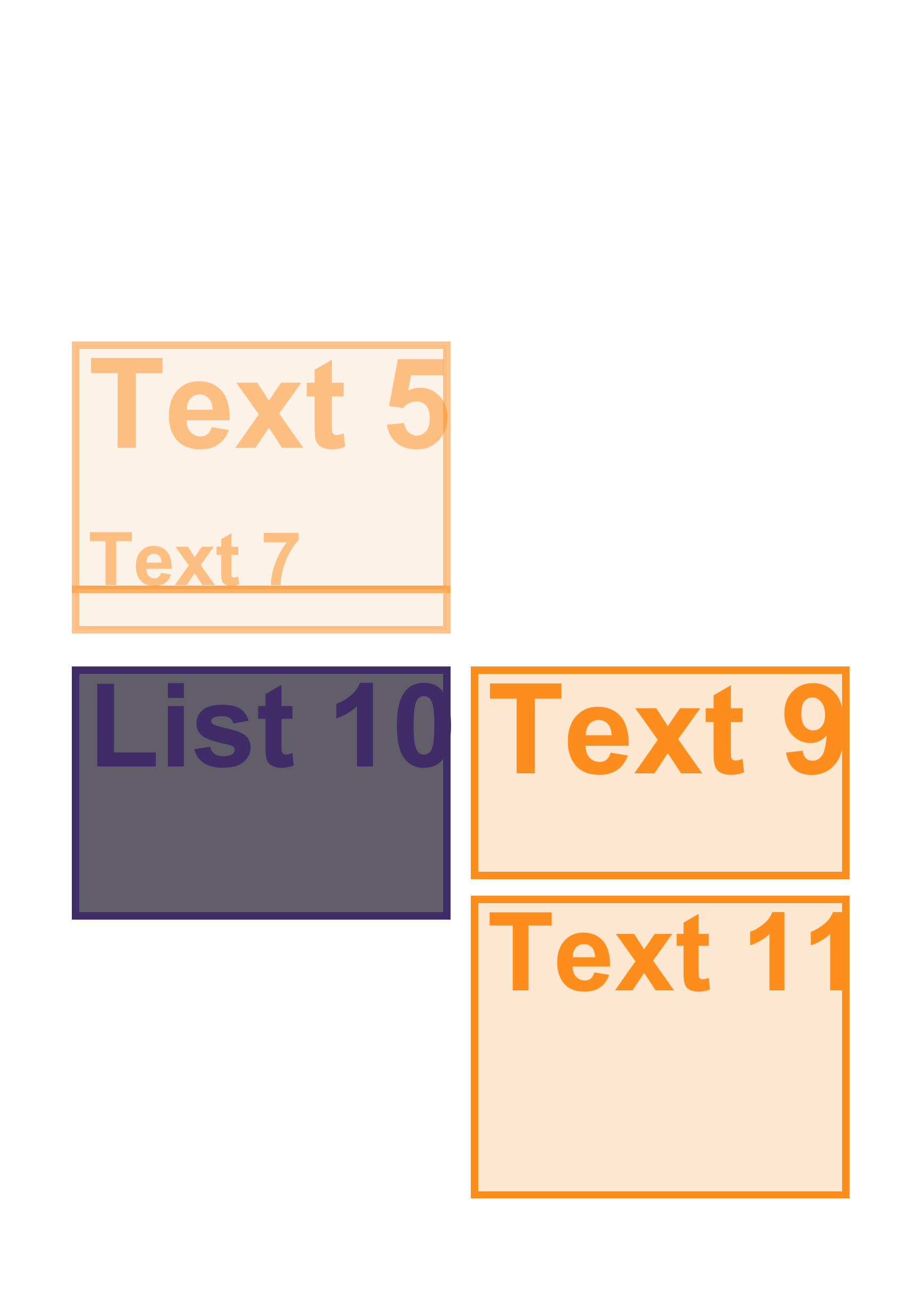} &
\includegraphics[width=\attentionDecoderVisDimmed,frame=0.1pt]{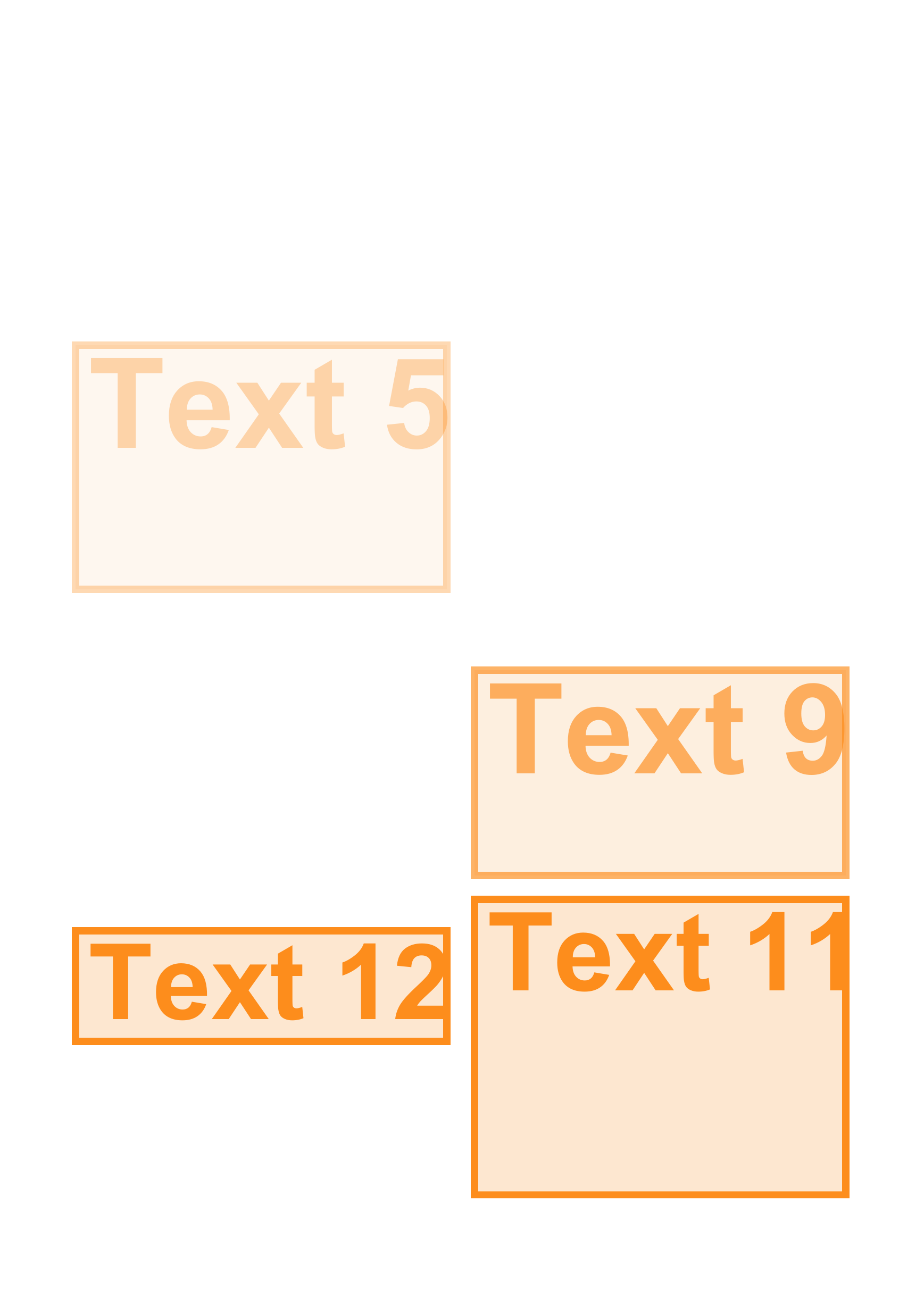} &
\includegraphics[width=\attentionDecoderVisDimmed,frame=0.1pt]{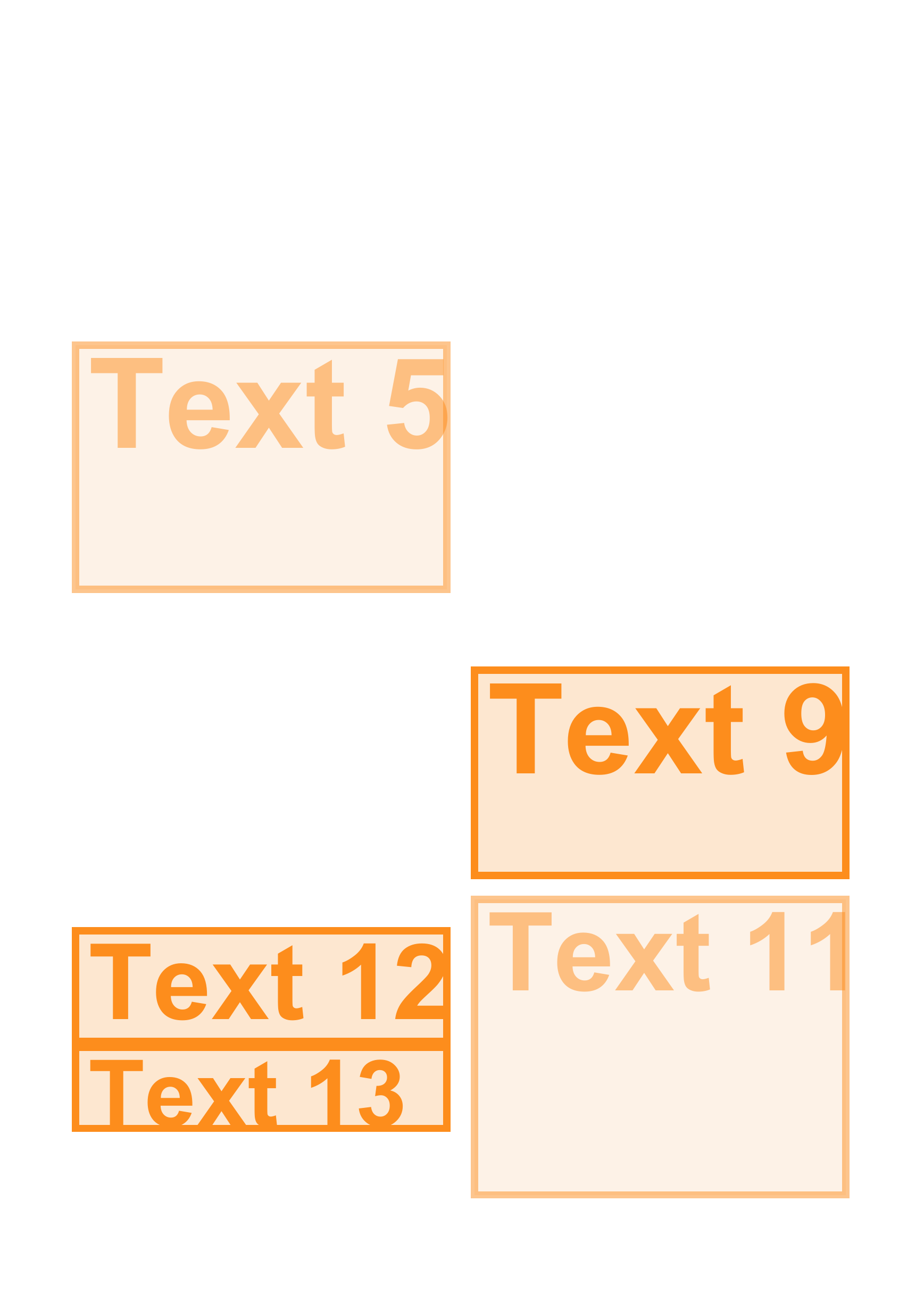} &
\includegraphics[width=\attentionDecoderVisDimmed,frame=0.1pt]{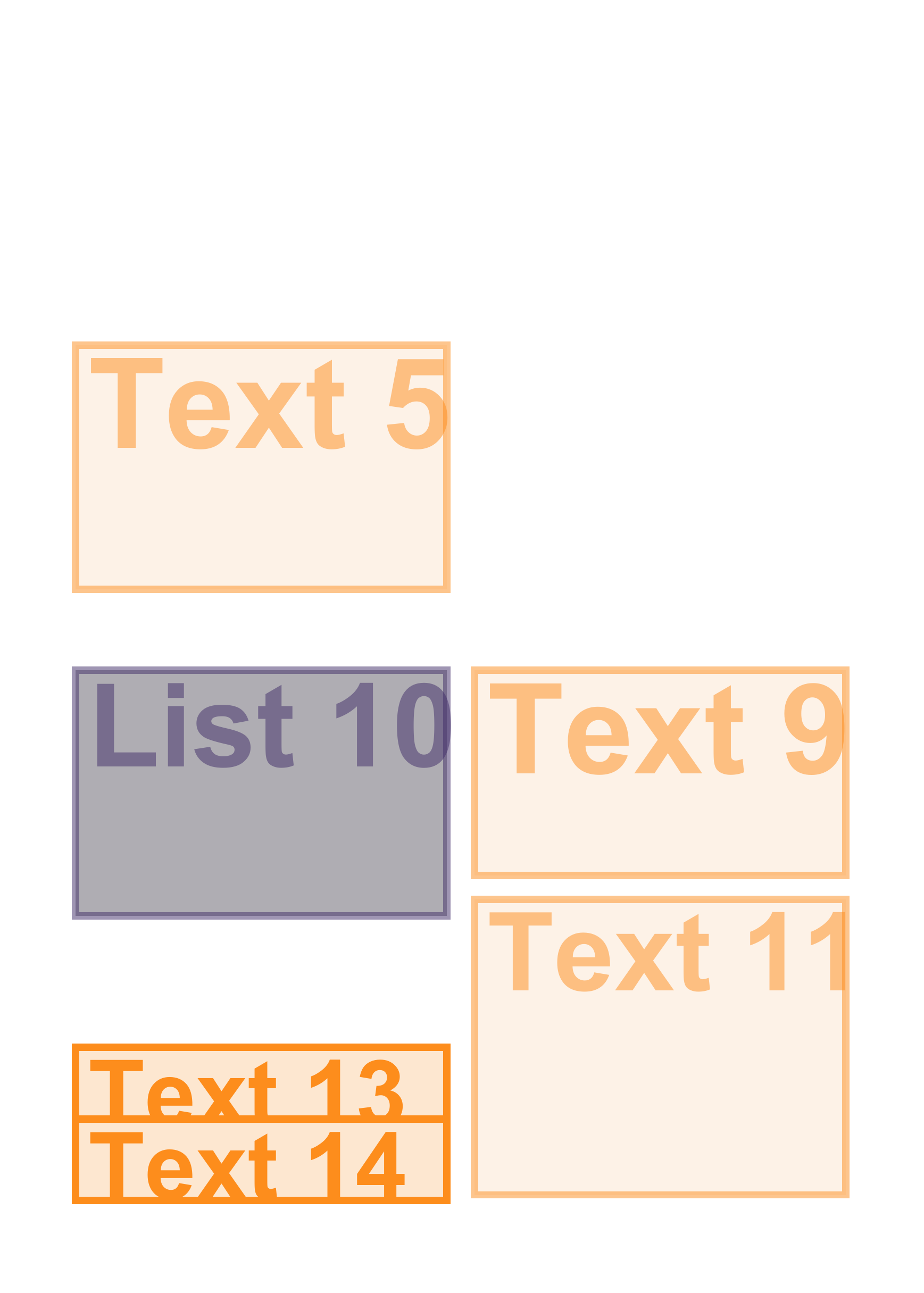} \\

\rotatebox{90}{\hspace{0.3cm}\footnotesize Layer 2}  &  
\includegraphics[width=\attentionDecoderVisDimmed,frame=0.1pt]{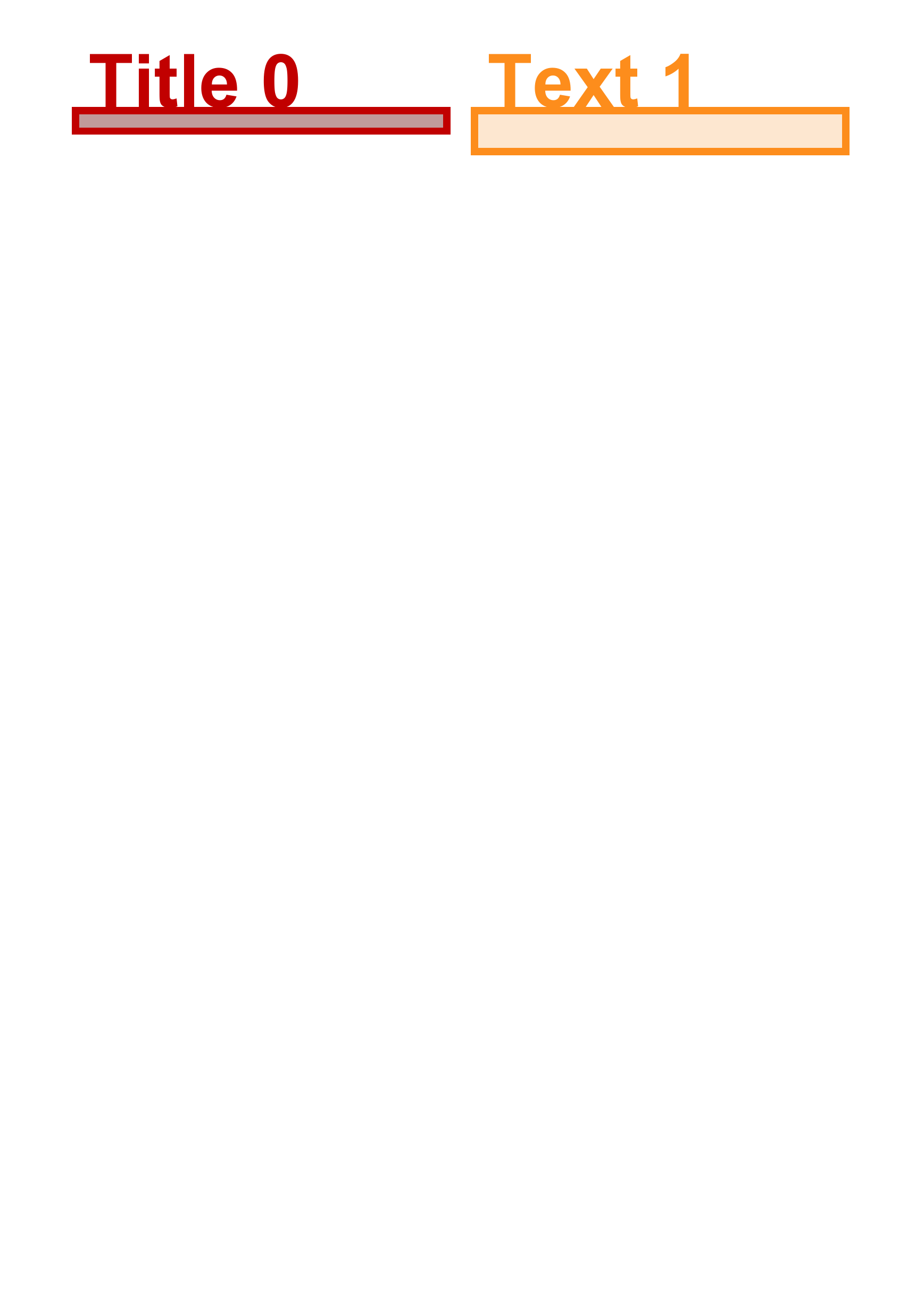} &
\includegraphics[width=\attentionDecoderVisDimmed,frame=0.1pt]{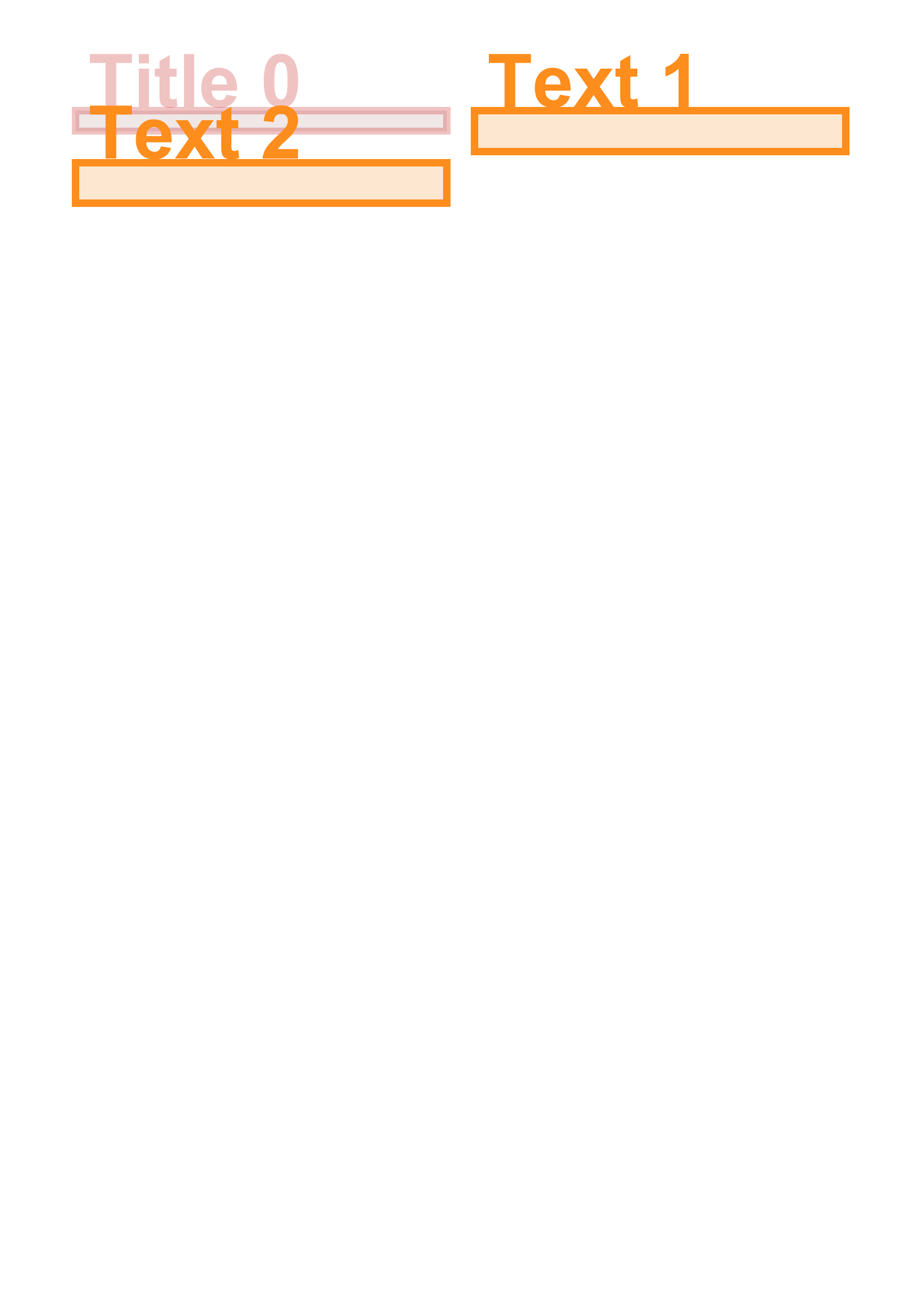} &
\includegraphics[width=\attentionDecoderVisDimmed,frame=0.1pt]{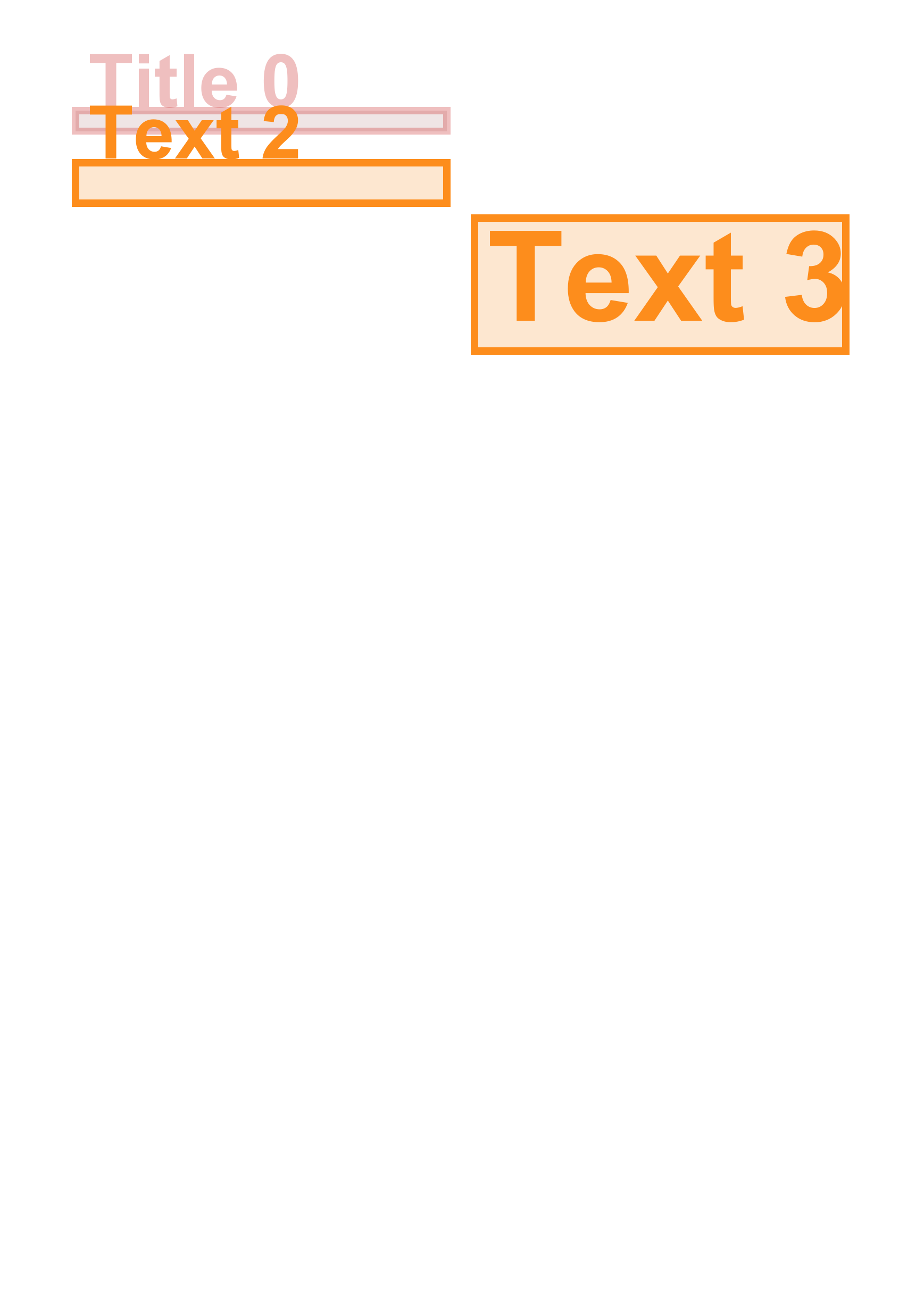} &
\includegraphics[width=\attentionDecoderVisDimmed,frame=0.1pt]{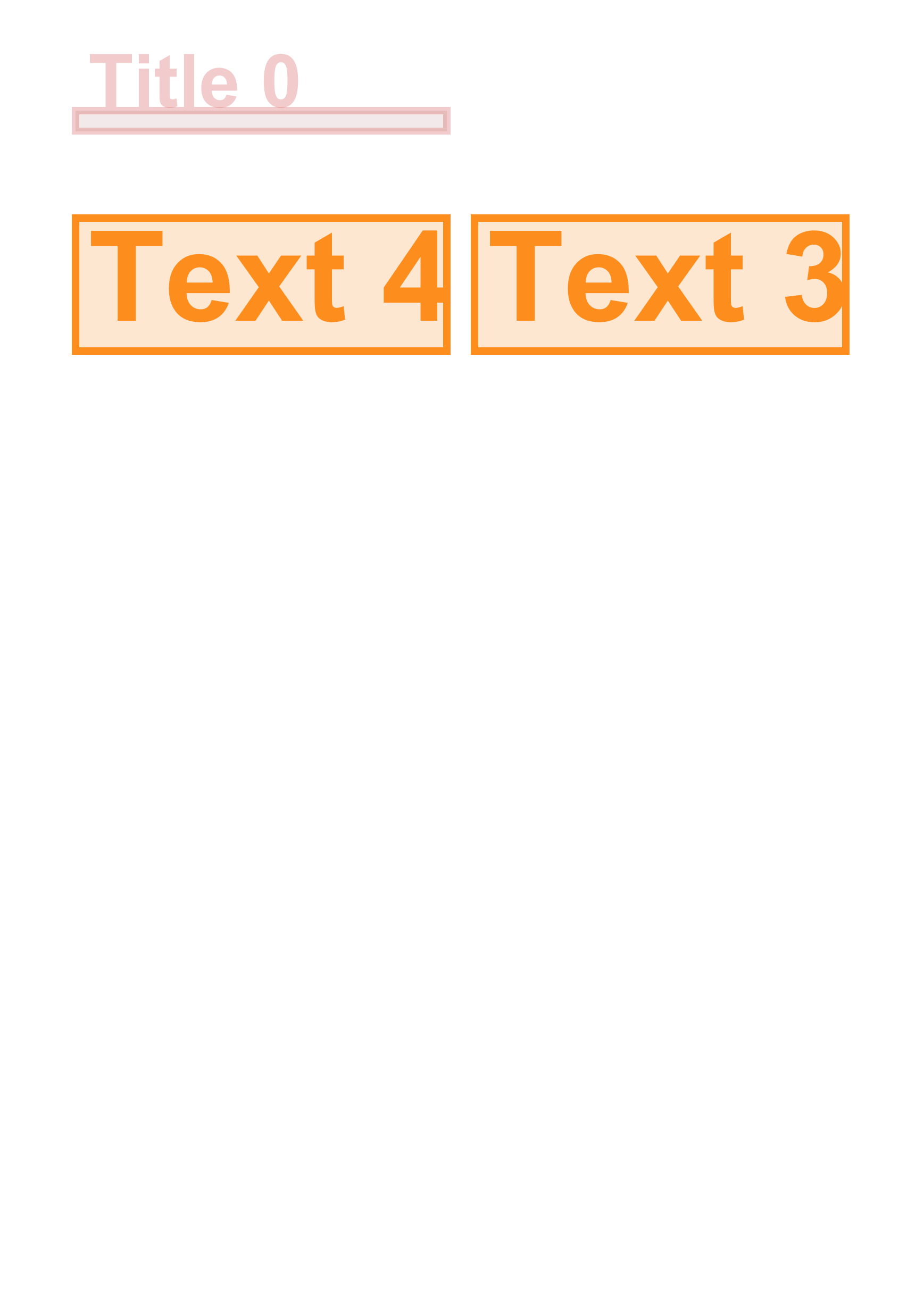} &
\includegraphics[width=\attentionDecoderVisDimmed,frame=0.1pt]{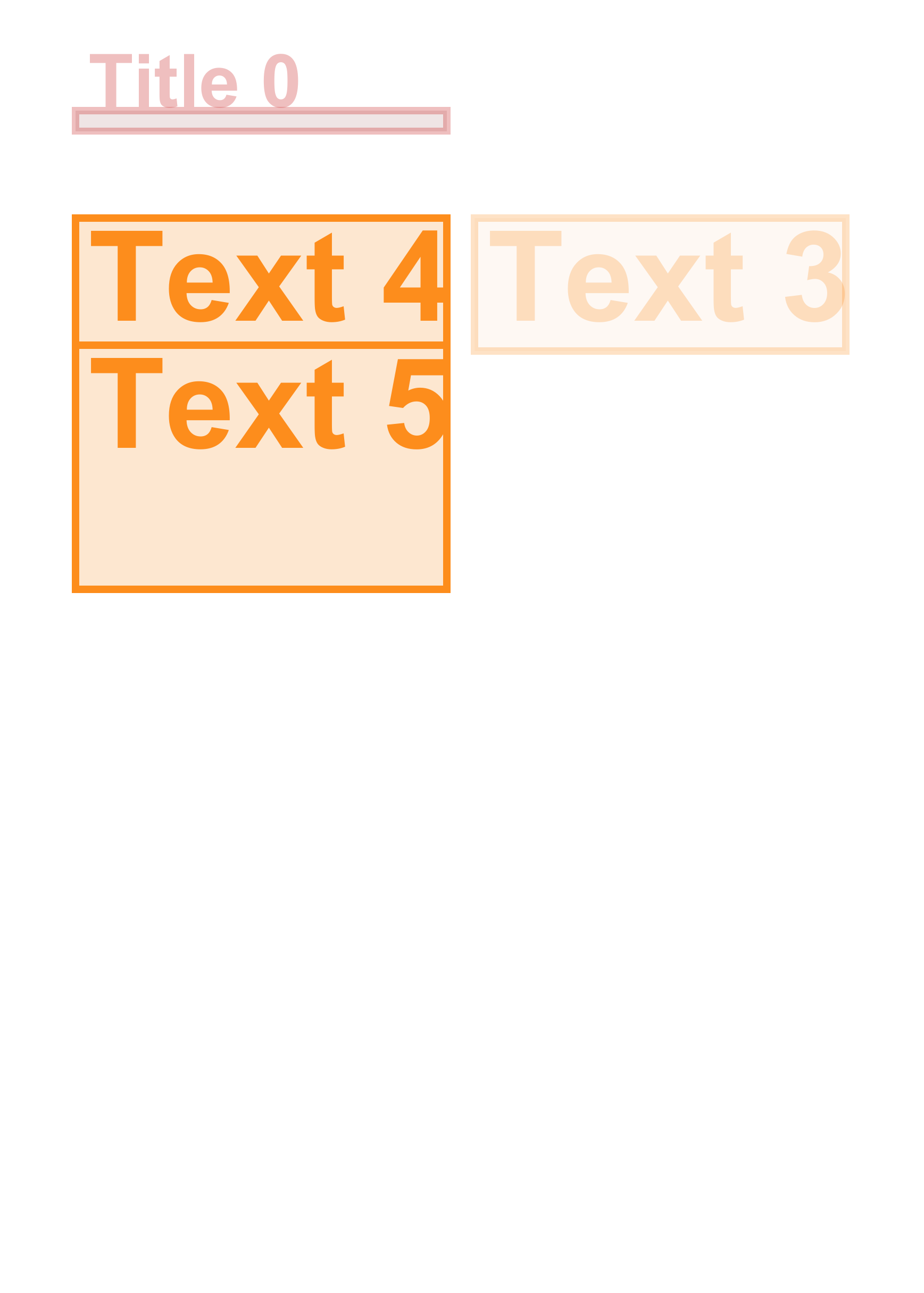} &
\includegraphics[width=\attentionDecoderVisDimmed,frame=0.1pt]{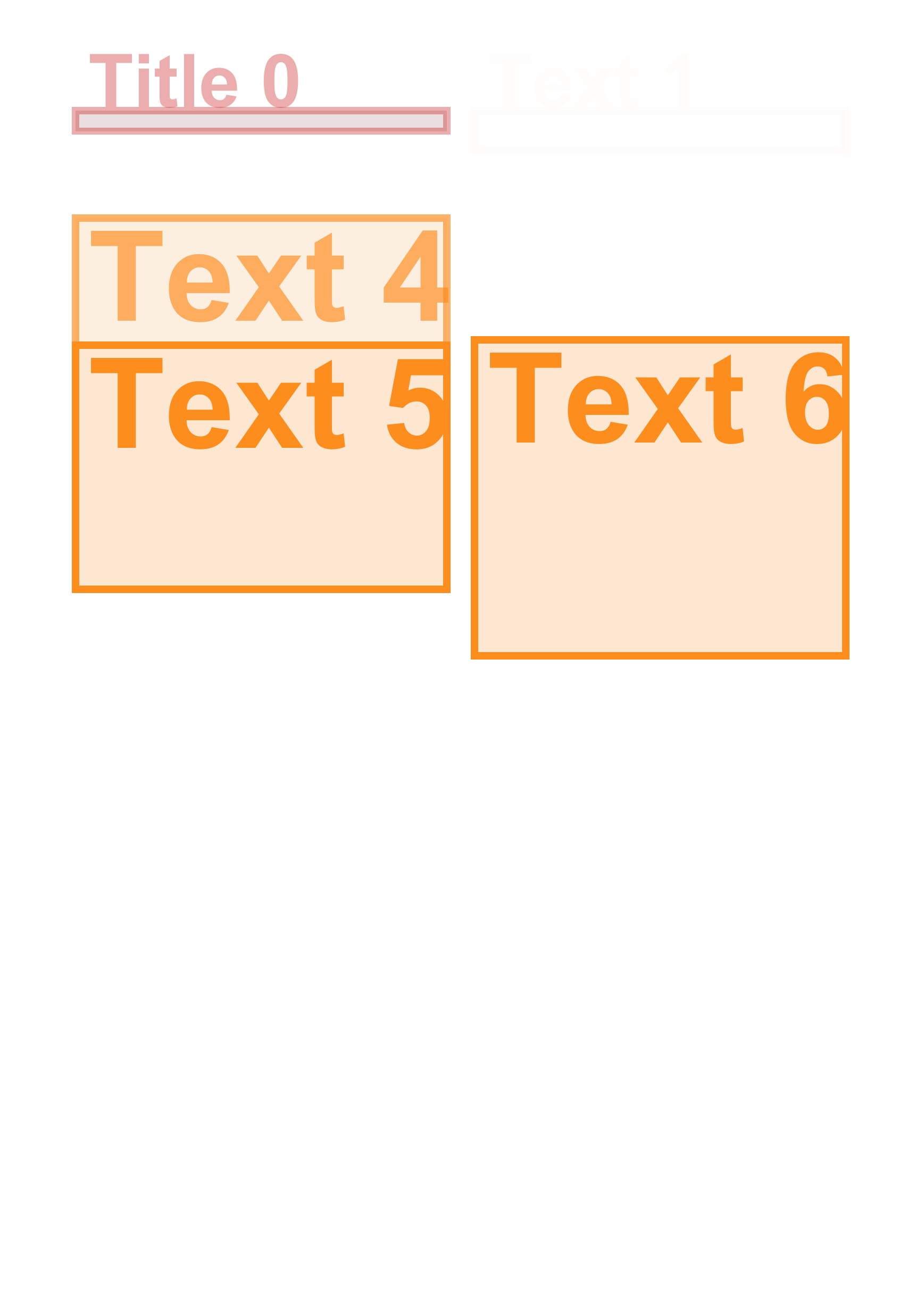} &
\includegraphics[width=\attentionDecoderVisDimmed,frame=0.1pt]{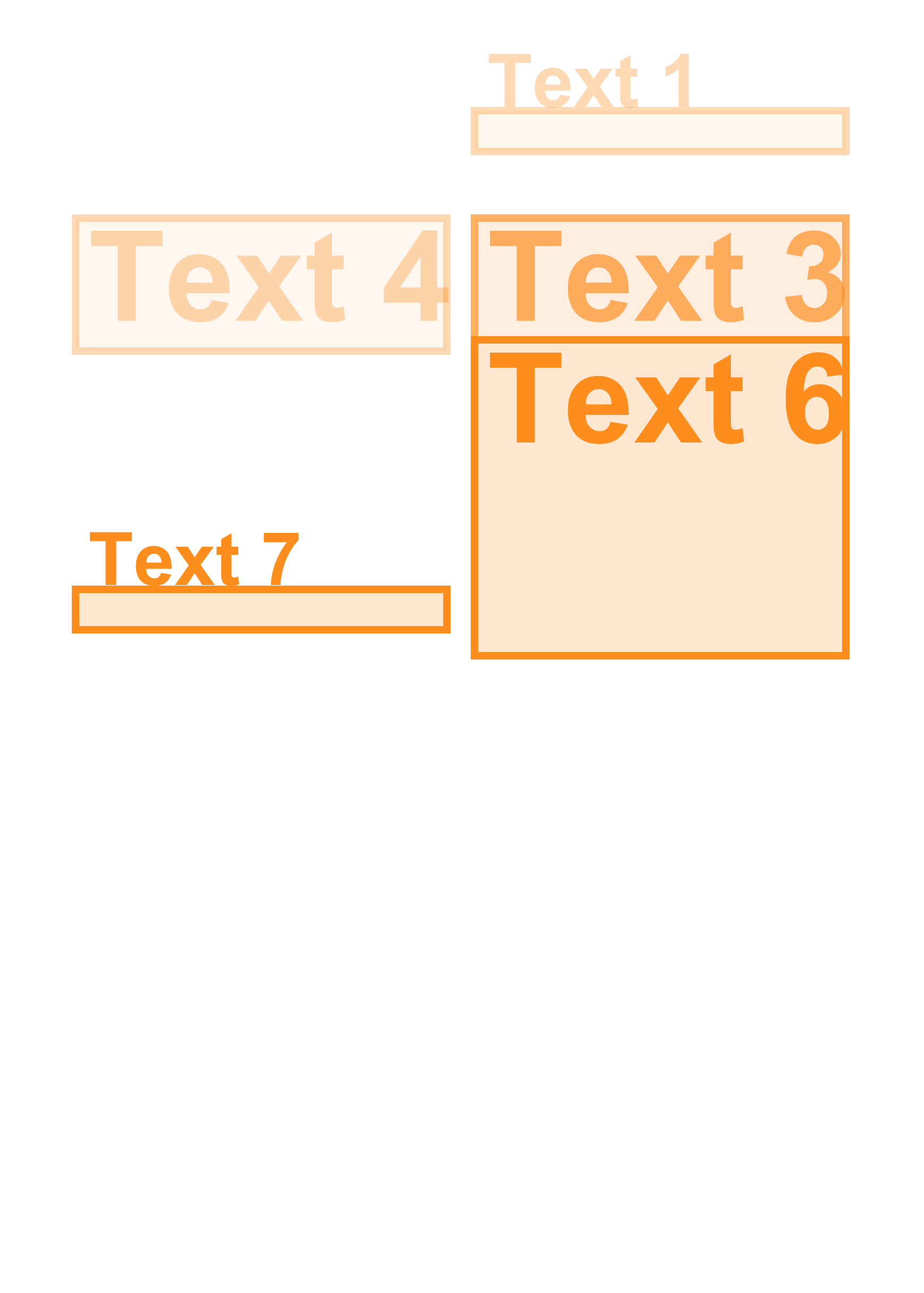} &
\includegraphics[width=\attentionDecoderVisDimmed,frame=0.1pt]{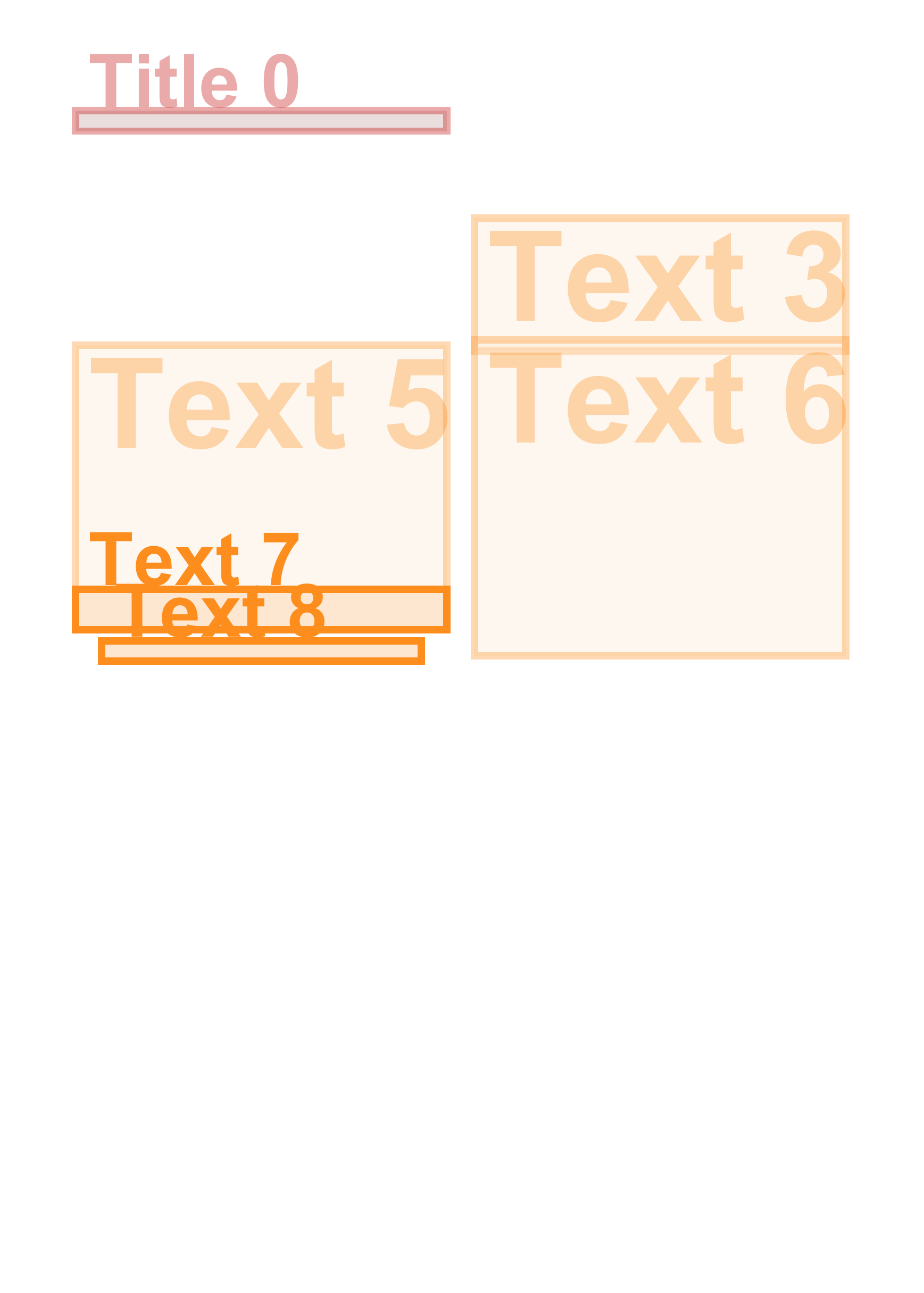} &
\includegraphics[width=\attentionDecoderVisDimmed,frame=0.1pt]{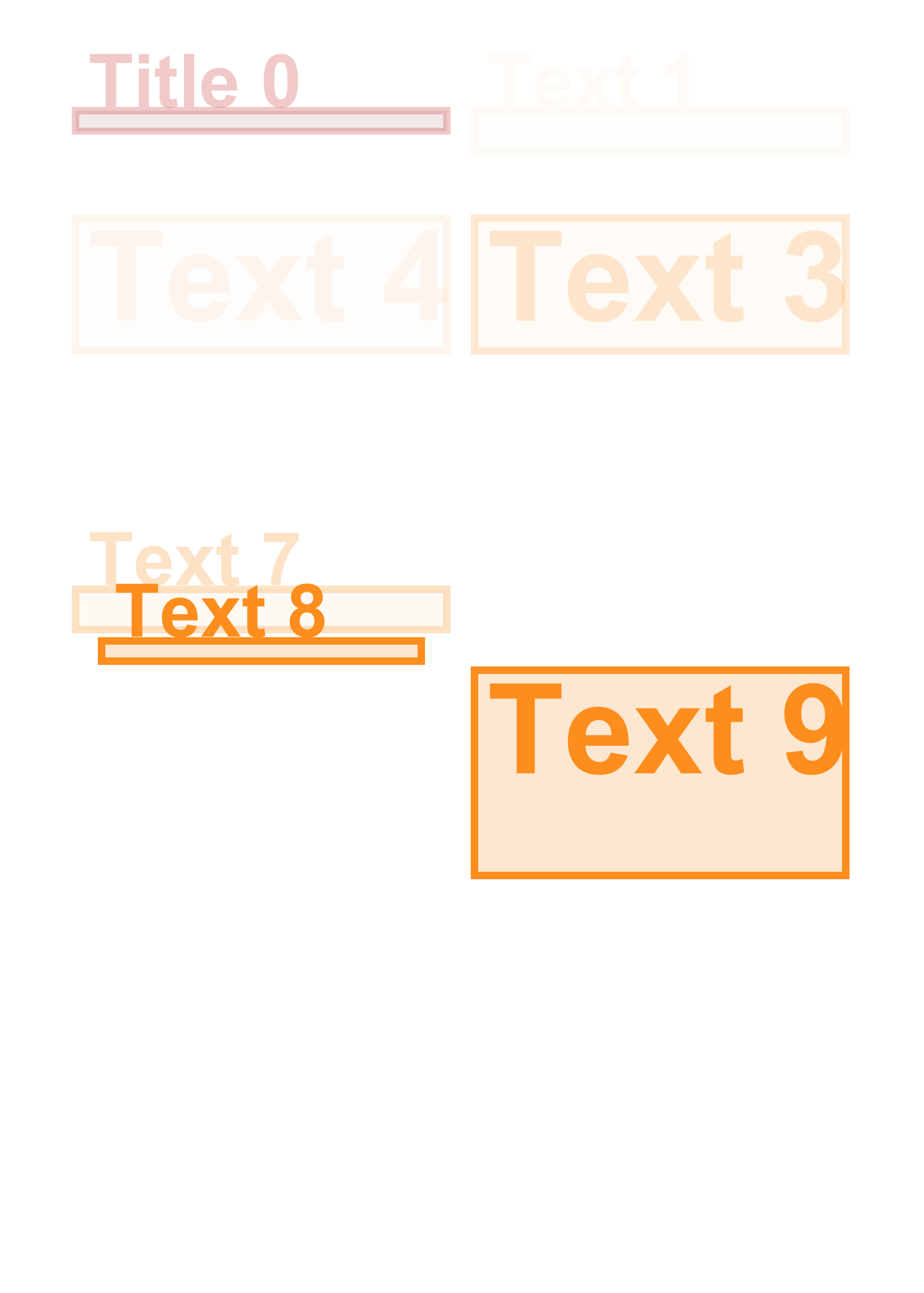} &
\includegraphics[width=\attentionDecoderVisDimmed,frame=0.1pt]{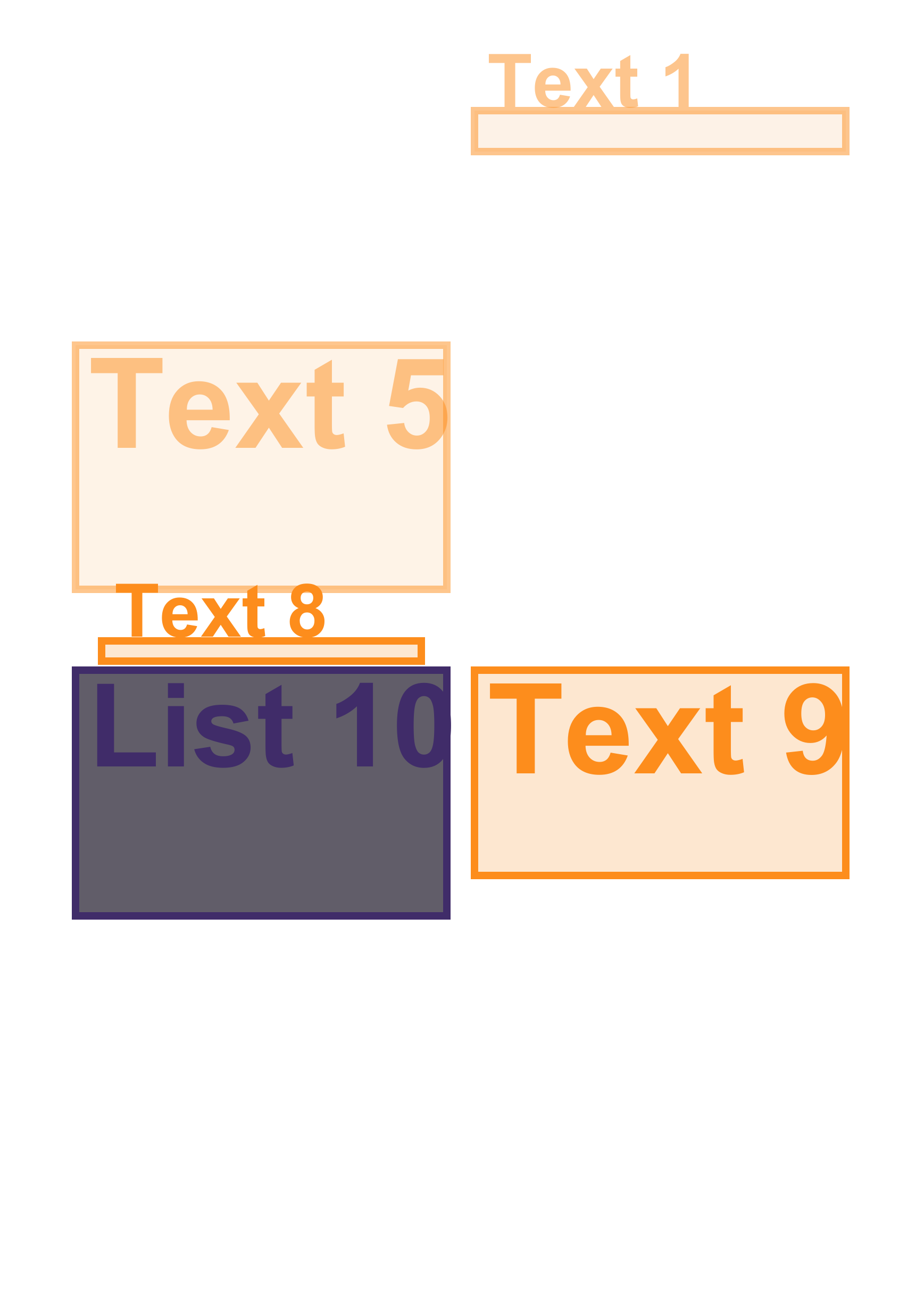} &
\includegraphics[width=\attentionDecoderVisDimmed,frame=0.1pt]{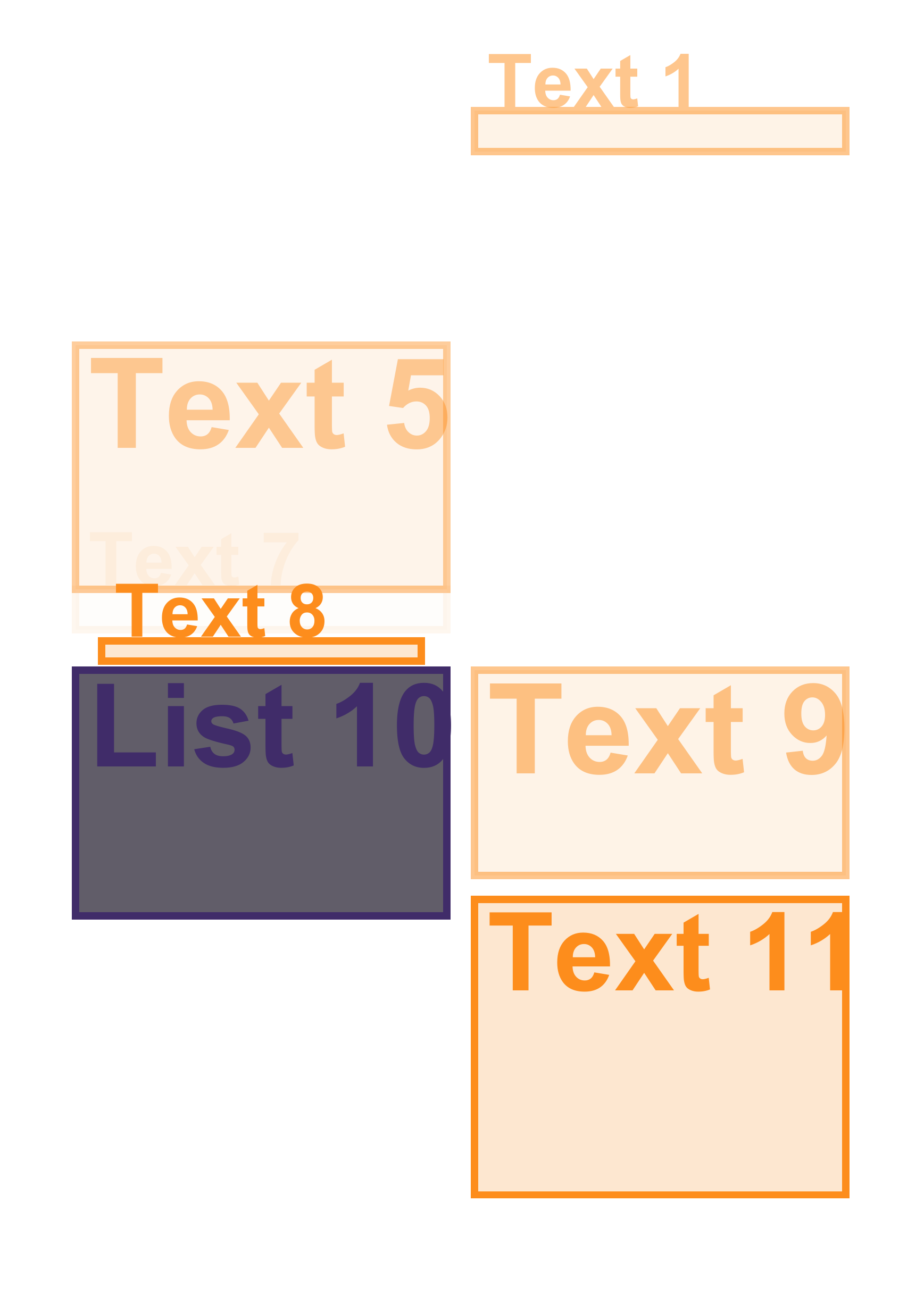} &
\includegraphics[width=\attentionDecoderVisDimmed,frame=0.1pt]{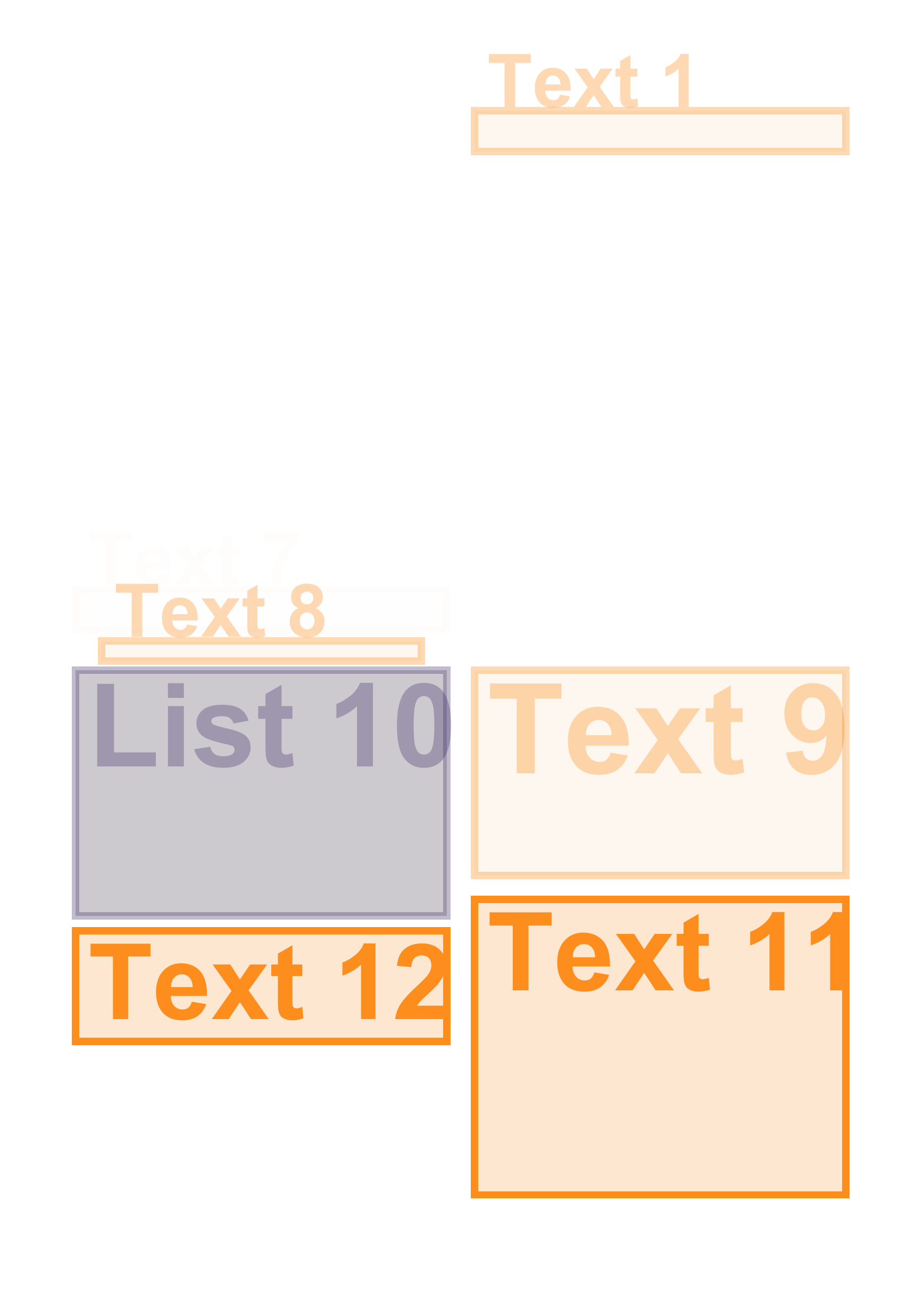} &
\includegraphics[width=\attentionDecoderVisDimmed,frame=0.1pt]{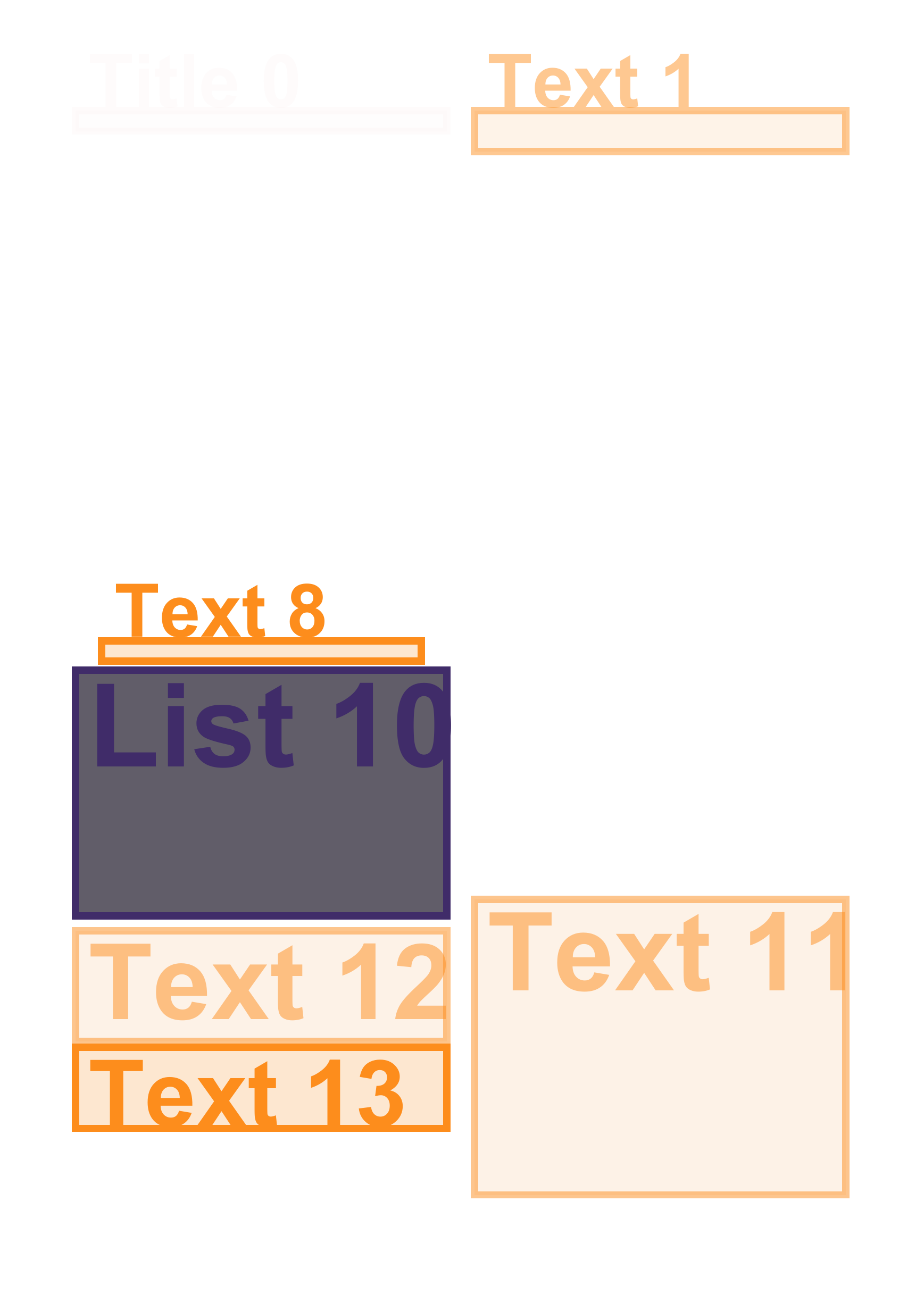} &
\includegraphics[width=\attentionDecoderVisDimmed,frame=0.1pt]{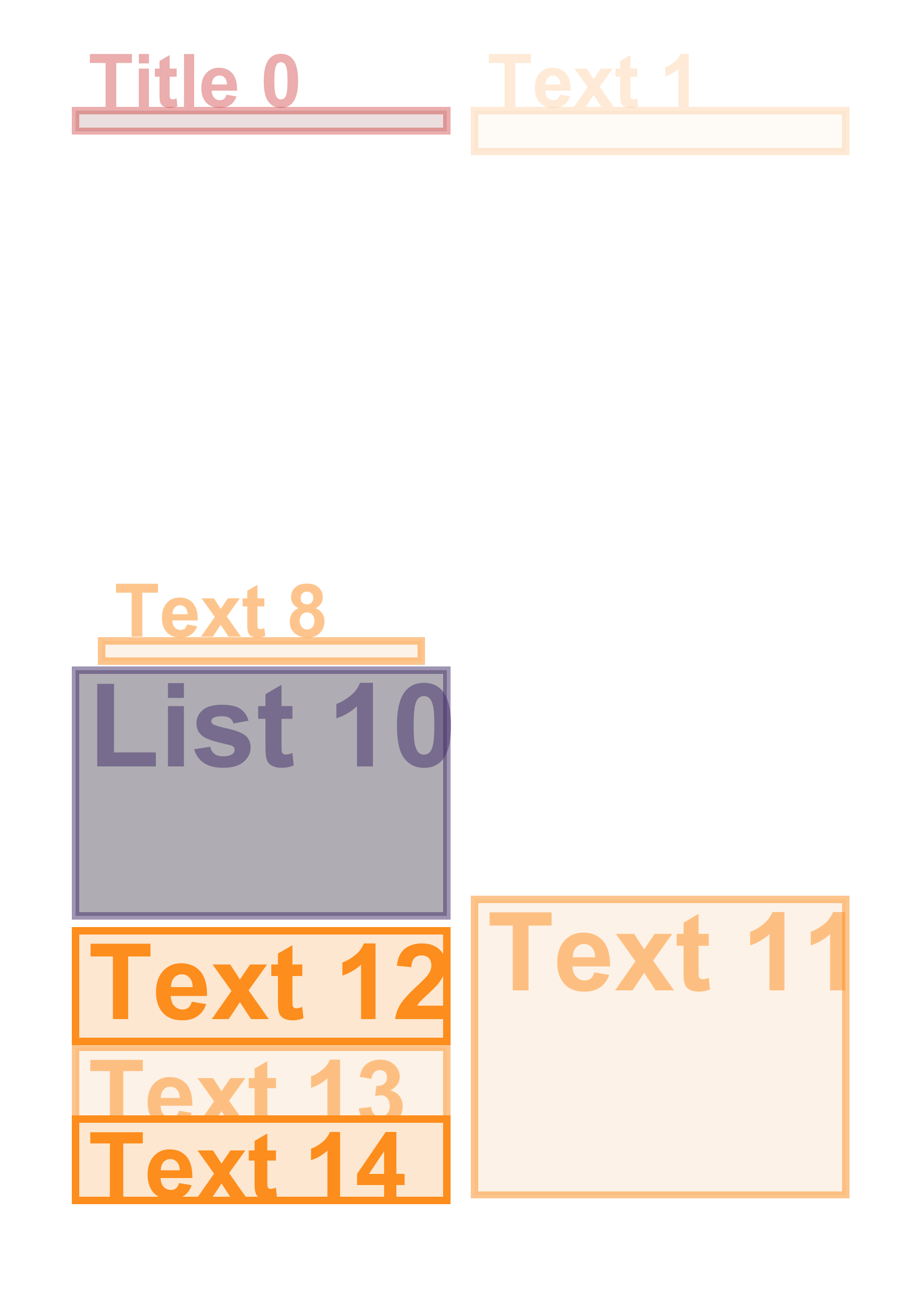} \\

\rotatebox{90}{\hspace{0.3cm}\footnotesize Layer 3}  &  
\includegraphics[width=\attentionDecoderVisDimmed,frame=0.1pt]{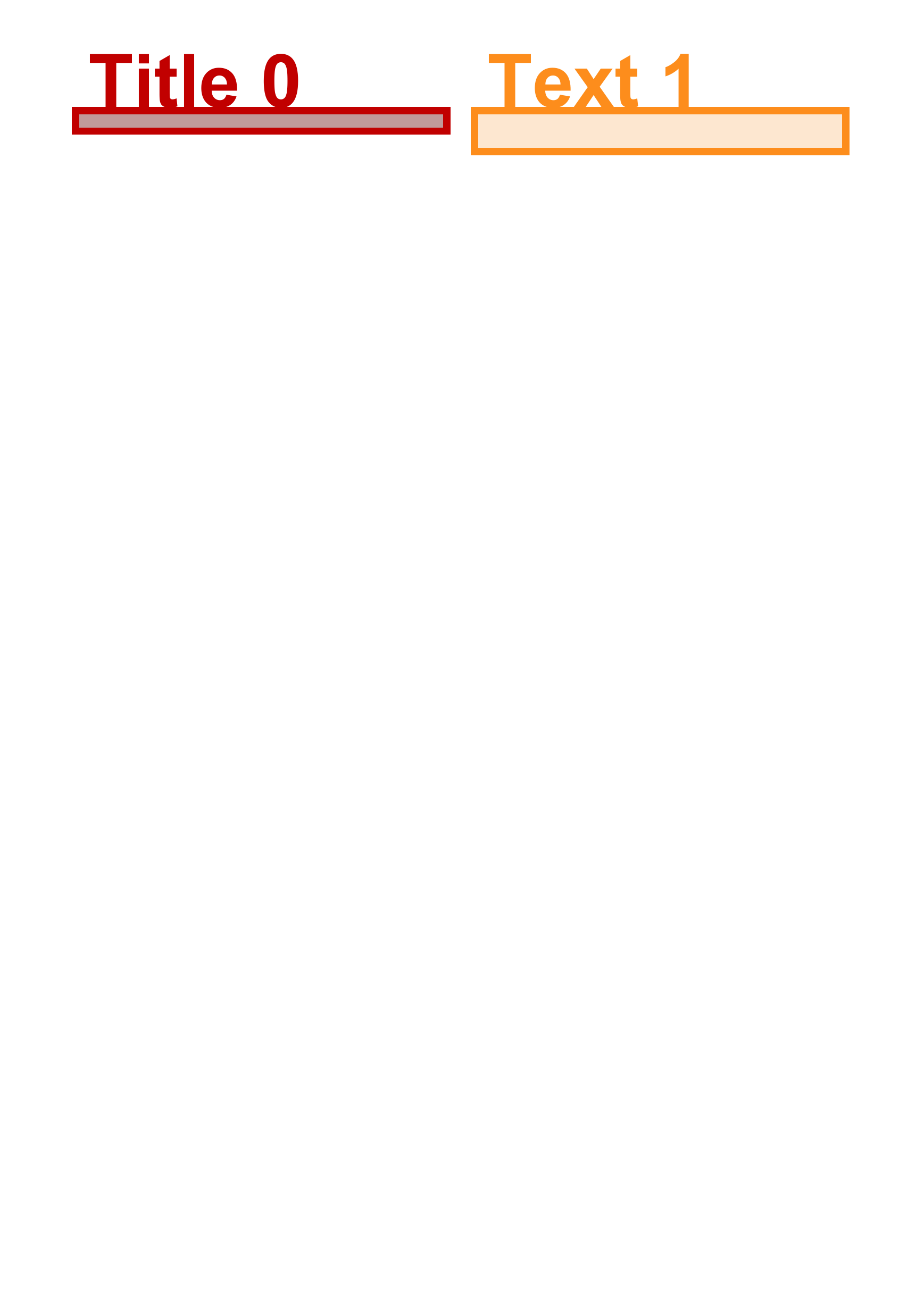} &
\includegraphics[width=\attentionDecoderVisDimmed,frame=0.1pt]{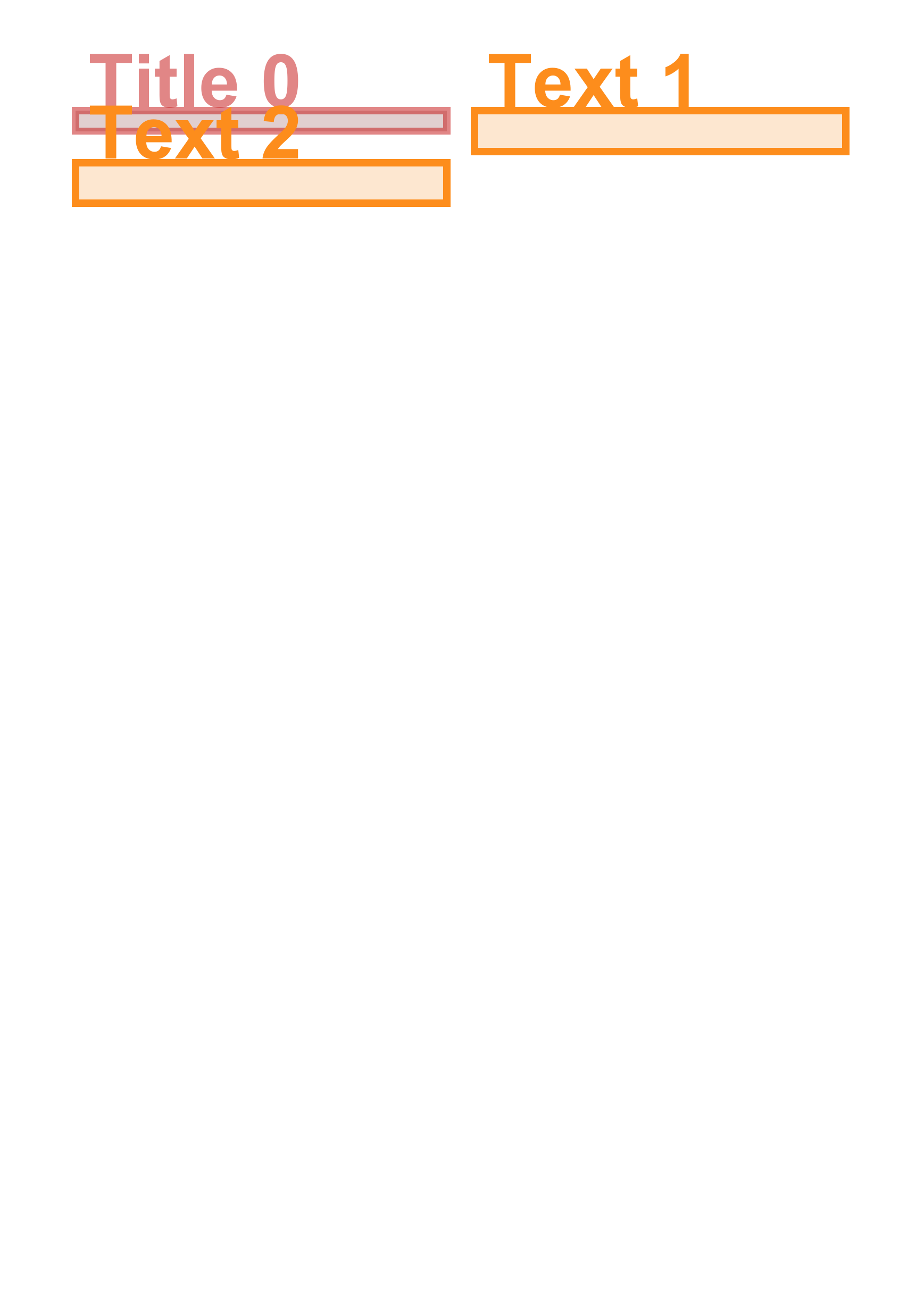} &
\includegraphics[width=\attentionDecoderVisDimmed,frame=0.1pt]{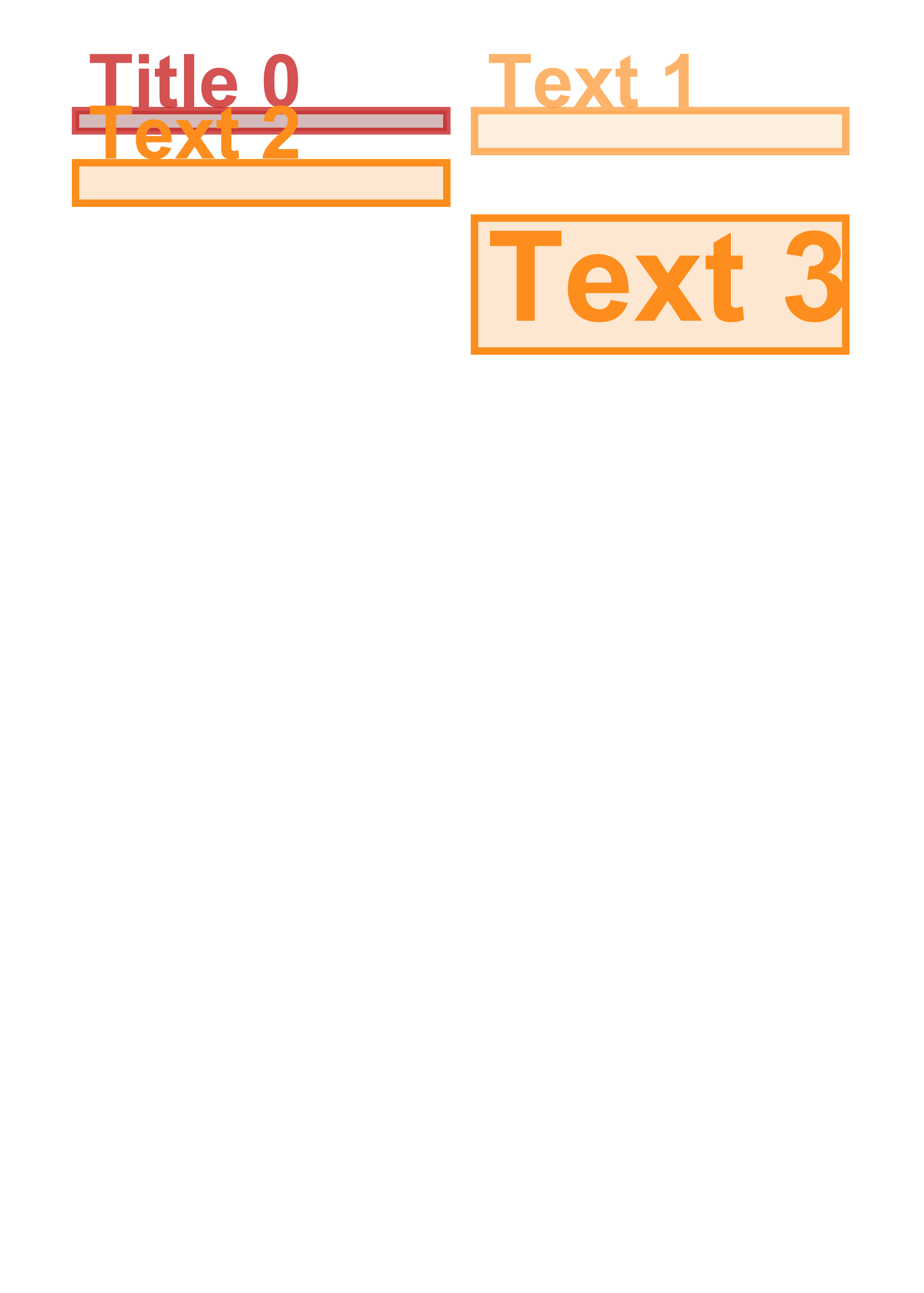} &
\includegraphics[width=\attentionDecoderVisDimmed,frame=0.1pt]{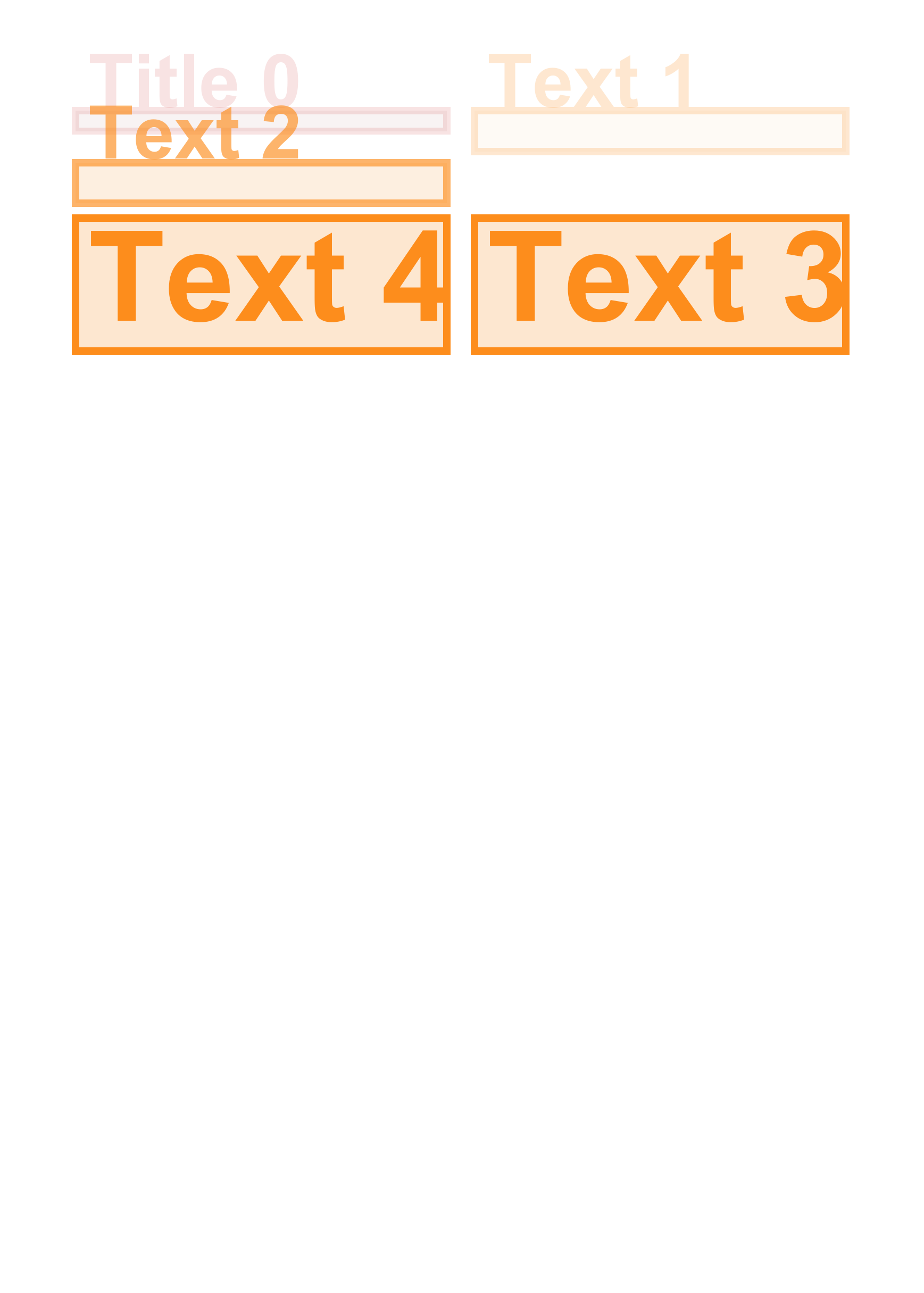} &
\includegraphics[width=\attentionDecoderVisDimmed,frame=0.1pt]{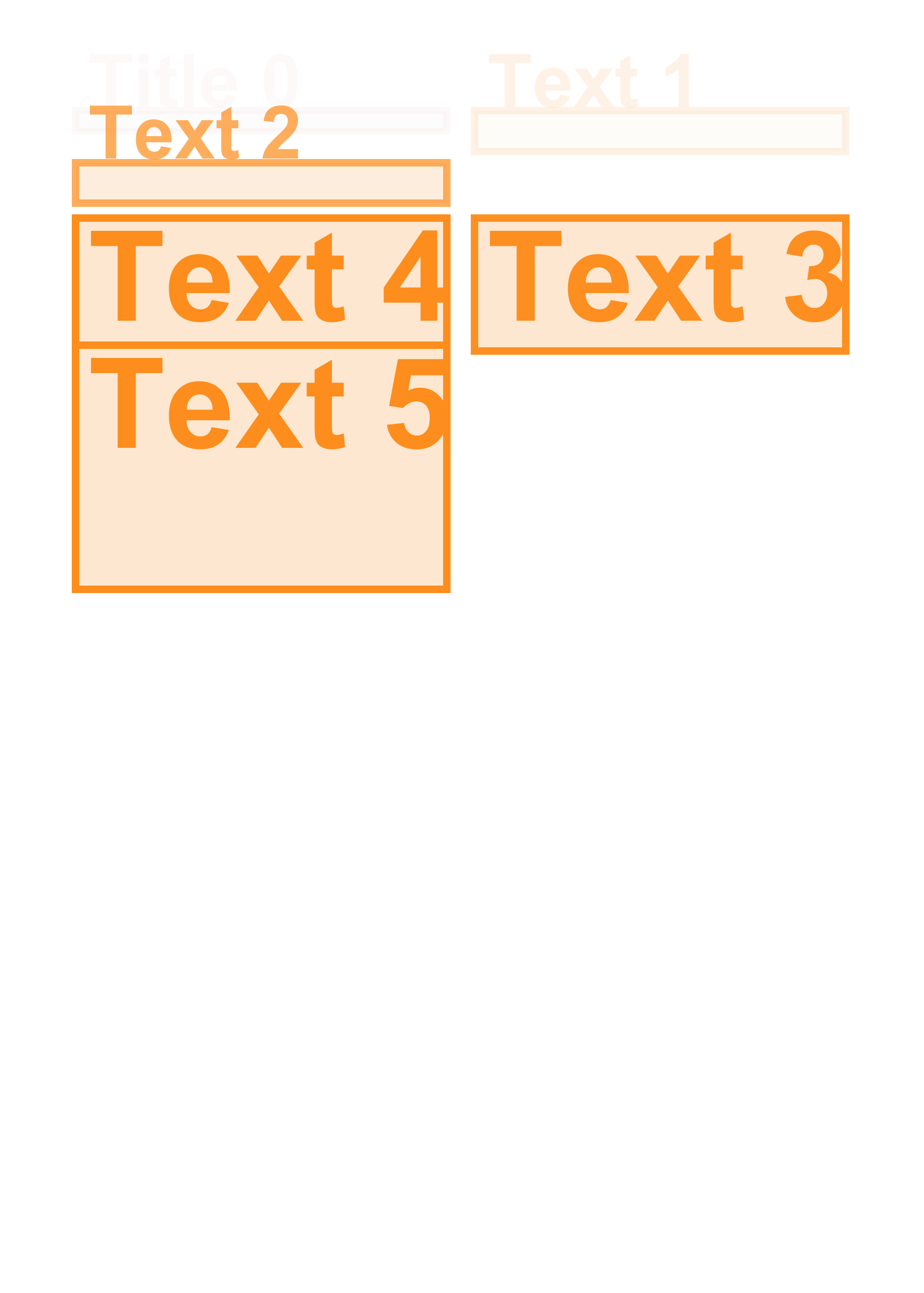} &
\includegraphics[width=\attentionDecoderVisDimmed,frame=0.1pt]{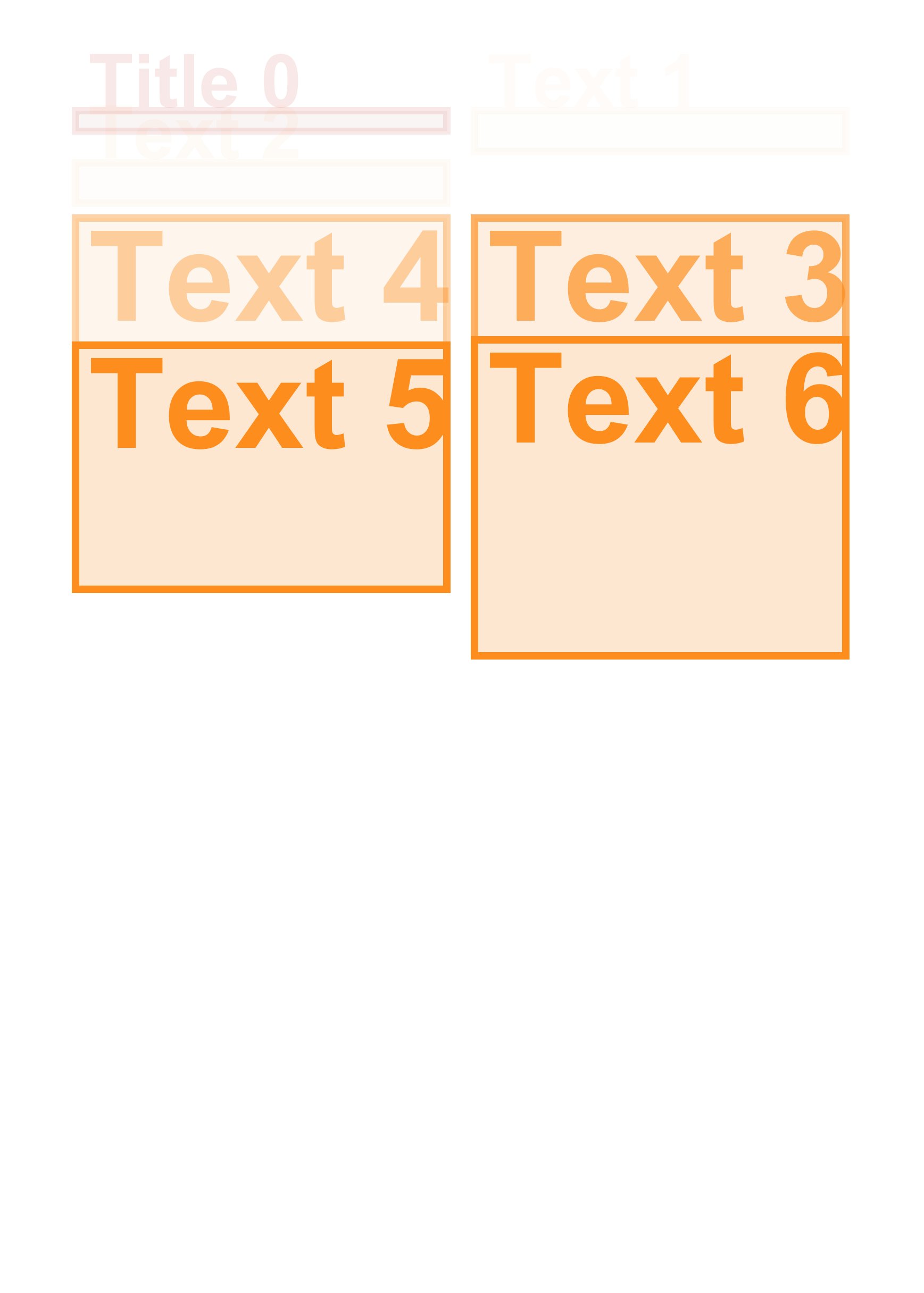} &
\includegraphics[width=\attentionDecoderVisDimmed,frame=0.1pt]{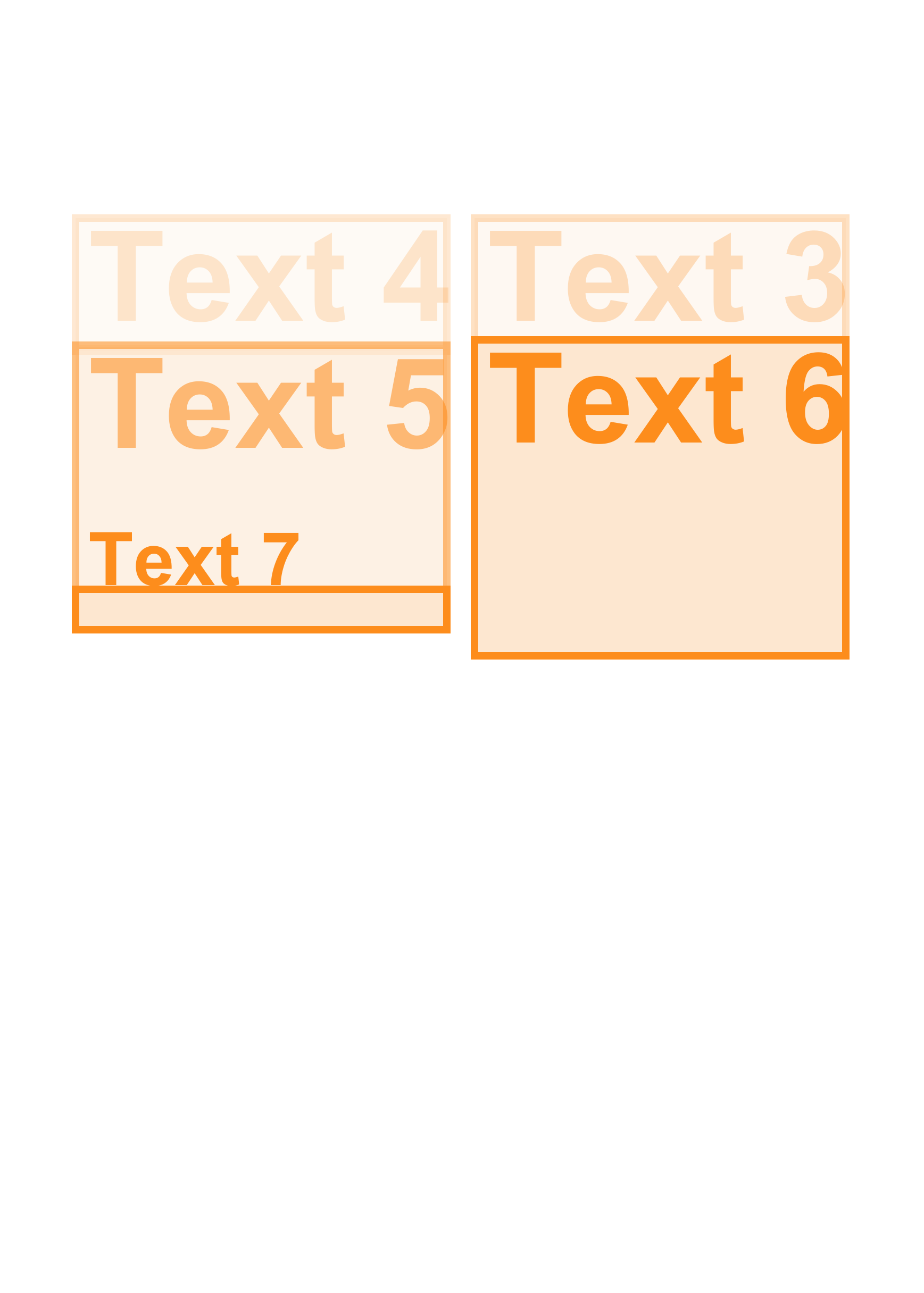} &
\includegraphics[width=\attentionDecoderVisDimmed,frame=0.1pt]{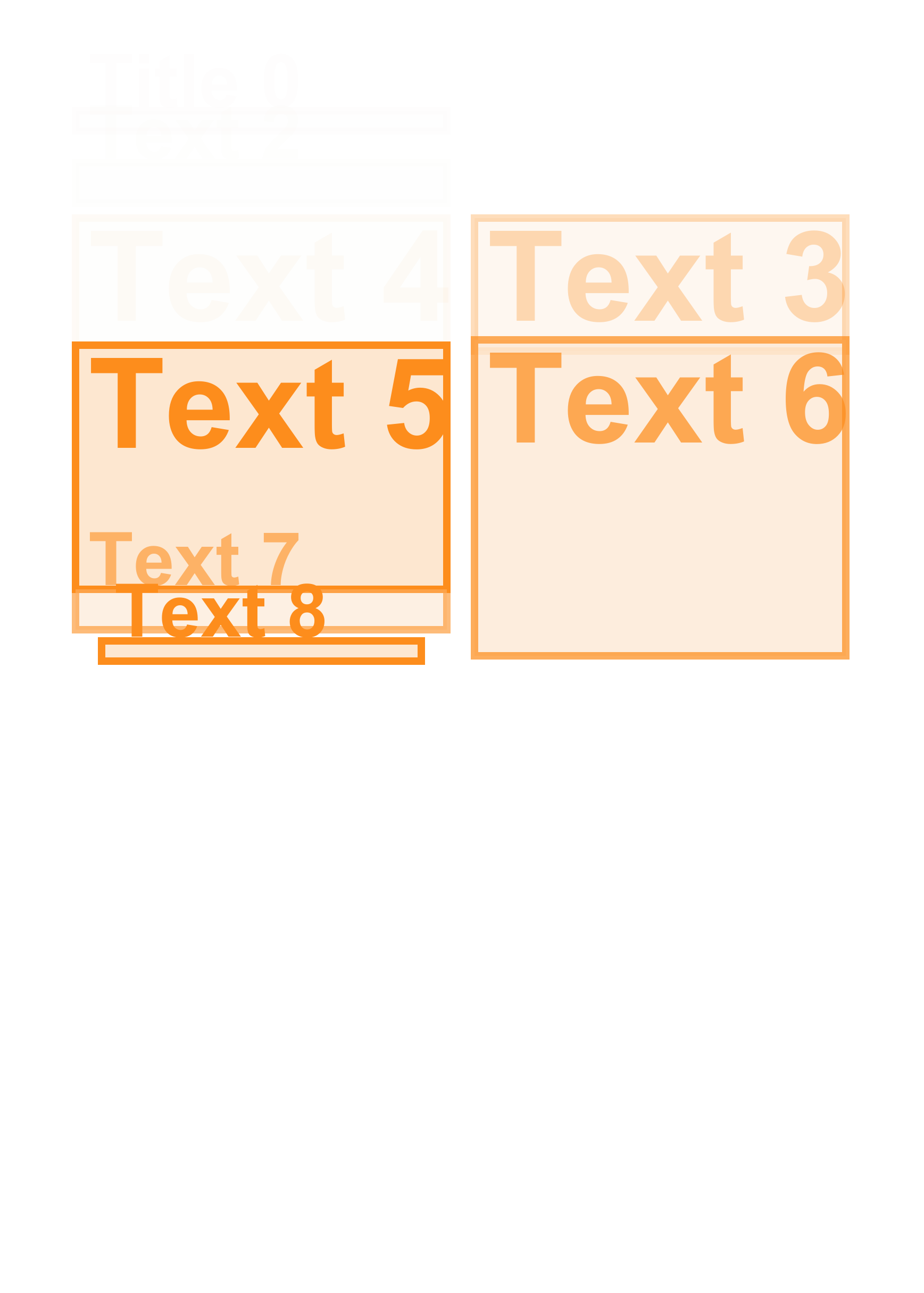} &
\includegraphics[width=\attentionDecoderVisDimmed,frame=0.1pt]{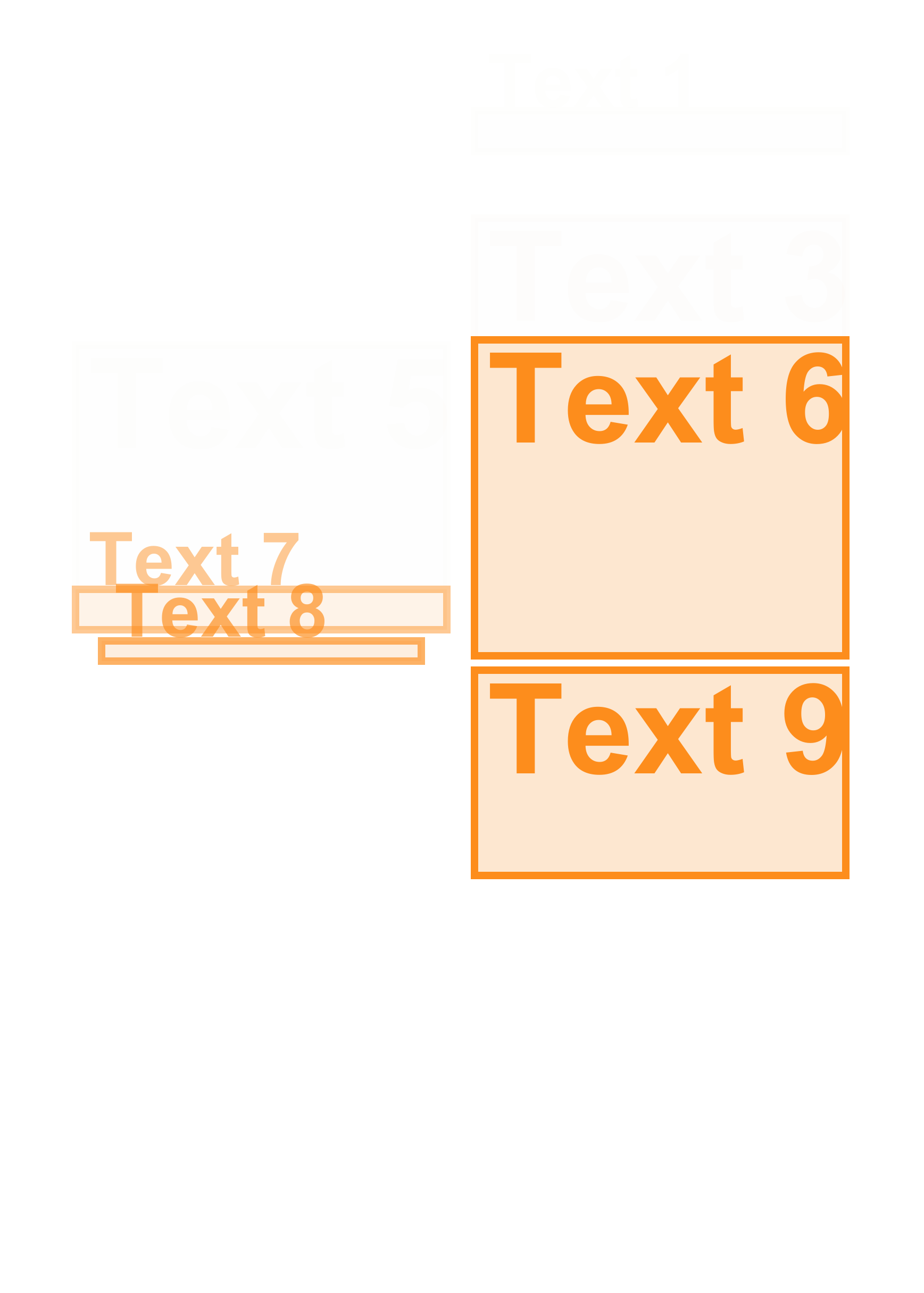} &
\includegraphics[width=\attentionDecoderVisDimmed,frame=0.1pt]{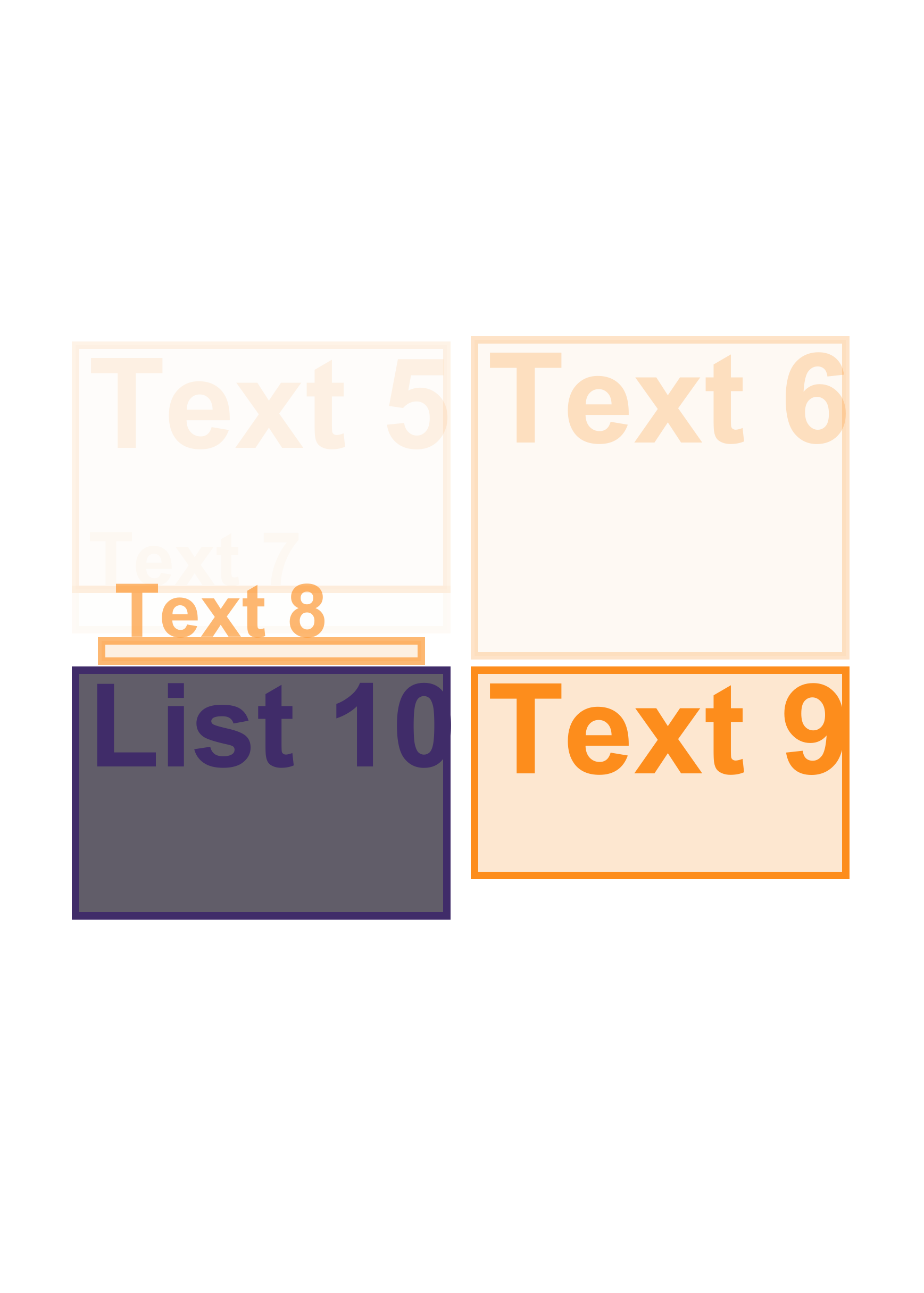} &
\includegraphics[width=\attentionDecoderVisDimmed,frame=0.1pt]{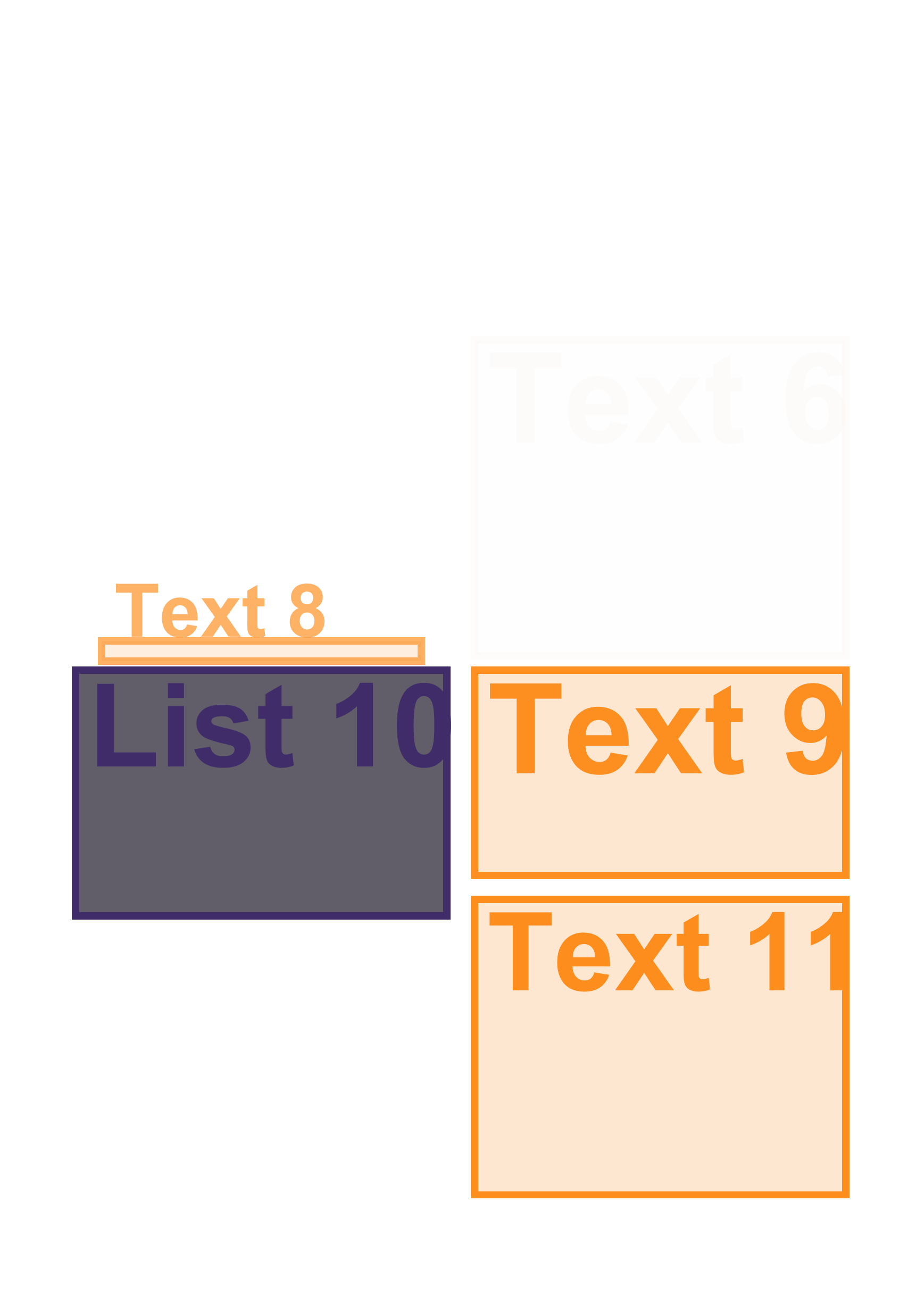} &
\includegraphics[width=\attentionDecoderVisDimmed,frame=0.1pt]{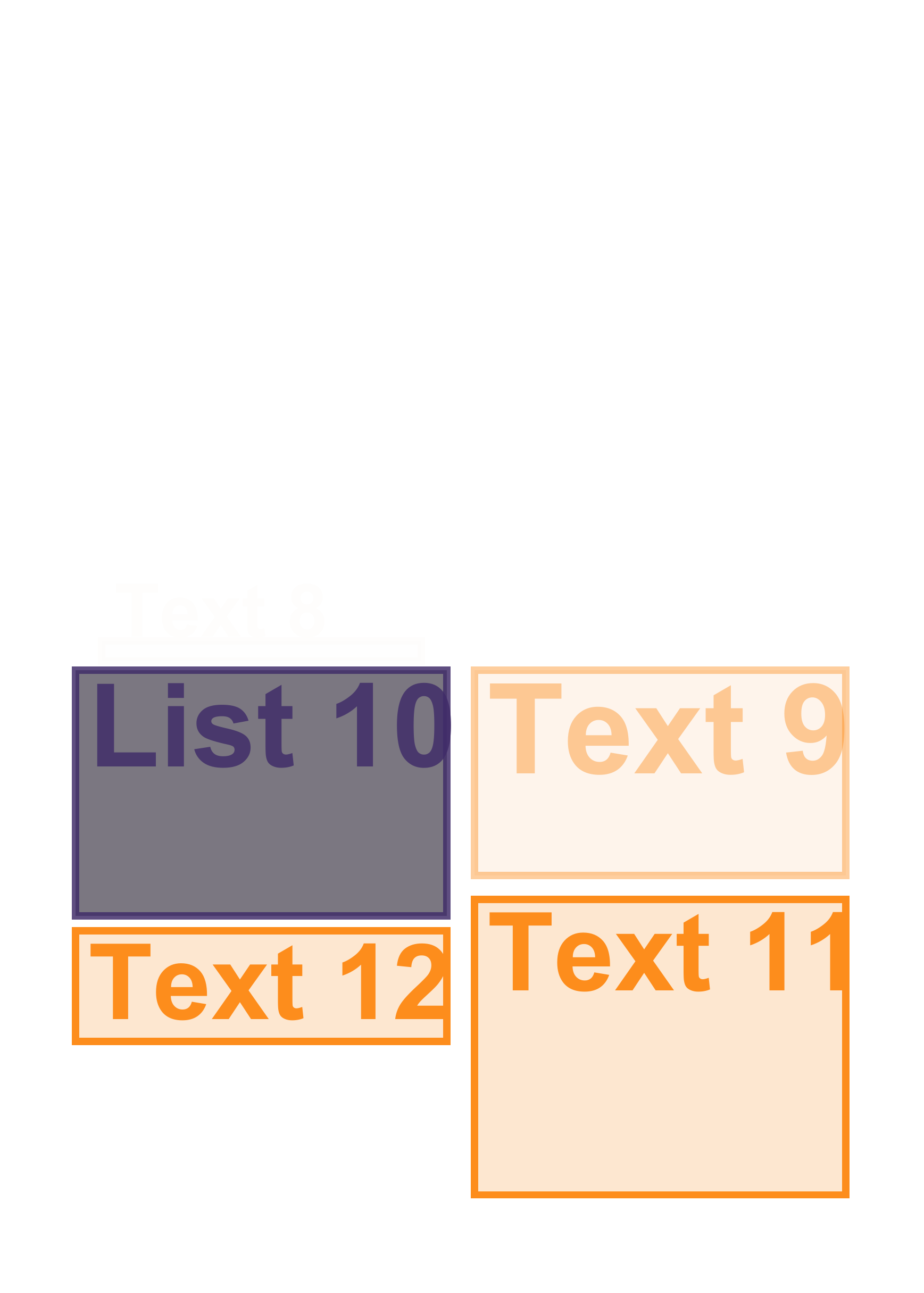} &
\includegraphics[width=\attentionDecoderVisDimmed,frame=0.1pt]{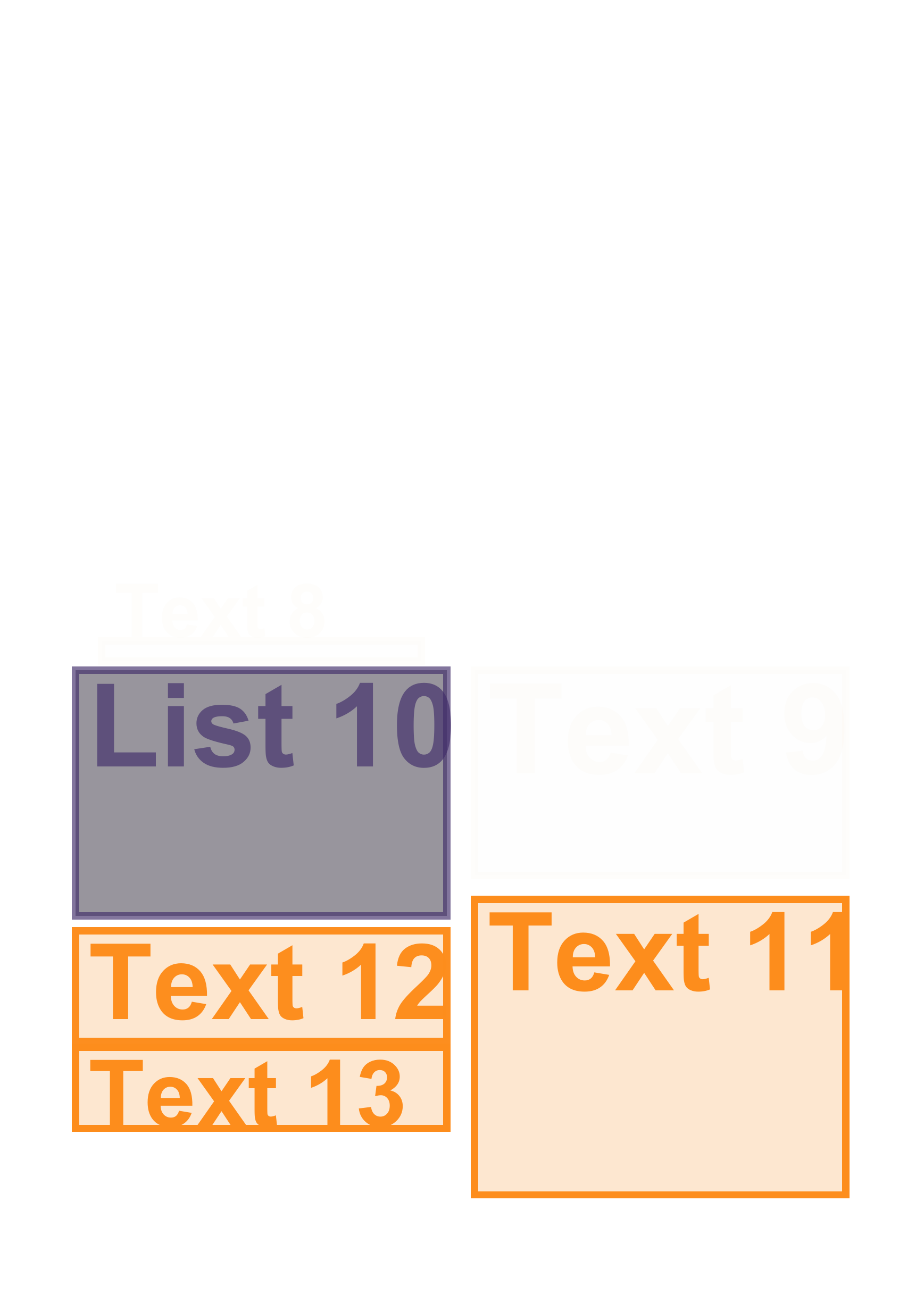} &
\includegraphics[width=\attentionDecoderVisDimmed,frame=0.1pt]{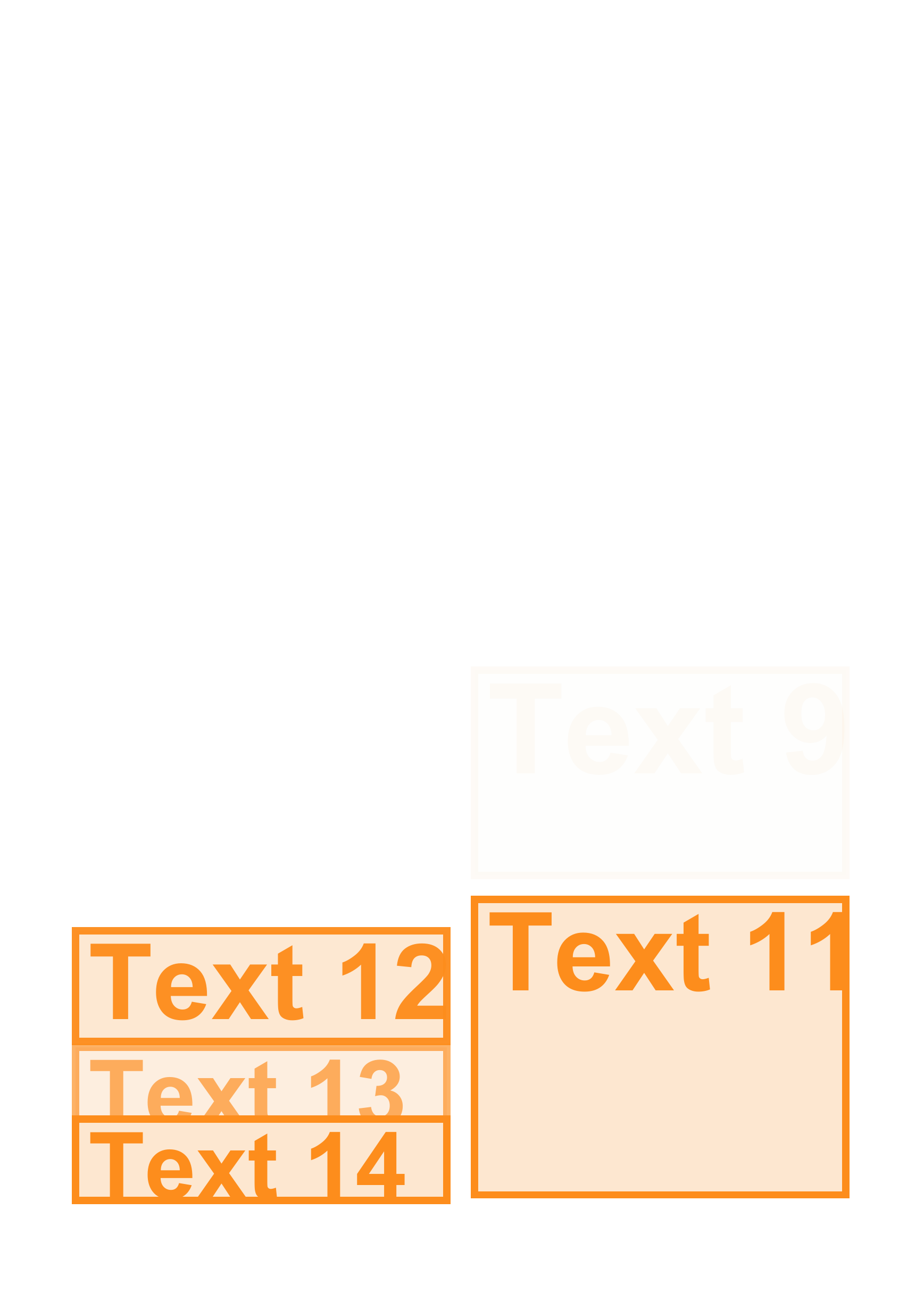} \\

\rotatebox{90}{\hspace{0.4cm}\footnotesize Layer 4}  &  
\includegraphics[width=\attentionDecoderVisDimmed,frame=0.1pt]{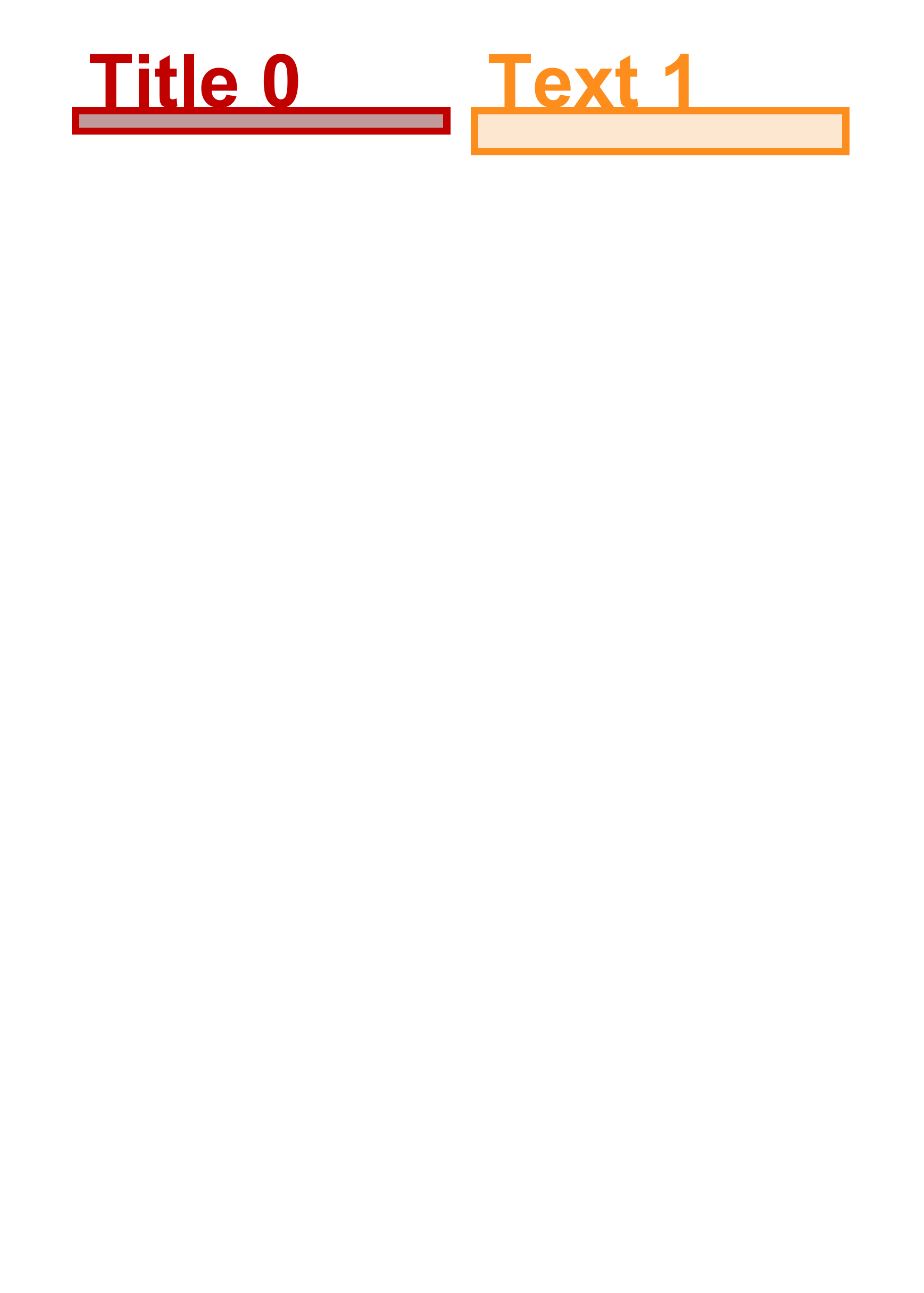} &
\includegraphics[width=\attentionDecoderVisDimmed,frame=0.1pt]{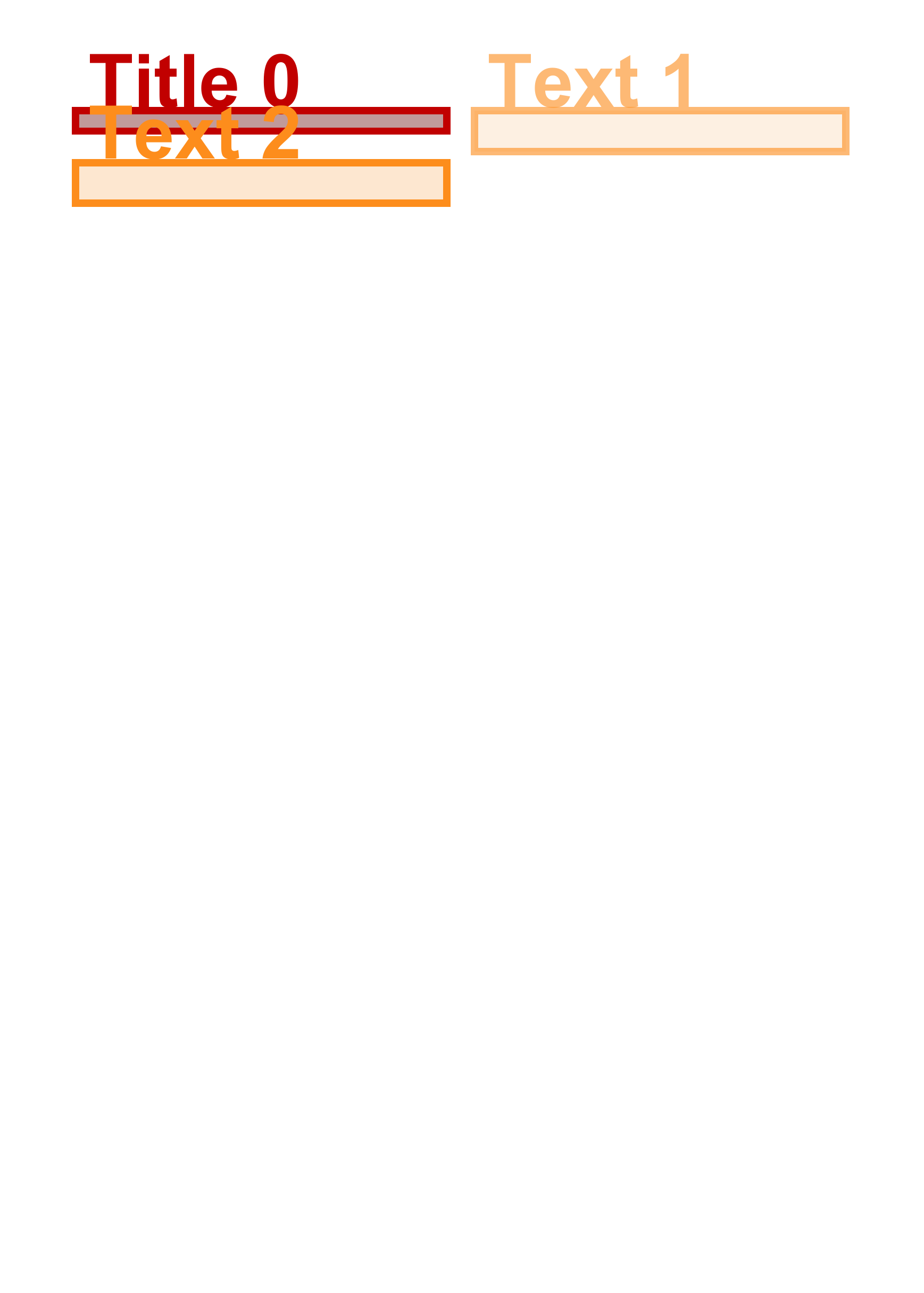} &
\includegraphics[width=\attentionDecoderVisDimmed,frame=0.1pt]{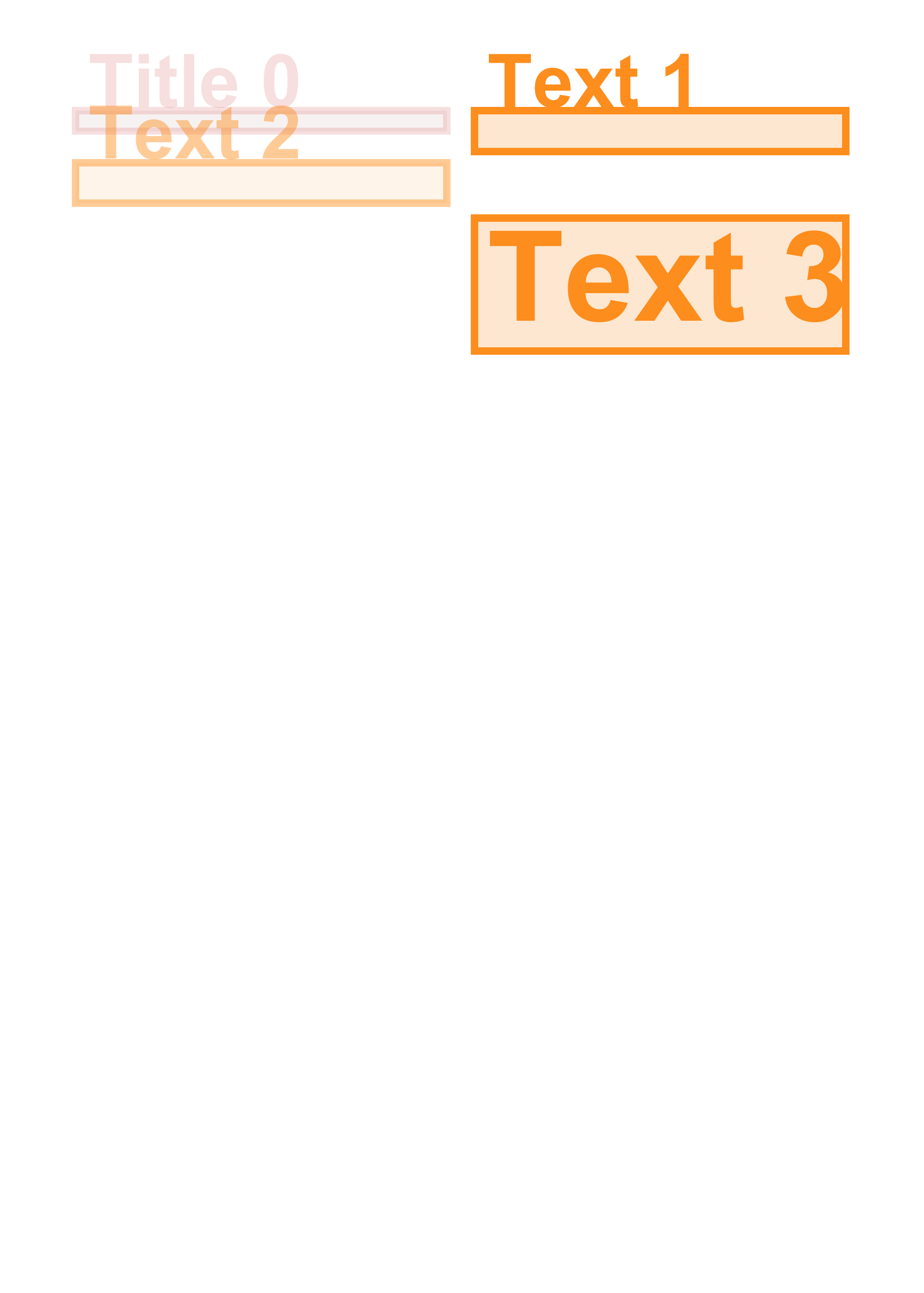} &
\includegraphics[width=\attentionDecoderVisDimmed,frame=0.1pt]{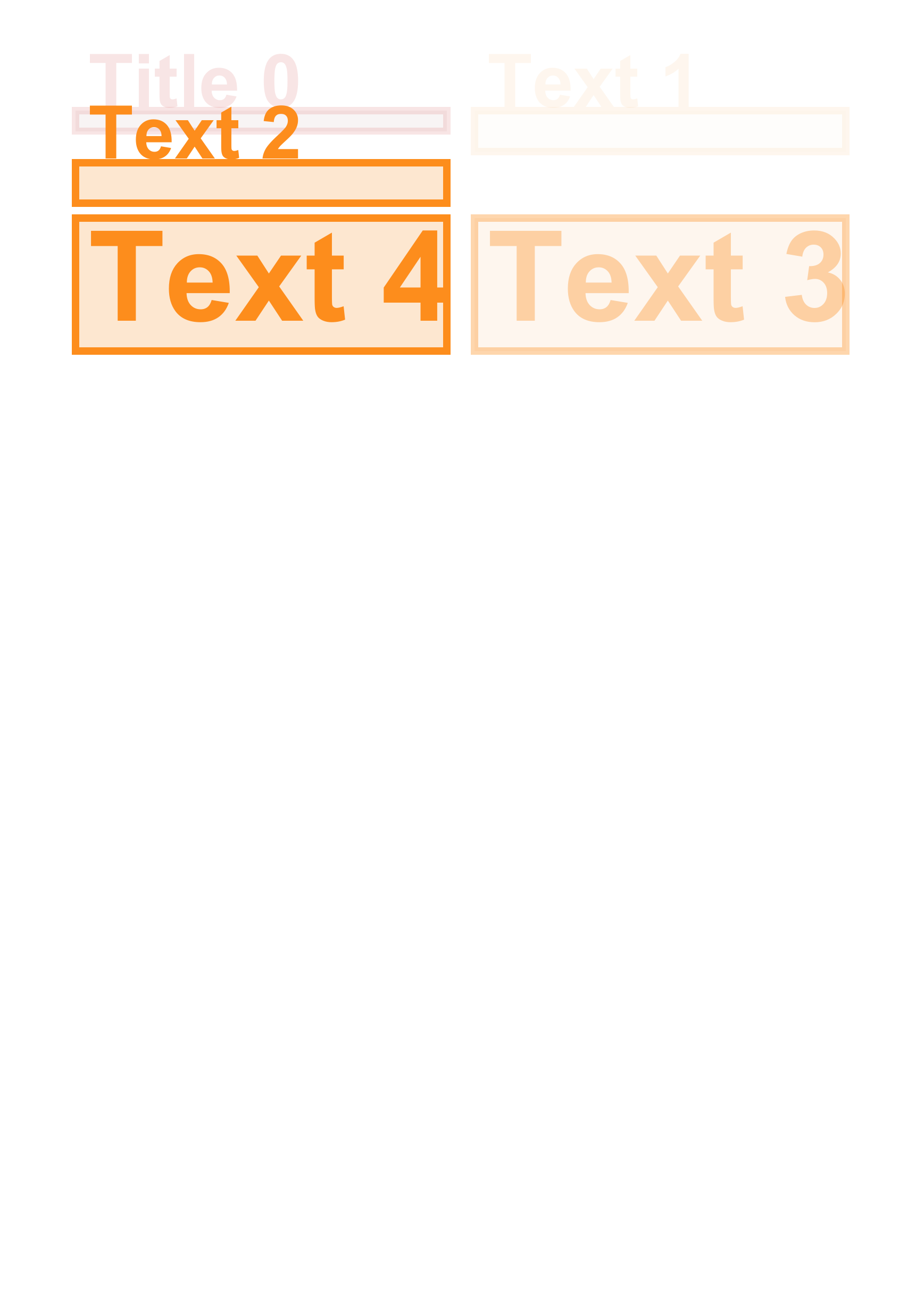} &
\includegraphics[width=\attentionDecoderVisDimmed,frame=0.1pt]{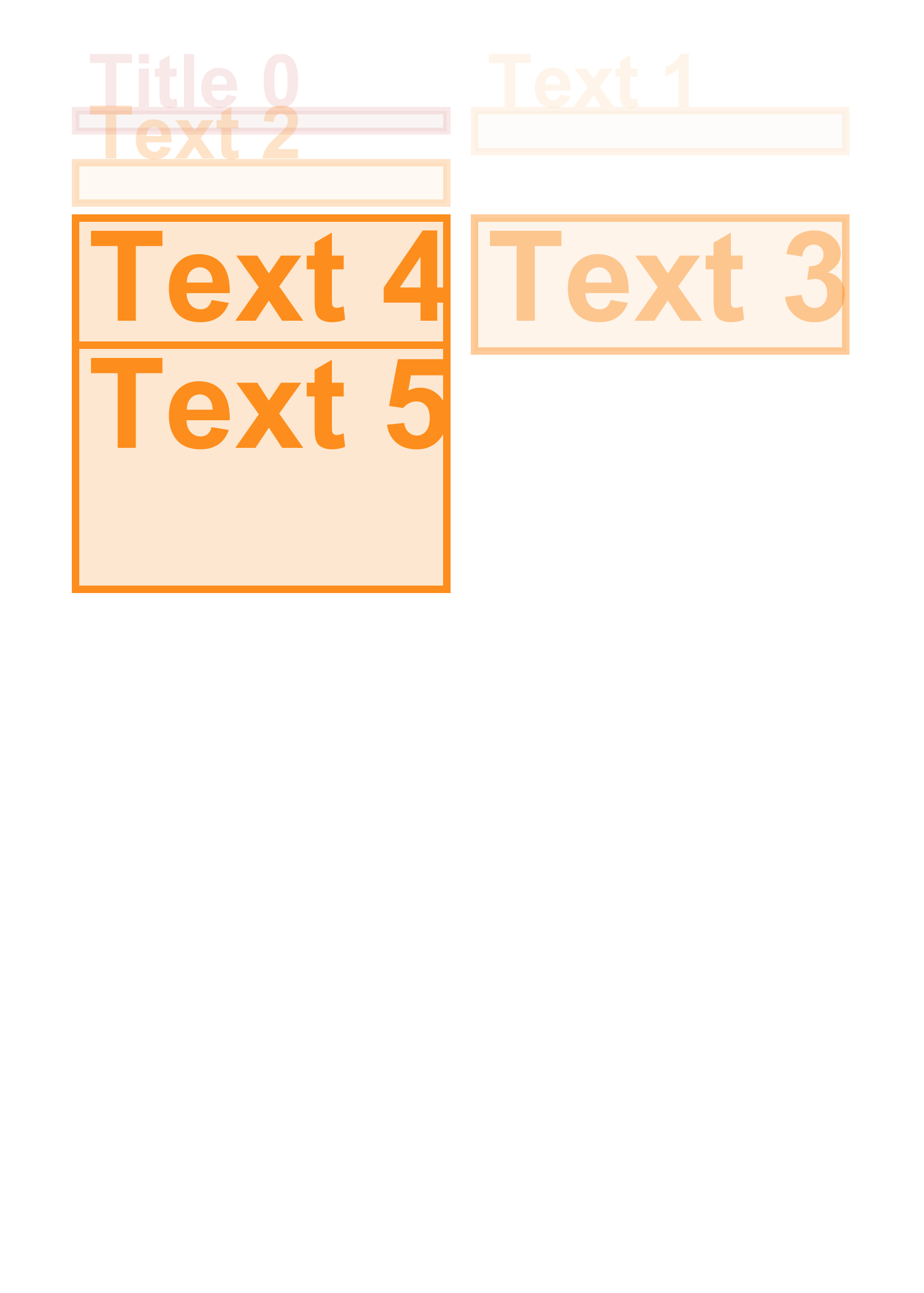} &
\includegraphics[width=\attentionDecoderVisDimmed,frame=0.1pt]{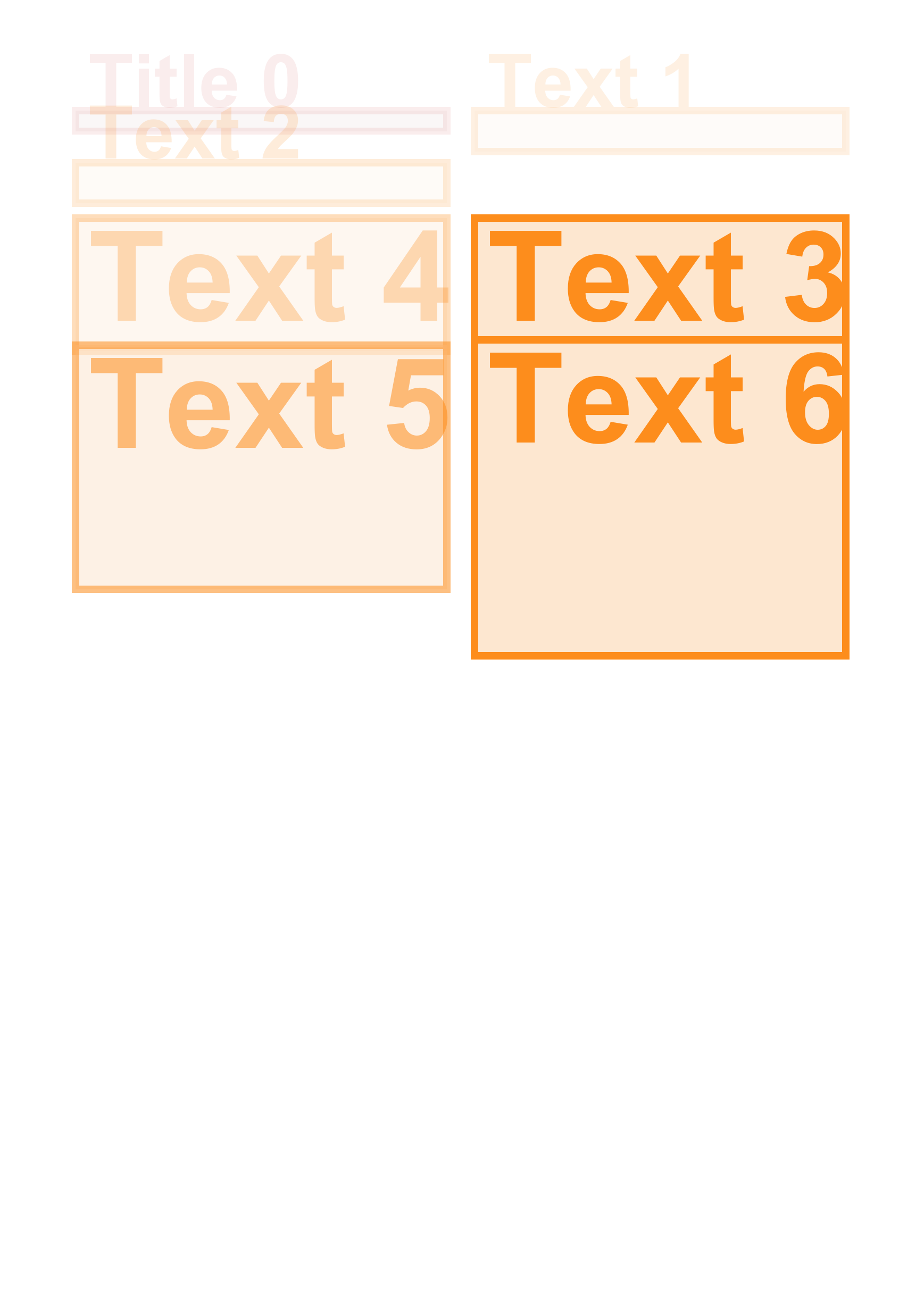} &
\includegraphics[width=\attentionDecoderVisDimmed,frame=0.1pt]{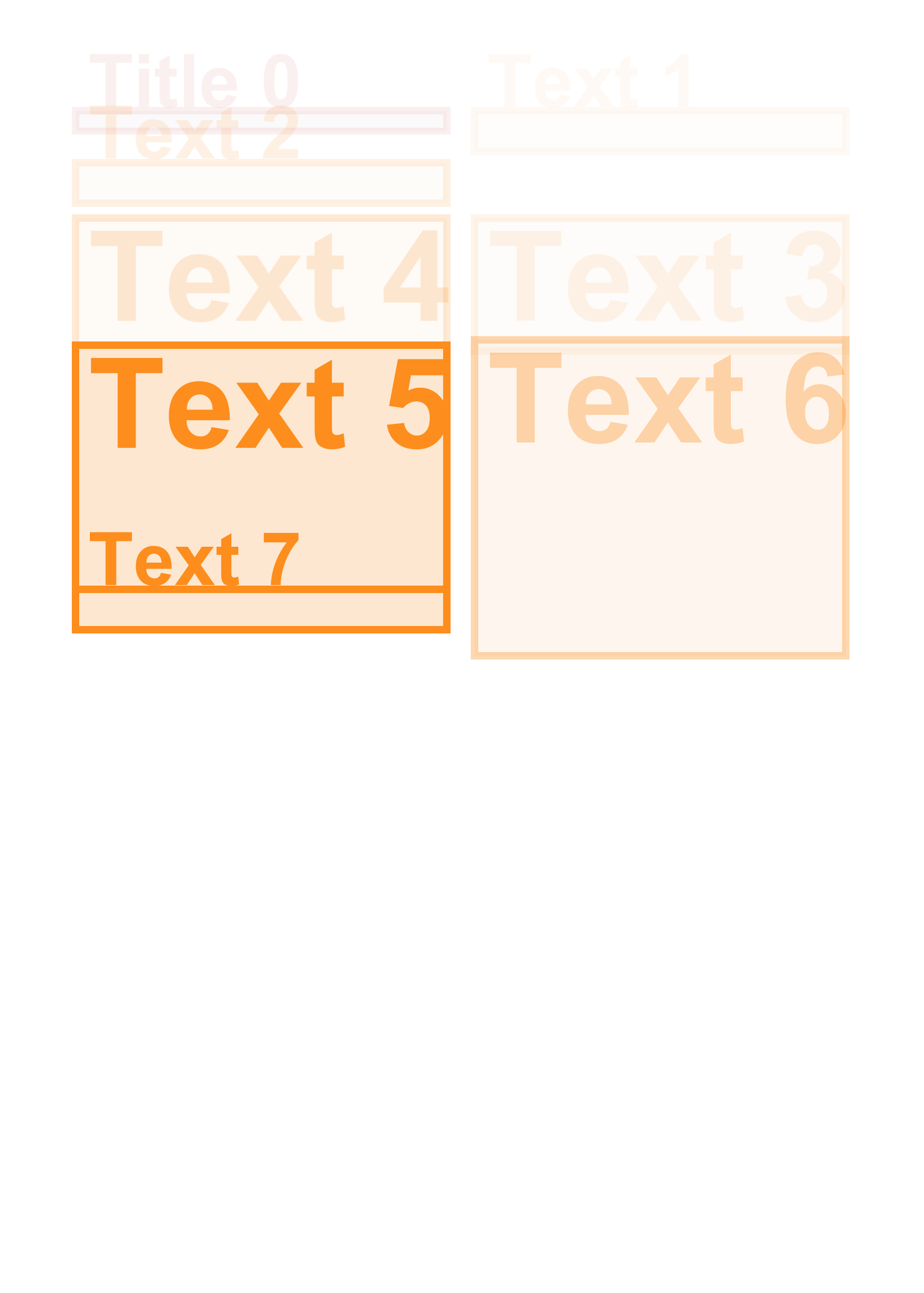} &
\includegraphics[width=\attentionDecoderVisDimmed,frame=0.1pt]{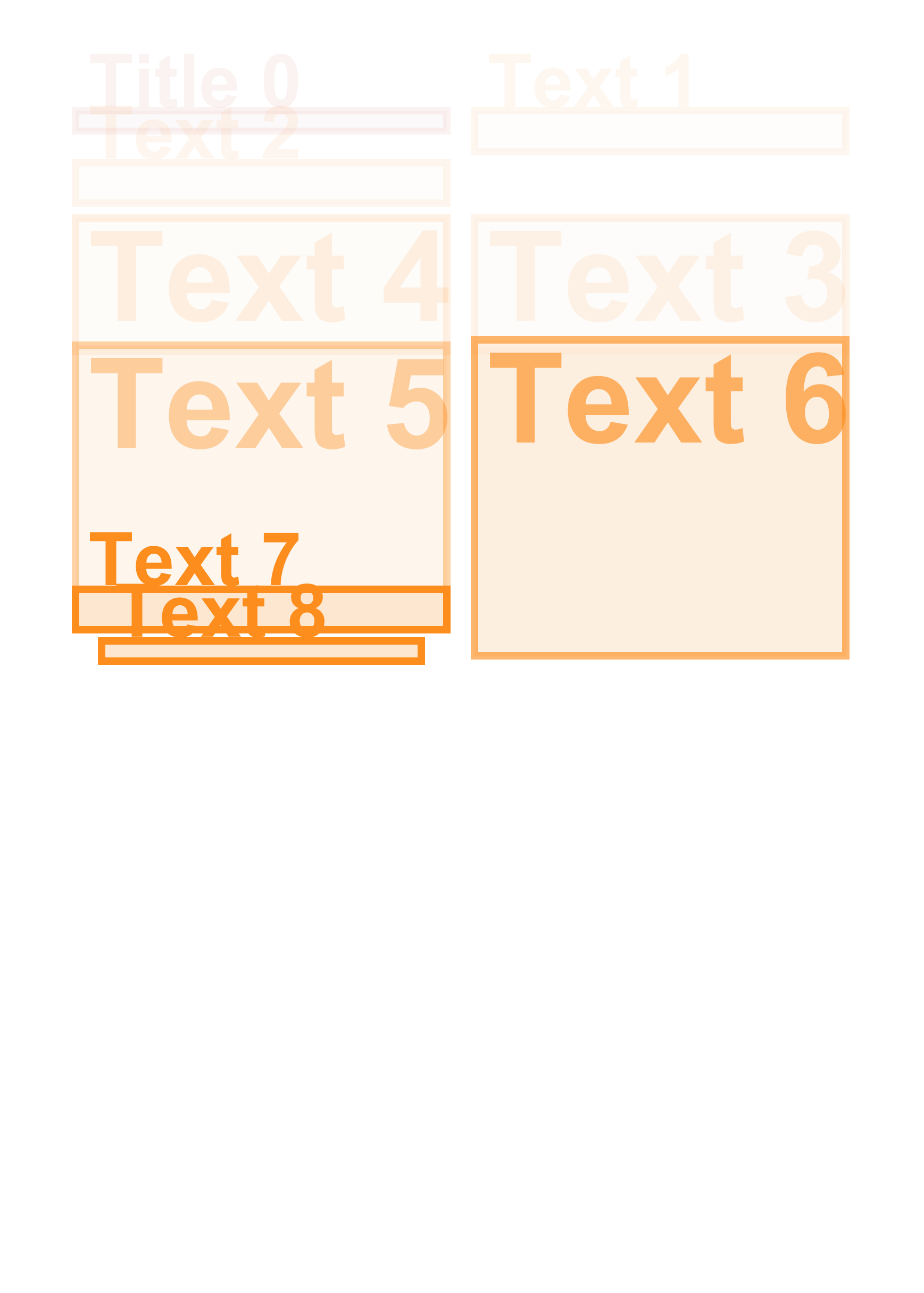} &
\includegraphics[width=\attentionDecoderVisDimmed,frame=0.1pt]{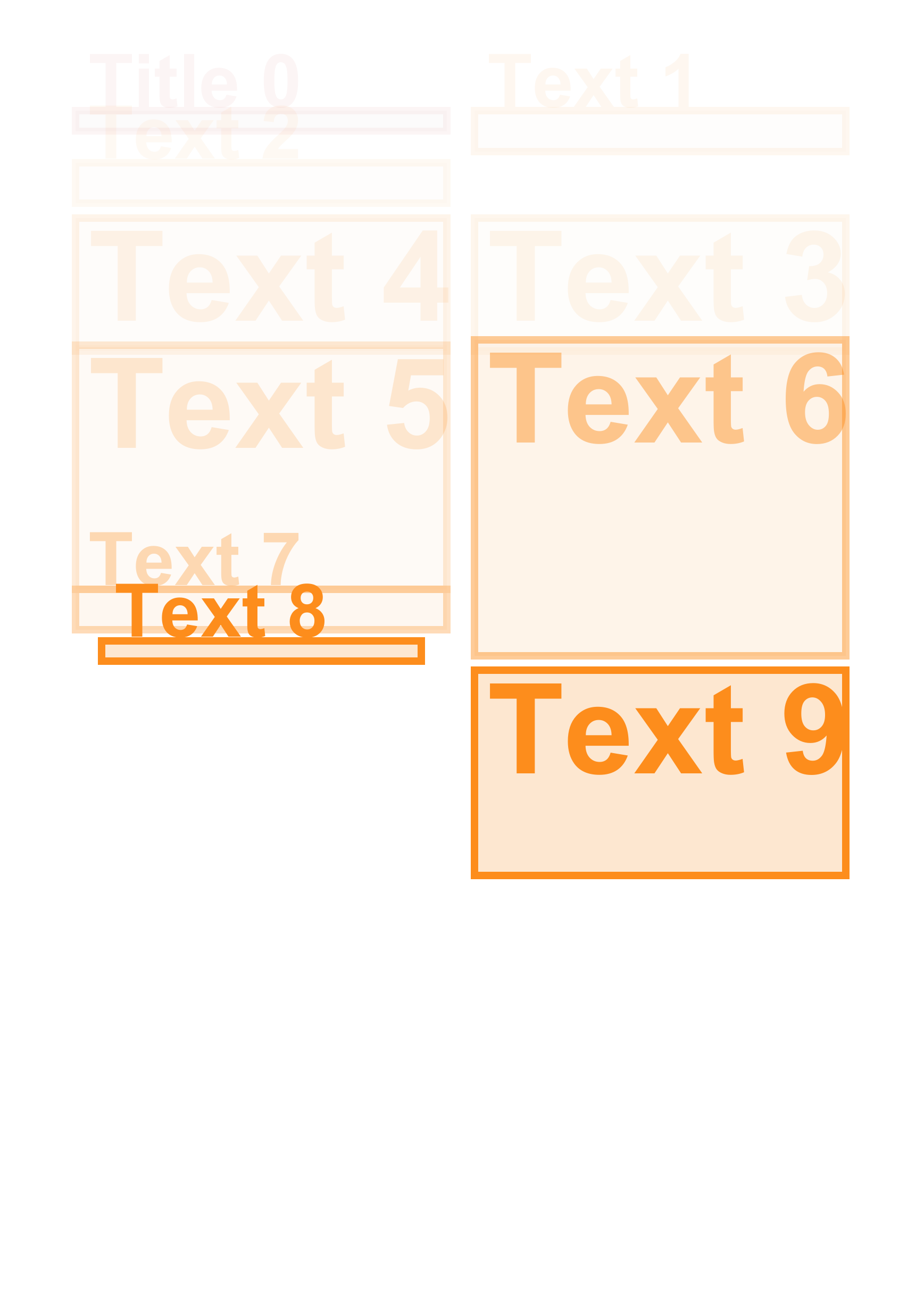} &
\includegraphics[width=\attentionDecoderVisDimmed,frame=0.1pt]{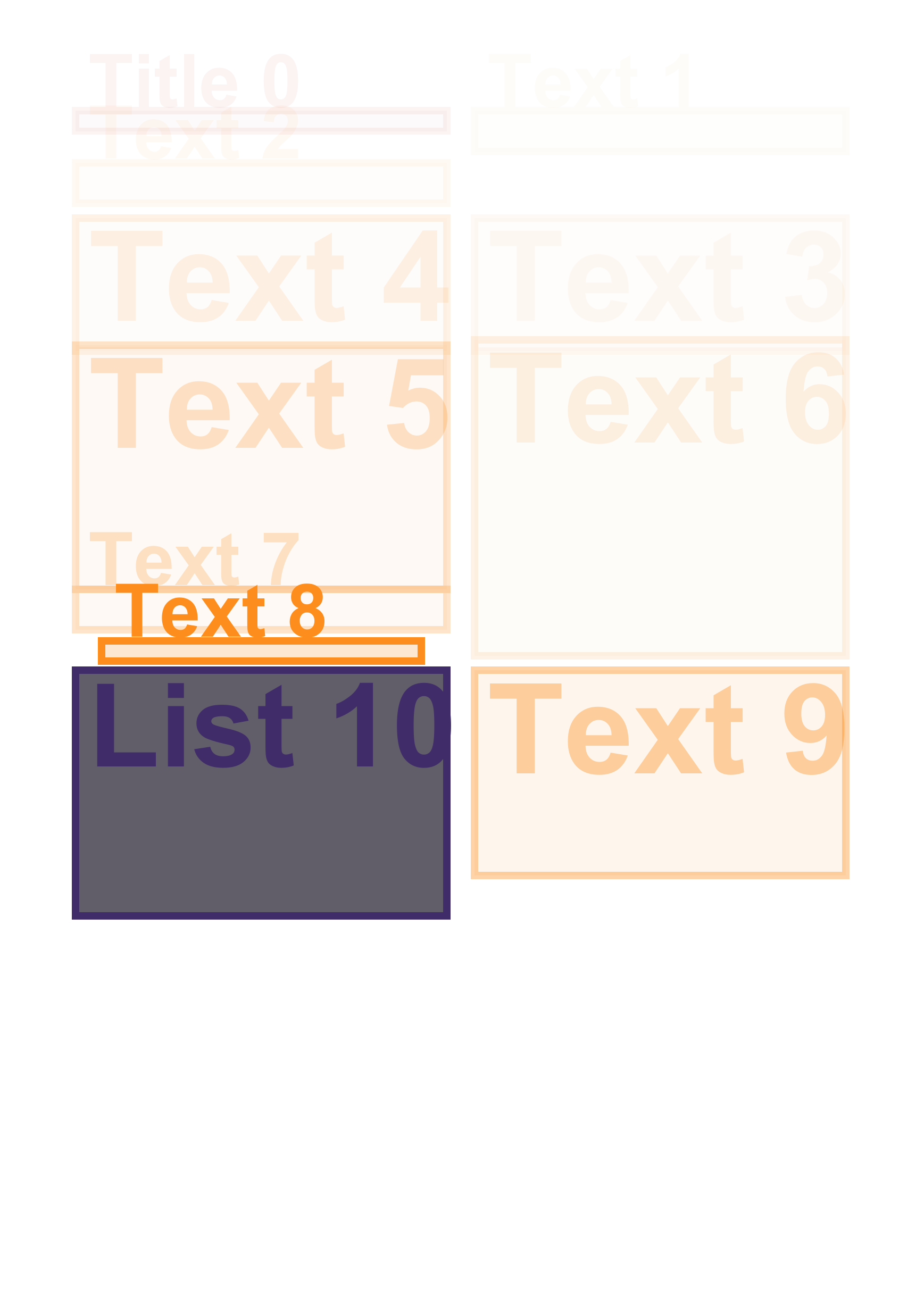} &
\includegraphics[width=\attentionDecoderVisDimmed,frame=0.1pt]{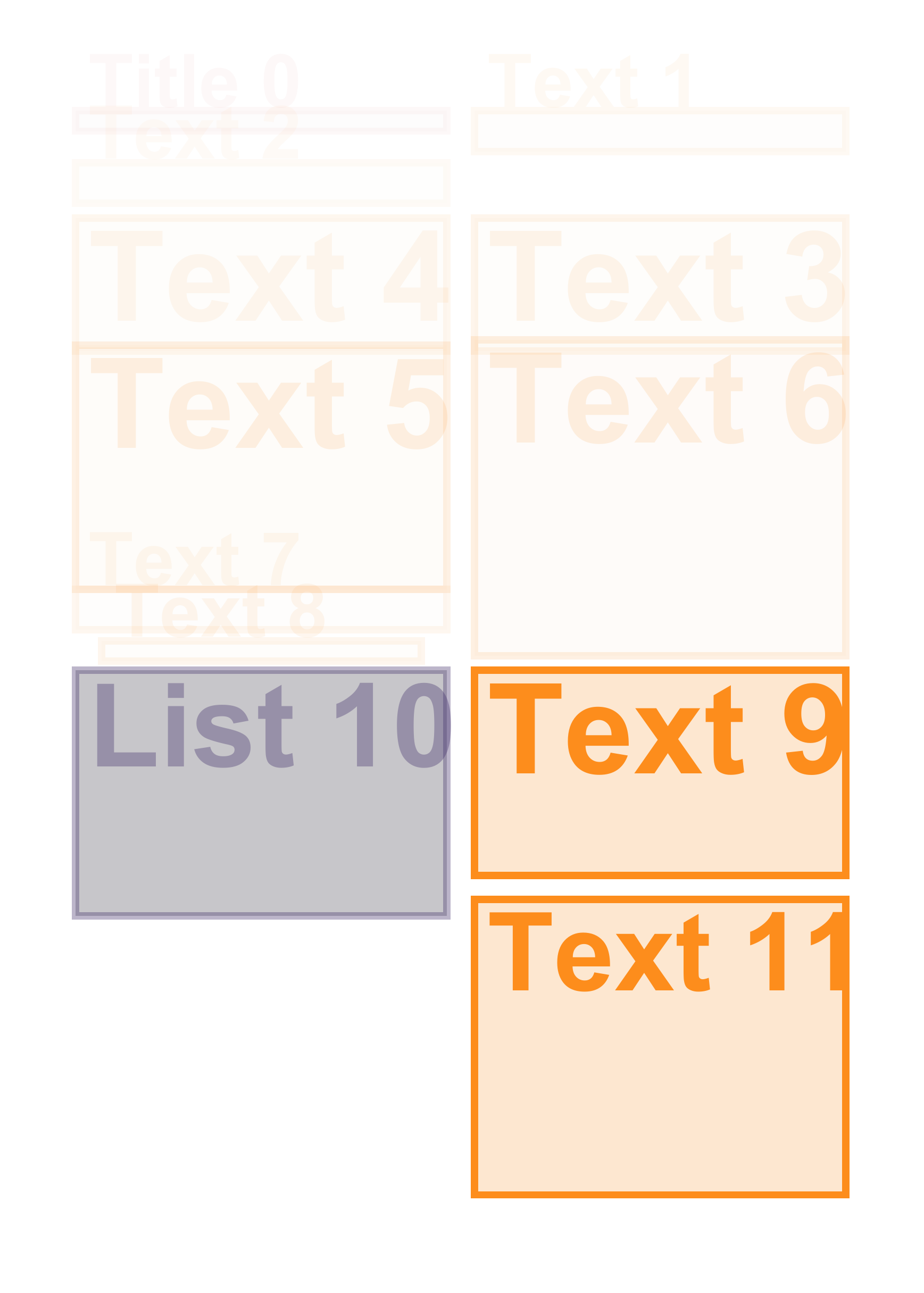} &
\includegraphics[width=\attentionDecoderVisDimmed,frame=0.1pt]{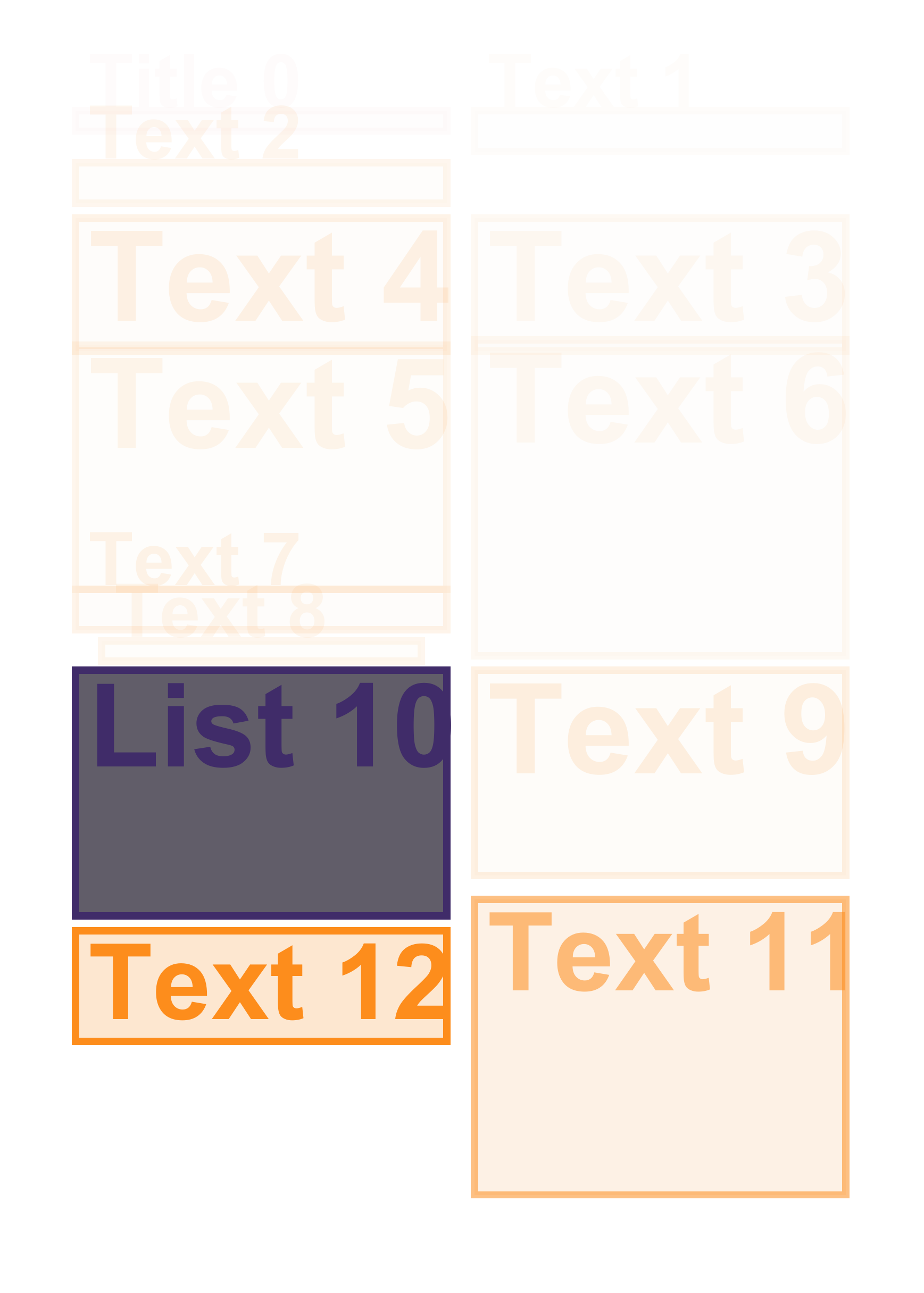} &
\includegraphics[width=\attentionDecoderVisDimmed,frame=0.1pt]{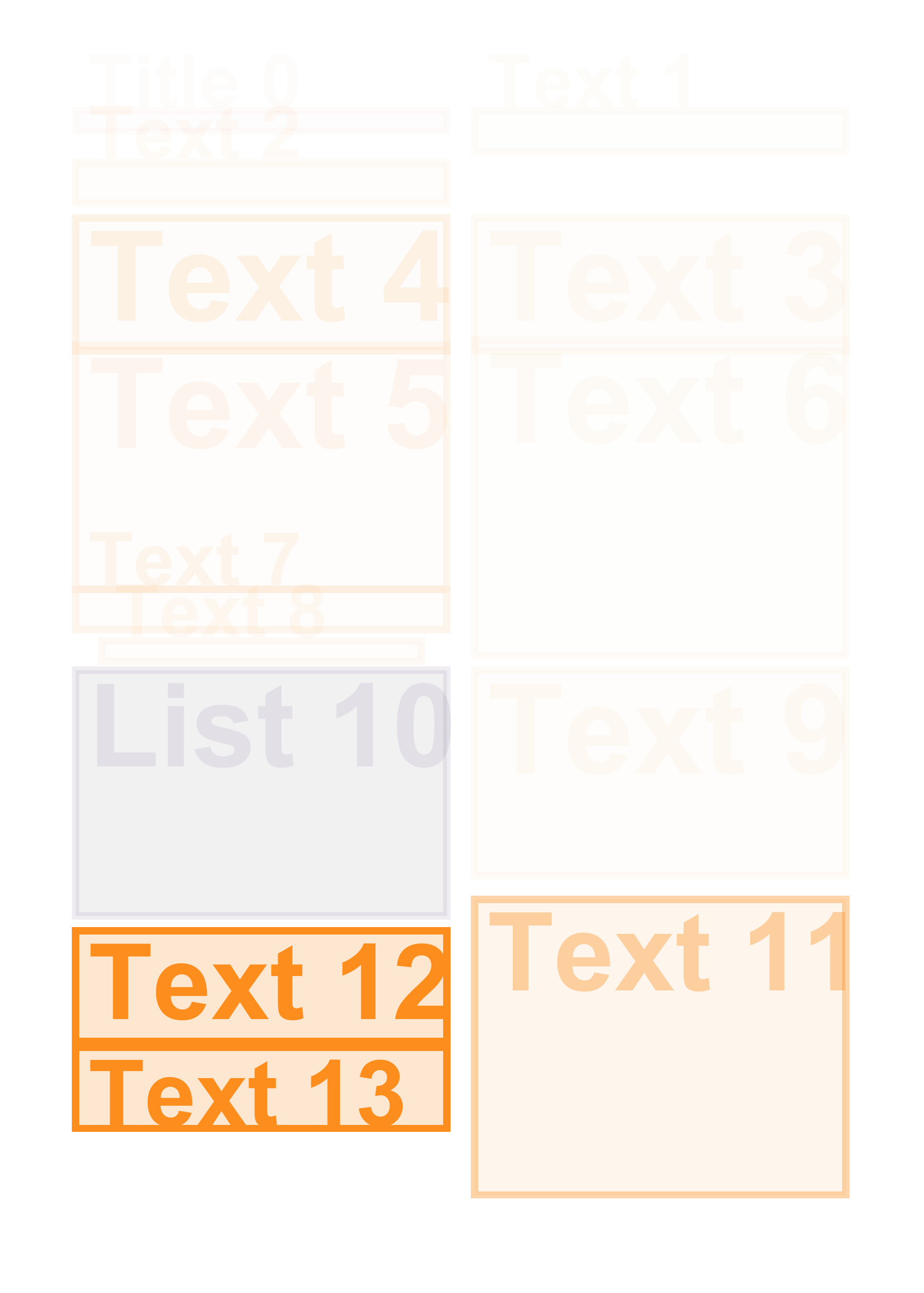} &
\includegraphics[width=\attentionDecoderVisDimmed,frame=0.1pt]{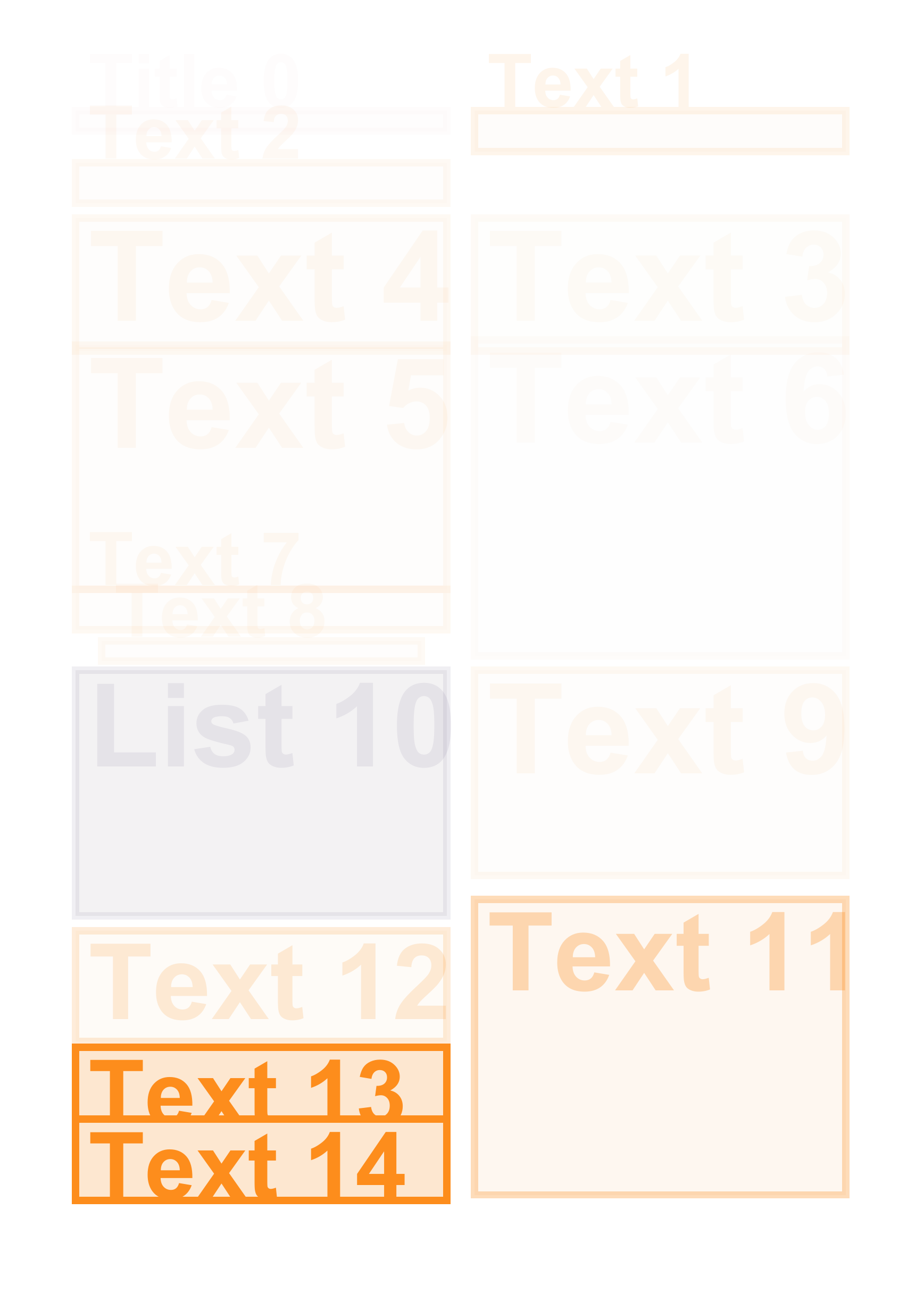} \\

    \end{tabular}
    \caption{Visualization of the relevance of existing elements during the autoregressive decoding.}
    \label{fig:attention_decoder_vis_dimmed}
\end{figure}

During the autoregressive decoding, the network not only relies on the encoded document vector $z$ to determine the location and size of the next element, but also on the result of the previous iterations via self-attention. In fig. \ref{fig:attention_decoder_vis_dimmed} this process is shown.

\clearpage
\onecolumn

\section{Latent Space Analysis}
The properties of its latent space are an important aspect of any VAE. In this section, we show several experiments to analyze the results of interpolating in latent space as well as the effect of each individual latent vector in the non-autoregressive decoder setting.

\subsection{Latent Space Interpolations}
\begin{figure}[h]
    \centering
    \setlength{\tabcolsep}{0.5pt}
    \newlength{\interpolationWidth}
    \setlength{\interpolationWidth}{0.092\linewidth}
    \begin{tabular}{ccccccccccc}
\footnotesize $z_1$&&&&&&&&&&\footnotesize $z_2$\\
\includegraphics[width=\interpolationWidth,frame=0.1pt]{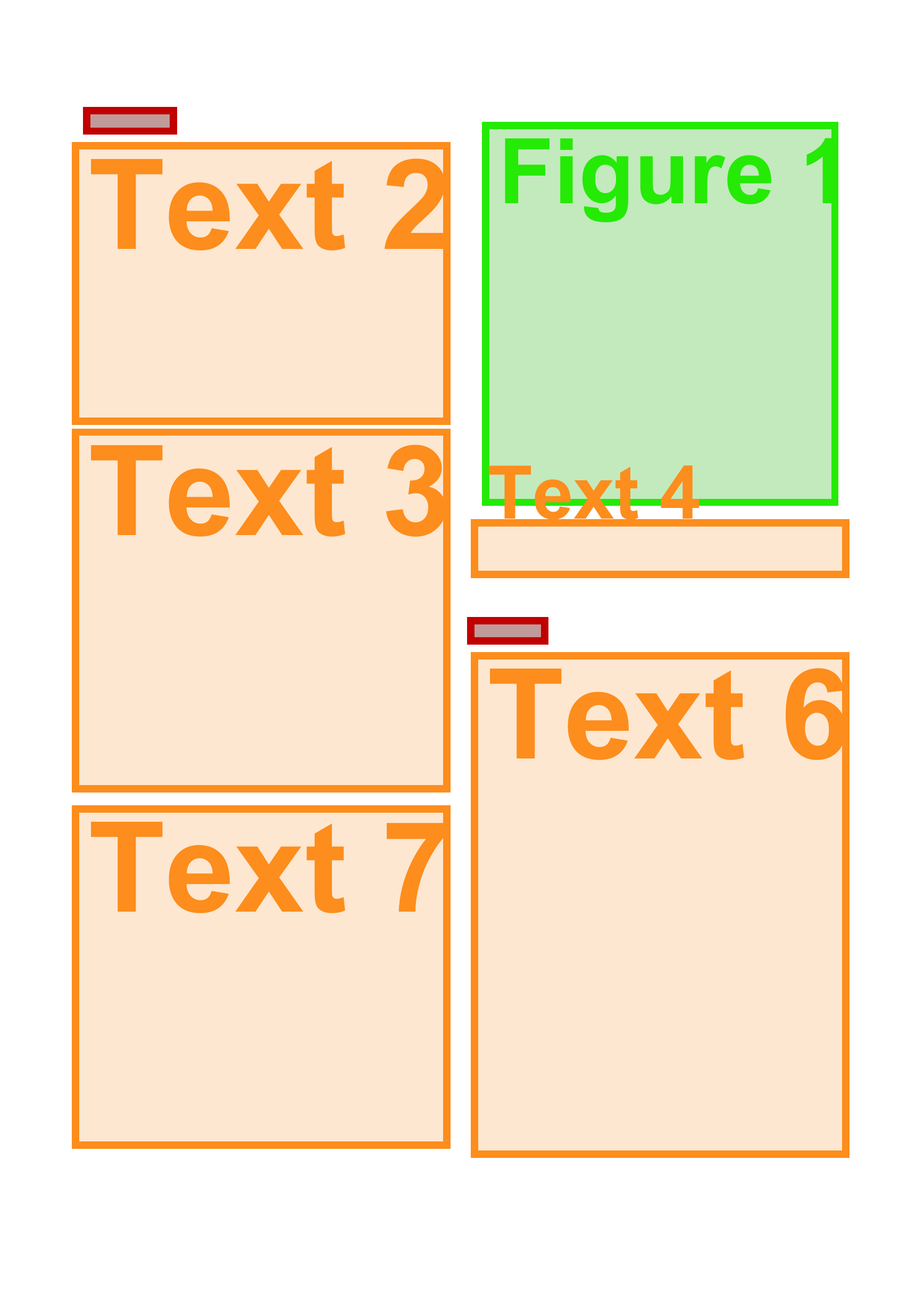} &
\includegraphics[width=\interpolationWidth,frame=0.1pt]{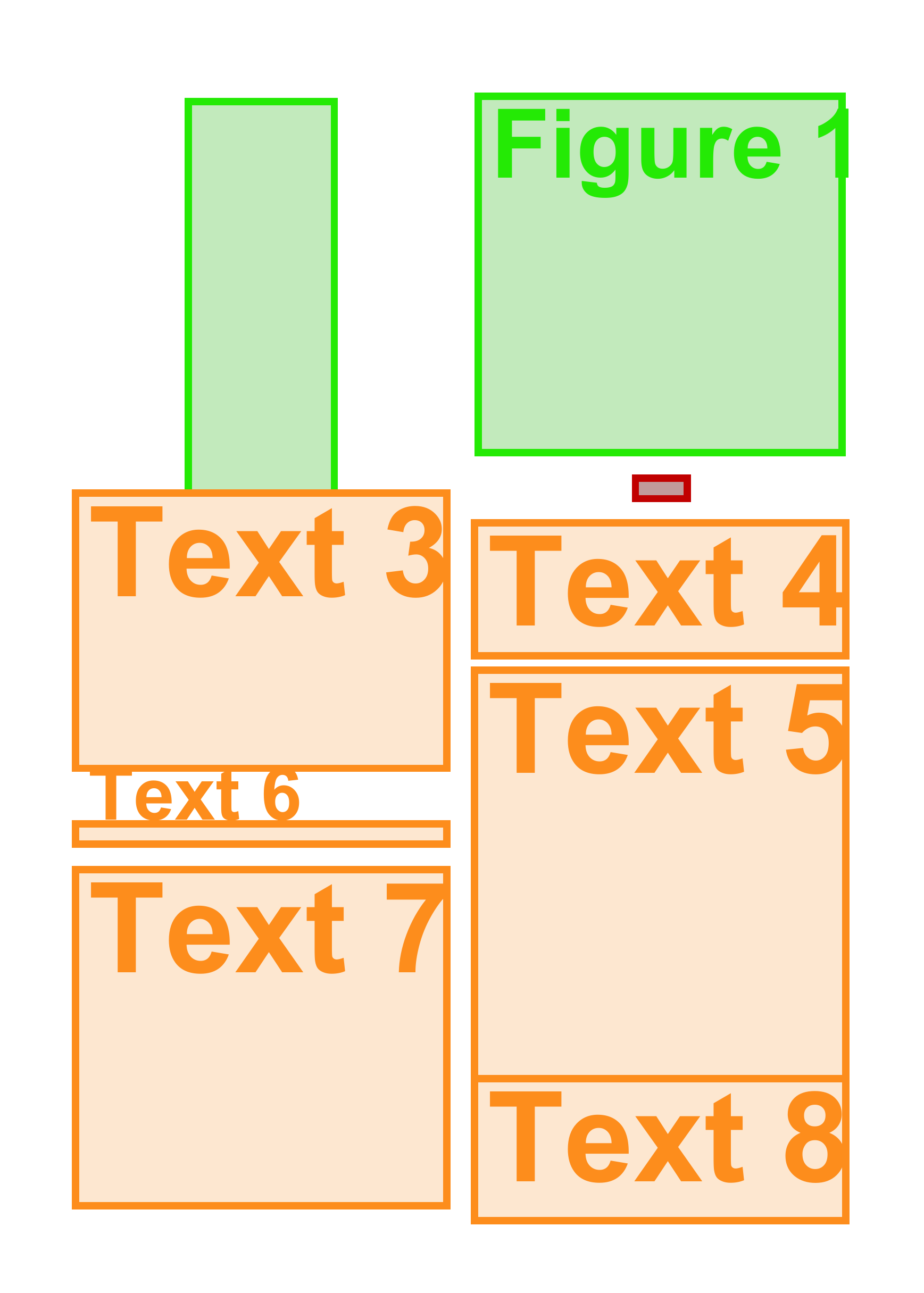} &
\includegraphics[width=\interpolationWidth,frame=0.1pt]{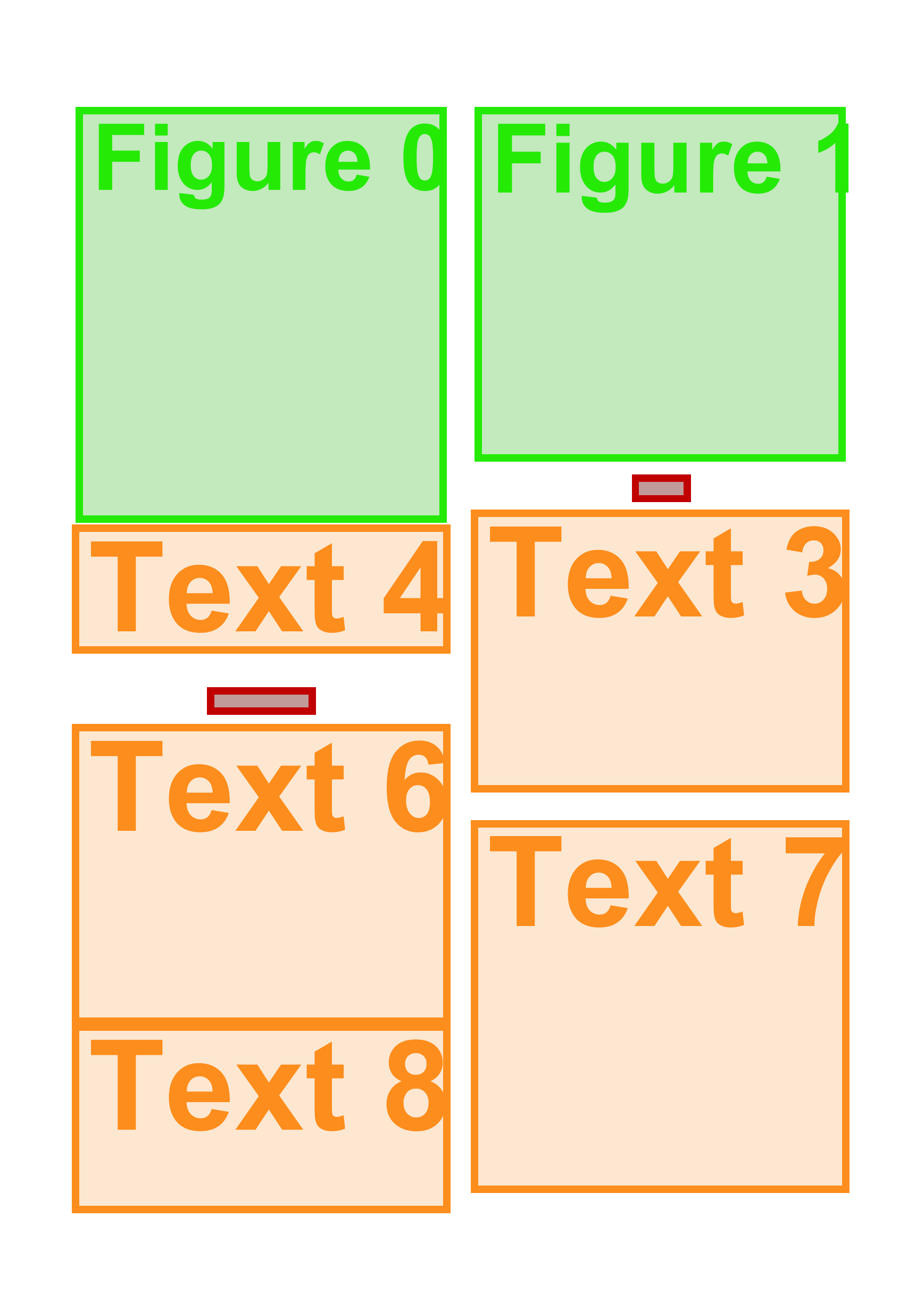} &
\includegraphics[width=\interpolationWidth,frame=0.1pt]{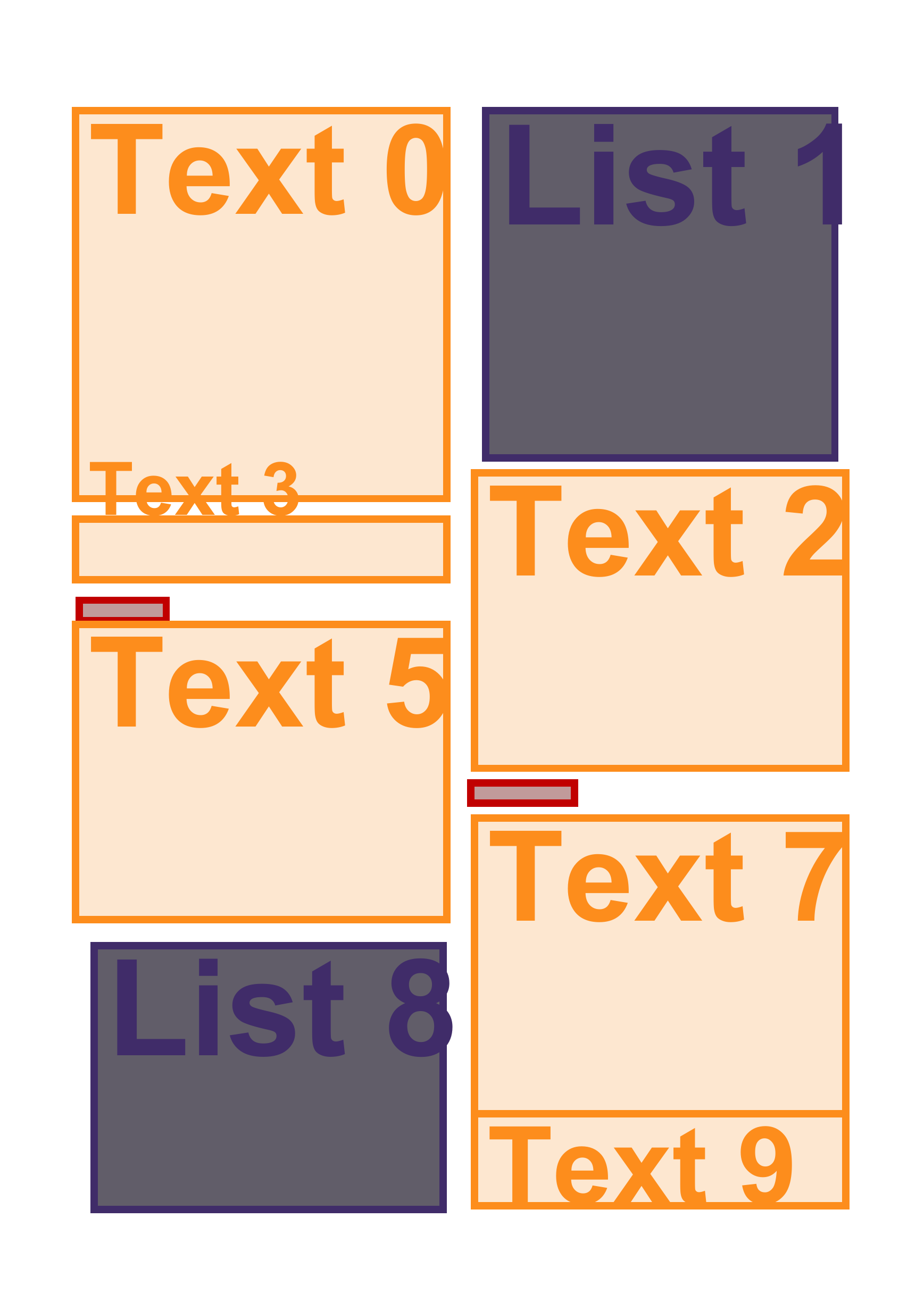} &
\includegraphics[width=\interpolationWidth,frame=0.1pt]{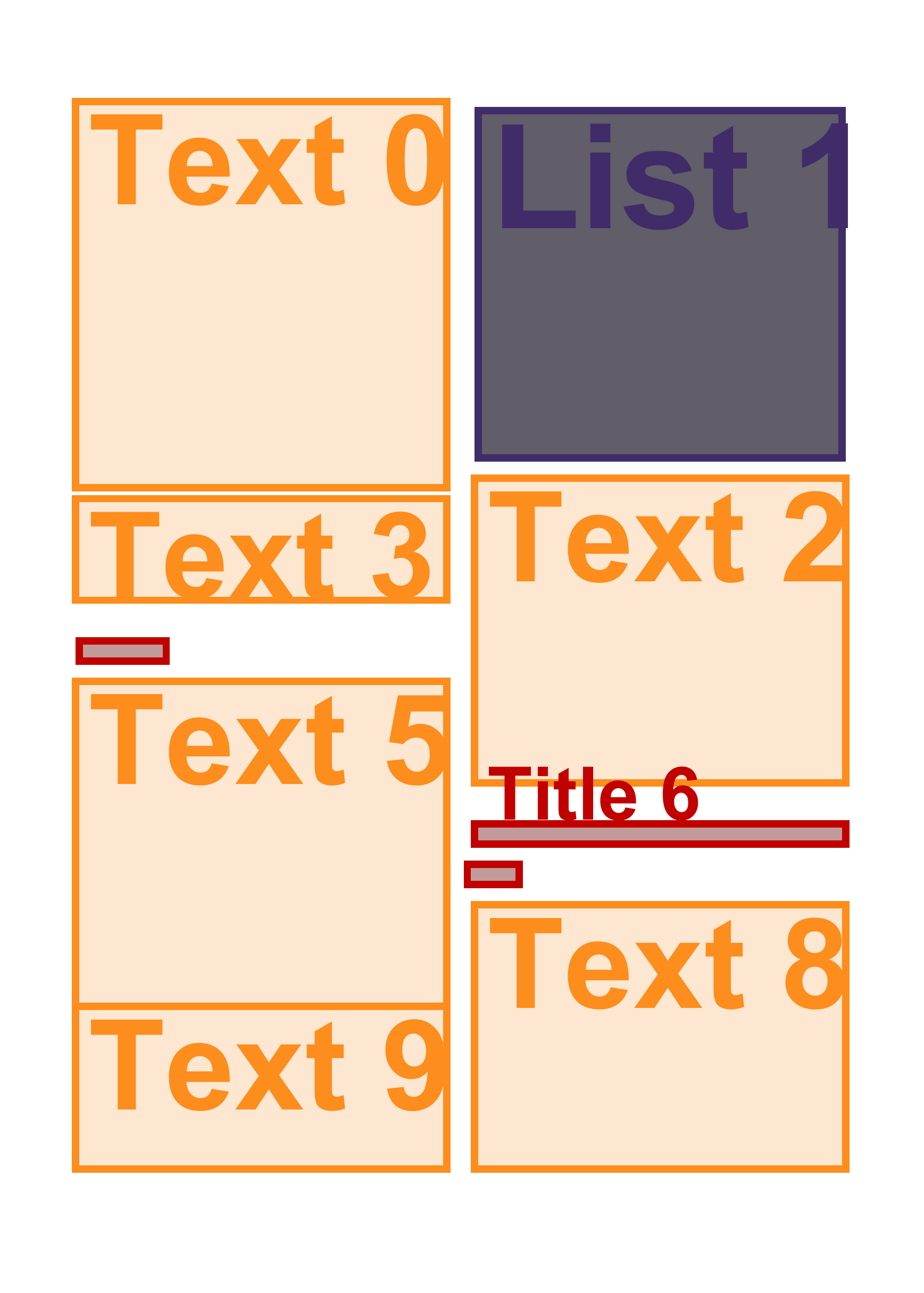} &
\includegraphics[width=\interpolationWidth,frame=0.1pt]{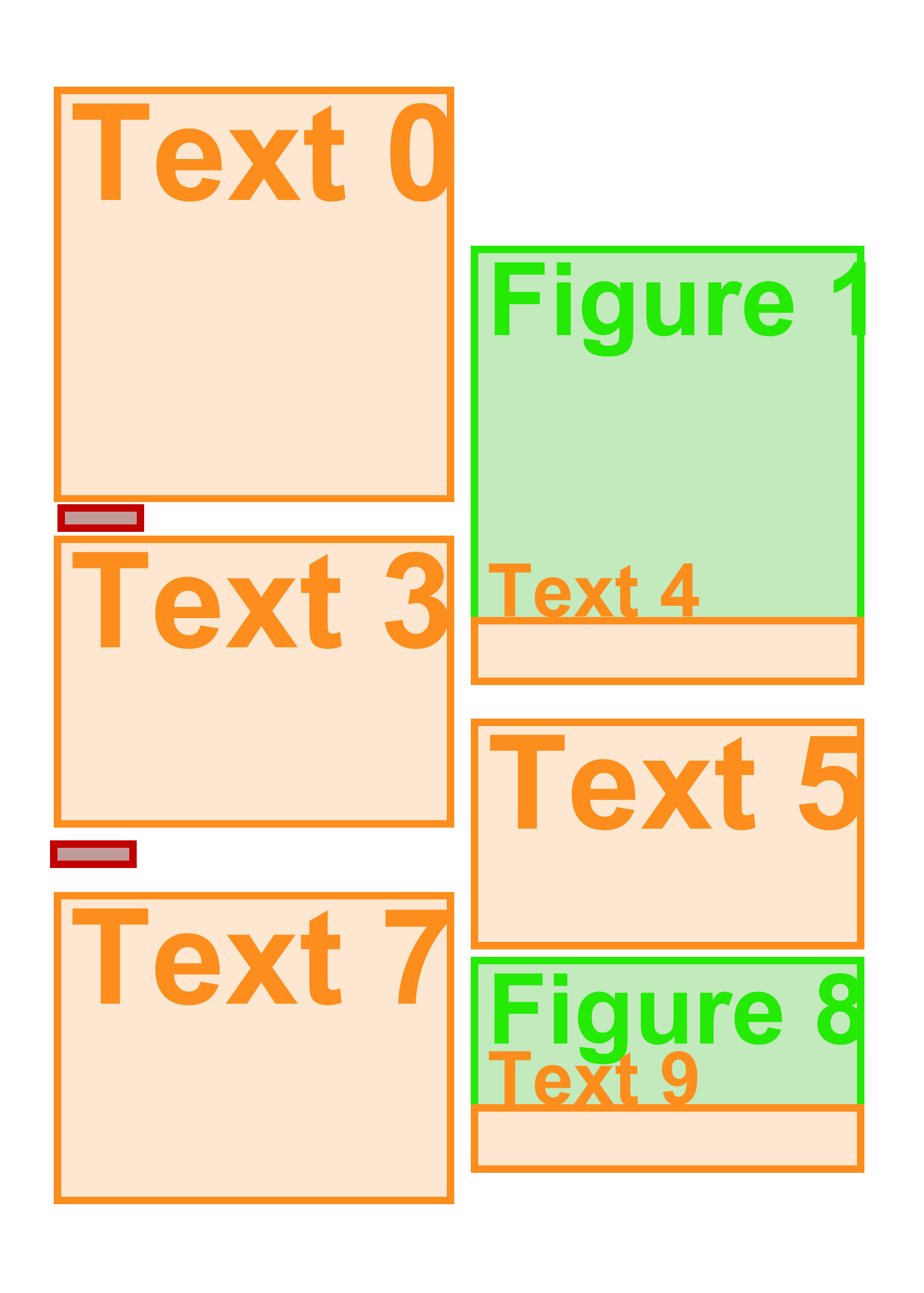} &
\includegraphics[width=\interpolationWidth,frame=0.1pt]{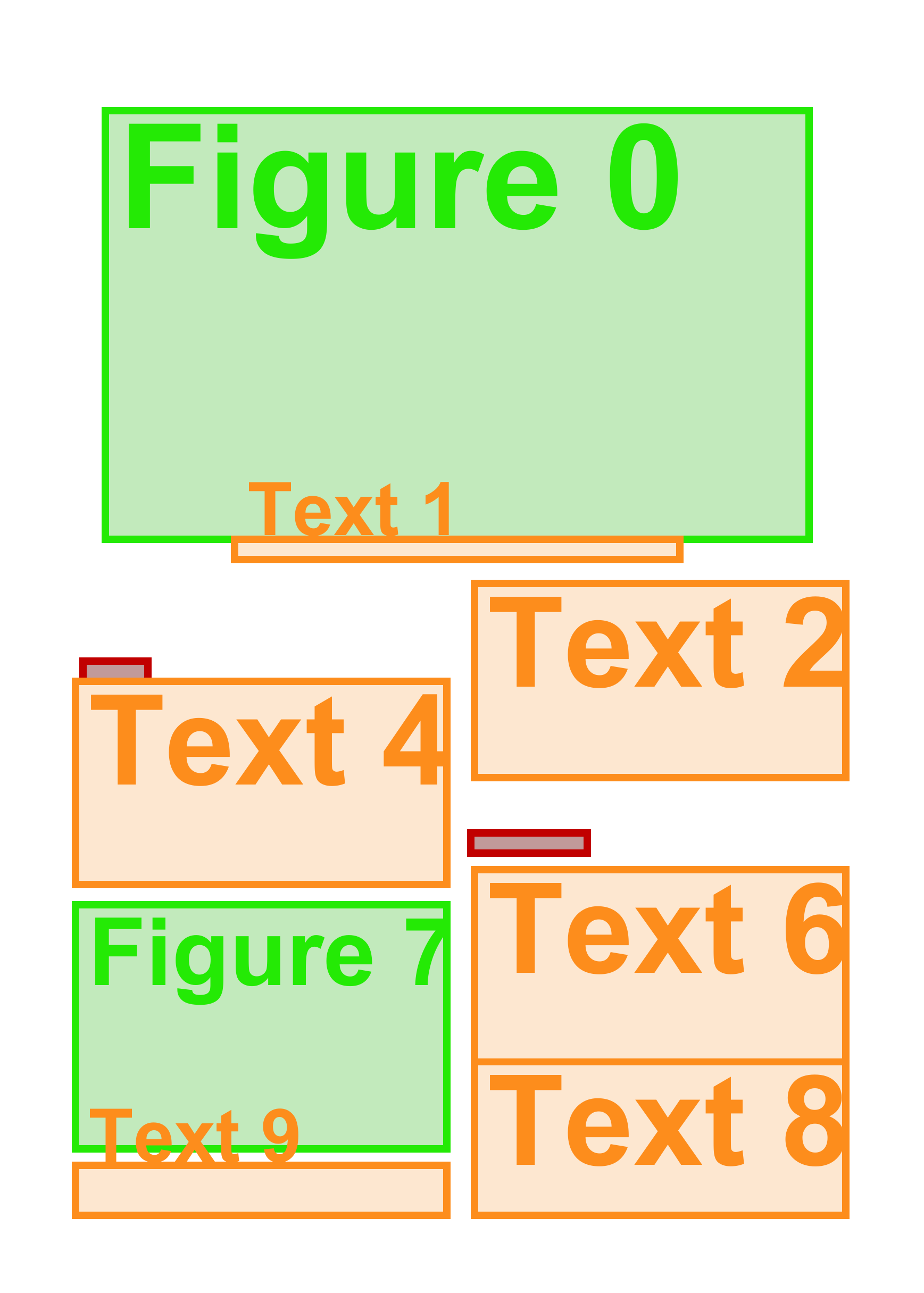} &
\includegraphics[width=\interpolationWidth,frame=0.1pt]{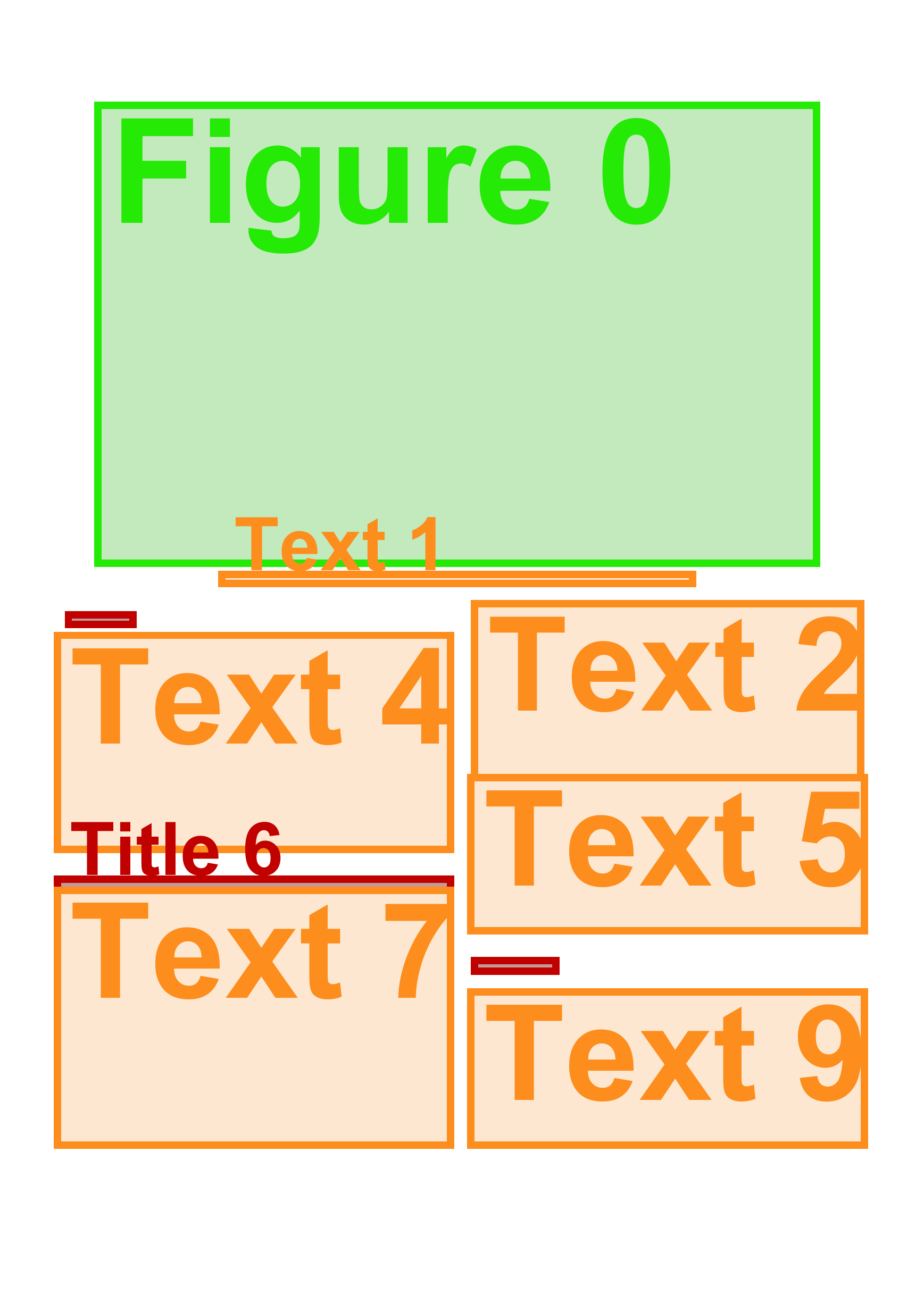} &
\includegraphics[width=\interpolationWidth,frame=0.1pt]{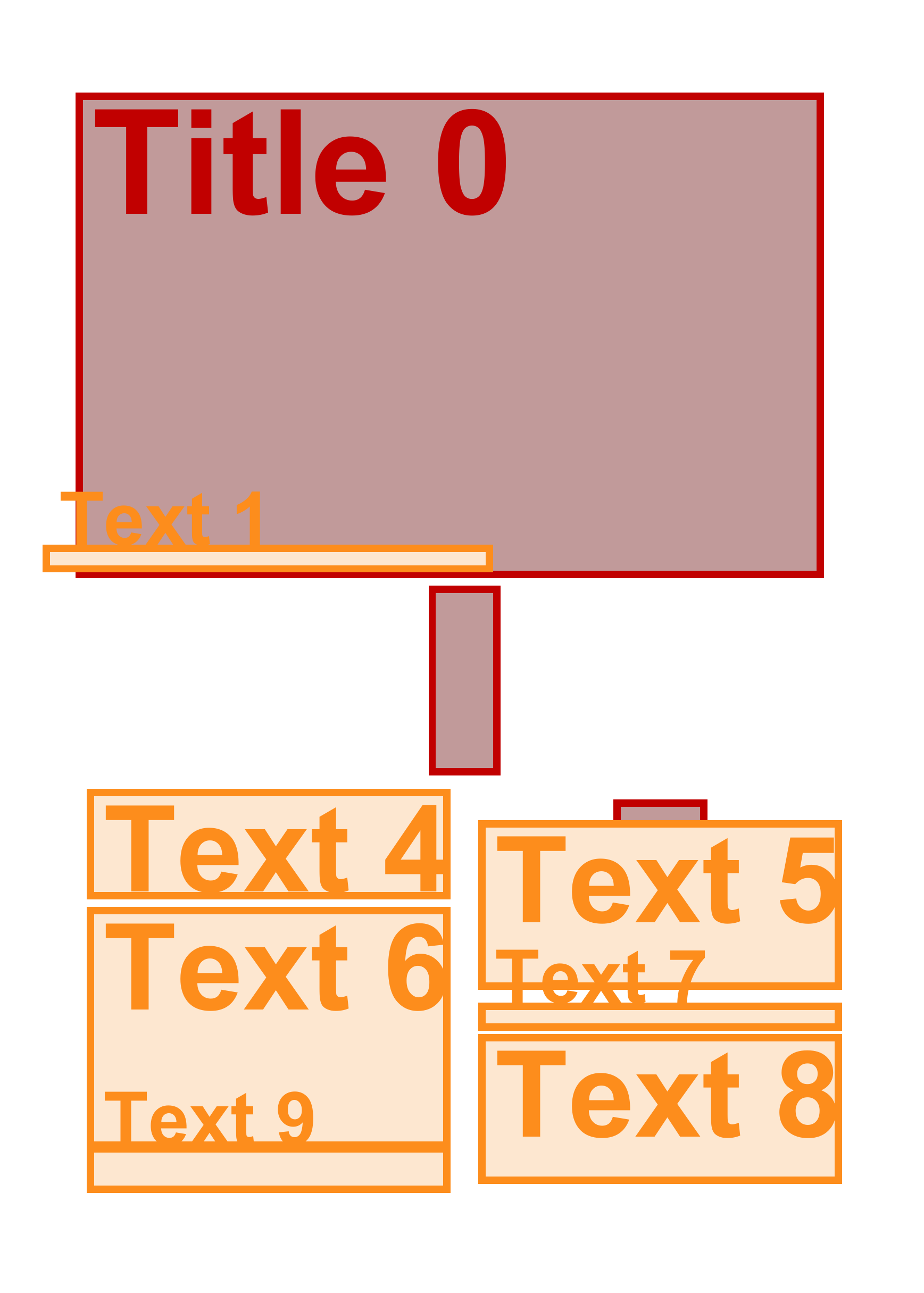} &
\includegraphics[width=\interpolationWidth,frame=0.1pt]{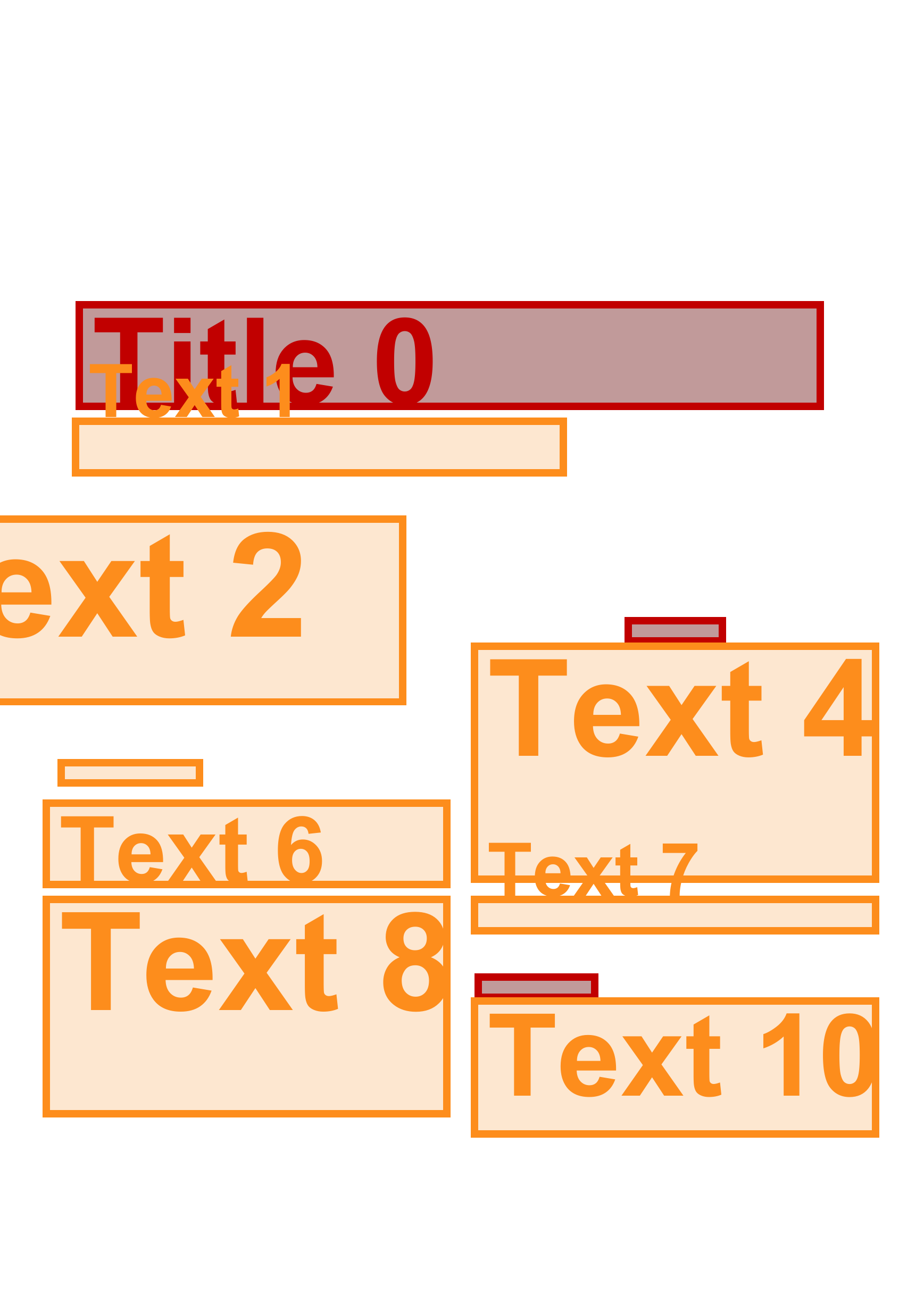} &
\includegraphics[width=\interpolationWidth,frame=0.1pt]{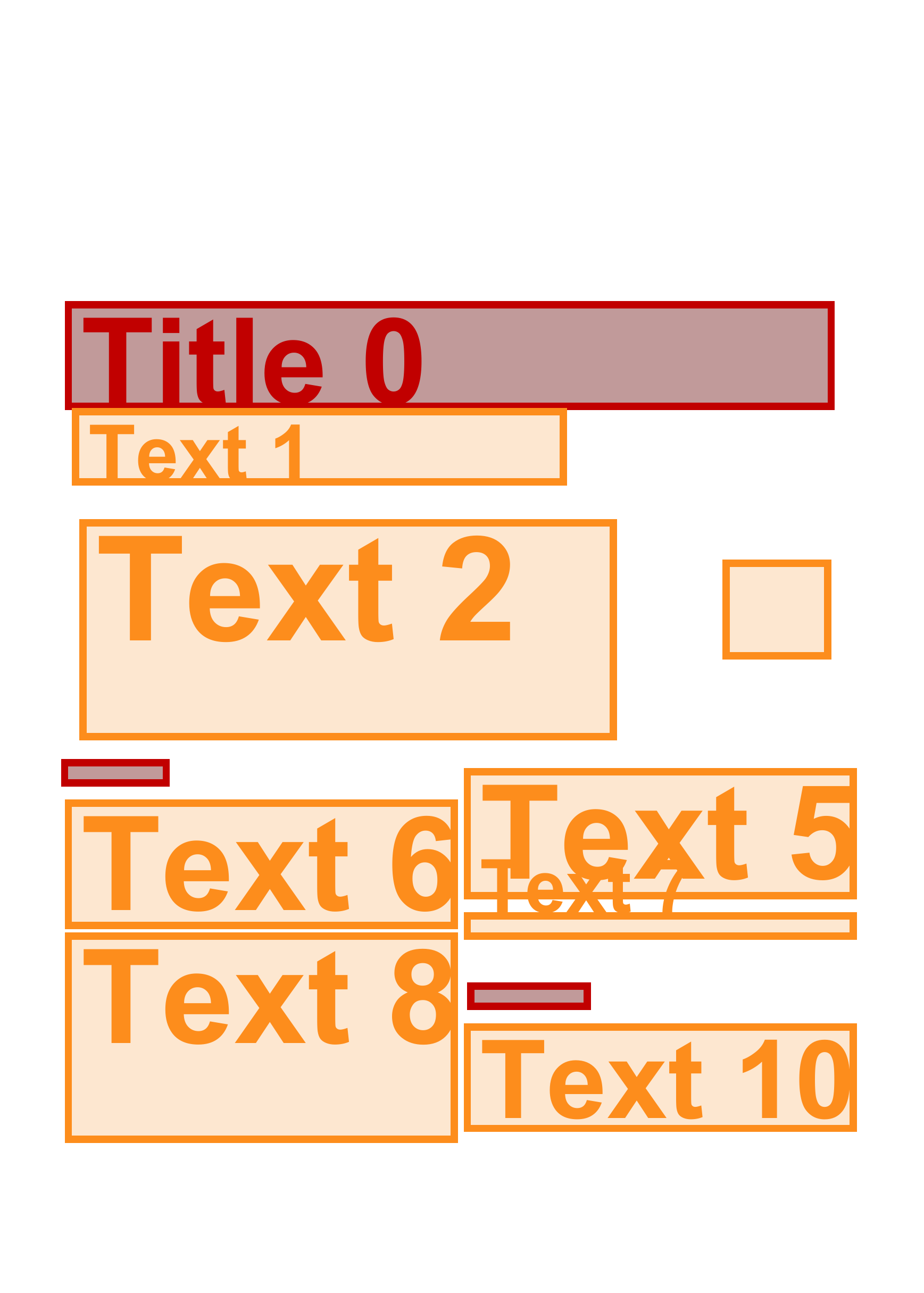} \\

\includegraphics[width=\interpolationWidth,frame=0.1pt]{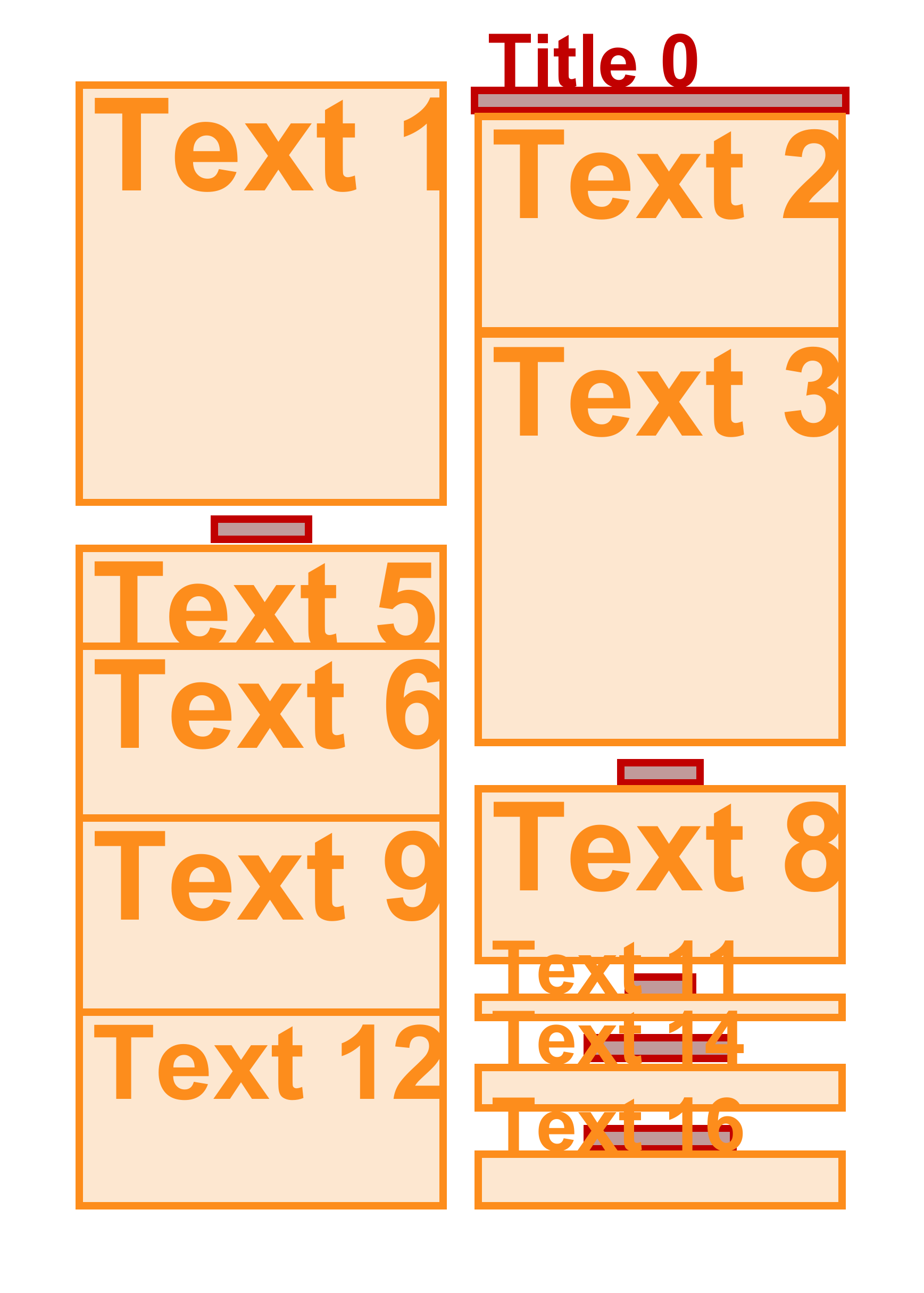} &
\includegraphics[width=\interpolationWidth,frame=0.1pt]{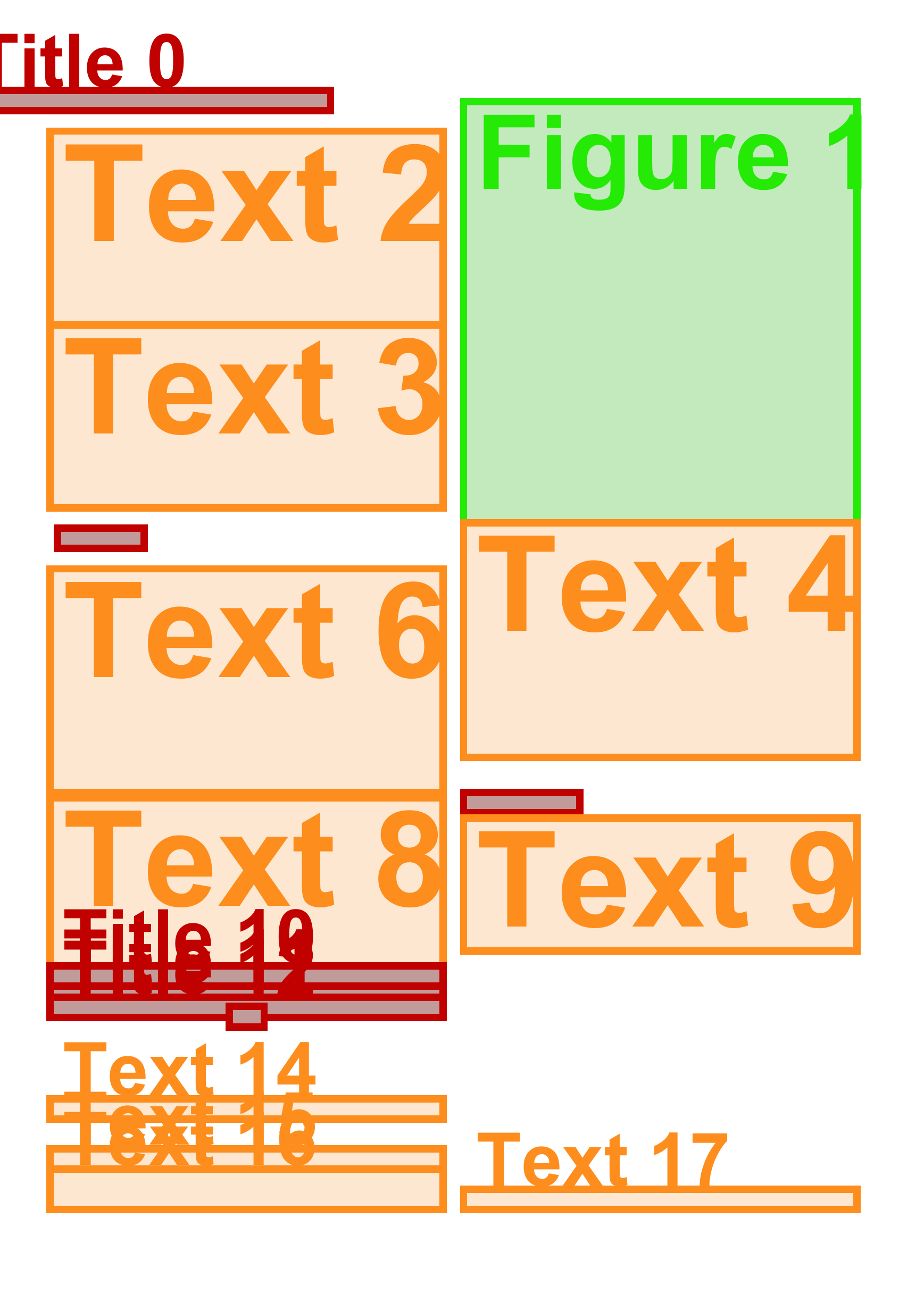} &
\includegraphics[width=\interpolationWidth,frame=0.1pt]{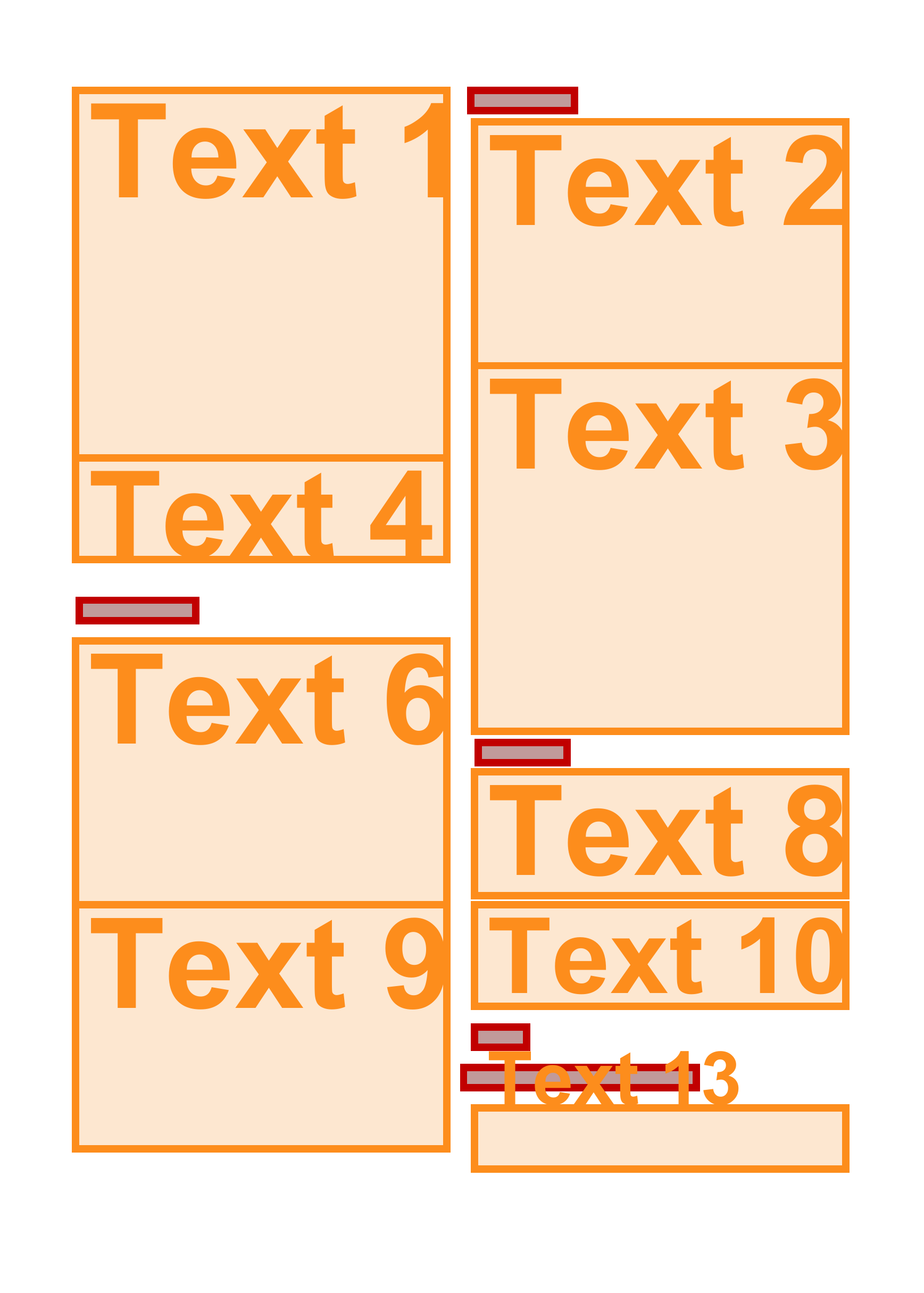} &
\includegraphics[width=\interpolationWidth,frame=0.1pt]{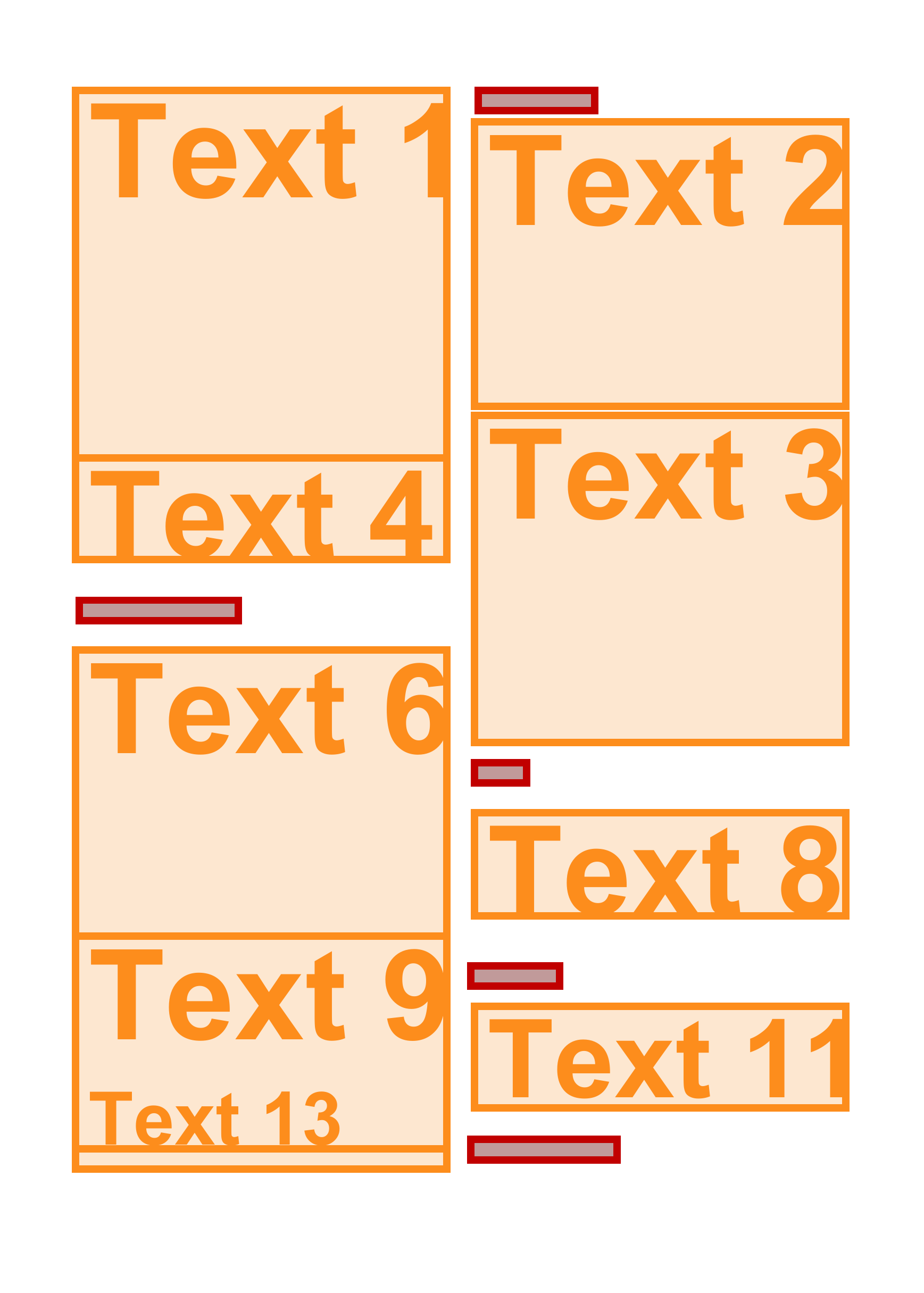} &
\includegraphics[width=\interpolationWidth,frame=0.1pt]{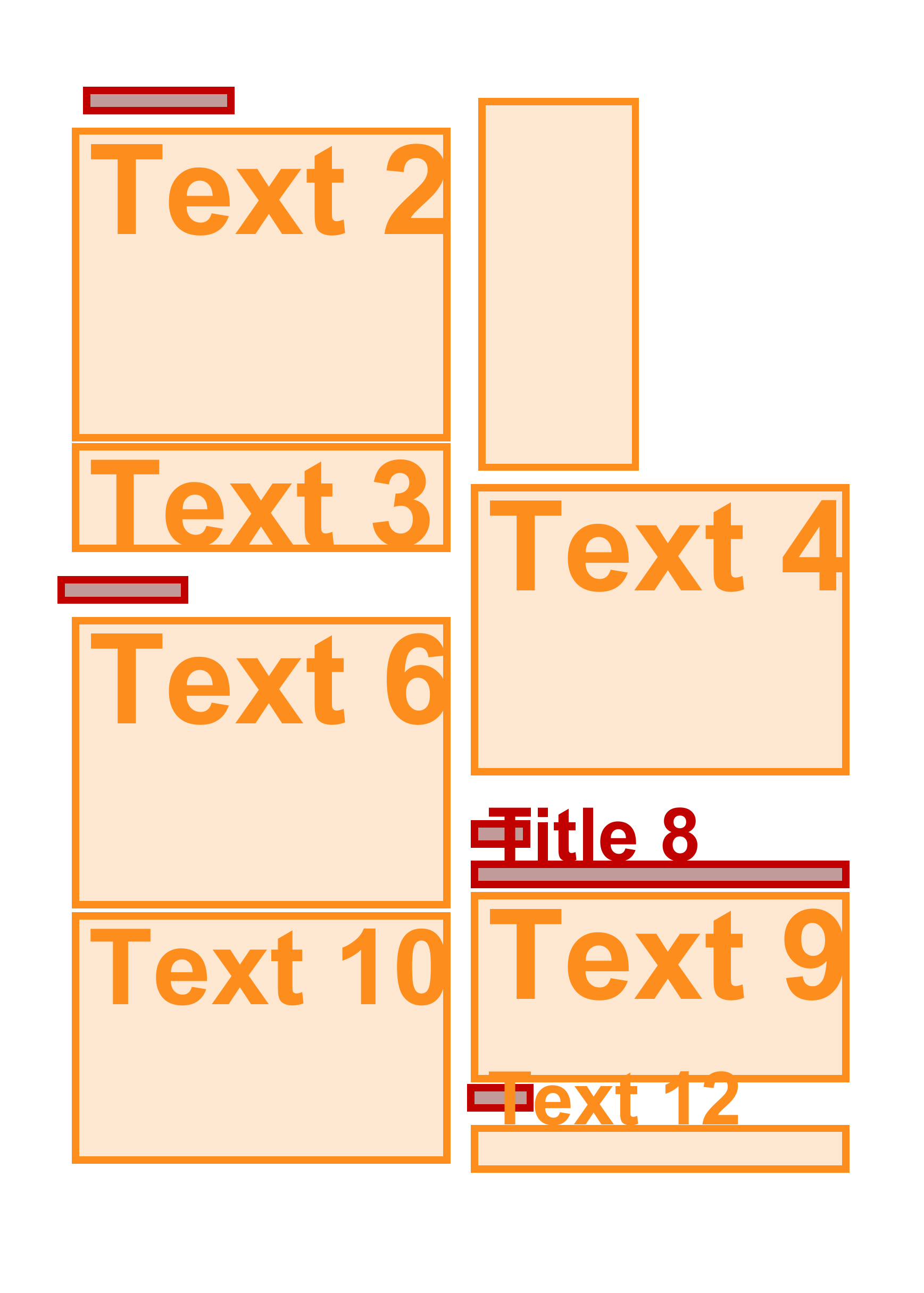} &
\includegraphics[width=\interpolationWidth,frame=0.1pt]{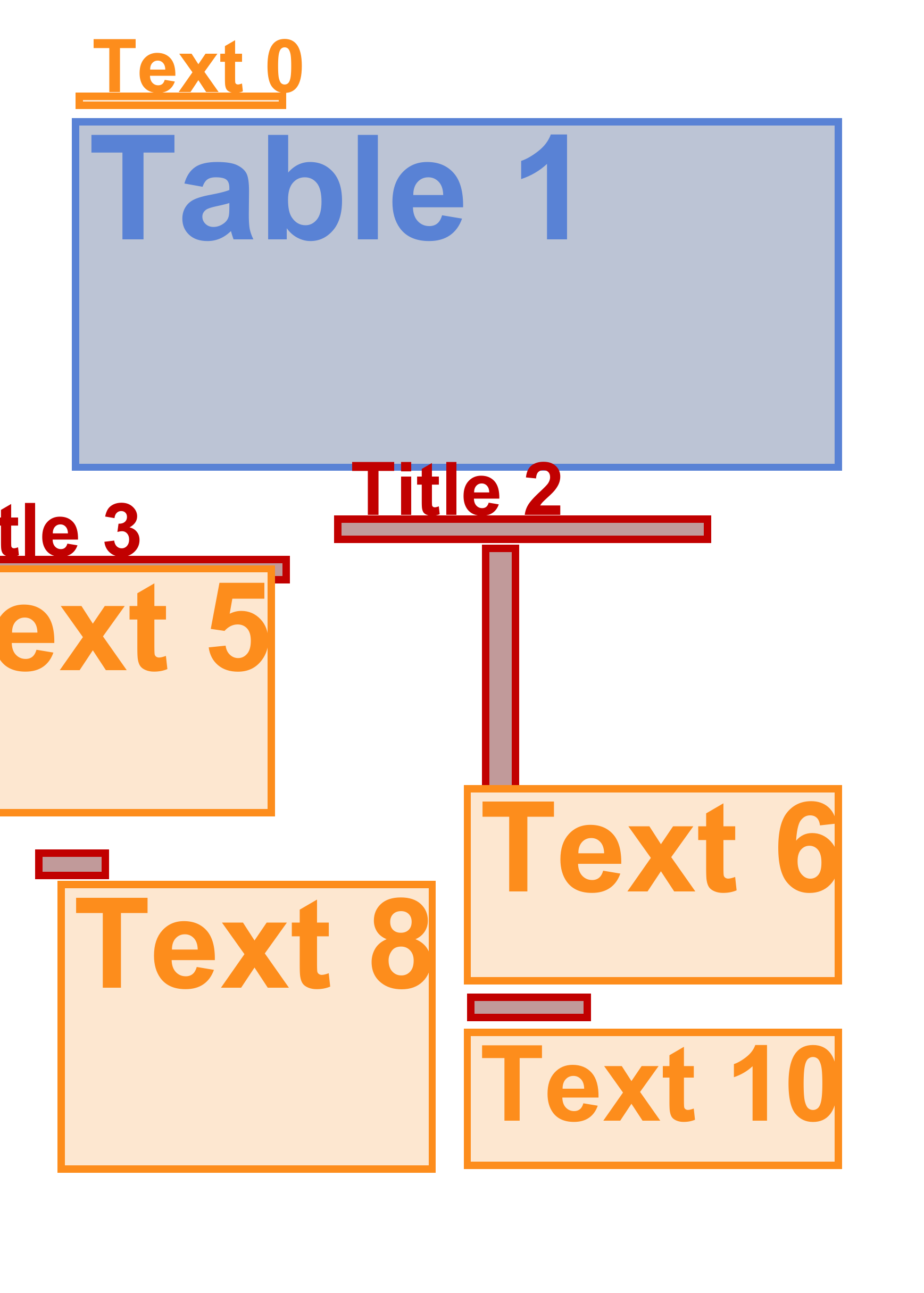} &
\includegraphics[width=\interpolationWidth,frame=0.1pt]{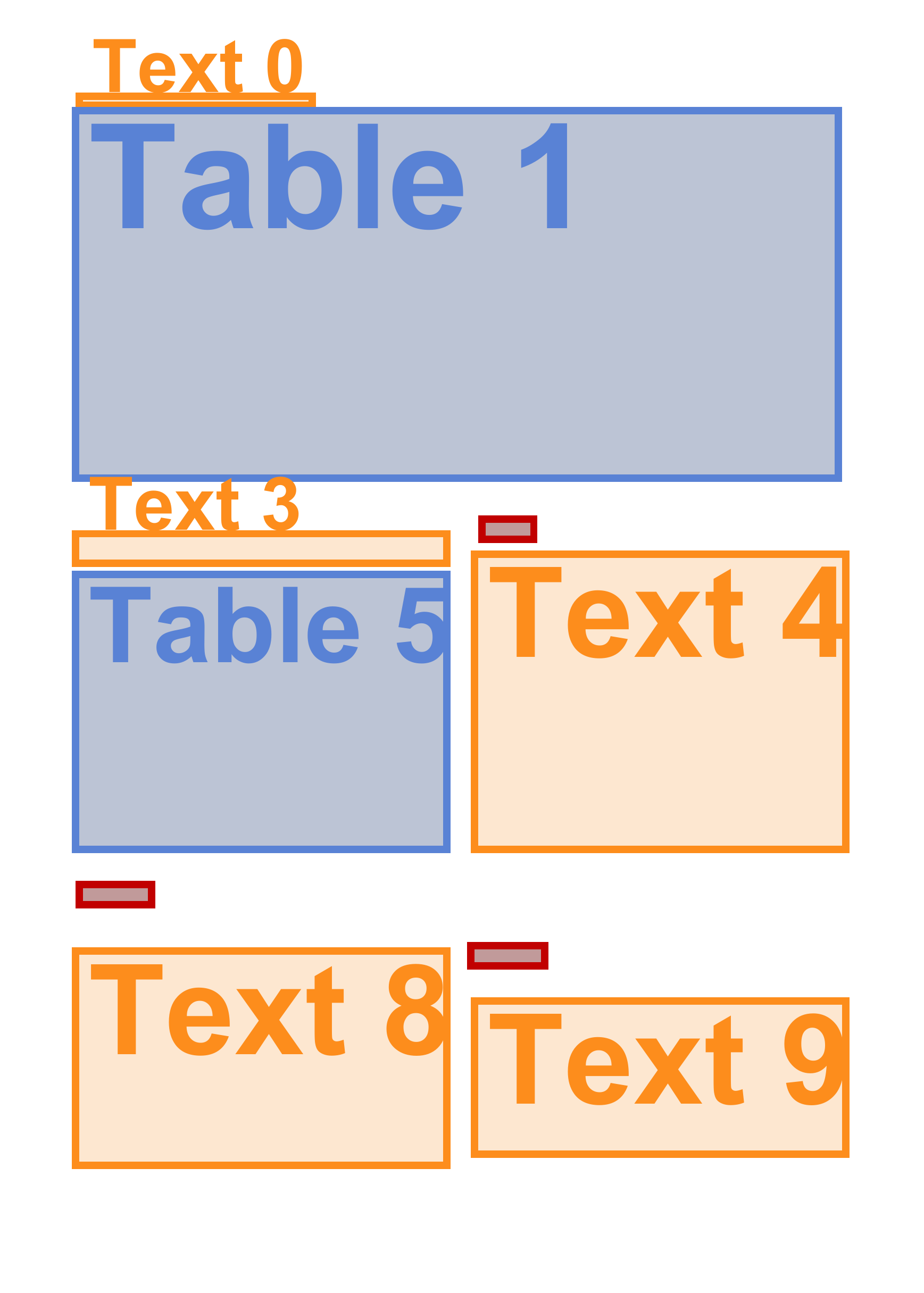} &
\includegraphics[width=\interpolationWidth,frame=0.1pt]{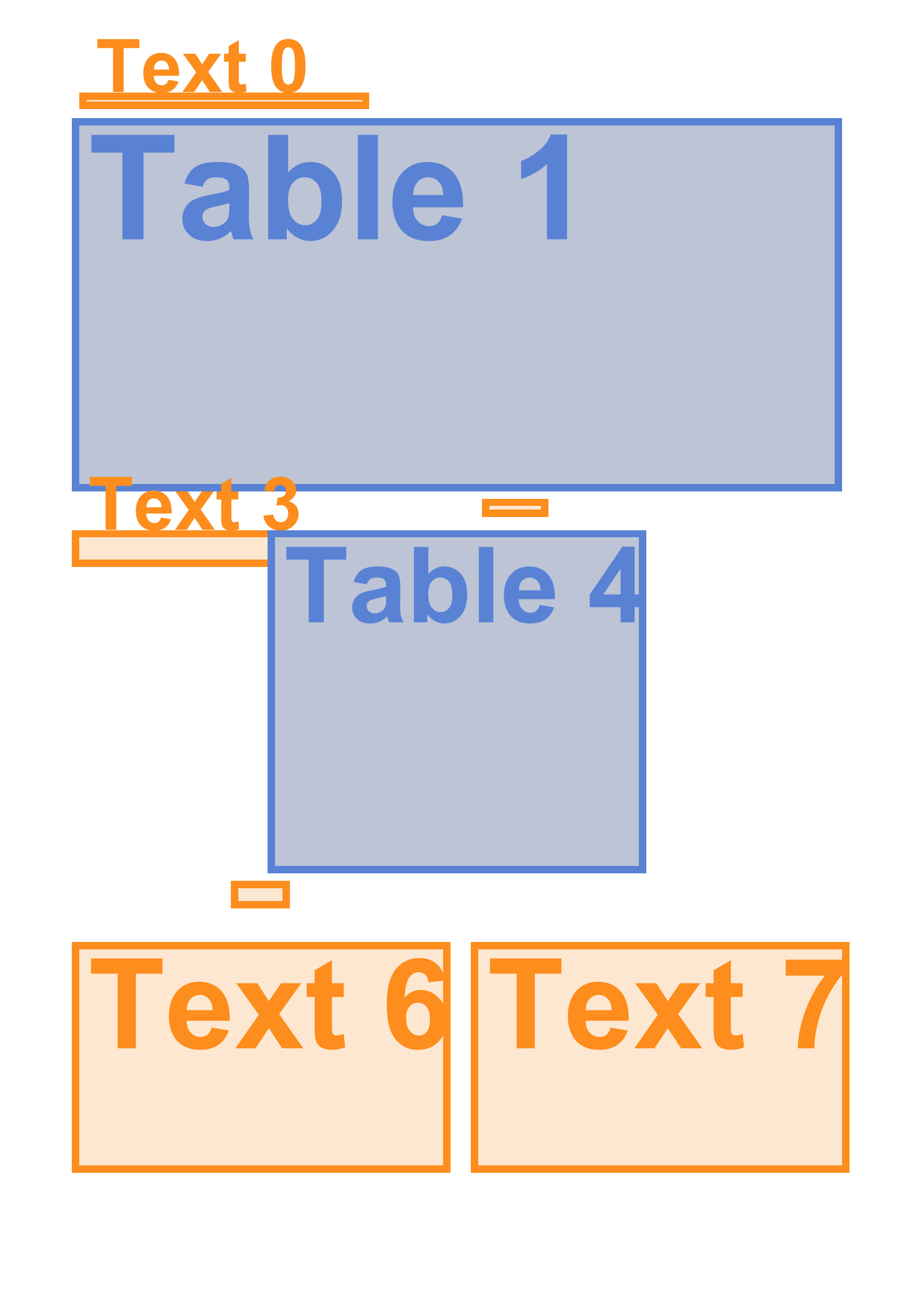} &
\includegraphics[width=\interpolationWidth,frame=0.1pt]{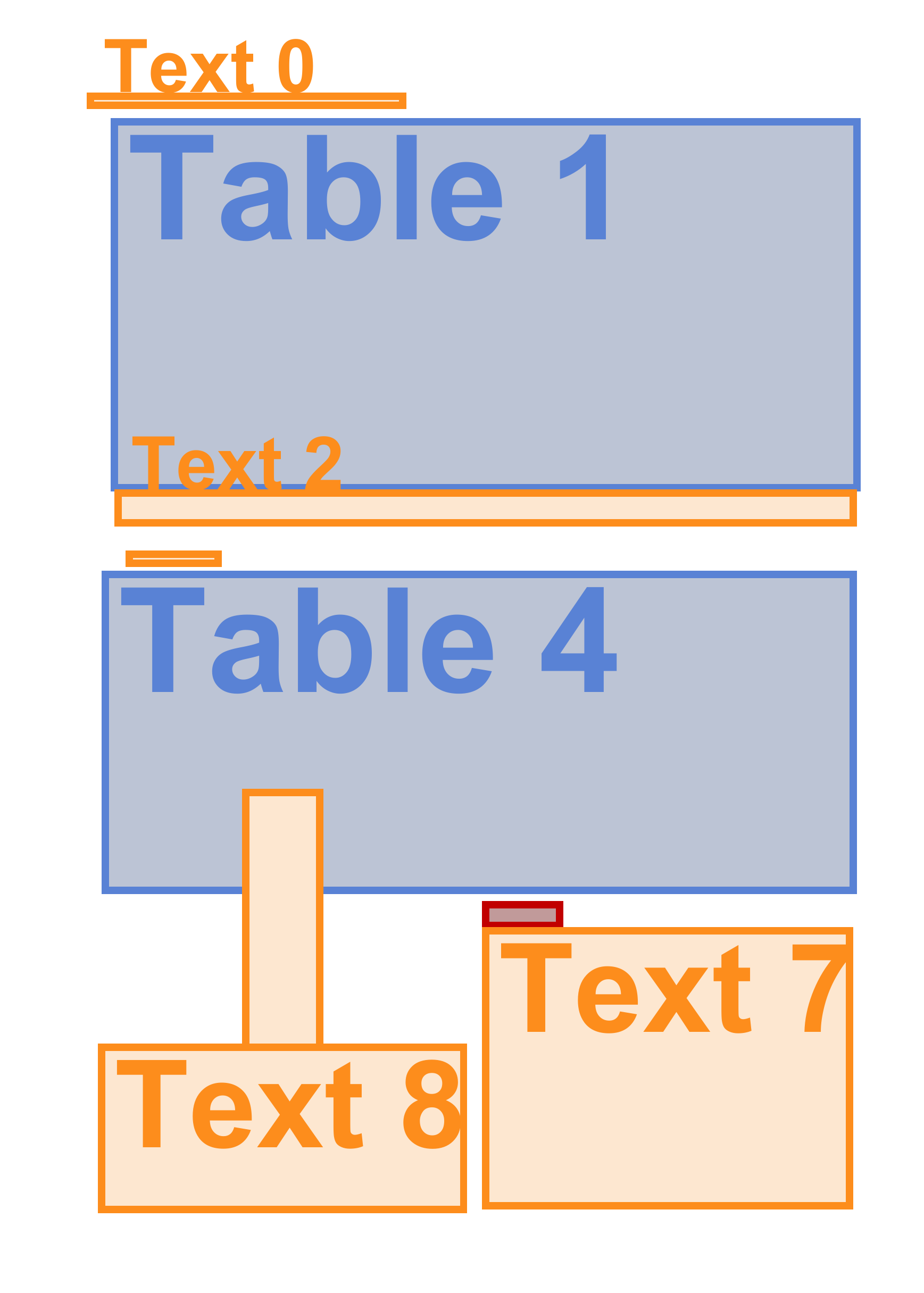} &
\includegraphics[width=\interpolationWidth,frame=0.1pt]{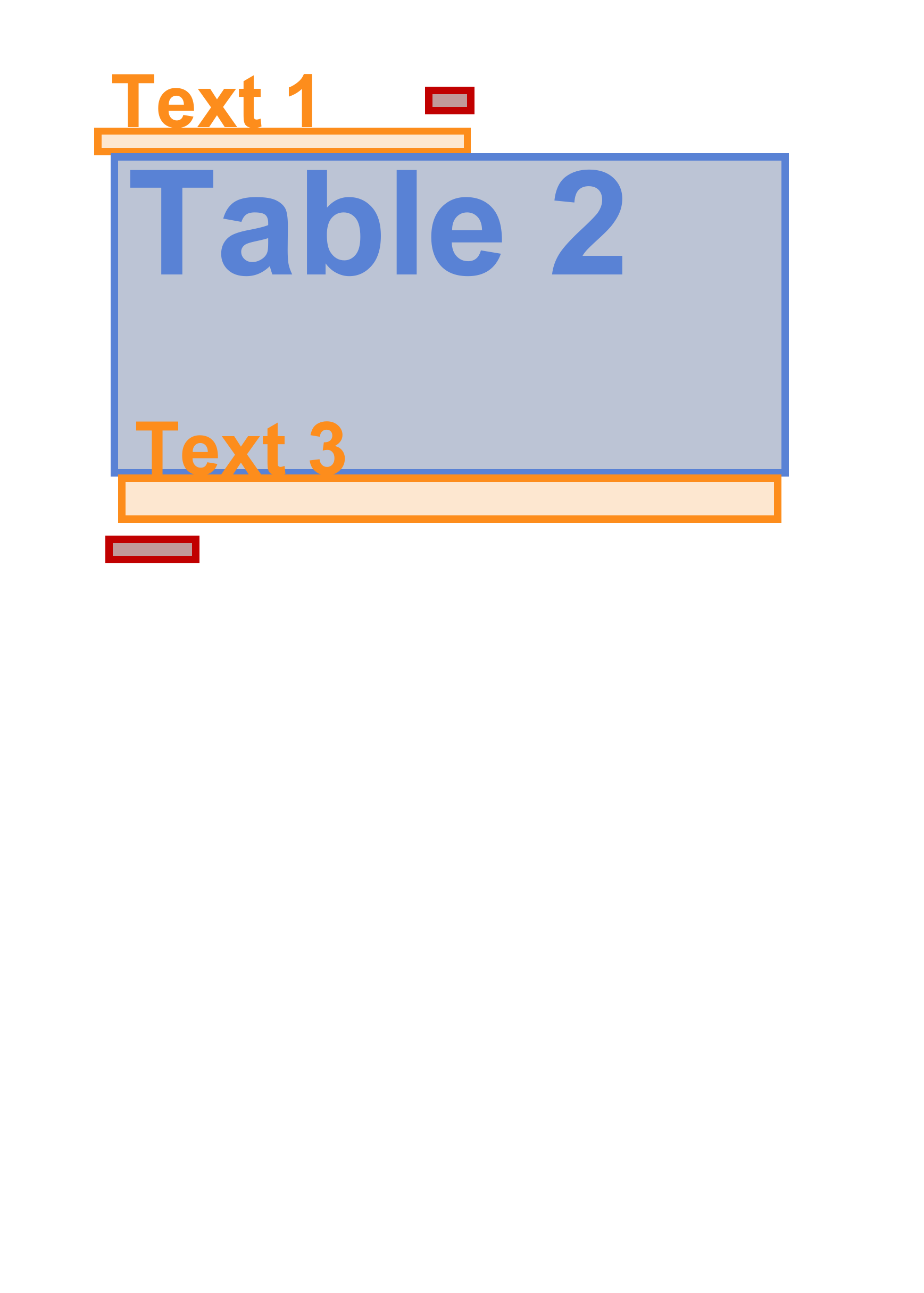} &
\includegraphics[width=\interpolationWidth,frame=0.1pt]{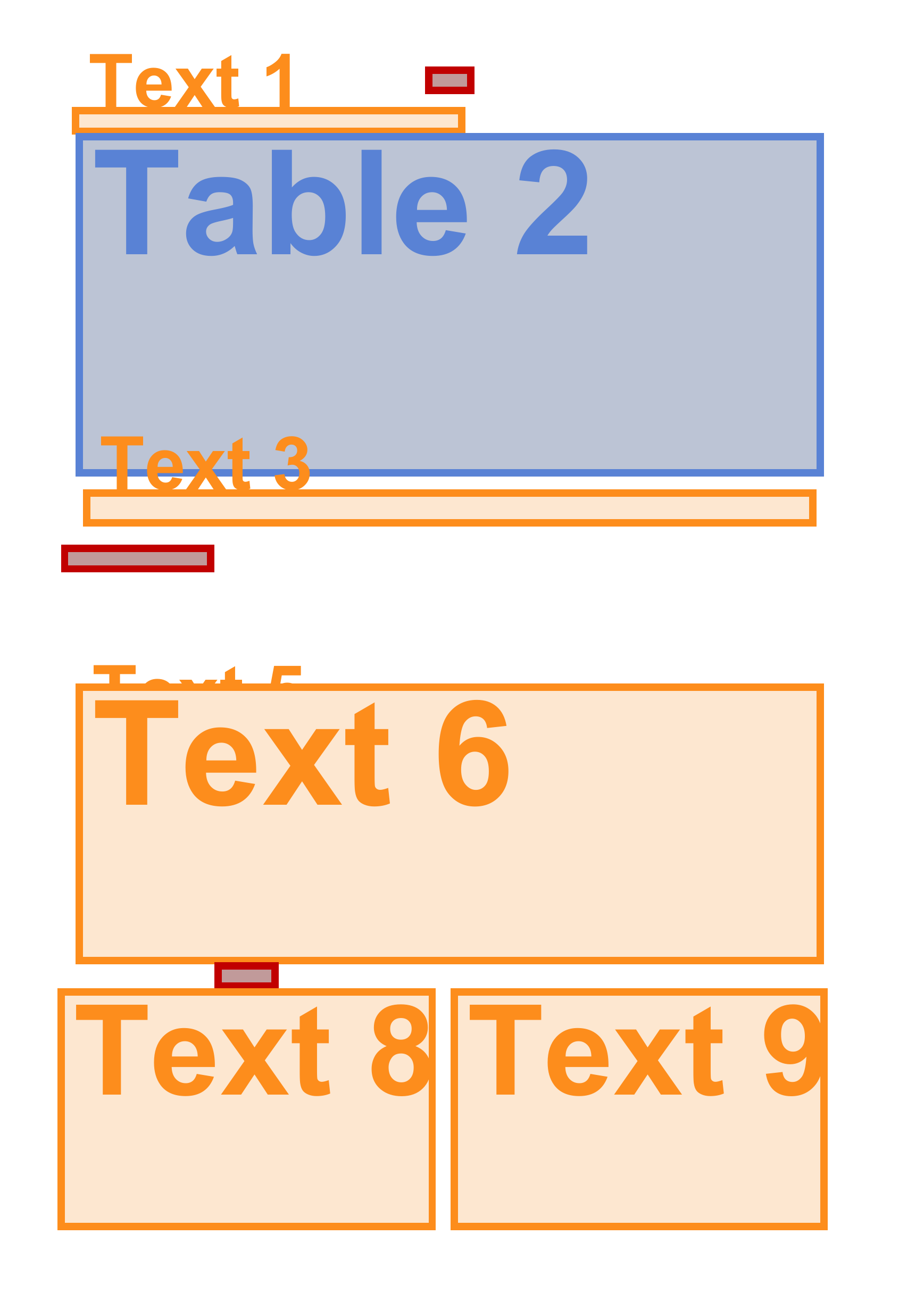} \\

\includegraphics[width=\interpolationWidth,frame=0.1pt]{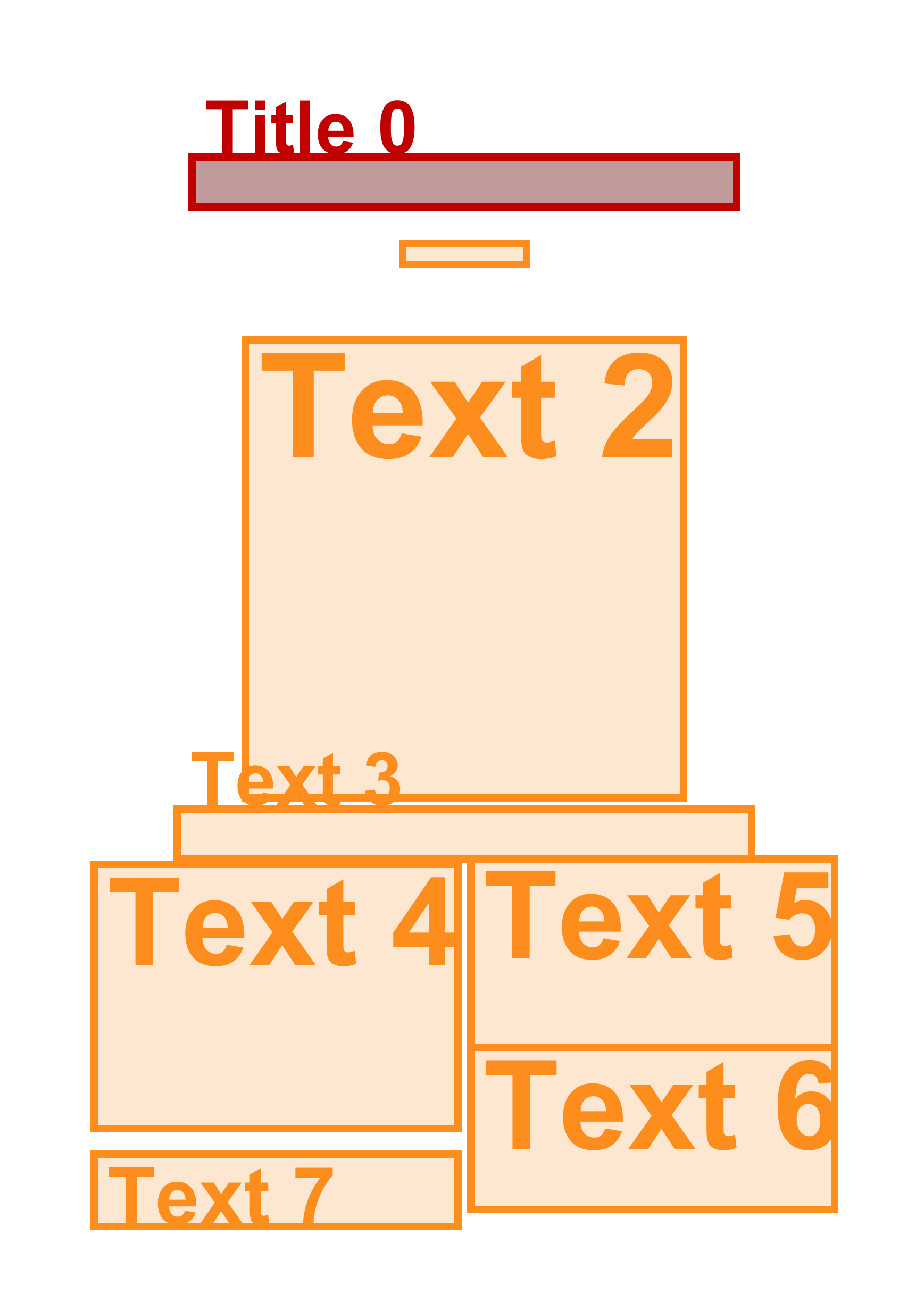} &
\includegraphics[width=\interpolationWidth,frame=0.1pt]{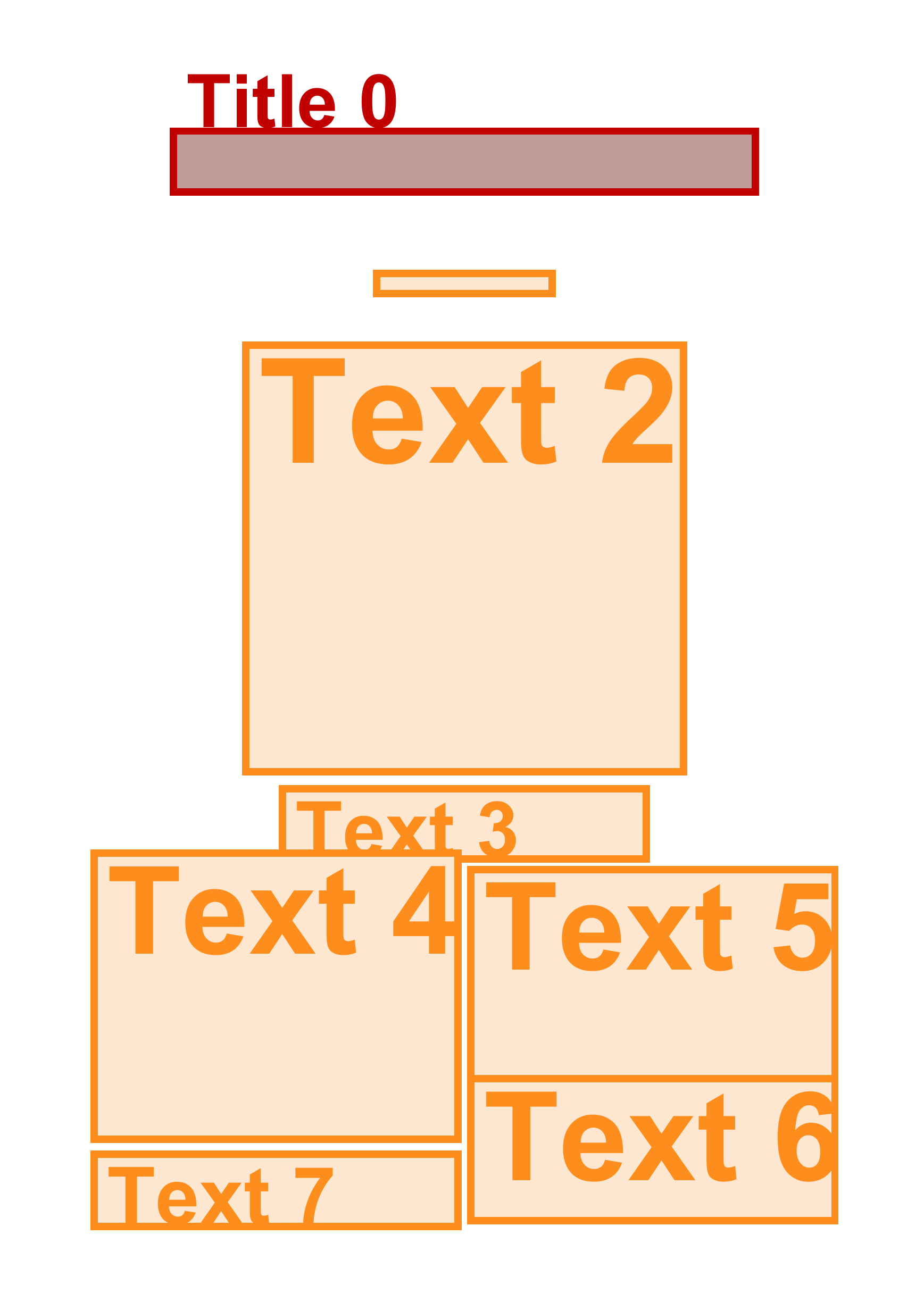} &
\includegraphics[width=\interpolationWidth,frame=0.1pt]{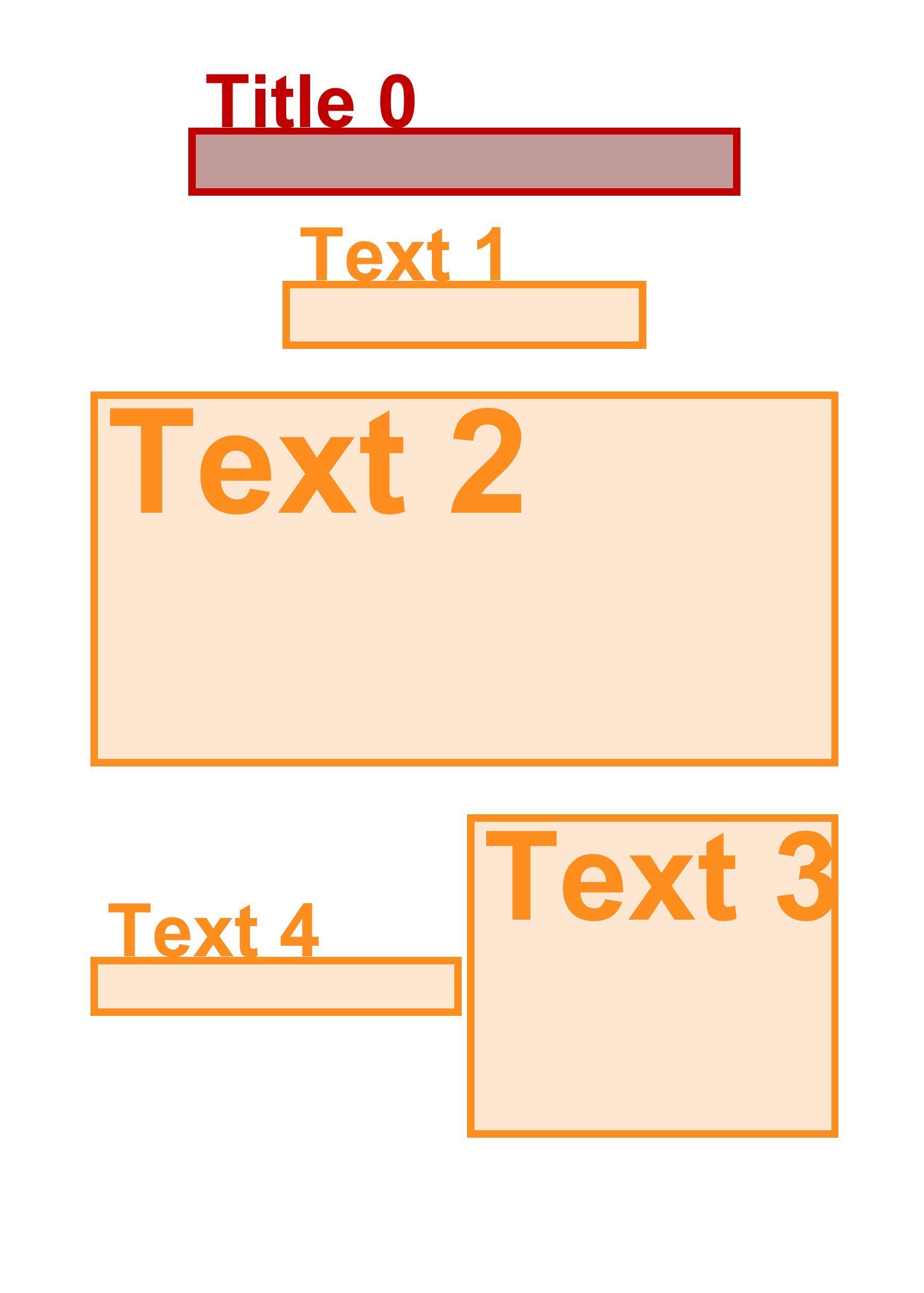} &
\includegraphics[width=\interpolationWidth,frame=0.1pt]{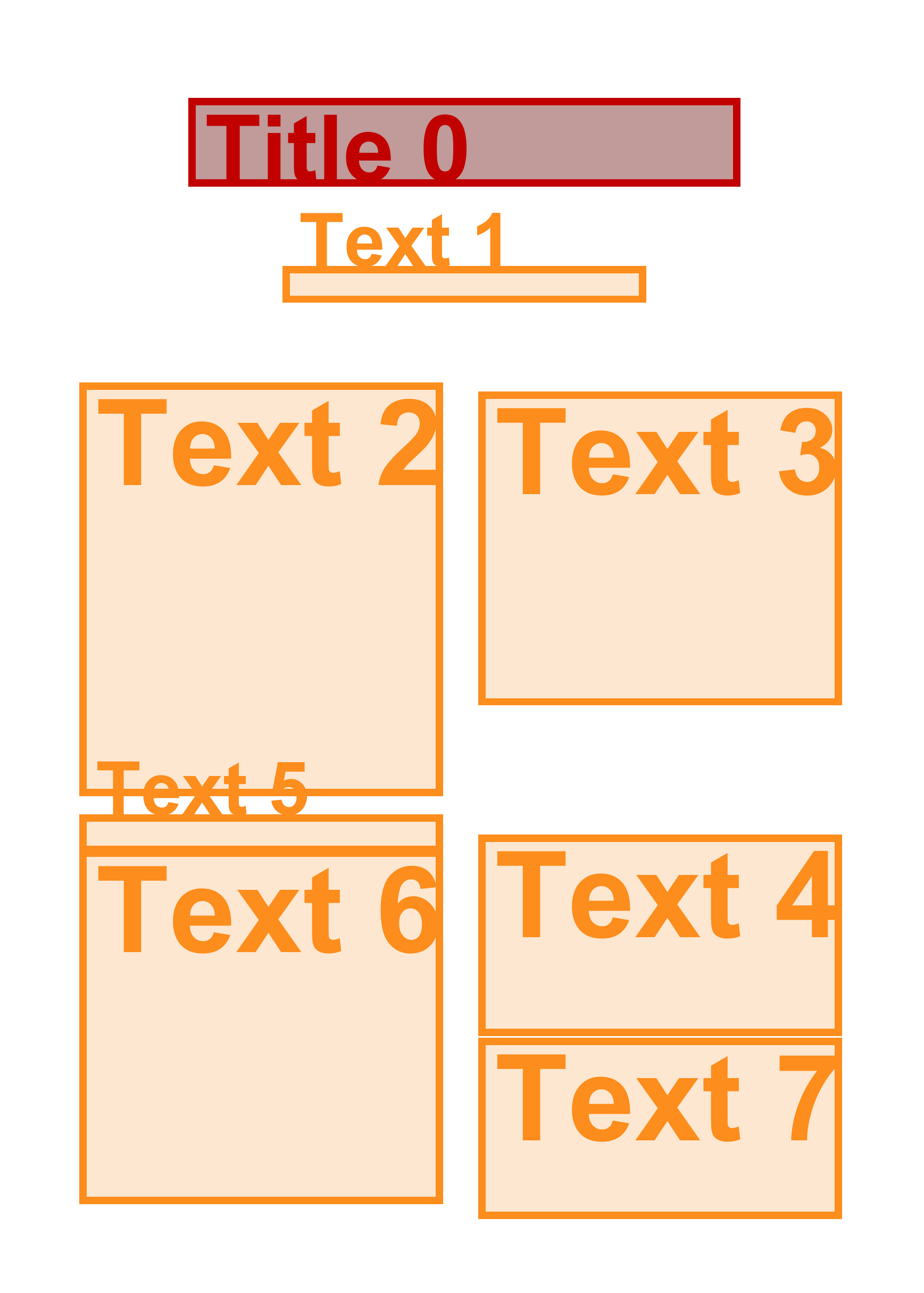} &
\includegraphics[width=\interpolationWidth,frame=0.1pt]{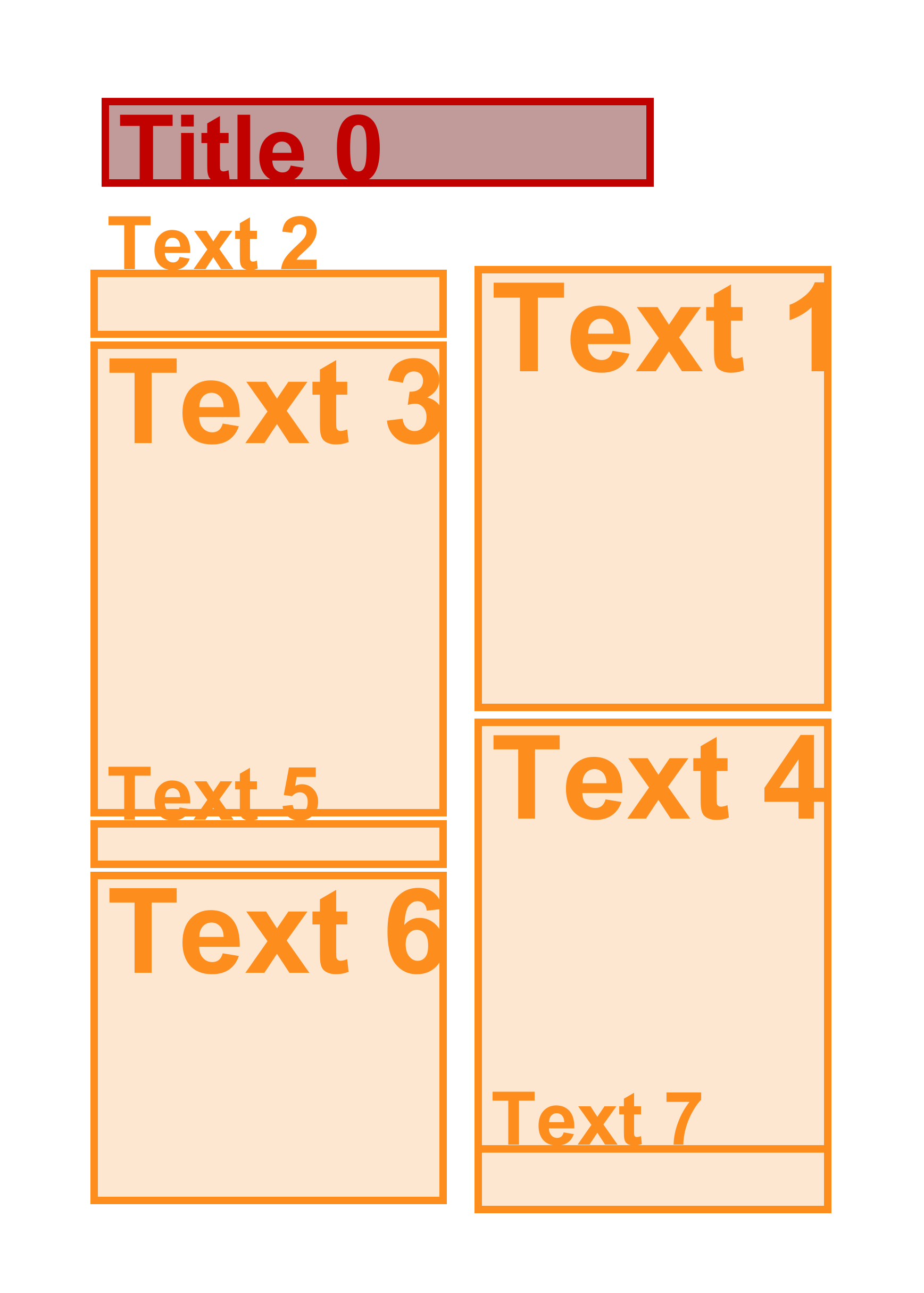} &
\includegraphics[width=\interpolationWidth,frame=0.1pt]{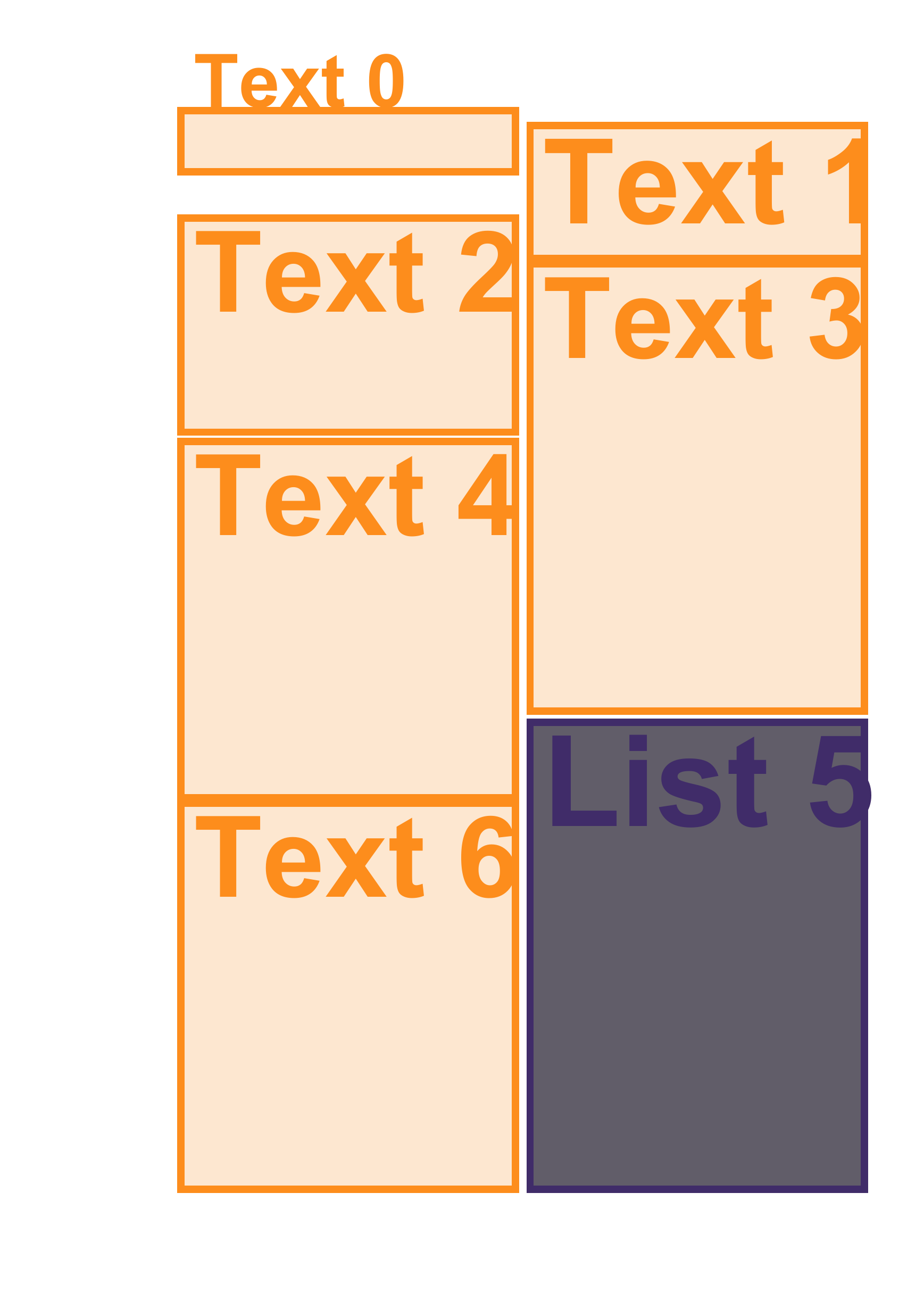} &
\includegraphics[width=\interpolationWidth,frame=0.1pt]{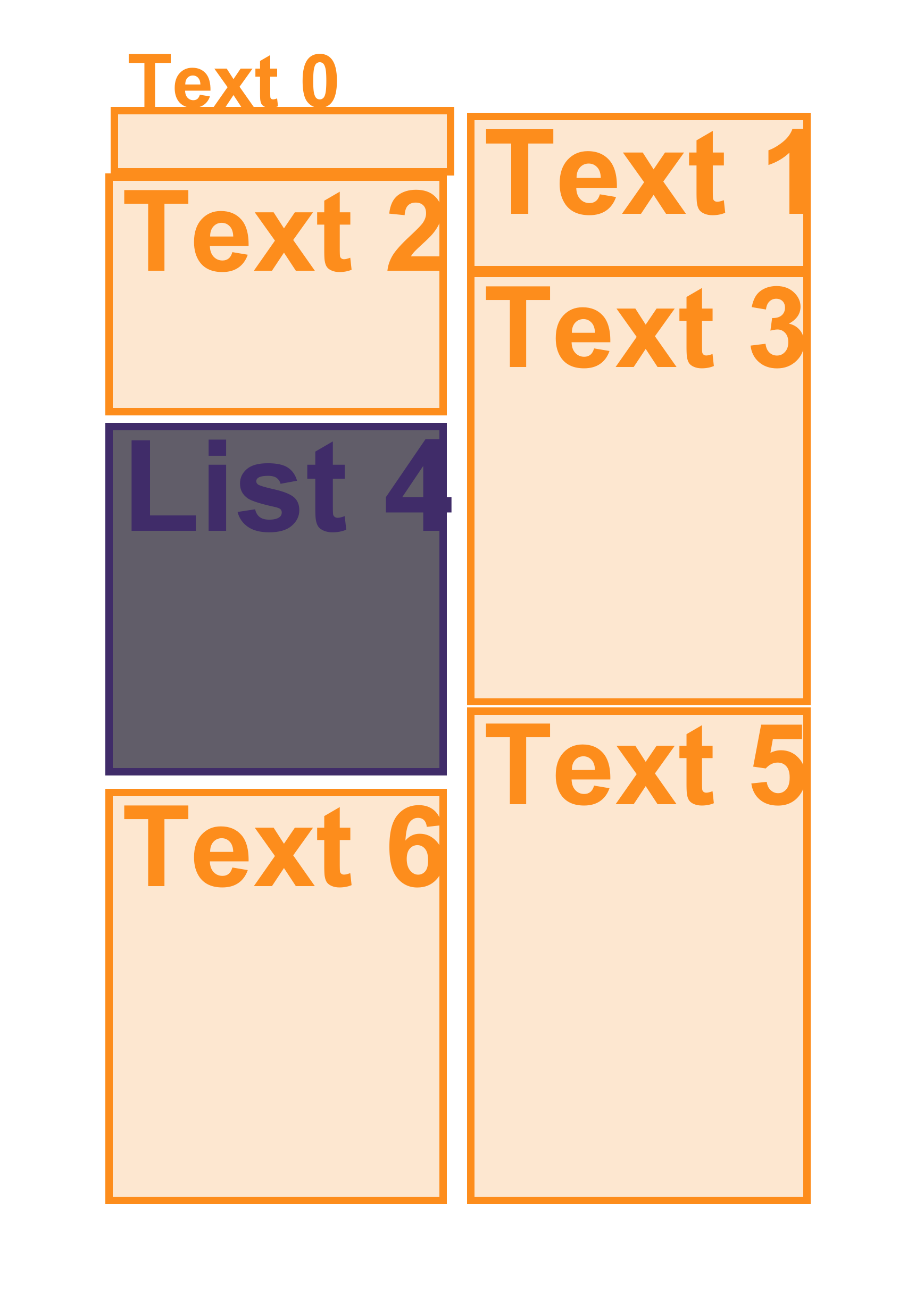} &
\includegraphics[width=\interpolationWidth,frame=0.1pt]{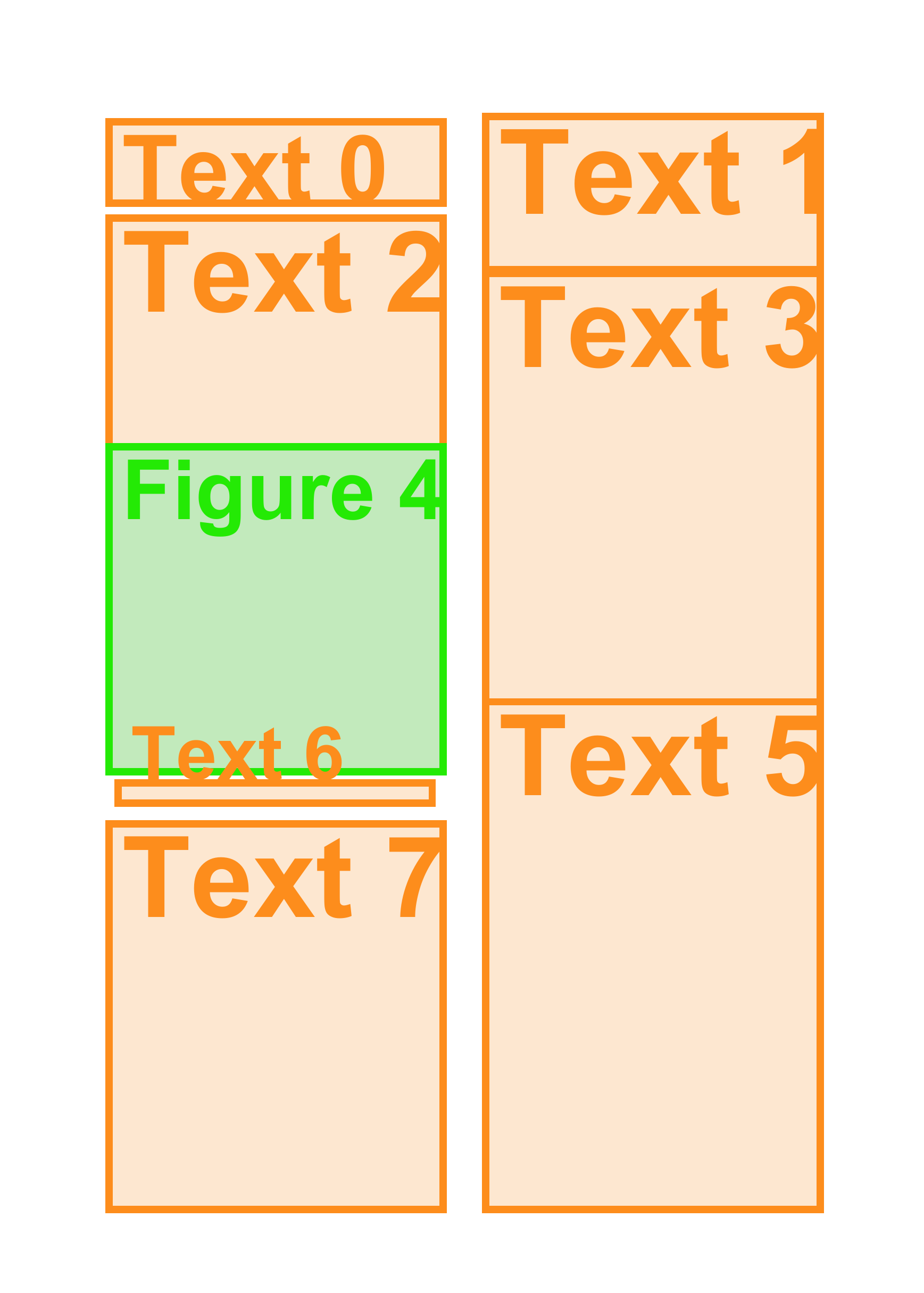} &
\includegraphics[width=\interpolationWidth,frame=0.1pt]{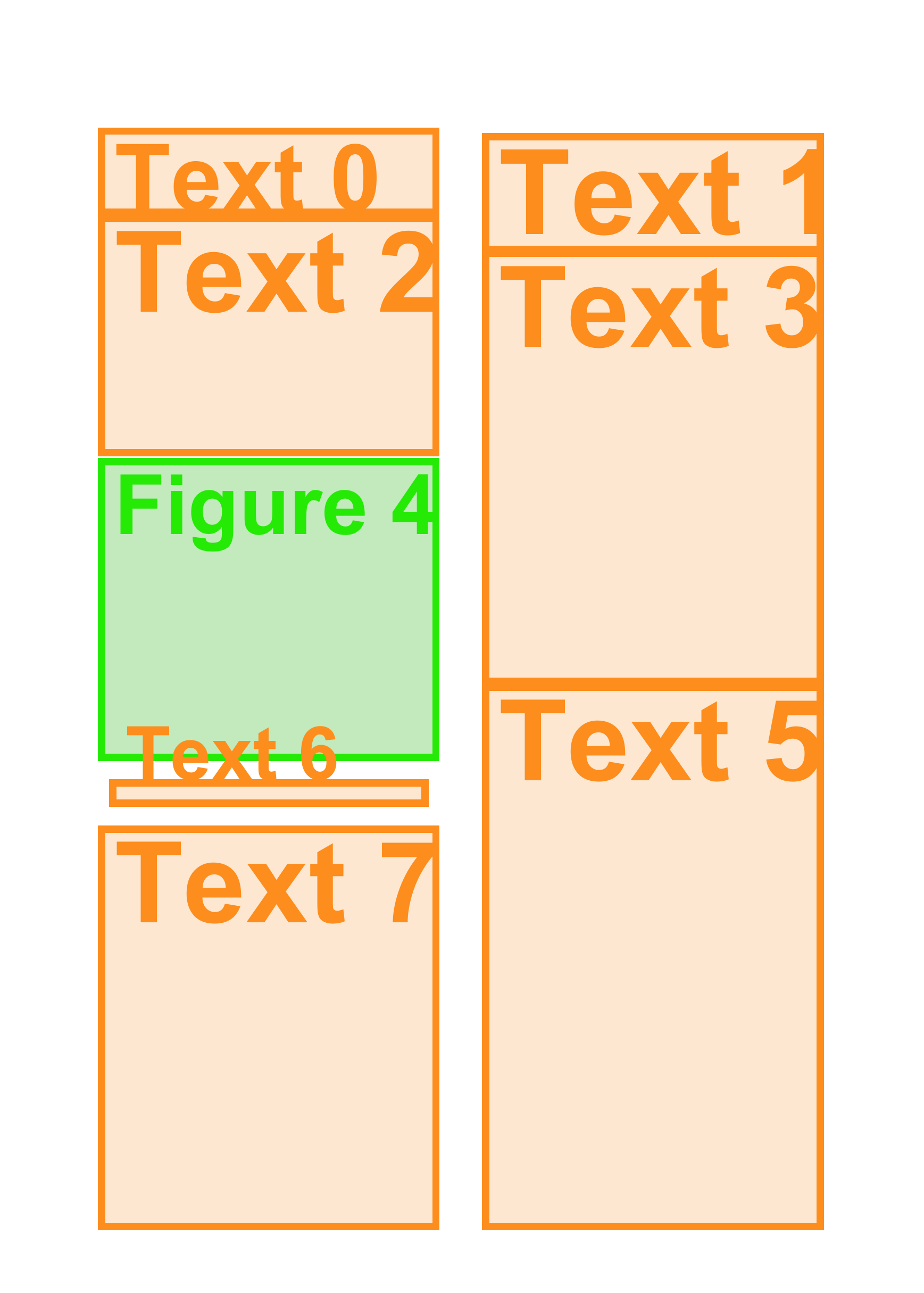} &
\includegraphics[width=\interpolationWidth,frame=0.1pt]{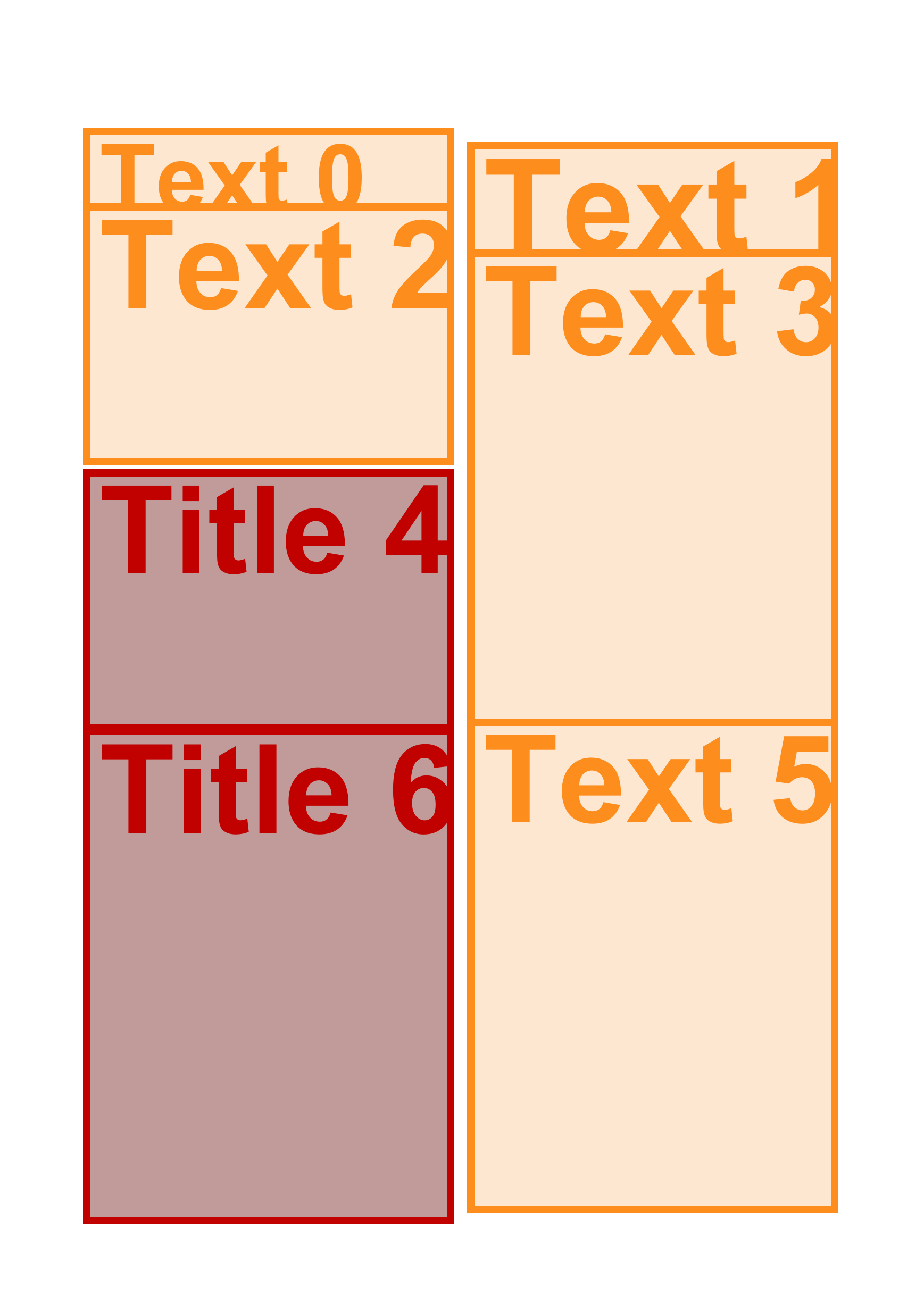} &
\includegraphics[width=\interpolationWidth,frame=0.1pt]{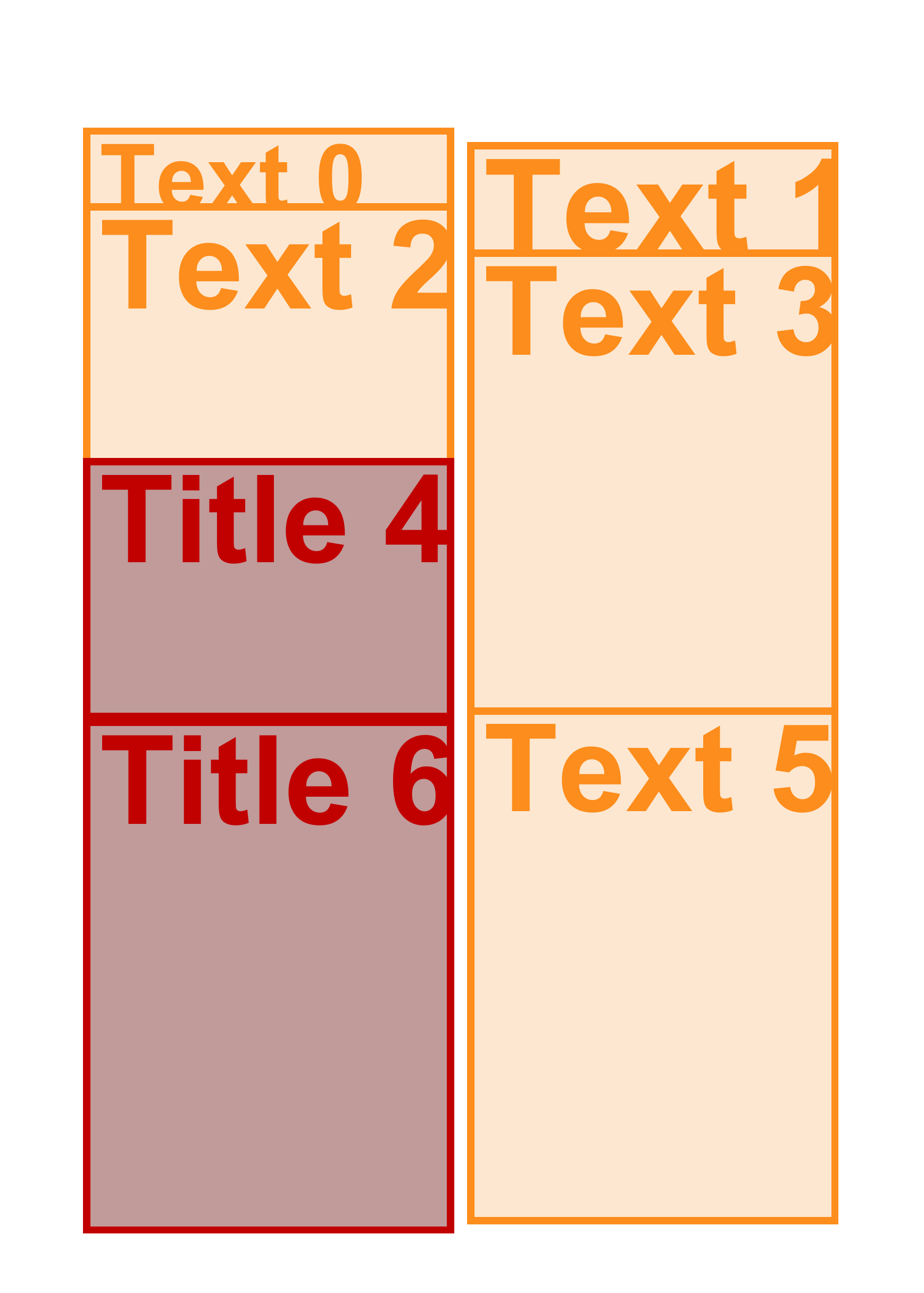} \\

\includegraphics[width=\interpolationWidth,frame=0.1pt]{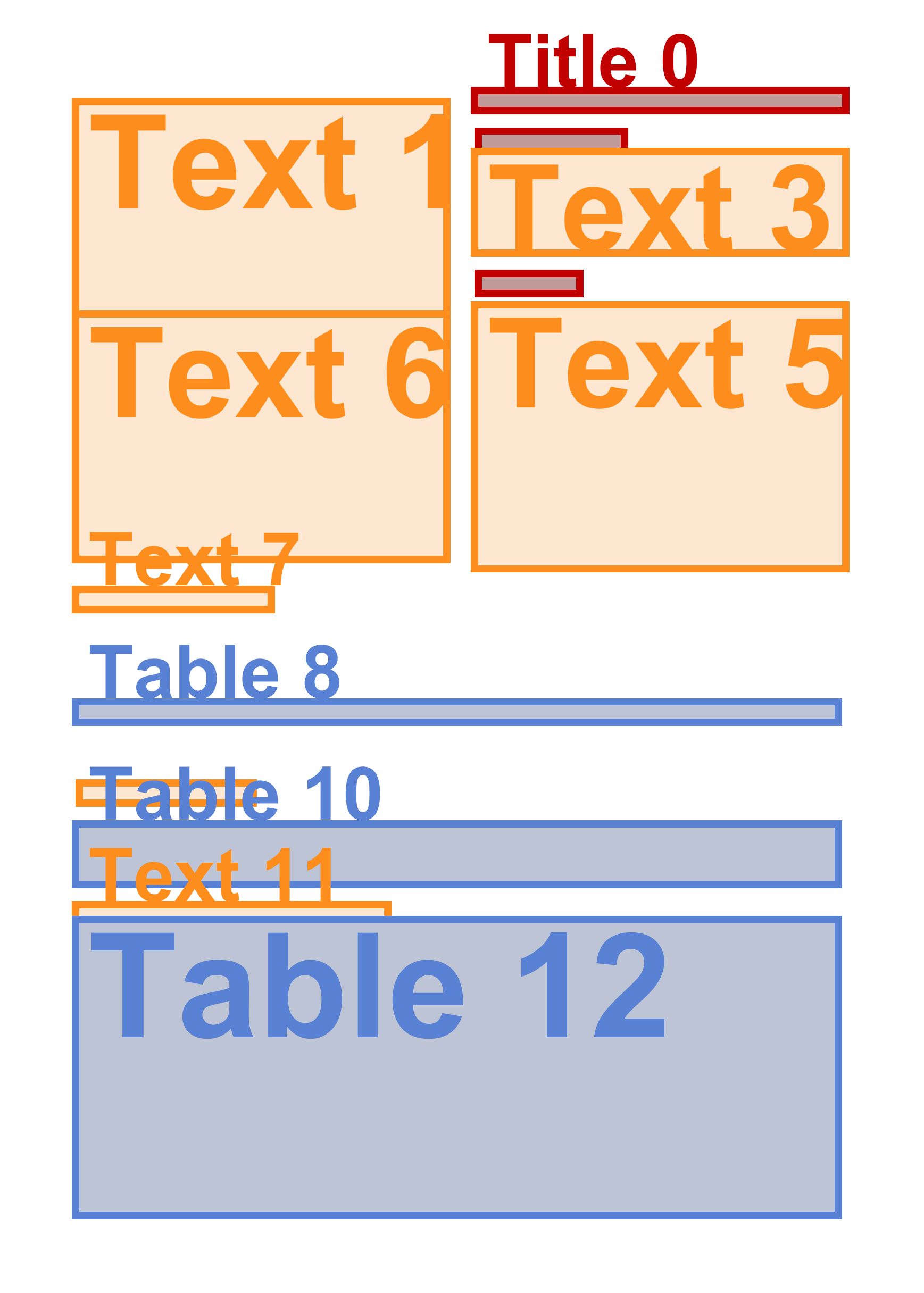} &
\includegraphics[width=\interpolationWidth,frame=0.1pt]{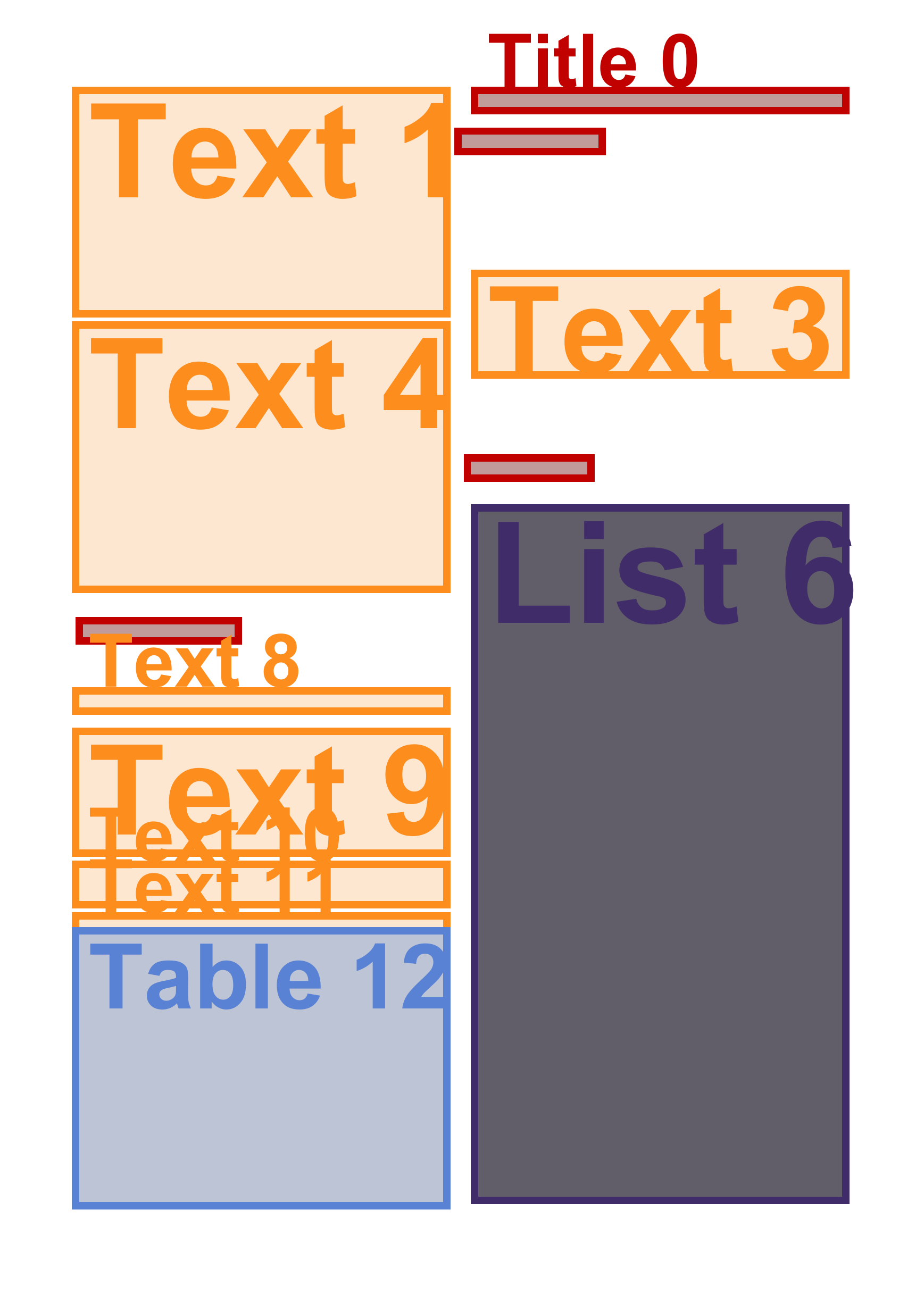} &
\includegraphics[width=\interpolationWidth,frame=0.1pt]{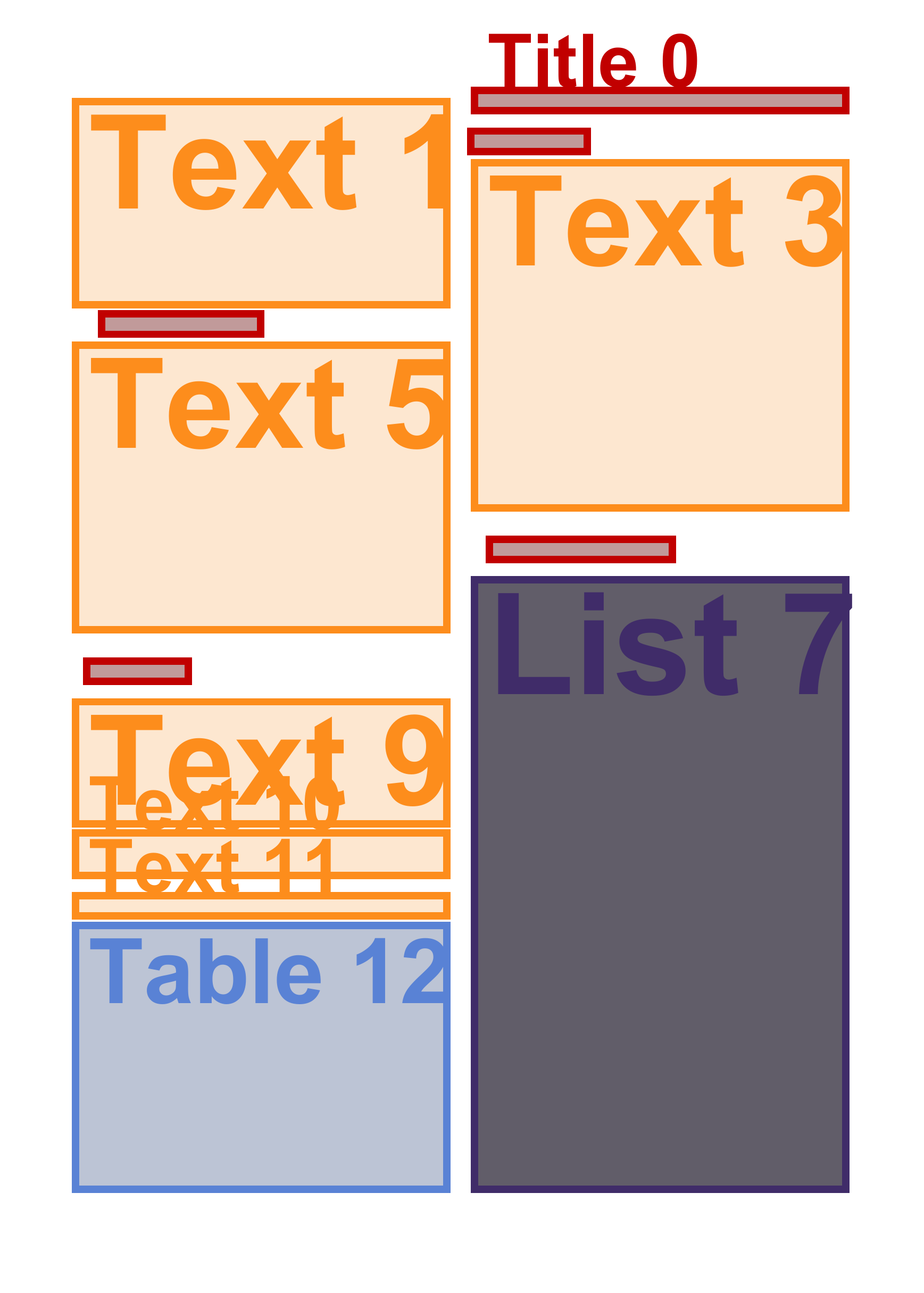} &
\includegraphics[width=\interpolationWidth,frame=0.1pt]{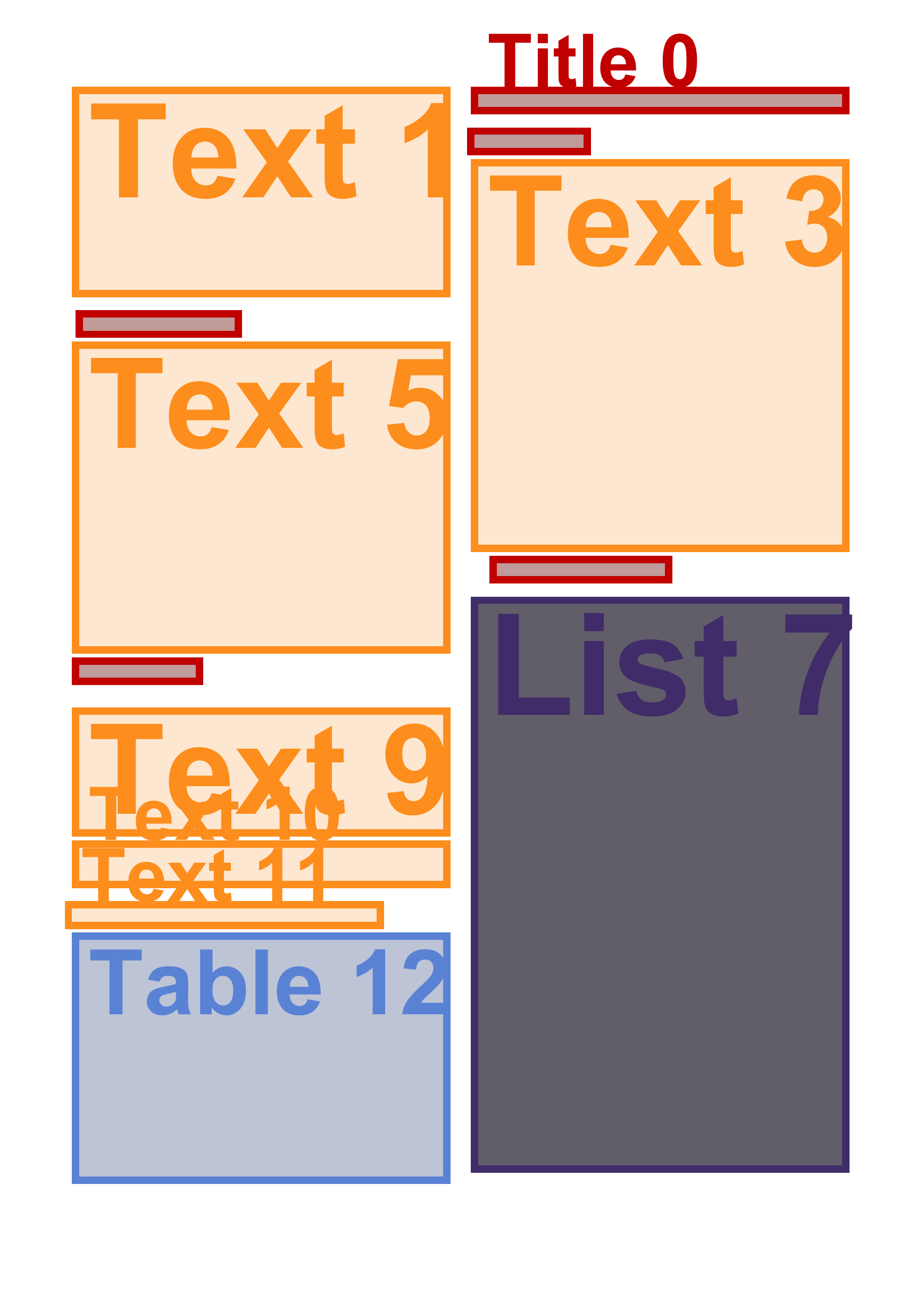} &
\includegraphics[width=\interpolationWidth,frame=0.1pt]{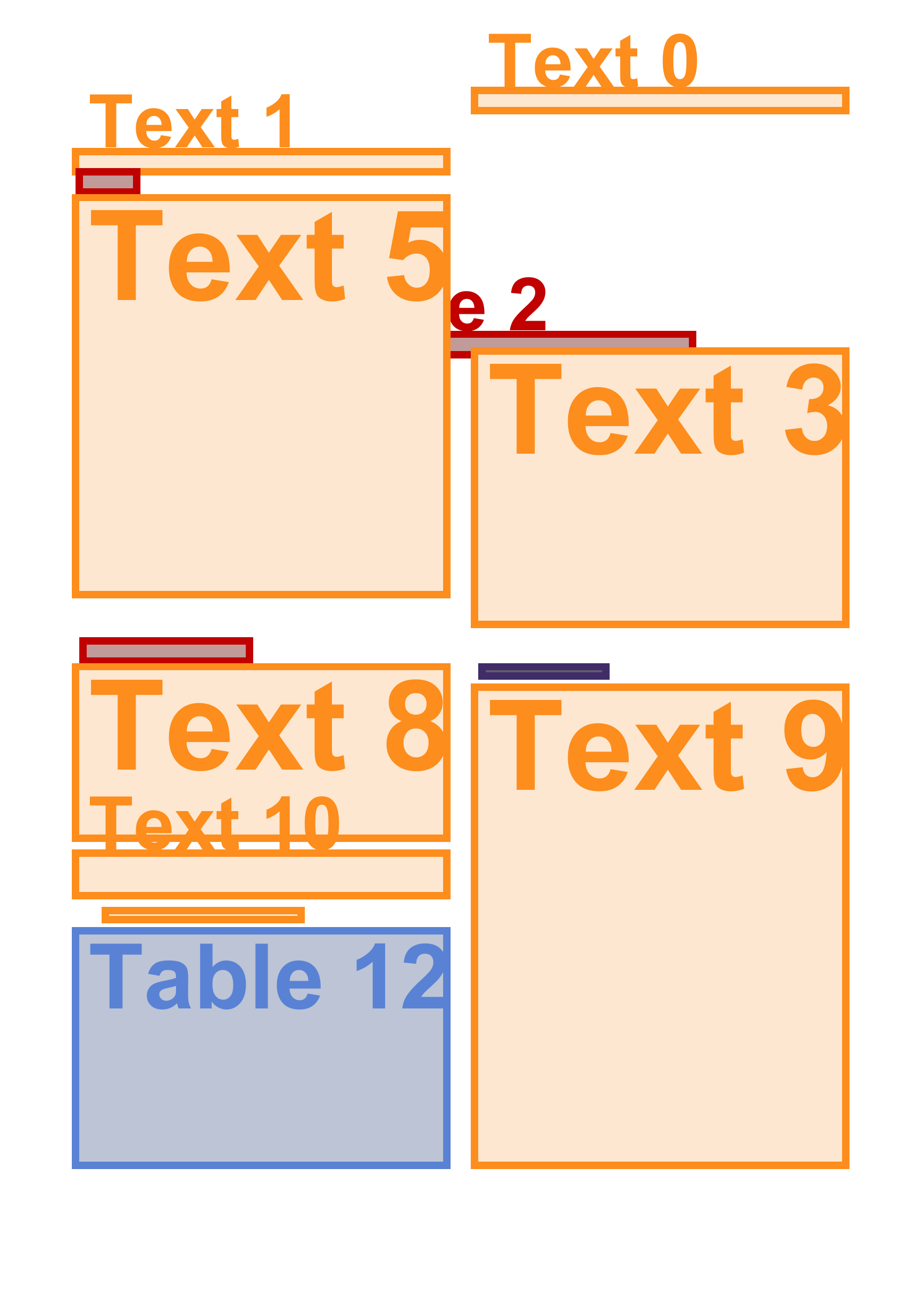} &
\includegraphics[width=\interpolationWidth,frame=0.1pt]{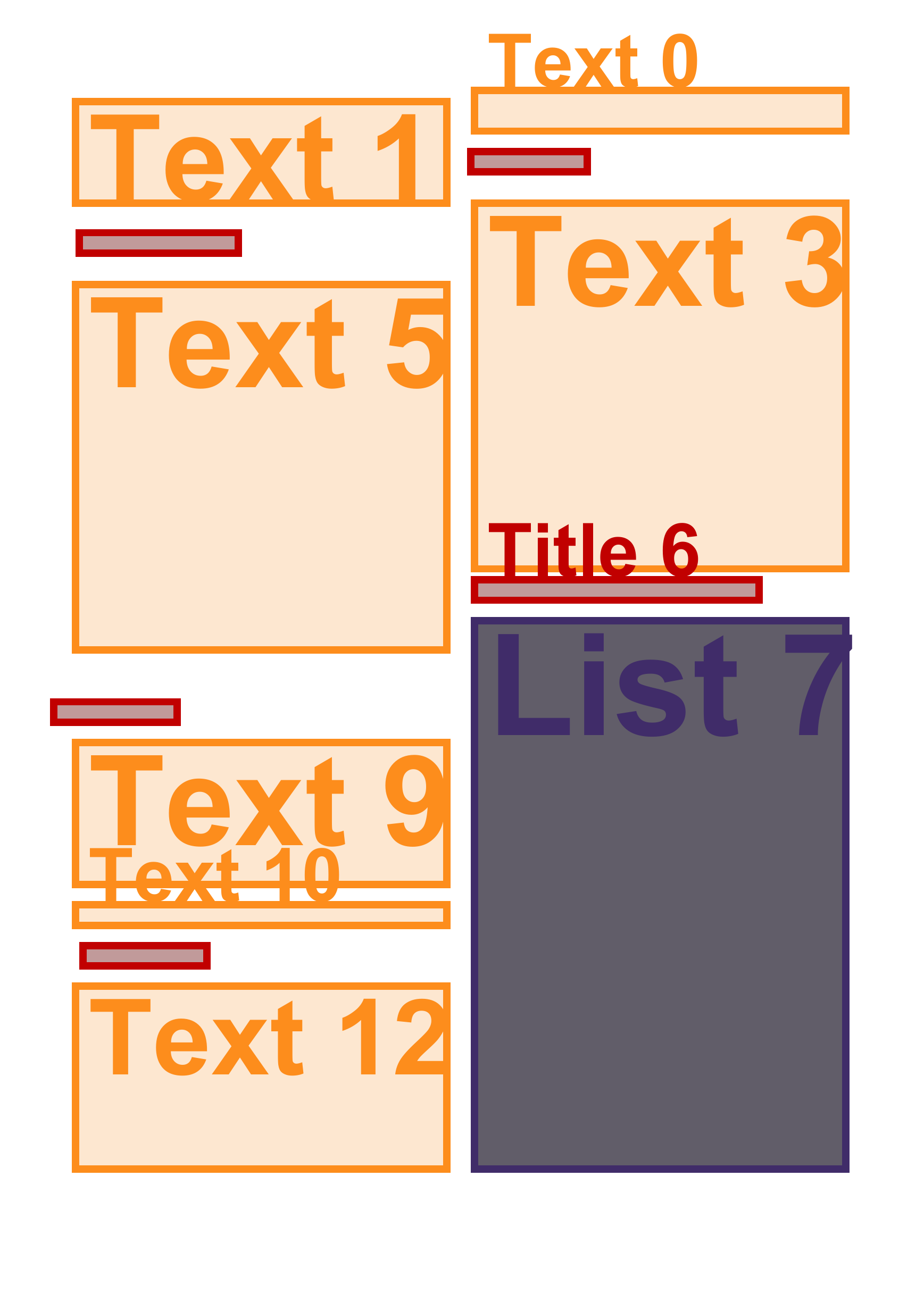} &
\includegraphics[width=\interpolationWidth,frame=0.1pt]{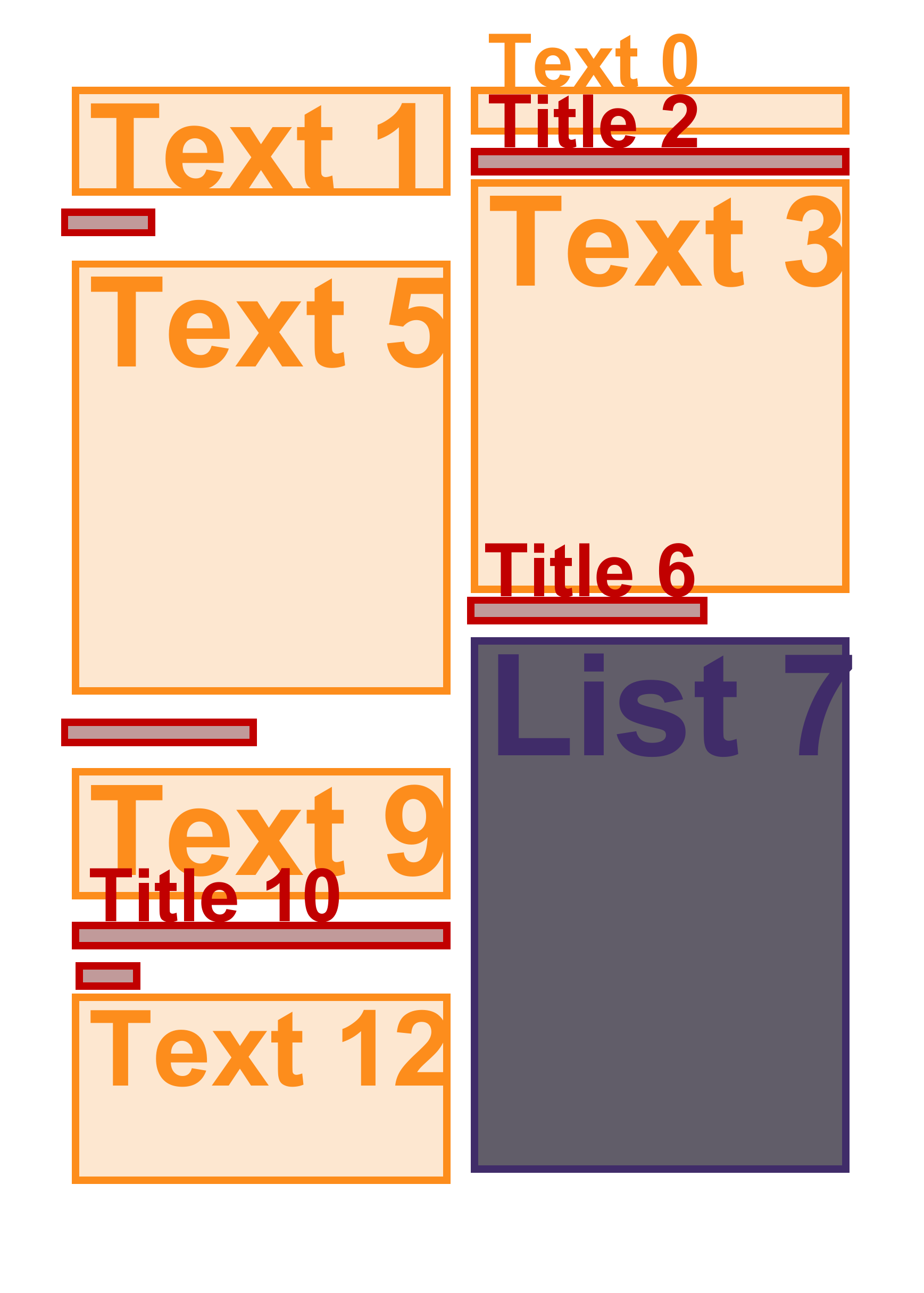} &
\includegraphics[width=\interpolationWidth,frame=0.1pt]{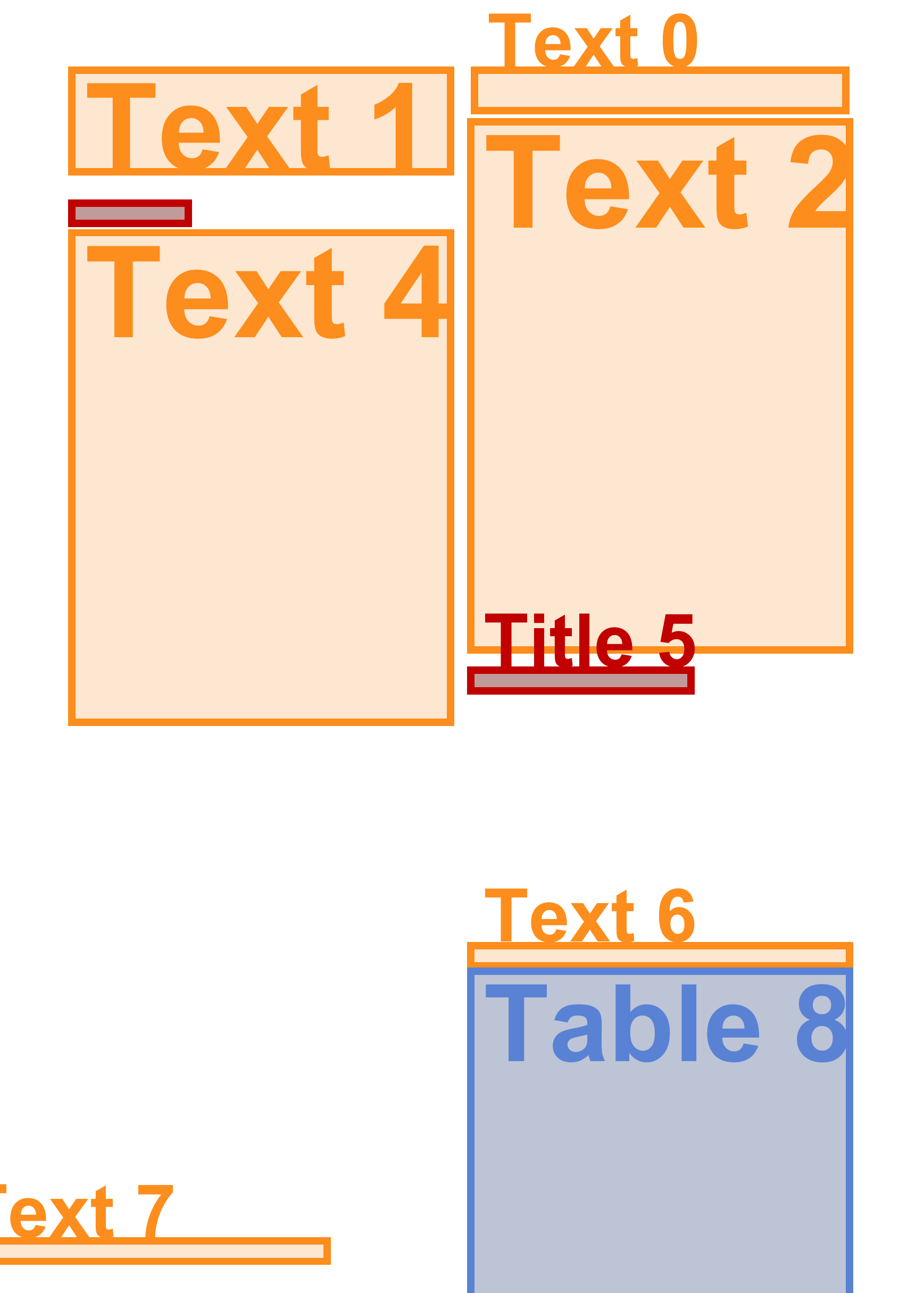} &
\includegraphics[width=\interpolationWidth,frame=0.1pt]{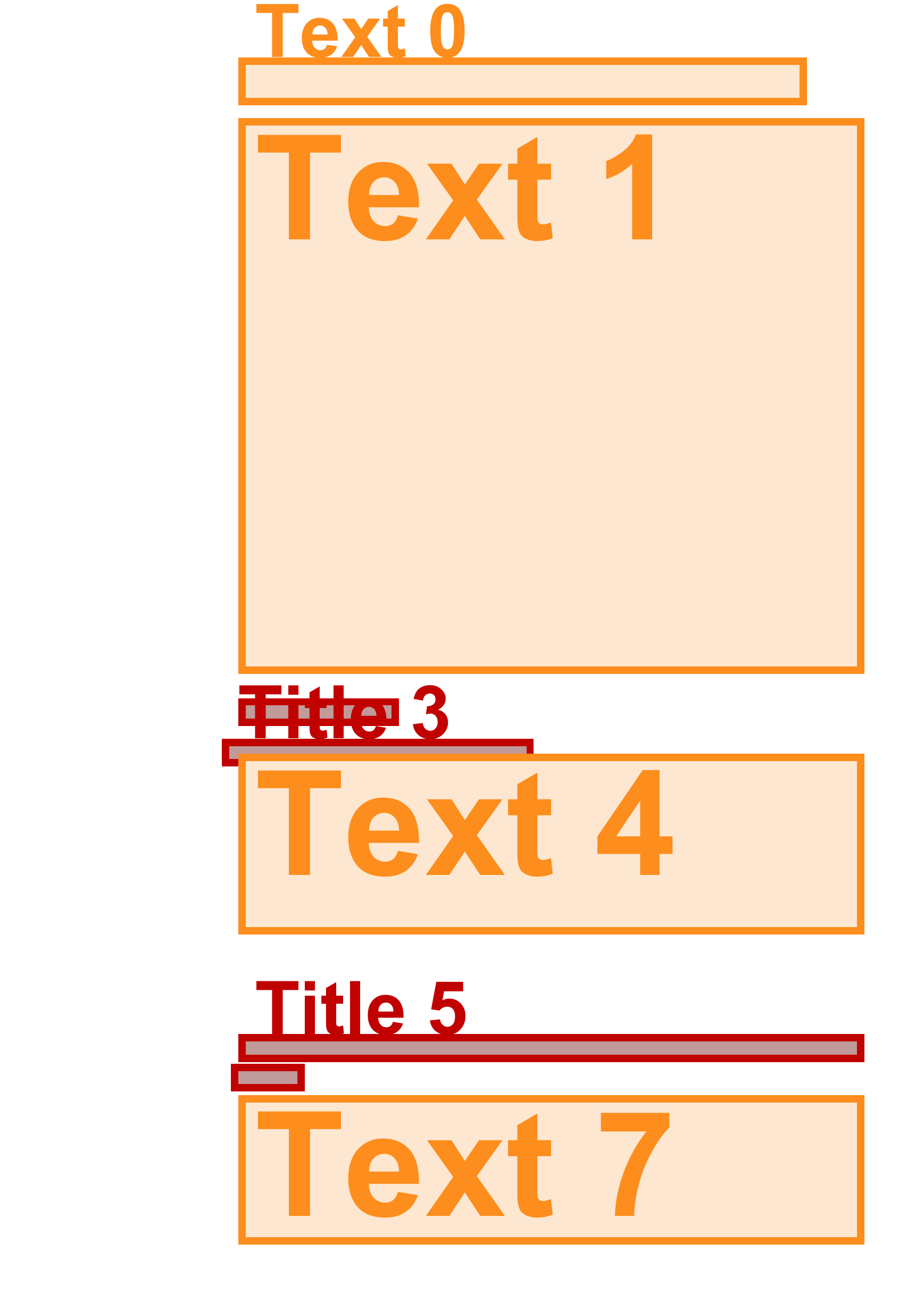} &
\includegraphics[width=\interpolationWidth,frame=0.1pt]{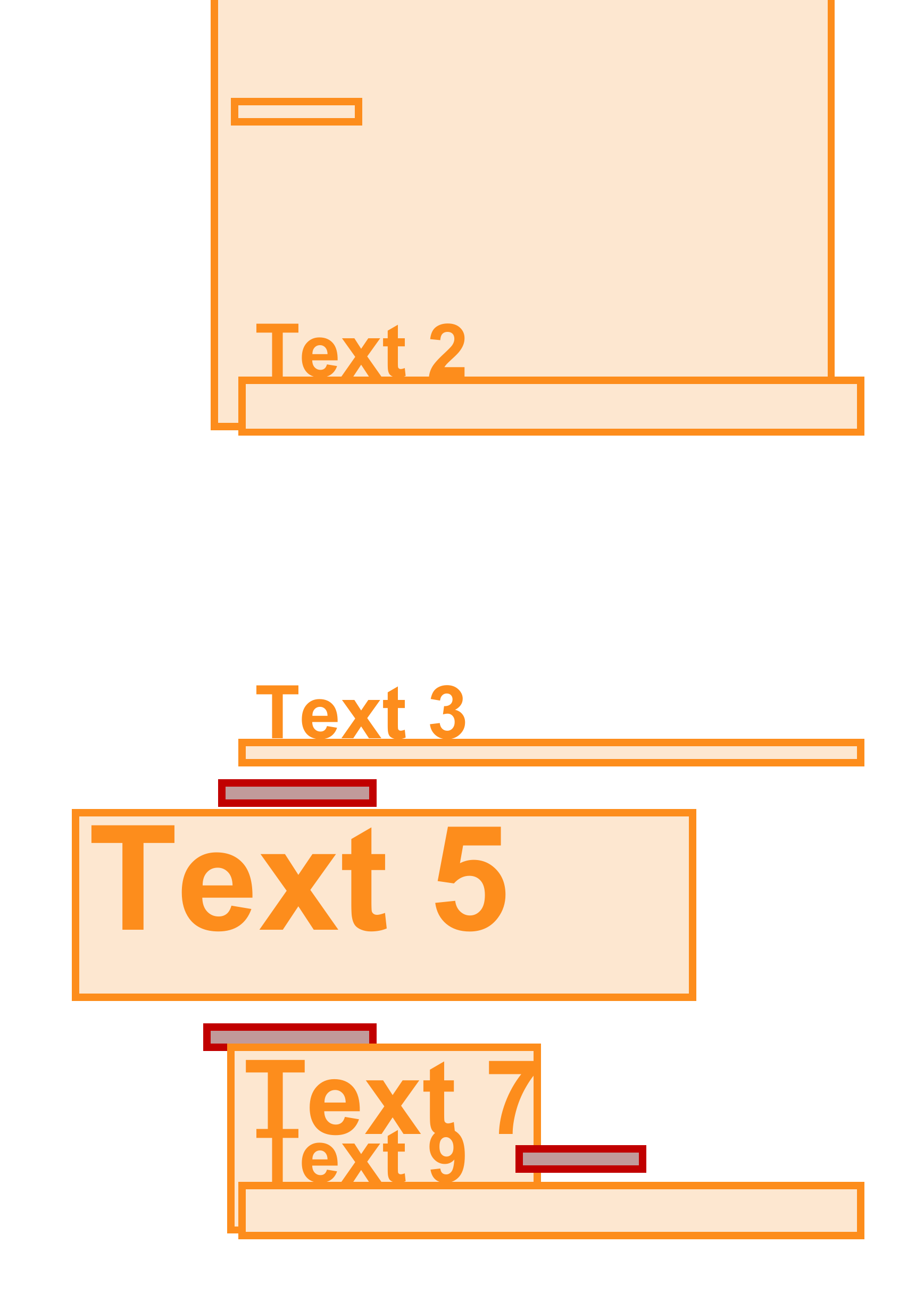} &
\includegraphics[width=\interpolationWidth,frame=0.1pt]{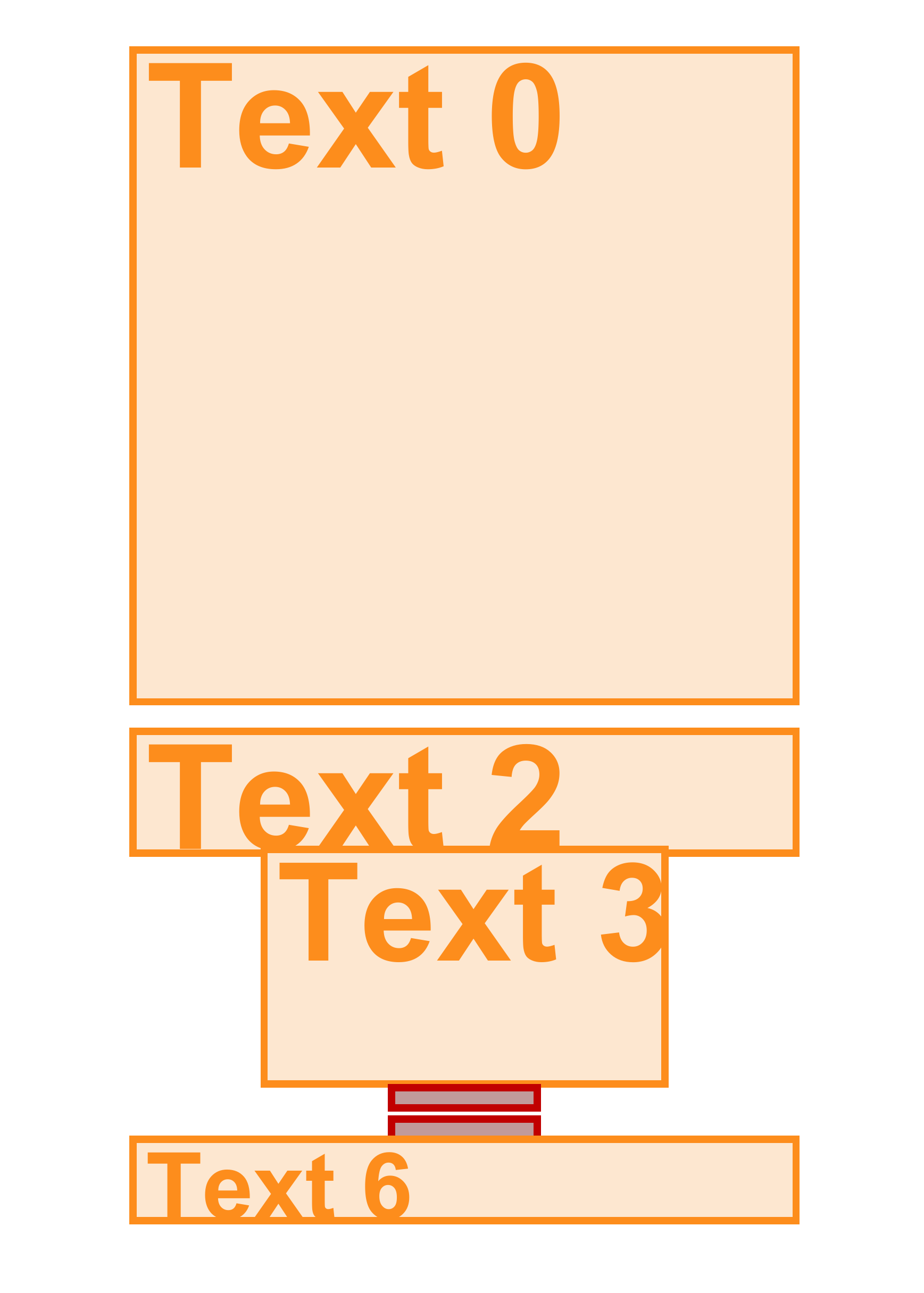} \\

\includegraphics[width=\interpolationWidth,frame=0.1pt]{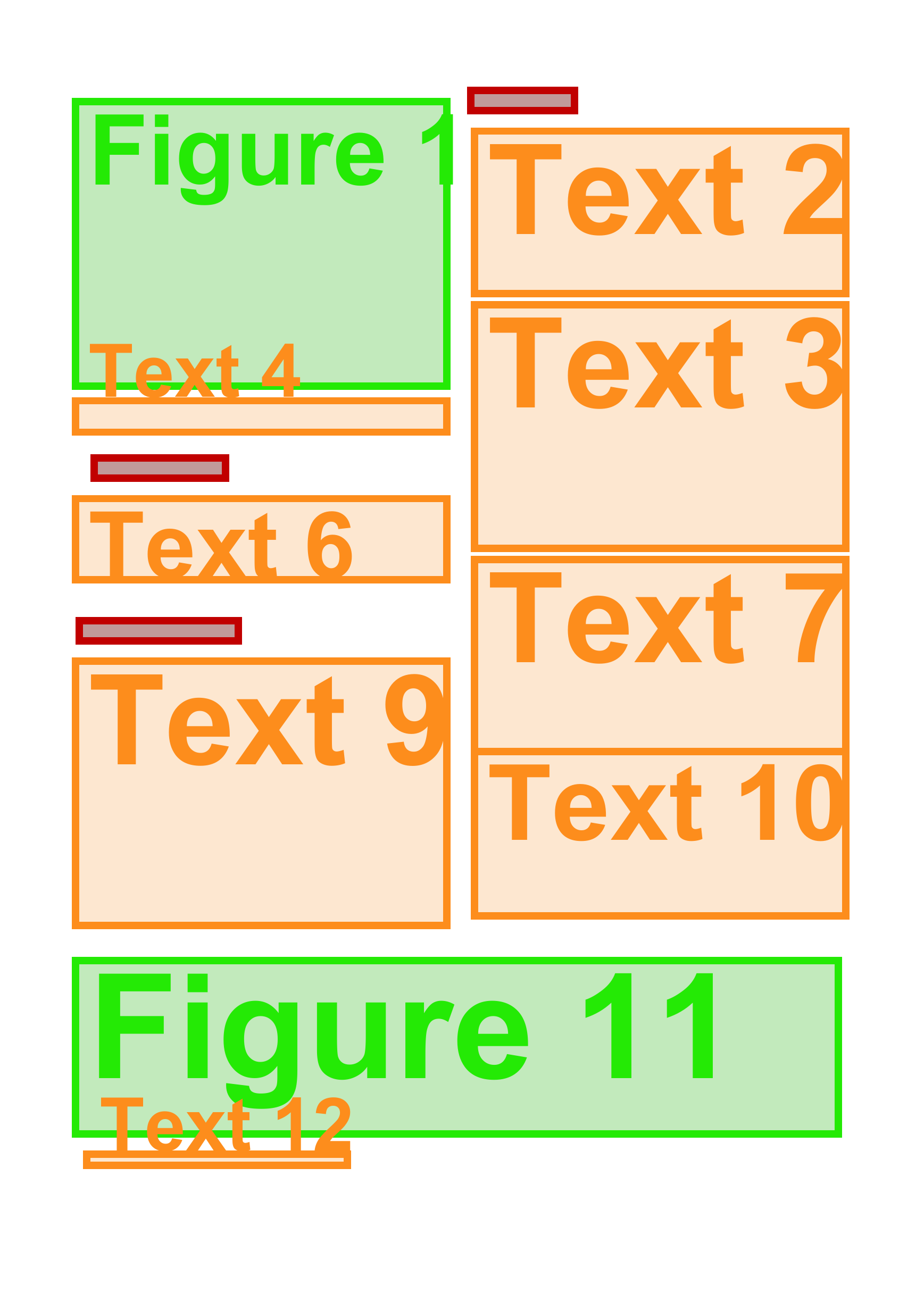} &
\includegraphics[width=\interpolationWidth,frame=0.1pt]{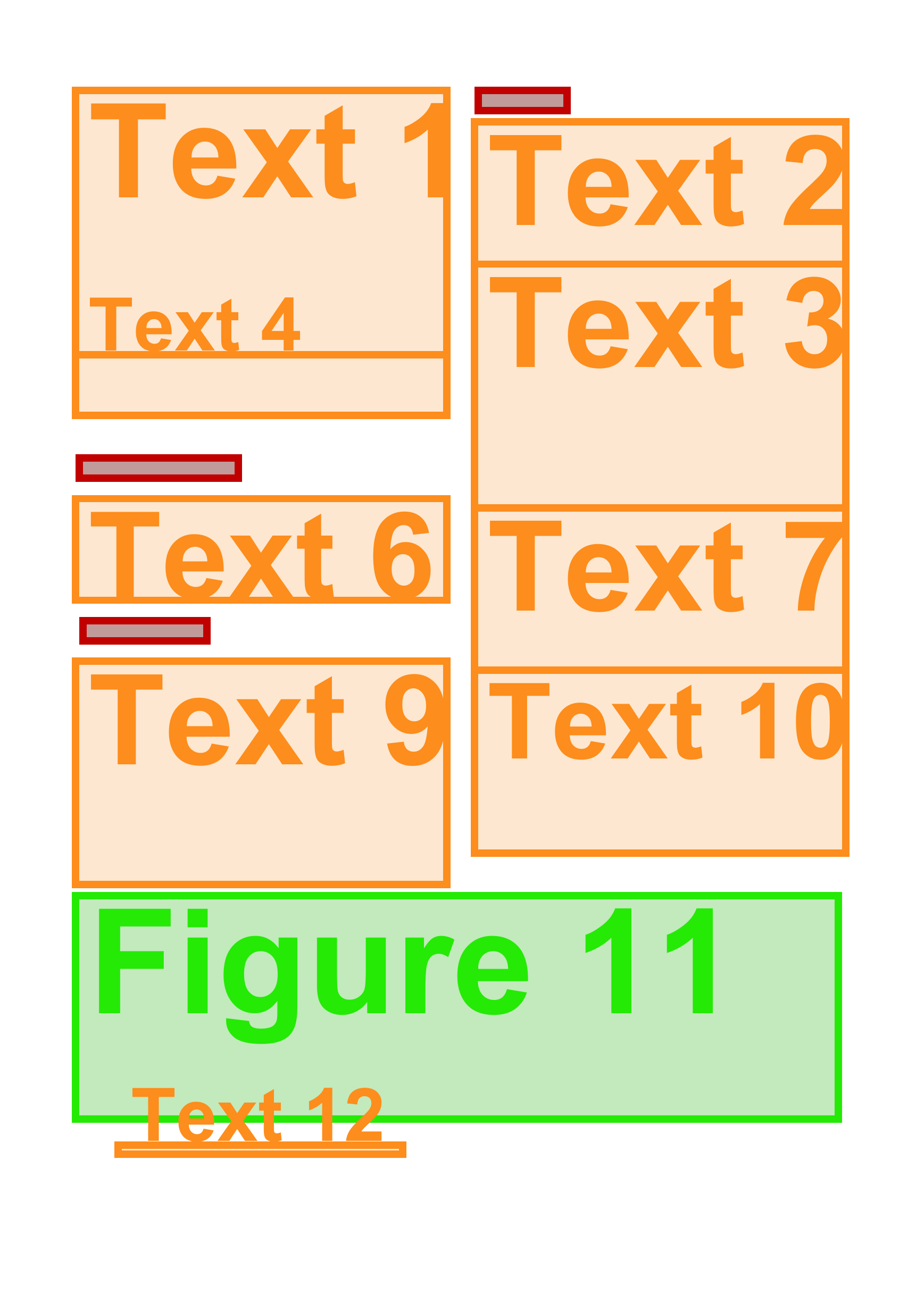} &
\includegraphics[width=\interpolationWidth,frame=0.1pt]{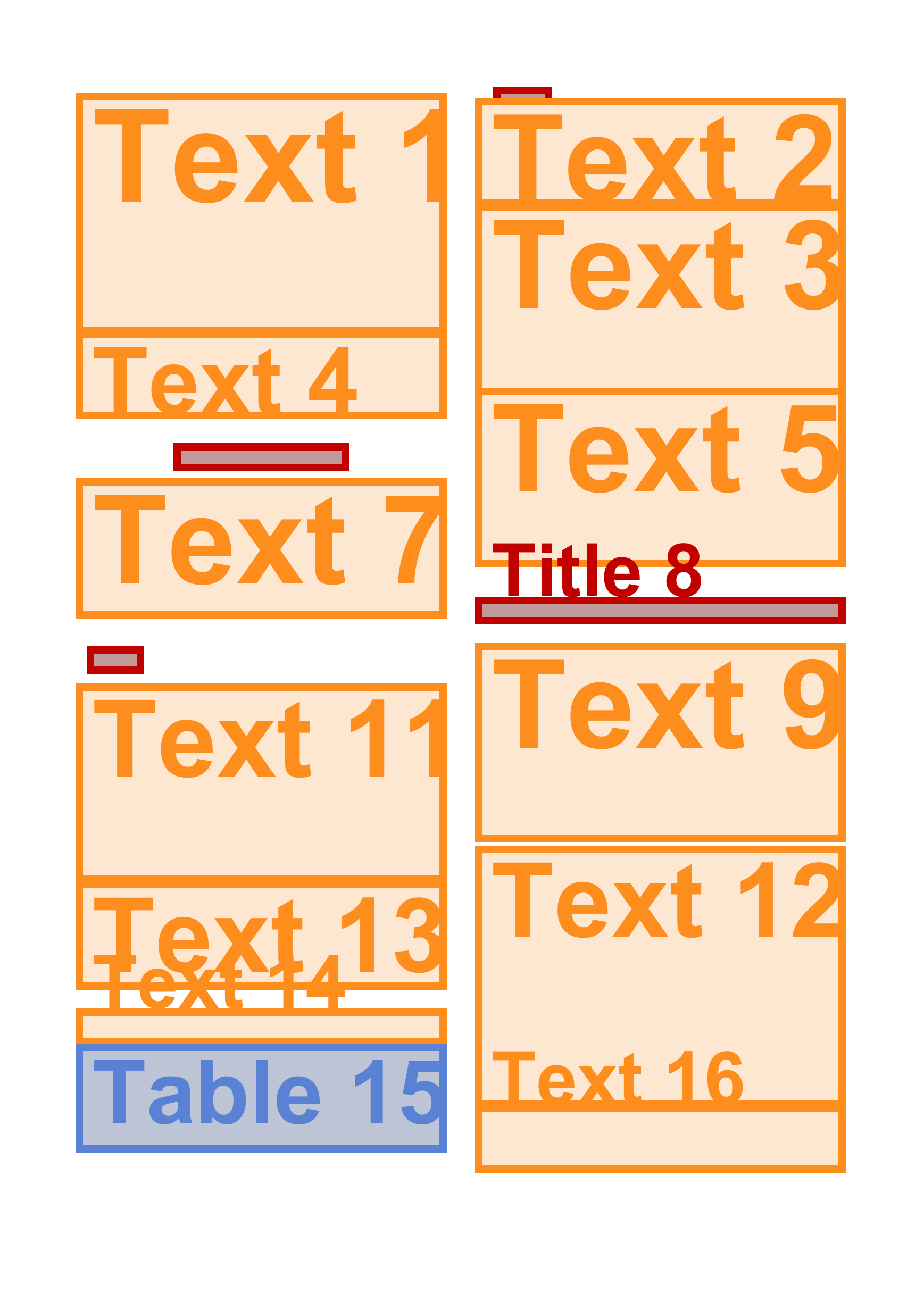} &
\includegraphics[width=\interpolationWidth,frame=0.1pt]{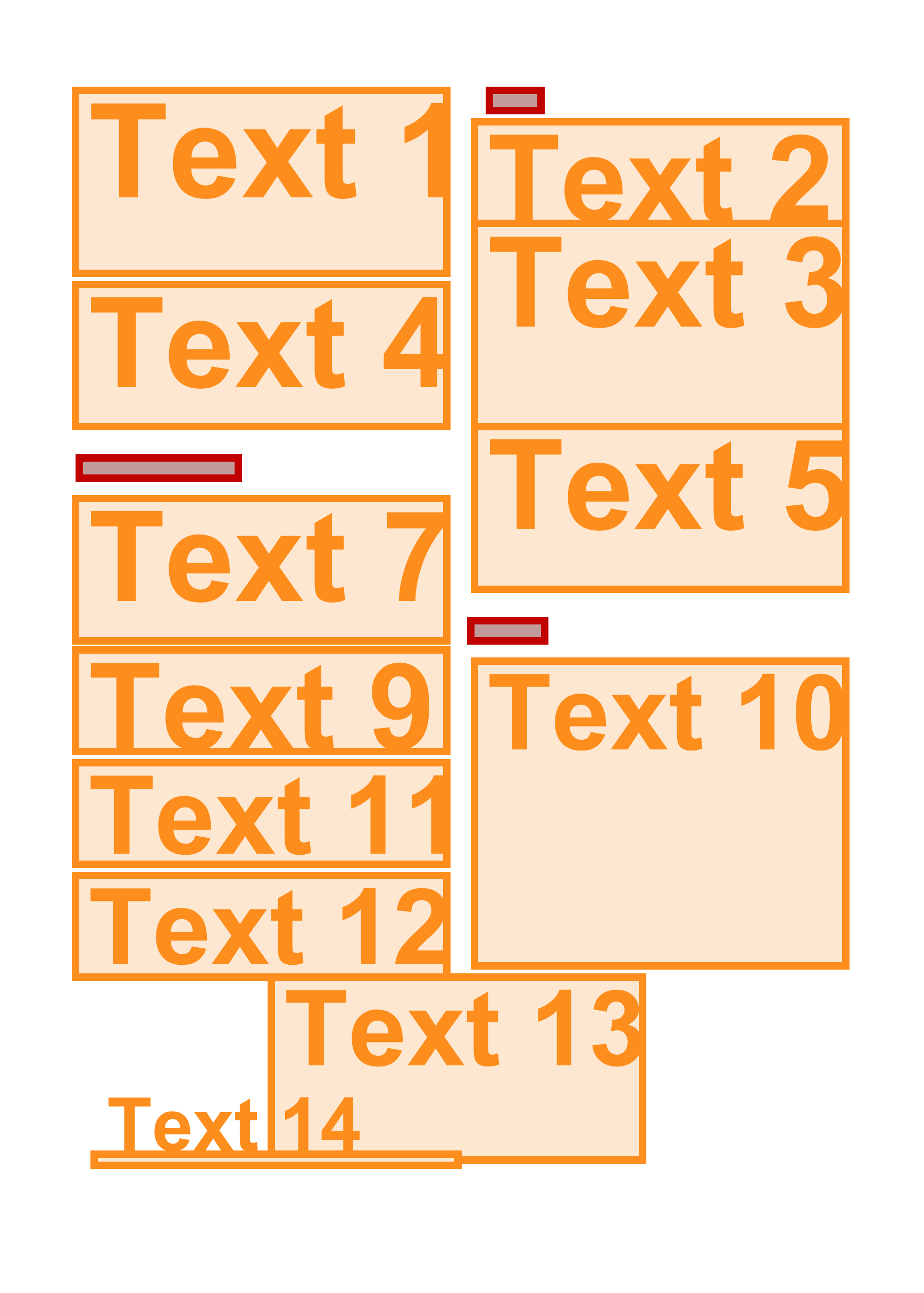} &
\includegraphics[width=\interpolationWidth,frame=0.1pt]{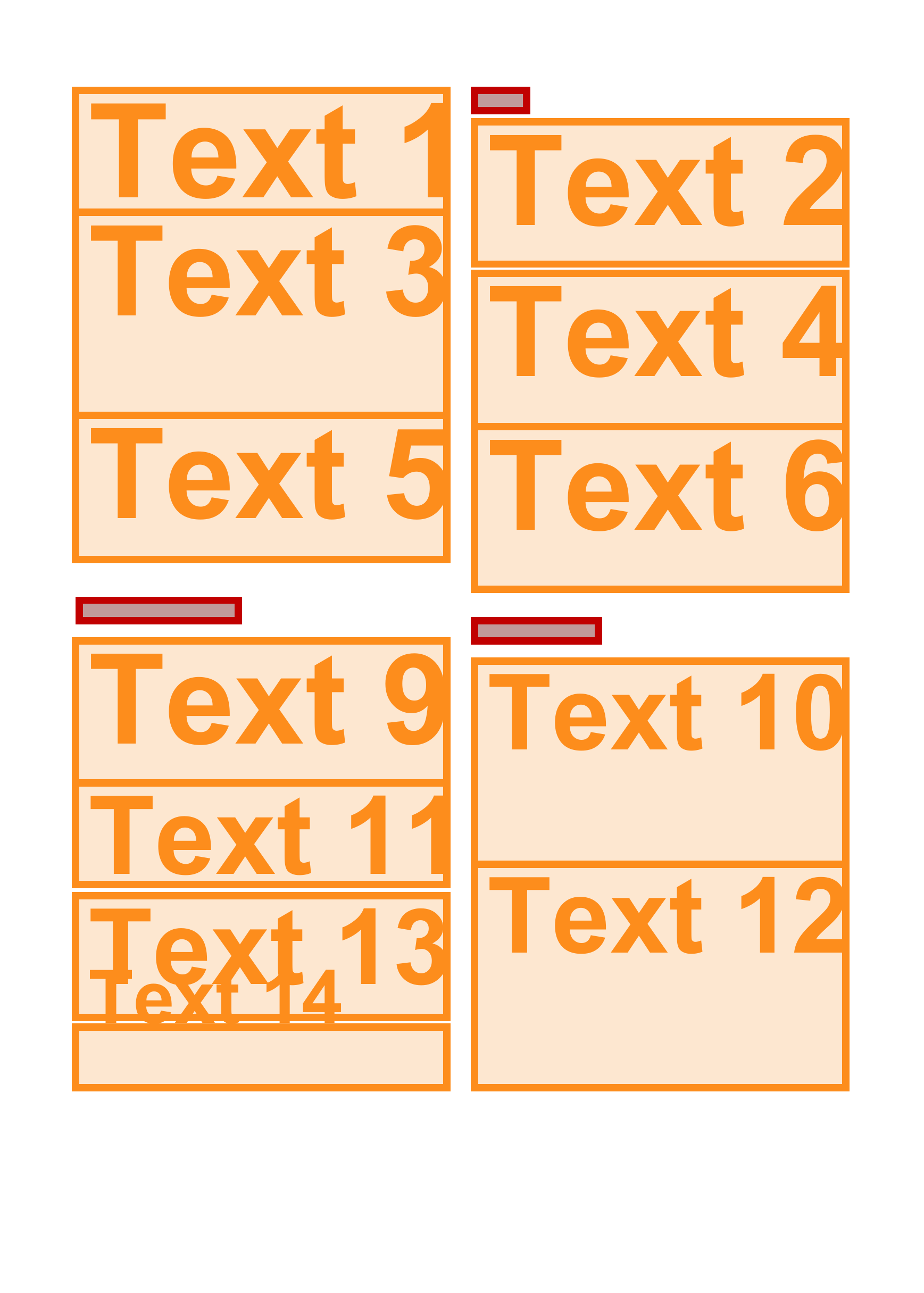} &
\includegraphics[width=\interpolationWidth,frame=0.1pt]{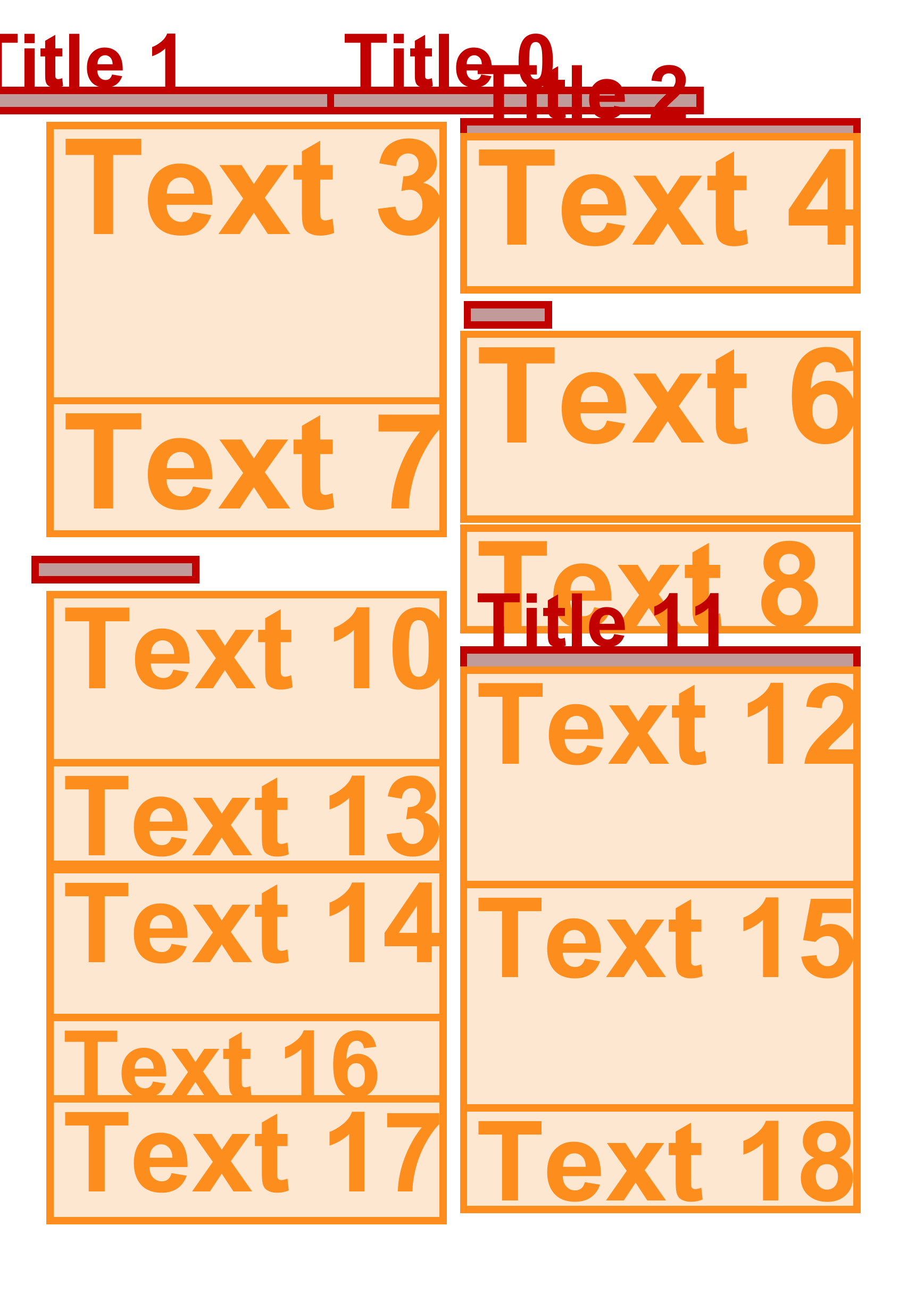} &
\includegraphics[width=\interpolationWidth,frame=0.1pt]{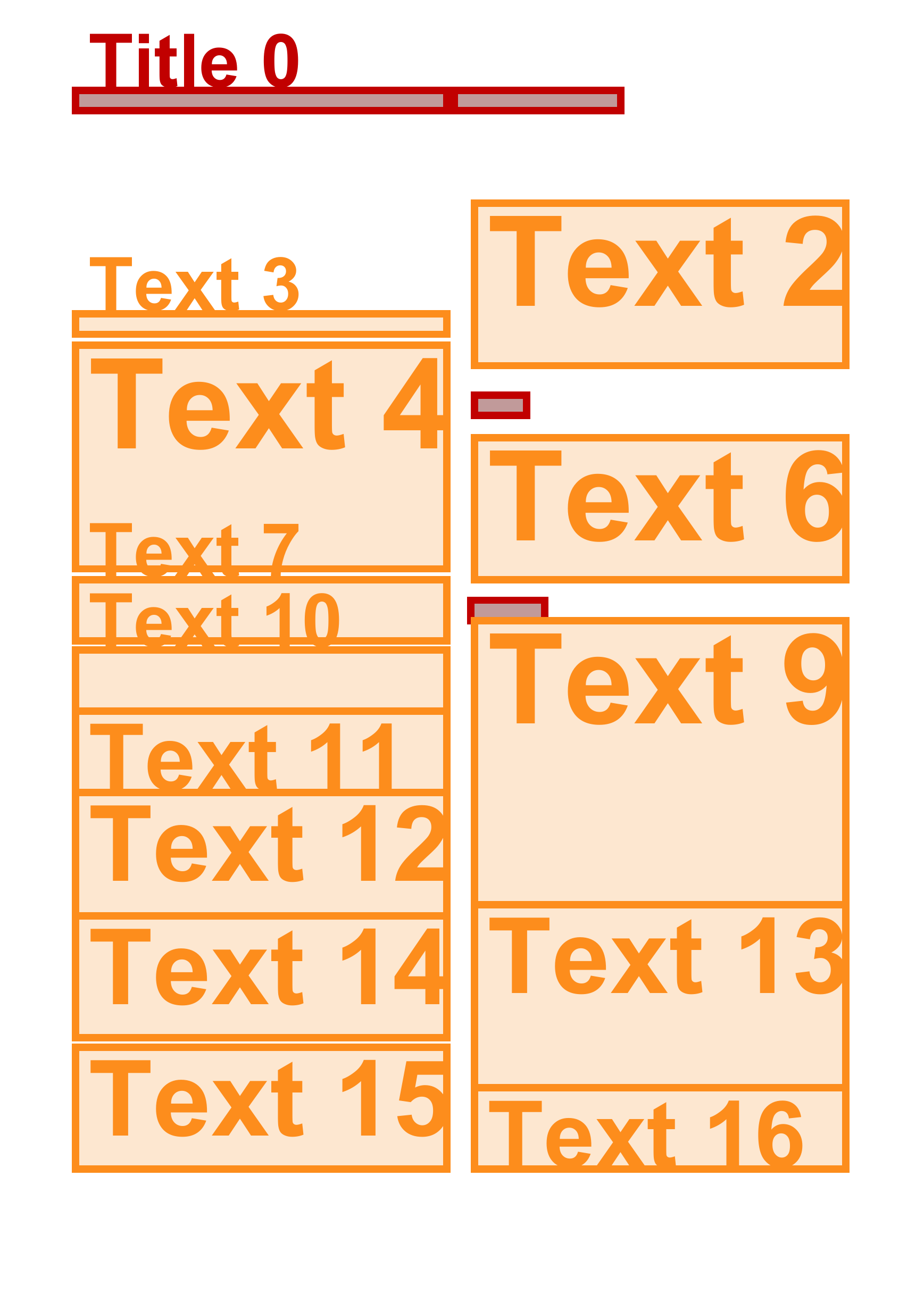} &
\includegraphics[width=\interpolationWidth,frame=0.1pt]{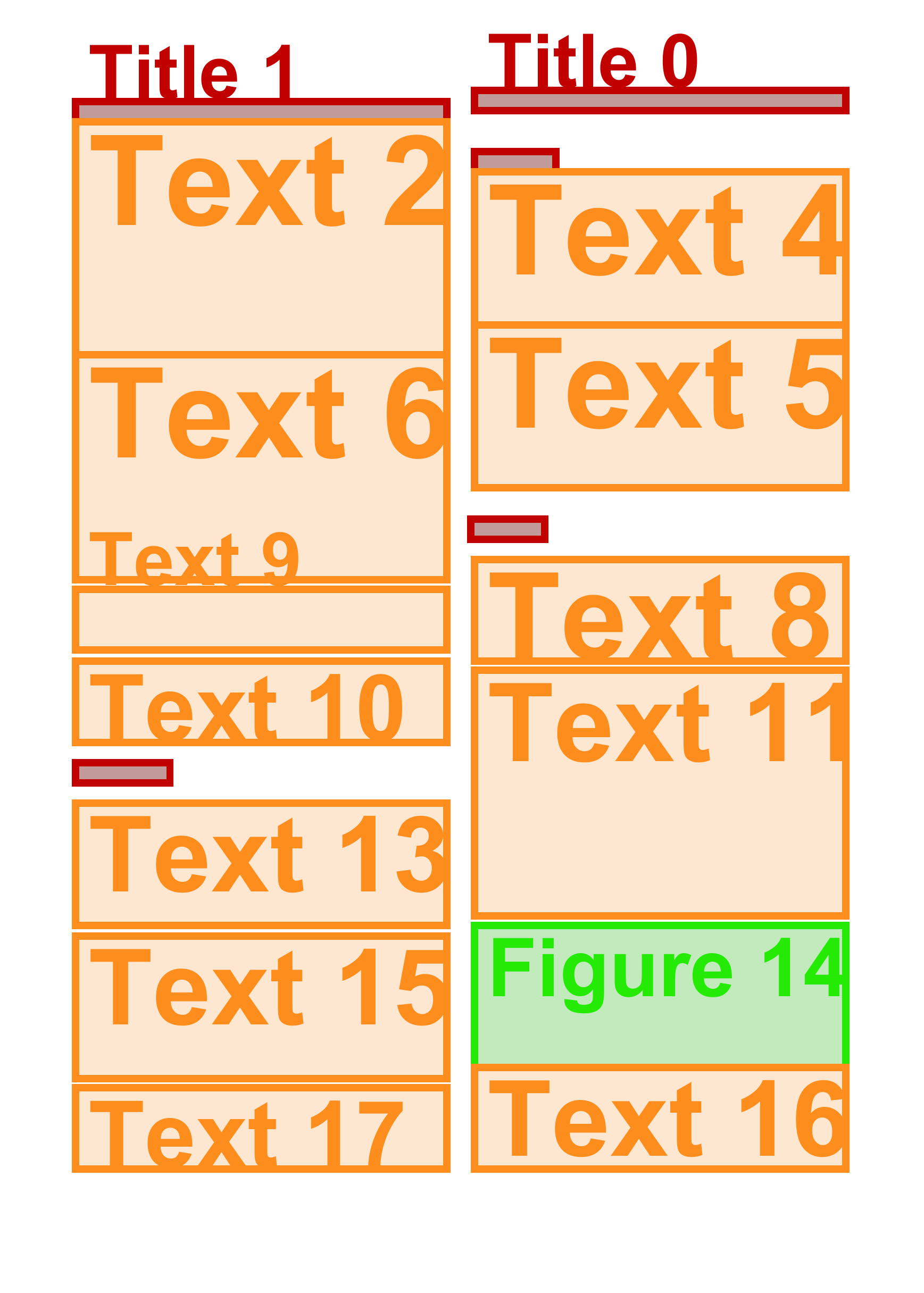} &
\includegraphics[width=\interpolationWidth,frame=0.1pt]{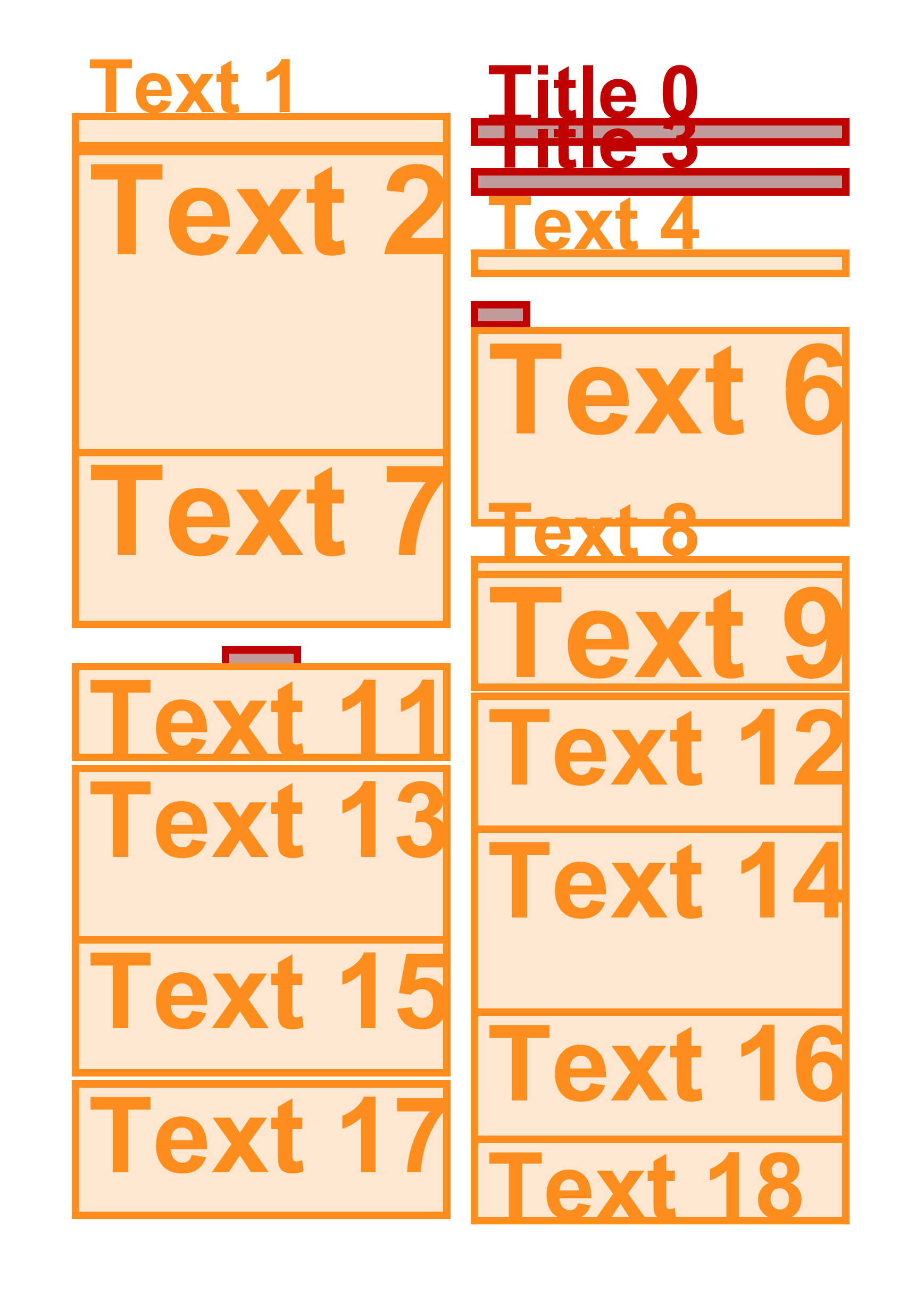} &
\includegraphics[width=\interpolationWidth,frame=0.1pt]{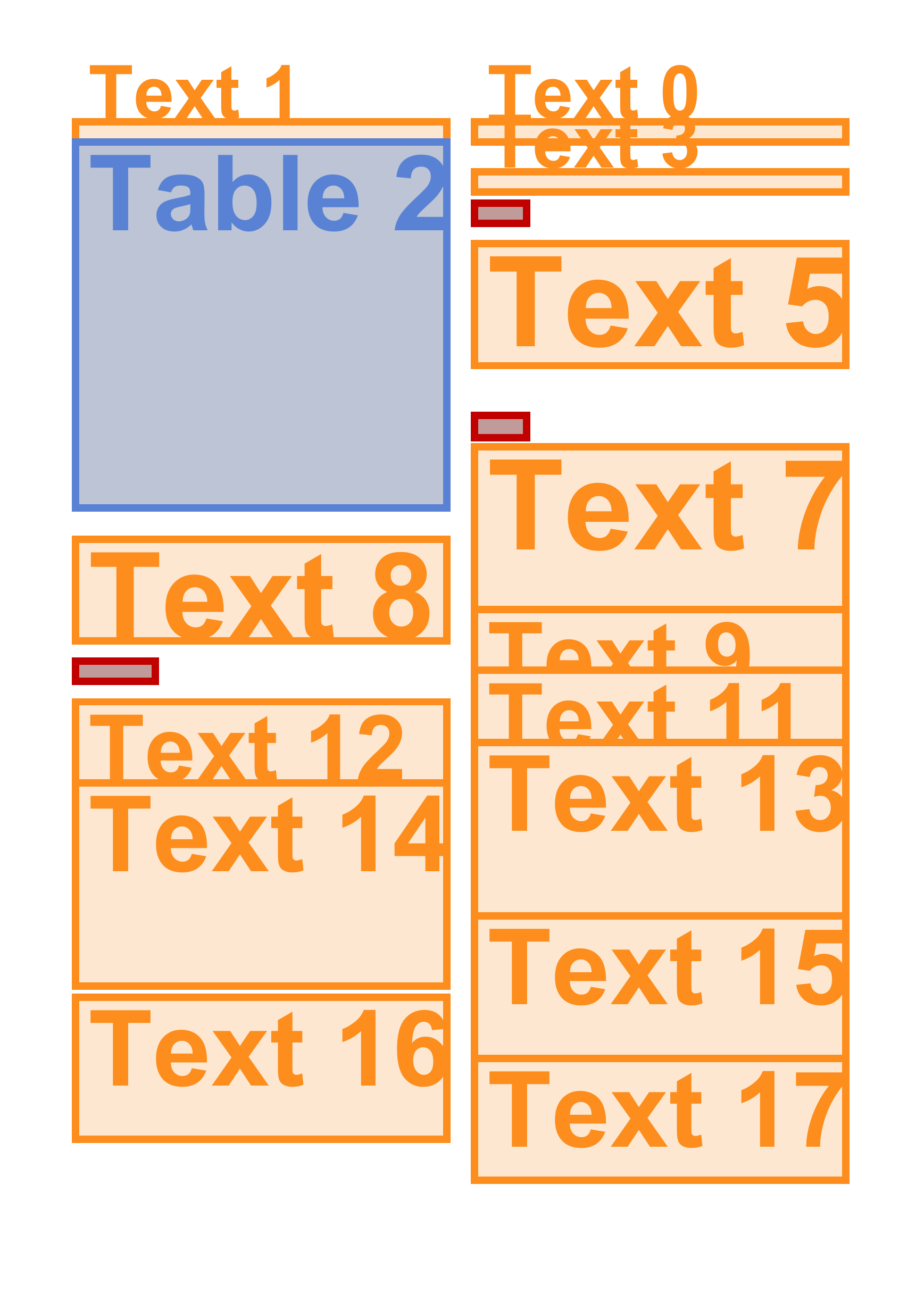} &
\includegraphics[width=\interpolationWidth,frame=0.1pt]{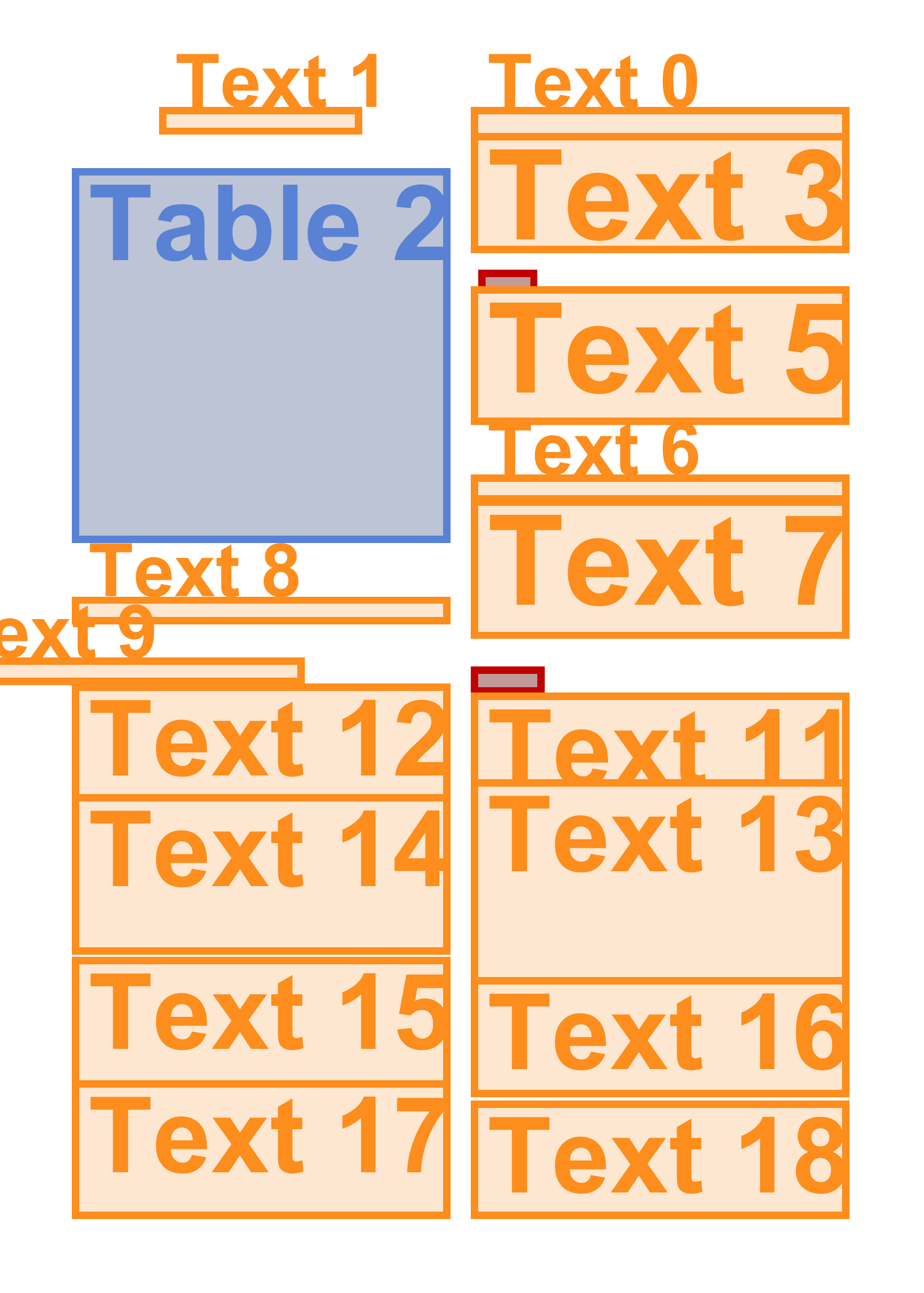} \\

\includegraphics[width=\interpolationWidth,frame=0.1pt]{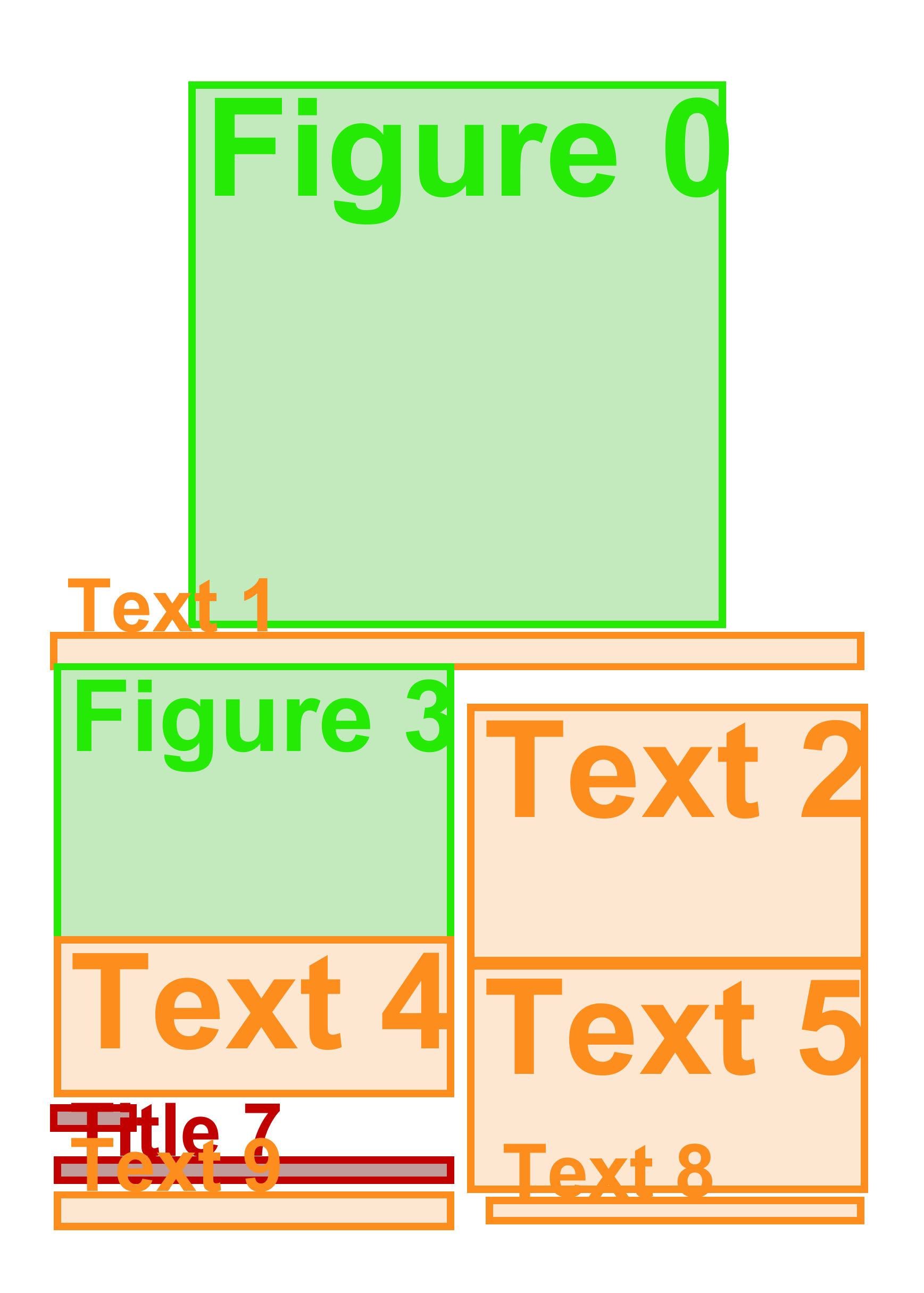} &
\includegraphics[width=\interpolationWidth,frame=0.1pt]{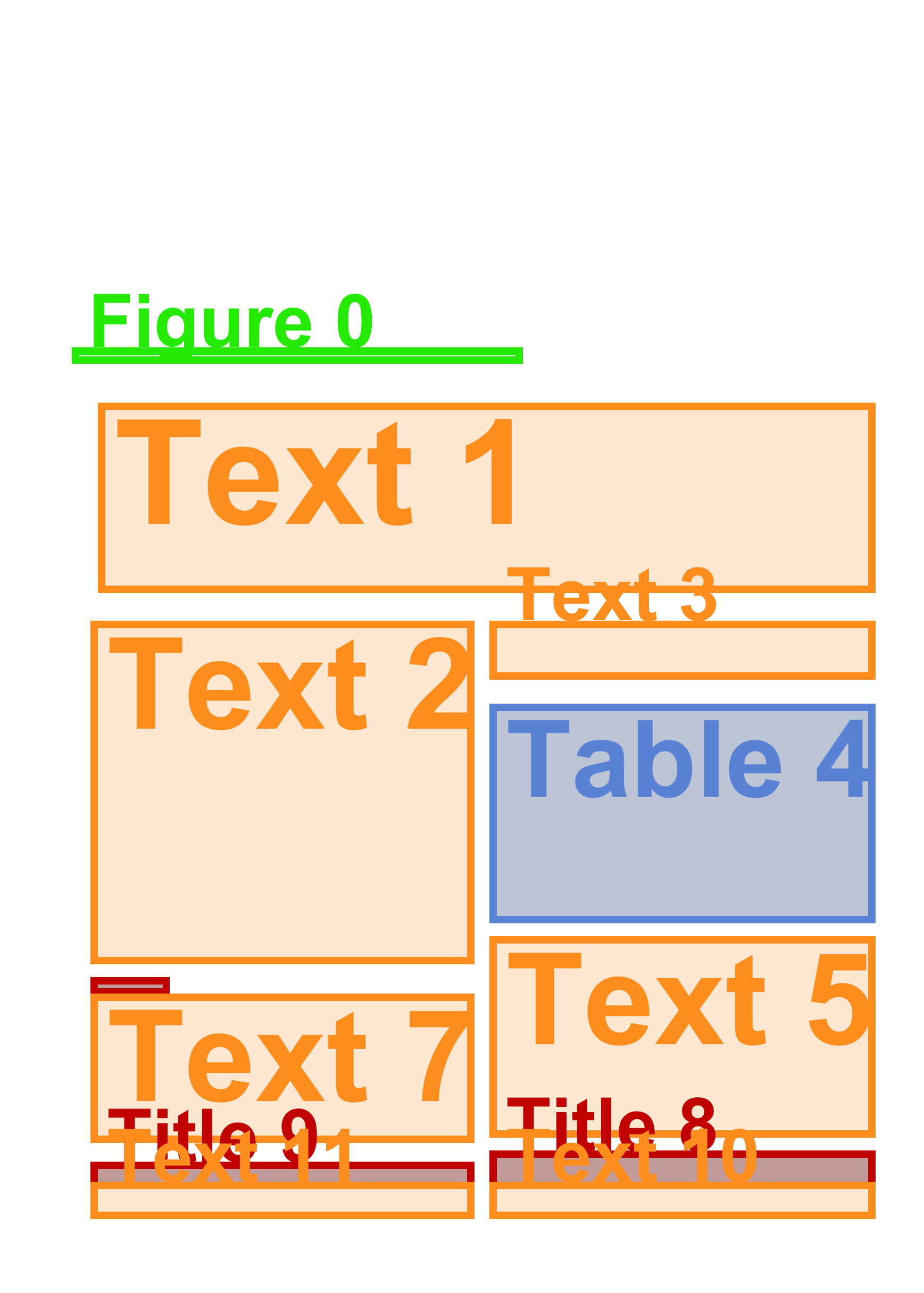} &
\includegraphics[width=\interpolationWidth,frame=0.1pt]{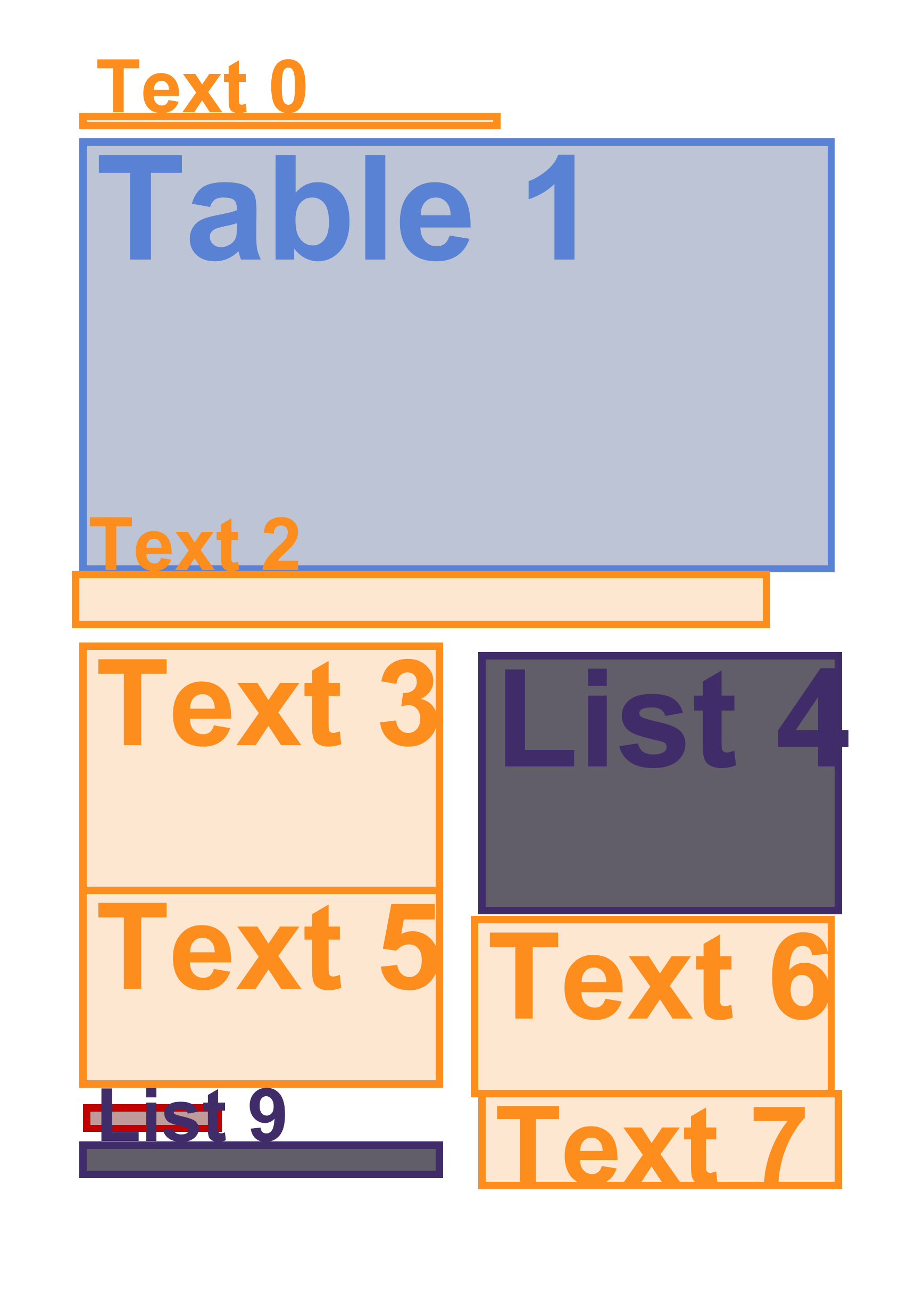} &
\includegraphics[width=\interpolationWidth,frame=0.1pt]{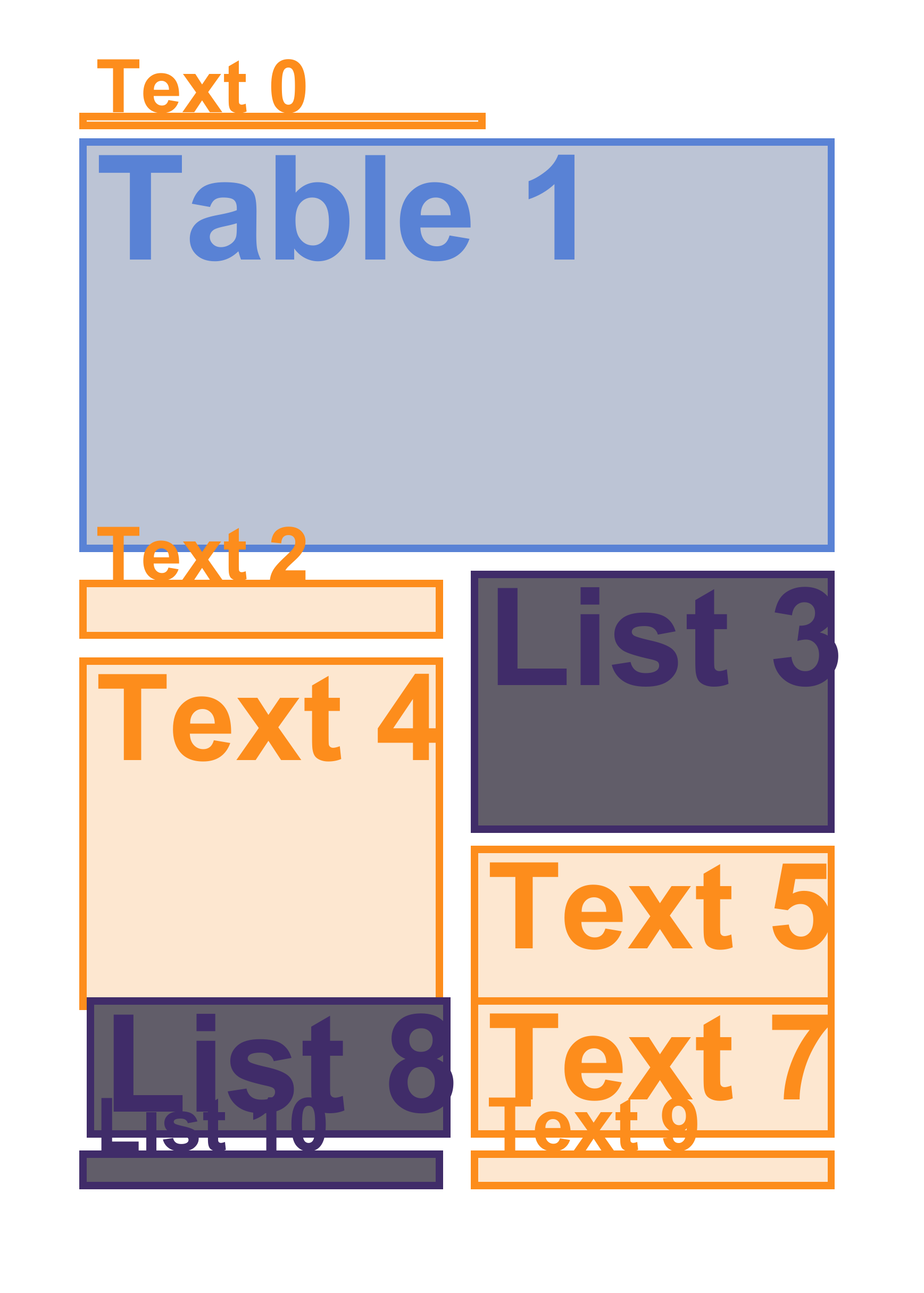} &
\includegraphics[width=\interpolationWidth,frame=0.1pt]{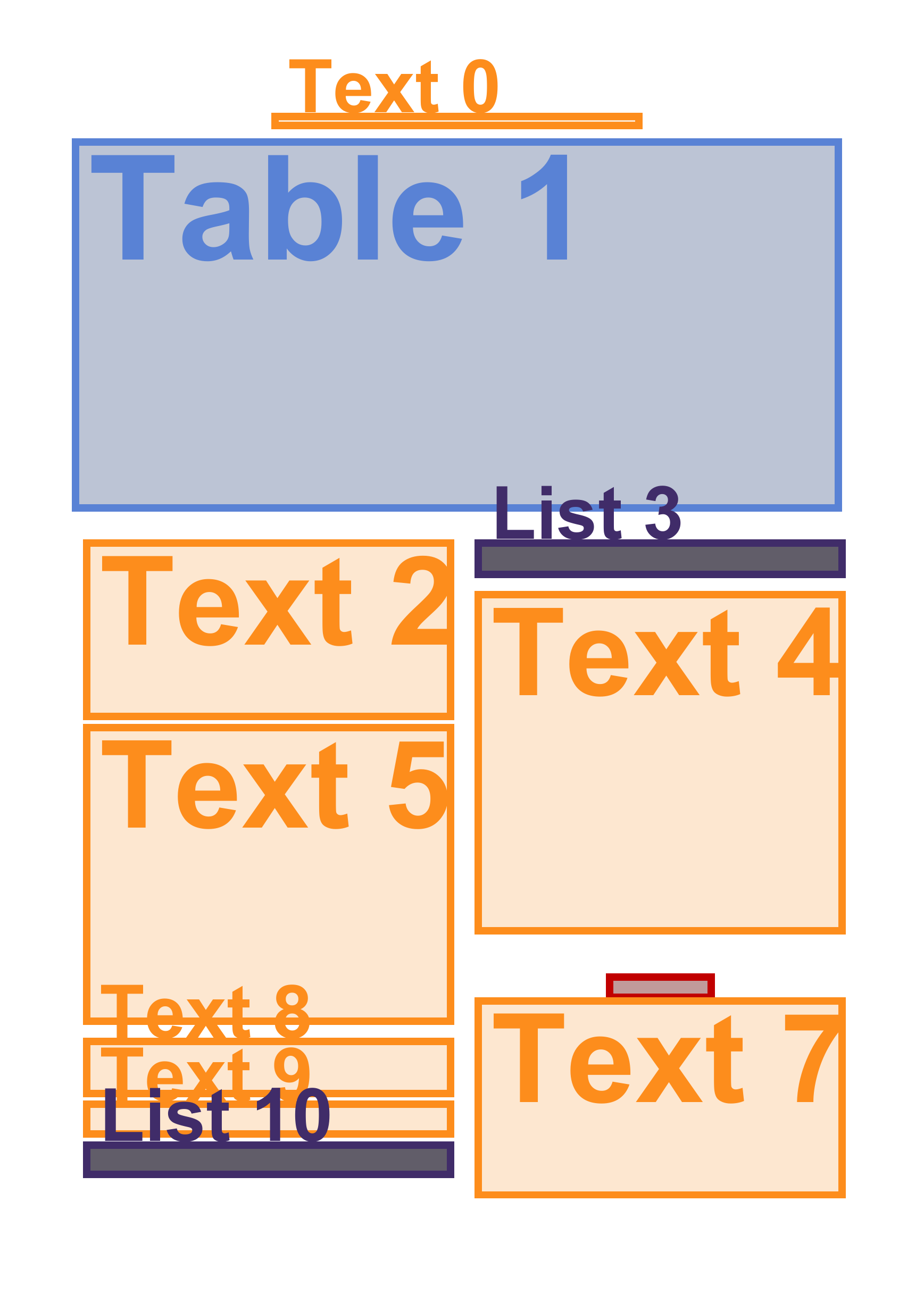} &
\includegraphics[width=\interpolationWidth,frame=0.1pt]{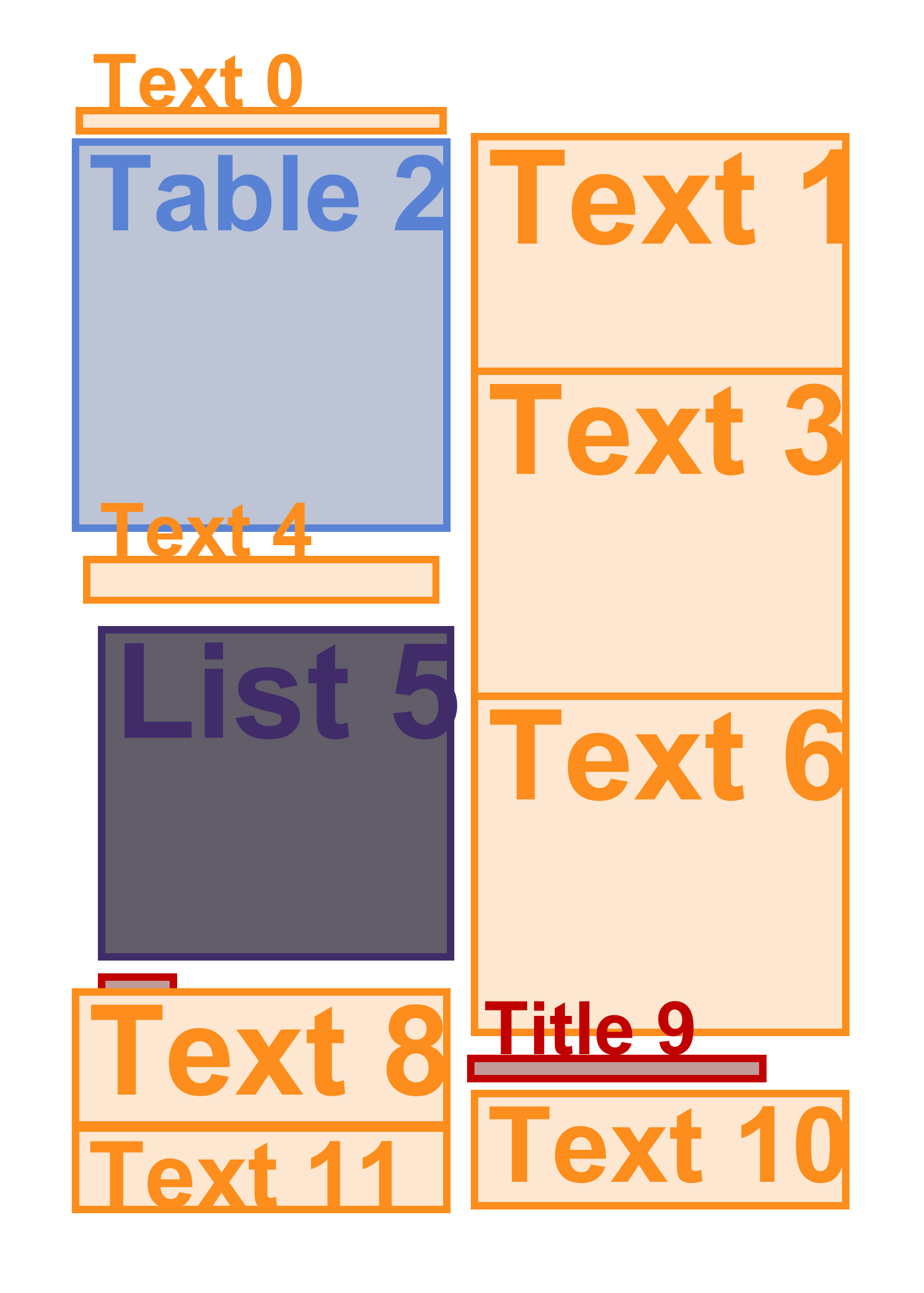} &
\includegraphics[width=\interpolationWidth,frame=0.1pt]{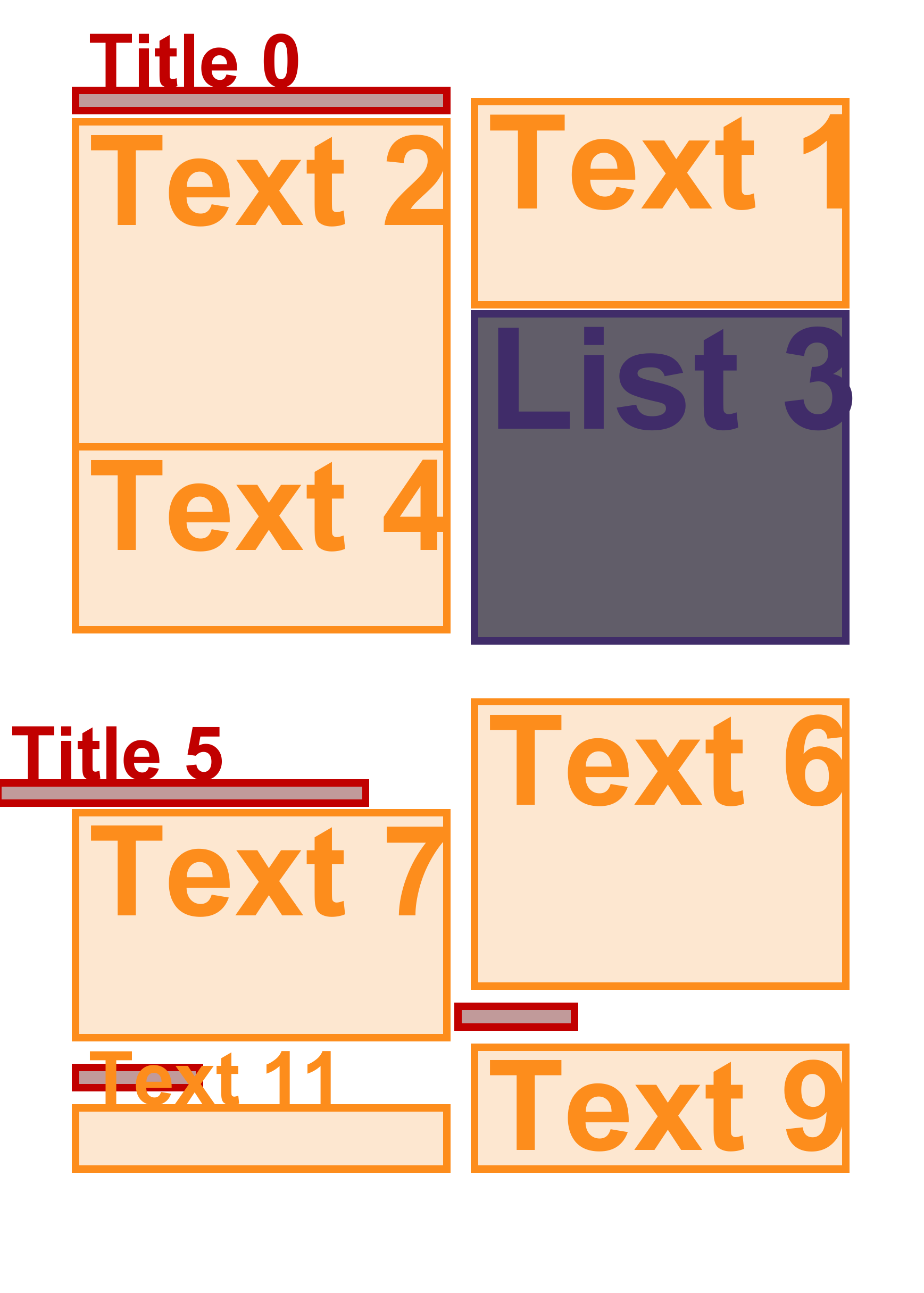} &
\includegraphics[width=\interpolationWidth,frame=0.1pt]{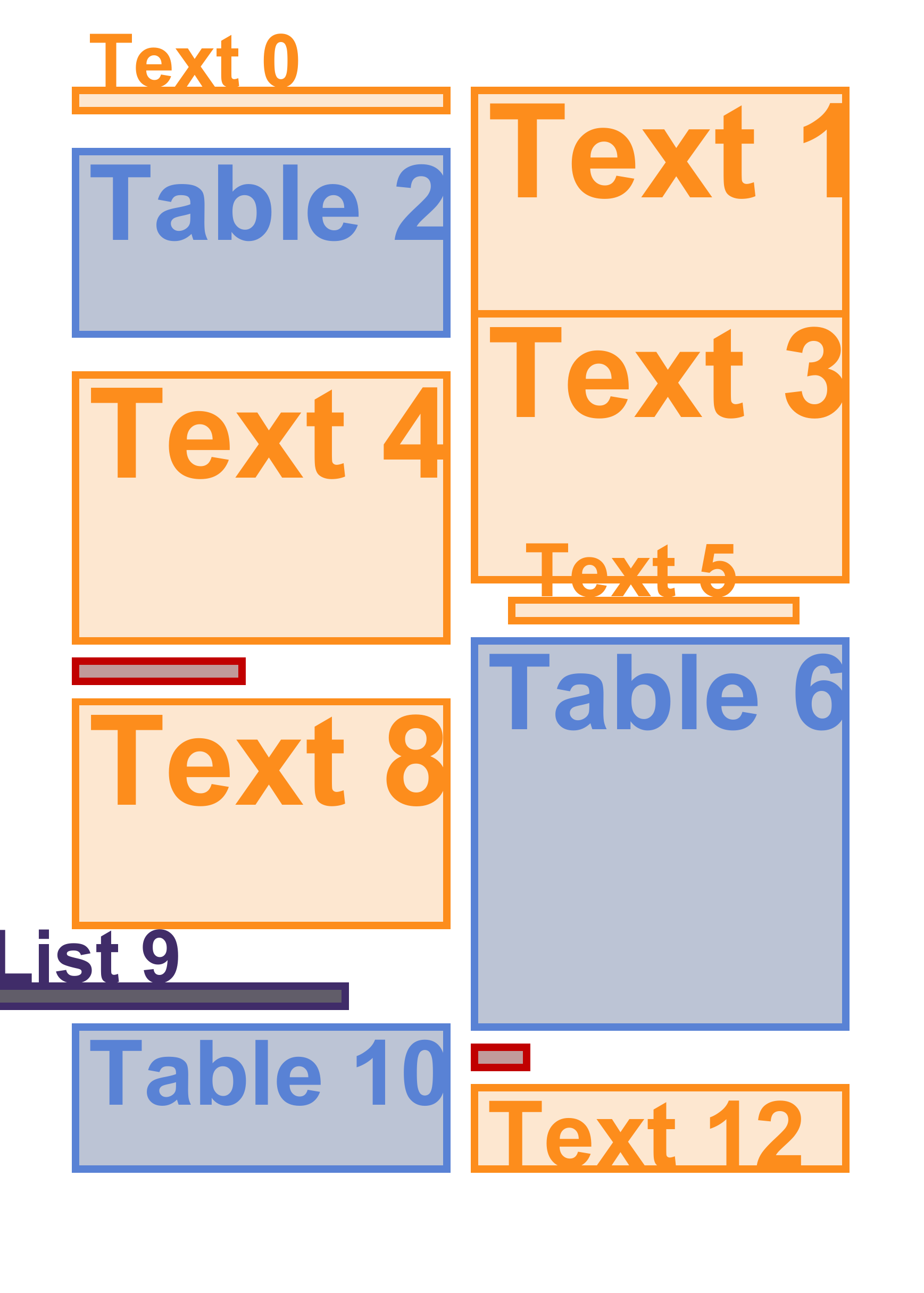} &
\includegraphics[width=\interpolationWidth,frame=0.1pt]{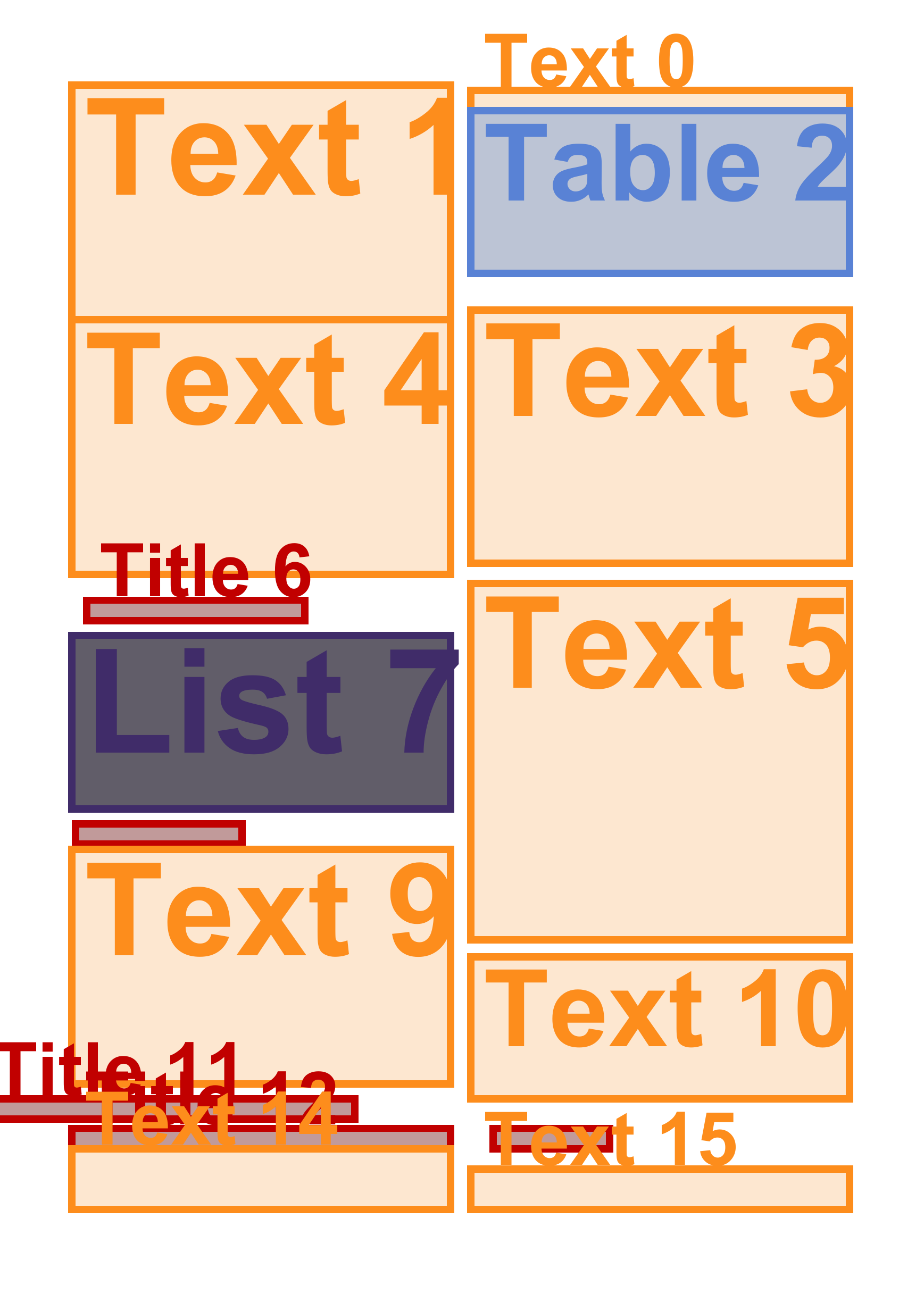} &
\includegraphics[width=\interpolationWidth,frame=0.1pt]{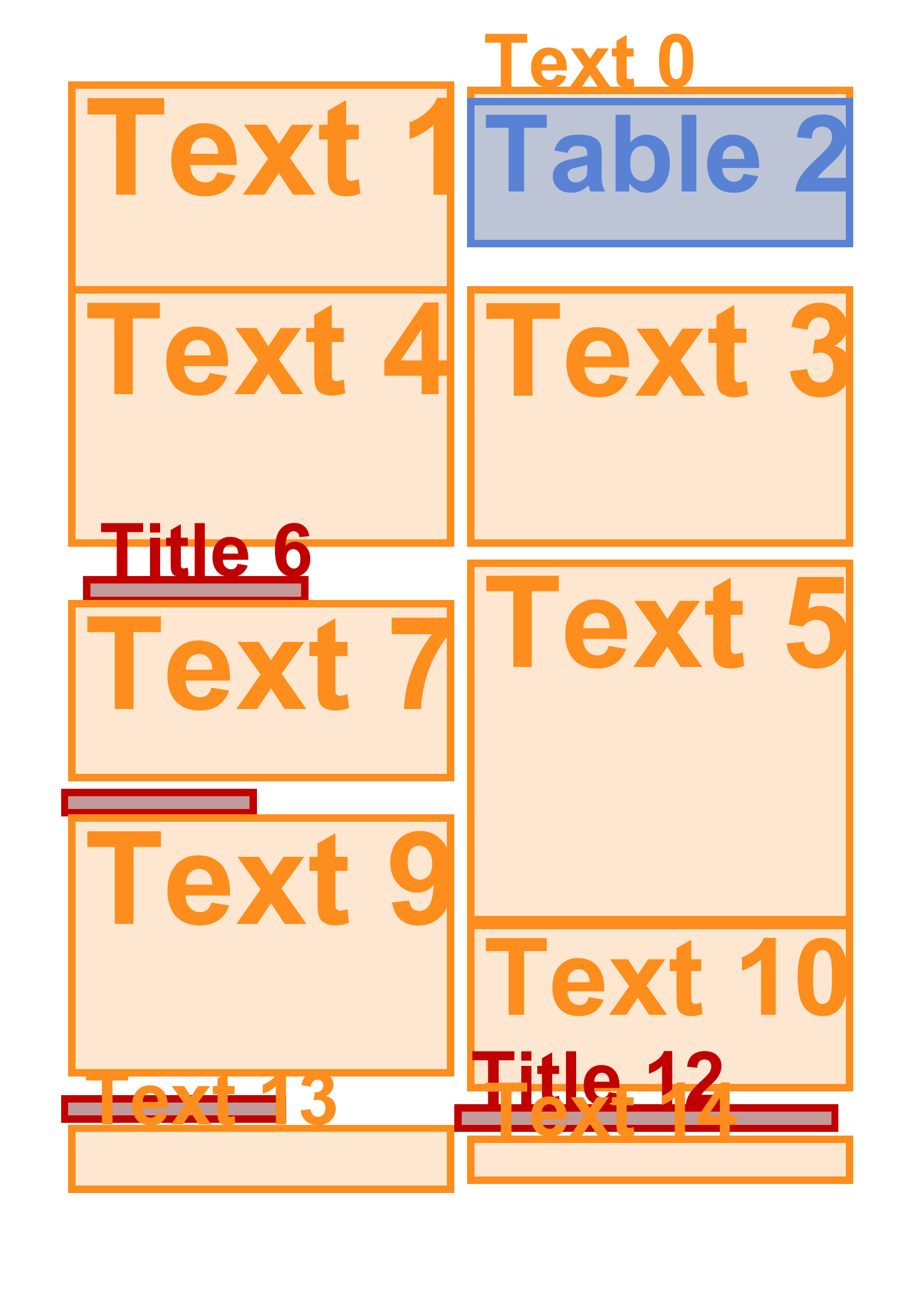} &
\includegraphics[width=\interpolationWidth,frame=0.1pt]{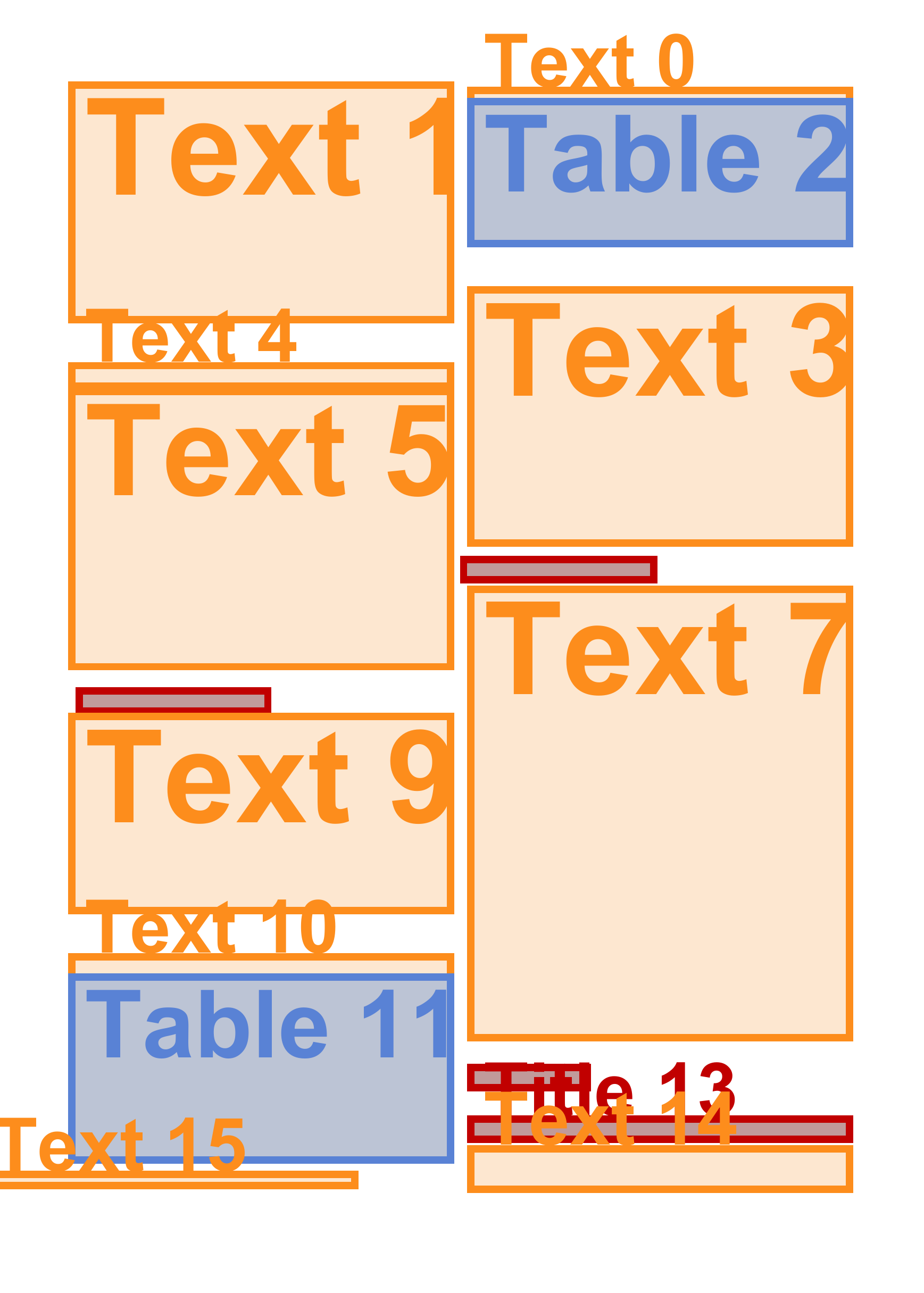} \\

\includegraphics[width=\interpolationWidth,frame=0.1pt]{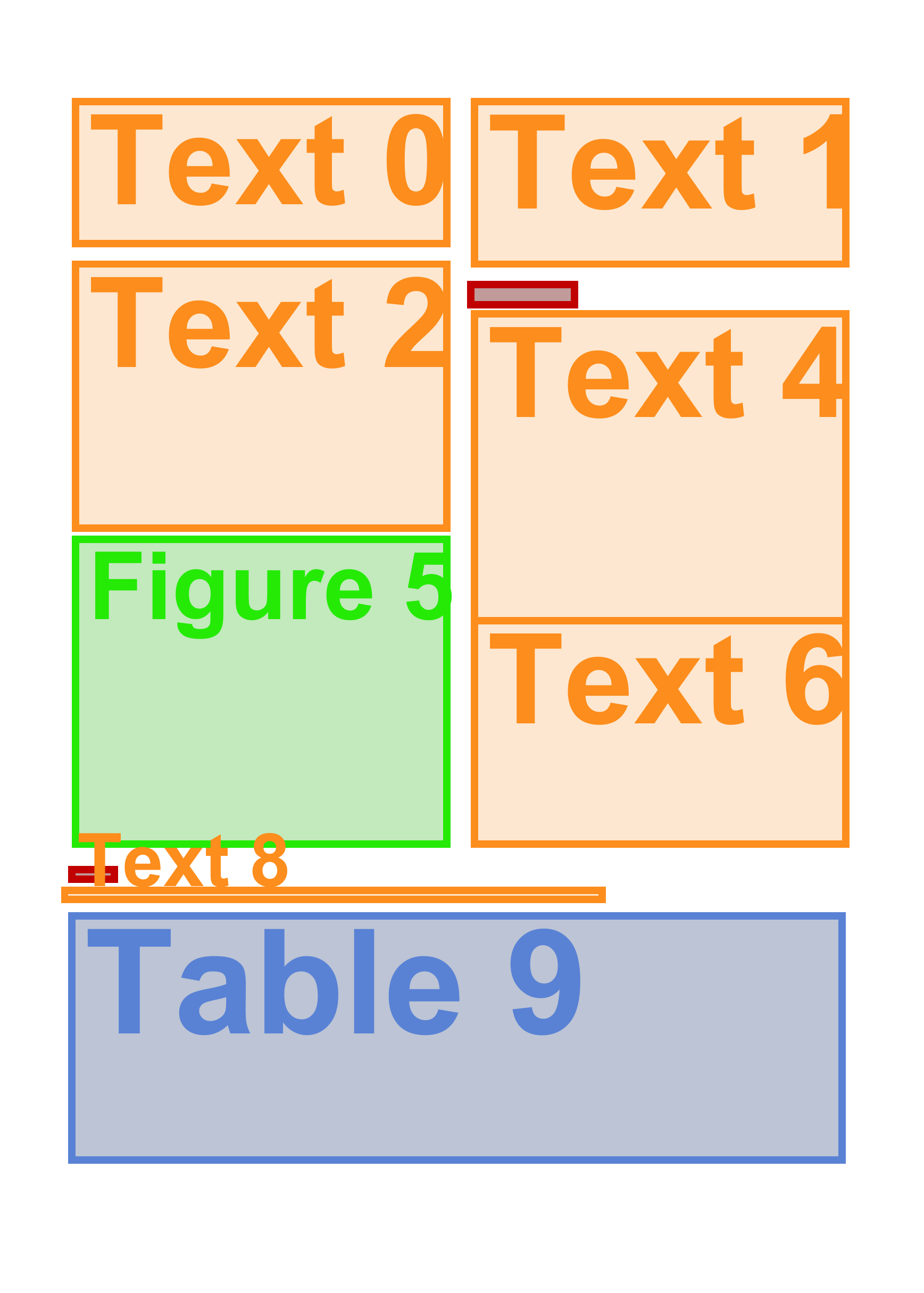} &
\includegraphics[width=\interpolationWidth,frame=0.1pt]{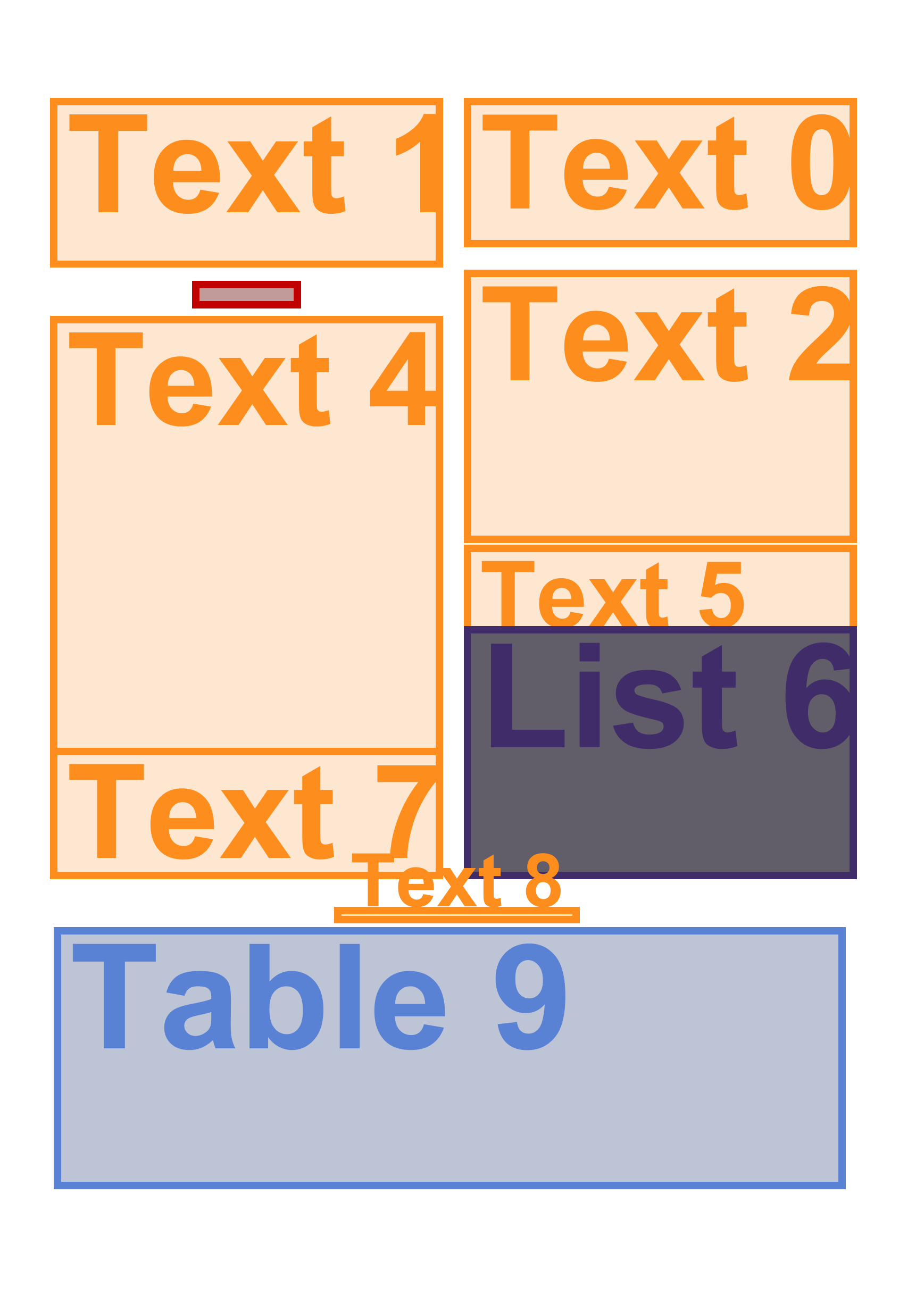} &
\includegraphics[width=\interpolationWidth,frame=0.1pt]{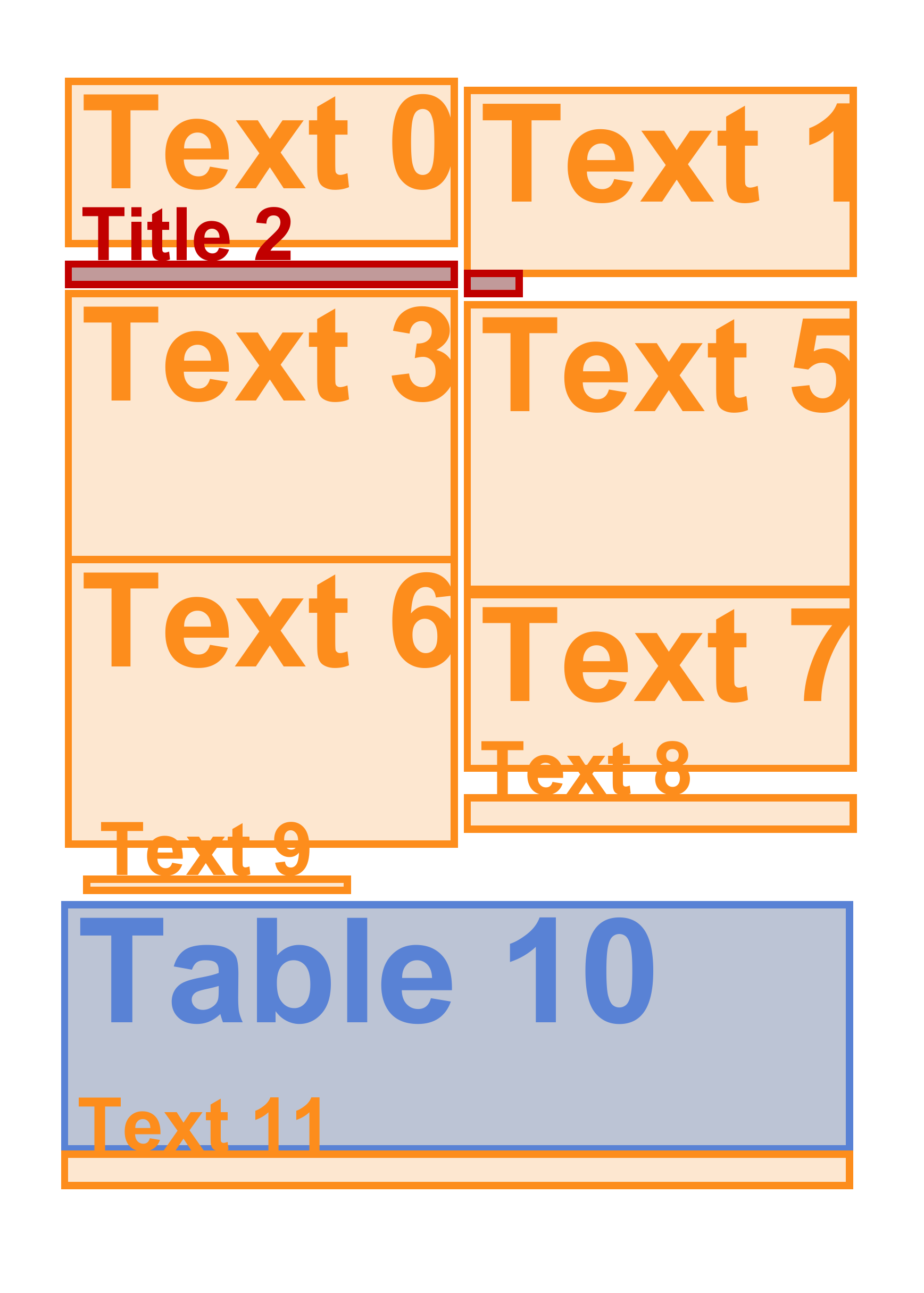} &
\includegraphics[width=\interpolationWidth,frame=0.1pt]{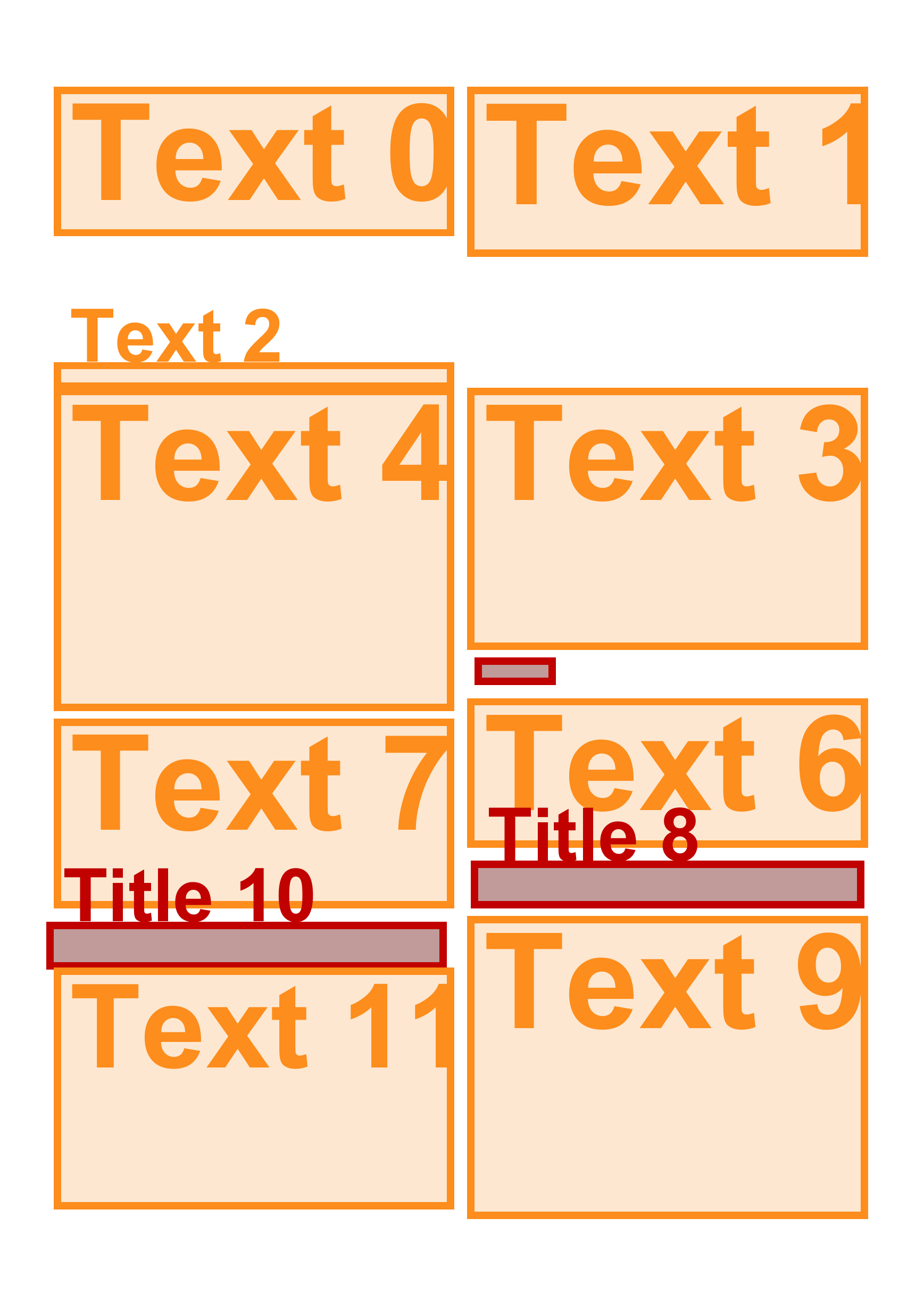} &
\includegraphics[width=\interpolationWidth,frame=0.1pt]{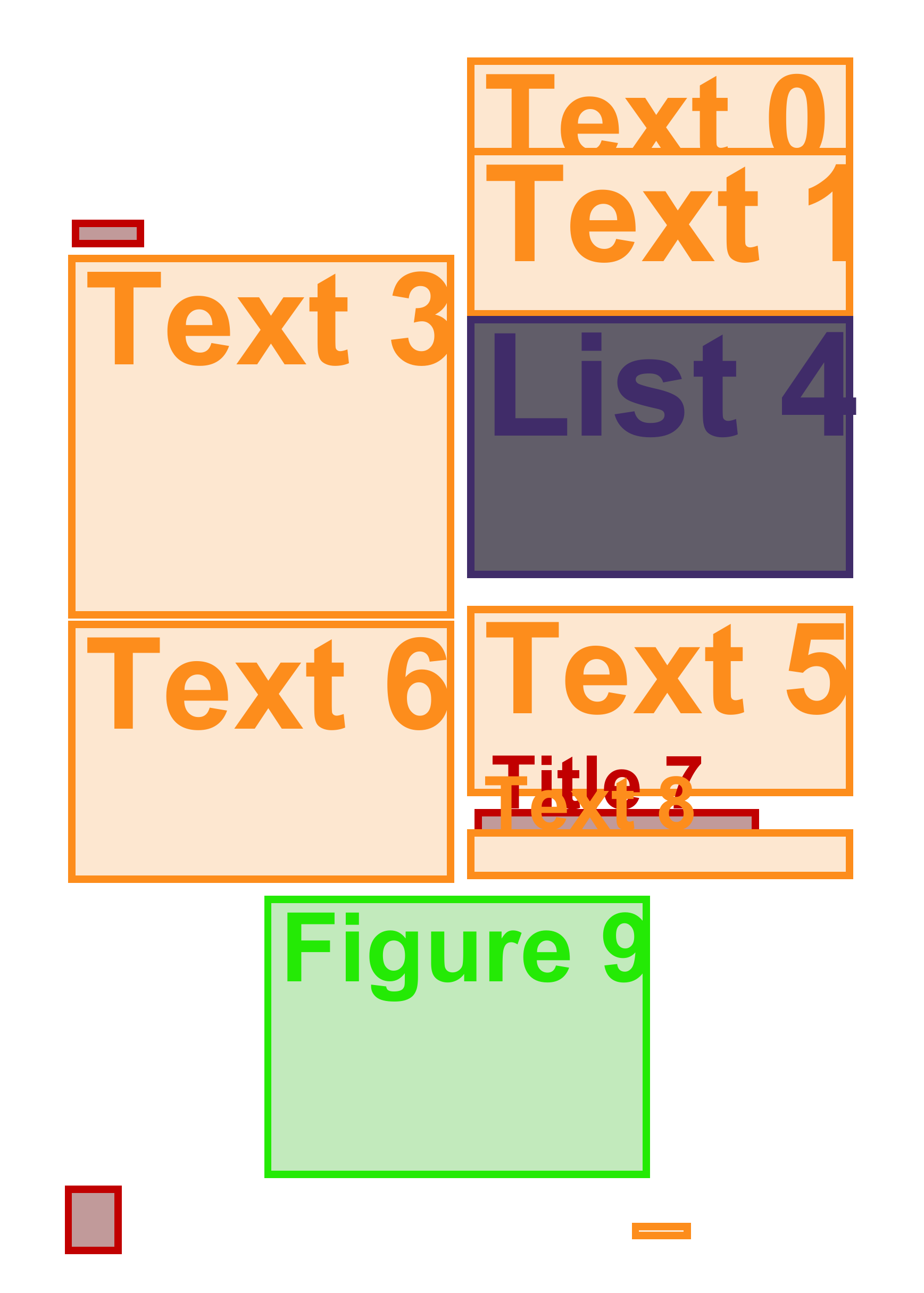} &
\includegraphics[width=\interpolationWidth,frame=0.1pt]{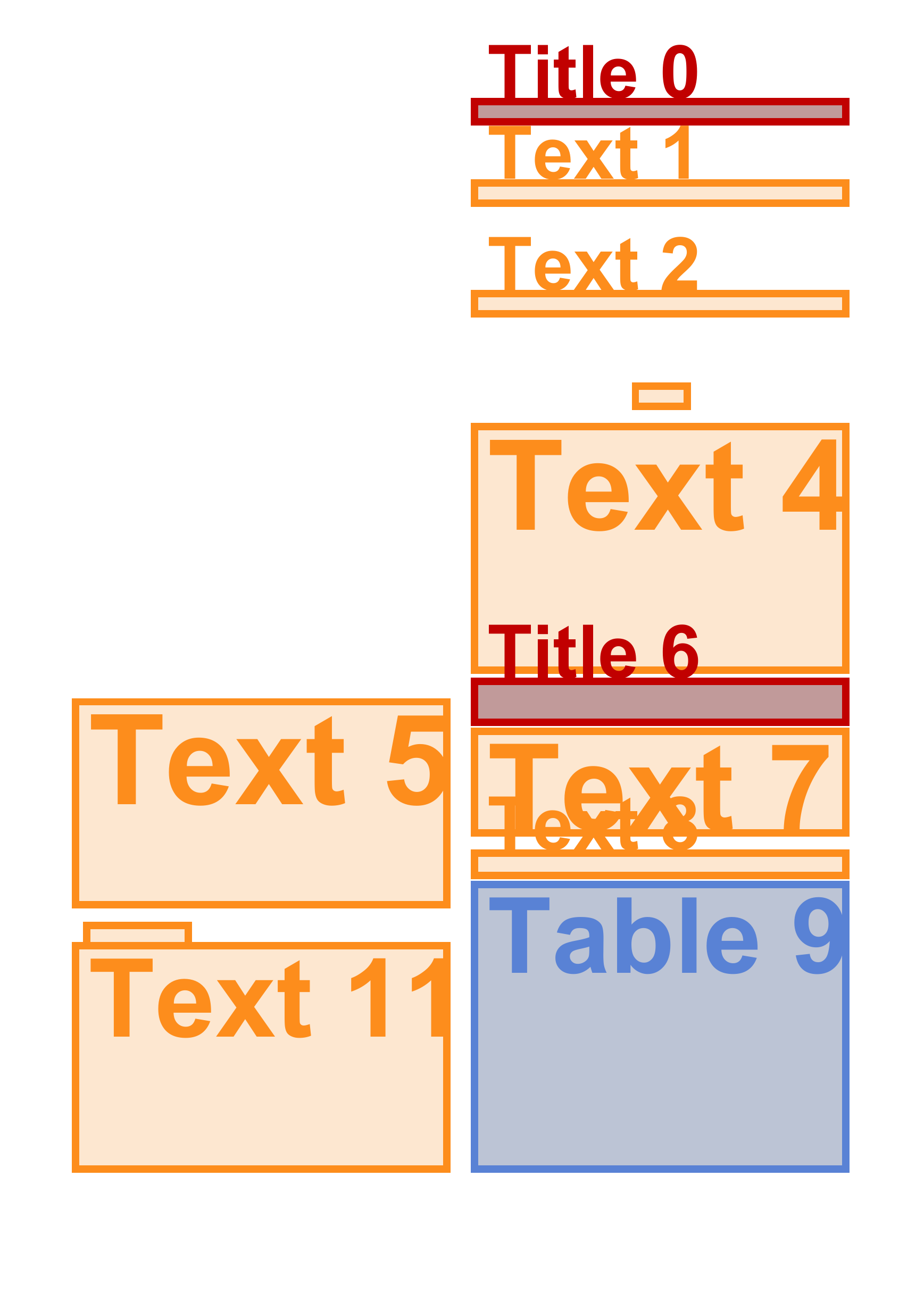} &
\includegraphics[width=\interpolationWidth,frame=0.1pt]{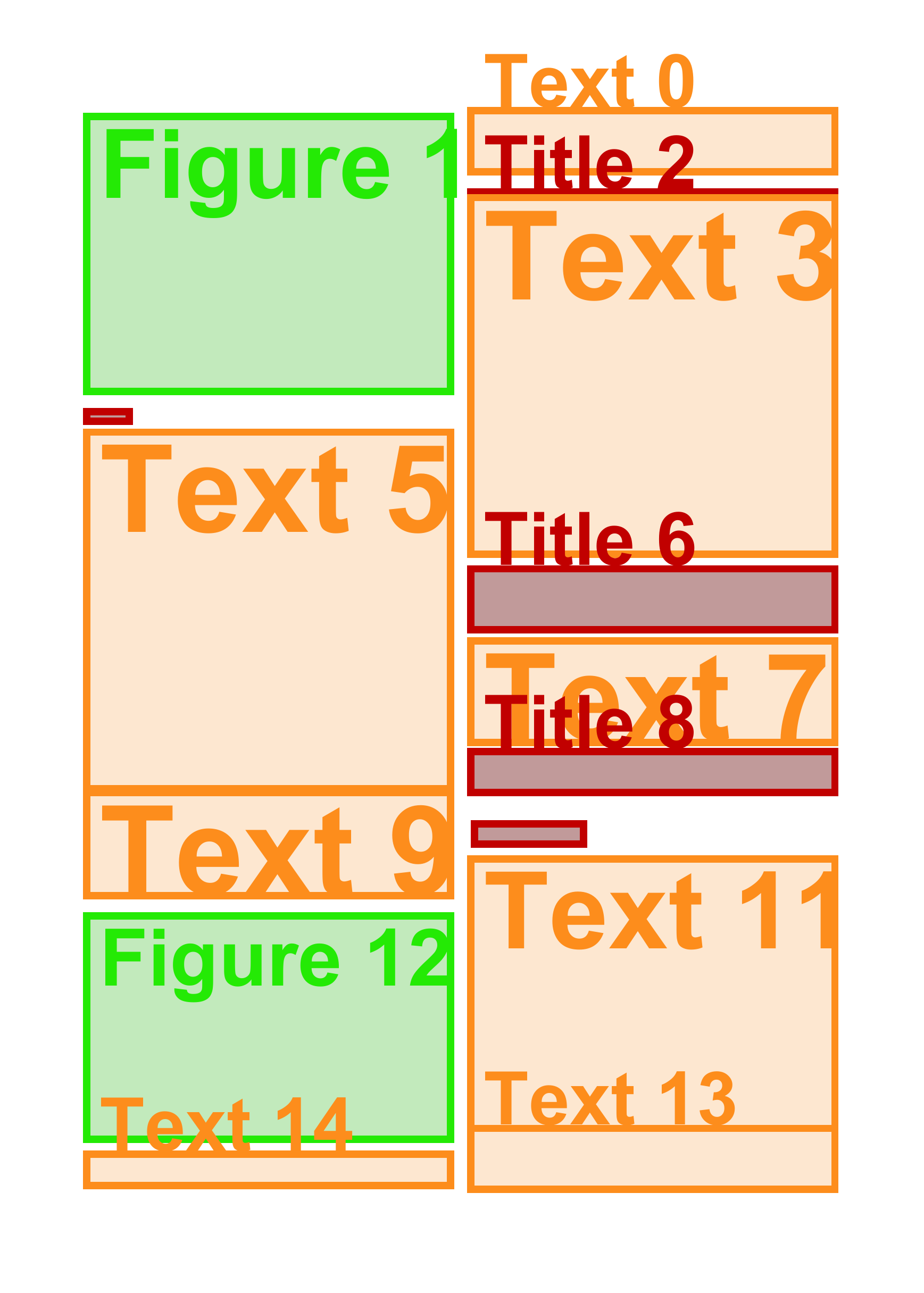} &
\includegraphics[width=\interpolationWidth,frame=0.1pt]{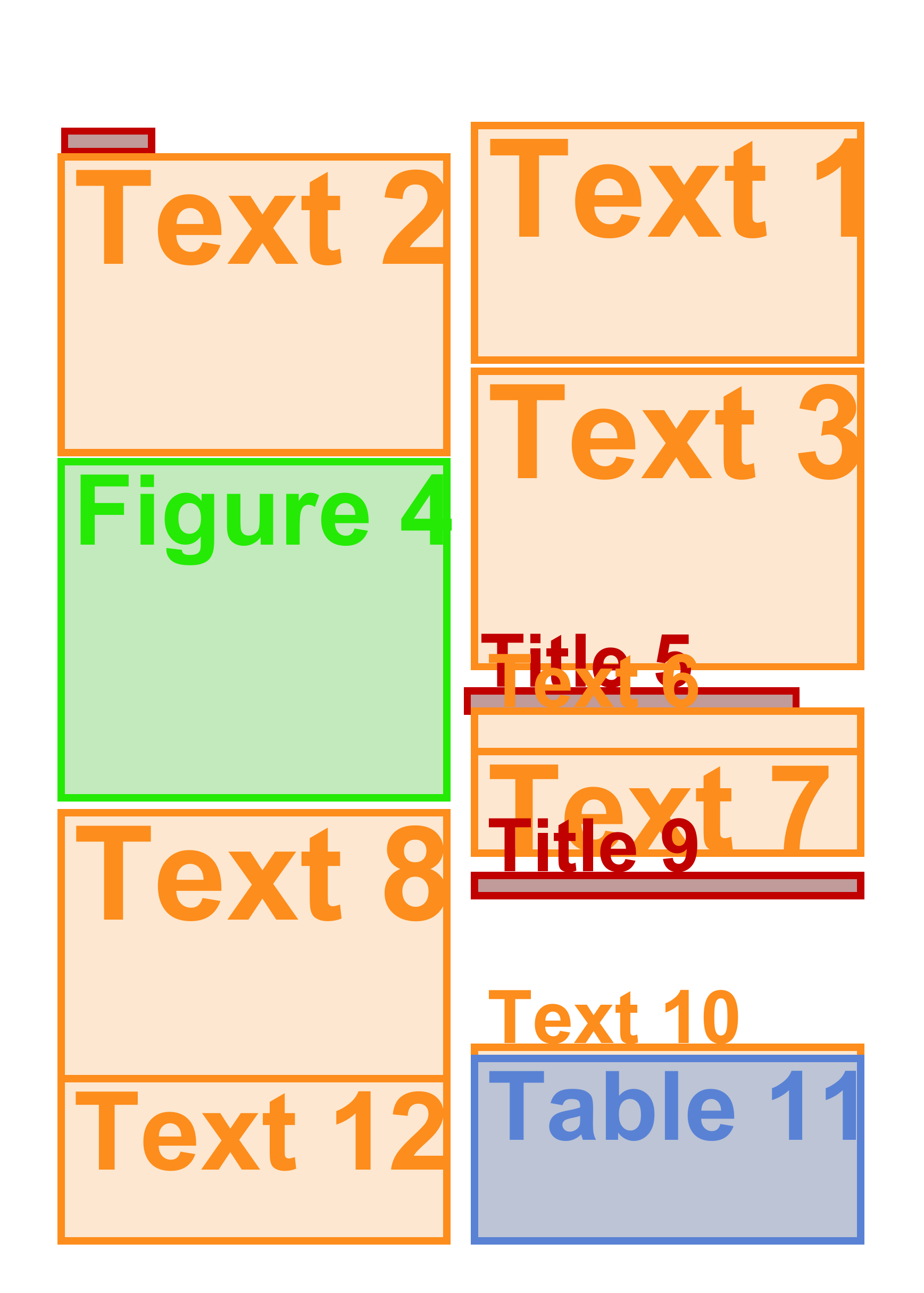} &
\includegraphics[width=\interpolationWidth,frame=0.1pt]{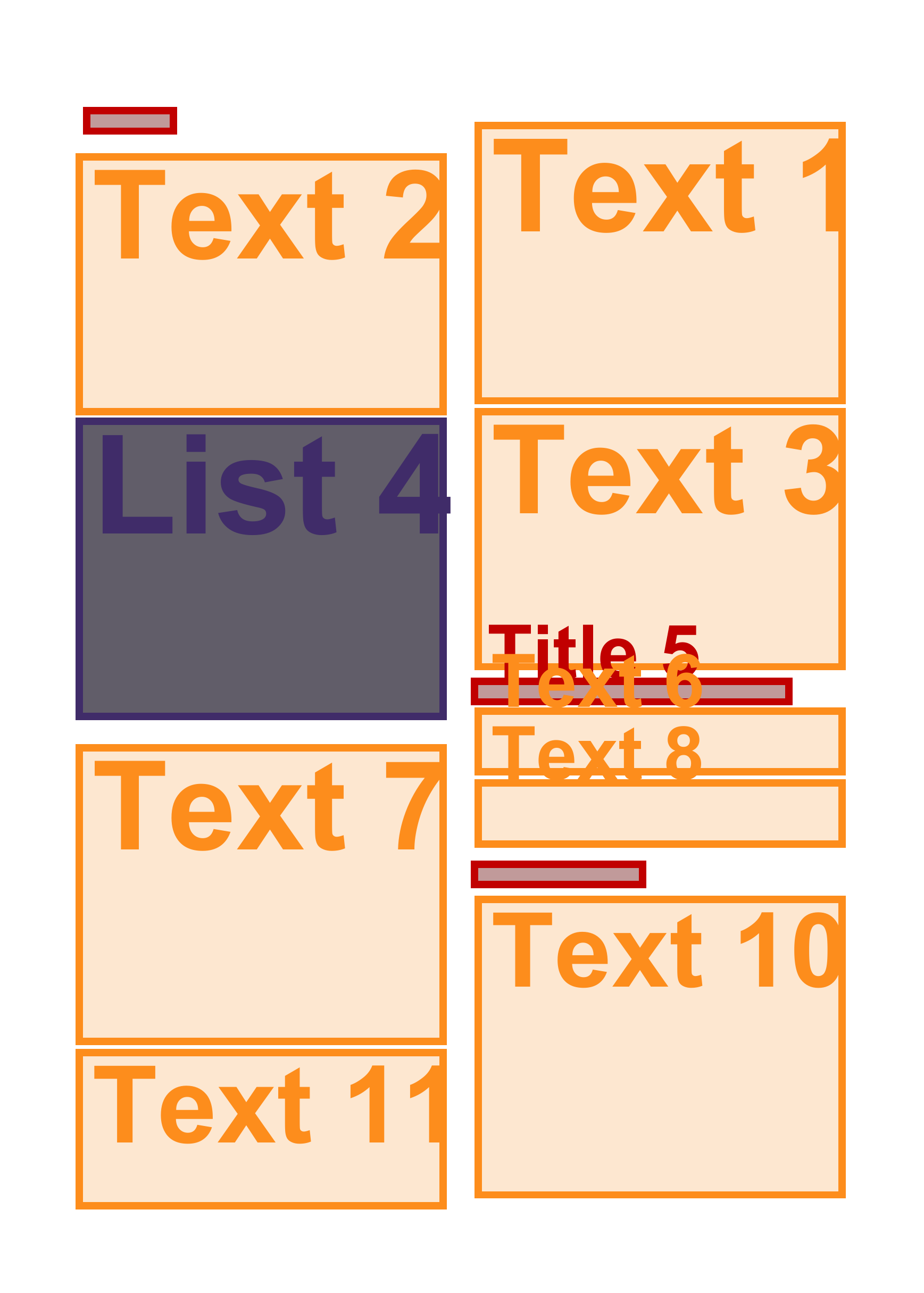} &
\includegraphics[width=\interpolationWidth,frame=0.1pt]{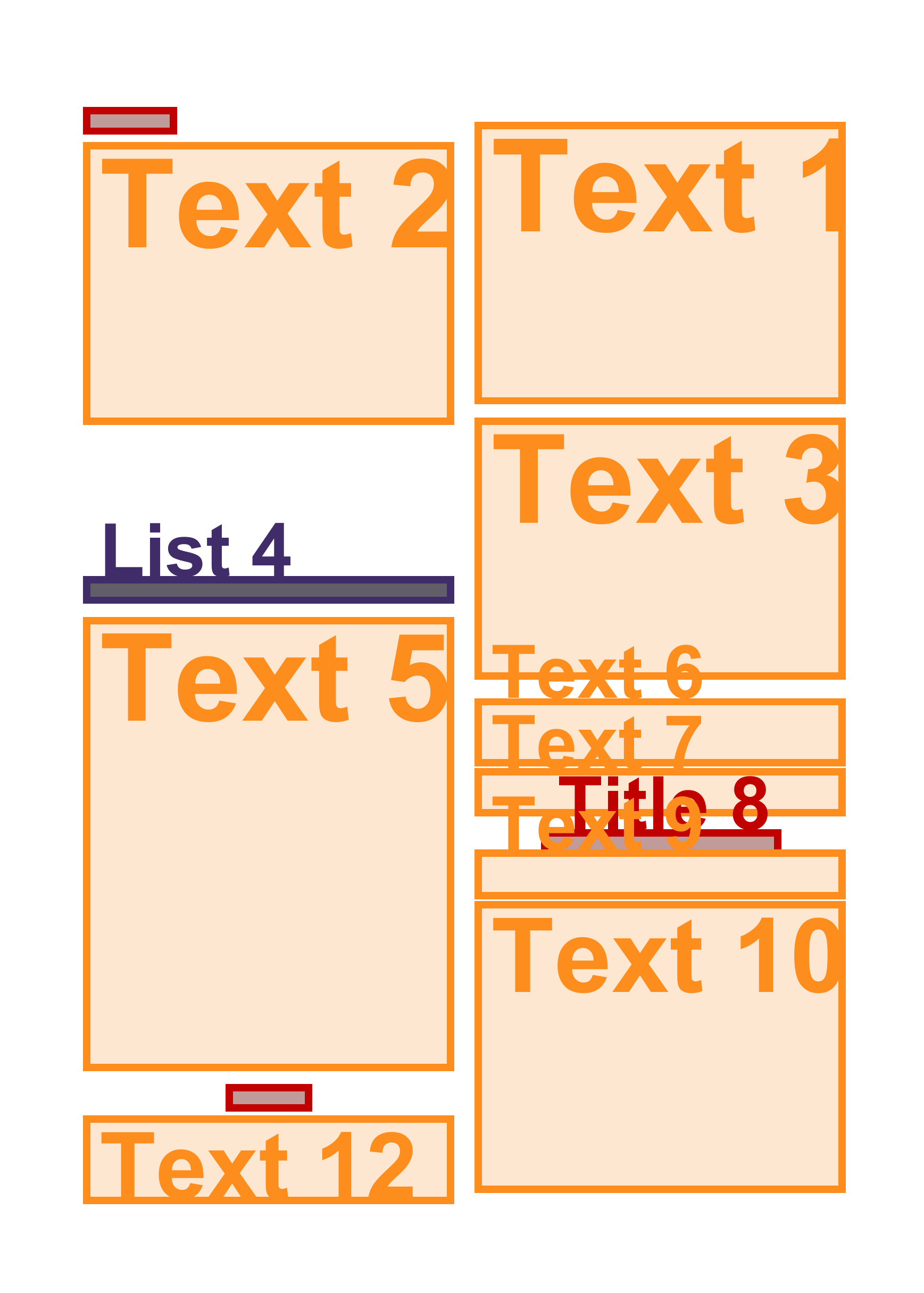} &
\includegraphics[width=\interpolationWidth,frame=0.1pt]{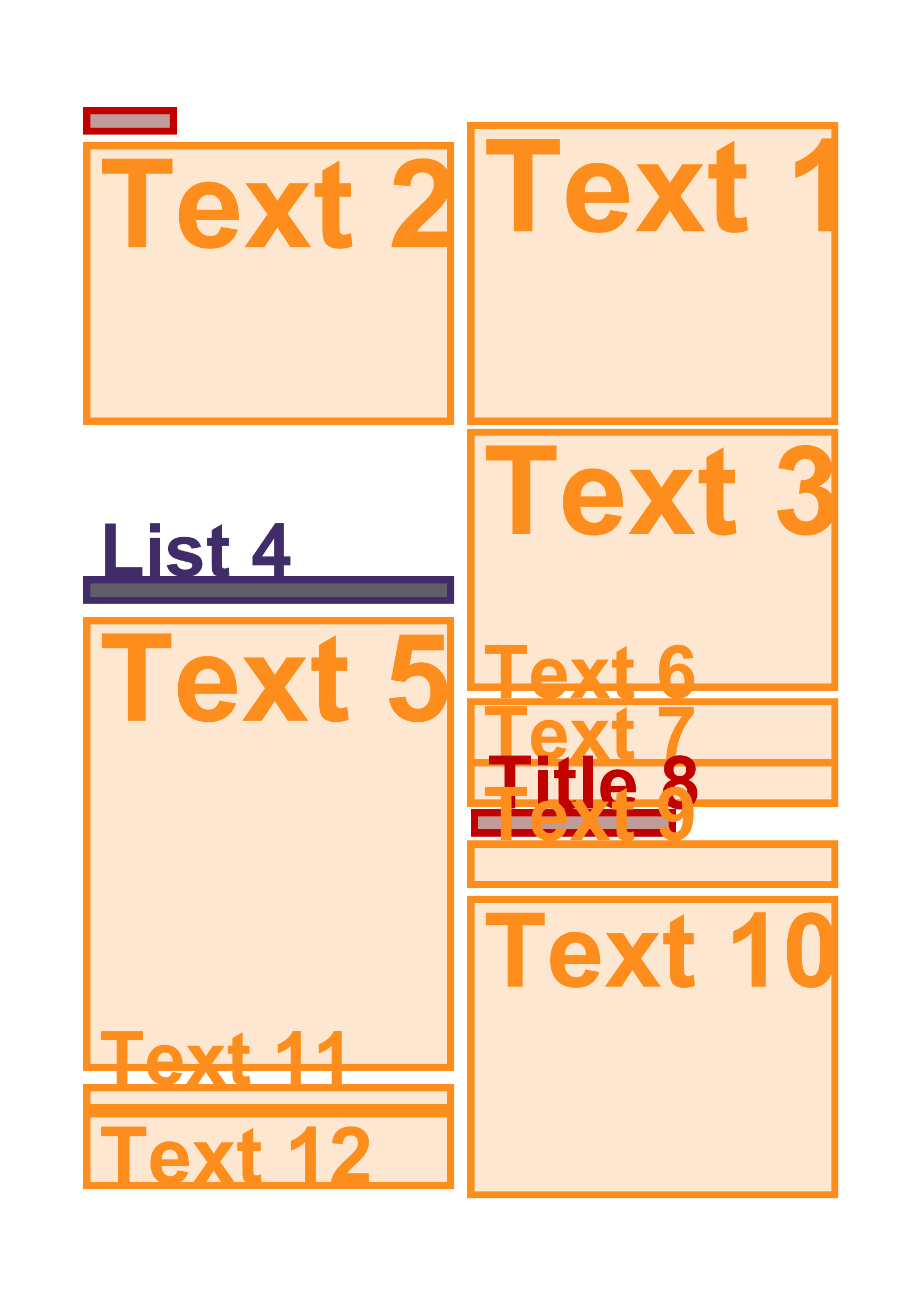} \\

    \end{tabular}
    \caption{Linear interpolations between two vectors in latent space, and evenly-sampled interpolated values.}
    \label{fig:latent_interpolations}
\end{figure}

In fig. \ref{fig:latent_interpolations} we show linear interpolations between two random vectors $z_1,z_2\sim\mathcal{N}(0, 1)$ and the intermediate results between them on PubLayNet. While the space is not perfectly smooth (some elements only appear in the intermediate samples), the results are not completely arbitrary, and each intermediate value $z' = z_1 + \lambda \cdot (z_2 - z_1), \lambda \in (0, 1)$ produces a valid output.

\clearpage
\onecolumn
\subsection{Adding/Removing Latent Vectors using Non-Autoregressive Decoder}

\begin{figure}[h]
    \newlength{\latentSpaceWidth}
    \setlength{\latentSpaceWidth}{0.15\linewidth}
    \setlength{\tabcolsep}{2pt}
    \centering
    \begin{tabular}{cccc}
    \includegraphics[width=\latentSpaceWidth]{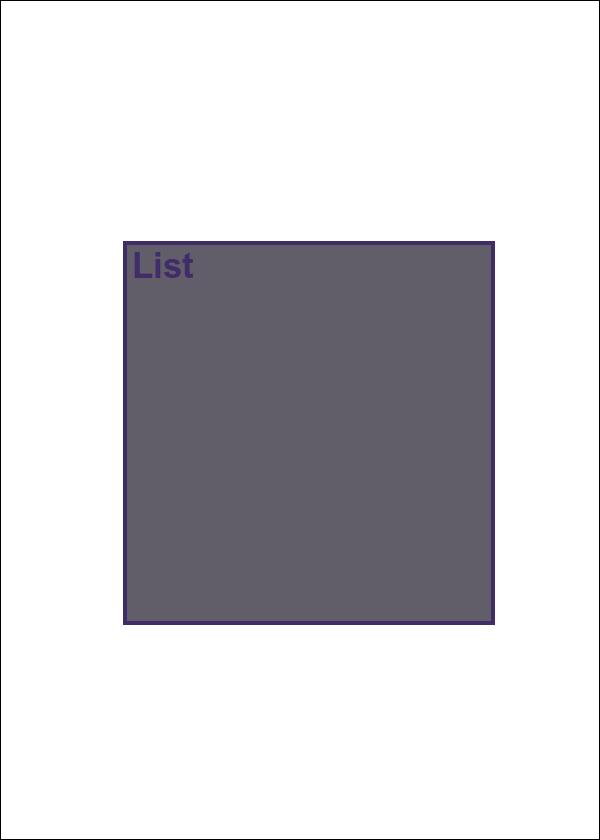} & 
    \includegraphics[width=\latentSpaceWidth]{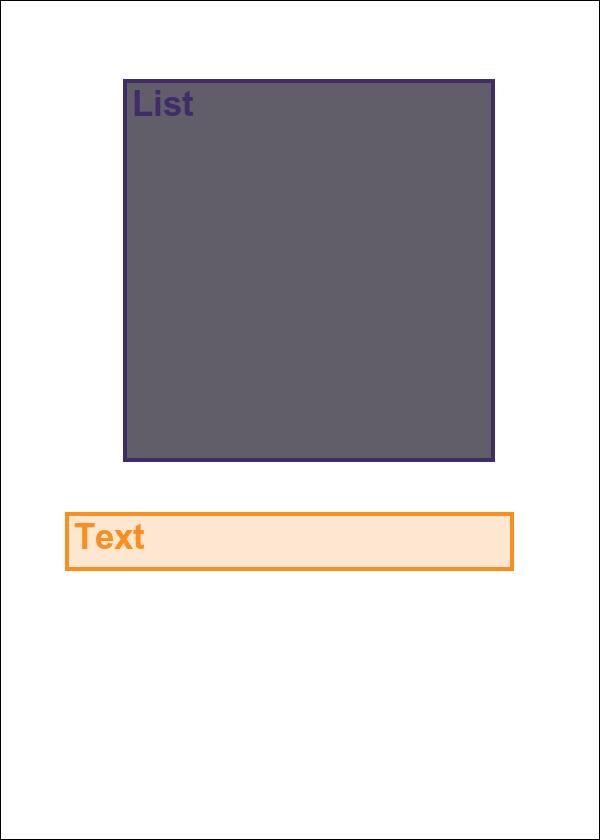} & 
    \includegraphics[width=\latentSpaceWidth]{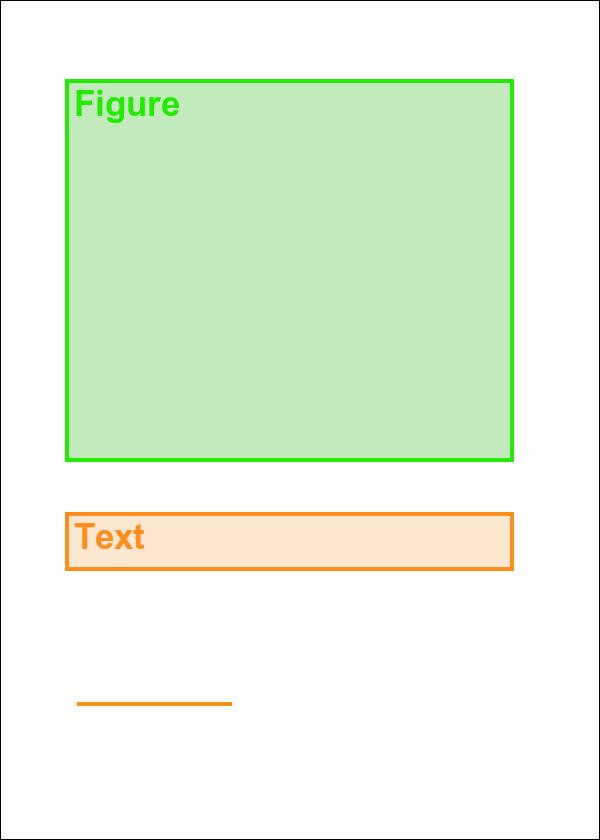} & 
    \includegraphics[width=\latentSpaceWidth]{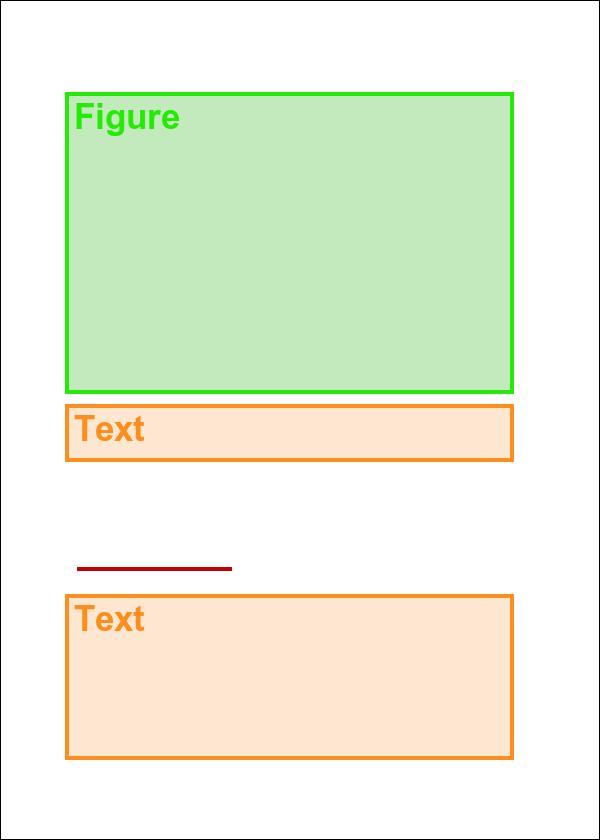} \\\hline
    \includegraphics[width=\latentSpaceWidth]{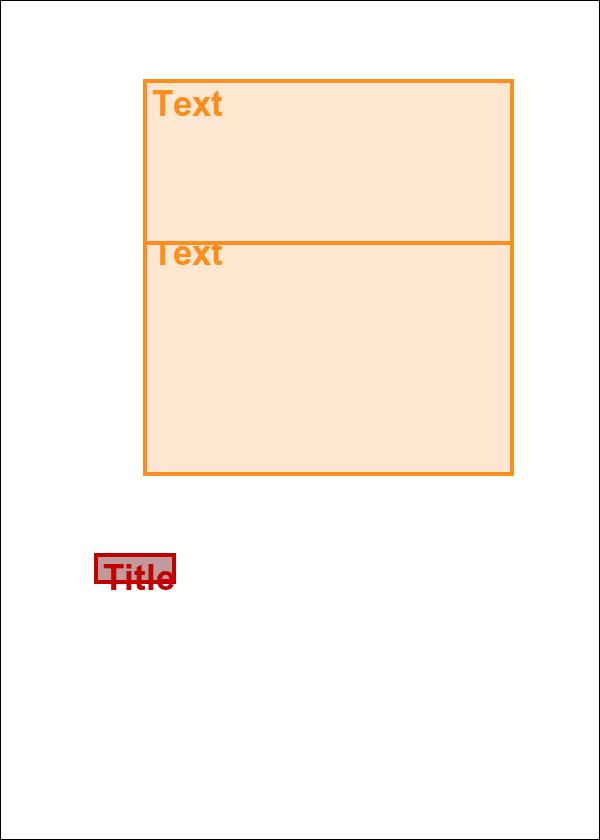} & 
    \includegraphics[width=\latentSpaceWidth]{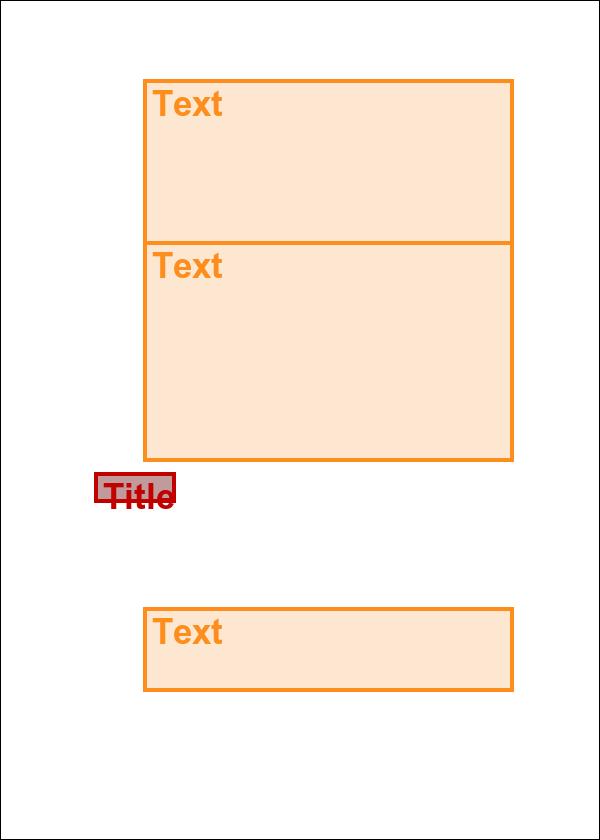} & 
    \includegraphics[width=\latentSpaceWidth]{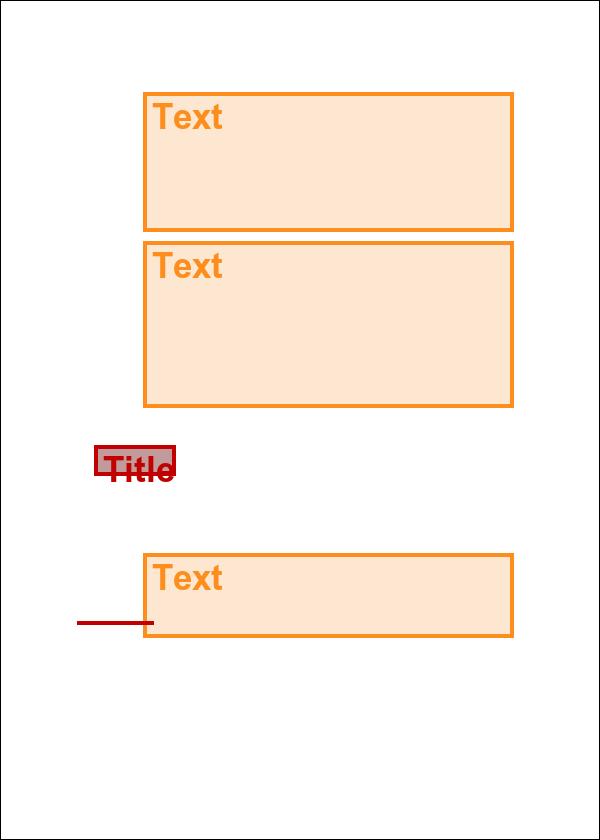} & 
    \includegraphics[width=\latentSpaceWidth]{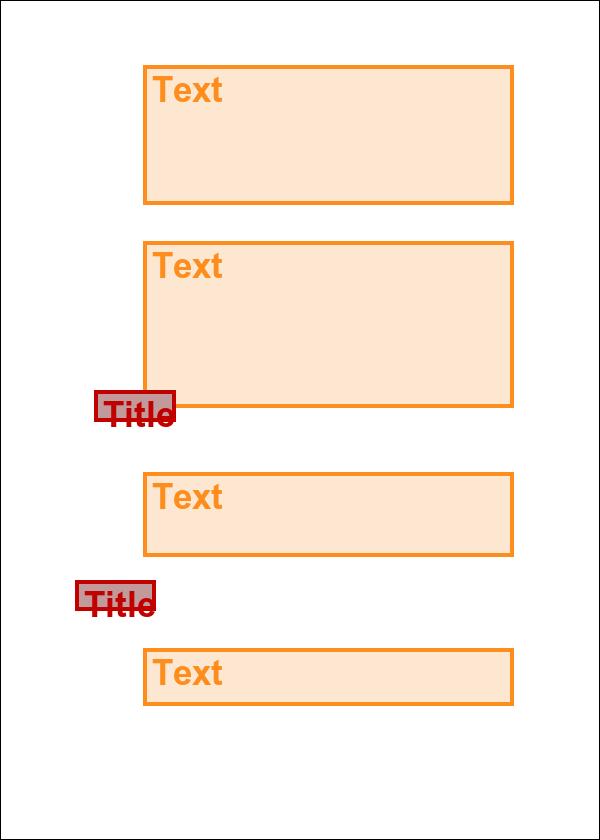} \\\hline
    \includegraphics[width=\latentSpaceWidth]{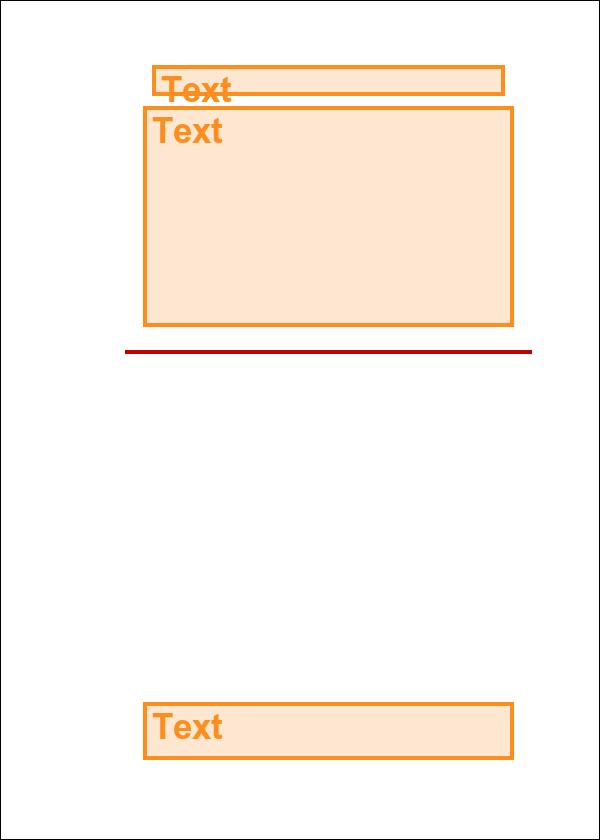} & 
    \includegraphics[width=\latentSpaceWidth]{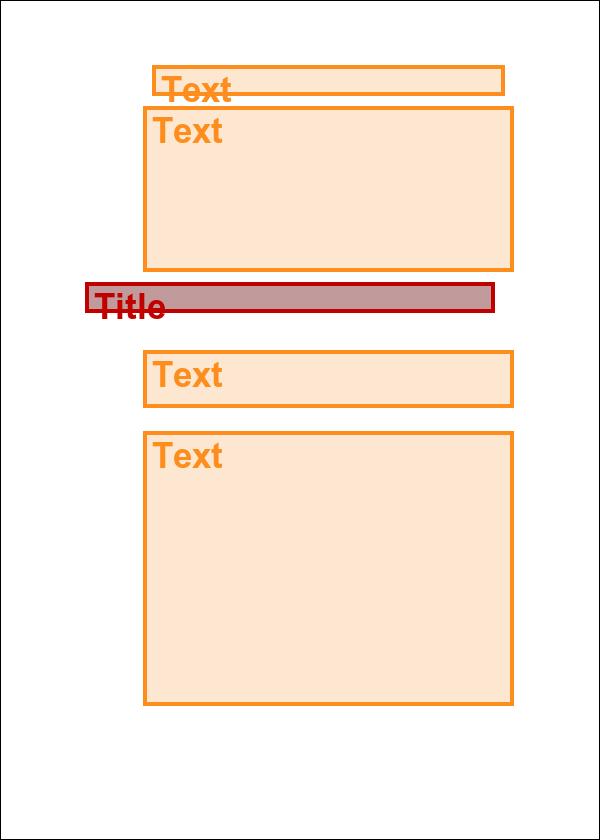} & 
    \includegraphics[width=\latentSpaceWidth]{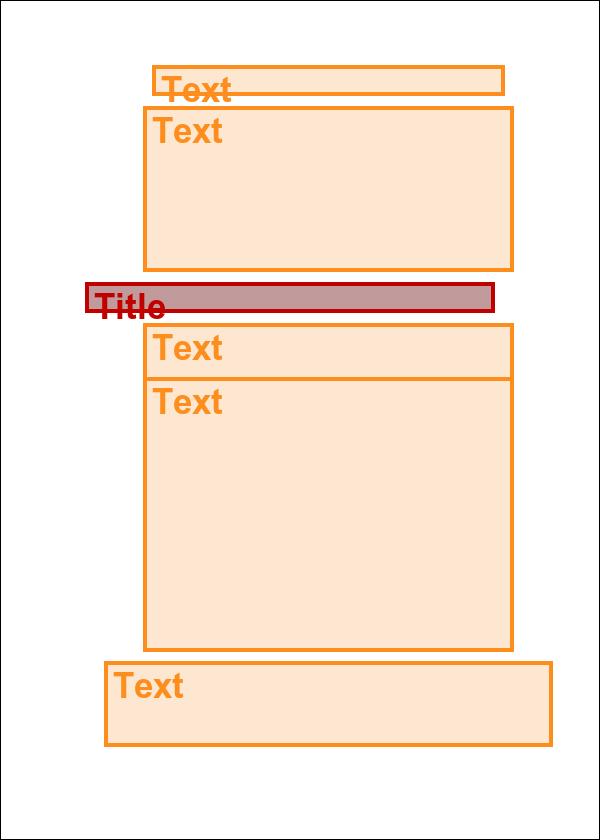} & 
    \includegraphics[width=\latentSpaceWidth]{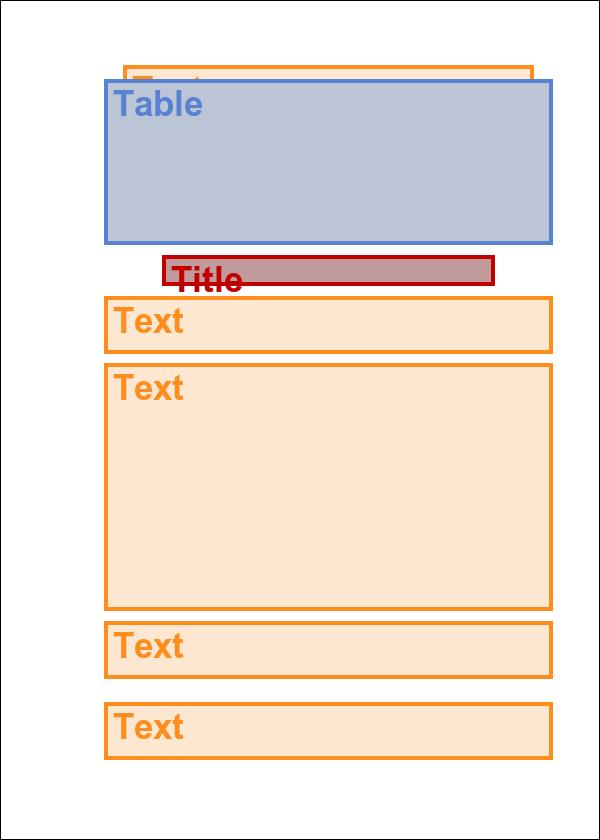} \\\hline
    \includegraphics[width=\latentSpaceWidth]{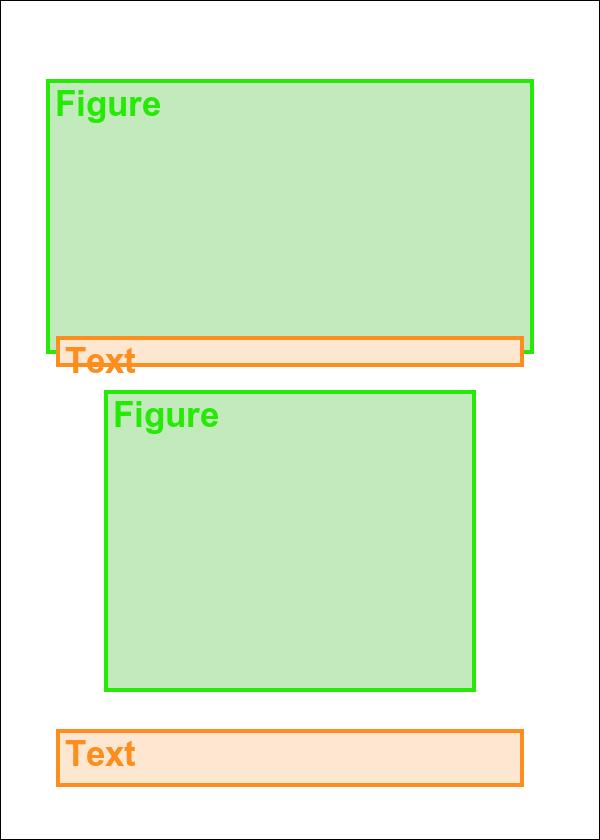} & 
    \includegraphics[width=\latentSpaceWidth]{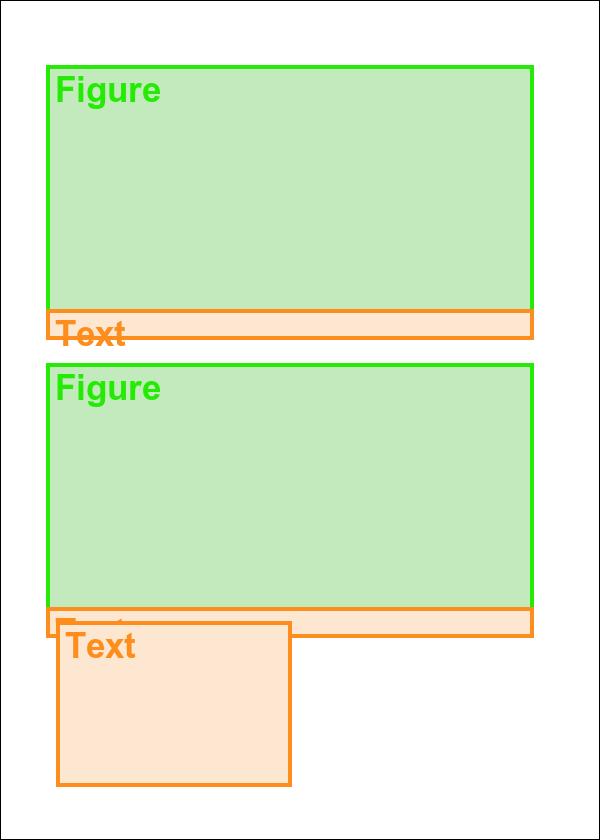} & 
    \includegraphics[width=\latentSpaceWidth]{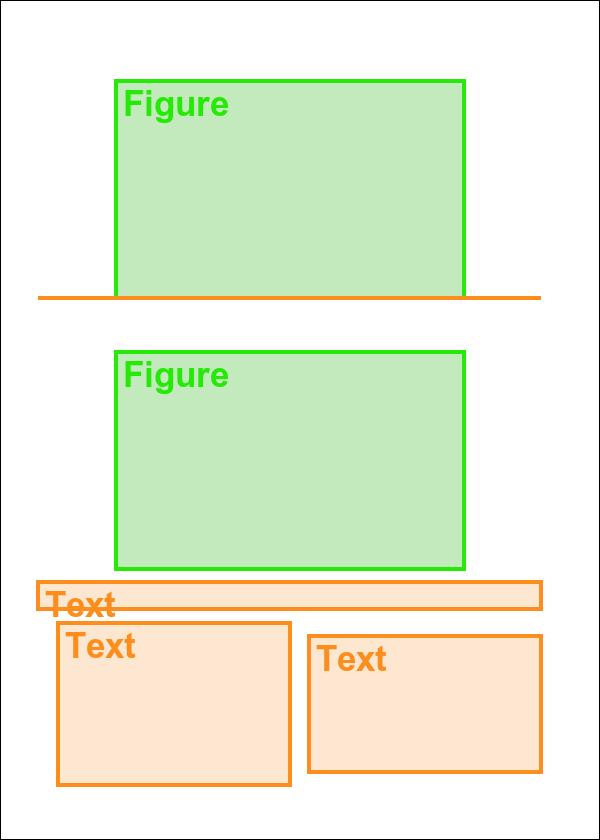} & 
    \includegraphics[width=\latentSpaceWidth]{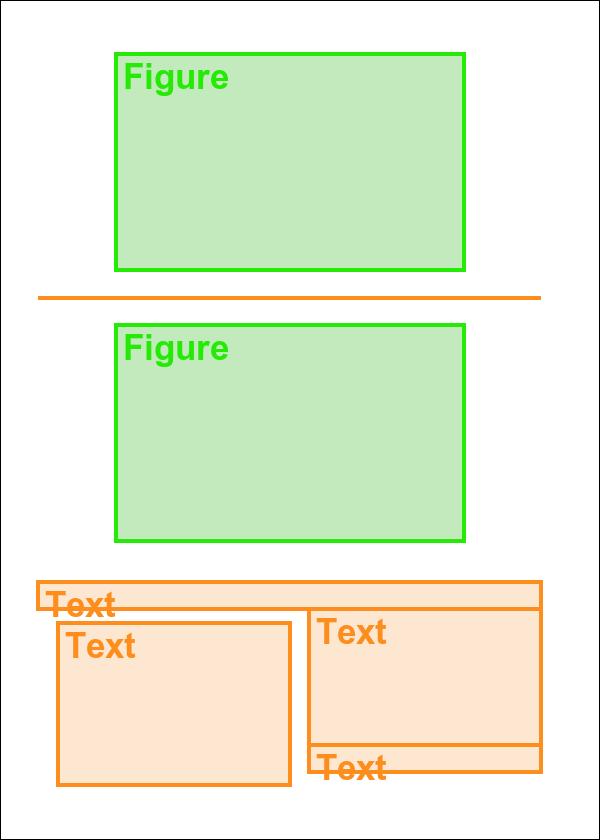} \\\hline
    \end{tabular}
    \caption{Investigation of stability of the latent space. From left to right, we add elements to the latent vector. We qualitatively observe that this results in reasonable extensions of the layout.}
    \label{fig:stability_samples}
\end{figure}

When training, the non-autoregressive decoder layouts $x\in \mathbb{R}^{l\times d_1}$ are encoded as latent representations $z\in \mathbb{R}^{l\times d_2}$\footnote{Technically the encoder parameterizes a distribution over $z$. However, for the sake of simplicity we shall consider only the mean in this experiment}. This experiment aims to investigate the consistency of these latent representations by removing elements from the latent code element by element - \ie $z\in \mathbb{R}^{l\times d_2} \rightarrow z'\in \mathbb{R}^{l\times d_2-1}$. By consistency we mean that removing one latent vector does not drastically change the decoded layout. Qualitative results can be found in fig. \ref{fig:stability_samples}. We observe that the latent code in fact appears largely consistent. Each new latent code appears to only introduce one new element in the layout while minimally changing the relative arrangement. This also implies that our model could be applied to the task of layout completion.

\clearpage
\onecolumn

\section{Qualitative results}
The small amount of samples presented in the main paper is not capable of conveying the diversity and quality of the synthesized data. In order to provide a clearer understanding, in the following we show a larger amount of samples for each dataset. These results are rendered using the network output coordinates without any postprocessing or cherry-picking applied to them to hide imperfections or failure cases.
The first rows show samples from other methods for comparison.

\subsection{PubLayNet}

\begin{figure}[h]
    \newlength{\publaynetBulkWidth}
    \setlength{\publaynetBulkWidth}{0.1125\linewidth}
    \setlength{\tabcolsep}{1pt}
    \centering
    \begin{tabular}{cccccccc}
\rotatebox{90}{\hspace{2mm}LayoutVAE [16]}&
\includegraphics[width=\publaynetBulkWidth]{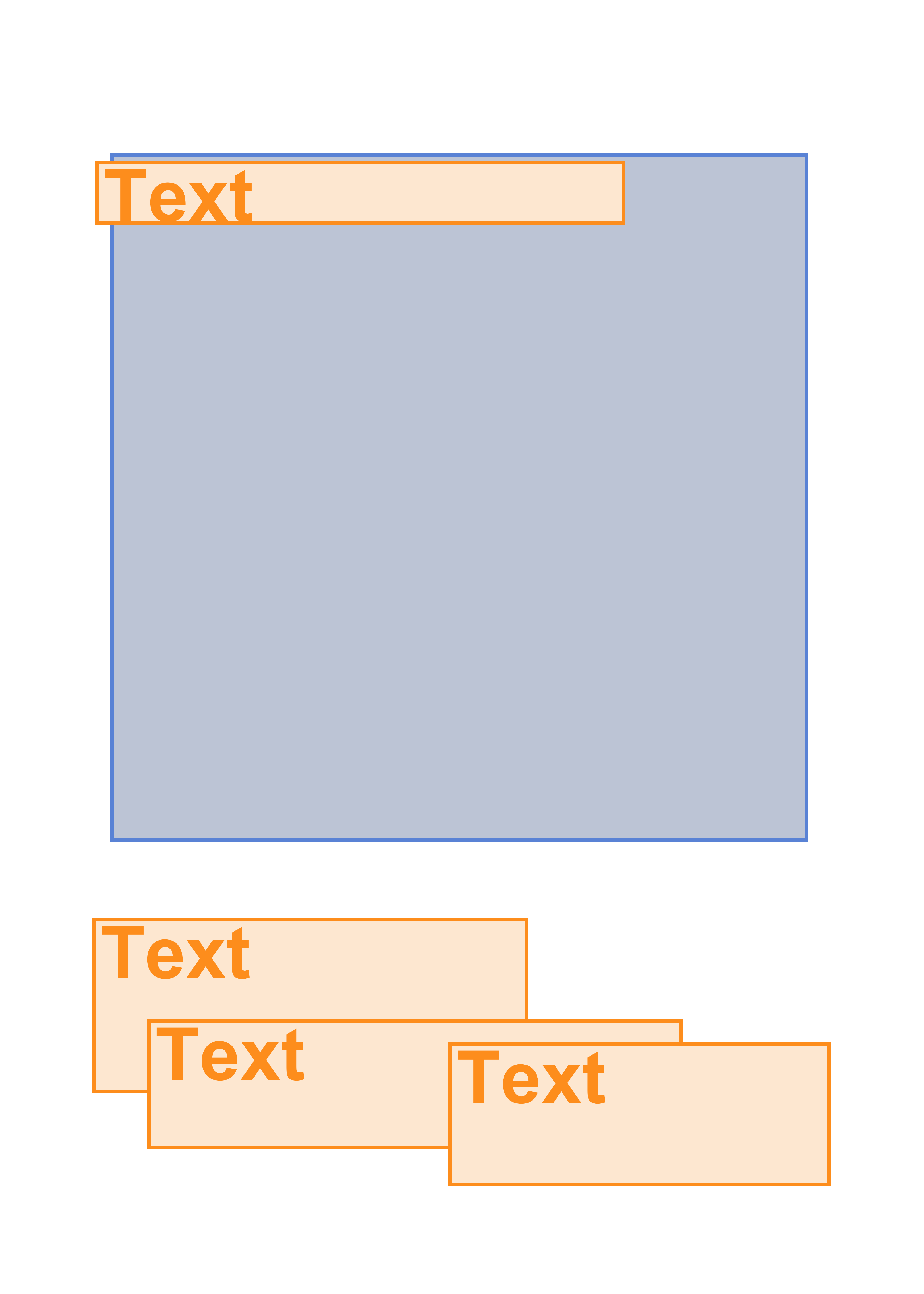} &
\includegraphics[width=\publaynetBulkWidth]{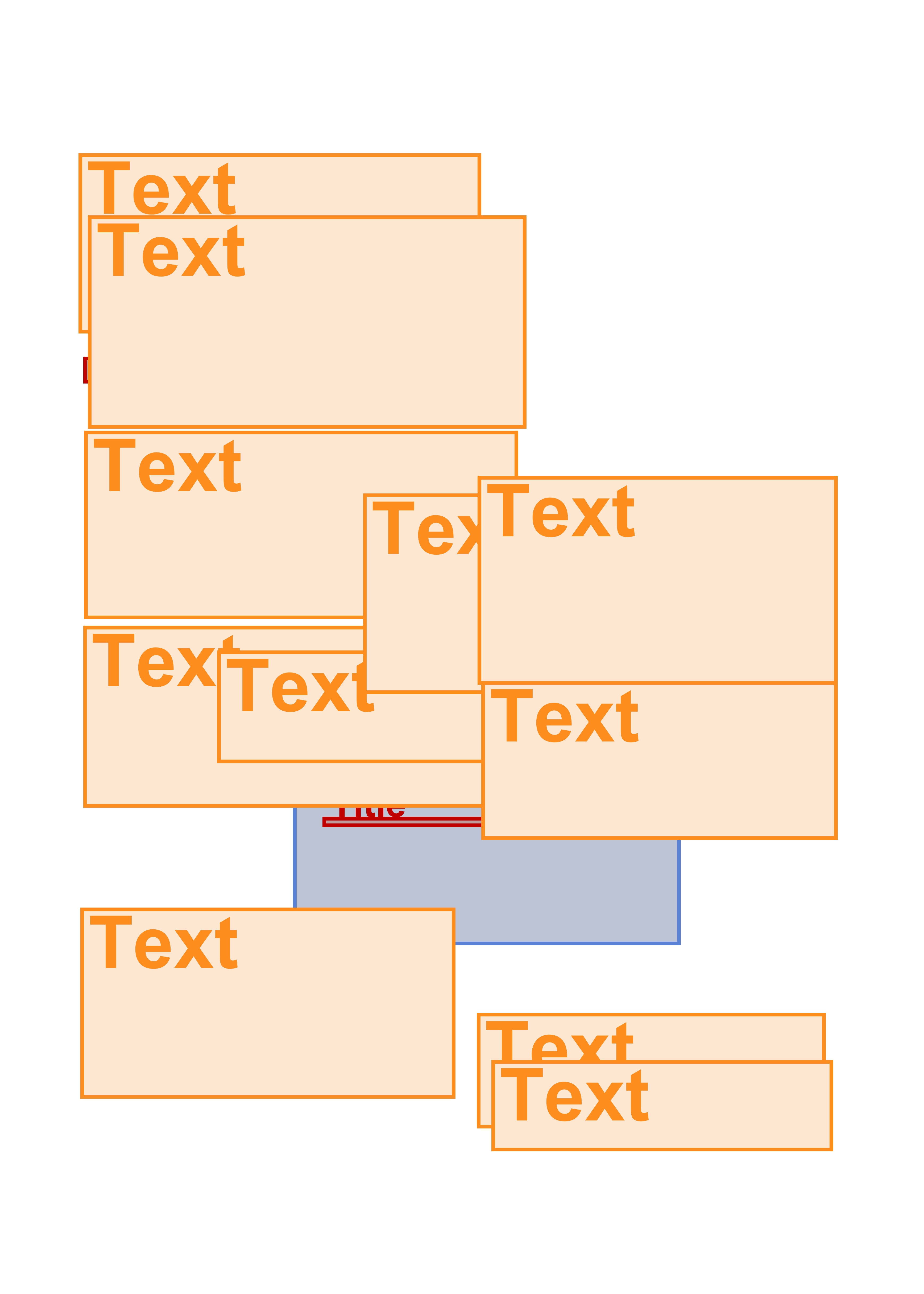} &
\includegraphics[width=\publaynetBulkWidth]{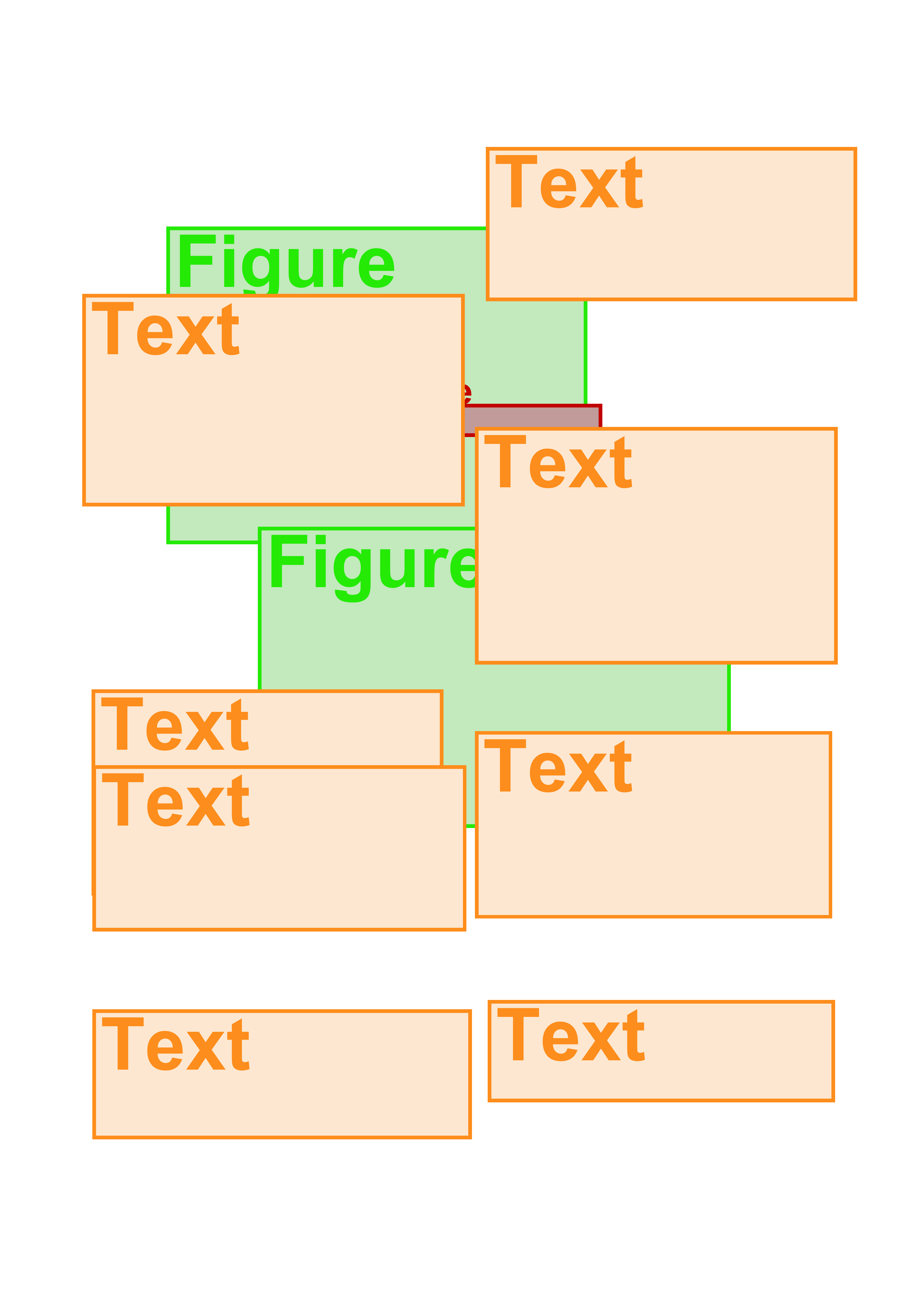} &
\includegraphics[width=\publaynetBulkWidth]{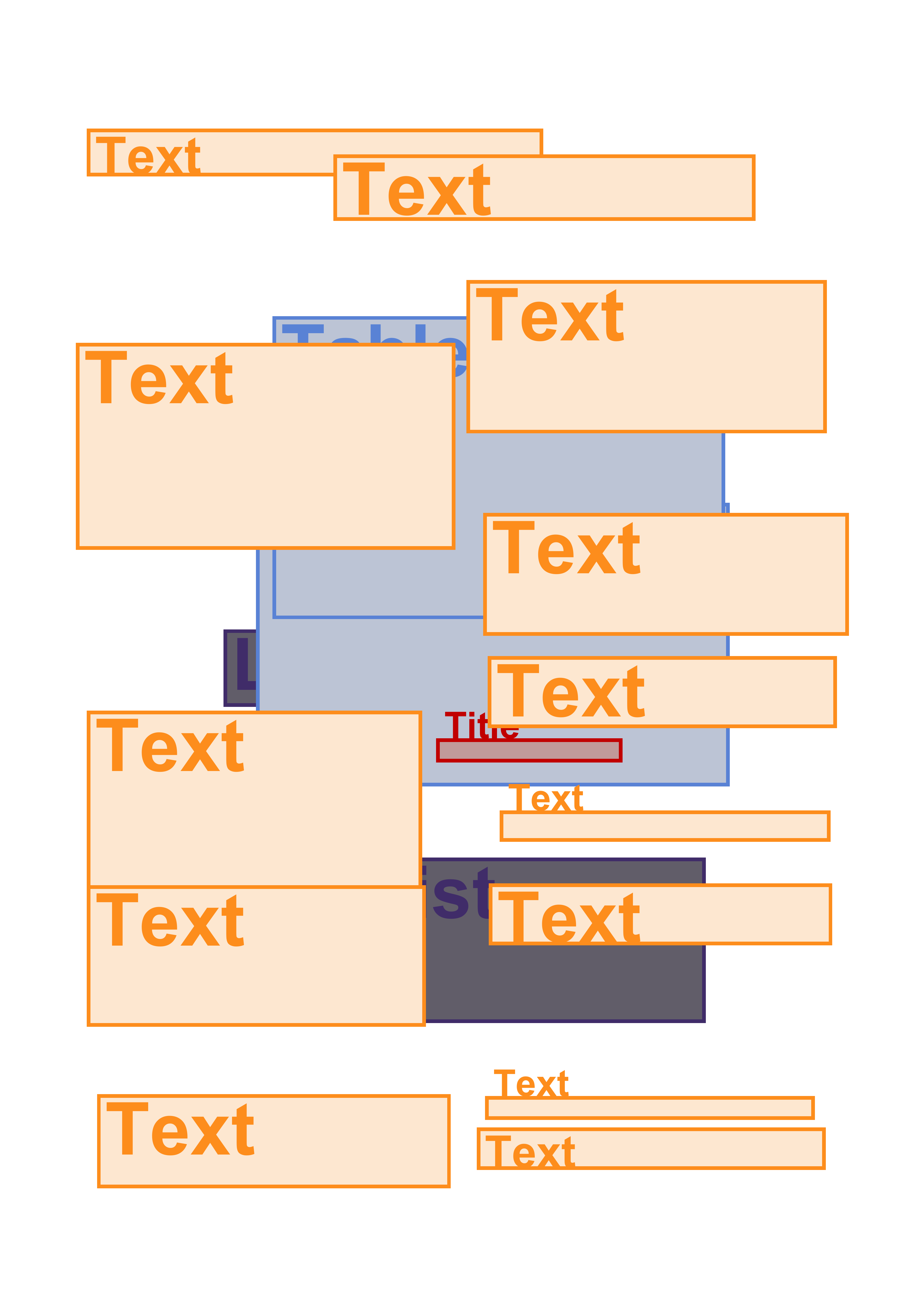} &
\includegraphics[width=\publaynetBulkWidth]{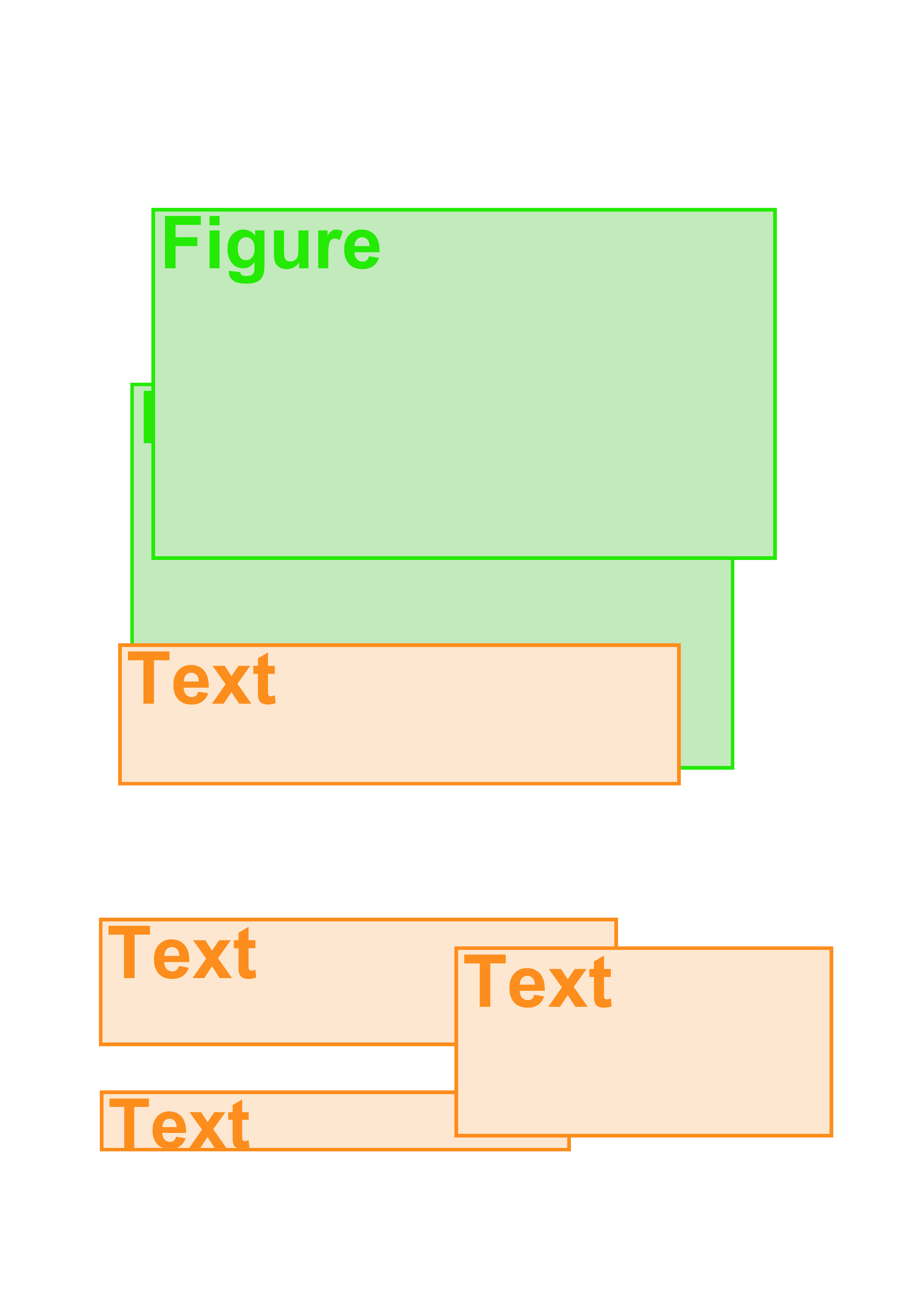} &
\includegraphics[width=\publaynetBulkWidth]{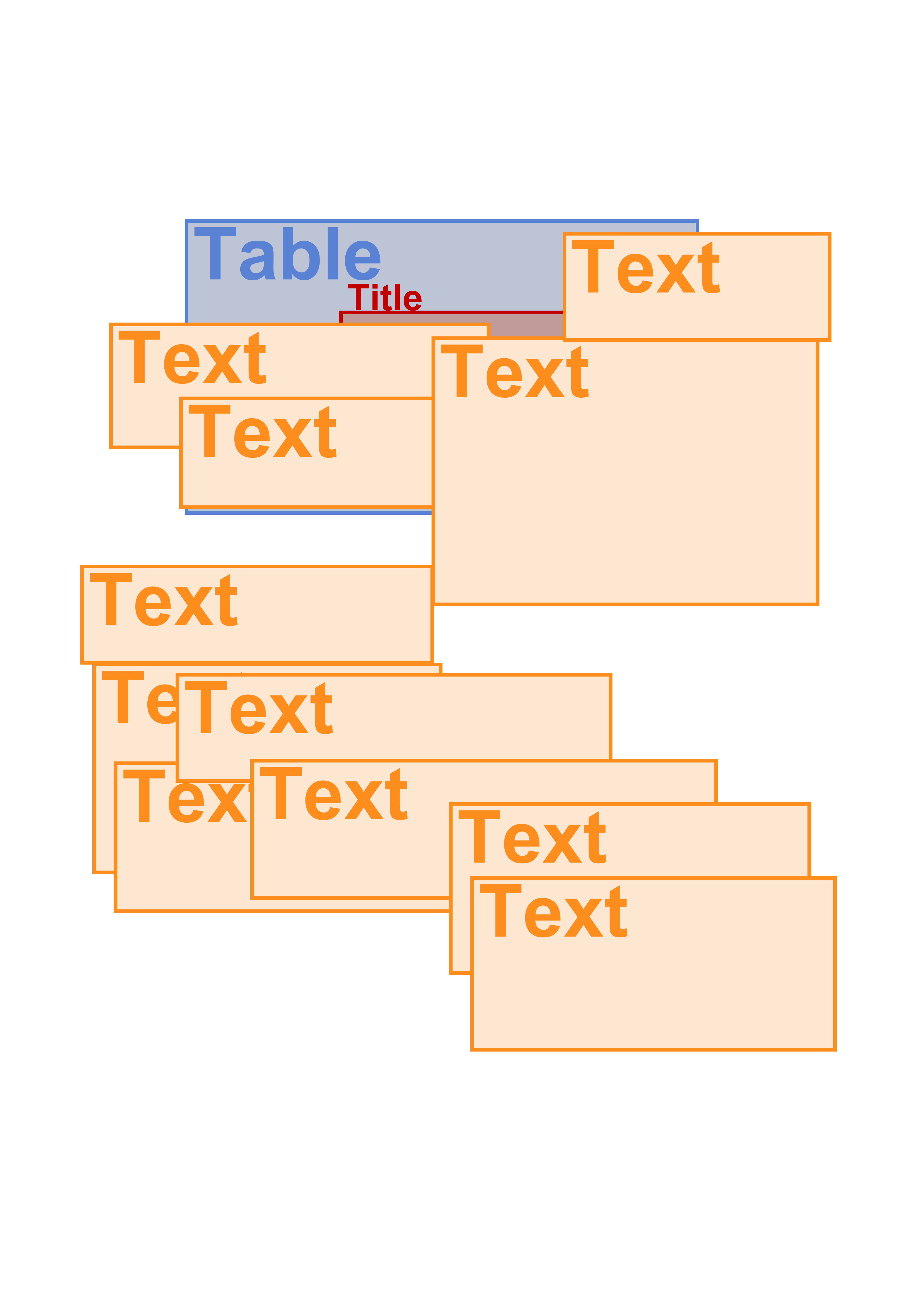} &
\includegraphics[width=\publaynetBulkWidth]{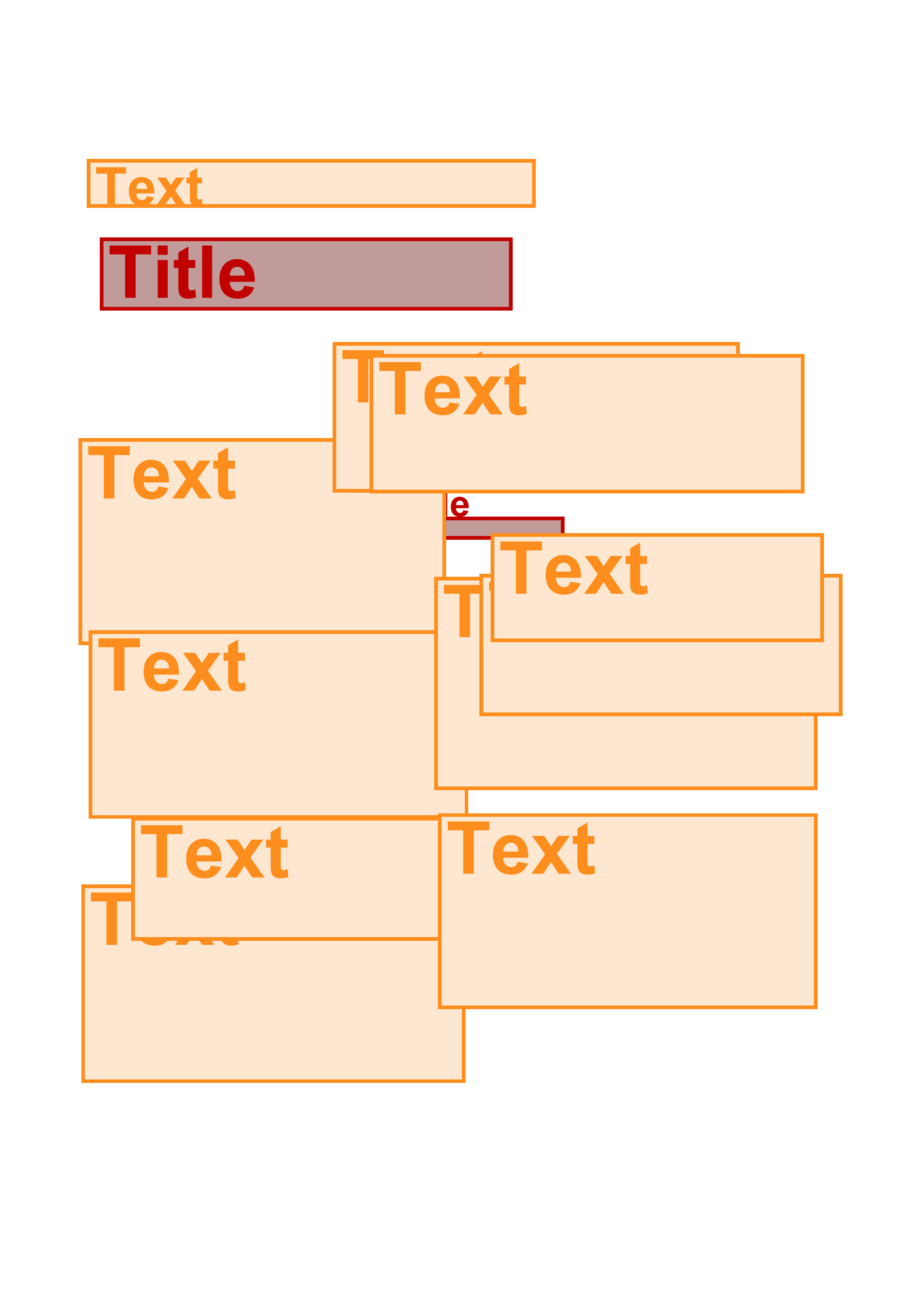} \\
\rotatebox{90}{\hspace{3mm}Gupta \etal [9]}&
\includegraphics[width=\publaynetBulkWidth]{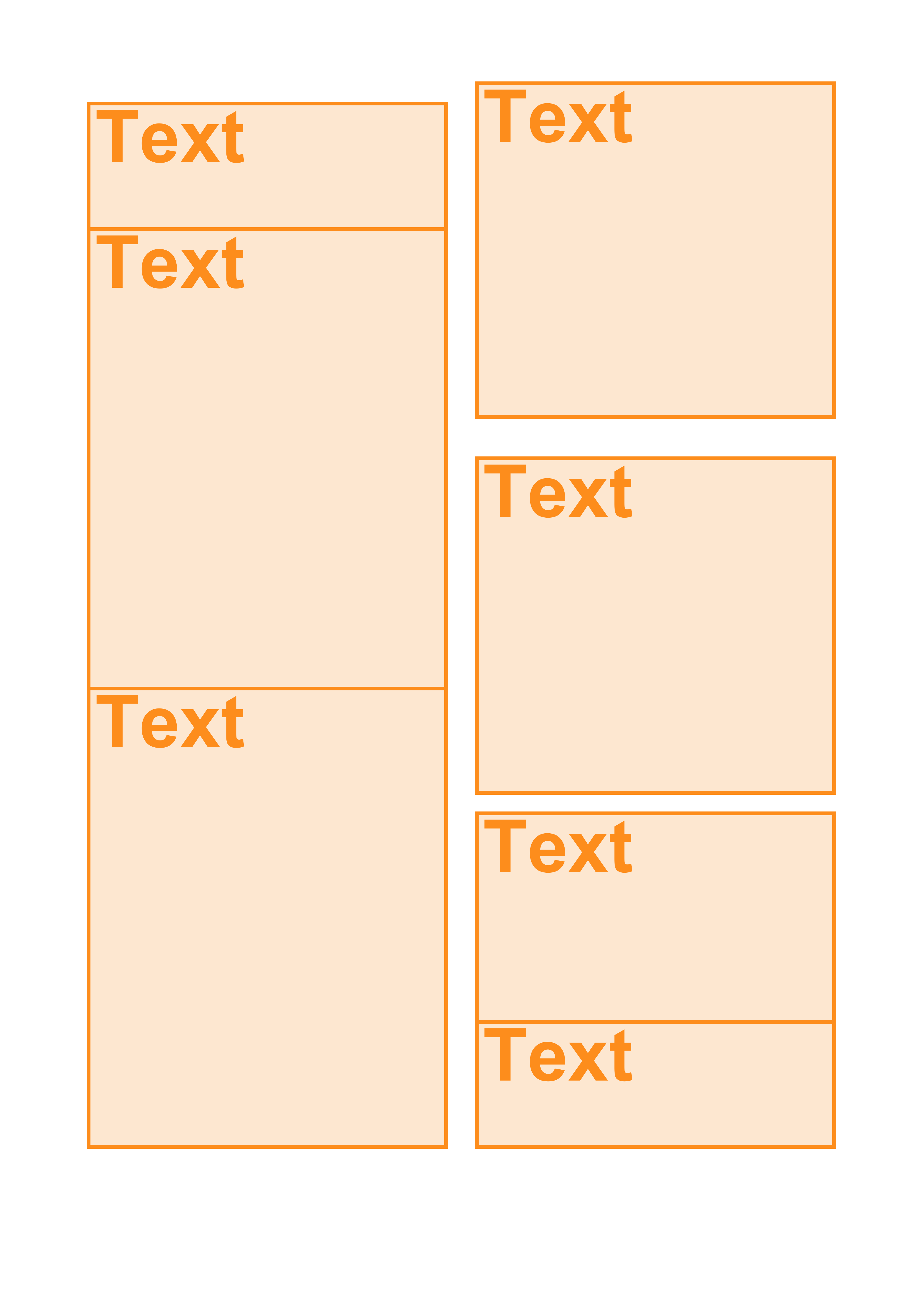} &
\includegraphics[width=\publaynetBulkWidth]{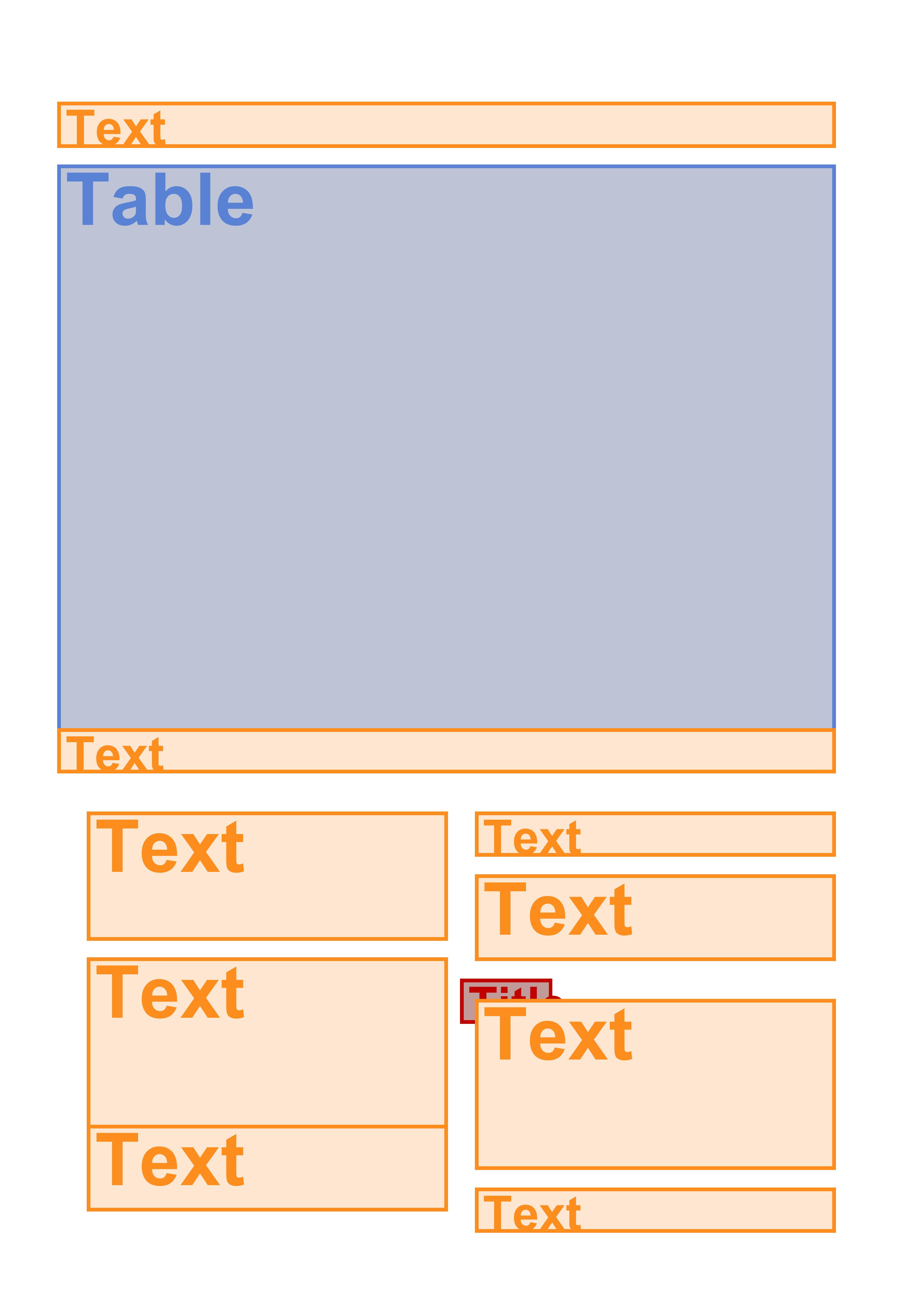} &\includegraphics[width=\publaynetBulkWidth]{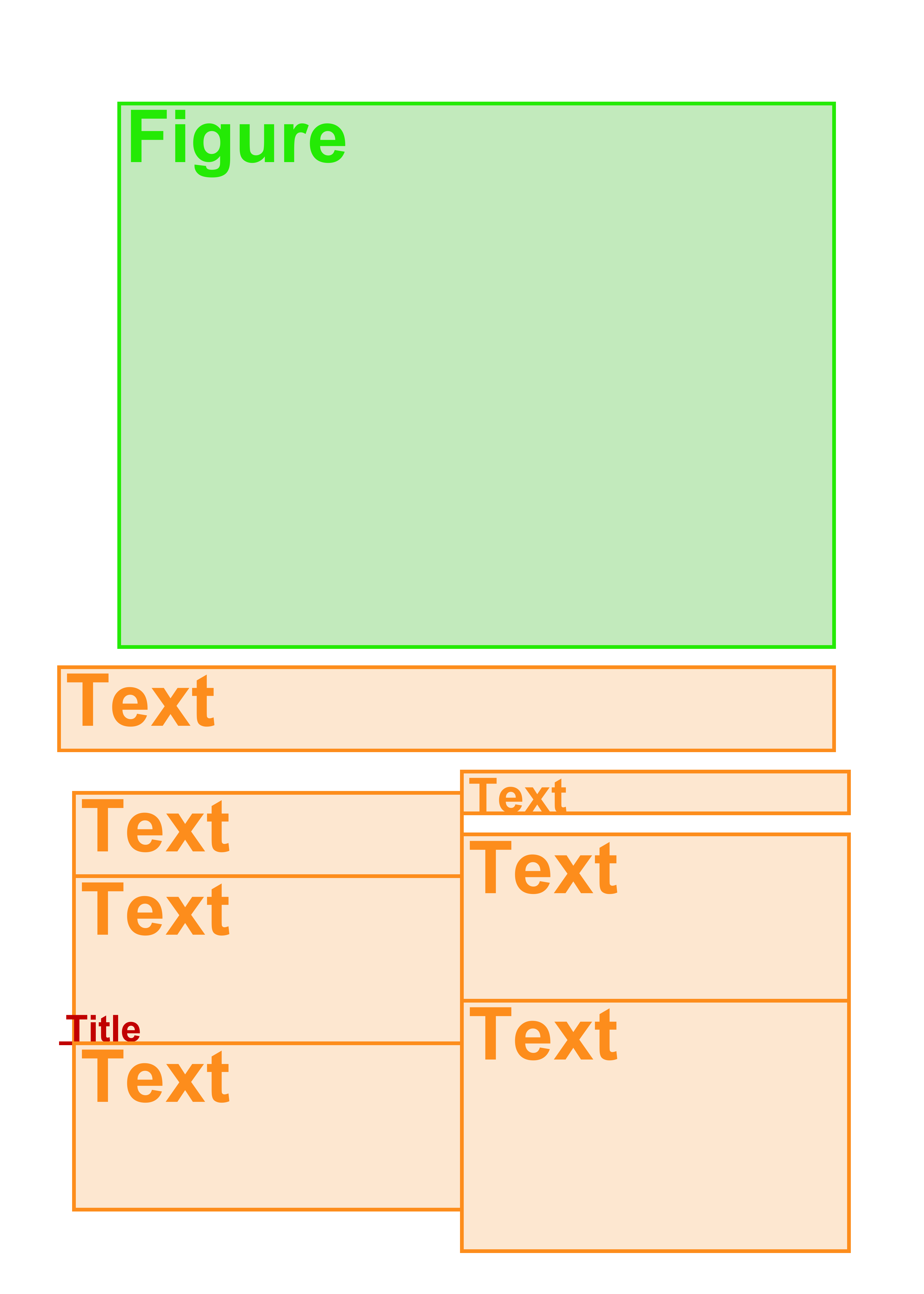} &\includegraphics[width=\publaynetBulkWidth]{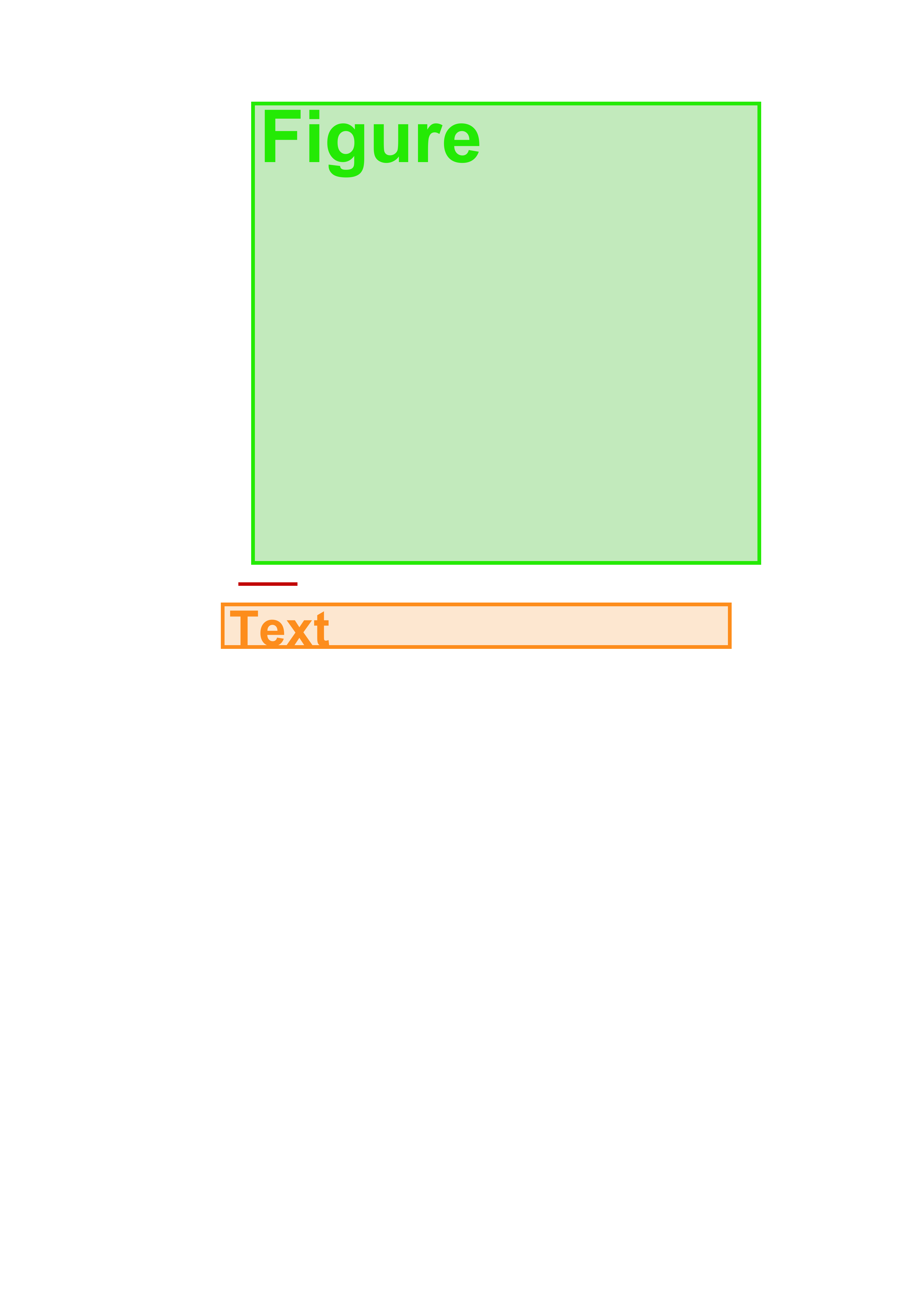} &\includegraphics[width=\publaynetBulkWidth]{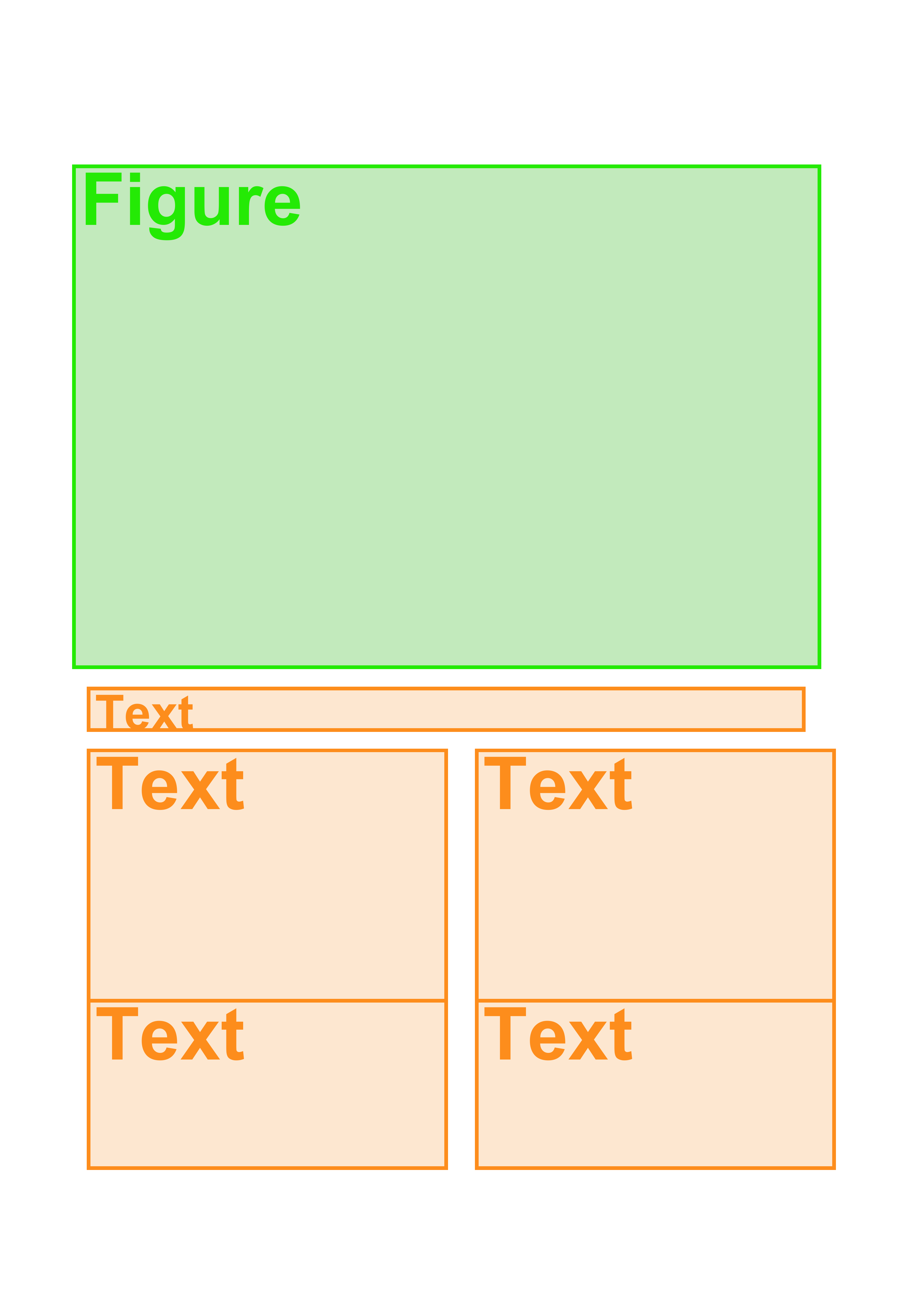} &\includegraphics[width=\publaynetBulkWidth]{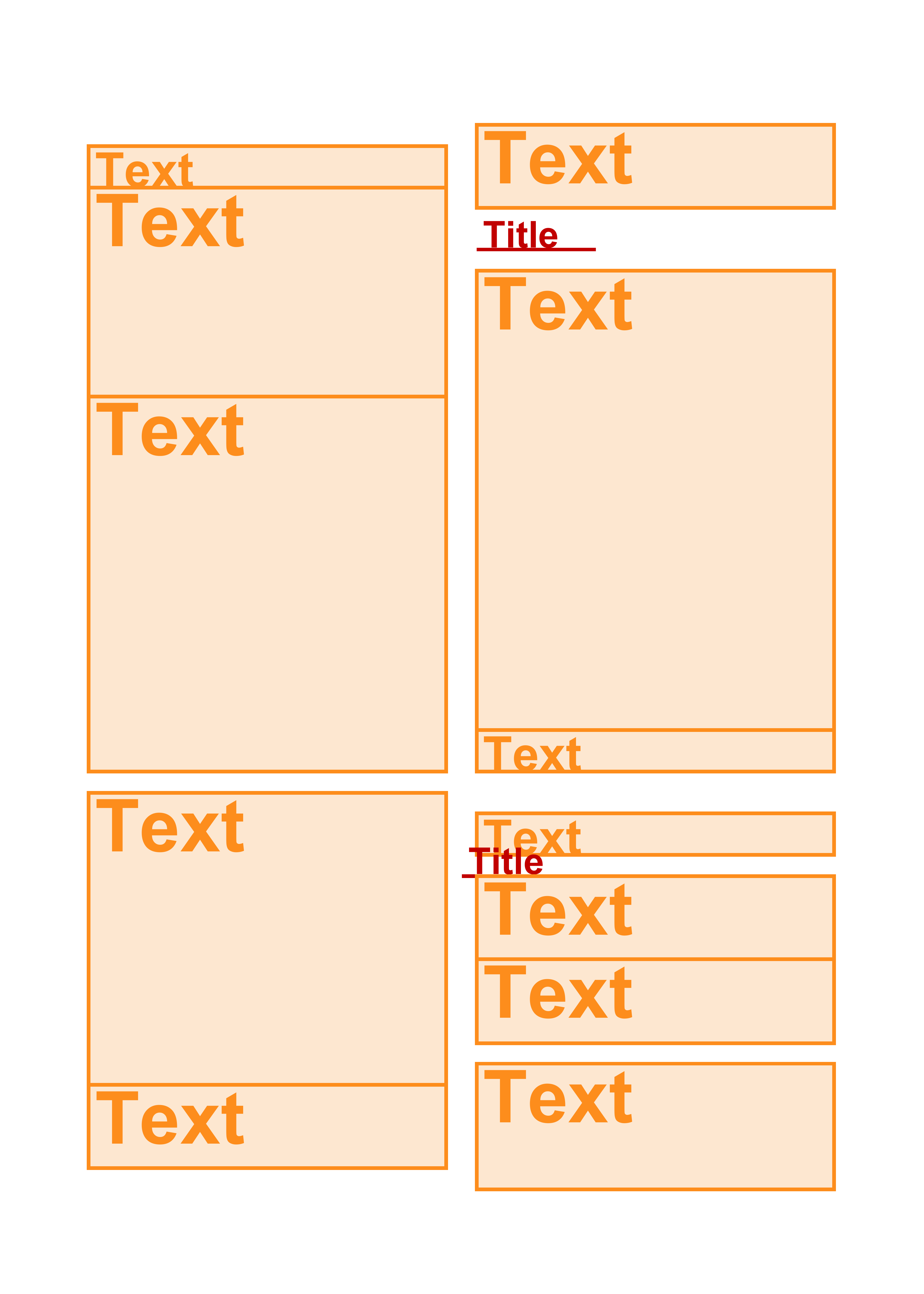} &\includegraphics[width=\publaynetBulkWidth]{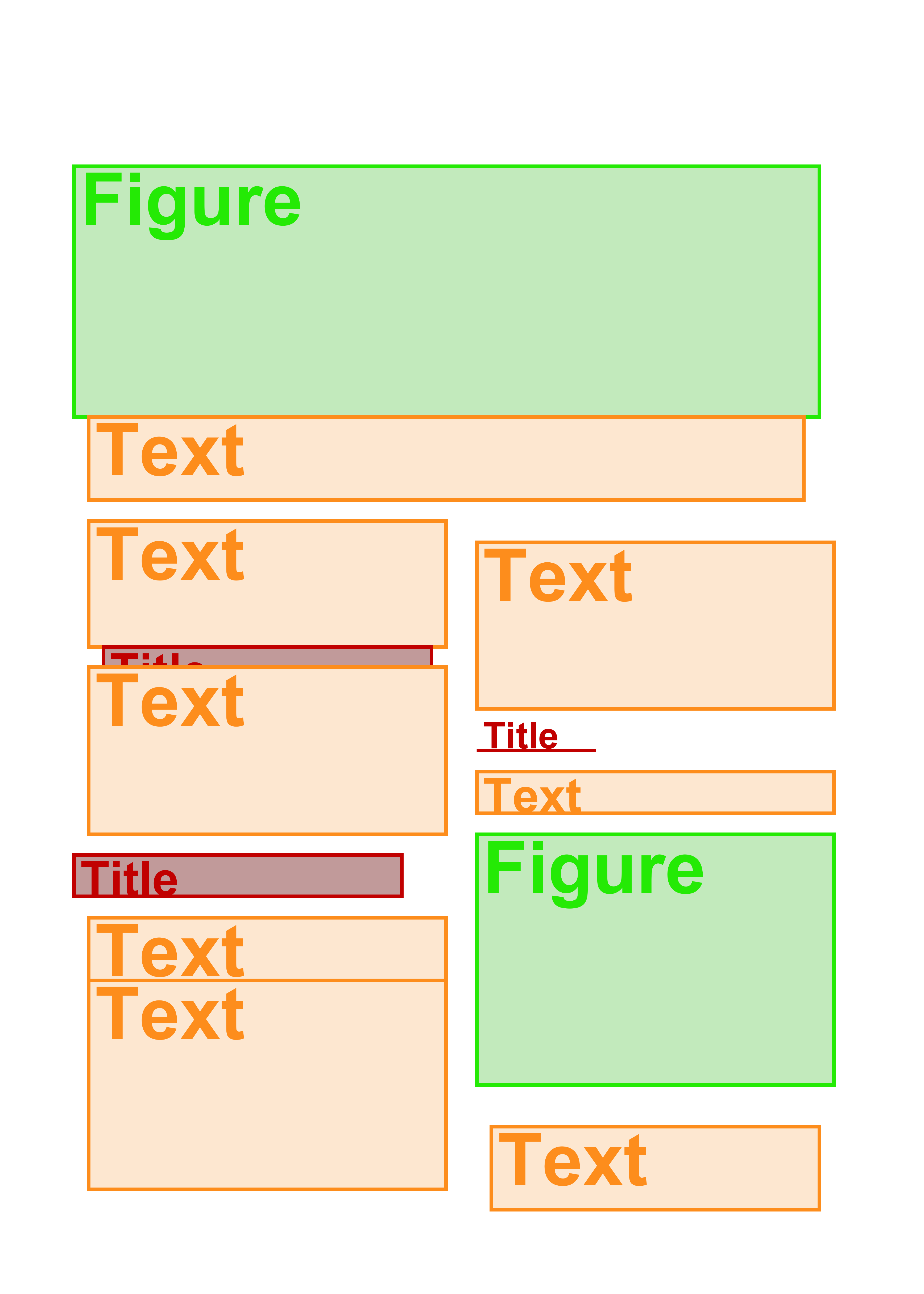} \\
\toprule
\rotatebox{90}{\hspace{10mm}Ours}&\includegraphics[width=\publaynetBulkWidth]{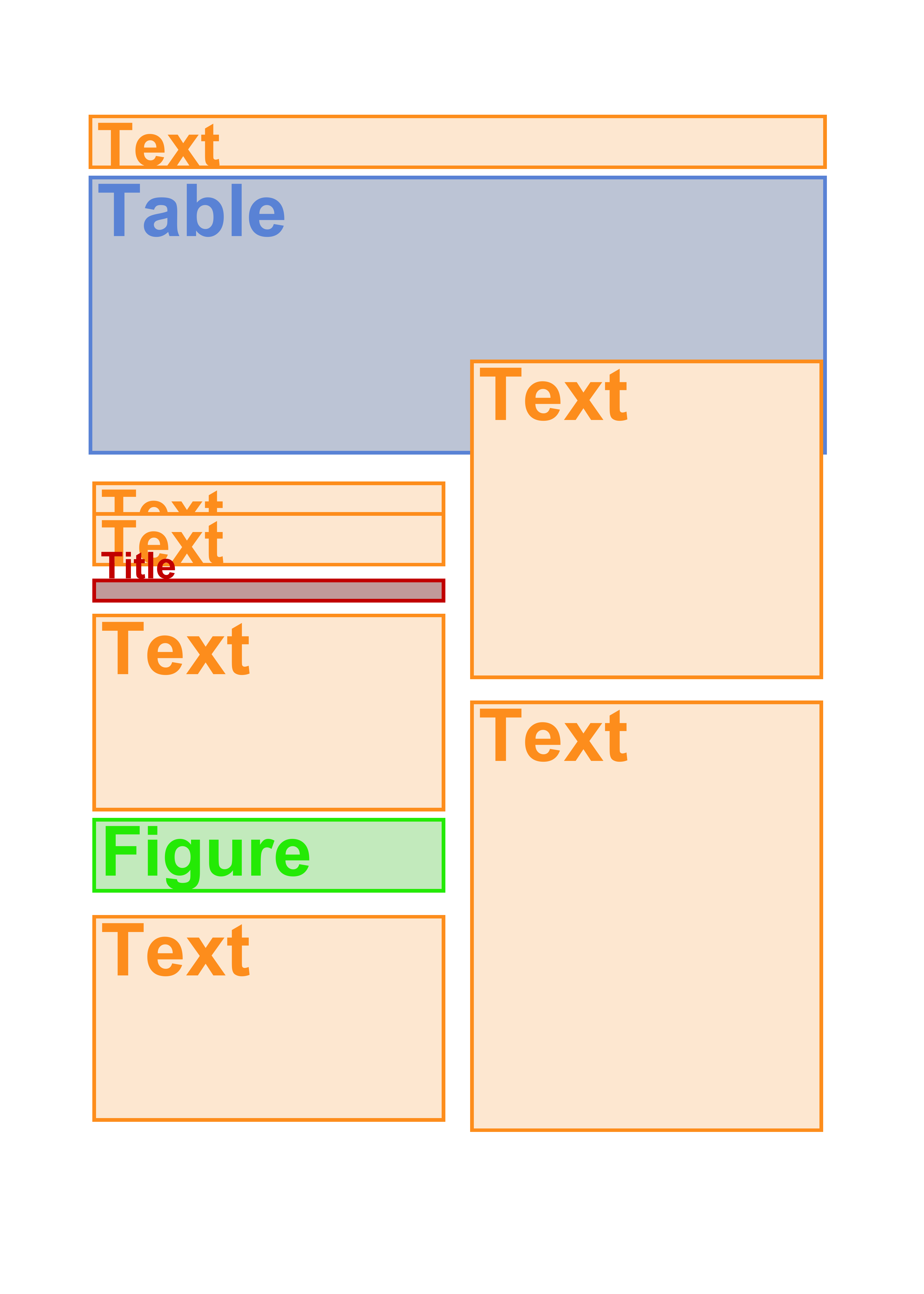} & 
\includegraphics[width=\publaynetBulkWidth]{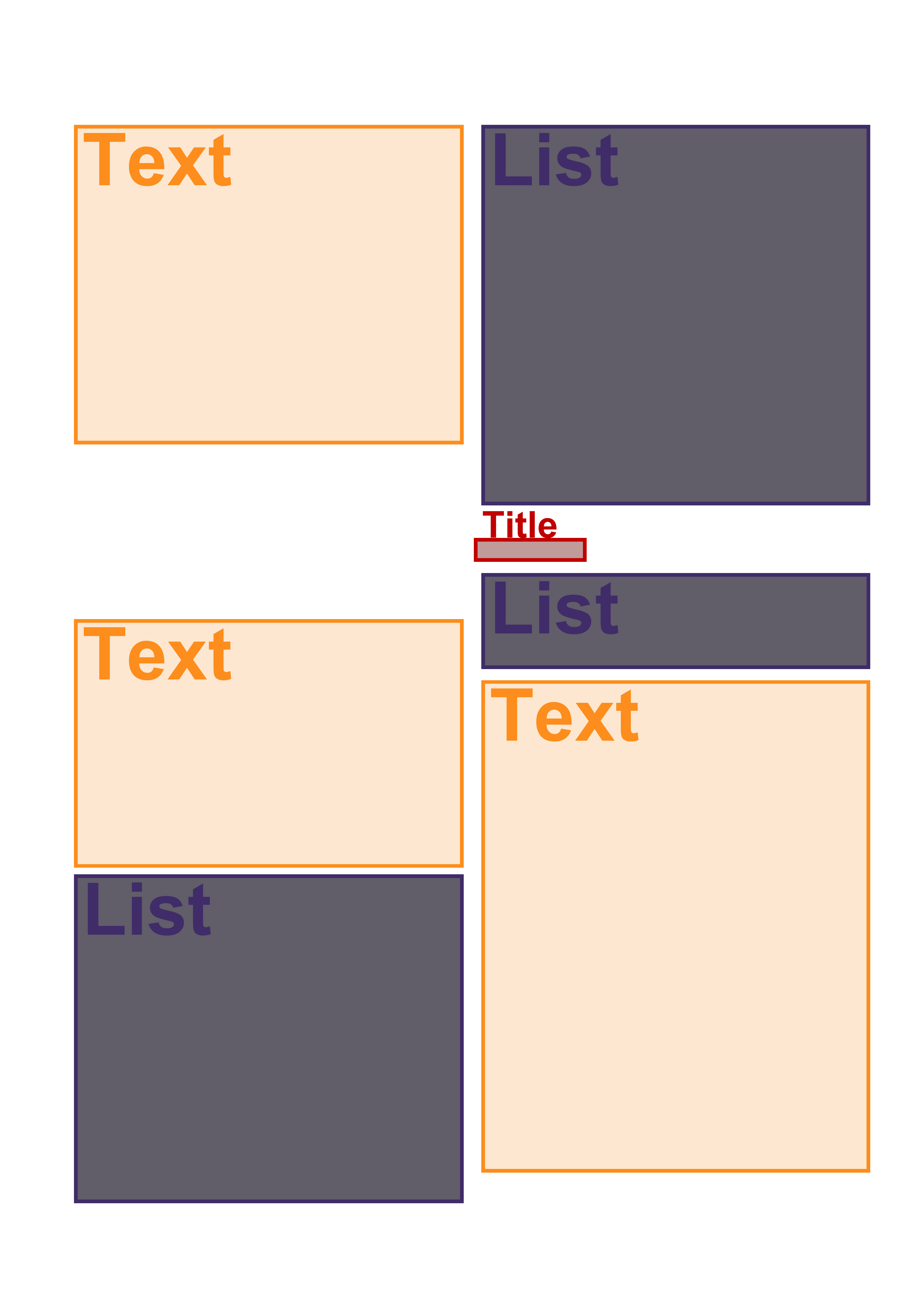} & 
\includegraphics[width=\publaynetBulkWidth]{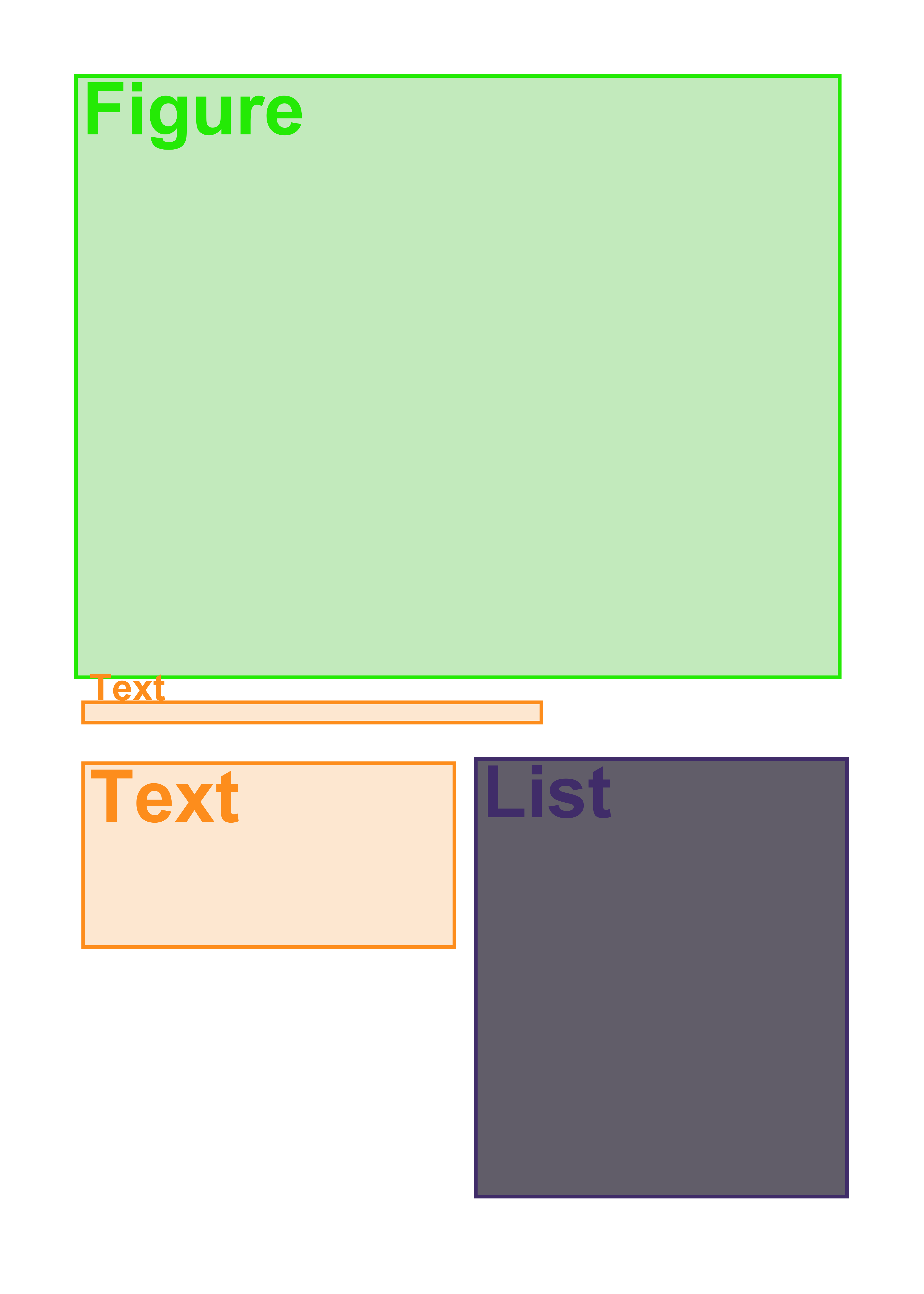} & 
\includegraphics[width=\publaynetBulkWidth]{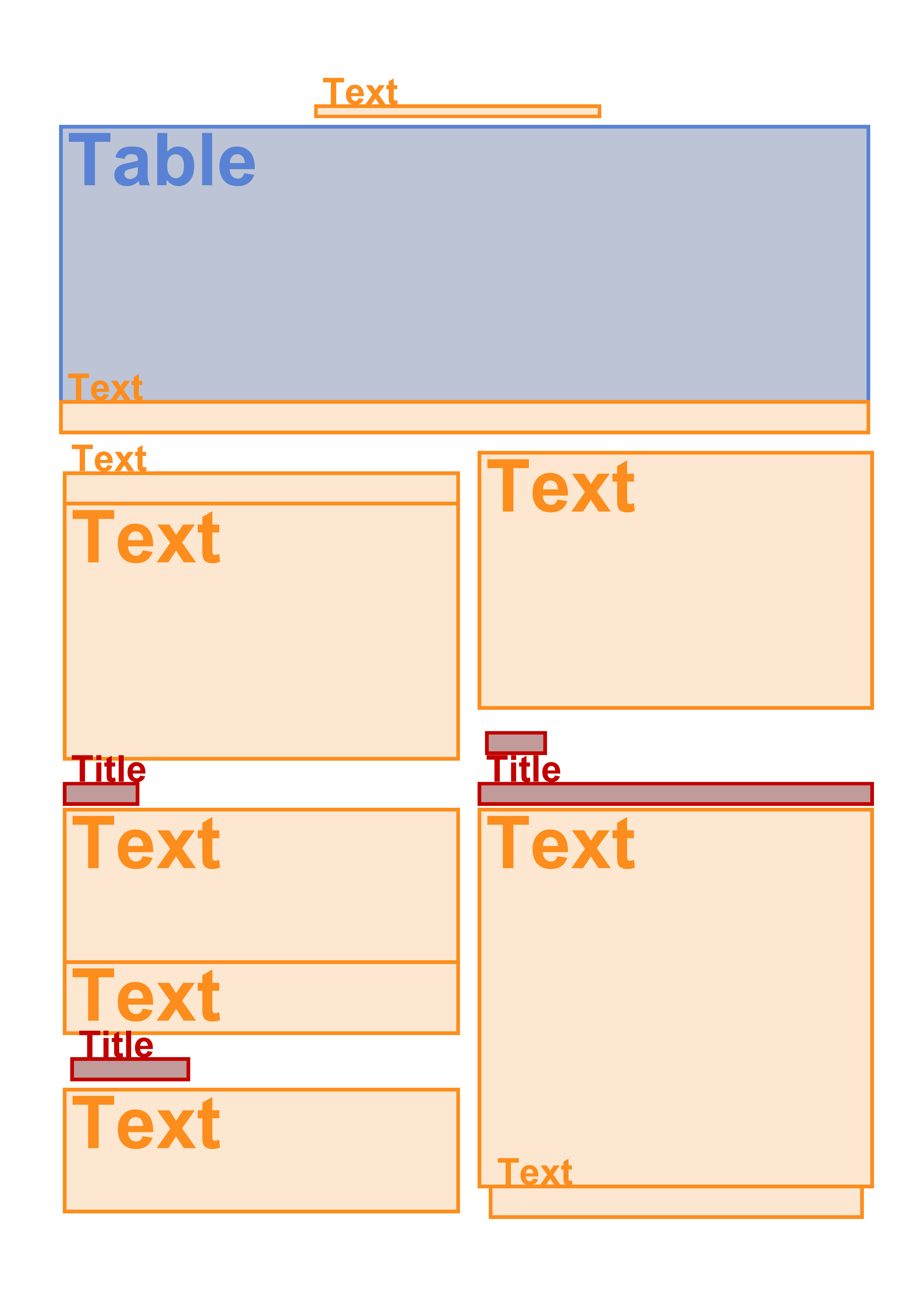} & 
\includegraphics[width=\publaynetBulkWidth]{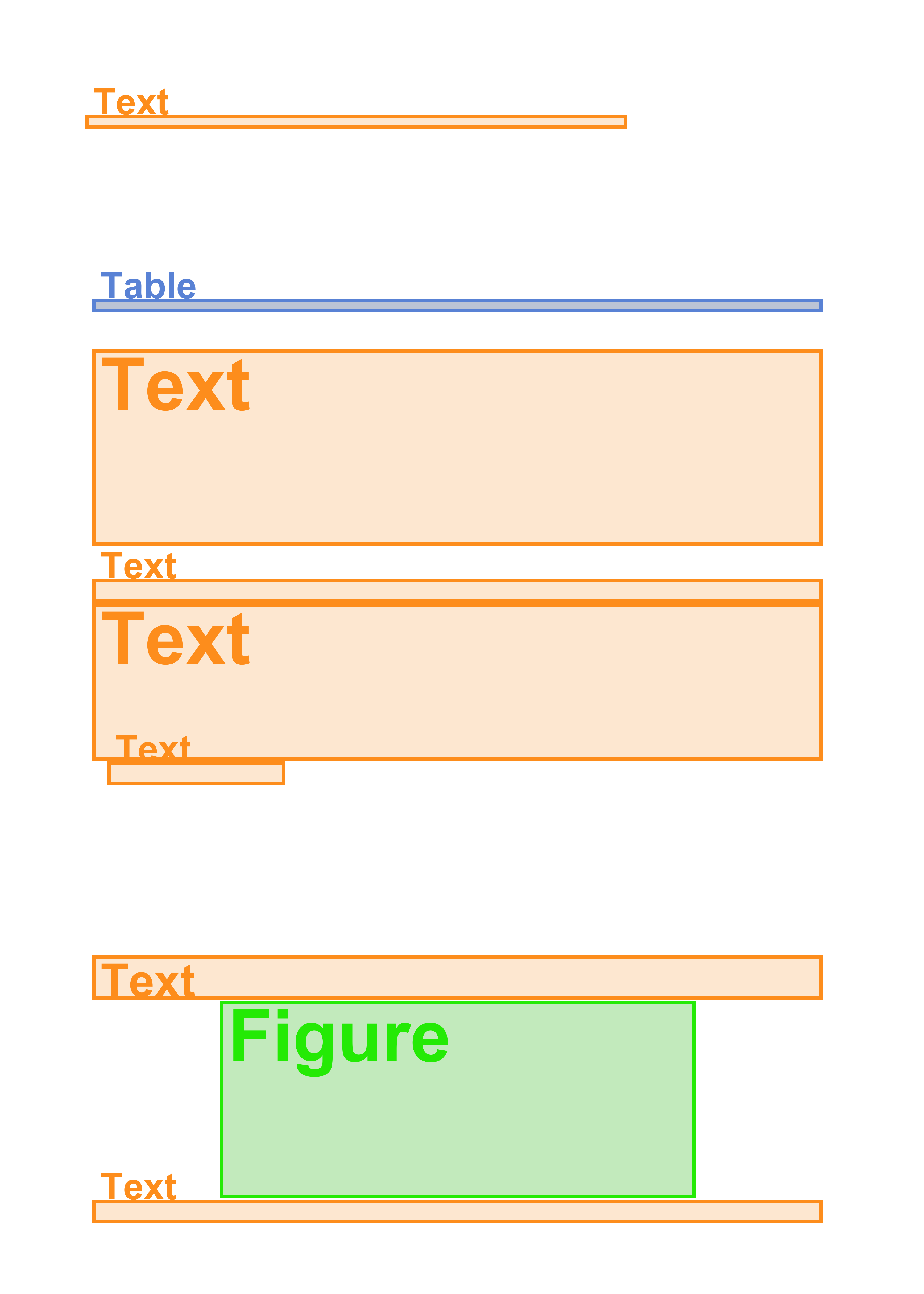} & 
\includegraphics[width=\publaynetBulkWidth]{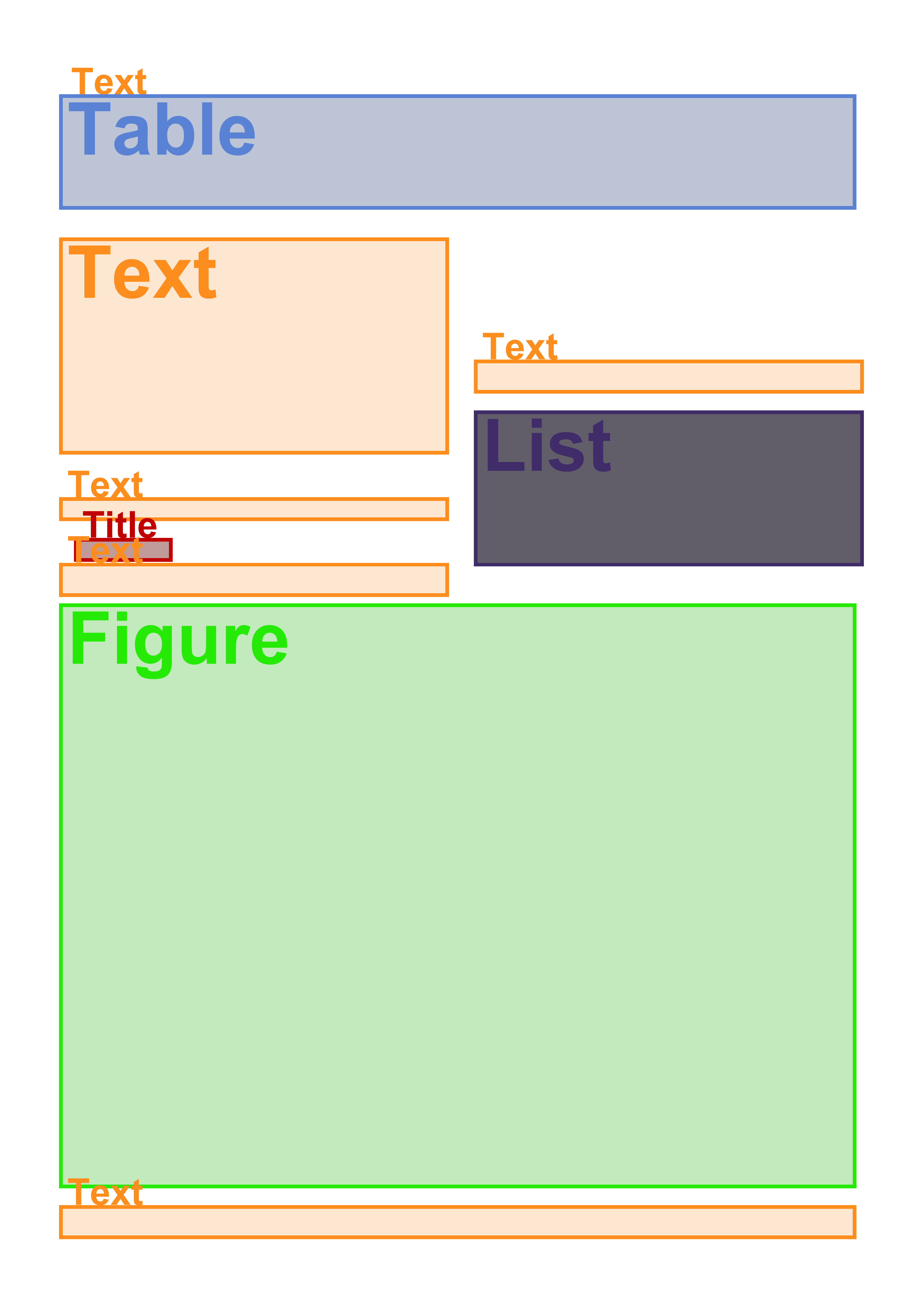} & 
\includegraphics[width=\publaynetBulkWidth]{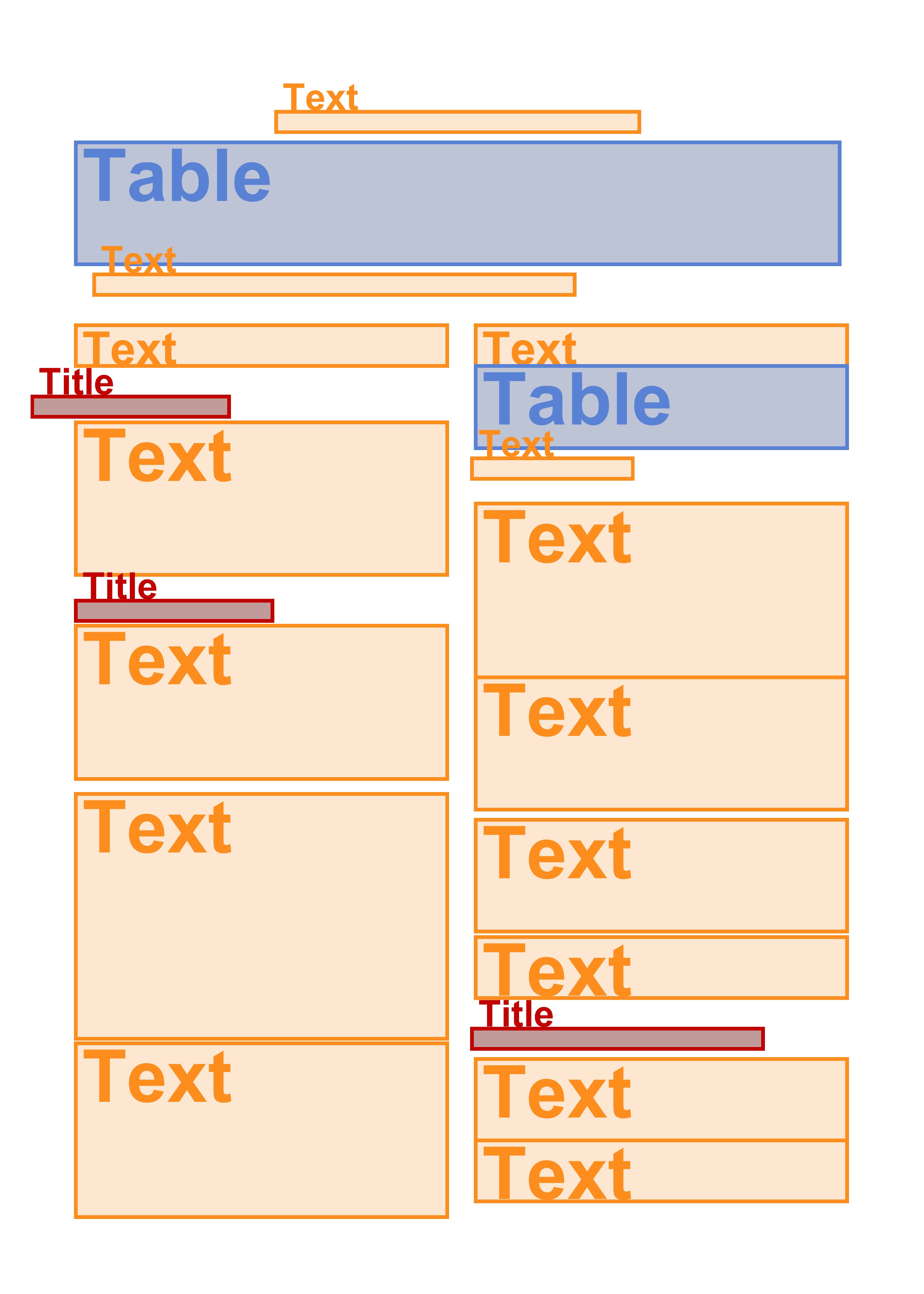} \\
&\includegraphics[width=\publaynetBulkWidth]{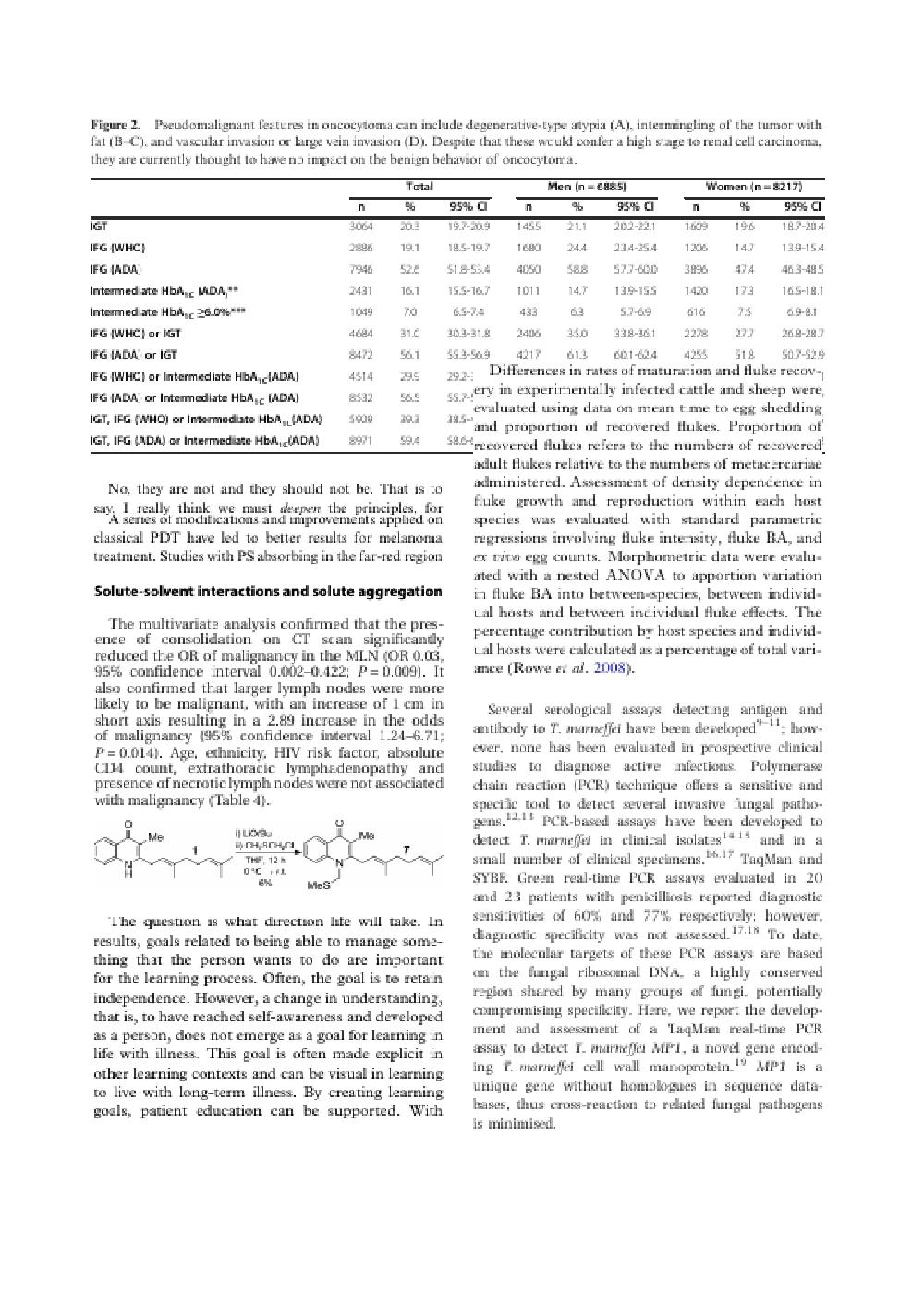} &
\includegraphics[width=\publaynetBulkWidth]{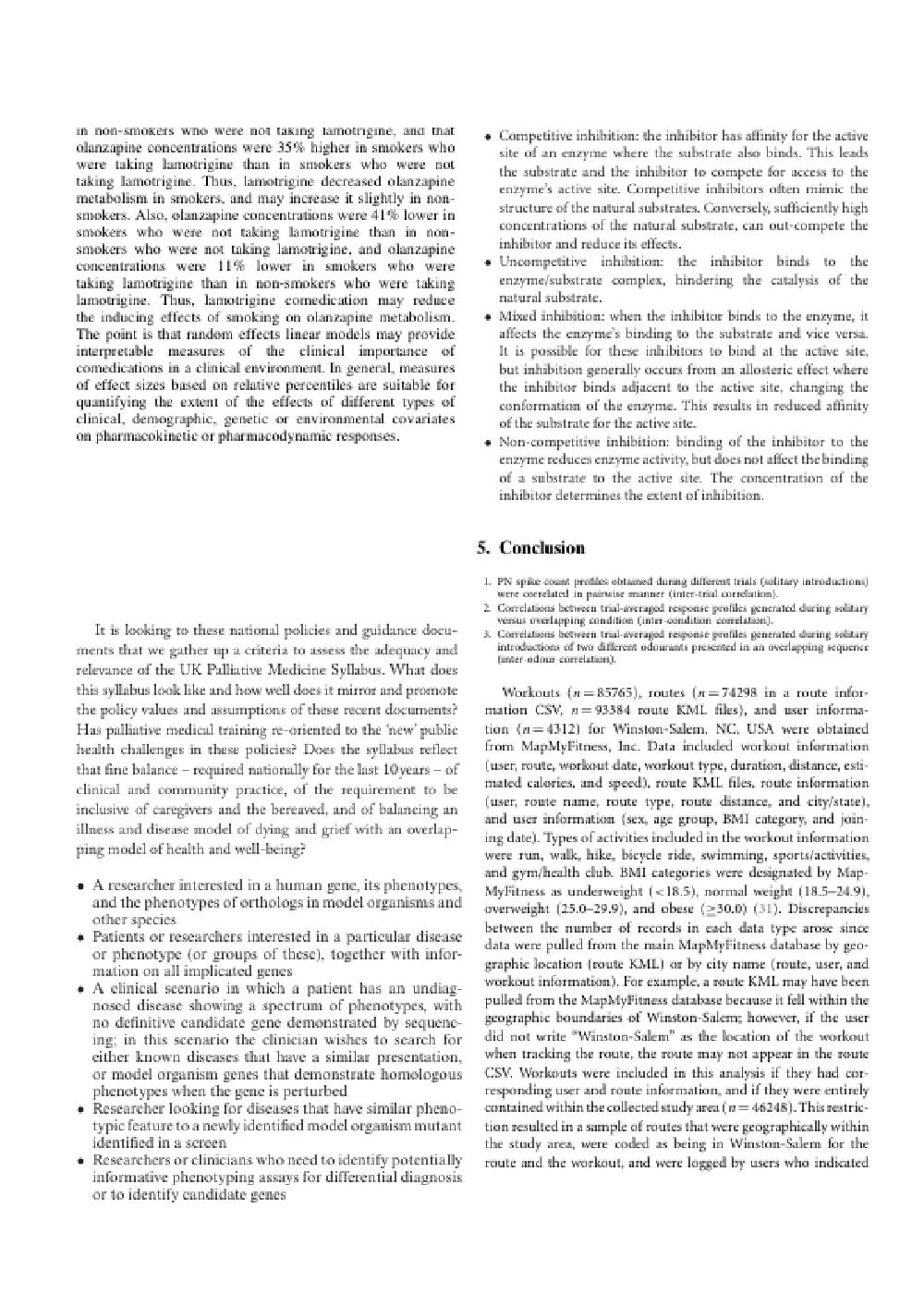} &
\includegraphics[width=\publaynetBulkWidth]{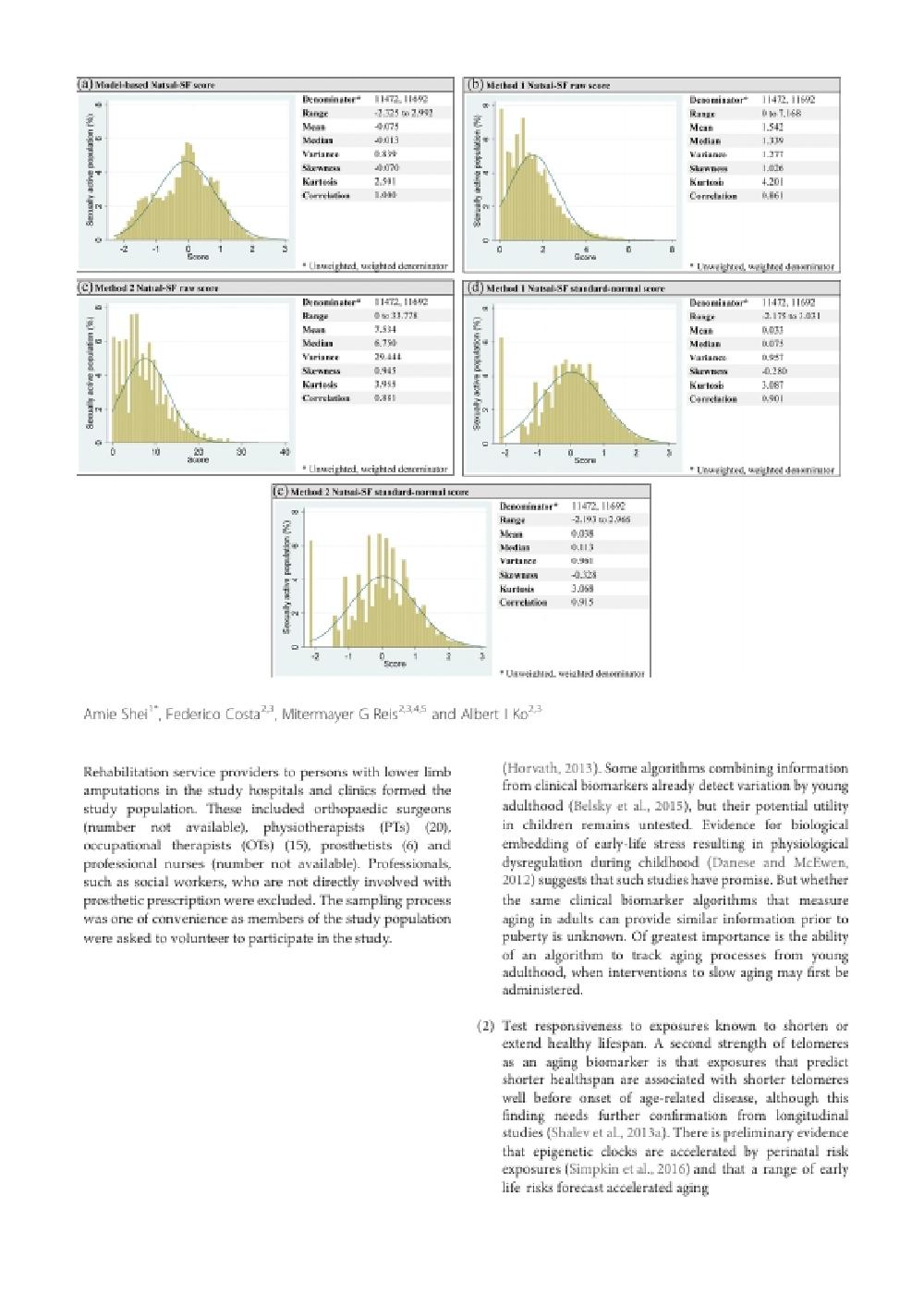} &
\includegraphics[width=\publaynetBulkWidth]{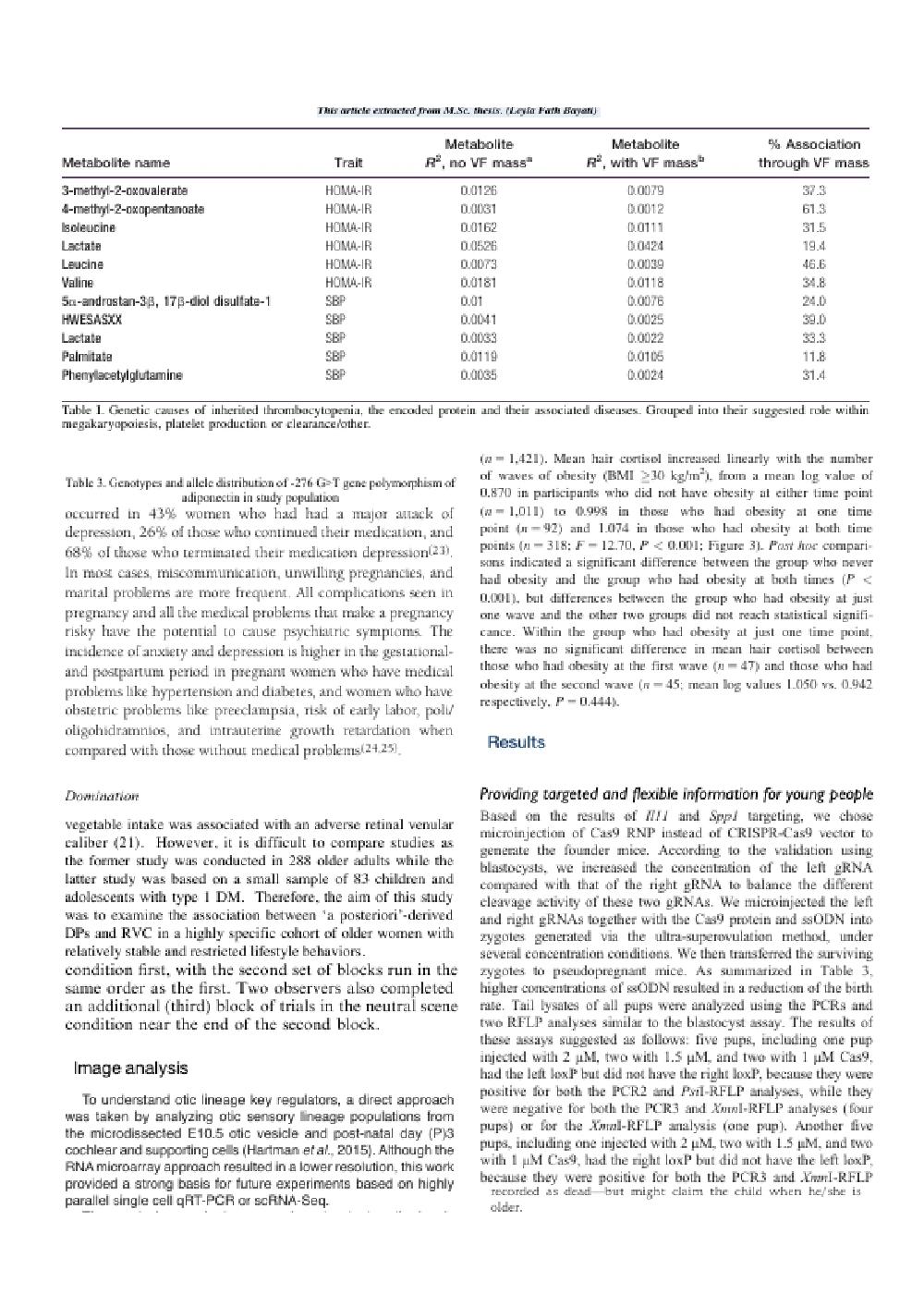} &
\includegraphics[width=\publaynetBulkWidth]{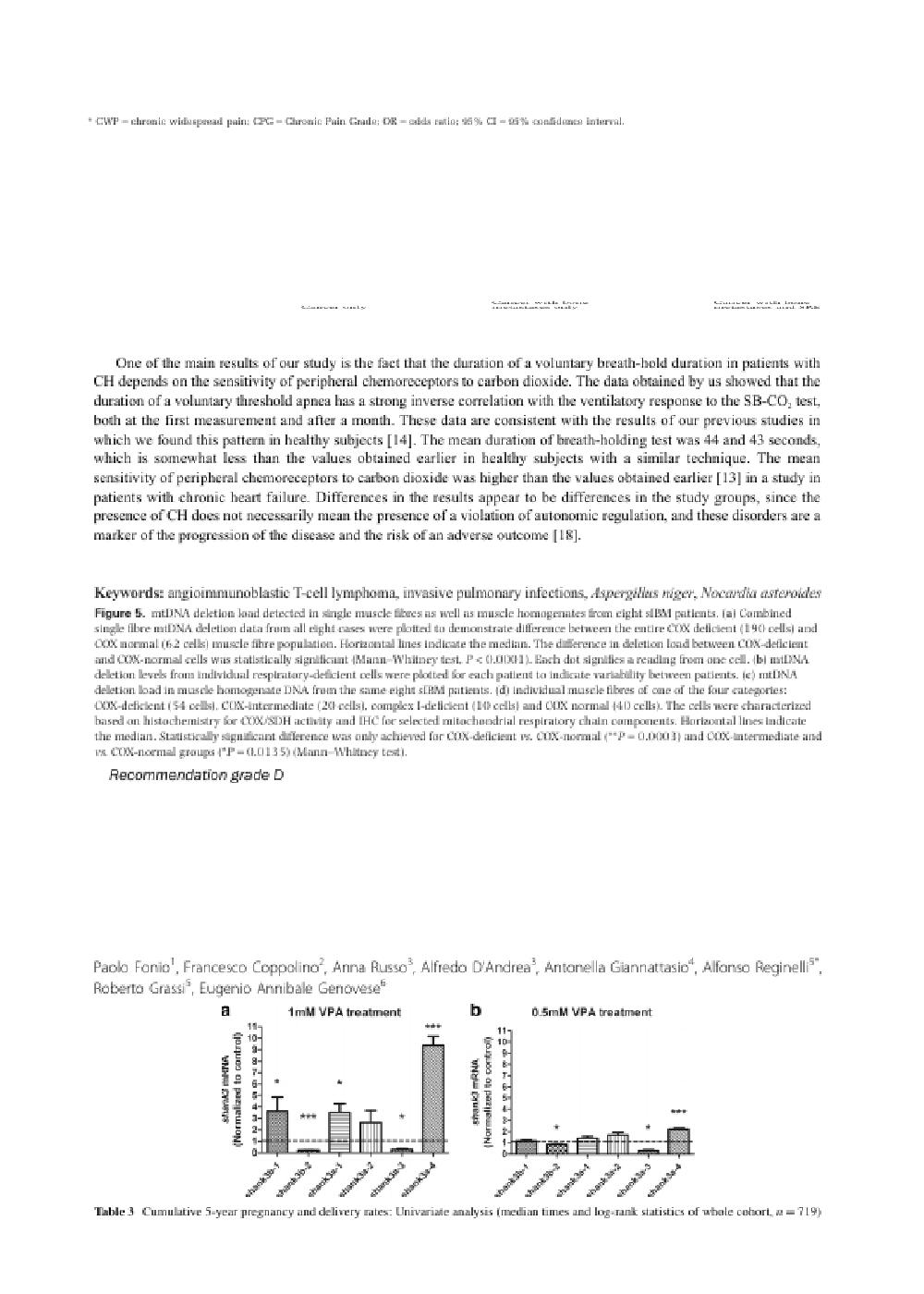} &
\includegraphics[width=\publaynetBulkWidth]{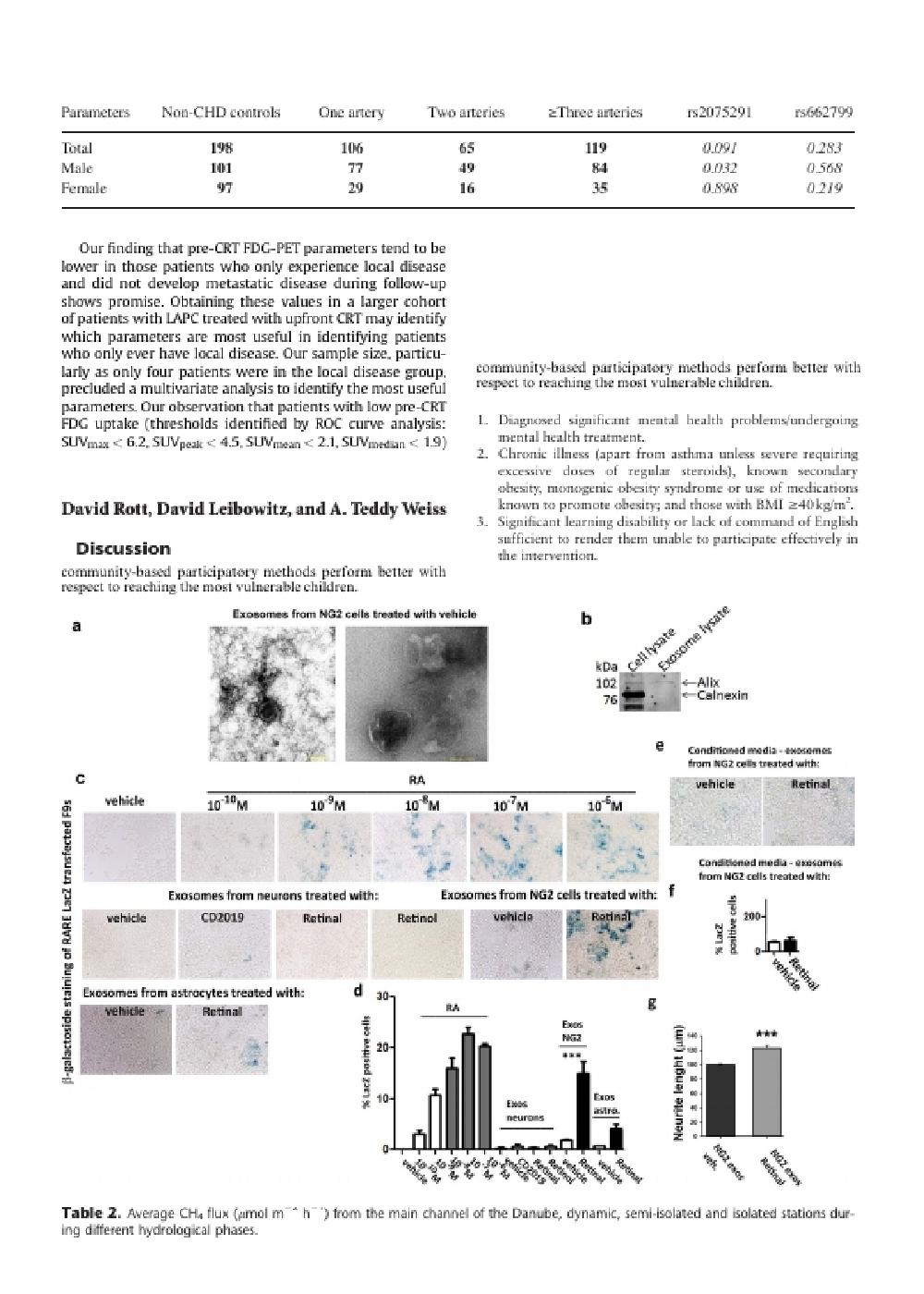} &
\includegraphics[width=\publaynetBulkWidth]{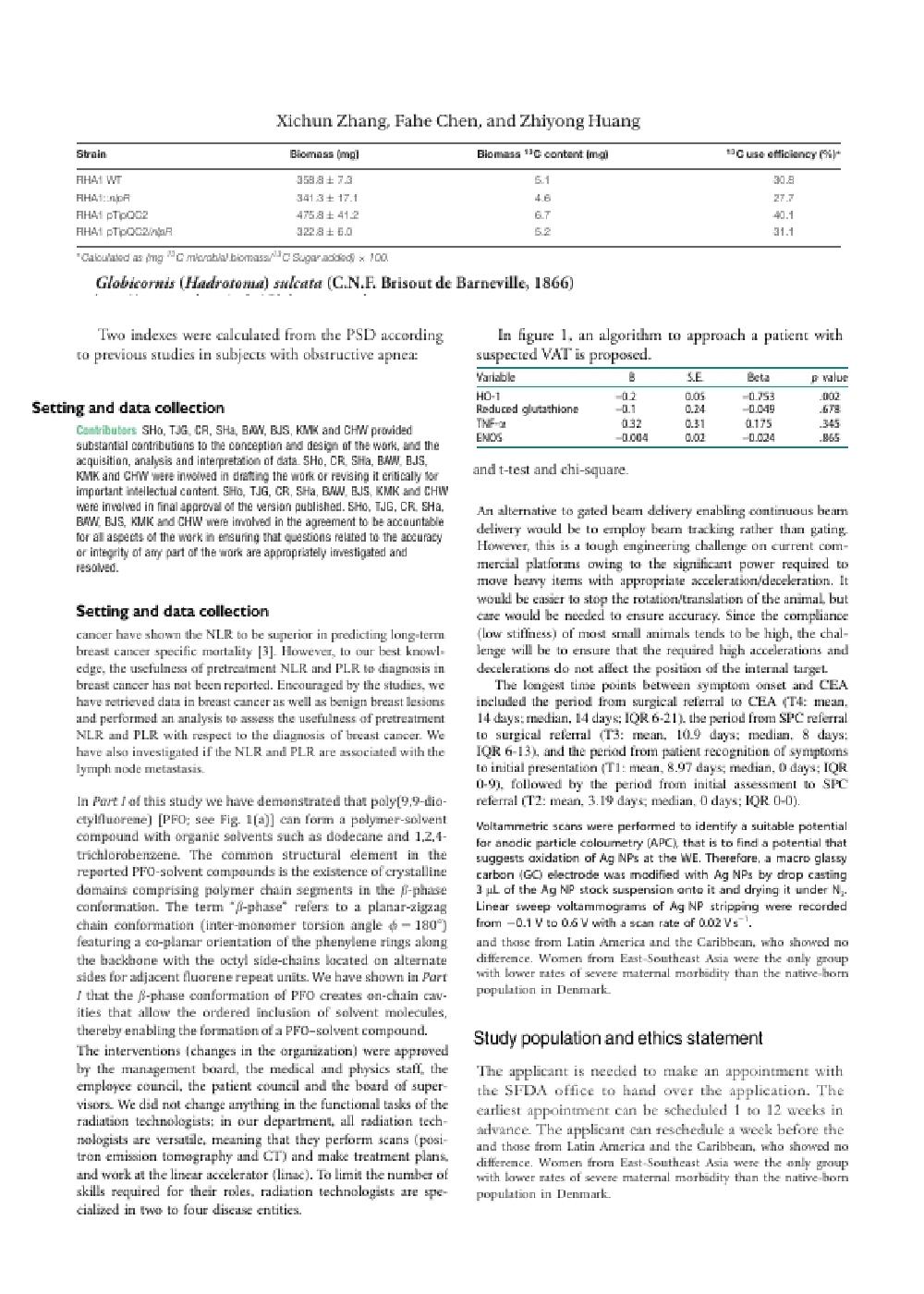} \\

&\includegraphics[width=\publaynetBulkWidth]{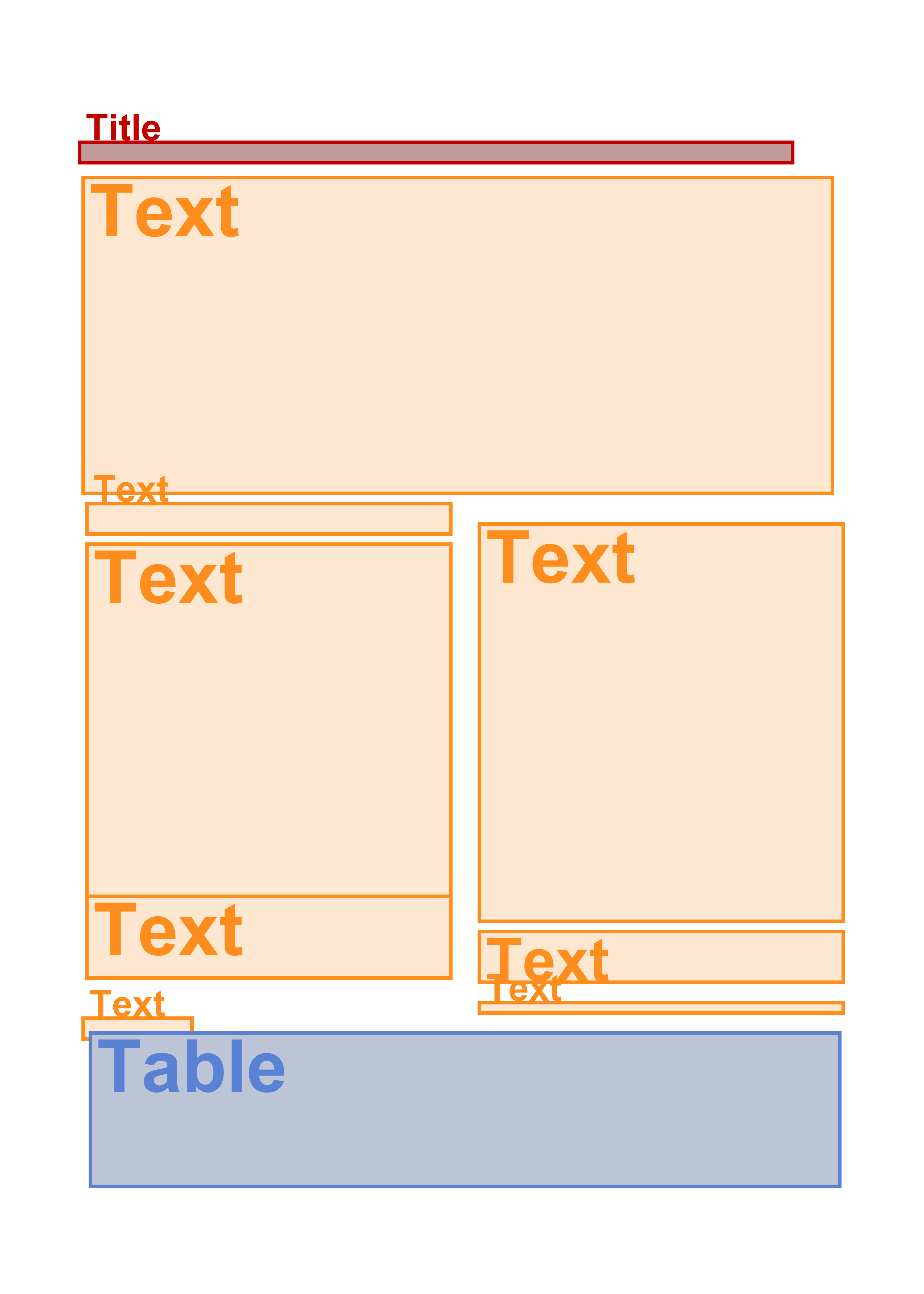} & 
\includegraphics[width=\publaynetBulkWidth]{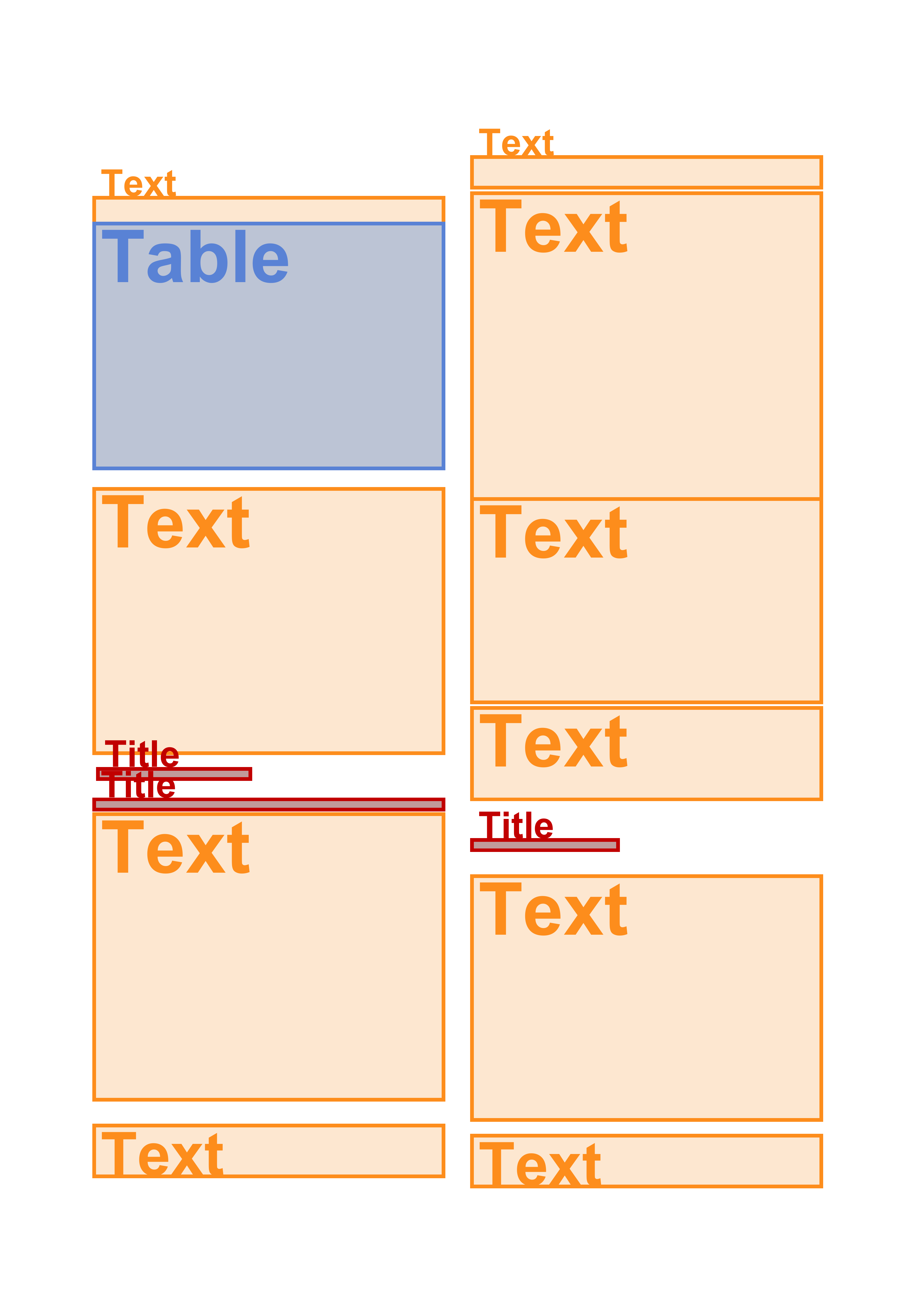} & 
\includegraphics[width=\publaynetBulkWidth]{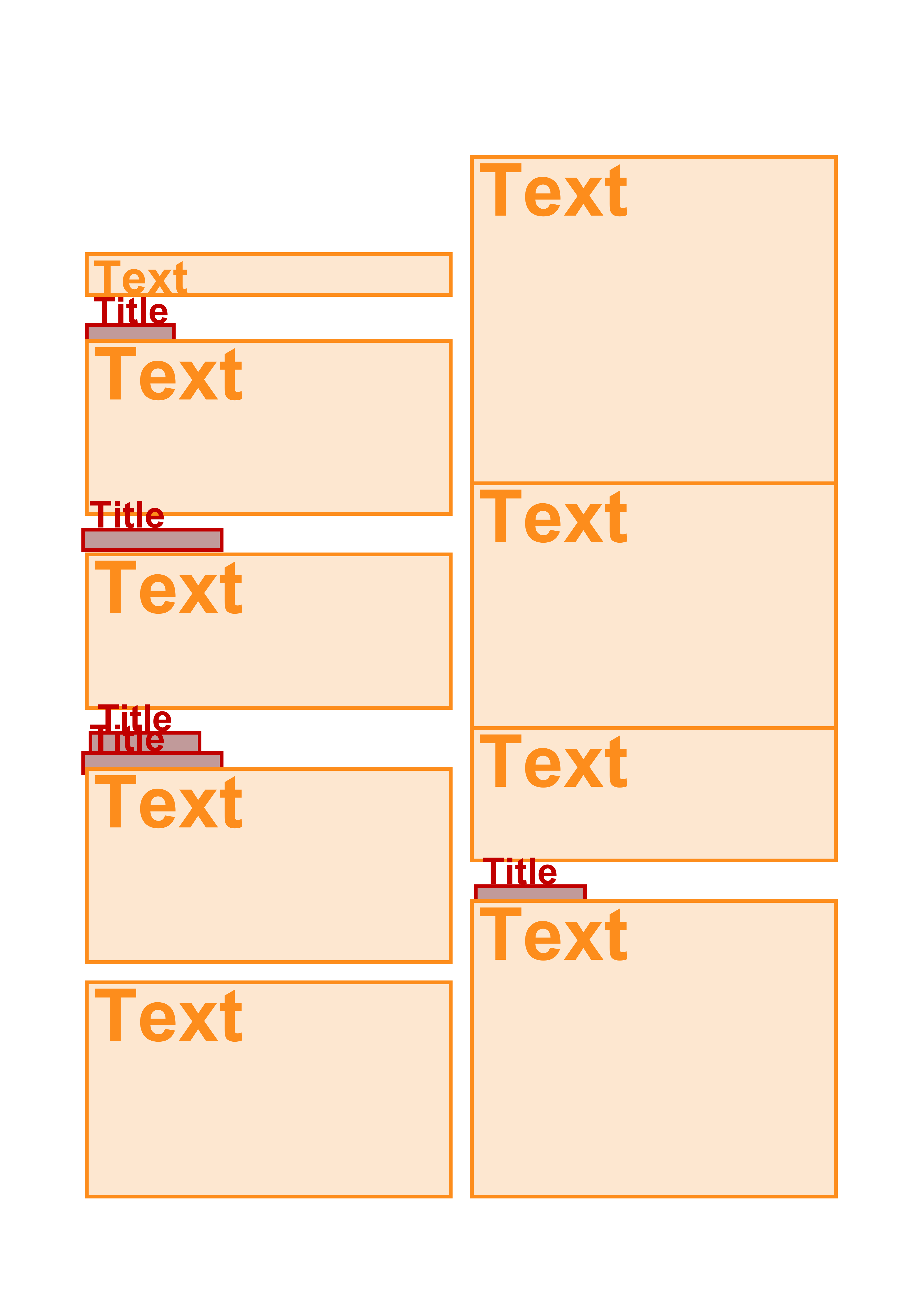} & 
\includegraphics[width=\publaynetBulkWidth]{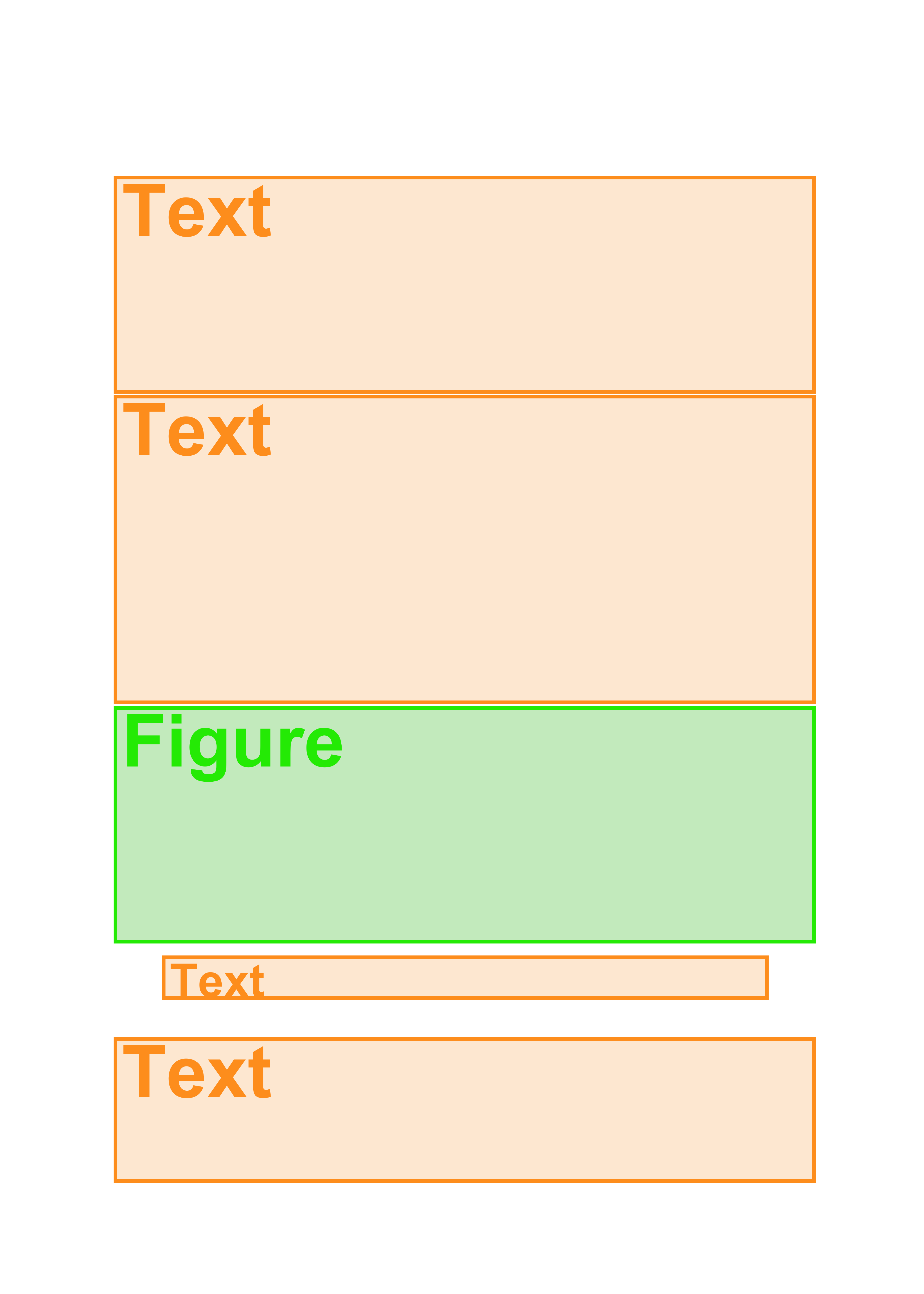} & 
\includegraphics[width=\publaynetBulkWidth]{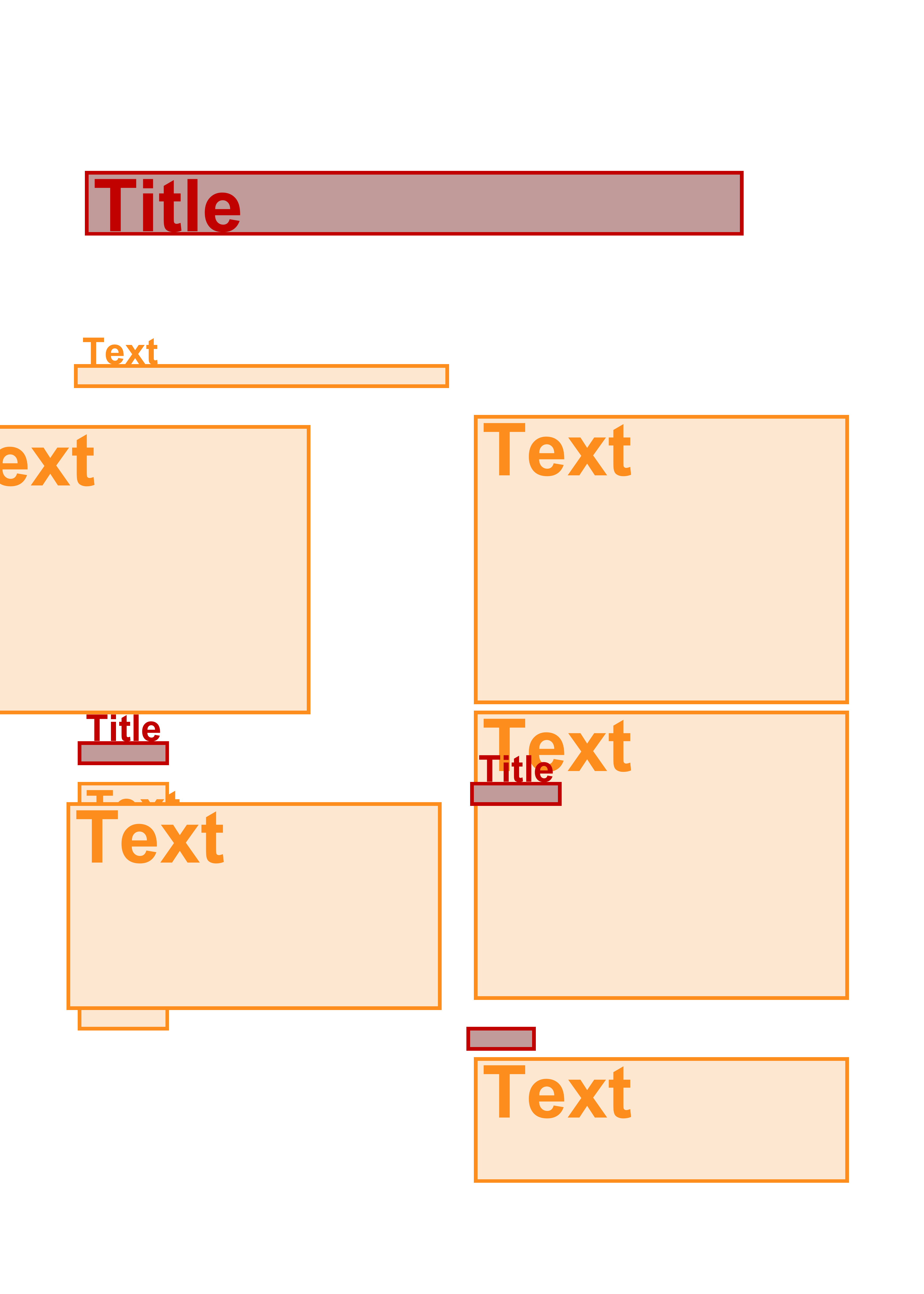} & 
\includegraphics[width=\publaynetBulkWidth]{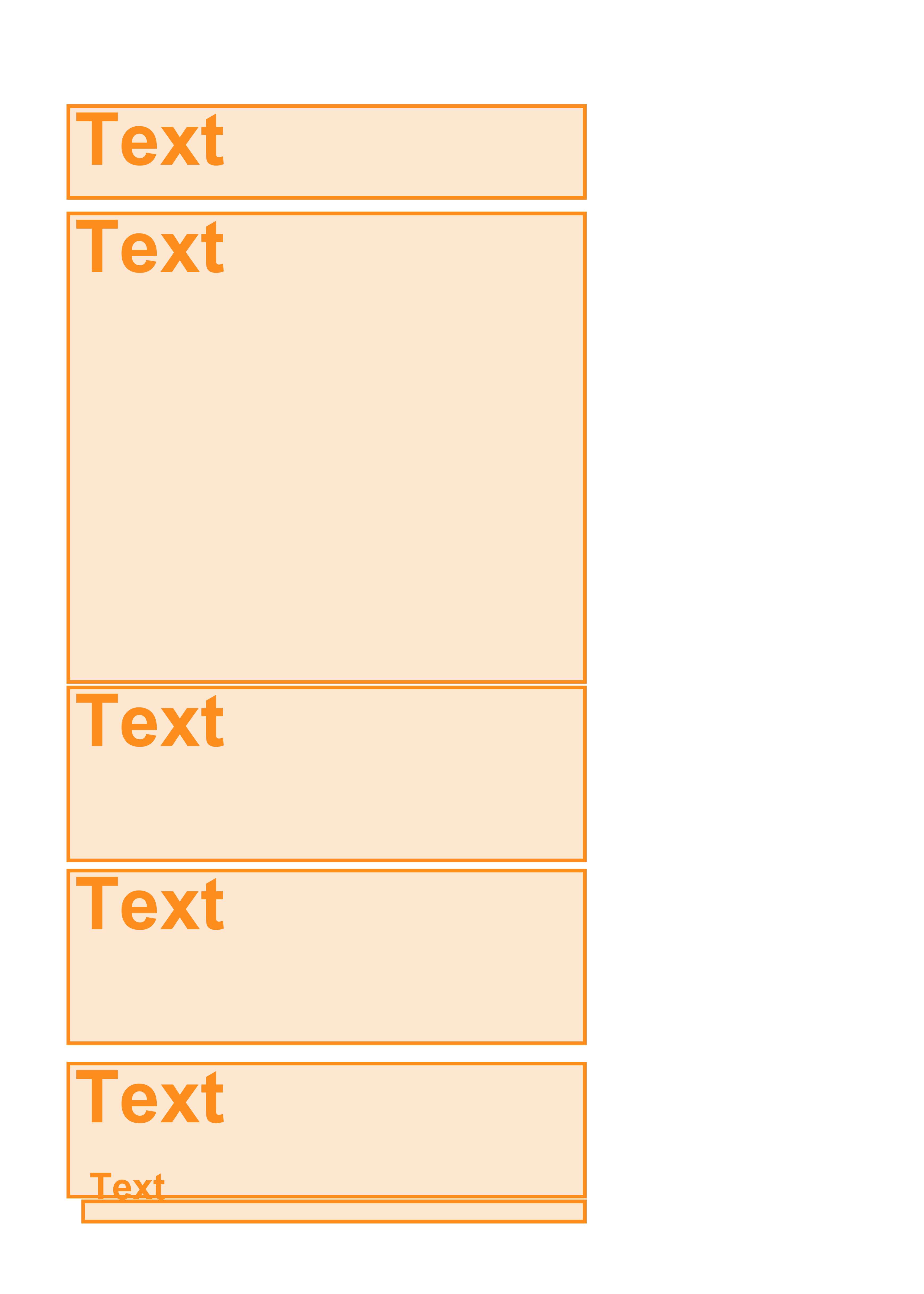} &
\includegraphics[width=\publaynetBulkWidth]{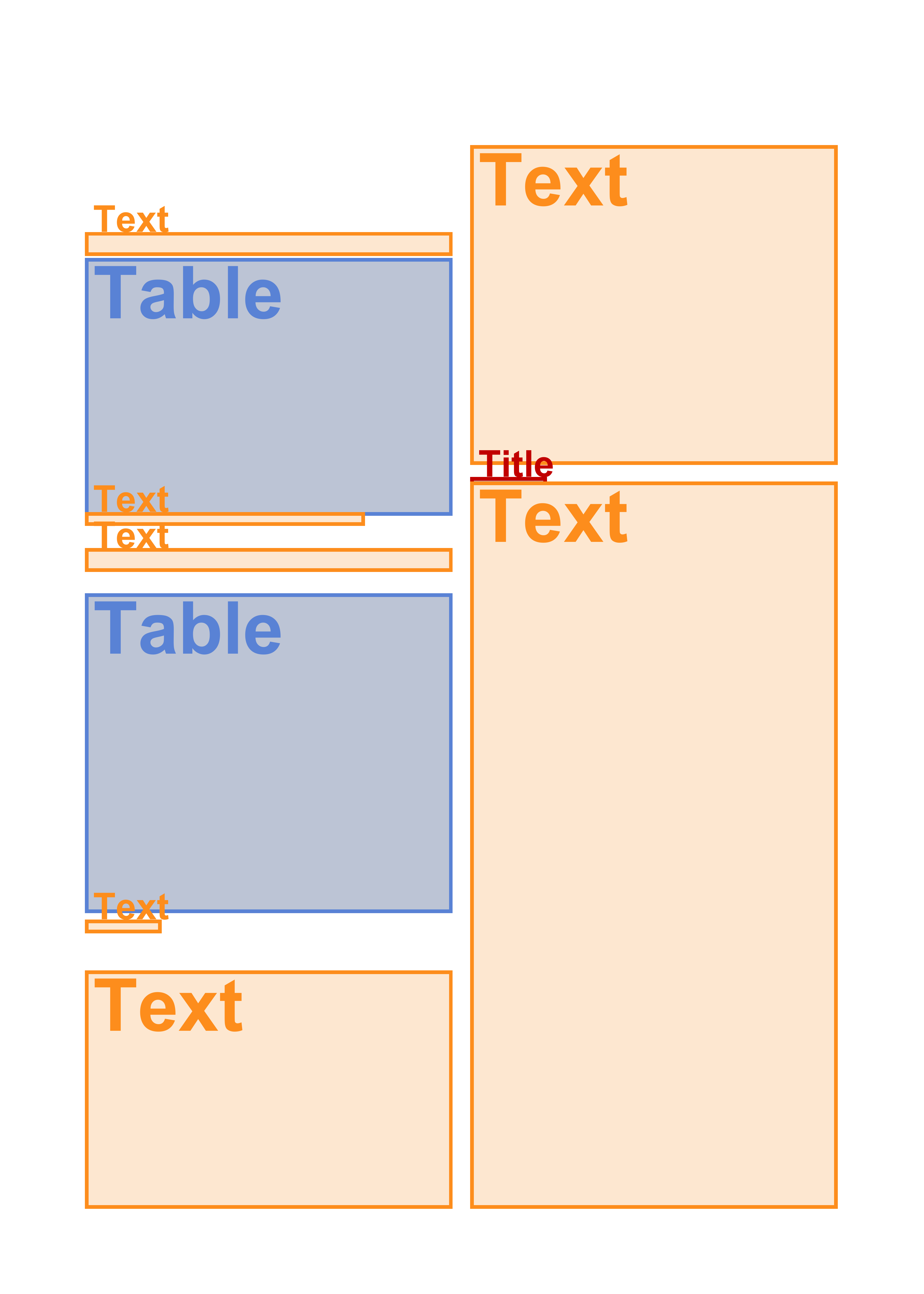}\\ 

&\includegraphics[width=\publaynetBulkWidth]{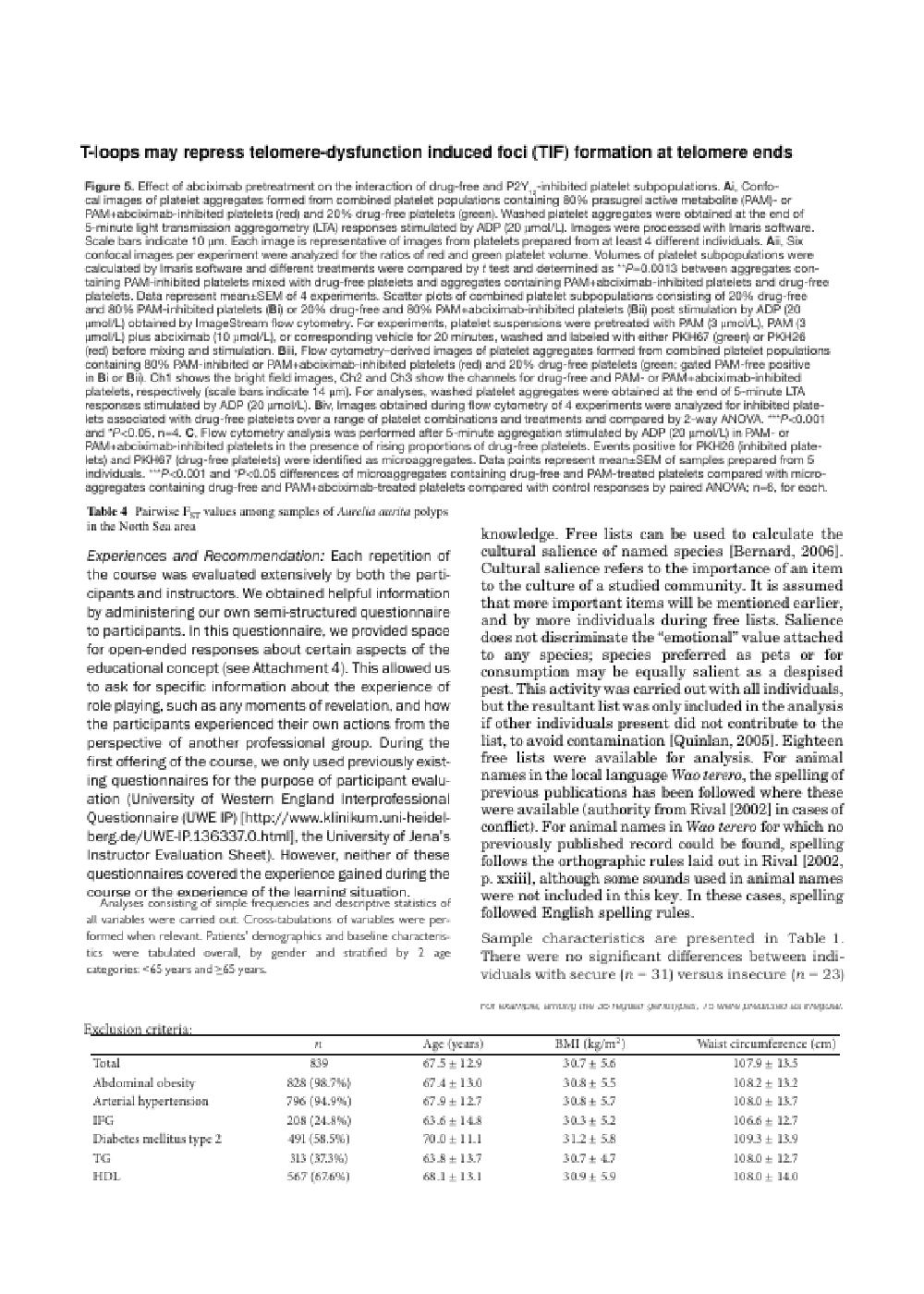} &
\includegraphics[width=\publaynetBulkWidth]{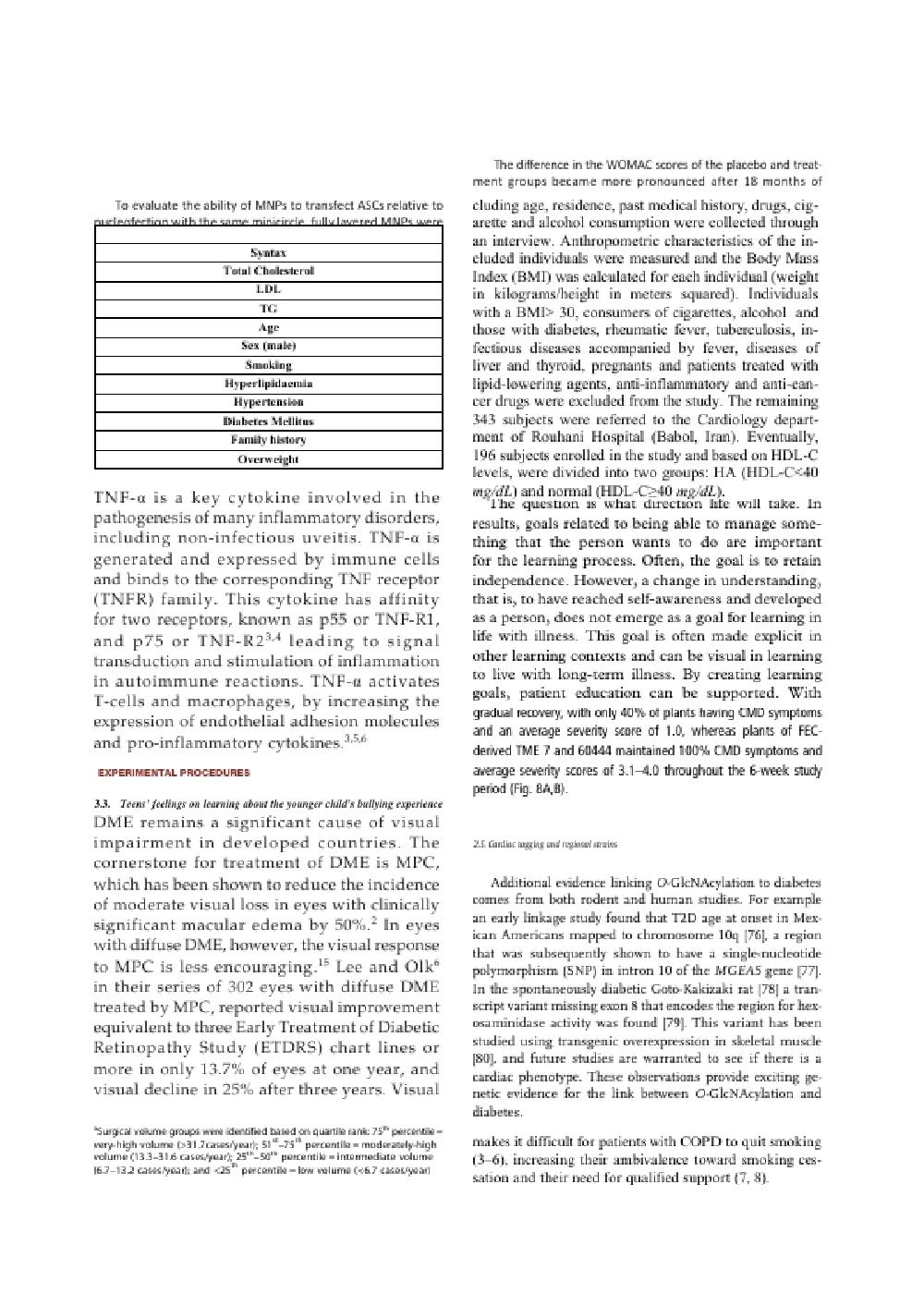} &
\includegraphics[width=\publaynetBulkWidth]{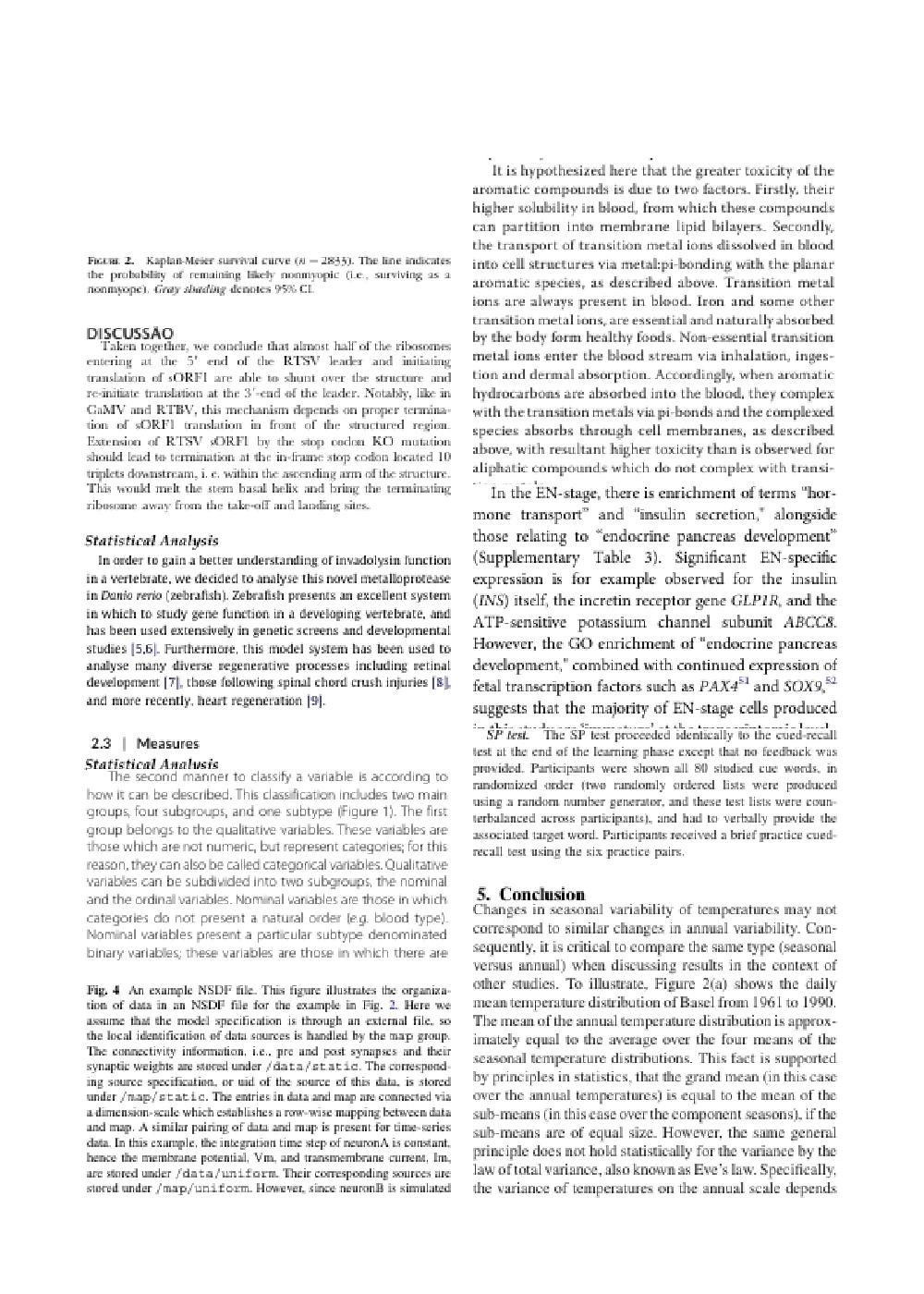} &
\includegraphics[width=\publaynetBulkWidth]{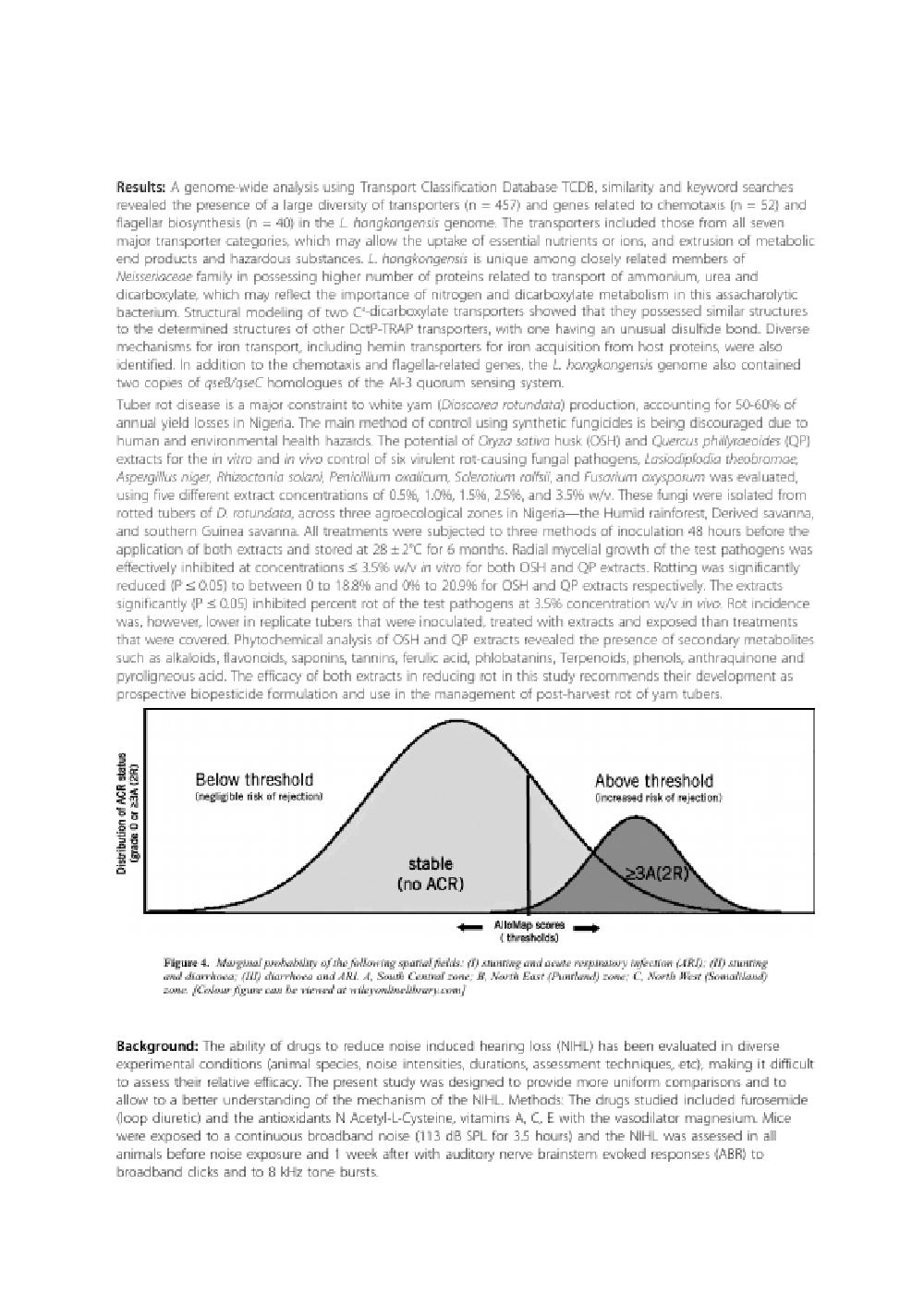} &
\includegraphics[width=\publaynetBulkWidth]{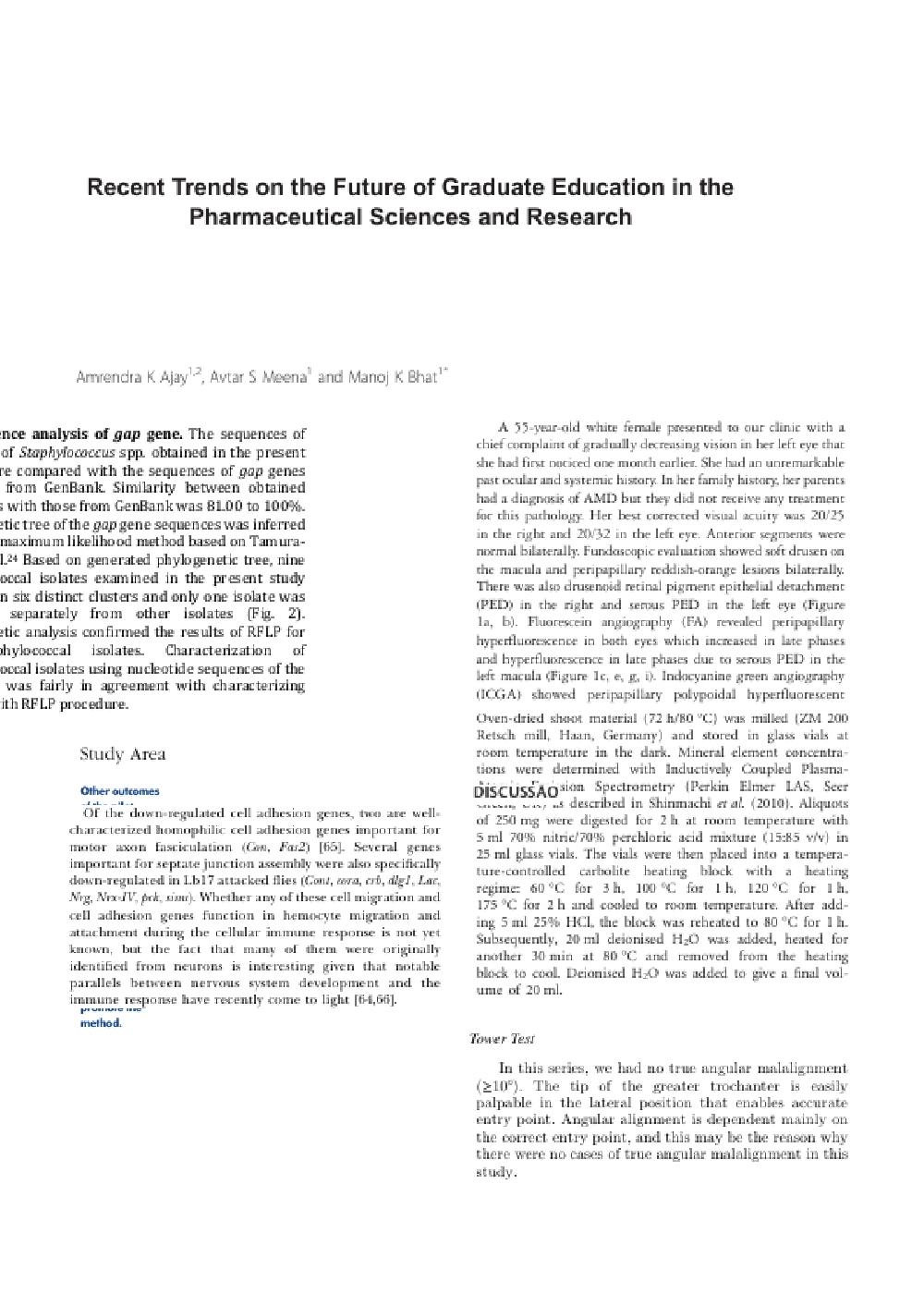} &
\includegraphics[width=\publaynetBulkWidth]{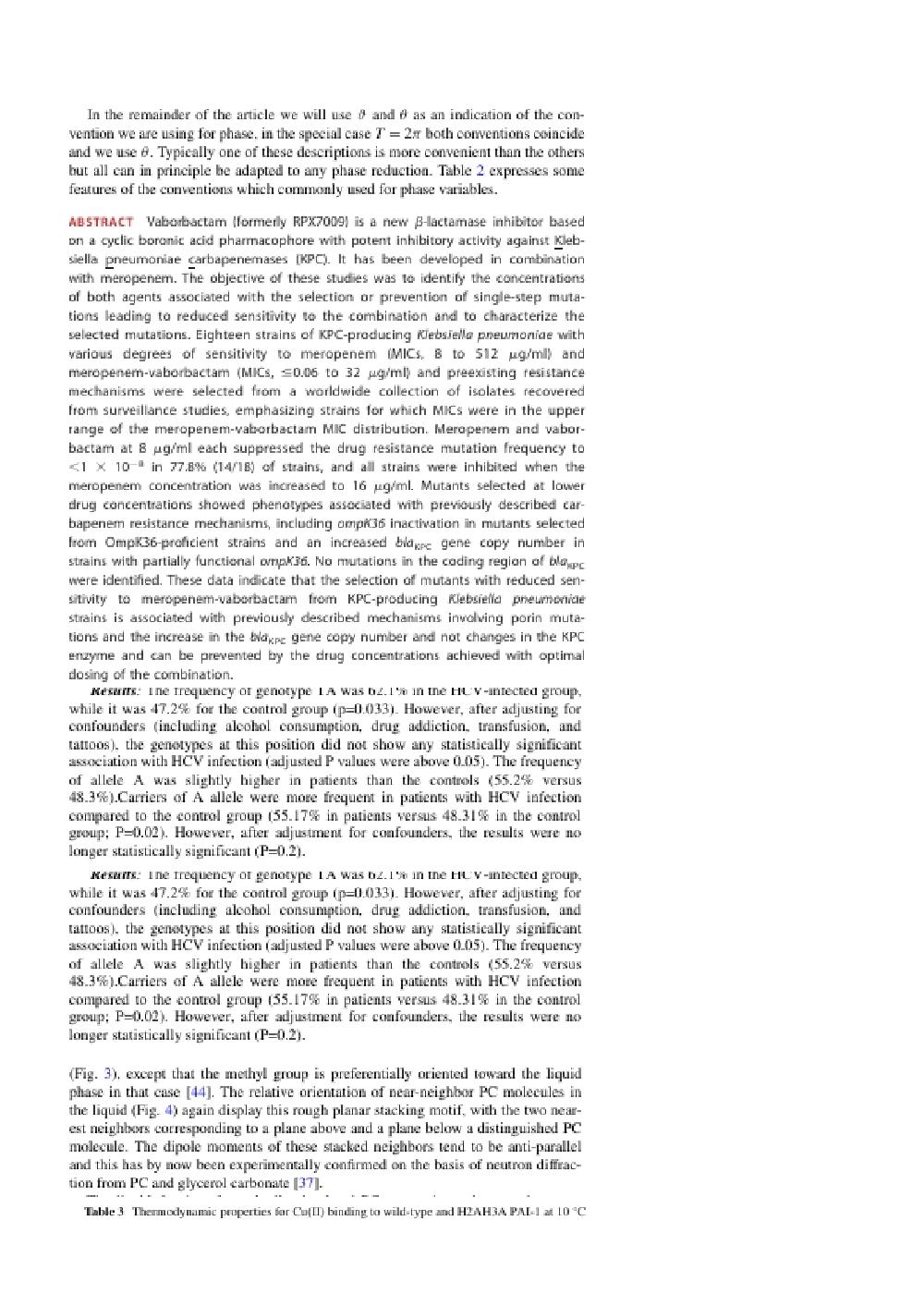} &
\includegraphics[width=\publaynetBulkWidth]{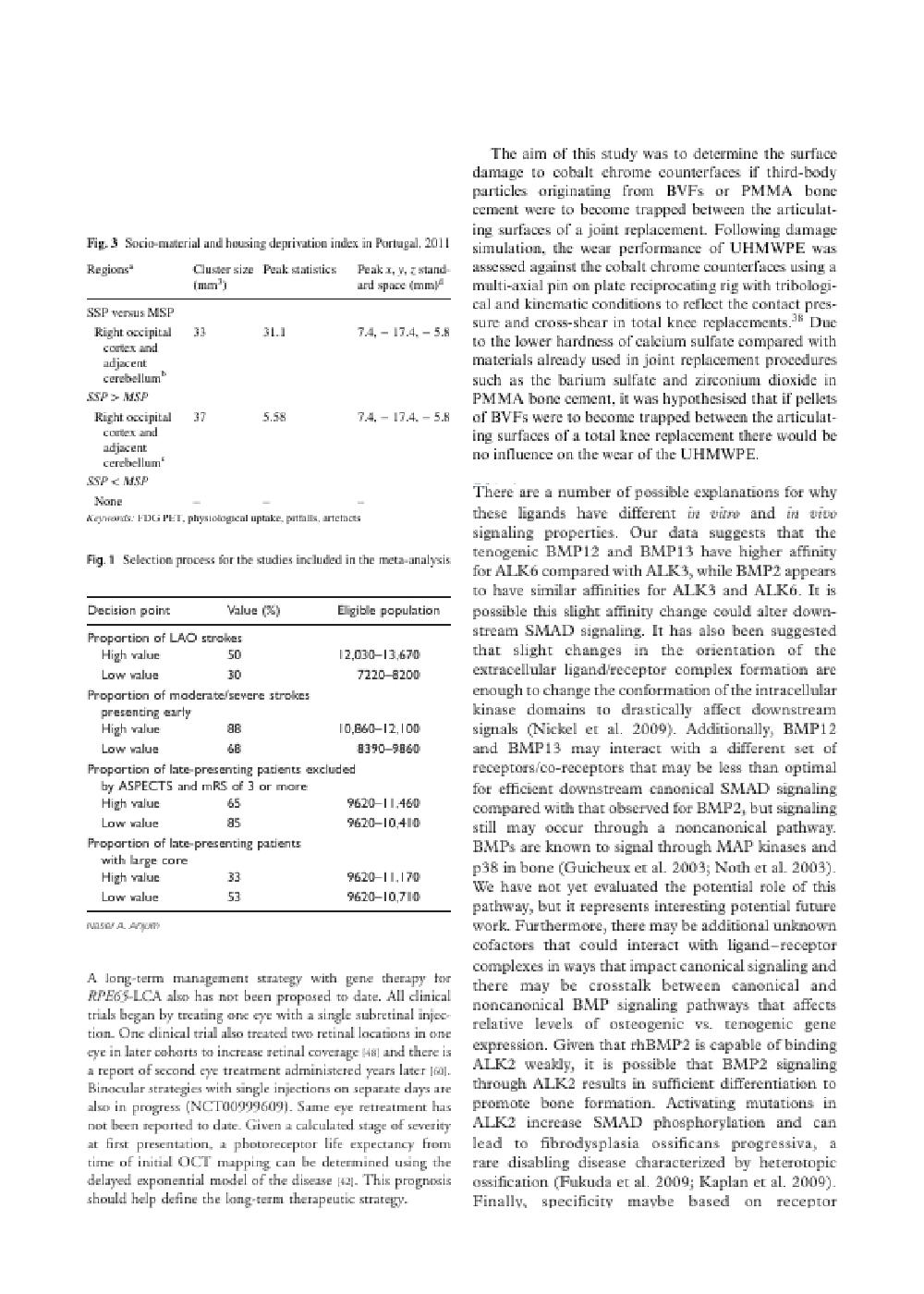}\\

    \end{tabular}
    \caption{Additional synthesized layouts on PubLayNet using an autoregressive decoder and the result of feeding the layout to a document renderer.}
    \label{fig:bulk_publaynet}
\end{figure}
\begin{figure}\ContinuedFloat
    \setlength{\publaynetBulkWidth}{0.15\linewidth}
    \setlength{\tabcolsep}{2pt}
    \centering
    \begin{tabular}{cccccc}
\includegraphics[width=\publaynetBulkWidth]{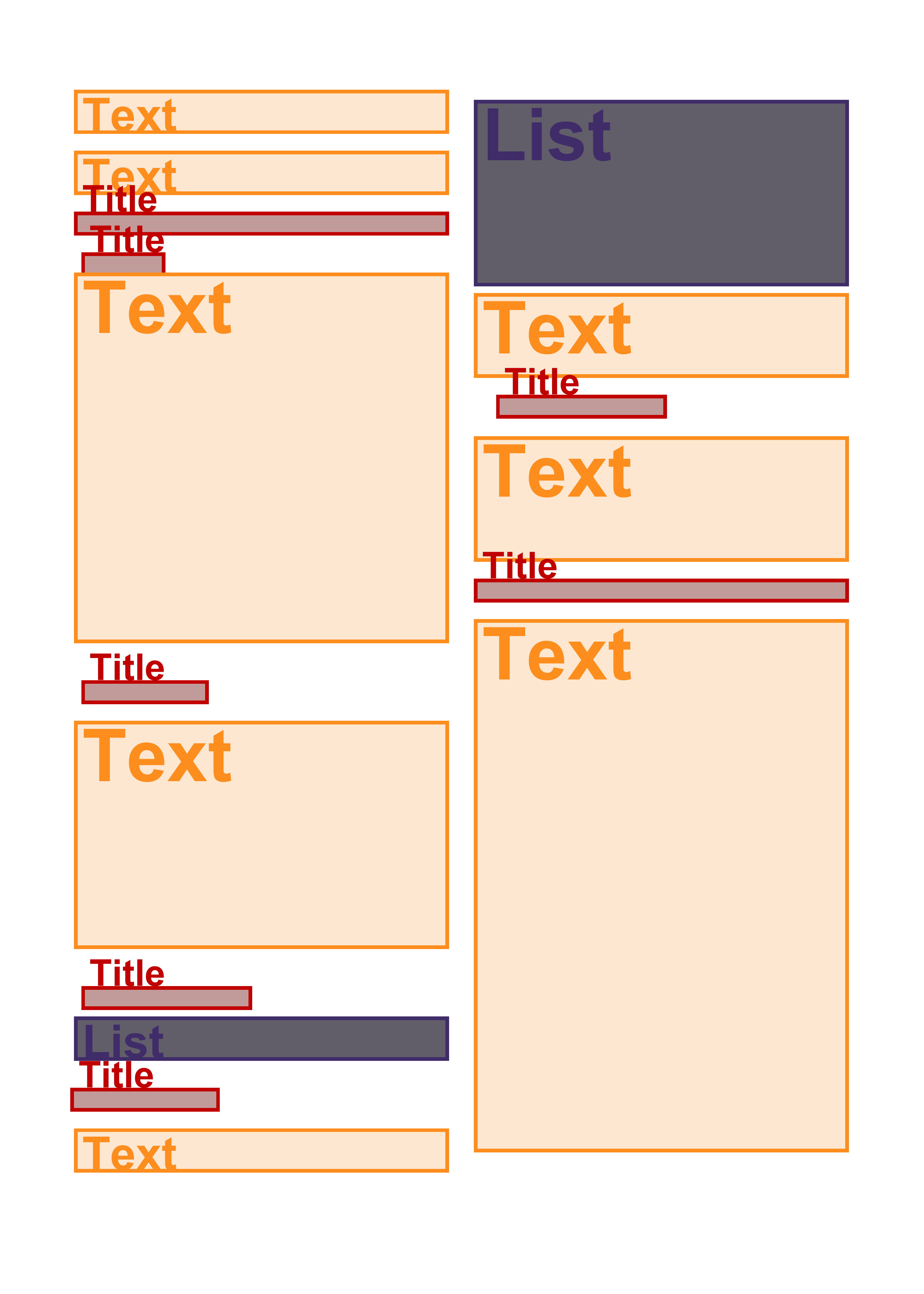} & 
\includegraphics[width=\publaynetBulkWidth]{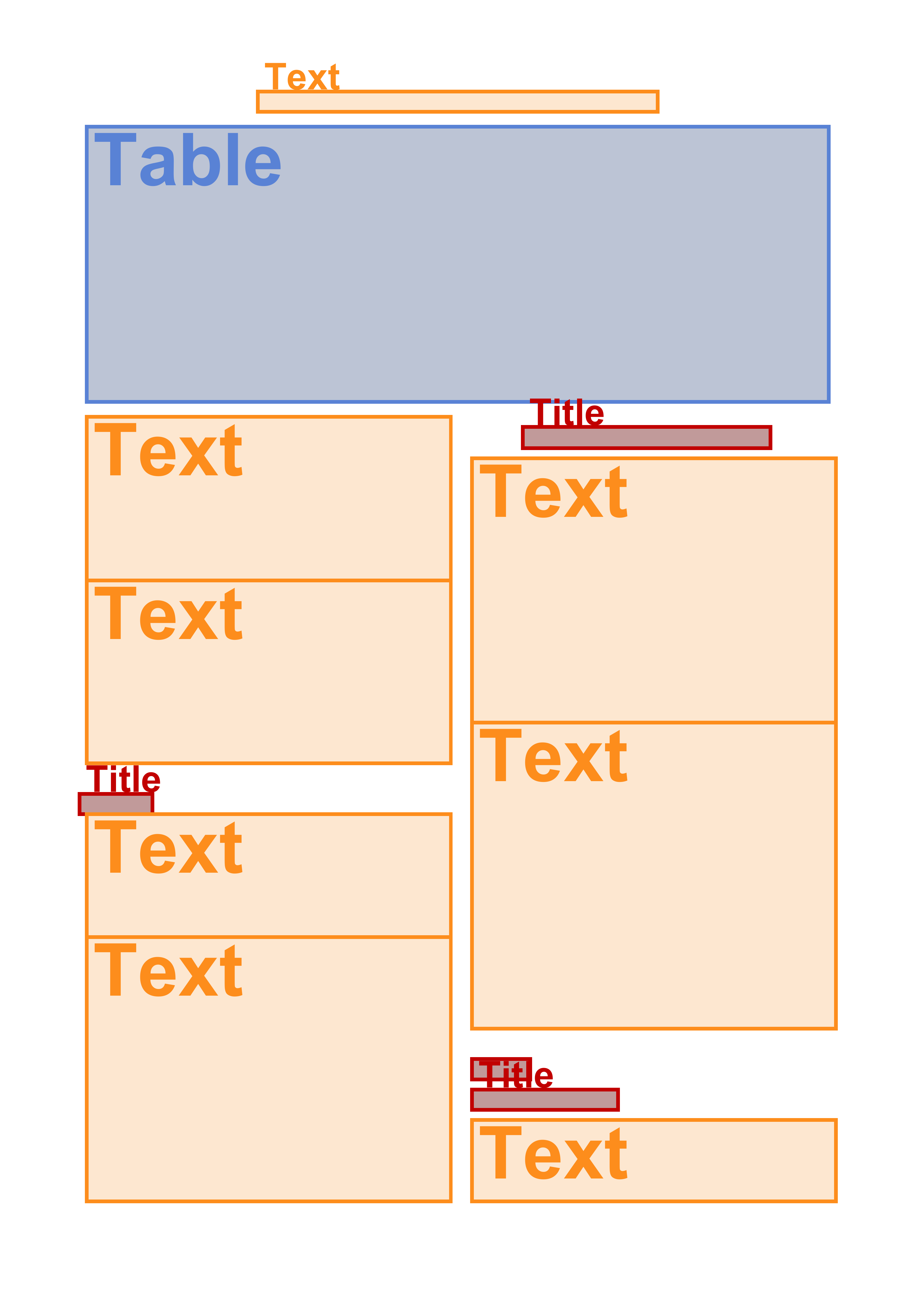} & 
\includegraphics[width=\publaynetBulkWidth]{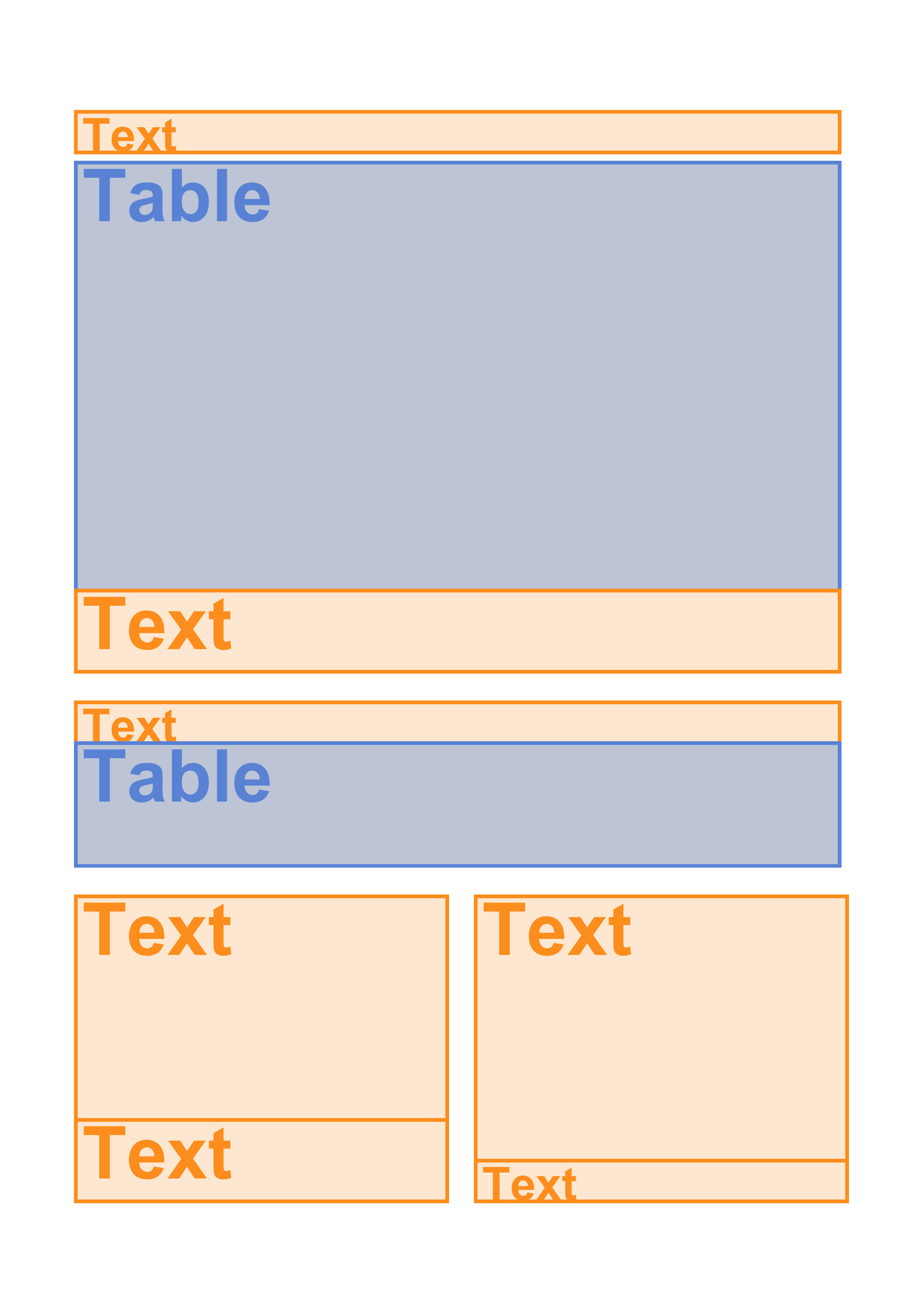} & 
\includegraphics[width=\publaynetBulkWidth]{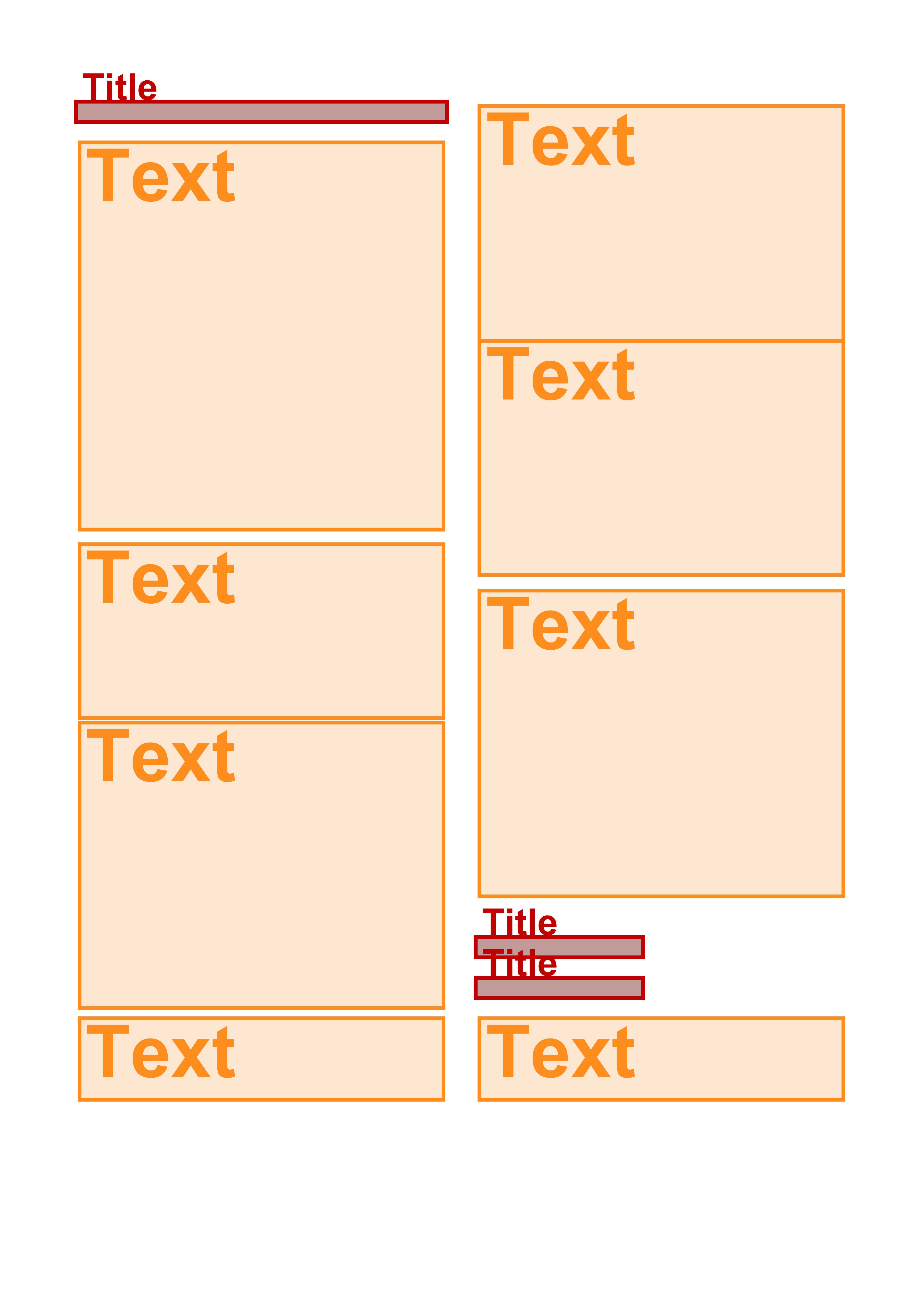} & 
\includegraphics[width=\publaynetBulkWidth]{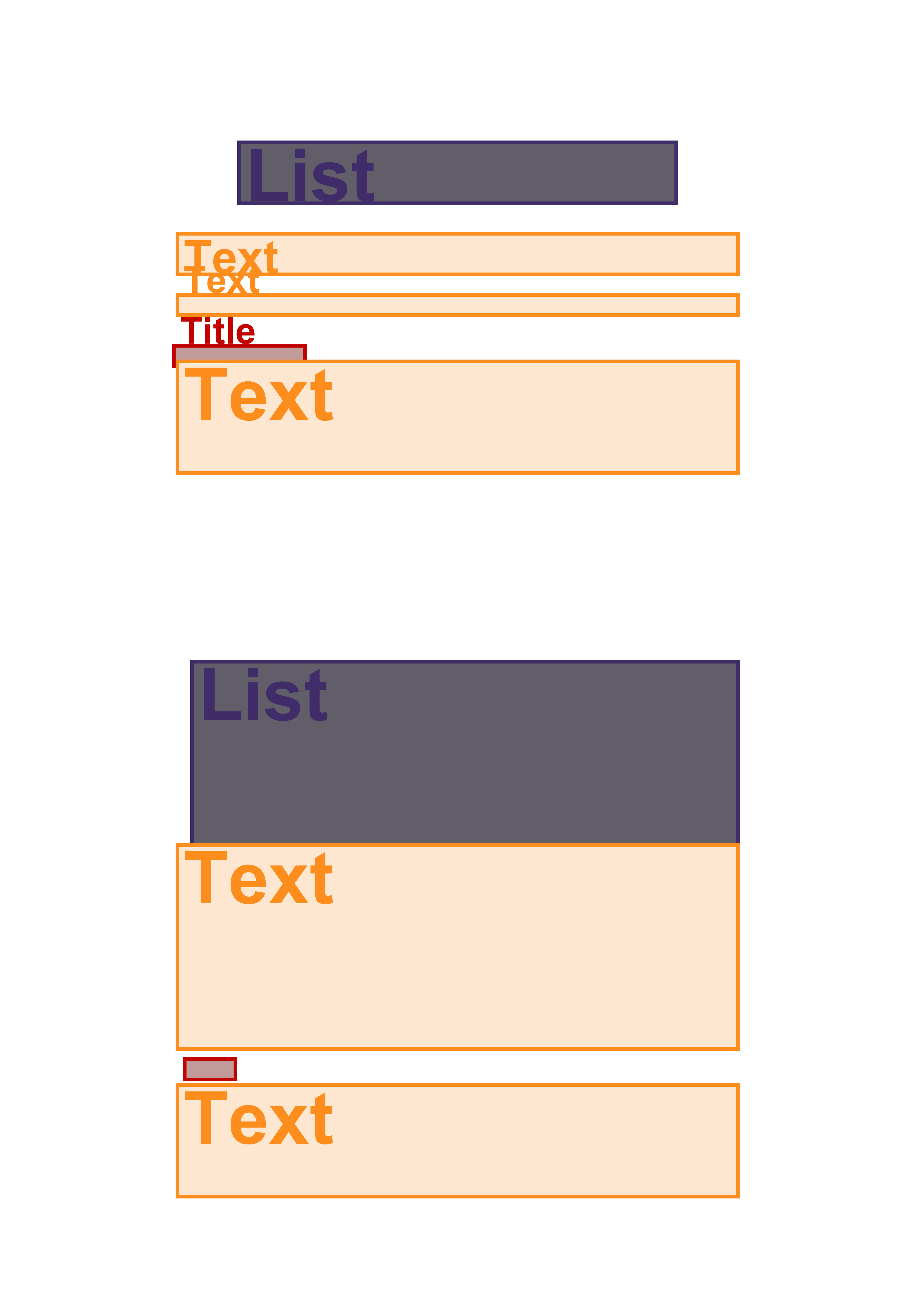} & 
\includegraphics[width=\publaynetBulkWidth]{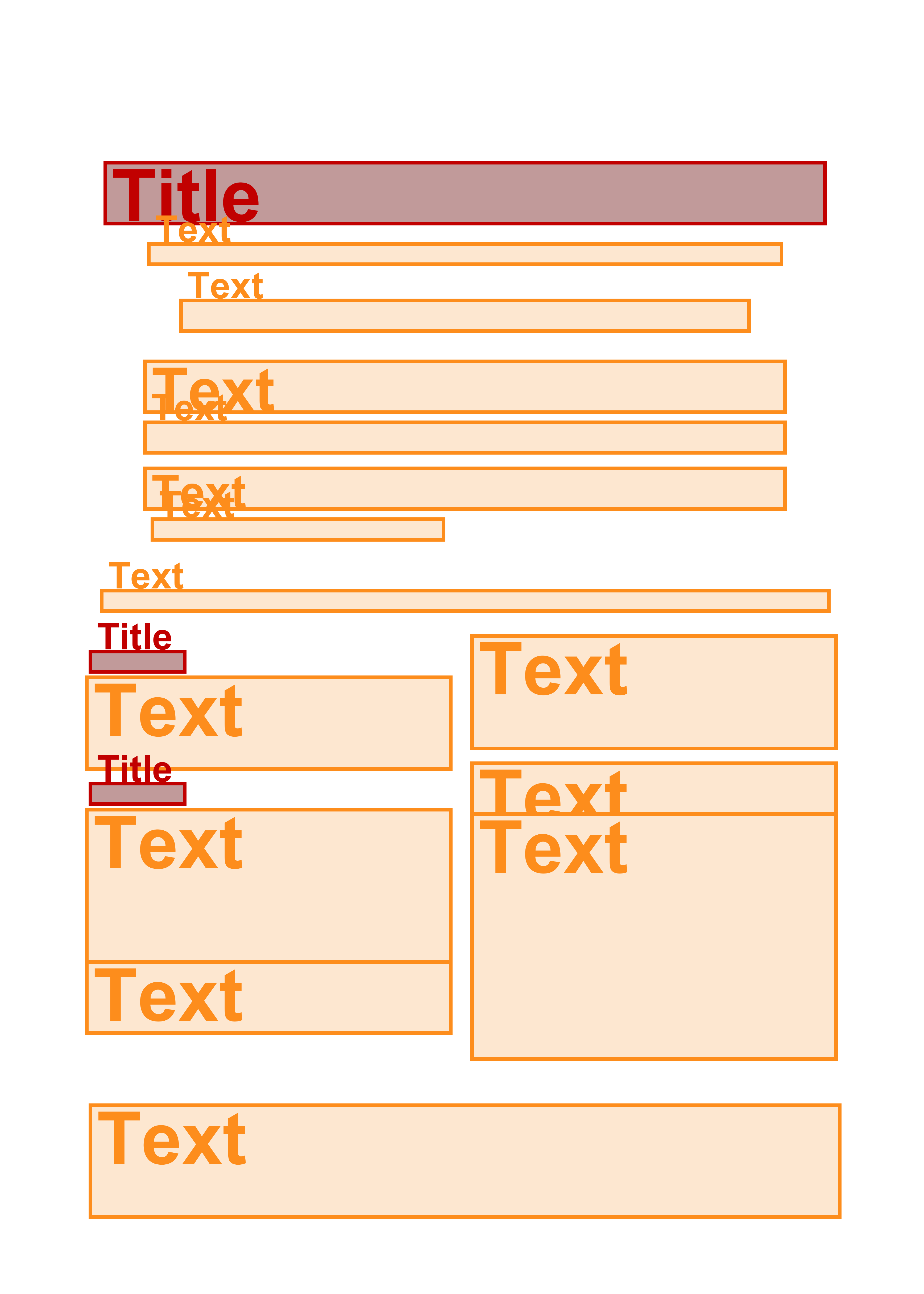} \\ 
\includegraphics[width=\publaynetBulkWidth]{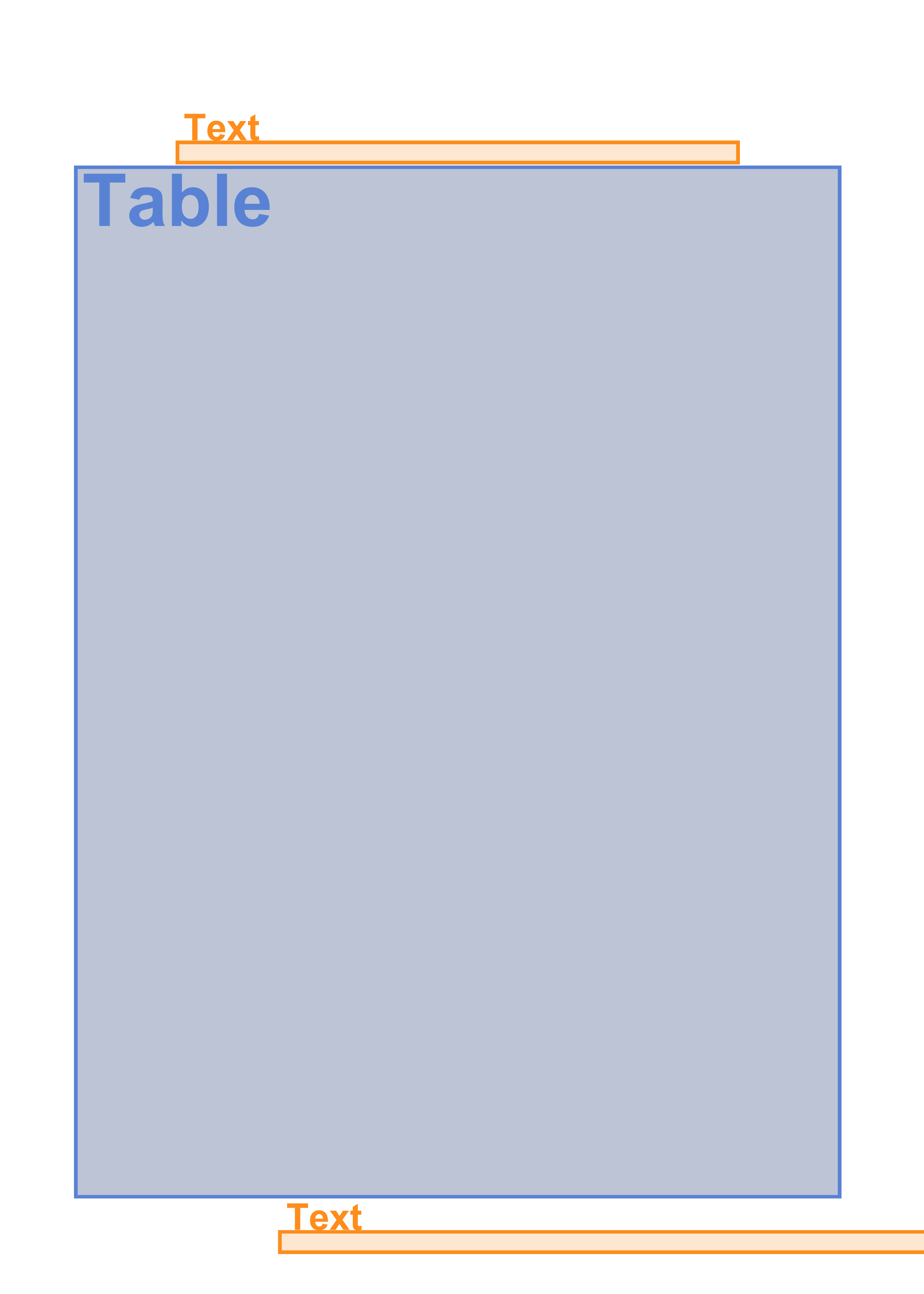} & 
\includegraphics[width=\publaynetBulkWidth]{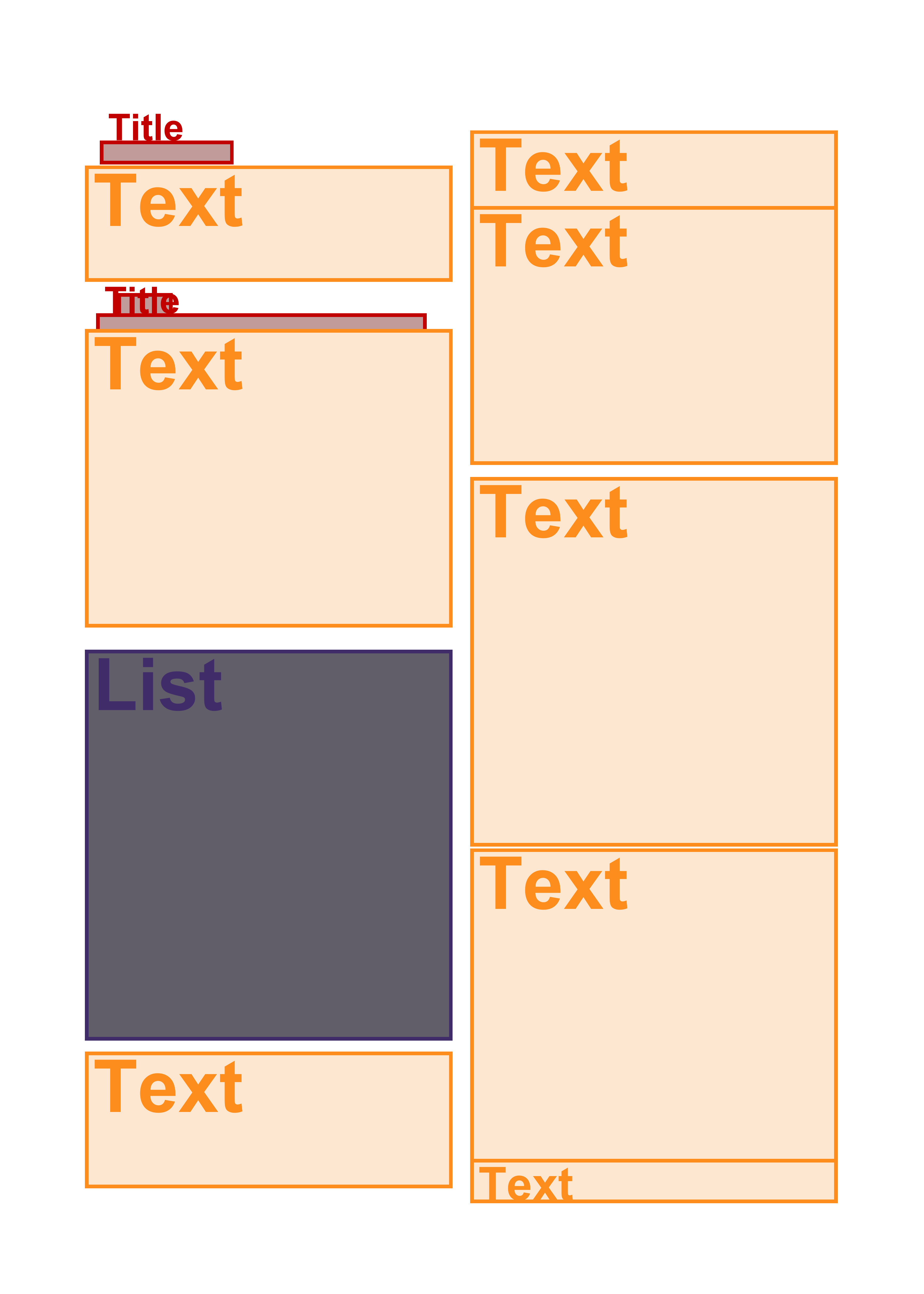} & 
\includegraphics[width=\publaynetBulkWidth]{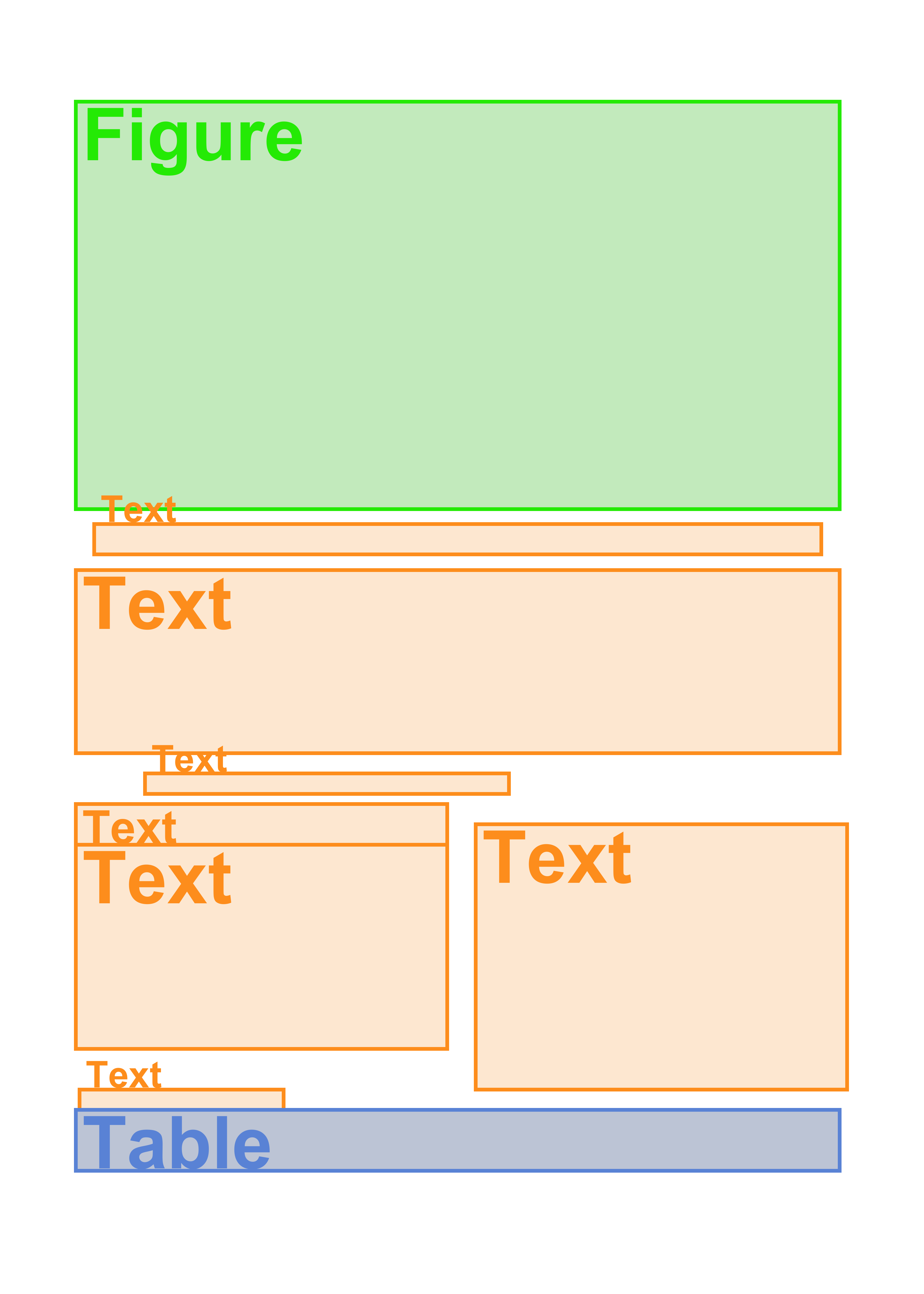} & 
\includegraphics[width=\publaynetBulkWidth]{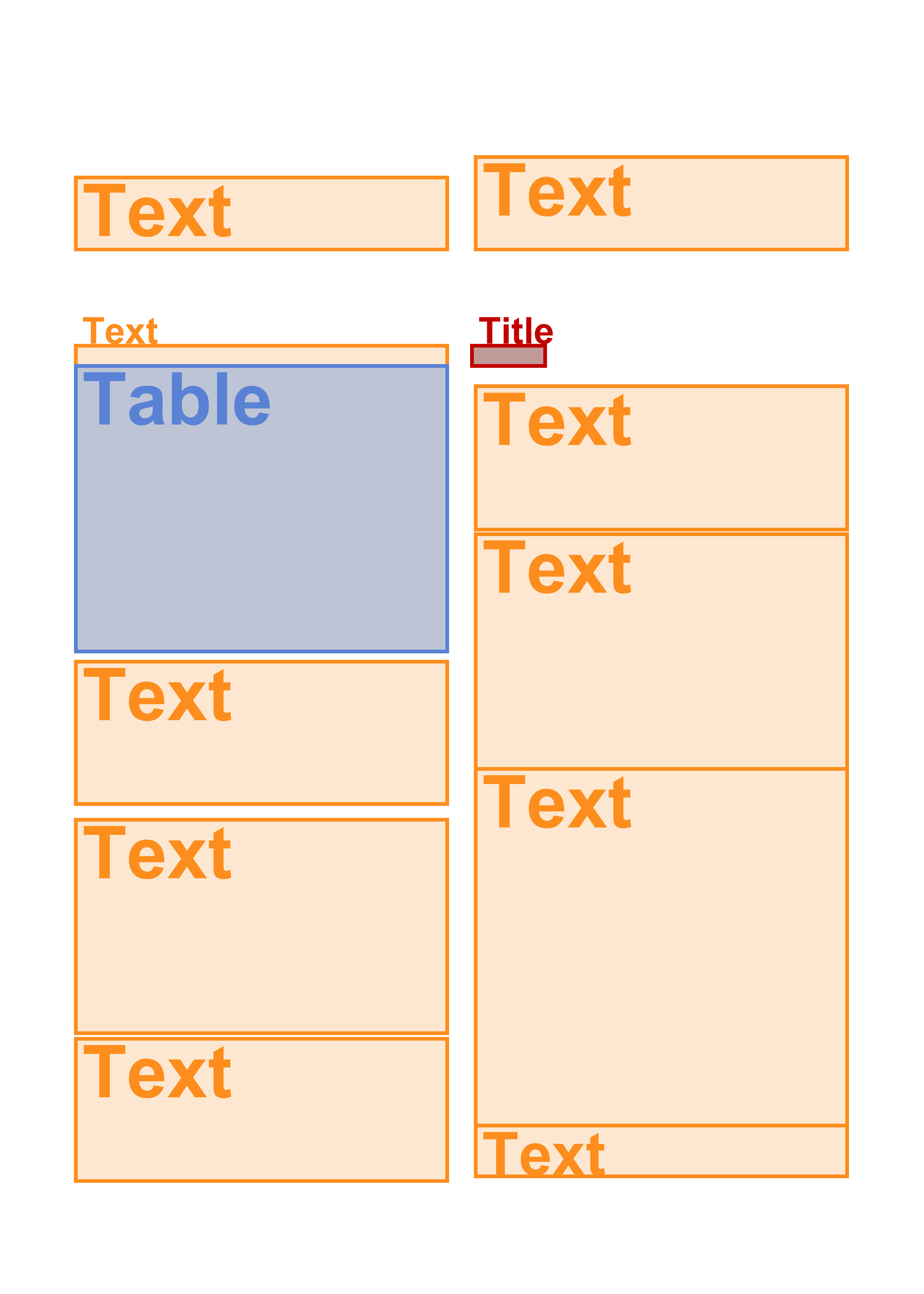} & 
\includegraphics[width=\publaynetBulkWidth]{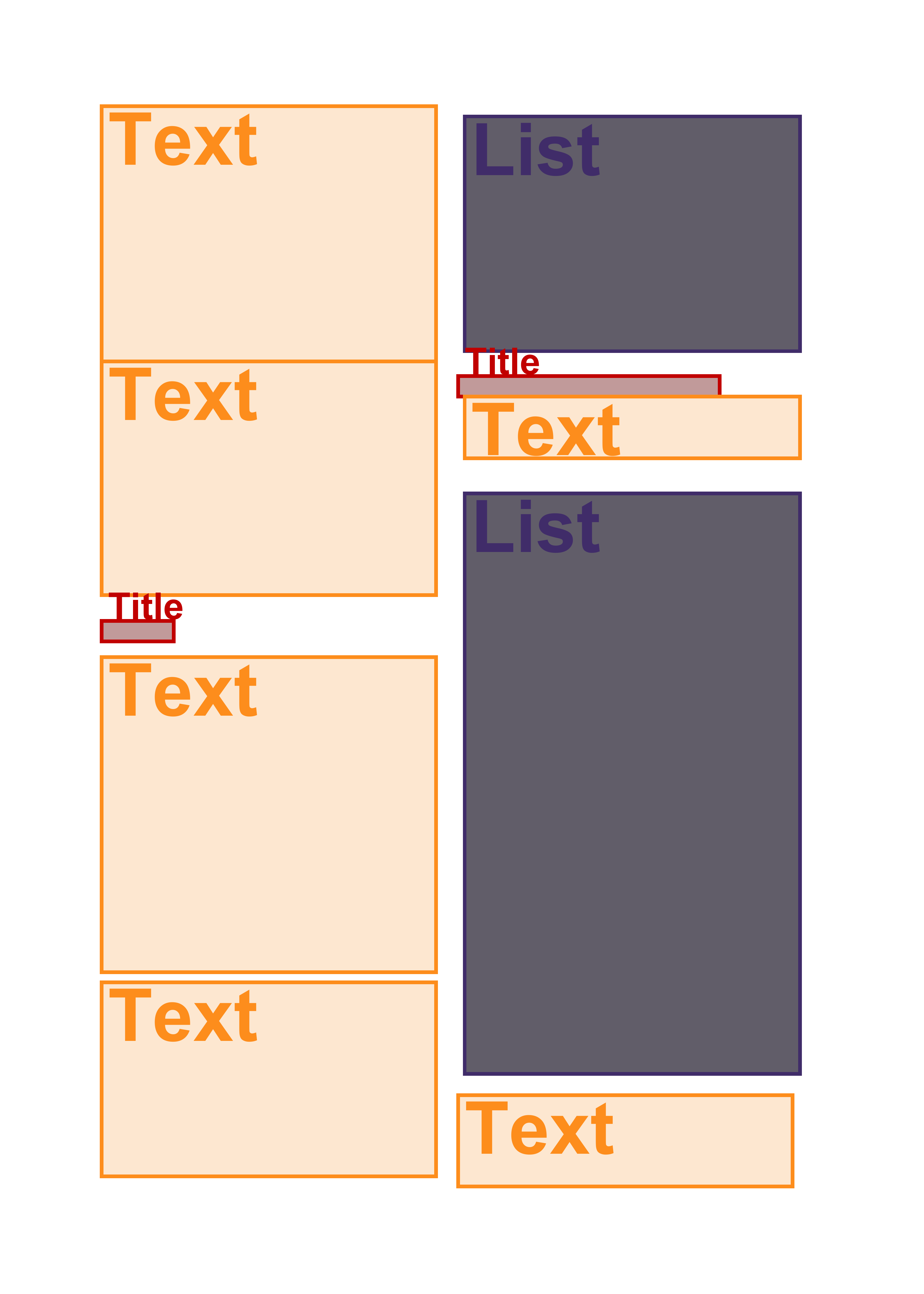} & 
\includegraphics[width=\publaynetBulkWidth]{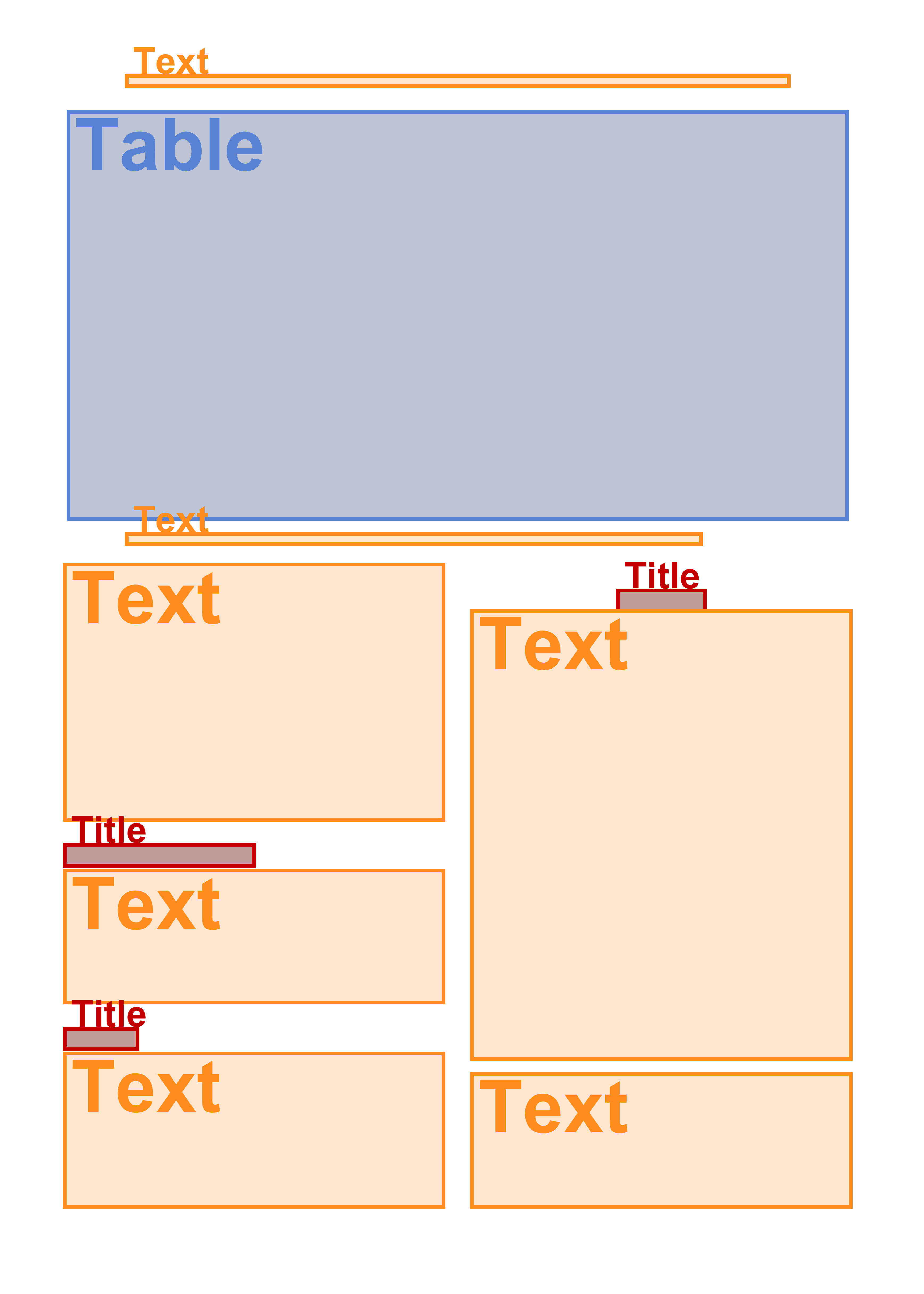} \\ 
\includegraphics[width=\publaynetBulkWidth]{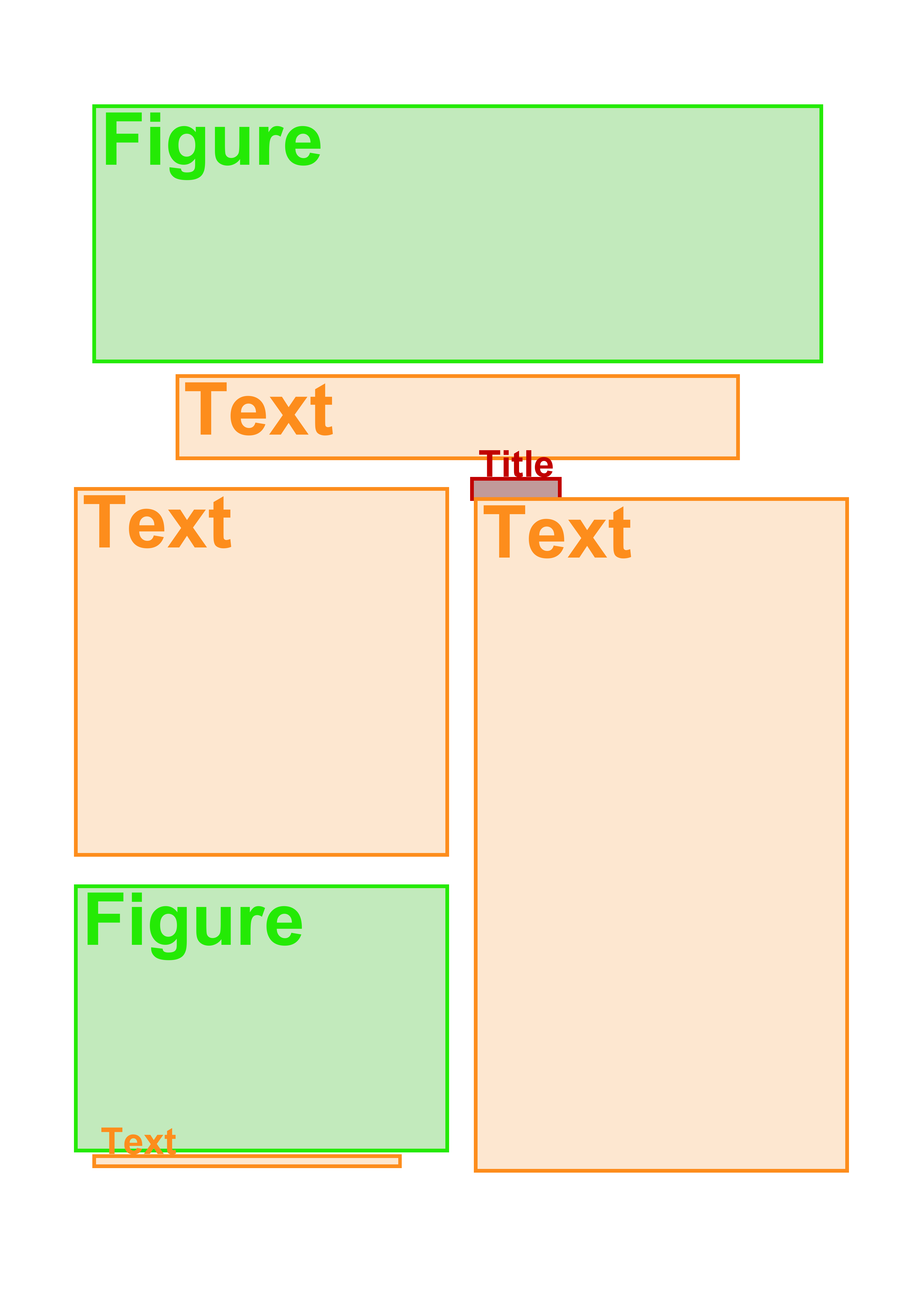} & 
\includegraphics[width=\publaynetBulkWidth]{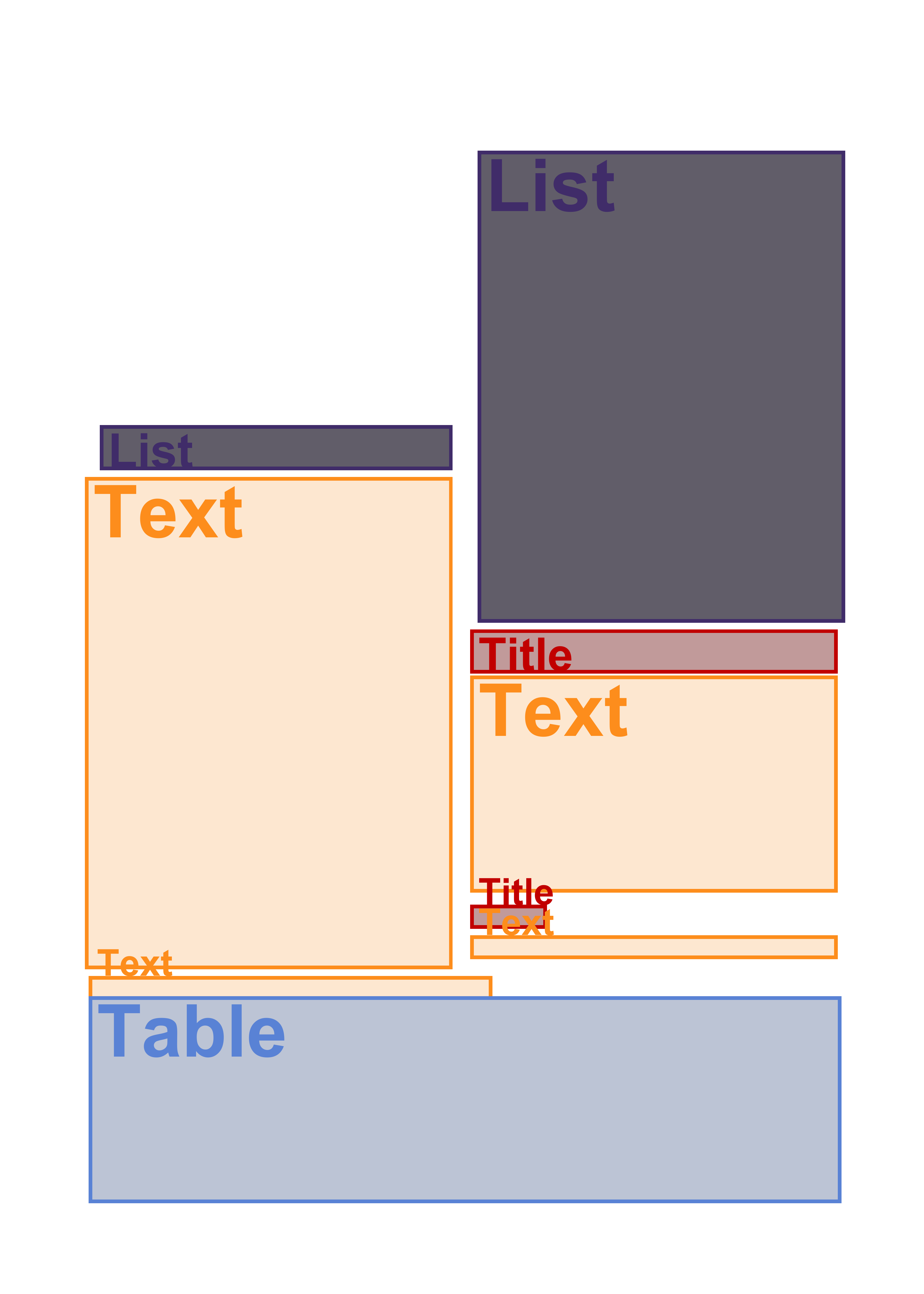} & 
\includegraphics[width=\publaynetBulkWidth]{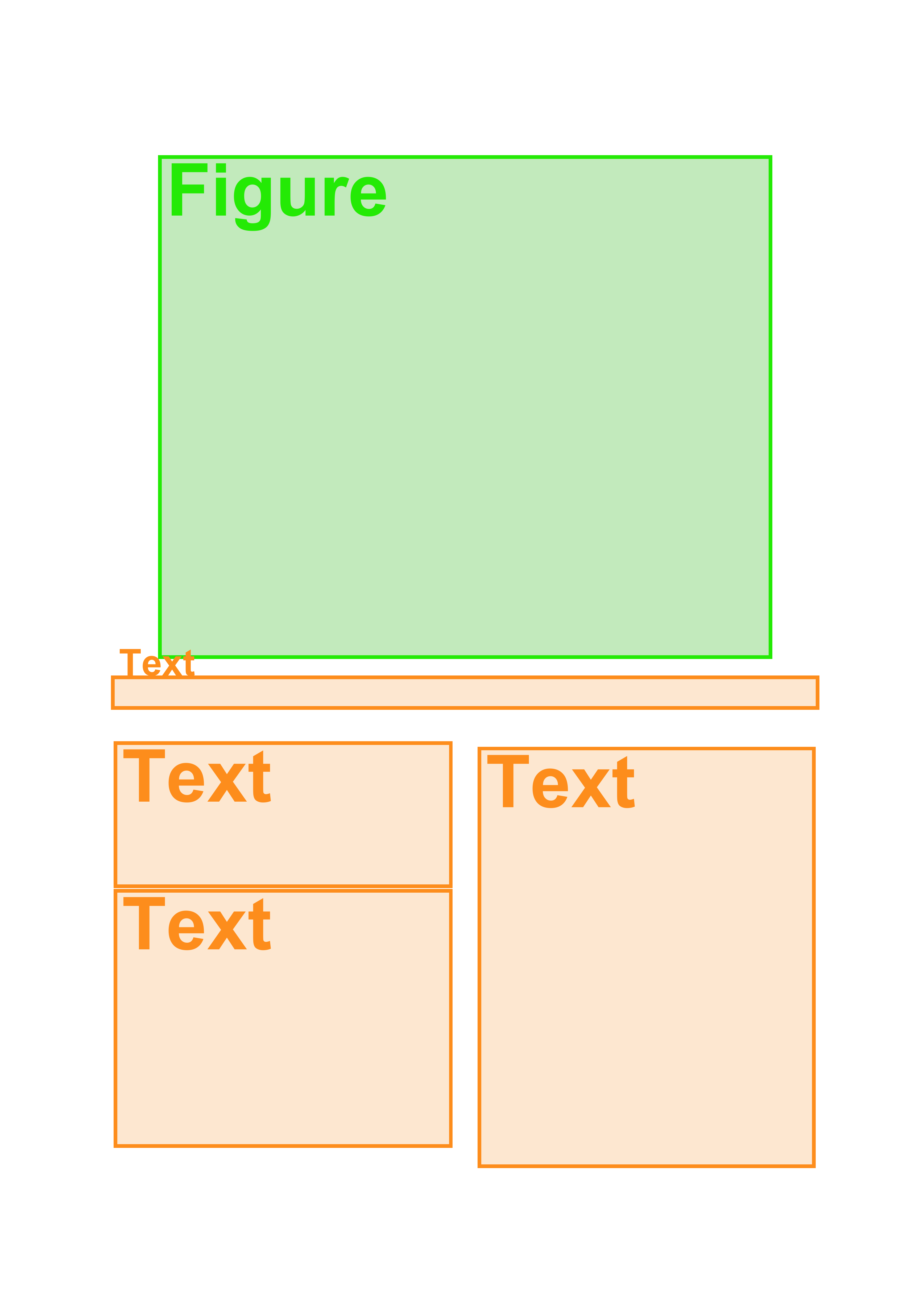} & 
\includegraphics[width=\publaynetBulkWidth]{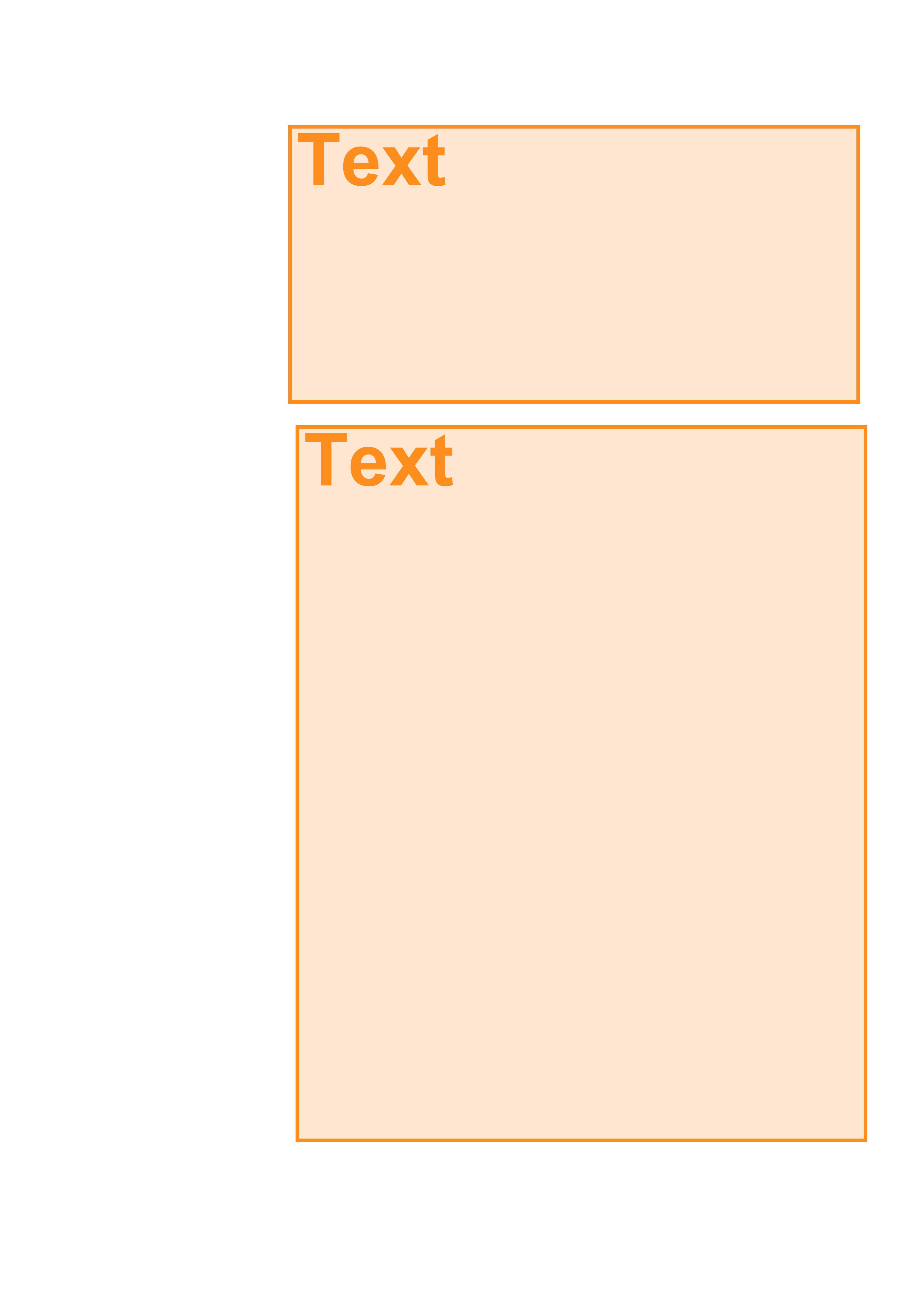} & 
\includegraphics[width=\publaynetBulkWidth]{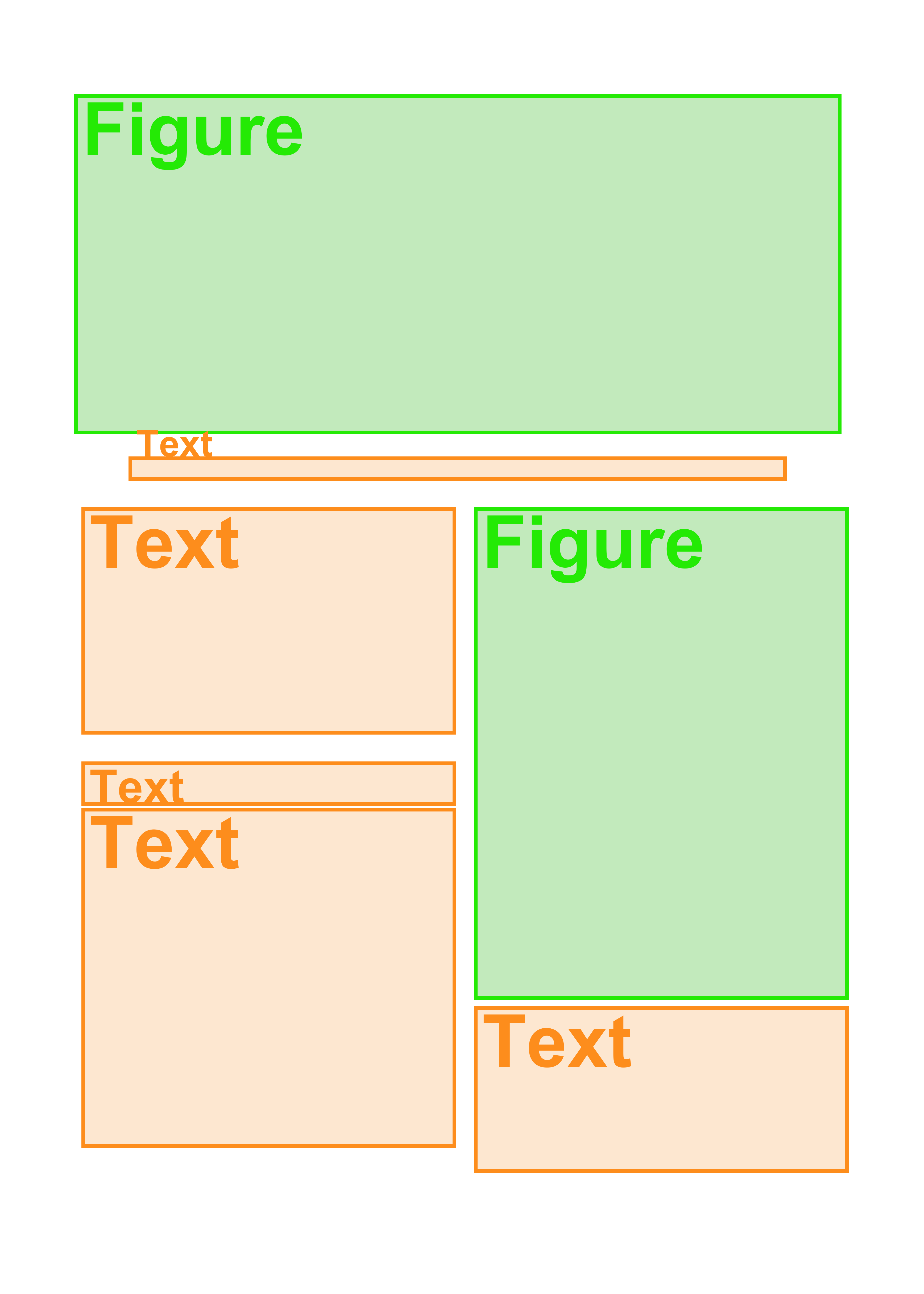} & 
\includegraphics[width=\publaynetBulkWidth]{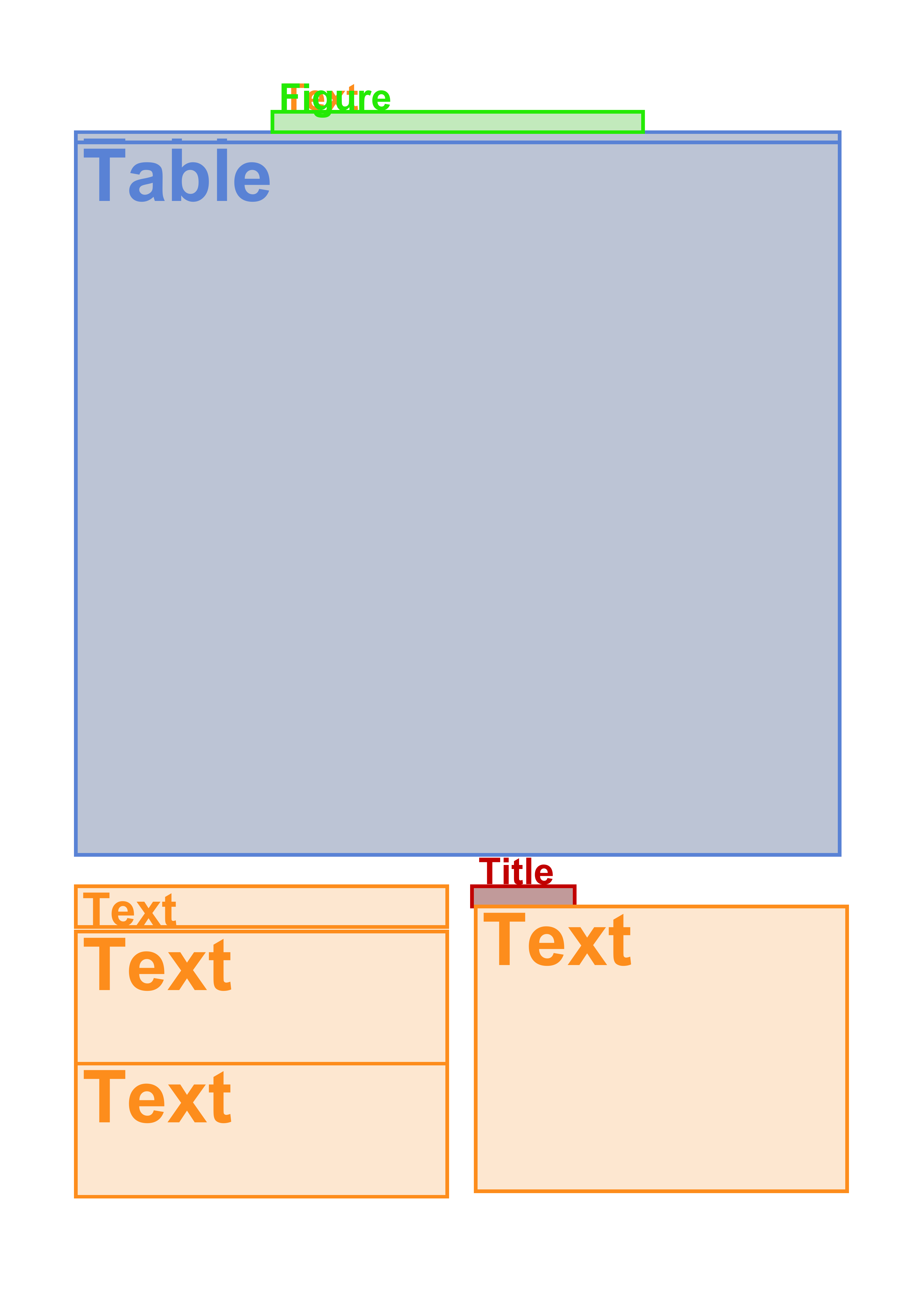} \\ 
\includegraphics[width=\publaynetBulkWidth]{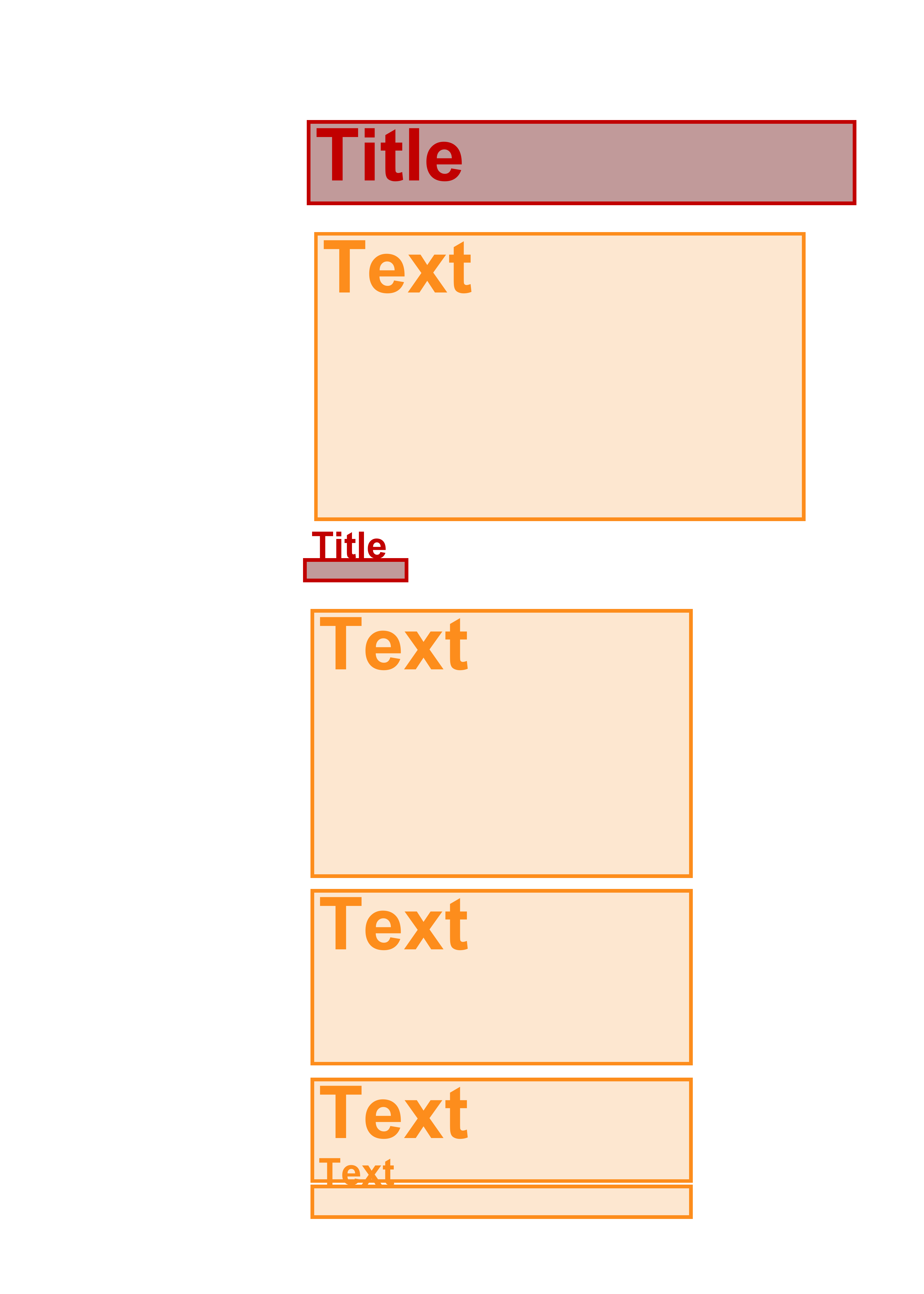} & 
\includegraphics[width=\publaynetBulkWidth]{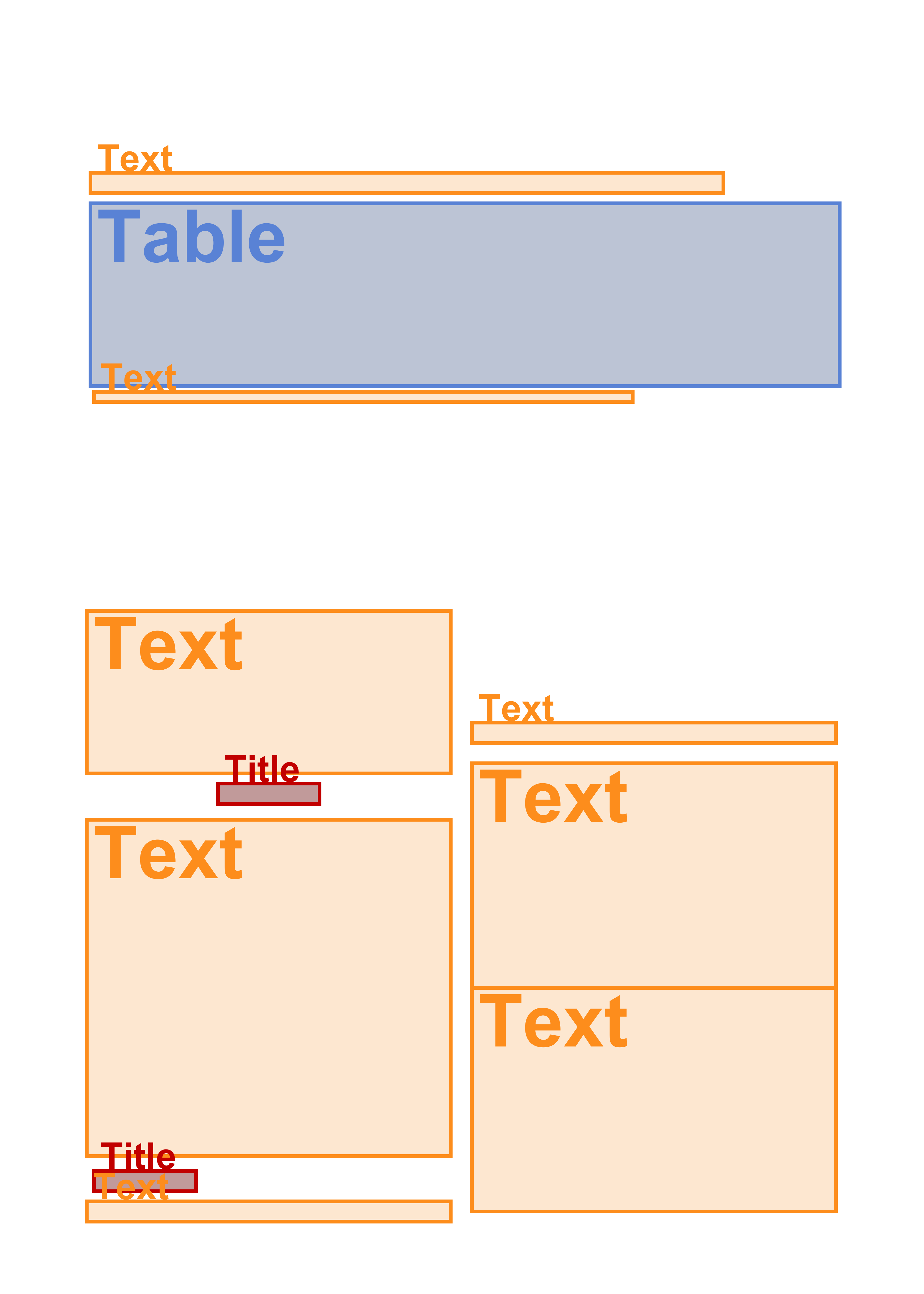} & 
\includegraphics[width=\publaynetBulkWidth]{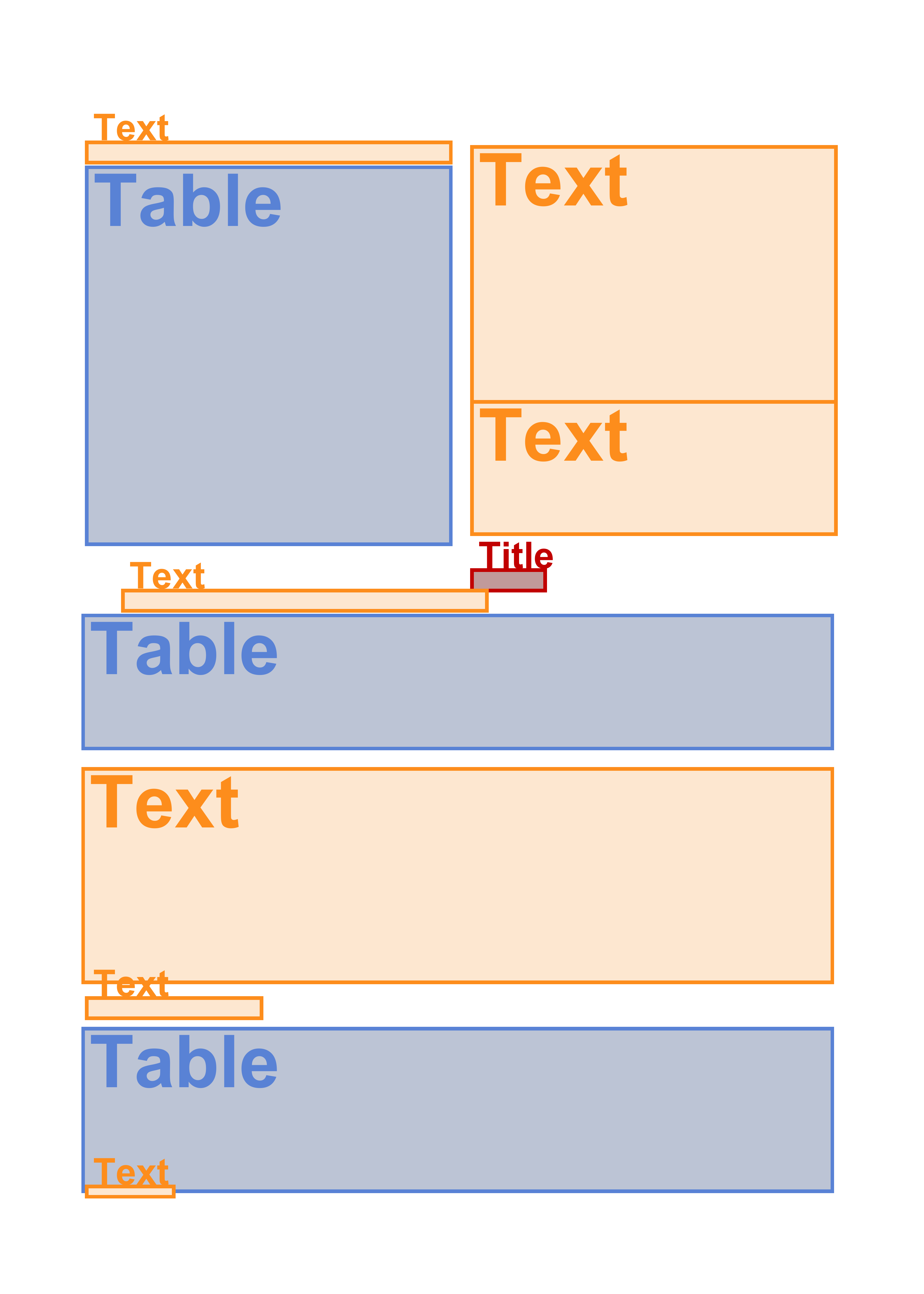} & 
\includegraphics[width=\publaynetBulkWidth]{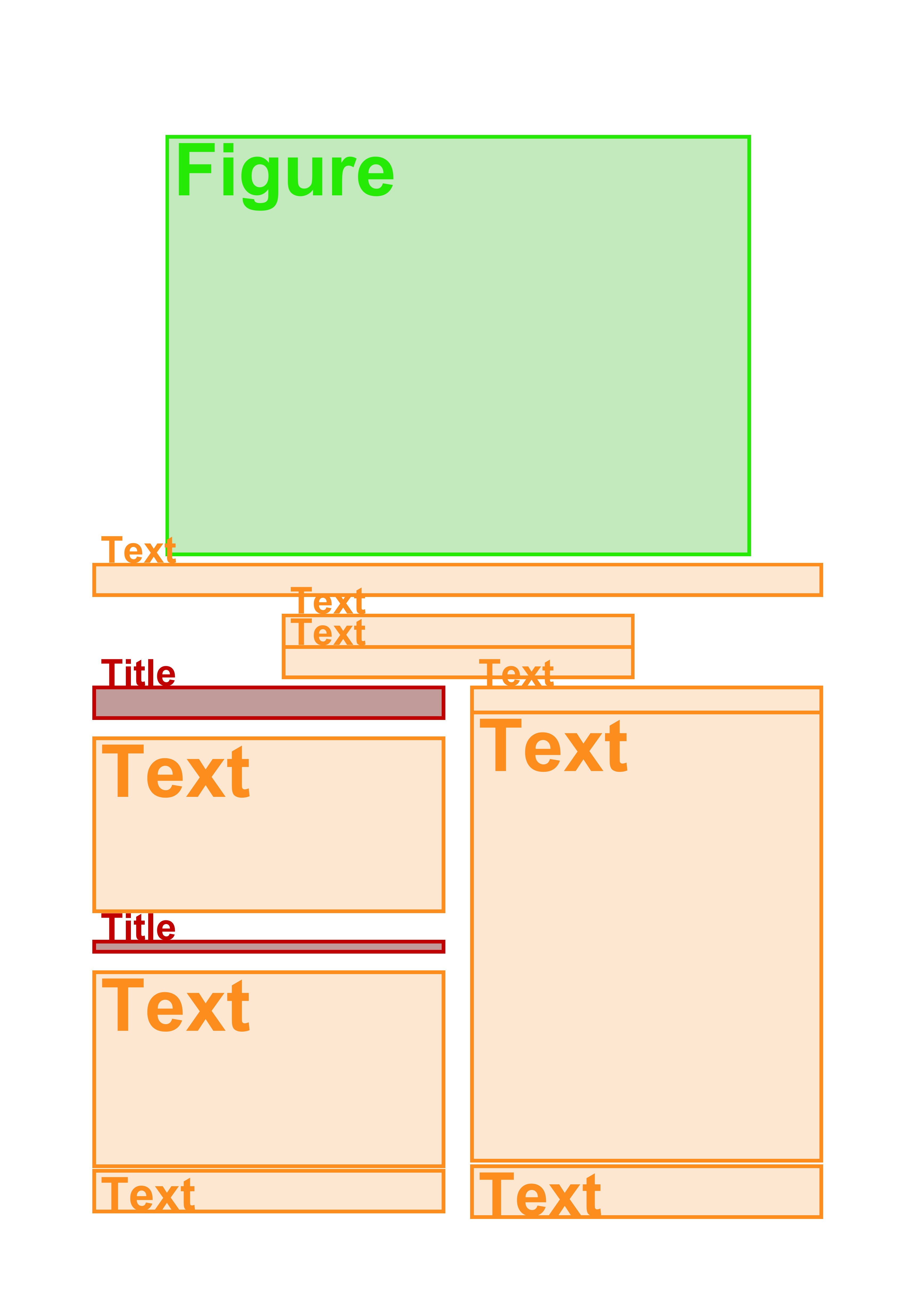} & 
\includegraphics[width=\publaynetBulkWidth]{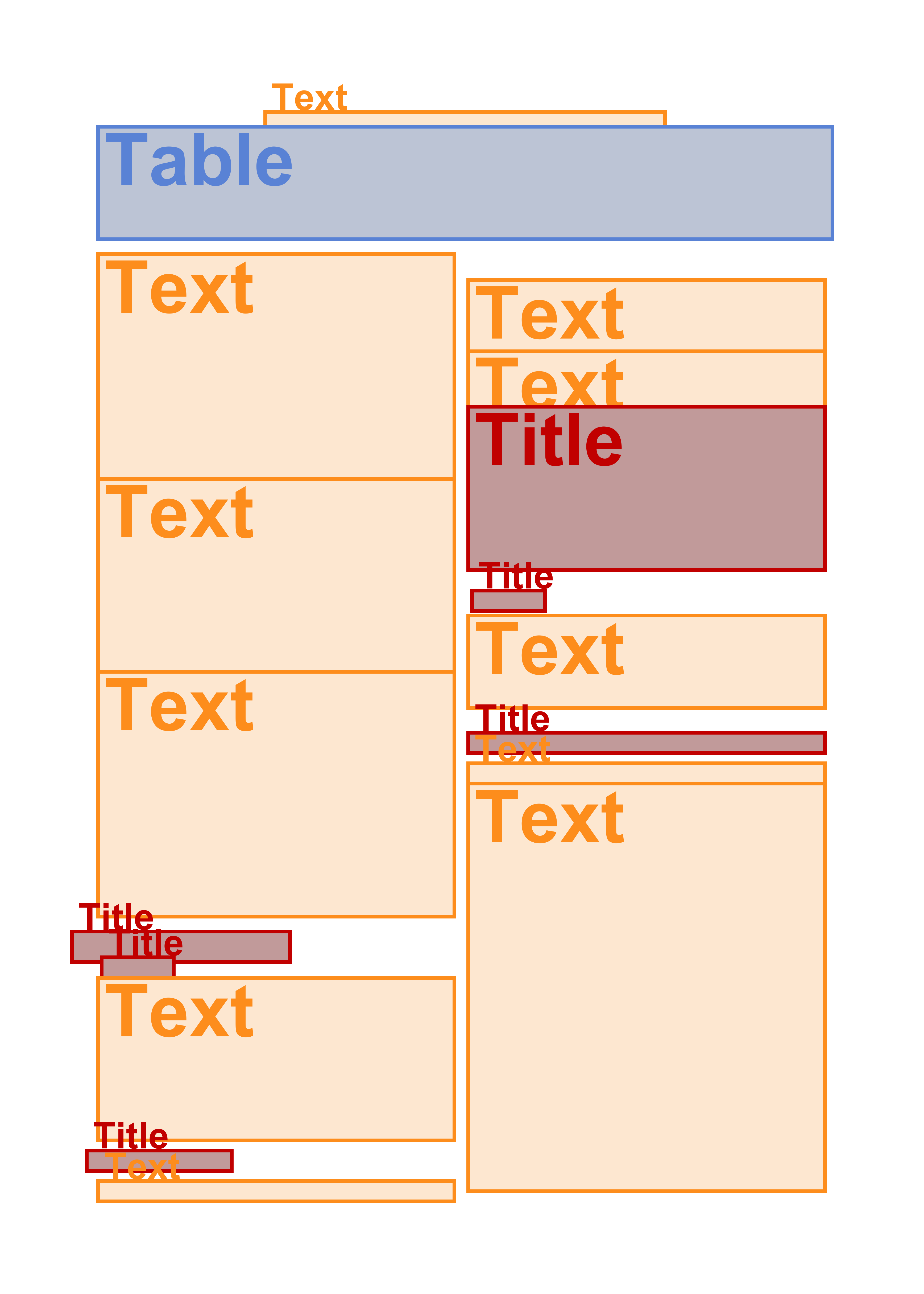} & 
\includegraphics[width=\publaynetBulkWidth]{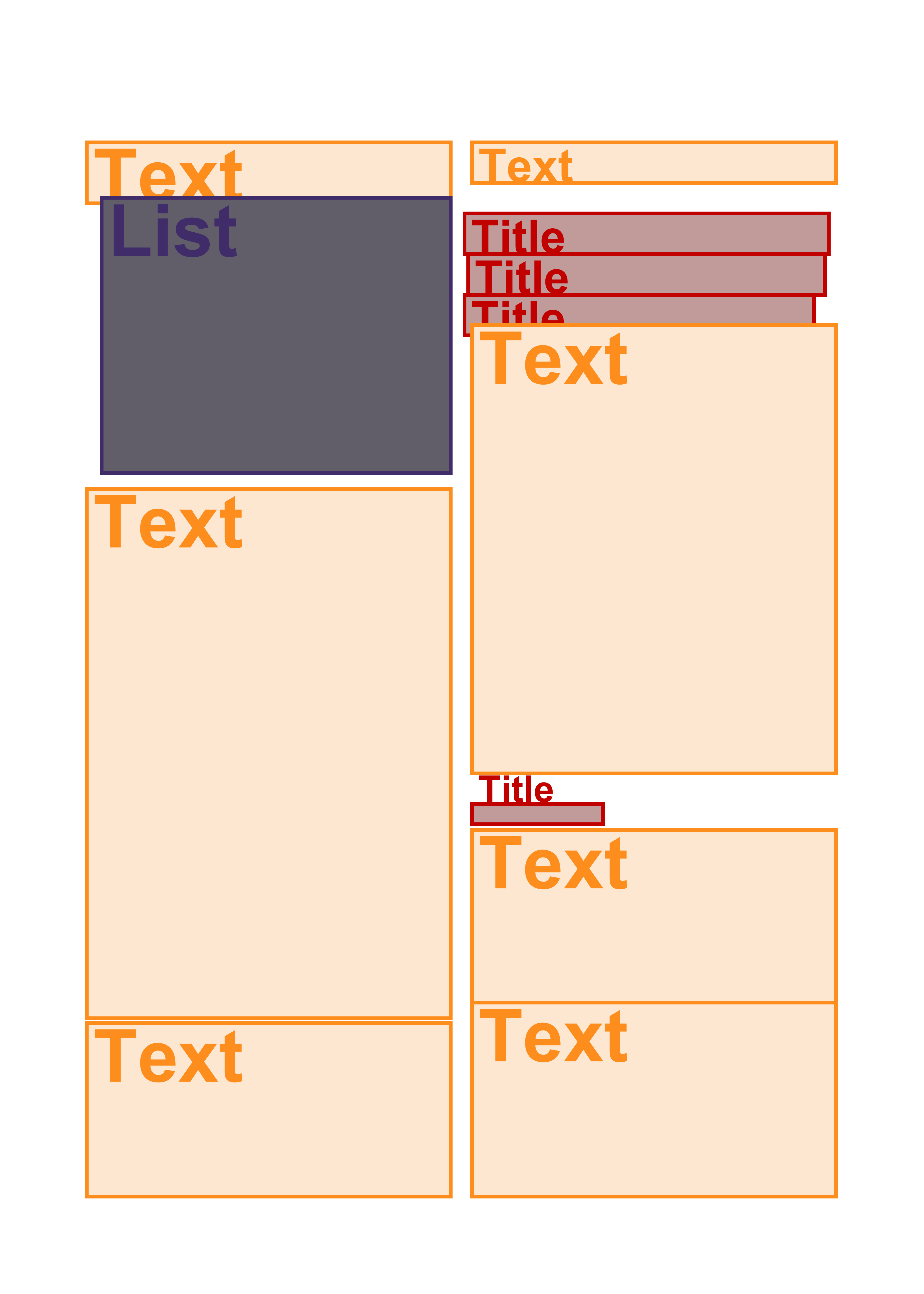} \\ 
\includegraphics[width=\publaynetBulkWidth]{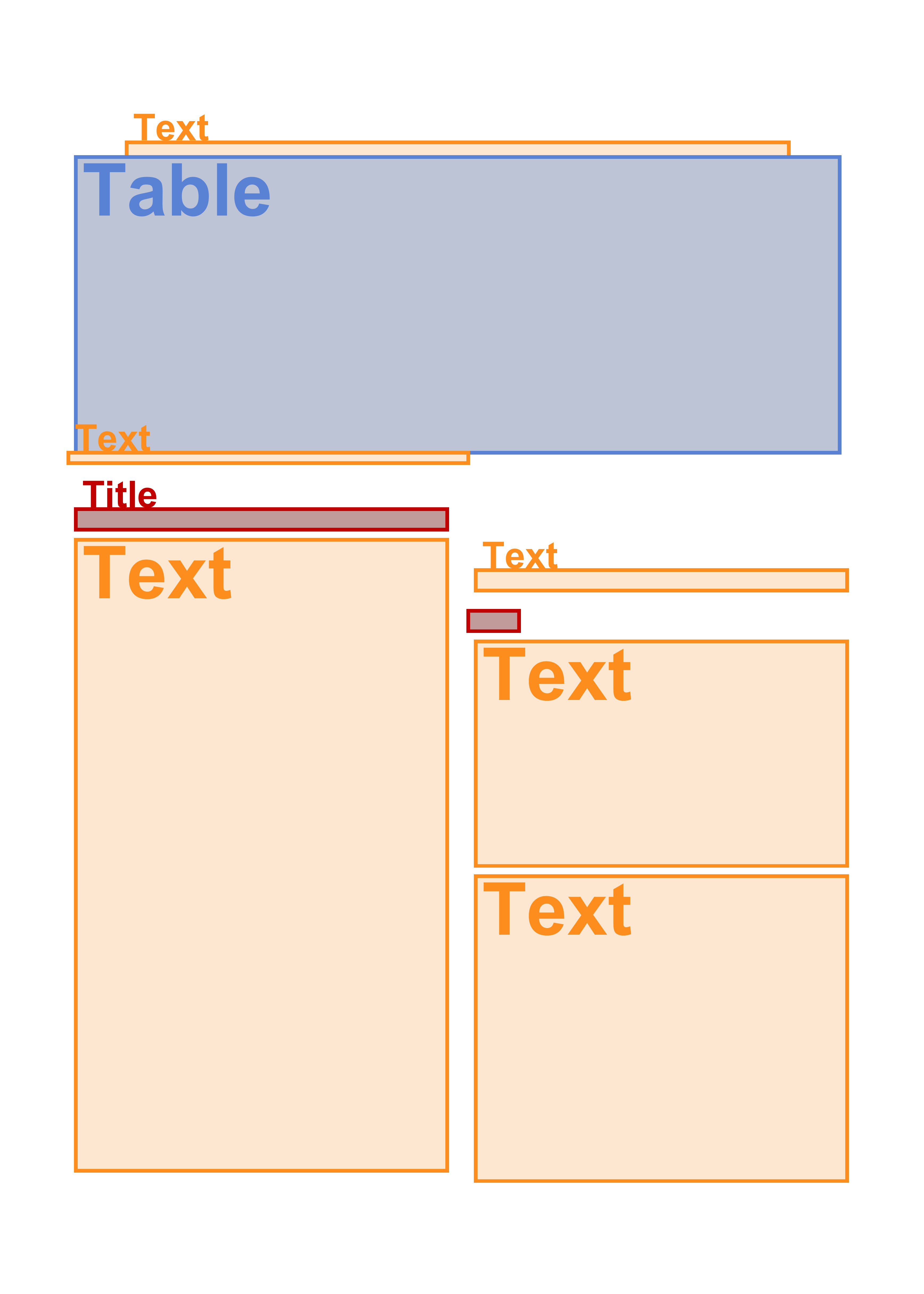} & 
\includegraphics[width=\publaynetBulkWidth]{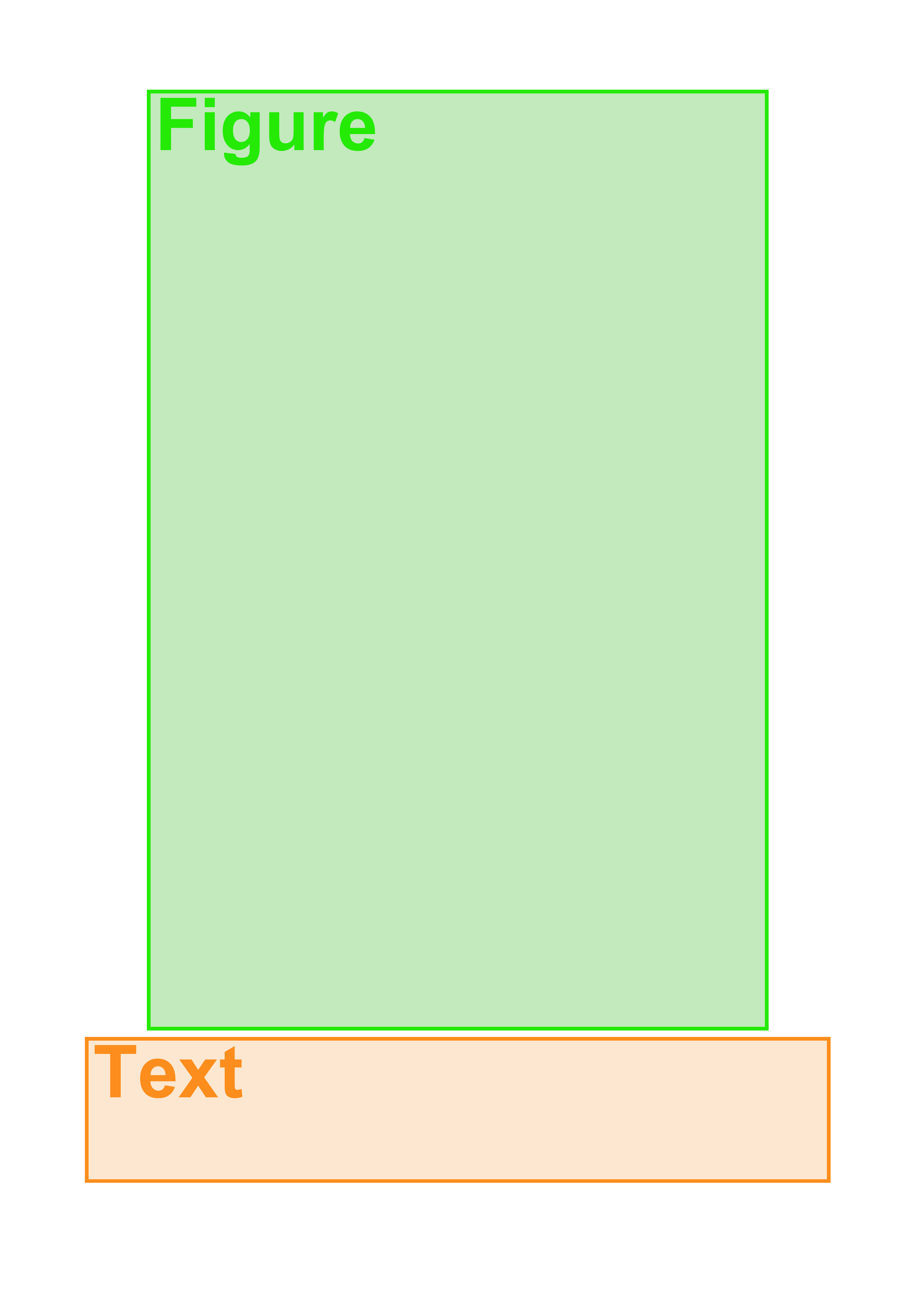} & 
\includegraphics[width=\publaynetBulkWidth]{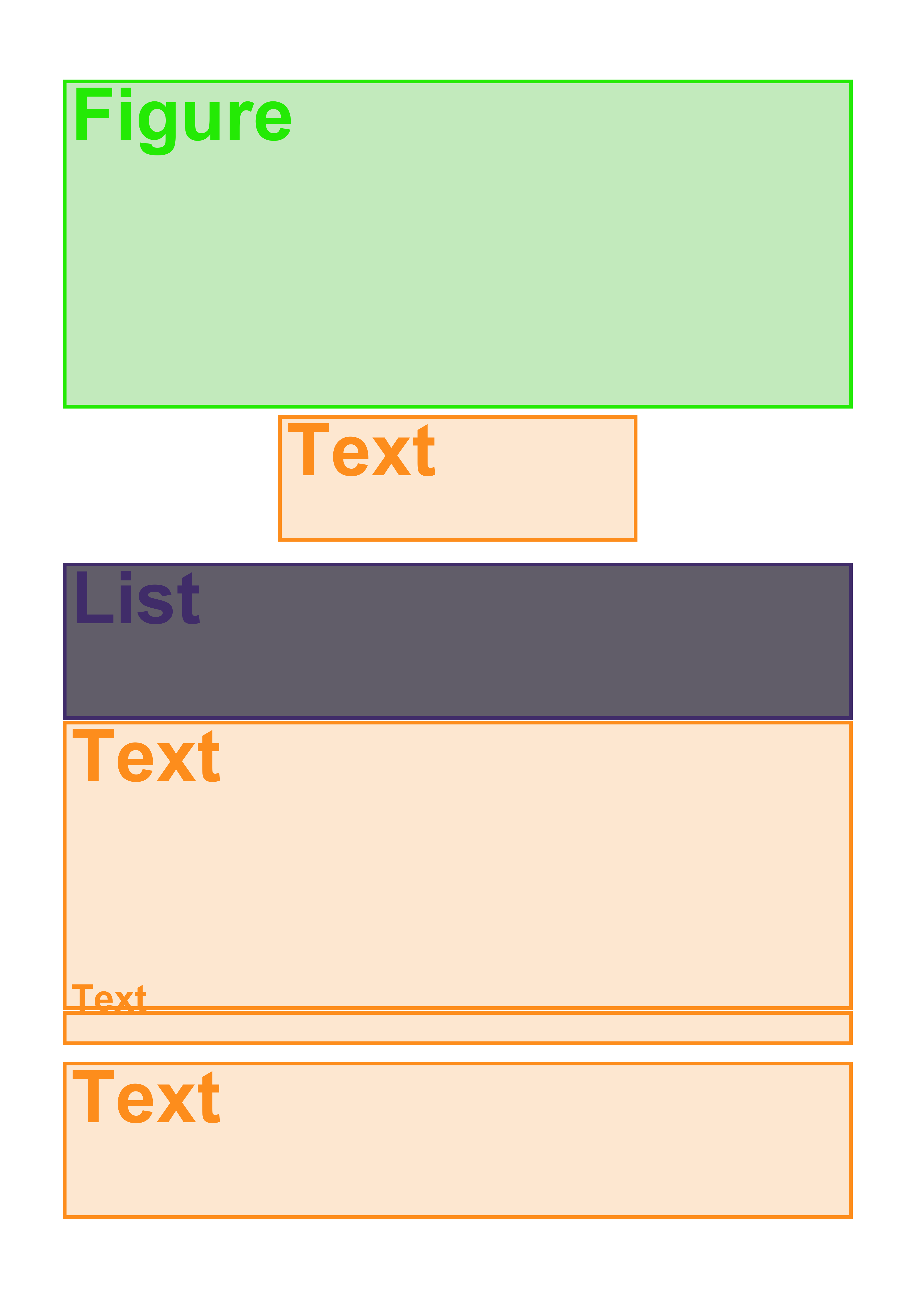} & 
\includegraphics[width=\publaynetBulkWidth]{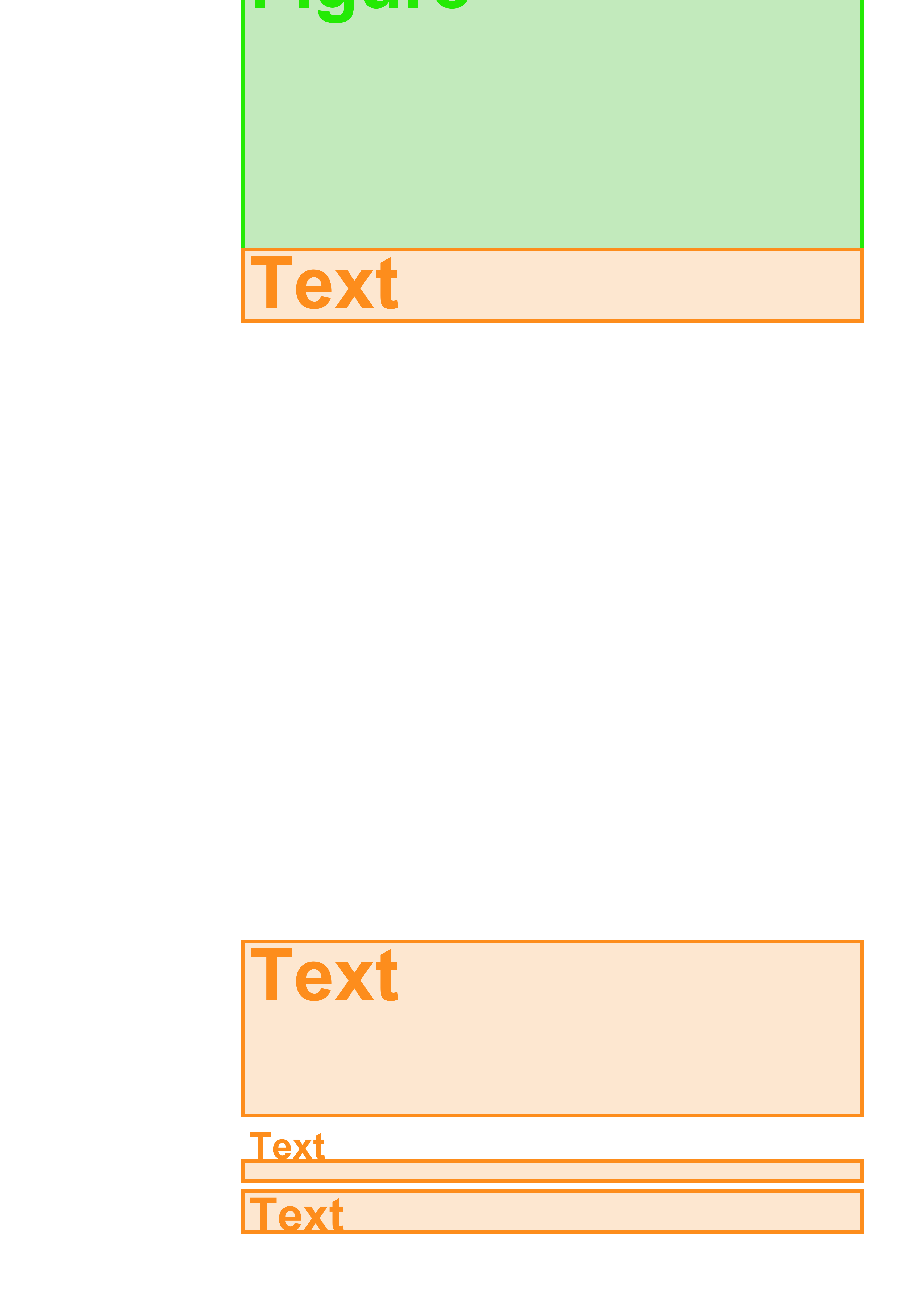} & 
\includegraphics[width=\publaynetBulkWidth]{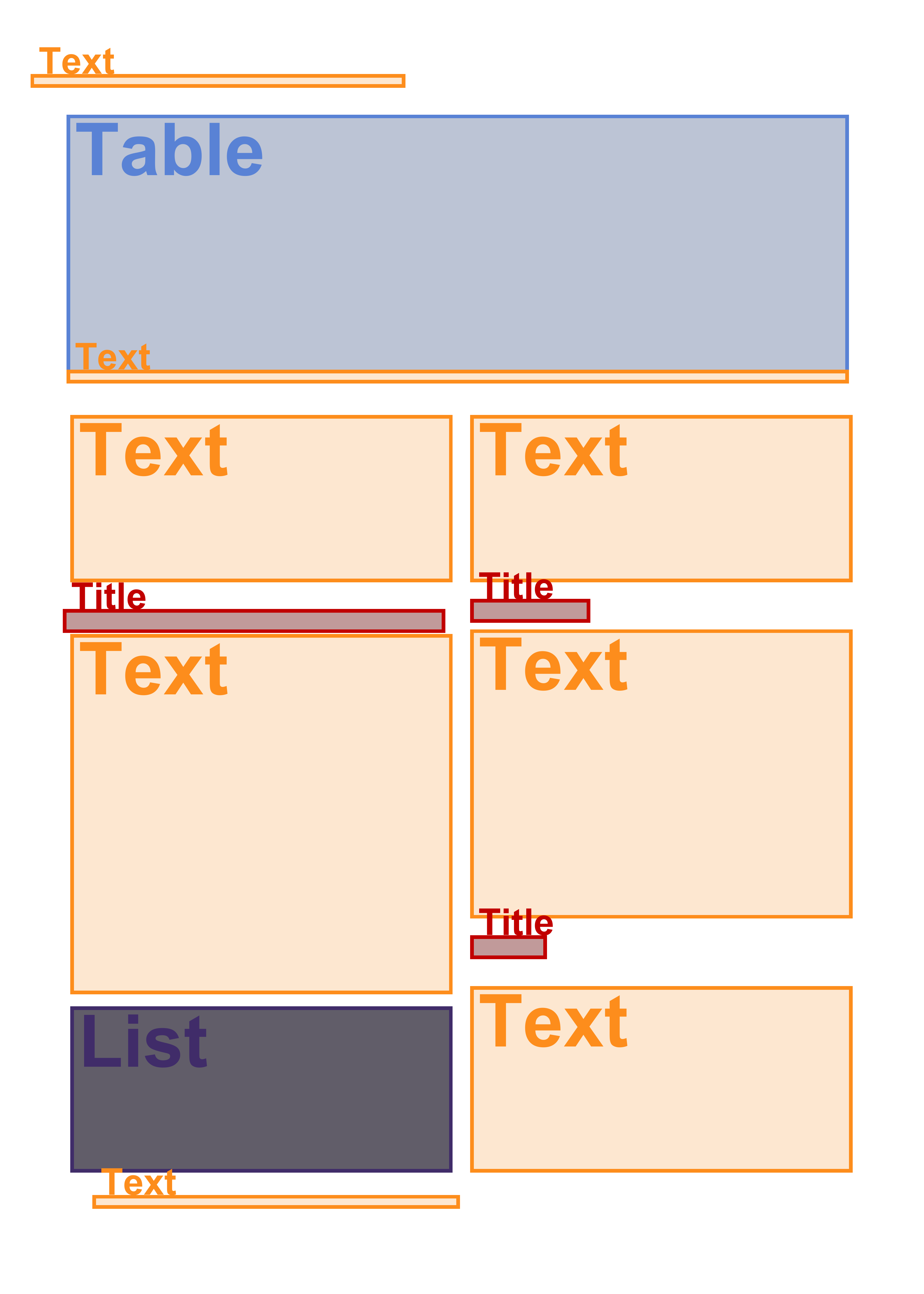} & 
\includegraphics[width=\publaynetBulkWidth]{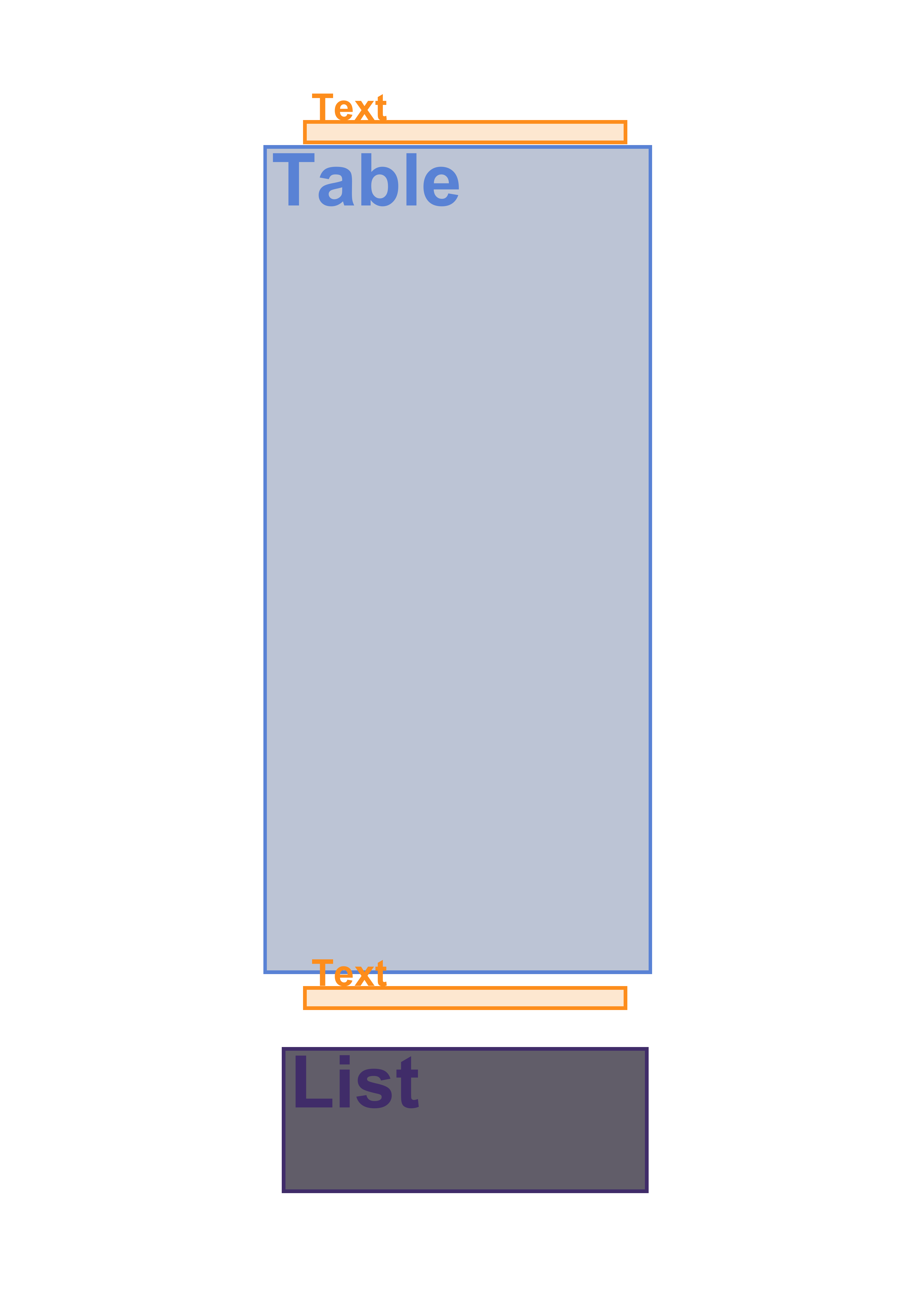} \\
    \end{tabular}
    \caption{(Cont.) Synthesized layouts on PubLayNet using an autoregressive decoder.}
    \label{fig:bulk_publaynet_cont}
\end{figure}
\begin{figure}\ContinuedFloat
    \setlength{\publaynetBulkWidth}{0.15\linewidth}
    \setlength{\tabcolsep}{2pt}
    \centering
    \begin{tabular}{cccccc}
\includegraphics[width=\publaynetBulkWidth]{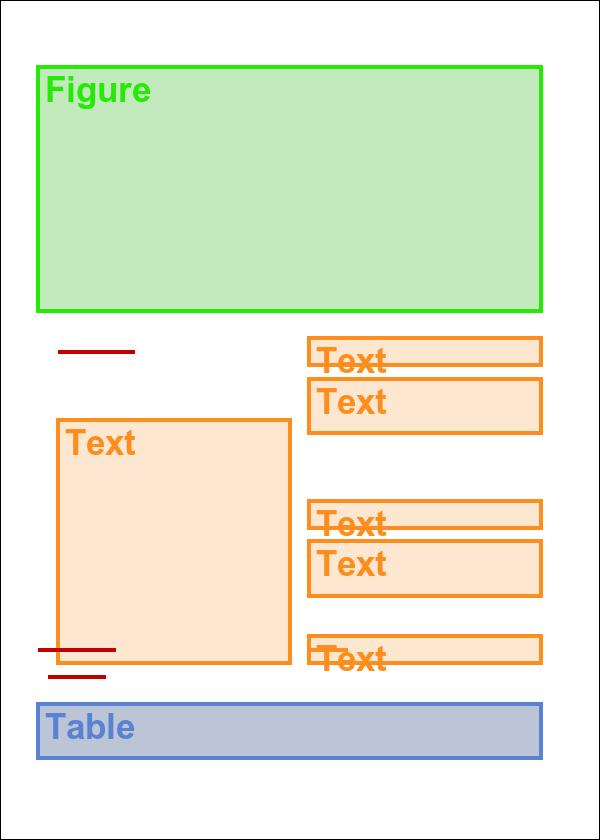} & 
\includegraphics[width=\publaynetBulkWidth]{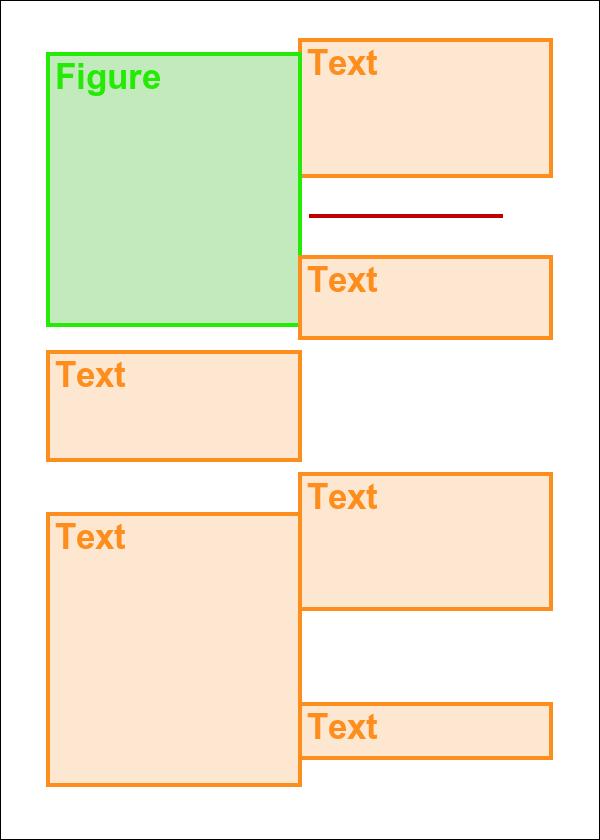} & 
\includegraphics[width=\publaynetBulkWidth]{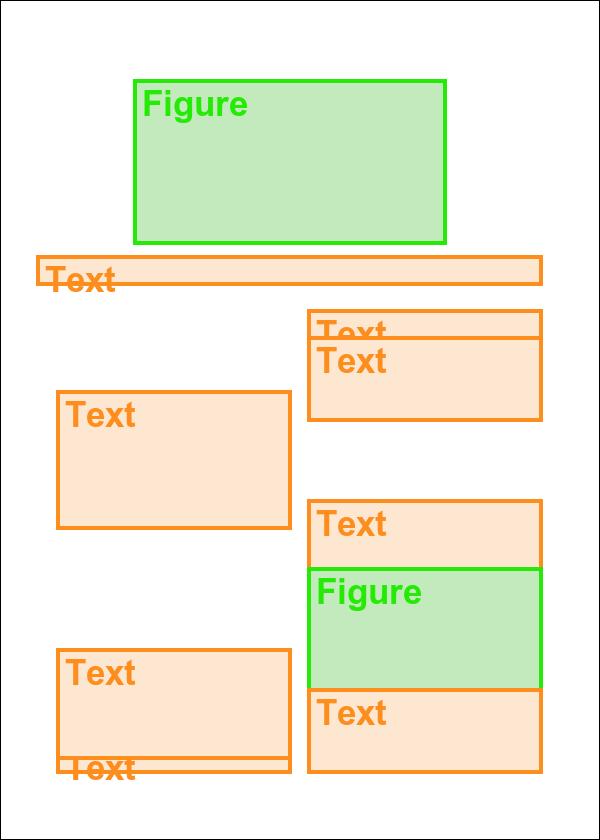} & 
\includegraphics[width=\publaynetBulkWidth]{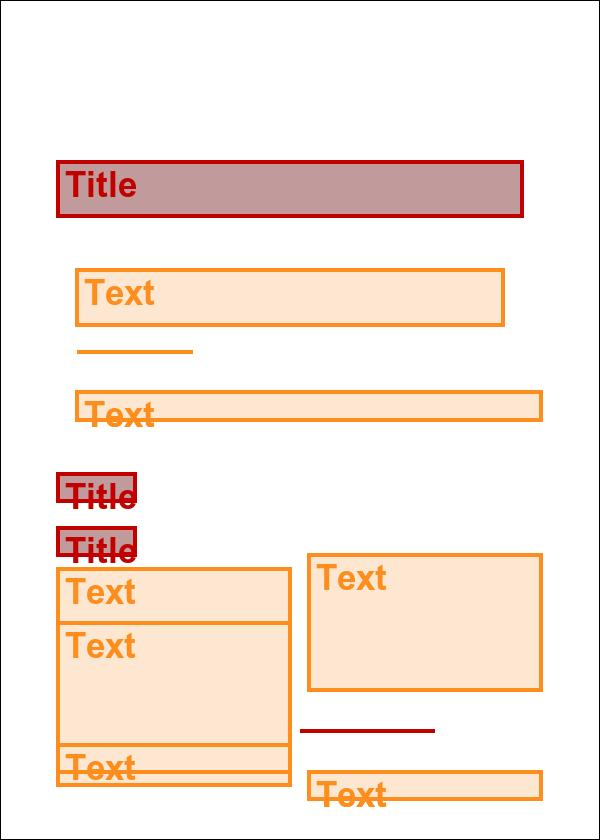} & 
\includegraphics[width=\publaynetBulkWidth]{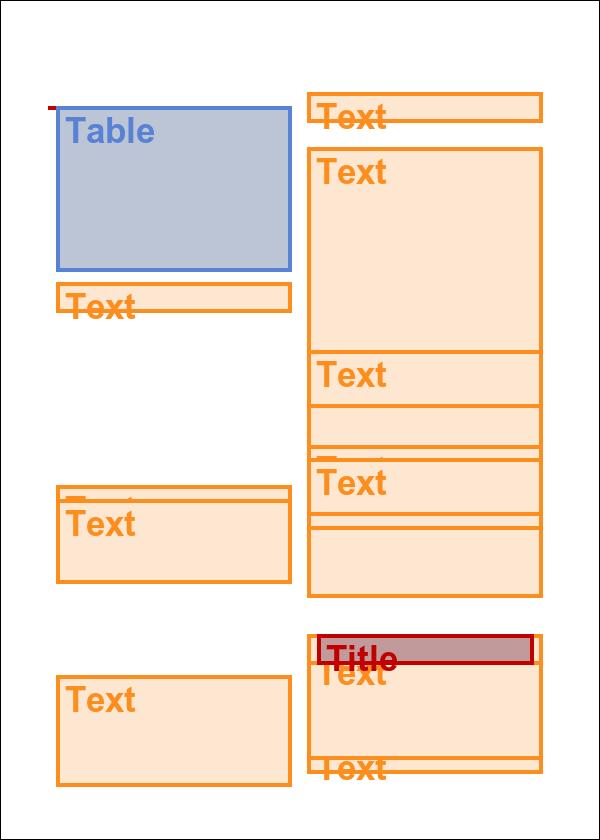} & 
\includegraphics[width=\publaynetBulkWidth]{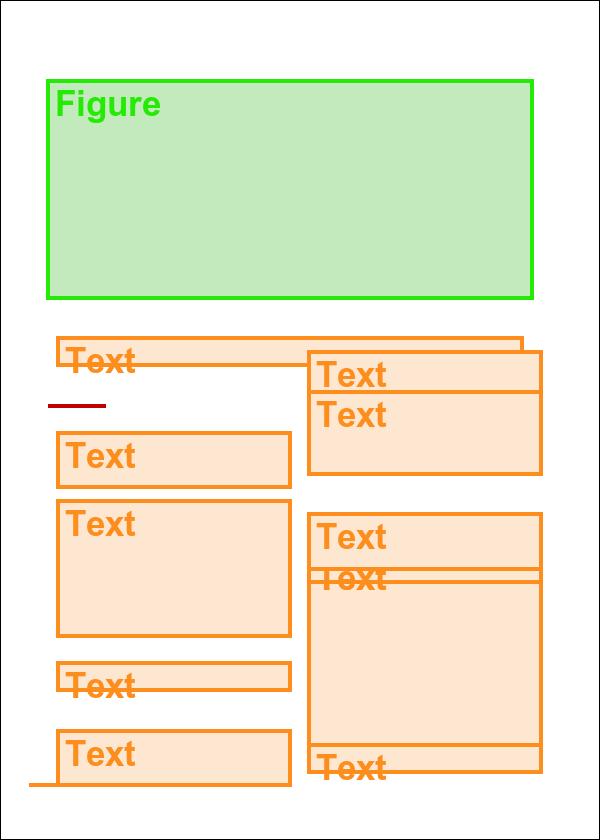} \\ 
\includegraphics[width=\publaynetBulkWidth]{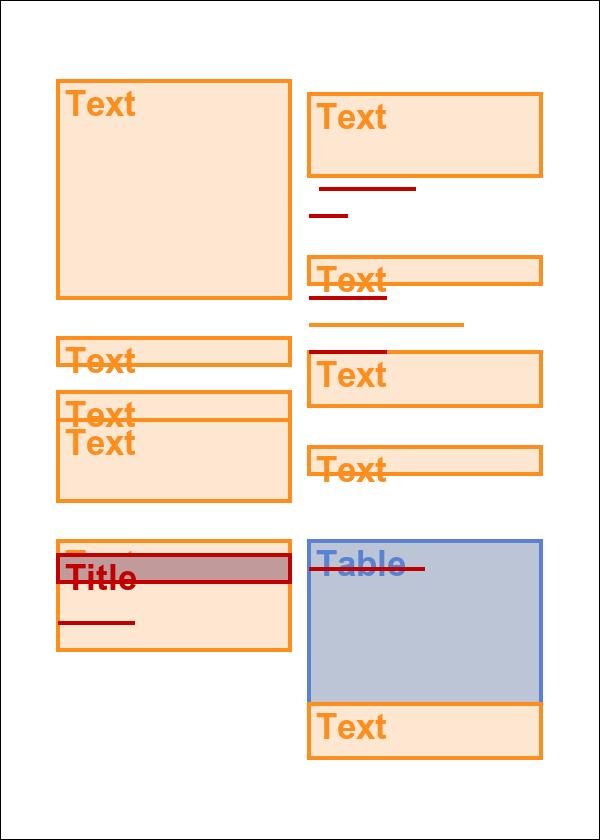} & 
\includegraphics[width=\publaynetBulkWidth]{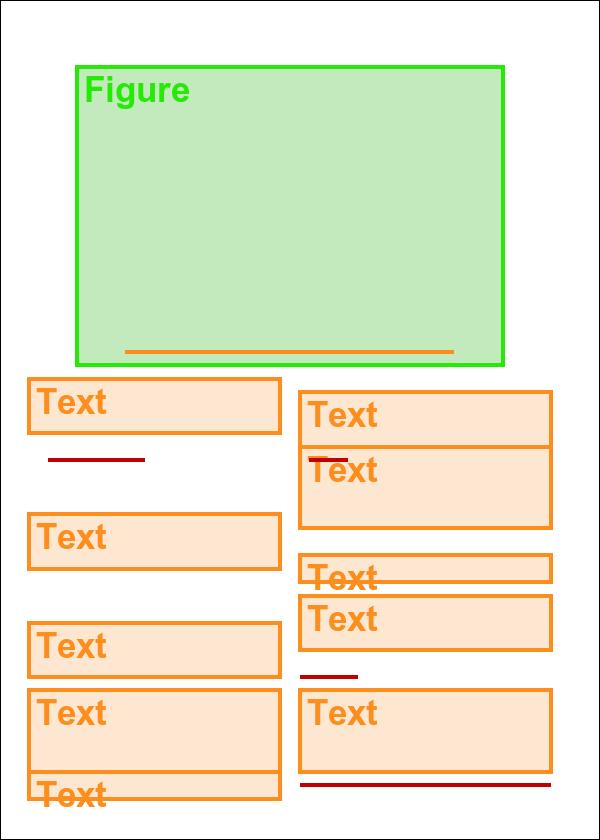} & 
\includegraphics[width=\publaynetBulkWidth]{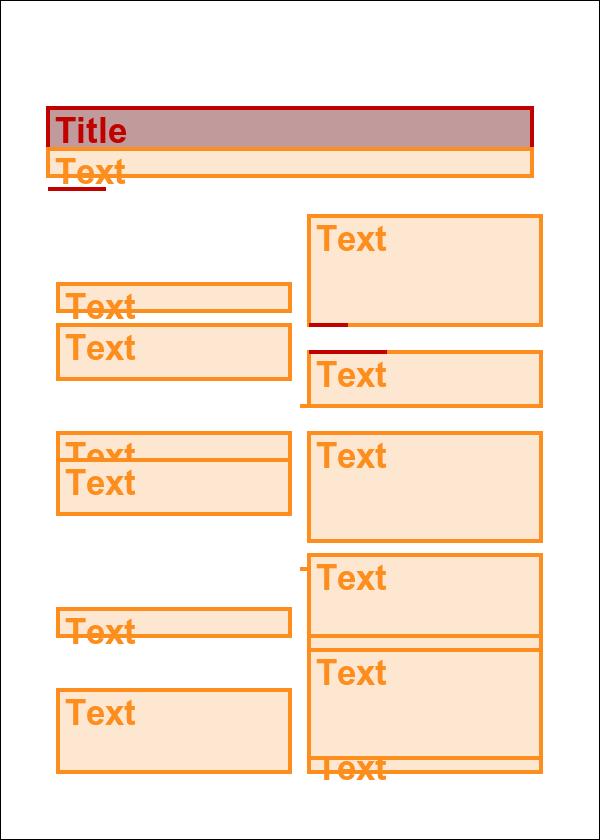} & 
\includegraphics[width=\publaynetBulkWidth]{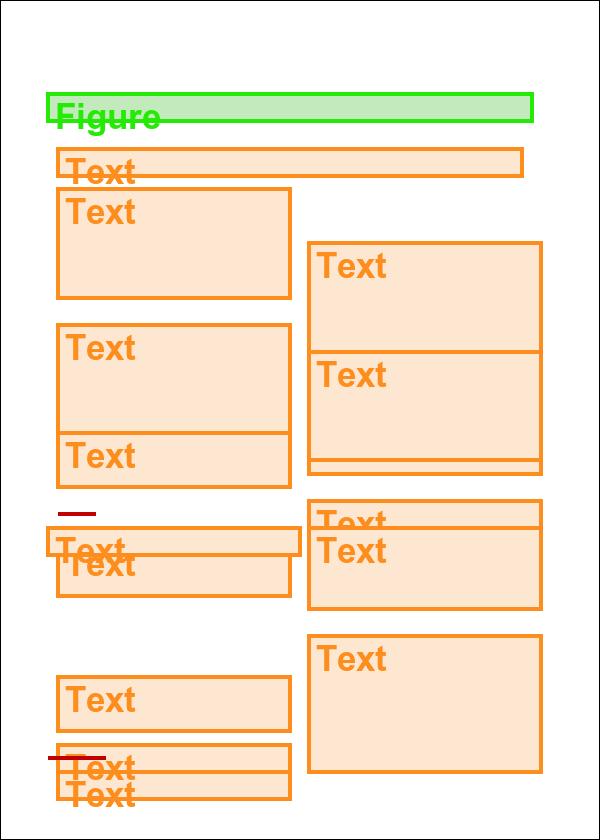} & 
\includegraphics[width=\publaynetBulkWidth]{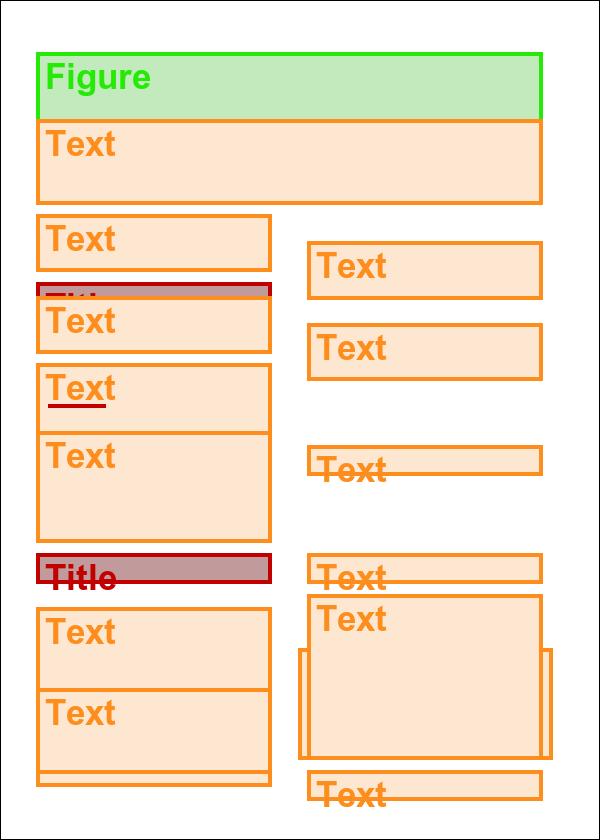} & 
\includegraphics[width=\publaynetBulkWidth]{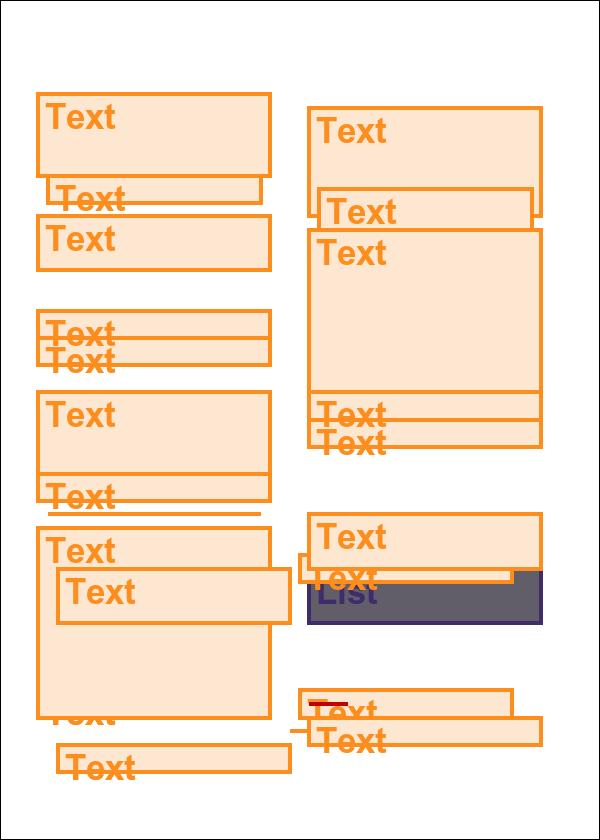} \\ 
\includegraphics[width=\publaynetBulkWidth]{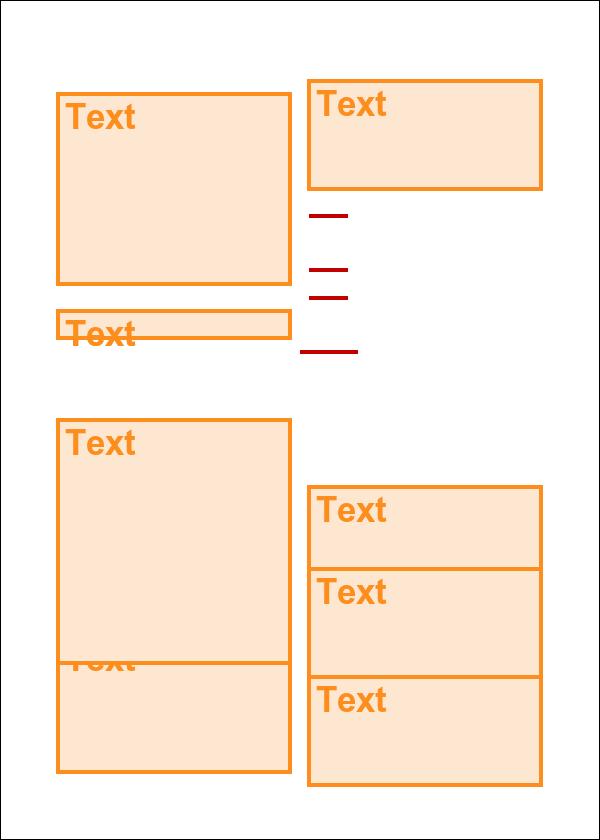} & 
\includegraphics[width=\publaynetBulkWidth]{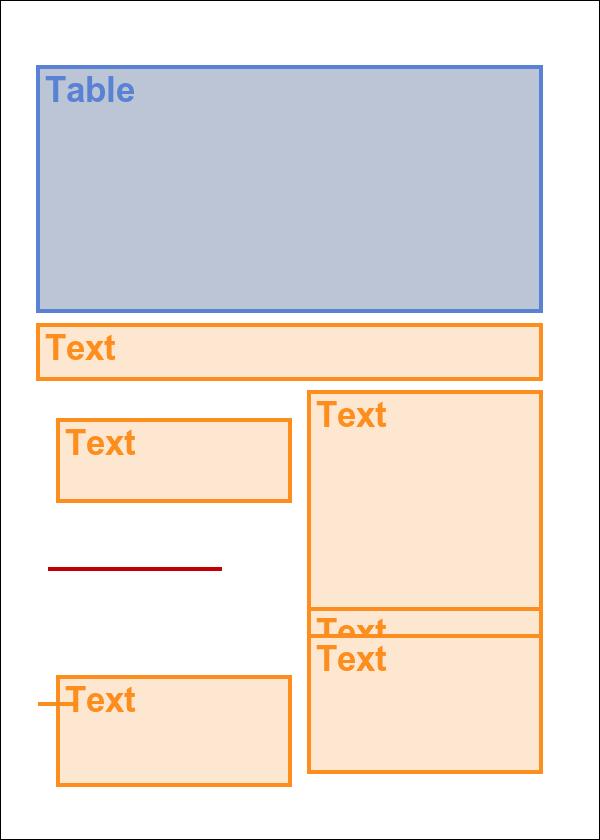} & 
\includegraphics[width=\publaynetBulkWidth]{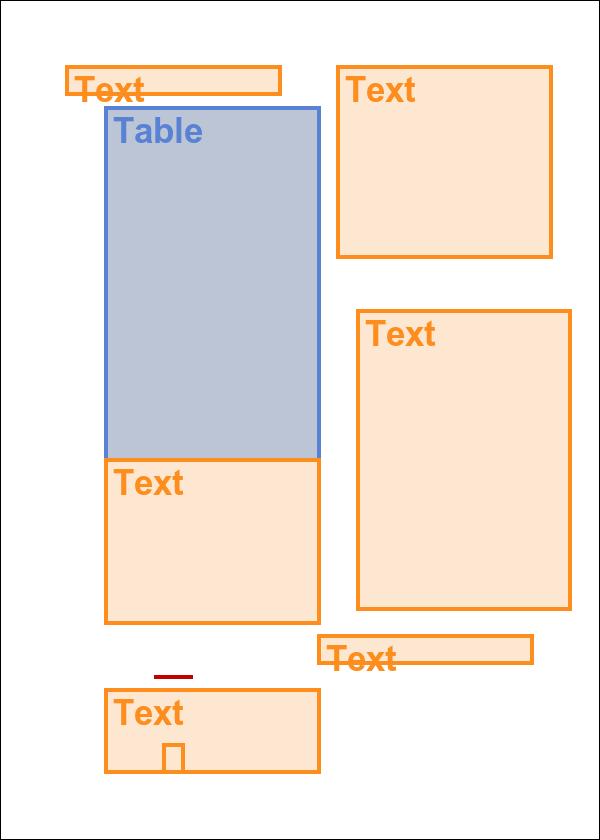} & 
\includegraphics[width=\publaynetBulkWidth]{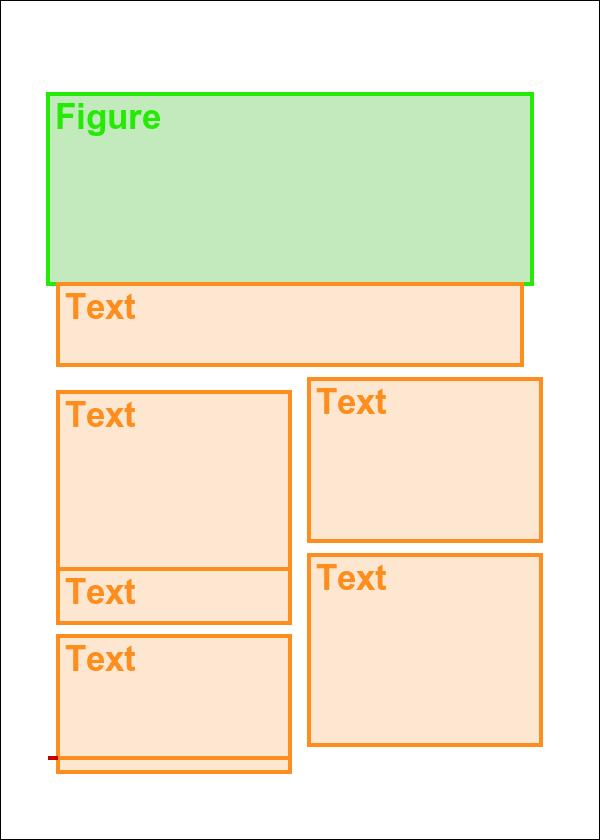} & 
\includegraphics[width=\publaynetBulkWidth]{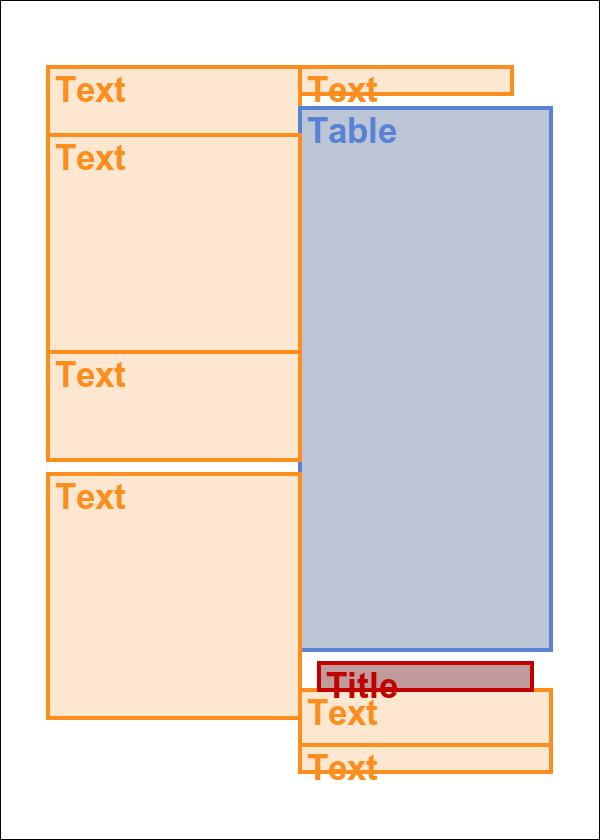} & 
\includegraphics[width=\publaynetBulkWidth]{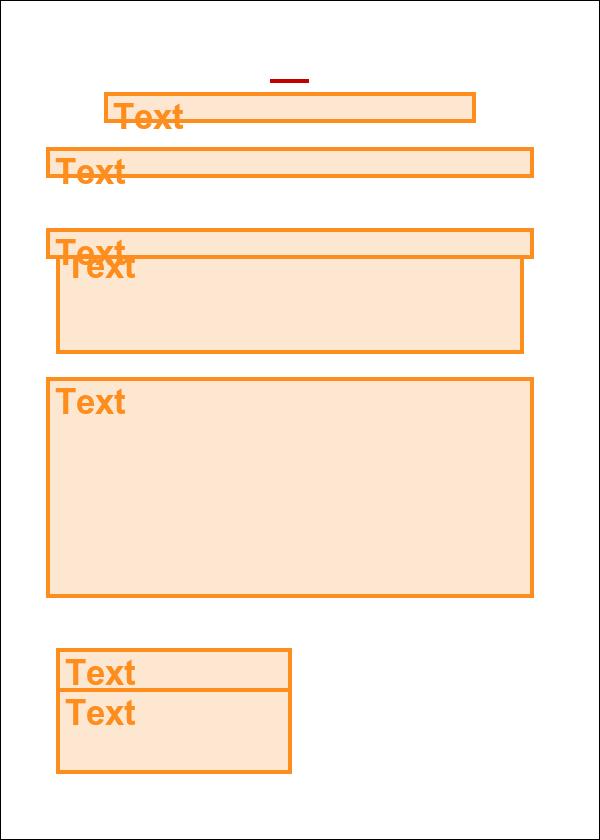} \\ 
\includegraphics[width=\publaynetBulkWidth]{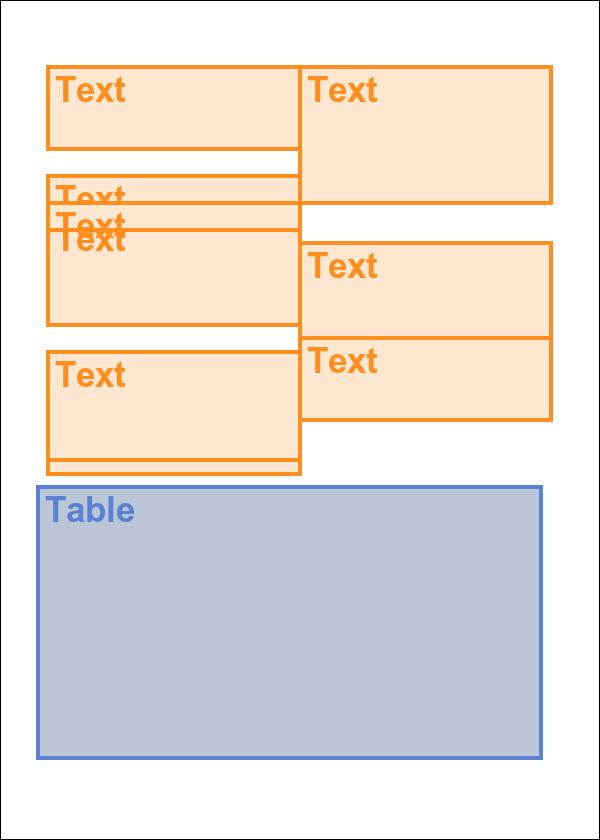} & 
\includegraphics[width=\publaynetBulkWidth]{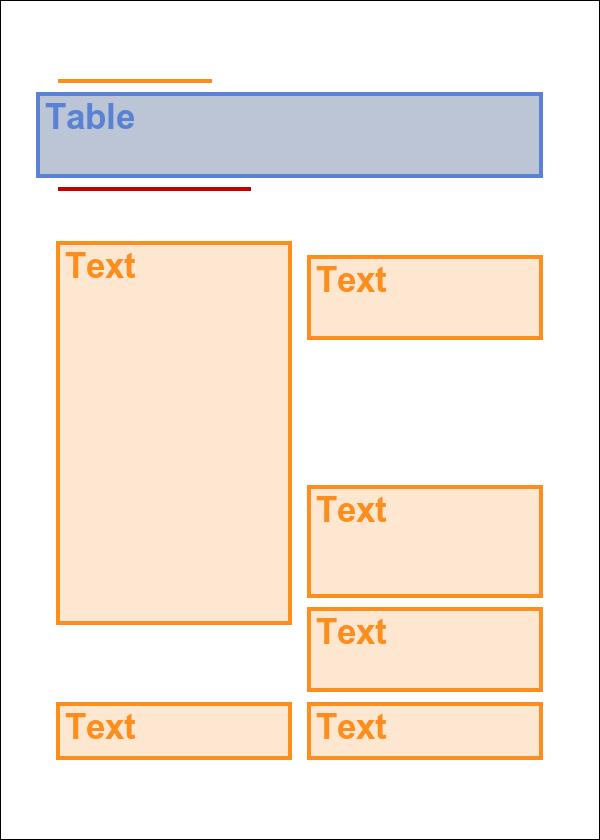} & 
\includegraphics[width=\publaynetBulkWidth]{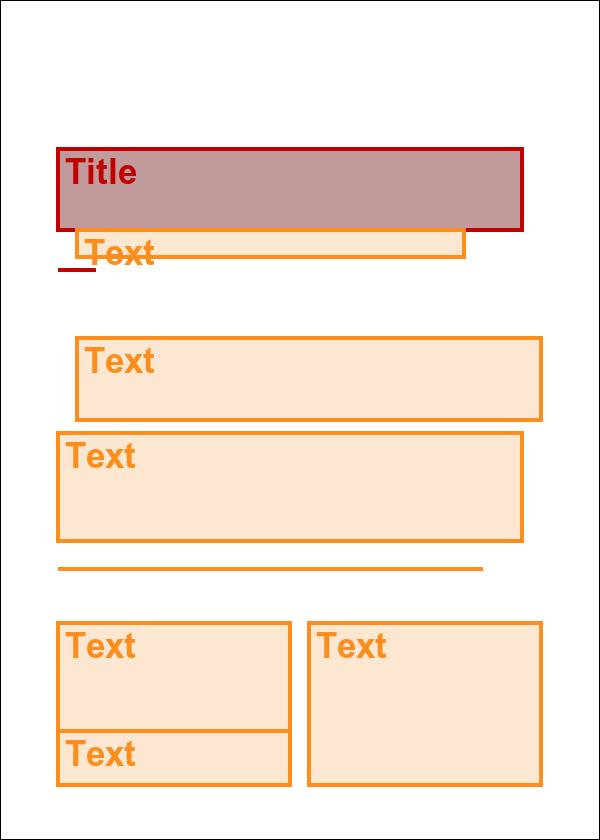} & 
\includegraphics[width=\publaynetBulkWidth]{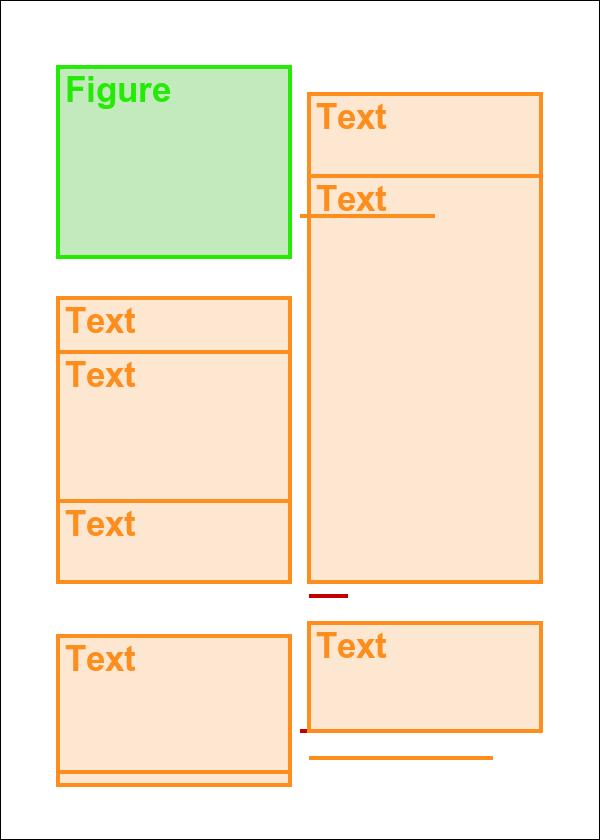} & 
\includegraphics[width=\publaynetBulkWidth]{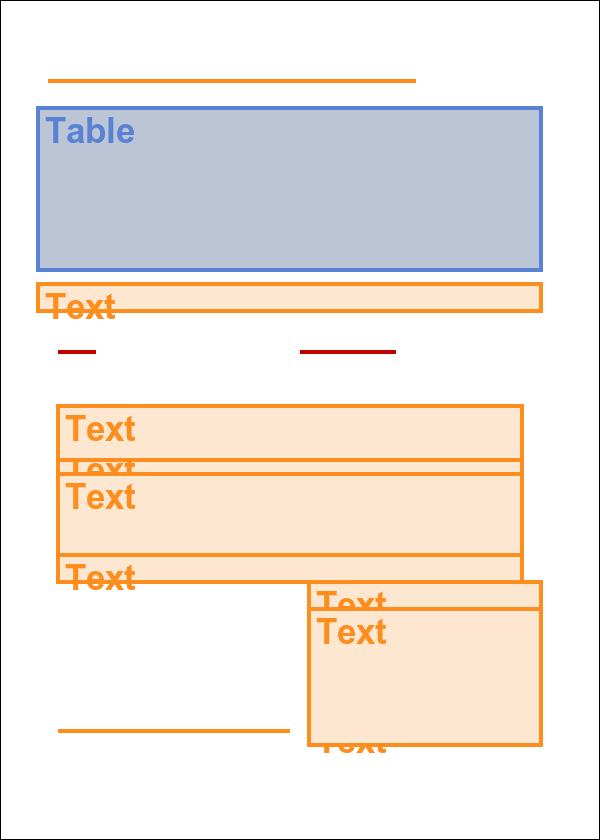} & 
\includegraphics[width=\publaynetBulkWidth]{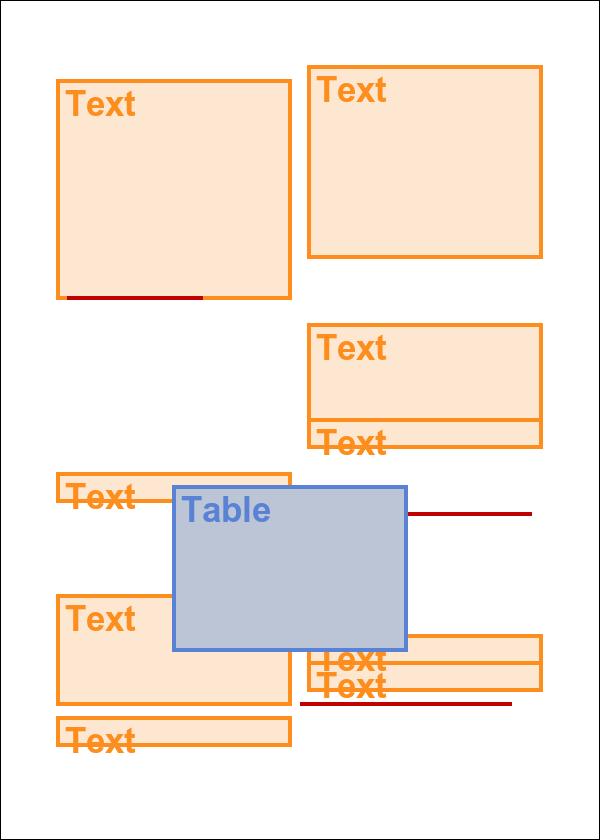} \\ 
\includegraphics[width=\publaynetBulkWidth]{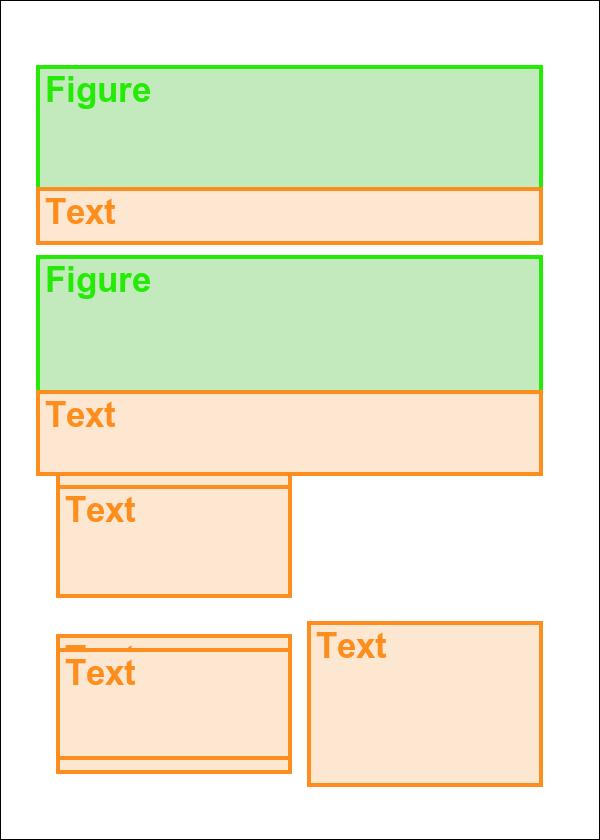} & 
\includegraphics[width=\publaynetBulkWidth]{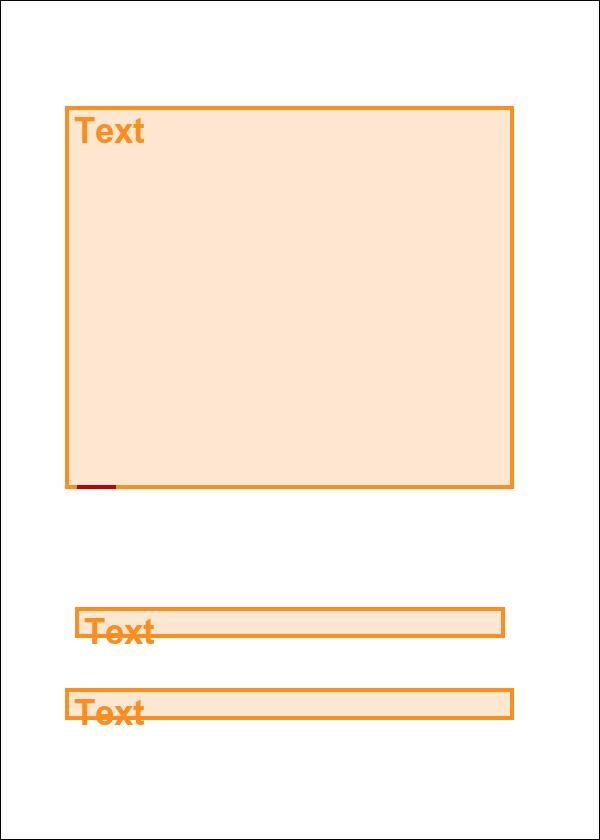} & 
\includegraphics[width=\publaynetBulkWidth]{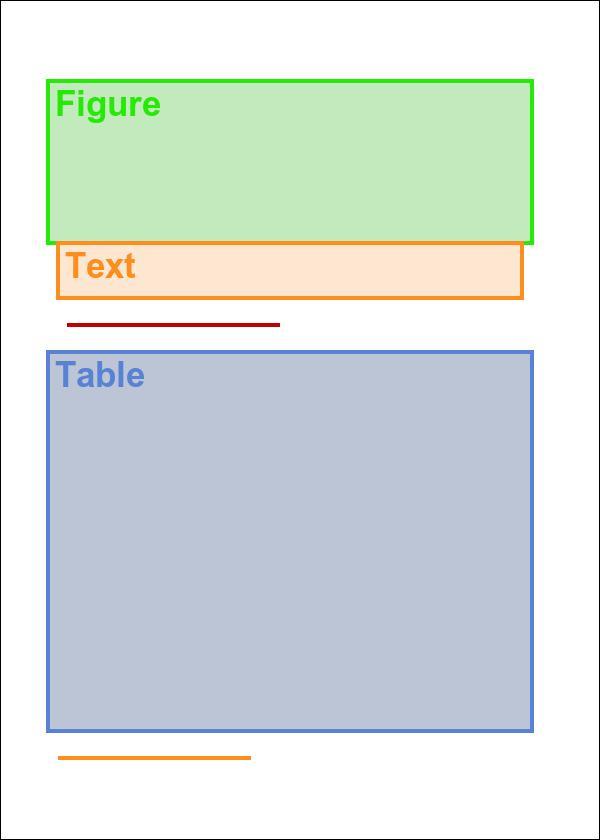} & 
\includegraphics[width=\publaynetBulkWidth]{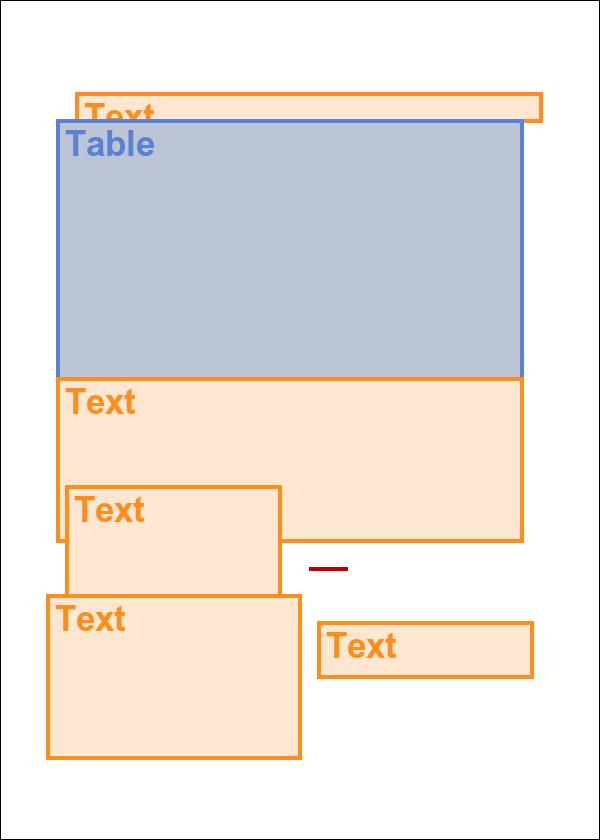} & 
\includegraphics[width=\publaynetBulkWidth]{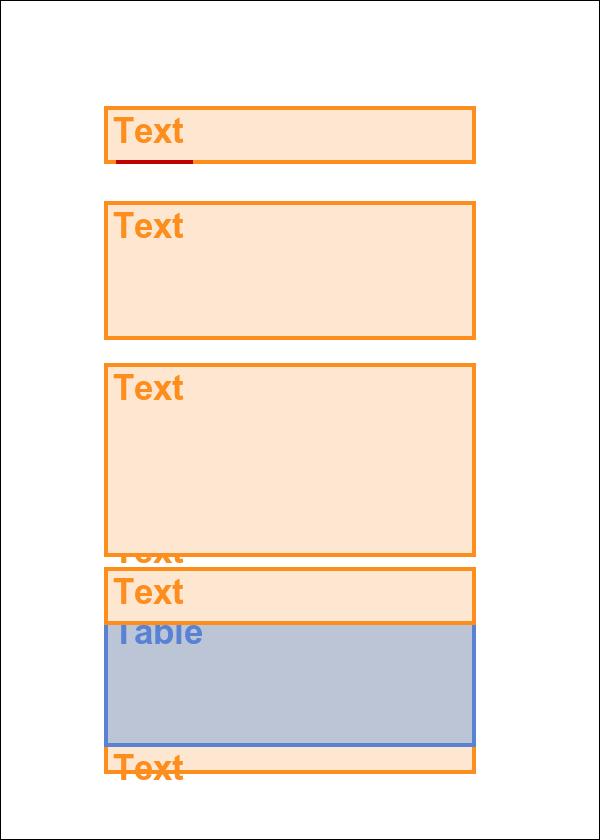} & 
\includegraphics[width=\publaynetBulkWidth]{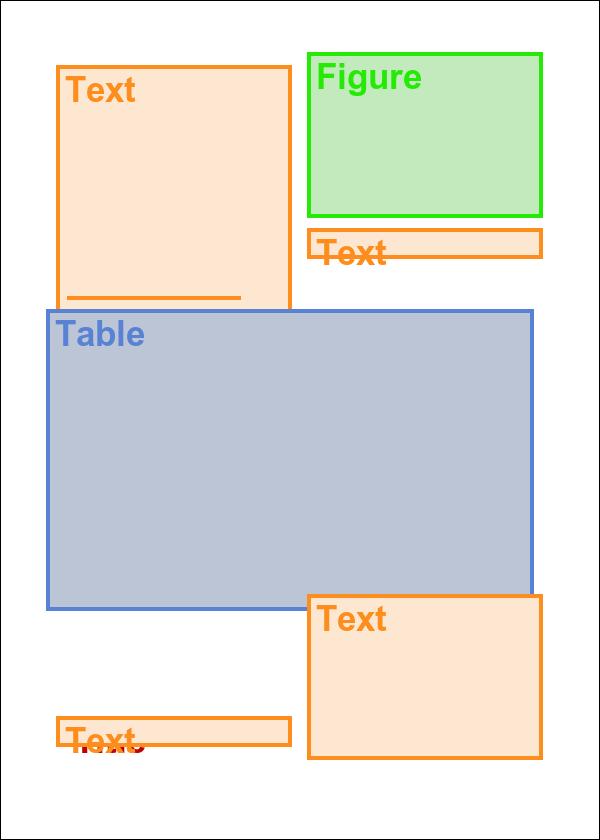} \\
    \end{tabular}
    \caption{Additional synthesized layouts on PubLayNet using a non-autoregressive decoder.}
    \label{fig:bulk_publaynet_non_autoregressive}
\end{figure}

\clearpage
\subsection{RICO}

\begin{figure}[h]
\setlength{\tabcolsep}{3pt}
\newlength{\ricoBulkWidth}
\setlength{\ricoBulkWidth}{0.12\linewidth}
    \centering
    \begin{tabular}{cccccccc}

\rotatebox{90}{\hspace{3mm}LayoutVAE [16]}&
\includegraphics[width=\ricoBulkWidth]{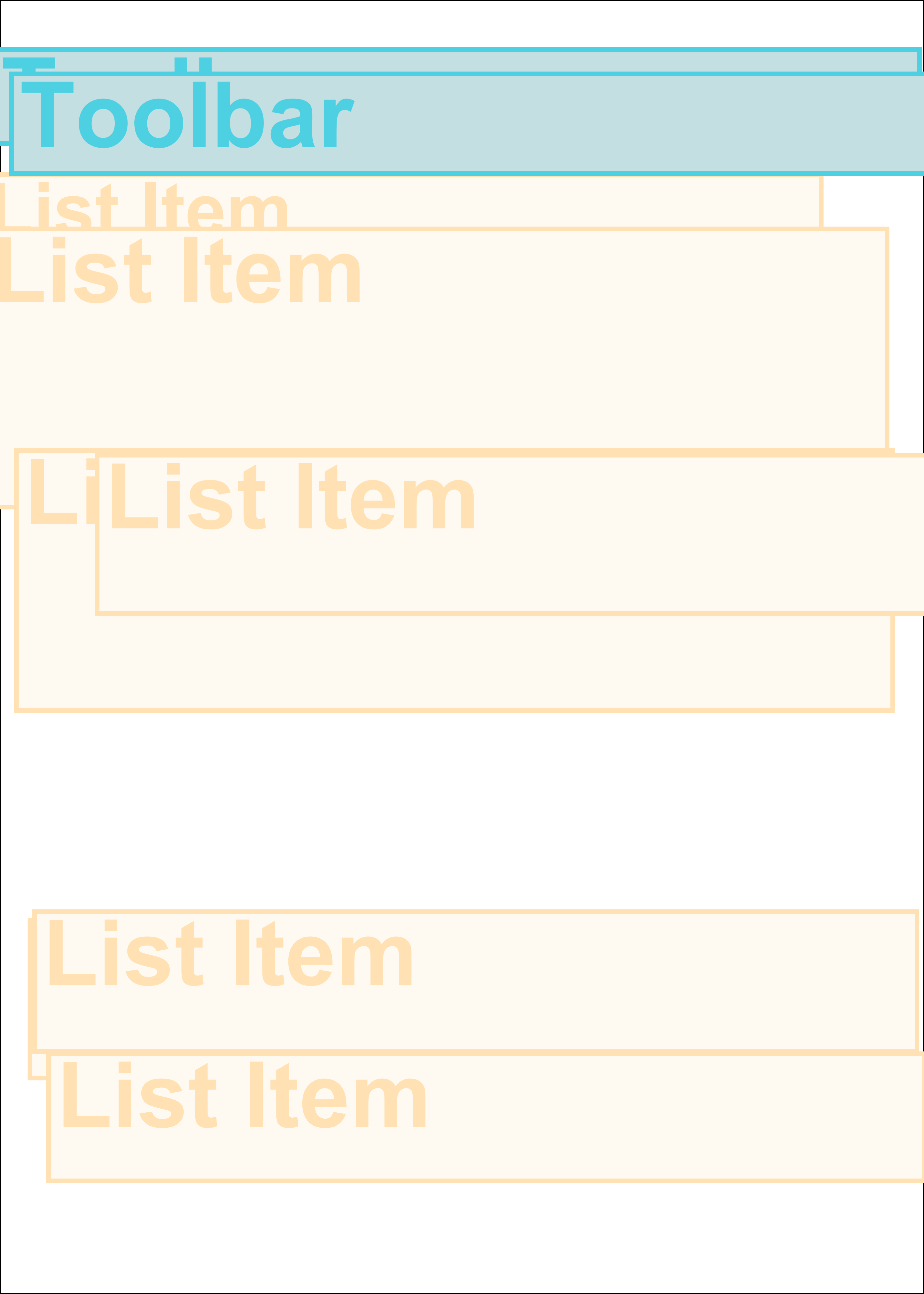} &
\includegraphics[width=\ricoBulkWidth]{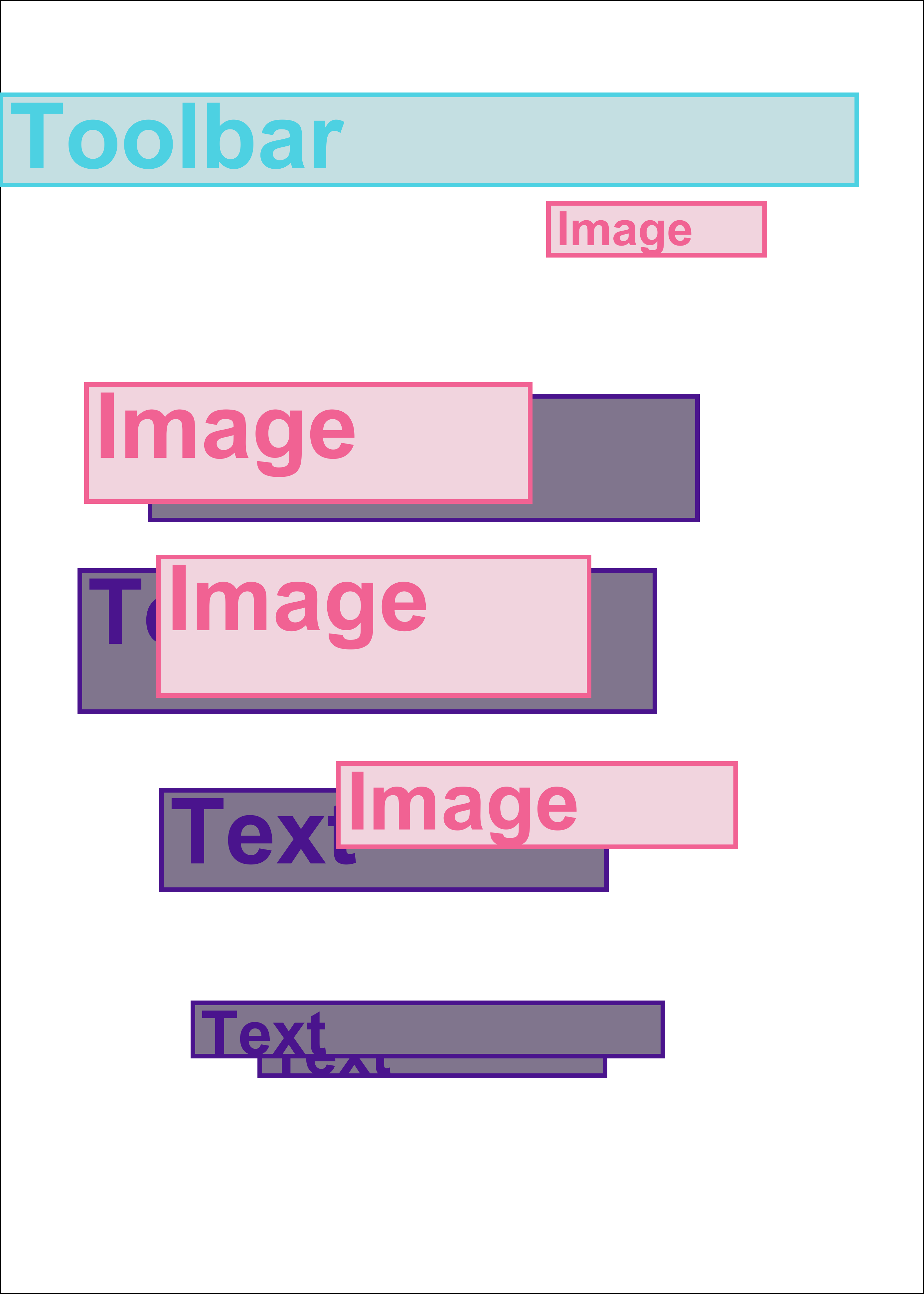} &
\includegraphics[width=\ricoBulkWidth]{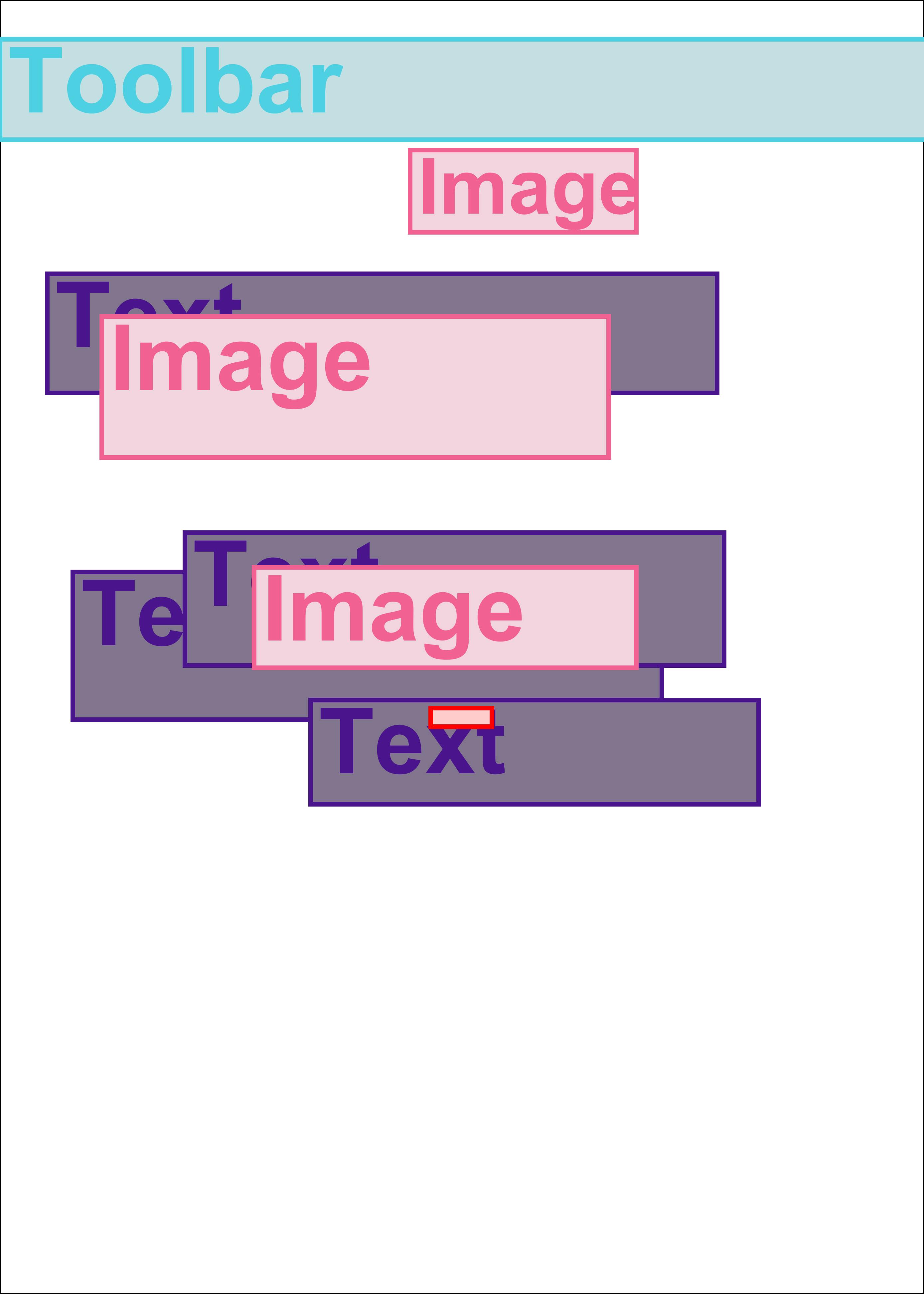} &
\includegraphics[width=\ricoBulkWidth]{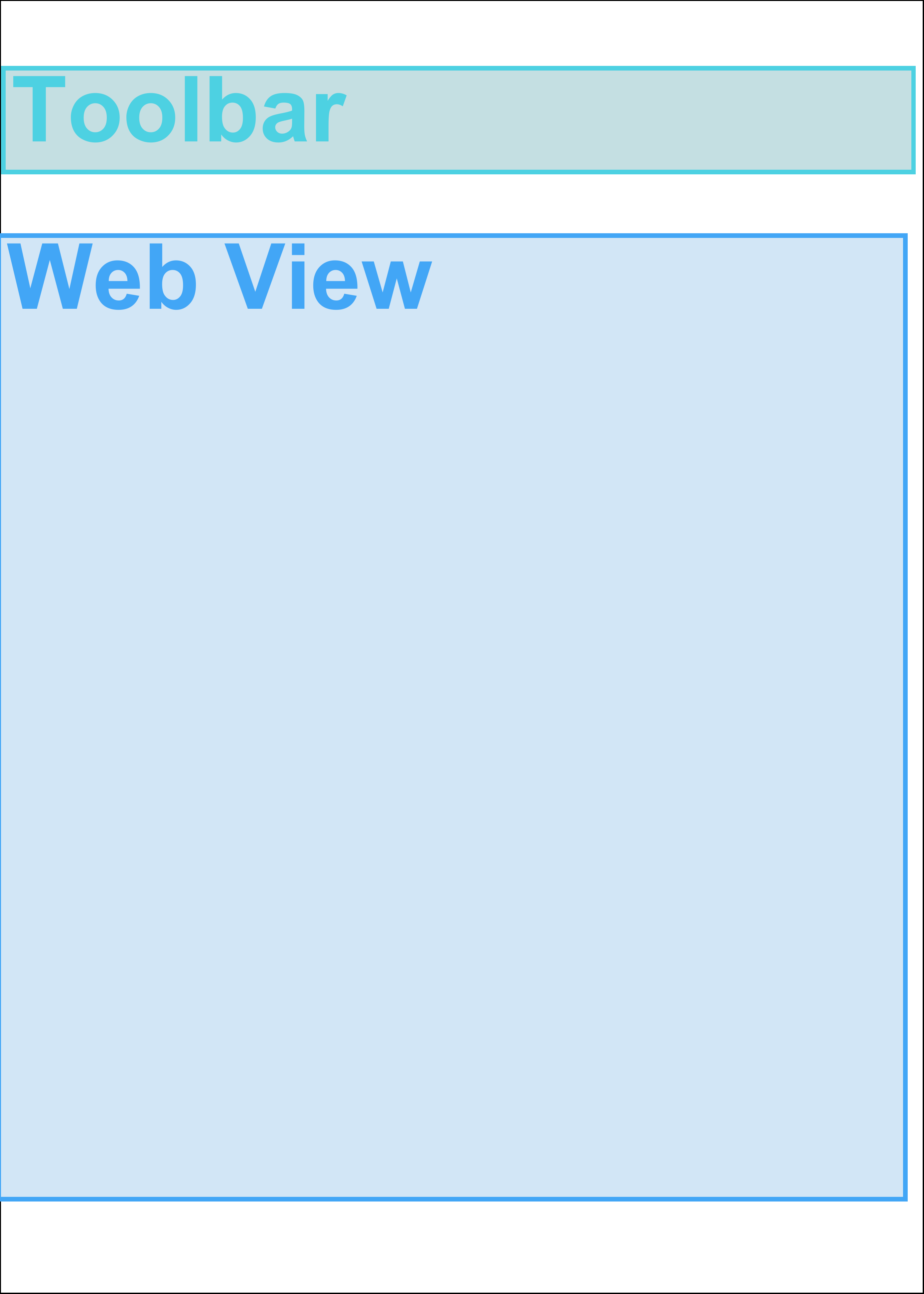} &
\includegraphics[width=\ricoBulkWidth]{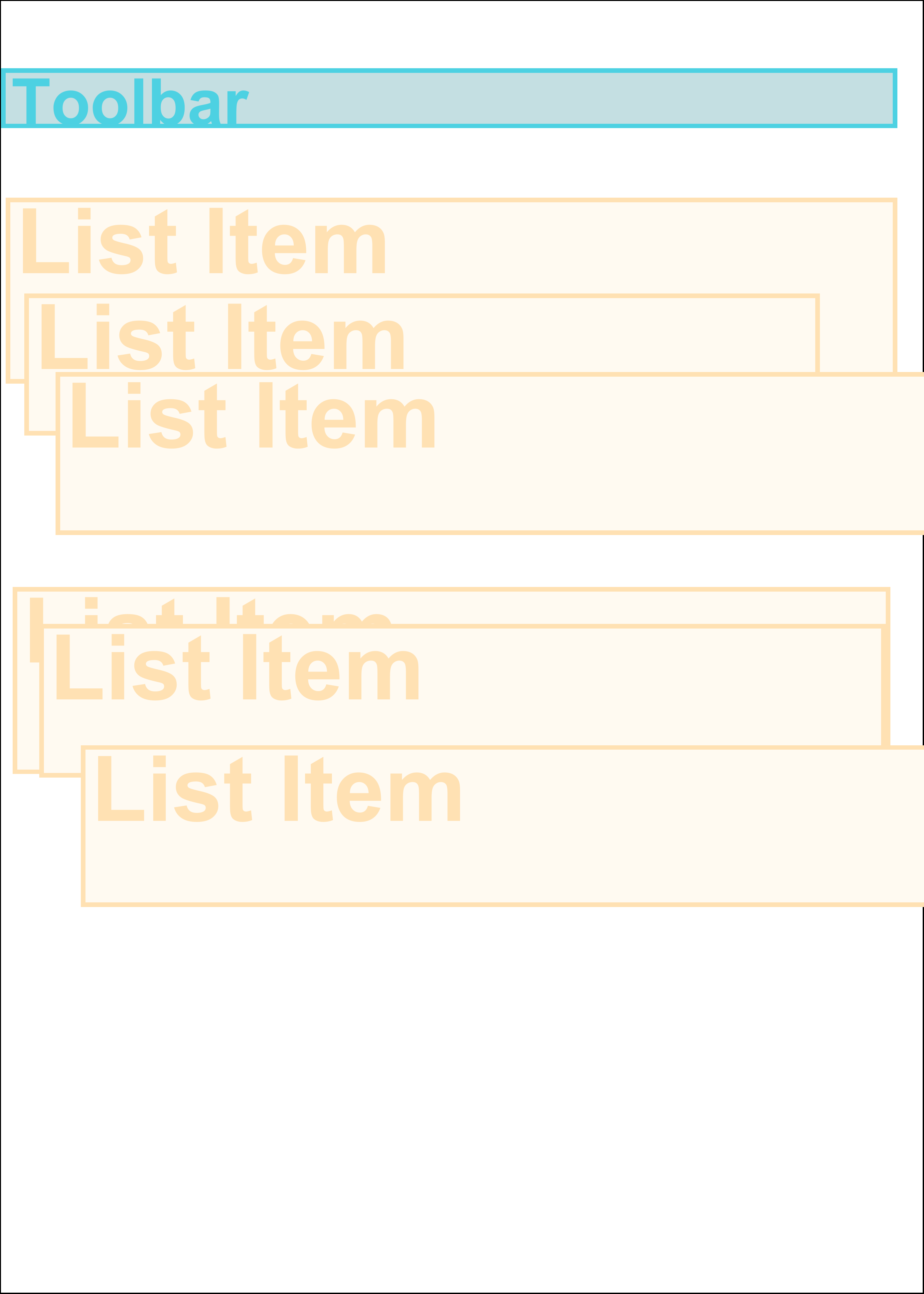} &
\includegraphics[width=\ricoBulkWidth]{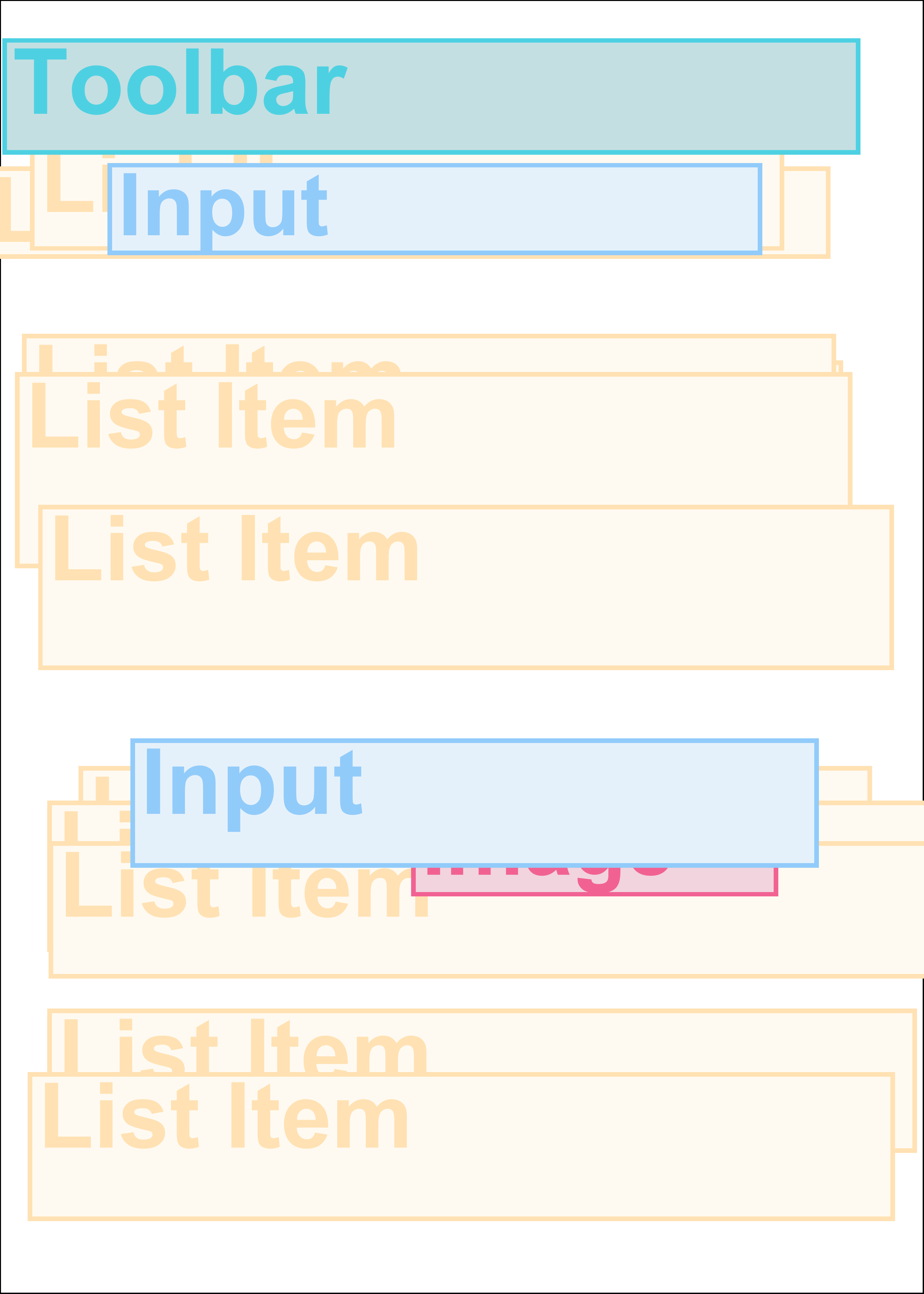} &
\includegraphics[width=\ricoBulkWidth]{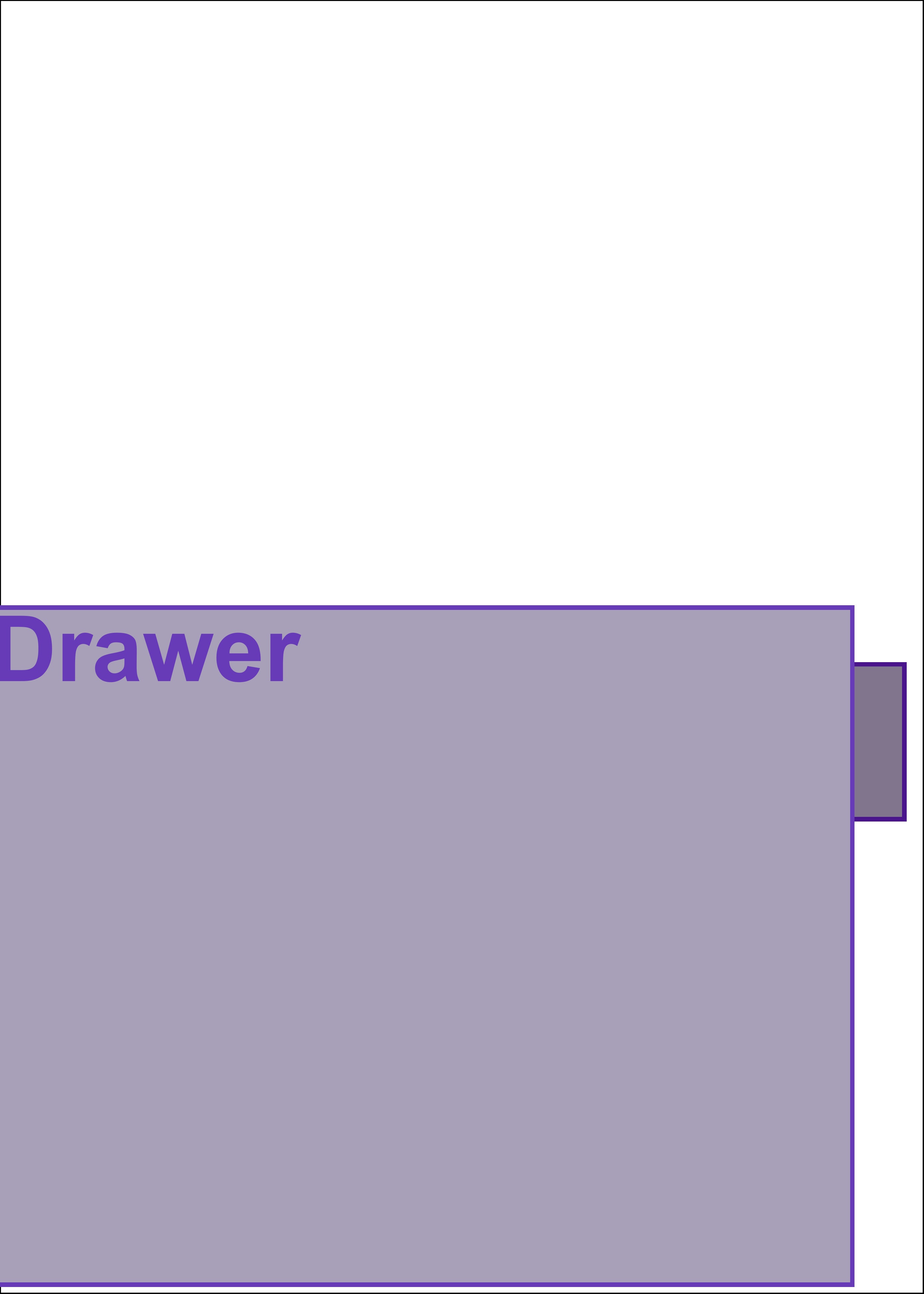}\\

\rotatebox{90}{\hspace{4mm}Gupta \etal [9]}&
\includegraphics[width=\ricoBulkWidth]{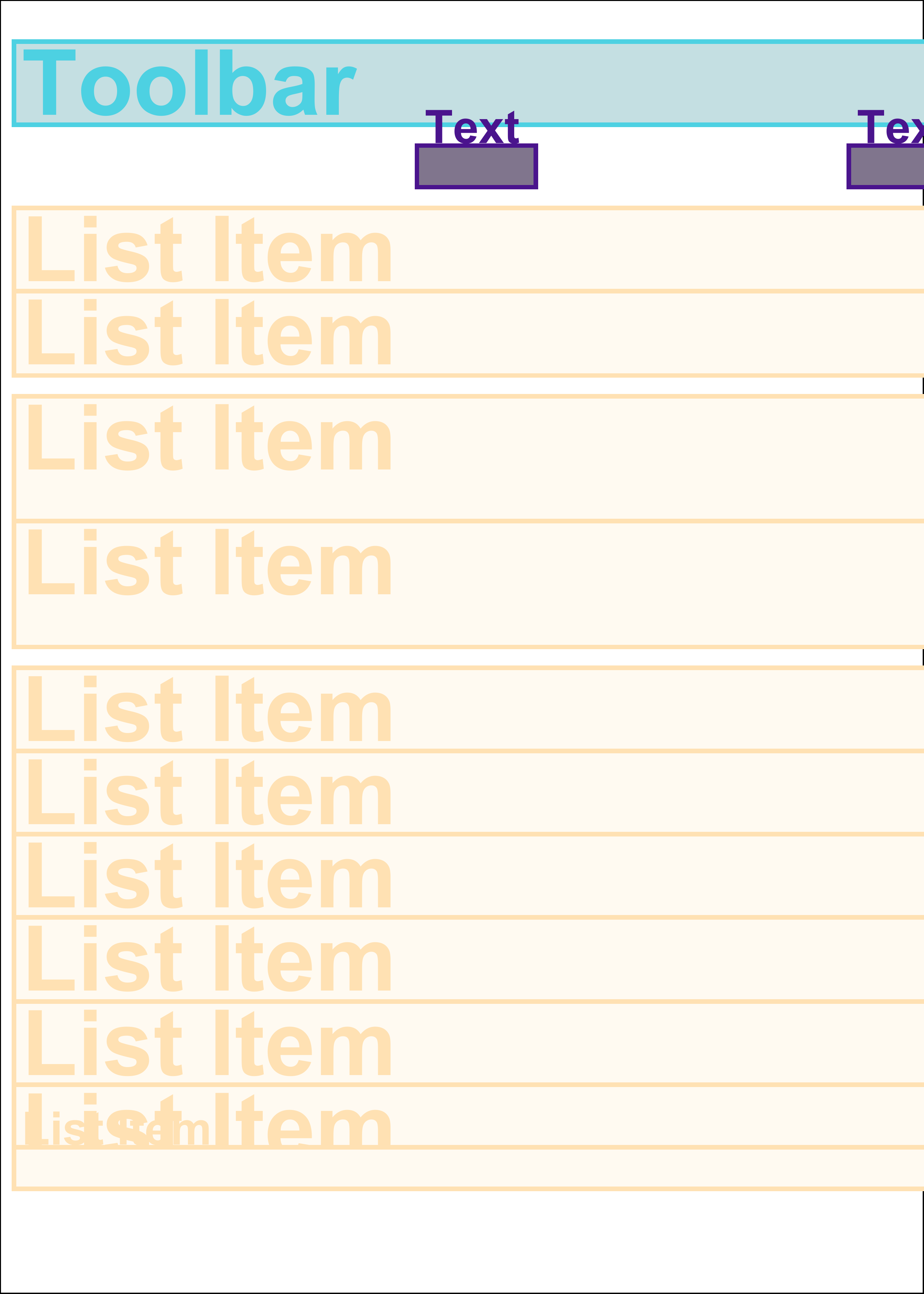}&
\includegraphics[width=\ricoBulkWidth]{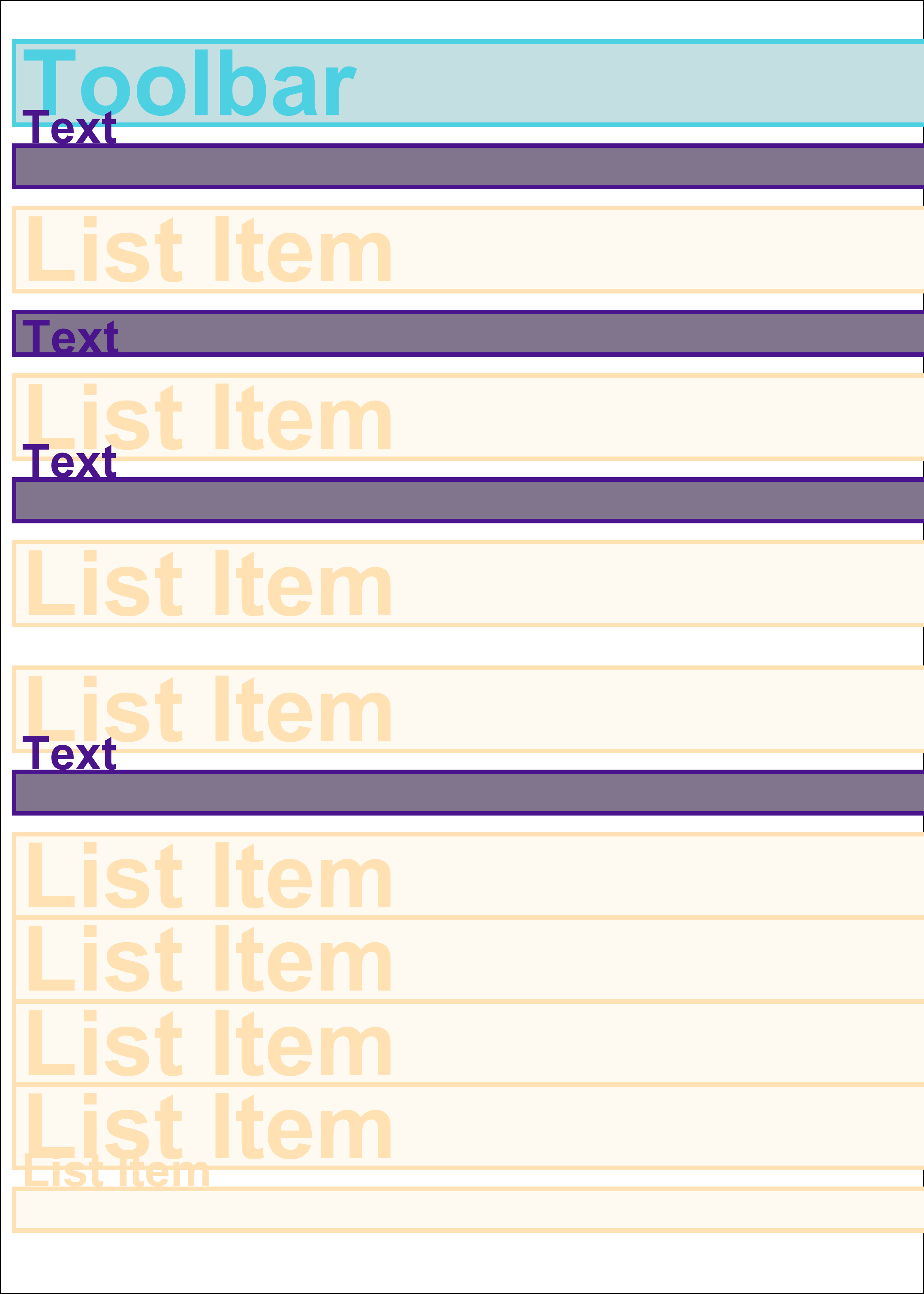}&
\includegraphics[width=\ricoBulkWidth]{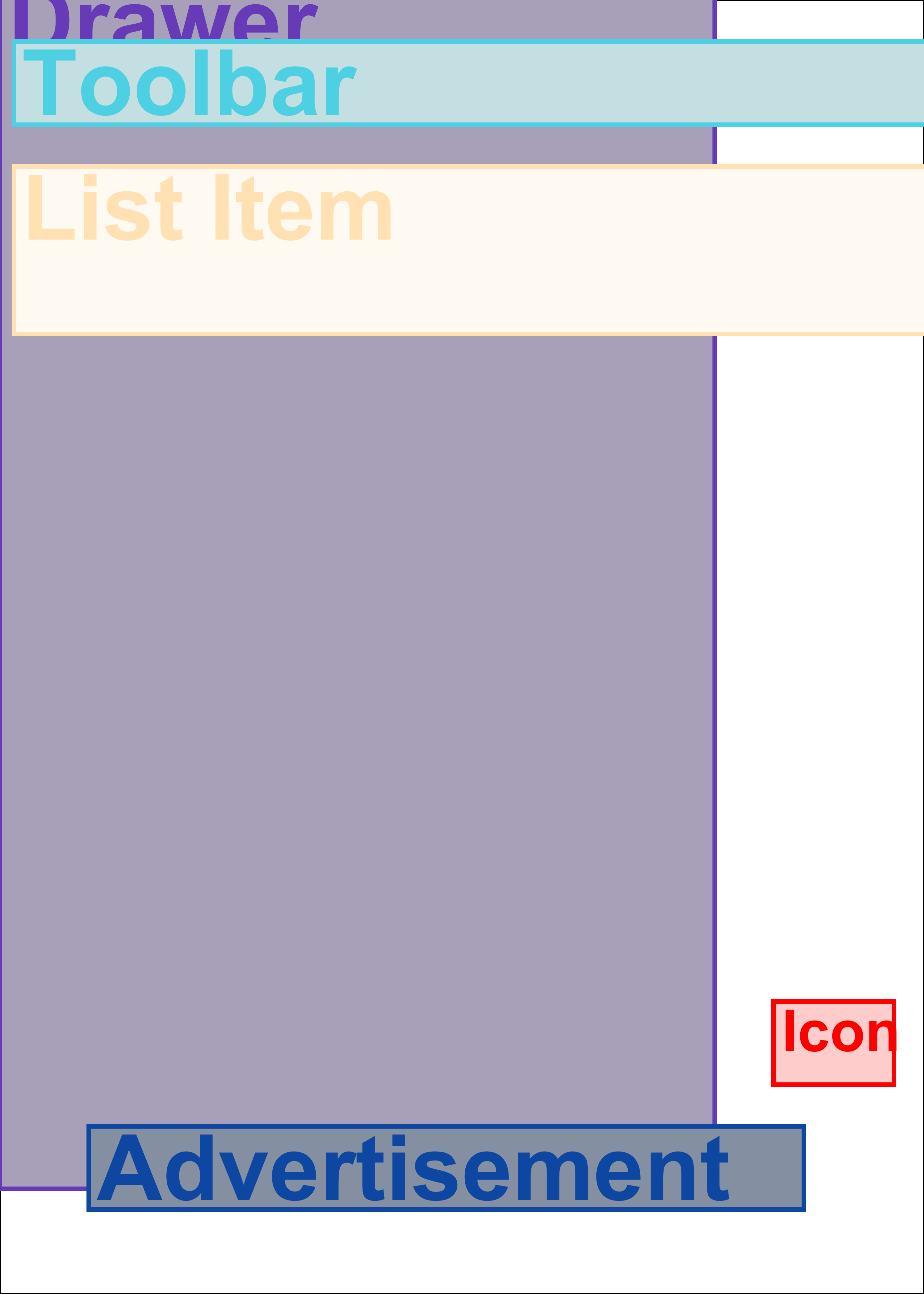}&
\includegraphics[width=\ricoBulkWidth]{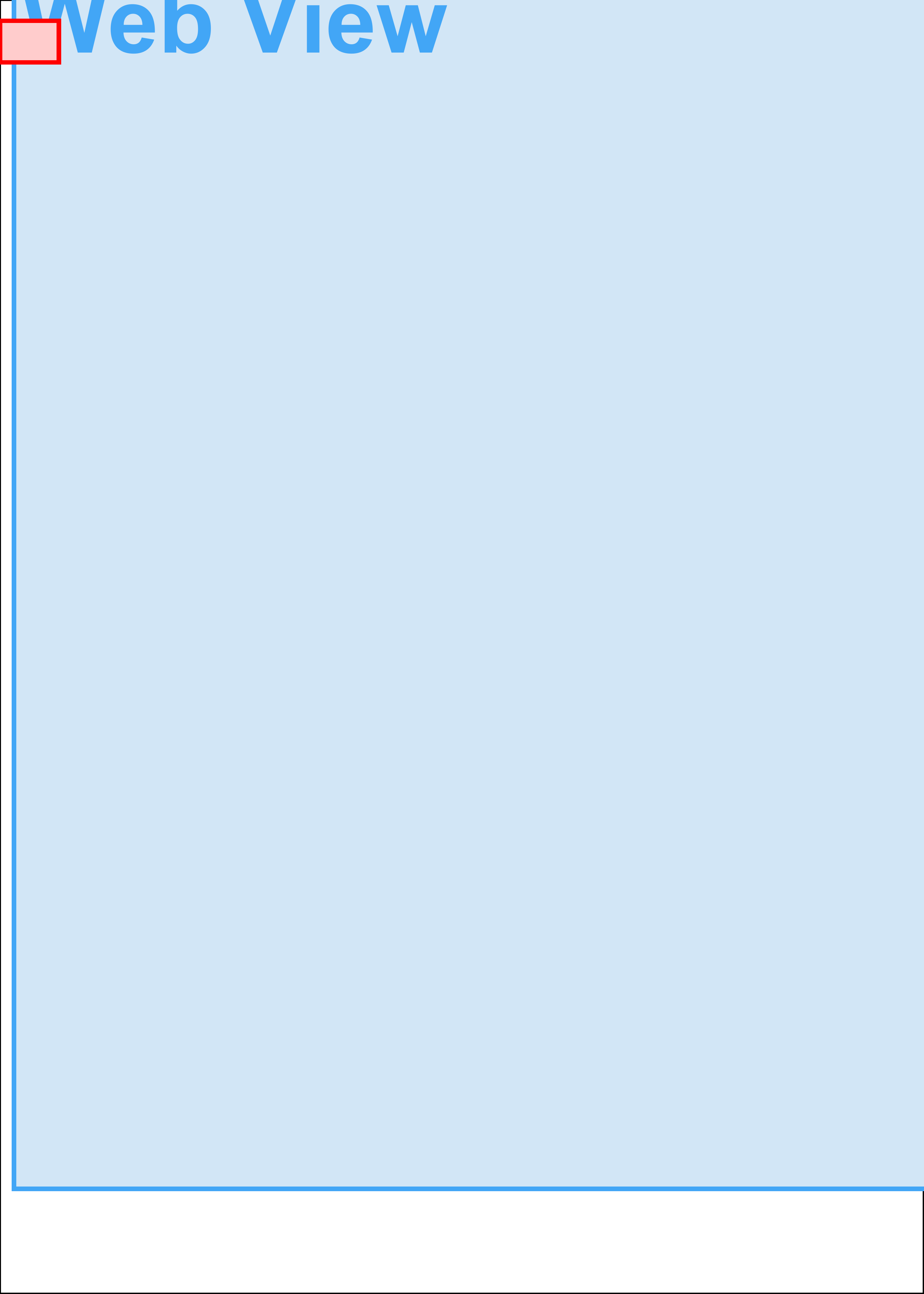}&
\includegraphics[width=\ricoBulkWidth]{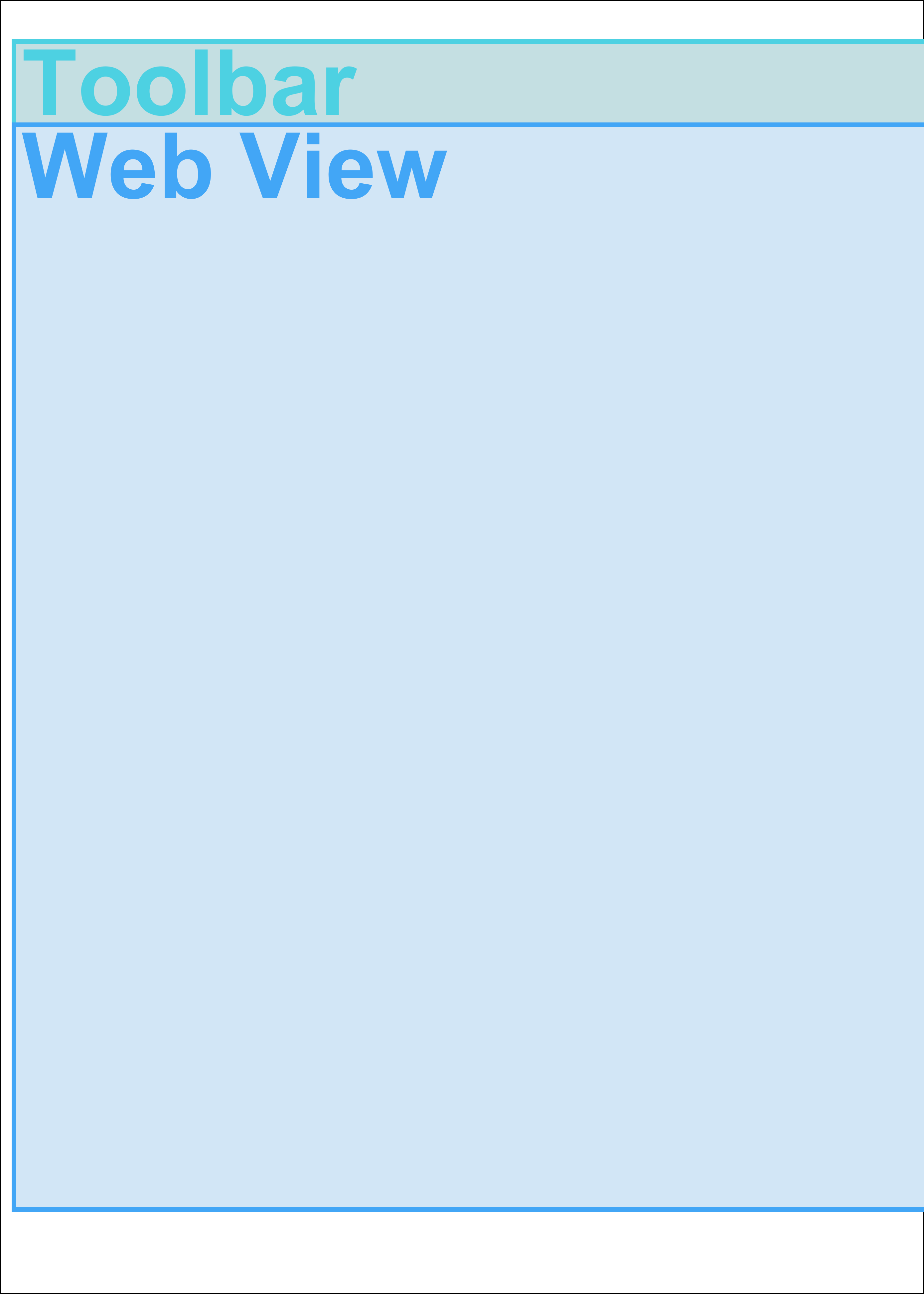}&
\includegraphics[width=\ricoBulkWidth]{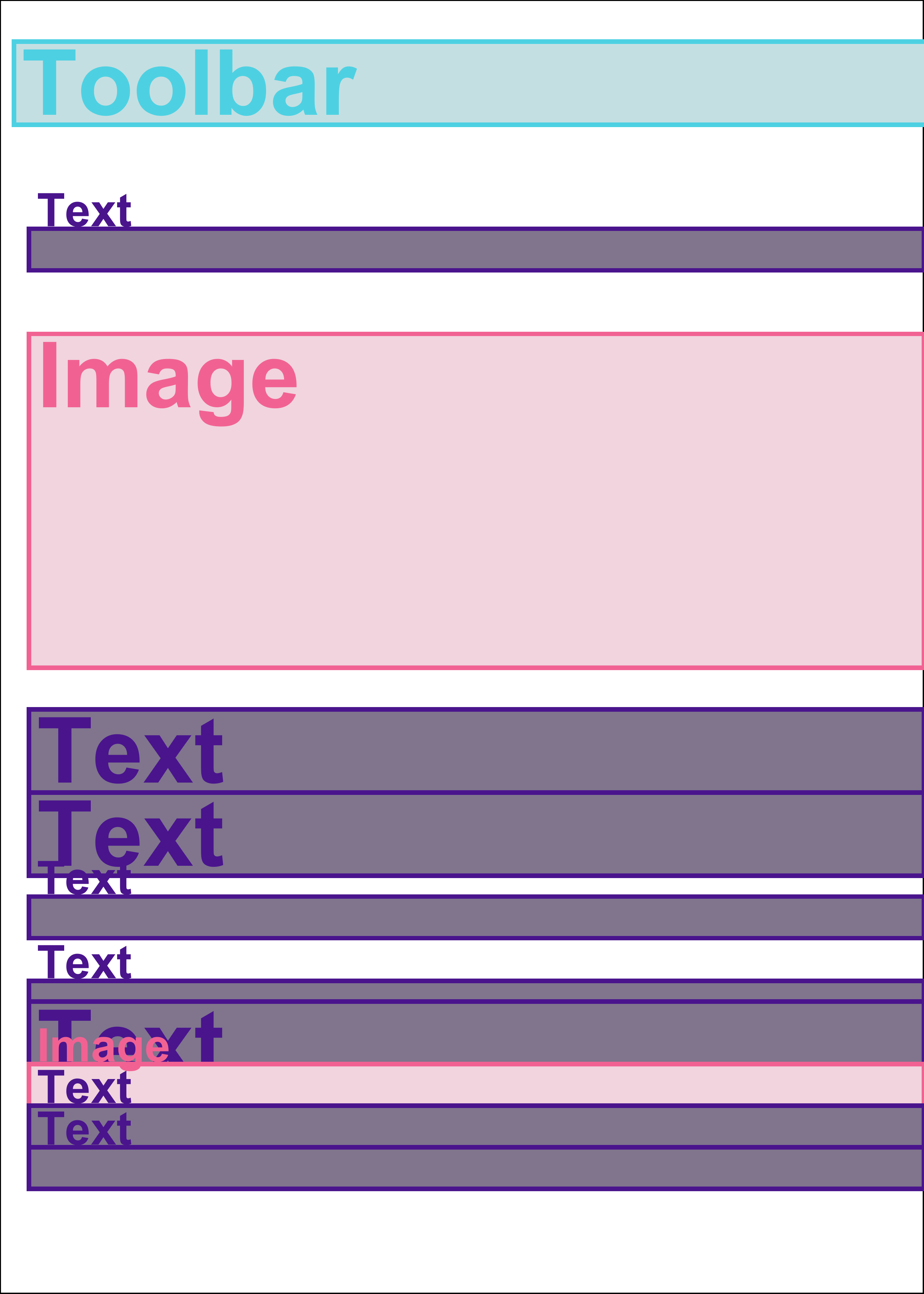}&
\includegraphics[width=\ricoBulkWidth]{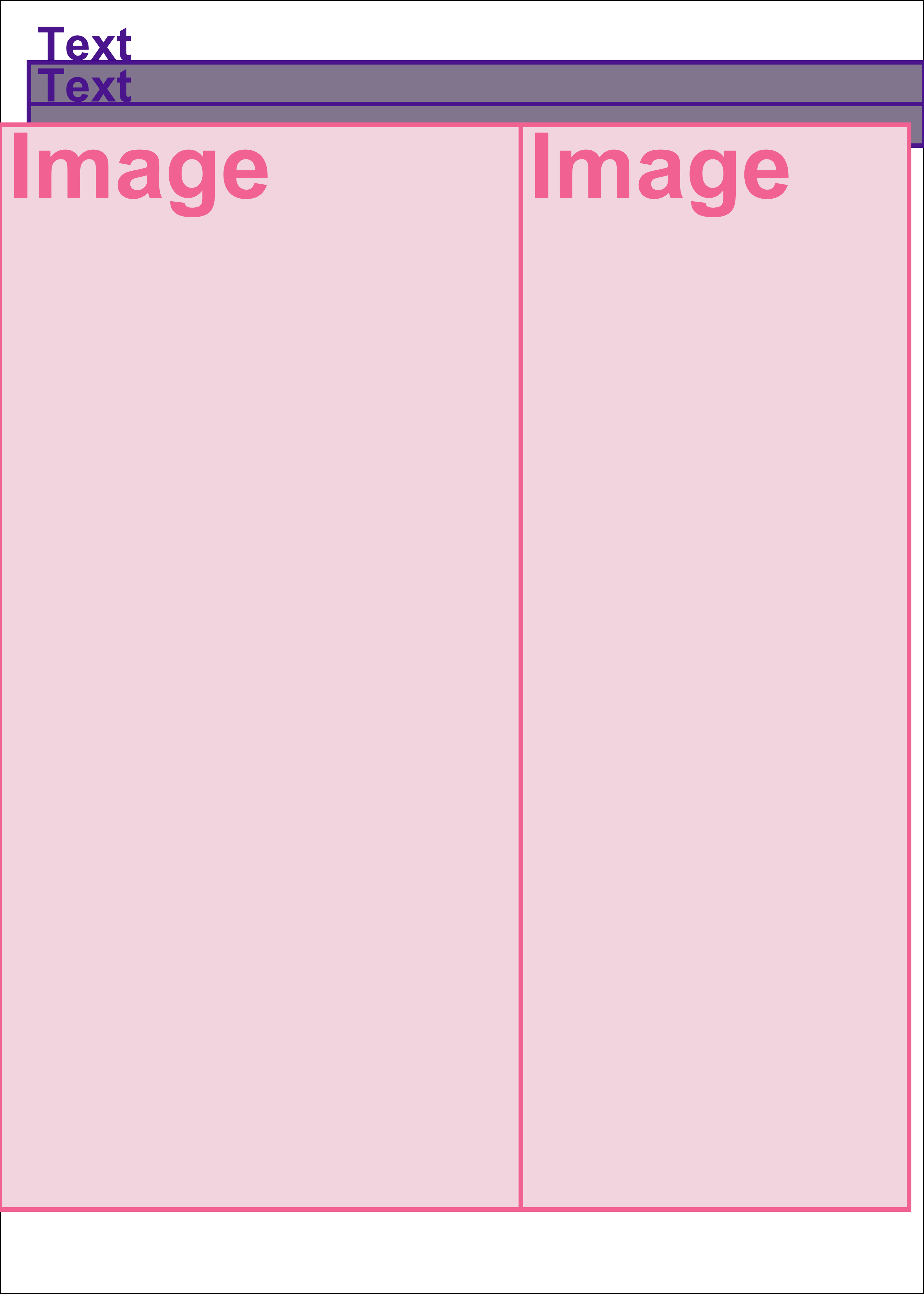}

\\\toprule

\rotatebox{90}{\hspace{10mm}Ours}&\includegraphics[width=\ricoBulkWidth]{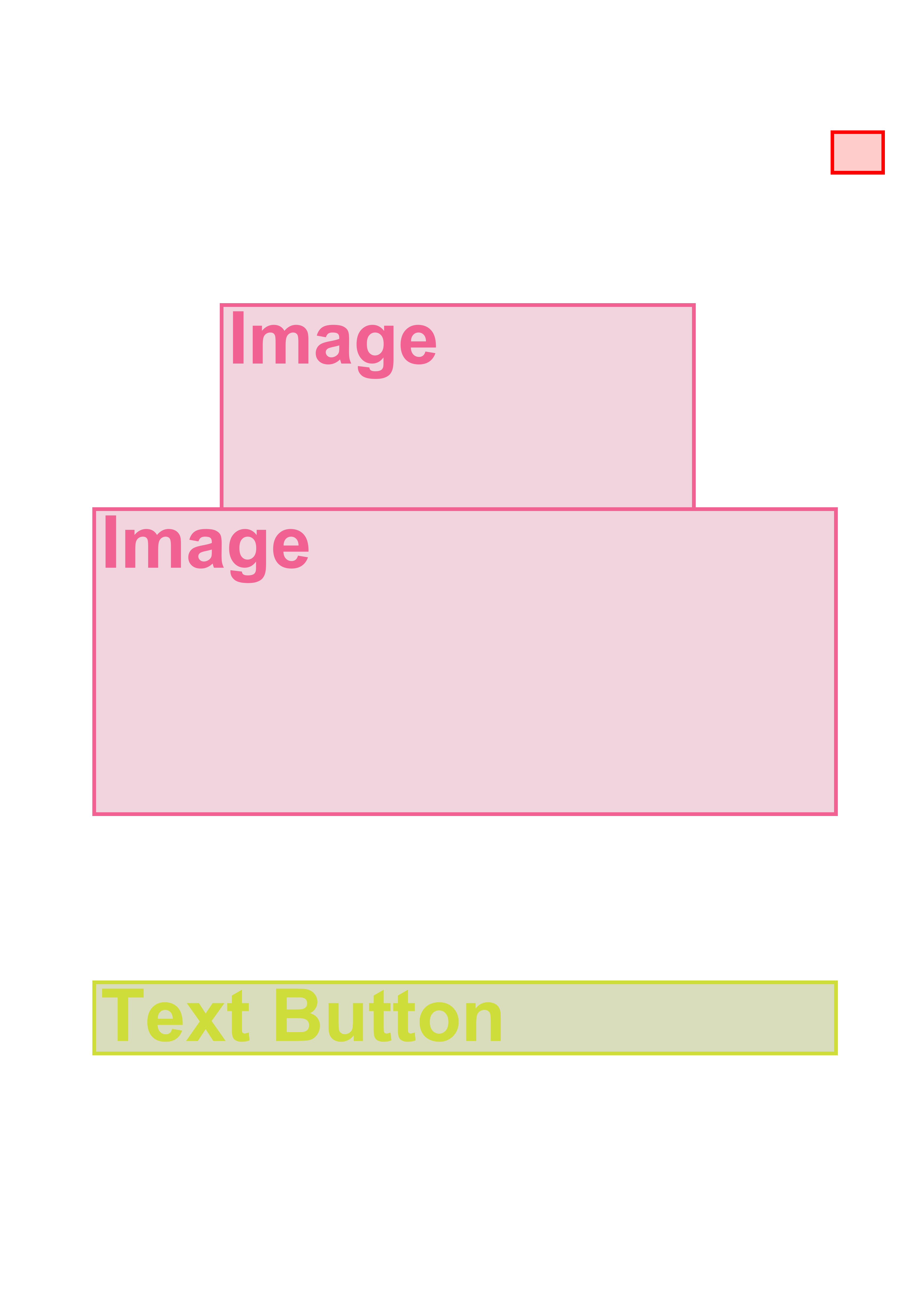} & 
\includegraphics[width=\ricoBulkWidth]{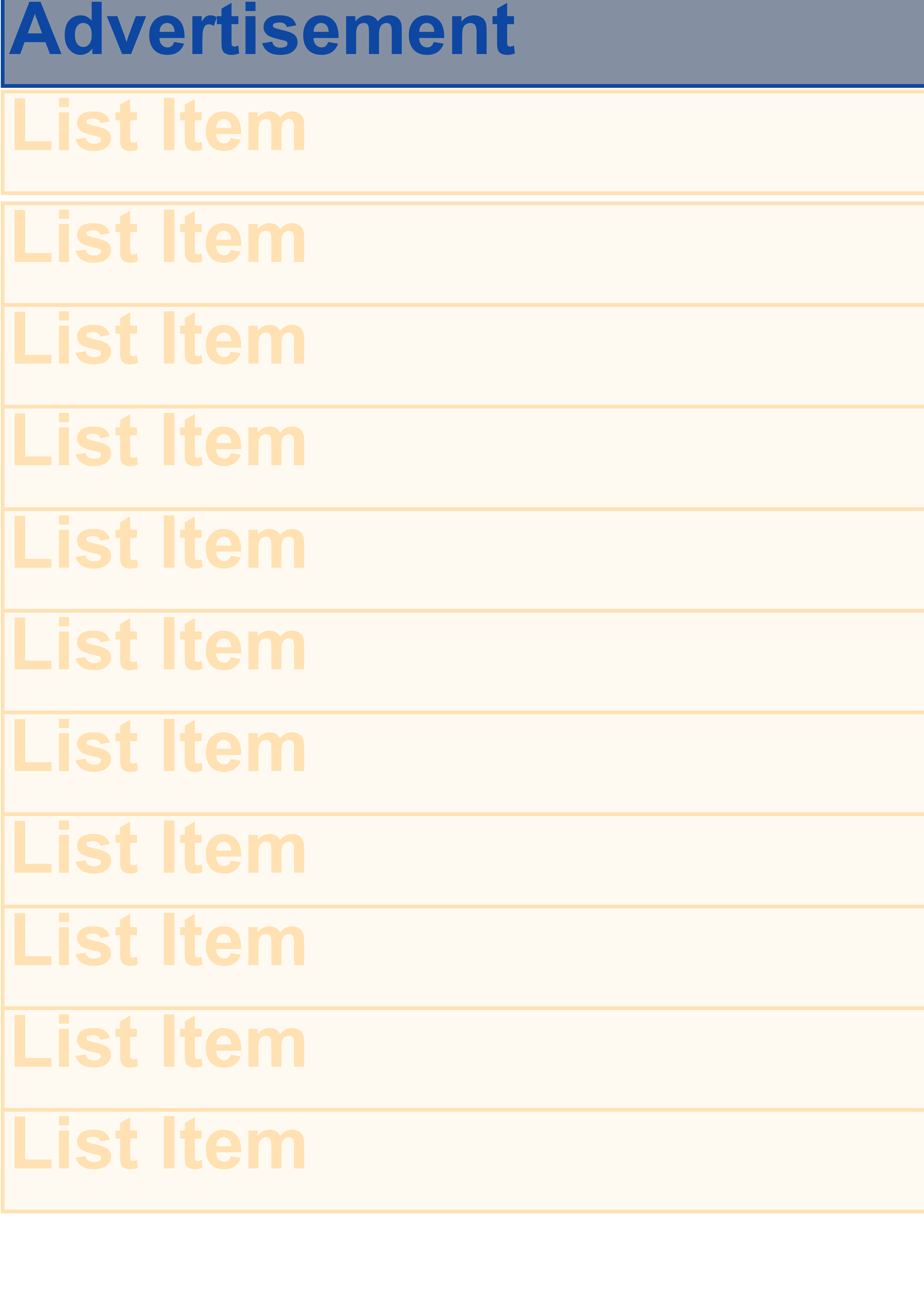} & 
\includegraphics[width=\ricoBulkWidth]{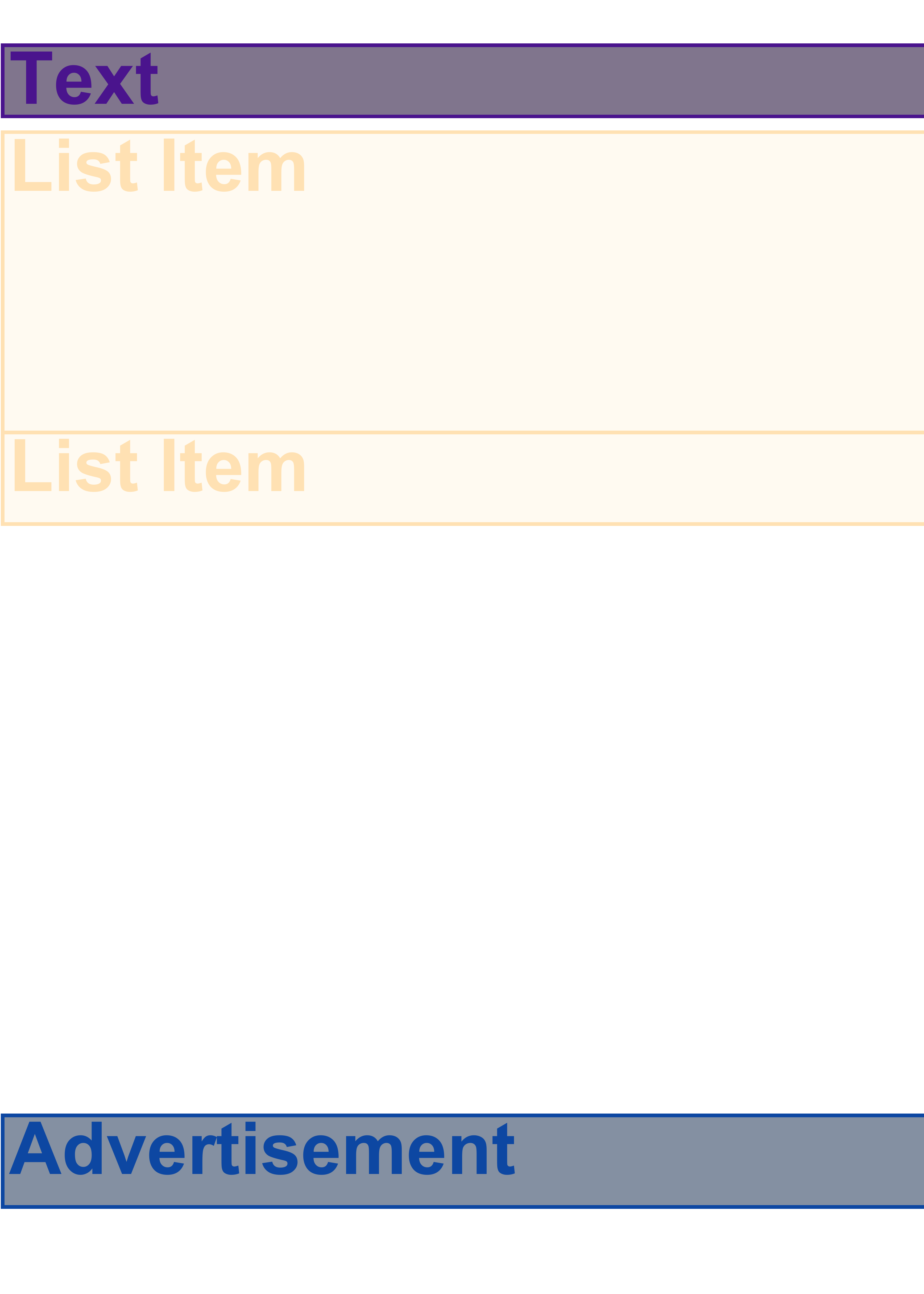} & 
\includegraphics[width=\ricoBulkWidth]{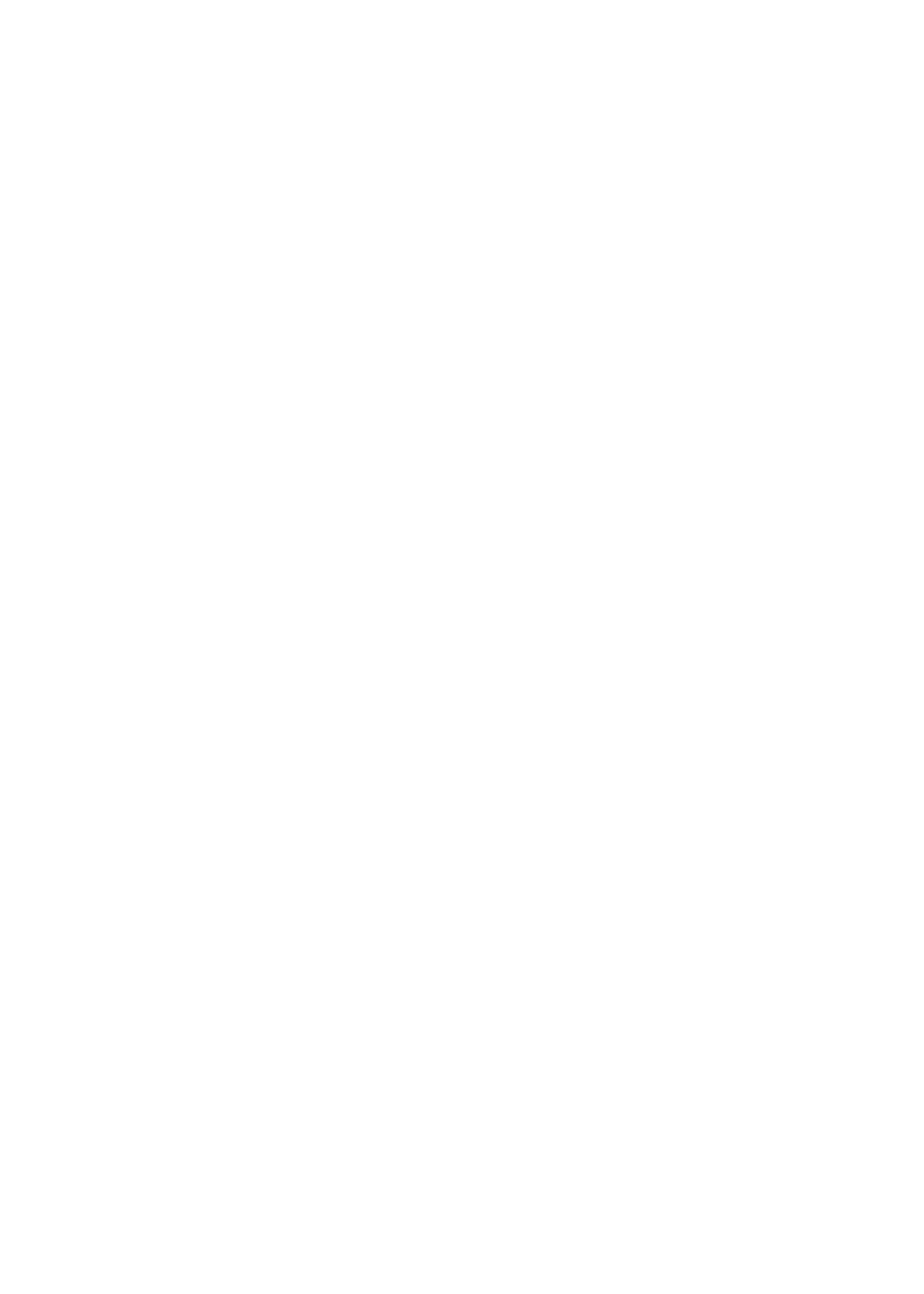} & 
\includegraphics[width=\ricoBulkWidth]{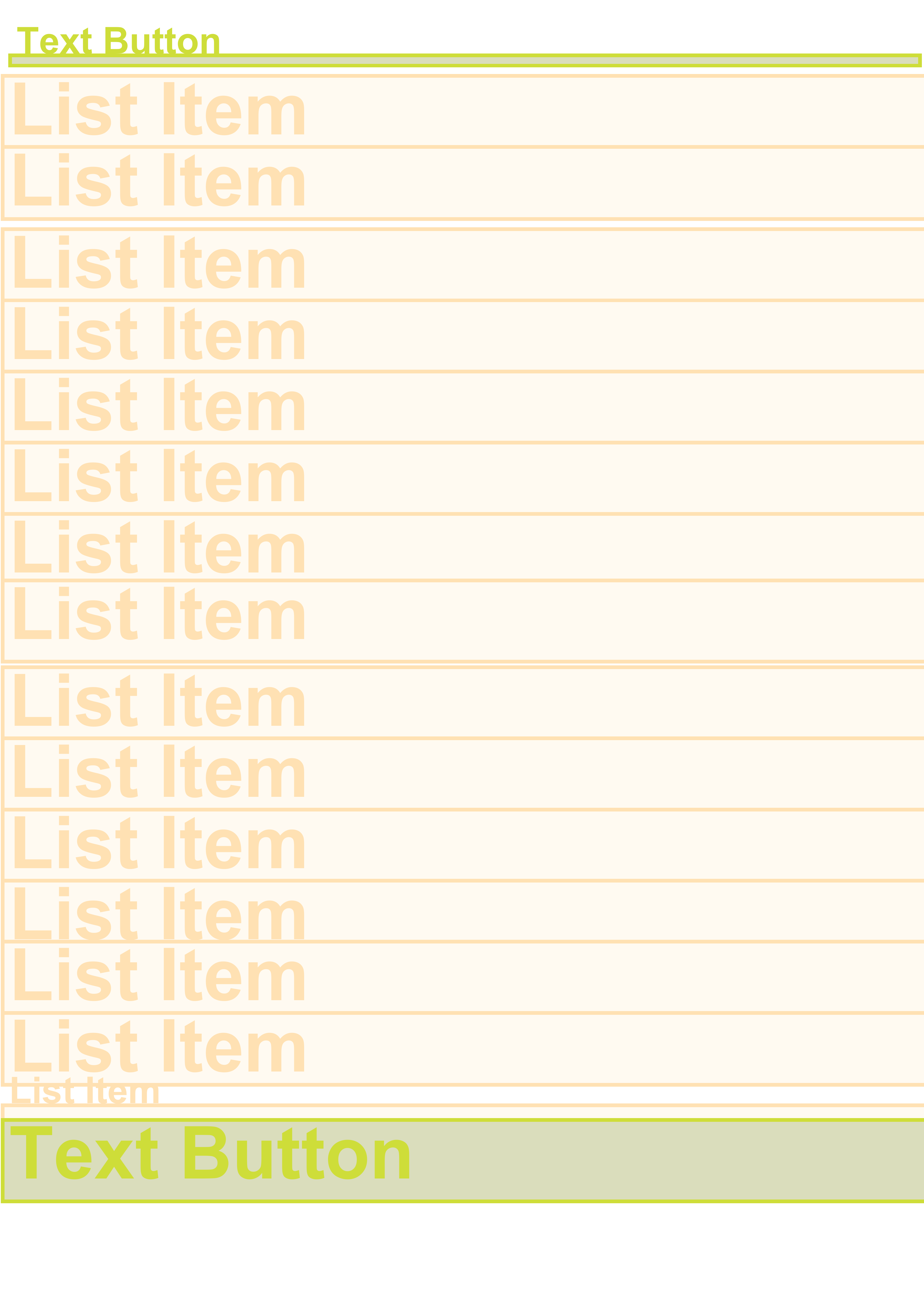} & 
\includegraphics[width=\ricoBulkWidth]{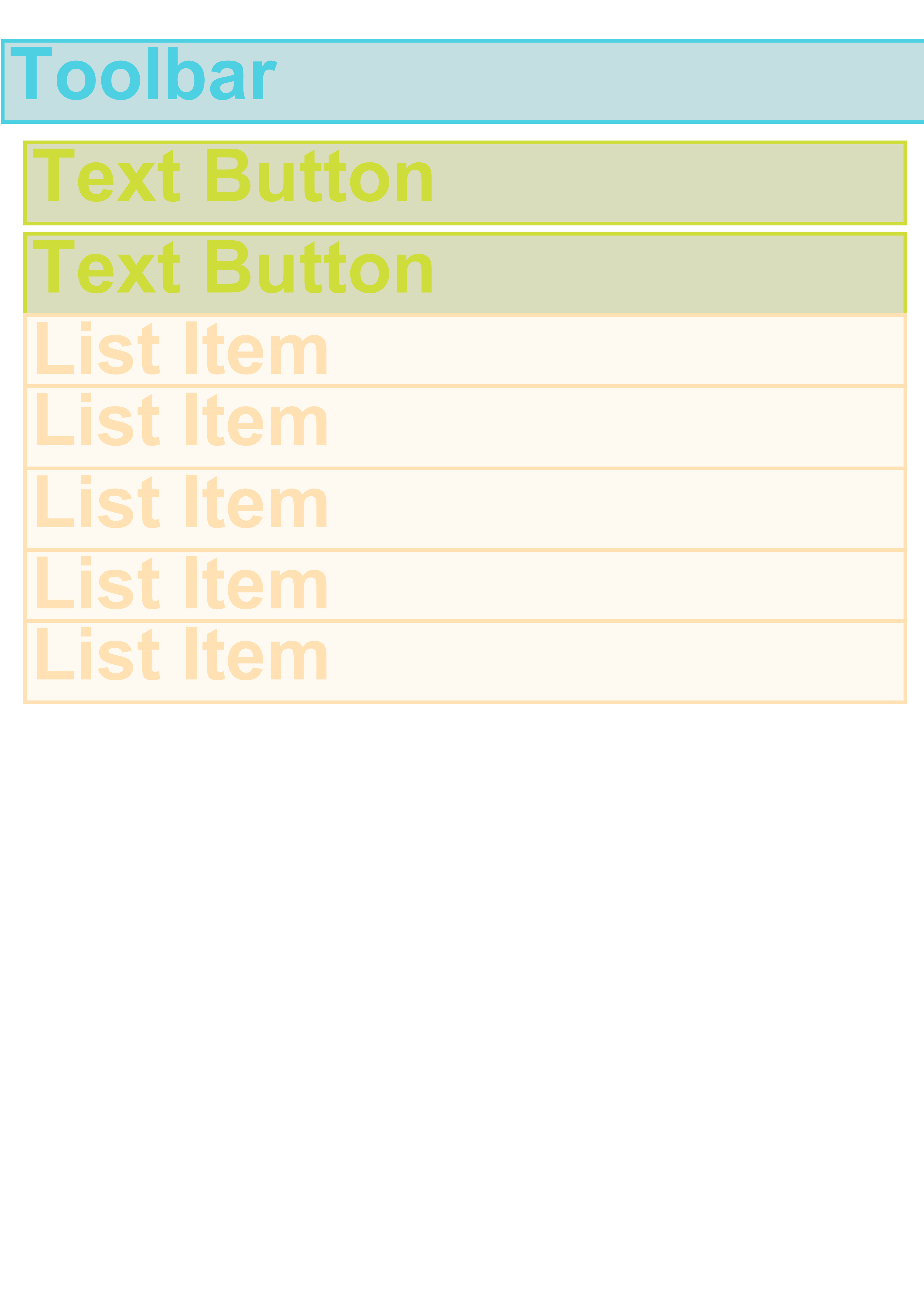} & 
\includegraphics[width=\ricoBulkWidth]{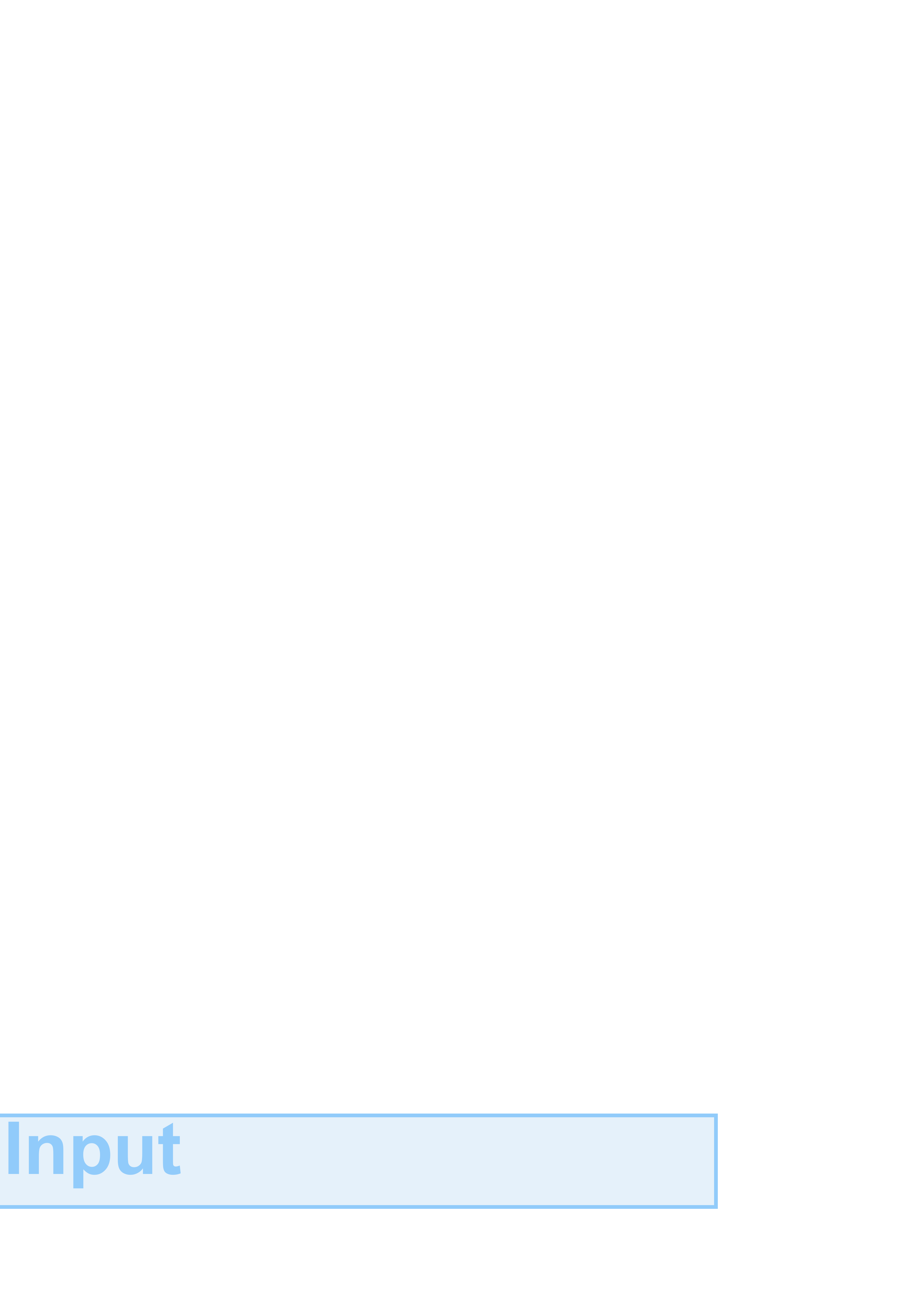} \\ 
&\includegraphics[width=\ricoBulkWidth]{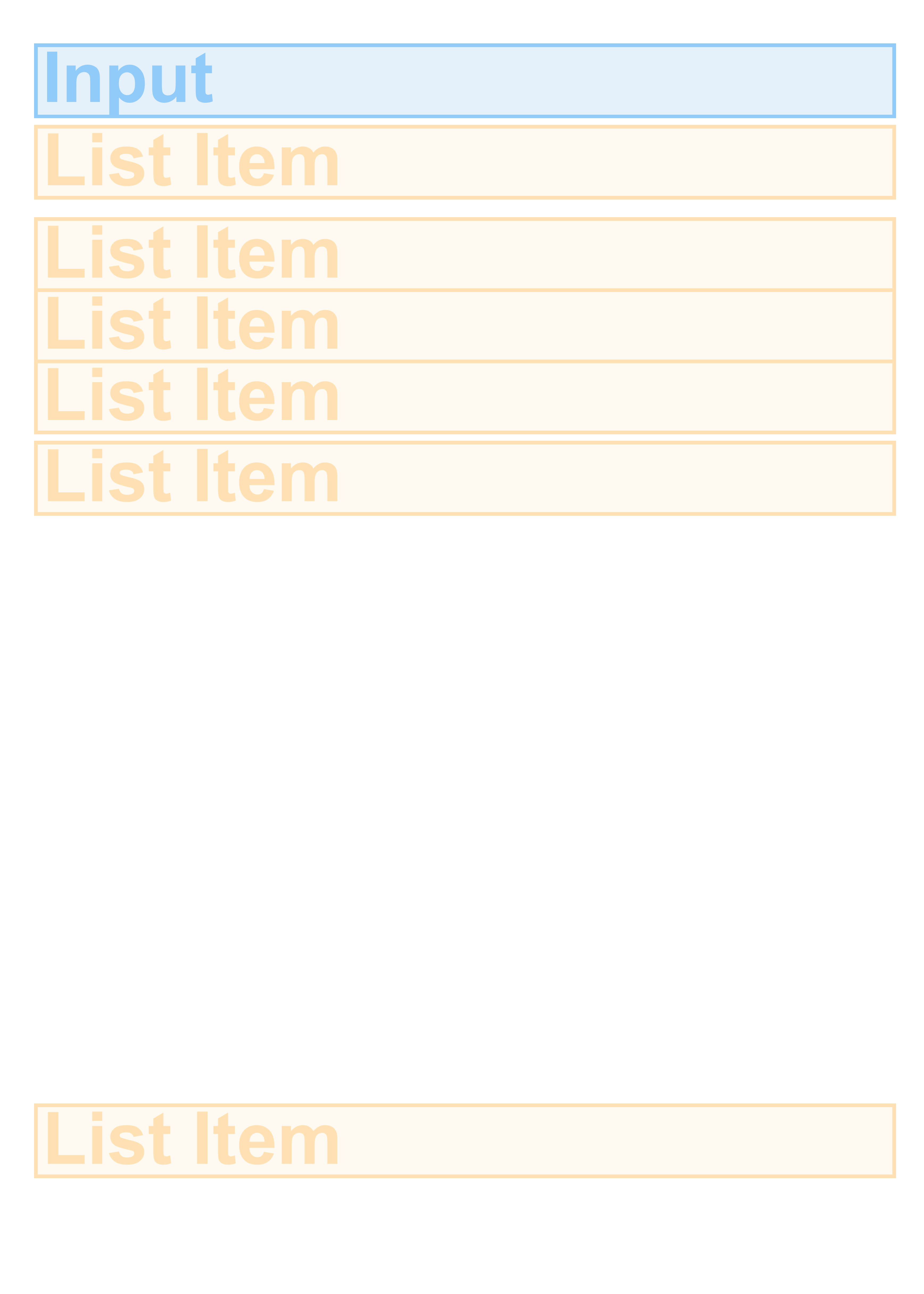} & 
\includegraphics[width=\ricoBulkWidth]{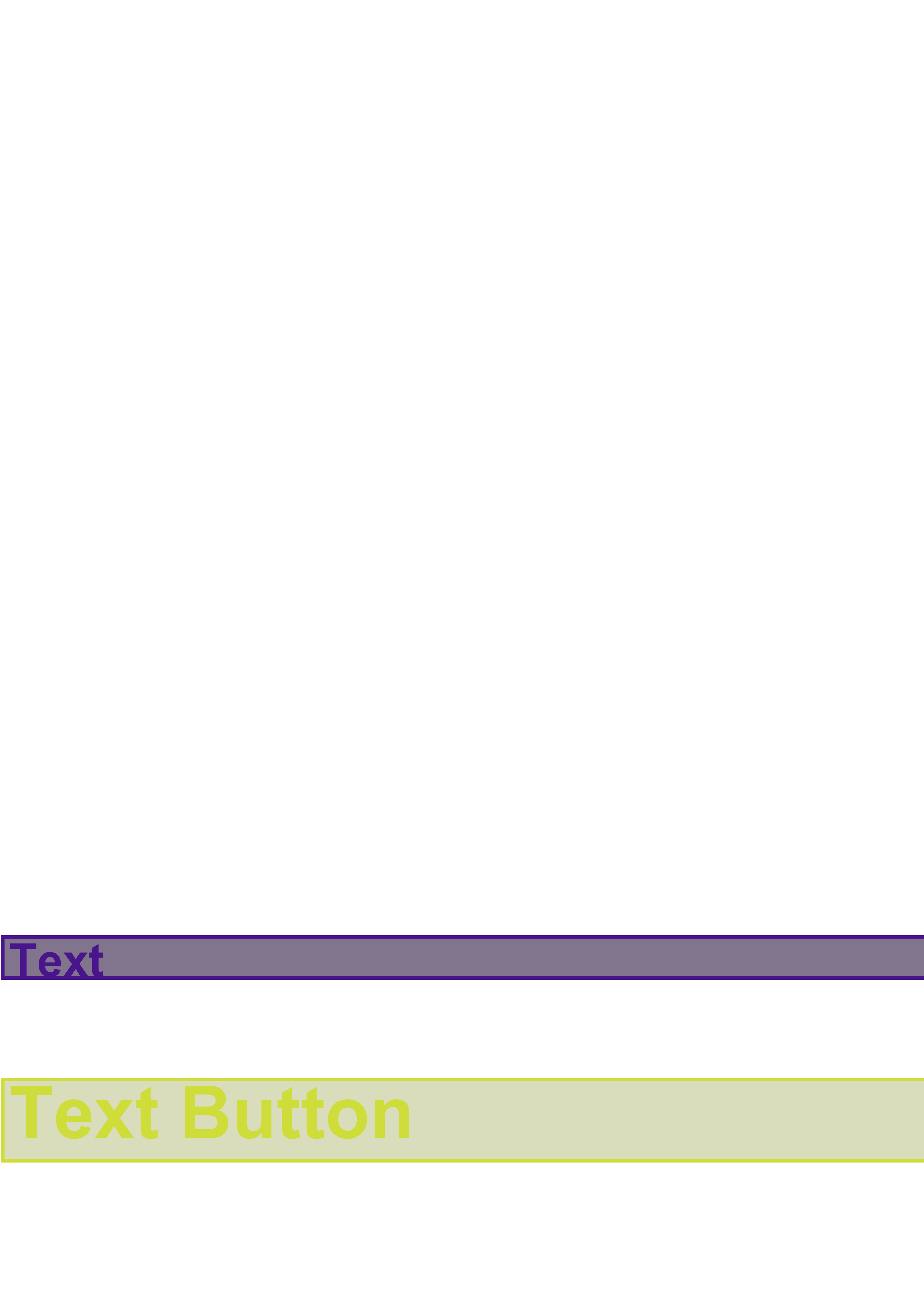} & 
\includegraphics[width=\ricoBulkWidth]{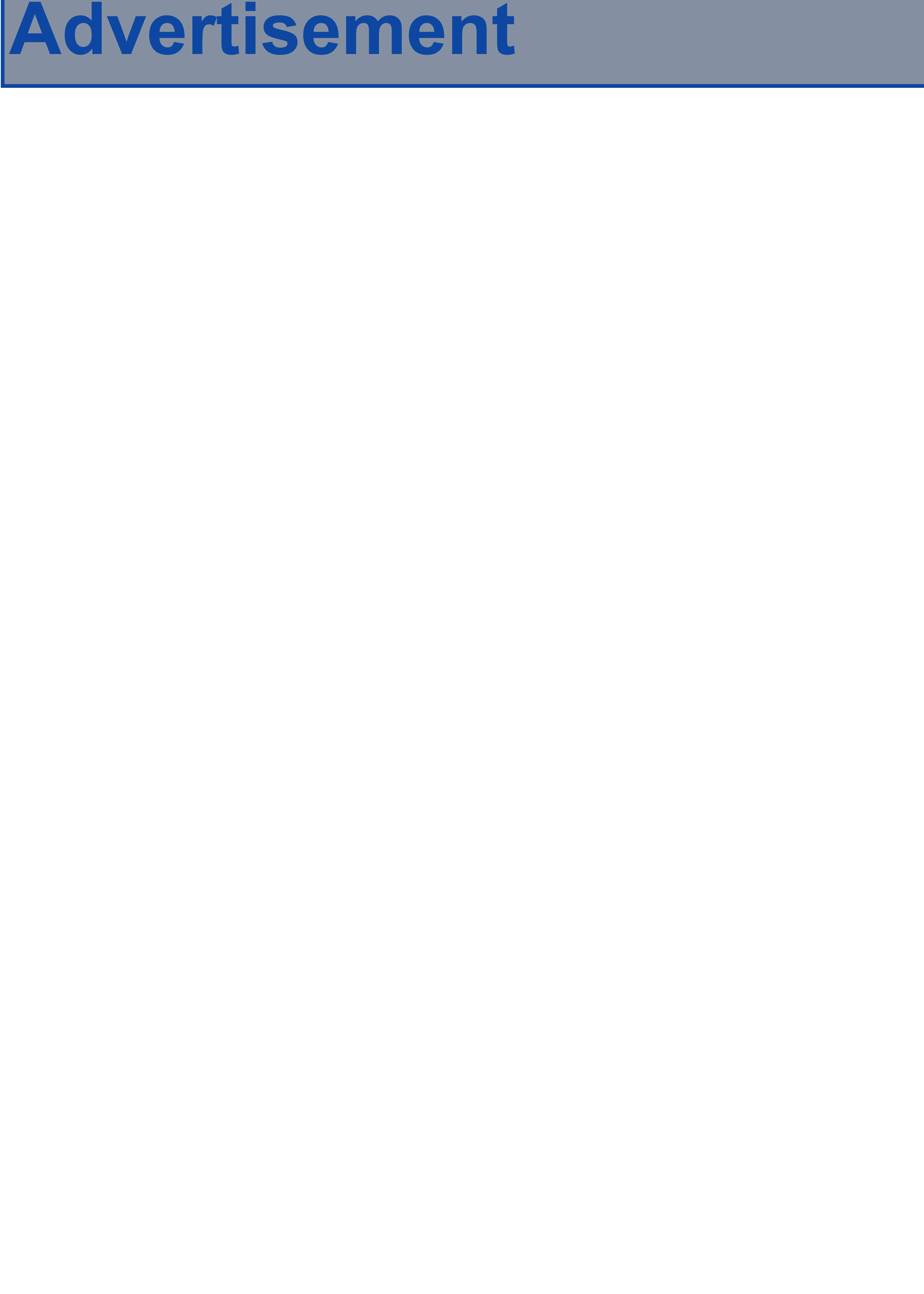} & 
\includegraphics[width=\ricoBulkWidth]{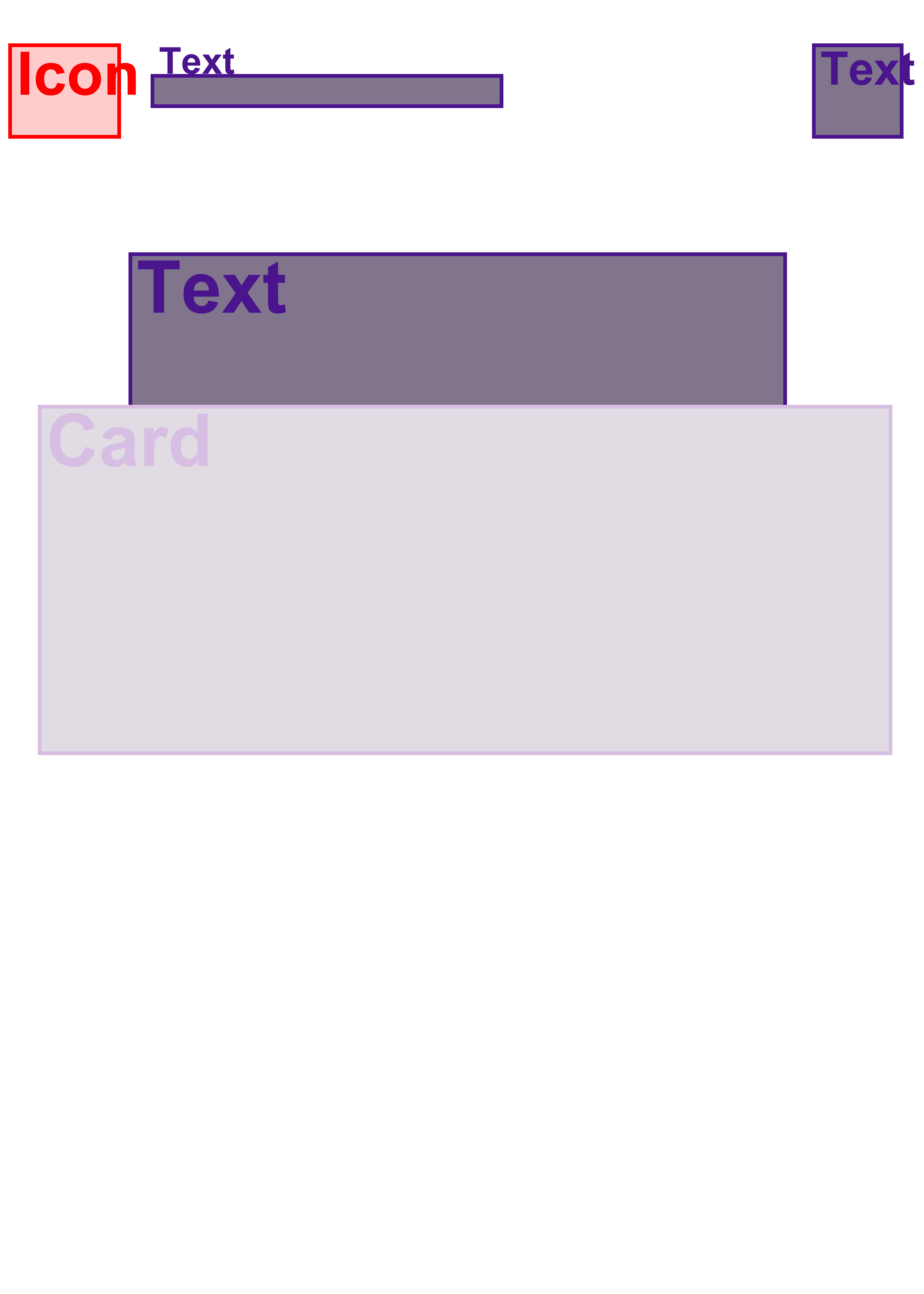} & 
\includegraphics[width=\ricoBulkWidth]{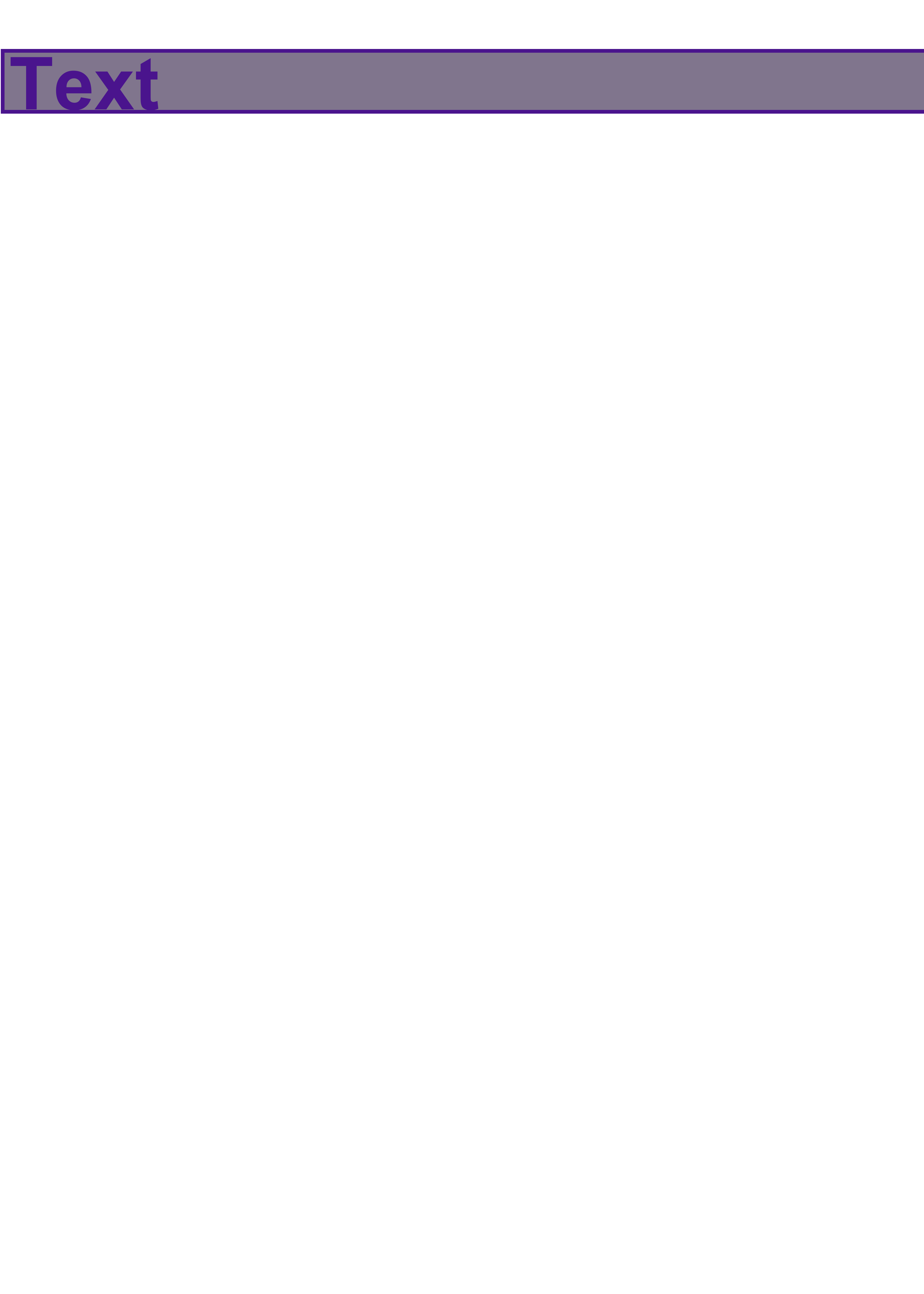} & 
\includegraphics[width=\ricoBulkWidth]{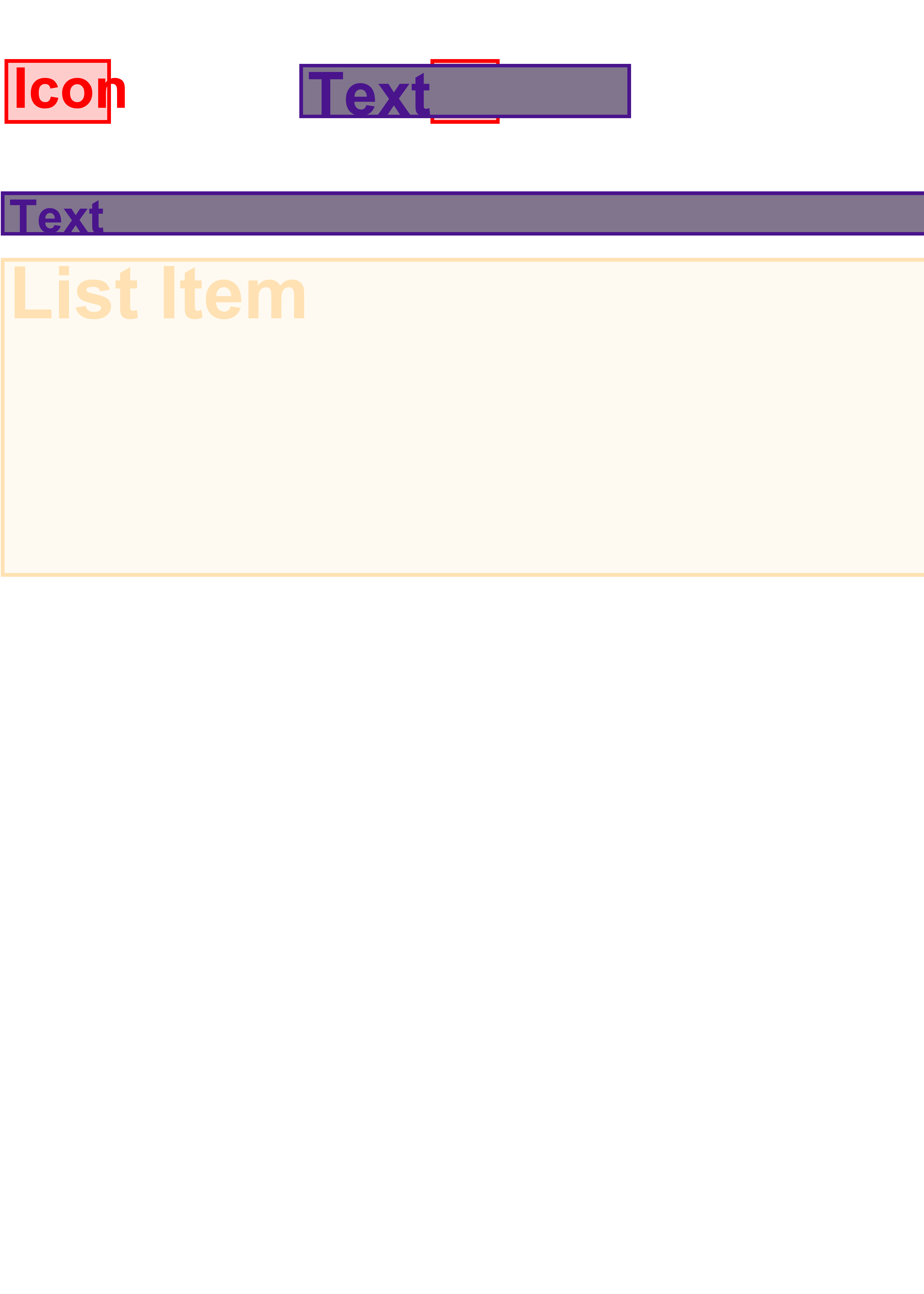} & 
\includegraphics[width=\ricoBulkWidth]{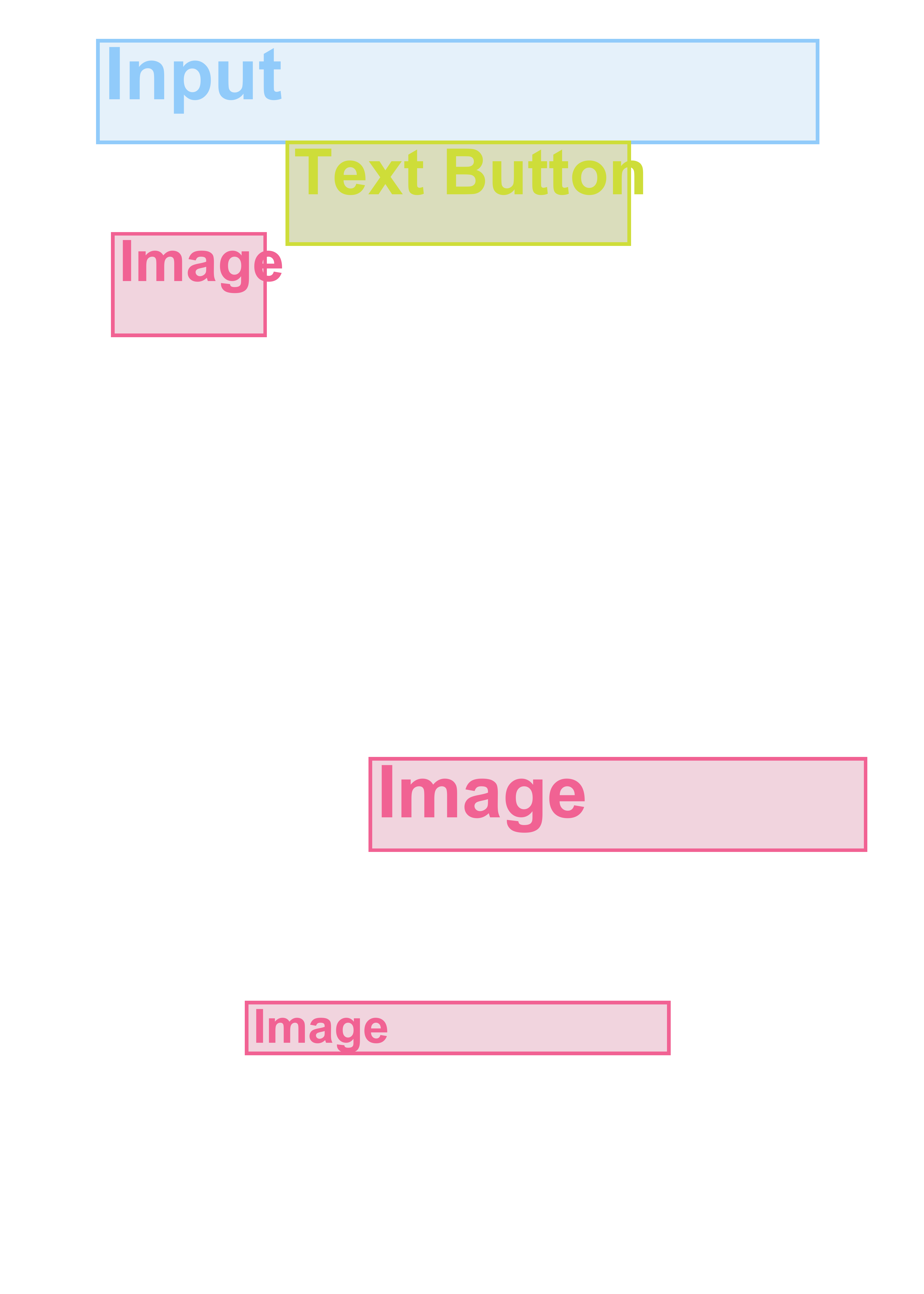} \\ 
&\includegraphics[width=\ricoBulkWidth]{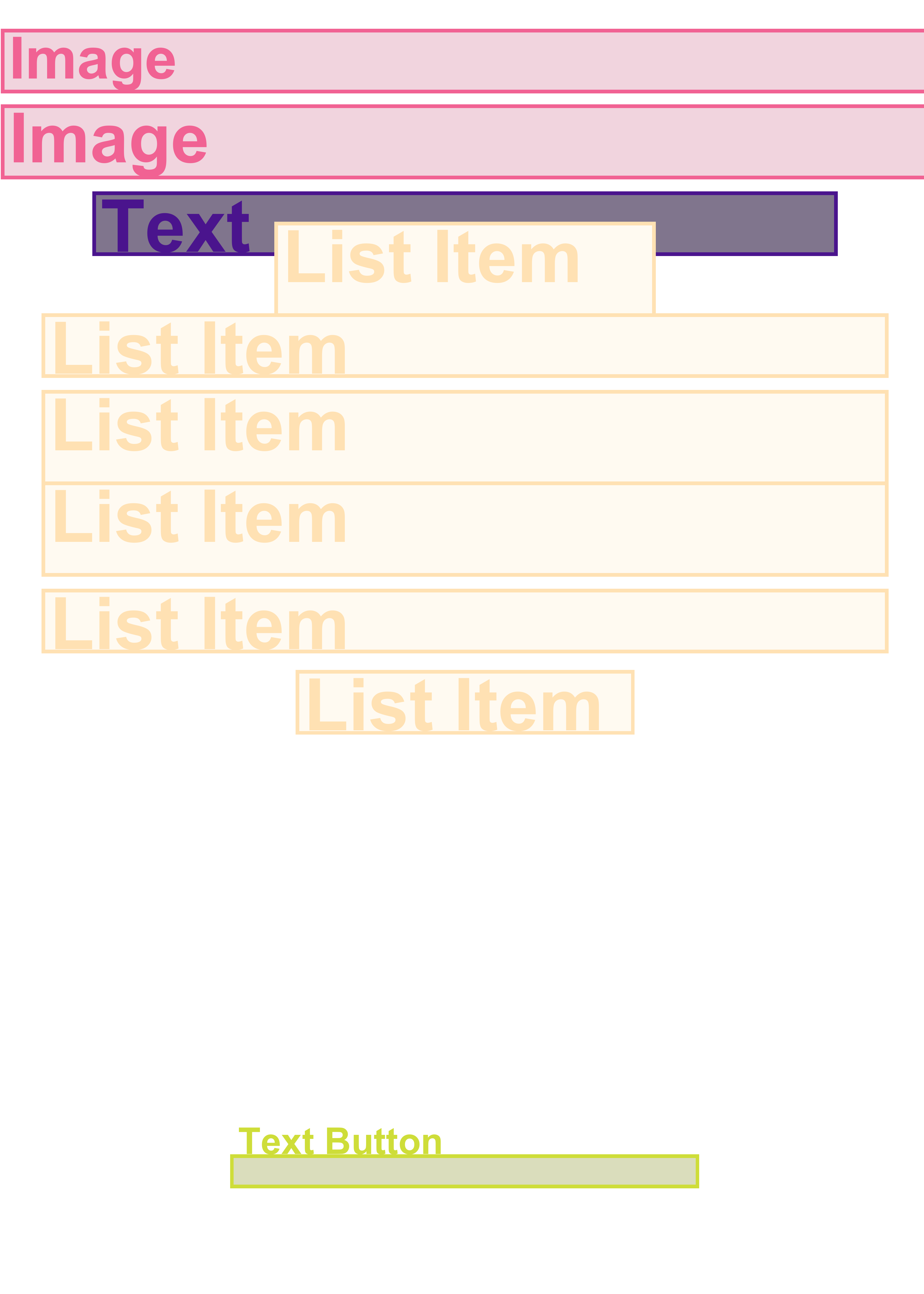} & 
\includegraphics[width=\ricoBulkWidth]{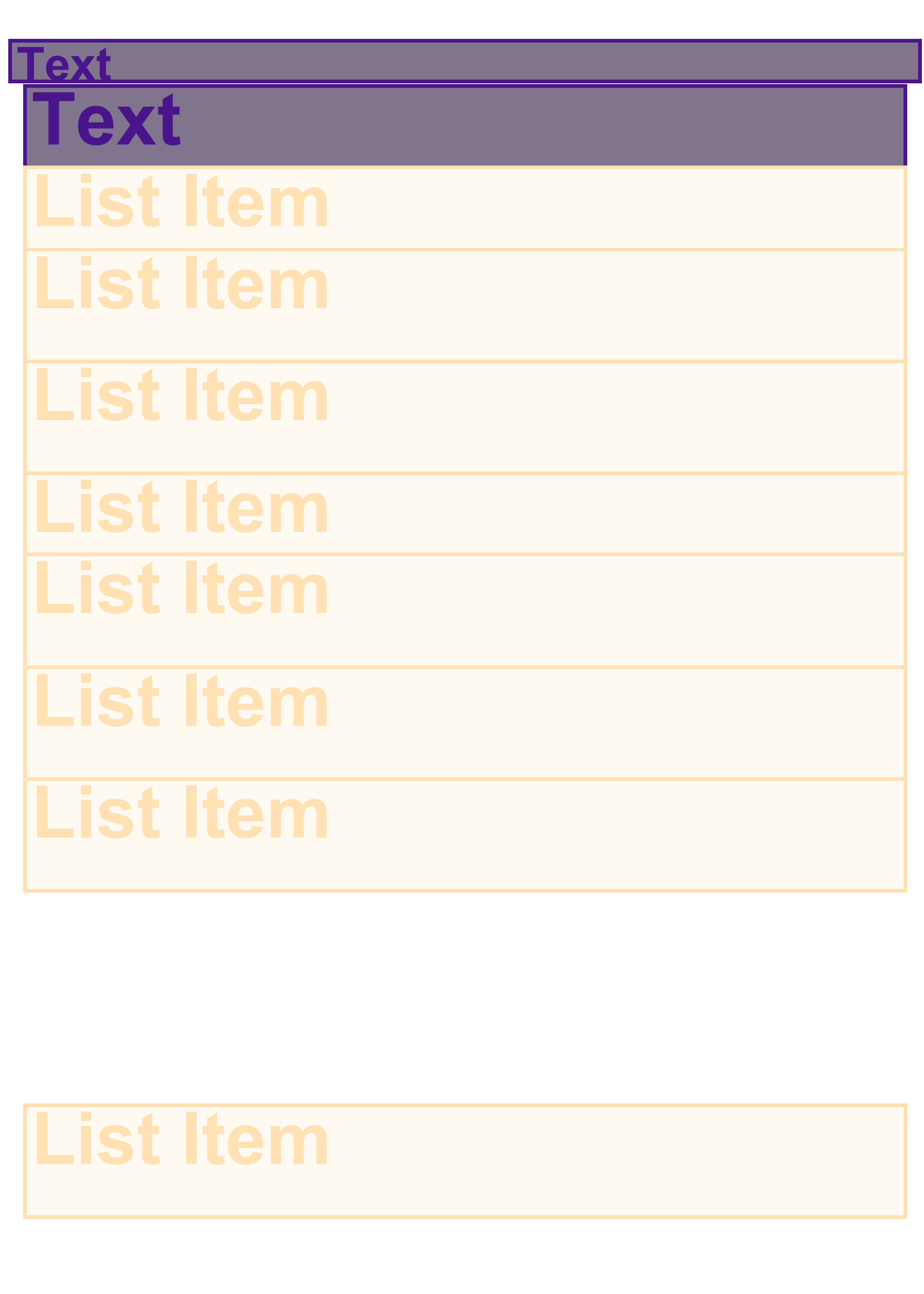} & 
\includegraphics[width=\ricoBulkWidth]{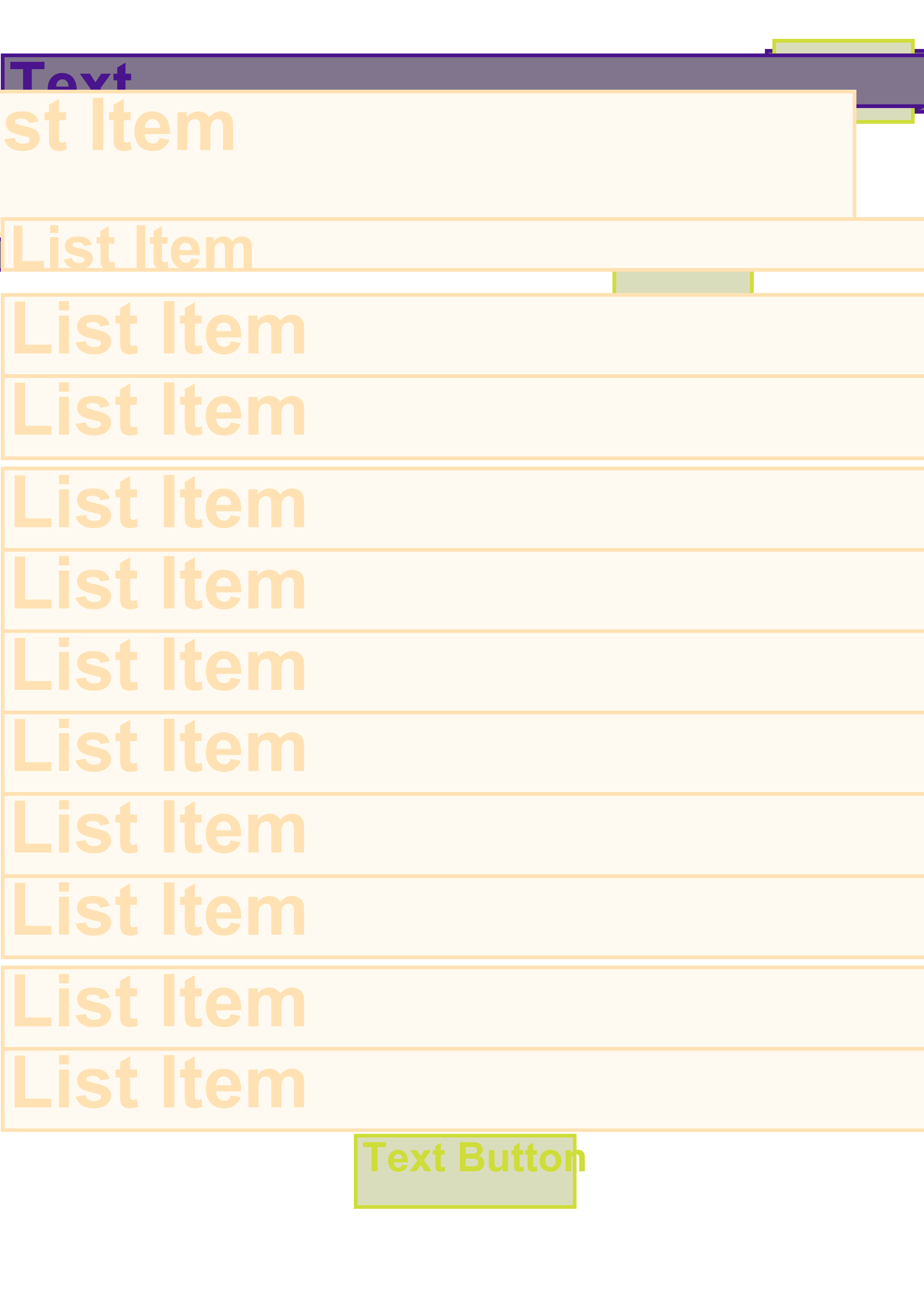} & 
\includegraphics[width=\ricoBulkWidth]{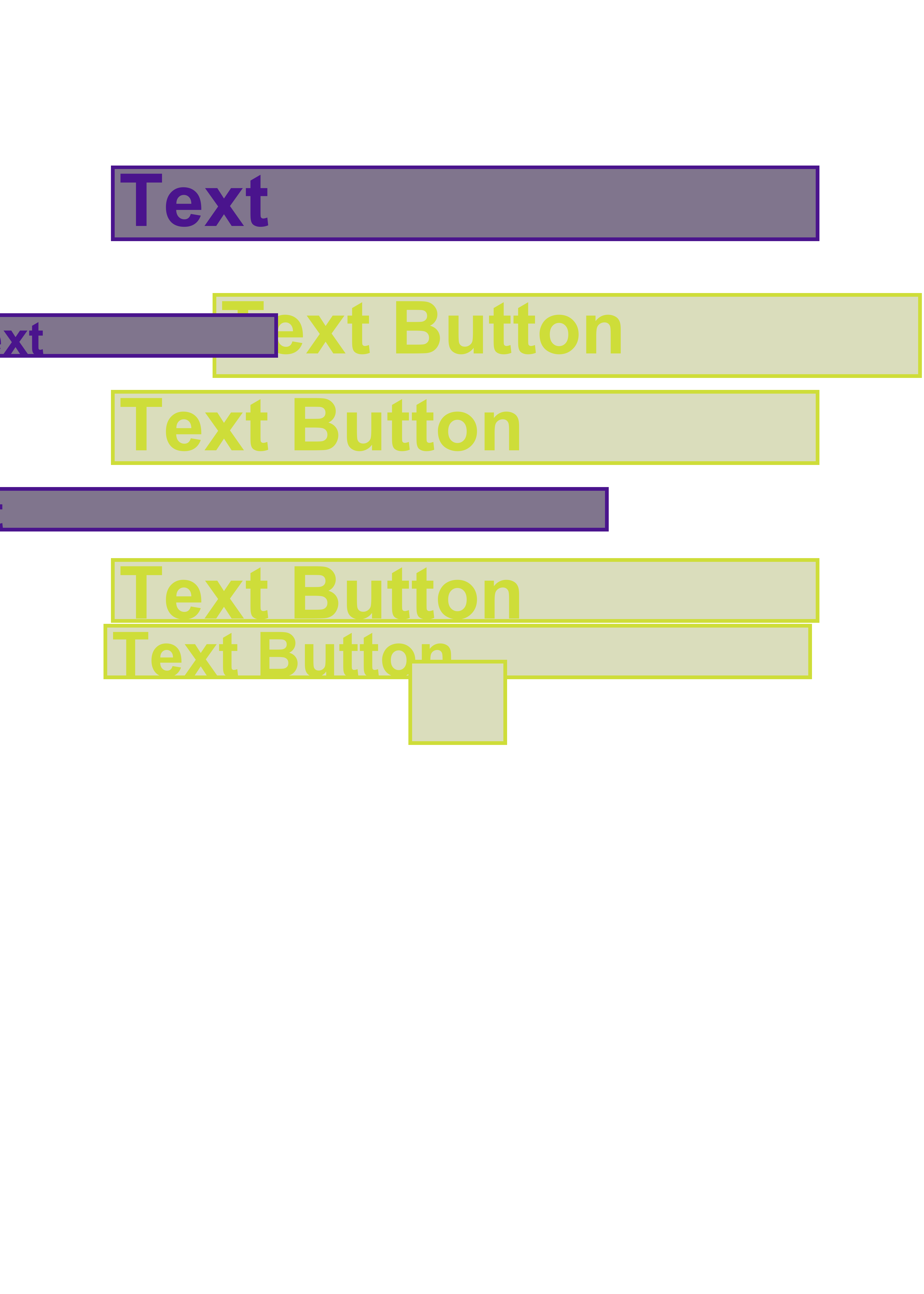} & 
\includegraphics[width=\ricoBulkWidth]{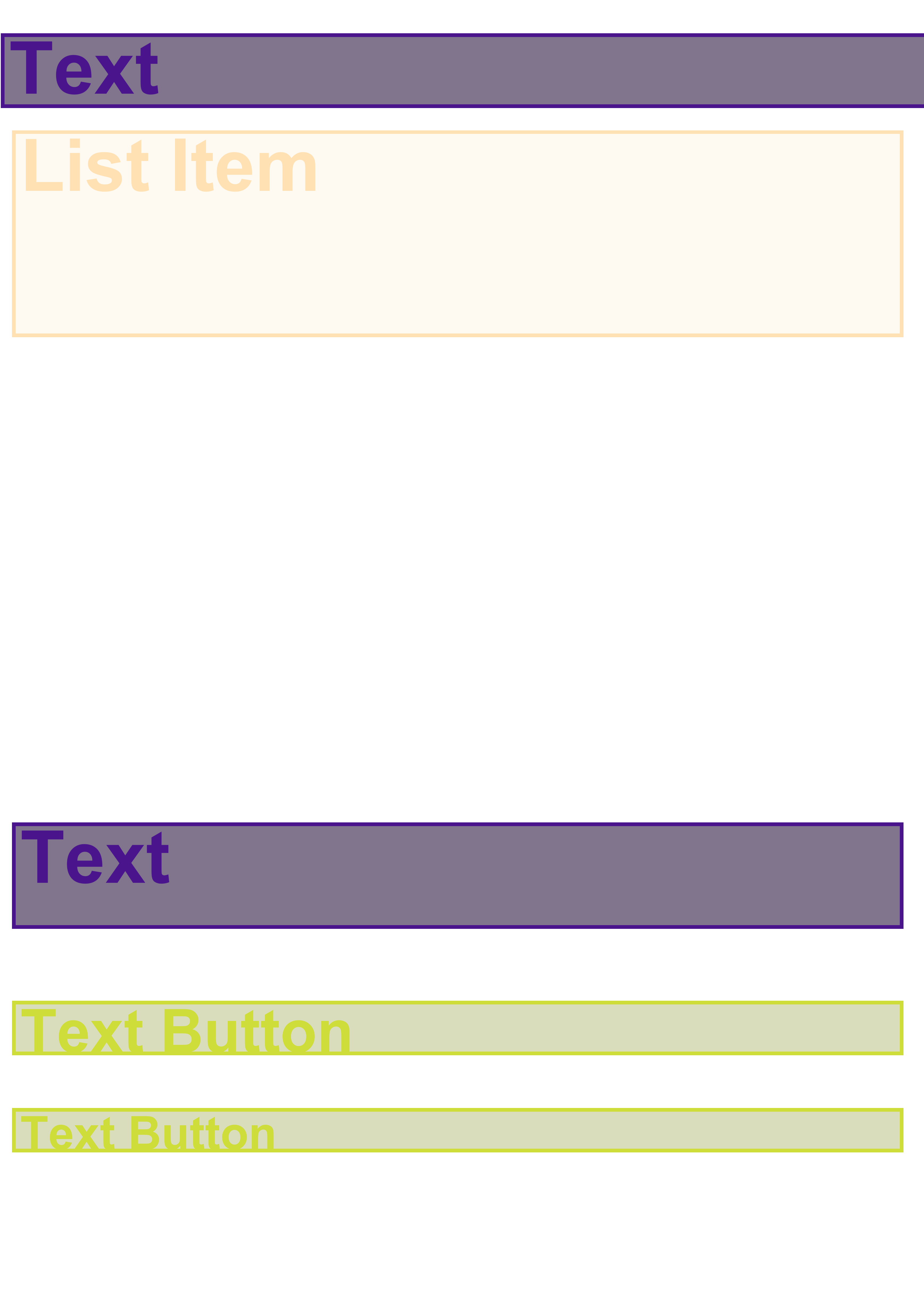} & 
\includegraphics[width=\ricoBulkWidth]{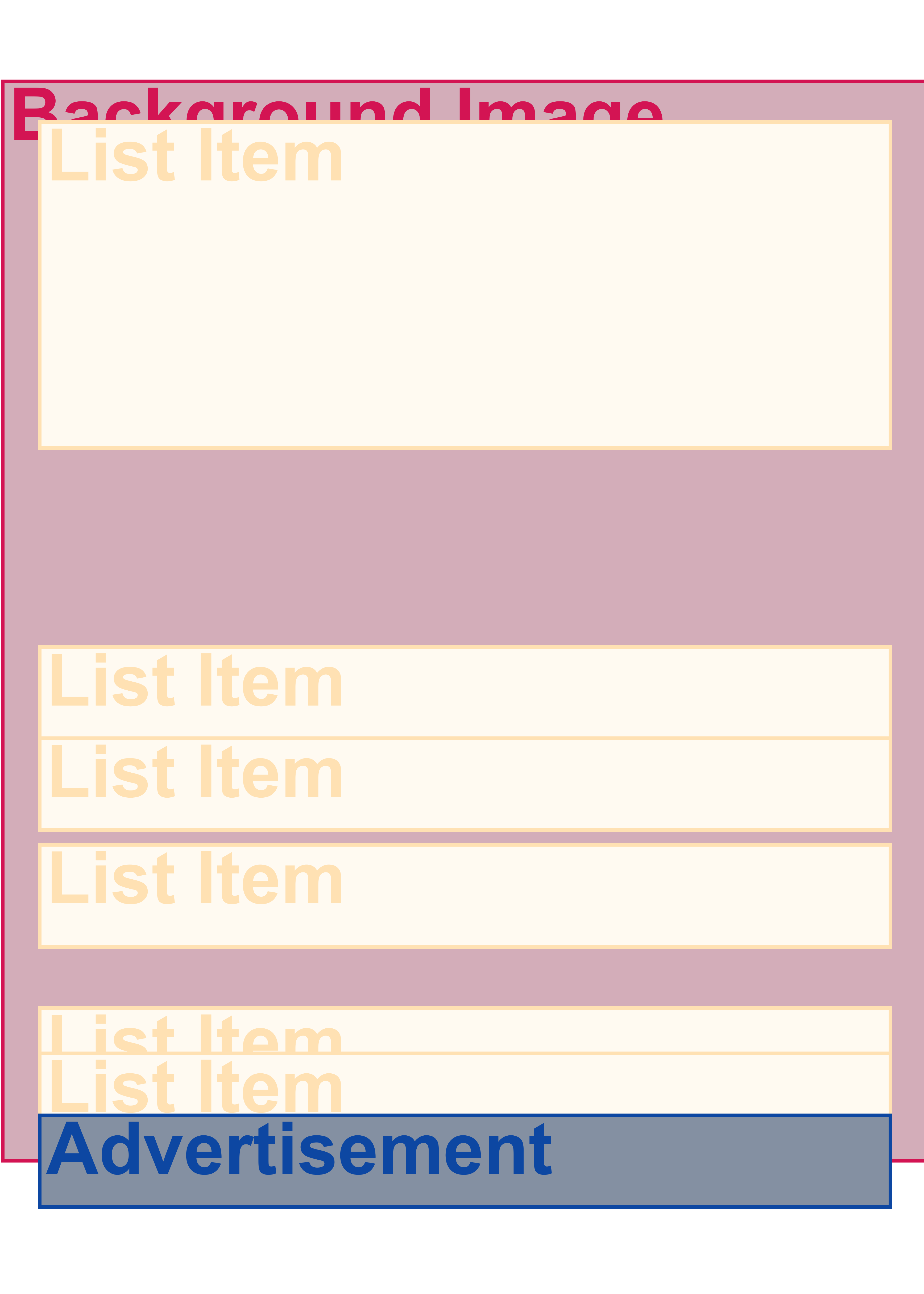} & 
\includegraphics[width=\ricoBulkWidth]{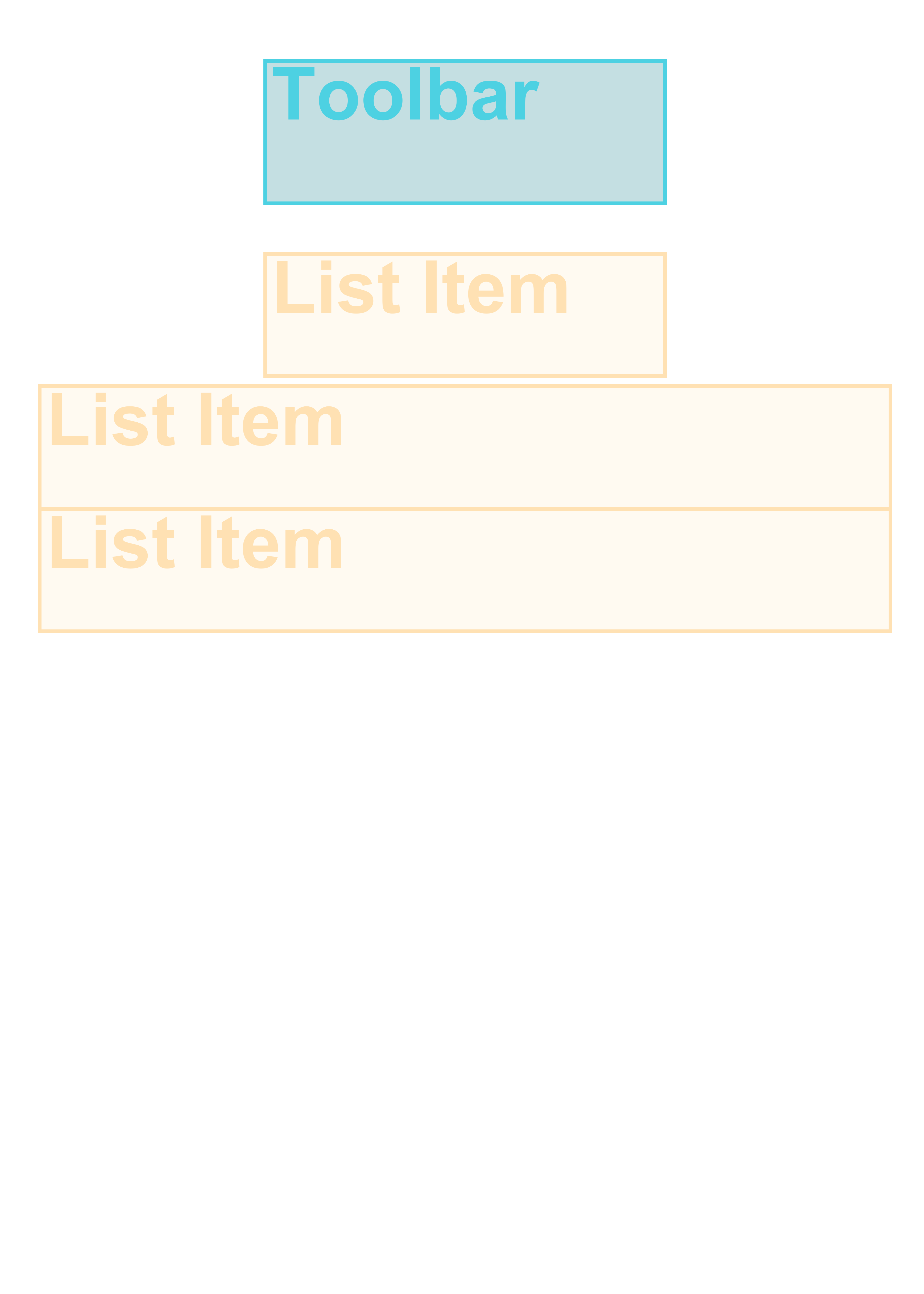} \\ 
&\includegraphics[width=\ricoBulkWidth]{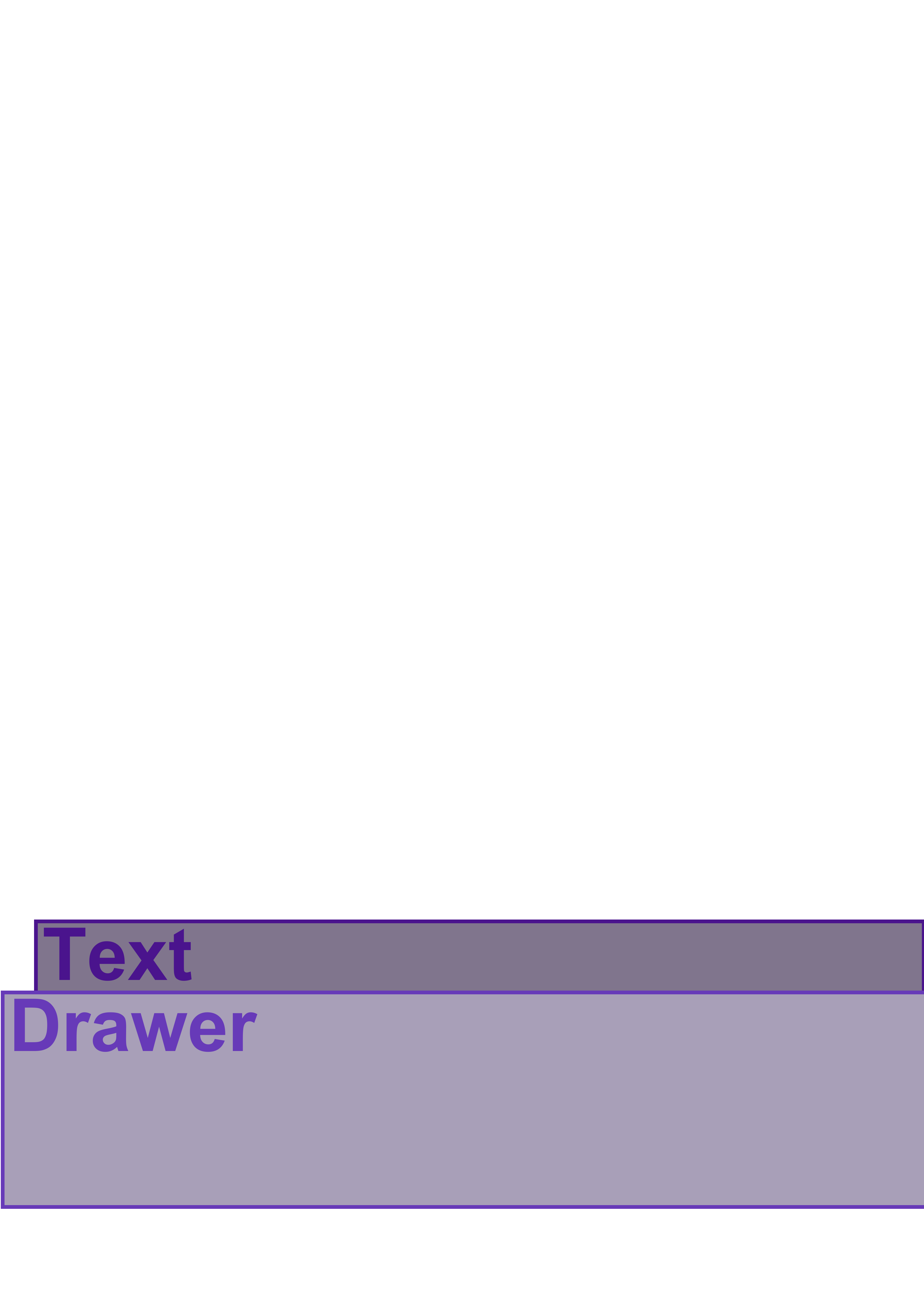} & 
\includegraphics[width=\ricoBulkWidth]{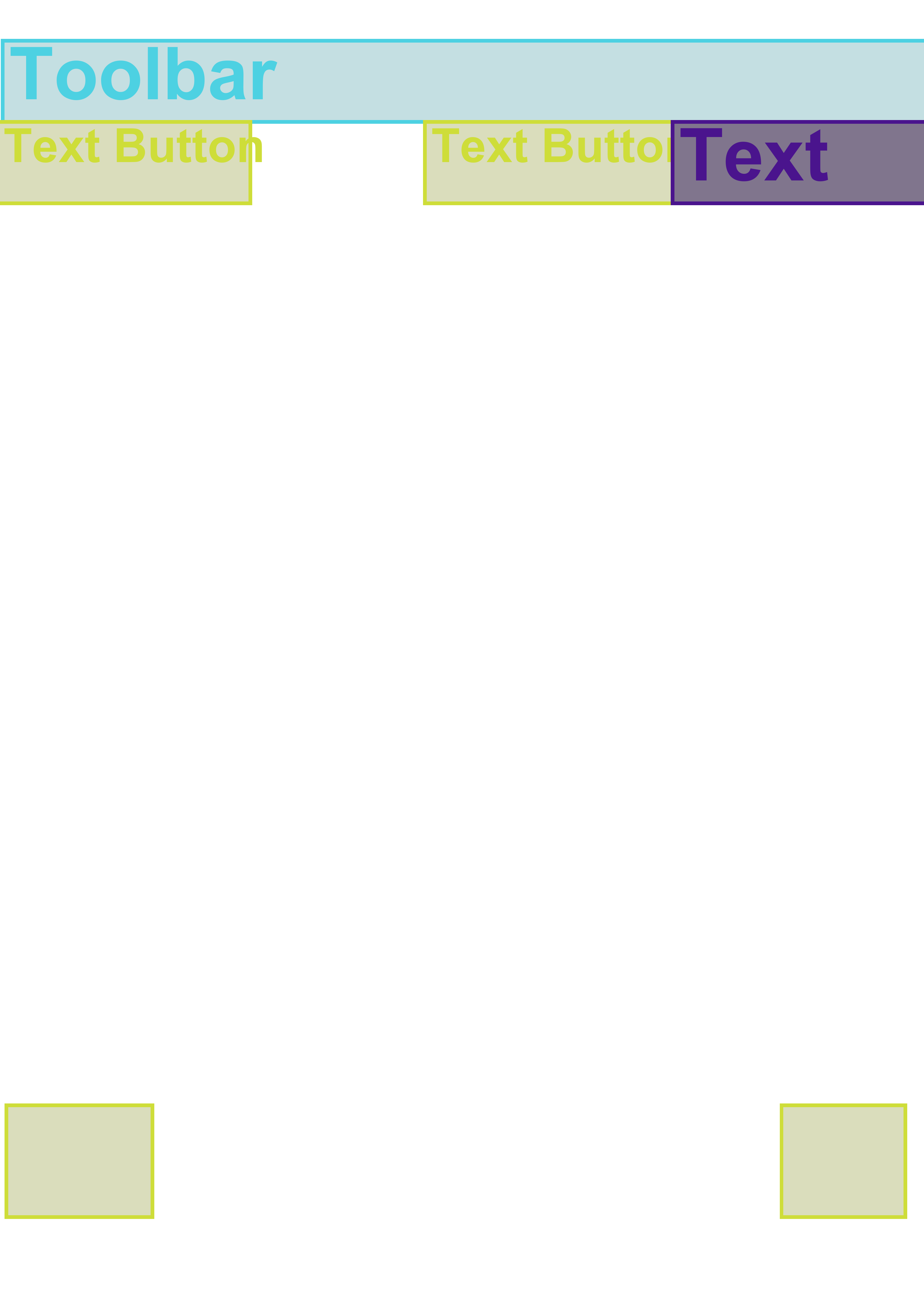} & 
\includegraphics[width=\ricoBulkWidth]{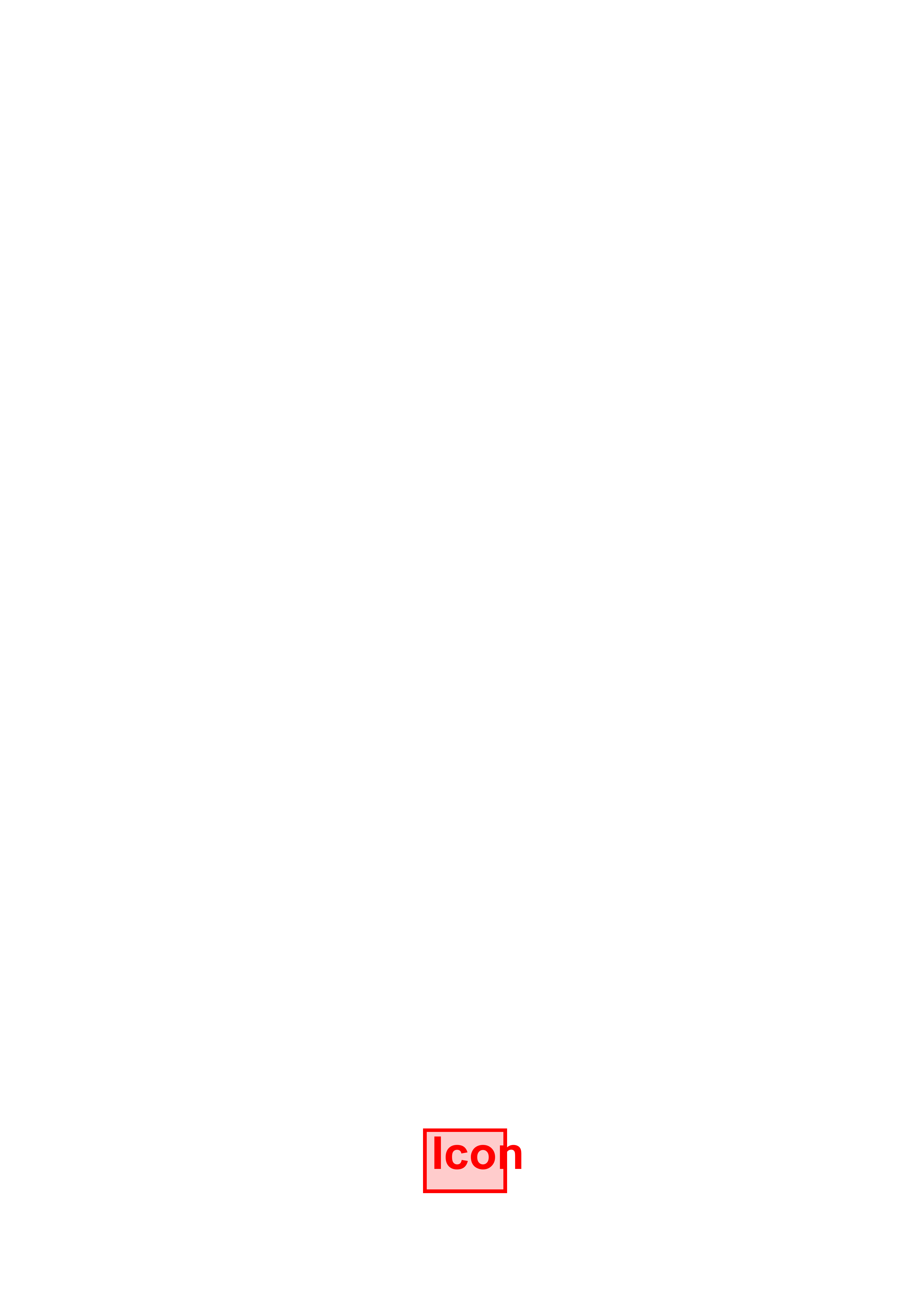} & 
\includegraphics[width=\ricoBulkWidth]{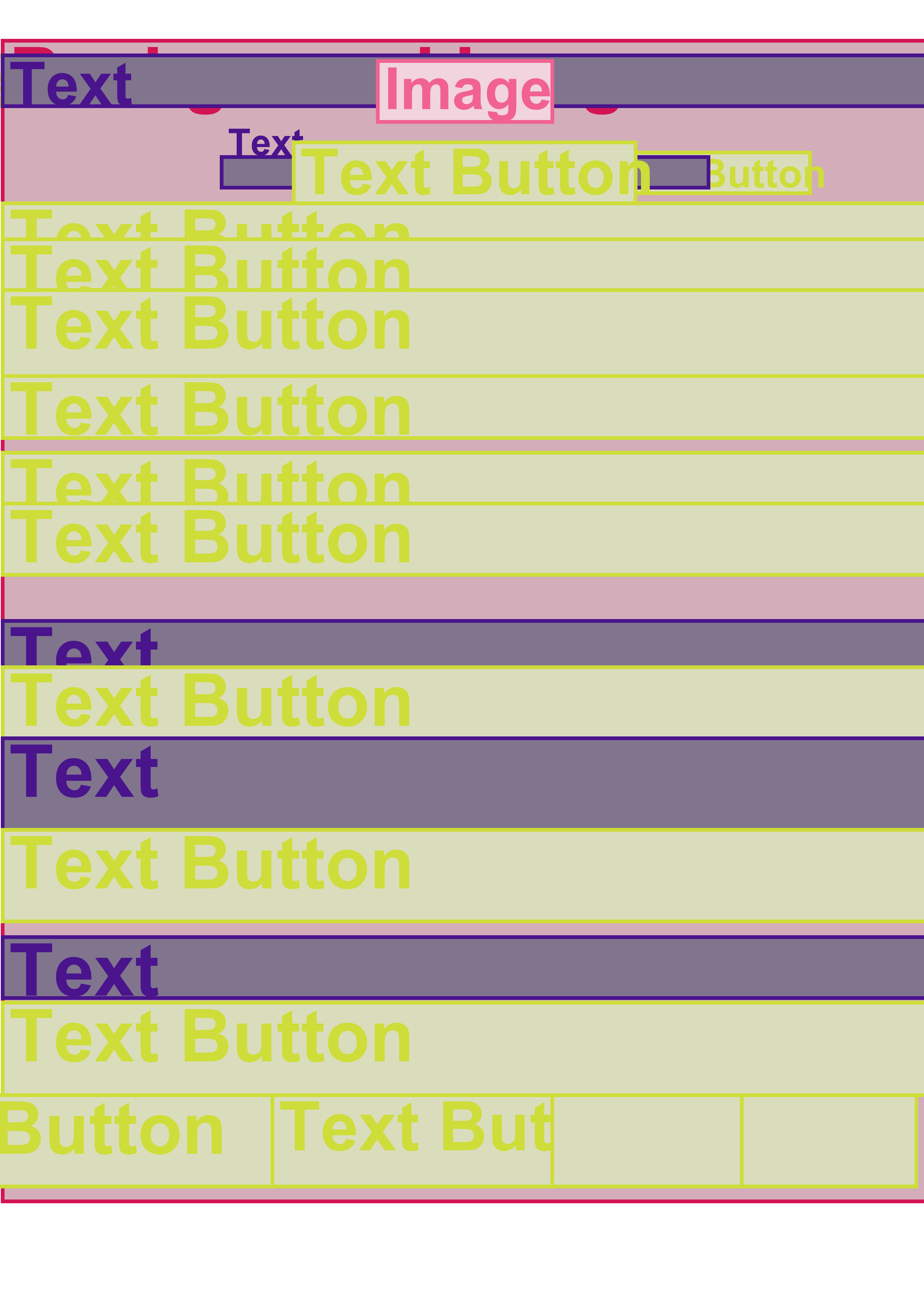} & 
\includegraphics[width=\ricoBulkWidth]{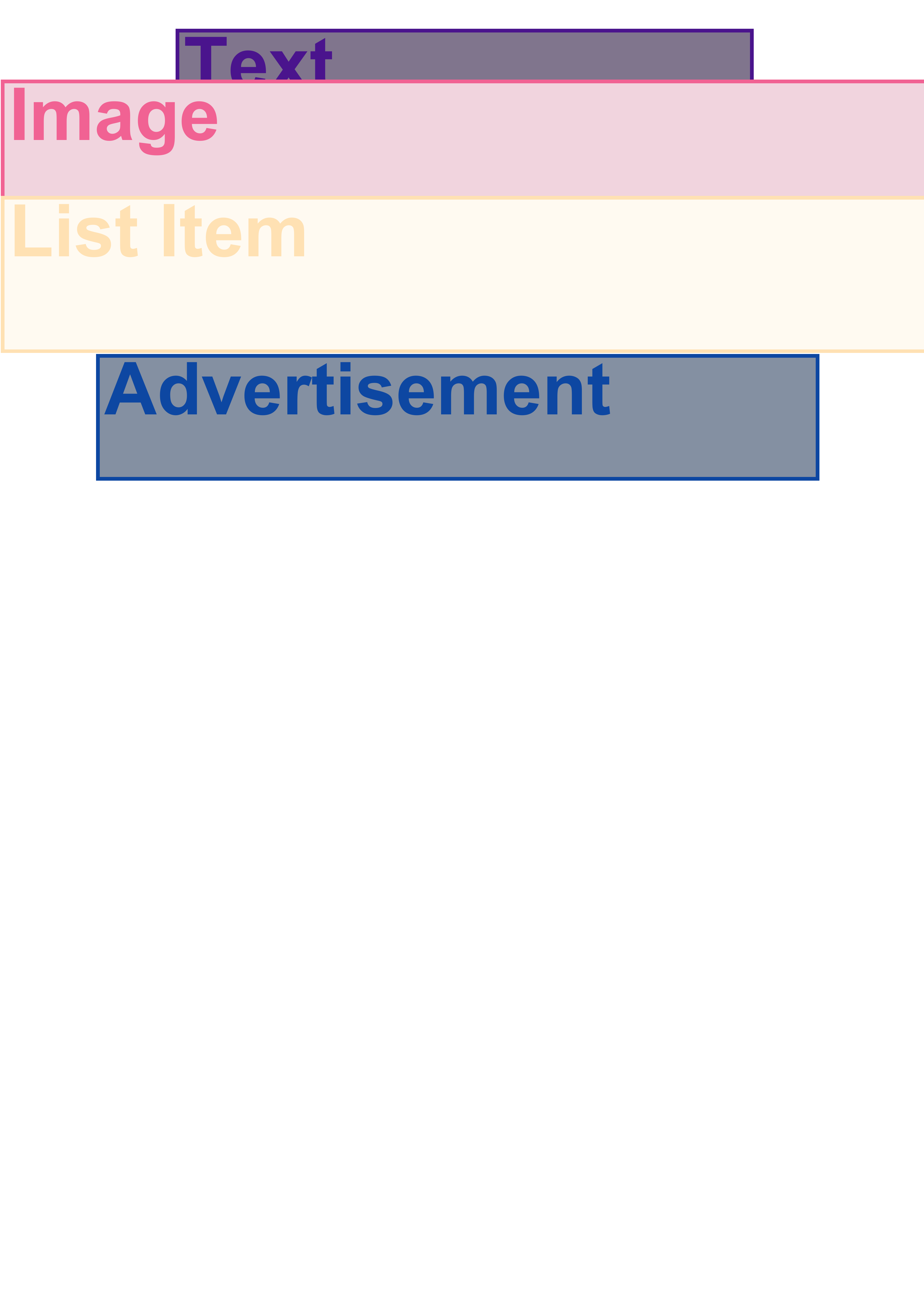} & 
\includegraphics[width=\ricoBulkWidth]{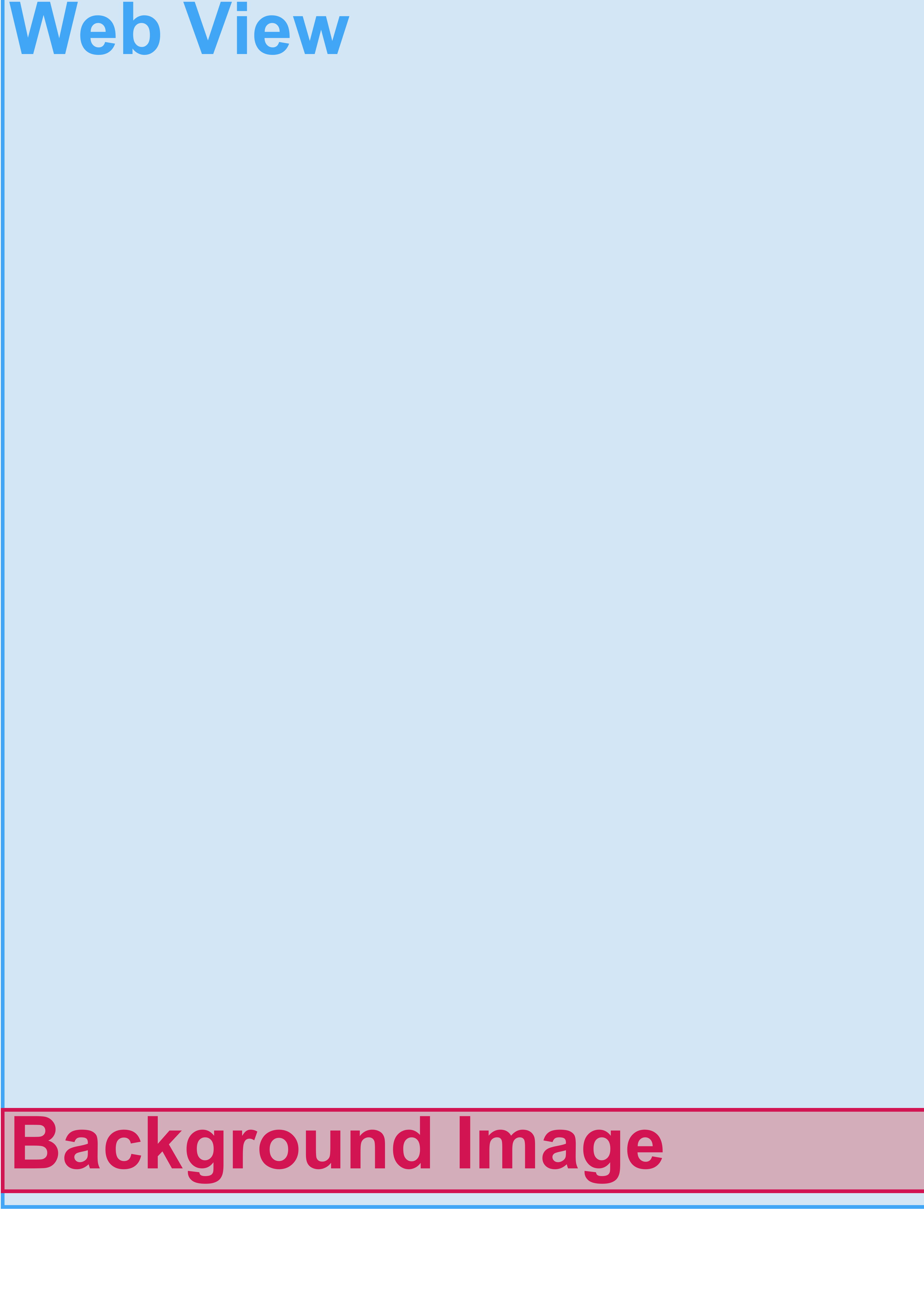} & 
\includegraphics[width=\ricoBulkWidth]{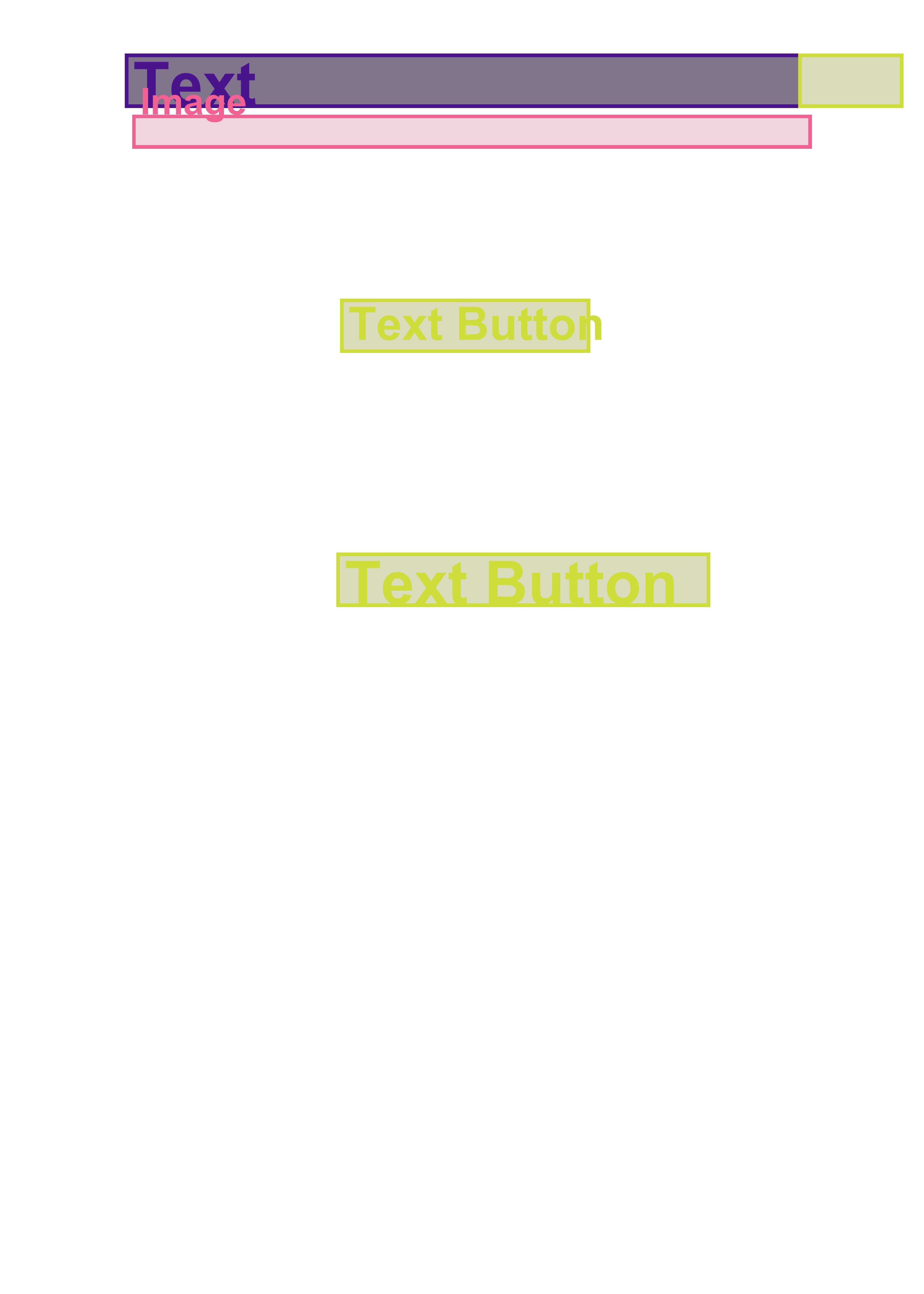} \\ 
    \end{tabular}
    \caption{Synthesized RICO examples.}
    \label{fig:bulk_rico}
\end{figure}

\begin{figure}\ContinuedFloat
    \setlength{\ricoBulkWidth}{0.12\linewidth}
    \centering
    \begin{tabular}{ccccccc}
    \includegraphics[width=\ricoBulkWidth]{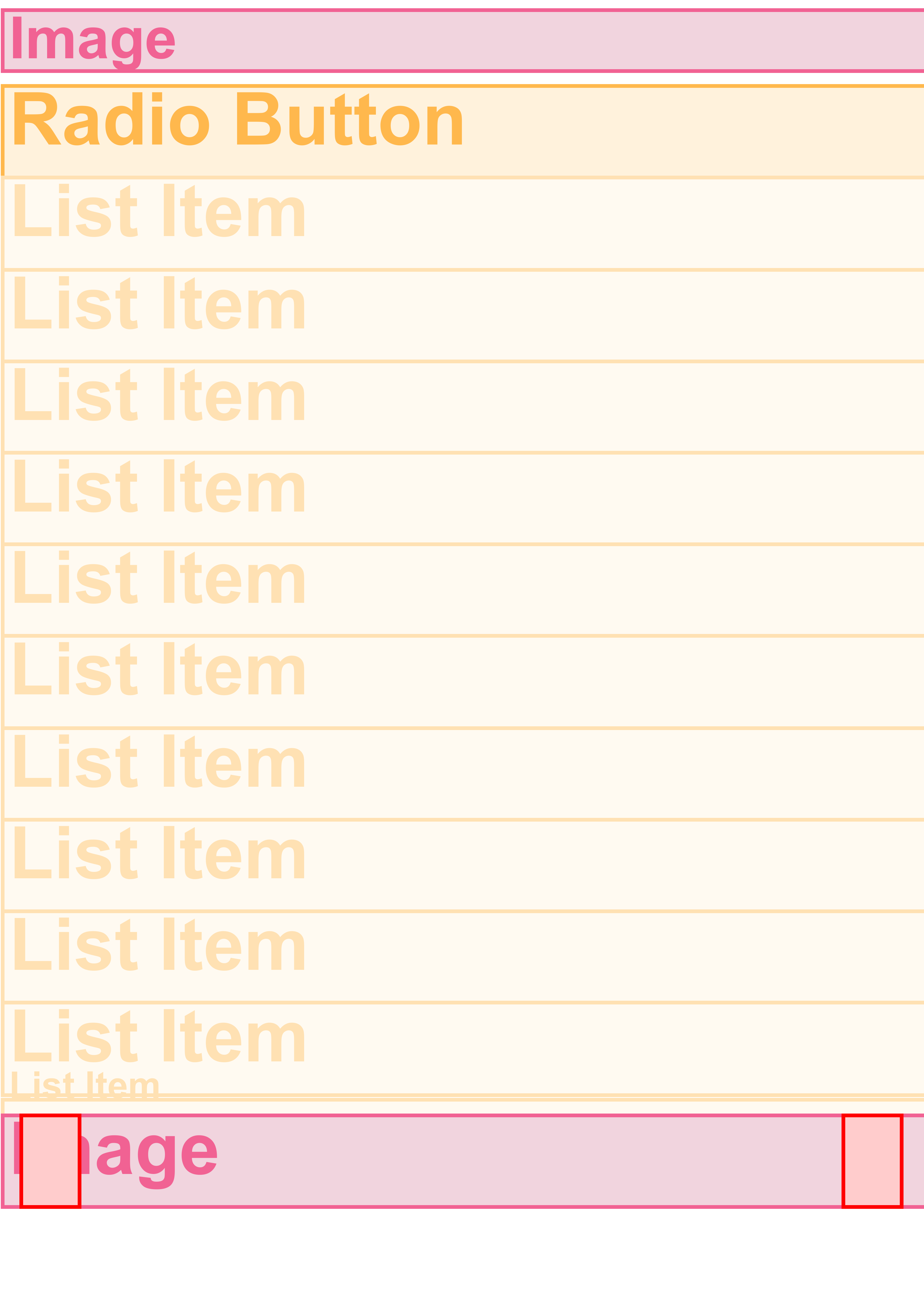} & 
    \includegraphics[width=\ricoBulkWidth]{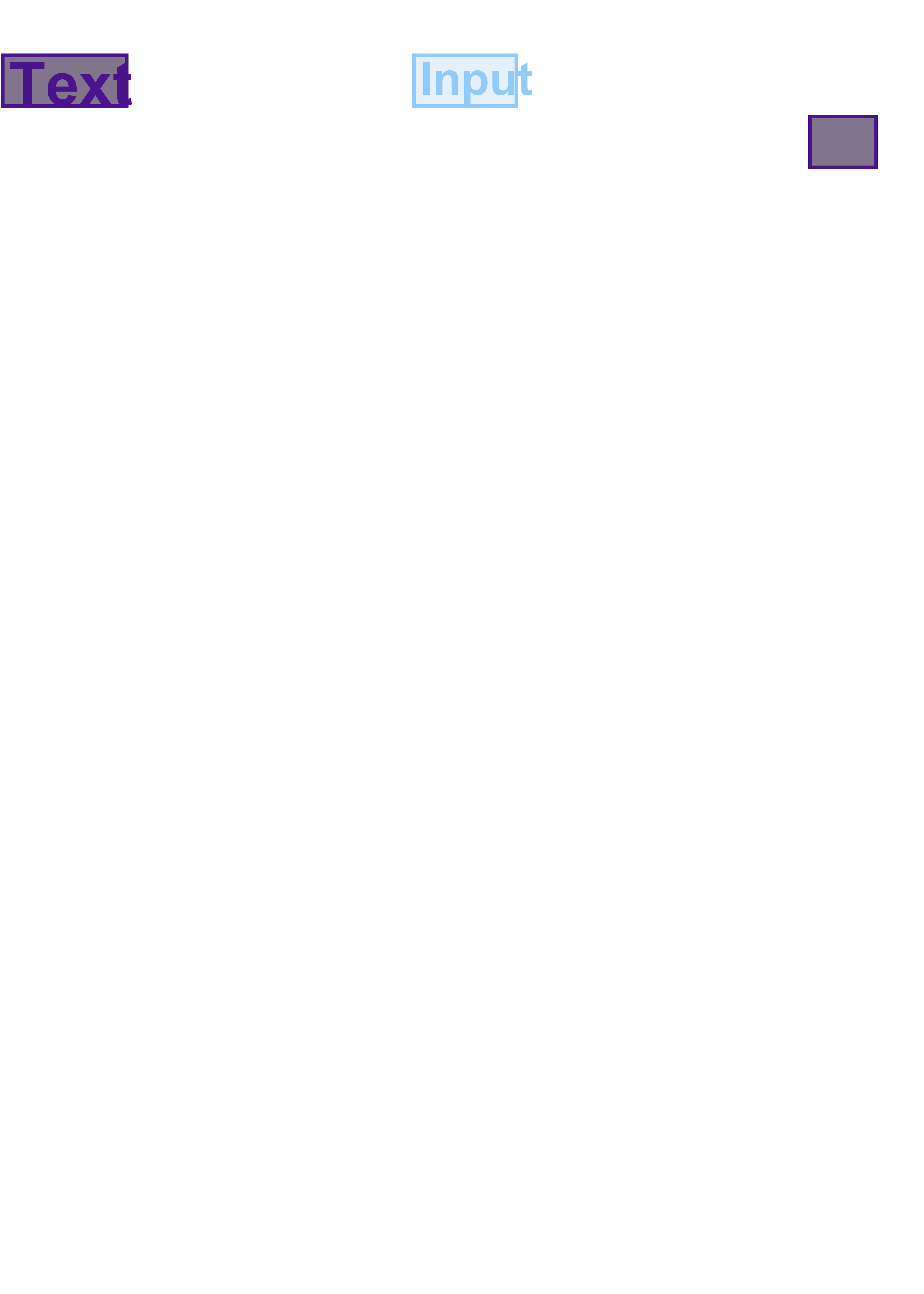} & 
    \includegraphics[width=\ricoBulkWidth]{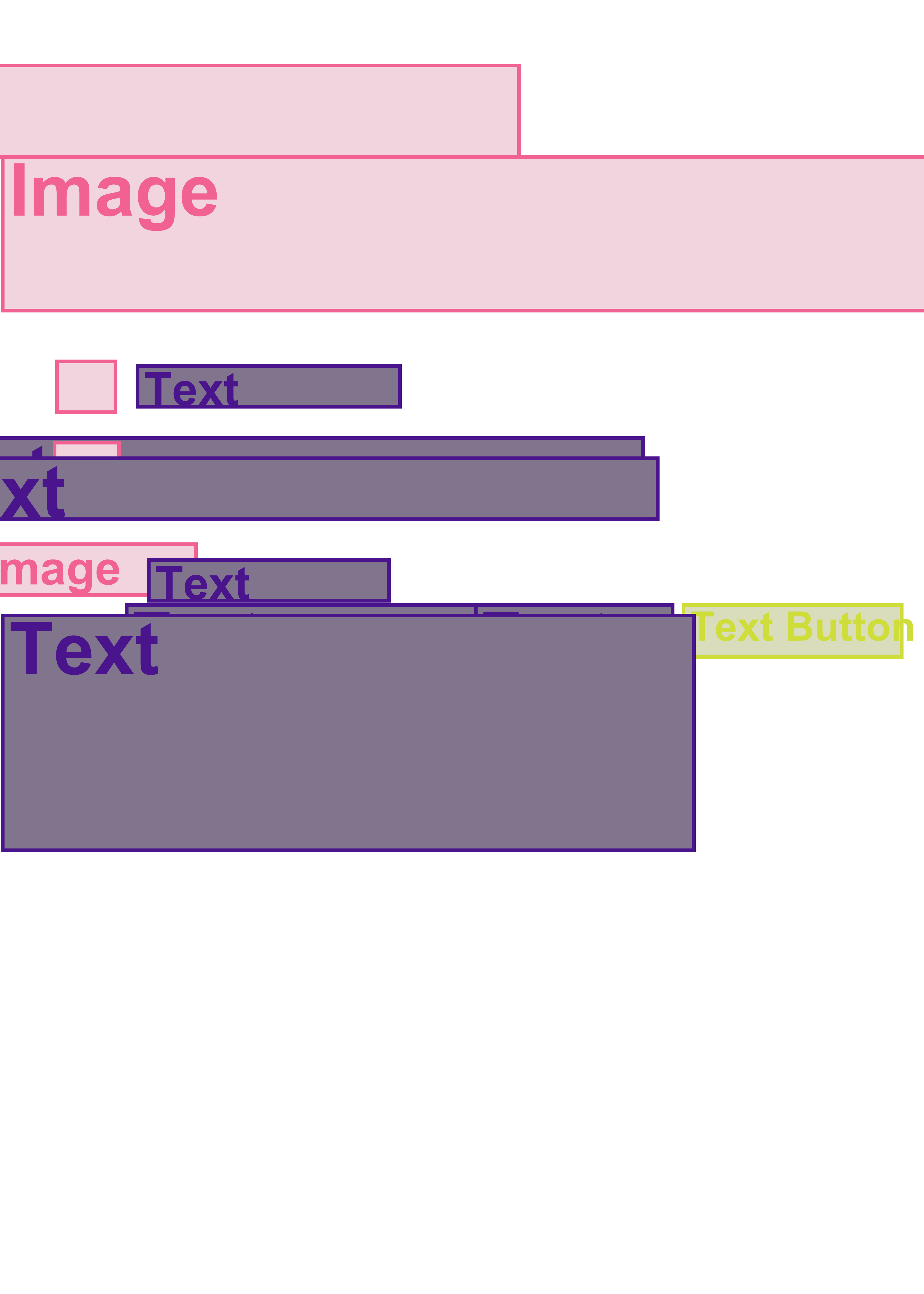} & 
    \includegraphics[width=\ricoBulkWidth]{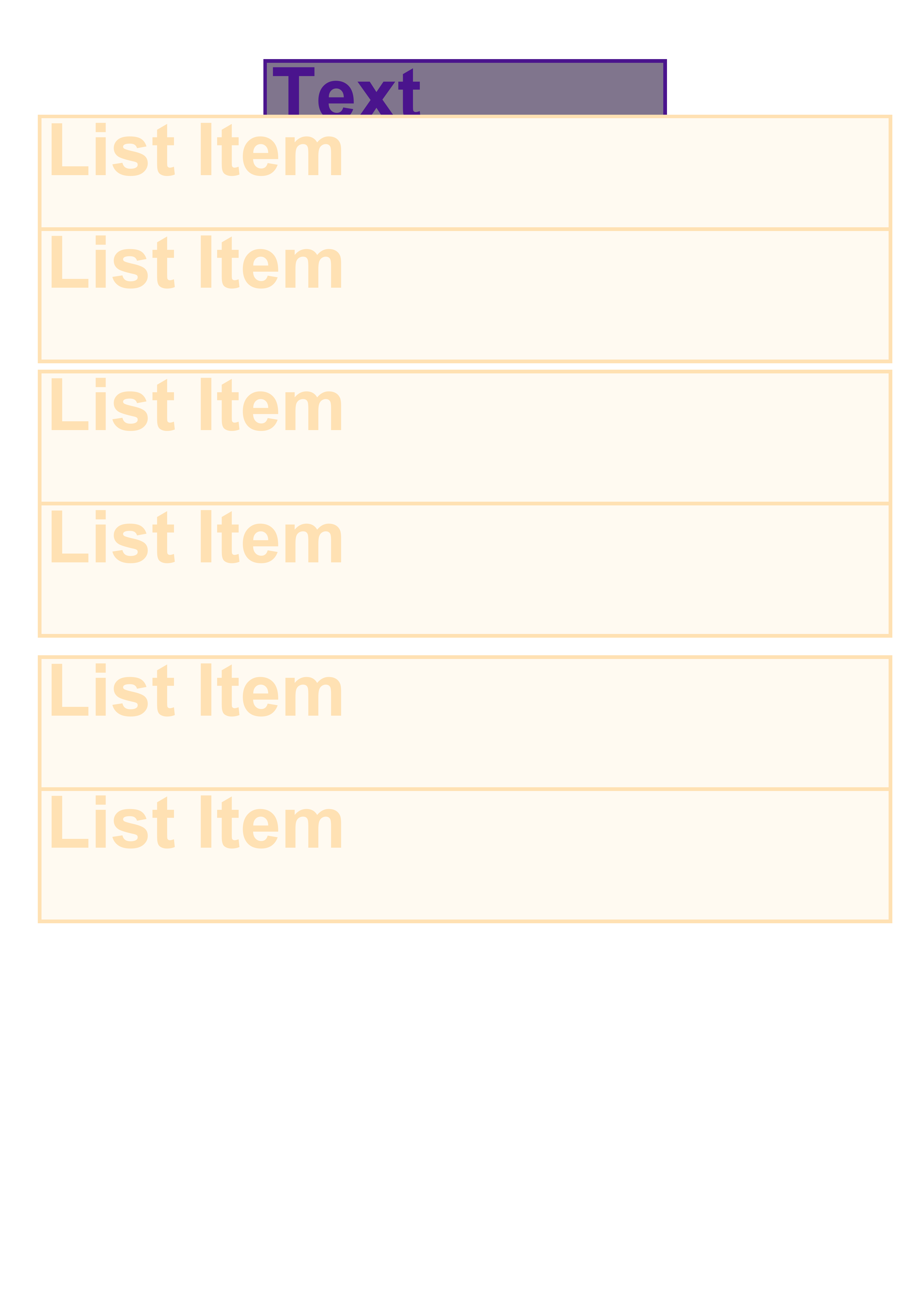} & 
    \includegraphics[width=\ricoBulkWidth]{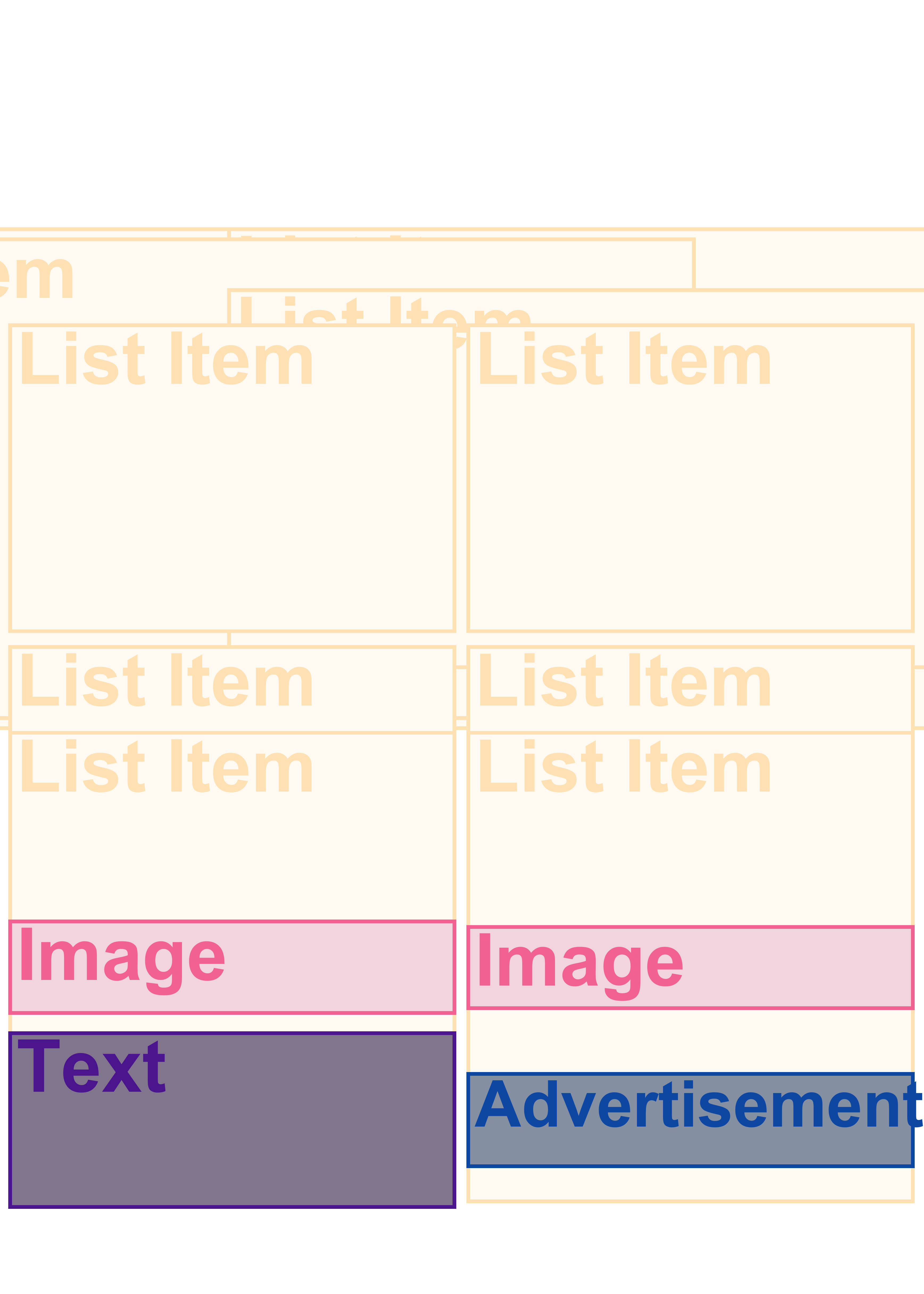} & 
    \includegraphics[width=\ricoBulkWidth]{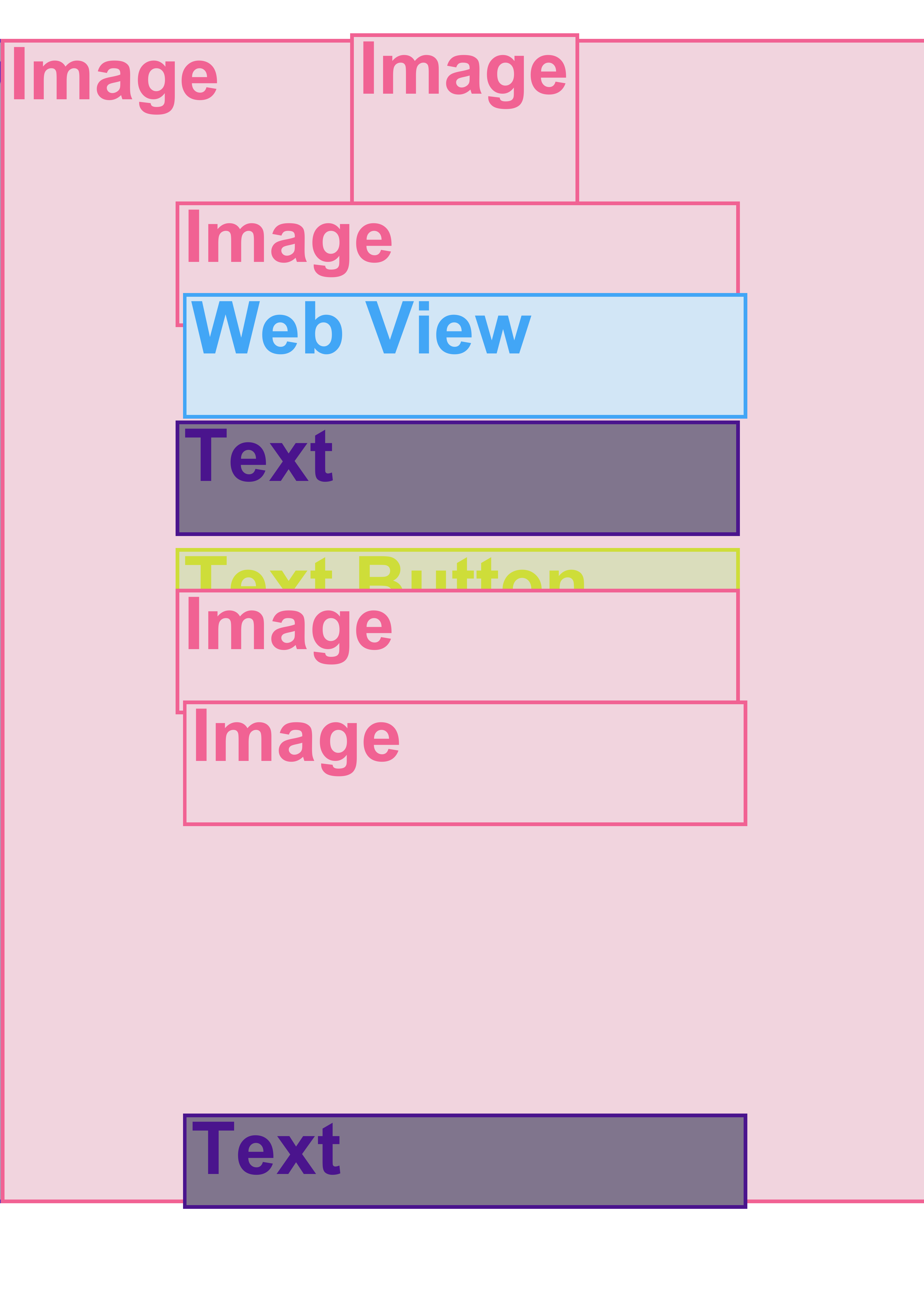} & 
    \includegraphics[width=\ricoBulkWidth]{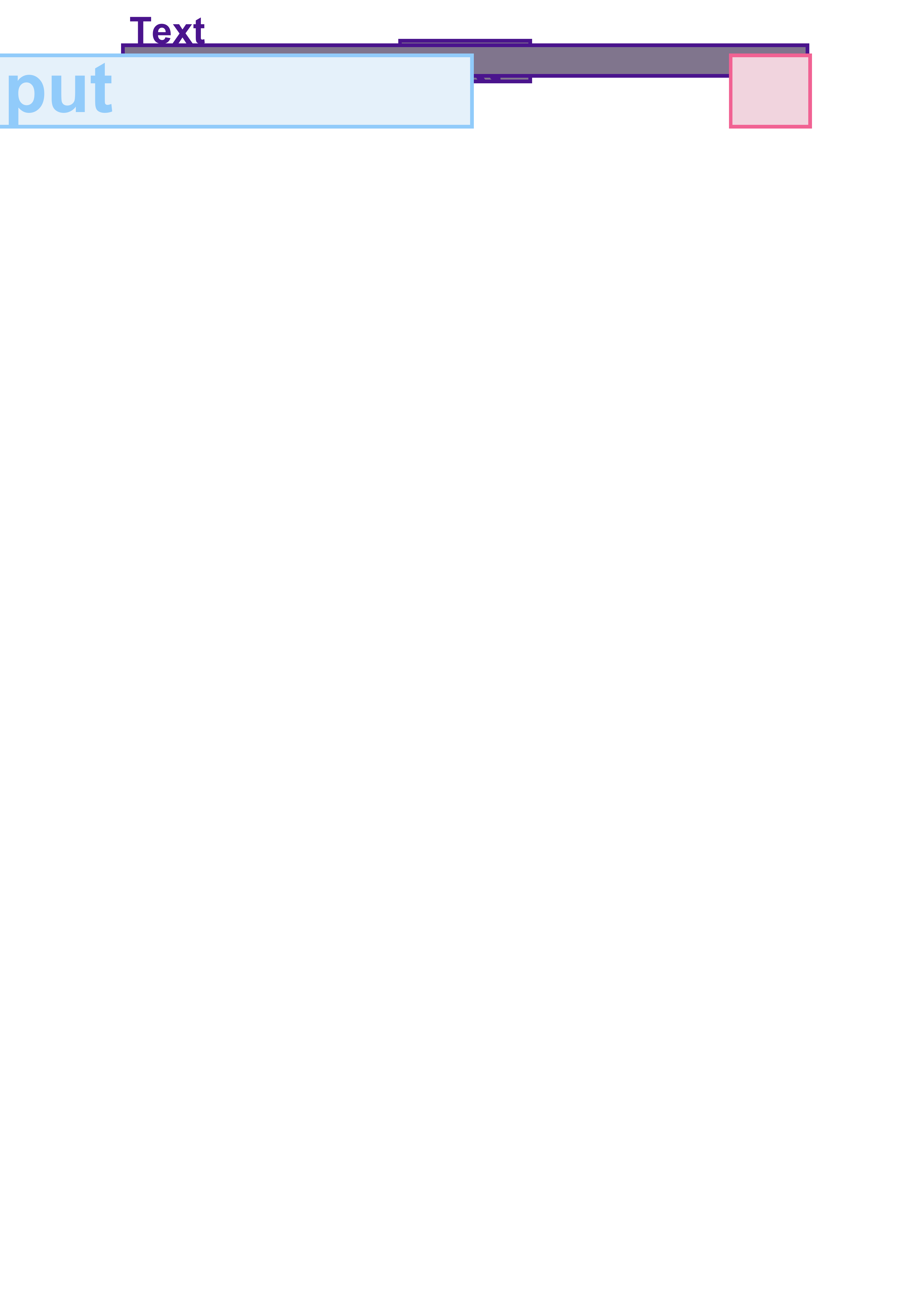} \\ 
    \includegraphics[width=\ricoBulkWidth]{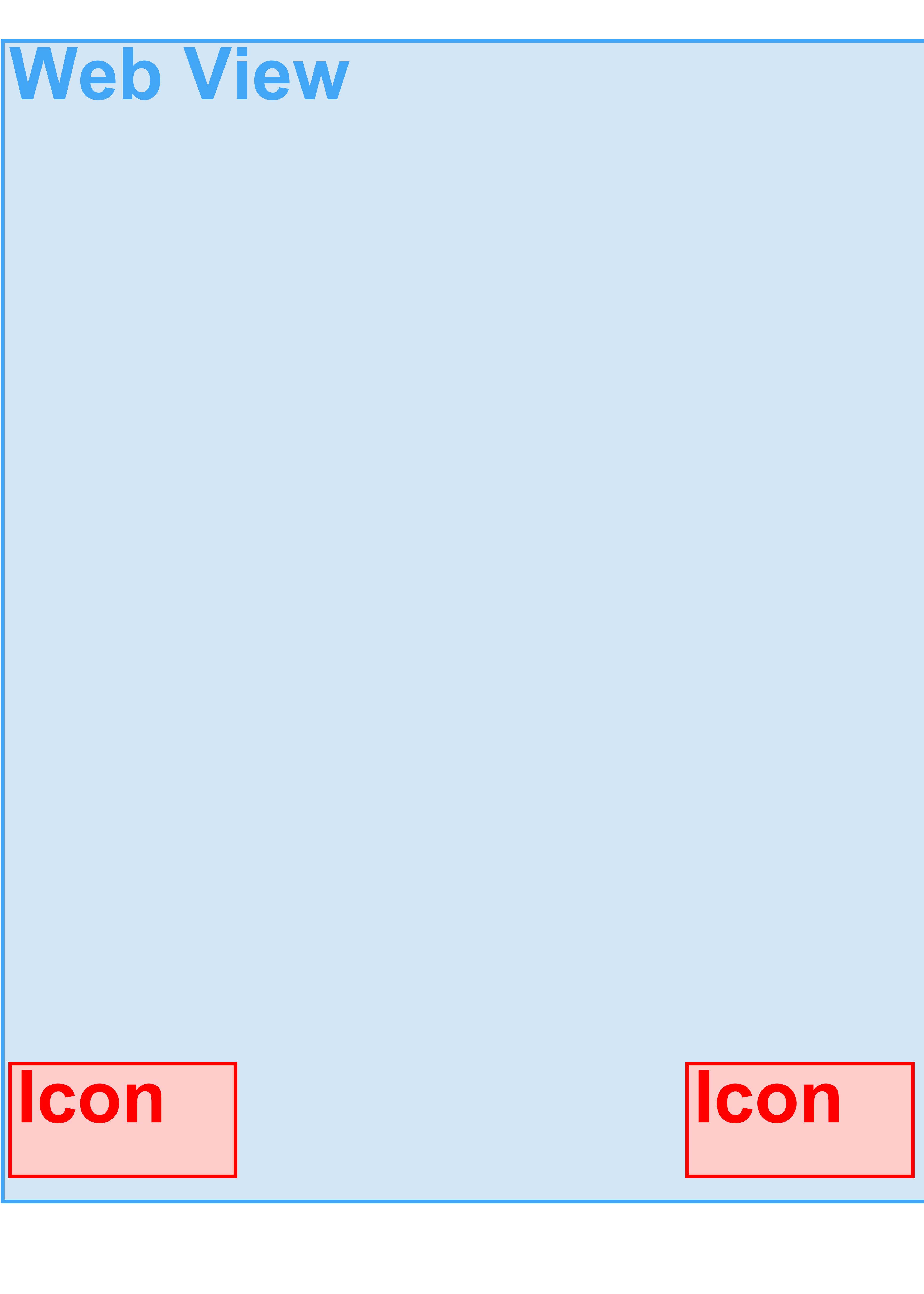} & 
    \includegraphics[width=\ricoBulkWidth]{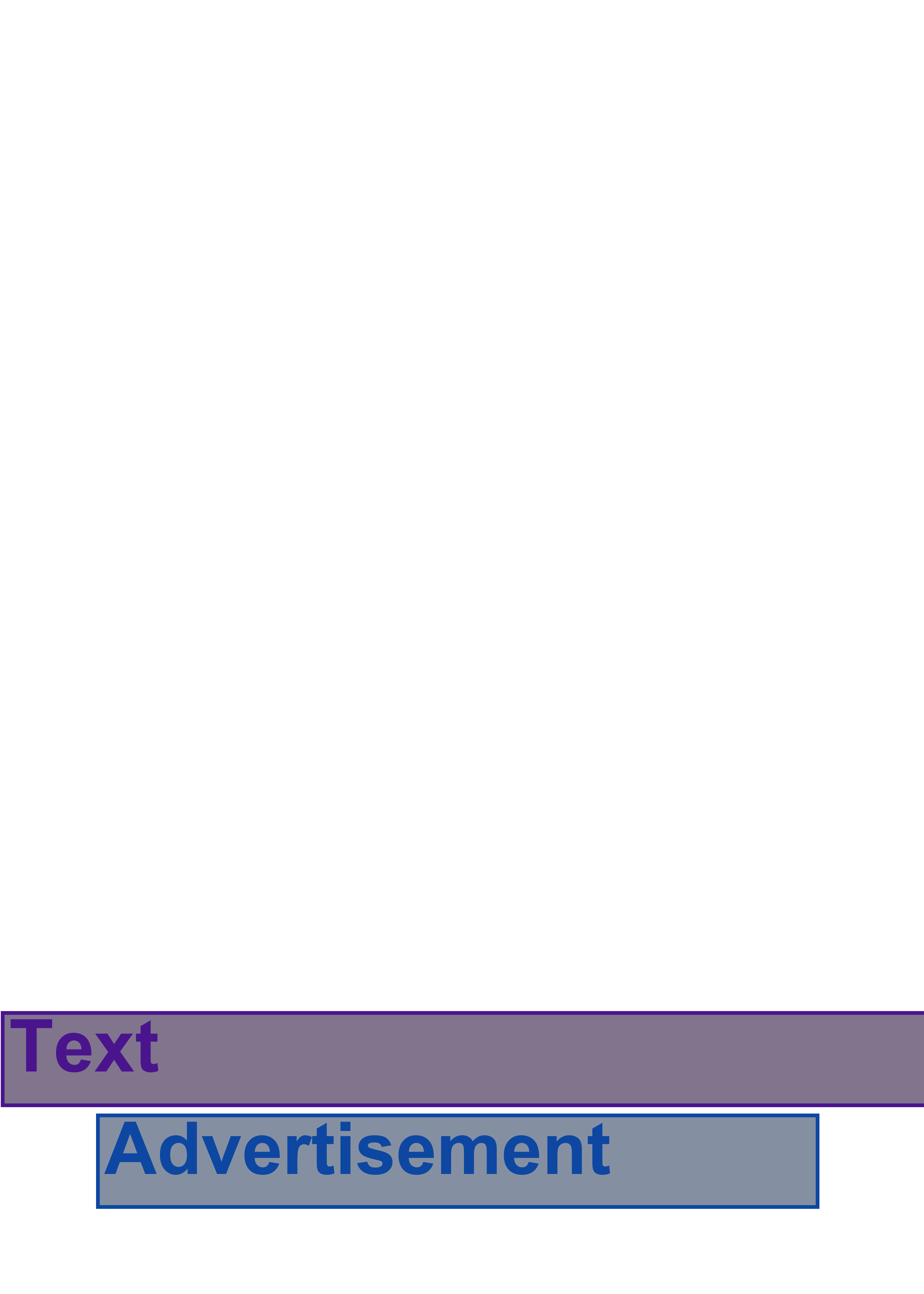} & 
    \includegraphics[width=\ricoBulkWidth]{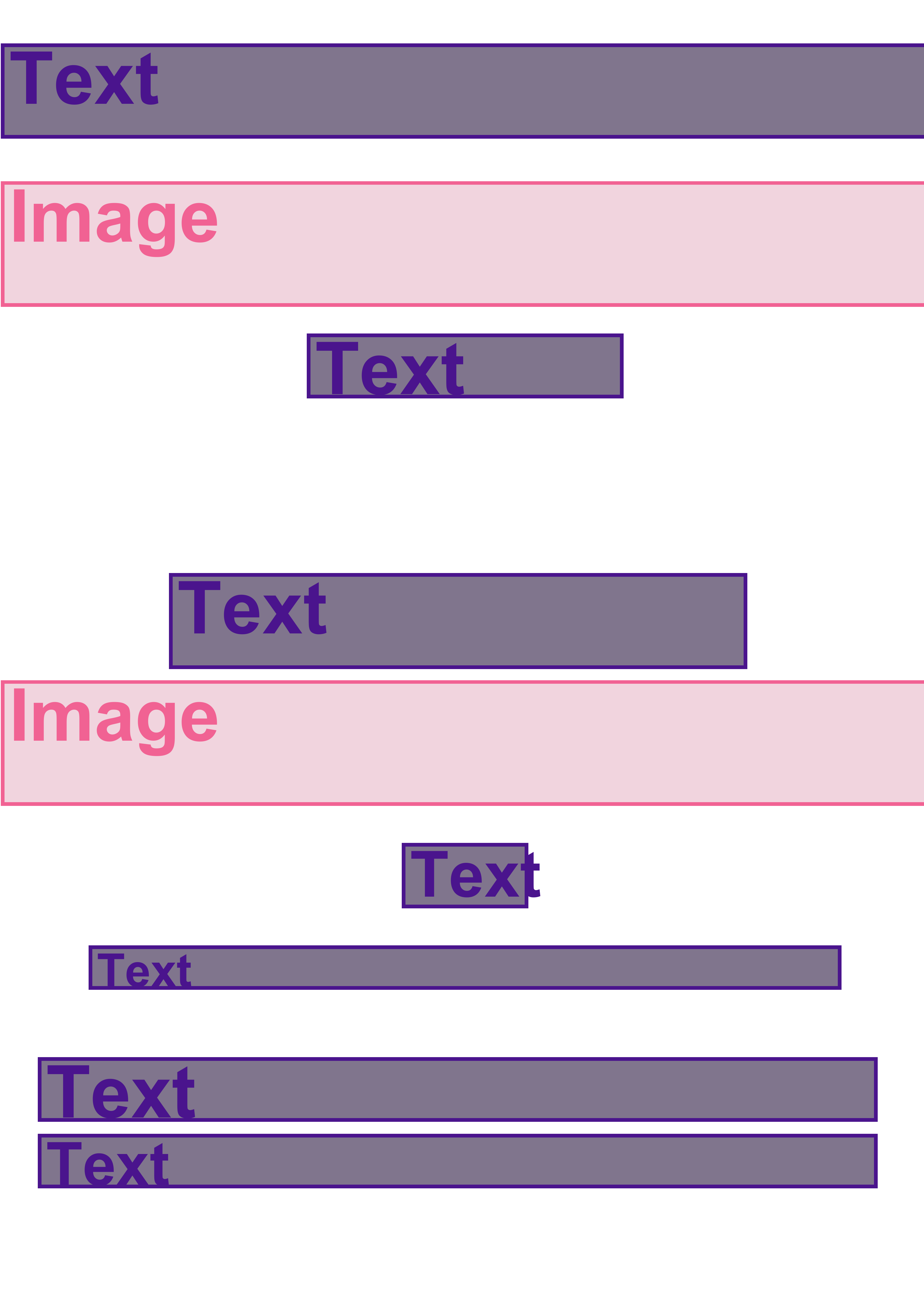} & 
    \includegraphics[width=\ricoBulkWidth]{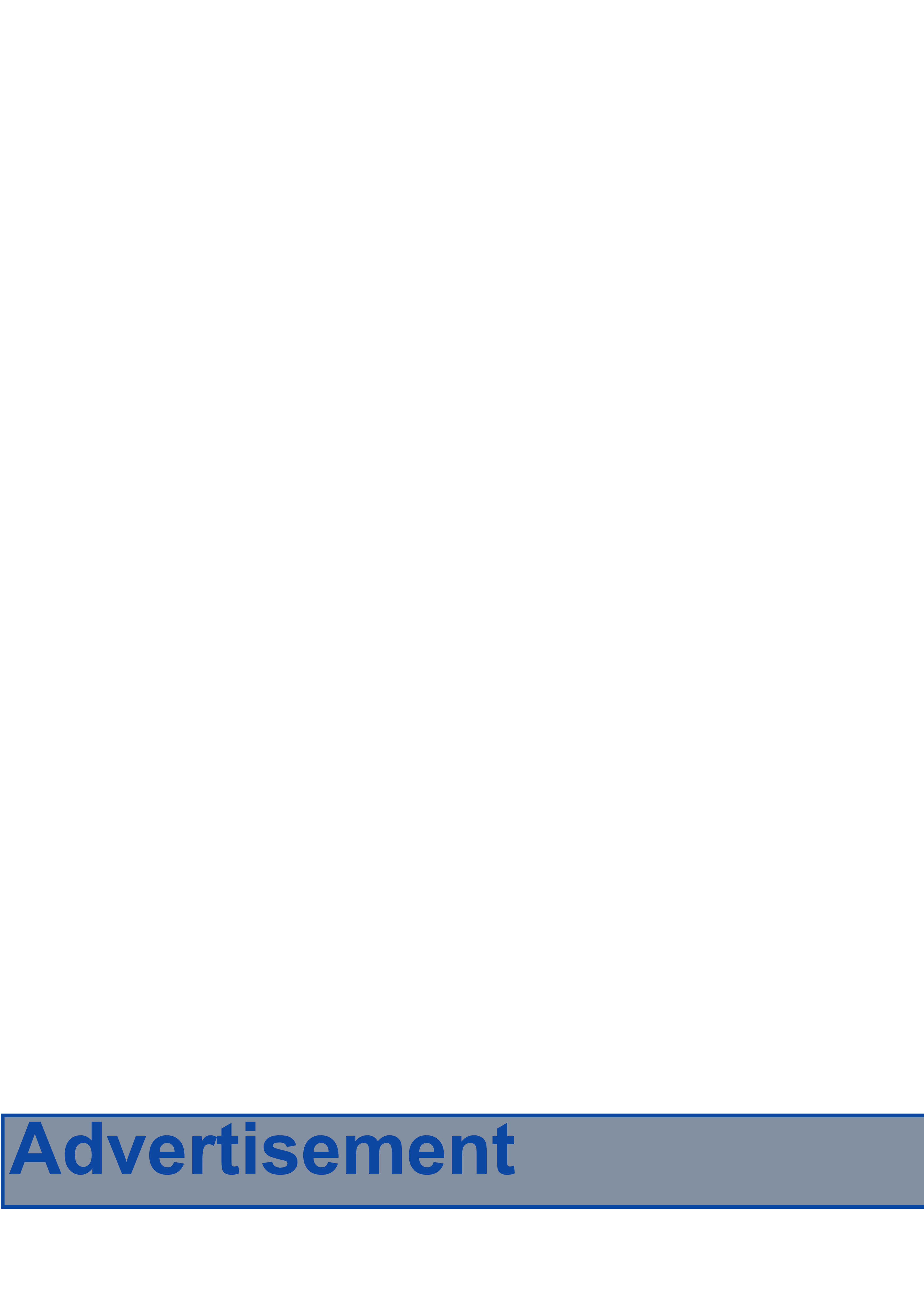} & 
    \includegraphics[width=\ricoBulkWidth]{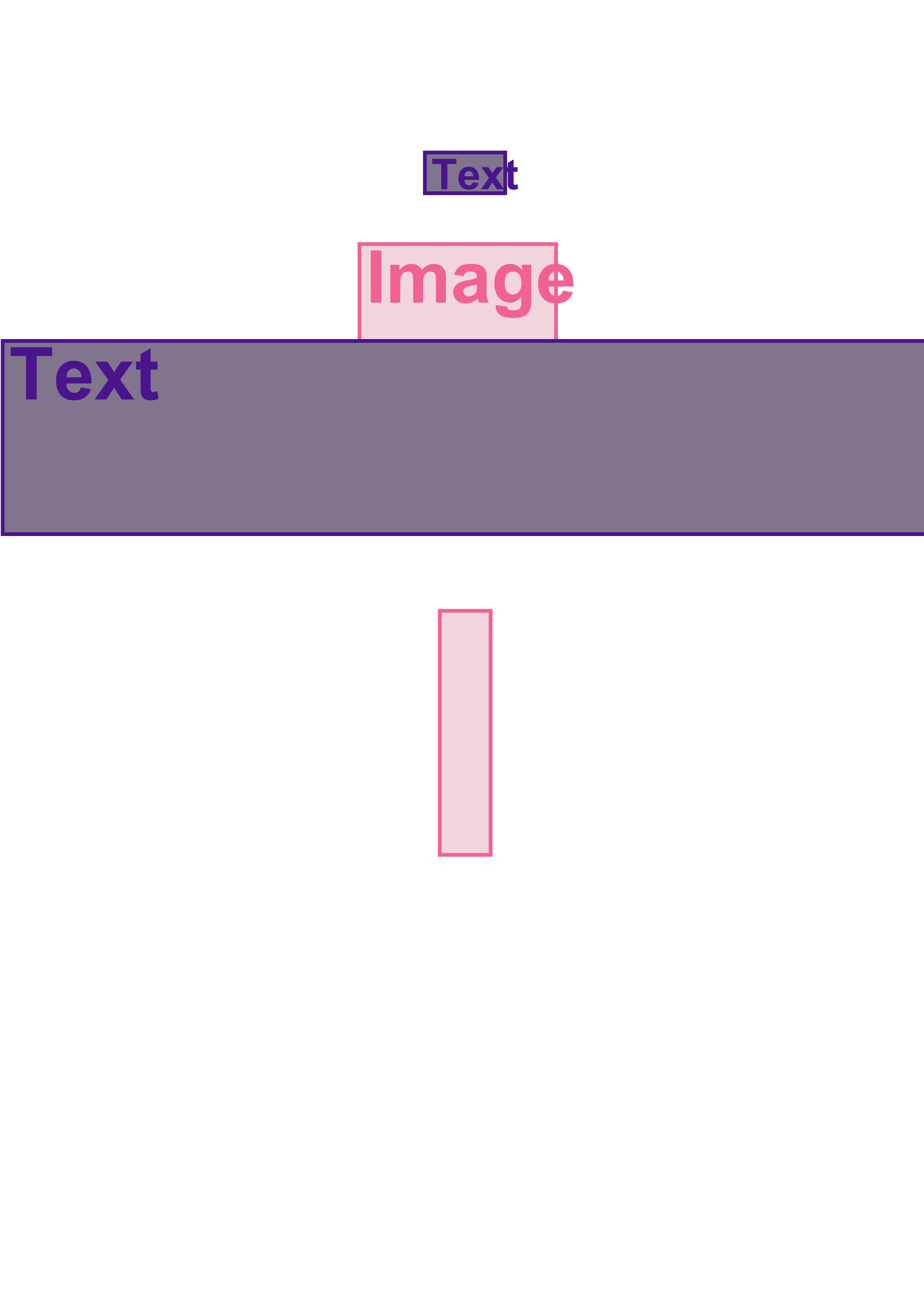} & 
    \includegraphics[width=\ricoBulkWidth]{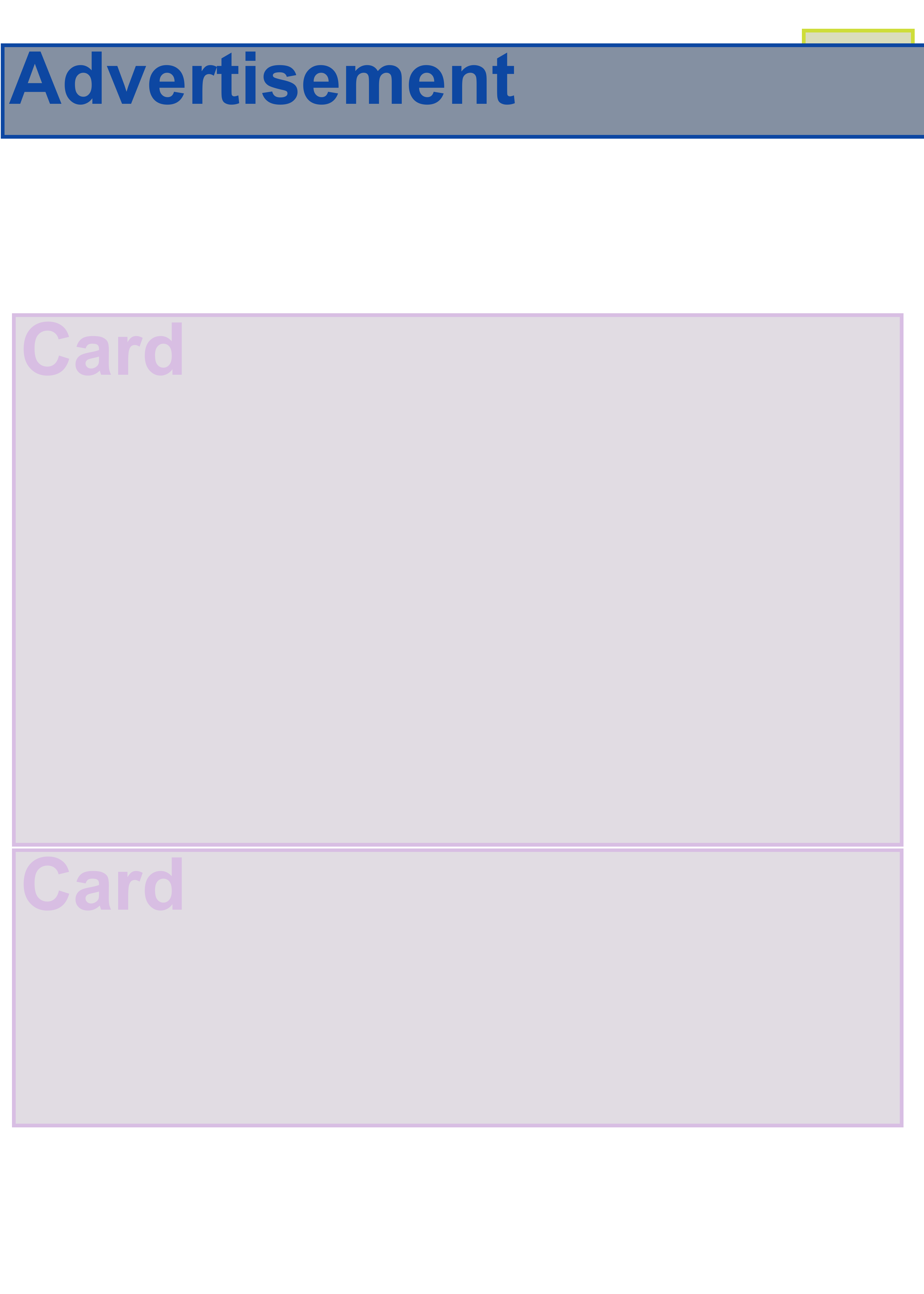} & 
    \includegraphics[width=\ricoBulkWidth]{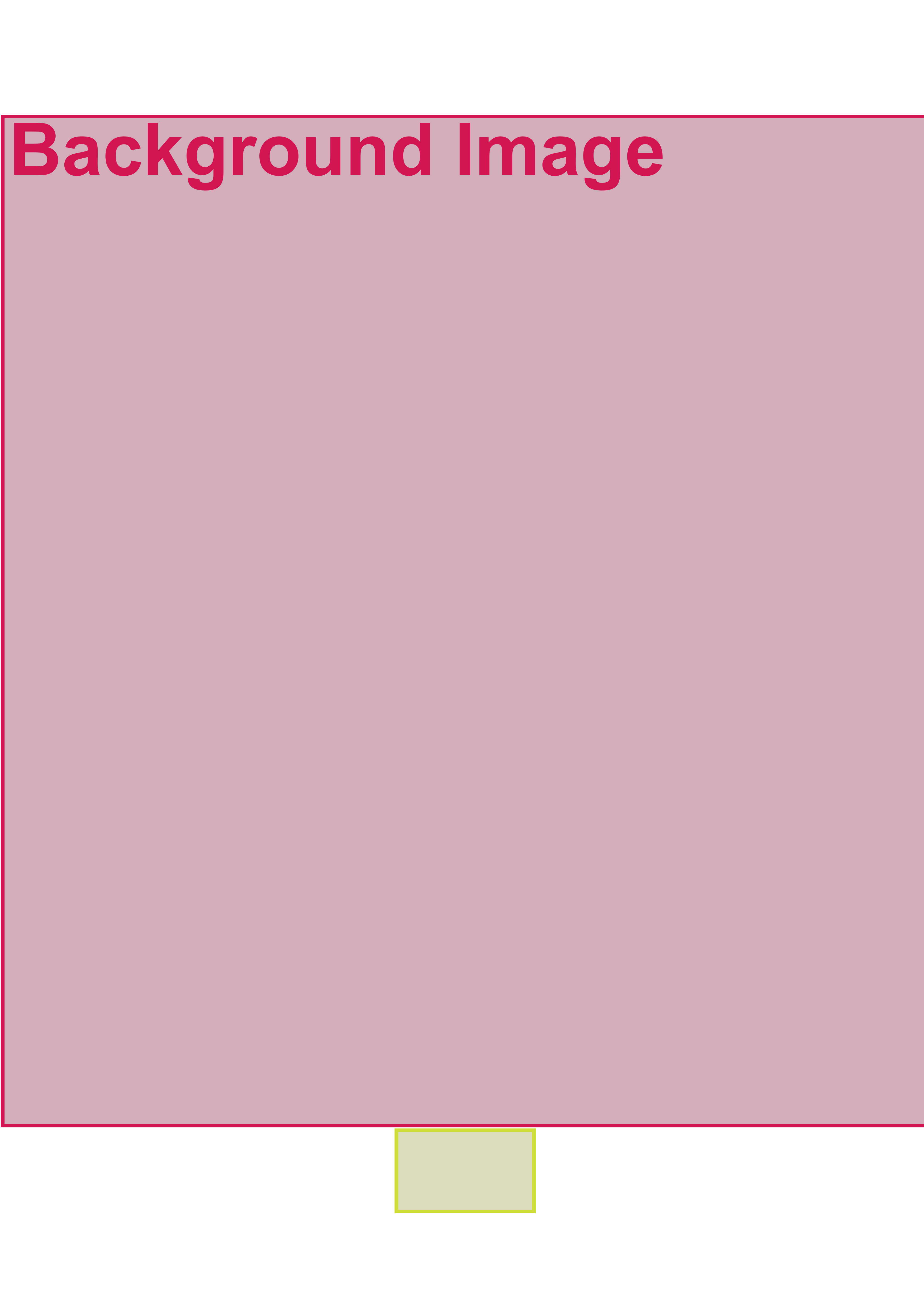} \\ 
    \includegraphics[width=\ricoBulkWidth]{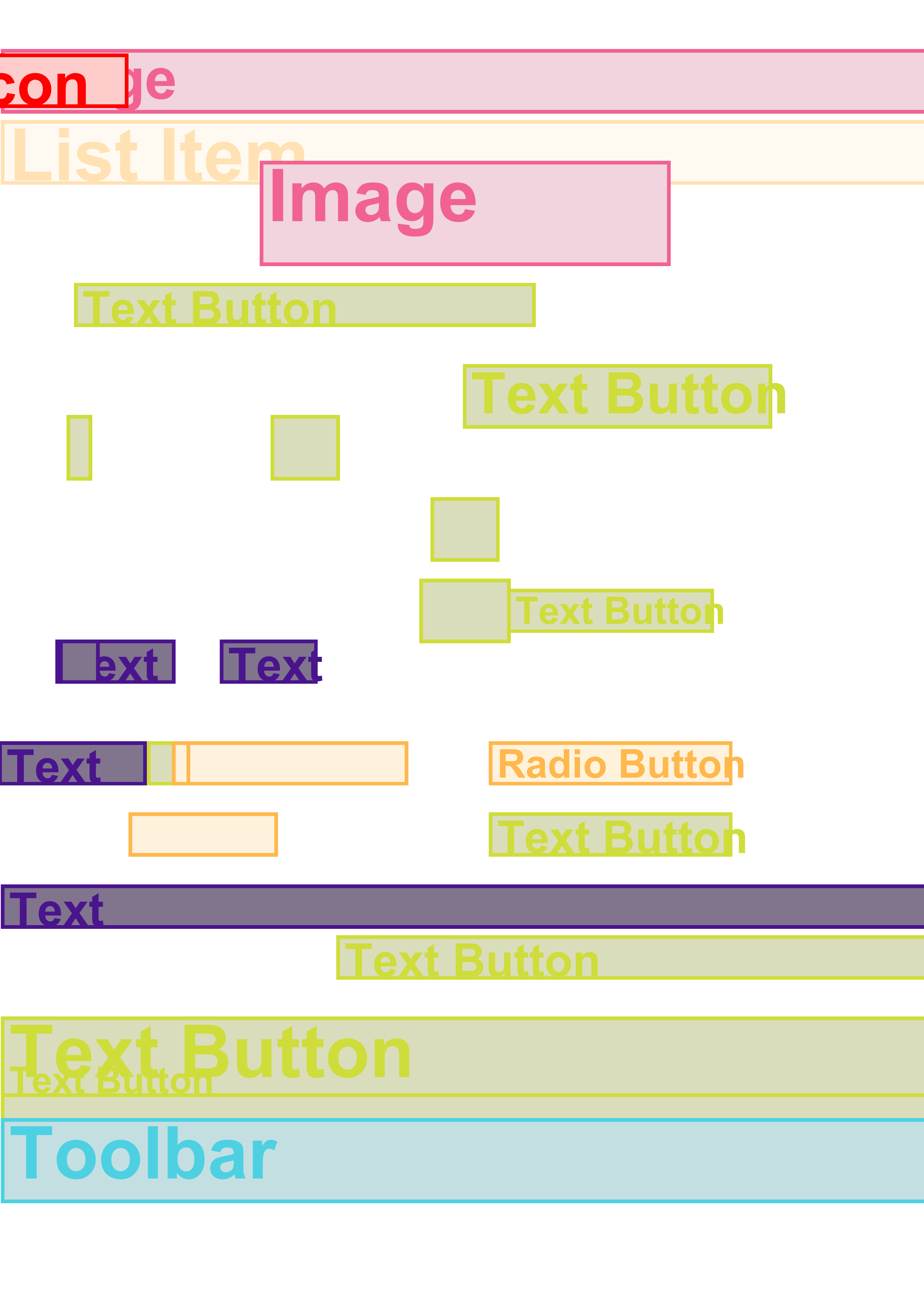} & 
    \includegraphics[width=\ricoBulkWidth]{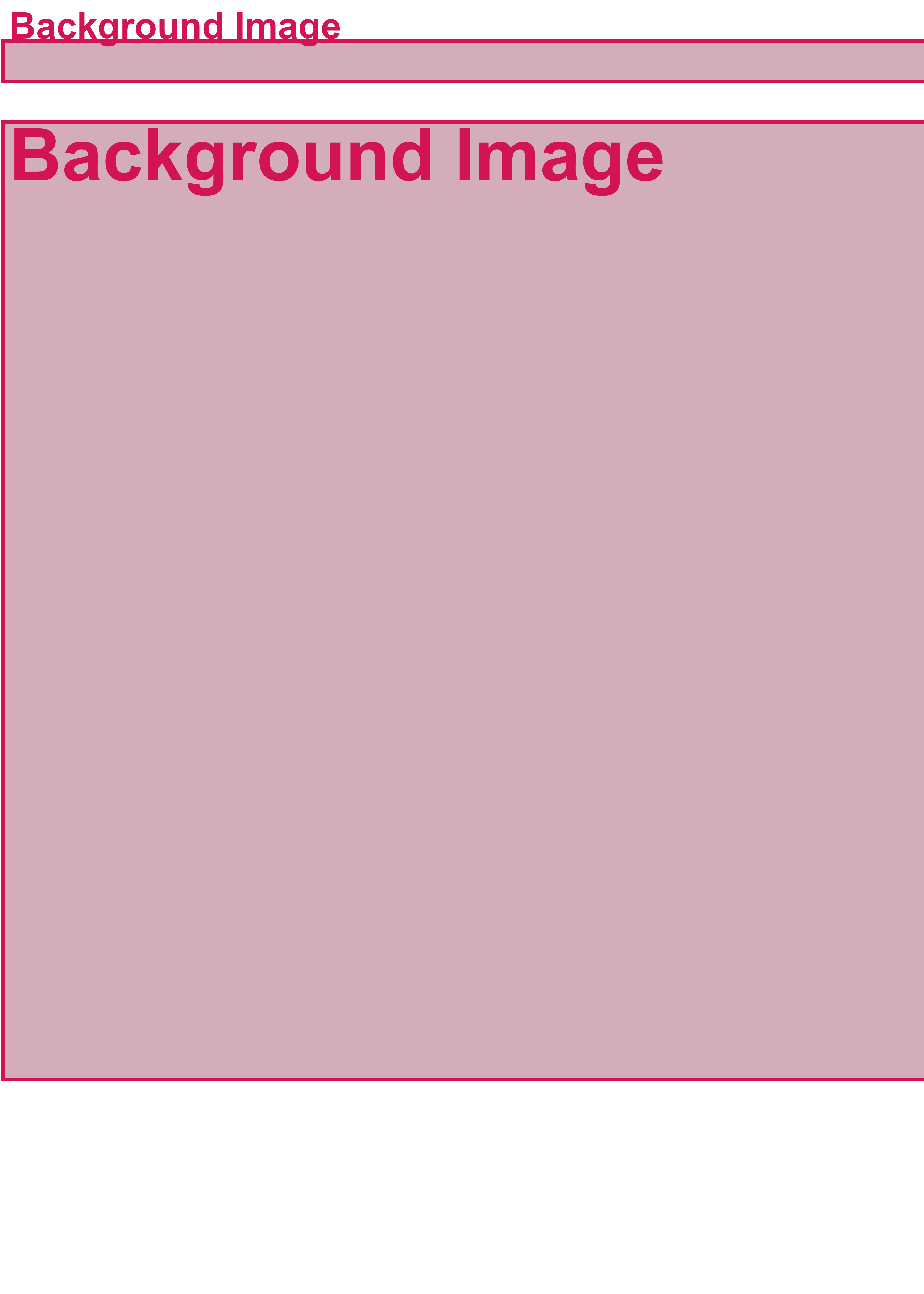} & 
    \includegraphics[width=\ricoBulkWidth]{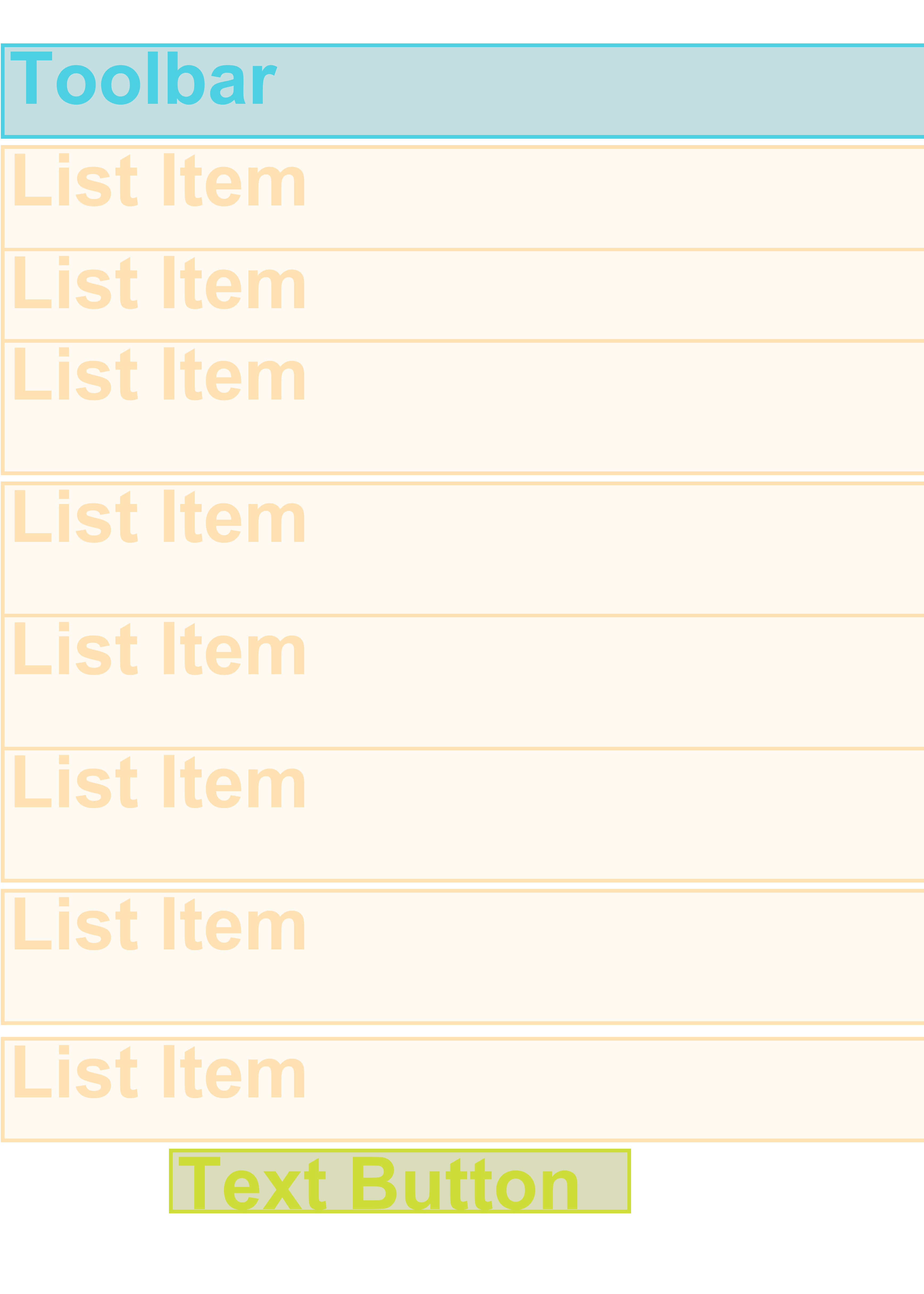} & 
    \includegraphics[width=\ricoBulkWidth]{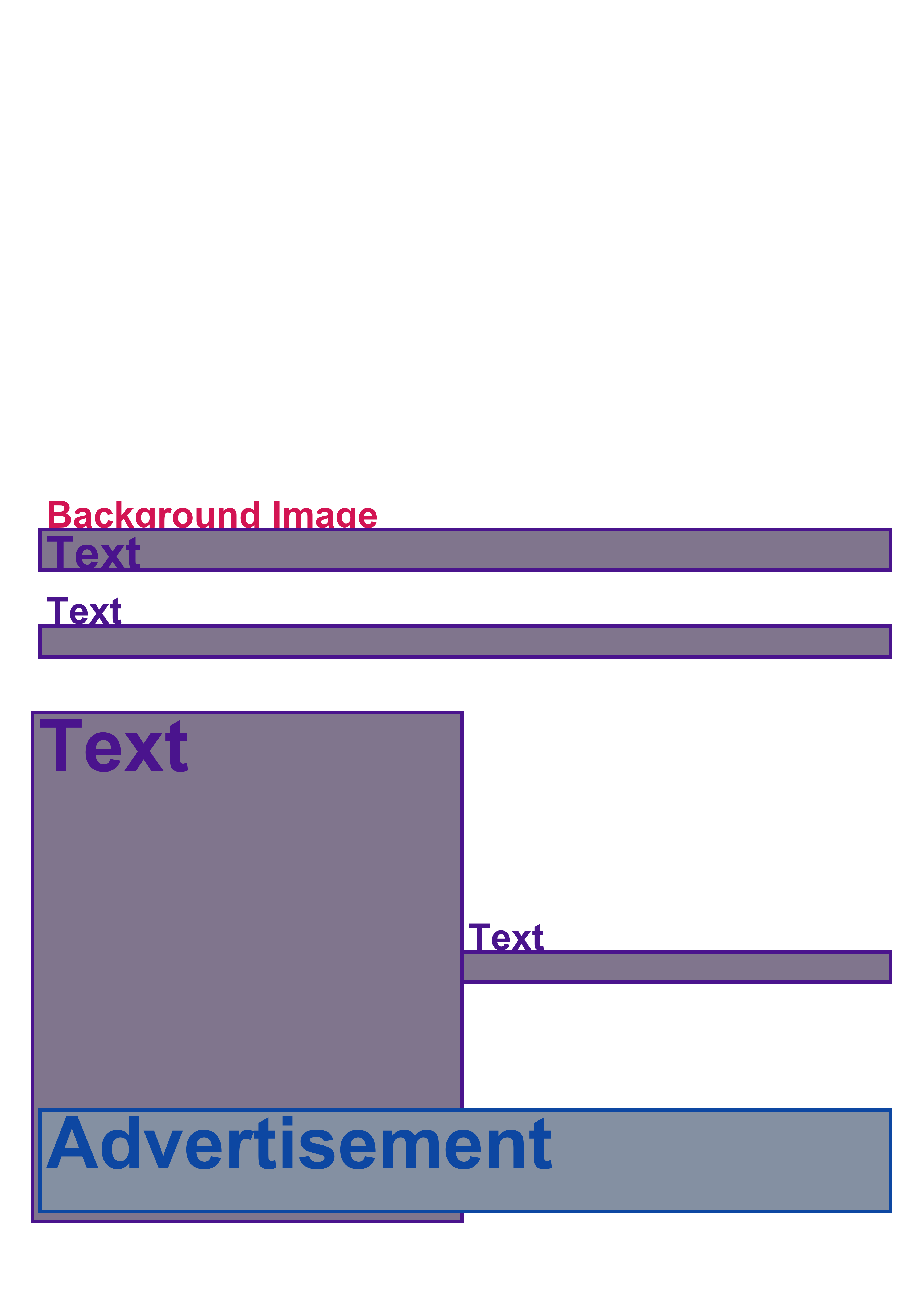} & 
    \includegraphics[width=\ricoBulkWidth]{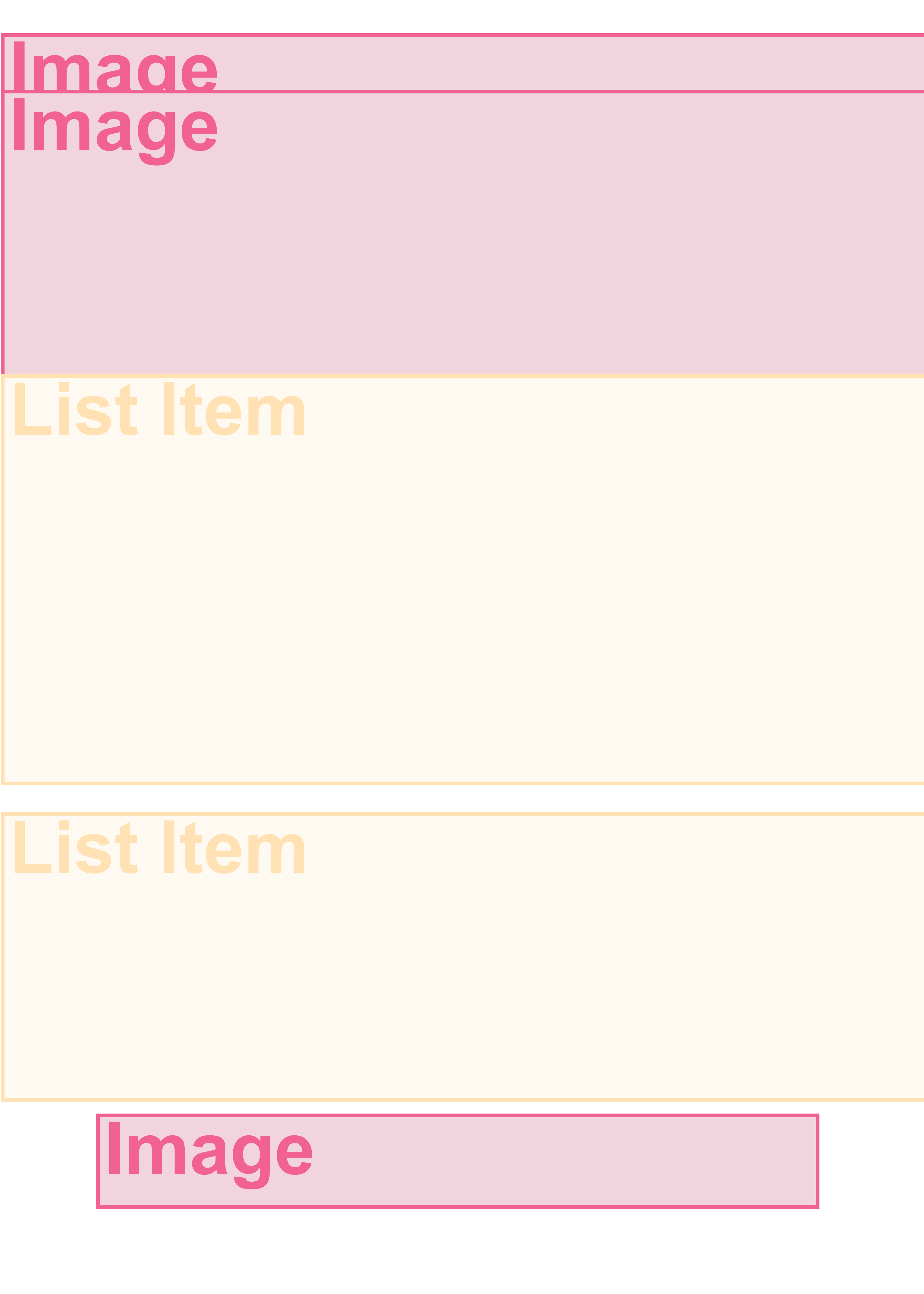} & 
    \includegraphics[width=\ricoBulkWidth]{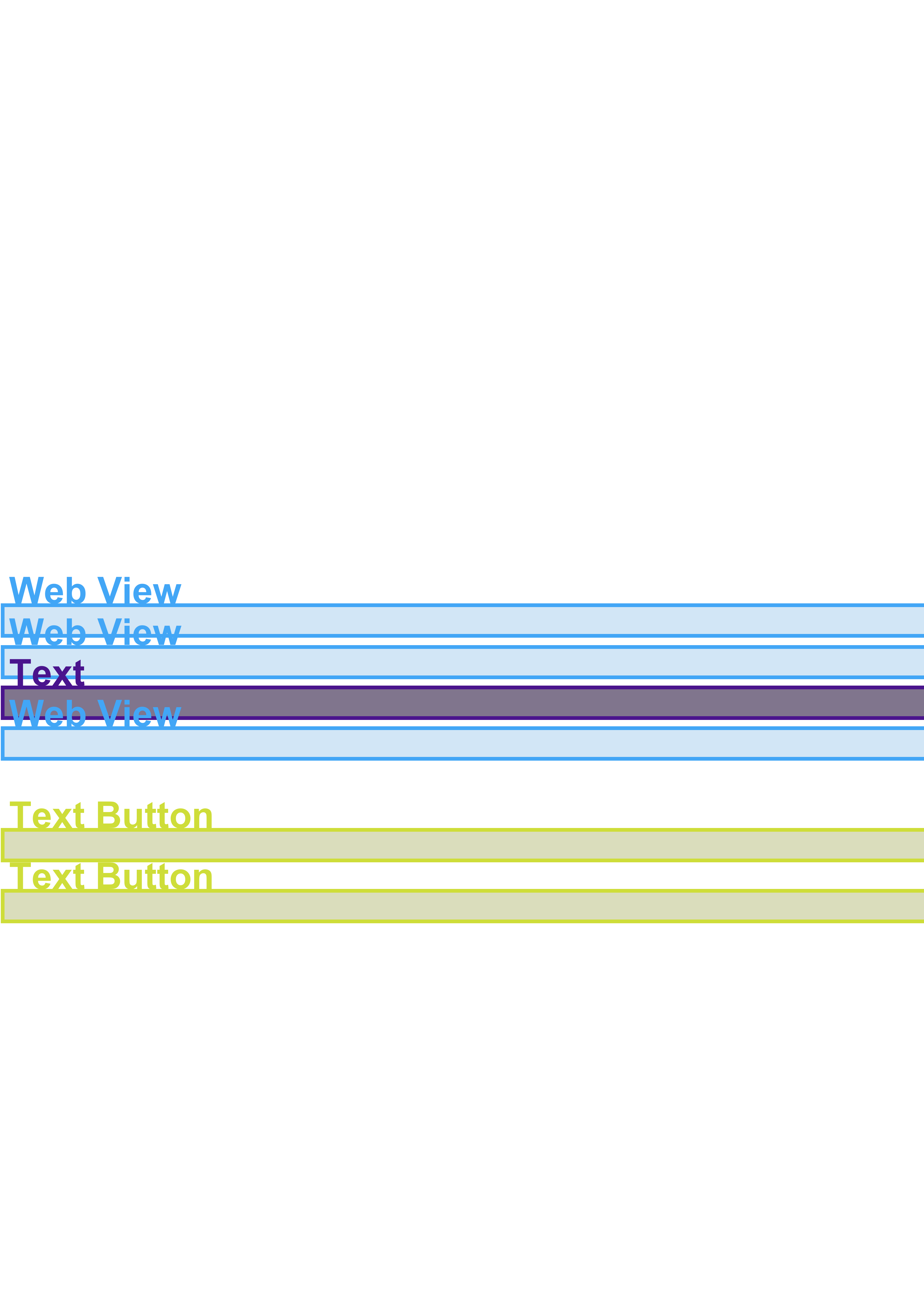} & 
    \includegraphics[width=\ricoBulkWidth]{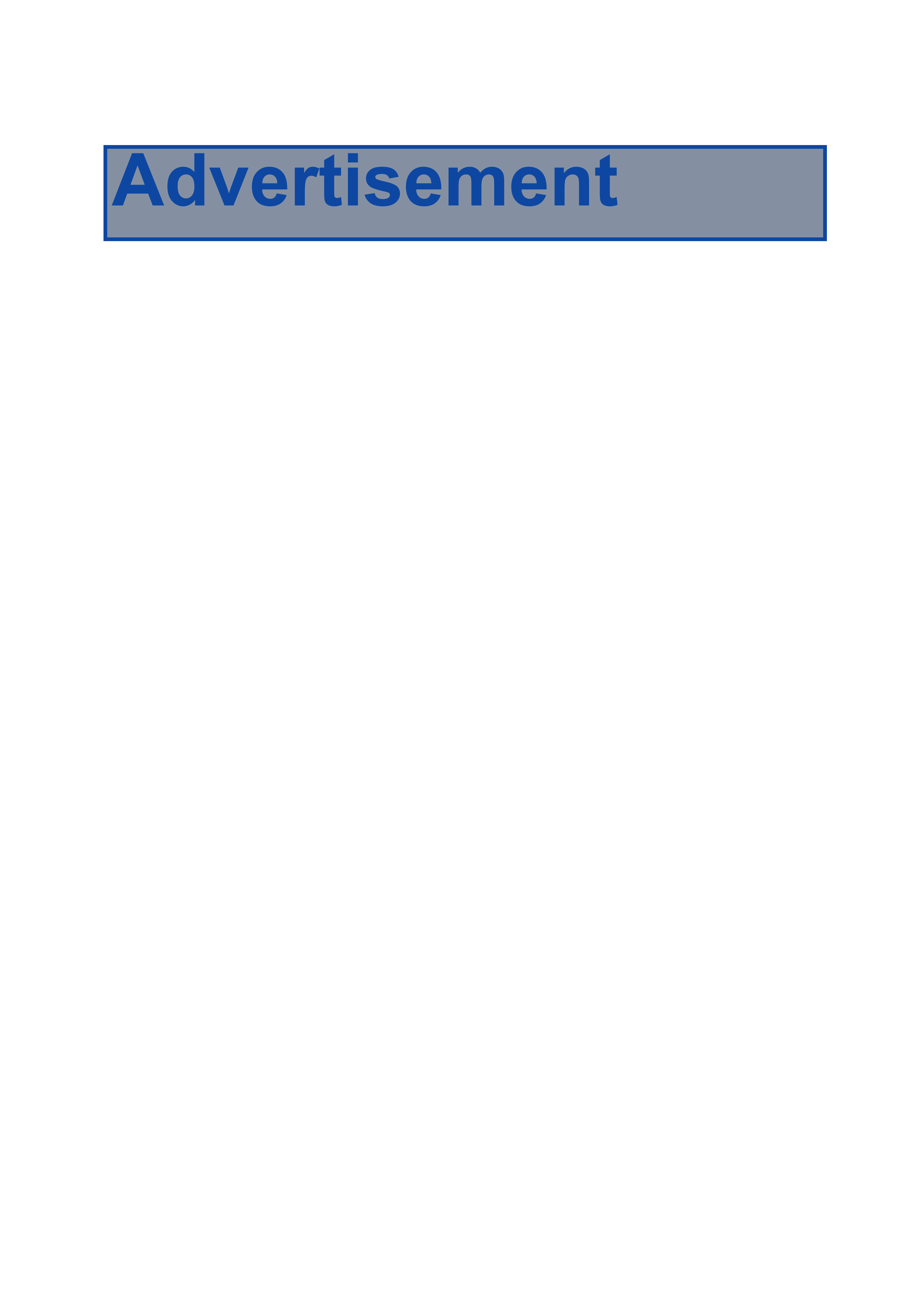} \\ 
    \includegraphics[width=\ricoBulkWidth]{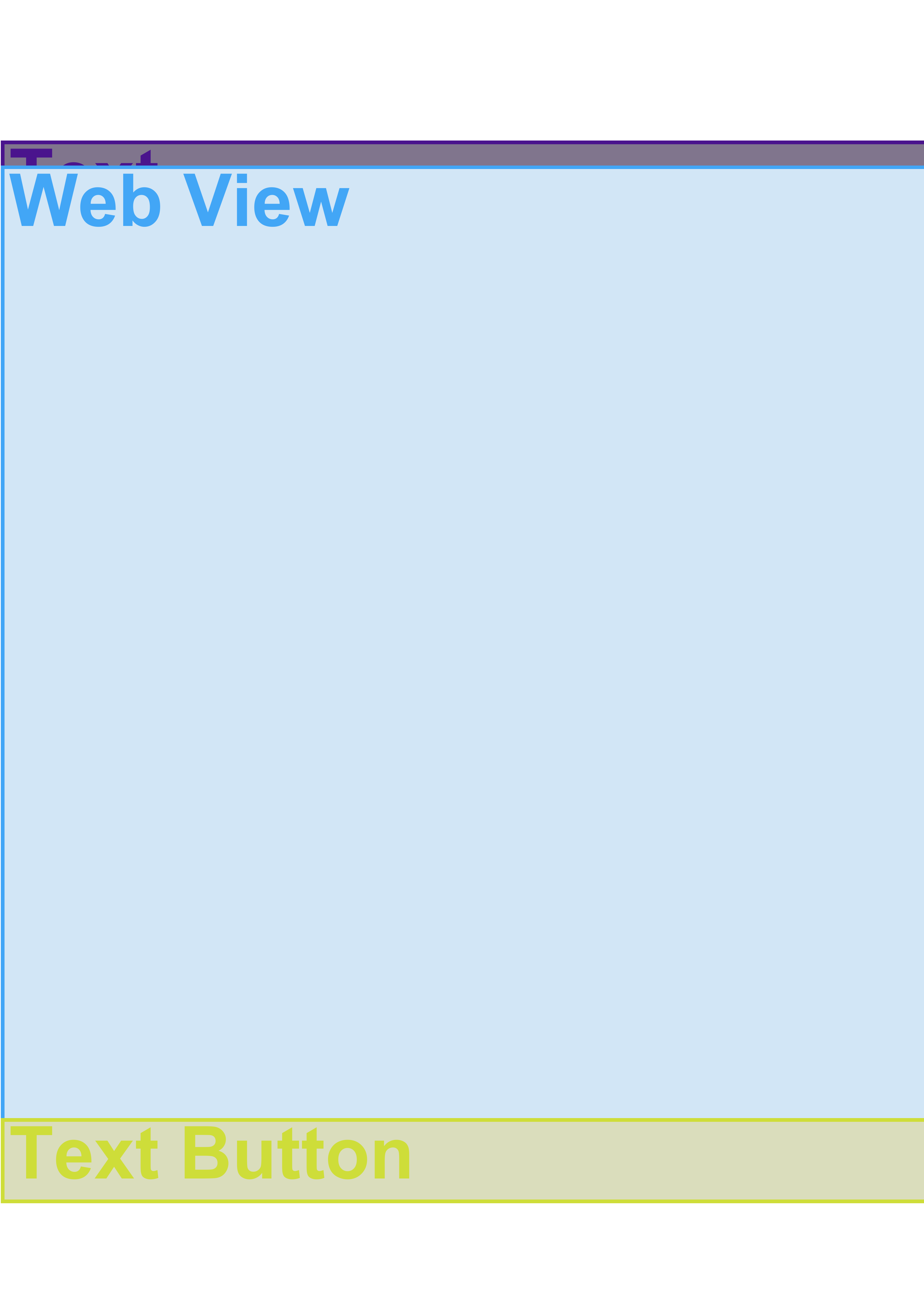} & 
    \includegraphics[width=\ricoBulkWidth]{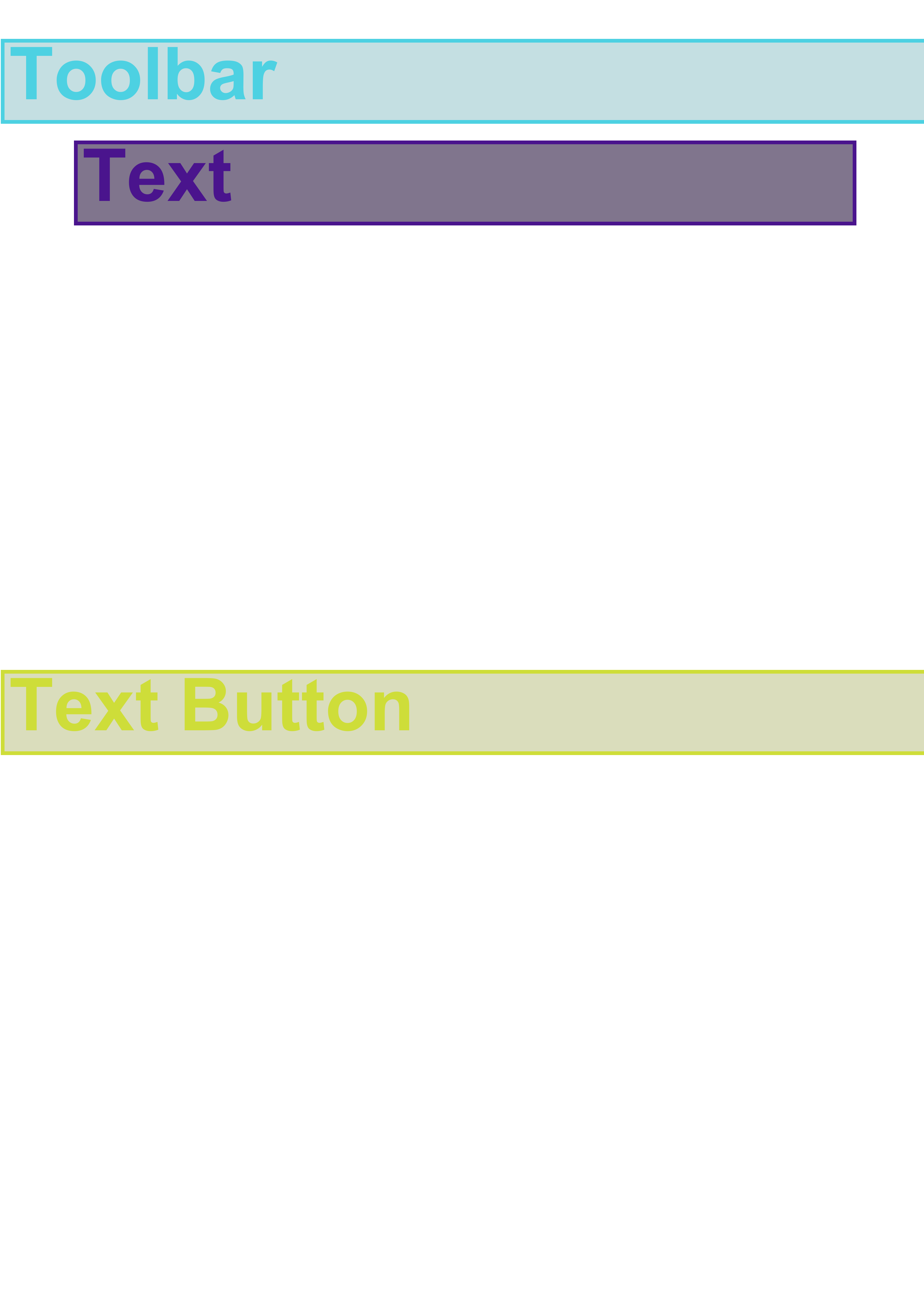} & 
    \includegraphics[width=\ricoBulkWidth]{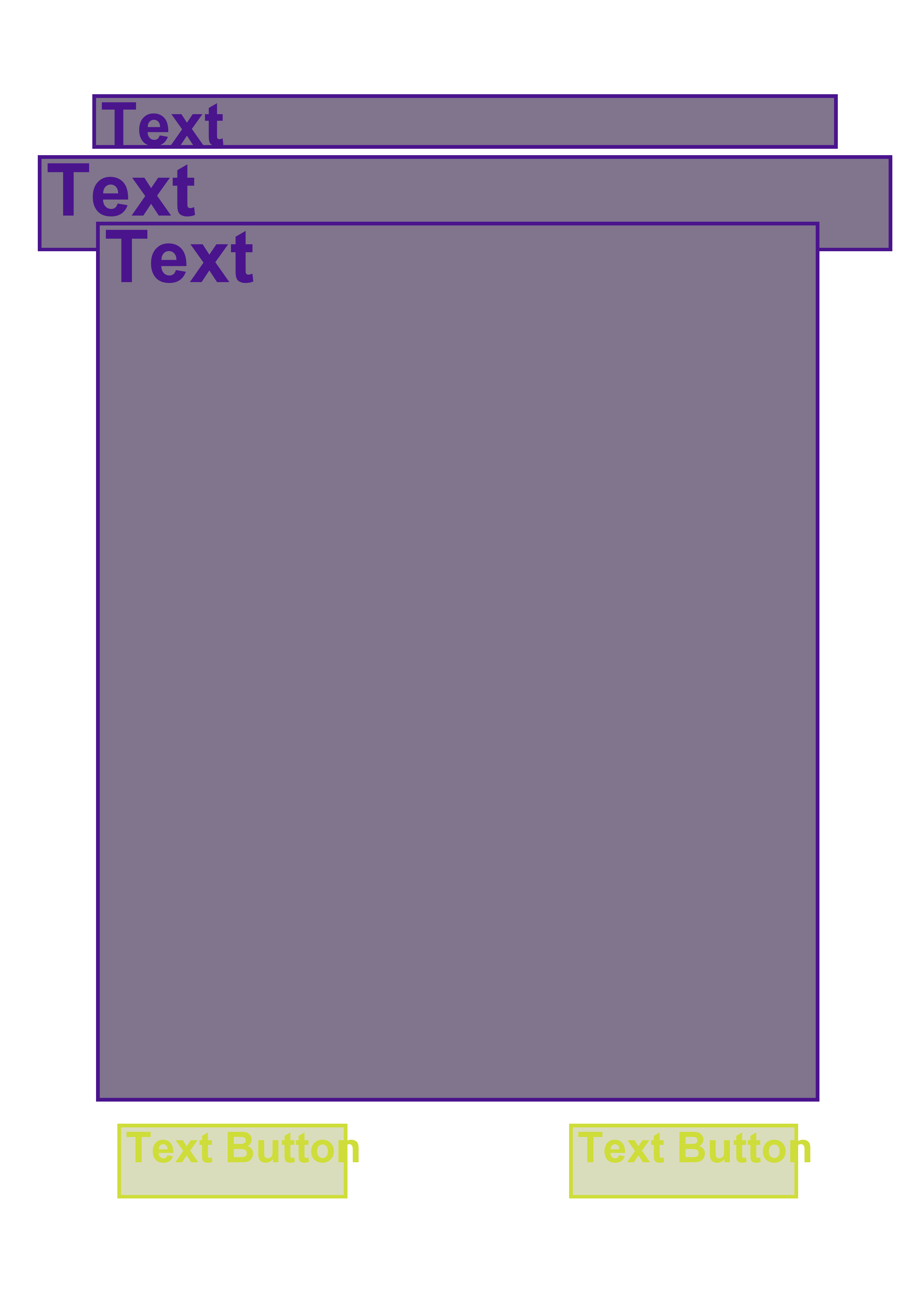} & 
    \includegraphics[width=\ricoBulkWidth]{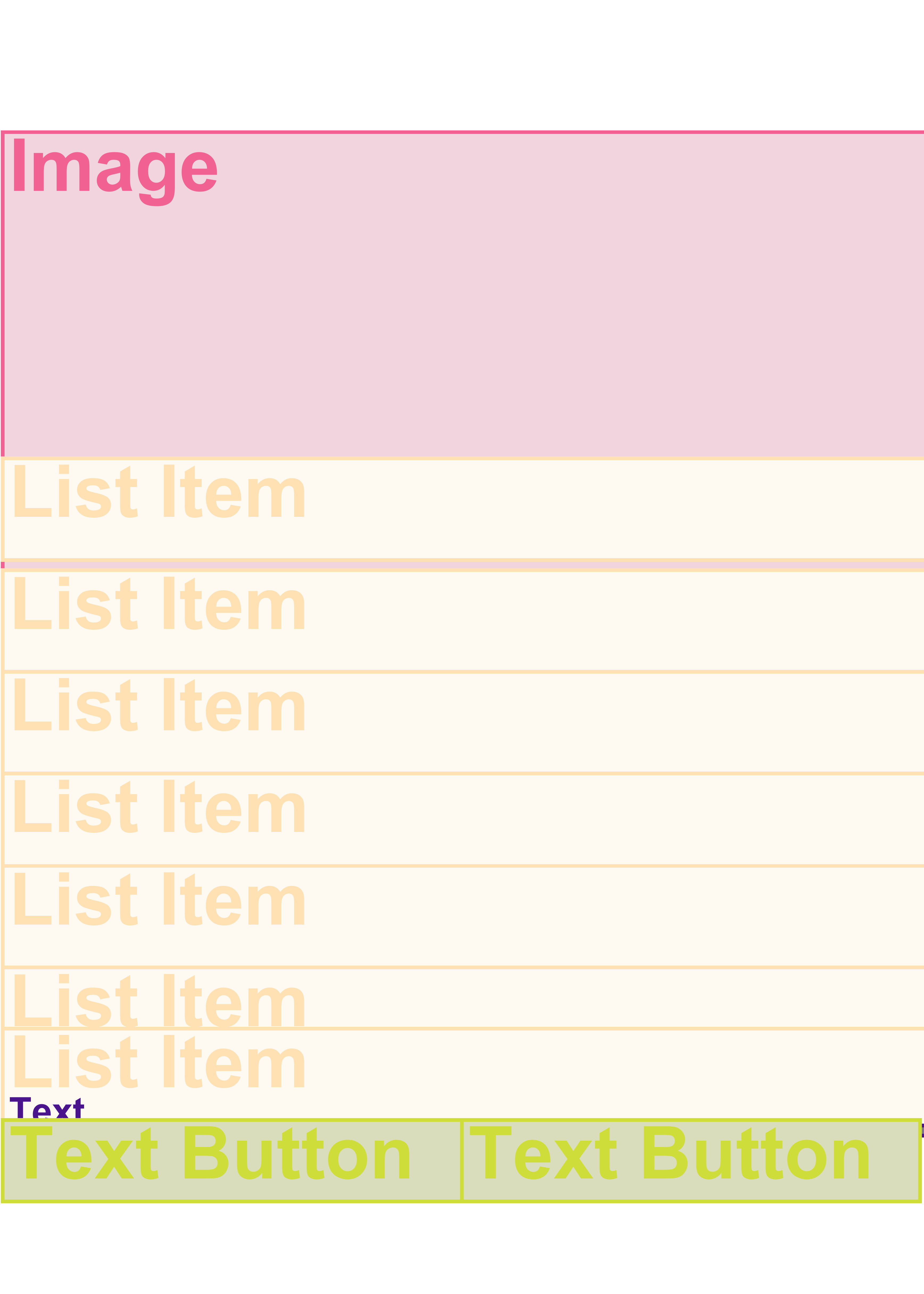} & 
    \includegraphics[width=\ricoBulkWidth]{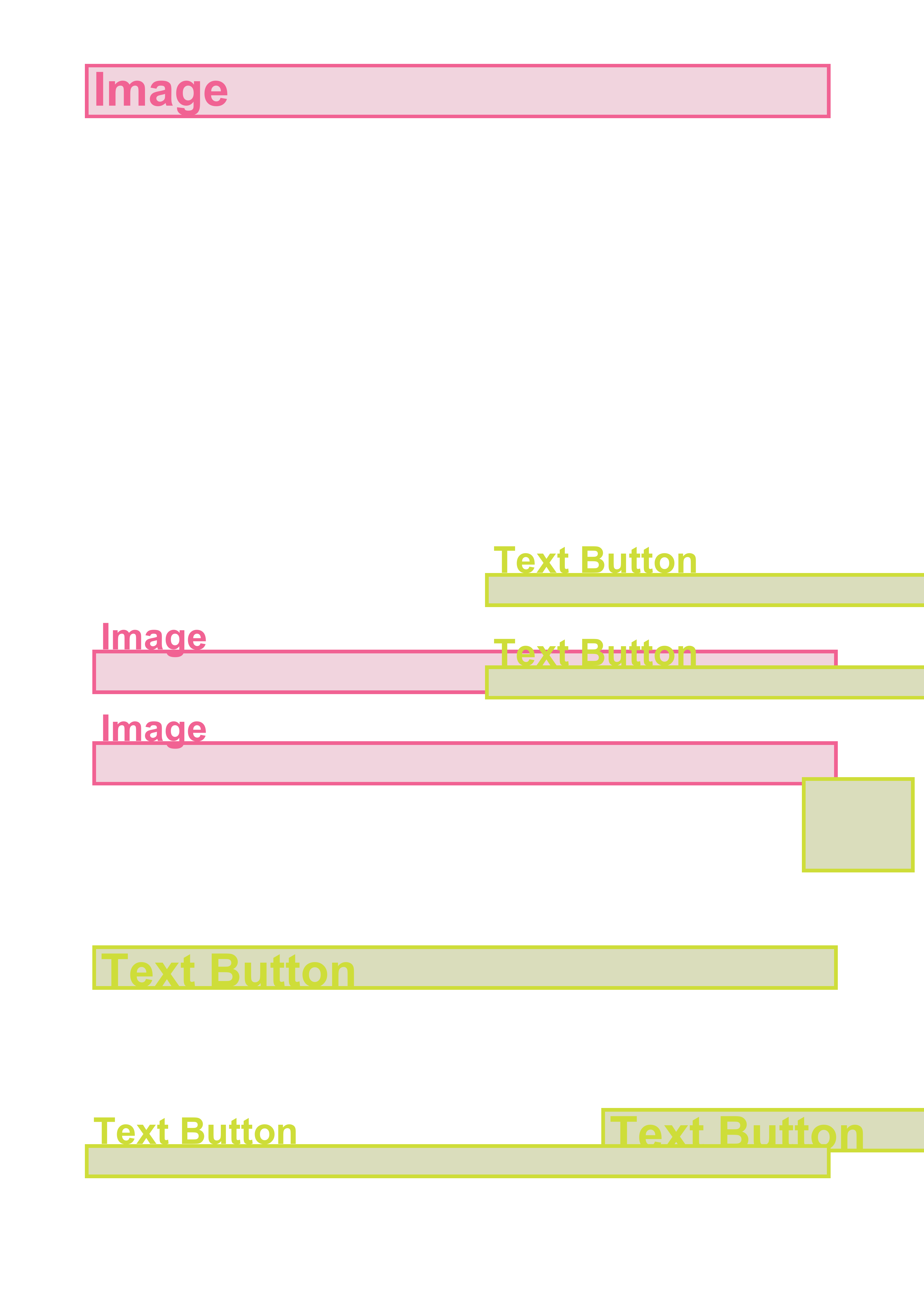} & 
    \includegraphics[width=\ricoBulkWidth]{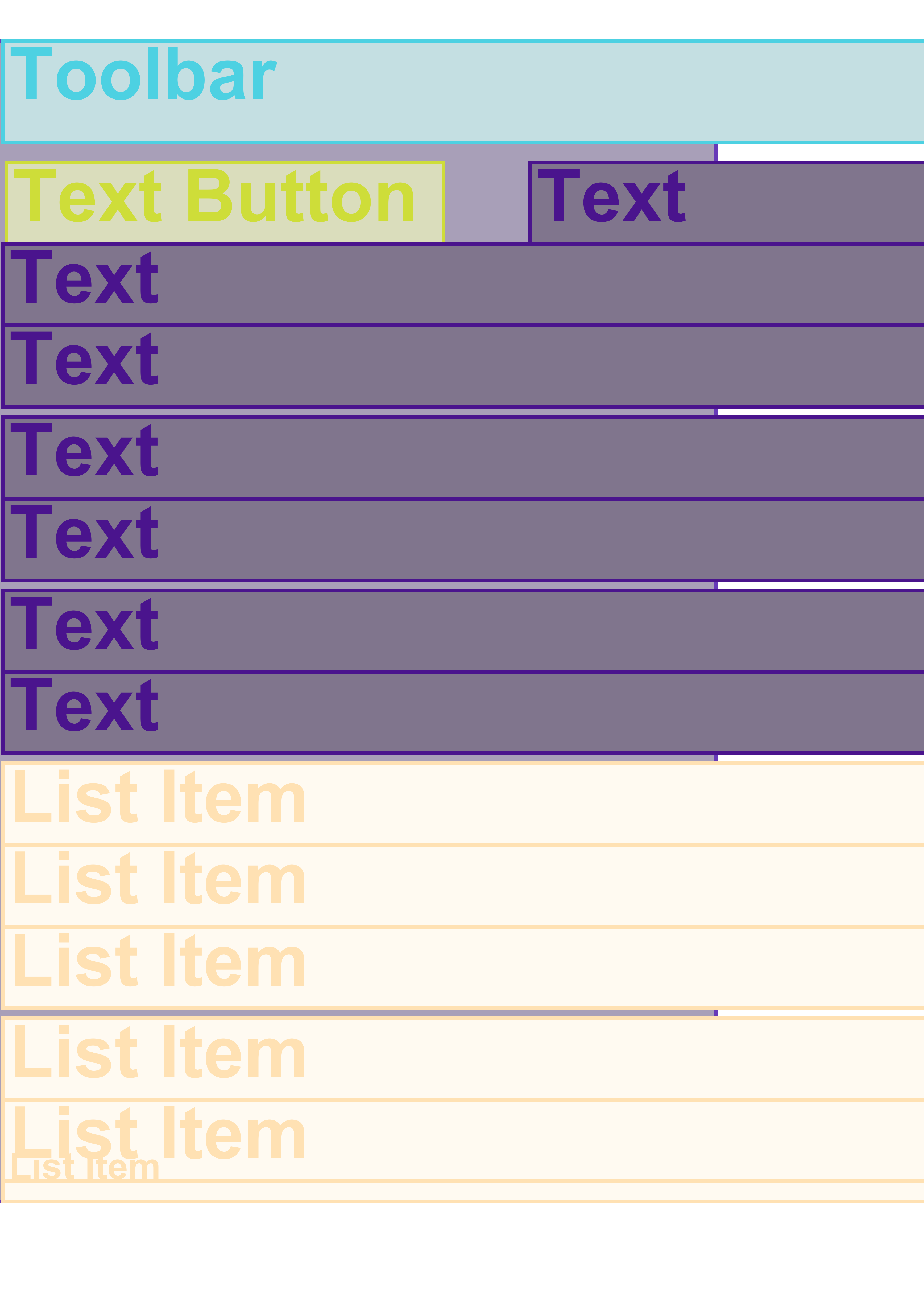} & 
    \includegraphics[width=\ricoBulkWidth]{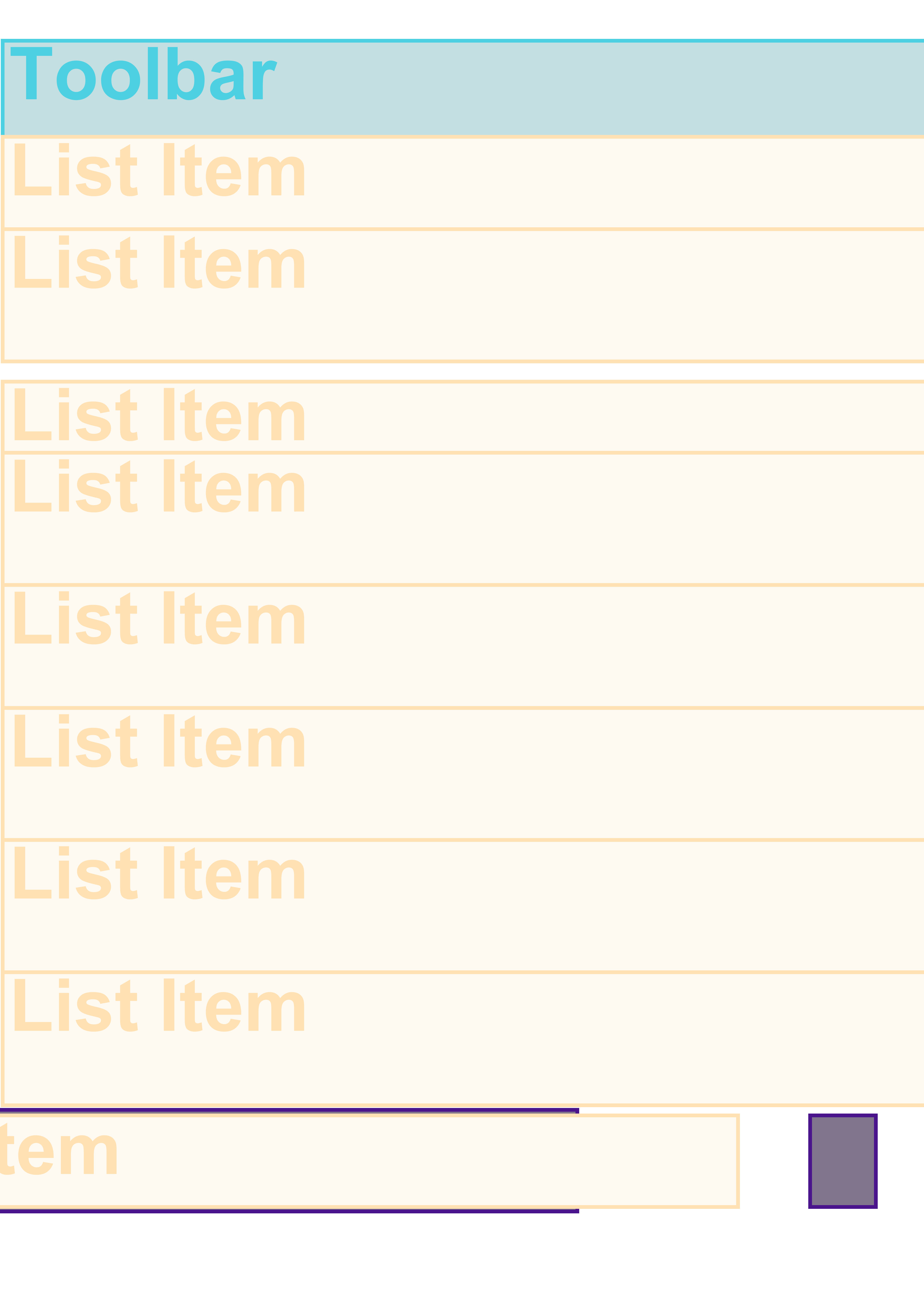} \\ 
    \includegraphics[width=\ricoBulkWidth]{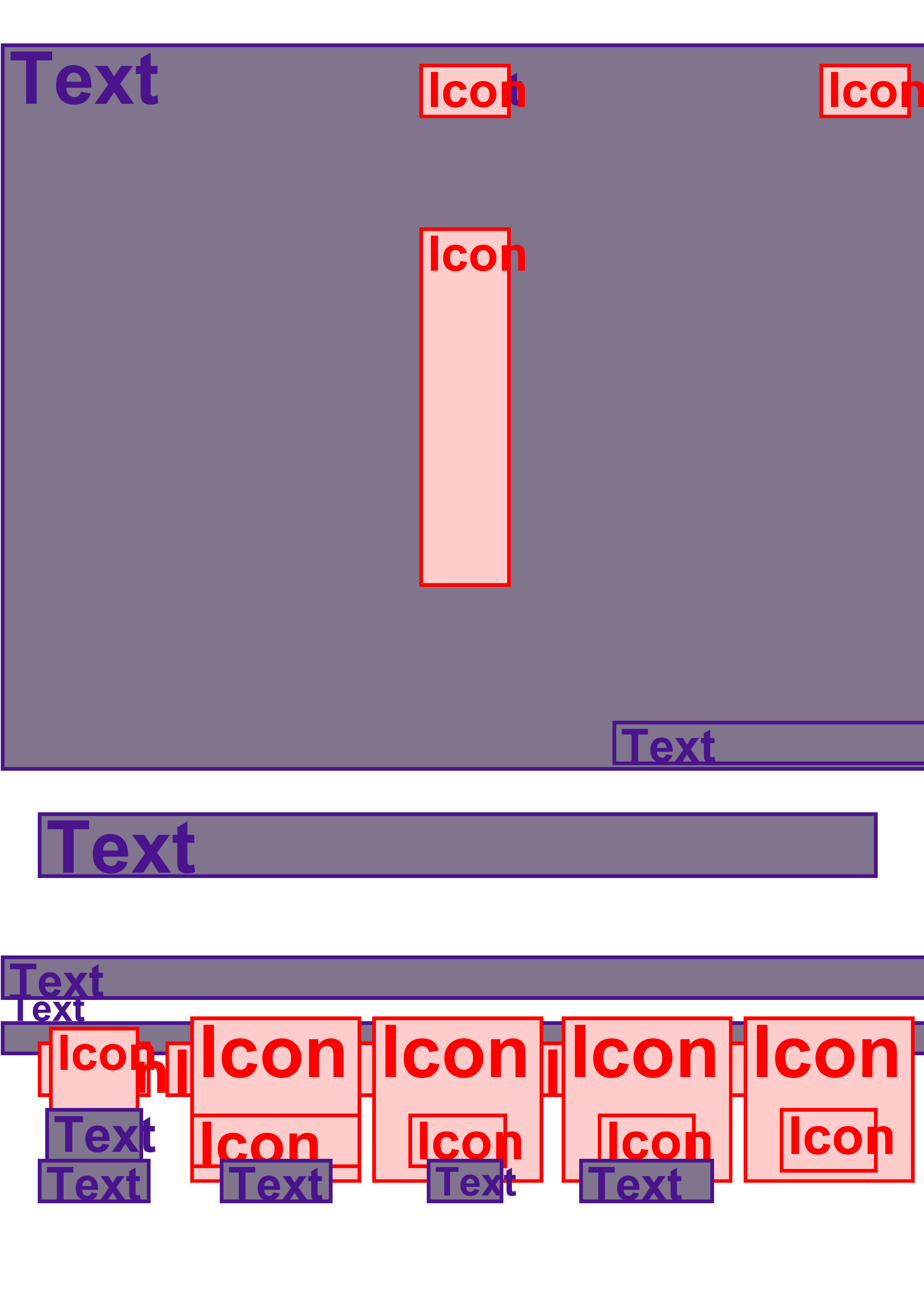} & 
    \includegraphics[width=\ricoBulkWidth]{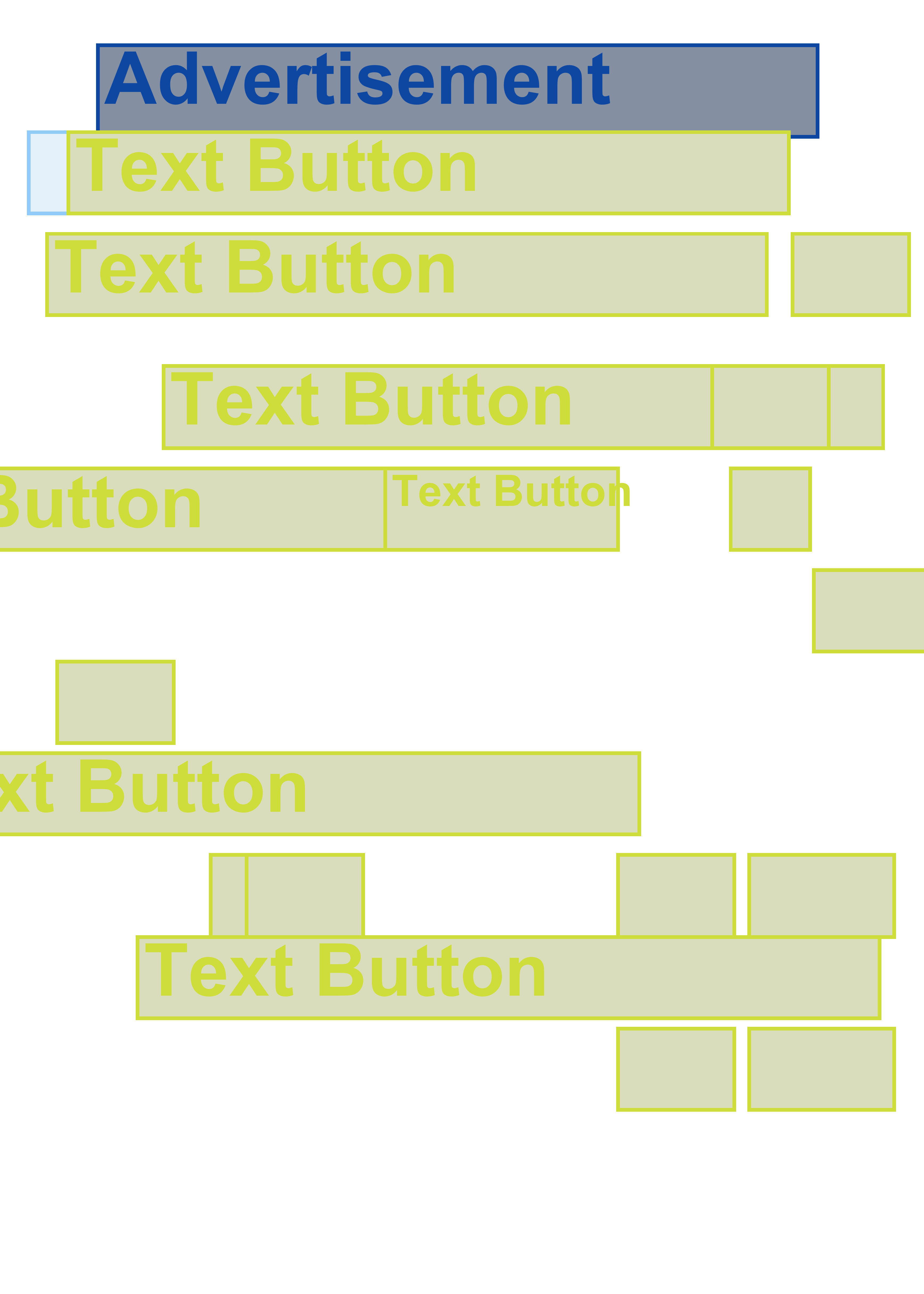} & 
    \includegraphics[width=\ricoBulkWidth]{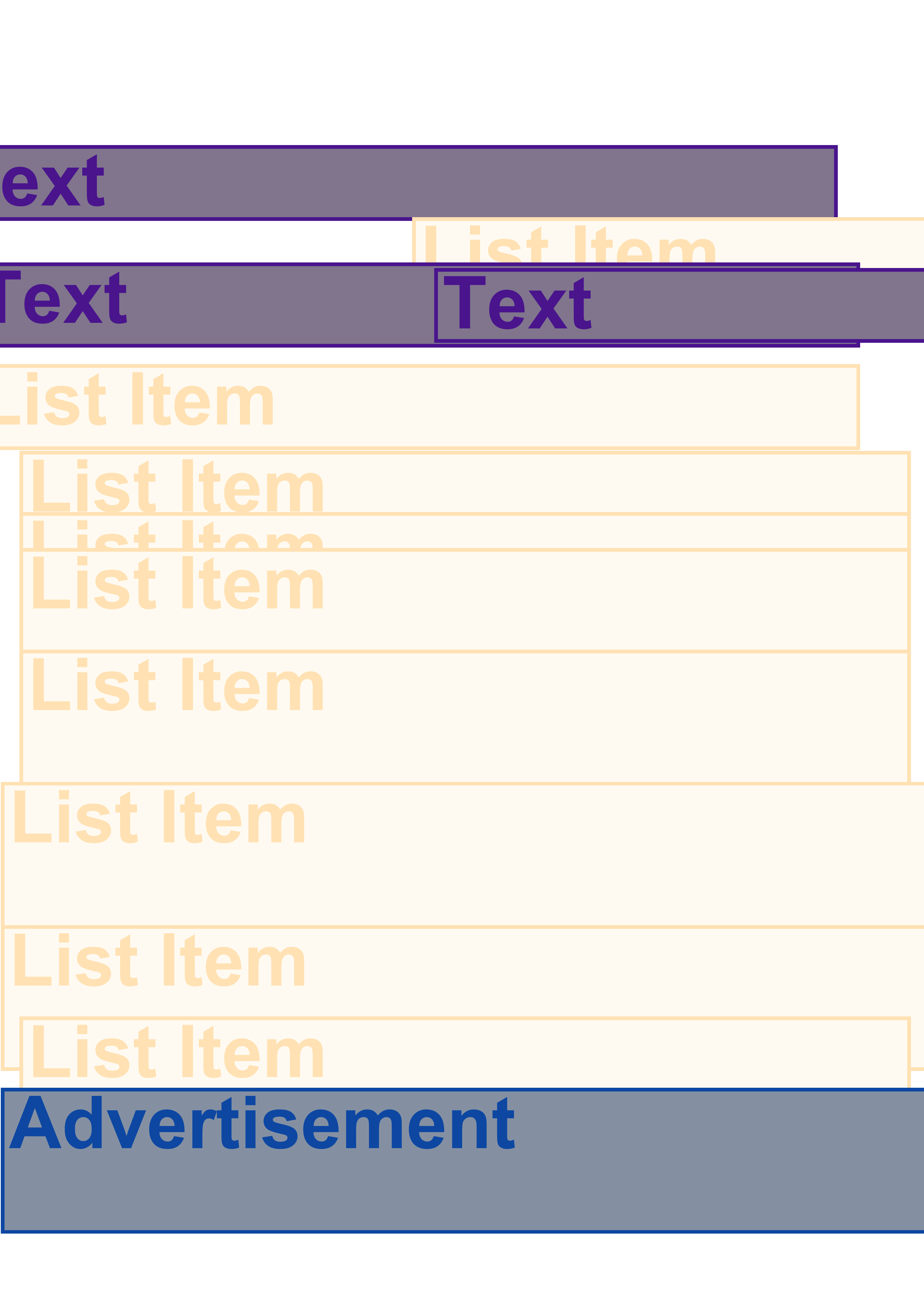} & 
    \includegraphics[width=\ricoBulkWidth]{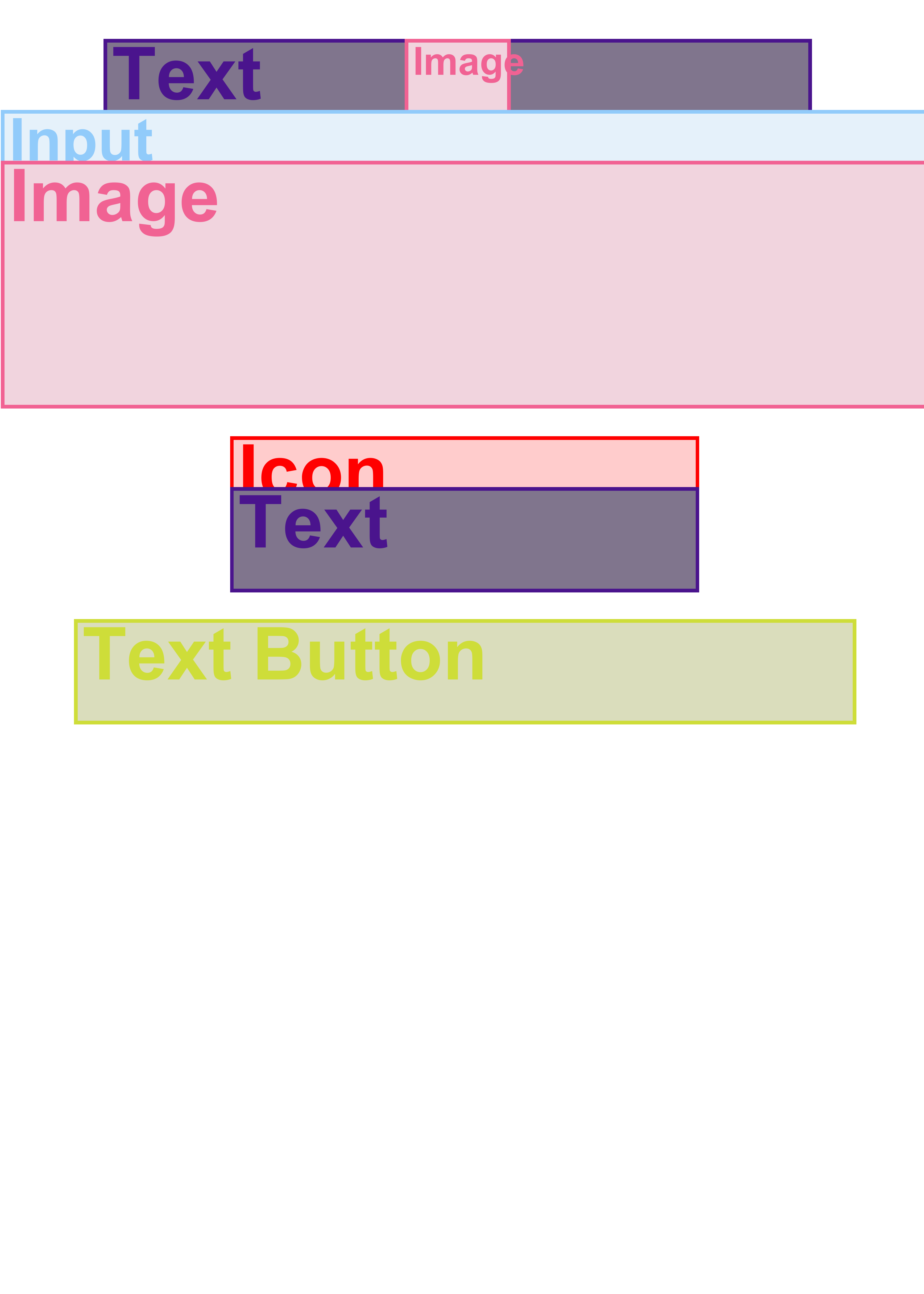} & 
    \includegraphics[width=\ricoBulkWidth]{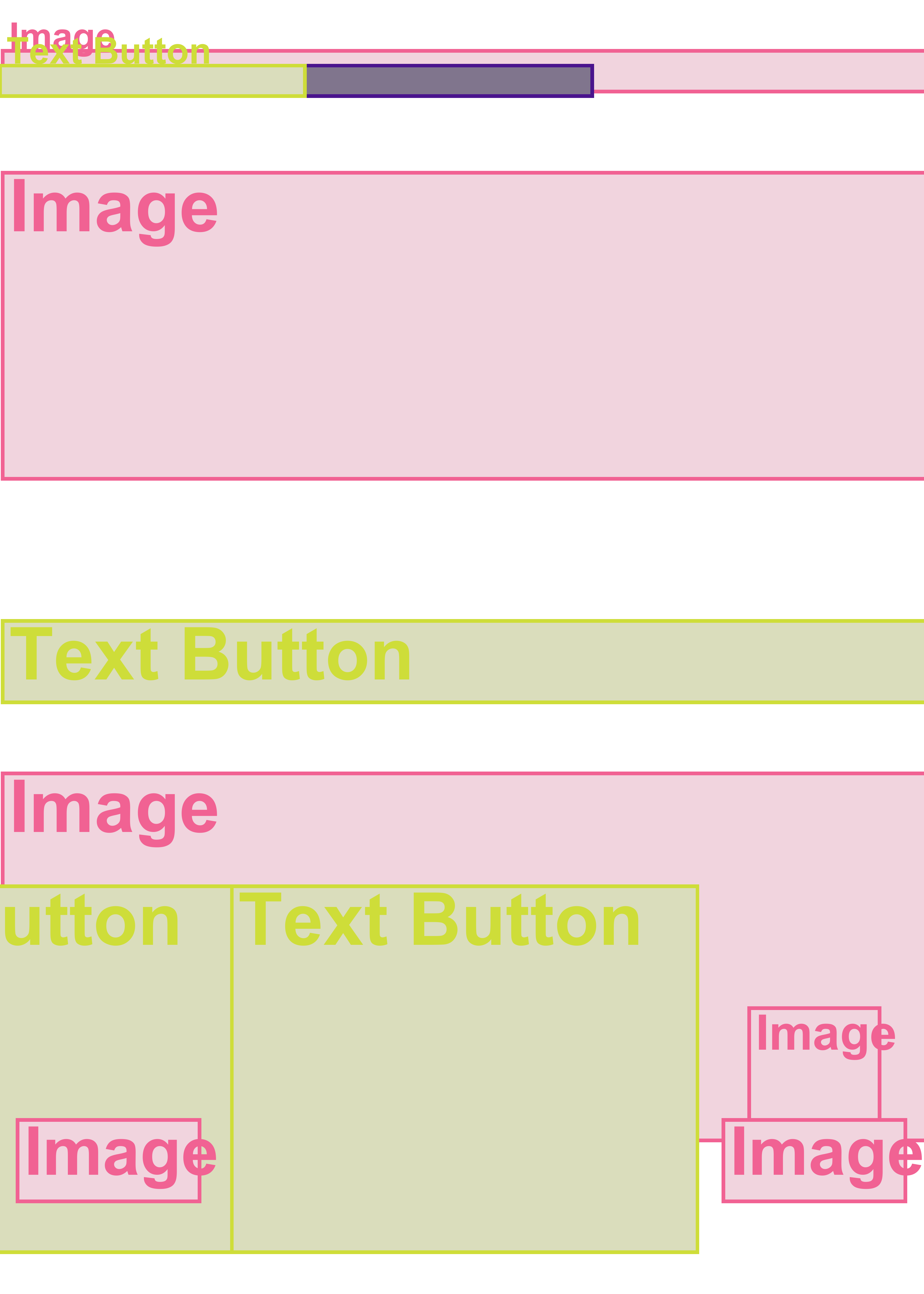} & 
    \includegraphics[width=\ricoBulkWidth]{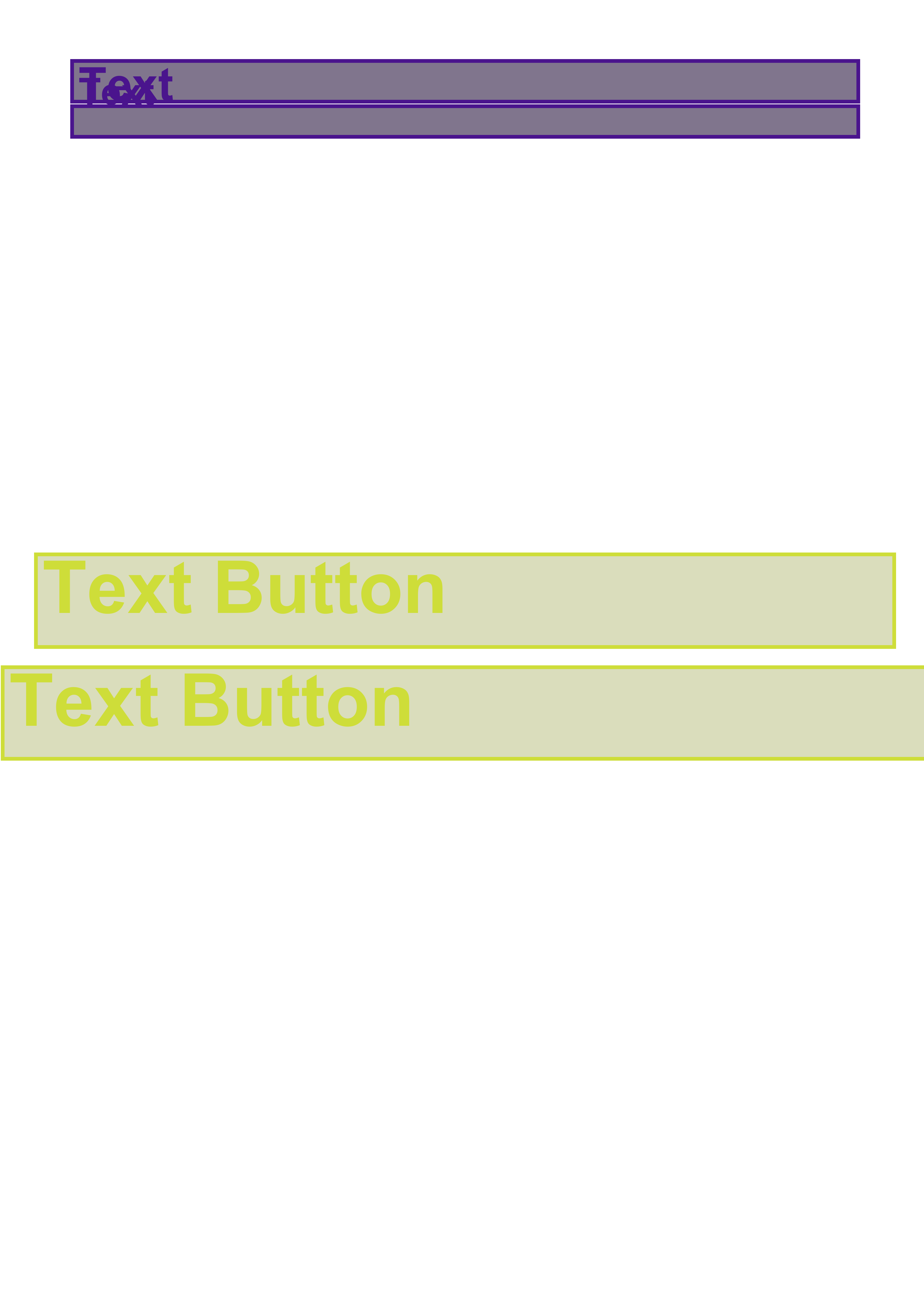} & 
    \includegraphics[width=\ricoBulkWidth]{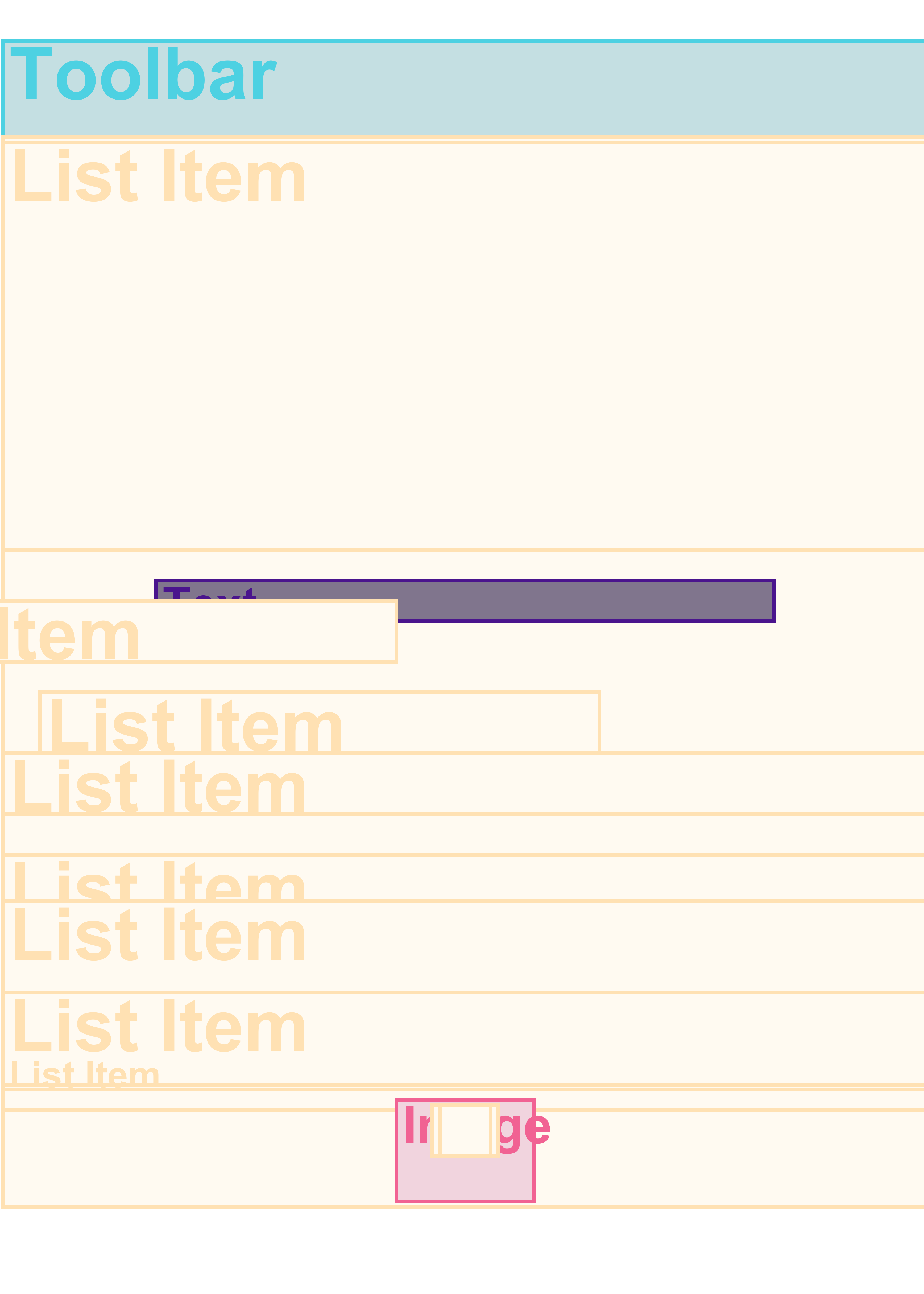} \\
    \includegraphics[width=\ricoBulkWidth]{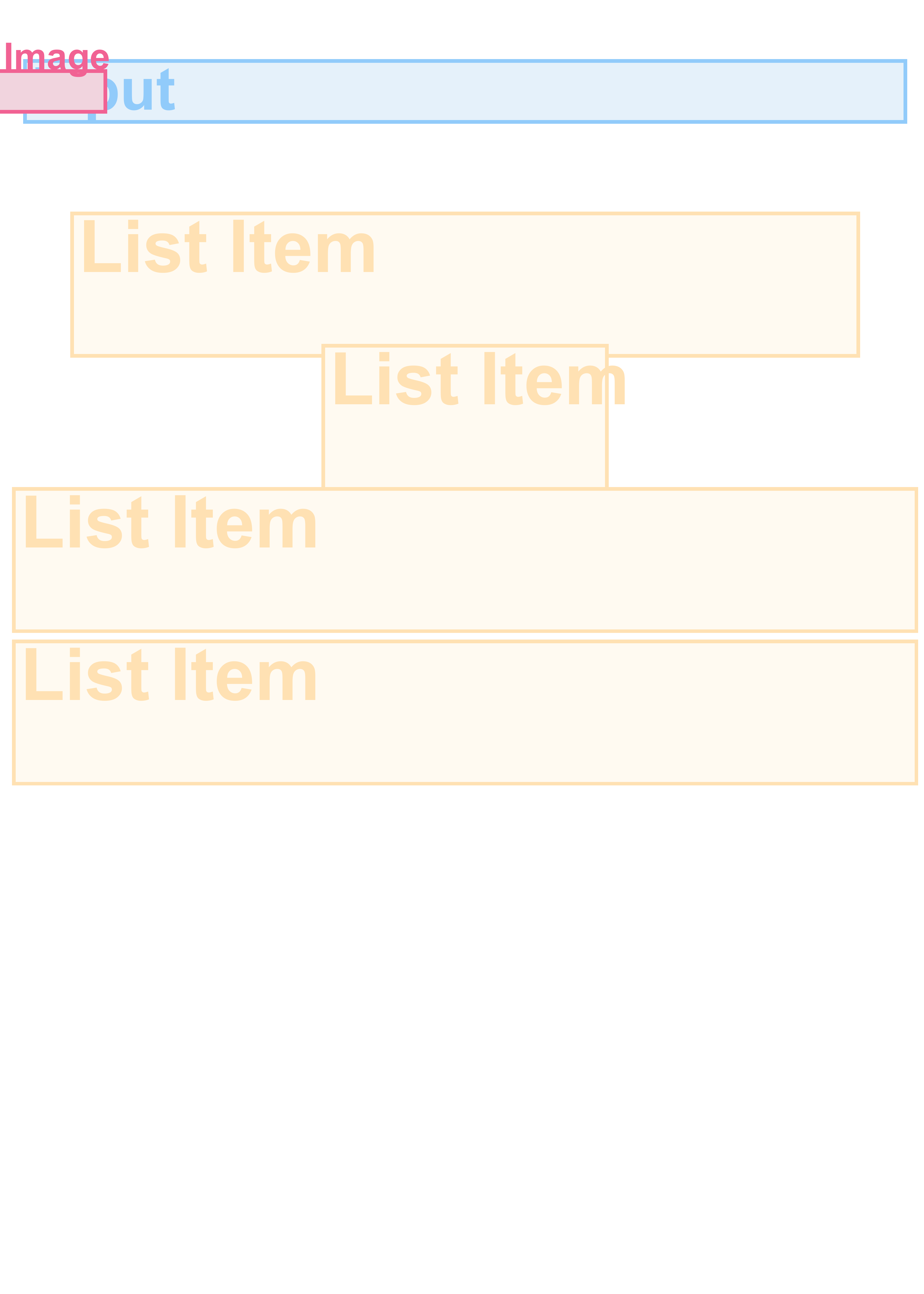} & 
    \includegraphics[width=\ricoBulkWidth]{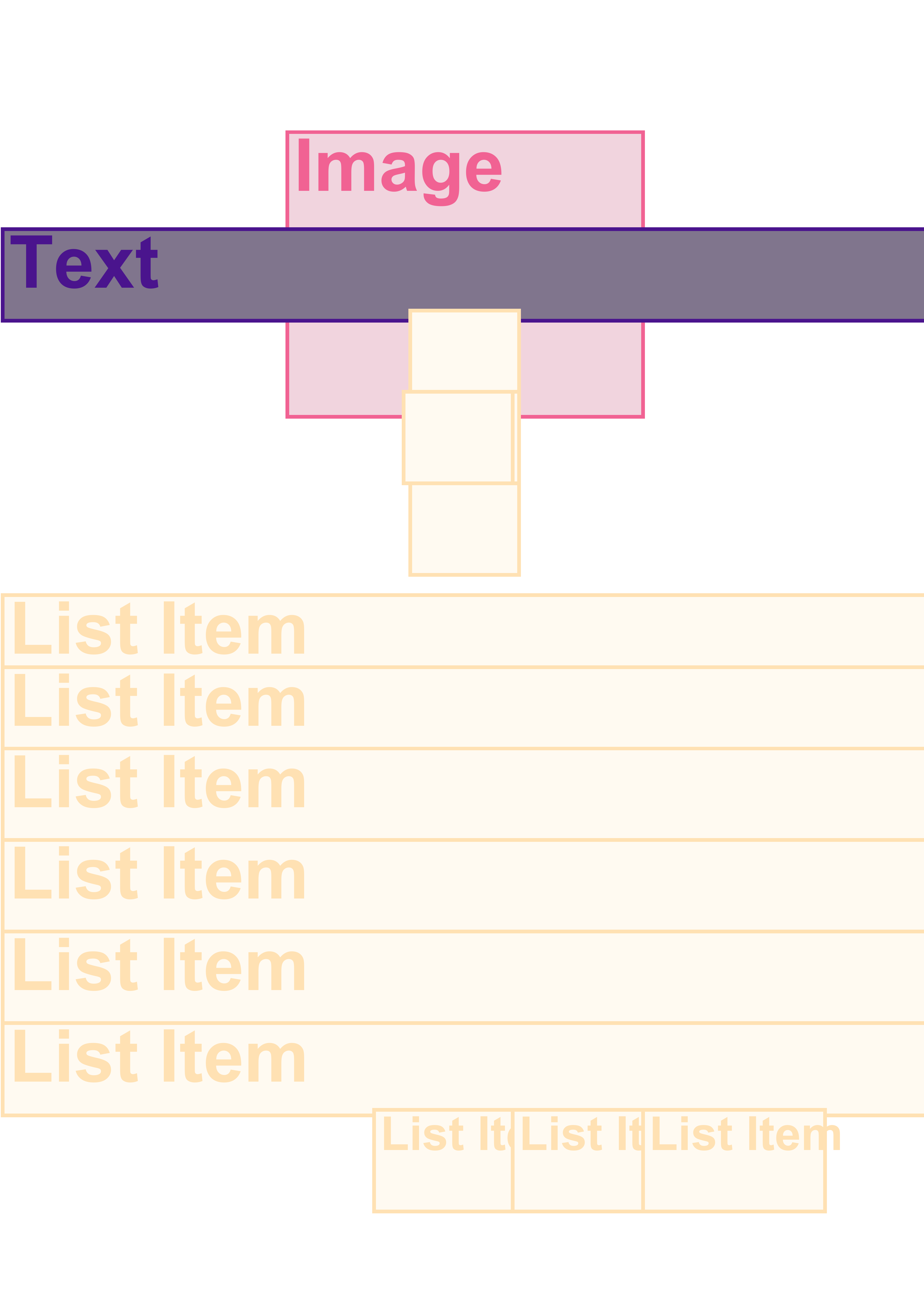} & 
    \includegraphics[width=\ricoBulkWidth]{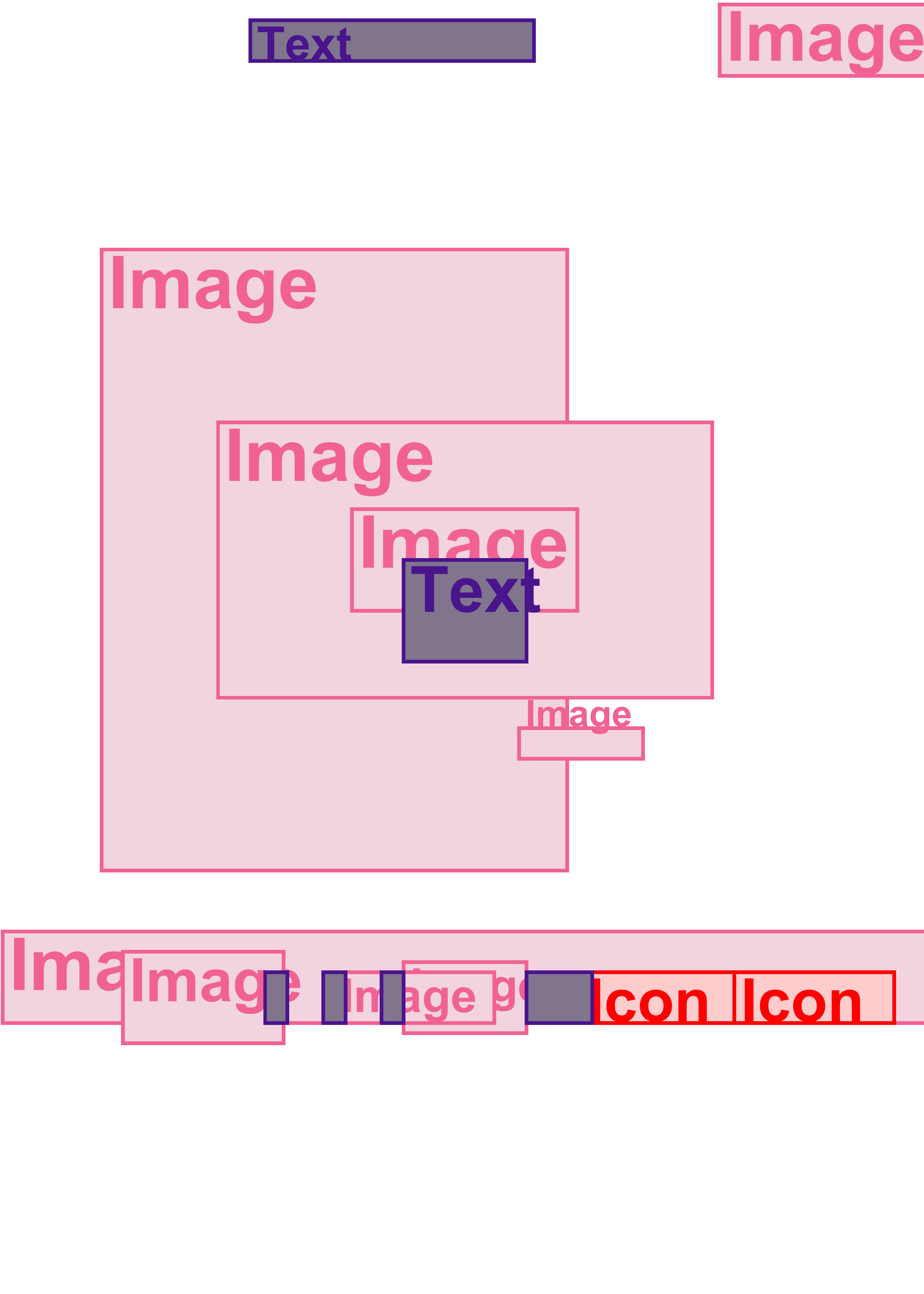} & 
    \includegraphics[width=\ricoBulkWidth]{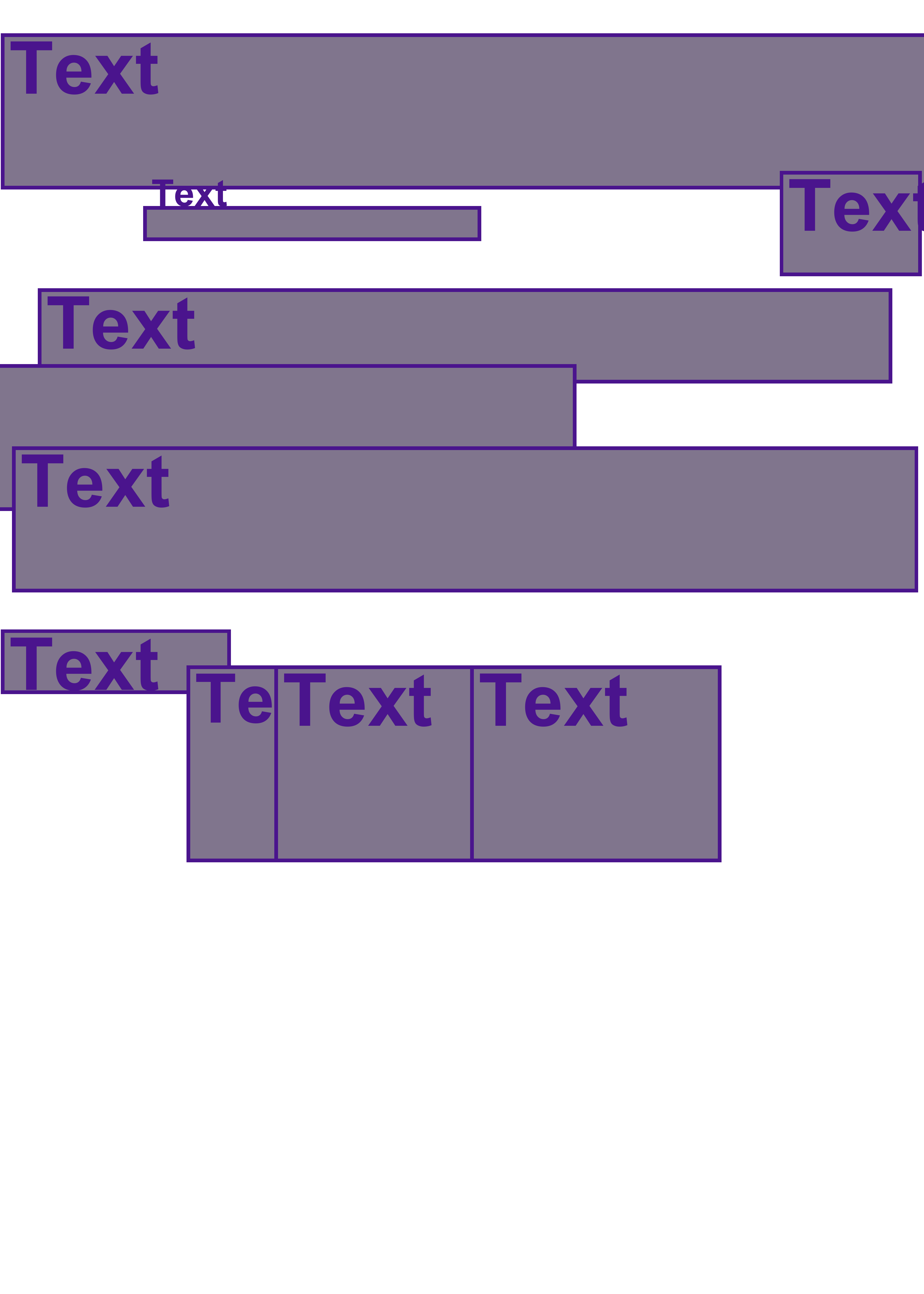} & 
    \includegraphics[width=\ricoBulkWidth]{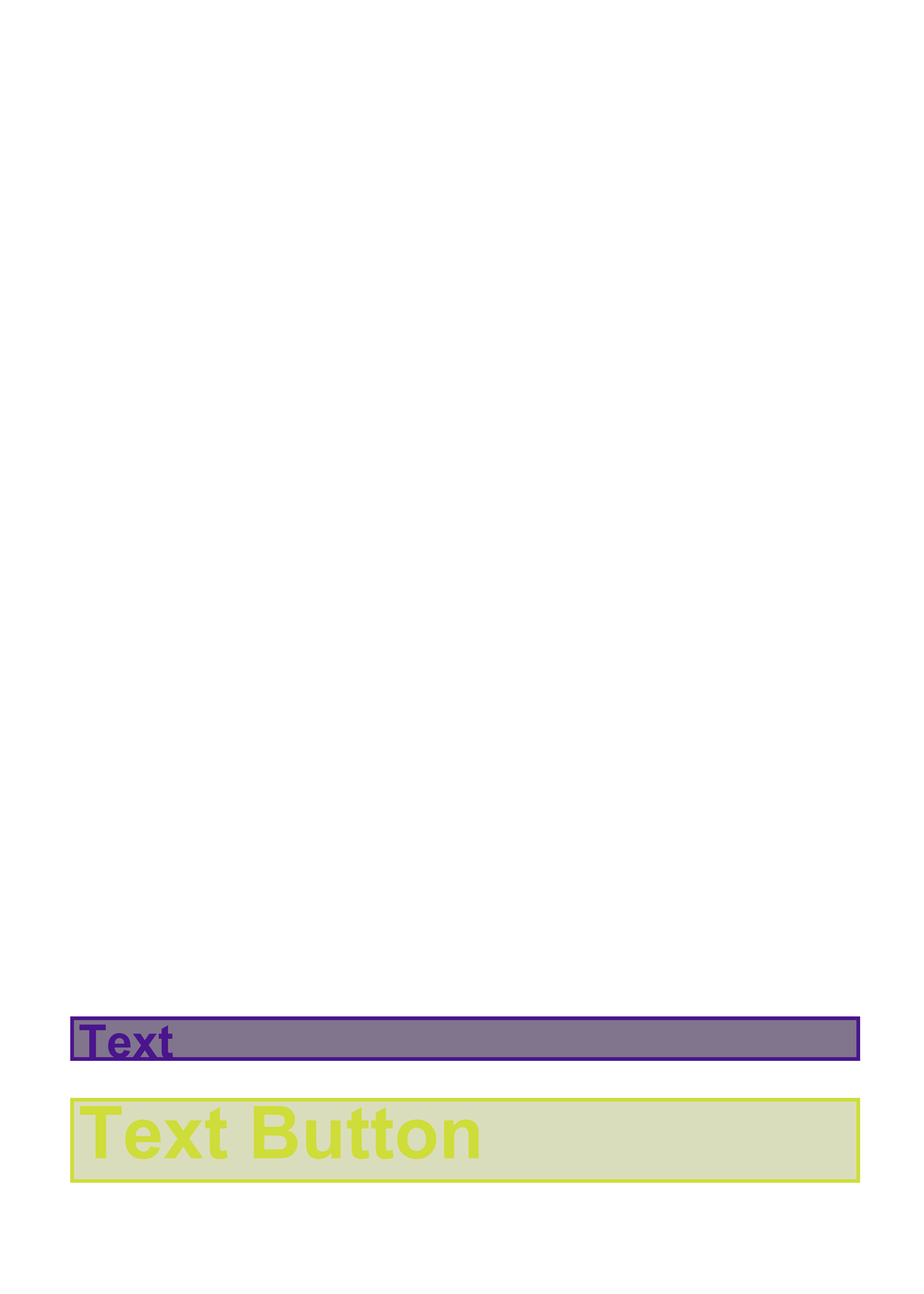} & 
    \includegraphics[width=\ricoBulkWidth]{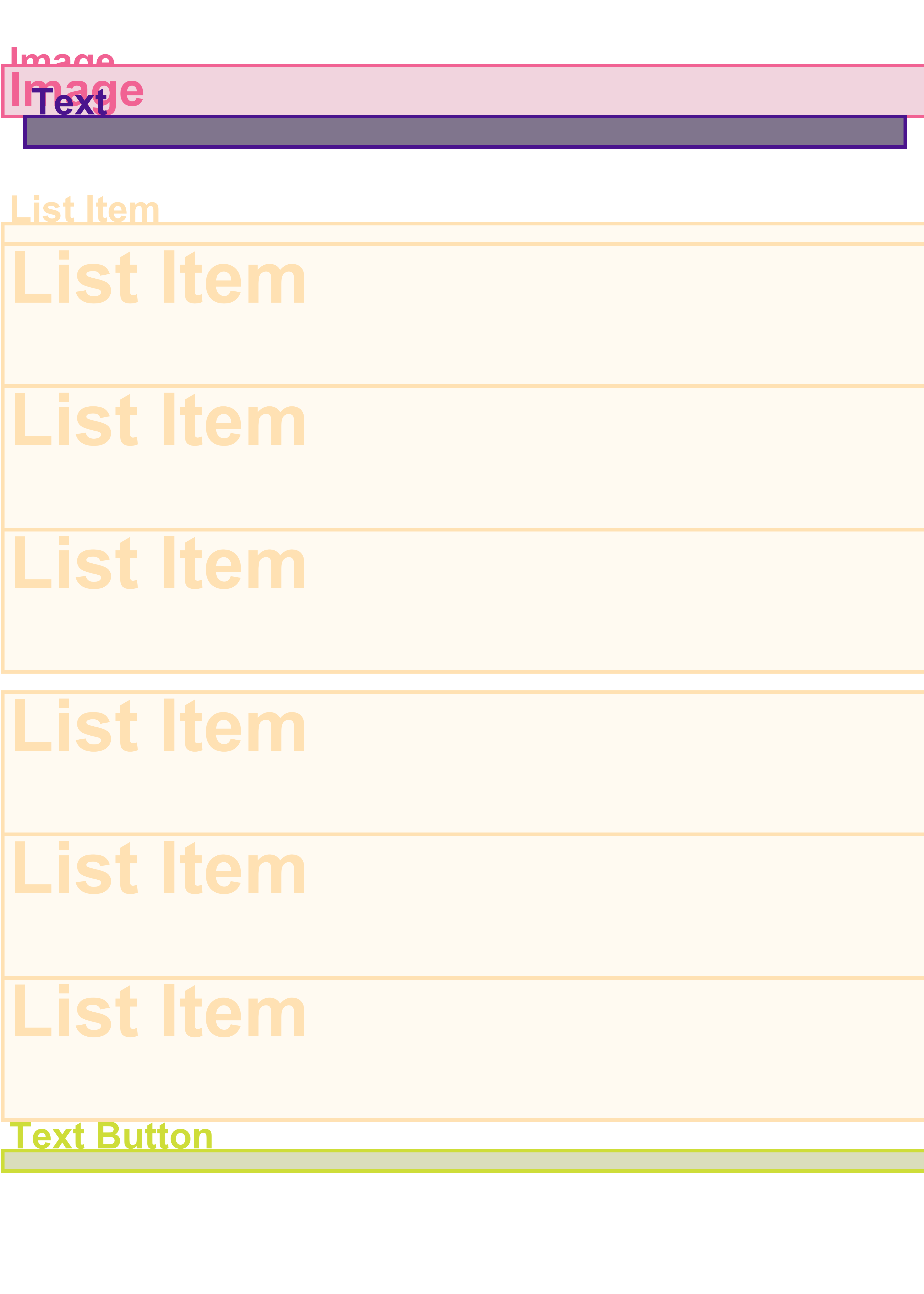} & 
    \includegraphics[width=\ricoBulkWidth]{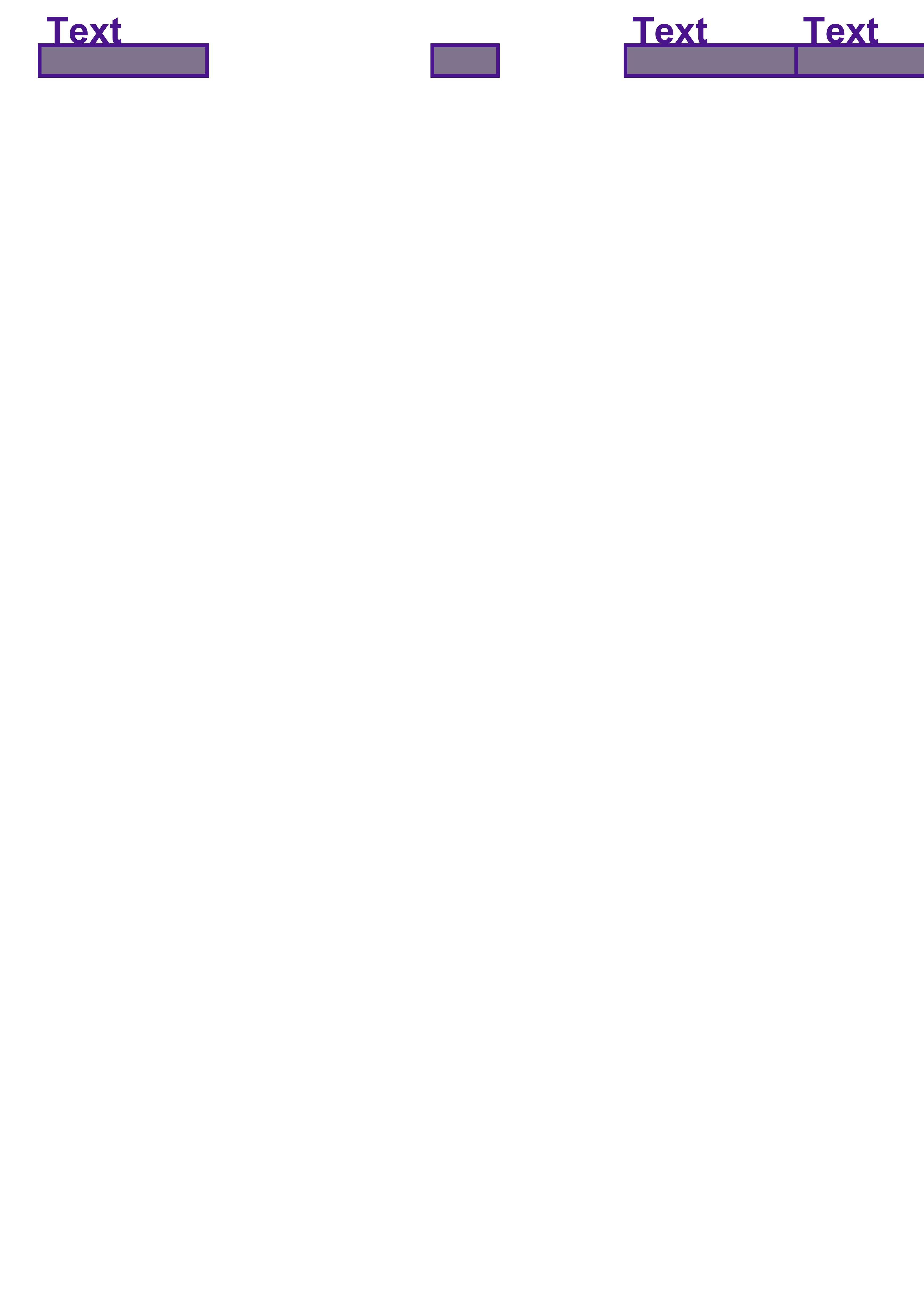} \\ 
    \includegraphics[width=\ricoBulkWidth]{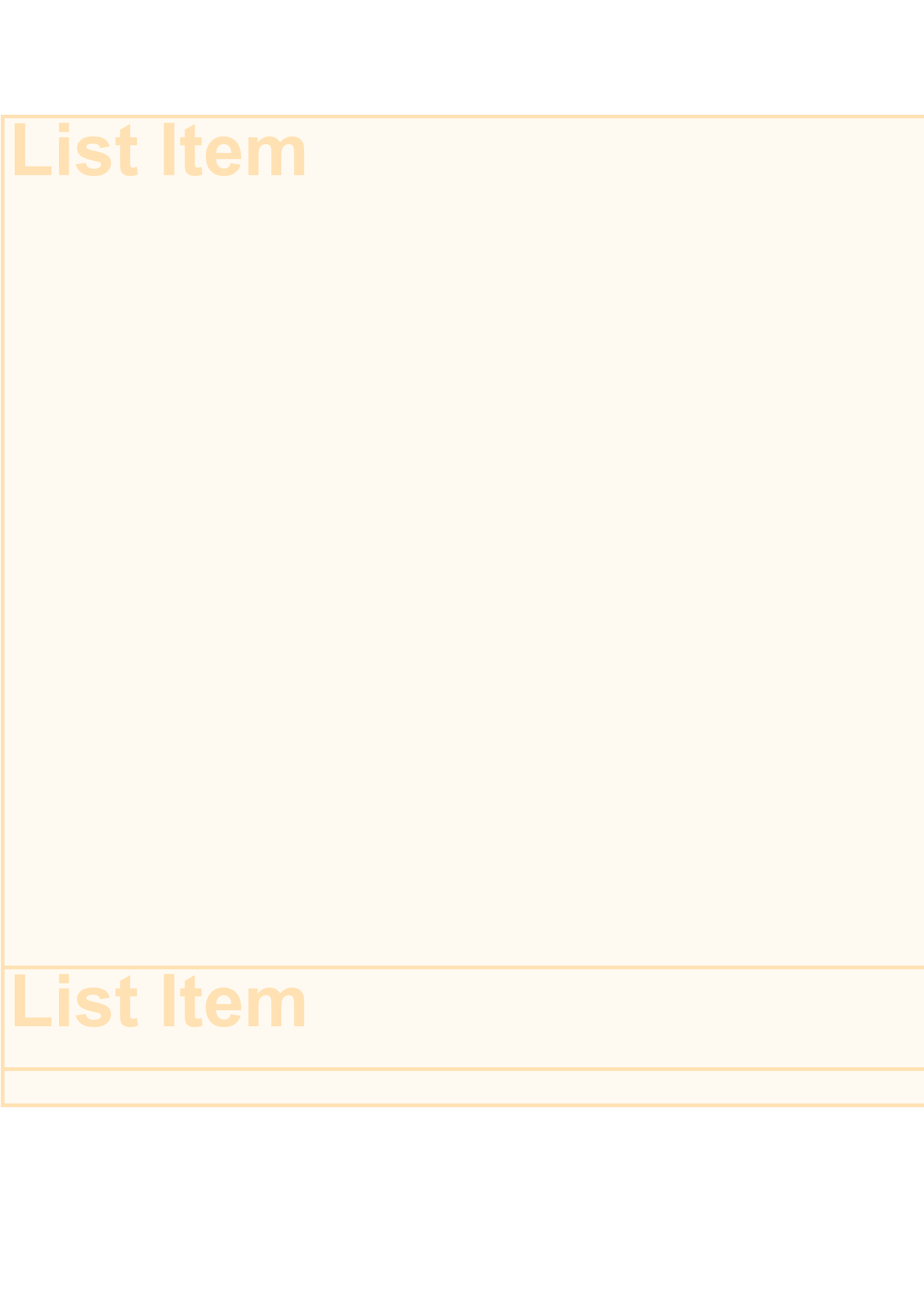} & 
    \includegraphics[width=\ricoBulkWidth]{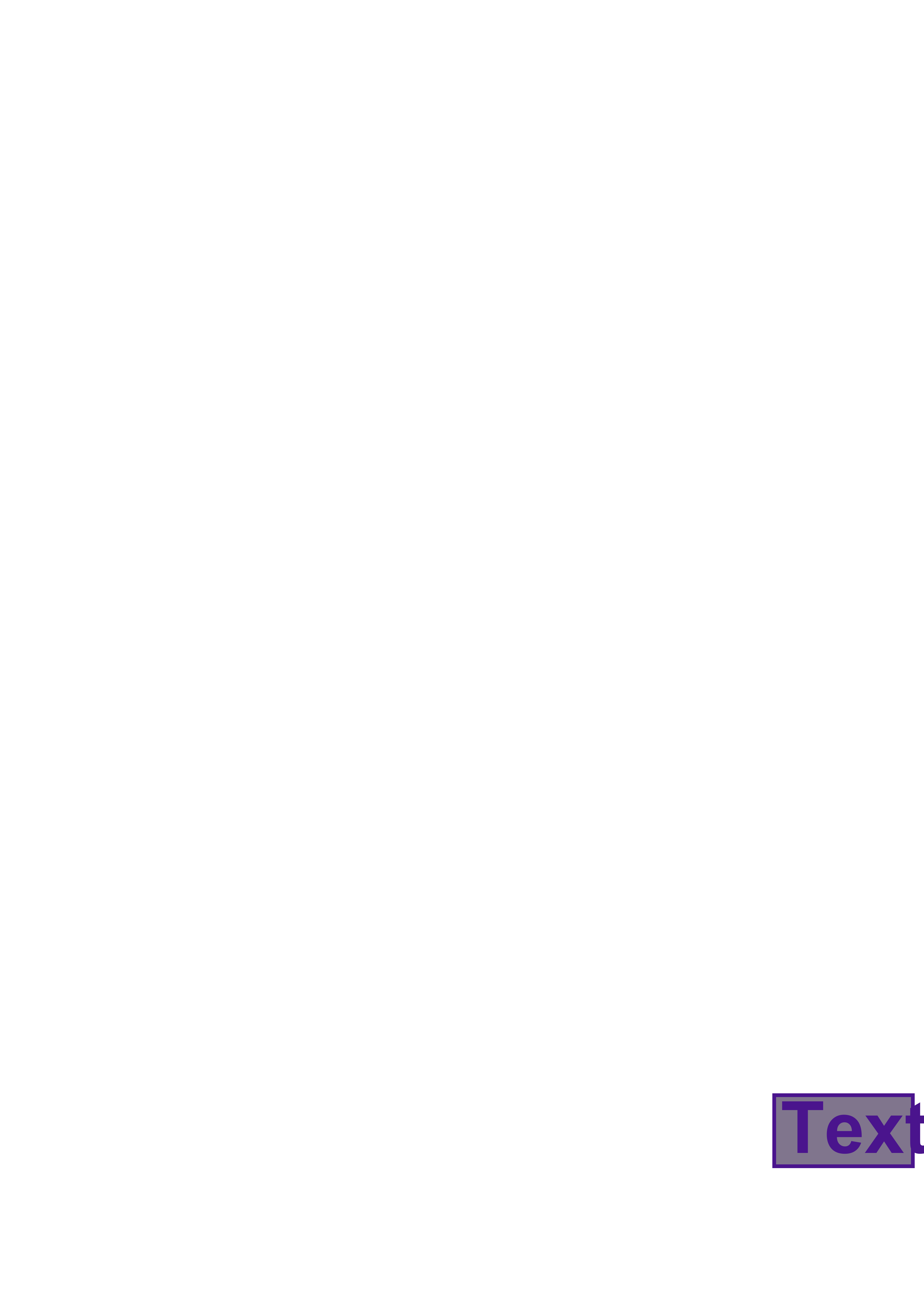} & 
    \includegraphics[width=\ricoBulkWidth]{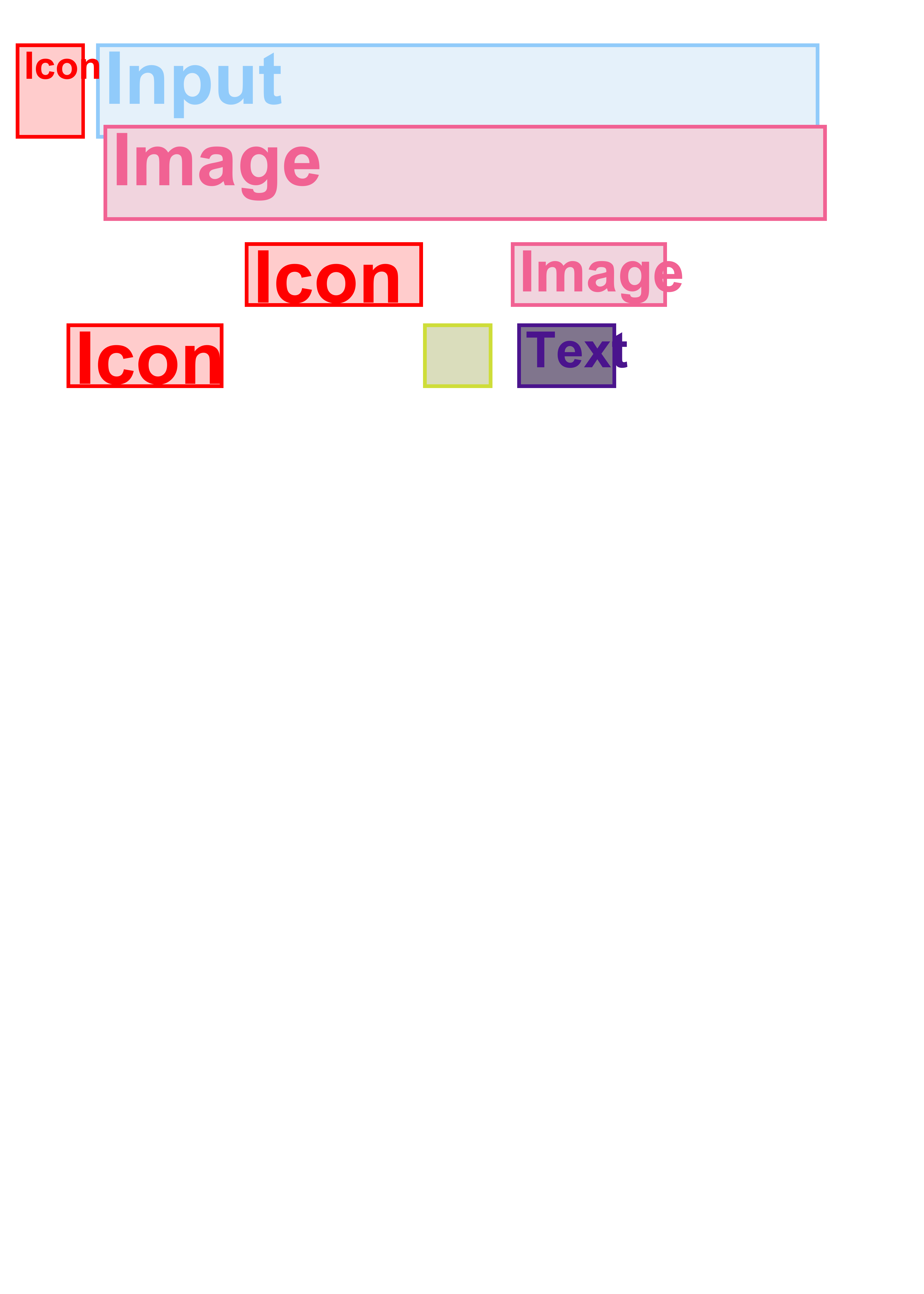} & 
    \includegraphics[width=\ricoBulkWidth]{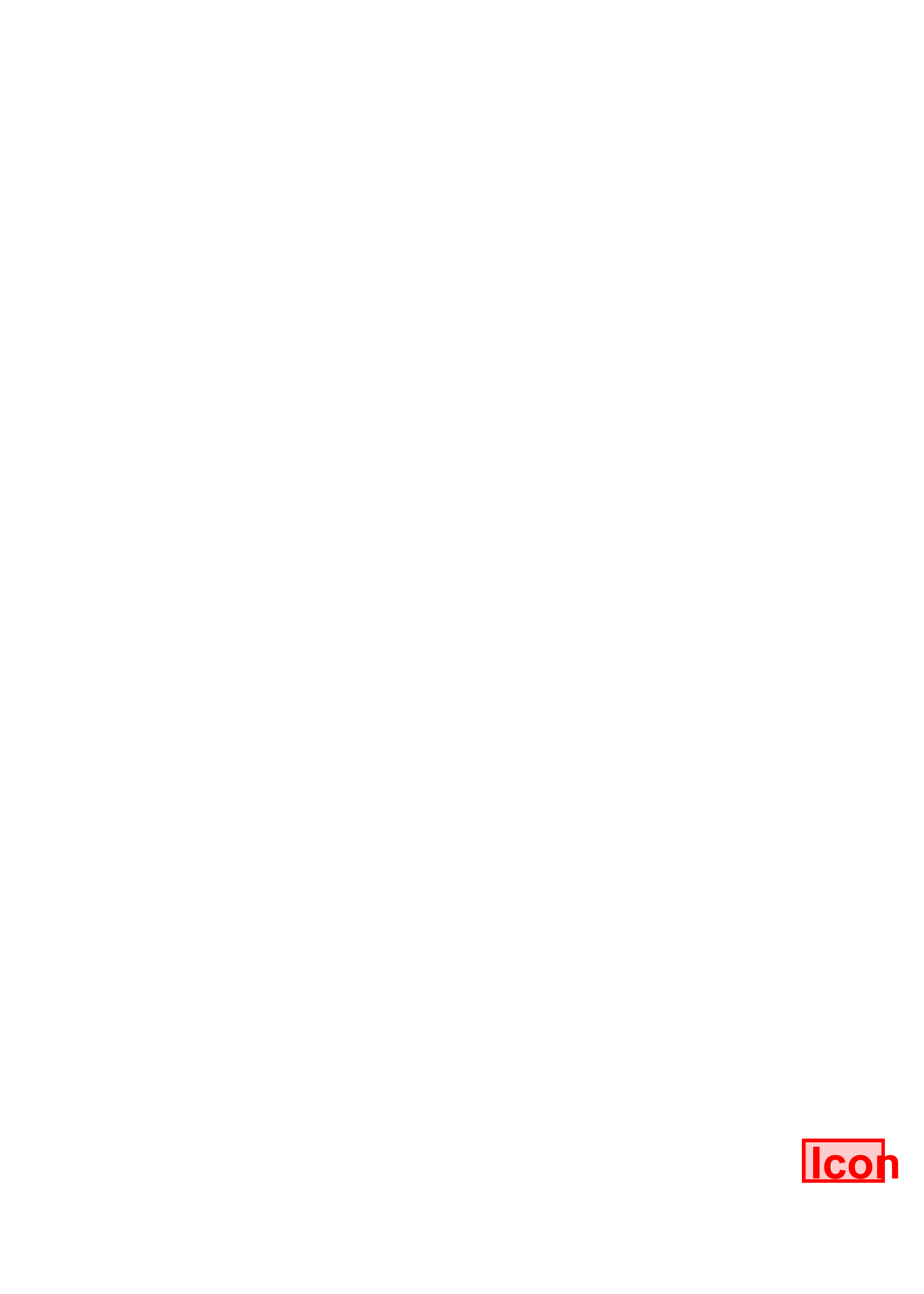} & 
    \includegraphics[width=\ricoBulkWidth]{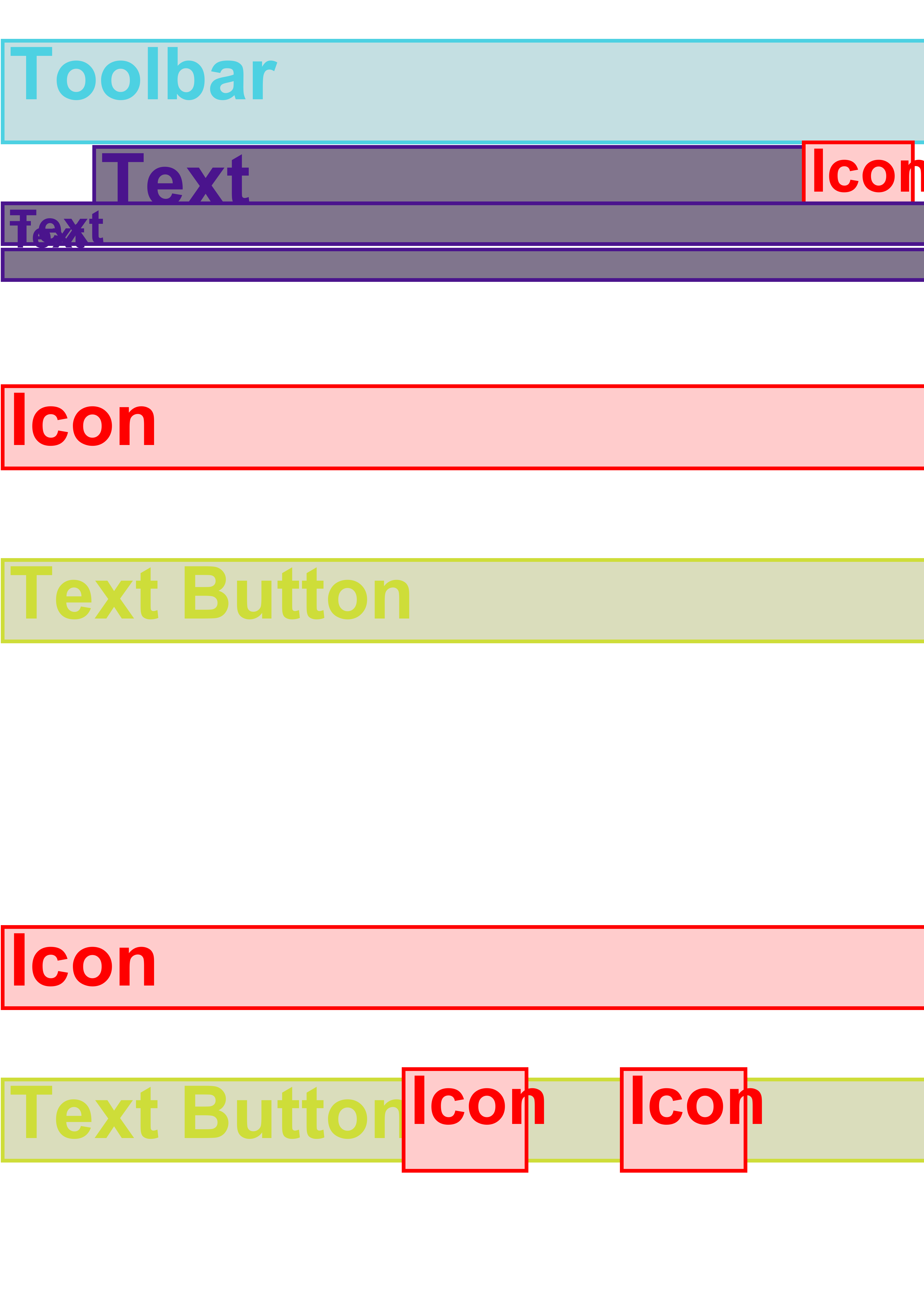} & 
    \includegraphics[width=\ricoBulkWidth]{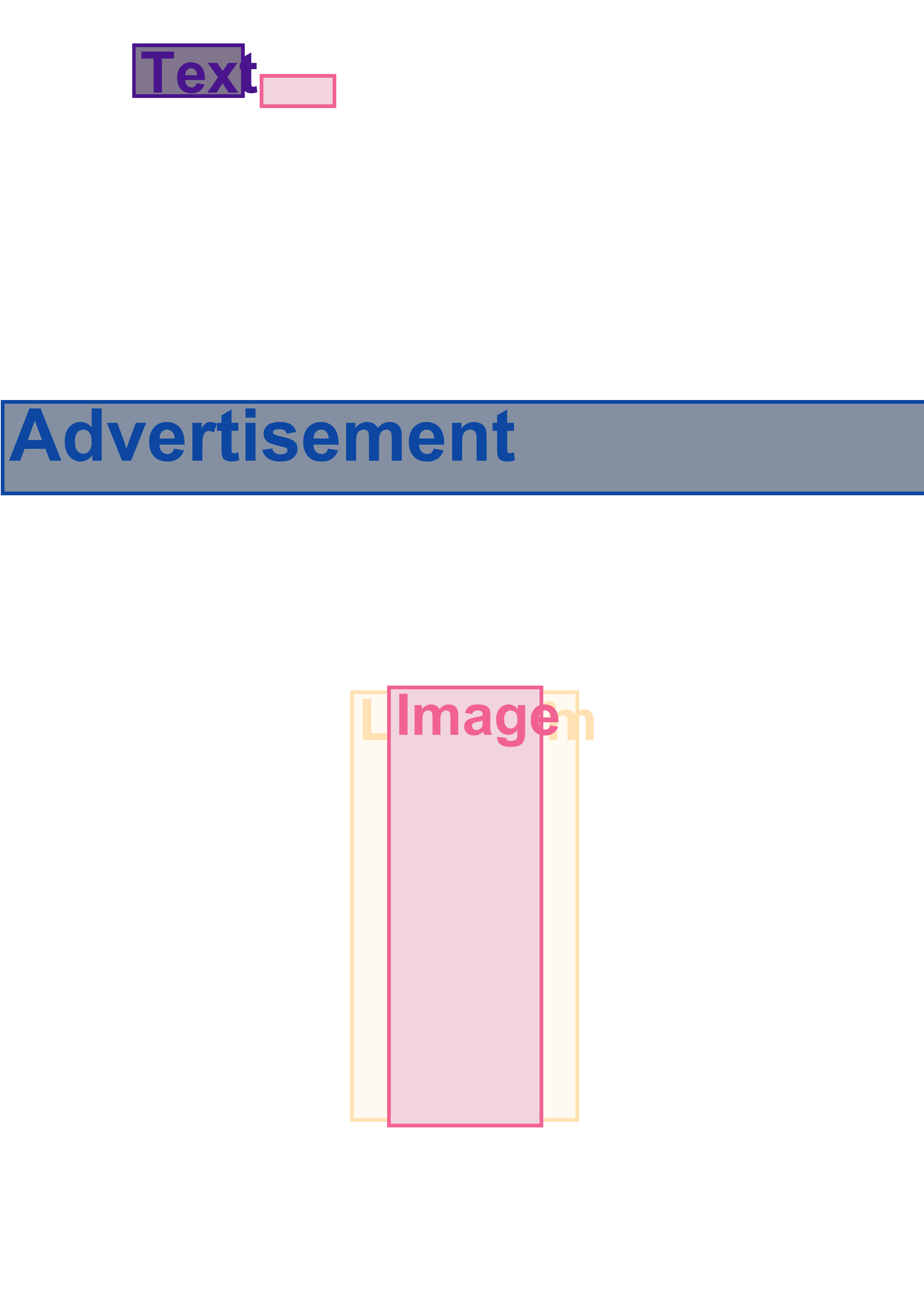} & 
    \includegraphics[width=\ricoBulkWidth]{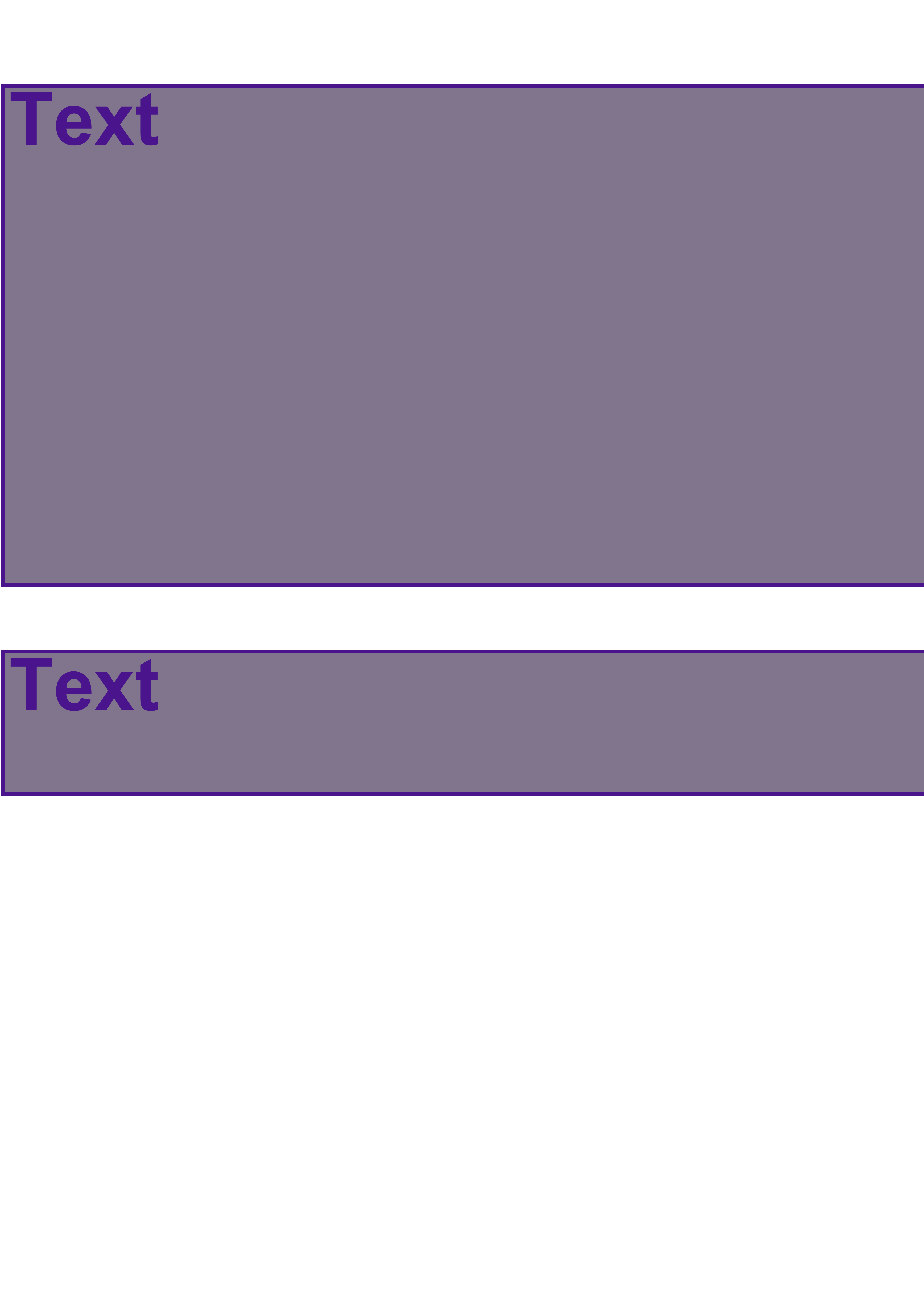} \\ 

    \end{tabular}
    \caption{(Cont.) Synthesized RICO examples.}
    \label{fig:bulk_rico_cont}
\end{figure}

\clearpage
\subsection{COCO-Stuff}

\begin{figure}[h]
    \centering
    \setlength{\tabcolsep}{2pt}
    \newlength{\cocoBulkWidth}
    \setlength{\cocoBulkWidth}{0.16\linewidth}
    \begin{tabular}{ccccccc}
\rotatebox{90}{\hspace{2mm}LayoutVAE [16]}&
\includegraphics[width=\cocoBulkWidth]{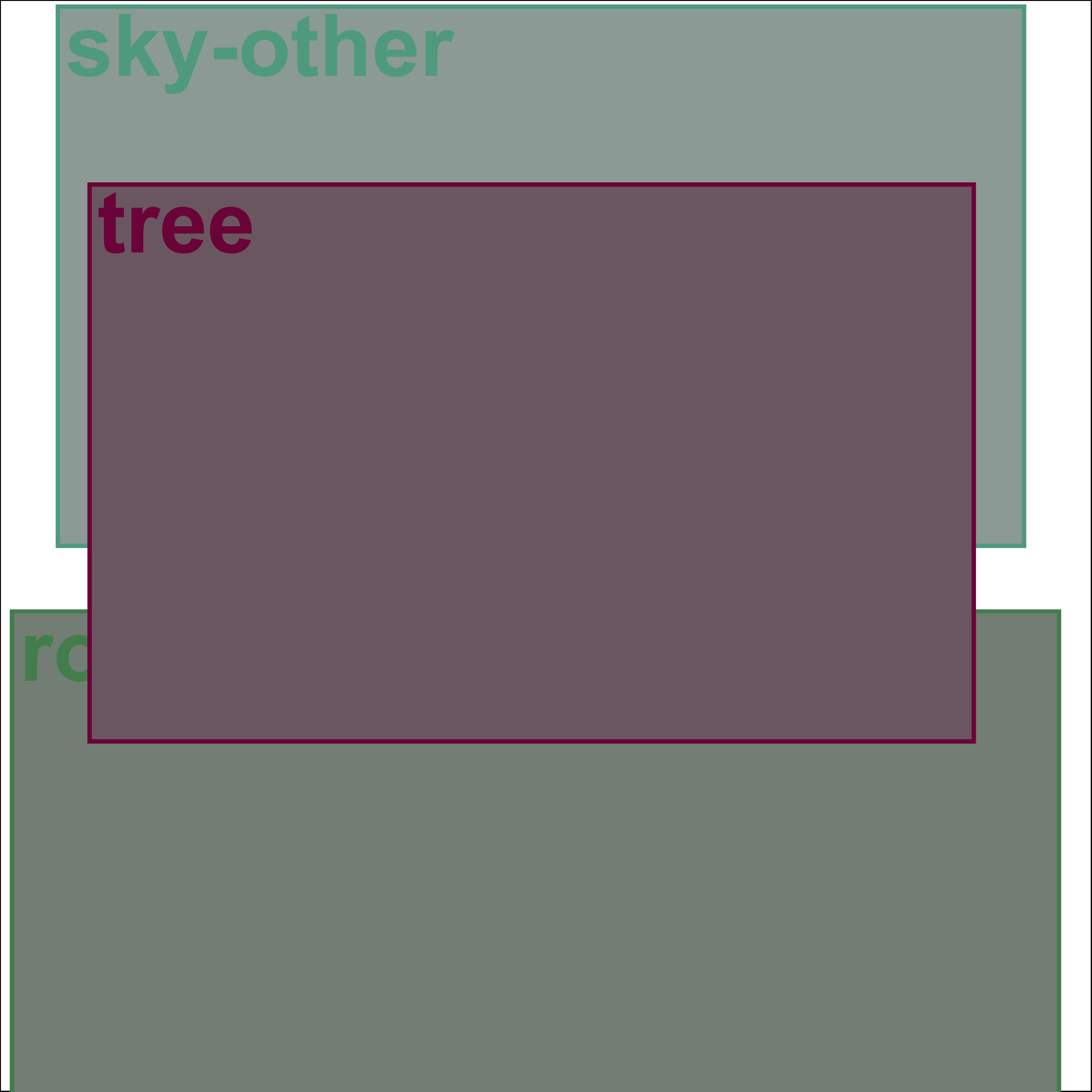} &
\includegraphics[width=\cocoBulkWidth]{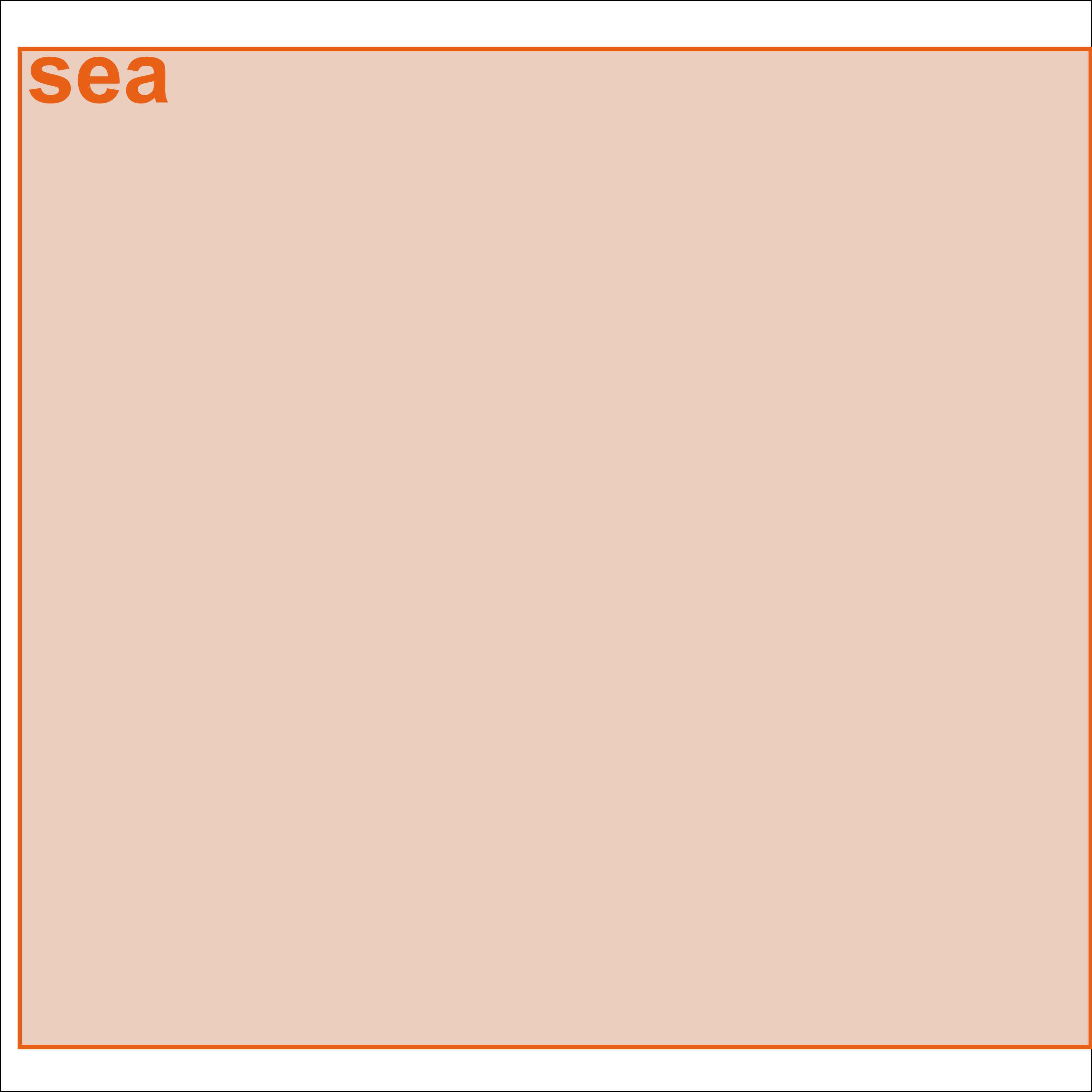} &
\includegraphics[width=\cocoBulkWidth]{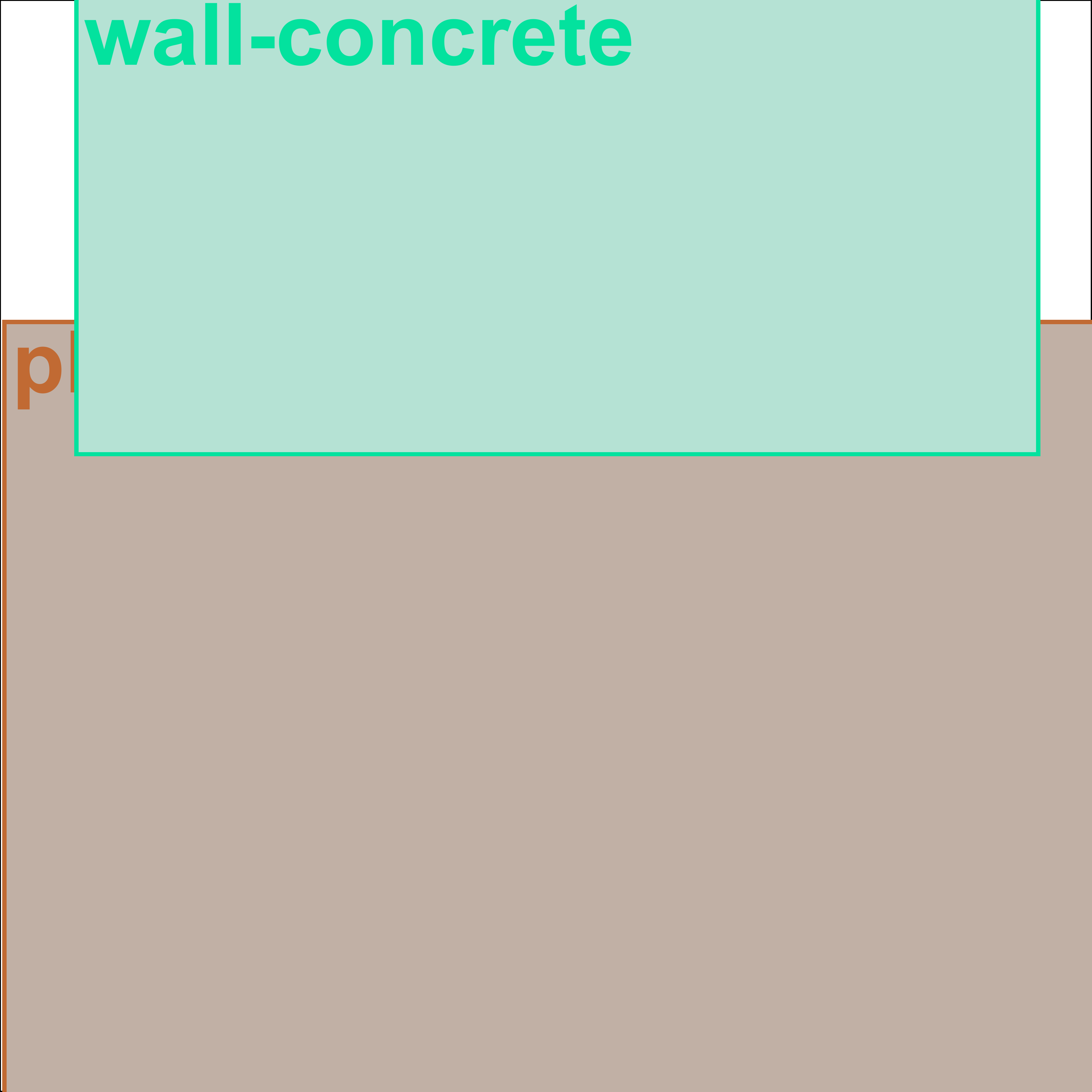} &
\includegraphics[width=\cocoBulkWidth]{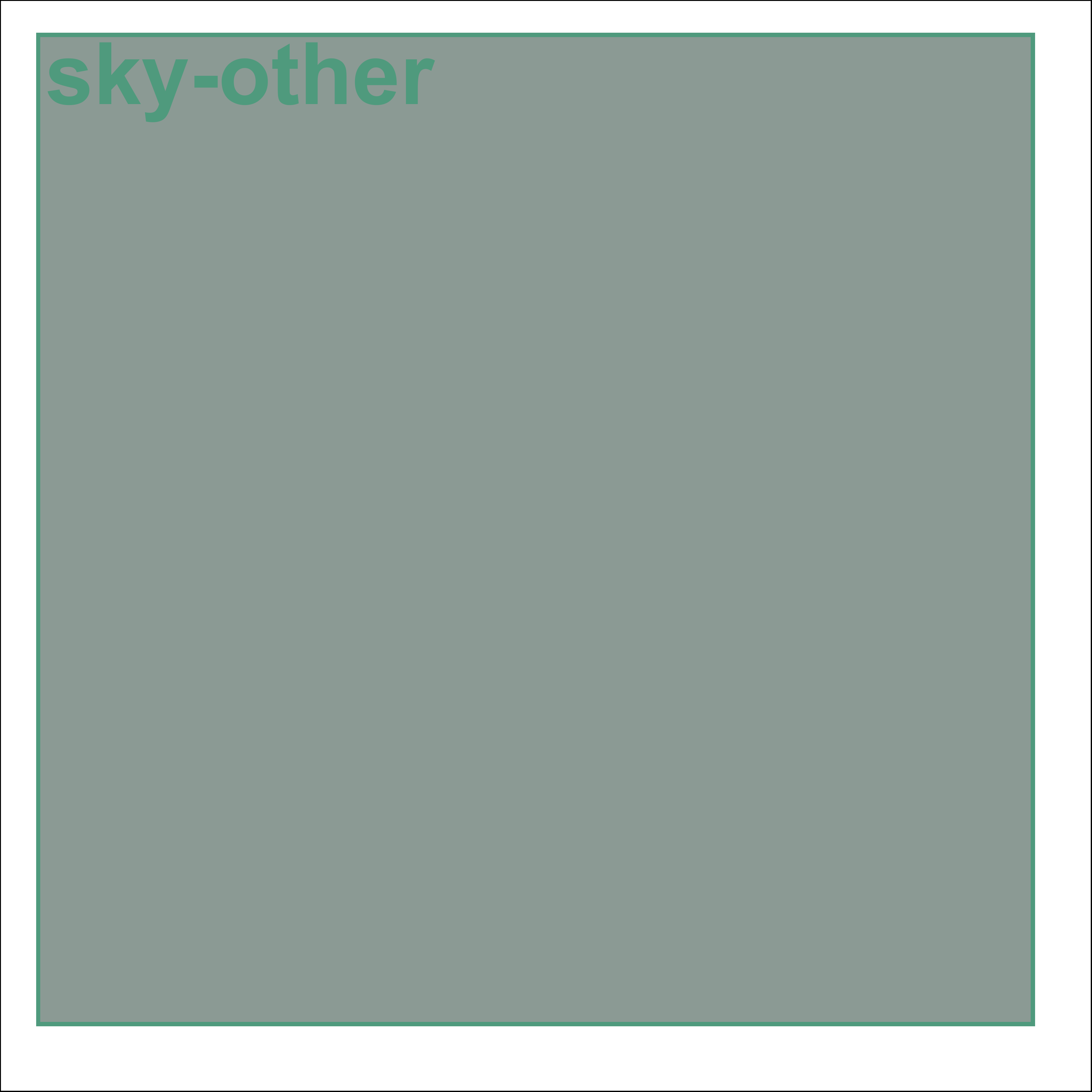} &
\includegraphics[width=\cocoBulkWidth]{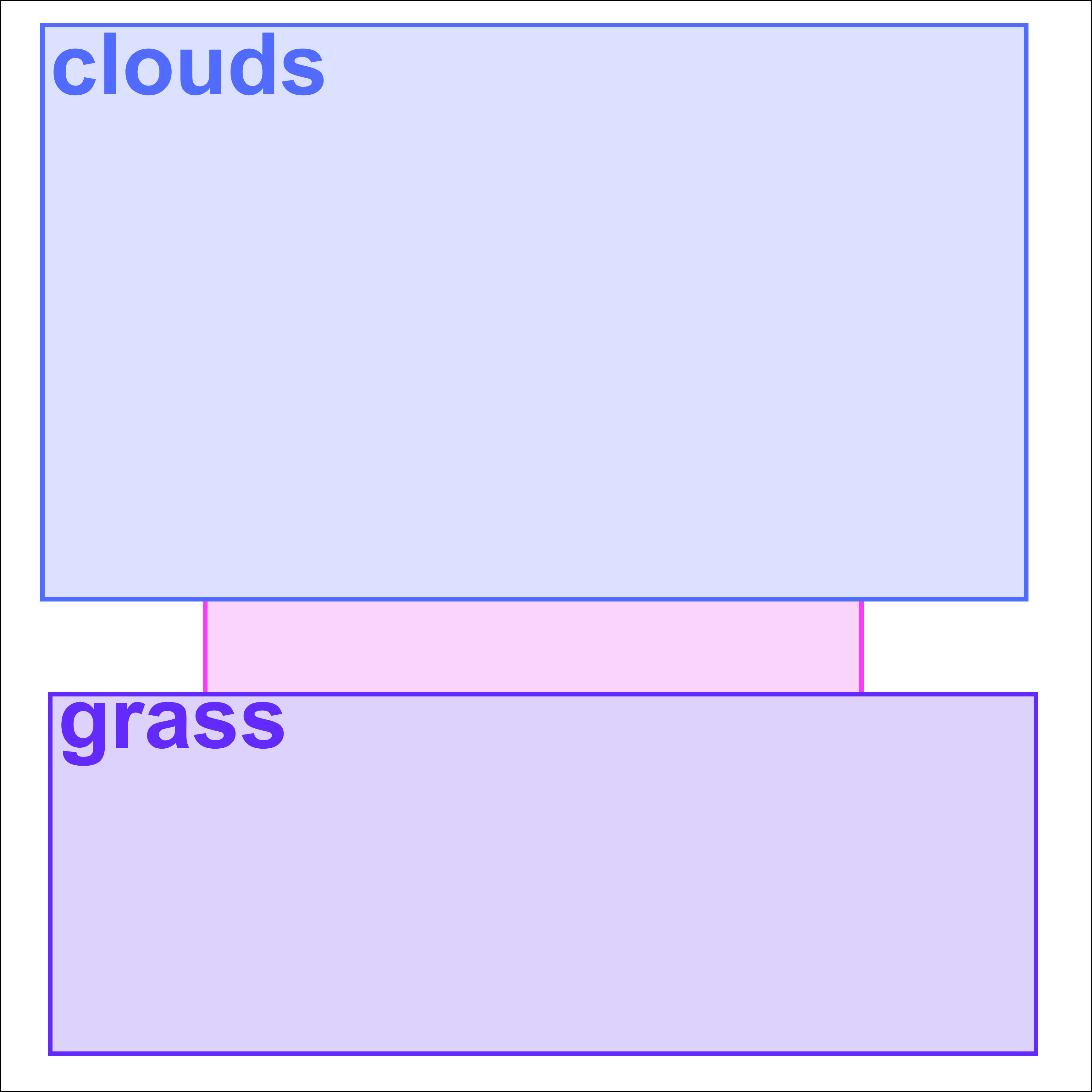} &
\includegraphics[width=\cocoBulkWidth]{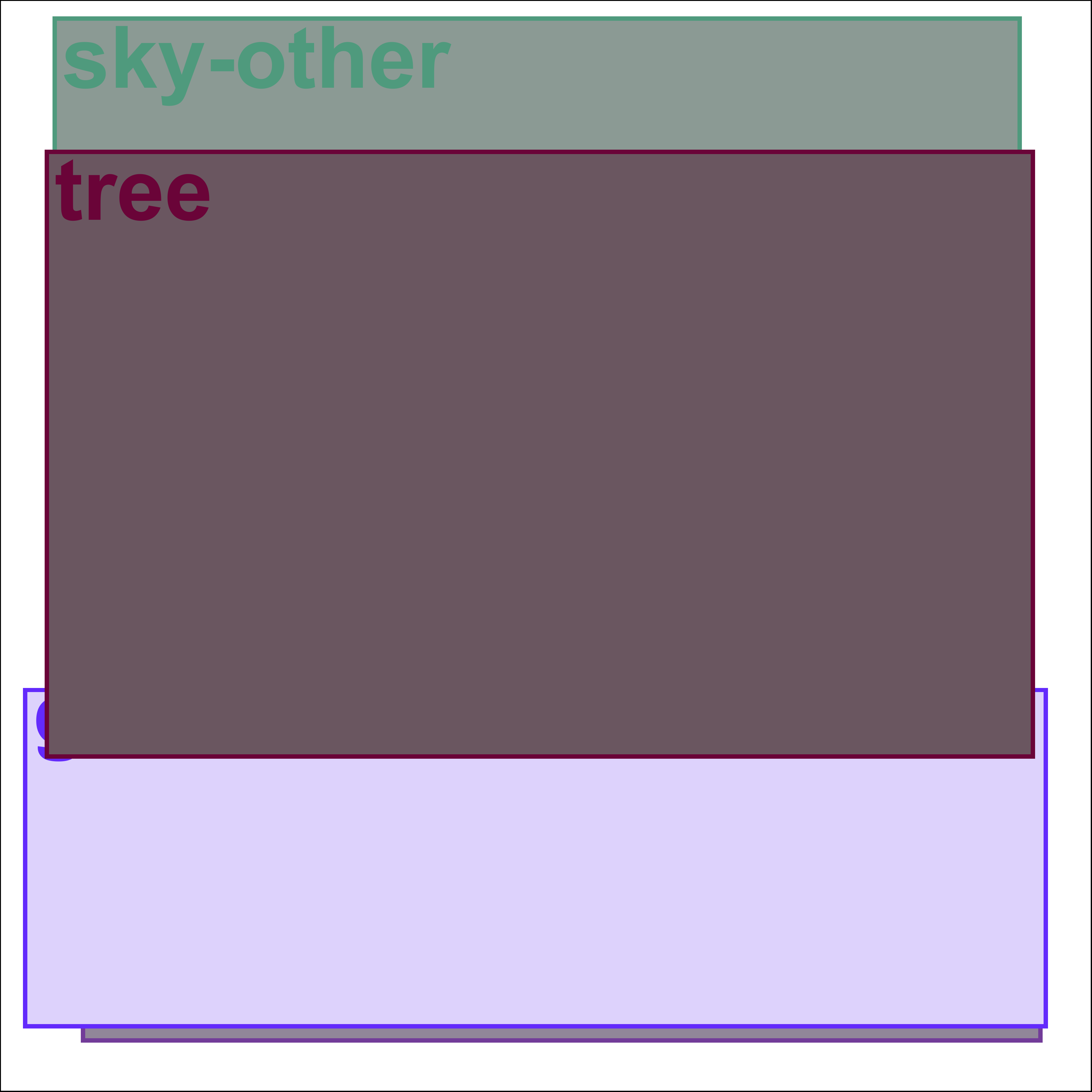} \\

\rotatebox{90}{\hspace{3mm}Gupta \etal [9]}&
\includegraphics[width=\cocoBulkWidth]{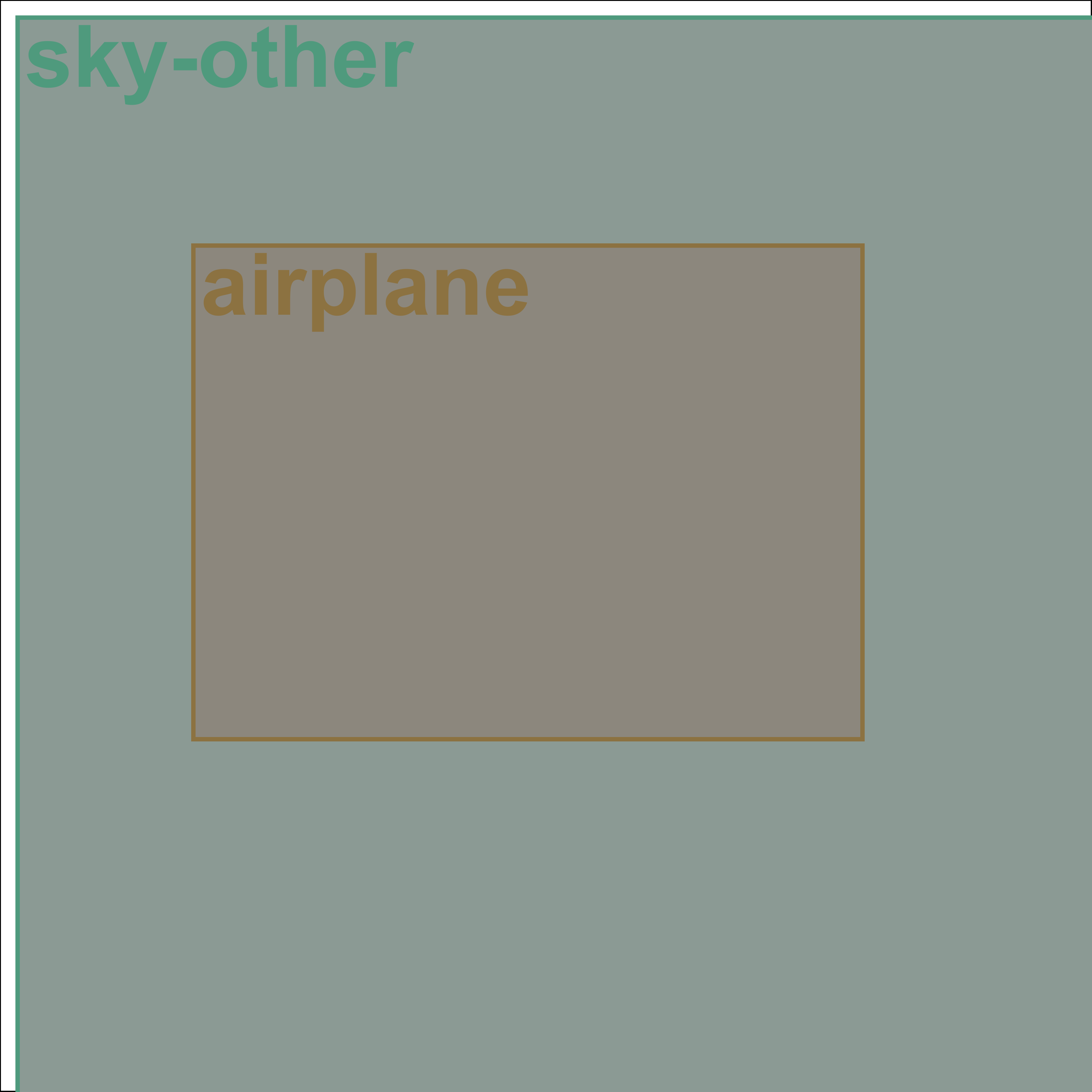}&
\includegraphics[width=\cocoBulkWidth]{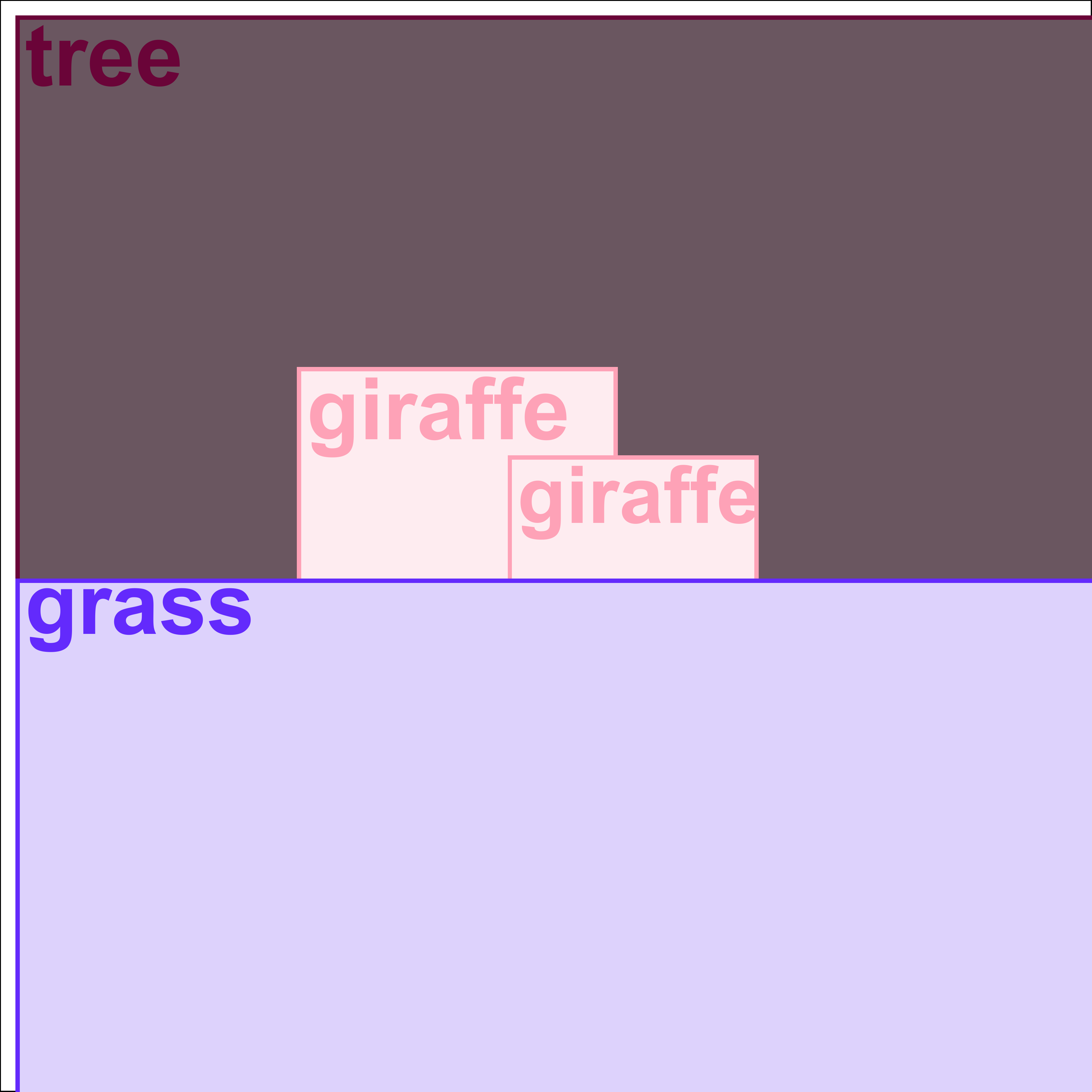}&
\includegraphics[width=\cocoBulkWidth]{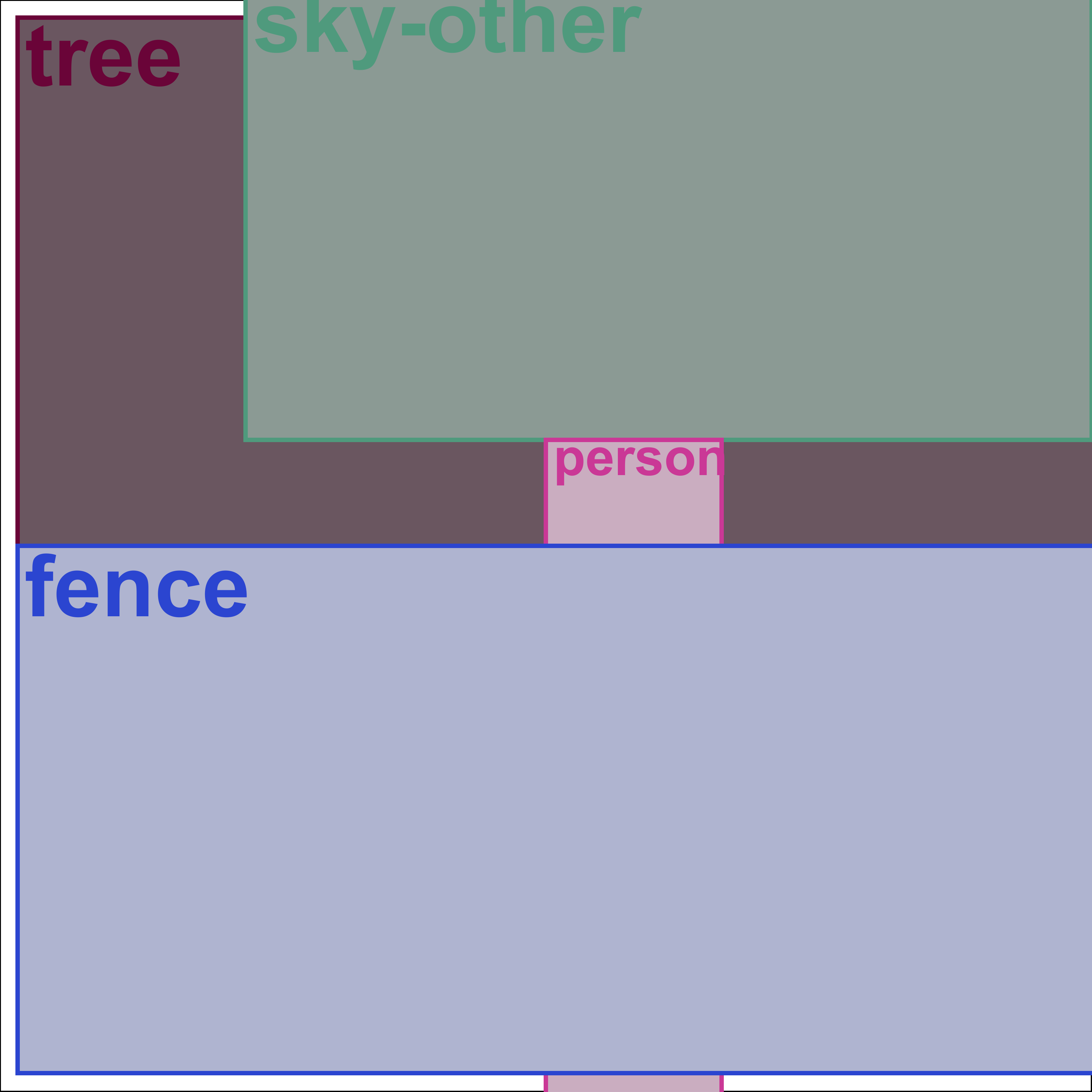}&
\includegraphics[width=\cocoBulkWidth]{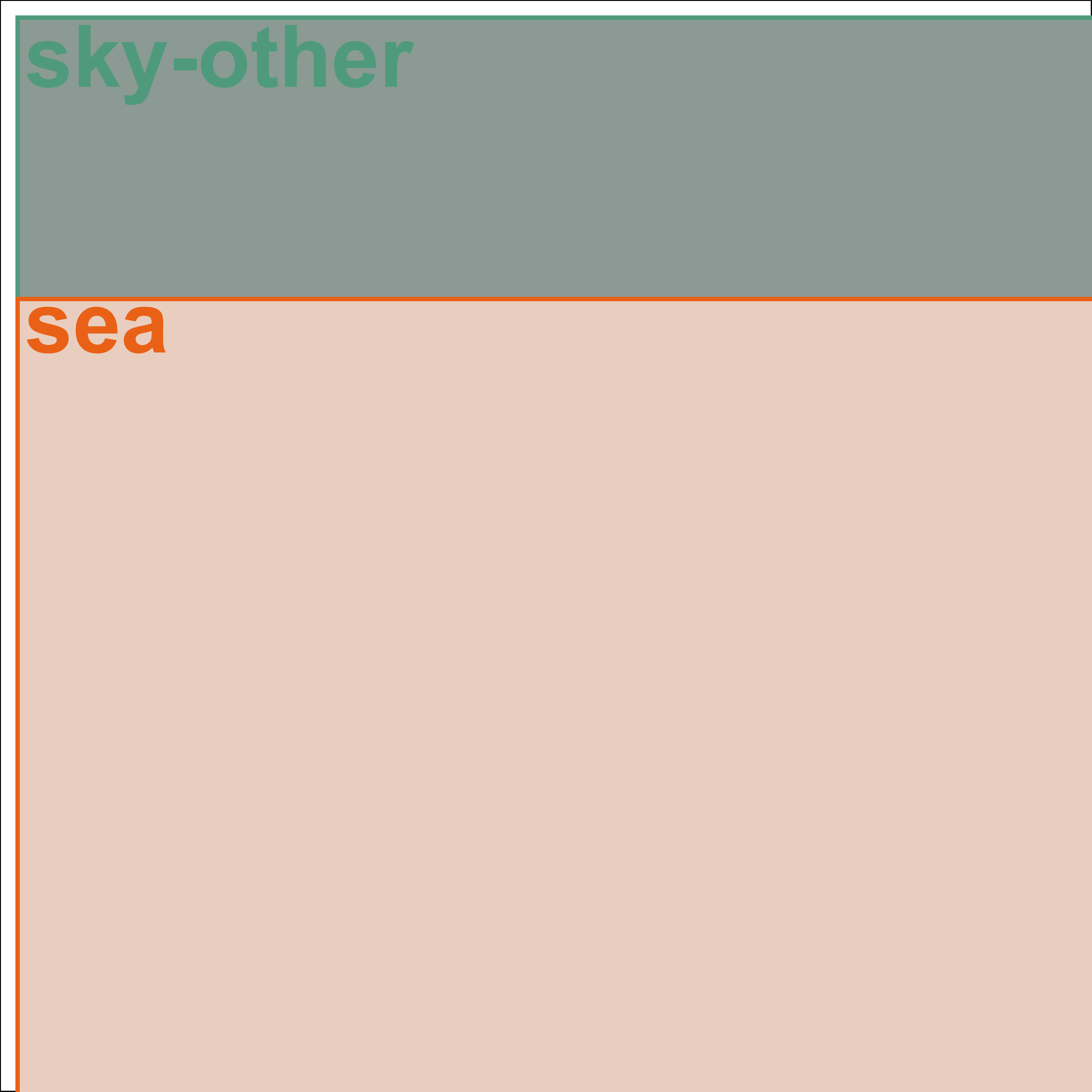}&
\includegraphics[width=\cocoBulkWidth]{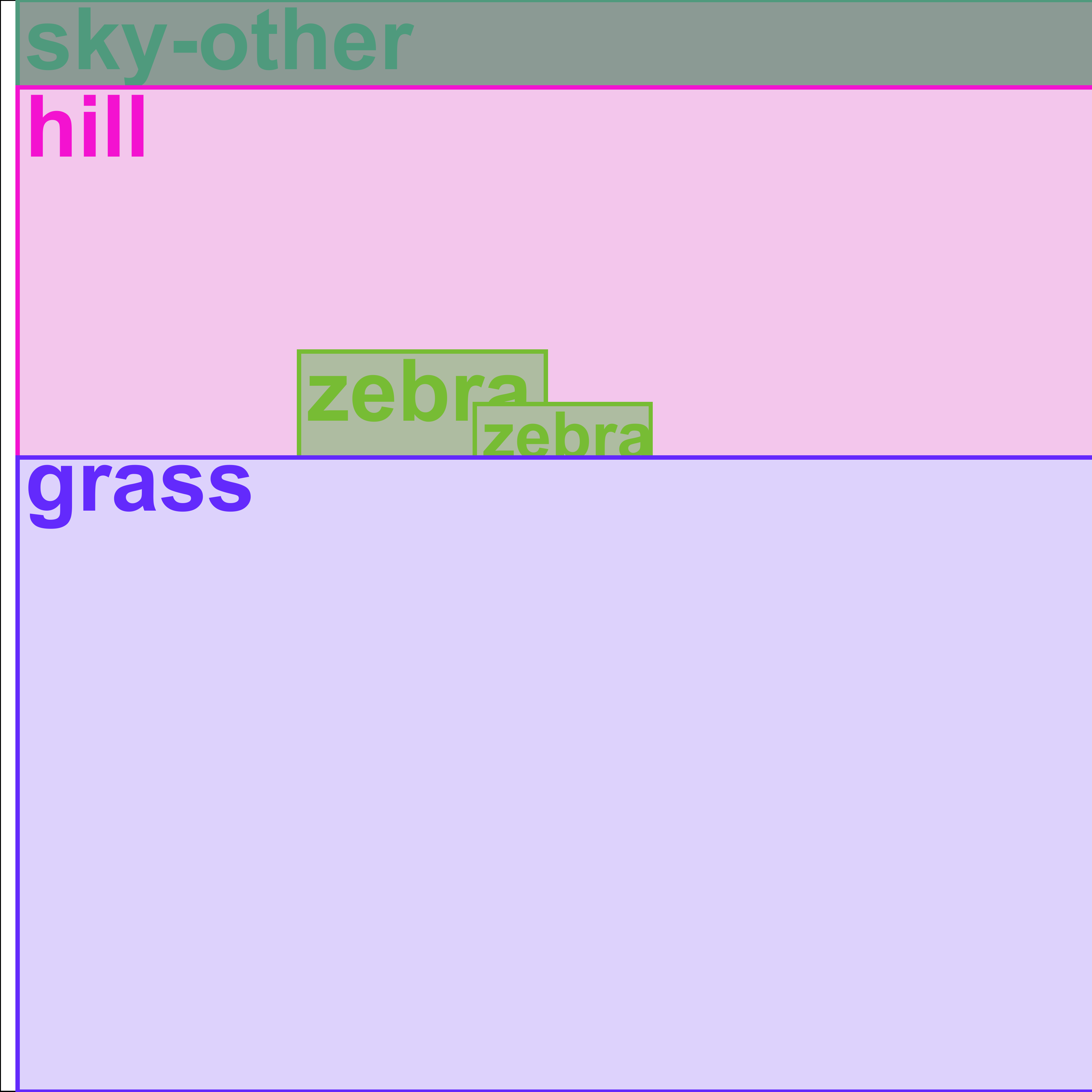}&
\includegraphics[width=\cocoBulkWidth]{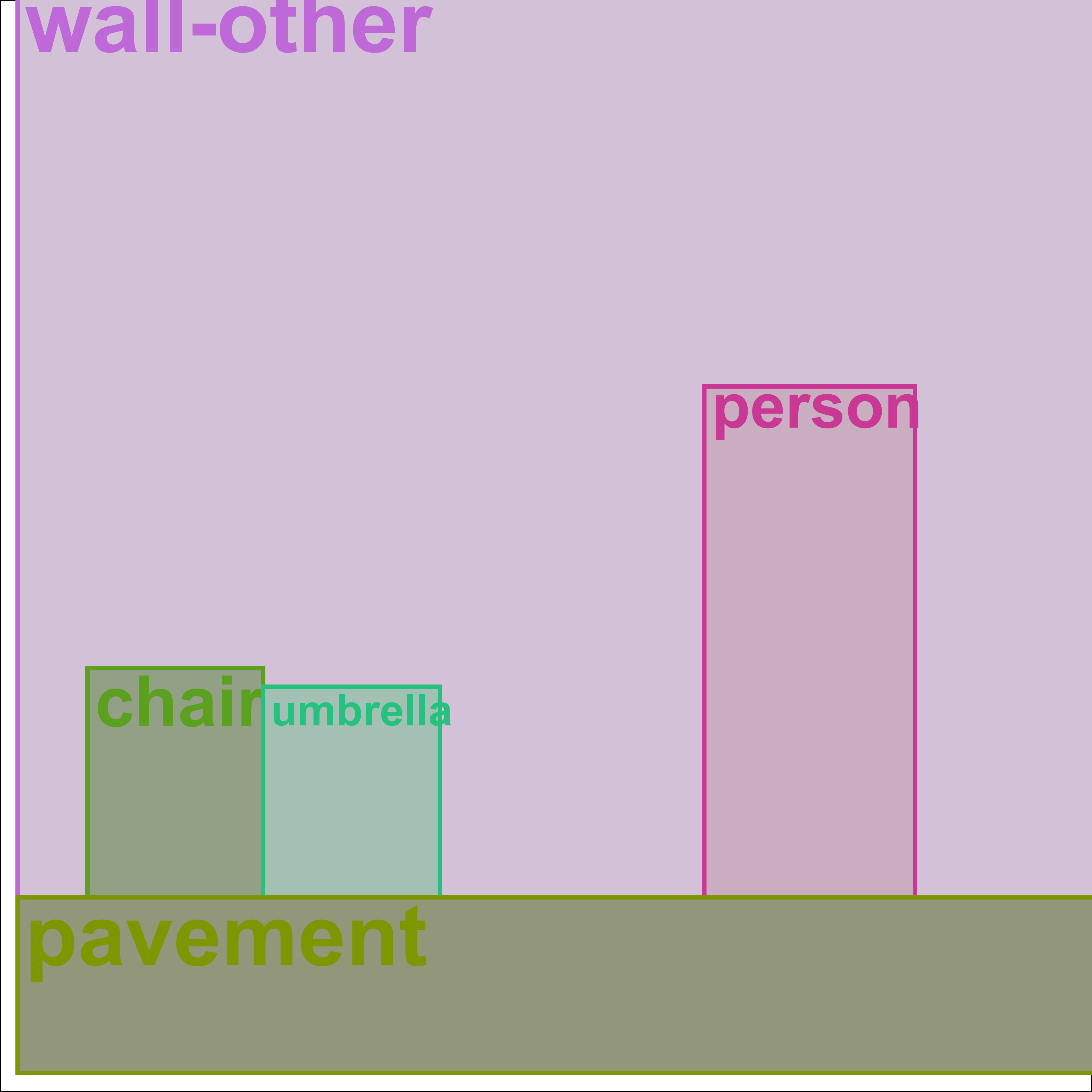}\\\toprule

\rotatebox{90}{\hspace{9mm}Ours}&\includegraphics[width=\cocoBulkWidth]{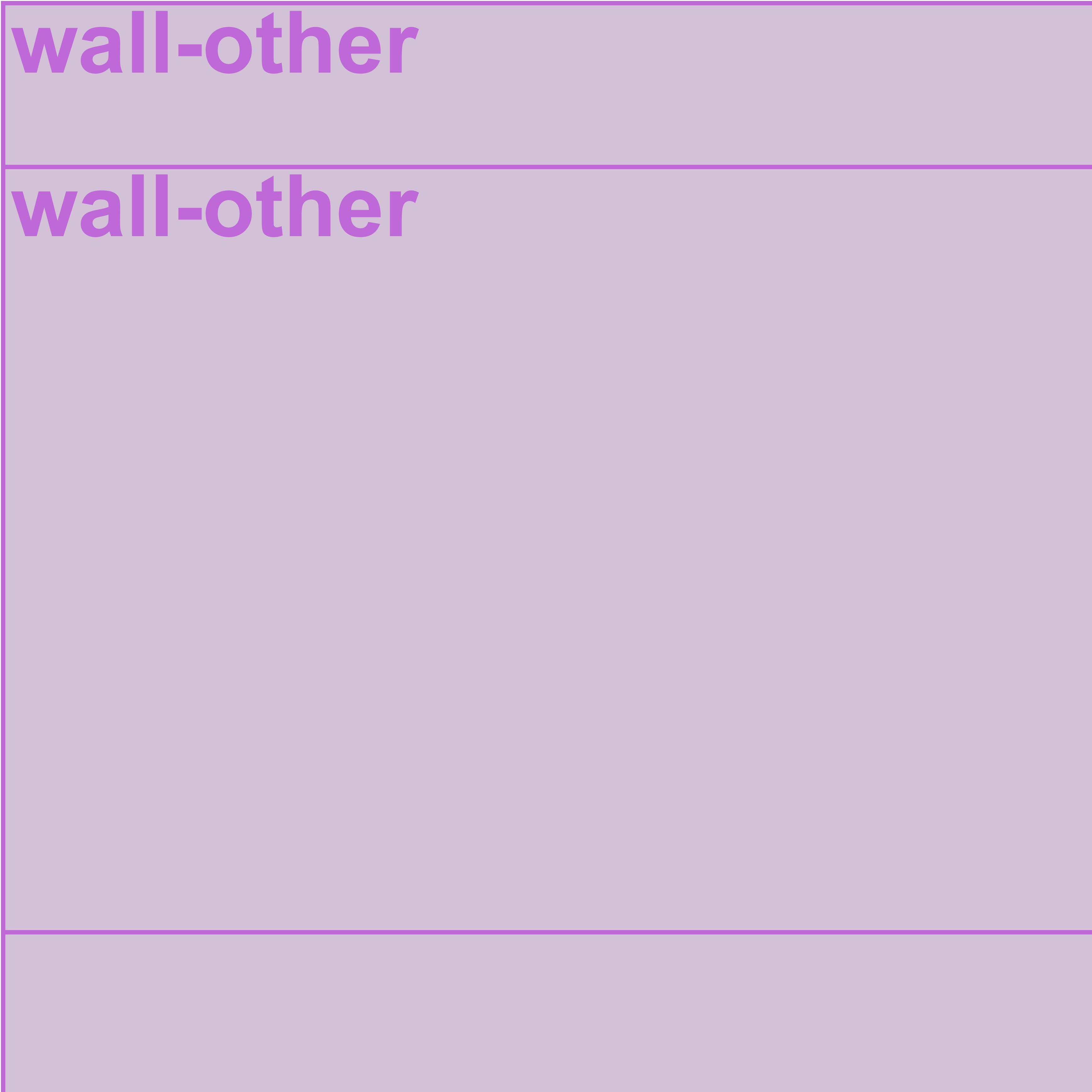} &
\includegraphics[width=\cocoBulkWidth]{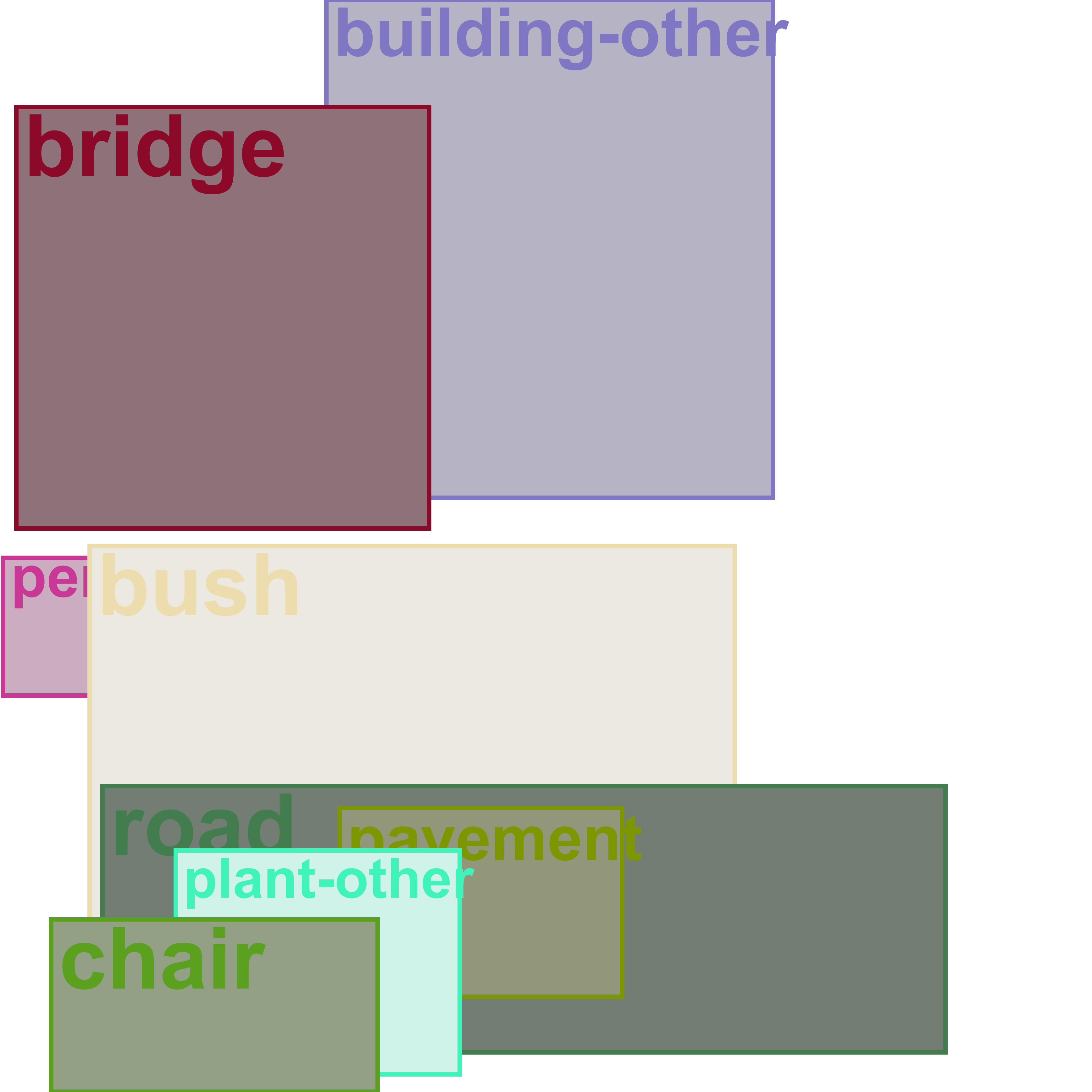} &
\includegraphics[width=\cocoBulkWidth]{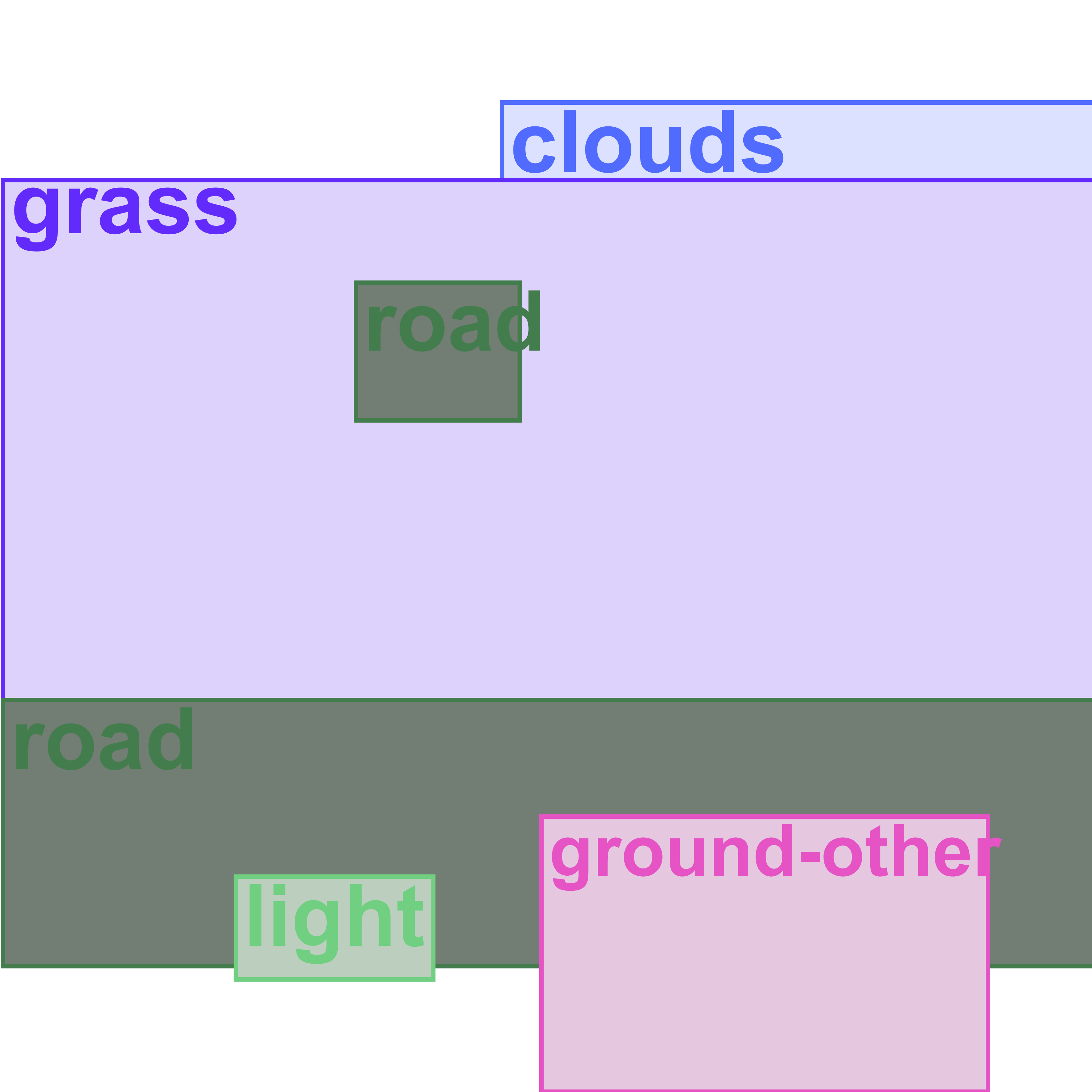} &
\includegraphics[width=\cocoBulkWidth]{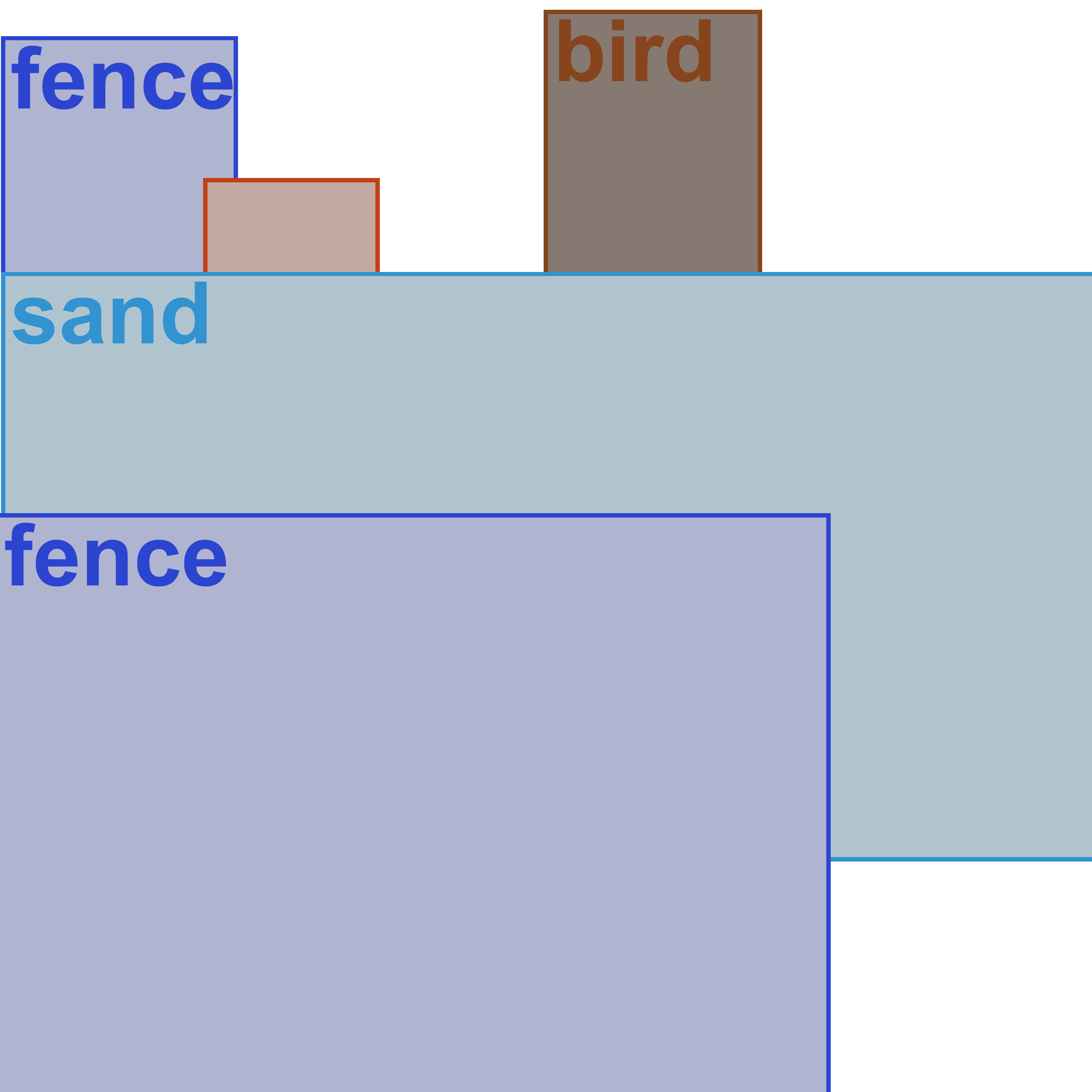} &
\includegraphics[width=\cocoBulkWidth]{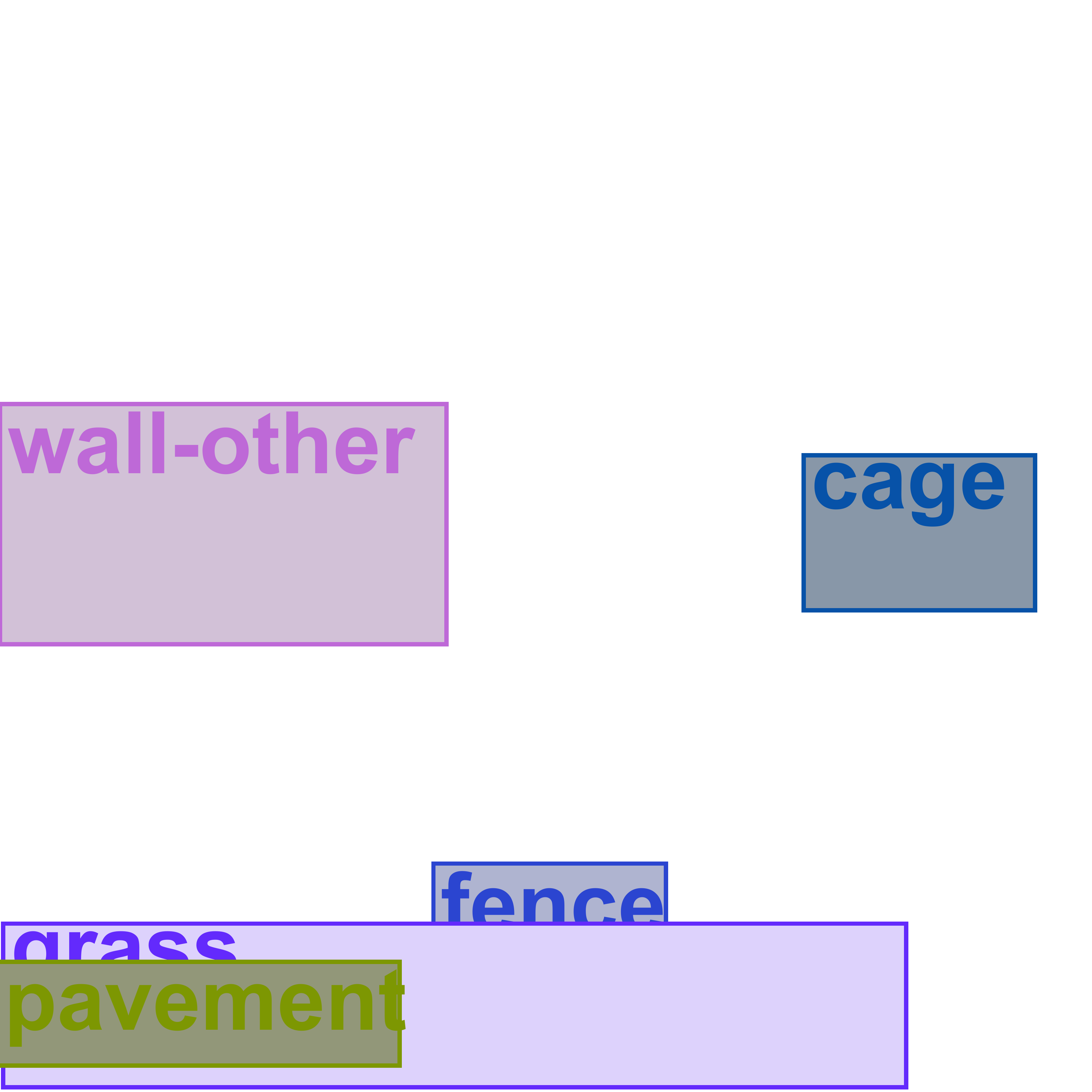} &
\includegraphics[width=\cocoBulkWidth]{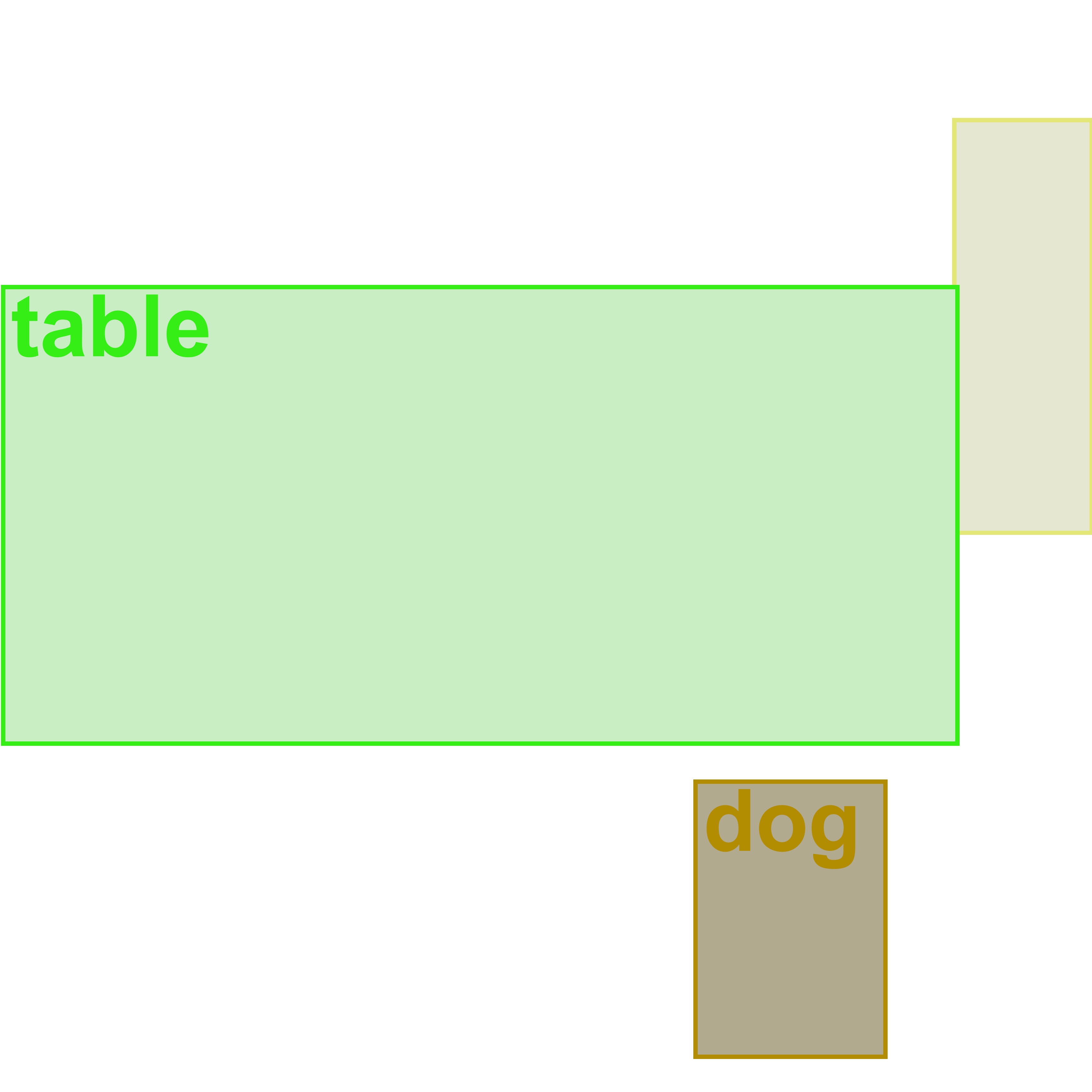} \\
&\includegraphics[width=\cocoBulkWidth]{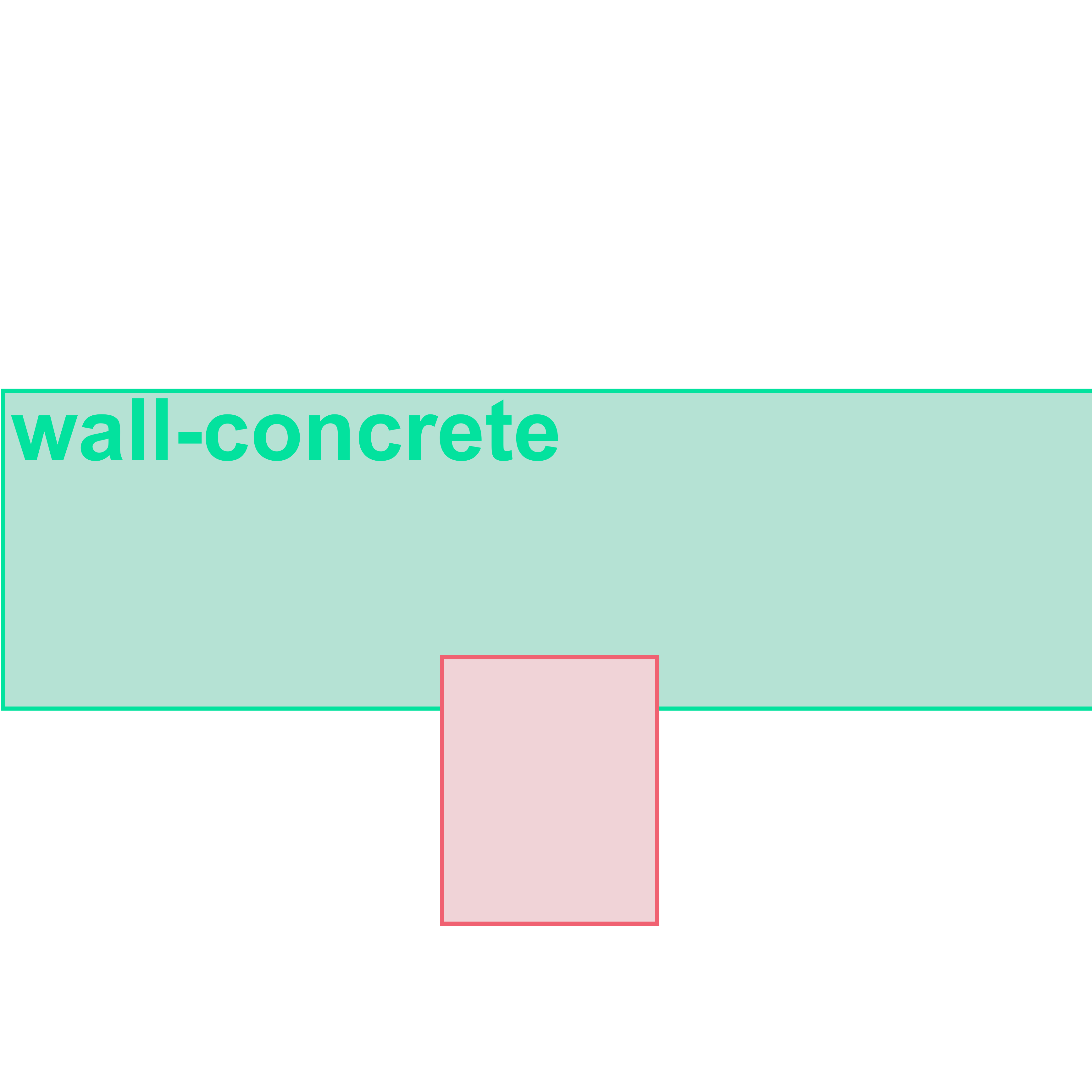} &
\includegraphics[width=\cocoBulkWidth]{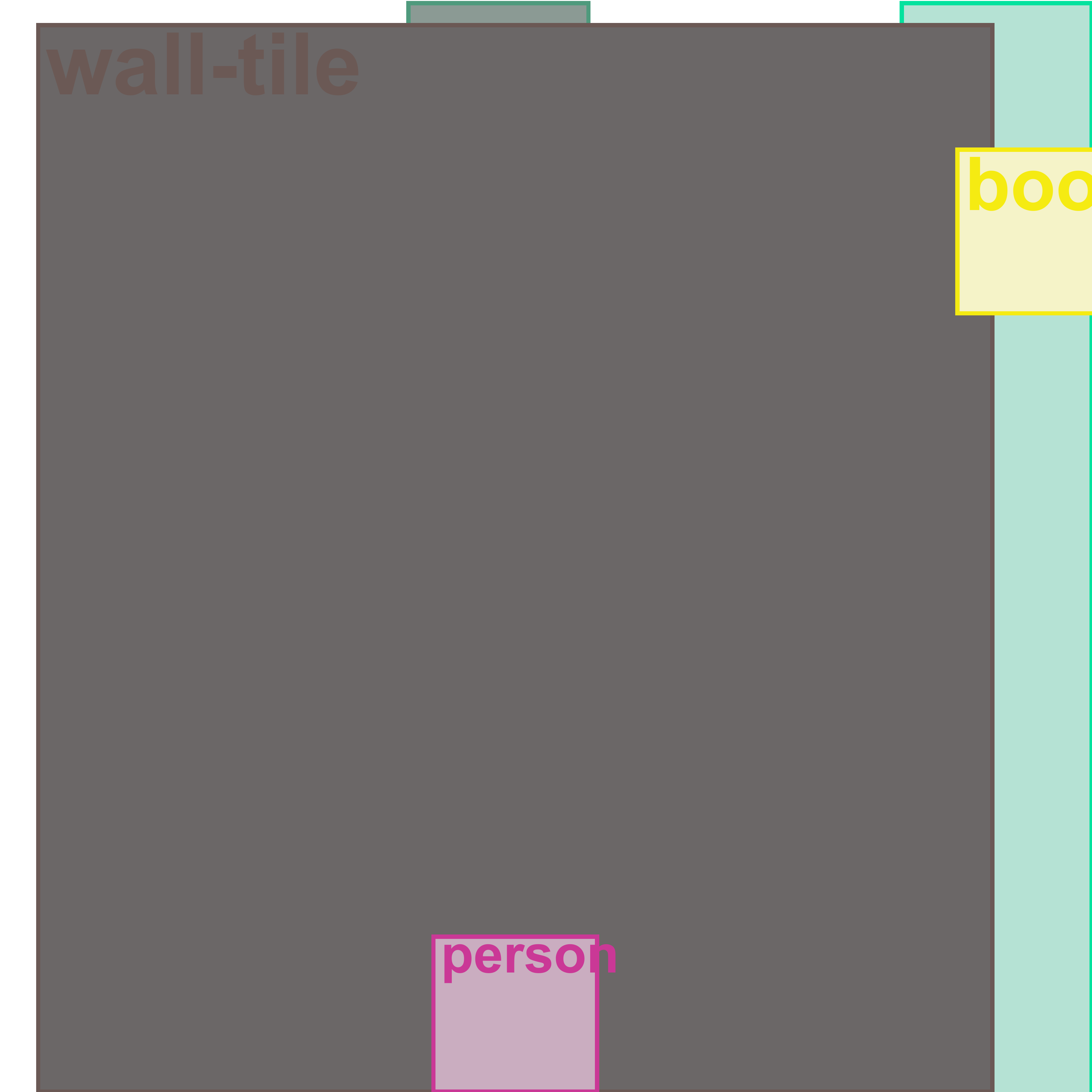} &
\includegraphics[width=\cocoBulkWidth]{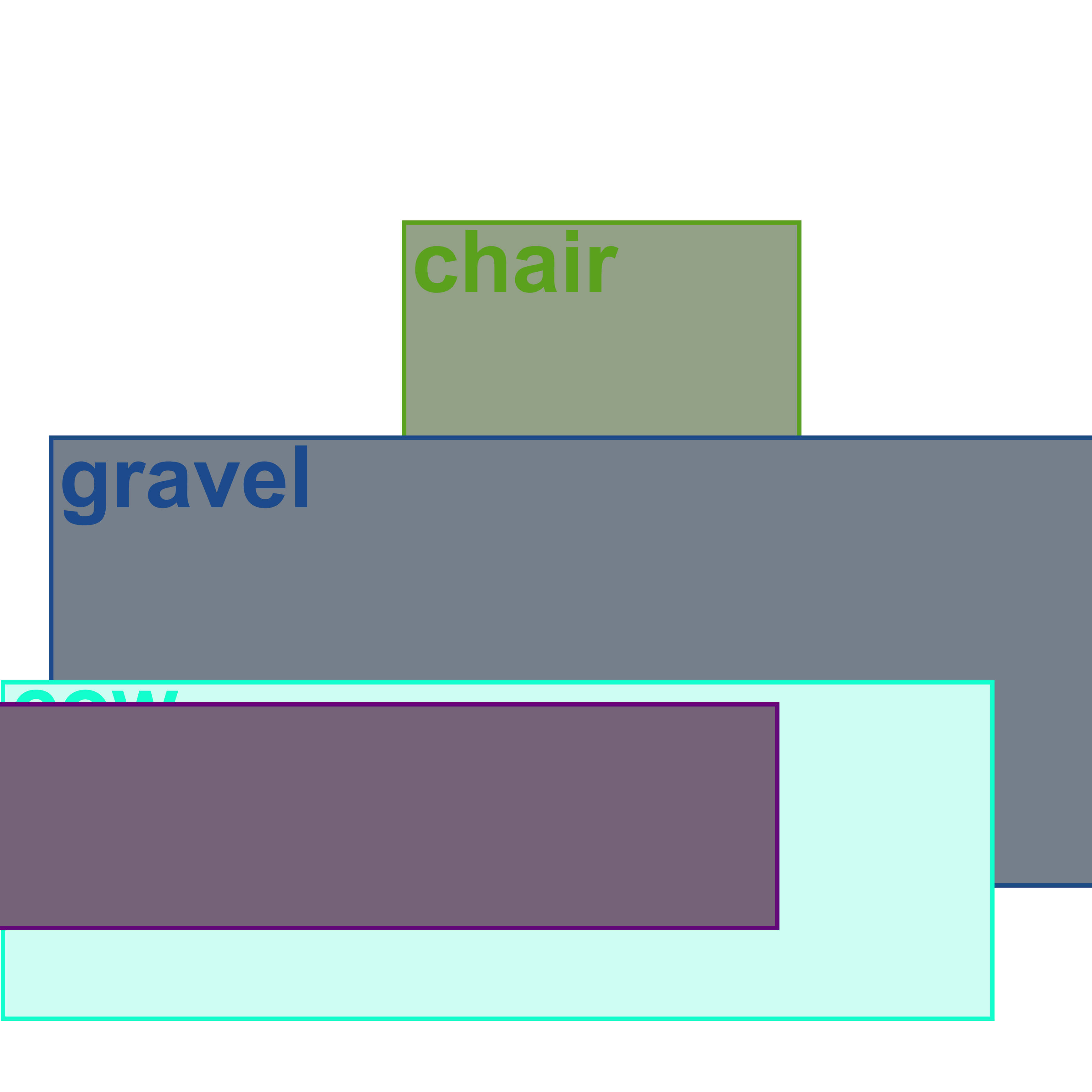} &
\includegraphics[width=\cocoBulkWidth]{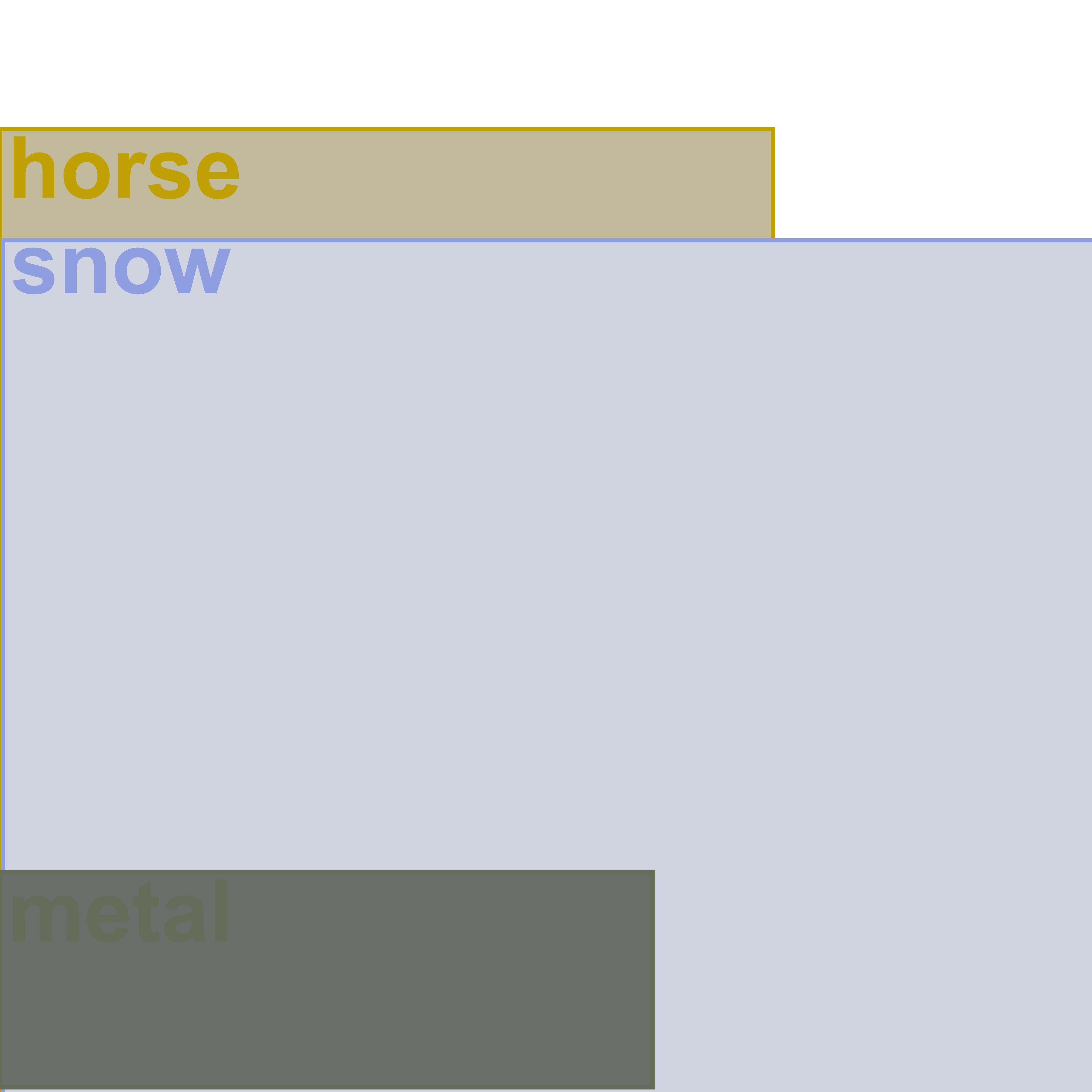} &
\includegraphics[width=\cocoBulkWidth]{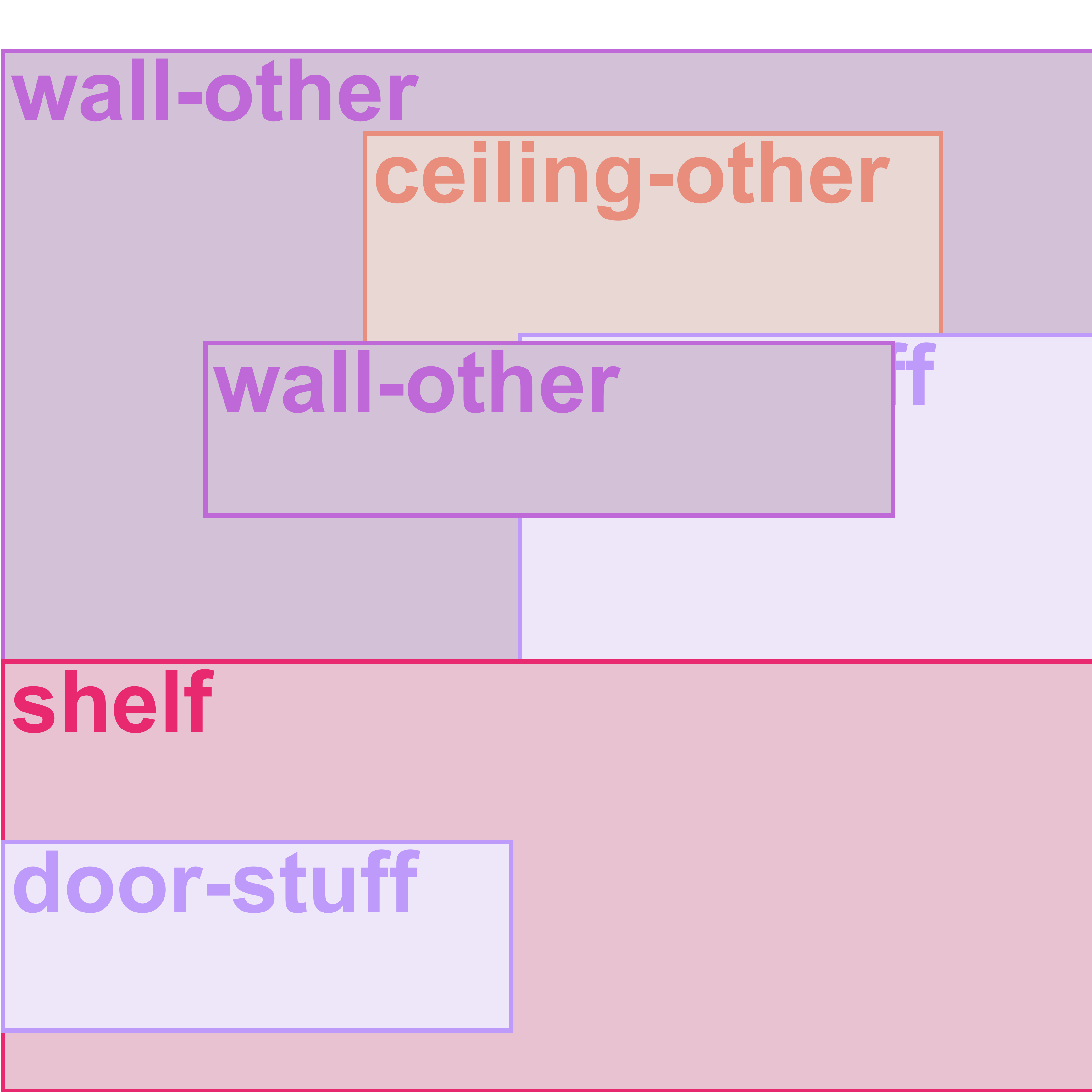} &
\includegraphics[width=\cocoBulkWidth]{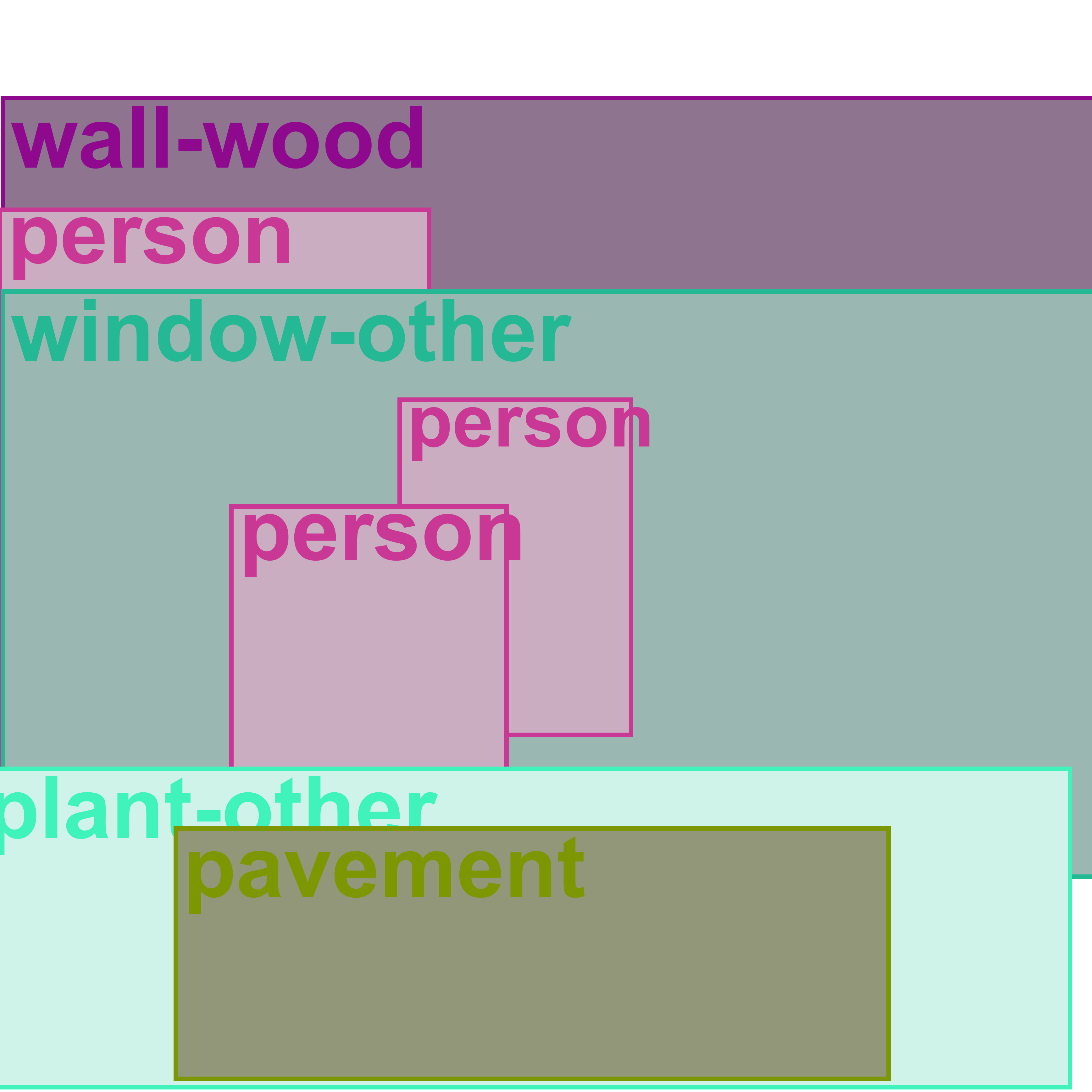} \\
&\includegraphics[width=\cocoBulkWidth]{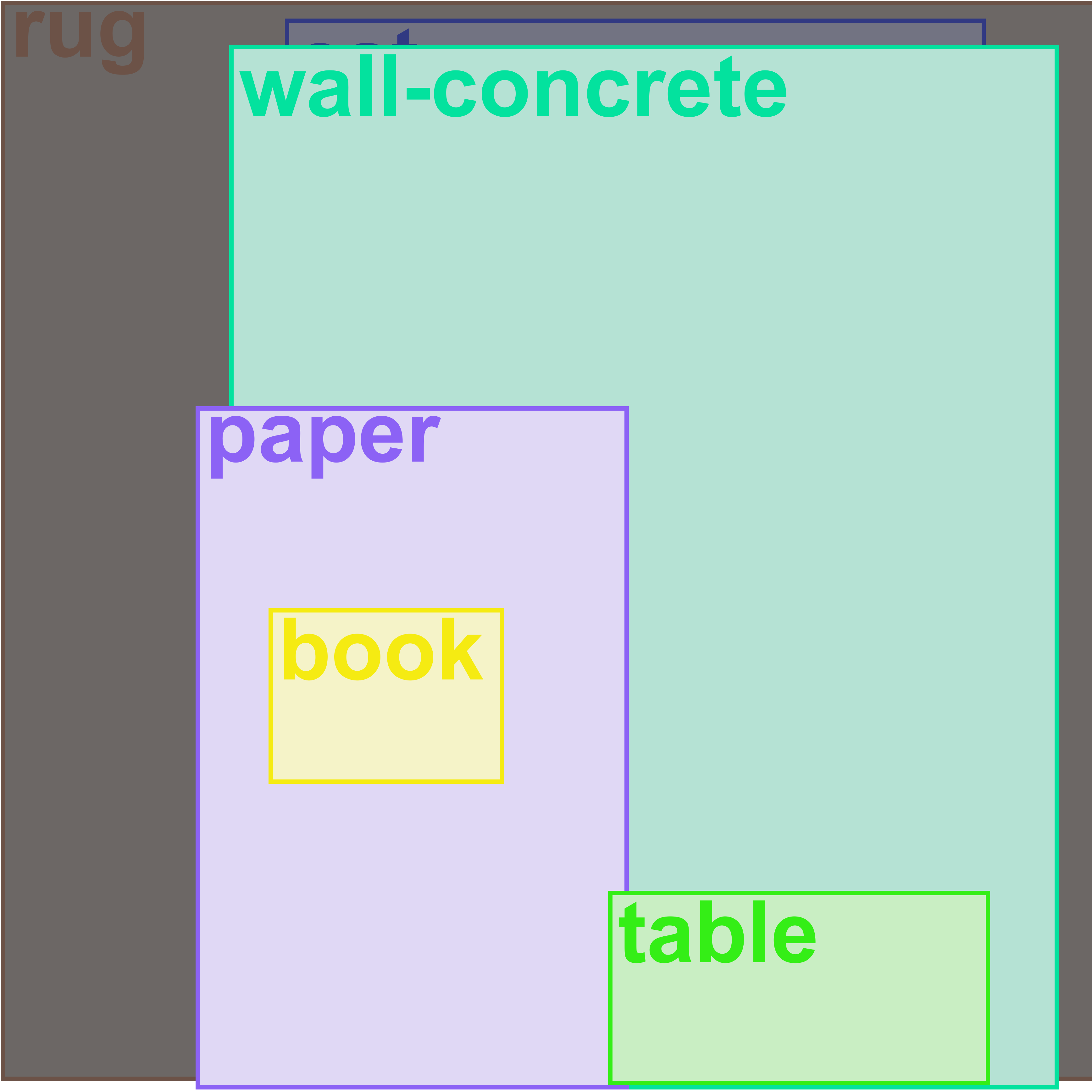} &
\includegraphics[width=\cocoBulkWidth]{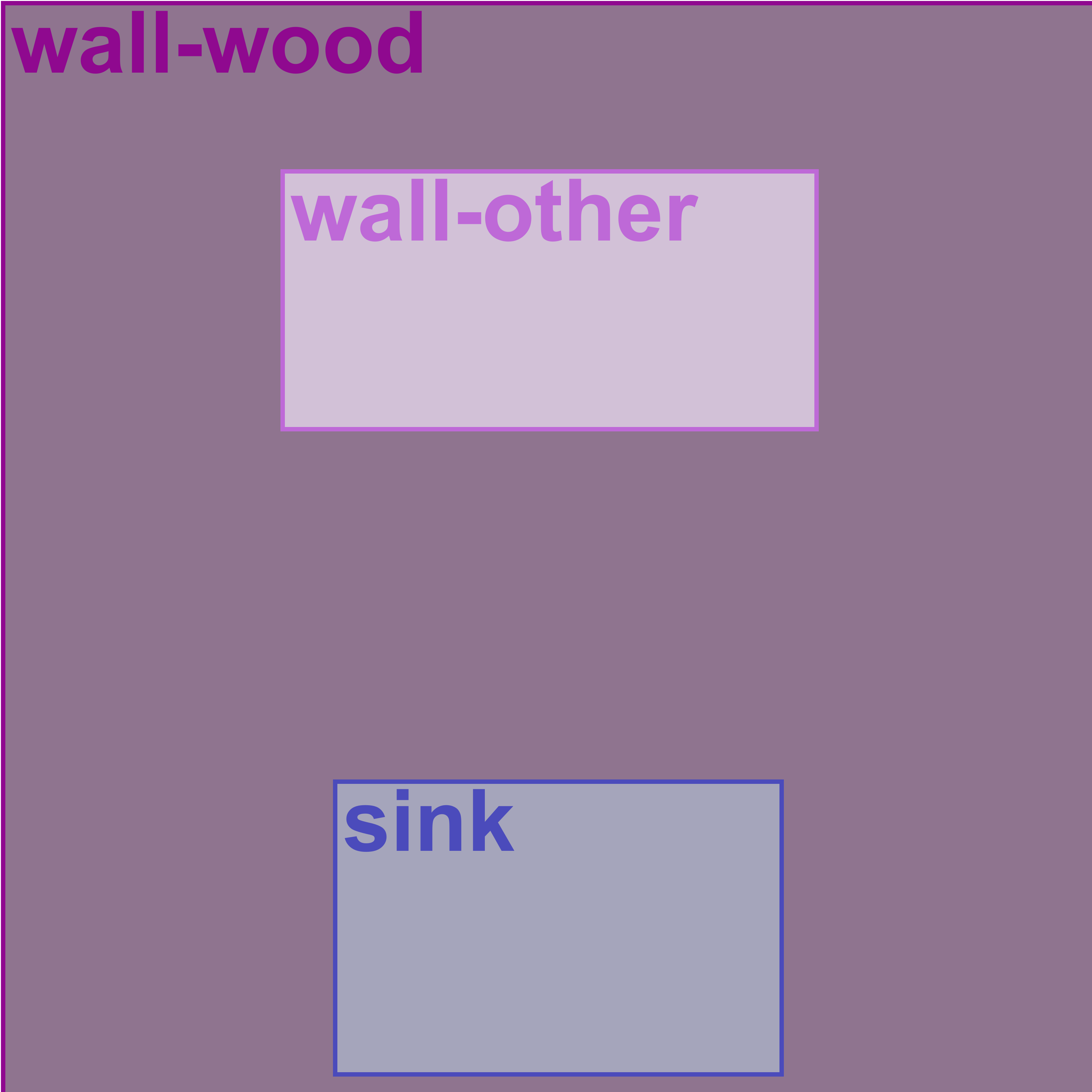} &
\includegraphics[width=\cocoBulkWidth]{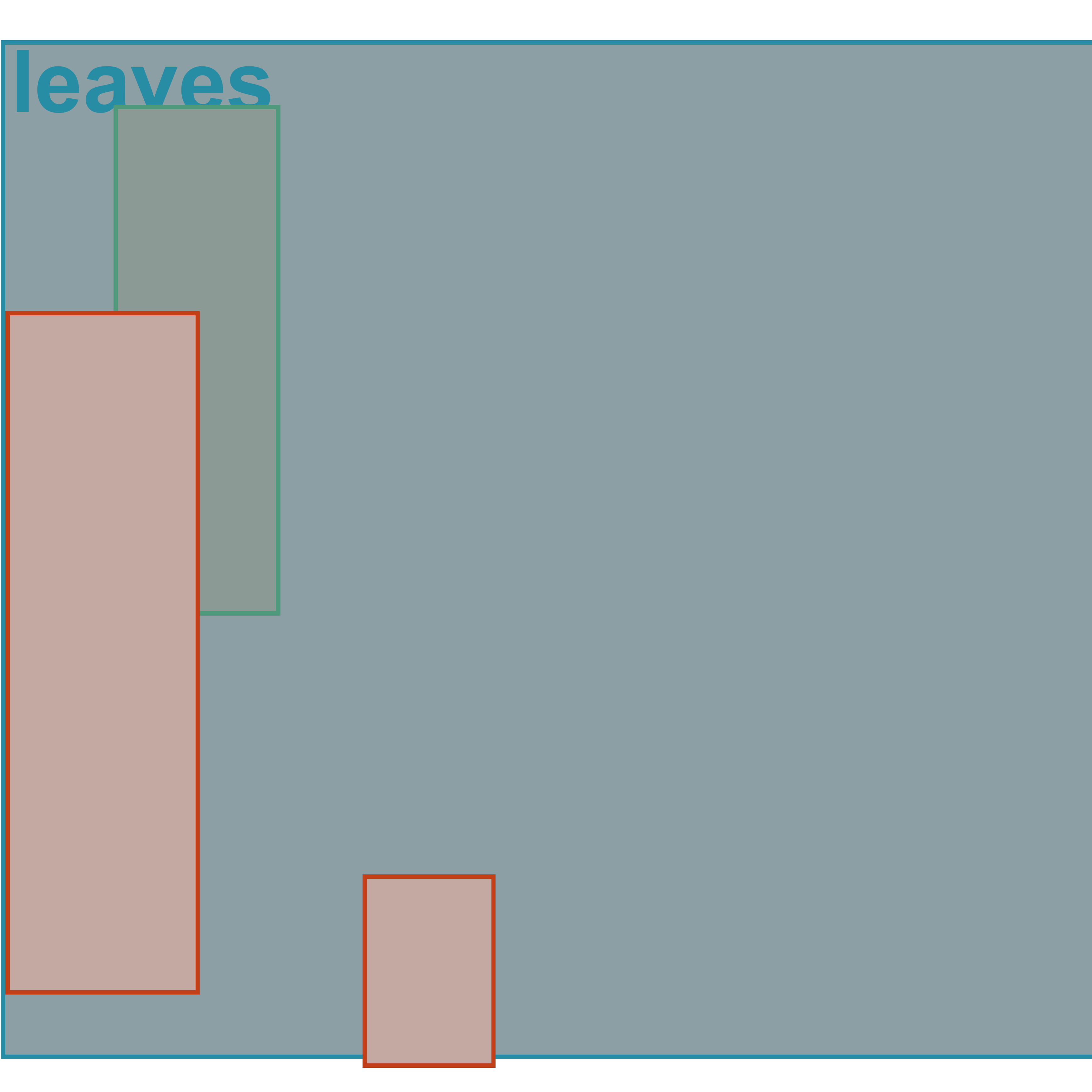} &
\includegraphics[width=\cocoBulkWidth]{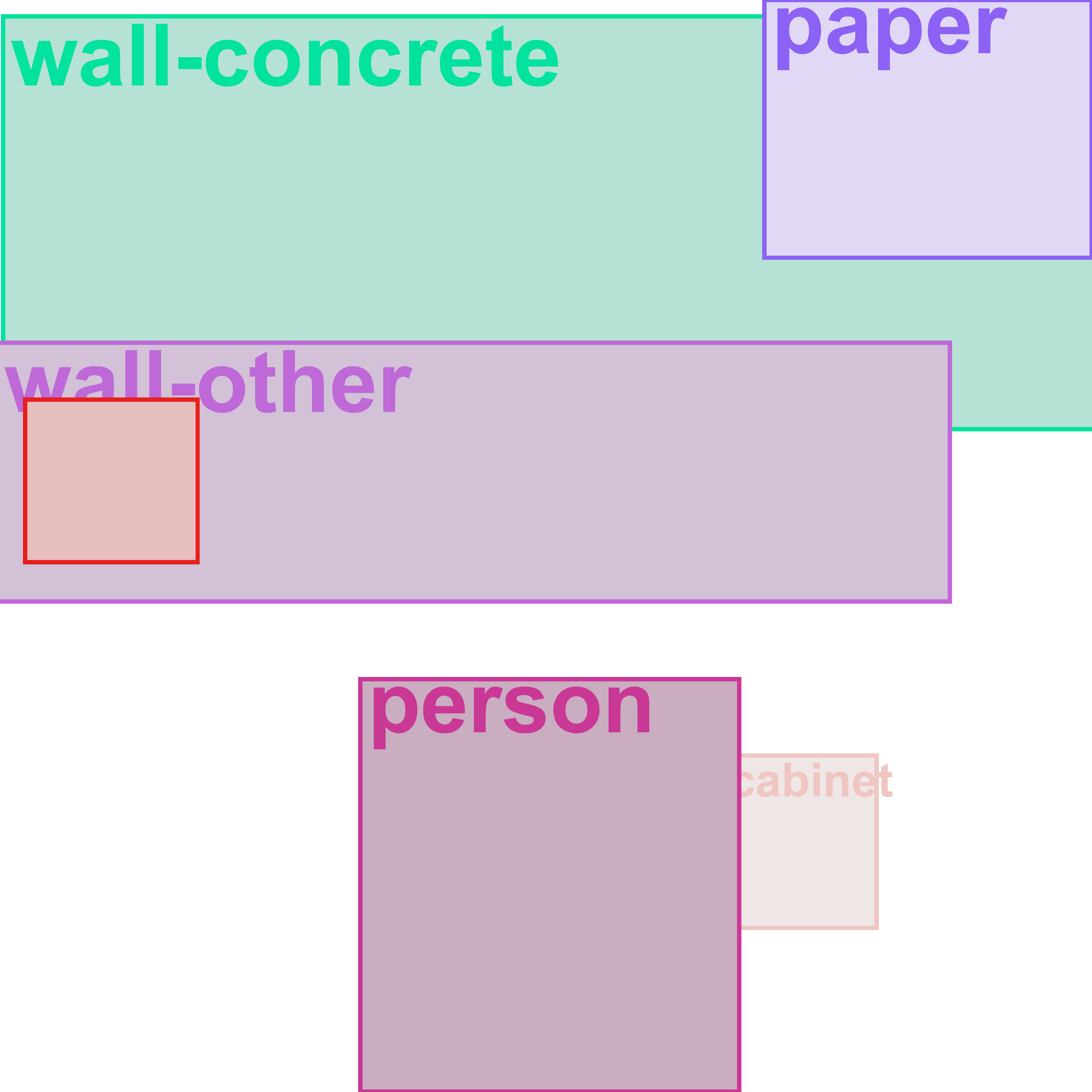} &
\includegraphics[width=\cocoBulkWidth]{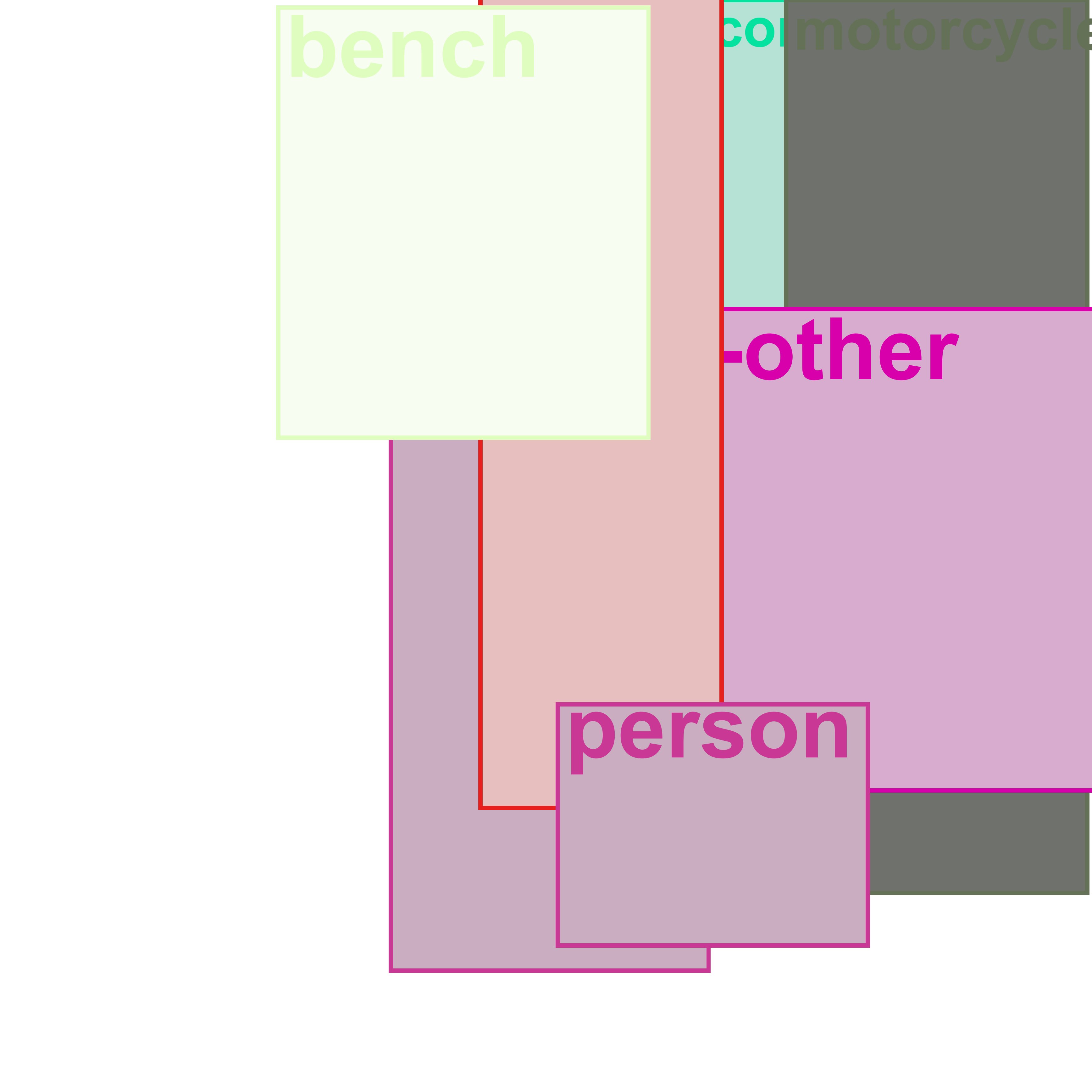} &
\includegraphics[width=\cocoBulkWidth]{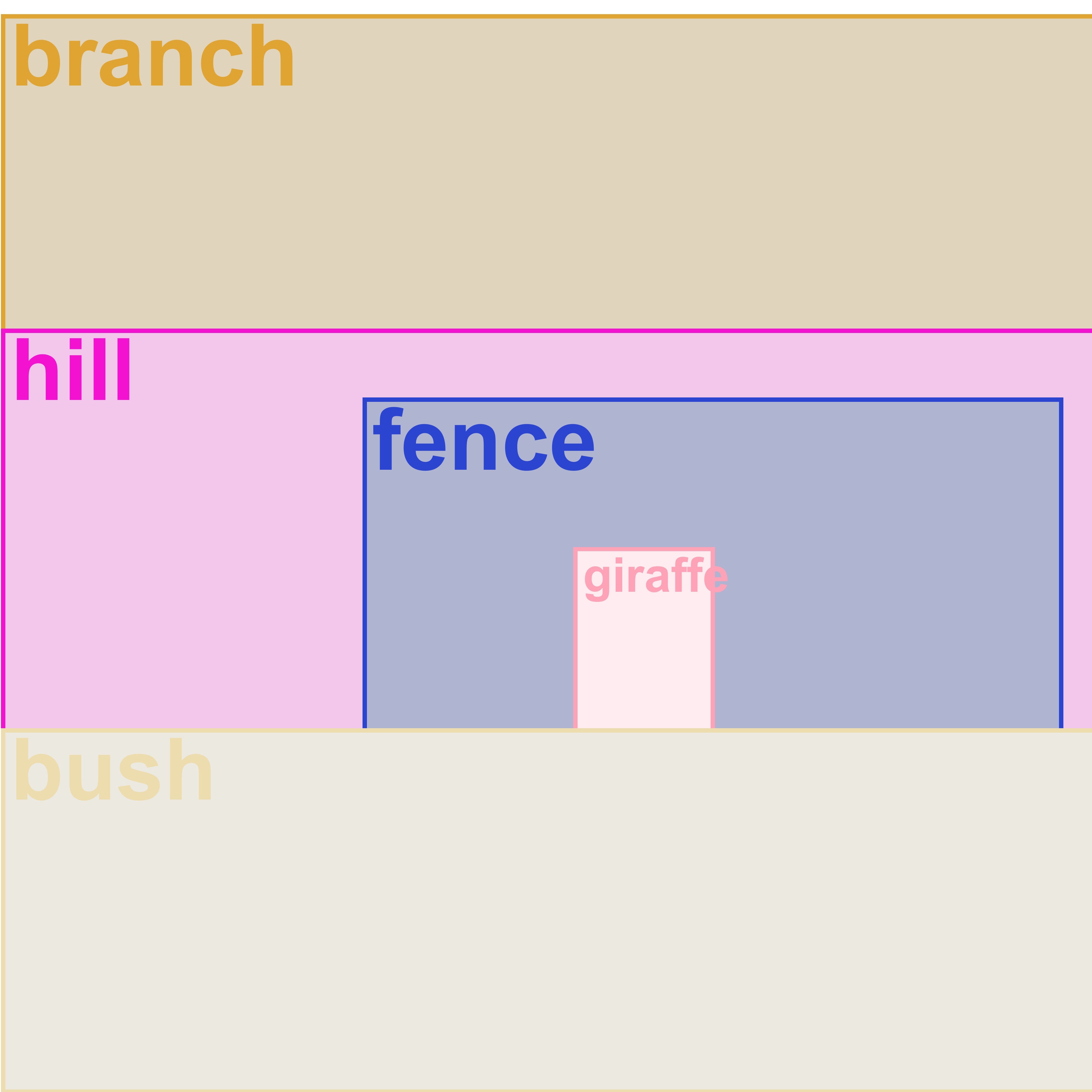} \\
&\includegraphics[width=\cocoBulkWidth]{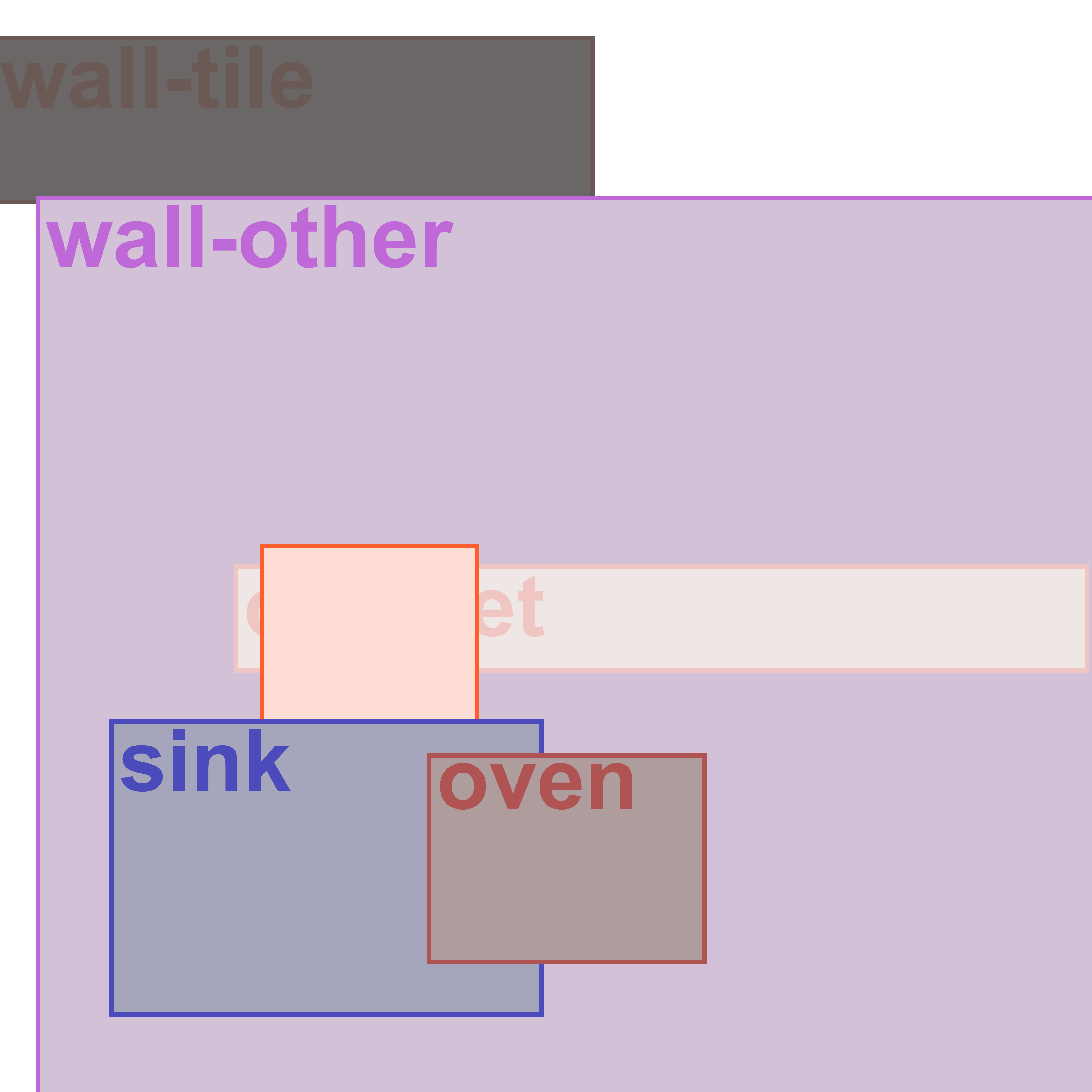} &
\includegraphics[width=\cocoBulkWidth]{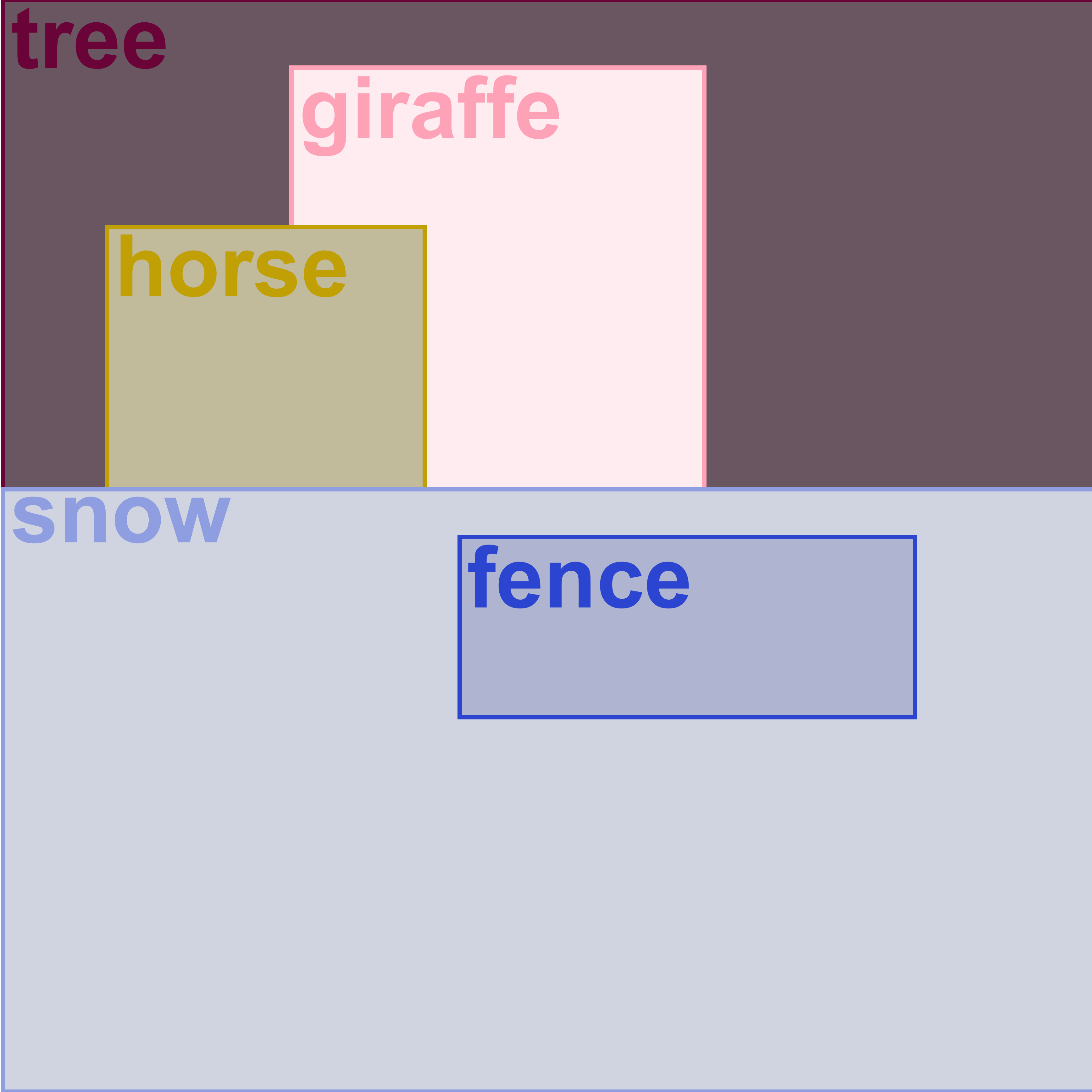} &
\includegraphics[width=\cocoBulkWidth]{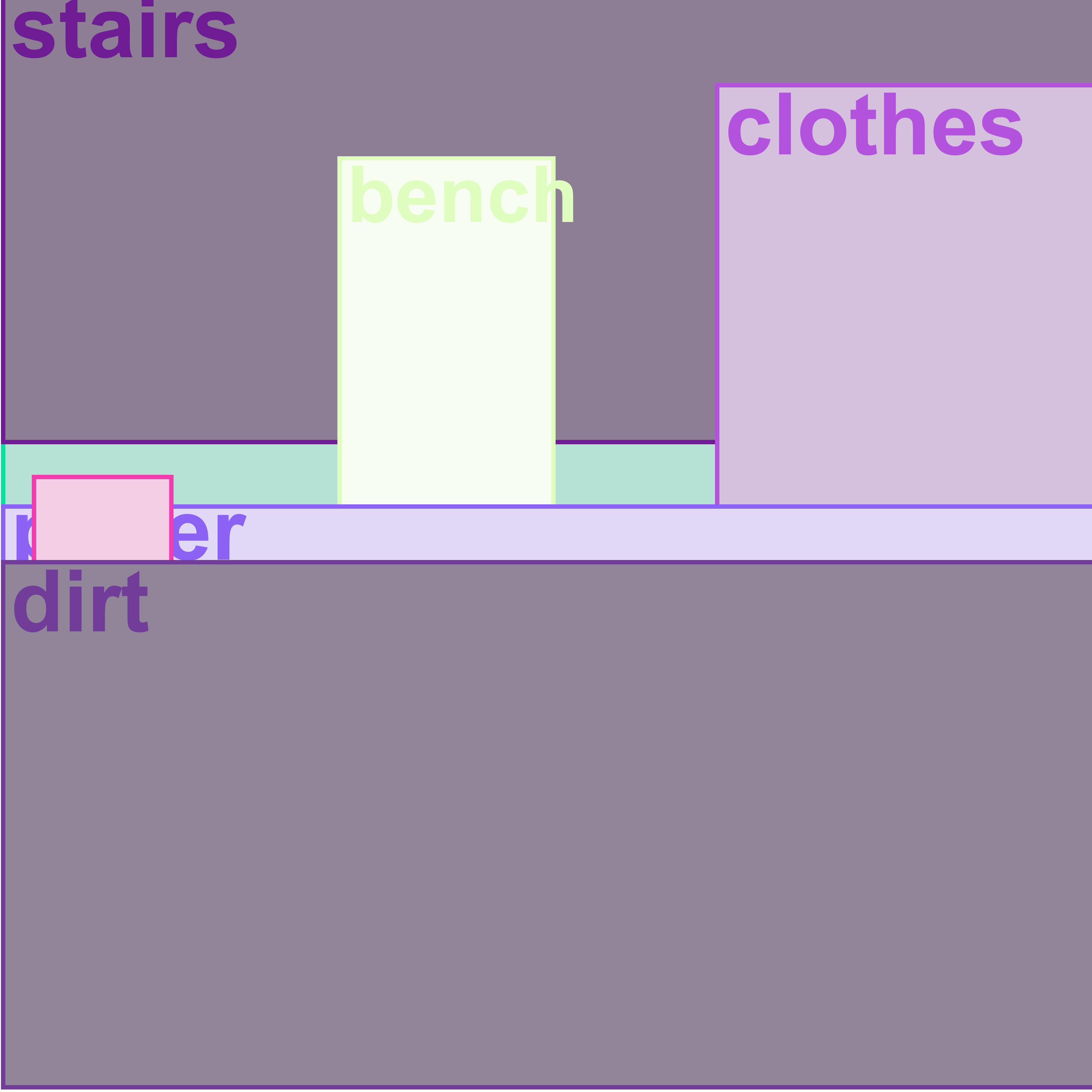} &
\includegraphics[width=\cocoBulkWidth]{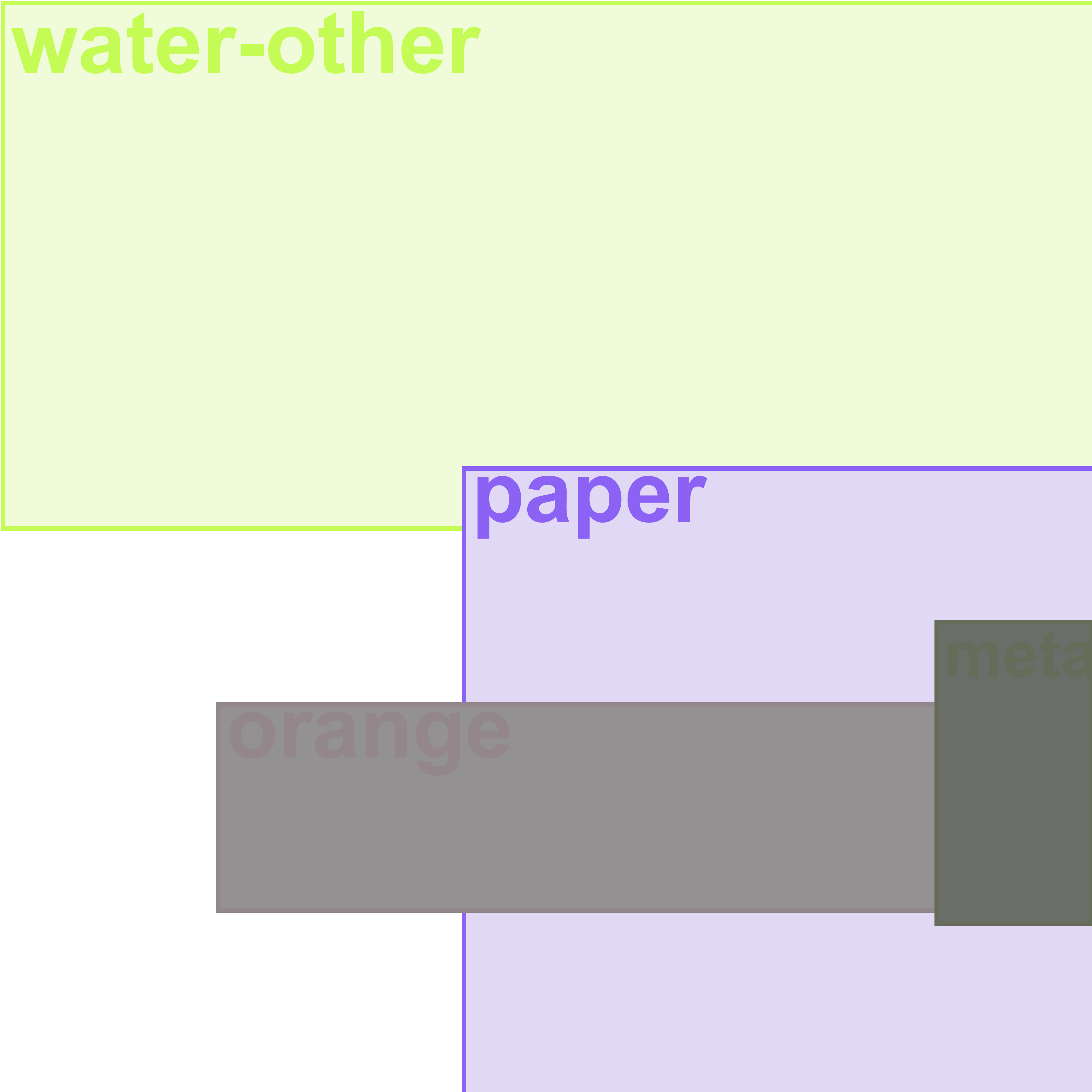} &
\includegraphics[width=\cocoBulkWidth]{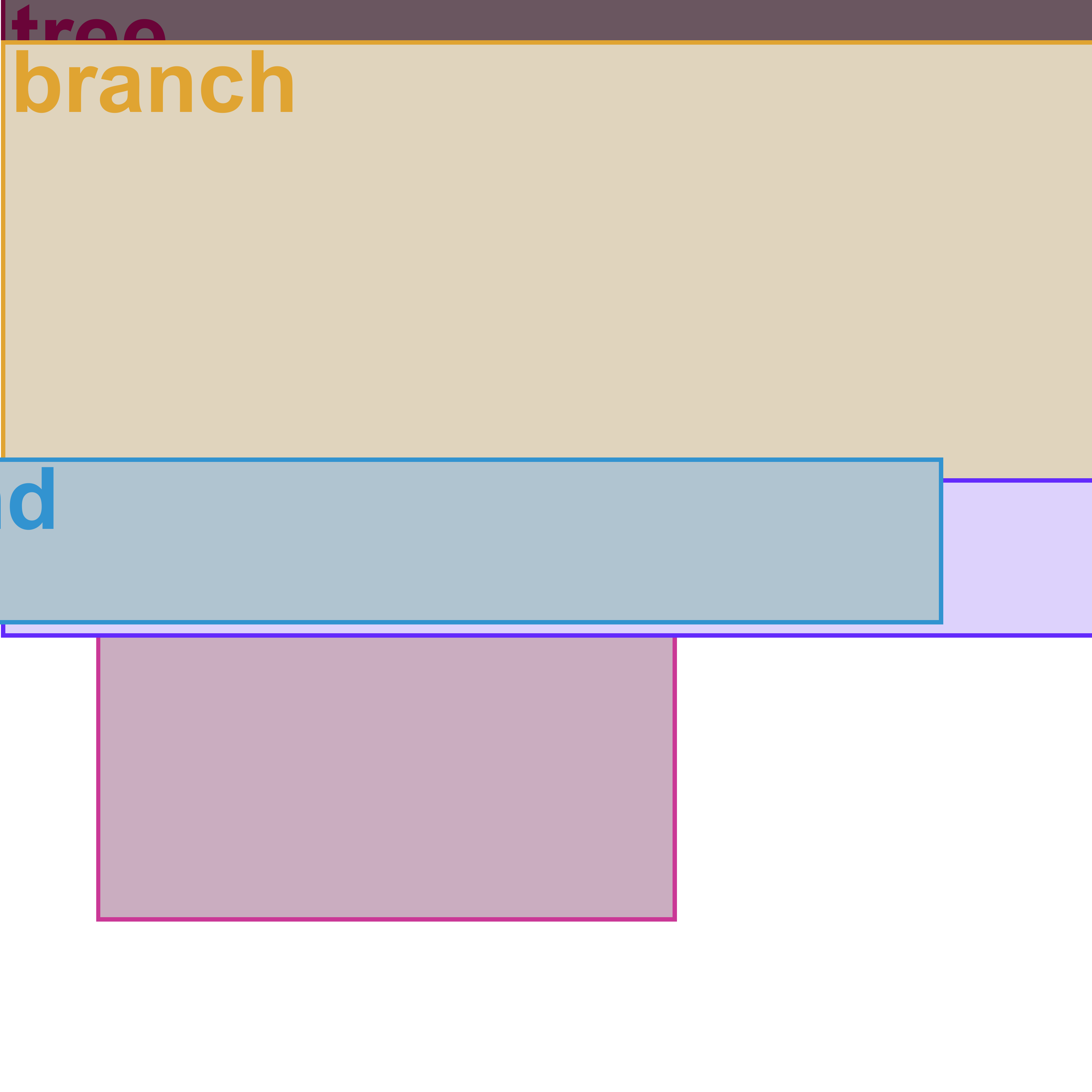} &
\includegraphics[width=\cocoBulkWidth]{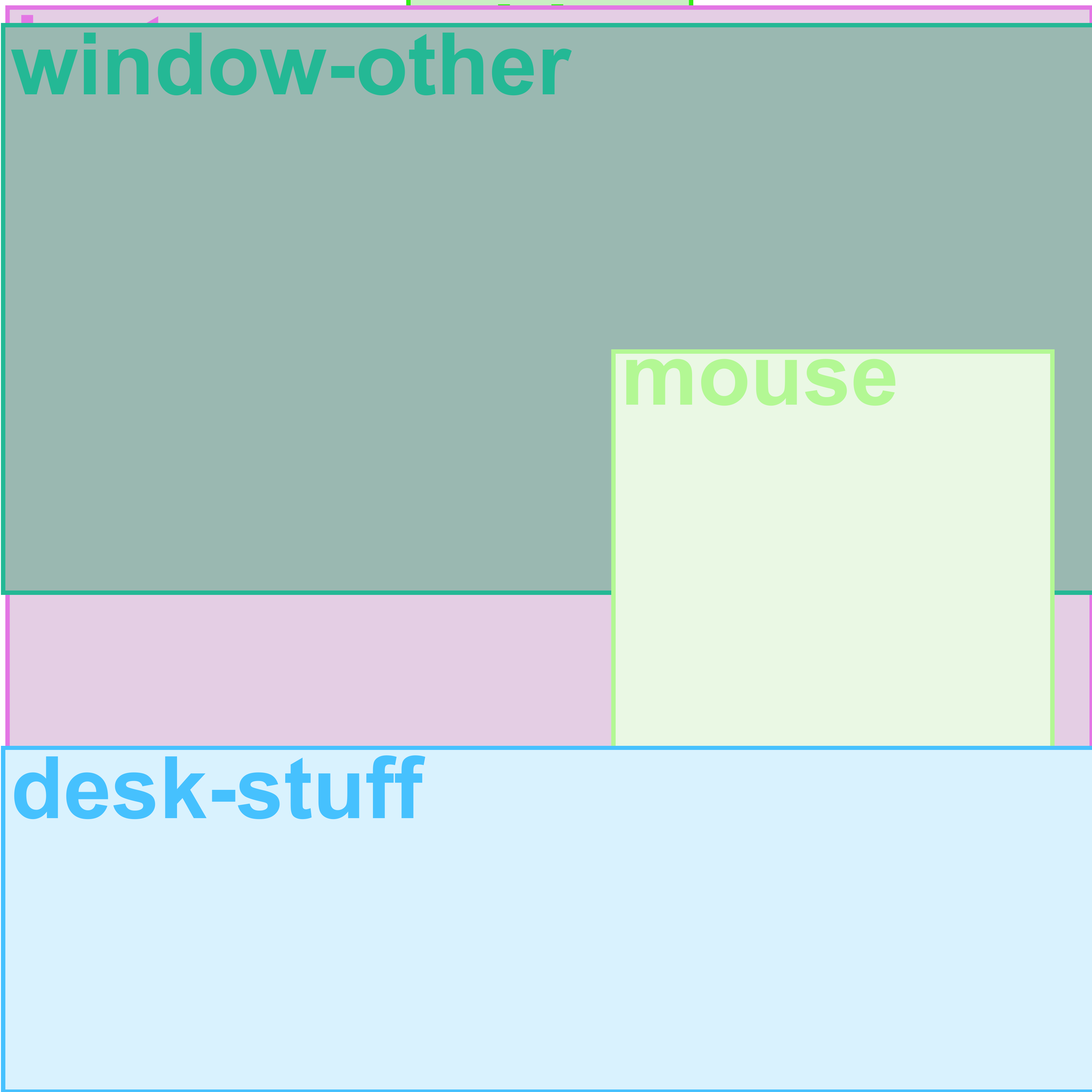} \\
&\includegraphics[width=\cocoBulkWidth]{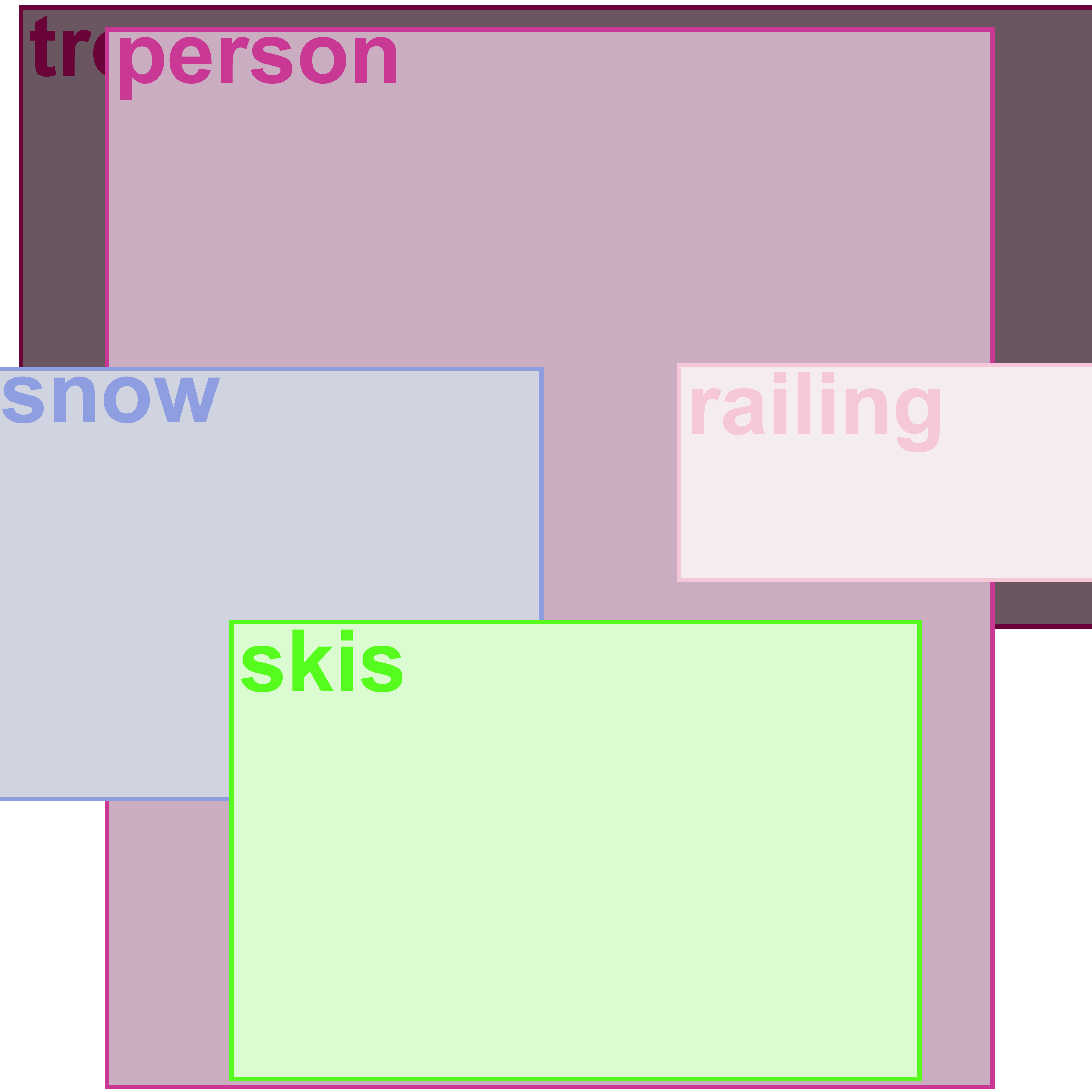} &
\includegraphics[width=\cocoBulkWidth]{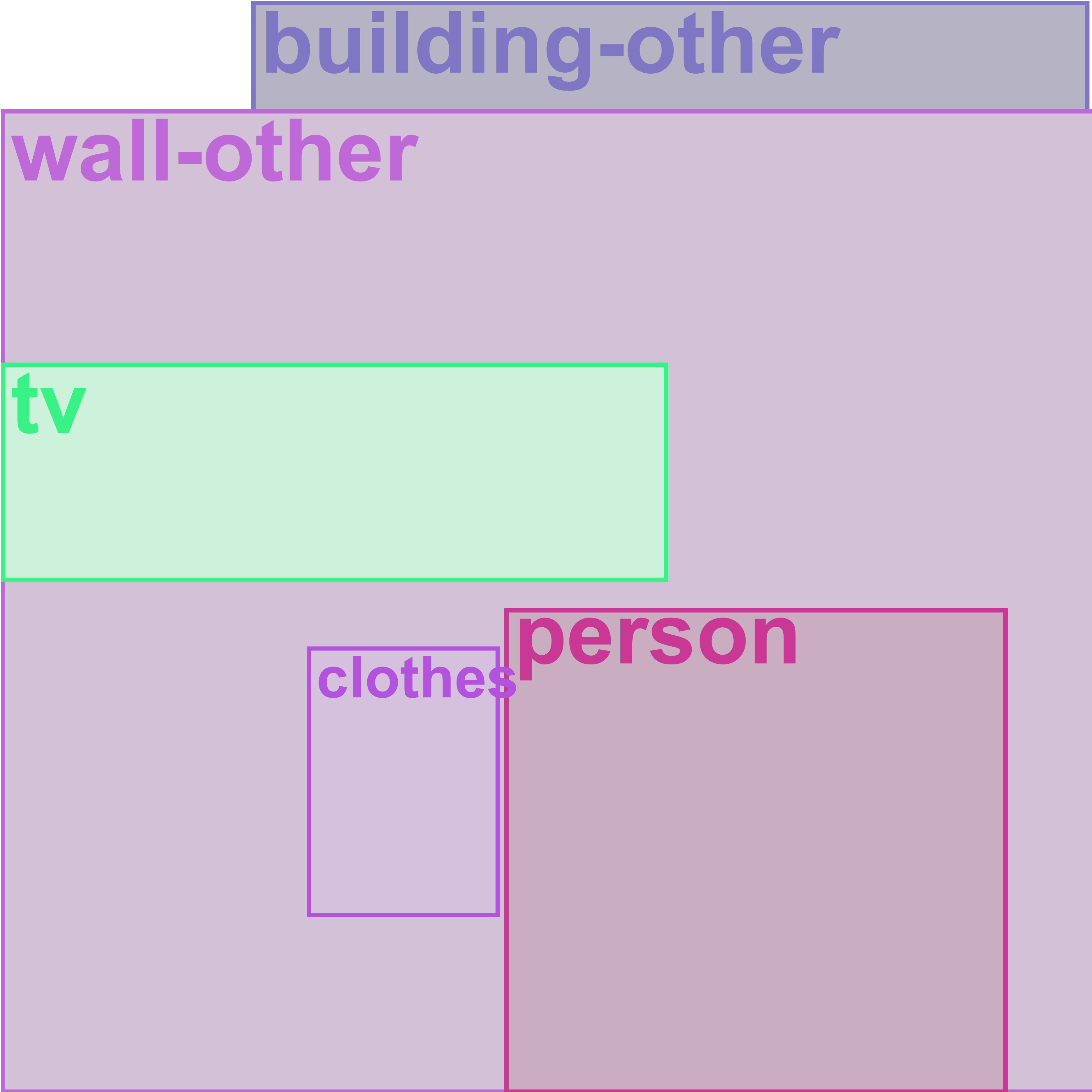} &
\includegraphics[width=\cocoBulkWidth]{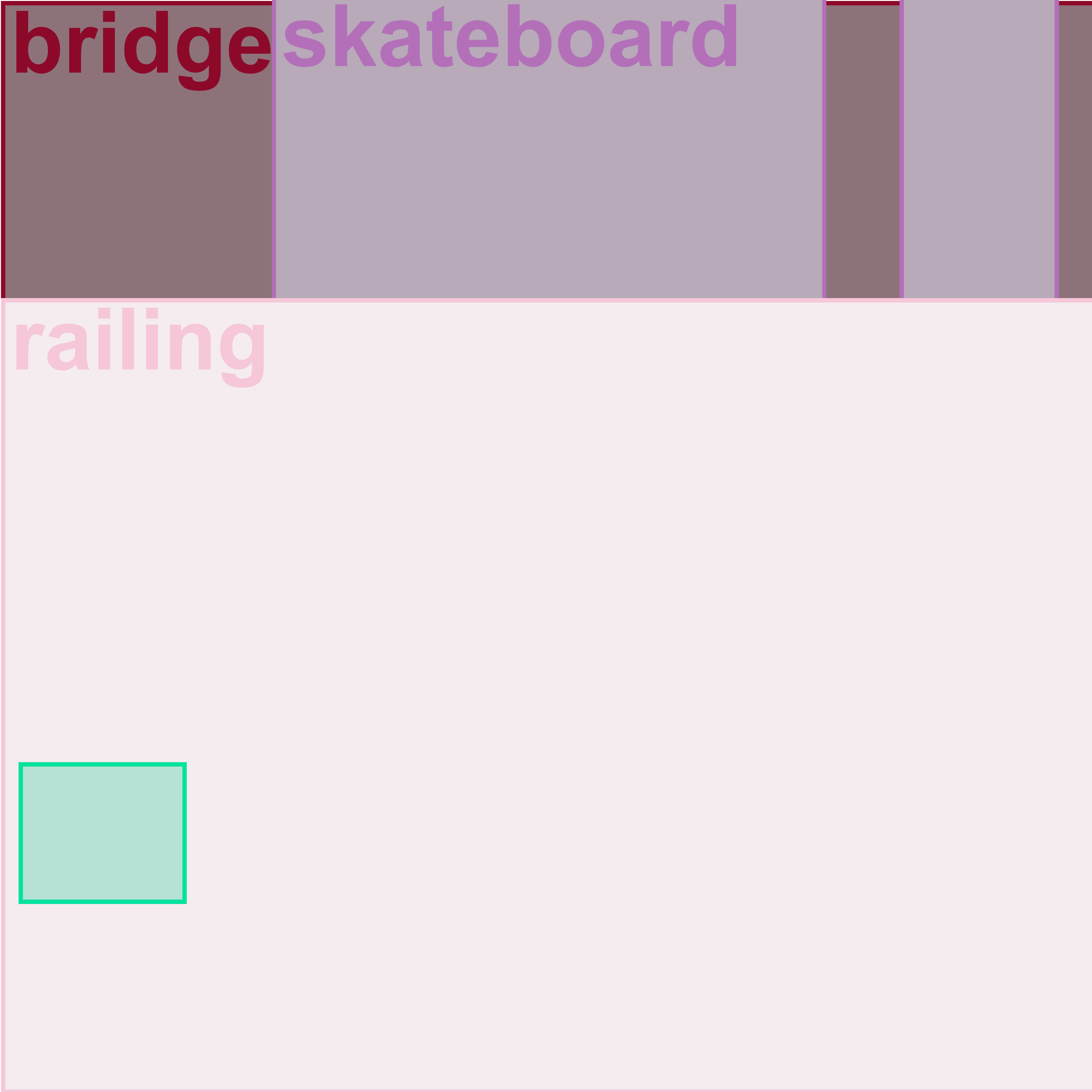} &
\includegraphics[width=\cocoBulkWidth]{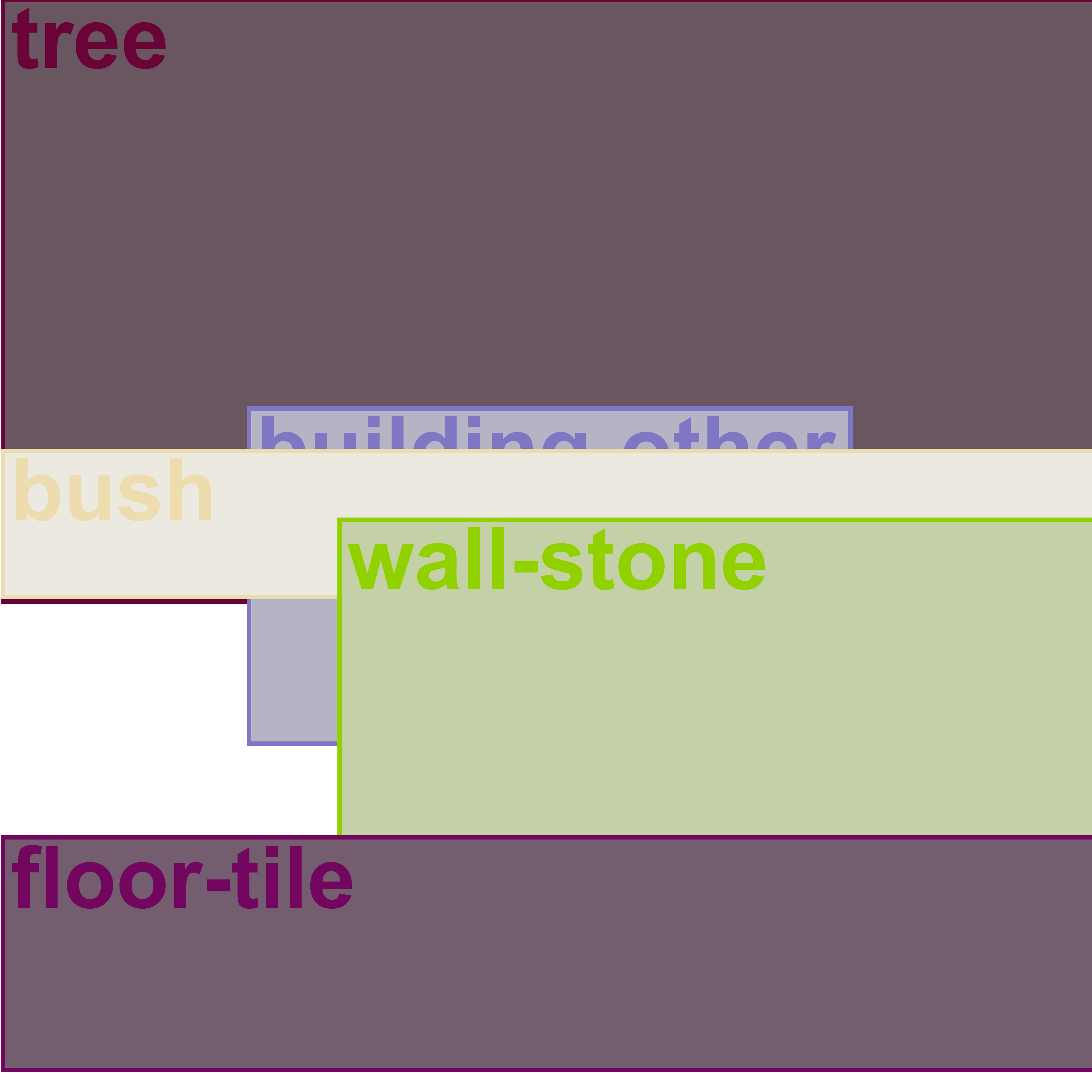} &
\includegraphics[width=\cocoBulkWidth]{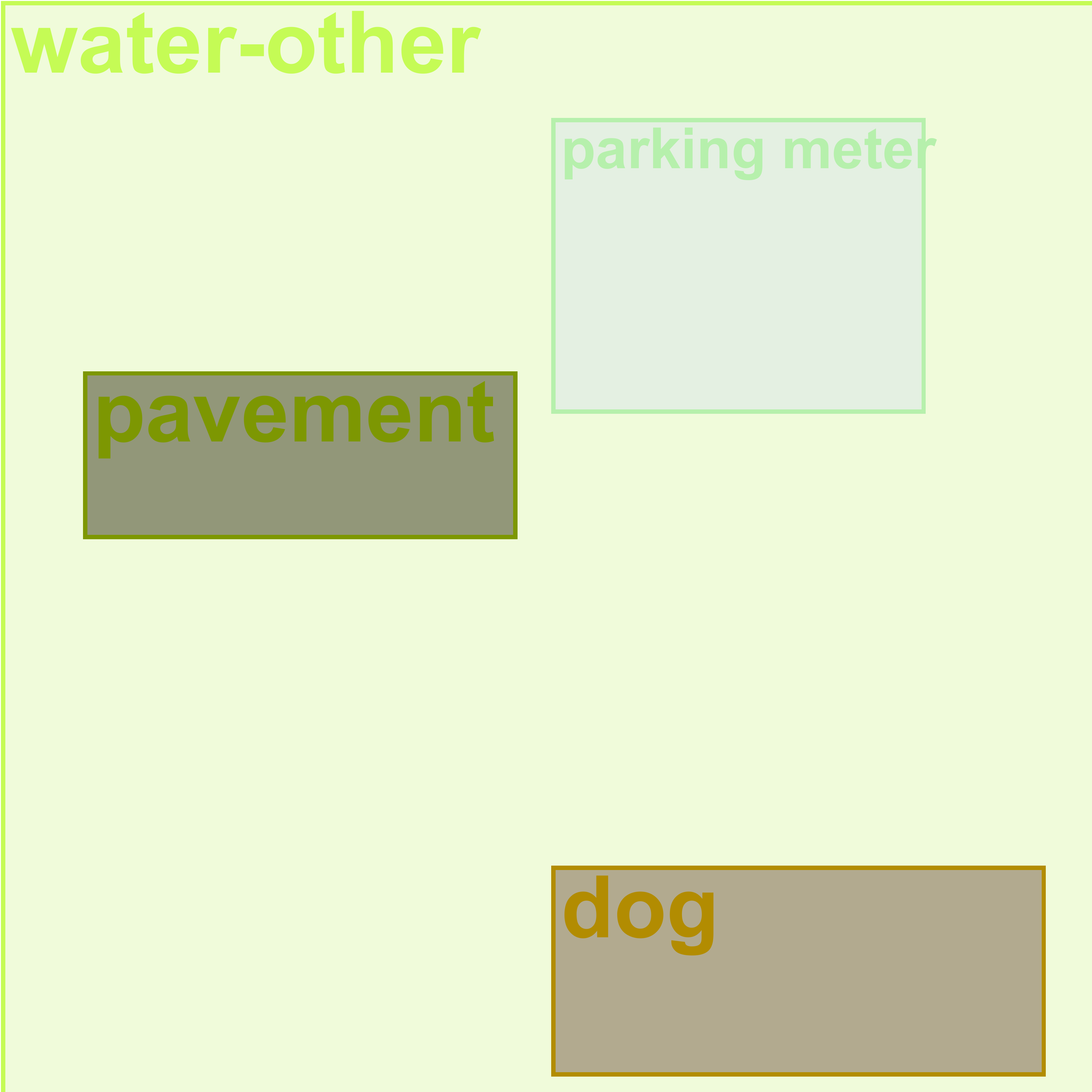} &
\includegraphics[width=\cocoBulkWidth]{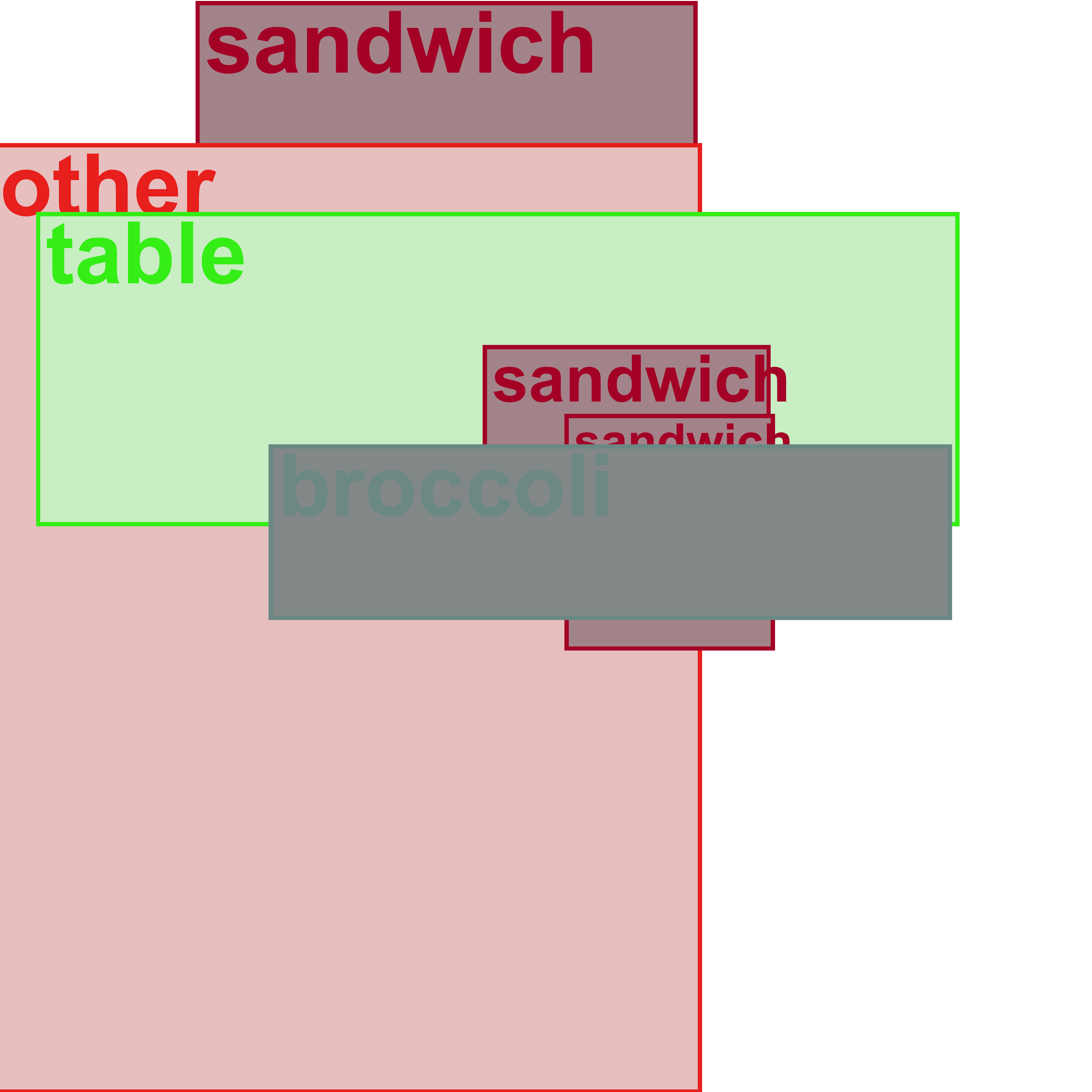} \\

    \end{tabular}
    \caption{Synthesized COCO-Stuff examples.}
    \label{fig:coco_bulk}
\end{figure}

\clearpage
\subsection{SUN RGB-D}

\begin{figure}[h]
    \centering
    \setlength{\tabcolsep}{1pt}
    \newlength{\sunrgbdWidth}
    \setlength{\sunrgbdWidth}{0.16\linewidth}
    \begin{tabular}{cccccc}
    
\includegraphics[width=\sunrgbdWidth,frame=.1pt]{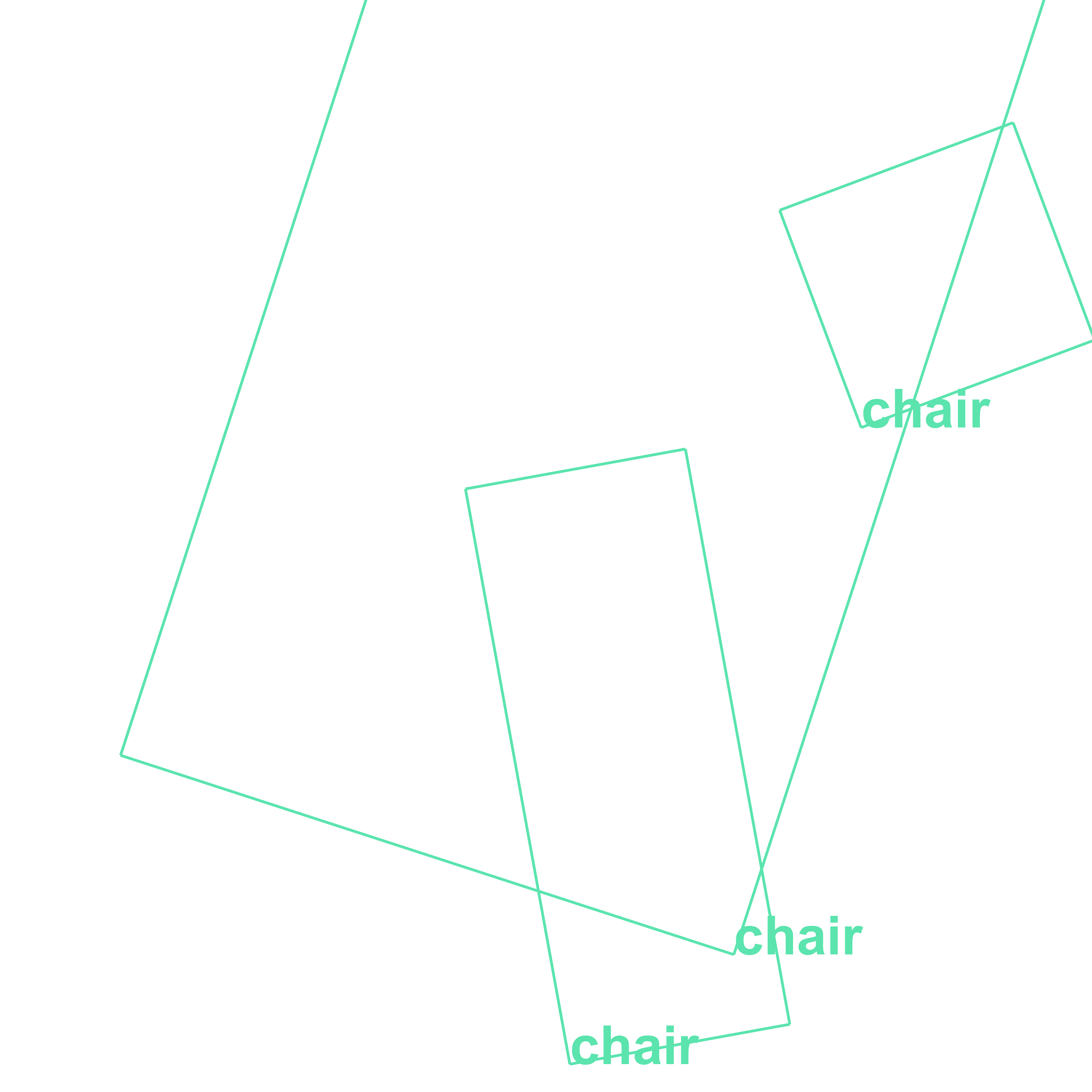} &
\includegraphics[width=\sunrgbdWidth,frame=.1pt]{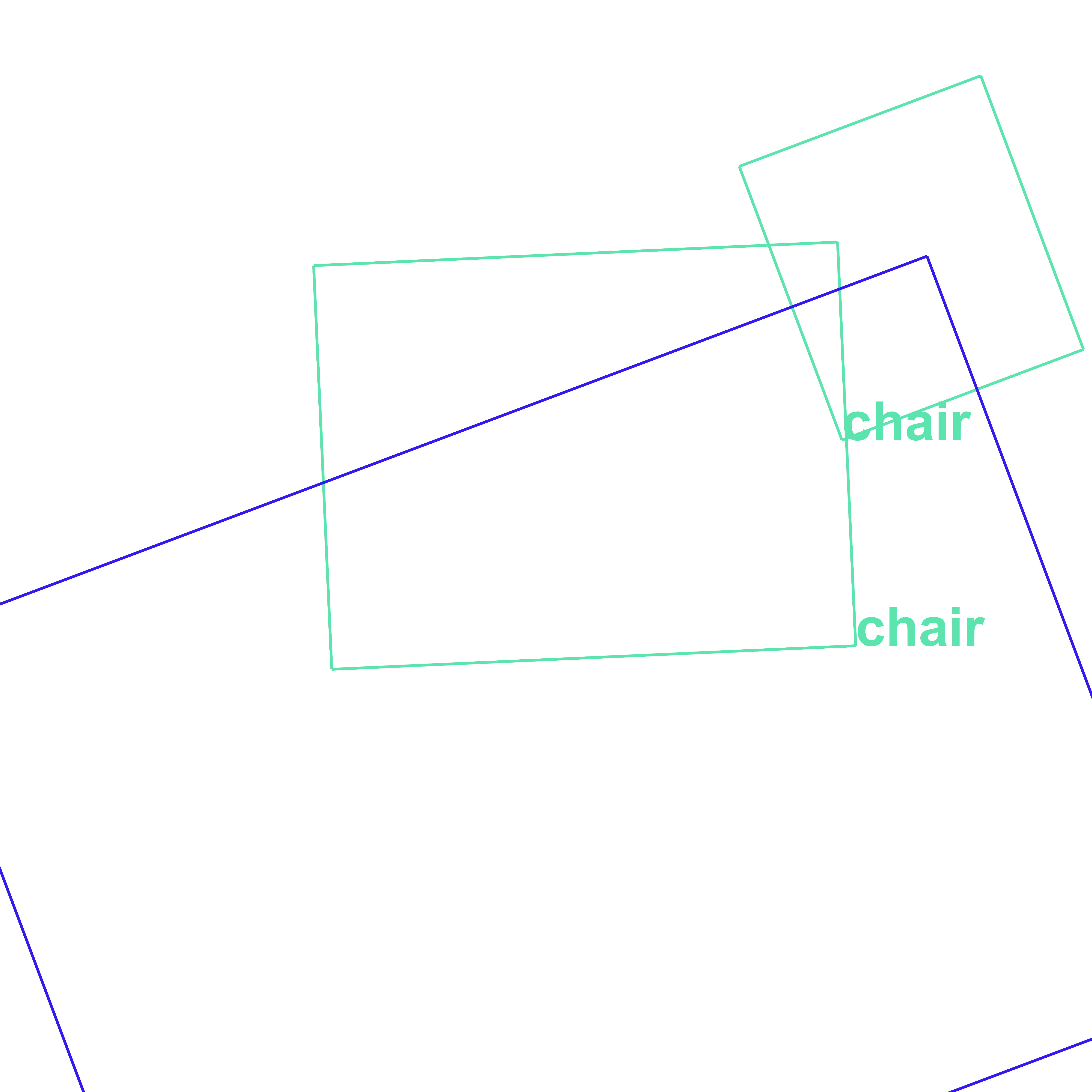} &
\includegraphics[width=\sunrgbdWidth,frame=.1pt]{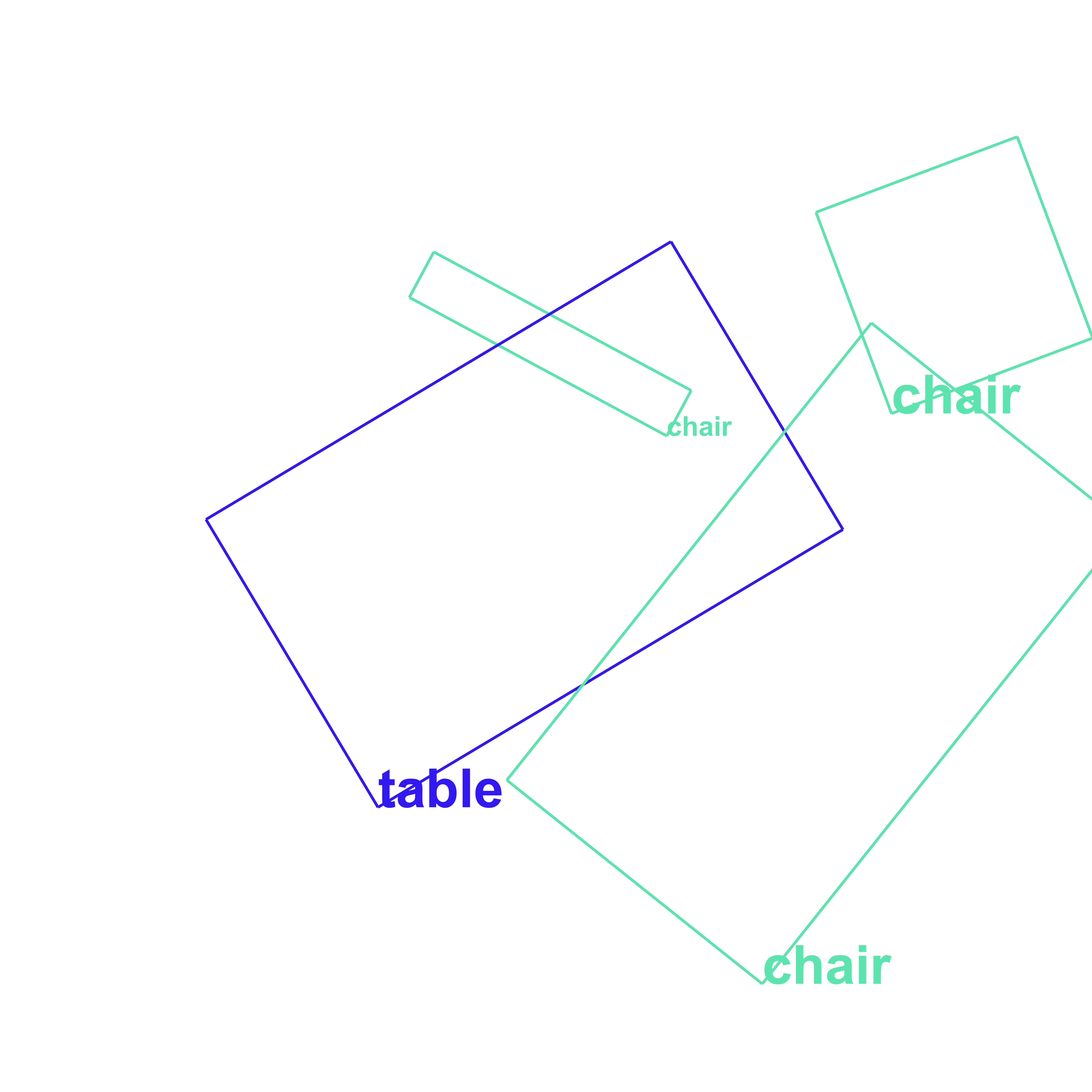} &
\includegraphics[width=\sunrgbdWidth,frame=.1pt]{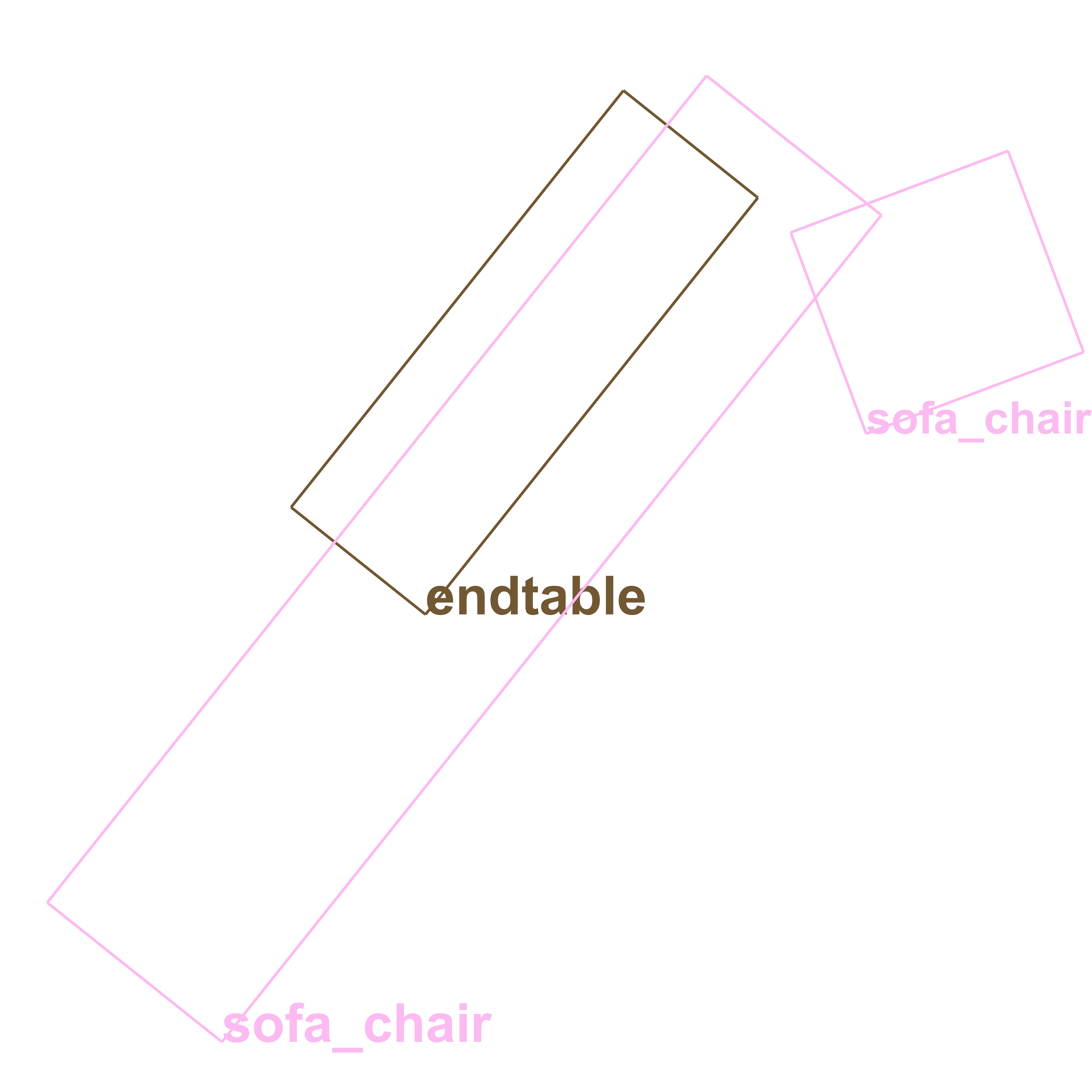} &
\includegraphics[width=\sunrgbdWidth,frame=.1pt]{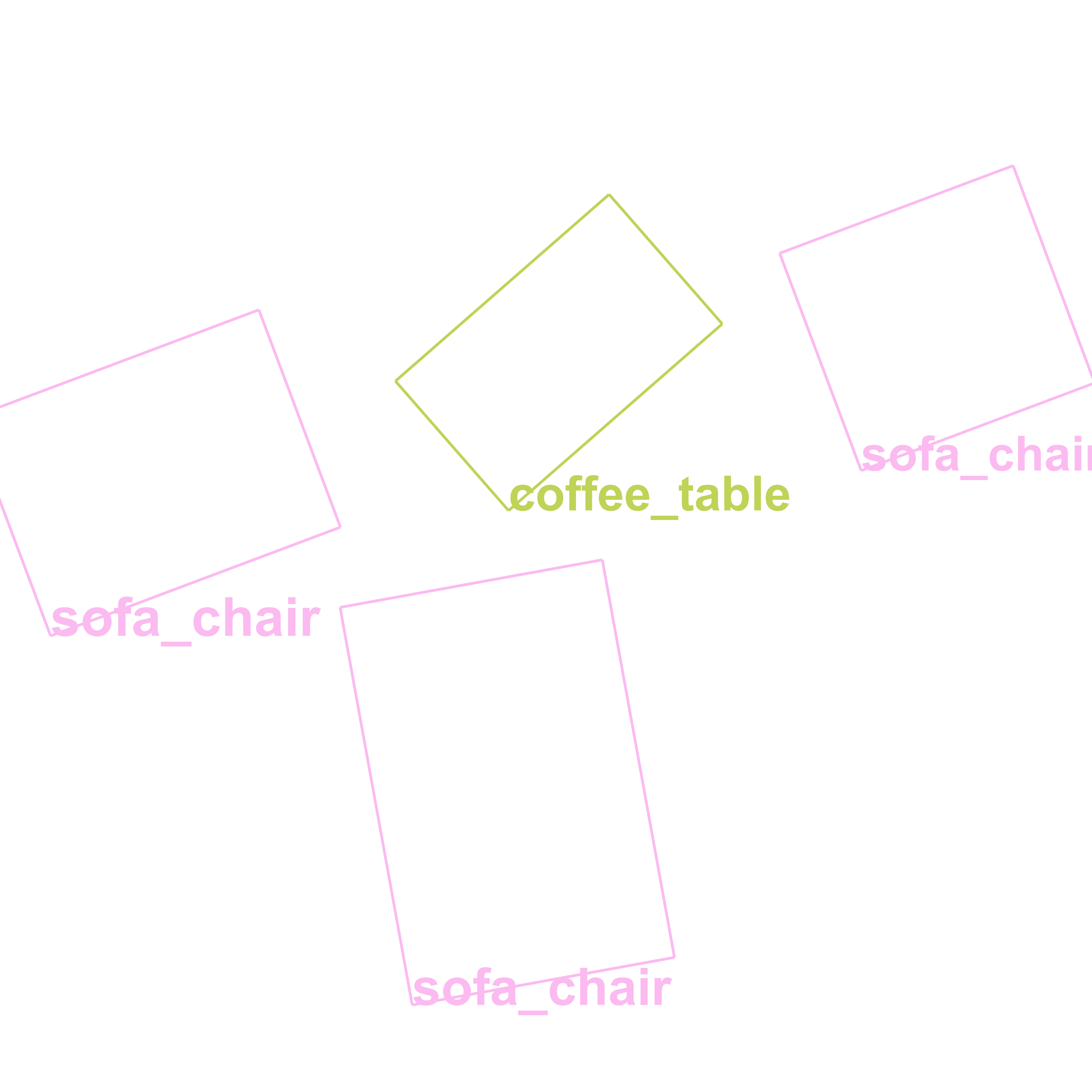} &
\includegraphics[width=\sunrgbdWidth,frame=.1pt]{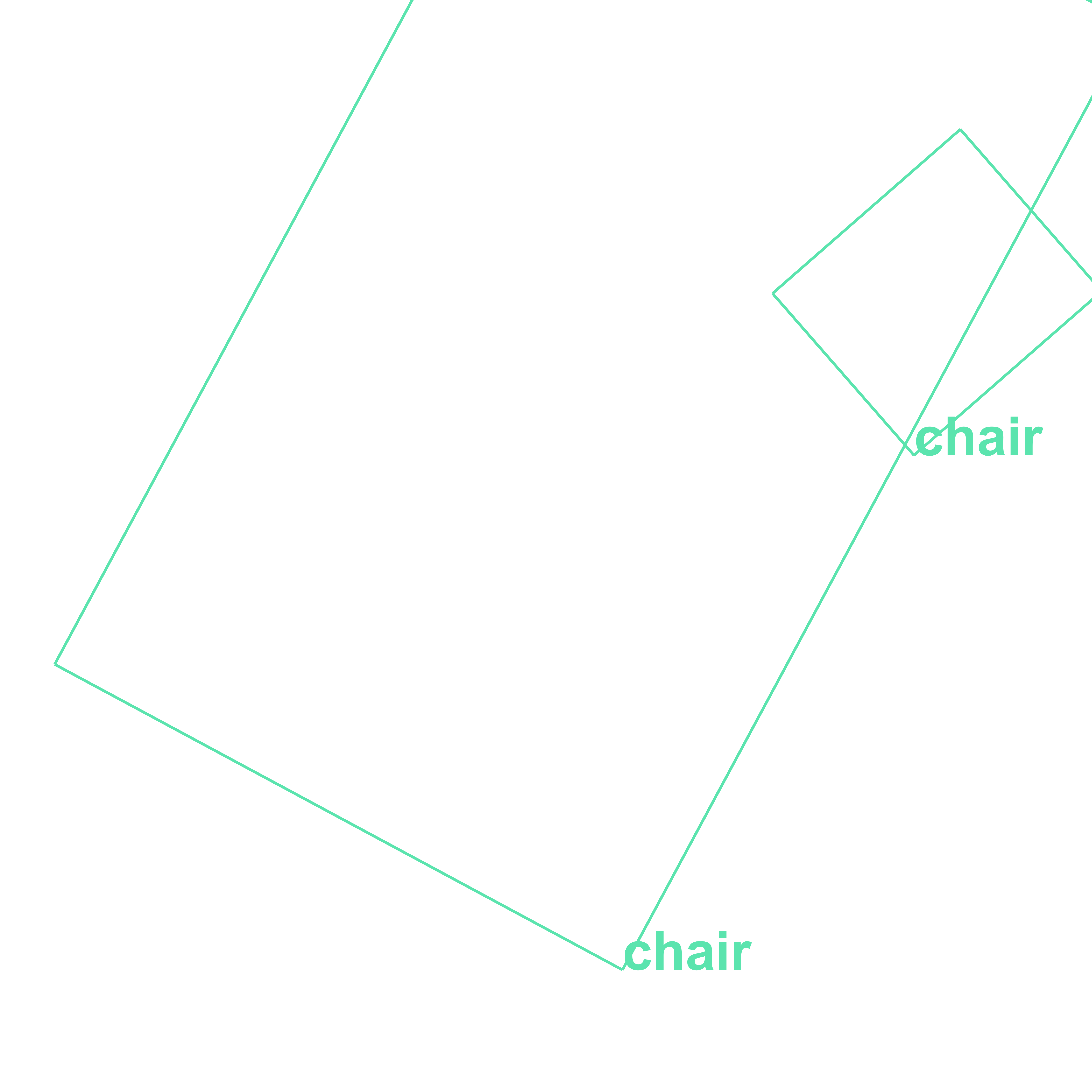} \\
\includegraphics[width=\sunrgbdWidth,frame=.1pt]{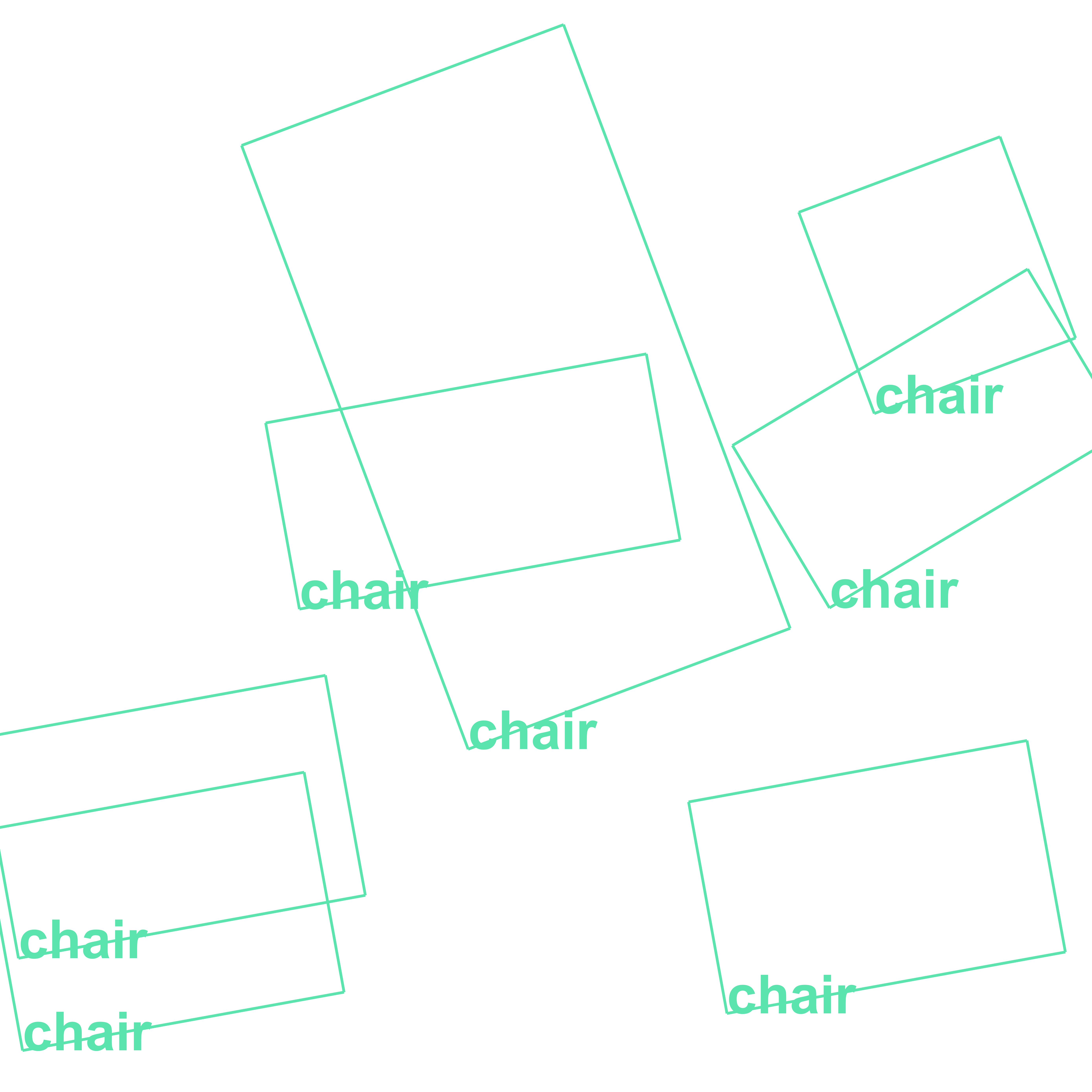} &
\includegraphics[width=\sunrgbdWidth,frame=.1pt]{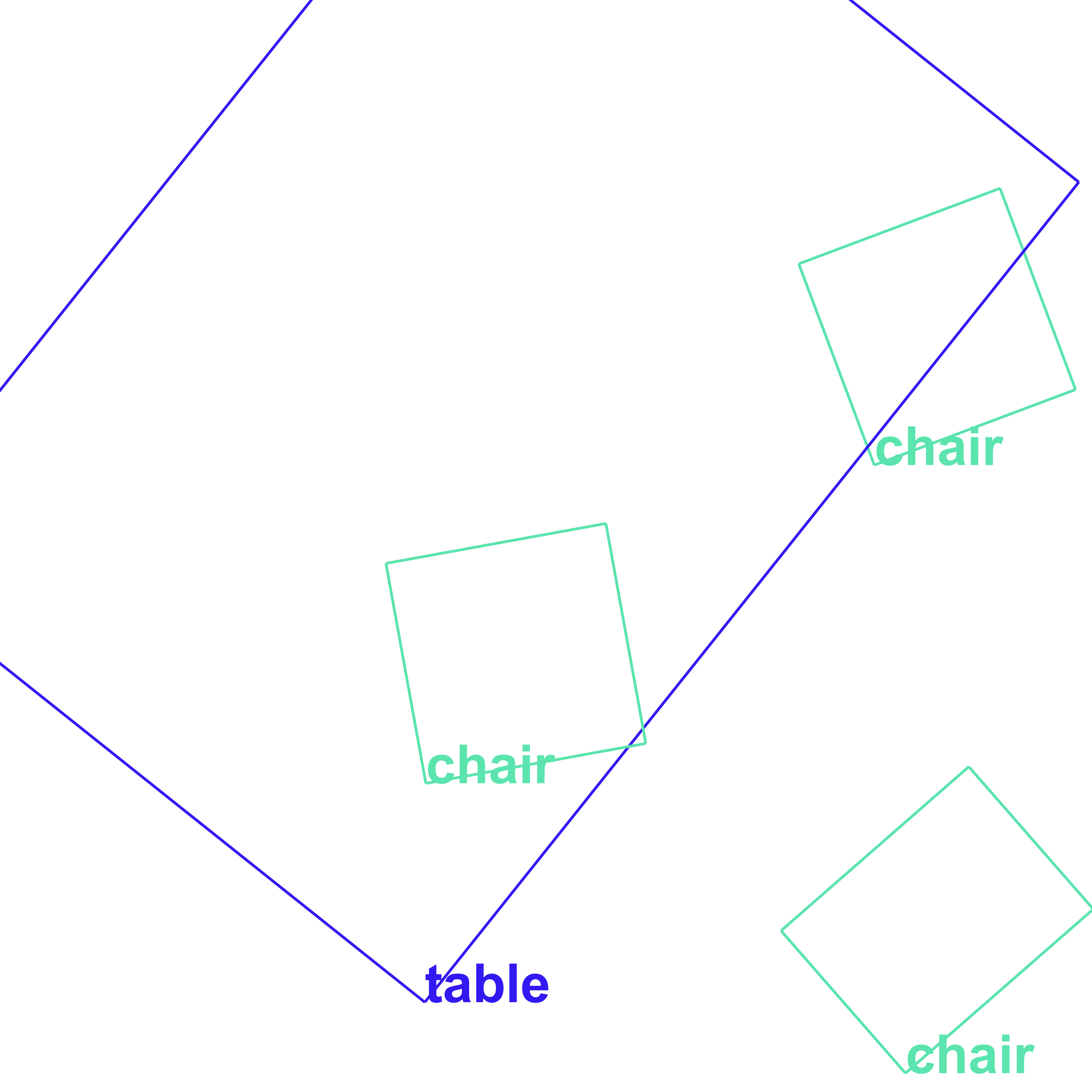} &
\includegraphics[width=\sunrgbdWidth,frame=.1pt]{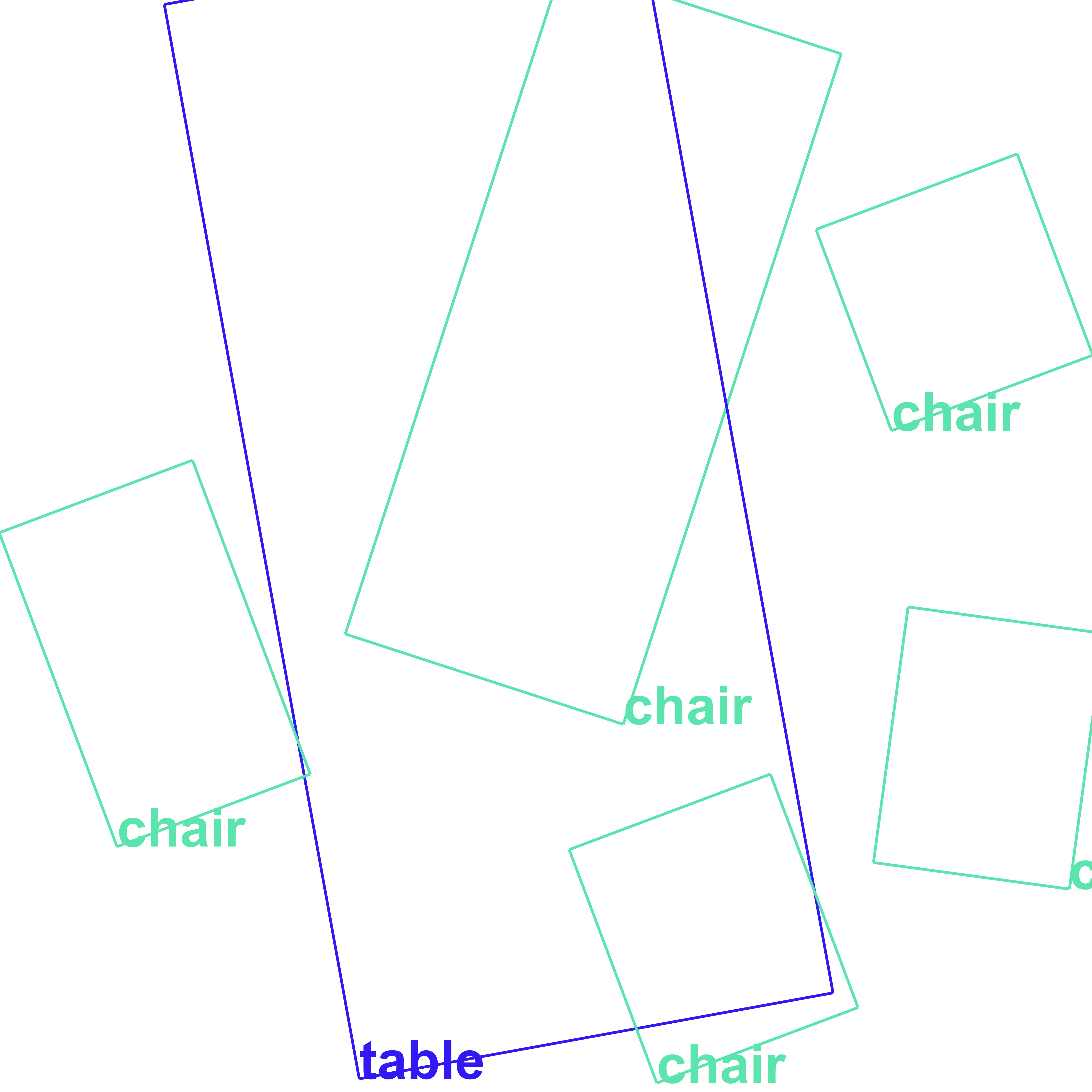} &
\includegraphics[width=\sunrgbdWidth,frame=.1pt]{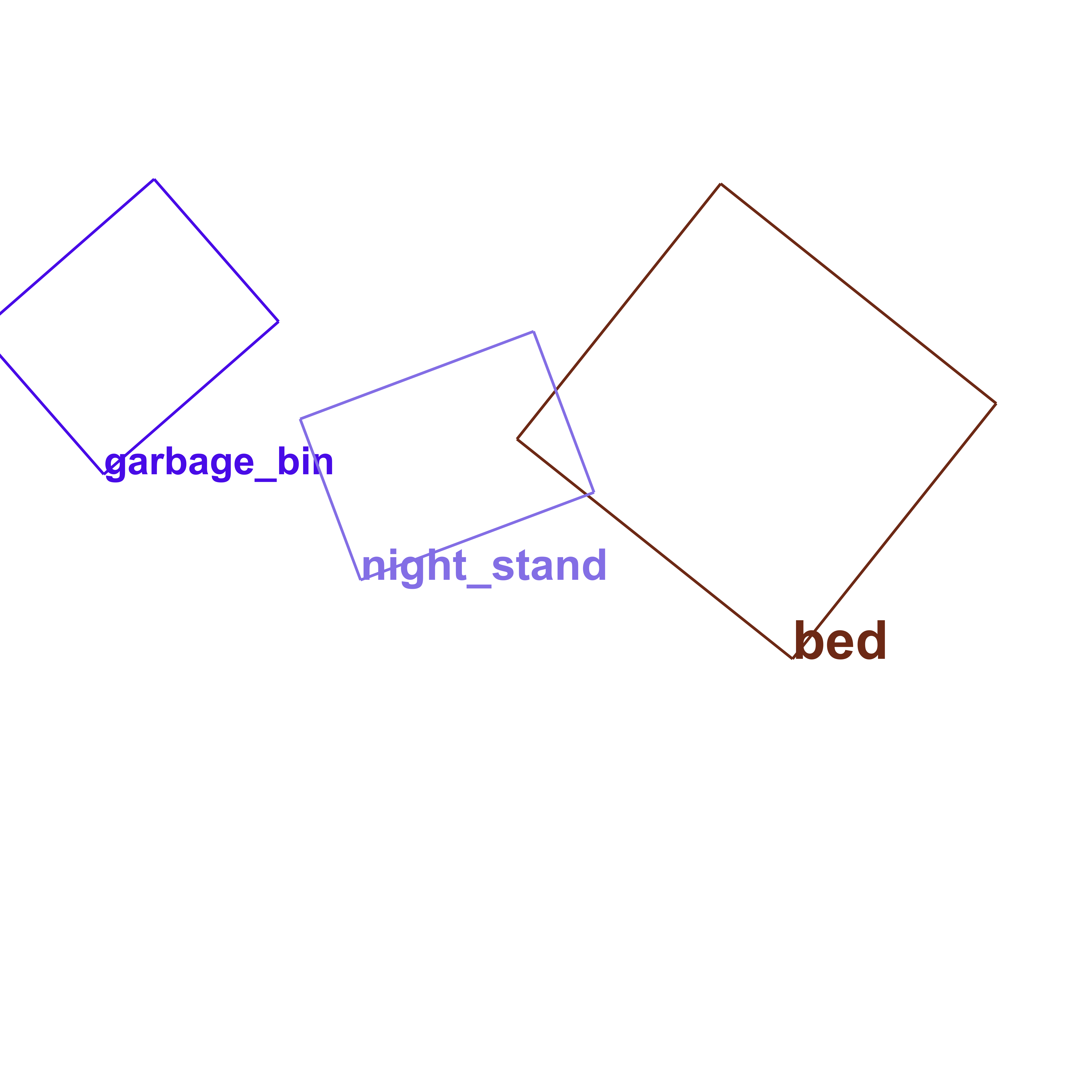} &
\includegraphics[width=\sunrgbdWidth,frame=.1pt]{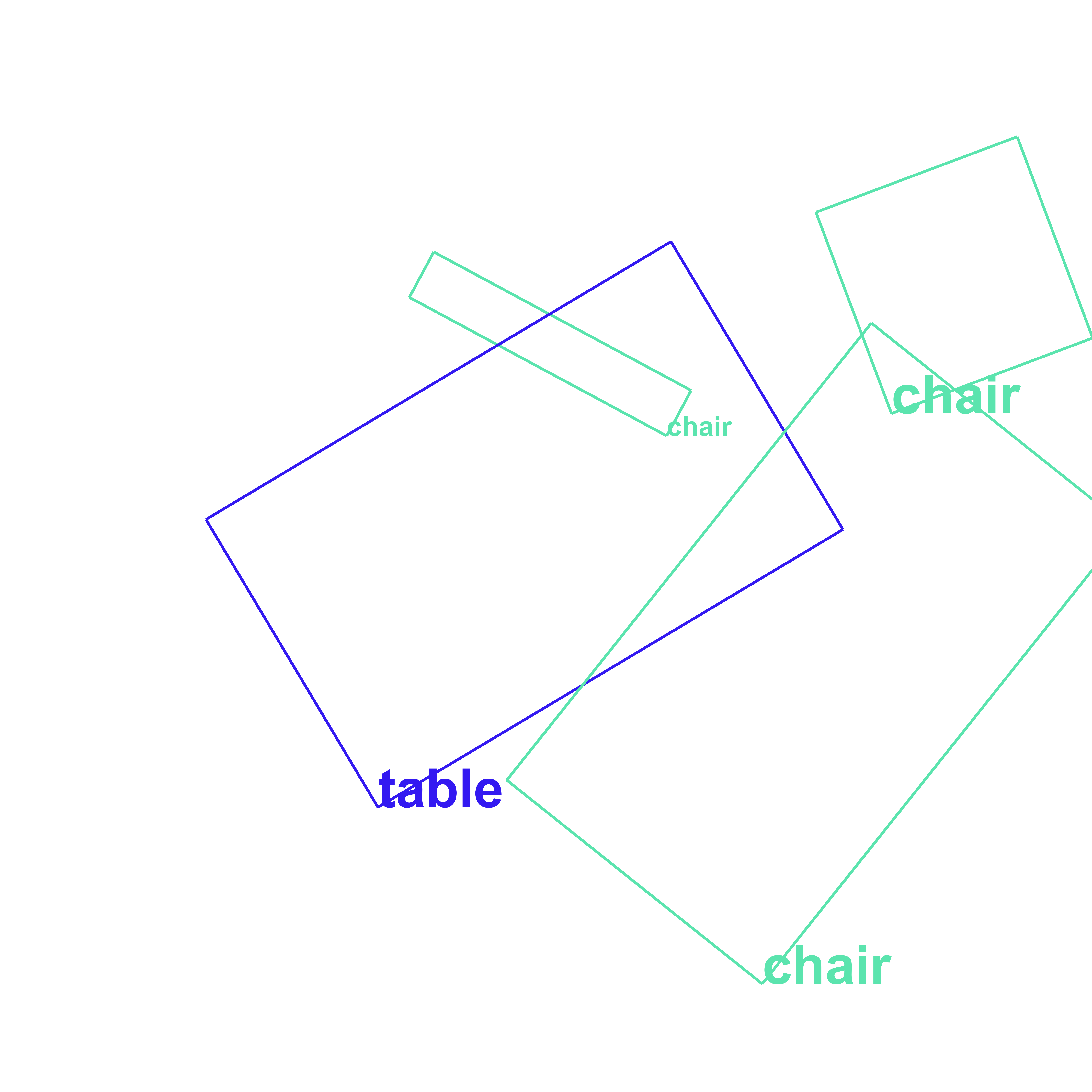} &
\includegraphics[width=\sunrgbdWidth,frame=.1pt]{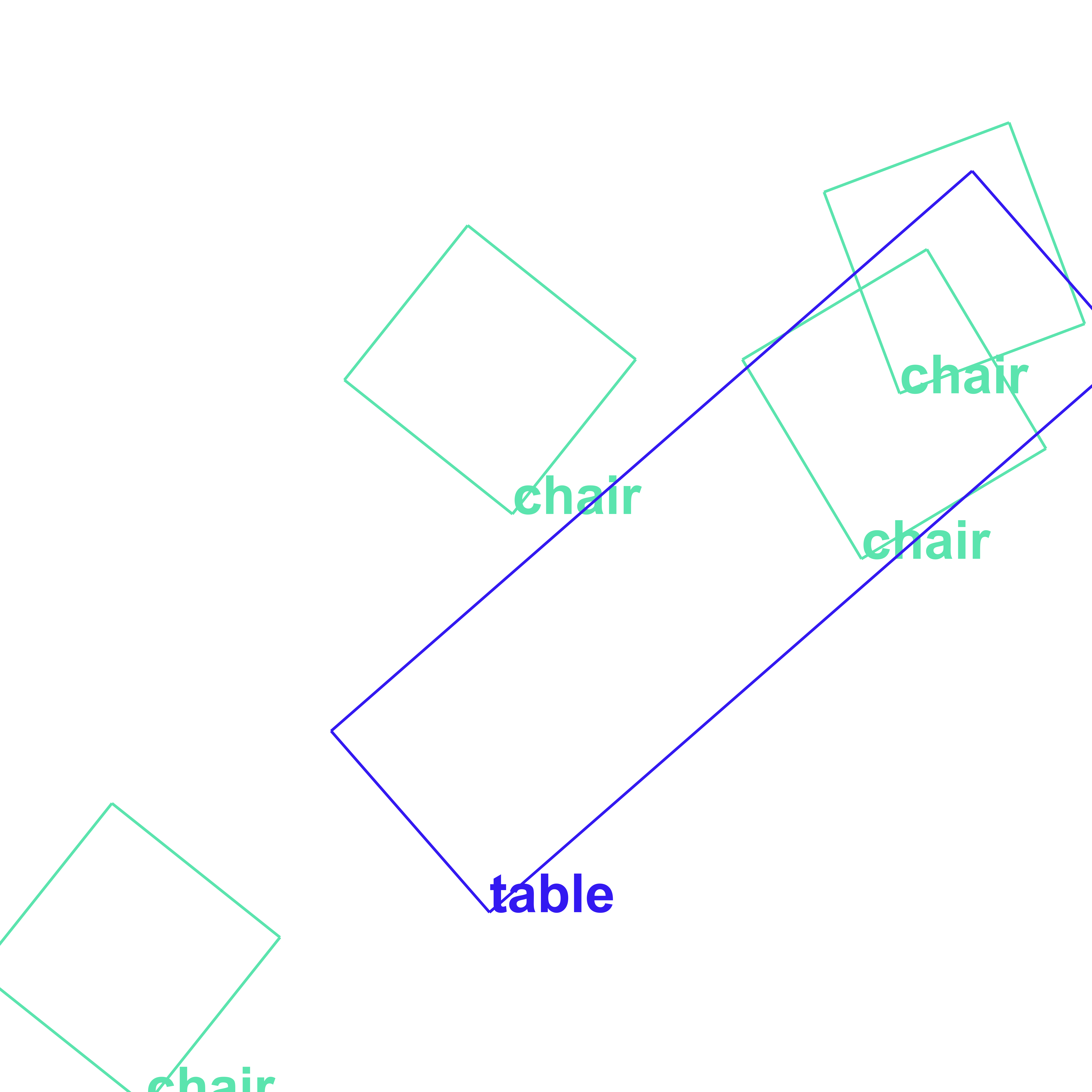} \\
\includegraphics[width=\sunrgbdWidth,frame=.1pt]{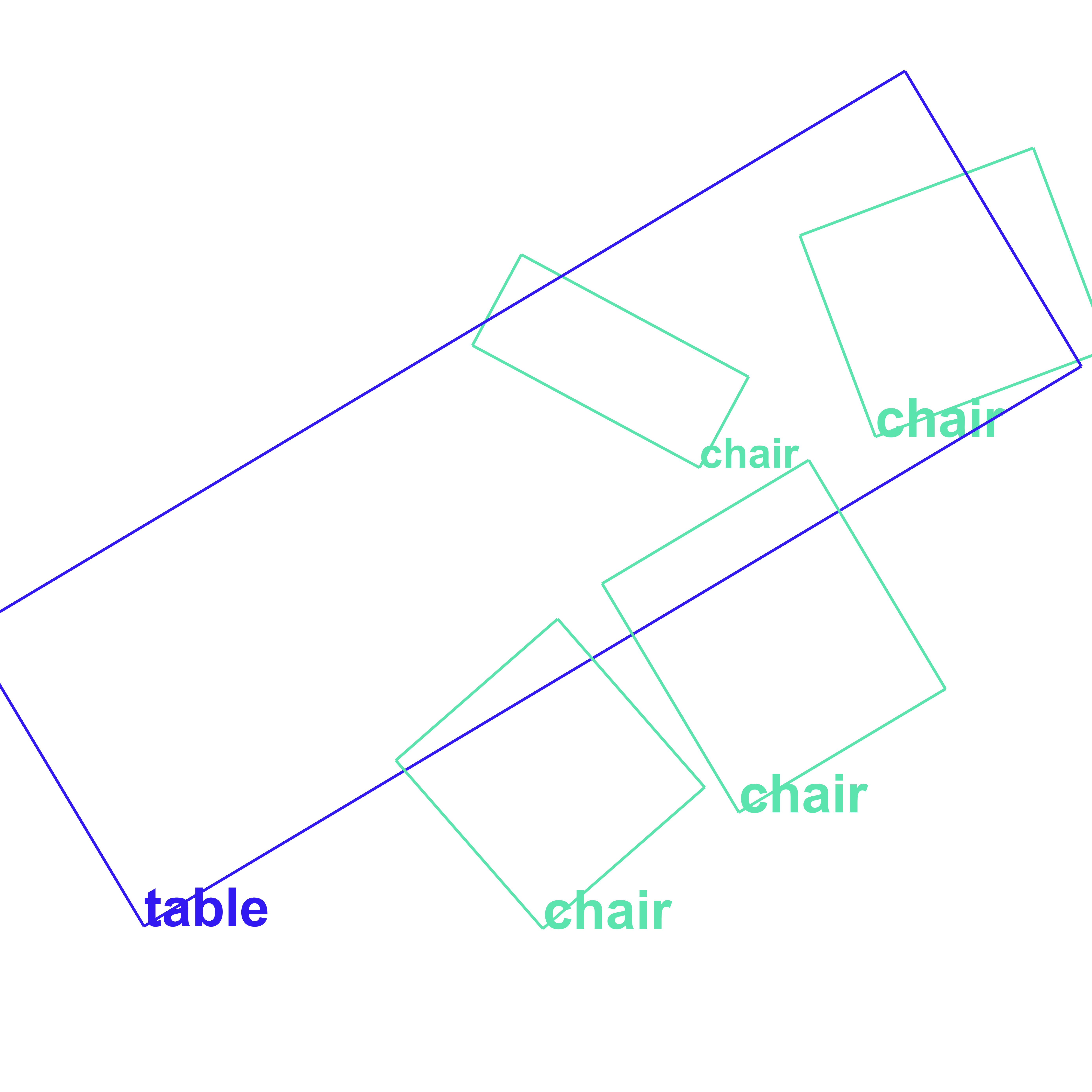} &
\includegraphics[width=\sunrgbdWidth,frame=.1pt]{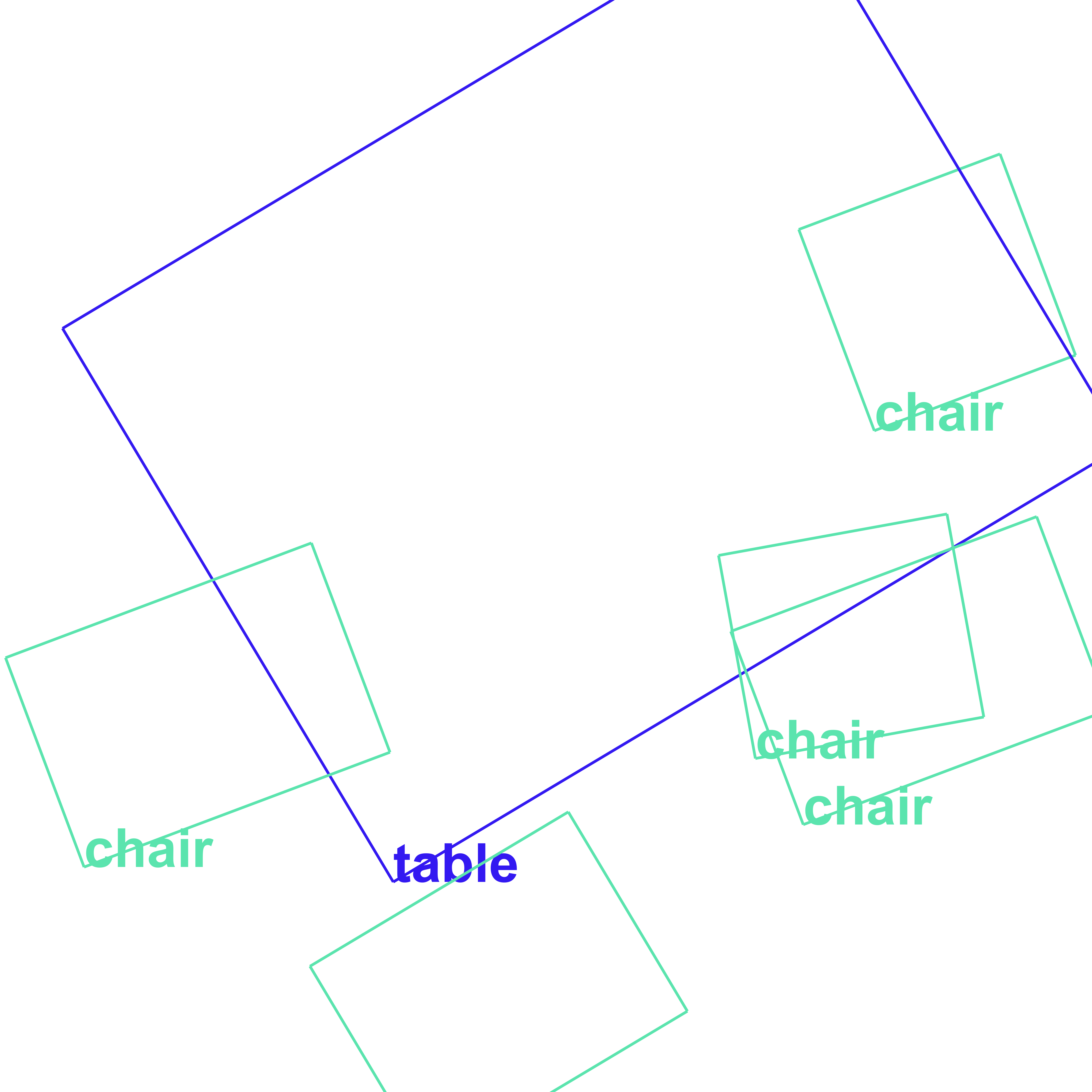} &
\includegraphics[width=\sunrgbdWidth,frame=.1pt]{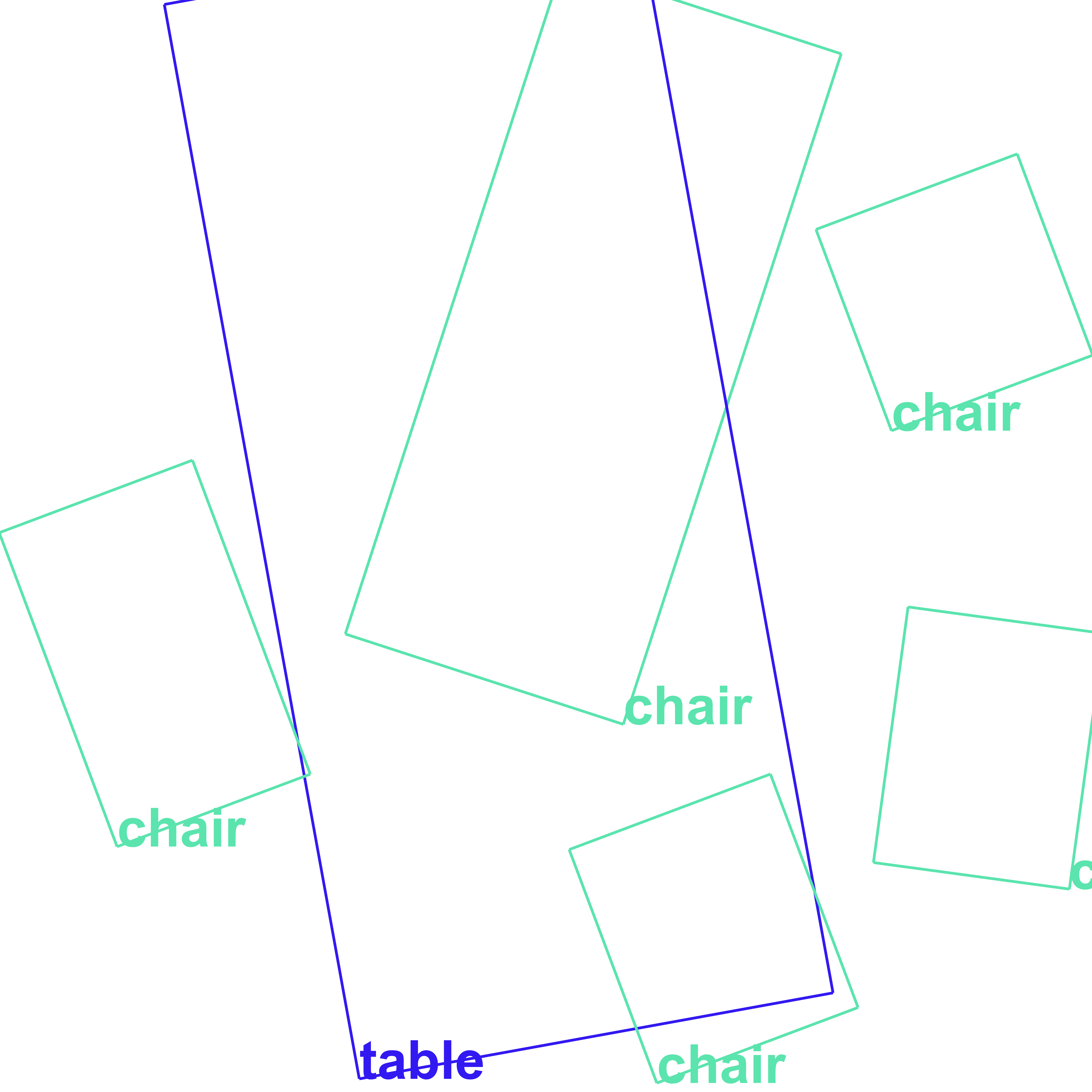} &
\includegraphics[width=\sunrgbdWidth,frame=.1pt]{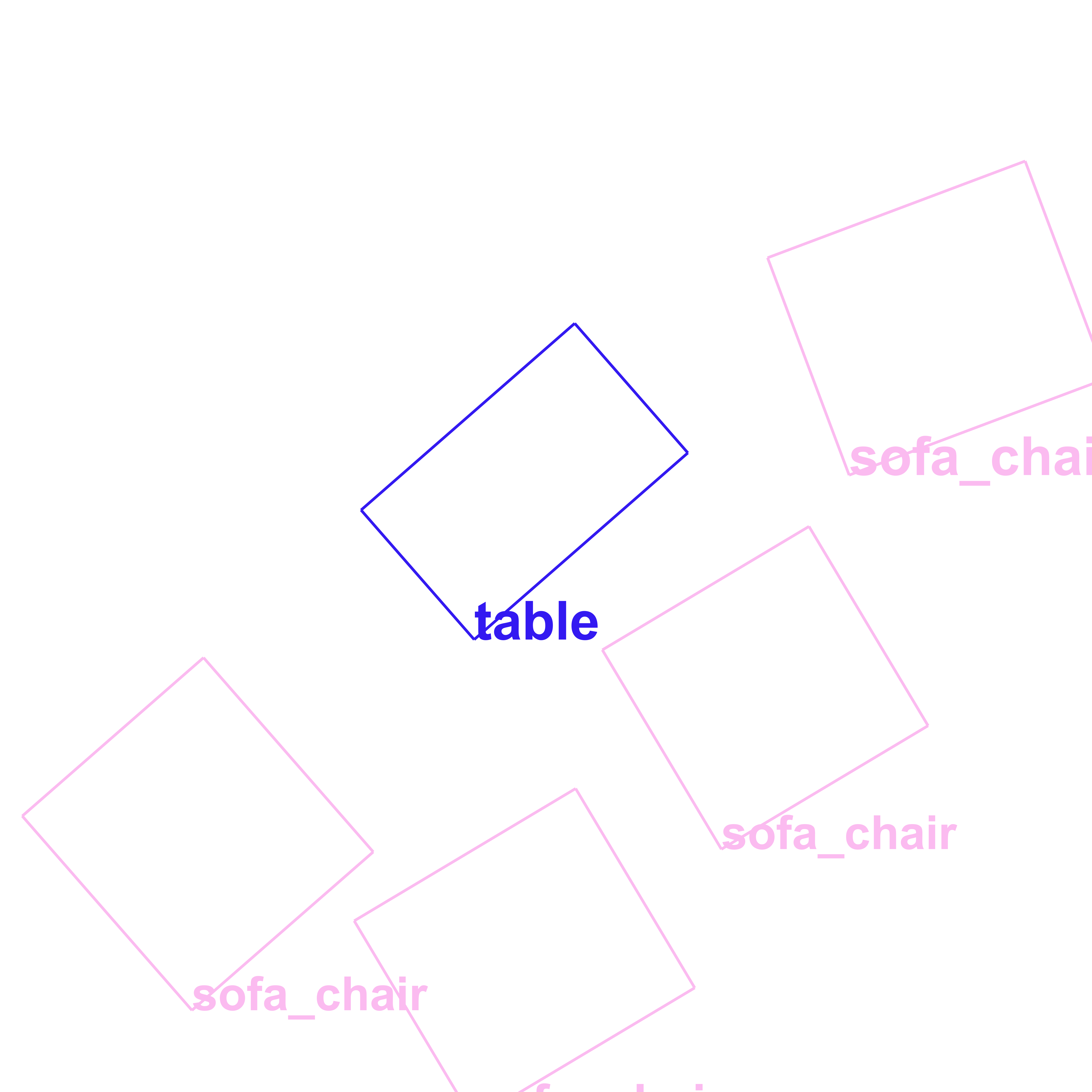} &
\includegraphics[width=\sunrgbdWidth,frame=.1pt]{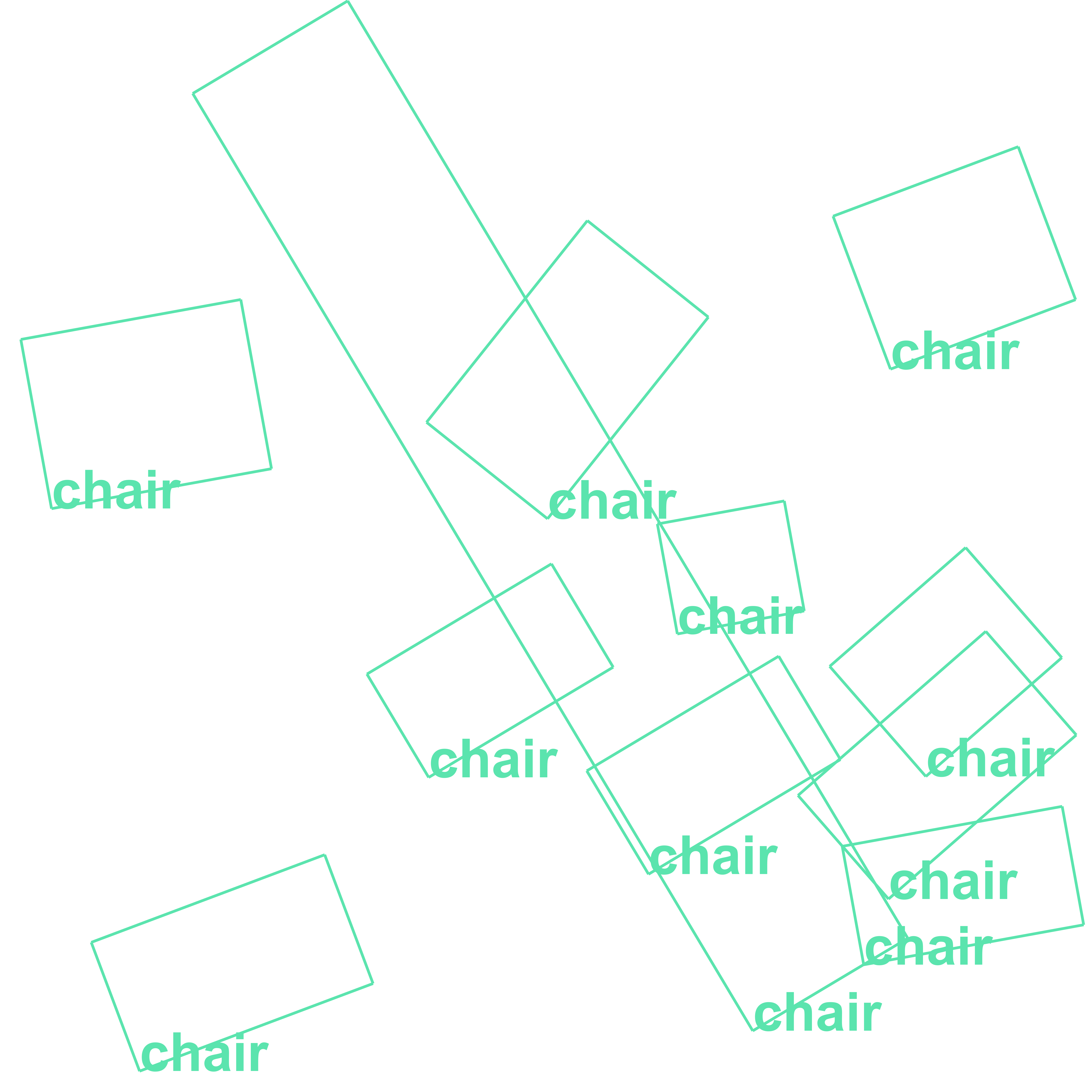} &
\includegraphics[width=\sunrgbdWidth,frame=.1pt]{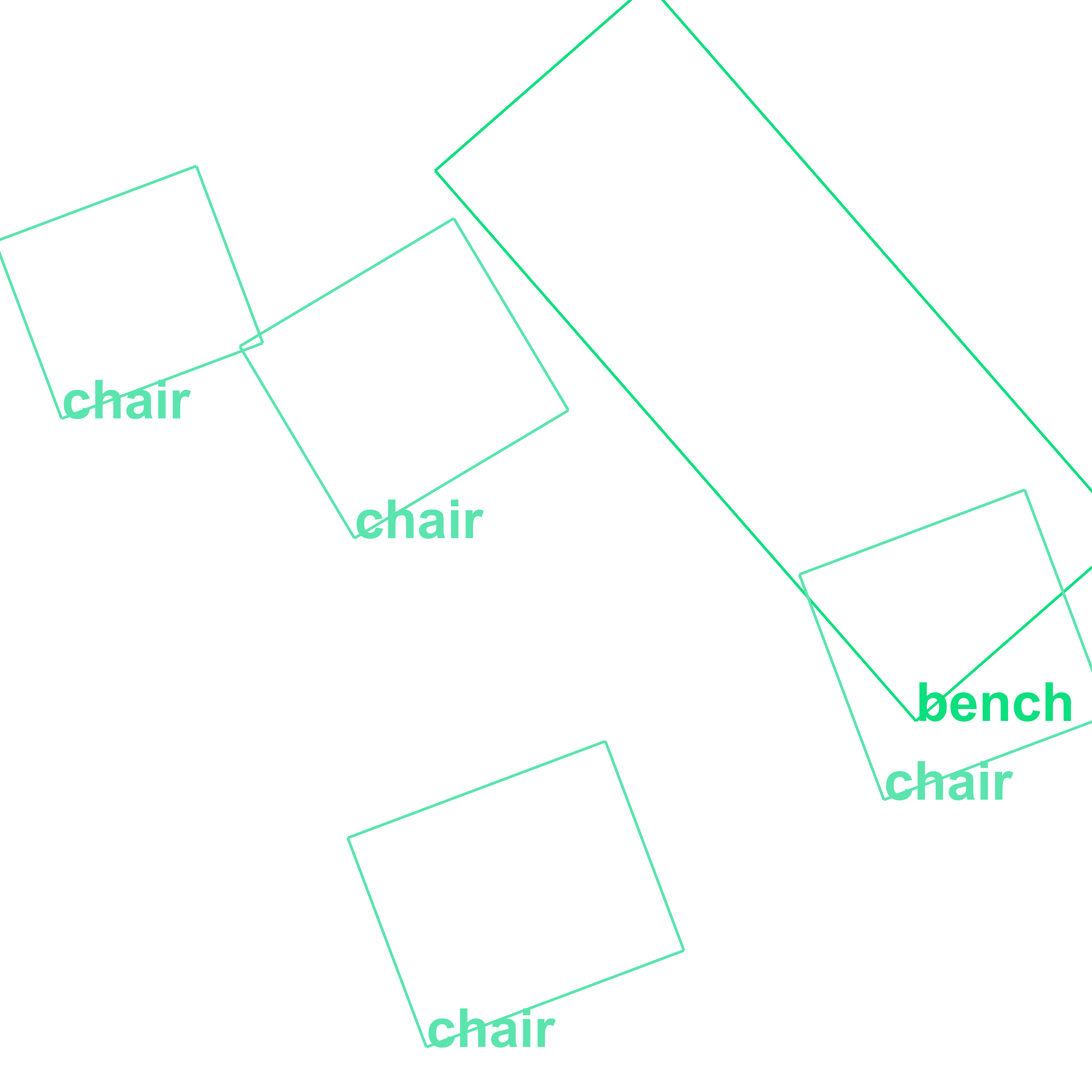} \\
\includegraphics[width=\sunrgbdWidth,frame=.1pt]{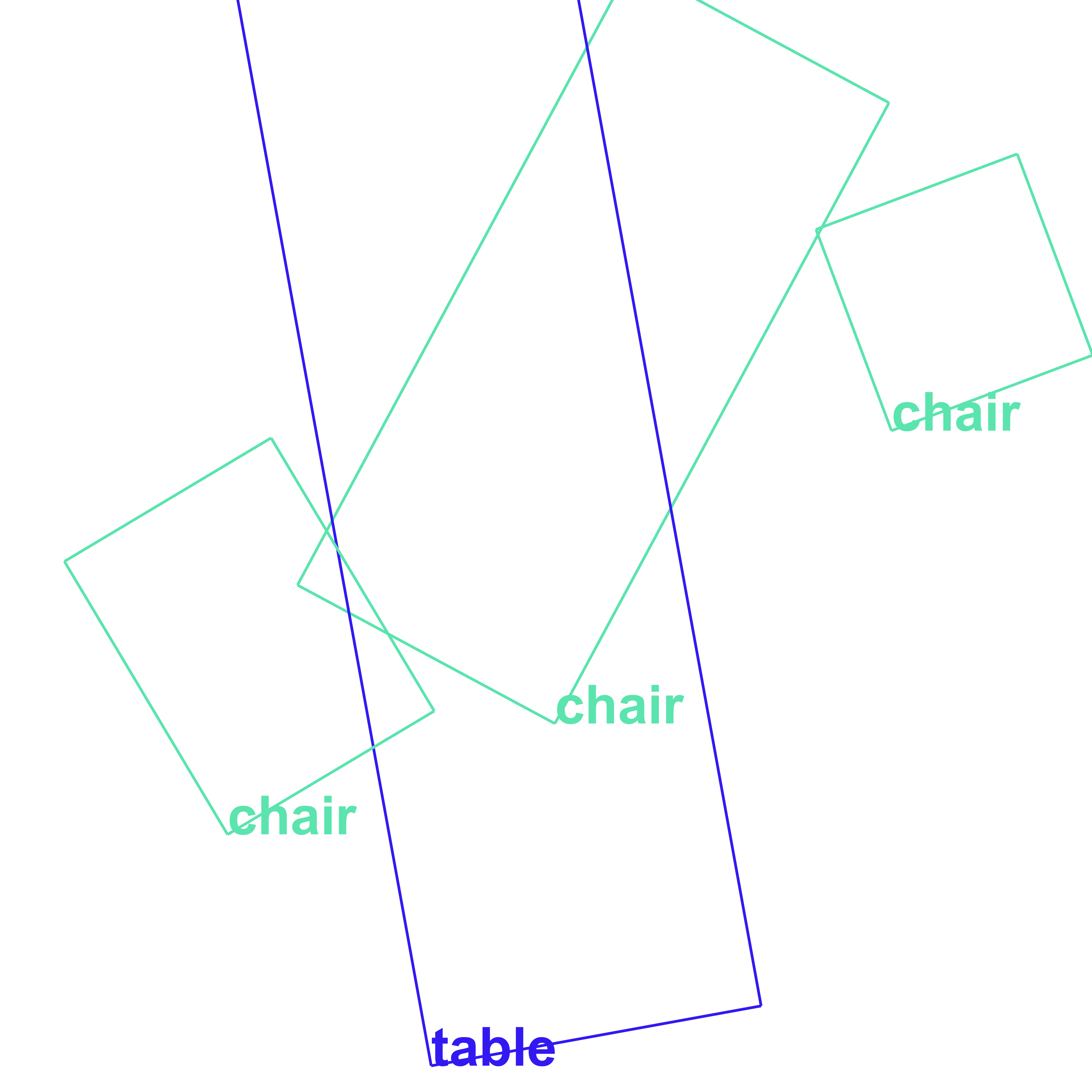} &
\includegraphics[width=\sunrgbdWidth,frame=.1pt]{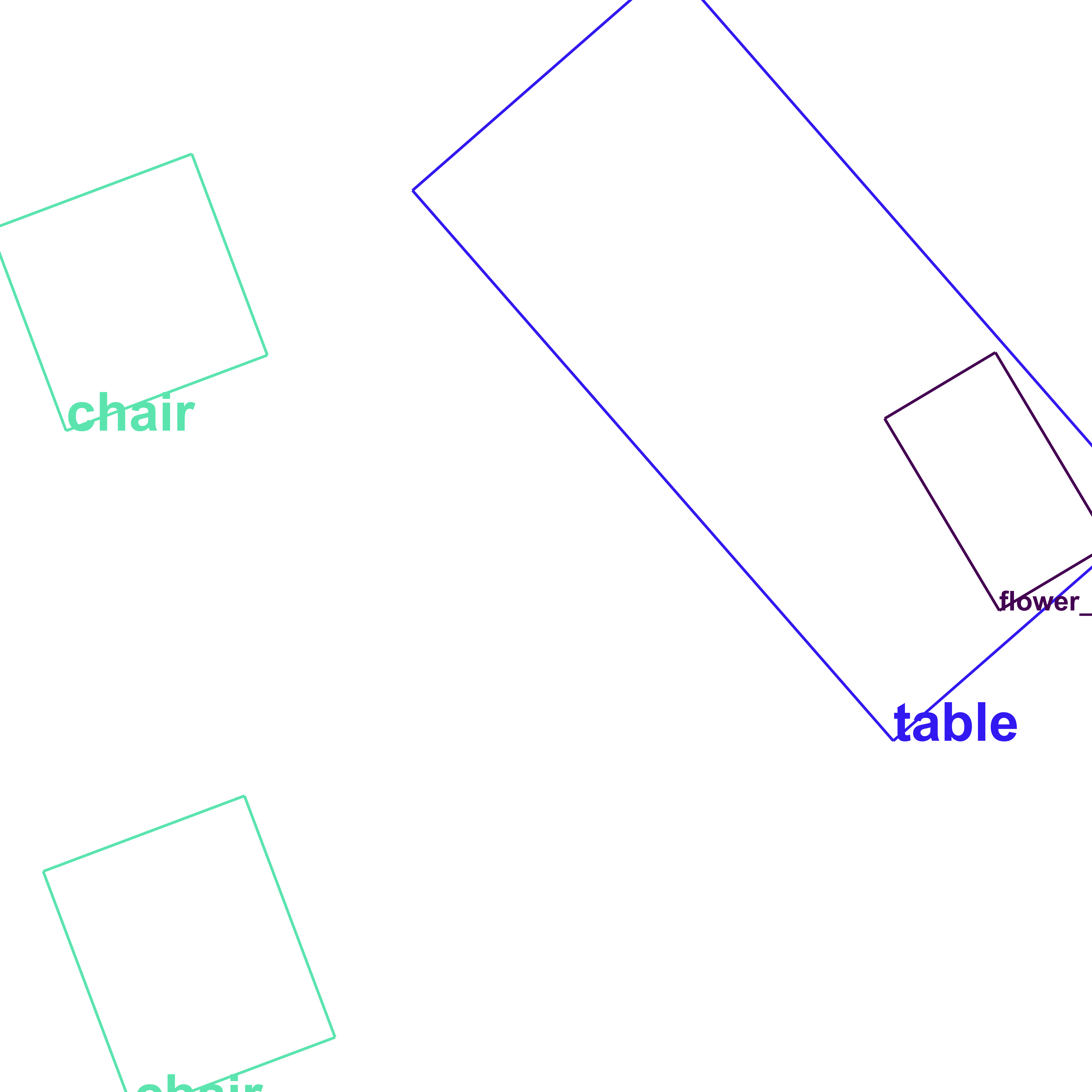} &
\includegraphics[width=\sunrgbdWidth,frame=.1pt]{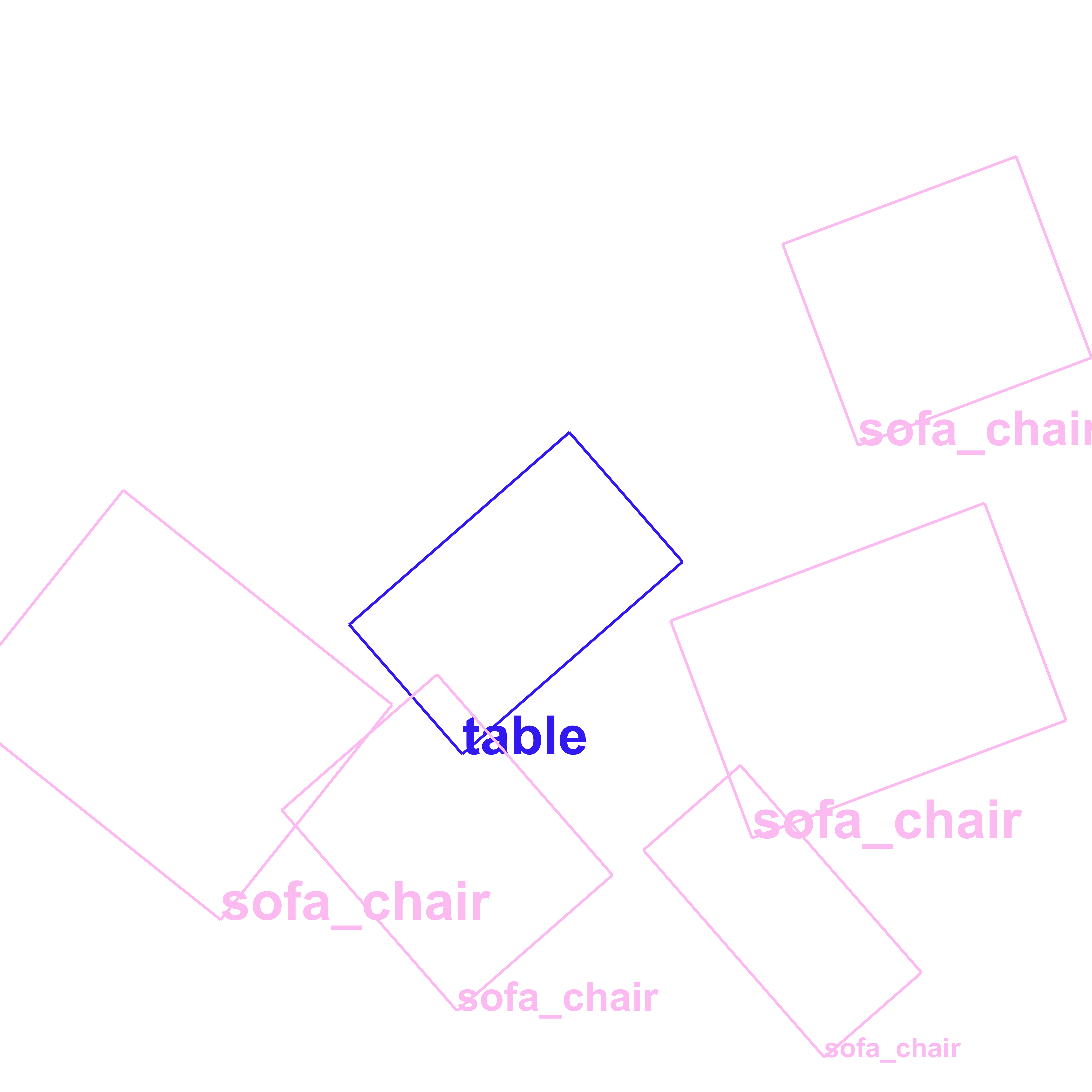} &
\includegraphics[width=\sunrgbdWidth,frame=.1pt]{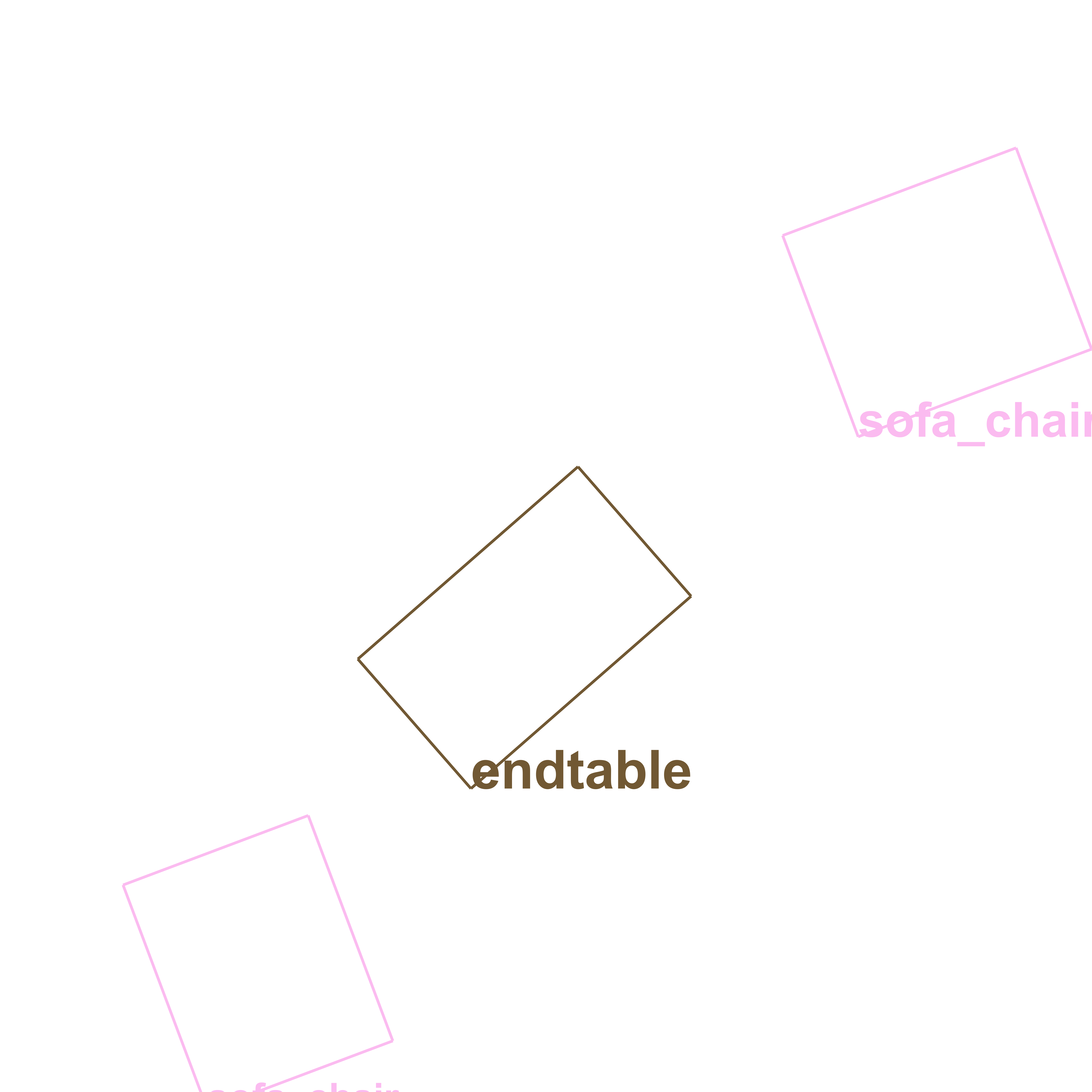} &
\includegraphics[width=\sunrgbdWidth,frame=.1pt]{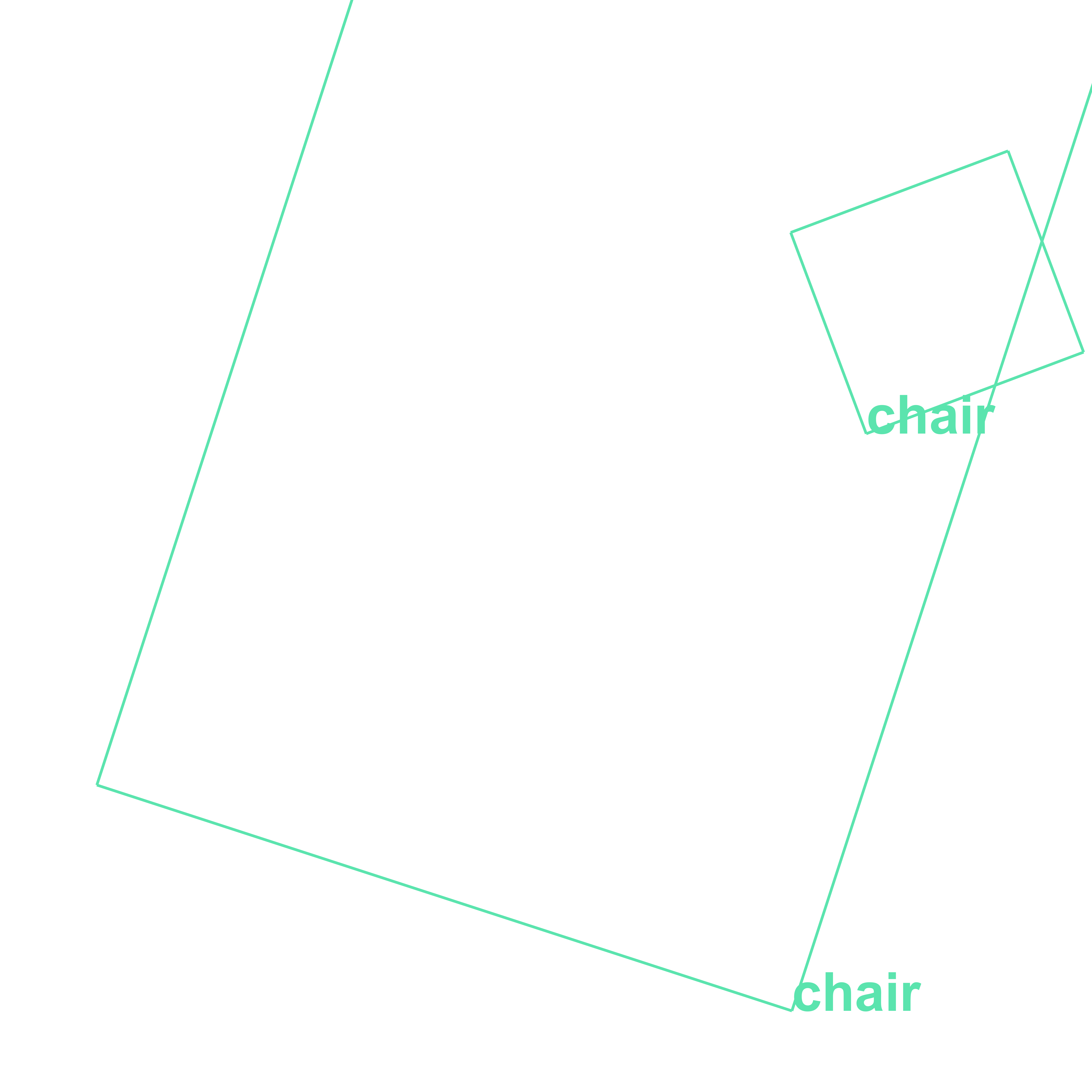} &
\includegraphics[width=\sunrgbdWidth,frame=.1pt]{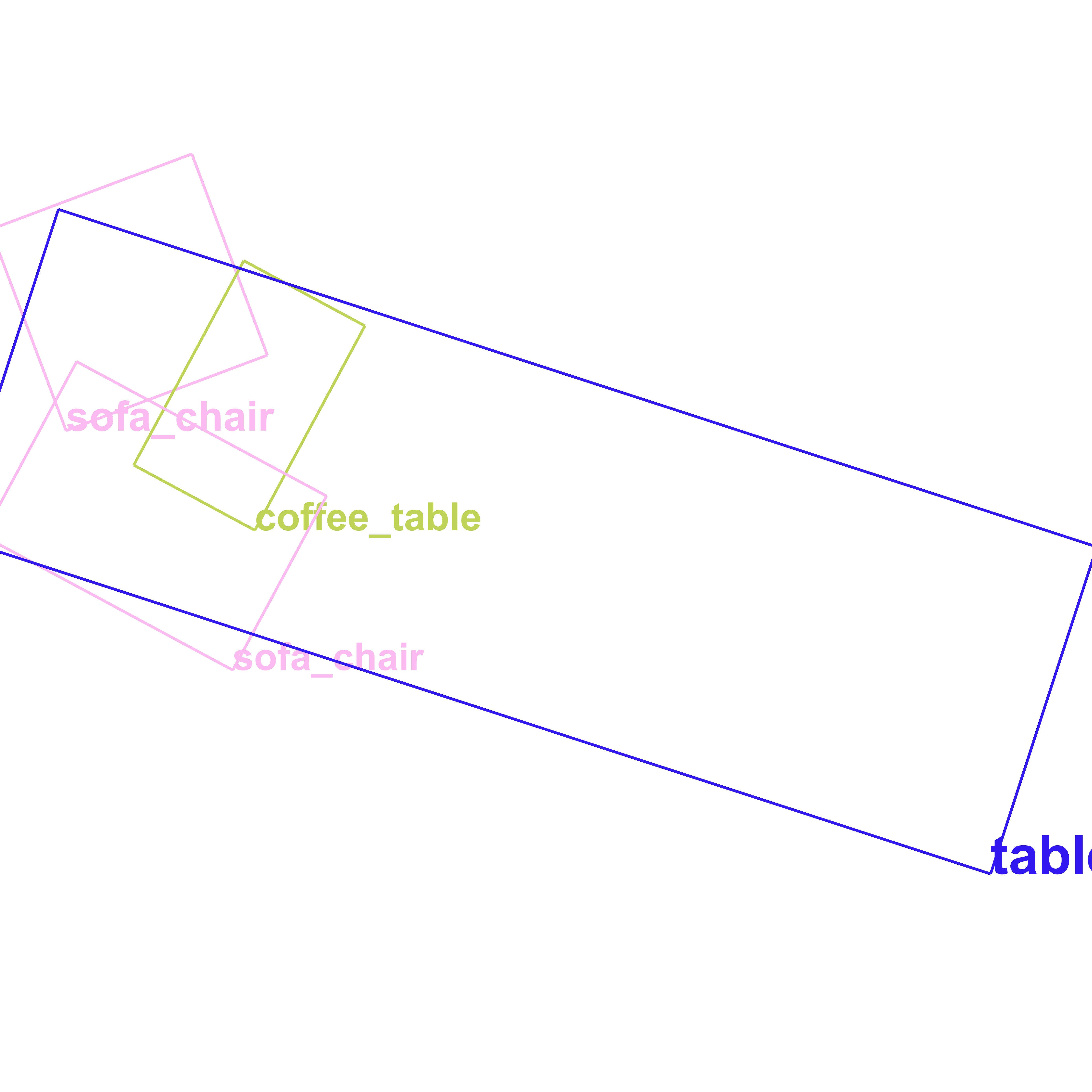} \\
\includegraphics[width=\sunrgbdWidth,frame=.1pt]{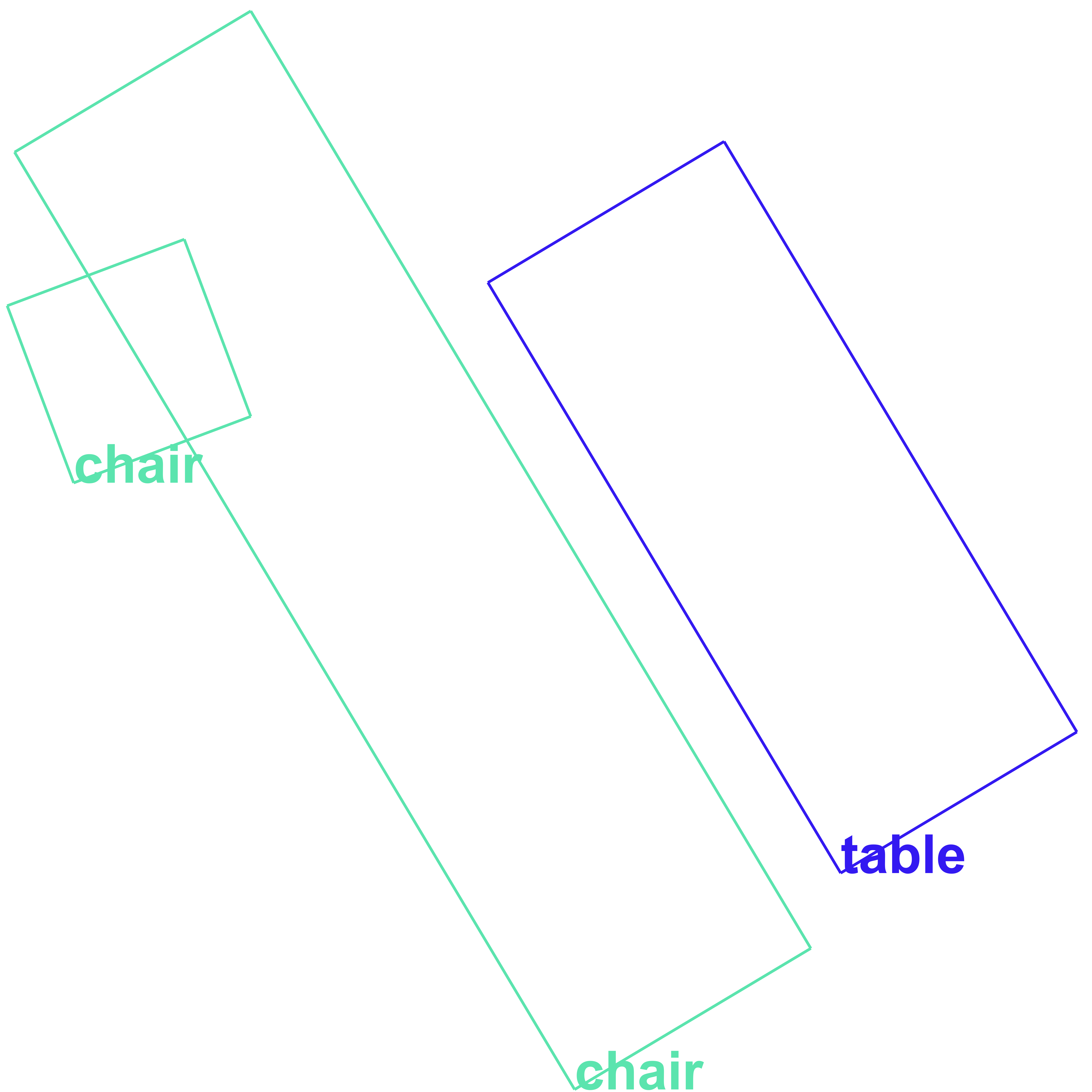} &
\includegraphics[width=\sunrgbdWidth,frame=.1pt]{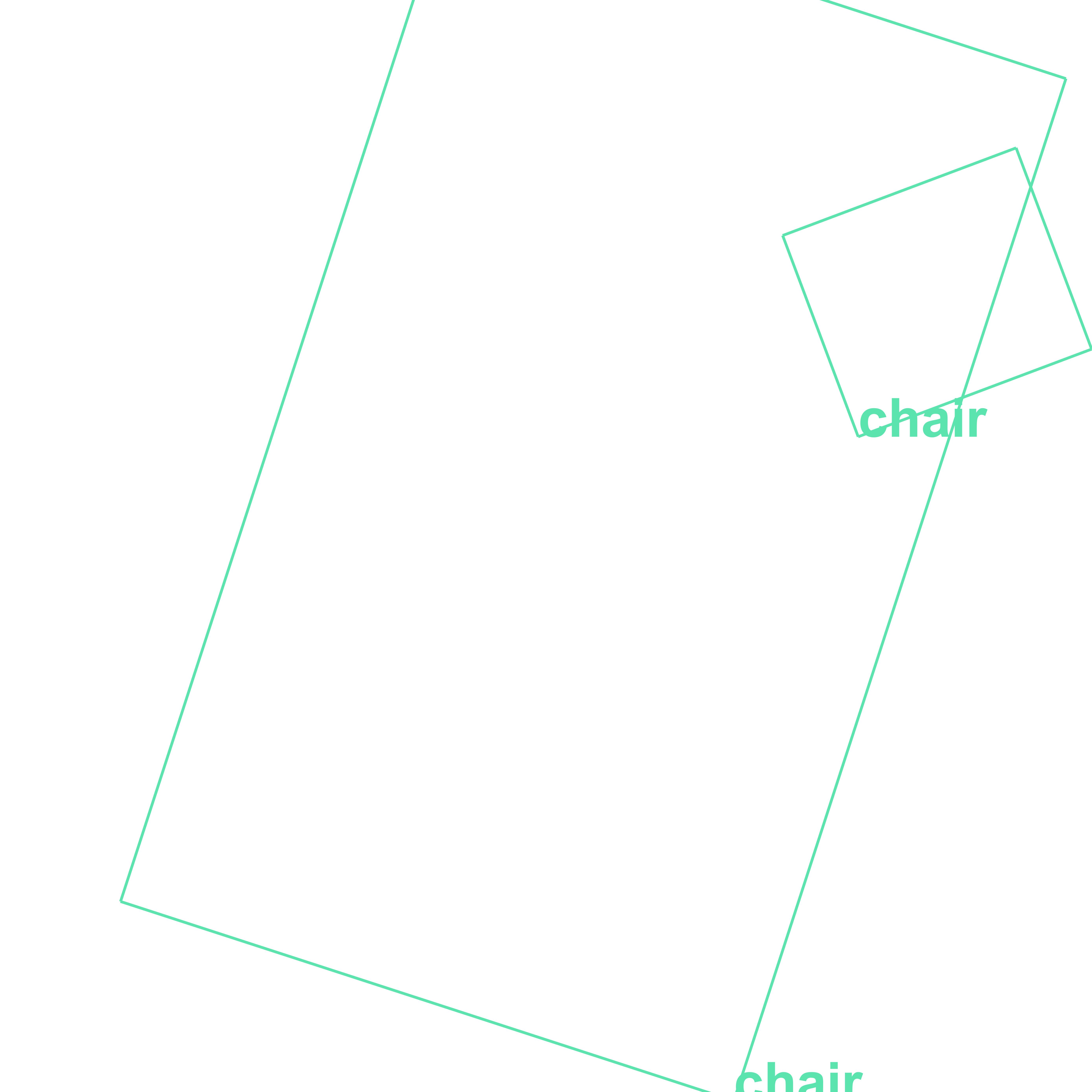} &
\includegraphics[width=\sunrgbdWidth,frame=.1pt]{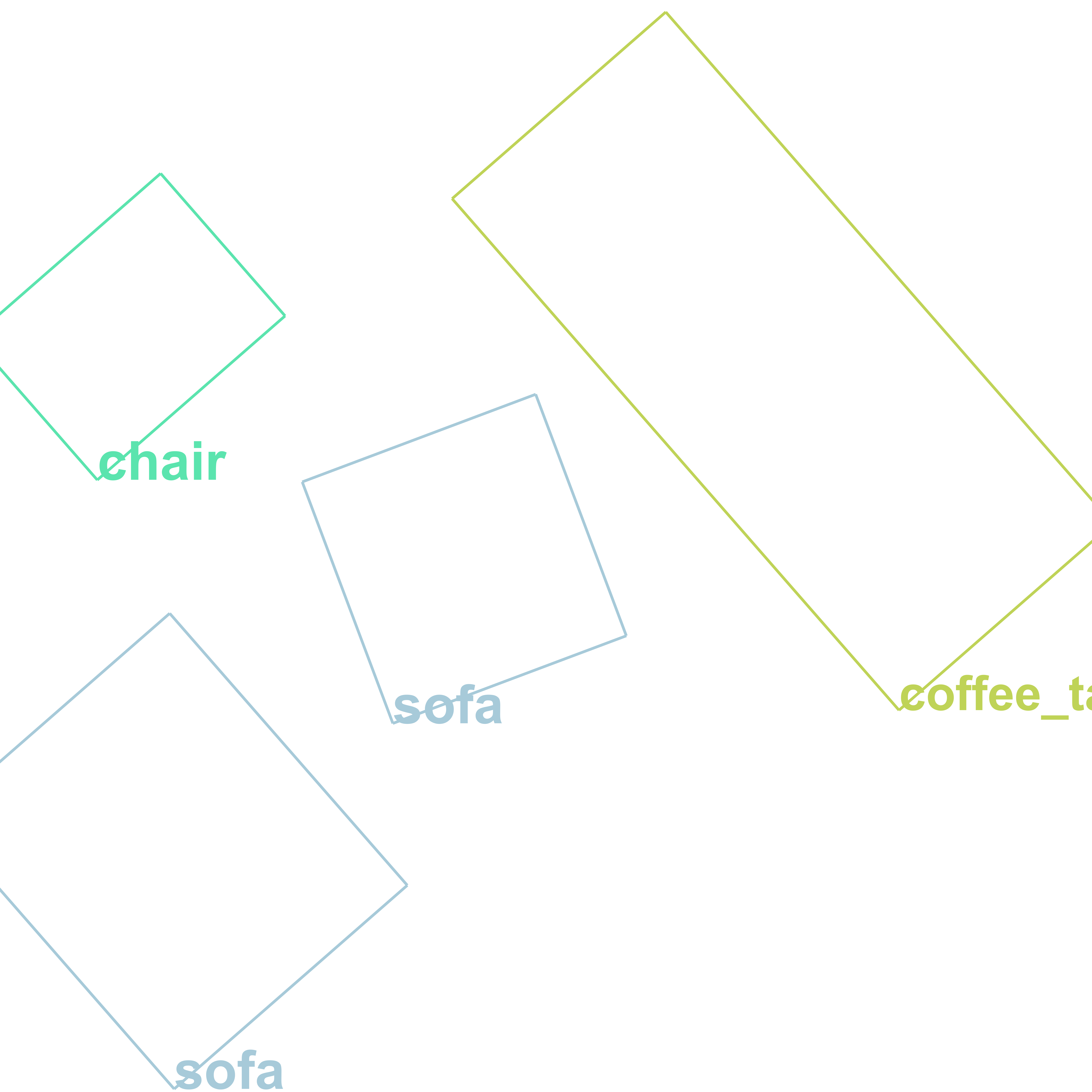} &
\includegraphics[width=\sunrgbdWidth,frame=.1pt]{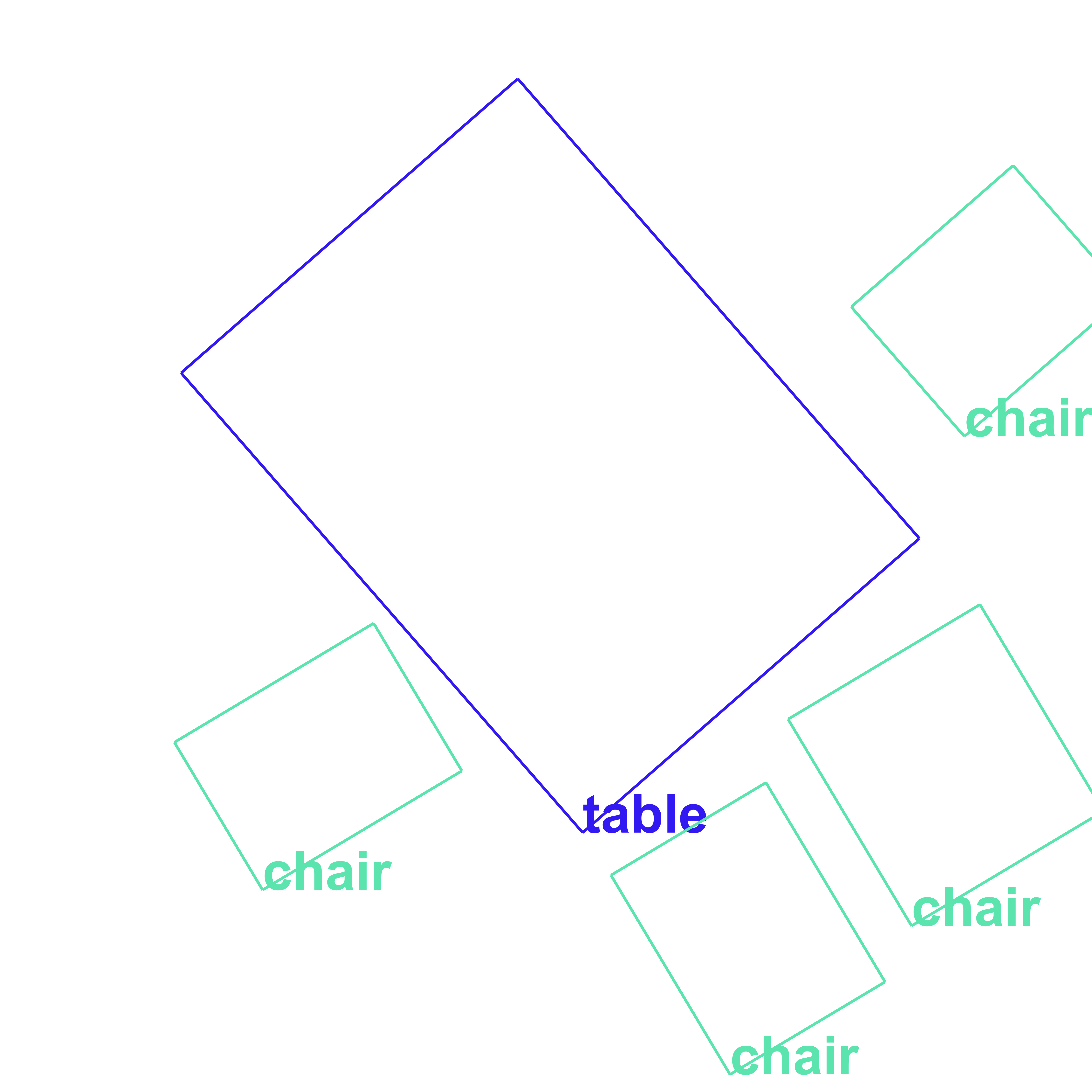} &
\includegraphics[width=\sunrgbdWidth,frame=.1pt]{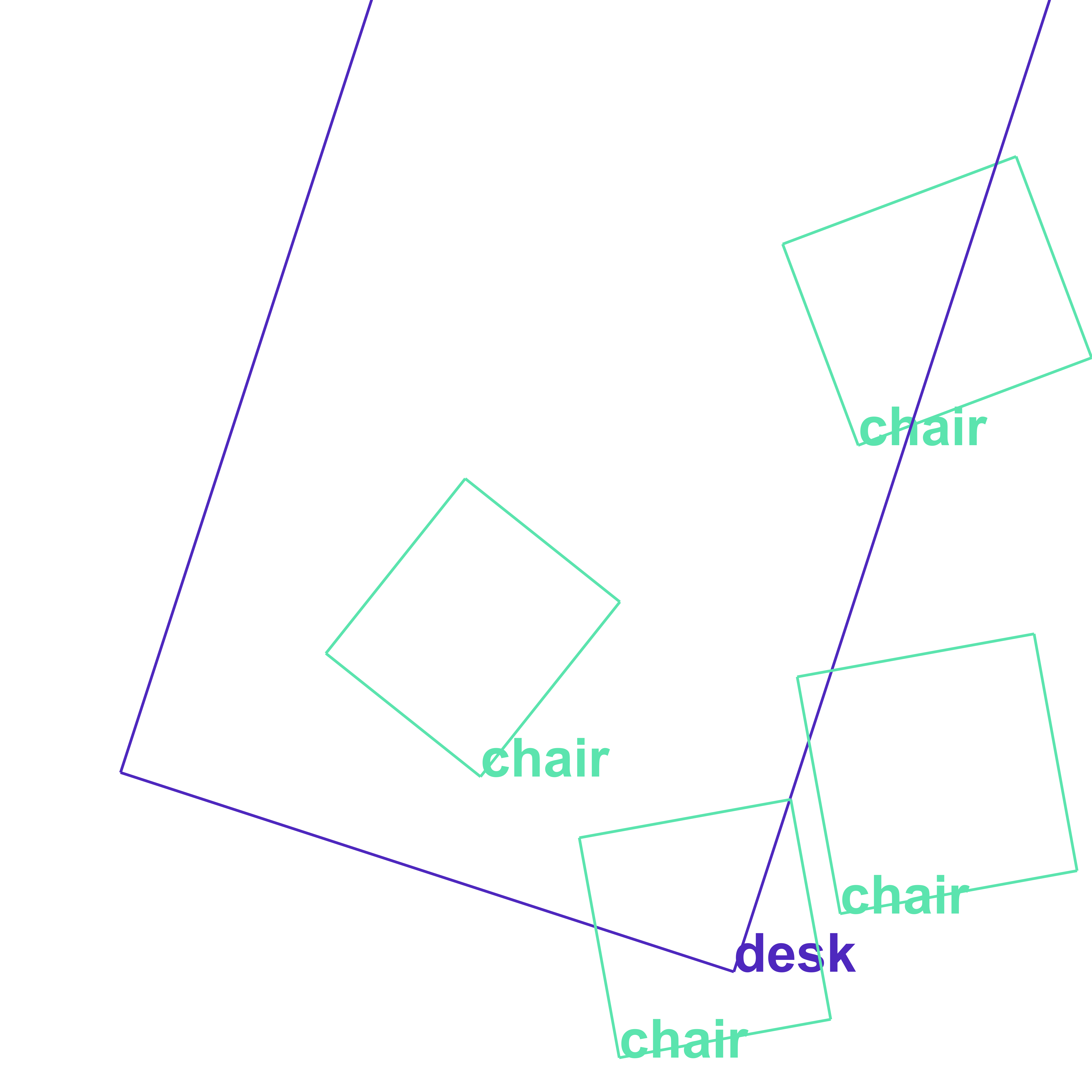} &
\includegraphics[width=\sunrgbdWidth,frame=.1pt]{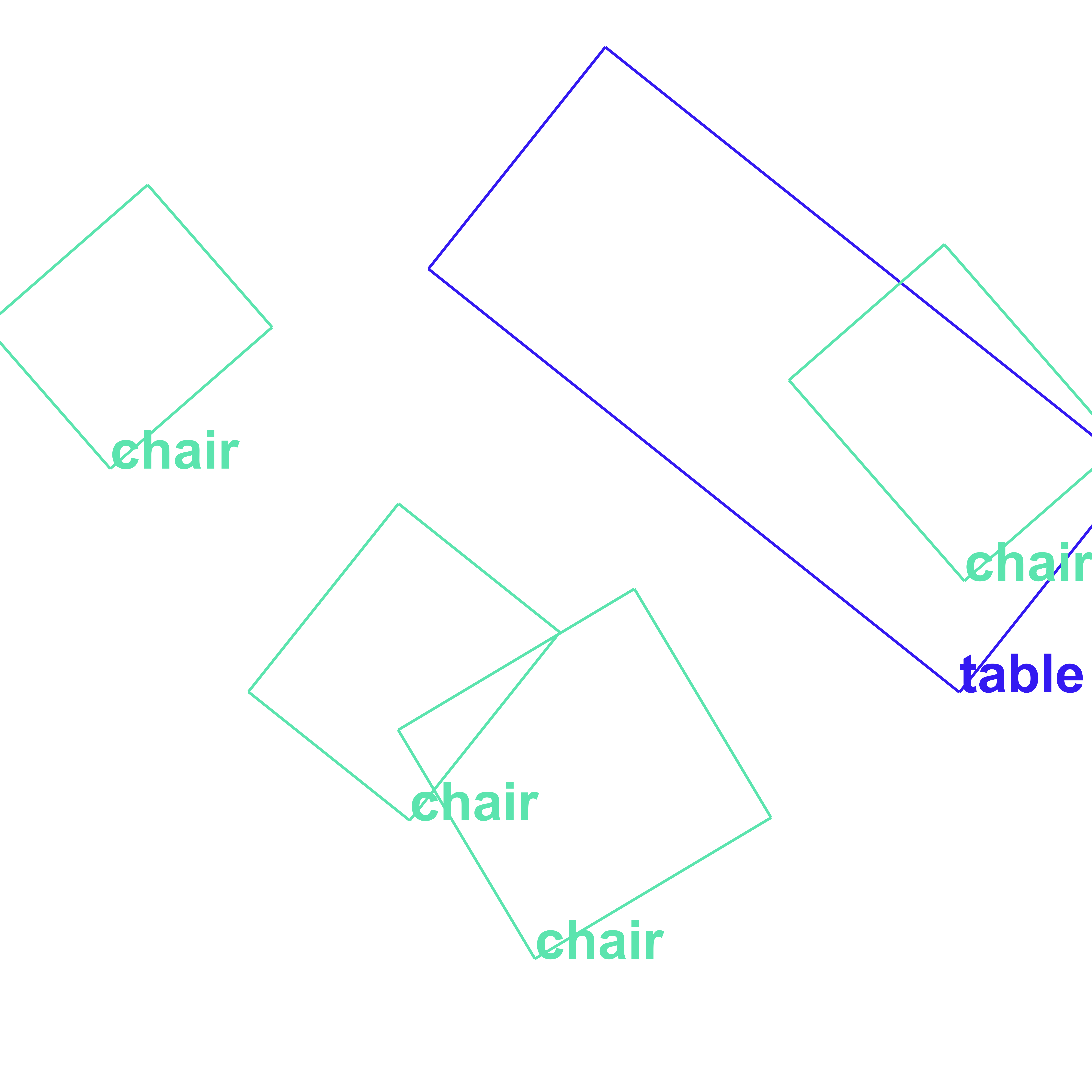} \\
\includegraphics[width=\sunrgbdWidth,frame=.1pt]{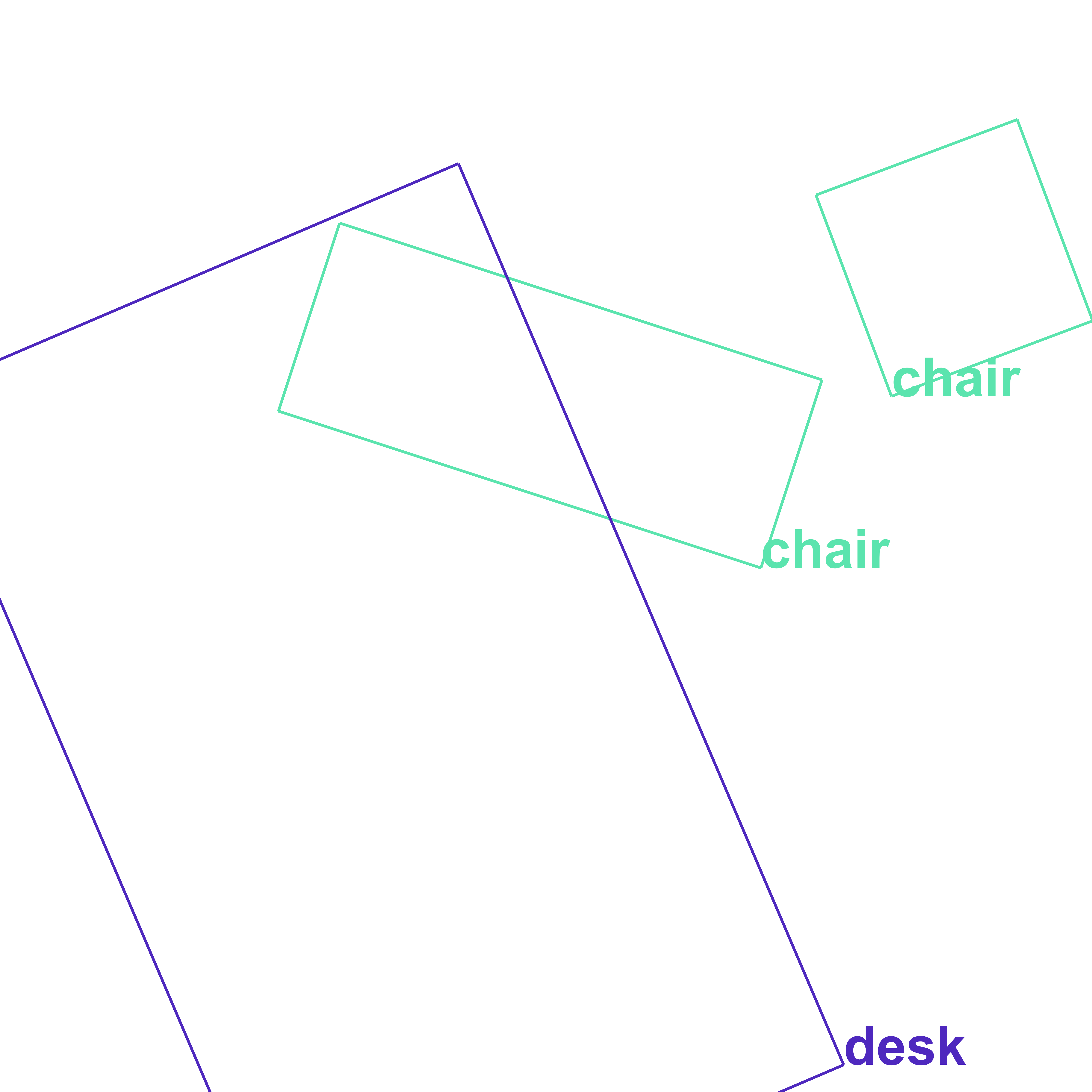} &
\includegraphics[width=\sunrgbdWidth,frame=.1pt]{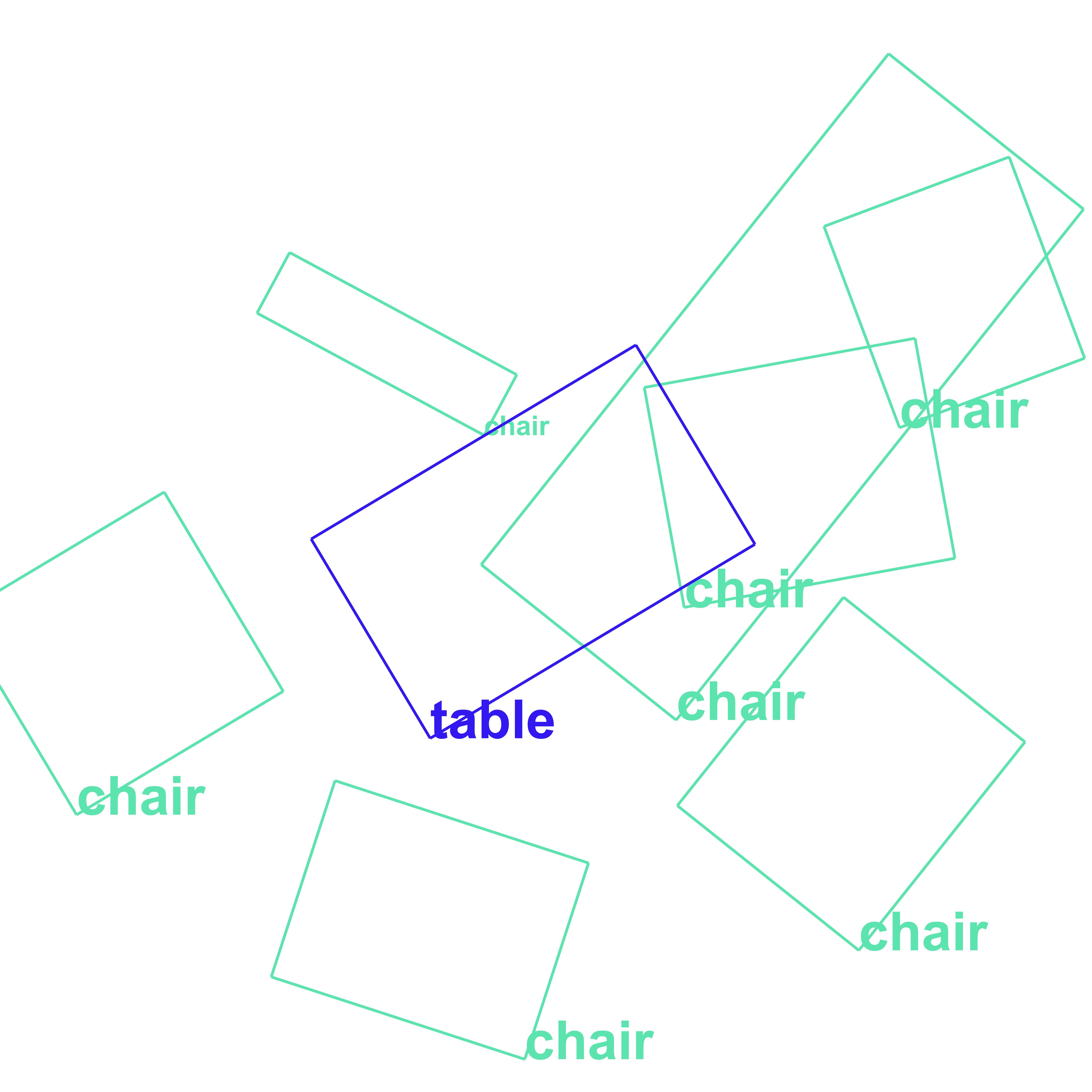} &
\includegraphics[width=\sunrgbdWidth,frame=.1pt]{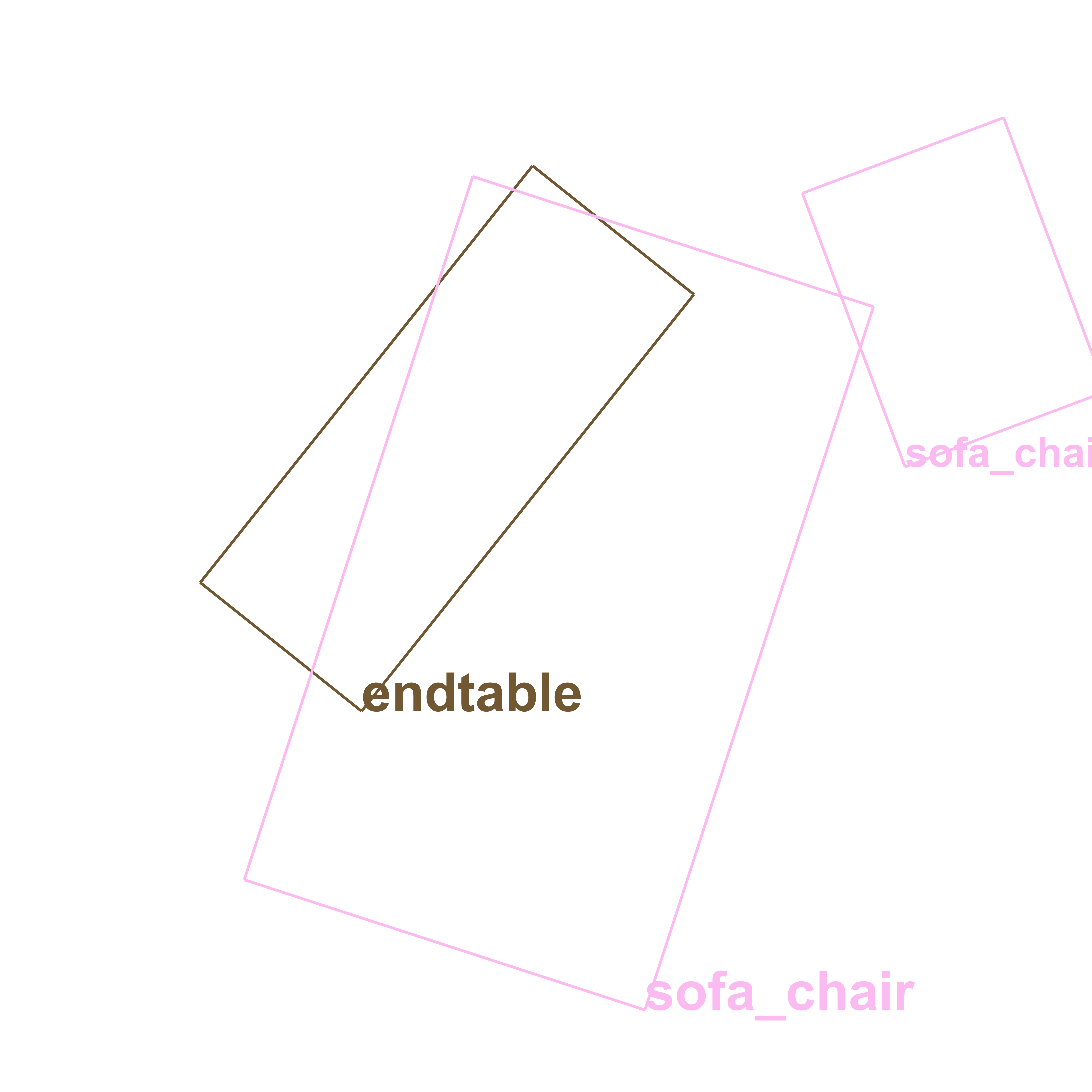} &
\includegraphics[width=\sunrgbdWidth,frame=.1pt]{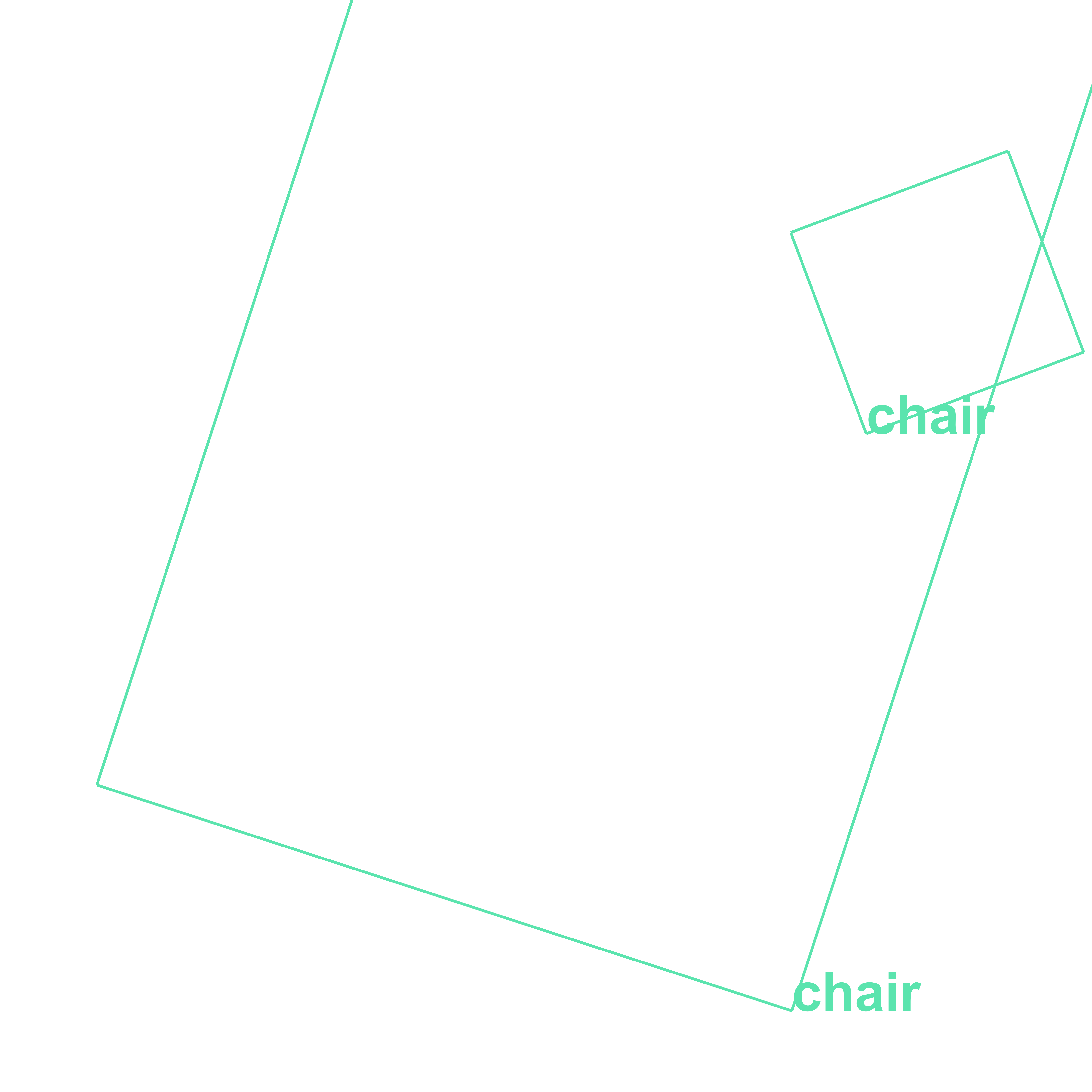} &
\includegraphics[width=\sunrgbdWidth,frame=.1pt]{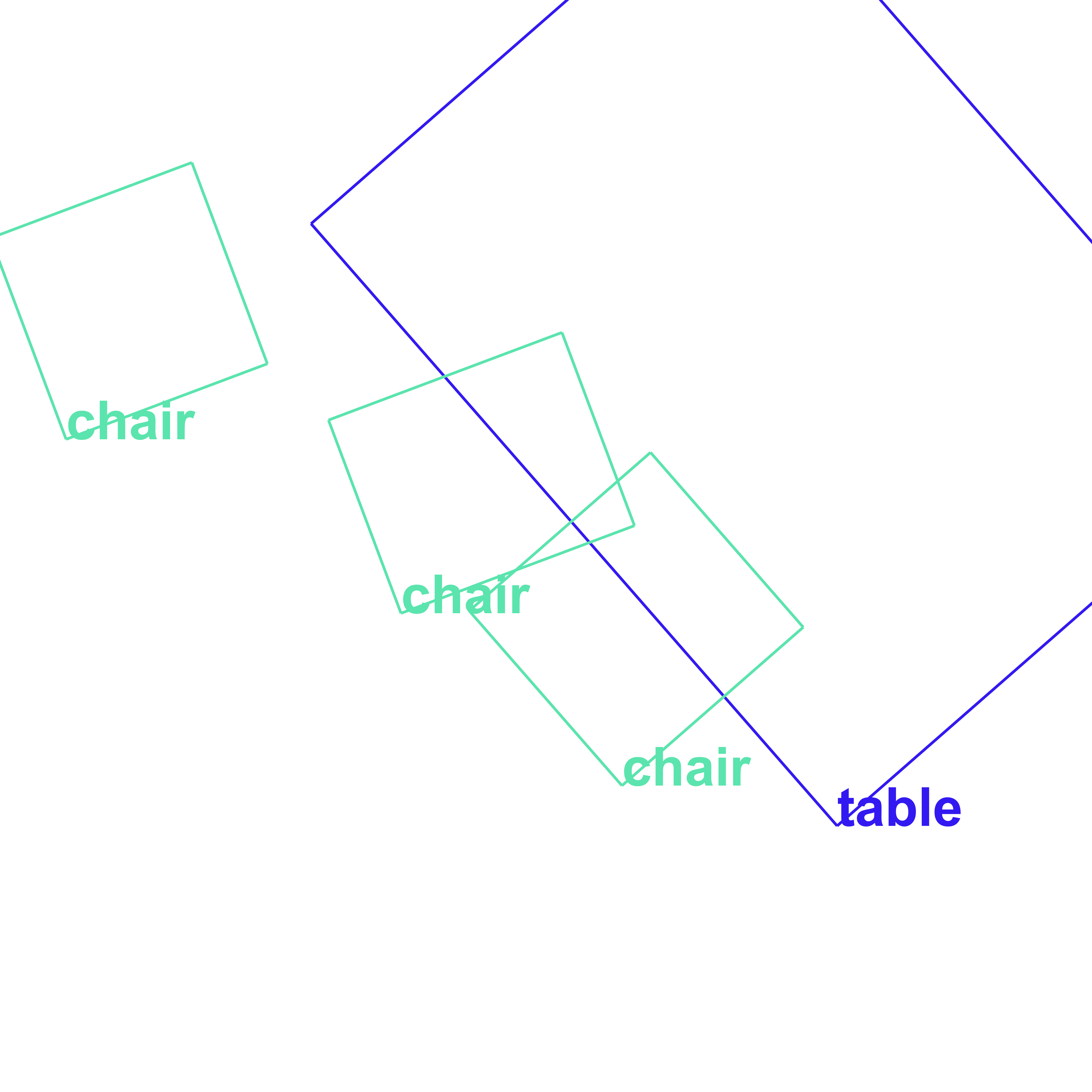} &
\includegraphics[width=\sunrgbdWidth,frame=.1pt]{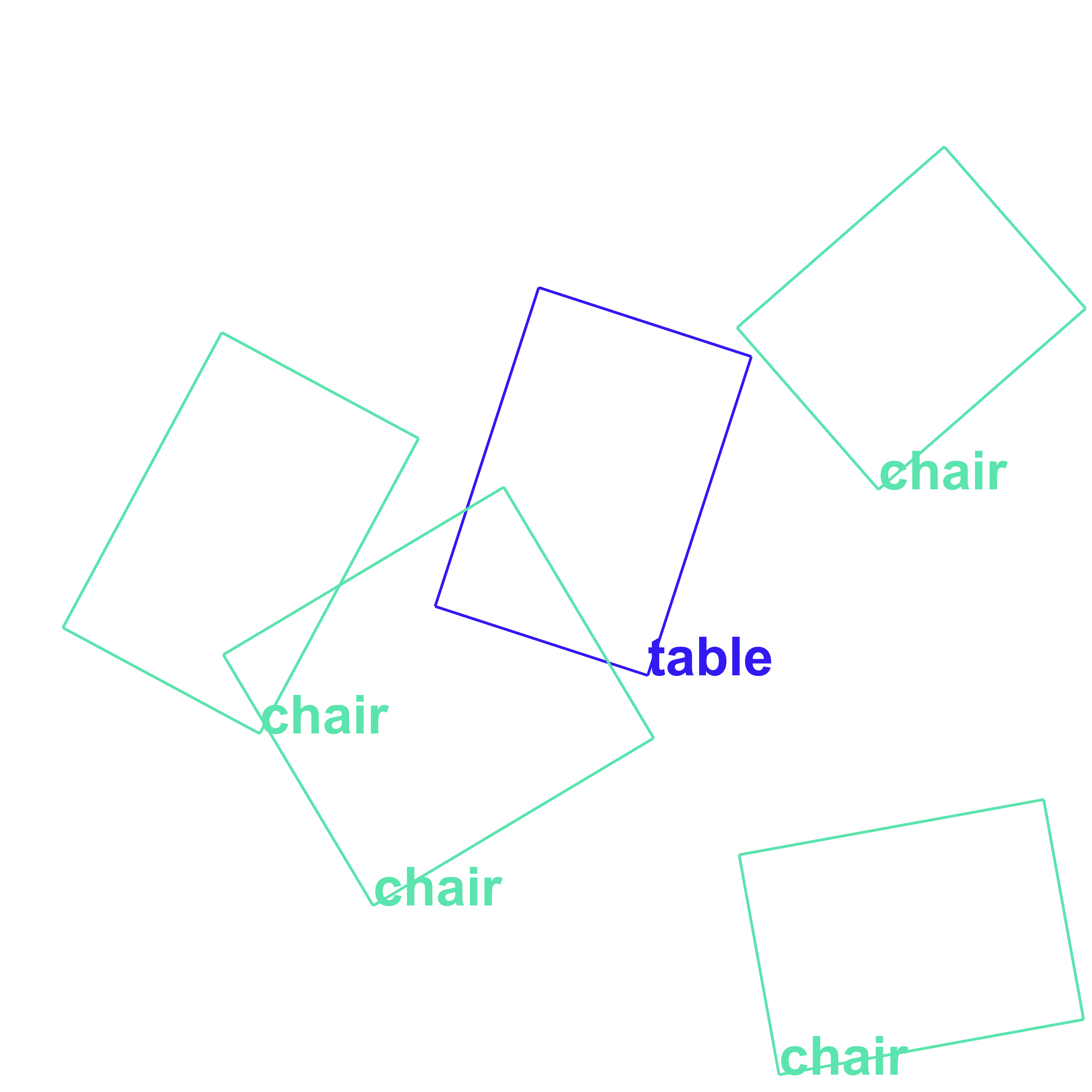} \\

    \end{tabular}
    \caption{Synthesized SUN RGB-D samples.}
    \label{fig:sunrgbd}
\end{figure}
\clearpage
\subsection{Variational Transformer Networks for Language Modeling}

The main motivation for our architectural choices is the success of self-attention in the field of natural language processing. It begs the question, how well does our method perform on text generation tasks? We train our autoregressive model on a dataset of 800K Amazon book reviews from \cite{ni-etal-2019-justifying} with no major modifications to the architecture: we simply replace the first fully connected layer with a word embedding (learned from scratch) and the output shape to that of our target vocabulary (30K words). For this experiment we use 6 self-attention layers on both encoder and decoder. In tab. \ref{tab:amazon_samples} we show the results of our method. The reviews have correct grammar and are highly realistic. This shows that our method can be applied to other tasks where the input is a sequence of data.

\begin{table}[h]
    \centering
    \setlength{\tabcolsep}{2pt}
    \begin{tabular}{l|l}\toprule
\rotatebox{90}{\hspace{-1cm}Real samples} & \begin{tabular}[c]{@{}p{17cm}@{}}A great read focusing on the main character named Shadow. The gods mentioned in the title are a dark and fading lot, and the story could be seen as a sort of parable about the old making way for the new. Good entertainment, though.\\ Just finished this book. I very much liked it. It kept me engaged throughout the entire reading. It sometimes seems as if you're not fully immersed in Shadow's like and persona, however, at the end of the book, I felt as though I knew enough to feel satisfied.\\ Long  winded and no excitement. It just plodded. On until it ended. complete waste of time, do t plan on reading anymore of this author's material\\ This is a really imaginative story that takes you totally by surprise and all over several dimentions.\\ Although I spent most of this book being not quite sure what was going on, once I finished it and took the whole thing in I loved it. The revelations at the end are what really sealed the deal for me, finding them very clever and fulfilling.\\ Great book. It felt as though the book moved both slowly and quickly as only a true classic can.\\ Great book! I love the Blossom Street Series! I am hooked on that series! Got the book in a reasonable amount of time! Very happy with the book! Came in great condition!\\ I have enjoyed Debbie Macomber's other book series so I decided to try this Blossom a Shop series. I love it and lose myself in the lives of the characters!\\ Loved it.\\ This series is awesome as good as all her books .I really enjoyed these books a lot\\ Anyone who enjoys reading about women's friendships will love this book. As a knitter I really like reading about the knitting and have taken away some ideas to use in my knitting projects, especially for the charitable knitting group I'm in!
\end{tabular} \\\hline
\rotatebox{90}{Ours} & \begin{tabular}[c]{@{}p{17cm}@{}}interesting ride of danger and suspense i think clancy has been on our lookout to clean ways\\ these books are awesome and can't wait to read more about the history of the city going through some presidential topped\\ a really good novel i couldn't put down there are many ups and downs yet so uplifting\\ very good book about an amazing woman this one leaves you begging and be entertained\\ therapy and exciting parts should be read by everyone real life persons an approach thanks\\ thanks for opening my eyes to everyone i've never read anything like it was very great\\ a great read one learns and never boring facts in school so that you found it very believable\\ a few of the characters were not particularly compelling but they never got any better it let me down with many people\\ good suspense and great writing will open a person's eyes on other subject matter\\ great read and interesting ending chapters are sad and far away from you\\ very interesting read hard to put down lehane does have really researched his subject matter but he made it realistic plus the ending in a quit\\ we learn about africa and from this novel every time we really need a man and makes it\\ a must read for any real good reason it's been years since the author came to life of your face\\ great book just like divergent and always worth reading this year messages of speech has been thorough\\ love it all my life truly a page turner and what else is happening great stuff about endurance\\ excellent end to an amazing trilogy energy and suspense can't wait there is another life\\ good read but takes forever to write about things and its our real lives\\ another success for john o'donohue and the same day i meet him but excellent a lot of true emotions\\ unexpected turn in surprises and surprise ending was a UNK think again and again about how worthwhile\\ a great read and i love his writing and it made me think about leader of tennis authorities\\ yes it is thinking me things i never knew about glad you were written or heard from author\\ good read makes you think and feel how it was affecting the people throughout every new books
\end{tabular}\\\bottomrule
\end{tabular}
\caption{\textbf{Top}: Real Amazon book reviews from \cite{ni-etal-2019-justifying}. \textbf{Bottom}: reviews generated with our method.}
    \label{tab:amazon_samples}
\end{table}
\clearpage
\section{Convergence tests}

\begin{figure}[h]
    \centering
    \includegraphics[width=\linewidth]{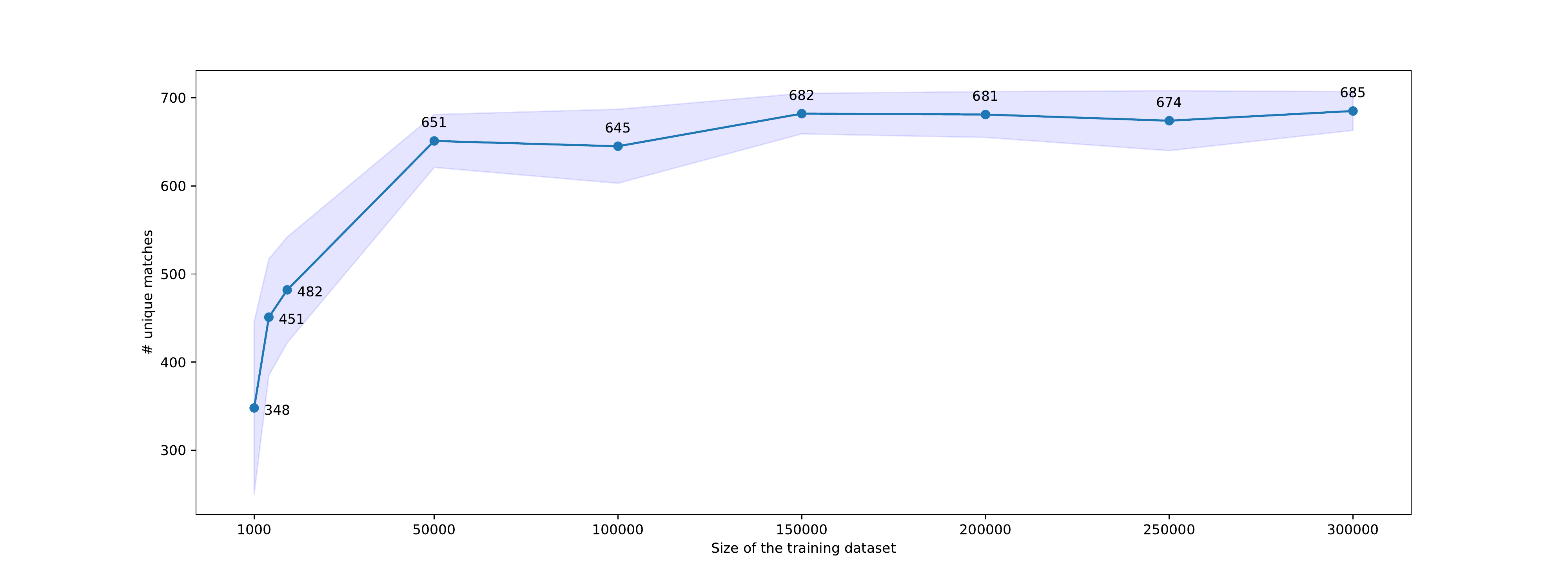}
    \caption{Convergence results on PubLayNet $\pm 1 \sigma$.}
    \label{fig:convergence_test}
\end{figure}

In order to determine the required number of elements for our model to generalize, we train our autoregressive model on various subsets of the PubLayNet training data, and evaluate the number of unique DocSim matches on 1000 samples. We average the results across 5 identical trainings. In fig. \ref{fig:convergence_test} we show the results. For this particular dataset, 50K training samples are enough to generalize well.

\section{Distribution analysis}

\begin{figure}[h]
    \centering
    \setlength{\tabcolsep}{1pt}
    \newlength{\distributionWidth}
    \setlength{\distributionWidth}{0.15\linewidth}
    \begin{tabular}{ccccccc}
    &All & Text & Title & Figure & List & Table \\
    \rotatebox{90}{\hspace{1.3cm}\footnotesize Synthetic} &
    \includegraphics[width=\distributionWidth]{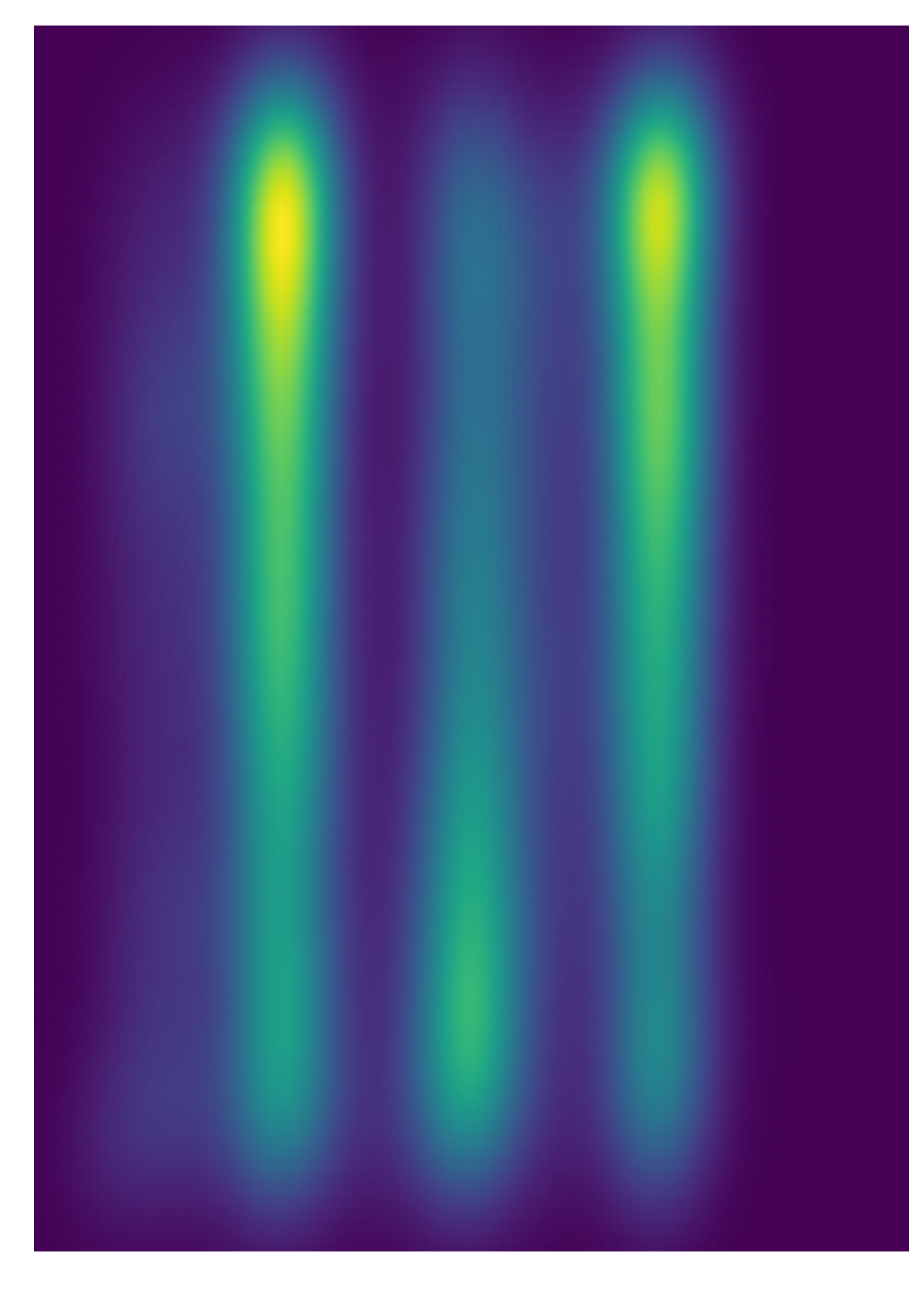} &
    \includegraphics[width=\distributionWidth]{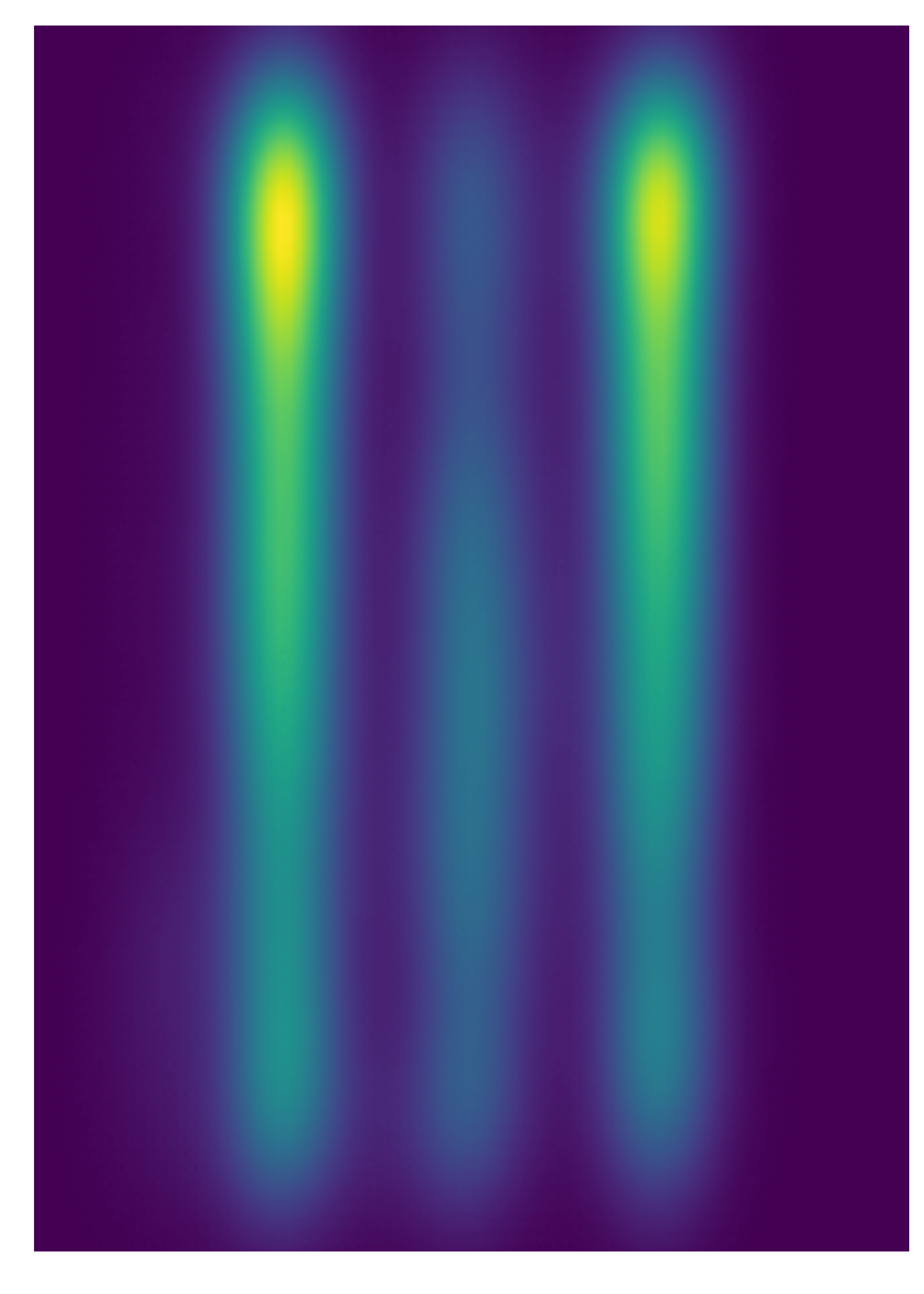} &
    \includegraphics[width=\distributionWidth]{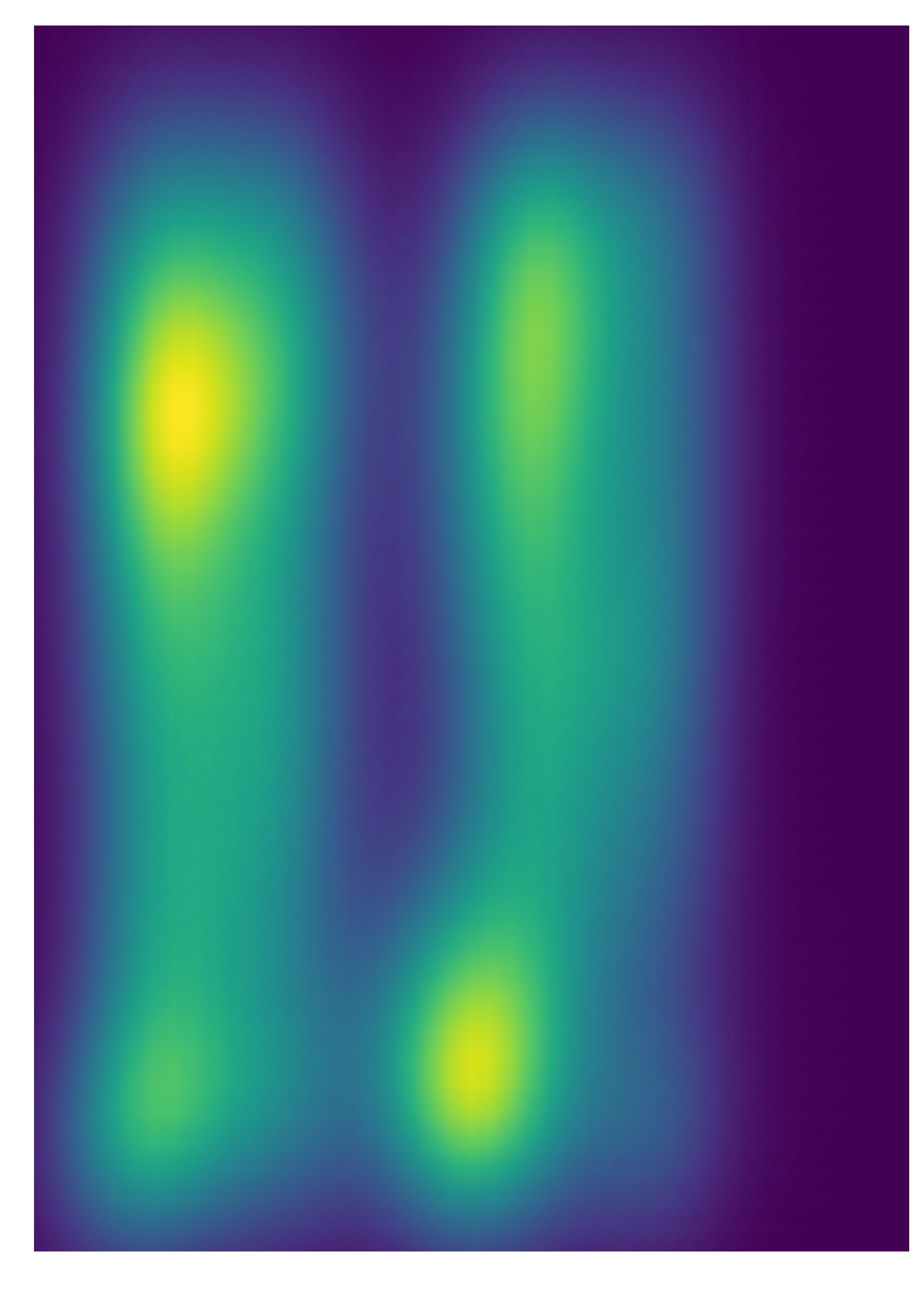} &
    \includegraphics[width=\distributionWidth]{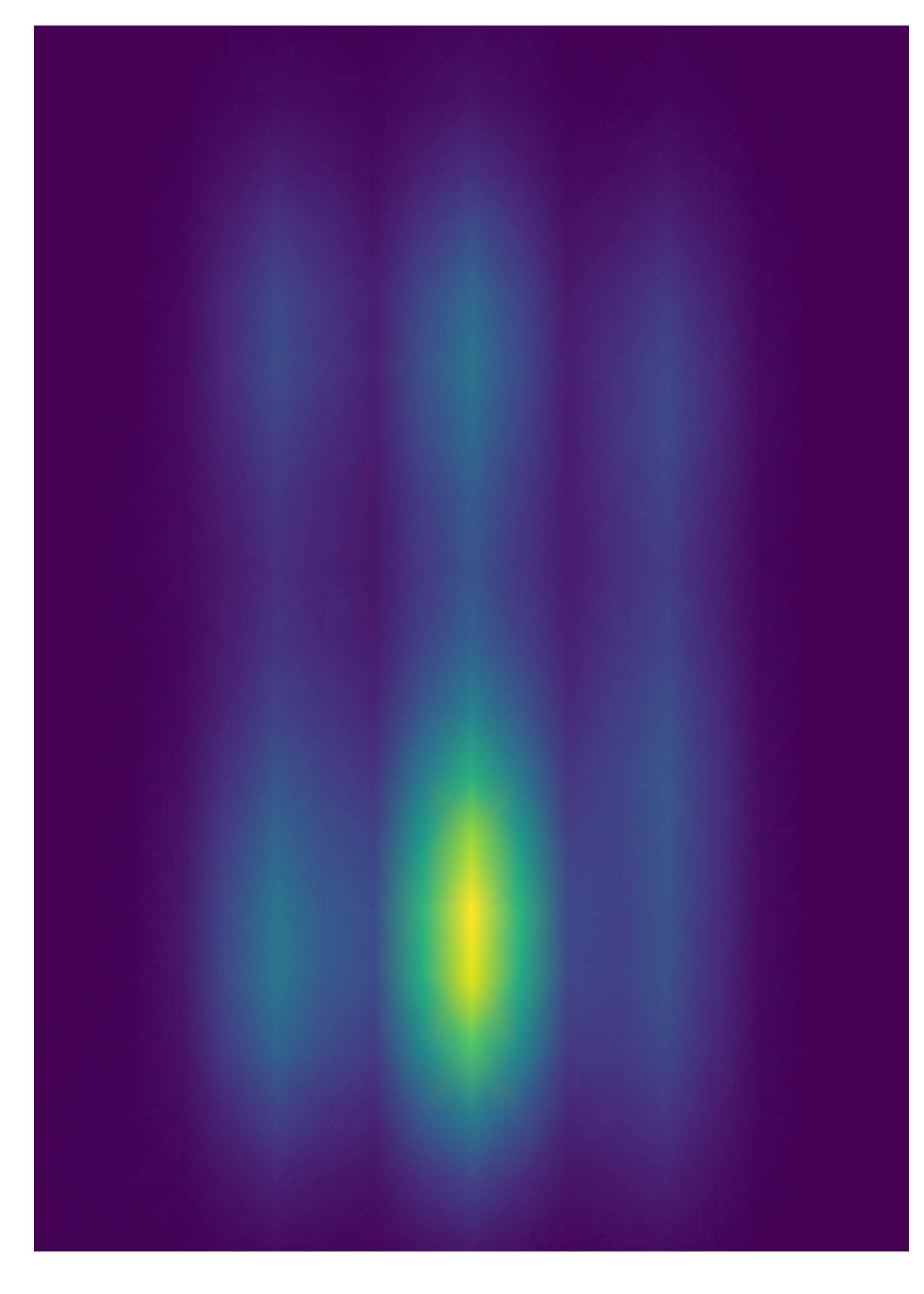} &
    \includegraphics[width=\distributionWidth]{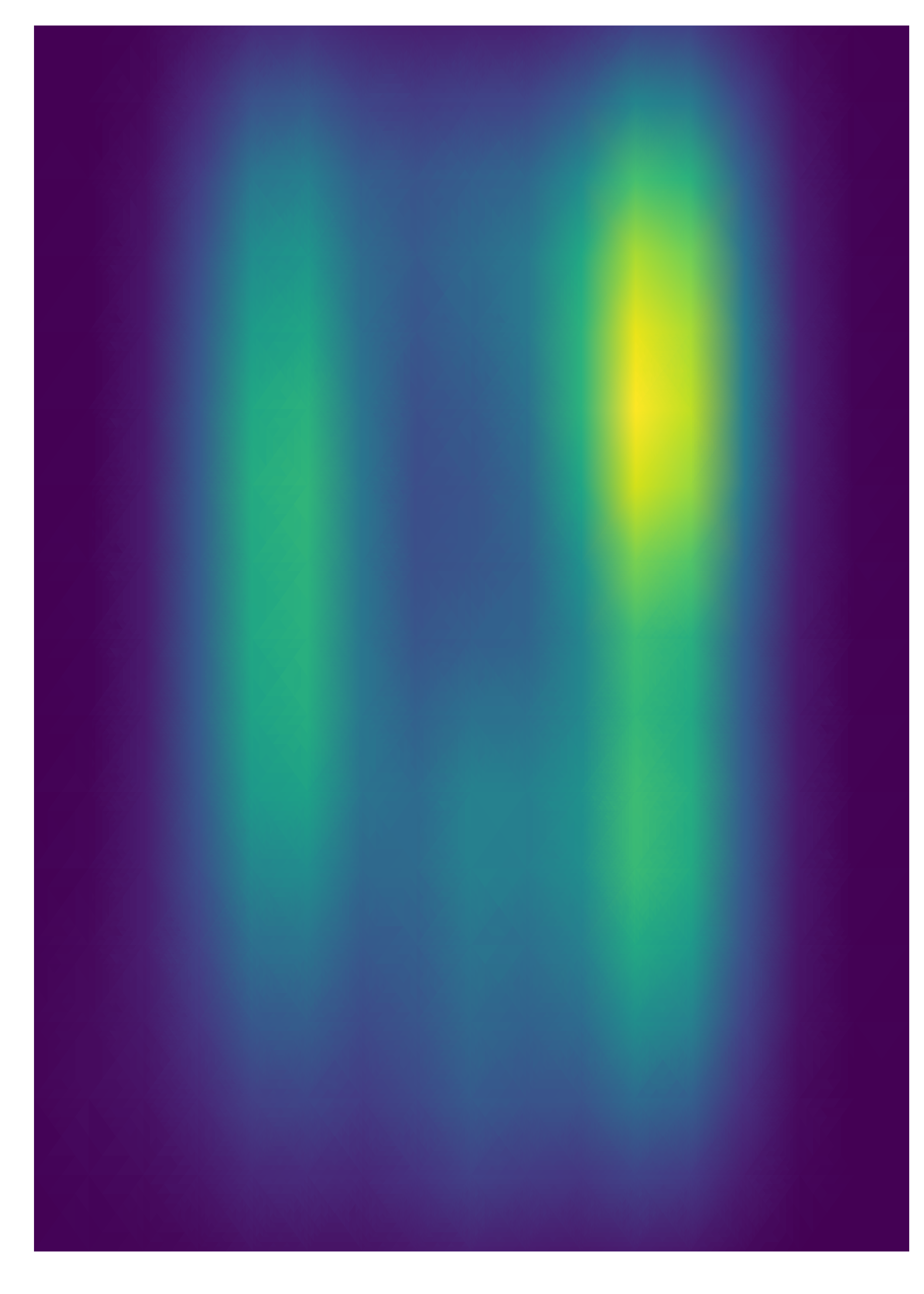} &
    \includegraphics[width=\distributionWidth]{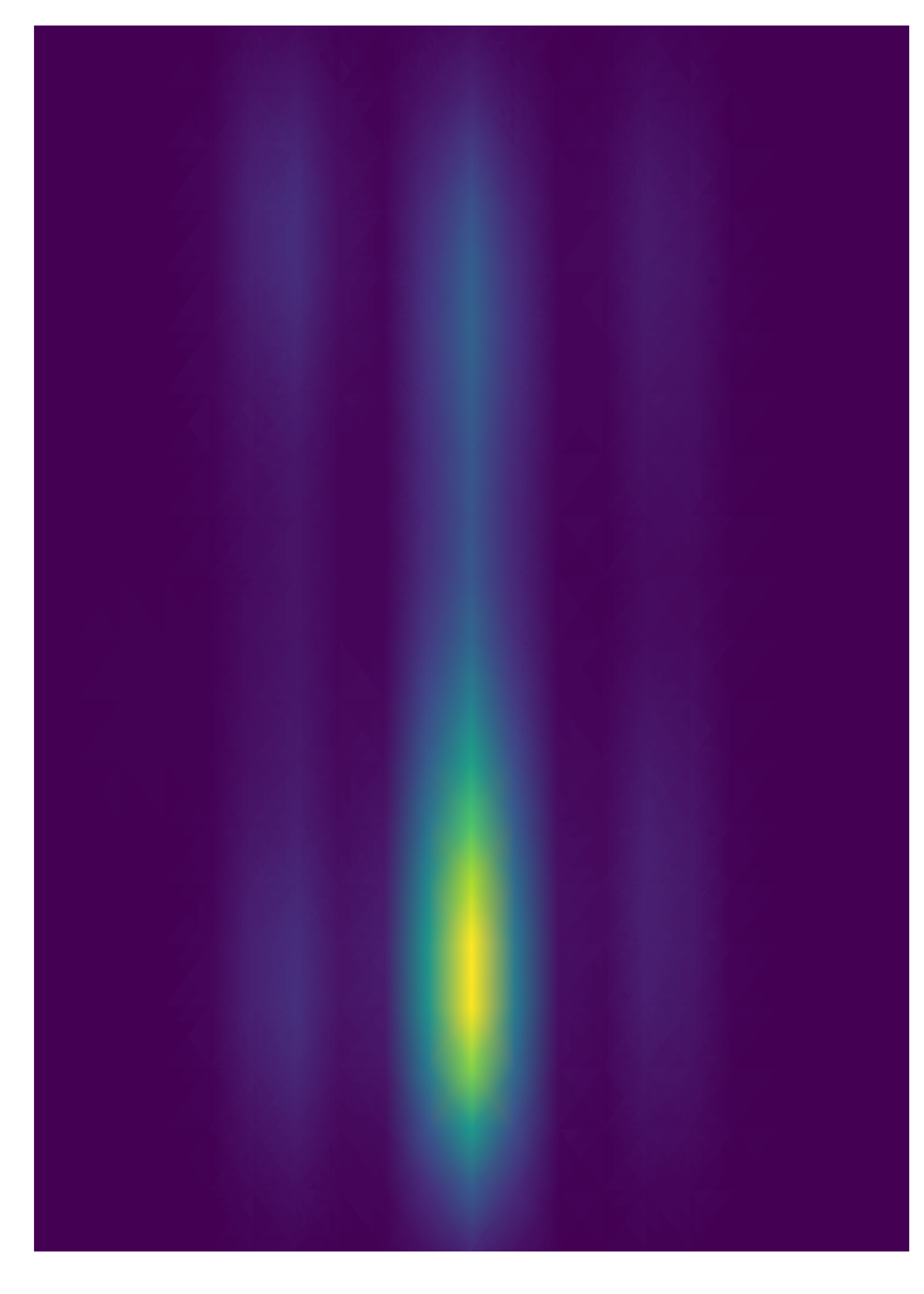} \\
    \rotatebox{90}{\hspace{1.5cm}\footnotesize Real} &
    \includegraphics[width=\distributionWidth]{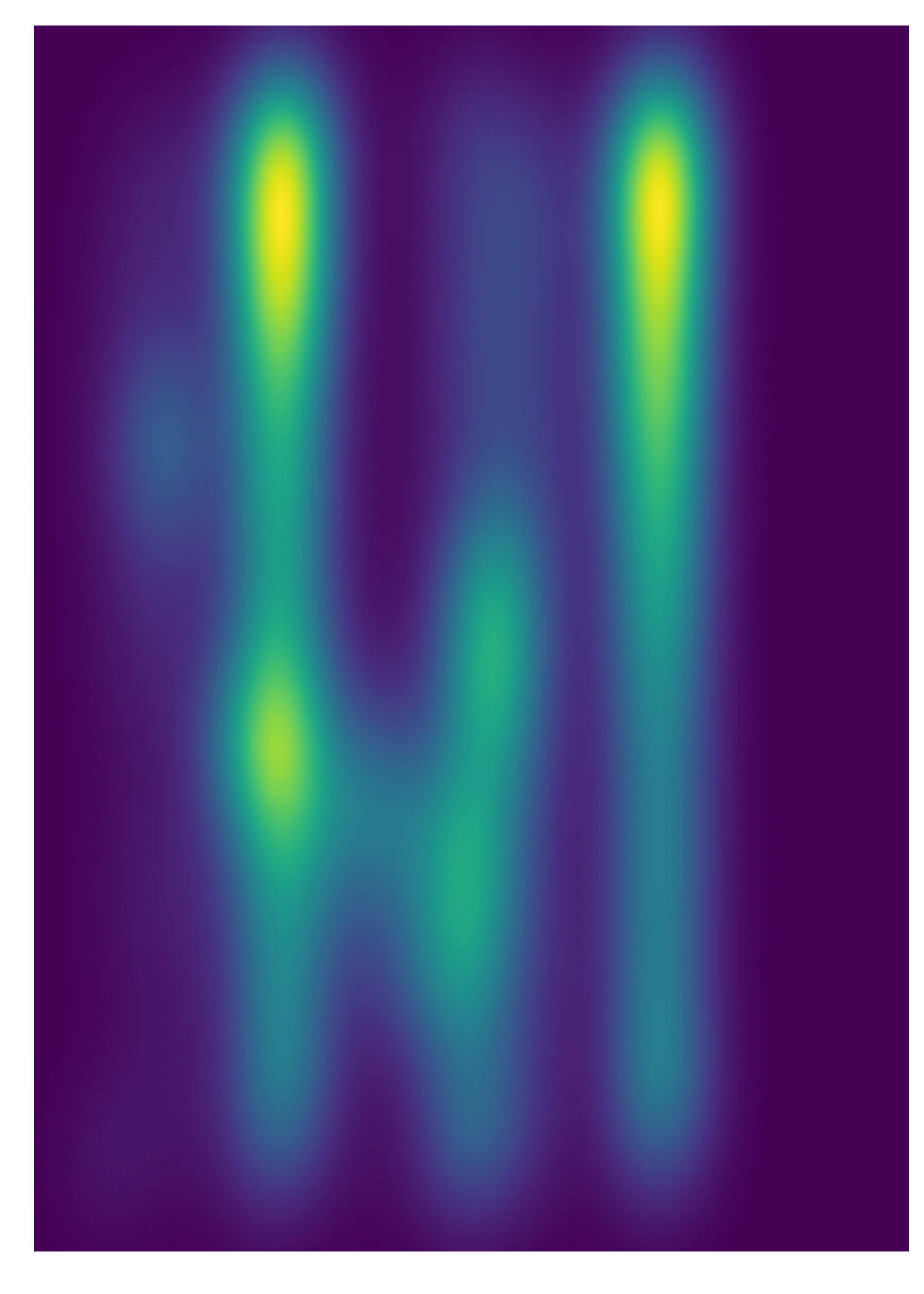} &
    \includegraphics[width=\distributionWidth]{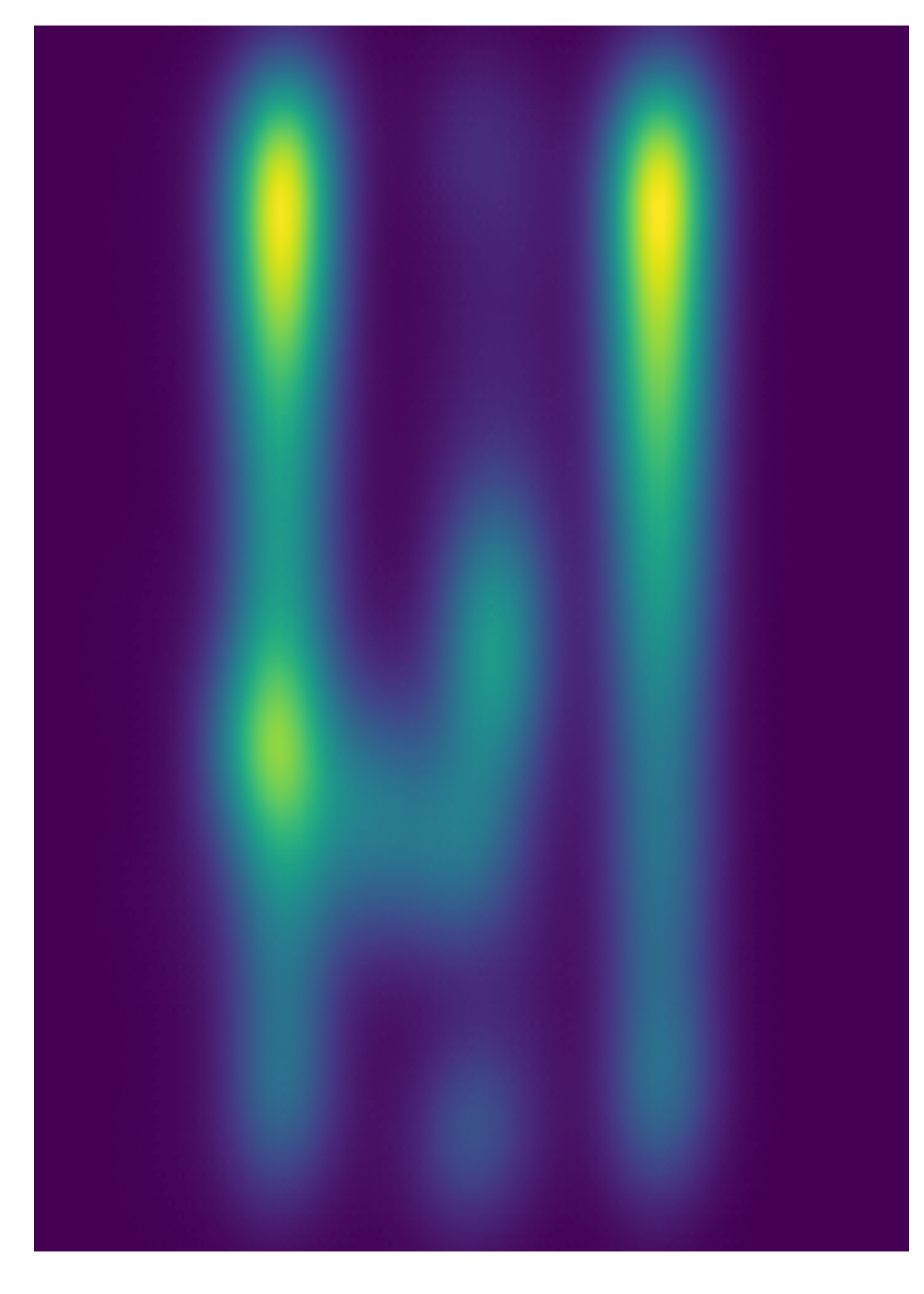} &
    \includegraphics[width=\distributionWidth]{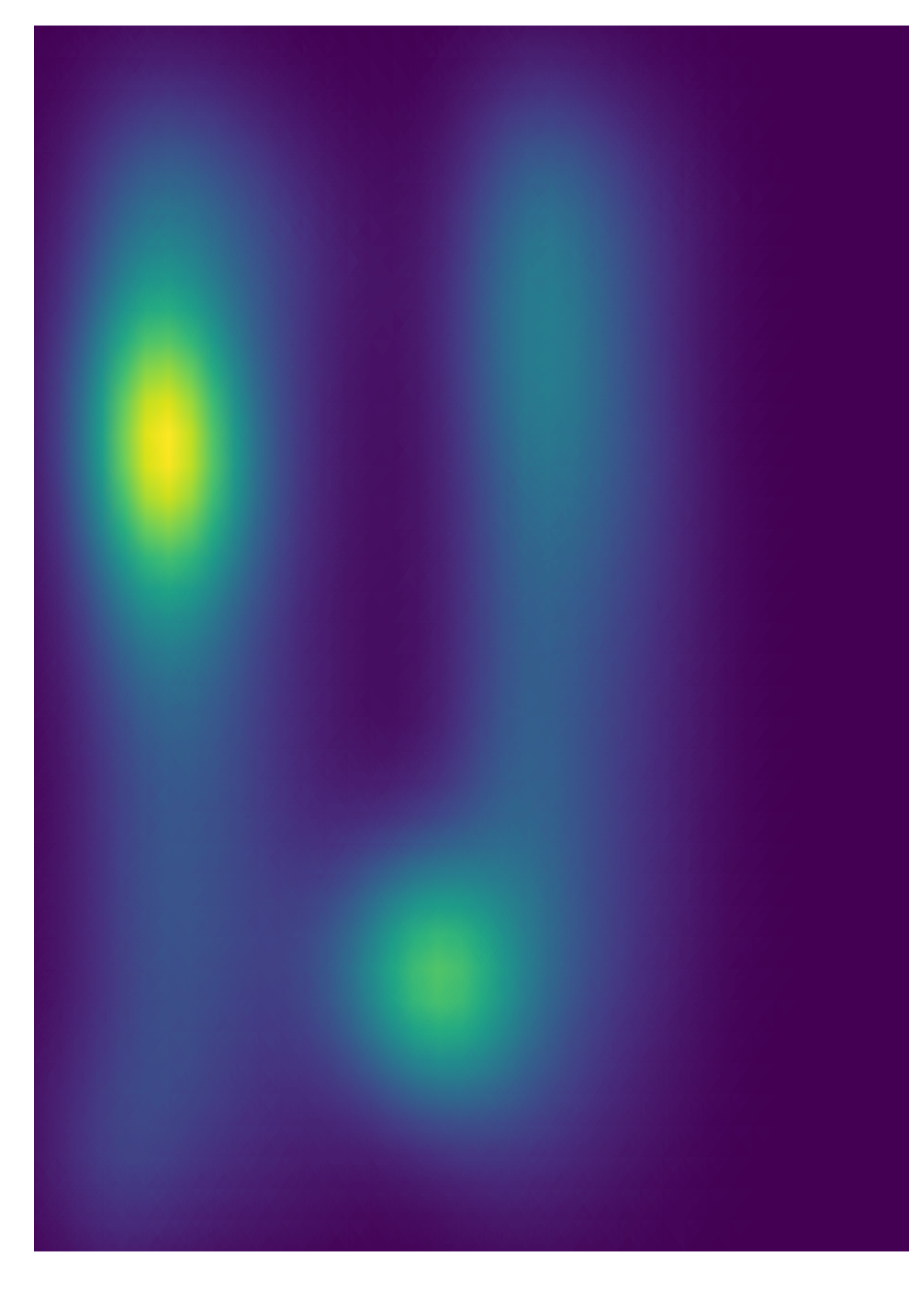} &
    \includegraphics[width=\distributionWidth]{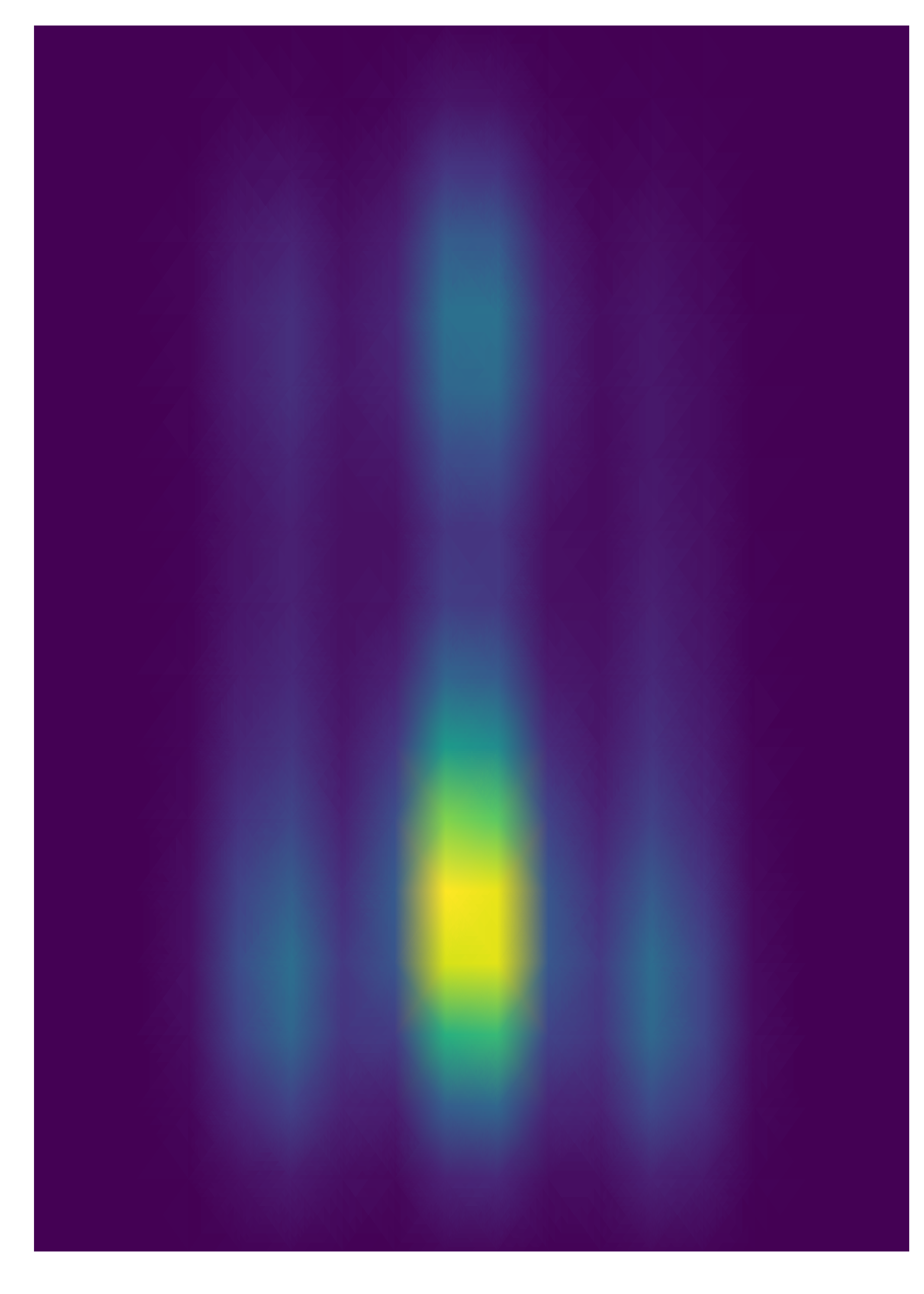} &
    \includegraphics[width=\distributionWidth]{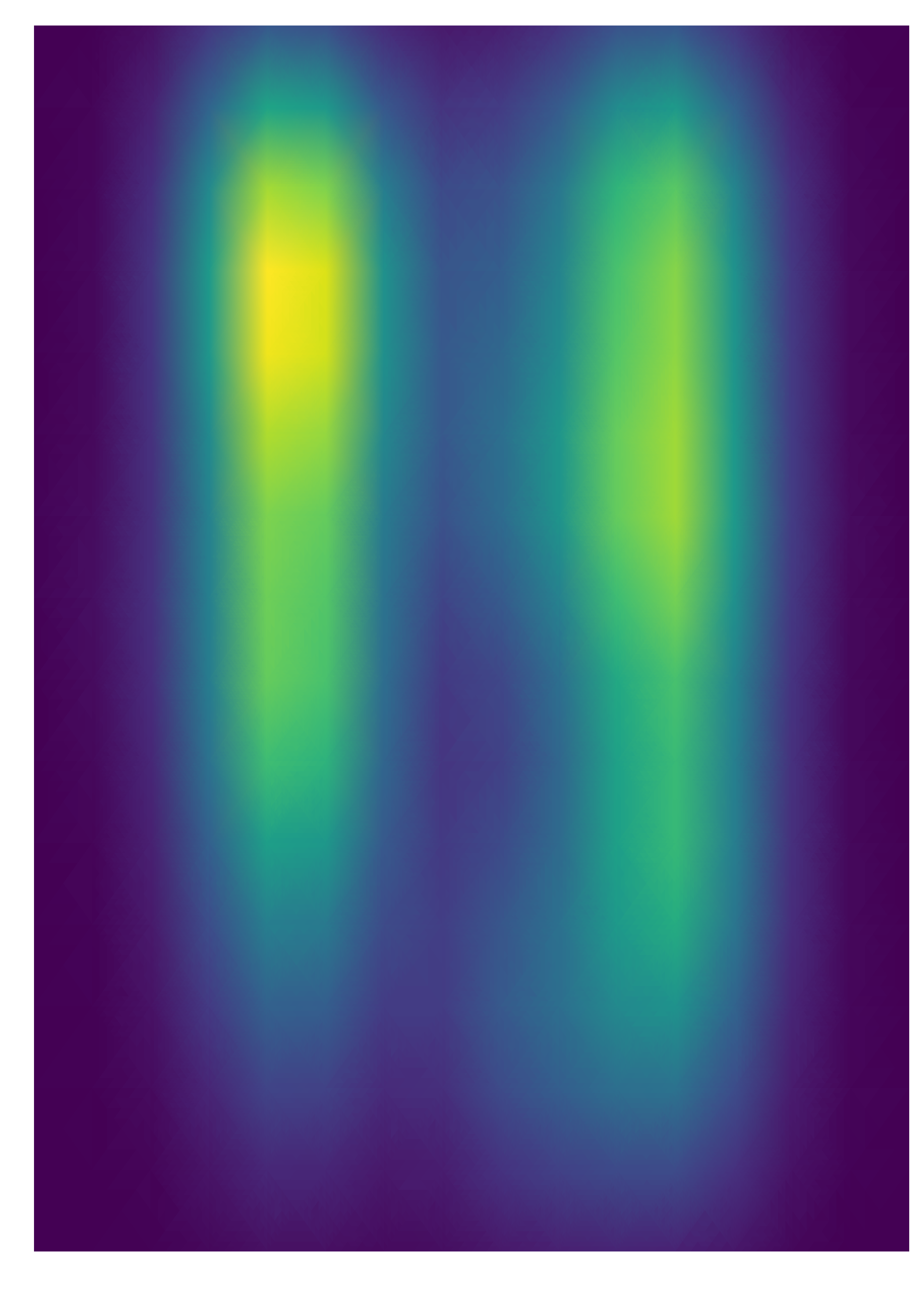} &
    \includegraphics[width=\distributionWidth]{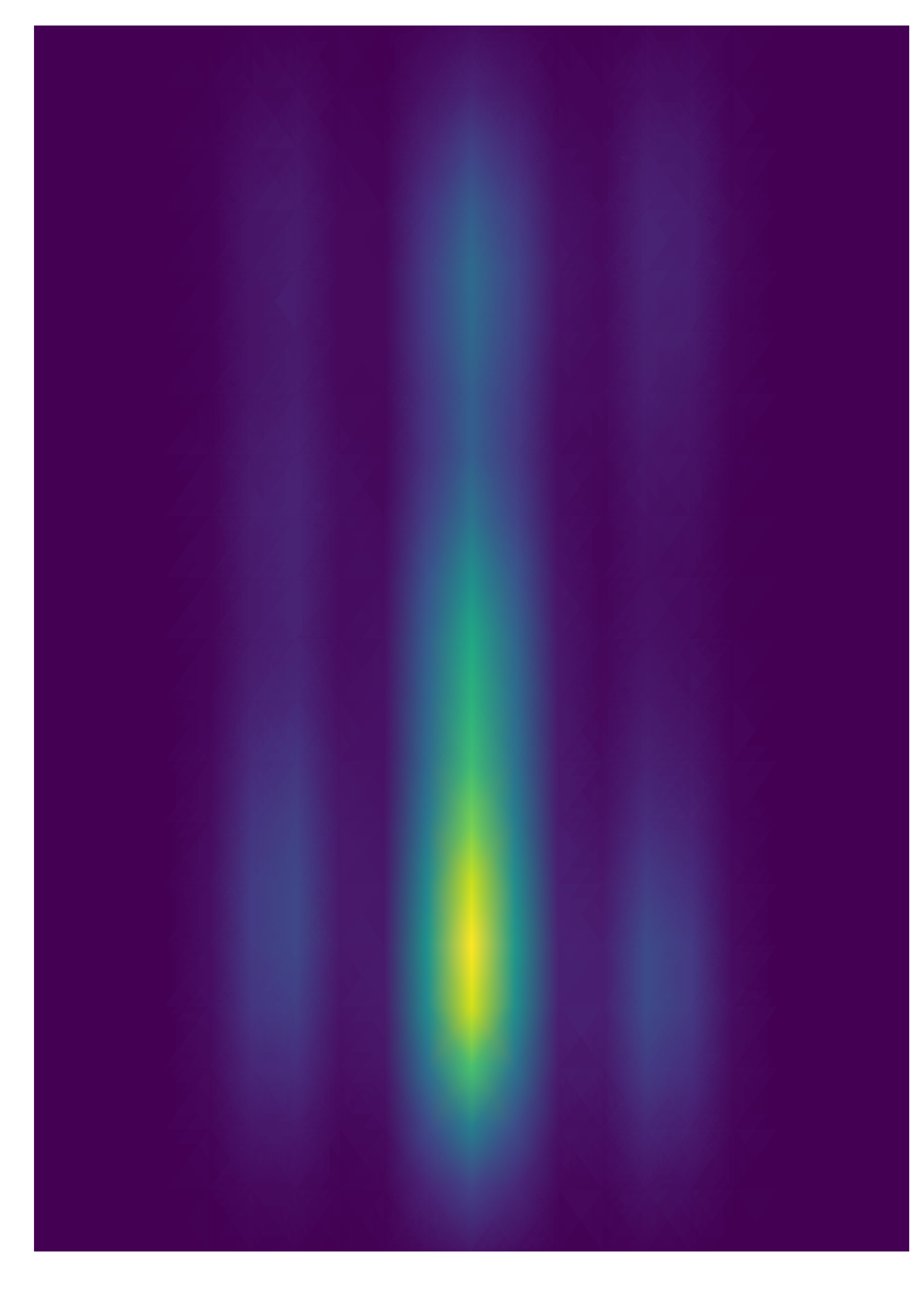} \\
    \end{tabular}
    \caption{\textbf{Top}: distribution of the bounding box center for synthetic data. \textbf{Bottom}: real data.}
    \label{fig:distribution_analysis}
\end{figure}

As an additional metric for the ability of our method to capture the layout distribution, in fig. \ref{fig:distribution_analysis} we show the frequency of each location as the center of a bounding box on 1000 samples of the PubLayNet test dataset and our model.

\clearpage
{\small
\bibliographystyle{ieee_fullname}
\bibliography{egbib}
}

\clearpage